\def\isarxiv{1} %%% for icml submission version, we comment this line

\ifdefined\isarxiv
\documentclass[11pt]{article}

\usepackage[numbers]{natbib}

\else
\documentclass[10pt,twocolumn,letterpaper]{article}

\usepackage[review]{cvpr}      % To produce the REVIEW version
\usepackage{xcolor}
\definecolor{cvprblue}{rgb}{0.21,0.49,0.74}
\usepackage[pagebackref,breaklinks,colorlinks,citecolor=cvprblue]{hyperref}

\fi

\usepackage{amsmath}
\usepackage{amsthm}
\usepackage{amssymb}
\usepackage{algorithm}
\usepackage{algpseudocode}
\usepackage{graphicx}
\usepackage{grffile}
\usepackage{wrapfig,epsfig}
\usepackage{url}
\usepackage{xcolor}
\usepackage{epstopdf}
\usepackage{hyperref}

\usepackage{bbm}
\usepackage{dsfont}

\allowdisplaybreaks

\ifdefined\isarxiv

\usepackage{tikz}
\usepackage{hyperref}  %%% arxiv don't allow this.
\hypersetup{colorlinks=true,citecolor=blue,linkcolor=blue} %%% Zhao : maybe we should comment this in submission.
\usetikzlibrary{arrows}
\usepackage[margin=1in]{geometry}

\else

\fi
\graphicspath{{./figs/}}

\newtheorem{theorem}{Theorem}[section]
\newtheorem{lemma}[theorem]{Lemma}
\newtheorem{definition}[theorem]{Definition}

\newtheorem{corollary}[theorem]{Corollary}

\newtheorem{fact}[theorem]{Fact}
\newtheorem{remark}[theorem]{Remark}

\newcommand{\wt}{\widetilde}
\newcommand{\ov}{\overline}

\newcommand{\R}{\mathbb{R}}

\renewcommand{\d}{\mathrm{d}}

\DeclareMathOperator*{\E}{{\mathbb{E}}}

\DeclareMathOperator{\poly}{poly}

\DeclareMathOperator{\nnz}{nnz}

\DeclareMathOperator{\diag}{diag}

\makeatletter
\newcommand*{\RN}[1]{\expandafter\@slowromancap\romannumeral #1@}
\makeatother
\usepackage{lineno}

\ifdefined\isarxiv

\else

\title{A Theoretical Insight into Attack and Defense of Gradient Leakage in Transformer}

\author{First Author\\
Institution1\\
Institution1 address\\
{\tt\small firstauthor@i1.org}
\and
Second Author\\
Institution2\\
First line of institution2 address\\
{\tt\small secondauthor@i2.org}
}

\fi

\begin{document}

\ifdefined\isarxiv

\date{}

\title{A Theoretical Insight into Attack and Defense of Gradient Leakage in Transformer}%{Theoretical Insights into Gradient Leakage and Safeguarding Data with Differential Privacy in Distributed Transformer Models}
%\iffalse
\author{ 
Chenyang Li\thanks{\texttt{lchenyang550@gmail.com}. Fuzhou University.} 
\and 
Zhao Song\thanks{\texttt{zsong@adobe.com}. Adobe Research.}
\and
Weixin Wang\thanks{\texttt{wwang176@jh.edu}. Johns Hopkins University.}
\and
Chiwun Yang\thanks{\texttt{christiannyang37@gmail.com}. Sun Yat-sen University.}
}
%\fi

\else

\maketitle
\fi

\ifdefined\isarxiv
\begin{titlepage}
  \maketitle
  \begin{abstract}
The Deep Leakage from Gradient (DLG) attack has emerged as a prevalent and highly effective method for extracting sensitive training data by inspecting exchanged gradients. This approach poses a substantial threat to the privacy of individuals and organizations alike. This research presents a comprehensive analysis of the gradient leakage method when applied specifically to transformer-based models. Through meticulous examination, we showcase the capability to accurately recover data solely from gradients and rigorously investigate the conditions under which gradient attacks can be executed, providing compelling evidence. Furthermore, we reevaluate the approach of introducing additional noise on gradients as a protective measure against gradient attacks. To address this, we outline a theoretical proof that analyzes the associated privacy costs within the framework of differential privacy. Additionally, we affirm the convergence of the Stochastic Gradient Descent (SGD) algorithm under perturbed gradients. The primary objective of this study is to augment the understanding of gradient leakage attack and defense strategies while actively contributing to the development of privacy-preserving techniques specifically tailored for transformer-based models. By shedding light on the vulnerabilities and countermeasures associated with gradient leakage, this research aims to foster advancements in safeguarding sensitive data and upholding privacy in the context of transformer-based models.

  \end{abstract}
  \thispagestyle{empty}
\end{titlepage}

{\hypersetup{linkcolor=black}
\tableofcontents
}
\newpage

\else

\begin{abstract}
The Deep Leakage from Gradient (DLG) attack has emerged as a prevalent and highly effective method for extracting sensitive training data by inspecting exchanged gradients. This approach poses a substantial threat to the privacy of individuals and organizations alike. This research presents a comprehensive analysis of the gradient leakage method when applied specifically to transformer-based models. Through meticulous examination, we showcase the capability to accurately recover data solely from gradients and rigorously investigate the conditions under which gradient attacks can be executed, providing compelling evidence. Furthermore, we reevaluate the approach of introducing additional noise on gradients as a protective measure against gradient attacks. To address this, we outline a theoretical proof that analyzes the associated privacy costs within the framework of differential privacy. Additionally, we affirm the convergence of the Stochastic Gradient Descent (SGD) algorithm under perturbed gradients. The primary objective of this study is to augment the understanding of gradient leakage attack and defense strategies while actively contributing to the development of privacy-preserving techniques specifically tailored for transformer-based models. By shedding light on the vulnerabilities and countermeasures associated with gradient leakage, this research aims to foster advancements in safeguarding sensitive data and upholding privacy in the context of transformer-based models.
\end{abstract}

\fi

\section{Introduction}

\begin{figure*}[ht]
    \centering
    \includegraphics[width=0.8\textwidth]{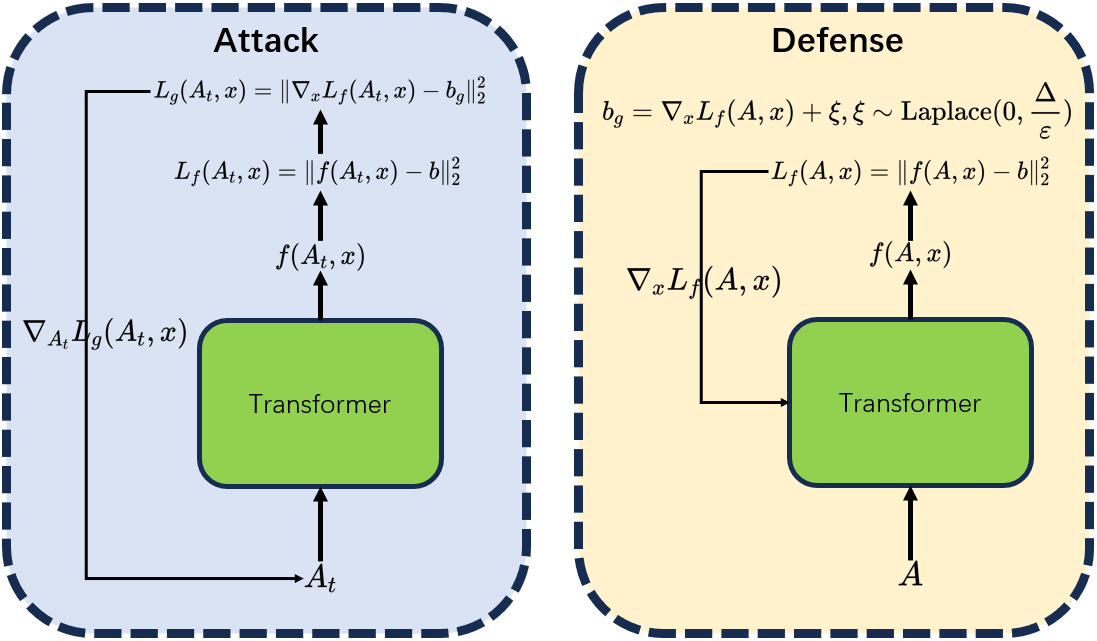}
    \caption{The visualization of attack and defense of gradient leakage in transformer.}
    \label{fig:architect}
\end{figure*}

Transformer models, such as Transformer, BERT, ViT, DETR, ChatGPT, CLIP, BLIP, Llama, GPT-4, PaLM, and Bard \cite{vsp+17, dclt19, dbk+20, cms+20, rkh+21, llxh22, cha22, llsh23, tli+23, tms+23, cnd+22, adf+23, bce+23, bar23}, have showcased remarkable performances in various domains, including Natural Language Processing (NLP), Computer Vision (CV), and Multimodal Learning \cite{gxl+22, har+22, yan20, glt20, hu20, vutf19, nzy21, lyyr20, cyz21, kgz+17}. These models rely on attention \cite{wcb14, bcb14, vsp+17} as the core computational mechanism. However, the effectiveness of the transformer models heavily relies on the quality of the training data \cite{kmh+20, hbm+22}. Unfortunately, a significant portion of high-quality data remains inaccessible to the public, posing a critical challenge.

To address this challenge, federated learning has emerged as a promising machine learning approach that addresses the issue of data isolation while preserving data privacy \cite{mmr+17, lftl20, rhl+20, zxb+21, mam21, kma+21, zll+18, ndp+21, lhs21}. In a decentralized machine learning setting, federated learning involves multiple clients, such as mobile devices, institutions, and organizations, collaborating with one or more central servers. Instead of sharing raw data, federated learning allows clients to exchange model weights and gradients, enabling them to enhance model performance while leveraging sensitive yet high-quality data. By adopting this strategy, federated learning enables improved performance while maintaining data privacy and confidentiality.

However, despite the perceived security and privacy advantages associated with the exchange of gradients in machine learning, there is a potential risk of privacy leakage in practical implementations \cite{zlh19, yzh21, hgs+21, gbdm20, lyy20, zmb20, wwc+21, lym+22, bds+23, lzll22, swyz23, ghd+22, kyl+23}. One prominent privacy attack method, known as Deep Leakage from Gradient (DLG) \cite{zlh19}, enables participants within a federated learning framework to potentially recover private data solely from the exchanged gradients. In formal terms, when given a model function $f$ and its weight $x$, the DLG method can reverse-engineer the training data by utilizing a gradient $g$. Previous research has also demonstrated that sharing gradients can potentially expose sensitive training data, whether it is in the form of images or text.

In this paper, we present a novel approach that focuses on the theoretical analysis of the attack and defense mechanisms related to gradient leakage in transformer-based models. To facilitate calculation and analysis, we adopt a transformation technique, inspired by \cite{dls23, gsx23, lsx+23, gsyz23, csy23a, csy23b, dsz23, dsxy23}, which converts the optimization problem of the attention mechanism in back-propagation into a regression problem known as {\it Softmax Regression}. This mathematical formulation enables us to analyze the behavior of the attention matrix in a more tractable manner. Specifically, we define Softmax Regression as follows:
\begin{definition}[Softmax regression: decomposition of attention matrix]\label{def:softmax_regression}
Let $A \in \mathbb{R}^{n \times d}$ be the input matrix and $b_f \in \mathbb{R}^d$ be the target vector. We introduce the weight vector $x \in \mathbb{R}^d$ to represent the parameters of softmax regression. The softmax regression function is defined as:
\begin{align*}
    f(A, x) := \langle \exp(Ax), {\bf 1}_n \rangle^{-1} \exp(Ax)
\end{align*}
where $n$ denotes the number of vectors in input matrix $A$, $d$ denotes the number of dimensions.
We define the loss of softmax regression as:
\begin{align*}
    L_f(A, x) = 0.5 \| f(A, x) - b_f \|_2^2
\end{align*}
\end{definition}

As Figure~\ref{fig:architect} shows, our contribution is including two parts
\begin{itemize}
    \item {\bf Attack.} We establish the guarantee of a successful attack by leveraging the information contained in the gradient and the parameters of softmax regression. Given an obtained gradient $b_g \in \R^d$ and weight $x \in \R^d$ of softmax regression, we transform the gradient attack to an optimization problem since our objective is to find the input matrix $A^* = \min_{A \in \R^{n \times d}} \| g(A, x) - b_g \|_2^2$, where $g(A, x) = \frac{\d L_f(A, x)}{\d x}$. We first provide strong results that confirm the Lipschitz-continuity and positive definiteness (PSD) of Hessian of $\| g(A, x) - b_g \|_2^2$. We employ a second-order method \cite{ans00, lsz19, cls21, dls23, dsxy23} to approximate a desirable $\wt{A} \in \R^{n \times d}$ that satisfying $\|\wt{A} - A^*\|_F\leq \epsilon$ within $O(\log(\|A_0 - A^*\|_F/\epsilon))$ times executions. It demonstrates the potential for successful attacks on the data privacy of transformer-based models by leveraging gradients in theoretical form.
    \item {\bf Defense.} We reconsider the effectiveness of gradient perturbation as a defense mechanism against gradient leakage. Given reasonable privacy cost settings, we propose that this method offers a balanced trade-off between preserving training convergence and providing effective protection for sensitive training data. Given a gradient $g(A, x) \in \R^d$, we introduce a noise perturbation $\xi \in \R^d$ that is sampled from the Laplace distribution, accordingly, the gradient for optimization becomes $\mathcal{M}(A, x) = g(A, x) + \xi$ (here $\mathcal{M}(A, x)$ is also the observed gradient $b_g$ in gradient leakage). We analyze the associated privacy costs within the differential privacy framework \cite{dwo06, gkn17, bps19, ks14, km11, kl10, mprv09, kov15, dwo08, gsy23, syyz23_dp}: given any neighboring batch data $A \in \R^{n \times d}$ and $A' \in \R^{n \times d}$, we always have $\frac{\Pr[\mathcal{M}(A)=b_g]}{\Pr[\mathcal{M}(A')=b_g]}\leq \exp(\varepsilon)$, where the $\varepsilon$ is the privacy cost. Additionally, we comprehensively examine the convergence properties of the Stochastic Gradient Descent (SGD) algorithm when perturbed gradients are utilized, subject to specific conditions.
\end{itemize}
Through these contributions, we aim to provide valuable insights into the attack and defense mechanism of gradient leakage in transformer-based models, with the ultimate goal of enhancing the understanding of this critical issue in training distributed models. By delving into the intricacies of gradient leakage, we aim to deepen awareness and knowledge surrounding the potential risks associated with this phenomenon. 

\section{Related Work}

In this section, we conduct a comprehensive review of the relevant literature pertaining to three key areas: Attention Computation Theory, AI Security and Optimization, and Convergence of Deep Neural Networks. These topics are closely interconnected with our work and provide essential background knowledge.

\paragraph{Attention Computation Theory.}
Following the rise of LLM, numerous studies have emerged on attention computation \cite{kkl20, tda+20, clp+21, zhdk23, tlto23, sht23, pmxa23, zkv+20, ag23, tbm+20, dls23, xgzc23, dgs23, kmz23, as23, bsz23, dms23, gms23, lsz23, llr23, ssz23, hjk+23, as23_tensor, gswy23}, revealing deeper nuances of attention models. \cite{dgtt23} investigates the potential optimization and generalization advantages of using multiple attention heads, they derive convergence and generalization guarantees for gradient-descent training of a single-layer multi-head self-attention model. \cite{lwd+23} proposes Deja Vu, a system that leverages contextual sparsity to reduce the computational cost of large language models (LLMs) at inference time. By identifying small, input-dependent sets of attention heads and MLP parameters that yield similar outputs as the dense model, they can speed up LLM inference without sacrificing model quality or in-context learning ability. Their low-cost algorithm predicts contextual sparsity on the fly, and their hardware-aware implementation further enhances performance. In \cite{gsx23}, they address regression problems in the context of in-context learning for large language models (LLMs). They formulate the regression problems using matrices and introduce a vectorization technique to expand the dimension. This approach is distinct from previous methods and allows for a comprehensive Lipschitz analysis.

\paragraph{AI Security.}
The rapid development and application of artificial intelligence (AI) bring the harm of abusing it, there are many works studying to overcome this challenge \cite{pzjy20, dwl+21, blm+22, kwr22, kgw+23, vkb23, csy23b, xza+23, gsy23, kgw+23, hxl+22, hxz+22, gsy23_coin, swx+23, dsxy23, buk23, bar23_review, lyt+23, zwkf23, bbb+23, rkv+23, rwhp23}. \cite{pzjy20} assesses the privacy risks of capturing sensitive data with eight models and introduces defensive strategies, balancing performance and privacy. \cite{blm+22} asserts that current methods fall short in guaranteeing comprehensive privacy for language models, recommending training on publicly intended text. \cite{kwr22} reveals that the vulnerability of large language models to privacy attacks is significantly tied to data duplication in training sets, emphasizing that deduplicating this data greatly boosts their resistance to such breaches. \cite{kgw+23} devised a way to watermark LLM output without compromising quality or accessing LLM internals. Meanwhile, \cite{vkb23} introduced near access-freeness (NAF), ensuring generative models, like transformers and image diffusion models, don't closely mimic copyrighted content by over $k$-bits.

\paragraph{Optimization and Convergence of Deep Neural Networks.}
 \cite{zsj+17,zsd17,ll18, dzps18, azls19a, azls19b, adh+19a, adh+19b, sy19, cgh+19, zmg19, cg19, zg19, os20, jt19, lss+20, hlsy21, zpd+20, bpsw20, zkv+20, szz21, als+22, mosw22, zha22, gms23, lsz23, qsy23, sy23} on the optimization and convergence of deep neural networks has been crucial in understanding their exceptional performance across various tasks. These studies have also contributed to enhancing the safety and efficiency of AI systems. In \cite{gms23} they define a neural function using an exponential activation function and apply the gradient descent algorithm to find optimal weights. In \cite{lsz23}, they focus on the exponential regression problem inspired by the attention mechanism in large language models. They address the non-convex nature of standard exponential regression by considering a regularization version that is convex. They propose an algorithm that leverages input sparsity to achieve efficient computation. The algorithm has a logarithmic number of iterations and requires nearly linear time per iteration, making use of the sparsity of the input matrix.

  %%% Section 1. Introduction
% \input{related_work}
% \input{preli}
\section{Background}

In this section, we provide introductions to the backgrounds of our research that form the foundation of our paper. Firstly, we propose using Softmax Regression to simplify the theoretical description of the transformer in Section~\ref{sub:softmax_regression}. Next, we introduce how to recover training data Deep Gradient Leakage from the obtained gradient in Section~\ref{sub:dlg}.

\subsection{Softmax Regression: Regression Problem Inspired by Attention Computation}\label{sub:softmax_regression}

The attention scheme \cite{wcb14, bcb14, vsp+17,rns+18,dclt19}, is a key component in transformer-based models. It is designed to map a sequence of input vectors, denoted as $X = [x_1, ..., x_n]$, $x_i \in \R^d$, $\forall i \in [n]$, to a corresponding sequence of output vectors, denoted as $Y = [y_1, ..., y_n]$, $y_i \in \R^d$, $\forall i \in [n]$. In each layer of the Transformer, a matrix $X_l$ is transformed into a new sequence $X^{(l+1)}$. Here, the self-attention function ${\sf Attention}: \R^{n \times d} \rightarrow \R^{n \times d}$ with sequence size $n$ and dimension $d$ parameterized by key, query, and value matrices $K \in \R^{d \times d}, Q \in \R^{d \times d}, V \in \R^{d \times d}$, computes as:
\begin{align*}
    {\sf Attention}(X_l) = D(X)^{-1} \exp (X_l Q K^\top X_l^\top) X_l V
\end{align*}
where $D(X) := \diag( \exp( X_l Q K^\top X_l^\top ) {\bf 1}_n)$. Then each row of $D^{-1} \exp()$ is a softmax.

Specifically, we examine the training procedure for a particular layer ($l$-th layer) and single-headed attention, where we have an input matrix denoted as $X_l$ and a target matrix denoted as $X_{l+1}$. Our objective is to minimize the loss function through back-propagation in the training process. We train 
a back-propagation step to minimize
\begin{align*}
    \| D(X)^{-1} \exp(X_l Q K^\top X_l^\top) X_l V - X_{l+1} \|_F^2.
\end{align*}
We first optimize matrix $V$ that 
\begin{align*}
    V = V - \eta \frac{\d }{\d V} \| D(X)^{-1} \exp (X_l Q K^\top X_l^\top) X_l V - X_{l+1} \|_F^2
\end{align*}, where $\eta$ is the learning rate. After that, our training objective becomes
\begin{align*}
    \| D(X)^{-1} \exp (X_l Q K^\top X_l^\top) - X_{l+1} (X_l V)^\dag \|_F^2
\end{align*}
where $^{\dag}$ is the Moore-penrose pseudo-inverse. We define $A := X_l \otimes X_{l+1}$. To further simplify the analysis, we flatten $X \in \R^{d \times d}$ by vectorization function ${\rm vec}(X)$. We have
\begin{align*}
    \| D(x)^{-1} \exp(Ax) - b \|_2^2
\end{align*}
where $x \in \R^{d^2}$ that $x = {\rm vec}(Q K^\top)$, $b \in \R^{d^2}$ that $b = {\rm vec}(X_{l+1} (X_l V)^\dag)$, and $D(x) = D(X) \otimes I_n$. 

\subsection{Deep Gradient Leakage}\label{sub:dlg}

In the context of gradient leakage attacks, described in the work by \cite{zlh19}, the goal is to recover the training data by leveraging the observed gradient. This can be formulated as an optimization problem using the following notation: Let $g$ represent the observed gradient, $f$ denotes the model function with model weight $\theta$, and $\ell$ represents the loss function. The input is denoted as $x$ and the corresponding label as $y$. Our objective is to minimize the loss function:
\begin{align*}
    L(x, y) = \| \frac{\d \ell(f(x), y)}{\d \theta} - g \|_2^2
\end{align*}
To approximate the training data, we initialize $x_0$ and $y_0$, and iteratively optimize them using the following update rules at step t:
\begin{align*}
    x_{t+1} = x_t - \eta \frac{\d L(x_t, y_t)}{\d x_t} \\
    y_{t+1} = y_t - \eta \frac{\d L(x_t, y_t)}{\d y_t}
\end{align*}
Here, $\eta$ represents the learning rate, and the derivatives are computed with respect to $x_t$ and $y_t$.

By iteratively updating $x$ and $y$ using these optimization rules, gradient leakage aims to approximate the original training data. This process allows us to recover meaningful information about the data that was used to train the model, which highlights the potential vulnerabilities of gradient exchange.

\section{Preliminary}

In this section, we present the preliminary information. We commence by introducing the notations used, which are further elaborated upon in Section~\ref{sub:notations}. Subsequently, we provide formal definitions of attack and defense mechanisms in the context of gradient leakage in Section~\ref{sub:problem_defs}.

\subsection{Notations}\label{sub:notations}

We used $\R$ to denote real numbers. We use $A \in \R^{n \times d}$ to denote an $n \times d$ size matrix where each entry is a real number. For any positive integer $n$, we use $[n]$ to denote $\{1,2,\cdots, n\}$. For a matrix $A \in \R^{n \times d}$, we use $a_{i,j}$ to denote the an entry of $A$ which is in $i$-th row and $j$-th column of $A$, for each $i \in [n]$, $j \in [d]$. We use $A_{i,j} \in \R^{n \times d}$ to denote a matrix such that all of its entries equal to $0$ except for $a_{i,j}$. We use ${\bf 1}_n$ to denote a length-$n$ vector where all the entries are ones. For a vector $w \in \R^n$, we use $\diag(w) \in \R^{n \times n}$ denote a diagonal matrix where $(\diag(w))_{i,i} = w_i$ and all other off-diagonal entries are zero. Let $D \in \R^{n \times n}$ be a diagonal matrix, we use $D^{-1} \in \R^{n \times n}$ to denote a diagonal matrix where $i$-th entry on the diagonal is $D_{i,i}$ and all the off-diagonal entries are zero. Given two vectors $a,b \in \R^n$, we use $(a \circ b) \in \R^n$ to denote the length-$n$ vector where $i$-th entry is $a_i b_i$. For a matrix $A \in \R^{n \times d}$, we use $A^\top \in \R^{d \times n}$ to denote the transpose of matrix $A$. For a vector $x \in \R^n$, we use $\exp(x) \in \R^n$ to denote a length-$n$ vector where $\exp(x)_i = \exp(x_i)$ for all $i \in [n]$. For a matrix $X \in \R^{n \times n}$, we use $\exp(X) \in \R^{n \times n}$ to denote matrix where $\exp(X)_{i,j} = \exp(X_{i,j})$. For any matrix $A \in \R^{n \times d}$, we define $\| A \|_F := ( \sum_{i=1}^n \sum_{j=1}^d A_{i,j}^2 )^{1/2}$. For a vector $a, b \in \R^n$, we use $\langle a, b \rangle$ to denote $\sum_{i=1}^n a_i b_i$.

\subsection{Problem Definitions}\label{sub:problem_defs}

Now we formally provide the definitions of the problem we aim to solve in this paper. It can briefly be divided into the following two parts:

\paragraph{Attack.}
Following \cite{zlh19}, the gradient leakage attack can be transformed into an optimization problem, here we provide our definition of the attack problem below.
\begin{definition}[Training objective of attack]\label{def:attack}
    Given the observed gradient $b_g$ and public weight of softmax regression $x \in \R^d$. Suppose given input matrix $A \in \R^{n \times d}$ and target vector $b \in \R^d$, we denote the truth gradient computation as:
    \begin{align*}
        g(A, x) = \frac{\d L_f(A, x)}{\d x}.
    \end{align*}
    Since we have the public shared gradient, denote as $b_g \in \R^d$, we define our training objective
    \begin{align*}
        L_g(A, x) = 0.5 \| g(A, x) - b_g \|_2^2.
    \end{align*}
    We consider the following problem:
    $
        \min_{A \in \R^{n \times d}} L_g(A, x)
    $.
\end{definition}

\paragraph{Defense.}
Following \cite{acg+16, zlh19}, we aim to find a defense method that protects data from gradient leakage. We formulize our goal into two key problems: 1) Differential privacy, 2) Minimal training effect. We provide the formal definitions as follows:
\begin{definition}[Differential privacy of noise perturbation]
    Since we have the public shared gradient, denote as $b_g \in \R^d$. Given two neighboring batch data $A \in \R^{n \times d}$ and $A' \in \R^{n \times d}$. Denote $\varepsilon$ as the privacy cost. If the protected gradient $\mathcal{M}(A)$ satisfies
    \begin{align*}
        \frac{\Pr[\mathcal{M}(A) = b_g]}{\Pr[\mathcal{M}(A') = b_g]} \leq \exp(\varepsilon)
    \end{align*}
    then $\mathcal{M}(A)$ is said to provide $\varepsilon$-differential privacy.
\end{definition}

\begin{definition}[Convergence under protected gradient]
    Given a training dataset $\mathcal{D} = \{ (A_i, b_i), \forall i \in [N] \}$, where $N$ is the size of dataset. At step $t$, we run SGD algorithm with learning rate $\eta$ to optimize $x$ as follows:
    \begin{align*}
        x_{t+1} = x_t - \eta \mathcal{M}(A_t)
    \end{align*}
    where $\mathcal{M}(A)$ is the protected gradient.

    For any training error $\epsilon > 0$, there exists a positive integer $T$ that
    \begin{align*}
        \min_{t \in [T]}\{ \| \nabla L_f(x_t) \|_2^2 \} \leq \epsilon
    \end{align*}
    where $L_f(x_t) = \frac{1}{N} \sum_{i=1}^N L_f(A_i, x_t)$, then we said the training is convergent, 
\end{definition}

\section{Recover Data from Gradient}

In this section, we present our findings regarding the use of a second-order method \cite{ans00, lsz19, cls21, dls23, dsxy23} to recover training data from observed gradients. We provide compelling evidence to support our results, starting with the confirmation of the Lipschitz continuity of the Hessian of the training objective $L_g(A, x)$ in Section~\ref{sub:lipschitz}. Additionally, we demonstrate another important result, which establishes the positive definiteness of the Hessian of the training objective $L_g(A, x)$ in Section~\ref{sub:psd}. In Section~\ref{sub:main_result}, we present the main outcome of our research, which is a convergence guarantee that ensures the generation of pixel-wise accurate images and token-wise matching texts in the training data. 

We provide our definition of Hessian below.
\begin{definition}\label{def:hessian_for_A}
    Given the obtained gradient $b_g \in \R^d$ and public weight of softmax regression $x \in \R^d$. We initialize $A \in \R^{n \times d}$, let $L_g(A, x)$ be defined as Definition~\ref{def:attack}, we define Hessian matrix of $L_g(A, x)$ as follows:
    \begin{align*}
        H(A) := \nabla^2_A L_g(A, x)
    \end{align*}
\end{definition}

Please refer to Lemma~\ref{lem:hessian_main_result} for the computation result of the Hessian matrix of $L_g(A, x)$.

\subsection{Hessian is Lipschitz-continuous}\label{sub:lipschitz}

We present our findings that establish the Lipschitz continuity property of the Hessian of $L_g(A, x)$,
which is a highly desirable characteristic in optimization. This property signifies that the second
derivatives of $L_g(A, x)$ exhibit smooth changes within a defined range. Leveraging this Lipschitz property enables us to employ gradient-based methods with guaranteed convergence rates and enhanced
stability. Consequently, our results validate the feasibility of utilizing the proposed training objective to achieve convergence in the gradient leakage attack. We show our result below.

\begin{lemma}[Hessian is Lipschitz-continuous, informal version of Lemma~\ref{lem:lipschitz:formal}]\label{lem:lipschitz:informal}
Given the obtained gradient $b_g \in \R^d$ and public weight of softmax regression $x \in \R^d$. Let $H(A)$ be defined as Definition~\ref{def:hessian_for_A}. 
Let $M = \poly(n,d) \exp(O(R^2))$, where we denote $R > 0$ as a scalar that $\|A\|_F \leq R$ and $\|\wt{A}\|_F\leq R$ for all $A \in \R^{n \times d}$ and $\wt{A} \in \R^{n \times d} $. We have
\begin{align*}
    \| H(A) - H(\wt{A}) \| \leq M \cdot \| A - \wt{A} \|_F
\end{align*}
\end{lemma}

Please refer to Appendix~\ref{app:lipschitz_hessian} for the proof of Lemma~\ref{lem:lipschitz:informal}.

\subsection{Hessian is Positive Definite}\label{sub:psd}

After computing the Hessian of $L_g(A, x)$, we now show our result confirming it is positive definite under proper regularization. Therefore, we can apply a modified Newton’s method to approach the optimal solution.

We introduce an additional regularization term on $L_g(A, x)$ below.
\begin{definition}[Regularization]\label{def:L_reg}
    Given a scalar $\tau > \frac{n^2 \sqrt{d} R^5}{2}$, we define an additional regularization term as follows:
    $
        L_{\rm reg}(A) = \tau \| {\rm vec}(A) \|_2^2
    $.
\end{definition}

\begin{lemma}[Hessian is positive definite, 
informal version of Lemma~\ref{lem:psd:formal}]\label{lem:psd:informal}
Let $L_g(A, x)$ be defined as Definition~\ref{def:attack}, by adding a regularization term $L_{\rm reg}(A)$ as Definition~\ref{def:L_reg}, we denote our training objective as $L(A) = L_g(A, x) + L_{\rm reg}(A)$. Then, we can show that $\nabla^2 L(A) \succ l \cdot I_{nd}$. Here $I_{nd}$ is a size $nd \times nd$ identity matrix and $l := -n^2 \sqrt{d} R^5 + 2\tau$.
\end{lemma}

The positive definiteness of the Hessian matrix of the training objective not only confirms the convexity of our training objective but also satisfies a crucial condition for the application of Newton's method, which guarantees fast convergence. For the proof of Lemma~\ref{lem:psd:informal}, demonstrating the positive semidefiniteness of the Hessian matrix, please refer to Appendix~\ref{app:psd_hessian}.

\subsection{Main Result}\label{sub:main_result}

Since we have Lipschitz property and positive definite property for the hessian of the training objective, we now provide our main result that guarantees the successful attack to recover training data from gradients, where we apply a modified Newton's method to the $L_g(A, x)$, and can recover training data within $T = O(\log(\|A_0 - A^*\|_F / \epsilon))$ executions.

\begin{theorem}[Main result, informal version of Theorem~\ref{thm:main_result:formal}]\label{thm:main_result:informal}
Given an observed gradient vector $b_g \in \R^d$. We let $L_g(A, x)$ be defined as Definition~\ref{def:attack}. We denote the optimal solution for $L_g(A, x)$ as $A^*$.

Next, we choose a good initial point $A_0$ that is close enough to $A^*$. Assume that there exists a scalar $R > 1$ such that $\|A_t\| \leq F$ for any $t \in [T]$.

Then, for any accuracy parameter $\epsilon \in (0,0.1)$ and a failure probability $\delta \in (0,0.1)$, an algorithm based on the Newton method can be employed to recover the initial data. The result of this algorithm guarantee within $T = O(\log(\|A_0 - A^*\|_F / \epsilon))$ executions, it outputs a matrix $\Tilde{A} \in \mathbb{R}^{n \times d}$ satisfying $\|\wt{A} - A^*\|_F \leq \epsilon$ with a probability of at least $1 - \delta$.
\end{theorem}

To the best of our understanding, our principal result, outlined in Theorem~\ref{thm:main_result:informal}, provides a comprehensive explanation of why gradient leakage can facilitate the easy recovery of sensitive training data. Furthermore, it offers substantial evidence supporting the correlation between the success of an attack and the specific conditions of the training data. Please see Appendix~\ref{app:main_result} for the proof of Theorem~\ref{thm:main_result:informal}.

\begin{figure*}
    \centering
    \includegraphics[width=\textwidth]{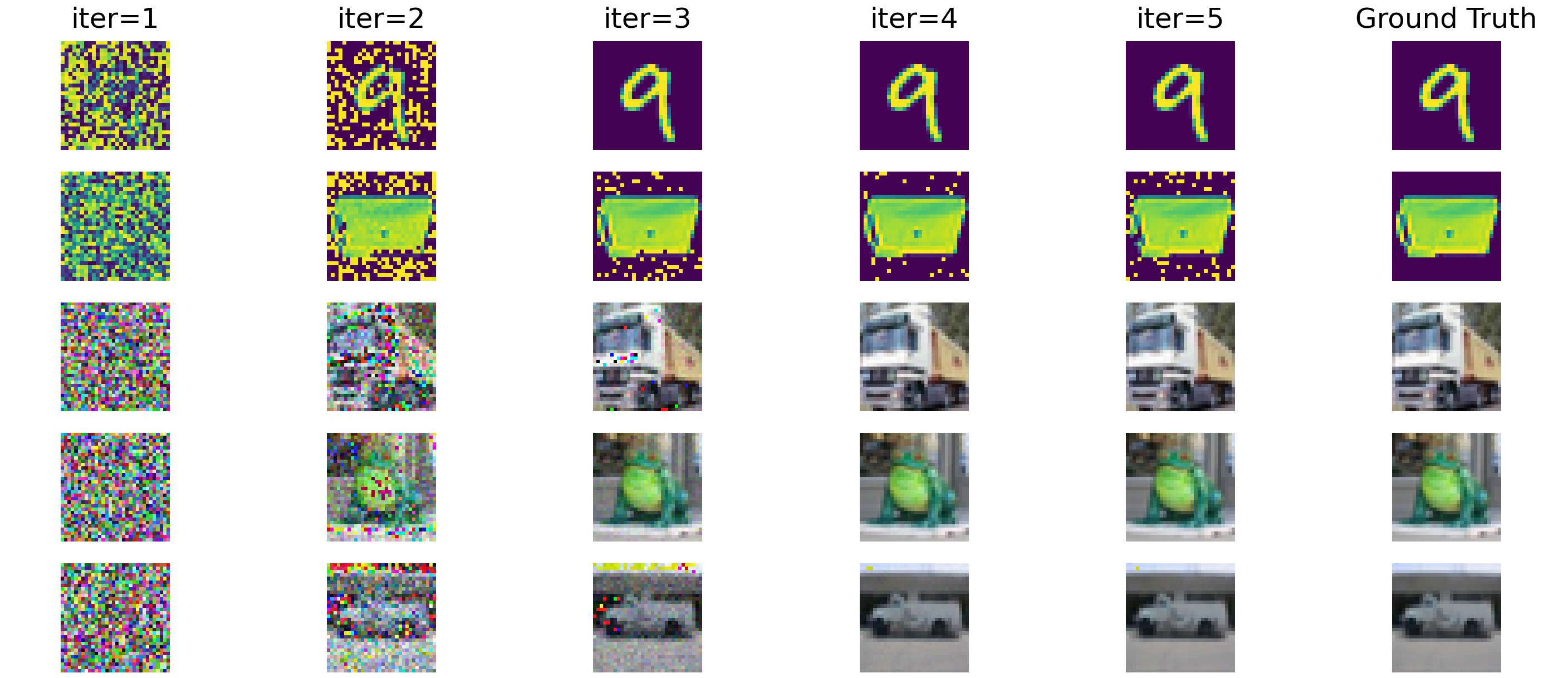}
    \caption{The visualization showing the deep leakage on images from MNIST, FMNIST, and Cifar-10 on our ViT model.}
    \label{fig:vit_recovery}
\end{figure*}

\section{Laplacian Perturbation Protection}

In this section, we revisit the strategy of implementing an additional layer of noisy perturbation as a means of protecting gradients from potential leakage. We commence with a thorough explanation of Laplacian noise in Section~\ref{sub:laplacian_noise}. Following this, in Section~\ref{sub:diff_privacy}, we engage in an exploration of the correlation between the privacy cost and the variance of the sampled Laplace distribution. It is essential to ensure that the adopted safeguarding mechanism does not detrimentally affect the convergence of the training process. Therefore, in Section~\ref{sub:convergence_thm}, we present our key finding, which assures that the training process can continue to converge dependably even when gradients are perturbed.

\subsection{Laplacian Noise}\label{sub:laplacian_noise}

We first provide the definition of the additional Laplacian distribution noise vector below.
\begin{definition}[Laplacian distribution noise vector]\label{def:laplace_noise:informal}
    We say a noise $\xi \in \R^{d}$ vector is sampled from ${\rm Laplace}(\mu, b)$, where the $i$-th element of $\xi_i$ that satisfies 
    \begin{align*}
        \Pr[\xi_i=x] = \frac{1}{2b} \exp( - \frac{| x - \mu |}{b})
    \end{align*}
\end{definition}

Hence, we show our definition of perturbed gradient as follows:

\begin{definition}[Perturbed gradient]\label{def:M:informal}
    Let $g(A, x) \in \R^d$ be defined as Definition~\ref{def:attack}, we sample $\xi$ from ${\rm Laplace}(\mu, b)$ as Definition~\ref{def:laplace_noise:informal}, we define
    \begin{align*}
        \mathcal{M}(A) := g(A, x) + \xi
    \end{align*}
\end{definition}

\subsection{Differential Privacy}\label{sub:diff_privacy}

Next, we present the computation result of the $\ell_1$ sensitivity of $g(A, x)$, which serves as a quantitative measure of the susceptibility of gradients to potential abuse and successful attacks.

\begin{lemma}[$\ell_1$ sensitivity, informal version of Lemma~\ref{lem:ell_sensitivity:formal}]\label{lem:ell_sensitivity:informal}
    Let $g(A, x)$ be defined as Definition~\ref{def:attack}. Given $A \in \R^{n \times d}$ and $A' \in \R^{n \times d}$ are neighbouring batch data. We have $\Delta$ as the $\ell_1$ sensitivity of $g(A, x)$:
    \begin{align*}
        \Delta = & ~ \max_{A \in \R^{n \times d}, A' \in \R^{n \times d}} \| g(A, x) - g(A', x) \|_1 \\
        \leq & ~ 20 \beta^{-2} n^{1.5} d \exp(10R^2) 
    \end{align*}
    where $R \geq 4$ and $\beta \in (0, 0.1)$.
\end{lemma}

We now present our findings that provide evidence of the differential privacy achieved through the perturbation of gradients. These results indicate that attackers are unable to recover the training data through the utilization of differential methods.

\begin{theorem}[Differential privacy, informal version of Theorem~\ref{thm:differential_privacy:formal}]\label{thm:differential_privacy:informal}
    We denote $\varepsilon > 0$ as our privacy cost, let $\Delta := 20 \beta^{-2} n^{1.5} d \exp(10R^2)$ be denoted as Lemma~\ref{lem:ell_sensitivity:informal}, where $R \geq 4$ and $\beta \in (0, 0.1)$. We sample $\xi \in \R^d$ from ${\rm Laplace}(0, \frac{\Delta}{\varepsilon})$ as Definition~\ref{def:laplace_noise:informal}. Let perturbed gradient be defined as $\mathcal{M}(A) \in \R^d$ as Definition~\ref{def:M:informal}, denote $b_g \in \R^d$ as the observed gradient. Suppose given $A \in \R^{n \times d}$ and $A' \in \R^{n \times d}$ are neighboring batch data. We have
    \begin{align*}
        \frac{\Pr[\mathcal{M}(A) = b_g]}{\Pr[\mathcal{M}(A') = b_g]} \leq \exp(\varepsilon)
    \end{align*}
\end{theorem}

Please refer to Appendix~\ref{app:diff_privacy} for the proof of Lemma~\ref{lem:ell_sensitivity:informal} and Theorem~\ref{thm:differential_privacy:informal}.

\subsection{Convergence Theorem}\label{sub:convergence_thm}

After we introduce Laplacian noise to the gradients. However, this action raises a critical question: can we still ensure the convergence of our training process under these conditions of perturbed gradients? Theorem~\ref{thm:training_convergence:informal} answers this question affirmatively. This theorem is significant as it not only upholds the robustness of our approach but also ensures its practicality. 

\begin{theorem}[Convergence under perturbed gradients, informal version of Theorem~\ref{thm:training_convergence:formal}]\label{thm:training_convergence:informal}
    Suppose for all $A_i \in \R^{n \times d}$ and $b_i \in \R^d$ in training dataset $\mathcal{D}$ satisfies $\| A_i \|_F \leq R$ and $\|b_i \|_1 \leq 1$ for $i \in |\mathcal{D}|$, where $|\mathcal{D}|$ is the size of $\mathcal{D}$. Denote $\beta \in (0, 0.1)$ and $R \geq 4$. We define $L(x) = 0.5 \sum_{i=1}^{|\mathcal{D}|} \|f(A_i, x) - b_i\|_2^2$ as the sum of $L_f(A, x)$ that is defined in Definition~\ref{def:softmax_regression}. Let $\mathcal{M}(x)$ be defined as Definition~\ref{def:M:informal}. Denote $\gamma := 7R\beta^{-2}n^{1.5}\exp(3R^2)$. Denote $\Delta := 20 \beta^{-2} n^{1.5} d \exp(10R^2)$. Denote $\epsilon > 0$ as the training error. We run SGD algorithm with learning rate $\eta = \sqrt{\frac{L(x_1) - L(x^*)}{T(8\gamma R^2 + d \gamma (\frac{\Delta}{\varepsilon})^2)}}$, at most 
    \begin{align*}
        T = \frac{4 (L(x_1) - L(x^*)) \cdot (8\gamma R^2 + d \gamma (\frac{\Delta}{\varepsilon})^2)}{\epsilon^2}
    \end{align*}
    iterations, we have
    \begin{align*}
        \min_t \{ (\nabla L(x_t))^2 \} \leq \epsilon
    \end{align*}
\end{theorem}

As per Theorem~\ref{thm:training_convergence:informal}, it becomes evident that the $\ell_1$ sensitivity $\Delta$ and the privacy cost $\varepsilon$ are the key variables influencing the speed of training convergence. However, it's important to note that there is a trade-off between protecting data from gradient leakage and achieving fast training convergence. A lower privacy cost $\epsilon$ results in better privacy protection but at the expense of slower convergence speed. Theorem~\ref{thm:training_convergence:informal} elucidates this relationship between convergence and protection, and our results in Section~\ref{sub:impact_noise_perturbation} further substantiate this point. In essence, achieving optimal privacy protection and fast training convergence simultaneously remains a challenge, and understanding this trade-off is crucial for effective model training in privacy-sensitive scenarios.

For the proof of Theorem~\ref{thm:training_convergence:informal}, please see Appendix~\ref{app:convergence}.

\section{Experiments}

We conduct a series of experiments to rigorously validate our theory. Specifically, we employ a ViT model and evaluate its performance using the MNIST, FMNIST, and Cifar-10 datasets. Consistent with our theoretical analysis, we divide the experiments into two main components: attack and defense. In Section~\ref{sub:setup}, we outline the experimental setup utilized in this study. To assess the vulnerability of the model to gradient leakage, we present the results of the attack in Section~\ref{sub:recover_cifar}, which aligns with the findings reported in \cite{zlh19}. Furthermore, in Section~\ref{sub:impact_noise_perturbation}, we present the results of our defense mechanism against gradient leakage, where we introduce Laplacian noise. We also demonstrate the convergence of the training process in the presence of perturbed noisy gradients.

\subsection{Setup}\label{sub:setup}

\paragraph{Model.} Primarily, we employ the Vision Transformer (ViT) \cite{dbk+20} for our theoretical evaluation. ViT is an innovative application of the pretrained transformer architecture specifically applied to computer vision tasks. We load the weights of the model from the Huggingface community \cite{wds+19}, where we set ${\rm hidden\_size} = 768$, ${\rm num\_attention\_heads} = 12$, ${\rm num\_channels} = 3$, ${\rm num\_hidden\_layers} = 12$, ${\rm patch\_size} = 16$.

\paragraph{Datasets.} 
Our methods are evaluated using three distinct datasets: MNIST \cite{lcb09}, Fashion-MNIST \cite{xrv17}, and Cifar-10 \cite{kh09}. These datasets are widely used in the field of computer vision and provide a comprehensive basis for our evaluation.

\paragraph{Platform and Device.} 
For our deep learning experiments, we utilize the PyTorch library \cite{pgm+19} for training our models. All of our experiments are conducted on an RTX 3090 GPU device.

\subsection{Recover Cifar-10 from Gradient of ViT}\label{sub:recover_cifar}

In our experiment, we apply the Deep Gradient Leakage method to the ViT model, using image data from the MNIST, Fashion-MNIST, and Cifar-10 datasets. As illustrated in Figure~\ref{fig:vit_recovery}, we found that gradient leakage could effectively recover pixel-level images within a mere 5 optimization steps. These results align with the experimental findings reported in \cite{zlh19}, thereby reinforcing the efficacy and potential risks of gradient leakage in transformer-based models. Interestingly, when compared to the results in \cite{zlh19}, we noted that the ViT model required fewer iterations to recover data from the gradient compared to other CNN models such as LeNet. This observation suggests that transformer-based visual models might be more susceptible to gradient leakage attacks than their CNN counterparts. However, a comprehensive exploration of this potential vulnerability is beyond the scope of the current study and is earmarked for future research.

\subsection{Impact of Noise Perturbation}\label{sub:impact_noise_perturbation}

In this experiment, we revisited the impact of noise perturbation on both the protection of training data from gradient leakage and the convergence of training. We adopted the term "scale" to denote the 'b' parameter in the Laplace distribution, ${\rm Laplace}(\mu, b)$, with $\mu$ set to 0. For instance, when we set "scale = 0.001", it implies that we are sampling noise, denoted as $\xi$, from $\mathrm{Laplace}(0, 0.001)$ to add to the gradients. A "scale" of 0 signifies no noise addition to the gradients.

\ifdefined\isarxiv
\begin{figure}[!ht]
\centering
    \includegraphics[width=0.8\textwidth]{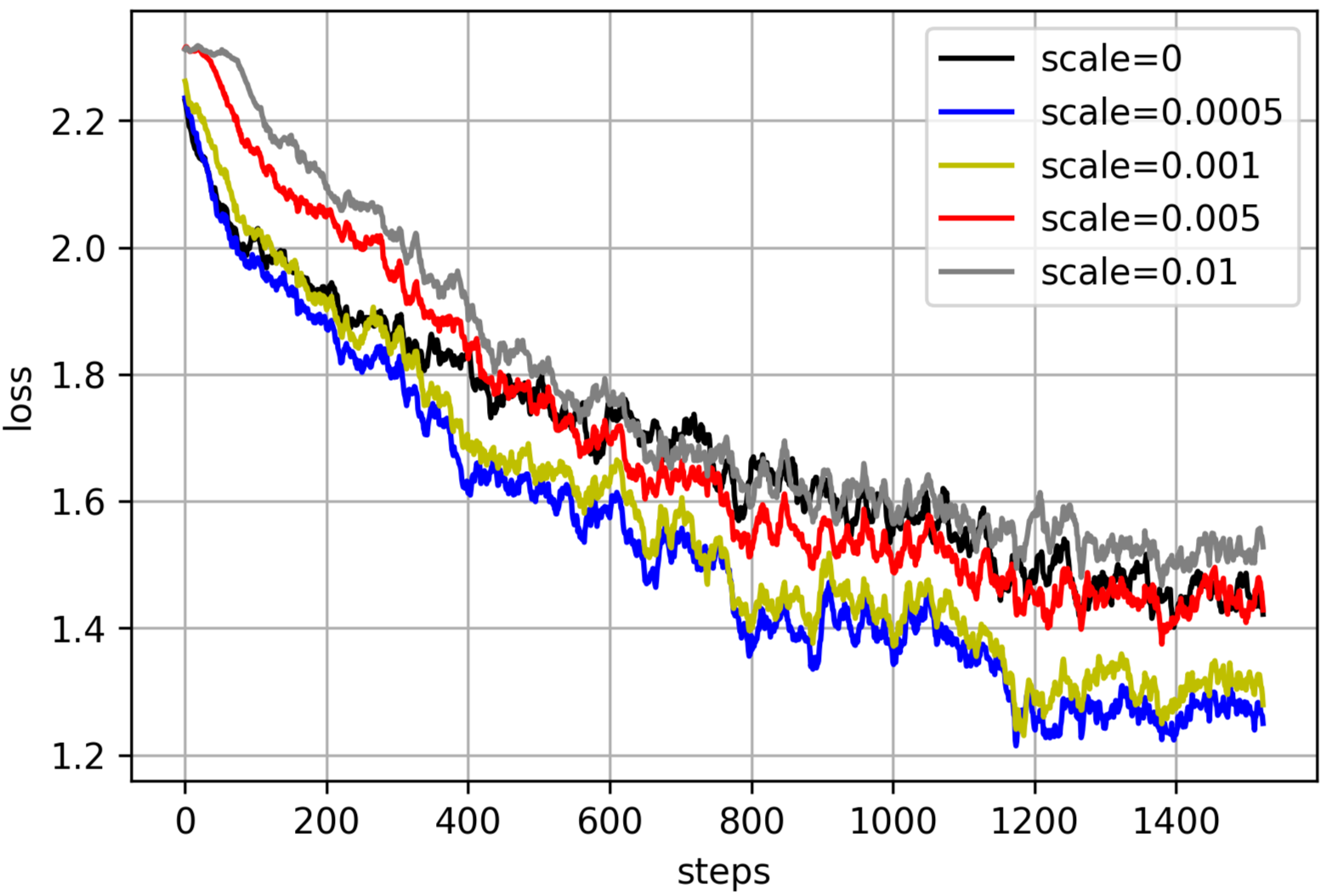}
    \caption{The loss of ViT training on Cifar-10 dataset under perturbed gradient. }
    \label{fig:noisy_training}
\end{figure}
\else
\begin{figure}[!ht]
    \includegraphics[width=0.46\textwidth]{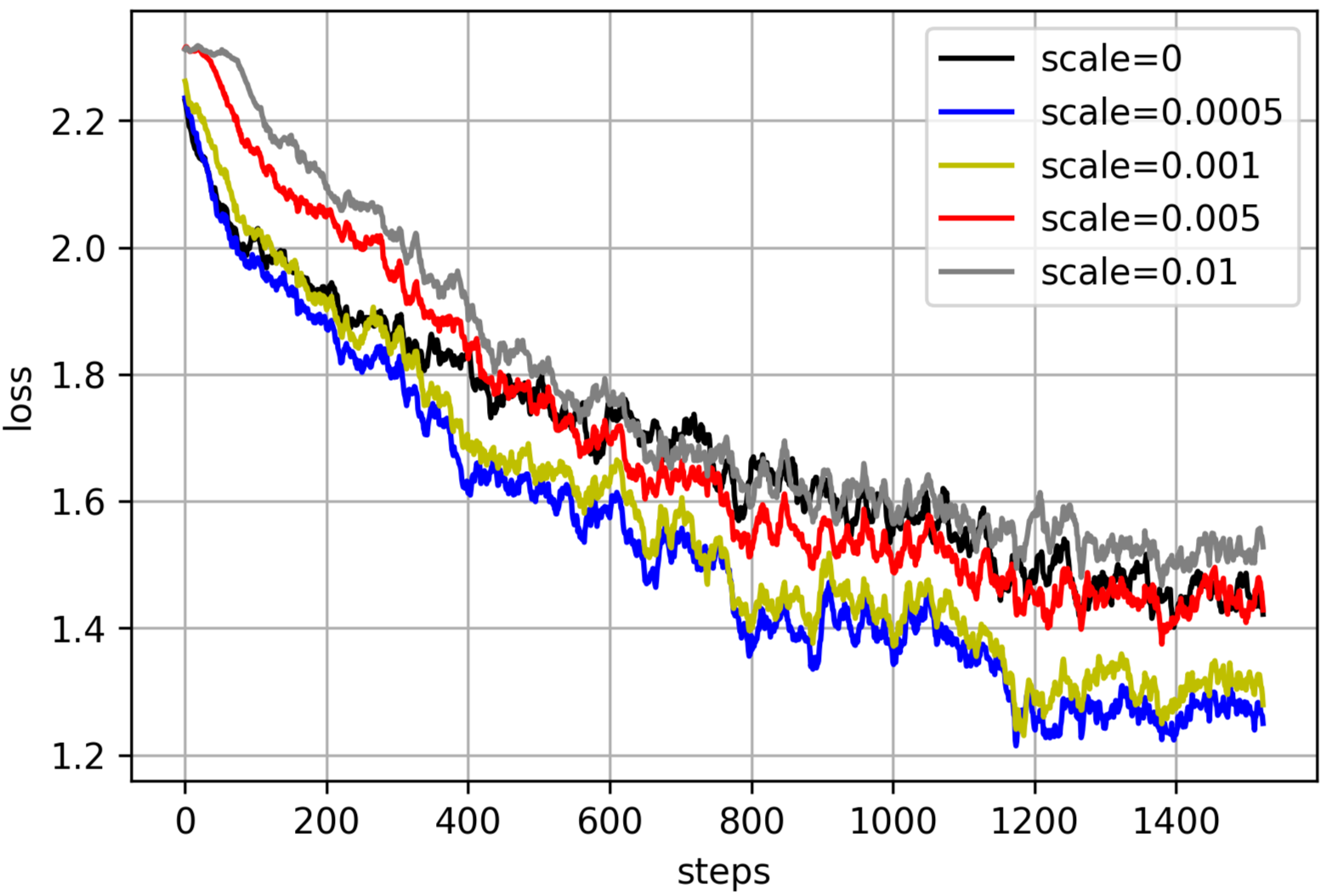}
    \caption{The loss of ViT training on Cifar-10 dataset under perturbed gradient. }
    \label{fig:noisy_training}
\end{figure}
\fi

As illustrated in Figure~\ref{fig:protection}, when the "scale" is set to or exceeds 0.001, the noise perturbation method effectively safeguards the image data (the "scale" threshold for data protection may vary depending on the model architecture). Figure~\ref{fig:noisy_training} displays the loss reduction process during the initial 1475 steps of fine-tuning the pretrained ViT model on the Cifar-10 dataset. The results confirm that when the "scale" is set to or exceeds 0.005, the noise protection does impact the speed of training convergence. This observation aligns with our theoretical results on the convergence under perturbed gradients (Theorem~\ref{thm:training_convergence:informal}).

\ifdefined\isarxiv
\begin{figure}[!ht]
\centering
    \includegraphics[width=0.8\textwidth]{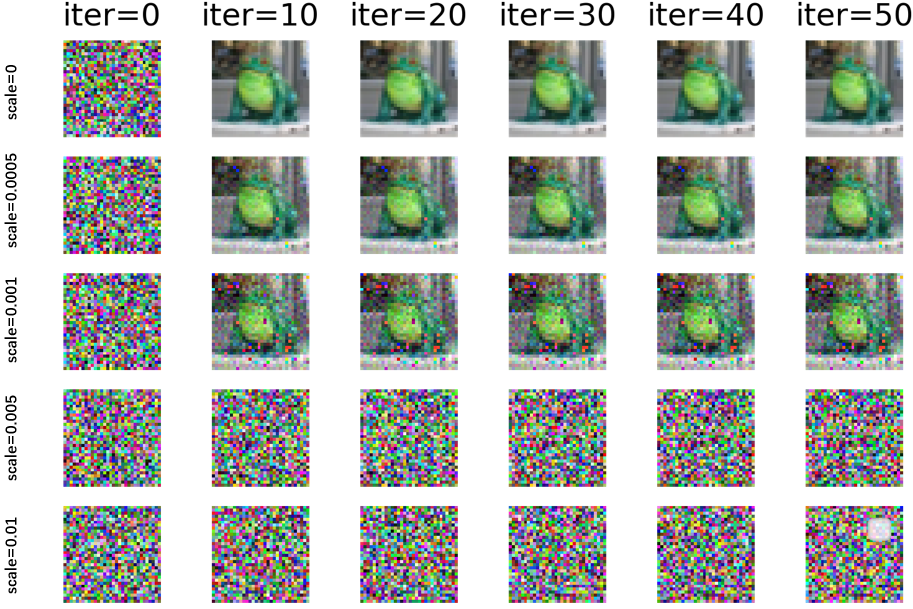}
    \caption{The protection of noise perturbation with differential "scale" of Laplace distribution. }
    \label{fig:protection}
\end{figure}
\else
\begin{figure}[!ht]
    \includegraphics[width=0.46\textwidth]{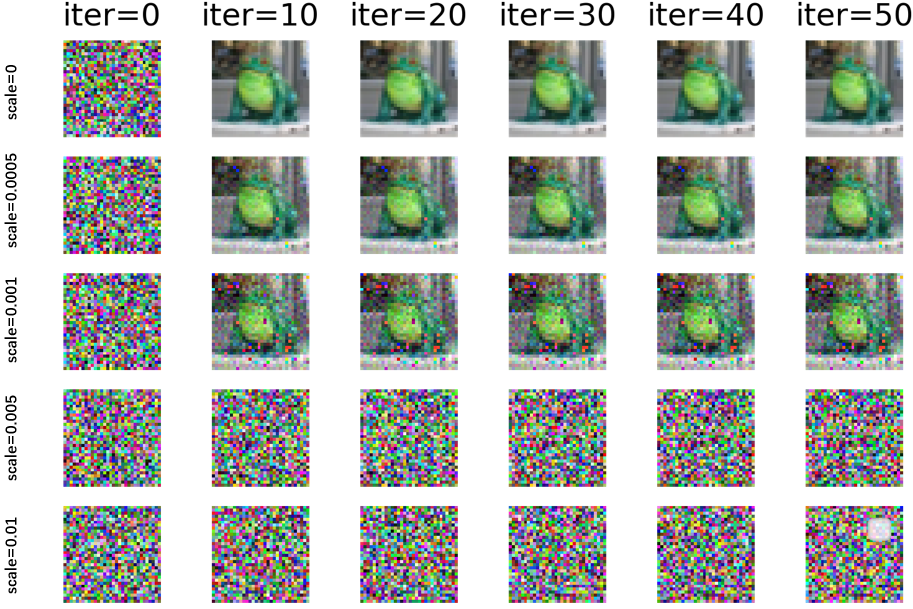}
    \caption{The protection of noise perturbation with differential "scale" of Laplace distribution. }
    \label{fig:protection}
\end{figure}
\fi

\section{Conclusion}\label{sec:conclusion}

In this paper, we present a comprehensive analysis of the attack and defense mechanisms related to gradient leakage in transformer-based models. However, we highlight the potential risk of privacy leakage in practical implementations, particularly through a method known as Deep Leakage from Gradient (DLG). Our novel approach involves the transformation of the optimization problem of the attention mechanism in back-propagation into a regression problem, termed as Softmax Regression. This enable a more tractable analysis of the behavior of the attention matrix. We establish the theoretical possibility of a successful attack by leveraging the information contained in the gradient and the parameters of softmax regression. Furthermore, we reconsider the effectiveness of gradient perturbation as a defense mechanism against gradient leakage, providing a balanced trade-off between preserving training convergence and protecting sensitive training data. Through our research, we provide a comprehensive understanding of the attack and defense mechanisms associated with gradient leakage in transformer-based models. This understanding is crucial for enhancing the security and privacy of distributed models, particularly in the context of federally distributed learning.

\ifdefined\isarxiv
%\section*{Acknowledgments}
% \bibliographystyle{alpha}
% \bibliography{ref}
\else
\bibliography{ref}
\bibliographystyle{ieeenat_fullname}%{alpha}

\fi

\newpage
\onecolumn
\appendix
\section*{Appendix}
\ifdefined\isarxiv

\else
{\hypersetup{linkcolor=black}
\tableofcontents
}
\fi

{\bf Rodmap.}
In Section~\ref{app:preli}, we define some basic math-related stuff for the following proof. In Section~\ref{app:gradient}, we compute the gradient of basic functions. In Section~\ref{app:lips_gradient}, we prove the lipschitz property for gradient function. In Section~\ref{app:upper_bound_basic},
we compute the upper bound of basic functions. In Section~\ref{app:lips_basic}
, we prove the lipschitz property for some basic functions. In Section~\ref{app:hessian_main}, we present the main results related to Hessian and split it into seven parts. In Section~\ref{app:more_gradients}, we compute more gradients for the following proof. In Section~\ref{app:hessian_first}, we prove the first term of Hessian is Lipschitz and PSD. Section~\ref{app:hessian_second}, we prove the second term of Hessian is Lipschitz and PSD. Section~\ref{app:hessian_third}, we prove the third term of Hessian is Lipschitz and PSD. Section~\ref{app:hessian_fouth}, we prove the fouth term of Hessian is Lipschitz and PSD. Section~\ref{app:hessian_fifth}, we prove the fifth term of Hessian is Lipschitz and PSD. Section~\ref{app:hessian_sixth}, we prove the sixth term of Hessian is Lipschitz and PSD. Section~\ref{app:hessian_seventh}, we prove the seventh term of Hessian is Lipschitz and PSD. In Section~\ref{app:lipschitz_hessian}, we prove the lipschitz property of Hessian. 
In Section~\ref{app:psd_hessian}, we prove that the Hessian is PSD. In Section~\ref{app:main_result}, we introduce the approximate newton method. In Section~\ref{app:diff_privacy}, we show the differential privacy of laplacian perturbation. In Section~\ref{app:convergence}, we prove the convergence of the perturbed gradient. In Section~\ref{app:core_tool}, we present a core tool for the convergence proof.

\section{Preliminary}\label{app:preli}

\subsection{Notations}

We used $\R$ to denote real numbers. We use $A \in \R^{n \times d}$ to denote an $n \times d$ size matrix where each entry is a real number. For any positive integer $n$, we use $[n]$ to denote $\{1,2,\cdots, n\}$. For a matrix $A \in \R^{n \times d}$, we use $a_{i,j}$ to denote the an entry of $A$ which is in $i$-th row and $j$-th column of $A$, for each $i \in [n]$, $j \in [d]$. We use $A_{i,j} \in \R^{n \times d}$ to denote a matrix such that all of its entries equal to $0$ except for $a_{i,j}$. We use ${\bf 1}_n$ to denote a length-$n$ vector where all the entries are ones. For a vector $w \in \R^n$, we use $\diag(w) \in \R^{n \times n}$ denote a diagonal matrix where $(\diag(w))_{i,i} = w_i$ and all other off-diagonal entries are zero. Let $D \in \R^{n \times n}$ be a diagonal matrix, we use $D^{-1} \in \R^{n \times n}$ to denote a diagonal matrix where $i$-th entry on diagonal is $D_{i,i}$ and all the off-diagonal entries are zero. Given two vectors $a,b \in \R^n$, we use $(a \circ b) \in \R^n$ to denote the length-$n$ vector where $i$-th entry is $a_i b_i$. For a matrix $A \in \R^{n \times d}$, we use $A^\top \in \R^{d \times n}$ to denote the transpose of matrix $A$. For a vector $x \in \R^n$, we use $\exp(x) \in \R^n$ to denote a length-$n$ vector where $\exp(x)_i = \exp(x_i)$ for all $i \in [n]$. For a matrix $X \in \R^{n \times n}$, we use $\exp(X) \in \R^{n \times n}$ to denote matrix where $\exp(X)_{i,j} = \exp(X_{i,j})$. For any matrix $A \in \R^{n \times d}$, we define frobenius norm $\| A \|_F := ( \sum_{i=1}^n \sum_{j=1}^d A_{i,j}^2 )^{1/2}$, and we define spectral norm $\| A\| := \max_{x\in R^d} \|A x\|_2 / \| x\|_2$. For a vector $a, b \in \R^n$, we use $\langle a, b \rangle$ to denote $\sum_{i=1}^n a_i b_i$. 

\subsection{Facts}
\begin{fact}
    Let $A \in \R^{n \times d}$, only one element of $A_{j_1,i_1}$ have non-zero gradient with respect to $A$. 
    
    $e_{i_1}$ is a column vector. For $d$ dimensional space, it is a $d \times 1$ column vector, in which only the $i_1$-th element is 1, and the other elements are all 0. This vector represents the unit vector in the $i_1$-th dimension. Similarly, $e_{j_1}$ is also a column vector representing the unit vector in the $j_1$-th dimension.
\end{fact}
\begin{fact} \label{fac:exponential_der_rule}
Let $g, f: \R^d \to \R^n$ and $q: \R^d \to \R$. 

Let $x \in \R^d$ be an arbitrary vector.

Let $a \in \R$ be an arbitrary real number.

Then, we have
    \begin{itemize}
        \item $\frac{\d q(x)^a}{\d x} =  a\cdot q(x)^{a-1} \cdot \frac{\d q(x)}{\d x}$
        \item $\frac {\d \|f(x) \|^2_2}{\d t} = 2 \langle f(x) , \frac{\d f(x)}{\d t} \rangle $
        \item $\frac{\d \langle f(x), g(x) \rangle}{\d t} = \langle \frac{\d f(x)}{\d t} , g(x) \rangle + \langle f(x), \frac{\d g(x)}{\d t} \rangle$
        \item $\frac{\d (g(x) \circ f(x))}{\d t} = \frac{\d g(x)}{\d t} \circ f(x) + g(x) \circ \frac{\d f(x)}{\d t}$ (product rule for Hadamard product)
    \end{itemize}
\end{fact}
\begin{fact}\label{fac:circ_rules}
    Let $a, b \in \R$.
    
    For all vectors $u,v,w \in \R^{n}$, we have 
    \begin{itemize}
        \item $\langle u,v \rangle = \langle u \circ v, {\bf 1}_n \rangle =  u^\top \mathrm{diag}(v)  {\bf 1}_n $
        \item $\langle u \circ v, w\rangle = \langle u \circ w, v\rangle$
        \item $\langle u \circ v, w \rangle =  \langle u \circ v \circ w, {\bf 1}_n  \rangle = u^\top \diag(v) w$
        \item $\langle u \circ v \circ w \circ z , {\bf 1}_n \rangle = u^\top \diag(v \circ w) z$
        \item $u \circ  v = v \circ u = \diag (u) \cdot v = \diag (v) \cdot u$ 
        \item $u^{\top}(v \circ w) = v^{\top}(u \circ w) = w^{\top}(u \circ v)= u^{\top}\diag(v) w = v^{\top}\diag(u) w = w^{\top}\diag(u) v$
        \item $ \diag (u)^{\top} = \diag (u)$
        \item $\diag (u) \cdot \diag (v) \cdot {\bf 1}_n = \diag(u) v$
        \item $\diag (u \circ v) = \diag (u) \diag (v)$
        \item $\diag (u) + \diag (v) = \diag (u +v)$
        \item $\langle u,v \rangle = \langle v,u \rangle$
        \item $\langle u,v \rangle = u^\top v = v^\top u$
        \item $a\langle w, v \rangle + b\langle u, v \rangle = \langle aw + bu, v \rangle = \langle v, aw + bu \rangle = a\langle v, w \rangle + b\langle v, u \rangle$.
    \end{itemize}
\end{fact}

\begin{fact}\label{fac:vector_norm}
Let $R > 0$.

For vectors $x,y,a,b,c,d \in \R^n$, and a constant $\alpha \in \R$ we have
\begin{itemize}
        \item $ \|\diag(x) \|_F =  \|x\|_2$
        \item $\| ab - cd \|_2 = \| ab - cb + cb - cd\|_2 \leq \| ab - cb \|_2 + \| cb - cd\|_2$
        \item $\|x\|_{2} = \|-x\|_{2}$
        \item $\langle u,v \rangle \leq \|u\|_{2} \cdot \|v\|_{2}$ (Cauchy-Schwarz inequality)
        \item $\| \diag(u) \| \leq \|u\|_{\infty}$
        \item $\| x \circ y \|_{2} \leq \| x \|_{\infty} \cdot \|y\|_{2}$
        \item $\| x \|_{\infty} \leq \| x \|_{2} \leq \sqrt{n} \|x\|_{\infty}$
        \item $\| x \|_{2} \leq \| x \|_{1} \leq \sqrt{n} \|x\|_{2}$
        \item $\| \exp(x)\|_{\infty} \leq \exp({\|x\|_{\infty}}) \leq \exp({\|x\|_{2}})$
        \item $\| x + y \|_{2} \leq \| x \|_{2} + \|y\|_{2}$
        \item  Let $\alpha$ be a scalar, then $\| \alpha x \|_{2} \leq |\alpha| \cdot \|x\|_{2}$
        \item  For any $\|x\|_{2}$, $\|y\|_{2} \leq R$, we have $\| \exp(x) - \exp(y) \|_{2} \leq \exp(R) \cdot \|x-y\|_{2}$
\end{itemize}
\end{fact}

\begin{fact}\label{fac:matrix_norm}
Let $R >0$.

For matrices $U, V \in \R^{n \times d}$ and $x \in \R^d$ denotes to a vector, we have
\begin{itemize}
        \item $\| U x \|_2 \leq \| U \|_F \cdot \| x \|_2$
        \item $\| U^{\top} \| = \| U \|$
        \item $\| U \| \geq \| V \| - \|U - V\|$
        \item $\| U + V \| \leq \| U \| + \|V\|$
        \item $\| U \cdot V \| \leq \| U \| \cdot \|V\|$
        \item If $U \preceq \alpha \cdot V$, then $\|U \| \leq \alpha \cdot \|V\|$
        \item For scalar $\alpha \in R$, we have $\| \alpha \cdot U\| \leq |\alpha| \cdot \| U \|$
        \item For any vector $v$, we have $\|Uv\| \leq \|U\| \cdot \|v\|_{2}$
        \item Let $u, v \in R^{n}$ denote two vectors, then we have $\|u \cdot v^{\top}\| \leq \|u\|_{2} \cdot \|v\|_{2}$
        \item For Frobenius norm and $A \in \R^{n \times d}$, we have $\|A\|_{F} = \sqrt{\sum_{i,j} |A_{i,j}|^{2}}$
    \end{itemize}
\end{fact}
\begin{fact}\label{fac:psd_rule}
    For any vectors $u,v \in \R^n$, we have
    \begin{itemize}
        \item Part 1. $u u^{\top} \preceq \| u\|_2^2 \cdot I_n $
        \item Part 2. $\diag(u) \preceq \|u\|_2 \cdot I_n$
        \item Part 3. $\diag(u \circ u) \preceq \|u\|_2^2 \cdot I_n$
        \item Part 4. $uv^{\top} + vu^{\top} \preceq uu^
        \top + vv^{\top}$
        \item Part 5. $uv^{\top} + vu^{\top} \succeq -( uu^
        \top + vv^{\top})$
        \item Part 6. $(v \circ u) (v \circ u)^{\top} \preceq \| v\|^2_{\infty} u u^{\top}$
        \item  Part 7. $\diag(u \circ v) \preceq \|u\|_2\|v\|_2 \cdot I_n$
    \end{itemize}
\end{fact}

\begin{fact}\label{fac:I_d_1_d}
We have
\begin{itemize}
    \item Part 1. $\| I_d \| = 1$
    \item Part 2. $\| I_d \|_F = \sqrt{d}$ (We remark that for a matrix $\| \cdot \|$ (spectral norm) is NOT same as $\| \cdot \|_F$ (Frobenius norm) )
    \item Part 3. $\| {\bf 1}_d \|_2 = \sqrt{d}$
\end{itemize}
\end{fact}

\begin{fact}
Let $a, b \in \R^d$ denote two vectors, then we have
\begin{align*}
   \| a b^\top \| \leq \| a\|_2 \cdot \| b \|_2
\end{align*}
\begin{align*}
    \| \diag(a ) \| \leq \| a \|_{\infty} \leq \| a \|_2 \leq \| a \|_1.
\end{align*}
\end{fact}

\subsection{Functions}

\begin{definition}\label{def:u}
 We define $u(A,x)$ as follows
\begin{align*}
    u(A,x) := \exp(Ax)
\end{align*}
\end{definition}

\begin{definition}\label{def:alpha}
We define $\alpha(A,x)$ as follows
\begin{align*}
    \alpha(A,x):= \langle u(A,x) , {\bf 1}_n \rangle
\end{align*}
\end{definition}

\begin{definition}\label{def:f}
We define $f(A,x)$ as follows
\begin{align*}
    f(A,x):= \alpha(A,x)^{-1} u(A,x).
\end{align*}
\end{definition}

\begin{definition}\label{def:c}
Given $b_f \in \R^n$. 
We define $c(A,x):= f(A,x) - b_f$.
\end{definition}

\begin{definition}\label{def:L_exp}

Let $A \in \R^{n \times d}$ be the input, and let $x \in \R^d$ denote the weights. 
Let $L(A,x)$ denote the loss function. 
\begin{align*}
    L_f(A,x) := 0.5 \| c(A,x) \|_2^2
\end{align*}

Let $g(A,x) \in \R^d$ denote the gradient of $L_f(A,x)$ with respect to $x \in \R^d$, i.e.,
\begin{align*}
 g(A,x) := \frac{\d L_f(A,x)}{ \d x}
\end{align*}
\end{definition}

\subsection{Gradient Closed-Form}

In this section, we use $g(A,x)$ to denote the gradient of $L_{f}(A,x)$.
\begin{definition}\label{def:g}
If the following conditions hold
\begin{itemize}
\item Let $A_{*,i} \in \R^n$ denote the $i$-th column of $A \in \R^{n \times d}$.
\item Let $c(A,x)$ be defined as Definition~\ref{def:c}.
\item Let $f(A,x)$ be defined as Definition~\ref{def:f}.
\end{itemize} 
We define $g(x) \in \R^d$ as follows
\begin{align*}
g(A,x) :=  &  ~\underbrace{A^\top }_{d \times n} \cdot \Big( -\underbrace{f(A,x) }_{n \times 1} \underbrace{ \langle c(A, x), f(A, x) \rangle }_{ \mathrm{scalar} } \\
& + ~ \underbrace{ \diag(f(A, x)) }_{ n\times n} \underbrace{ c(A,x) }_{n \times 1} \Big) 
\end{align*}
Equivalently, for each $i \in [d]$, we define 
\begin{align*}
g(A,x)_i := & ~ -\underbrace{ \langle A_{*,i}, f(A,x) \rangle }_{\mathrm{scalar}} \cdot \underbrace{ \langle c(A, x), f(A, x) \rangle }_{ \mathrm{scalar} } \\
& + ~  \underbrace{ \langle A_{*,i}, f(A, x) \circ c(A, x)\rangle }_{ \mathrm{scalar} }.  
\end{align*}
\end{definition}

\begin{definition}\label{def:c_g}
Given $b_g \in \R^d$.

We define $c_g: \R^{n \times d}  \rightarrow \R^d $
\begin{align*}
    c_g(A) := g(A,x) - b_g
\end{align*}
\end{definition}

\begin{definition}\label{def:l_g}
Given the observed gradient $b_g \in \R^d$ and weight $x \in \R^d$, we define
\begin{align*}
    L_g(A,x) := 0.5 \| c_g(A) \|_2^2
\end{align*}
The goal is to solve a new optimization problem
\begin{align*}
    \min_{A \in \R^{n \times d}} L_g(A)
\end{align*}
\end{definition}

\section{Gradient Computation}\label{app:gradient}

\subsection{Gradient for \texorpdfstring{$u$}{} and \texorpdfstring{$\alpha$}{}}

\begin{lemma}\label{lem:gradient_u}
If the following conditions hold,
\begin{itemize}
    \item Let $A_{*,i} \in \R^n$ denote the $i$-th column of $A \in \R^{n \times d}$.
    \item Let $x \in \R^d$ 
    \item Let $u(A,x)$ be defined as Definition~\ref{def:u}
    \item Let $\alpha(A,x)$ be defined as Definition~\ref{def:alpha}
    \item Let $f(A,x)$ be defined as Definition~\ref{def:f}
    \item Let $c(A,x)$ be defined as Definition~\ref{def:c}
    \item Let $g(A,x)$ be defined as Definition~\ref{def:g} 
    \item Let $c_g(A,x)$ be defined as Definition~\ref{def:c_g} 
    \item Let $L_g(A,x)$ be defined as Definition~\ref{def:l_g}
\end{itemize}
Then, we can compute the following
\begin{itemize}
    \item {\bf Part 1.} Let $j_0 \in [n]$, let $i_1 \in [d]$, let $j_0 = j_1$
    \begin{align*}
        \frac{\d (Ax)_{j_0}}{ \d A_{j_1,i_1} } = x_{i_1}
    \end{align*}
    \item {\bf Part 2.} Let $j_0 \in [n]$, let $i_1 \in [d]$, let $j_0 \neq j_1$
    \begin{align*}
        \frac{\d (Ax)_{j_0}}{ \d A_{j_1,i_1} } = 0
    \end{align*}
    \item {\bf Part 3.} Let $j_0 \in [n]$, let $i_1 \in [d]$, let $j_0 = j_1$
    \begin{align*}
        \frac{\d u(A,x)_{j_0}}{ \d A_{j_1,i_1} } = u(A,x)_{j_1} \cdot x_{i_1}
    \end{align*}
    \item {\bf Part 4.} Let $j_0 \in [n]$, let $i_1 \in [d]$, let $j_0 \neq j_1$
    \begin{align*}
        \frac{\d u(A,x)_{j_0}}{ \d A_{j_1,i_1} } = 0
    \end{align*}
    \item {\bf Part 5.} 
    \begin{align*}
        \frac{\d u(A,x)}{ \d A_{j_1,i_1} } =  u(A,x)_{j_1} \cdot x_{i_1} \cdot e_{j_1}
    \end{align*}
\end{itemize}
\end{lemma}
\begin{proof}
{\bf Proof of Part 1}
For $j_0= j_1$. 
We have
\begin{align*}
    \frac{\d (Ax)_{j_0}}{ \d A_{j_1,i_1} }
    = & ~ \frac{\d \langle A_{j_0,*} ,x \rangle }{ \d A_{j_0,i_1} } \\
    = & ~  \langle \frac{\d  A_{j_0,*}}{ \d A_{j_0,i_1}} , x \rangle \\
    = & ~ \langle e_{i_1}, x \rangle \\
    = & ~ x_{i_1}
\end{align*}
where the first step follows from the multiplication rules of matrices, the second step comes from the product rule of derivative, the third step follows from only one element of $A_{j_0,*}$ have non-zero gradient with respect to $A_{j_0,i_1}$, and the last step follows from simple algebra.

{\bf Proof of Part 2}

For $j_0 \neq j_1$,

\begin{align*}
    \frac{\d (Ax)_{j_0}}{ \d A_{j_1,i_1} }
    = & ~ \frac{\d \langle A_{j_0,*} ,x \rangle }{ \d A_{j_1,i_1} } \\
    = & ~  \langle \frac{\d  A_{j_0,*}}{ \d A_{j_1,i_1}} , x \rangle \\
    = & ~ \langle {\bf 0}_d, x \rangle \\
    = & ~ 0
\end{align*}
where the first step follows from the multiplication rules of matrices, the second step comes from the product rule of derivative, the third step follows from  $A_{j_0,*}$ have all $0$ gradient with respect to $A_{j_1,i_1}$, and the last step follows from simple algebra.

{\bf Proof of Part 3.}

For $j_0 = j_1$,
\begin{align*}
    \frac{\d u(A,x)_{j_0}}{ \d A_{j_1,i_1} }
    = & ~ \frac{\d (\exp(Ax))_{j_1}}{ \d A_{j_1,i_1} } \\
    = & ~ (\exp(Ax))_{j_1} \frac{\d (Ax)_{j_1}}{ \d A_{j_1,i_1} } \\
    = & ~ u(A,x)_{j_1} \cdot x_{i_1}
\end{align*}
where the first step follows from the definition of $u(A,x)$ (see Definition~\ref{def:u}), the second step follows from chain rule of derivative, and the last step is due to {\bf Part 1}.

{\bf Proof of Part 4.}

For $j_0 \neq j_1$
\begin{align*}
    \frac{\d u(A,x)_{j_0}}{ \d A_{j_1,i_1} }
    = & ~ \frac{\d (\exp(Ax))_{j_0}}{ \d A_{j_1,i_1} } \\
    = & ~ (\exp(Ax))_{j_0} \frac{\d (Ax)_{j_0}}{ \d A_{j_1,i_1} } \\
    = & ~ 0
\end{align*}
where the first step follows from the definition of $u(A,x)$ (see Definition~\ref{def:u}), the second step follows from chain rule of derivative, and the last step is due to {\bf Part 2}.

{\bf Proof of Part 5.}
\begin{align*}
     \frac{\d u(A,x)}{ \d A_{j_1,i_1} } 
     = & ~ u(A,x)_{j_1} \cdot x_{i_1} \cdot e_{j_1}
\end{align*}

where the first step follows from {\bf Part 3 and 4}.
\end{proof}

\subsection{Gradient for \texorpdfstring{$\alpha$}{} and \texorpdfstring{$\alpha^{-1}$}{}}

\begin{lemma}\label{lem:gradient_alpha}
If the following conditions hold
\begin{itemize}
    \item Let $x \in \R^d$ 
    \item Let $u(A,x) \in \R^n$ be defined as Definition~\ref{def:u}
    \item Let $\alpha(A,x) \in \R$ be defined as Definition~\ref{def:alpha}
    \item Let $f(A,x) \in \R^n$ be defined as Definition~\ref{def:f}
\end{itemize}
Then, let $j_1 \in [n], i_1 \in [d]$, we have
\begin{itemize}
    \item {\bf Part 1}.
    \begin{align*}
        \frac{\d \alpha(A,x) }{ \d A_{j_1,i_1} } =  u(A,x)_{j_1} \cdot x_{i_1} 
    \end{align*}
    \item {\bf Part 2}.
    \begin{align*}
        \frac{\d \alpha(A,x)^{-1} }{ \d A_{j_1,i_1} } = - \alpha(A,x)^{-1} \cdot  f(A,x)_{j_1} \cdot x_{i_1} 
    \end{align*}
\end{itemize}
\end{lemma}
\begin{proof}

{\bf Proof of Part 1.}

\begin{align*}
    \frac{\d \alpha(A,x) }{ \d A_{j_1,i_1} } 
    = & ~ \frac{ \d \langle u(A, x), {\bf 1}_n \rangle } { \d A_{j_1,i_1} } \\
    = & ~ \langle \frac{\d u(A, x)}{\d A_{j_1,i_1}},  {\bf 1}_n  \rangle \\
    = & ~  \langle u(A,x)_{j_1} \cdot x_{i_1} \cdot e_{i_1},{\bf 1}_n  \rangle \\
    = & ~ u(A,x)_{j_1} \cdot x_{i_1} 
\end{align*}
where the first step follows from the definition of $\alpha(A,x)$ (see Definition~\ref{def:alpha}), the second step comes from the product rule of derivative, the third step is based on {\bf Part 3, 4} of Lemma~\ref{lem:gradient_u}, and the last step follows from the simple algebra.

{\bf Proof of Part 2.}
We can show that
\begin{align*}
     \frac{\d \alpha(A,x)^{-1} }{ \d A_{j_1,i_1} } 
     = & ~ - \alpha(A,x)^{-2} \cdot \frac{\d \alpha(A,x) }{ \d A_{j_1,i_1} } \\
     = & ~ - \alpha(A,x)^{-2} \cdot u(A,x)_{j_1} \cdot x_{i_1} \\
     = & ~ - \alpha(A,x)^{-1} \cdot f(A,x)_{j_1} \cdot x_{i_1} 
\end{align*}
where the first step follows from the chain rule of derivative, the second step follows from {\bf Part 1}, and the last step is based on the definition of $f(A,x)$ (see Definition~\ref{def:f}).

\end{proof}

\subsection{Gradient for \texorpdfstring{$f(A,x)$}{}}

\begin{lemma}\label{lem:gradient_f}
If the following conditions hold
\begin{itemize}
    \item Let $x \in \R^d$
    \item Let $u(A,x) \in \R^n$ be defined as Definition~\ref{def:u}
    \item Let $\alpha(A,x) \in \R$ be defined as Definition~\ref{def:alpha}
    \item Let $f(A,x) \in \R^n$ be defined as Definition~\ref{def:f}
\end{itemize}
Then, let $i_1 \in [d]$, we have
\begin{itemize}
    \item {\bf Part 1.} For $j_0 = j_1 \in [n]$
    \begin{align*}
        \frac{\d f(A,x)_{j_0} }{\d A_{j_1,i_1}} = ~- f(A,x)_{j_1}^2 \cdot x_{i_1} + f(A,x)_{j_1} \cdot x_{i_1}
    \end{align*}
    \item {\bf Part 2.} For $j_0 \neq j_1 \in [n]$
    \begin{align*}
        \frac{\d f(A,x)_{j_0} }{ \d A_{j_1,i_1} } = ~- f(A,x)_{j_1} \cdot  f(A,x)_{j_0} \cdot x_{i_1}
    \end{align*}
    \item {\bf Part 3.} Let $j_1 \in [n]$
    \begin{align*}
        \underbrace{ \frac{\d f(A,x) }{ \d A_{j_1,i_1} } }_{n \times 1} = - \underbrace{ f(A,x) }_{n \times 1} \cdot \underbrace{ f(A,x)_{j_1} }_{ \mathrm{scalar} } \cdot \underbrace{ x_{i_1} }_{ \mathrm{scalar} } + \underbrace{ e_{j_1} }_{n \times 1} \cdot \underbrace{ f(A,x)_{j_1} }_{\mathrm{scalar}} \cdot \underbrace{ x_{i_1} }_{\mathrm{scalar}}
    \end{align*}
\end{itemize}
\end{lemma}
\begin{proof}
{\bf Proof of Part 1.}
For $j_0 = j_1$
\begin{align*}
    \frac{\d f(A,x)_{j_0} }{\d A_{j_1,i_1}}
    = & ~  \frac{\d \alpha(A,x)^{-1} u(A,x)_{j_1} }{\d A_{j_1,i_1}} \\
    = & ~ \frac{\d \alpha(A,x)^{-1}  }{\d A_{j_1,i_1}} \cdot  u(A,x)_{j_1}  + \alpha(A,x)^{-1} \cdot\frac{\d  u(A,x)_{j_1} }{\d A_{j_1,i_1}} \\
    = & ~ -\alpha(A,x)^{-1} \cdot  f(A,x)_{j_1} \cdot x_{i_1} \cdot  u(A,x)_{j_1}\\ & + ~\alpha(A,x)^{-1} u(A,x)_{j_1} \cdot x_{i_1}\\
    = & ~ - f(A,x)_{j_1}^2 \cdot x_{i_1} + f(A,x)_{j_1} \cdot x_{i_1}
\end{align*}
where the first step follows from the the definition of $f(A,x)$ (see Definition~\ref{def:f}), the second step is based on the product rule of derivative, the third step is due to {\bf Part 3} of Lemma~\ref{lem:gradient_u} and {\bf Part 2} of Lemma~\ref{lem:gradient_alpha}, and the last step follows from the definition of $f(A,x)$ (see Definition~\ref{def:f}).

{\bf Proof of Part 2.}
For $j_0 \neq j_1$
\begin{align*}
    \frac{\d f(A,x)_{j_0} }{\d A_{j_1,i_1}}
    = & ~  \frac{\d \alpha(A,x)^{-1} u(A,x)_{j_0} }{\d A_{j_1,i_1}} \\
    = & ~ \frac{\d \alpha(A,x)^{-1}  }{\d A_{j_1,i_1}} \cdot  u(A,x)_{j_0}  + \alpha(A,x)^{-1} \cdot\frac{\d  u(A,x)_{j_0} }{\d A_{j_1,i_1}} \\
    = & ~ -\alpha(A,x)^{-1} \cdot  f(A,x)_{j_1} \cdot x_{i_1} \cdot u(A,x)_{j_0} \\
    = & ~  - f(A,x)_{j_1} \cdot  f(A,x)_{j_0} \cdot x_{i_1}
\end{align*}
where the first step follows from the definition of $f(A,x)$ (see Definition~\ref{def:f}), the second step is based on the product rule of derivative, the third step is due to {\bf Part 4} of Lemma~\ref{lem:gradient_u} and {\bf Part 2} of Lemma~\ref{lem:gradient_alpha}, and the last step follows from the definition of $f(A,x)$ (see Definition~\ref{def:f}).

{\bf Proof of Part 3.}
\begin{align*}
    \frac{\d f(A,x) }{ \d A_{j_1,i_1}}
    = & ~ \frac{\d \alpha(A,x)^{-1} u(A,x) }{ \d A_{j_1,i_1}} \\
    = & ~ \frac{\d \alpha(A,x)^{-1}  }{\d A_{j_1,i_1}} \cdot  u(A,x)  + \alpha(A,x)^{-1} \cdot\frac{\d  u(A,x) }{\d A_{j_1,i_1}} \\
    = & ~  -\alpha(A,x)^{-1} \cdot  f(A,x)_{j_1} \cdot x_{i_1} \cdot u(A,x) \\
    & + ~ \alpha(A,x)^{-1}u(A,x)_{j_1} \cdot x_{i_1} \cdot e_{i_1} \\
    = & ~ f(A,x)_{j_1} \cdot x_{i_1} (e_{j_1}- f(A,x) )
\end{align*}
where the first step follows from the definition of $f(A,x)$ (see Definition~\ref{def:f}), the second step follows from the product rule of derivative, the third step follows from {\bf Part 5} of Lemma~\ref{lem:gradient_u} and {\bf Part 2} of Lemma~\ref{lem:gradient_alpha}, and the last step follows from the definition of $f(A,x)$ (see Definition~\ref{def:f}).
\end{proof}

\subsection{Gradient for \texorpdfstring{$c(A,x)$}{}}

\begin{lemma}\label{lem:gradient_c}
If the following conditions hold
\begin{itemize}
    \item Let $x \in \R^d$ 
    \item Let $u(A,x) \in \R^n$ be defined as Definition~\ref{def:u}
    \item Let $\alpha(A,x) \in \R$ be defined as Definition~\ref{def:alpha}
    \item Let $f(A,x) \in \R^n$ be defined as Definition~\ref{def:f}
    \item Let $c(A,x) \in \R^n$ be defined as Definition~\ref{def:c}
    \item Let $b_f \in \R^n$, let $(b_f)_{j_1}$ denote the $j_1$-th coordinate of $b_f$, for each $j_1 \in [n]$.
\end{itemize}
Then, let $i_1\in [d]$ we have
\begin{itemize}
    \item {\bf Part 1.} For $j_0 = j_1 \in [n]$ 
    \begin{align*}
        \frac{\d c(A,x)_{j_0} }{\d A_{j_1,i_1}} = - f(A,x)_{j_1}^2 \cdot x_{i_1} + f(A,x)_{j_1} \cdot x_{i_1}
    \end{align*}
    \item {\bf Part 2.} For $j_0 \neq j_1 \in [n]$
    \begin{align*}
        \frac{\d c(A,x)_{j_0} }{ \d A_{j_1,i_1} } = - f(A,x)_{j_1} \cdot  f(A,x)_{j_0} \cdot x_{i_1}
    \end{align*}
    \item {\bf Part 3.}
    \begin{align*}
        \frac{\d c(A,x) }{ \d A_{j_1,i_1} } = &  ~ - f(A,x) \cdot f(A,x)_{j_1} \cdot x_{i_1} \\
        & + ~  e_{j_1}  \cdot f(A,x)_{j_1} \cdot x_{i_1} 
    \end{align*}
\end{itemize}
\end{lemma}
\begin{proof}
{\bf Proof of Part 1.}
For $j_0 = j_1$
\begin{align*}
    \frac{\d c(A,x)_{j_0} }{\d A_{j_1,i_1}}
    = & ~ \frac{\d (f(A,x)_{j_1} - (b_{f})_{j_1} ) }{\d A_{j_1,i_1}} \\
    = & ~ \frac{\d f(A,x)_{j_1}}{\d A_{j_1,i_1}} \\
    = & ~ - f(A,x)_{j_1}^2 \cdot x_{i_1} + f(A,x)_{j_1} \cdot x_{i_1}
\end{align*}
where the first step follows from the definition of $c(A,x)$ (see Definition~\ref{def:c}), the second step comes from the chain rule of derivative, and the last step is due to {\bf Part 2} of Lemma~\ref{lem:gradient_f}.

{\bf Proof of Part 2.}
For $j_0 \neq j_1$
\begin{align*}
    \frac{\d c(A,x)_{j_0} }{\d A_{j_1,i_1}}
    = & ~ \frac{\d (f(A,x)_{j_0} -(b_{f})_{j_1})}{\d A_{j_1,i_1}} \\
    = & ~ \frac{\d f(A,x)_{j_0}}{\d A_{j_1,i_1}} \\
    = & ~  - f(A,x)_{j_1} \cdot  f(A,x)_{j_0} \cdot x_{i_1}
\end{align*}
where the first step follows from the definition of $c(A,x)$ (see Definition~\ref{def:c}), the second step comes from the chain rule of derivative, and the last step is due to {\bf Part 1} of Lemma~\ref{lem:gradient_f}.

{\bf Proof of Part 3.}
\begin{align*}
    \frac{\d c(A,x) }{\d A_{j_1,i_1}}
     = & ~ \frac{\d (f(A,x) - b_f)}{\d A_{j_1,i_1}} \\
      = & ~ \frac{\d f(A,x)}{\d A_{j_1,i_1}} \\
      = & ~ - f(A,x) \cdot  f(A,x)_{j_1} \cdot x_{i_1} +  e_{j_1}  \cdot f(A,x)_{j_1} \cdot x_{i_1} 
\end{align*}
where the first step follows from the definition of $c(A,x)$ (see Definition~\ref{def:c}), the second step comes from the chain rule of derivative, and the last step is due to {\bf Part 3} of Lemma~\ref{lem:gradient_f}.
\end{proof}

\subsection{Gradient for \texorpdfstring{$\langle f(A,x), v \rangle$}{}}

\begin{lemma}\label{lem:gradient_fv}
If the following conditions hold
\begin{itemize}
    \item Let $x \in \R^d$ 
    \item Let $f(A,x) \in \R^n$ be defined as Definition~\ref{def:f}
    \item Let $v \in \R^n$ denote a fixed vector that is independent of $A$
\end{itemize}
Then, we have
\begin{itemize}
    \item {\bf Part 1.} 
    \begin{align*}
        \frac{\d \langle f(A,x) , v \rangle}{ \d A_{j_1,i_1}} = \underbrace{f(A,x)_{j_1}}_{\mathrm{scalar}} \cdot \underbrace{x_{i_1}}_{\mathrm{scalar}} (\langle - \underbrace{f(A,x)}_{n \times 1}, \underbrace{v}_{n \times 1} \rangle +  \underbrace{v_{j_1}}_{\mathrm{scalar}})
    \end{align*}
\end{itemize}
\end{lemma}
\begin{proof}
\begin{align*}
    \frac{\d \langle f(A,x) , v \rangle}{ \d A_{j_1,i_1}} 
    = & ~ \langle  \frac{\d  f(A,x)}{ \d A_{j_1,i_1}} , v  \rangle\\
    = & ~ \langle - f(A,x) \cdot  f(A,x)_{j_1} \cdot x_{i_1}, v \rangle \\
    & +~ \langle e_{j_1}  \cdot f(A,x)_{j_1} \cdot x_{i_1}, v \rangle \\
    = & ~ f(A,x)_{j_1} \cdot x_{i_1} (\langle - f(A,x), v \rangle +  v_{j_1})
\end{align*}
where the first step follows from the product rule of derivative, the second step comes from {\bf Part 3} of Lemma~\ref{lem:gradient_f}, and the last step is based on simple algebra.
\end{proof}

\subsection{Gradient for \texorpdfstring{$\langle c(A,x), v \rangle$}{}}

\begin{lemma}\label{lem:gradient_cv}
If the following conditions hold
\begin{itemize}
    \item Let $x \in \R^d$ 
    \item Let $f(A,x) \in \R^n$ be defined as Definition~\ref{def:f}
    \item Let $c(A,x) \in \R^n$ be defined as Definition~\ref{def:c}
    \item Let $v \in \R^n$ denote a fixed vector that is independent of $A$
\end{itemize}
Then, we have
\begin{itemize}
    \item {\bf Part 1.} 
    \begin{align*}
        \frac{\d \langle c(A,x) , v \rangle}{ \d A_{j_1,i_1}} = \underbrace{f(A,x)_{j_1}}_{\mathrm{scalar}} \cdot \underbrace{x_{i_1}}_{\mathrm{scalar}} \cdot (\langle - \underbrace{f(A,x)}_{n \times 1}, \underbrace{v}_{n \times 1} \rangle +  \underbrace{v_{j_1}}_{\mathrm{scalar}})
    \end{align*}
\end{itemize}
\end{lemma}
\begin{proof}
\begin{align*}
    \frac{\d \langle c(A,x) , v \rangle}{ \d A_{j_1,i_1}} 
    = & ~ \langle  \frac{\d  c(A,x)}{ \d A_{j_1,i_1}} , v  \rangle\\
    = & ~ \langle - f(A,x) \cdot  f(A,x)_{j_1} \cdot x_{i_1} +  e_{j_1}  \cdot f(A,x)_{j_1} \cdot x_{i_1}, v \rangle \\
    = & ~ f(A,x)_{j_1} \cdot x_{i_1} \cdot (\langle - f(A,x), v \rangle +  v_{j_1})
\end{align*}
where the first step follows from the product rule of derivative, the second step comes from {\bf Part 3} of Lemma~\ref{lem:gradient_f}, and the last step is based on simple algebra.
\end{proof}

\subsection{Gradient for \texorpdfstring{$\langle c(A,x), f(A,x) \rangle$}{}}

\begin{lemma}\label{lem:gradient_c_f}
If the following conditions hold,
\begin{itemize}
    \item Let $x \in \R^d$
    \item Let $f(A,x) \in \R^n$ be defined as Definition~\ref{def:f}
    \item Let $c(A,x) \in \R^n$ be defined as Definition~\ref{def:c}
\end{itemize}
then we have
\begin{itemize}
    \item {\bf Part 1.} 
    \begin{align*}
   \frac{\d \langle c(A,x), f(A,x) \rangle}{\d A_{j_1,i_1}} = & ~ \underbrace{f(A,x)_{j_1}}_{\mathrm{scalar}} \cdot \underbrace{x_{i_1}}_{\mathrm{scalar}} \cdot (\langle - \underbrace{f(A,x)}_{n \times 1}, \underbrace{f(A,x)}_{n \times 1} \rangle \\
   & + ~  \underbrace{f(A,x)_{j_1}}_{\mathrm{scalar}} \cdot \underbrace{x_{i_1}}_{\mathrm{scalar}} \cdot \underbrace{f(A,x)_{j_1}}_{\mathrm{scalar}} \\
        & ~ + \underbrace{f(A,x)_{j_1}}_{\mathrm{scalar}} \cdot \underbrace{x_{i_1}}_{\mathrm{scalar}} \cdot (\langle - \underbrace{f(A,x)}_{n \times 1}, \underbrace{c(A,x)}_{n \times 1} \rangle +  \underbrace{c(A,x)_{j_1}}_{\mathrm{scalar}})
   \end{align*}
\end{itemize}
\end{lemma}
\begin{proof}
    \begin{align*}
        \frac{\d \langle c(A,x), f(A,x) \rangle}{\d A_{j_1,i_1}} 
        = & ~  \langle \frac{\d c(A,x) }{\d A_{j_1,i_1}}, f(A,x) \rangle\\
        & + ~ \langle \frac{\d f(A,x) }{\d A_{j_1,i_1}}, c(A,x) \rangle \\
        = & ~ \langle f(A,x)_{j_1} \cdot x_{i_1} (e_{j_1}- f(A,x) ), f(A,x) \rangle \\
        & + ~\langle f(A,x)_{j_1} \cdot x_{i_1} (e_{j_1}- f(A,x) ), c(A,x)\rangle \\
        = & ~ f(A,x)_{j_1} \cdot x_{i_1} \cdot \langle - f(A,x), f(A,x) \rangle\\
        & + ~  f(A,x)_{j_1} \cdot x_{i_1} \cdot f(A,x)_{j_1} \\
        & +~ f(A,x)_{j_1} \cdot x_{i_1} \cdot (\langle - f(A,x), c(A,x) \rangle )\\
        & + ~  f(A,x)_{j_1} \cdot x_{i_1} \cdot c(A,x)_{j_1}
    \end{align*}
where the first step follows from the product rule of derivative,  the second step comes from {\bf Part 1} of Lemma~\ref{lem:gradient_fv} and {\bf Part 1} of Lemma~\ref{lem:gradient_cv}, and the last step follows from simple algebra.
\end{proof}

\subsection{Gradient for \texorpdfstring{$\langle A_{*,i_0}, f(A,x) \rangle$}{}}

\begin{lemma}\label{lem:gradient_a_f}
If the following conditions hold,
\begin{itemize}
    \item Let $A_{*,i} \in \R^n$ denote the $i$-th column of $A \in \R^{n \times d}$.
    \item Let $x \in \R^d$ 
    \item Let $f(A,x) \in \R^n$ be defined as Definition~\ref{def:f}
    \item Let $c(A,x) \in \R^n$ be defined as Definition~\ref{def:c} 
\end{itemize}
then we have
\begin{itemize}
    \item {\bf Part 1.} For $i_0 = i_1 \in [d]$
    \begin{align*}
    \frac{\d \langle A_{*,i_0}, f(A,x) \rangle}{\d A_{j_1,i_1}} = \underbrace{f(A,x)_{j_1}}_{\mathrm{scalar}} + \underbrace{f(A,x)_{j_1}}_{\mathrm{scalar}} \underbrace{x_{i_1}}_{\mathrm{scalar}} \underbrace{A_{j_1,i_1}}_{\mathrm{scalar}} - \underbrace{f(A,x)_{j_1}}_{\mathrm{scalar}} \underbrace{x_{i_1}}_{\mathrm{scalar}} \cdot \langle \underbrace{A_{*,i_1}}_{n \times 1} ,\underbrace{f(A,x)}_{n \times 1} \rangle
    \end{align*}
    \item {\bf Part 2.} For $i_0 \neq i_1 \in [d]$
    \begin{align*}
    \frac{\d \langle A_{*,i_0}, f(A,x) \rangle}{\d A_{j_1,i_1}} = \underbrace{f(A,x)_{j_1}}_{\mathrm{scalar}} \underbrace{x_{i_1}}_{\mathrm{scalar}} \underbrace{A_{j_1,i_1}}_{\mathrm{scalar}} - \underbrace{f(A,x)_{j_1}}_{\mathrm{scalar}} \underbrace{x_{i_1}}_{\mathrm{scalar}} \cdot \langle \underbrace{A_{*,i_0}}_{n \times 1} ,\underbrace{f(A,x)}_{n \times 1} \rangle
    \end{align*}
\end{itemize}
\end{lemma}

\begin{proof}
{\bf Proof of Part 1.}
For $i_0 = i_1 \in [d]$
    \begin{align*}
        \frac{\d \langle A_{*,i_0}, f(A,x) \rangle}{\d A_{j_1,i_1}} 
        = & ~\langle \frac{\d  A_{*,i_1}}{\d A_{j_1,i_1}},  f(A,x) \rangle + \langle A_{*,i_1},  \frac{\d   f(A,x)}{\d A_{j_1,i_1}}\rangle \\
        = & ~ \langle e_{j_1}, f(A,x) \rangle +\langle A_{*,i_1},  f(A,x)_{j_1} \cdot x_{i_1} (e_{j_1}- f(A,x) ) \rangle \\
        = & ~ f(A,x)_{j_1} + A_{j_1,i_1} f(A,x)_{j_1} x_{i_1} - \langle A_{*,i_1} , f(A,x) \rangle \cdot f(A,x)_{j_1} \cdot x_{i_1}
    \end{align*}
where the first step follows from the product rule of derivative, the second step follows from {\bf Part 3} of Lemma~\ref{def:f}, and the last step follows from simple algebra.

{\bf Proof of Part 2.}
For $i_0 \neq i_1 \in [d]$
    \begin{align*}
        \frac{\d \langle A_{*,i_0}, f(A,x) \rangle}{\d A_{j_1,i_1}} 
        = & ~\langle \frac{\d  A_{*,i_0}}{\d A_{j_1,i_1}},  f(A,x) \rangle + \langle A_{*,i_0},  \frac{\d   f(A,x)}{\d A_{j_1,i_1}}\rangle \\
        = & ~ \langle A_{*,i_0},  f(A,x)_{j_1} \cdot x_{i_1} (e_{j_1}- f(A,x) ) \\
        = & ~ f(A,x)_{j_1} x_{i_1} A_{j_1,i_1} - f(A,x)_{j_1} x_{i_1} \cdot \langle A_{*,i_0} , f(A,x) \rangle
    \end{align*}
where the first step follows from the product rule of derivative, the second step follows from {\bf Part 3} of Lemma~\ref{def:f}, and the last step follows from simple algebra
\end{proof}
\subsection{Gradient for \texorpdfstring{$\langle A_{*,i_0}, f(A,x) \circ c(A,x) \rangle$}{}}

\begin{lemma}\label{lem:gradient_a_f_circ_c}
If the following conditions hold,
\begin{itemize}
    \item Let $A_{*,i} \in \R^n$ denote the $i$-th column of $A \in \R^{n \times d}$.
    \item Let $x \in \R^d$ 
    \item Let $f(A,x) \in \R^n$ be defined as Definition~\ref{def:f}
    \item Let $c(A,x) \in \R^n$ be defined as Definition~\ref{def:c}
    \item Let $q(A,x) = c(A,x) + f(A,x) \in \R^n$
\end{itemize}
then we have
\begin{itemize}
    \item {\bf Part 1.}
    \begin{align*}
         \frac{\d f(A,x) \circ c(A,x)}{\d A_{j_1,i_1}} = \underbrace{f(A,x)_{j_1}}_{\mathrm{scalar}} \cdot \underbrace{x_{i_1}}_{\mathrm{scalar}} (\underbrace{e_{j_1}}_{n \times 1}- \underbrace{f(A,x)}_{n \times 1} ) \circ \underbrace{q(A,x)}_{n \times 1}
    \end{align*}
    \item {\bf Part 2.} For $i_0 = i_1 \in [d]$
    \begin{align*}
        \frac{\d \langle A_{*,i_0}, f(A,x) \circ c(A,x) \rangle}{\d A_{j_1,i_1}} = \underbrace{f(A,x)_{j_1}}_{\mathrm{scalar}} \underbrace{c(A,x)_{j_1}}_{\mathrm{scalar}} + \underbrace{f(A,x)_{j_1}}_{\mathrm{scalar}} \cdot \underbrace{x_{i_1}}_{\mathrm{scalar}} \langle \underbrace{A_{*,i_1}}_{n \times 1}, (\underbrace{e_{j_1}}_{n \times 1}- \underbrace{f(A,x)}_{n \times 1} ) \circ \underbrace{q(A,x)}_{n \times 1} \rangle
    \end{align*}
    \item {\bf Part 3.} For $i_0 \neq i_1 \in [d]$
    \begin{align*}
        \frac{\d \langle A_{*,i_0}, f(A,x) \circ c(A,x) \rangle}{\d A_{j_1,i_1}} = \underbrace{f(A,x)_{j_1}}_{\mathrm{scalar}} \cdot \underbrace{x_{i_1}}_{\mathrm{scalar}} \langle \underbrace{A_{*,i_0}}_{n \times 1}, (\underbrace{e_{j_1}}_{n \times 1}- \underbrace{f(A,x)}_{n \times 1} ) \circ \underbrace{q(A,x)}_{n \times 1}\rangle
    \end{align*}
\end{itemize}
\end{lemma}
\begin{proof}
{\bf Proof of Part 1.}
\begin{align*}
    \frac{\d f(A,x) \circ c(A,x)}{\d A_{j_1,i_1}} 
    = & ~\frac{\d f(A,x)}{\d A_{j_1,i_1}} \circ c(A,x) + \frac{\d c(A,x)}{\d A_{j_1,i_1}} \circ f(A,x) \\
    = & ~ f(A,x)_{j_1} \cdot x_{i_1} (e_{j_1}- f(A,x) ) \circ q(A,x) 
\end{align*}
where the first step follows from {\bf Part 3} of Lemma~\ref{lem:gradient_f} and {\bf Part 3} of Lemma~\ref{lem:gradient_c}.

{\bf Proof of Part 2.}
\begin{align*}
    & ~ \frac{\d \langle A_{*,i_0}, f(A,x) \circ c(A,x) \rangle}{\d A_{j_1,i_1}} \\
    = & ~ \langle \frac{\d  A_{*,i_1}}{\d A_{j_1,i_1}}, f(A,x) \circ c(A,x) \rangle + 
    \langle A_{*,i_1}, \frac{\d f(A,x) \circ c(A,x)}{\d A_{j_1,i_1}}\rangle \\ 
    = & ~ \langle e_{j_1}, f(A,x) \circ c(A,x) \rangle +  \langle A_{*,i_1}, f(A,x)_{j_1} \cdot x_{i_1} (e_{j_1}- f(A,x) ) \circ (c(A,x) + f(A,x)) \rangle \\
    = & ~f(A,x)_{j_1} c(A,x)_{j_1} +  f(A,x)_{j_1} \cdot x_{i_1} \langle A_{*,i_1}, (e_{j_1}- f(A,x) ) \circ q(A,x)  \rangle
\end{align*}
where the first step follows from the Fact~\ref{fac:exponential_der_rule}, the second step follows from {\bf Part 1}, and the last step follows from the definition of $q(A,x)$ and simple algebra.

{\bf Proof of Part 3.}
\begin{align*}
    & ~ \frac{\d \langle A_{*,i_0}, f(A,x) \circ c(A,x) \rangle}{\d A_{j_1,i_1}} \\
    = & ~ \langle \frac{\d  A_{*,i_0}}{\d A_{j_1,i_1}}, f(A,x) \circ c(A,x) \rangle + 
    \langle A_{*,i_0}, \frac{\d f(A,x) \circ c(A,x)}{\d A_{j_1,i_1}}\rangle \\ 
    = & ~ \langle {\bf 0}_n, f(A,x) \circ c(A,x) \rangle +  \langle A_{*,i_0}, f(A,x)_{j_1} \cdot x_{i_1} (e_{j_1}- f(A,x) ) \circ (c(A,x) + f(A,x)) \rangle \\
    = & ~  f(A,x)_{j_1} \cdot x_{i_1} \langle A_{*,i_0}, (e_{j_1}- f(A,x) ) \circ q(A,x)  \rangle
\end{align*}
where the first step follows from the Fact~\ref{fac:exponential_der_rule}, the second step follows from {\bf Part 1}, and the last step follows from the definition of $q(A,x)$ and simple algebra.

\end{proof}
\subsection{Gradient for \texorpdfstring{$q(A,x)$}{}} 
\begin{lemma}\label{lem:gradient_q}

If the following conditions hold
\begin{itemize}
    \item Let $A_{*,i} \in \R^n$ denote the $i$-th column of $A \in \R^{n \times d}$.
    \item Let $x \in \R^d$ 
    \item Let $u(A,x) \in \R^n$ be defined as Definition~\ref{def:u}
    \item Let $\alpha(A,x) \in \R$ be defined as Definition~\ref{def:alpha}
    \item Let $f(A,x) \in \R^n$ be defined as Definition~\ref{def:f}
    \item Let $c(A,x) \in \R^n$ be defined as Definition~\ref{def:c}
    \item Let $g(A,x) \in \R^d$ be defined as Definition~\ref{def:g} 
    \item Let $q(A,x) = c(A,x) + f(A,x) \in \R^n$
\end{itemize}
Then, we have
\begin{itemize}
    \item {\bf Part 1.}
\begin{align*}
    \frac{\d q(A,x)}{\d A_{j_1,i_1}} = 2 f(A,x)_{j_1} \cdot x_{i_1} (e_{j_1} - f(A,x))
\end{align*}
\end{itemize}
\end{lemma}
\begin{proof}
{\bf Proof of Part 1.}
    \begin{align*}
        \frac{\d q(A,x)}{\d A_{j_1,i_1}} 
        = & ~ \frac{\d f(A,x) + c(A,x)}{ \d A_{j_1,i_1}} \\
        = & ~ \frac{\d f(A,x)}{\d A_{j_1,i_1}} + \frac{\d c(A,x)}{\d A_{j_1,i_1}} \\
        = & ~2 f(A,x)_{j_1} \cdot x_{i_1} (e_{j_1} - f(A,x))
    \end{align*}
where the first step follows from the definition of $q(A,x)$, the second step follows from the simple algebra, and the last step follows from Lemma~\ref{lem:gradient_f} and Lemma\ref{lem:gradient_c}.
    \end{proof}
\subsection{Gradient for \texorpdfstring{$g(A,x)$}{}}

\begin{lemma}\label{lem:gradient_g}
If the following conditions hold
\begin{itemize}
    \item Let $A_{*,i} \in \R^n$ denote the $i$-th column of $A \in \R^{n \times d}$.
    \item Let $x \in \R^d$ 
    \item Let $u(A,x) \in \R^n$ be defined as Definition~\ref{def:u}
    \item Let $\alpha(A,x) \in \R$ be defined as Definition~\ref{def:alpha}
    \item Let $f(A,x) \in \R^n$ be defined as Definition~\ref{def:f}
    \item Let $c(A,x) \in \R^n$ be defined as Definition~\ref{def:c}
    \item Let $g(A,x) \in \R^d$ be defined as Definition~\ref{def:g} 
    \item Let $q(A,x) = c(A,x) + f(A,x) \in \R^n$
\end{itemize}
Then, we have
\begin{itemize}
    \item {\bf Part 1.} For $i_0 = i_1 \in [d]$
    \begin{align*}
        & ~ \frac{\d g(A,x)_{i_0} }{\d A_{j_1,i_1}} \\
        = & ~ - (\underbrace{f(A,x)_{j_1}}_{\mathrm{scalar}} + \underbrace{f(A,x)_{j_1}}_{\mathrm{scalar}} \underbrace{x_{i_1}}_{\mathrm{scalar}} \underbrace{A_{j_1,i_1}}_{\mathrm{scalar}} - \underbrace{f(A,x)_{j_1}}_{\mathrm{scalar}} \underbrace{x_{i_1}}_{\mathrm{scalar}} \cdot \langle \underbrace{A_{*,i_1}}_{n \times 1} ,\underbrace{f(A,x)}_{n \times 1} \rangle)\cdot \langle \underbrace{c(A, x)}_{n \times 1}, \underbrace{f(A, x)}_{n \times 1} \rangle \\
    & -  ~  \langle \underbrace{A_{*,i_1}}_{n \times 1}, \underbrace{f(A,x)}_{n \times 1} \rangle \cdot \underbrace{f(A,x)_{j_1}}_{\mathrm{scalar}} \cdot \underbrace{x_{i_1}}_{\mathrm{scalar}} \cdot (\langle - \underbrace{f(A,x)}_{n \times 1}, \underbrace{f(A,x)}_{n \times 1} \rangle +  \underbrace{f(A,x)_{j_1}}_{\mathrm{scalar}}) \\
    & - ~ \langle \underbrace{A_{*,i_1}}_{n \times 1}, \underbrace{f(A,x)}_{n \times 1} \rangle \cdot  \underbrace{f(A,x)_{j_1}}_{\mathrm{scalar}} \cdot \underbrace{x_{i_1}}_{\mathrm{scalar}} \cdot (\langle - \underbrace{f(A,x)}_{n \times 1}, \underbrace{c(A,x)}_{n \times 1} \rangle +  \underbrace{c(A,x)_{j_1}}_{\mathrm{scalar}}) \\
    & - ~ (\underbrace{f(A,x)_{j_1}}_{\mathrm{scalar}} \underbrace{c(A,x)_{j_1}}_{\mathrm{scalar}} + \underbrace{f(A,x)_{j_1}}_{\mathrm{scalar}} \cdot \underbrace{x_{i_1}}_{\mathrm{scalar}} \langle \underbrace{A_{*,i_1}}_{n \times 1}, (\underbrace{e_{j_1}}_{n \times 1}- \underbrace{f(A,x)}_{n \times 1} ) \circ \underbrace{q(A,x)}_{n \times 1} \rangle)
    \end{align*}
    \item {\bf Part 2.} For $i_0 \neq i_1 \in [d]$
    \begin{align*}
        \frac{\d g(A,x)_{i_0} }{ \d A_{j_1,i_1} } = & ~ - (A_{j_1,i_1} f(A,x)_{j_1} x_{i_1} - \langle A_{*,i_0} , f(A,x) \rangle \cdot f(A,x)_{j_1} \cdot x_{i_1} )\cdot \langle c(A, x), f(A, x) \rangle \\
    & -  ~  \langle A_{*,i_0}, f(A,x) \rangle \cdot f(A,x)_{j_1} \cdot x_{i_1} \cdot (\langle - f(A,x), f(A,x) \rangle +  f(A,x)_{j_1}) \\
    & - ~ \langle A_{*,i_0}, f(A,x) \rangle \cdot  f(A,x)_{j_1} \cdot x_{i_1} \cdot (\langle - f(A,x), c(A,x) \rangle +  c(A,x)_{j_1}) \\
    & - ~ f(A,x)_{j_1} \cdot x_{i_1} \langle A_{*,i_0}, (e_{j_1}- f(A,x) ) \circ q(A,x)  \rangle
    \end{align*}
\end{itemize}
\end{lemma}
\begin{proof}
{\bf Proof of Part 1.}
For $i_0 = i_1 \in [d]$

\begin{align*}
    \frac{\d g(A,x)_{i_0} }{\d A_{j_1,i_1}}
    = & ~ - \frac{d (\langle A_{*,i_1}, f(A,x) \rangle  \cdot \langle c(A, x), f(A, x) \rangle +  \langle A_{*,i_1}, f(A, x) \circ c(A, x)\rangle)}{\d A_{j_1,i_1}} \\
    = & ~- \frac{\d \langle A_{*,i}, f(A,x) \rangle}{\d A_{j_1,i_1}} \cdot \langle c(A, x), f(A, x) \rangle + \langle A_{*,i_1}, f(A,x) \rangle \cdot \frac{\d \langle c(A, x), f(A, x) \rangle}{\d A_{j_1,i_1}} \\
    & - ~ \frac{d \langle A_{*,i_1}, f(A, x) \circ c(A, x)\rangle}{\d A_{j_1,i_1}} \\ 
    = & ~ - (f(A,x)_{j_1} + A_{j_1,i_1} f(A,x)_{j_1} x_{i_1} - \langle A_{*,i_1} , f(A,x) \rangle \cdot f(A,x)_{j_1} \cdot x_{i_1} )\cdot \langle c(A, x), f(A, x) \rangle \\
    & -  ~  \langle A_{*,i_1}, f(A,x) \rangle \cdot f(A,x)_{j_1} \cdot x_{i_1} \cdot (\langle - f(A,x), f(A,x) \rangle +  f(A,x)_{j_1}) \\
    & - ~ \langle A_{*,i_1}, f(A,x) \rangle \cdot  f(A,x)_{j_1} \cdot x_{i_1} \cdot (\langle - f(A,x), c(A,x) \rangle +  c(A,x)_{j_1}) \\
    & - ~ (f(A,x)_{j_1} c(A,x)_{j_1} +  f(A,x)_{j_1} \cdot x_{i_1} \langle A_{*,i_1}, (e_{j_1}- f(A,x) ) \circ q(A,x)  \rangle)
\end{align*}
where the first step follows from the definition of $g(A,x)$ (see Definition~\ref{def:g}), the second step follows from product rule of derivative, and the last step comes from {\bf Part 1} of Lemma~\ref{lem:gradient_c_f}, {\bf Part 1} of Lemma~\ref{lem:gradient_a_f}, and {\bf Part 1} of Lemma~\ref{lem:gradient_a_f_circ_c}.

{\bf Proof of Part 2.} For $i_0 \neq i_1 \in [d]$

\begin{align*}
    \frac{\d g(A,x)_{i_0} }{\d A_{j_1,i_1}}
    = & ~ - \frac{d (\langle A_{*,i_0}, f(A,x) \rangle  \cdot \langle c(A, x), f(A, x) \rangle +  \langle A_{*,i_0}, f(A, x) \circ c(A, x)\rangle)}{\d A_{j_1,i_1}} \\
    = & ~- \frac{\d \langle A_{*,i_0}, f(A,x) \rangle}{\d A_{j_1,i_1}} \cdot \langle c(A, x), f(A, x) \rangle + \langle A_{*,i_0}, f(A,x) \rangle \cdot \frac{\d c(A, x), f(A, x)}{\d A_{j_1,i_1}} \\
    & - ~ \frac{d \langle A_{*,i_0}, f(A, x) \circ c(A, x)\rangle}{\d A_{j_1,i_1}} \\
    = & ~ - (A_{j_1,i_1} f(A,x)_{j_1} x_{i_1} - \langle A_{*,i_0} , f(A,x) \rangle \cdot f(A,x)_{j_1} \cdot x_{i_1} )\cdot \langle c(A, x), f(A, x) \rangle \\
    & -  ~  \langle A_{*,i_0}, f(A,x) \rangle \cdot f(A,x)_{j_1} \cdot x_{i_1} \cdot (\langle - f(A,x), f(A,x) \rangle +  f(A,x)_{j_1}) \\
    & - ~ \langle A_{*,i_0}, f(A,x) \rangle \cdot  f(A,x)_{j_1} \cdot x_{i_1} \cdot (\langle - f(A,x), c(A,x) \rangle +  c(A,x)_{j_1}) \\
    & - ~ f(A,x)_{j_1} \cdot x_{i_1} \langle A_{*,i_0}, (e_{j_1}- f(A,x) ) \circ q(A,x)  \rangle
\end{align*}
where the first step follows from the definition of $g(A,x)$ (see Definition~\ref{def:g}), the second step follows from product rule of derivative, and the last step comes from {\bf Part 1} of Lemma~\ref{lem:gradient_c_f}, {\bf Part 2} of Lemma~\ref{lem:gradient_a_f}, and {\bf Part 3} of Lemma~\ref{lem:gradient_a_f_circ_c}.
\end{proof}

\subsection{Gradient for \texorpdfstring{$c_g(A,x)$}{}}

\begin{fact}\label{fac:vectorize_rule}
\begin{itemize}
    \item {\bf Part 1.} If for each $i_0 \in [d]$ $F_{i_0} = \langle A_{*,i_0}, h \rangle$ where $h \in \R^{n}$ is a vector and $A \in \R^{n \times d}$, then we have
    \begin{align*}
        F = \underbrace{ A^\top }_{d \times n} \underbrace{h }_{n \times 1}
    \end{align*}
    \item {\bf Part 2.} If for each $i_0 \in [d]$, $F_{i_0}= A_{j_1,i_0} x_{i_0}$, then
    \begin{align*}
        F = \underbrace{ A_{j_1,*}^\top }_{d \times 1} \circ \underbrace{x }_{ d \times 1}
    \end{align*}
    \item {\bf Part 3.} If for each $i_0 \in [d]$, $F_{i_0}= \langle A_{*,i_0}, h \rangle \cdot x_{i_0}$, then
    \begin{align*}
        F = & ~ \underbrace{\diag(x)}_{d \times d} \underbrace{A^{\top}}_{d \times n} \underbrace{h}_{n \times 1 }
    \end{align*}
\end{itemize}
\end{fact}

\begin{lemma}\label{lem:gradient_c_g}
If the following conditions hold
\begin{itemize}
    \item Let $A_{*,i} \in \R^n$ denote the $i$-th column of $A \in \R^{n \times d}$.
    \item Let $A_{j_1,*} \in \R^d$ for all $j_1 \in [n]$
    \item Let $x \in \R^d$
    \item Let $f(A,x) \in \R^n$ be defined as Definition~\ref{def:f}
    \item Let $c(A,x) \in \R^n$ be defined as Definition~\ref{def:c}
    \item Let $g(A,x) \in \R^d$ be defined as Definition~\ref{def:g} 
    \item Let $q(A,x) = c(A,x) + f(A,x) \in \R^n$
    \item Let $c_g(A,x) \in \R^d$ be defined as Definition~\ref{def:c_g}.
\end{itemize}
Then, we have
\begin{itemize}
    \item {\bf Part 1.} For $i_0 = i_1 \in [d]$
    \begin{align*}
        \frac{\d c_g(A,x)_{i_0} }{\d A_{j_1,i_1}} = & ~ \underbrace{ - (f(A,x)_{j_1} ) \cdot \langle c(A,x) , f(A,x) \rangle }_{ \mathrm{unique~term~1} } \\
        & ~ \underbrace{- ( A_{j_1,i_1} f(A,x)_{j_1} x_{i_1} ) \cdot \langle c(A, x), f(A, x) \rangle }_{ \mathrm{shared~term~1} }\\
        & ~ \underbrace{+ \langle A_{*,i_1} , f(A,x) \rangle \cdot f(A,x)_{j_1} \cdot x_{i_1} \cdot \langle c(A, x), f(A, x) \rangle }_{ \mathrm{shared~term~2} } \\
    & ~ \underbrace{ -  \langle A_{*,i_1}, f(A,x) \rangle \cdot f(A,x)_{j_1} \cdot x_{i_1} \cdot (\langle - f(A,x), f(A,x) \rangle +  f(A,x)_{j_1}) }_{ \mathrm{shared~term~3} } \\
    & ~ \underbrace{ - \langle A_{*,i_1}, f(A,x) \rangle \cdot  f(A,x)_{j_1} \cdot x_{i_1} \cdot (\langle - f(A,x), c(A,x) \rangle +  c(A,x)_{j_1}) }_{ \mathrm{shared~term~4} } \\
    & ~ \underbrace{ - f(A,x)_{j_1} c(A,x)_{j_1} }_{ \mathrm{unique~term~2} } \\
    & ~ \underbrace{ - f(A,x)_{j_1} \cdot x_{i_1} \cdot \langle A_{*,i_1}, (e_{j_1}- f(A,x) ) \circ q(A,x)  \rangle }_{ \mathrm{shared~term~5} }
    \end{align*}
    \item {\bf Part 2.} For $i_0 \neq i_1 \in [d]$
    \begin{align*}
        \frac{\d c_g(A,x)_{i_0} }{ \d A_{j_1,i_1} } = & ~ - \underbrace{ (A_{j_1,i_1} f(A,x)_{j_1} x_{i_1} ) \cdot \langle c(A, x), f(A, x) \rangle}_{\mathrm{shared~term~1}} \\
        & ~ + \underbrace{\langle A_{*,i_0} , f(A,x) \rangle \cdot f(A,x)_{j_1} \cdot x_{i_1} \cdot \langle c(A, x), f(A, x) \rangle}_{\mathrm{shared~term~2}} \\
    & -  ~  \underbrace{\langle A_{*,i_0}, f(A,x) \rangle \cdot f(A,x)_{j_1} \cdot x_{i_1} \cdot (\langle - f(A,x), f(A,x) \rangle +  f(A,x)_{j_1})}_{\mathrm{shared~term~3}} \\
    & - ~ \underbrace{\langle A_{*,i_0}, f(A,x) \rangle \cdot  f(A,x)_{j_1} \cdot x_{i_1} \cdot (\langle - f(A,x), c(A,x) \rangle +  c(A,x)_{j_1})}_{\mathrm{shared~term~4}} \\
    & - ~ 
    \underbrace{f(A,x)_{j_1} \cdot x_{i_1} \langle A_{*,i_0}, (e_{j_1}- f(A,x) ) \circ q(A,x)  \rangle}_{\mathrm{shared~term~5}}
    \end{align*}
    \item {\bf Part 3.} 
    \begin{align*}
        \frac{ \d c_g(A,x) }{ \d A_{j_1,i_1} } 
        = & ~ \underbrace{ - f(A,x)_{j_1}  \cdot \langle c(A,x), f(A,x) \rangle \cdot e_{i_1} }_{ \mathrm{unique~term~1} } \\
        & ~ \underbrace{ - f(A,x)_{j_1} \cdot c(A,x)_{j_1} \cdot e_{i_1} }_{\mathrm{unique~term~2}} \\
        & ~ \underbrace{- f(A,x)_{j_1} \cdot \langle c(A,x), f(A,x) \rangle \cdot ( (A_{j_1,*})^\top \circ x ) }_{ \mathrm{shared~term~1} } \\
        & ~ \underbrace{+ \diag (x) A^{\top} \cdot   f(A,x)    \cdot f(A,x)_{j_1} \cdot    \langle c(A, x), f(A, x) \rangle }_{ \mathrm{shared~term~2} } \\
        & ~ \underbrace{ -  f(A,x)_{j_1}\diag(x) A^{\top} f(A,x) \cdot ( \langle -f(A,x), f(A,x) \rangle + f(A,x)_{j_1}) }_{ \mathrm{shared~term~3}  }\\
        & ~ \underbrace{ -   f(A,x)_{j_1}\diag(x) A^{\top} f(A,x) \cdot(- \langle f(A,x), c(A,x) \rangle + f(A,x)_{j_1}) }_{ \mathrm{shared~term~4}  }\\
        & ~ \underbrace{ - f(A,x)_{j_1}\diag(x) A^{\top} ((e_{j_1} - f(A,x)) \circ q(A,x)) }_{ \mathrm{shared~term~5}  }
    \end{align*}
\end{itemize}
\end{lemma}
\begin{proof}
{\bf Proof of Part 1}
For $i_0 = i_1 \in [d]$
\begin{align*}
    \frac{\d c_g(A,x)_{i_0} }{\d A_{j_1,i_1}} 
    = &  ~ \frac{\d g(A,x)_{i_1} -(b_g)_{i_1}}{\d A_{j_1,i_1}}   \\
     = & ~ - (f(A,x)_{j_1} + A_{j_1,i_1} f(A,x)_{j_1} x_{i_1} - \langle A_{*,i_1} , f(A,x) \rangle \cdot f(A,x)_{j_1} \cdot x_{i_1} )\cdot \langle c(A, x), f(A, x) \rangle \\
    & -  ~  \langle A_{*,i_1}, f(A,x) \rangle \cdot f(A,x)_{j_1} \cdot x_{i_1} \cdot (\langle - f(A,x), f(A,x) \rangle +  f(A,x)_{j_1}) \\
    & - ~ \langle A_{*,i_1}, f(A,x) \rangle \cdot  f(A,x)_{j_1} \cdot x_{i_1} \cdot (\langle - f(A,x), c(A,x) \rangle +  c(A,x)_{j_1}) \\
    & - ~ (f(A,x)_{j_1} c(A,x)_{j_1} +  f(A,x)_{j_1} \cdot x_{i_1} \langle A_{*,i_1}, (e_{j_1}- f(A,x) ) \circ q(A,x)  \rangle)
\end{align*}
where the first step follows from the definition of $c_g(A,x)$ (see Definition~\ref{def:c_g}), and the second step comes from the {\bf Part 1} of Lemma~\ref{lem:gradient_g}

{\bf Proof of Part 2}
For $i_0 \neq i_1 \in [d]$
\begin{align*}
    \frac{\d c_g(A,x)_{i_0} }{\d A_{j_1,i_1}} 
    = & ~ \frac{\d g(A,x)_{i_0} -(b_g)_{i_0}}{\d A_{j_1,i_1}} \\
     = & ~ - (A_{j_1,i_1} f(A,x)_{j_1} x_{i_1} - \langle A_{*,i_0} , f(A,x) \rangle \cdot f(A,x)_{j_1} \cdot x_{i_1} )\cdot \langle c(A, x), f(A, x) \rangle \\
    & -  ~  \langle A_{*,i_0}, f(A,x) \rangle \cdot f(A,x)_{j_1} \cdot x_{i_1} \cdot (\langle - f(A,x), f(A,x) \rangle +  f(A,x)_{j_1}) \\
    & - ~ \langle A_{*,i_0}, f(A,x) \rangle \cdot  f(A,x)_{j_1} \cdot x_{i_1} \cdot (\langle - f(A,x), c(A,x) \rangle +  c(A,x)_{j_1}) \\
    & - ~ f(A,x)_{j_1} \cdot x_{i_1} \langle A_{*,i_0}, (e_{j_1}- f(A,x) ) \circ q(A,x)  \rangle
\end{align*}
where the first step follows from the definition of $c_g(A,x)$ (see Definition~\ref{def:c_g}), and the second step comes from the {\bf Part 2} of Lemma~\ref{lem:gradient_g}

{\bf Proof of Part 3}
For unique term 1, we can write its vector form as
\begin{align*}
    - \underbrace{f(A,x)_{j_1}}_{\mathrm{scalar}}  \cdot \langle \underbrace{c(A,x)}_{n \times 1}, \underbrace{f(A,x)}_{n \times 1} \rangle \cdot \underbrace{e_{i_1}}_{d \times 1}
\end{align*}

For unique term 2, 
\begin{align*}
    - \underbrace{f(A,x)_{j_1}}_{\mathrm{scalar}} \cdot \underbrace{c(A,x)_{j_1}}_{\mathrm{scalar}} \cdot \underbrace{e_{i_1}}_{d \times 1}
\end{align*}

For shared term 1, we have
\begin{align*}
    -\underbrace{f(A,x)_{j_1}}_{\mathrm{scalar}} \cdot \underbrace{\langle c(A,x), f(A,x)\rangle}_{\mathrm{scalar}} \cdot \underbrace{((A_{j_1,*})^{\top}}_{d \times 1} \circ \underbrace{x}_{d \times 1})
\end{align*} 
where the first step follows from {\bf Part 2} of Fact~\ref{fac:vectorize_rule}.

For shared term 2,
\begin{align*}
    \underbrace{f(A,x)_{j_1}}_{\mathrm{scalar}} \underbrace{\diag (x)}_{d \times d} \underbrace{A^{\top}}_{d \times n}   \underbrace{f(A,x)}_{n \times 1}  \cdot    \underbrace{\langle c(A, x), f(A, x) \rangle}_{\mathrm{scalar}} 
\end{align*}
where the first step follows from {\bf Part 3} of Fact~\ref{fac:vectorize_rule}.

For shared term 3, we have
\begin{align*}
    -
    \underbrace{f(A,x)_{j_1}}_{\mathrm{scalar}} \underbrace{\diag (x)}_{d \times d} \underbrace{A^{\top}}_{d \times n}   \underbrace{f(A,x)}_{n \times 1}  (\underbrace{\langle -f(A,x), f(A,x)\rangle}_{\mathrm{scalar}} + \underbrace{f(A,x)_{j_1}}_{\mathrm{scalar}}) 
\end{align*}
where the first step follows from {\bf Part 3} of Fact~\ref{fac:vectorize_rule}.

For shared term 4, we have
\begin{align*}
     -    \underbrace{f(A,x)_{j_1}}_{\mathrm{scalar}} \underbrace{\diag (x)}_{d \times d} \underbrace{A^{\top}}_{d \times n}   \underbrace{f(A,x)}_{n \times 1}   (\underbrace{\langle-f(A,x), c(A,x) \rangle}_{\mathrm{scalar}} + \underbrace{f(A,x)_{j_1}}_{\mathrm{scalar}}) 
\end{align*}
where the first step follows from {\bf Part 3} of Fact~\ref{fac:vectorize_rule}.

For shared term 5, we have
\begin{align*}
    - \underbrace{f(A,x)_{j_1}}_{\mathrm{scalar}} \underbrace{\diag (x)}_{d \times d} \underbrace{A^{\top}}_{d \times n}     ((\underbrace{e_{j_1}}_{n \times 1} - \underbrace{f(A,x)}_{n \times 1}) \circ \underbrace{q(A,x)}_{n \times 1}) 
\end{align*}
where the first step follows from {\bf Part 3} of Fact~\ref{fac:vectorize_rule}.

By combining above equations, we have
\begin{align*}
     \frac{ \d c_g(A,x) }{ \d A_{j_1,i_1} } 
      = & ~ - f(A,x)_{j_1}  \cdot \langle c(A,x), f(A,x) \rangle \cdot e_{i_1} \\
        & ~ - f(A,x)_{j_1} \cdot c(A,x)_{j_1} \cdot e_{i_1} \\
        & ~ - f(A,x)_{j_1} \cdot \langle c(A,x), f(A,x) \rangle \cdot ( (A_{j_1,*})^\top \circ x )\\
        & ~ +  f(A,x)_{j_1}\diag (x) A^{\top}   f(A,x)  \cdot    \langle c(A, x), f(A, x) \rangle \\
        & ~ -  f(A,x)_{j_1}\diag(x) A^{\top} f(A,x) \cdot (\langle -f(A,x), f(A,x)\rangle + f(A,x)_{j_1}) \\
        & ~ -   f(A,x)_{j_1}\diag(x) A^{\top} f(A,x) \cdot(\langle-f(A,x), c(A,x) \rangle + f(A,x)_{j_1}) \\
        & ~ - f(A,x)_{j_1}\diag(x) A^{\top} ((e_{j_1} - f(A,x)) \circ q(A,x)) 
\end{align*}
where the first step follows from {\bf Part 1 and 2}. 
\end{proof}

\subsection{Gradient for \texorpdfstring{$L_g(A,x)$}{}}

\begin{lemma}\label{lem:gradient_lg}
If the following conditions hold
\begin{itemize}
    \item Let $A_{j_1,*} \in \R^d$ for all $j_1 \in [n]$
    \item Let $x \in \R^d$
    \item Let $f(A,x) \in \R^n$ be defined as Definition~\ref{def:f}
    \item Let $c(A,x) \in \R^n$ be defined as Definition~\ref{def:c}
    \item Let $g(A,x) \in \R^d$ be defined as Definition~\ref{def:g} 
    \item Let $q(A,x) = c(A,x) + f(A,x) \in \R^n$
    \item Let $c_g(A,x) \in \R^d$ be defined as Definition~\ref{def:c_g}.
    \item Let $L_g(A,x) \in \R$ be defined as Definition~\ref{def:l_g}
\end{itemize}
Then, we have
\begin{itemize}
    \item
    \begin{align*}
        \frac{\d L_g(A,x) }{\d A_{j_1,i_1}} 
   = & ~ -  c_g(A)^{\top} \cdot f(A,x)_{j_1}  \cdot \langle c(A,x), f(A,x) \rangle \cdot e_{i_1} \\
        & ~ - c_g(A,x)^{\top} \cdot f(A,x)_{j_1} \cdot c(A,x)_{j_1} \cdot e_{i_1} \\
        & ~ - c_g(A,x)^{\top} \cdot f(A,x)_{j_1} \cdot \langle c(A,x), f(A,x) \rangle \cdot ( (A_{j_1,*})^\top \circ x )\\
        & ~ + c_g(A,x)^{\top} \cdot f(A,x)_{j_1}\diag (x) A^{\top}   f(A,x)  \cdot    \langle c(A, x), f(A, x) \rangle \\
        & ~ - c_g(A,x)^{\top} \cdot f(A,x)_{j_1}\diag(x) A^{\top} f(A,x) \cdot (\langle -f(A,x), f(A,x) \rangle + f(A,x)_{j_1}) \\
        & ~ - c_g(A,x)^{\top} \cdot  f(A,x)_{j_1}\diag(x) A^{\top} f(A,x) \cdot(\langle -f(A,x), c(A,x) \rangle + f(A,x)_{j_1}) \\
        & ~ - c_g(A,x)^{\top} \cdot f(A,x)_{j_1}\diag(x) A^{\top} ((e_{j_1} - f(A,x)) \circ q(A,x)) 
    \end{align*} 
\end{itemize}
\end{lemma}
\begin{proof}
\begin{align*}
   \frac{\d L_g(A,x) }{\d A_{j_1,i_1}} = & ~ 0.5 \| c_g(A,x) \|_2^2 \\
   = & ~  \langle c_g(A,x), \frac{ \d c_g(A,x) }{\d A_{j_1,i_1}} \rangle\\
   = & ~ -  c_g(A,x)^{\top} \cdot f(A,x)_{j_1}  \cdot \langle c(A,x), f(A,x) \rangle \cdot e_{i_1} \\
        & ~ - c_g(A,x)^{\top} \cdot f(A,x)_{j_1} \cdot c(A,x)_{j_1} \cdot e_{i_1} \\
        & ~ - c_g(A,x)^{\top} \cdot f(A,x)_{j_1} \cdot \langle c(A,x), f(A,x) \rangle \cdot ( (A_{j_1,*})^\top \circ x )\\
        & ~ + c_g(A,x)^{\top} \cdot f(A,x)_{j_1}\diag (x) A^{\top}   f(A,x)  \cdot    \langle c(A, x), f(A, x) \rangle \\
        & ~ - c_g(A,x)^{\top} \cdot f(A,x)_{j_1}\diag(x) A^{\top} f(A,x) \cdot (\langle -f(A,x), f(A,x) \rangle + f(A,x)_{j_1}) \\
        & ~ - c_g(A,x)^{\top} \cdot  f(A,x)_{j_1}\diag(x) A^{\top} f(A,x) \cdot(\langle -f(A,x), c(A,x) \rangle + f(A,x)_{j_1}) \\
        & ~ - c_g(A,x)^{\top} \cdot f(A,x)_{j_1}\diag(x) A^{\top} ((e_{j_1} - f(A,x)) \circ q(A,x)) 
\end{align*}
where the first step follows from the definition of $L_g(A,x)$ (see Definition~\ref{def:l_g}), the second step follows from Fact~\ref{fac:exponential_der_rule}, and the last step follows from {\bf Part 3} of Lemma~\ref{lem:gradient_c_g}.
\end{proof}

\section{Lipschitz for Gradient}\label{app:lips_gradient}

\begin{lemma}\label{lem:lipschitz_grad}
We can show
\begin{itemize}
    \item Part 1. For each $j_1 \in [n]$, for each $i_1 \in [d]$,
    \begin{align*}
        | \frac{\d L_g(A,x)}{\d A_{j_1,i_1}} - \frac{\d L_g(\wt{A},x)}{\d A_{j_1,i_1}} | \leq 10 \beta^{-2} \cdot n   \exp(6 R^2)\|A - \wt{A}\|_F
    \end{align*}
    \item Part 2.
    \begin{align*}
        \| \frac{\d L_g(A,x)}{\d A} - \frac{\d L_g(\wt{A},x)}{\d A} \|_F \leq 10 \beta^{-2} \cdot n^{1.5} \cdot d^{0.5}  \exp(6 R^2)\|A - \wt{A}\|_F
    \end{align*}
\end{itemize}
\end{lemma}
    
\begin{proof}
{\bf Proof of Part 1.}
\begin{align*}
    & ~ | \frac{\d L_g(A,x)}{\d A_{j_1,i_1}} - \frac{\d L_g(\wt{A},x)}{\d A_{j_1,i_1}}  | \\
    \leq & ~| - c_g(A,x)^{\top} \cdot f(A,x)_{j_1}  \cdot \langle c(A,x), f(A,x) \rangle \cdot e_{i_1} - (-c_g(\wt{A},x)^{\top} \cdot f(\wt{A},x)_{j_1}  \cdot \langle c(\wt{A},x), f(\wt{A},x) \rangle \cdot e_{i_1} )| \\
      +  & ~ |- c_g(A,x)^{\top} \cdot f(A,x)_{j_1} \cdot c(A,x)_{j_1} \cdot e_{i_1} -( - c_g(\wt{A},x)^{\top} \cdot f(\wt{A},x)_{j_1} \cdot c(\wt{A},x)_{j_1} \cdot e_{i_1}) | \\
       + & ~ | - c_g(A,x)^{\top} \cdot f(A,x)_{j_1} \cdot \langle c(A,x), f(A,x) \rangle \cdot ( (A_{j_1,*})^\top \circ x ) \\
       & - ~ (c_g(\wt{A},x)^{\top} \cdot f(\wt{A},x)_{j_1} \cdot \langle c(\wt{A},x), f(\wt{A},x) \rangle \cdot ( (\wt{A}_{j_1,*})^\top \circ x ))|\\
       + & ~ | c_g(A,x)^{\top} \cdot f(A,x)_{j_1}\diag (x) A^{\top}   f(A,x)  \cdot    \langle c(A, x), f(A, x) \rangle \\
       & ~ - c_g(\wt{A},x)^{\top} \cdot f(\wt{A},x)_{j_1}\diag (x) \wt{A}^{\top}   f(\wt{A},x)  \cdot    \langle c(\wt{A}, x), f(\wt{A}, x) \rangle | \\
        +  & ~ |- c_g(A,x)^{\top} \cdot f(A,x)_{j_1}\diag(x) A^{\top} f(A,x) \cdot (\langle -f(A,x), f(A,x) \rangle + f(A,x)_{j_1}) \\
        & - ~( -c_g(\wt{A},x)^{\top} \cdot f(\wt{A},x)_{j_1}\diag(x) \wt{A}^{\top} f(\wt{A},x) \cdot (\langle -f(\wt{A},x), f(\wt{A},x) \rangle + f(\wt{A},x)_{j_1}) )|\\
        + & ~ |- c_g(A,x)^{\top} \cdot  f(A,x)_{j_1}\diag(x) A^{\top} f(A,x) \cdot(\langle -f(A,x), c(A,x) \rangle + f(A,x)_{j_1}) \\
         & ~ -(- c_g(\wt{A},x)^{\top} \cdot  f(\wt{A},x)_{j_1}\diag(x) \wt{A}^{\top} f(\wt{A},x) \cdot(\langle -f(\wt{A},x), c(\wt{A},x) \rangle + f(\wt{A},x)_{j_1}))| \\
        + & ~ |- c_g(A,x)^{\top} \cdot f(A,x)_{j_1}\diag(x) A^{\top} ((e_{j_1} - f(A,x)) \circ q(A,x)) \\
        & ~ -(- c_g(\wt{A},x)^{\top} \cdot f(\wt{A},x)_{j_1}\diag(x) \wt{A}^{\top} ((e_{j_1} - f(\wt{A},x)) \circ q(\wt{A},x)))| \\
        \leq & ~ \beta^{-2} \cdot n   \exp(5R^2)\|A - \wt{A}\|_F\\
        + & ~ \beta^{-2} \cdot n   \exp(5R^2)\|A - \wt{A}\|_F\\
        + & ~ \beta^{-2} \cdot n   \exp(6 R^2)\|A - \wt{A}\|_F \\
        + & ~ \beta^{-2} \cdot n   \exp(6 R^2)\|A - \wt{A}\|_F \\
        + & ~ 2 \beta^{-2} \cdot n   \exp(6 R^2)\|A - \wt{A}\|_F \\
        + & ~ 2 \beta^{-2} \cdot n   \exp(6 R^2)\|A - \wt{A}\|_F \\
        + & ~ 2 \beta^{-2} \cdot n   \exp(6 R^2)\|A - \wt{A}\|_F \\
        \leq & ~  10 \beta^{-2} \cdot n   \exp(6 R^2)\|A - \wt{A}\|_F
\end{align*}
where the first step follows from Lemma~\ref{lem:product_rule}, and the second step follows from Lemma~\ref{lem:lips_gradient_1}, Lemma~\ref{lem:lips_gradient_2}, Lemma~\ref{lem:lips_gradient_3}, Lemma~\ref{lem:lips_gradient_4}, Lemma~\ref{lem:lips_gradient_5}, Lemma~\ref{lem:lips_gradient_6}, and Lemma~\ref{lem:lips_gradient_7}, and the last step follows from $\beta^{-1} > 1, n> 1, R > 4$ 

{\bf Proof of Part 2.}
\begin{align*}
    \| \frac{\d L_g(A,x)}{\d A} - \frac{\d L_g(\wt{A},x)}{\d A} \|_F^2 = & ~ \sum_{j_1=1}^n \sum_{i_1=1}^d | \frac{\d L_g(A,x)}{\d A_{j_1,i_1}} - \frac{\d L_g(\wt{A},x)}{\d A_{j_1,i_1}} |^2 \\
    \leq & ~ nd \cdot (10 \beta^{-2} \cdot n   \exp(6 R^2)\|A - \wt{A}\|_F)^2 .
\end{align*}
Taking the square root on both sides, then we complete the proof.
\end{proof}

\subsection{Product Tools}
\begin{lemma}\label{lem:product_rule}
Given a collection of functions $f_1(A), \cdots, f_n(A) \in \R$, we have
\begin{align*}
| \prod_{i=1}^n f_i(A) - \prod_{i=1}^n f_i(\wt{A}) \|_2 \leq \sum_{l=1}^{n} | ( \prod_{i=0}^l f_i(A) ) ( \prod_{j=l+1}^{n+1} f_j( \wt{A} ) ) - ( \prod_{i=0}^{l-1} f_i(A) ) ( \prod_{j=l}^{n+1} f_j( \wt{A} ) ) |
\end{align*}
For convenient, let $f_0(A)=  f_{n+1}(A) = 1$, for all $A$.
\end{lemma}
\begin{remark}
For example, when $n=2$, it is easy to see that the above equation becomes
\begin{align*}
|f_1(A) f_2(A) - f_1(\wt{A}) f_2( \wt{A}) |
\leq | f_1(A) f_2(A) - f_1(A) f_2(\wt{A}) | + | f_1(A) f_2(\wt{A}) - f_1(\wt{A}) f_2(\wt{A})  |
\end{align*}
\end{remark}

\subsection{First term \texorpdfstring{$-c_g(A,x)^{\top} \cdot f(A,x)_{j_1}  \cdot \langle c(A,x), f(A,x) \rangle \cdot e_{i_1}$}{}}

\begin{lemma}\label{lem:lips_gradient_1}
If the following conditions hold
\begin{itemize}
    \item Let $A_{j_1,*} \in \R^d$ for all $j_1 \in [n]$
    \item Let $x \in \R^d$
    \item Let $f(A,x) \in \R^n$ be defined as Definition~\ref{def:f}
    \item Let $c(A,x) \in \R^n$ be defined as Definition~\ref{def:c}
    \item Let $g(A,x) \in \R^d$ be defined as Definition~\ref{def:g} 
    \item Let $q(A,x) = c(A,x) + f(A,x) \in \R^n$
    \item Let $c_g(A,x) \in \R^d$ be defined as Definition~\ref{def:c_g}.
\end{itemize}
Then we have
\begin{align*}
    & ~ | -c_g(A,x)^{\top} \cdot f(A,x)_{j_1}  \cdot \langle c(A,x), f(A,x) \rangle \cdot e_{i_1} -(- c_g( \wt{A},x)^{\top} \cdot f( \wt{A},x)_{j_1}  \cdot \langle c( \wt{A},x), f( \wt{A} ,x) \rangle \cdot e_{i_1}) | \\
    \leq & ~ \beta^{-2} \cdot n   \exp(5R^2)\|A - \wt{A}\|_F
\end{align*}
\end{lemma}
\begin{proof}
Firstly, we have
    \begin{align*}
       & ~ | -c_g(A,x)^{\top} \cdot f(A,x)_{j_1}  \cdot \langle c(A,x), f(A,x) \rangle \cdot e_{i_1} - (-c_g( \wt{A},x)^{\top} \cdot f( \wt{A},x)_{j_1}  \cdot \langle c( \wt{A},x), f( \wt{A} ,x) \rangle \cdot e_{i_1} )| \\
       = & 
       ~| c_g(A,x)^{\top} \cdot f(A,x)_{j_1}  \cdot \langle c(A,x), f(A,x) \rangle \cdot e_{i_1} - c_g( \wt{A},x)^{\top} \cdot f( \wt{A},x)_{j_1}  \cdot \langle c( \wt{A},x), f( \wt{A} ,x) \rangle \cdot e_{i_1} | \\
       \leq & ~ | c_g(A,x)^{\top} \cdot f(A,x)_{j_1}  \cdot \langle c(A,x), f(A,x) \rangle \cdot e_{i_1} -c_g(\wt{A},x)^{\top} \cdot f(A,x)_{j_1}  \cdot \langle c(A,x), f(A,x) \rangle \cdot e_{i_1} | \\
       & + ~ | c_g(\wt{A},x)^{\top} \cdot f(A,x)_{j_1}  \cdot \langle c(A,x), f(A,x) \rangle \cdot e_{i_1} -c_g(\wt{A},x)^{\top} \cdot f(\wt{A},x)_{j_1}  \cdot \langle c(A,x), f(A,x) \rangle \cdot e_{i_1} | \\
       & + ~ | c_g(\wt{A},x)^{\top} \cdot f(\wt{A},x)_{j_1}  \cdot \langle c(A,x), f(A,x) \rangle \cdot e_{i_1} -c_g(\wt{A},x)^{\top} \cdot f(\wt{A},x)_{j_1}  \cdot \langle c(\wt{A},x), f(A,x) \rangle \cdot e_{i_1} | \\
       & + ~ | c_g(\wt{A},x)^{\top} \cdot f(\wt{A},x)_{j_1}  \cdot \langle c(\wt{A},x), f(A,x) \rangle \cdot e_{i_1} -c_g(\wt{A},x)^{\top} \cdot f(\wt{A},x)_{j_1}  \cdot \langle c(\wt{A},x), f(\wt{A},x) \rangle \cdot e_{i_1} | 
    \end{align*}
where the first step follows from Fact~\ref{fac:vector_norm}, and the second step follows from Lemma~\ref{lem:product_rule}.
    
For the first term, we have 
\begin{align*}
    & ~ | c_g(A,x)^{\top} \cdot f(A,x)_{j_1}  \cdot \langle c(A,x), f(A,x) \rangle \cdot e_{i_1} -c_g(\wt{A},x)^{\top} \cdot f(A,x)_{j_1}  \cdot \langle c(A,x), f(A,x) \rangle \cdot e_{i_1} | \\
    \leq & ~ |c_g(A,x)^{\top} - c_g(\wt{A},x)^{\top} | \cdot |f(A,x)_{j_1} | \cdot  \|c(A,x) \|_2  \cdot \|f(A,x) \|_2  \cdot  \|e_{i_1} \|_2 \\
    \leq & ~  40\beta^{-2} \cdot n \cdot R \cdot  \exp(3R^{2}) \cdot \|A - \wt{A}\|_F 
\end{align*}
where the first step follows from Fact~\ref{fac:vector_norm},  and the last step follows from {\bf Part 8} of \ref{lem:basic_lips}, and {\bf Part 3, 4, 7} of \ref{lem:upper_bound}.

For the second term, we have
\begin{align*}
    & ~ | c_g(\wt{A},x)^{\top} \cdot f(A,x)_{j_1}  \cdot \langle c(A,x), f(A,x) \rangle \cdot e_{i_1} -c_g(\wt{A},x)^{\top} \cdot f(\wt{A},x)_{j_1}  \cdot \langle c(A,x), f(A,x) \rangle \cdot e_{i_1} | \\
    \leq & ~ \| c_g(\wt{A},x)^{\top}\|_2 |f(A,x)_{j_1}  - f(\wt{A},x)_{j_1}|\cdot  \|c(A,x) \|_2  \cdot \|f(A,x) \|_2 \|e_{i_1} \|_2 \\
    \leq & ~ 20     \beta^{-2} \cdot n   R \exp(3R^2)\|A - \wt{A}\|_F  
\end{align*}
where the first step follows from Fact~\ref{fac:vector_norm}, and the second step follows from {\bf Part 10} of Lemma~\ref{lem:basic_lips} and {\bf Part 3, 4, and 6} of Lemma~\ref{lem:upper_bound}.

For the third term, we have 
\begin{align*}
    & ~| c_g(\wt{A},x)^{\top} \cdot f(\wt{A},x)_{j_1}  \cdot \langle c(A,x), f(A,x) \rangle \cdot e_{i_1} -c_g(\wt{A},x)^{\top} \cdot f(\wt{A},x)_{j_1}  \cdot \langle c(\wt{A},x), f(A,x) \rangle \cdot e_{i_1} | \\
    \leq & ~ \|c_g(\wt{A},x)^{\top} \|_2 \cdot|f(\wt{A},x)_{j_1}|\cdot \|c(A,x) - c(\wt{A},x) \|_2\cdot \|f(A,x) \|_2 \cdot \|e_{i_1} \|_2 \\
    \leq & ~ 10 \beta^{-2} \cdot n   R \exp(3R^2)\|A - \wt{A}\|_F  
\end{align*}
where the first step follows from Fact~\ref{fac:vector_norm}, and the second step follows from {\bf Part 5} of Lemma~\ref{lem:basic_lips} and {\bf Part 3, 6, and 7} of Lemma~\ref{lem:upper_bound}.

For the fourth term, we have 
\begin{align*}
    & ~ | c_g(\wt{A},x)^{\top} \cdot f(\wt{A},x)_{j_1}  \cdot \langle c(\wt{A},x), f(A,x) \rangle \cdot e_{i_1} -c_g(\wt{A},x)^{\top} \cdot f(\wt{A},x)_{j_1}  \cdot \langle c(\wt{A},x), f(\wt{A},x) \rangle \cdot e_{i_1} | \\
    \leq & ~ \|c_g(\wt{A},x)^{\top} \|_2 \cdot|f(\wt{A},x)_{j_1}|\cdot \| c(\wt{A},x) \|_2\cdot \|f(A,x)  - f(\wt{A},x)\|_2 \cdot \|e_{i_1} \|_2 \\
    \leq & ~ 20 \beta^{-2} \cdot n   R \exp(3R^2)\|A - \wt{A}\|_F 
\end{align*}
where the first step follows from Fact~\ref{fac:vector_norm}, and the second step follows from {\bf Part 5} of Lemma~\ref{lem:basic_lips} and {\bf Part 4, 6, and 7} of Lemma~\ref{lem:upper_bound}.

Then, we have
\begin{align*}
     &  ~| c_g(A,x)^{\top} \cdot f(A,x)_{j_1}  \cdot \langle c(A,x), f(A,x) \rangle \cdot e_{i_1} - c_g( \wt{A},x)^{\top} \cdot f( \wt{A},x)_{j_1}  \cdot \langle c( \wt{A},x), f( \wt{A} ,x) \rangle \cdot e_{i_1} | \\
     \leq & ~ 40\beta^{-2} \cdot n R \cdot  \exp(3R^{2}) \cdot \|A - \wt{A}\|_F \\
     & + ~ 20 \beta^{-2} \cdot n   R \exp(3R^2)\|A - \wt{A}\|_F \\
      & + ~ 10 \beta^{-2} \cdot n   R \exp(3R^2)\|A - \wt{A}\|_F \\
       & + ~ 20 \beta^{-2} \cdot n   R \exp(3R^2)\|A - \wt{A}\|_F \\ 
       \leq & ~  \beta^{-2} \cdot n   \exp(5R^2)\|A - \wt{A}\|_F
\end{align*}
where the second step follows from $R \leq \exp(R^2)$ and $R \geq 4$.
\end{proof}

\subsection{Second term \texorpdfstring{$- c_g(A,x)^{\top} \cdot f(A,x)_{j_1} \cdot c(A,x)_{j_1} \cdot e_{i_1} $}{}}
\begin{lemma}\label{lem:lips_gradient_2}
If the following conditions hold
\begin{itemize}
    \item Let $A_{j_1,*} \in \R^d$ for all $j_1 \in [n]$
    \item Let $x \in \R^d$
    \item Let $f(A,x) \in \R^n$ be defined as Definition~\ref{def:f}
    \item Let $c(A,x) \in \R^n$ be defined as Definition~\ref{def:c}
    \item Let $g(A,x) \in \R^d$ be defined as Definition~\ref{def:g} 
    \item Let $q(A,x) = c(A,x) + f(A,x) \in \R^n$
    \item Let $c_g(A,x) \in \R^d$ be defined as Definition~\ref{def:c_g}.
\end{itemize}
Then we have
\begin{align*}
 & ~|- c_g(A,x)^{\top} \cdot f(A,x)_{j_1} \cdot c(A,x)_{j_1} \cdot e_{i_1} - (- c_g(\wt{A},x)^{\top} \cdot f(\wt{A},x)_{j_1} \cdot c(\wt{A},x)_{j_1} \cdot e_{i_1}) | \\
 \leq & ~   \beta^{-2} \cdot n   \exp(5R^2)\|A - \wt{A}\|_F
\end{align*}
\end{lemma}
\begin{proof}
    Firstly, we have
    \begin{align*}
        & ~  |- c_g(A,x)^{\top} \cdot f(A,x)_{j_1} \cdot c(A,x)_{j_1} \cdot e_{i_1} - (- c_g(\wt{A},x)^{\top} \cdot f(\wt{A},x)_{j_1} \cdot c(\wt{A},x)_{j_1} \cdot e_{i_1}) | \\
        \leq & ~  |- c_g(A,x)^{\top} \cdot f(A,x)_{j_1} \cdot c(A,x)_{j_1} \cdot e_{i_1} - (- c_g(\wt{A},x)^{\top} \cdot f(A,x)_{j_1} \cdot c(A,x)_{j_1} \cdot e_{i_1}) | \\
        & + ~ |- c_g(\wt{A},x)^{\top} \cdot f(A,x)_{j_1} \cdot c(A,x)_{j_1} \cdot e_{i_1} - (- c_g(\wt{A},x)^{\top} \cdot f(\wt{A},x)_{j_1} \cdot c(A,x)_{j_1} \cdot e_{i_1}) |\\
        & + ~ |- c_g(\wt{A},x)^{\top} \cdot f(\wt{A},x)_{j_1} \cdot c(A,x)_{j_1} \cdot e_{i_1} - (- c_g(\wt{A},x)^{\top} \cdot f(\wt{A},x)_{j_1} \cdot c(\wt{A},x)_{j_1} \cdot e_{i_1}) |
    \end{align*}
where the first step follows from Lemma~\ref{lem:product_rule}.

For the first term, we have
\begin{align*}
    & ~ |- c_g(A,x)^{\top} \cdot f(A,x)_{j_1} \cdot c(A,x)_{j_1} \cdot e_{i_1} - (- c_g(\wt{A},x)^{\top} \cdot f(A,x)_{j_1} \cdot c(A,x)_{j_1} \cdot e_{i_1}) | \\
    \leq & ~ |c_g(A,x)^{\top} \cdot f(A,x)_{j_1} \cdot c(A,x)_{j_1} \cdot e_{i_1} - c_g(\wt{A},x)^{\top} \cdot f(A,x)_{j_1} \cdot c(A,x)_{j_1} \cdot e_{i_1} | \\
    \leq & ~ |c_g(A,x)^{\top} - c_g(\wt{A},x)^{\top} | \cdot |f(A,x)_{j_1} |  \cdot |c(A,x)_{j_1} |  \cdot  \|e_{i_1} \|_2 \\
    \leq & ~ 40\beta^{-2} \cdot n \cdot R \cdot  \exp(3R^{2}) \cdot \|A - \wt{A}\|_F
\end{align*}
where the first and second step follows from Fact~\ref{fac:vector_norm},  and the last step follows from {\bf Part 8} of \ref{lem:basic_lips}, and {\bf Part 7 and 9} of \ref{lem:upper_bound}.

For the second term, we have
\begin{align*}
    & ~|- c_g(\wt{A},x)^{\top} \cdot f(A,x)_{j_1} \cdot c(A,x)_{j_1} \cdot e_{i_1} - (- c_g(\wt{A},x)^{\top} \cdot f(\wt{A},x)_{j_1} \cdot c(A,x)_{j_1} \cdot e_{i_1}) | \\
    \leq & ~ | c_g(\wt{A},x)^{\top} \cdot f(A,x)_{j_1} \cdot c(A,x)_{j_1} \cdot e_{i_1} -  c_g(\wt{A},x)^{\top} \cdot f(\wt{A},x)_{j_1} \cdot c(A,x)_{j_1} \cdot e_{i_1} | \\
    \leq & ~ | c_g(\wt{A},x)^{\top} | \cdot |f(A,x)_{j_1} - f(\wt{A},x)_{j_1} |  \cdot |c(A,x)_{j_1} |  \cdot  \|e_{i_1} \|_2 \\
    \leq & ~  20\beta^{-2} \cdot n \cdot R\cdot \exp(3R^2) \cdot \|A - \wt{A}\|_F
\end{align*}
where the first and second step follows from Fact~\ref{fac:vector_norm}, and the third step follows from {\bf Part 10} of Lemma~\ref{lem:basic_lips} and {\bf Part 4 and 6} of Lemma~\ref{lem:upper_bound}.

For the third term, we have 
\begin{align*}
    & ~|- c_g(\wt{A},x)^{\top} \cdot f(\wt{A},x)_{j_1} \cdot c(A,x)_{j_1} \cdot e_{i_1} - (- c_g(\wt{A},x)^{\top} \cdot f(\wt{A},x)_{j_1} \cdot c(\wt{A},x)_{j_1} \cdot e_{i_1}) | \\
    \leq & ~ | c_g(\wt{A},x)^{\top} \cdot f(\wt{A},x)_{j_1} \cdot c(A,x)_{j_1} \cdot e_{i_1} - c_g(\wt{A},x)^{\top} \cdot f(\wt{A},x)_{j_1} \cdot c(\wt{A},x)_{j_1} \cdot e_{i_1}| \\
    \leq & ~ | c_g(\wt{A},x)^{\top} | \cdot | f(\wt{A},x)_{j_1}| \cdot |c(A,x)_{j_1} -c(\wt{A},x)_{j_1}|  \cdot  \|e_{i_1} \|_2 \\
    \leq & ~ 10\beta^{-2} \cdot n \cdot R \cdot \exp(3R^2) \cdot \|A - \wt{A}\|_F
\end{align*}
where the first and second step follows from Fact~\ref{fac:vector_norm}, and the third step follows from {\bf Part 5} of Lemma~\ref{lem:basic_lips} and {\bf Part 6 and 7} of Lemma~\ref{lem:upper_bound}.

Then, we have
    \begin{align*}
        & ~  |- c_g(A,x)^{\top} \cdot f(A,x)_{j_1} \cdot c(A,x)_{j_1} \cdot e_{i_1} - (- c_g(\wt{A},x)^{\top} \cdot f(\wt{A},x)_{j_1} \cdot c(\wt{A},x)_{j_1} \cdot e_{i_1}) | \\
        \leq & ~ 40\beta^{-2} \cdot n \cdot R \cdot  \exp(3R^{2}) \cdot \|A - \wt{A}\|_F \\ 
        & + ~ 20\beta^{-2} \cdot n \cdot R\cdot \exp(3R^2) \cdot \|A - \wt{A}\|_F \\
        & + ~ 10\beta^{-2} \cdot n \cdot R \cdot \exp(3R^2) \cdot \|A - \wt{A}\|_F \\
        \leq & ~\beta^{-2} \cdot n   \exp(5R^2)\|A - \wt{A}\|_F
    \end{align*}
where the second step follows from $\beta^{-1} \geq 1, n > 1, R \leq \exp(R^2), R^2 \leq \exp(R^2)$ and $R \geq 4$.
\end{proof}

\subsection{Third term \texorpdfstring{$- c_g(A,x)^{\top} \cdot f(A,x)_{j_1} \cdot \langle c(A,x), f(A,x) \rangle \cdot ( (A_{j_1,*})^\top \circ x )$}{}}
\begin{lemma}\label{lem:lips_gradient_3}
If the following conditions hold
\begin{itemize}
    \item Let $A_{j_1,*} \in \R^d$ for all $j_1 \in [n]$
    \item Let $x \in \R^d$
    \item Let $f(A,x) \in \R^n$ be defined as Definition~\ref{def:f}
    \item Let $c(A,x) \in \R^n$ be defined as Definition~\ref{def:c}
    \item Let $g(A,x) \in \R^d$ be defined as Definition~\ref{def:g} 
    \item Let $q(A,x) = c(A,x) + f(A,x) \in \R^n$
    \item Let $c_g(A,x) \in \R^d$ be defined as Definition~\ref{def:c_g}.
\end{itemize}
Then we have
\begin{align*}
 & ~|c_g(A,x)^{\top} \cdot f(A,x)_{j_1} \cdot \langle c(A,x), f(A,x) \rangle \cdot ( (A_{j_1,*})^\top \circ x ) \\ & ~ - c_g(\wt{A},x)^{\top} \cdot f(\wt{A},x)_{j_1} \cdot \langle c(\wt{A},x), f(\wt{A},x) \rangle \cdot ( (\wt{A}_{j_1,*})^\top \circ x ) | \\
 \leq & ~\beta^{-2} \cdot n \cdot \exp(6R^2) \cdot \|A - \wt{A}\|_F
\end{align*}
\end{lemma}
\begin{proof}
Firstly, we have 
\begin{align*}
    & ~ |c_g(A,x)^{\top} \cdot f(A,x)_{j_1} \cdot \langle c(A,x), f(A,x) \rangle \cdot ( (A_{j_1,*})^\top \circ x ) \\ & ~ - c_g(\wt{A},x)^{\top} \cdot f(\wt{A},x)_{j_1} \cdot \langle c(\wt{A},x), f(\wt{A},x) \rangle \cdot ( (\wt{A}_{j_1,*})^\top \circ x ) |\\
    \leq & ~  |c_g(A,x)^{\top} \cdot f(A,x)_{j_1} \cdot \langle c(A,x), f(A,x) \rangle \cdot ( (A_{j_1,*})^\top \circ x )\\ & ~  - c_g(\wt{A},x)^{\top} \cdot f(A,x)_{j_1} \cdot \langle c(A,x), f(A,x) \rangle \cdot ( (A_{j_1,*})^\top \circ x ) | \\
    & + ~ |c_g(\wt{A},x)^{\top} \cdot f(A,x)_{j_1} \cdot \langle c(A,x), f(A,x) \rangle \cdot ( (A_{j_1,*})^\top \circ x ) \\ & ~ - c_g(\wt{A},x)^{\top} \cdot f(\wt{A},x)_{j_1} \cdot \langle c(A,x), f(A,x) \rangle \cdot ( (A_{j_1,*})^\top \circ x ) | \\
    & + ~ |c_g(\wt{A},x)^{\top} \cdot f(\wt{A},x)_{j_1} \cdot \langle c(A,x), f(A,x) \rangle \cdot ( (A_{j_1,*})^\top \circ x ) \\ & ~ - c_g(\wt{A},x)^{\top} \cdot f(\wt{A},x)_{j_1} \cdot \langle c(\wt{A},x), f(A,x) \rangle \cdot ( (A_{j_1,*})^\top \circ x ) | \\
     & + ~ |c_g(\wt{A},x)^{\top} \cdot f(\wt{A},x)_{j_1} \cdot \langle c(\wt{A},x), f(A,x) \rangle \cdot ( (A_{j_1,*})^\top \circ x )\\ & ~  - c_g(\wt{A},x)^{\top} \cdot f(\wt{A},x)_{j_1} \cdot \langle c(\wt{A},x), f(\wt{A},x) \rangle \cdot ( (A_{j_1,*})^\top \circ x ) | \\
     & + ~ |c_g(\wt{A},x)^{\top} \cdot f(\wt{A},x)_{j_1} \cdot \langle c(\wt{A},x), f(\wt{A},x) \rangle \cdot ( (A_{j_1,*})^\top \circ x )\\ & ~  - c_g(\wt{A},x)^{\top} \cdot f(\wt{A},x)_{j_1} \cdot \langle c(\wt{A},x), f(\wt{A},x) \rangle \cdot ( (\wt{A}_{j_1,*})^\top \circ x ) |
\end{align*}
where the first step follows from Lemma~\ref{lem:product_rule}.

For the first term, we have
\begin{align*}
     & ~|c_g(A,x)^{\top} \cdot f(A,x)_{j_1} \cdot \langle c(A,x), f(A,x) \rangle \cdot ( (A_{j_1,*})^\top \circ x )   \\
        & ~ - c_g(\wt{A},x)^{\top} \cdot f(A,x)_{j_1} \cdot \langle c(A,x), f(A,x) \rangle \cdot ( (A_{j_1,*})^\top \circ x ) | \\
    \leq & ~ |c_g(A,x)^{\top} - c_g(\wt{A},x)^{\top} | \cdot |f(A,x)_{j_1} |  \cdot \|c(A,x) \|_2  \cdot \|f(A,x) \|_2 \cdot  \| x \|_2  \cdot \|(A_{j_1,*})^\top \|_2 \\
    \leq & ~ 40\beta^{-2} \cdot n \cdot R^3 \cdot  \exp(3R^{2}) \cdot \|A - \wt{A}\|_F
\end{align*}
where the first step follows from Fact~\ref{fac:vector_norm}, and the second step follows from {\bf Part 8} of Lemma~\ref{lem:basic_lips}, {\bf Part 3, 4, and 7} of Lemma~\ref{lem:upper_bound}, $\|(A_{j_1,*})^\top \|_2 \leq \|A\|_F \leq R$, and $\|\diag(x)\|_F = \|x\|_2 \leq R$.

For the second term, we have
\begin{align*}
     & ~|c_g(\wt{A},x)^{\top} \cdot f(A,x)_{j_1} \cdot \langle c(A,x), f(A,x) \rangle \cdot ( (A_{j_1,*})^\top \circ x )   \\
        & ~- c_g(\wt{A},x)^{\top} \cdot f(\wt{A},x)_{j_1} \cdot \langle c(A,x), f(A,x) \rangle \cdot ( (A_{j_1,*})^\top \circ x ) | \\
    \leq & ~ |c_g(\wt{A},x)^{\top} | \cdot |f(A,x)_{j_1} - f(\wt{A},x)_{j_1} |  \cdot \|c(A,x) \|_2  \cdot \|f(A,x) \|_2 \cdot  \| x \|_2  \cdot \|(A_{j_1,*})^\top \|_2 \\
    \leq & ~  20\beta^{-2} \cdot n \cdot R^3 \cdot \exp(3R^2) \cdot \|A - \wt{A}\|_F
\end{align*}
where the first step follows from Fact~\ref{fac:vector_norm}, and the second step follows from {\bf Part 10} of Lemma~\ref{lem:basic_lips}, {\bf Part 3, 4, and 6} of Lemma~\ref{lem:upper_bound}, $\|A\|_F \leq R$, and $\|\diag(x)\|_F = \|x\|_2 \leq R$.

For the third term, we have
\begin{align*}
     & ~|c_g(\wt{A},x)^{\top} \cdot f(\wt{A},x)_{j_1} \cdot \langle c(A,x), f(A,x) \rangle \cdot ( (A_{j_1,*})^\top \circ x )  \\
        & ~ - c_g(\wt{A},x)^{\top} \cdot f(\wt{A},x)_{j_1} \cdot \langle c(\wt{A},x), f(A,x) \rangle \cdot ( (A_{j_1,*})^\top \circ x ) | \\
    \leq & ~ |c_g(\wt{A},x)^{\top} | \cdot | f(\wt{A},x)_{j_1} |  \cdot \|c(A,x) - c(\wt{A},x) \|_2  \cdot \|f(A,x) \|_2 \cdot  \| x \|_2  \cdot \|(A_{j_1,*})^\top \|_2 \\
    \leq & ~ 10\beta^{-2} \cdot n \cdot R^3 \cdot \exp(3R^2) \cdot \|A - \wt{A}\|_F
\end{align*}
where the first step follows from Fact~\ref{fac:vector_norm}, and the second step follows from {\bf Part 5} of Lemma~\ref{lem:basic_lips}, {\bf Part 3, 6, and 7} of Lemma~\ref{lem:upper_bound}, $\|A\|_F \leq R$, and $\|\diag(x)\|_F = \|x\|_2 \leq R$.

For the forth term, we have
\begin{align*}
     & ~|c_g(\wt{A},x)^{\top} \cdot f(\wt{A},x)_{j_1} \cdot \langle c(\wt{A},x), f(A,x) \rangle \cdot ( (A_{j_1,*})^\top \circ x )   \\
        & ~- c_g(\wt{A},x)^{\top} \cdot f(\wt{A},x)_{j_1} \cdot \langle c(\wt{A},x), f(\wt{A},x) \rangle \cdot ( (A_{j_1,*})^\top \circ x ) | \\
    \leq & ~ |c_g(\wt{A},x)^{\top} | \cdot |f(\wt{A},x)_{j_1} |  \cdot \|c(\wt{A},x) \|_2  \cdot \|f(A,x) - f(\wt{A},x) \|_2 \cdot  \| x \|_2  \cdot \|(A_{j_1,*})^\top \|_2 \\
    \leq & ~ 20\beta^{-2} \cdot n \cdot R^3 \cdot \exp(3R^2) \cdot \|A - \wt{A}\|_F
\end{align*}
where the first step follows from Fact~\ref{fac:vector_norm}, and the second step follows from {\bf Part 4} of Lemma~\ref{lem:basic_lips}, {\bf Part 4, 6, and 7} of Lemma~\ref{lem:upper_bound}, $\|A\|_F \leq R$, and $\|\diag(x)\|_F = \|x\|_2 \leq R$.

For the fifth term, we have
\begin{align*}
     & ~|c_g(\wt{A},x)^{\top} \cdot f(\wt{A},x)_{j_1} \cdot \langle c(\wt{A},x), f(\wt{A},x) \rangle \cdot ( (A_{j_1,*})^\top \circ x )  \\
        & ~- c_g(\wt{A},x)^{\top} \cdot f(\wt{A},x)_{j_1} \cdot \langle c(\wt{A},x), f(\wt{A},x) \rangle \cdot ( (\wt{A}_{j_1,*})^\top \circ x ) | \\
    \leq & ~ |c_g(\wt{A},x)^{\top} | \cdot |f(\wt{A},x)_{j_1} |  \cdot \|c(\wt{A},x) \|_2  \cdot \| f(\wt{A},x) \|_2 \cdot  \| x \|_2  \cdot \|(A_{j_1,*})^\top - (\wt{A}_{j_1,*})^\top \|_2 \\
    \leq & ~ 10 R^2  \cdot\|A - \wt{A}\|_F 
\end{align*}
where the first step follows from Fact~\ref{fac:vector_norm}, and the second step follows from {\bf Part 3, 4, 6, and 7} of Lemma~\ref{lem:upper_bound}, $\|(A_{j_1,*})^\top - (\wt{A}_{j_1,*})^\top \|_2 \leq \|A - \wt{A}\|_F$, and $\|\diag(x)\|_F = \|x\|_2 \leq R$.

Then, we have
\begin{align*}
 & ~|c_g(A,x)^{\top} \cdot f(A,x)_{j_1} \cdot \langle c(A,x), f(A,x) \rangle \cdot ( (A_{j_1,*})^\top \circ x ) \\ & ~ - c_g(\wt{A},x)^{\top} \cdot f(\wt{A},x)_{j_1} \cdot \langle c(\wt{A},x), f(\wt{A},x) \rangle \cdot ( (\wt{A}_{j_1,*})^\top \circ x ) | \\
 \leq & ~ 40\beta^{-2} \cdot n \cdot R^3 \cdot \exp(3R^2) \cdot \|A - \wt{A}\|_F \\
 & +  ~  20\beta^{-2} \cdot n \cdot R^3 \cdot \exp(3R^2) \cdot \|A - \wt{A}\|_F \\
 & +  ~  10\beta^{-2} \cdot n \cdot R^3 \cdot \exp(3R^2) \cdot \|A - \wt{A}\|_F\\
  & +  ~ 20\beta^{-2} \cdot n \cdot R^3 \cdot \exp(3R^2) \cdot \|A - \wt{A}\|_F \\
 & +  ~ 10 R^2 \cdot\|A - \wt{A}\|_F \\
 \leq & ~ \beta^{-2} \cdot n \cdot \exp(6R^2) \cdot \|A - \wt{A}\|_F
\end{align*}
where the second step follows from $\beta^{-1} \geq 1, n > 1, R \leq \exp(R^2), R^2 \leq \exp(R^2)$ and $R \geq 4$.
\end{proof}

\subsection{Forth term \texorpdfstring{$c_g(A,x)^{\top} \cdot f(A,x)_{j_1}\diag (x) A^{\top}   f(A,x)  \cdot    \langle c(A, x), f(A, x) \rangle$}{}}
\begin{lemma}\label{lem:lips_gradient_4}
If the following conditions hold
\begin{itemize}
    \item Let $A_{j_1,*} \in \R^d$ for all $j_1 \in [n]$
    \item Let $x \in \R^d$
    \item Let $f(A,x) \in \R^n$ be defined as Definition~\ref{def:f}
    \item Let $c(A,x) \in \R^n$ be defined as Definition~\ref{def:c}
    \item Let $g(A,x) \in \R^d$ be defined as Definition~\ref{def:g} 
    \item Let $q(A,x) = c(A,x) + f(A,x) \in \R^n$
    \item Let $c_g(A,x) \in \R^d$ be defined as Definition~\ref{def:c_g}.
\end{itemize}
Then we have
\begin{align*}
 & ~|c_g(A,x)^{\top}  f(A,x)_{j_1}\diag (x) A^{\top}   f(A,x)      \langle c(A, x), f(A, x) \rangle   \\
        & ~- c_g(\wt{A},x)^{\top} f(\wt{A},x)_{j_1}\diag (x) \wt{A}^{\top}   f(\wt{A},x)     \langle c(\wt{A}, x), f(\wt{A}, x) \rangle | \\
 \leq & ~ \beta^{-2} \cdot n \cdot \exp(6R^2) \cdot \|A - \wt{A}\|_F
\end{align*}
\end{lemma}
\begin{proof}
Firstly, we have
\begin{align*}
     & ~|c_g(A,x)^{\top}  f(A,x)_{j_1}\diag (x) A^{\top}   f(A,x)      \langle c(A, x), f(A, x) \rangle  \\
        & ~ - c_g(\wt{A},x)^{\top}  f(\wt{A},x)_{j_1}\diag (x) \wt{A}^{\top}   f(\wt{A},x)      \langle c(\wt{A}, x), f(\wt{A}, x) \rangle | \\
     \leq & ~ |c_g(A,x)^{\top}  f(A,x)_{j_1}\diag (x) A^{\top}   f(A,x)      \langle c(A, x), f(A, x) \rangle  \\
        & ~ - c_g(\wt{A},x)^{\top}  f(A,x)_{j_1}\diag (x) A^{\top}   f(A,x)      \langle c(A, x), f(A, x) \rangle  | \\
     + &  ~ |c_g(\wt{A},x)^{\top}  f(A,x)_{j_1}\diag (x) A^{\top}   f(A,x)      \langle c(A, x), f(A, x) \rangle  \\
        & ~ - c_g(\wt{A},x)^{\top}  f(\wt{A},x)_{j_1}\diag (x) A^{\top}   f(A,x)      \langle c(A, x), f(A, x) \rangle  |\\
      + &  ~ |c_g(\wt{A},x)^{\top}  f(\wt{A},x)_{j_1}\diag (x) A^{\top}   f(A,x)      \langle c(A, x), f(A, x) \rangle  \\
        & ~ - c_g(\wt{A},x)^{\top}  f(\wt{A},x)_{j_1}\diag (x) \wt{A}^{\top}   f(A,x)      \langle c(A, x), f(A, x) \rangle  |\\
     +&  ~ |c_g(\wt{A},x)^{\top} f(\wt{A},x)_{j_1}\diag (x) \wt{A}^{\top}   f(A,x)      \langle c(A, x), f(A, x) \rangle  \\
        & ~ - c_g(\wt{A},x)^{\top}  f(\wt{A},x)_{j_1}\diag (x) \wt{A}^{\top}   f(\wt{A},x)      \langle c(A, x), f(A, x) \rangle  | \\
      +&  ~ |c_g(\wt{A},x)^{\top}  f(\wt{A},x)_{j_1}\diag (x) \wt{A}^{\top}   f(\wt{A},x)      \langle c(A, x), f(A, x) \rangle   \\
        & ~- c_g(\wt{A},x)^{\top}  f(\wt{A},x)_{j_1}\diag (x) \wt{A}^{\top}   f(\wt{A},x)      \langle c(\wt{A}, x), f(A, x) \rangle  | \\
     + &  ~ |c_g(\wt{A},x)^{\top}  f(\wt{A},x)_{j_1}\diag (x) \wt{A}^{\top}   f(\wt{A},x)      \langle c(\wt{A}, x), f(A, x) \rangle  \\
        & ~ - c_g(\wt{A},x)^{\top}  f(\wt{A},x)_{j_1}\diag (x) \wt{A}^{\top}   f(\wt{A},x)      \langle c(\wt{A}, x), f(\wt{A}, x) \rangle  |
\end{align*}
For the first term, we have
\begin{align*}
    & ~ |c_g(A,x)^{\top}  f(A,x)_{j_1}\diag (x) A^{\top}   f(A,x)      \langle c(A, x), f(A, x) \rangle  \\
        & ~ - c_g(\wt{A},x)^{\top}  f(A,x)_{j_1}\diag (x) A^{\top}   f(A,x)      \langle c(A, x), f(A, x) \rangle  | \\
    \leq & ~ \|c_g(A,x)^{\top} -  c_g(\wt{A},x)^{\top} \|_2 \cdot |f(A,x)_{j_1} | \cdot \| x \|_2 \cdot \| A^{\top} \|_F \cdot  \| f(A,x)\|_2 \cdot    \| c(A, x)\|_2 \cdot \|f(A, x)\|_2 \\
    \leq & ~ 40 \beta^{-2} \cdot n \cdot R^3 \cdot  \exp(3R^{2}) \cdot \|A - \wt{A}\|_F 
\end{align*}
where the first step follows from Fact~\ref{fac:vector_norm} and Fact~\ref{fac:matrix_norm}, and the second step follows from {\bf Part 8} of Lemma~\ref{lem:basic_lips}, {\bf 3, 4, and 7} of Lemma~\ref{lem:upper_bound}, and $\|x \|_2 \leq R, \| A\|_F \leq R$. 

For the second term, we have
\begin{align*}
    & ~ |c_g(\wt{A},x)^{\top}  f(A,x)_{j_1}\diag (x) A^{\top}   f(A,x)      \langle c(A, x), f(A, x) \rangle   \\
        & ~- c_g(\wt{A},x)^{\top}  f(\wt{A},x)_{j_1}\diag (x) A^{\top}   f(A,x)      \langle c(A, x), f(A, x) \rangle  | \\
    \leq & ~ \| c_g(\wt{A},x)^{\top} \|_2 \cdot |f(A,x)_{j_1} - f(\wt{A},x)_{j_1} |\cdot  \| x \|_2 \cdot \| A^{\top} \|_F \cdot  \| f(A,x)\|_2  \cdot   \| c(A, x)\|_2 \cdot \|f(A, x)\|_2 \\
    \leq & ~ 20 \beta^{-2} \cdot n \cdot R^3 \cdot \exp(3R^2) \cdot \|A - \wt{A}\|_F 
\end{align*}
where the first step follows from Fact~\ref{fac:vector_norm} and Fact~\ref{fac:matrix_norm}, and the second step follows from {\bf Part 10} of Lemma~\ref{lem:basic_lips}, {\bf 3, 4, and 6} of Lemma~\ref{lem:upper_bound}, and $\|x \|_2 \leq R, \| A\|_F \leq R$. 

For the third term, we have
\begin{align*}
    & ~ |c_g(\wt{A},x)^{\top}  f(\wt{A},x)_{j_1}\diag (x) A^{\top}   f(A,x)      \langle c(A, x), f(A, x) \rangle  \\
        & ~ - c_g(\wt{A},x)^{\top}  f(\wt{A},x)_{j_1}\diag (x) \wt{A}^{\top}   f(A,x)      \langle c(A, x), f(A, x) \rangle  | \\
    \leq & ~ \| c_g(\wt{A},x)^{\top} \|_2 \cdot |f(\wt{A},x)_{j_1} |\cdot \| x \|_2 \cdot \| A^{\top} - \wt{A}^{\top}\|_F \cdot  \| f(A,x)\|_2  \cdot   \| c(A, x)\|_2 \cdot \|f(A, x)\|_2 \\
    \leq & ~ 10 R^2 \cdot \| A - \wt{A}\|_F 
\end{align*}
where the first step follows from Fact~\ref{fac:vector_norm} and Fact~\ref{fac:matrix_norm}, and the second step follows from {\bf  3, 4, 6 and 7} of Lemma~\ref{lem:upper_bound} and $\|x \|_2 \leq R$. 

For the fourth term, we have
\begin{align*}
    & ~ |c_g(\wt{A},x)^{\top} f(\wt{A},x)_{j_1}\diag (x) \wt{A}^{\top}   f(A,x)      \langle c(A, x), f(A, x) \rangle  \\
        & ~ - c_g(\wt{A},x)^{\top}  f(\wt{A},x)_{j_1}\diag (x) \wt{A}^{\top}   f(\wt{A},x)      \langle c(A, x), f(A, x) \rangle  | \\
    \leq & ~ \| c_g(\wt{A},x)^{\top} \|_2 \cdot  |f(\wt{A},x)_{j_1} |\cdot \| x \|_2 \cdot \|  \wt{A}^{\top}\|_F \cdot  \| f(A,x) - f(\wt{A},x)\|_2  \cdot   \| c(A, x)\|_2 \cdot \|f(A, x)\|_2 \\
    \leq & ~ 20\beta^{-2} \cdot n \cdot R^3 \cdot\exp(3R^2) \cdot \|A - \wt{A}\|_F
\end{align*}
where the first step follows from Fact~\ref{fac:vector_norm} and Fact~\ref{fac:matrix_norm}, and the second step follows from {\bf  3, 4, 6 and 7} of Lemma~\ref{lem:upper_bound} and $\|x \|_2 \leq R, \|A \|_F \leq R$. 

For fifth and sixth term, it is similar to the fourth term.
Then, we have
\begin{align*}
    & ~|c_g(A,x)^{\top}  f(A,x)_{j_1}\diag (x) A^{\top}   f(A,x)      \langle c(A, x), f(A, x) \rangle   \\
        & ~- c_g(\wt{A},x)^{\top}  f(\wt{A},x)_{j_1}\diag (x) \wt{A}^{\top}   f(\wt{A},x)      \langle c(\wt{A}, x), f(\wt{A}, x) \rangle | \\
    \leq & ~ 40\beta^{-2} \cdot n \cdot R^3 \cdot  \exp(3R^{2}) \cdot \|A - \wt{A}\|_F  \\
    & + ~ 20\beta^{-2} \cdot n \cdot R^3 \cdot \exp(3R^2) \cdot \|A - \wt{A}\|_F  \\
    & + ~ 10 R^2 \cdot \| A - \wt{A}\|_F  \\
    & + ~ 20\beta^{-2} \cdot n \cdot R^3 \cdot\exp(3R^2) \cdot \|A - \wt{A}\|_F \\
     & + ~ 20\beta^{-2} \cdot n \cdot R^3 \cdot\exp(3R^2) \cdot \|A - \wt{A}\|_F \\
      & + ~ 20\beta^{-2} \cdot n \cdot R^3 \cdot\exp(3R^2) \cdot \|A - \wt{A}\|_F \\
      \leq & ~ \beta^{-2} \cdot n \cdot \exp(6R^2) \cdot \|A - \wt{A}\|_F
\end{align*}
where the second step follows from $\beta^{-1} \geq 1, n > 1, R \leq \exp(R^2), R^2 \leq \exp(R^2)$ and $R \geq 4$.
\end{proof}

\subsection{Fifth term \texorpdfstring{$- c_g(A,x)^{\top} \cdot f(A,x)_{j_1}\diag(x) A^{\top} f(A,x) \cdot (\langle -f(A,x), f(A,x) \rangle + f(A,x)_{j_1}) $}{}}
\begin{lemma}\label{lem:lips_gradient_5}
If the following conditions hold
\begin{itemize}
    \item Let $A_{j_1,*} \in \R^d$ for all $j_1 \in [n]$
    \item Let $x \in \R^d$
    \item Let $f(A,x) \in \R^n$ be defined as Definition~\ref{def:f}
    \item Let $c(A,x) \in \R^n$ be defined as Definition~\ref{def:c}
    \item Let $g(A,x) \in \R^d$ be defined as Definition~\ref{def:g} 
    \item Let $q(A,x) = c(A,x) + f(A,x) \in \R^n$
    \item Let $c_g(A,x) \in \R^d$ be defined as Definition~\ref{def:c_g}.
\end{itemize}
Then we have
\begin{align*}
 & ~ |- c_g(A,x)^{\top} \cdot f(A,x)_{j_1}\diag(x) A^{\top} f(A,x) \cdot (\langle -f(A,x), f(A,x) \rangle + f(A,x)_{j_1})   \\
 & - ~ (- c_g(\wt{A},x)^{\top} \cdot f(\wt{A},x)_{j_1}\diag(x) \wt{A}^{\top} f(\wt{A},x) \cdot (\langle -f(\wt{A},x), f(\wt{A},x) \rangle + f(\wt{A},x)_{j_1}) ) | \\
 \leq & ~ 2\beta^{-2} \cdot n \cdot \exp(6R^2) \cdot \|A - \wt{A}\|_F
\end{align*}
\end{lemma}
\begin{proof}
We define 
\begin{align*}
    G_1(A) := & ~  c_g(A,x)^{\top} \cdot f(A,x)_{j_1}\diag(x) A^{\top} f(A,x) \cdot \langle -f(A,x), f(A,x) \rangle \\
    G_2(A) := & ~  c_g(A,x)^{\top} \cdot f(A,x)_{j_1}\diag(x) A^{\top} f(A,x) \cdot f(A,x)_{j_1}
\end{align*}
      Firstly, we have
      \begin{align*}
          & ~ |- c_g(A,x)^{\top} \cdot f(A,x)_{j_1}\diag(x) A^{\top} f(A,x) \cdot (\langle -f(A,x), f(A,x) \rangle + f(A,x)_{j_1})   \\
  & - ~ (- c_g(\wt{A},x)^{\top} \cdot f(\wt{A},x)_{j_1}\diag(x) \wt{A}^{\top} f(\wt{A},x) \cdot (\langle -f(\wt{A},x), f(\wt{A},x) \rangle + f(\wt{A},x)_{j_1}))  | \\
 = & ~ | c_g(A,x)^{\top} \cdot f(A,x)_{j_1}\diag(x) A^{\top} f(A,x) \cdot (\langle -f(A,x), f(A,x) \rangle + f(A,x)_{j_1})   \\
  &  - ~ c_g(\wt{A},x)^{\top} \cdot f(\wt{A},x)_{j_1}\diag(x) \wt{A}^{\top} f(\wt{A},x) \cdot (\langle -f(\wt{A},x), f(\wt{A},x) \rangle + f(\wt{A},x)_{j_1}) | \\
  = & ~ |G_1(A) + G_2(A) - G_1(\wt{A}) -G_2(\wt{A}) |\\
  \leq & ~ |G_1(A) - G_1(\wt{A})| + |G_2(A) -G_2(\wt{A})| 
  \end{align*}

For the  $|G_1(A) - G_1(\wt{A})|$, we have
\begin{align*}
   & ~ |G_1(A) - G_1(\wt{A})|  \\
  =  &   ~ |c_g(A,x)^{\top} \cdot f(A,x)_{j_1}\diag(x) A^{\top} f(A,x) \cdot \langle -f(A,x), f(A,x) \rangle \\
  & - ~ c_g(\wt{A},x)^{\top} \cdot f(\wt{A},x)_{j_1}\diag(x) \wt{A}^{\top} f(\wt{A},x) \cdot \langle -f(\wt{A},x), f(\wt{A},x) \rangle |  \\
  \leq & ~ |c_g(A,x)^{\top} \cdot f(A,x)_{j_1}\diag(x) A^{\top} f(A,x) \cdot \langle -f(A,x), f(A,x) \rangle \\
  & - ~c_g(\wt{A},x)^{\top} \cdot f(A,x)_{j_1}\diag(x) A^{\top} f(A,x) \cdot \langle -f(A,x), f(A,x) \rangle |  \\
   + & ~ |c_g(\wt{A},x)^{\top} \cdot f(A,x)_{j_1}\diag(x) A^{\top} f(A,x) \cdot \langle -f(A,x), f(A,x) \rangle \\
  & - ~c_g(\wt{A},x)^{\top} \cdot f(\wt{A},x)_{j_1}\diag(x) A^{\top} f(A,x) \cdot \langle -f(A,x), f(A,x) \rangle |  \\
  + & ~ |c_g(\wt{A},x)^{\top} \cdot f(\wt{A},x)_{j_1}\diag(x) A^{\top} f(A,x) \cdot \langle -f(A,x), f(A,x) \rangle \\
  & - ~c_g(\wt{A},x)^{\top} \cdot f(\wt{A},x)_{j_1}\diag(x) \wt{A}^{\top} f(A,x) \cdot \langle -f(A,x), f(A,x) \rangle |  \\
   + & ~ |c_g(\wt{A},x)^{\top} \cdot f(\wt{A},x)_{j_1}\diag(x) \wt{A}^{\top} f(A,x) \cdot \langle -f(A,x), f(A,x) \rangle \\
  & - ~c_g(\wt{A},x)^{\top} \cdot f(\wt{A},x)_{j_1}\diag(x) \wt{A}^{\top} f(\wt{A},x) \cdot \langle -f(A,x), f(A,x) \rangle |  \\
   + & ~ |c_g(\wt{A},x)^{\top} \cdot f(\wt{A},x)_{j_1}\diag(x) \wt{A}^{\top} f(\wt{A},x) \cdot \langle -f(A,x), f(A,x) \rangle \\
  & - ~c_g(\wt{A},x)^{\top} \cdot f(\wt{A},x)_{j_1}\diag(x) \wt{A}^{\top} f(\wt{A},x) \cdot \langle -f(\wt{A},x), f(A,x) \rangle |  \\
   + & ~ |c_g(\wt{A},x)^{\top} \cdot f(\wt{A},x)_{j_1}\diag(x) \wt{A}^{\top} f(\wt{A},x) \cdot \langle -f(\wt{A},x), f(A,x) \rangle \\
  & - ~c_g(\wt{A},x)^{\top} \cdot f(\wt{A},x)_{j_1}\diag(x) \wt{A}^{\top} f(\wt{A},x) \cdot \langle -f(\wt{A},x), f(\wt{A},x) \rangle | 
\end{align*}
For the first term, we have 
\begin{align*}
   & ~ |c_g(A,x)^{\top} \cdot f(A,x)_{j_1}\diag(x) A^{\top} f(A,x) \cdot \langle -f(A,x), f(A,x) \rangle \\
  & - ~ c_g(\wt{A},x)^{\top} \cdot  f(A,x)_{j_1}\diag(x) A^{\top} f(A,x) \cdot \langle -f(A,x), f(A,x) \rangle |  \\ 
  \leq & ~  \| c_g(A,x)^{\top} - c_g(\wt{A},x)^{\top} \|_2 \cdot | f(A,x)_{j_1} |\cdot \|x \|_2 \cdot \|A^{\top}\|_F \cdot \|f(A,x) \|_2 \cdot \|f(A,x) \|_2 \cdot \|f(A,x) \|_2 \\
  \leq & ~ 20\beta^{-2} \cdot n \cdot R^3 \cdot  \exp(3R^{2}) \cdot \|A - \wt{A}\|_F 
\end{align*}
where the first step follows from Fact~\ref{fac:vector_norm} and Fact~\ref{fac:matrix_norm}, and the second step follows from {\bf Part 8} of Lemma~\ref{lem:basic_lips}, {\bf Part 3, 7} of Lemma~\ref{lem:upper_bound} and $\|x \|_2 \leq R, \|A \|_F \leq R$.

For the second term, we have
\begin{align*}
    & ~ |c_g(\wt{A},x)^{\top} \cdot f(A,x)_{j_1}\diag(x) A^{\top} f(A,x) \cdot \langle -f(A,x), f(A,x) \rangle \\
  & - ~c_g(\wt{A},x)^{\top} \cdot f(\wt{A},x)_{j_1}\diag(x) A^{\top} f(A,x) \cdot \langle -f(A,x), f(A,x) \rangle | \\
  \leq & ~ \| c_g(\wt{A},x)^{\top}\|_2 \cdot |f(A,x)_{j_1} - f(\wt{A},x)_{j_1}|\cdot \| x\|_2 \cdot \| A^{\top}\|_F \cdot \|f(A,x)\|_2 \cdot \|f(A,x) \|_2 \cdot \|f(A,x) \|_2 \\
  \leq & ~ 10\beta^{-2} \cdot n \cdot R^3 \cdot \exp(3R^2) \cdot \|A - \wt{A}\|_F 
\end{align*}
where the first step follows from Fact~\ref{fac:vector_norm} and Fact~\ref{fac:matrix_norm}, and the second step follows from {\bf Part 8} of Lemma~\ref{lem:basic_lips}, {\bf Part 3, 6, 7} of Lemma~\ref{lem:upper_bound}, and $\|x \|_2 \leq R, \|A \|_F \leq R$.
 
For the third term, we have 
\begin{align*}
    & ~ |c_g(\wt{A},x)^{\top} \cdot f(\wt{A},x)_{j_1}\diag(x) A^{\top} f(A,x) \cdot \langle -f(A,x), f(A,x) \rangle \\
  & - ~c_g(\wt{A},x)^{\top} \cdot f(\wt{A},x)_{j_1}\diag(x) \wt{A}^{\top} f(A,x) \cdot \langle -f(A,x), f(A,x) \rangle |  \\
  \leq & ~ \| c_g(\wt{A},x)^{\top}\|_2 \cdot |f(\wt{A},x)_{j_1}|\cdot \| x\|_2 \cdot \| A^{\top} - \wt{A}^{\top}\|_F \cdot \|f(A,x)\|_2 \cdot \|f(A,x) \|_2 \cdot \|f(A,x) \|_2 \\
  \leq & ~ 5R^2 \cdot \| A^{\top} - \wt{A}^{\top}\|_F
\end{align*}
where the first step follows from Fact~\ref{fac:vector_norm} and Fact~\ref{fac:matrix_norm}, and the second step follows from {\bf Part 3, 6, 7} of Lemma~\ref{lem:upper_bound}, and $\|x \|_2 \leq R, \|A \|_F \leq R$.

For the fourth term, we have 
\begin{align*}
    & ~ |c_g(\wt{A},x)^{\top} \cdot f(\wt{A},x)_{j_1}\diag(x) \wt{A}^{\top} f(A,x) \cdot \langle -f(A,x), f(A,x) \rangle \\
  & - ~c_g(\wt{A},x)^{\top} \cdot f(\wt{A},x)_{j_1}\diag(x) \wt{A}^{\top} f(\wt{A},x) \cdot \langle -f(A,x), f(A,x) \rangle |  \\
  \leq & ~ \| c_g(\wt{A},x)^{\top}\|_2 \cdot |f(\wt{A},x)_{j_1}|\cdot \| x\|_2 \cdot \| \wt{A}^{\top}\|_F \cdot \|f(A,x) - f(\wt{A},x)\|_2 \cdot \|f(A,x) \|_2 \cdot \|f(A,x) \|_2 \\
  \leq & ~ 10\beta^{-2} \cdot n \cdot R^3 \cdot \exp(3R^2) \cdot \|A - \wt{A}\|_F
\end{align*}
where the first step follows from Fact~\ref{fac:vector_norm} and Fact~\ref{fac:matrix_norm}, and the second step follows from {\bf Part 4} of Lemma~\ref{lem:basic_lips}, {\bf Part 3, 6, 7} of Lemma~\ref{lem:upper_bound}, and $\|x \|_2 \leq R, \|A \|_F \leq R$.

For fifth and sixth term, it is similar to the fourth term. Then, we have
\begin{align*}
   & ~ |G_1(A) - G_1(\wt{A})|  \\
  \leq  &   ~  20\beta^{-2} \cdot n \cdot R^3 \cdot  \exp(3R^{2}) \cdot \|A - \wt{A}\|_F \\
  & +   ~  10\beta^{-2} \cdot n \cdot R^3 \cdot \exp(3R^2) \cdot \|A - \wt{A}\|_F \\
  & +   ~  5R^2 \cdot \| A^{\top} - \wt{A}^{\top}\|_F \\
  & +   ~  10\beta^{-2} \cdot n \cdot R^3 \cdot \exp(3R^2) \cdot \|A - \wt{A}\|_F \\
  & +   ~  10\beta^{-2} \cdot n \cdot R^3 \cdot \exp(3R^2) \cdot \|A - \wt{A}\|_F \\
  & +   ~  10\beta^{-2} \cdot n \cdot R^3 \cdot \exp(3R^2) \cdot \|A - \wt{A}\|_F \\
  \leq & ~ \beta^{-2} \cdot n \cdot \exp(6R^2) \cdot \|A - \wt{A}\|_F
\end{align*}
where the second step follows from $\beta^{-1} \geq 1, n > 1, R \leq \exp(R^2), R^2 \leq \exp(R^2)$ and $R \geq 4$.

For the  $|G_2(A) - G_2(\wt{A})|$, we have
\begin{align*}
   & ~ |G_2(A) - G_2(\wt{A})|  \\
  =  &   ~ |c_g(A,x)^{\top} \cdot f(A,x)_{j_1}\diag(x) A^{\top} f(A,x) \cdot f(A,x)_{j_1} \\
  & - ~ c_g(\wt{A},x)^{\top} \cdot f(\wt{A},x)_{j_1}\diag(x) \wt{A}^{\top} f(\wt{A},x) \cdot f(\wt{A},x)_{j_1}|  \\
  \leq & ~   |c_g(A,x)^{\top} \cdot f(A,x)_{j_1}\diag(x) A^{\top} f(A,x) \cdot f(A,x)_{j_1} \\
  & - ~ c_g(\wt{A},x)^{\top} \cdot f(A,x)_{j_1}\diag(x) A^{\top} f(A,x) \cdot f(A,x)_{j_1}|  \\
  + & ~   |c_g(\wt{A},x)^{\top} \cdot f(A,x)_{j_1}\diag(x) A^{\top} f(A,x) \cdot f(A,x)_{j_1} \\
  & - ~ c_g(\wt{A},x)^{\top} \cdot f(\wt{A},x)_{j_1}\diag(x) A^{\top} f(A,x) \cdot f(A,x)_{j_1}|  \\
  + & ~   |c_g(\wt{A},x)^{\top} \cdot f(\wt{A},x)_{j_1}\diag(x) A^{\top} f(A,x) \cdot f(A,x)_{j_1} \\
  & - ~ c_g(\wt{A},x)^{\top} \cdot f(\wt{A},x)_{j_1}\diag(x) \wt{A}^{\top} f(A,x) \cdot f(A,x)_{j_1}|  \\
  + & ~   |c_g(\wt{A},x)^{\top} \cdot f(\wt{A},x)_{j_1}\diag(x) \wt{A}^{\top} f(A,x) \cdot f(A,x)_{j_1} \\
  & - ~ c_g(\wt{A},x)^{\top} \cdot f(\wt{A},x)_{j_1}\diag(x) \wt{A}^{\top} f(\wt{A},x) \cdot f(A,x)_{j_1}|  \\
  + & ~   |c_g(\wt{A},x)^{\top} \cdot f(\wt{A},x)_{j_1}\diag(x) \wt{A}^{\top} f(\wt{A},x) \cdot f(A,x)_{j_1} \\
  & - ~ c_g(\wt{A},x)^{\top} \cdot f(\wt{A},x)_{j_1}\diag(x) \wt{A}^{\top} f(\wt{A},x) \cdot f(\wt{A},x)_{j_1}|  \\
\end{align*}
For the first term, we have 
\begin{align*}
   & ~ |c_g(A,x)^{\top} \cdot f(A,x)_{j_1}\diag(x) A^{\top} f(A,x) \cdot f(A,x)_{j_1} \\
  & - ~ c_g(\wt{A},x)^{\top} \cdot f(A,x)_{j_1}\diag(x) A^{\top} f(A,x) \cdot f(A,x)_{j_1}| \\
  \leq & ~  \| c_g(A,x)^{\top} - c_g(\wt{A},x)^{\top} \|_2 \cdot | f(A,x)_{j_1} |\cdot \|x \|_2 \cdot \|A^{\top}\|_F \cdot \|f(A,x) \|_2 \cdot | f(A,x)_{j_1} |  \\
  \leq & ~ 20\beta^{-2} \cdot n \cdot R^3 \cdot  \exp(3R^{2}) \cdot \|A - \wt{A}\|_F 
\end{align*}
where the first step follows from Fact~\ref{fac:vector_norm} and Fact~\ref{fac:matrix_norm}, and the second step follows from {\bf Part 8} of Lemma~\ref{lem:basic_lips}, {\bf Part 3, 7} of Lemma~\ref{lem:upper_bound} and $\|x \|_2 \leq R, \|A \|_F \leq R$.

For the second term, we have
\begin{align*}
    & ~ |c_g(\wt{A},x)^{\top} \cdot f(A,x)_{j_1}\diag(x) A^{\top} f(A,x) \cdot f(A,x)_{j_1} \\
  & - ~c_g(\wt{A},x)^{\top} \cdot f(\wt{A},x)_{j_1}\diag(x) A^{\top} f(A,x) \cdot f(A,x)_{j_1} | \\
  \leq & ~ \| c_g(\wt{A},x)^{\top}\|_2 \cdot |f(A,x)_{j_1} - f(\wt{A},x)_{j_1}|\cdot \| x\|_2 \cdot \| A^{\top}\|_F \cdot \|f(A,x)\|_2 \cdot | f(A,x)_{j_1} | \\
  \leq & ~ 10\beta^{-2} \cdot n \cdot R^3 \cdot \exp(3R^2) \cdot \|A - \wt{A}\|_F 
\end{align*}
where the first step follows from Fact~\ref{fac:vector_norm} and Fact~\ref{fac:matrix_norm}, and the second step follows from {\bf Part 8} of Lemma~\ref{lem:basic_lips}, {\bf Part 3, 6, 7} of Lemma~\ref{lem:upper_bound}, and $\|x \|_2 \leq R, \|A \|_F \leq R$.

For the third term, we have 
\begin{align*}
    & ~ |c_g(\wt{A},x)^{\top} \cdot f(\wt{A},x)_{j_1}\diag(x) A^{\top} f(A,x) \cdot f(A,x)_{j_1} \\
  & - ~c_g(\wt{A},x)^{\top} \cdot f(\wt{A},x)_{j_1}\diag(x) \wt{A}^{\top} f(A,x) \cdot f(A,x)_{j_1}|  \\
  \leq & ~ \| c_g(\wt{A},x)^{\top}\|_2 \cdot |f(\wt{A},x)_{j_1}|\cdot \| x\|_2 \cdot \| A^{\top} - \wt{A}^{\top}\|_F \cdot \|f(A,x)\|_2 \cdot | f(A,x)_{j_1} | \\
  \leq & ~ 5R^2 \cdot \| A^{\top} - \wt{A}^{\top}\|_F
\end{align*}
where the first step follows from Fact~\ref{fac:vector_norm} and Fact~\ref{fac:matrix_norm}, and the second step follows from {\bf Part 3, 6, 7} of Lemma~\ref{lem:upper_bound}, and $\|x \|_2 \leq R, \|A \|_F \leq R$.

For the fourth term, we have 
\begin{align*}
    & ~ |c_g(\wt{A},x)^{\top} \cdot f(\wt{A},x)_{j_1}\diag(x) \wt{A}^{\top} f(A,x) \cdot f(A,x)_{j_1} \\
  & - ~c_g(\wt{A},x)^{\top} \cdot f(\wt{A},x)_{j_1}\diag(x) \wt{A}^{\top} f(\wt{A},x) \cdot f(A,x)_{j_1} |  \\
  \leq & ~ \| c_g(\wt{A},x)^{\top}\|_2 \cdot |f(\wt{A},x)_{j_1}|\cdot \| x\|_2 \cdot \| \wt{A}^{\top}\|_F \cdot \|f(A,x) - f(\wt{A},x)\|_2 \cdot | f(A,x)_{j_1} | \\
  \leq & ~ 10\beta^{-2} \cdot n \cdot R^3 \cdot \exp(3R^2) \cdot \|A - \wt{A}\|_F
\end{align*}
where the first step follows from Fact~\ref{fac:vector_norm} and Fact~\ref{fac:matrix_norm}, and the second step follows from {\bf Part 4} of Lemma~\ref{lem:basic_lips}, {\bf Part 3, 6, 7} of Lemma~\ref{lem:upper_bound}, and $\|x \|_2 \leq R, \|A \|_F \leq R$.

For fifth term, it is similar to the second term. Then, we have
\begin{align*}
   & ~ |G_2(A) - G_2(\wt{A})|  \\
  \leq  &   ~  20\beta^{-2} \cdot n \cdot R^3 \cdot  \exp(3R^{2}) \cdot \|A - \wt{A}\|_F  \\
  & +   ~  10\beta^{-2} \cdot n \cdot R^3 \cdot \exp(3R^2) \cdot \|A - \wt{A}\|_F \\
  & +   ~  5R^2 \cdot \| A^{\top} - \wt{A}^{\top}\|_F \\
  & +   ~  10\beta^{-2} \cdot n \cdot R^3 \cdot \exp(3R^2) \cdot \|A - \wt{A}\|_F \\
  & +   ~  10\beta^{-2} \cdot n \cdot R^3 \cdot \exp(3R^2) \cdot \|A - \wt{A}\|_F \\
  \leq & ~ \beta^{-2} \cdot n \cdot \exp(6R^2) \cdot \|A - \wt{A}\|_F
\end{align*}
where the second step follows from $\beta^{-1} \geq 1, n > 1, R \leq \exp(R^2), R^2 \leq \exp(R^2)$ and $R \geq 4$.

Then, we have
\begin{align*}
  & ~ |- c_g(A,x)^{\top} \cdot f(A,x)_{j_1}\diag(x) A^{\top} f(A,x) \cdot (\langle -f(A,x), f(A,x) \rangle + f(A,x)_{j_1})   \\
  & - ~ (- c_g(\wt{A},x)^{\top} \cdot f(\wt{A},x)_{j_1}\diag(x) \wt{A}^{\top} f(\wt{A},x) \cdot (\langle -f(\wt{A},x), f(\wt{A},x) \rangle + f(\wt{A},x)_{j_1}))  | \\
 \leq & ~ |G_1(A) - G_1(\wt{A})| + |G_2(A) -G_2(\wt{A})| \\
 \leq & ~ 2\beta^{-2} \cdot n \cdot \exp(6R^2) \cdot \|A - \wt{A}\|_F
  \end{align*}
\end{proof}

\subsection{Sixth term \texorpdfstring{$- c_g(A,x)^{\top} \cdot f(A,x)_{j_1}\diag(x) A^{\top} f(A,x) \cdot (\langle -f(A,x), c(A,x) \rangle + f(A,x)_{j_1}) $}{}}
\begin{lemma}\label{lem:lips_gradient_6}
If the following conditions hold
\begin{itemize}
    \item Let $A_{j_1,*} \in \R^d$ for all $j_1 \in [n]$
    \item Let $x \in \R^d$
    \item Let $f(A,x) \in \R^n$ be defined as Definition~\ref{def:f}
    \item Let $c(A,x) \in \R^n$ be defined as Definition~\ref{def:c}
    \item Let $g(A,x) \in \R^d$ be defined as Definition~\ref{def:g} 
    \item Let $q(A,x) = c(A,x) + f(A,x) \in \R^n$
    \item Let $c_g(A,x) \in \R^d$ be defined as Definition~\ref{def:c_g}.
\end{itemize}
Then we have
\begin{align*}
 & ~ |- c_g(A,x)^{\top} \cdot f(A,x)_{j_1}\diag(x) A^{\top} f(A,x) \cdot (\langle -f(A,x), c(A,x) \rangle + f(A,x)_{j_1})   \\
  & - ~ (- c_g(\wt{A},x)^{\top} \cdot f(\wt{A},x)_{j_1}\diag(x) \wt{A}^{\top} f(\wt{A},x) \cdot (\langle -f(\wt{A},x), c(\wt{A},x) \rangle + f(\wt{A},x)_{j_1})  | \\
 \leq & ~  2\beta^{-2} \cdot n \cdot \exp(6R^2) \cdot \|A - \wt{A}\|_F
\end{align*}
\end{lemma}
\begin{proof}
    We define 
\begin{align*}
    G_1(A) := & ~  c_g(A,x)^{\top} \cdot f(A,x)_{j_1}\diag(x) A^{\top} f(A,x) \cdot \langle -f(A,x), c(A,x) \rangle \\
    G_2(A) := & ~  c_g(A,x)^{\top} \cdot f(A,x)_{j_1}\diag(x) A^{\top} f(A,x) \cdot f(A,x)_{j_1}
\end{align*}
      Firstly, we have
      \begin{align*}
          & ~ |- c_g(A,x)^{\top} \cdot f(A,x)_{j_1}\diag(x) A^{\top} f(A,x) \cdot (\langle -f(A,x), c(A,x) \rangle + f(A,x)_{j_1})   \\
  & - ~ (- c_g(\wt{A},x)^{\top} \cdot f(\wt{A},x)_{j_1}\diag(x) \wt{A}^{\top} f(\wt{A},x) \cdot (\langle -f(\wt{A},x), c(\wt{A},x) \rangle + f(\wt{A},x)_{j_1}))  | \\
 = & ~ | c_g(A,x)^{\top} \cdot f(A,x)_{j_1}\diag(x) A^{\top} f(A,x) \cdot (\langle -f(A,x), c(A,x) \rangle + f(A,x)_{j_1})   \\
  &  - ~ c_g(\wt{A},x)^{\top} \cdot f(\wt{A},x)_{j_1}\diag(x) \wt{A}^{\top} f(\wt{A},x) \cdot (\langle -f(\wt{A},x), c(\wt{A},x) \rangle + f(\wt{A},x)_{j_1}) | \\
  = & ~ |G_1(A) + G_2(A) - G_1(\wt{A}) -G_2(\wt{A}) |\\
  \leq & ~ |G_1(A) - G_1(\wt{A})| + |G_2(A) -G_2(\wt{A})| 
  \end{align*}
  For the  $|G_1(A) - G_1(\wt{A})|$, we have
\begin{align*}
   & ~ |G_1(A) - G_1(\wt{A})|  \\
  =  &   ~ |c_g(A,x)^{\top} \cdot f(A,x)_{j_1}\diag(x) A^{\top} f(A,x) \cdot \langle -f(A,x), c(A,x) \rangle \\
  & - ~ c_g(\wt{A},x)^{\top} \cdot f(\wt{A},x)_{j_1}\diag(x) \wt{A}^{\top} f(\wt{A},x) \cdot \langle -f(\wt{A},x), c(\wt{A},x) \rangle |  \\
  \leq & ~ |c_g(A,x)^{\top} \cdot f(A,x)_{j_1}\diag(x) A^{\top} f(A,x) \cdot \langle -f(A,x), c(A,x) \rangle \\
  & - ~c_g(\wt{A},x)^{\top} \cdot f(A,x)_{j_1}\diag(x) A^{\top} f(A,x) \cdot \langle -f(A,x), c(A,x) \rangle |  \\
   + & ~ |c_g(\wt{A},x)^{\top} \cdot f(A,x)_{j_1}\diag(x) A^{\top} f(A,x) \cdot \langle -f(A,x), c(A,x) \rangle \\
  & - ~c_g(\wt{A},x)^{\top} \cdot f(\wt{A},x)_{j_1}\diag(x) A^{\top} f(A,x) \cdot \langle -f(A,x), c(A,x) \rangle |  \\
  + & ~ |c_g(\wt{A},x)^{\top} \cdot f(\wt{A},x)_{j_1}\diag(x) A^{\top} f(A,x) \cdot \langle -f(A,x), c(A,x) \rangle \\
  & - ~c_g(\wt{A},x)^{\top} \cdot f(\wt{A},x)_{j_1}\diag(x) \wt{A}^{\top} f(A,x) \cdot \langle -f(A,x), c(A,x) \rangle |  \\
   + & ~ |c_g(\wt{A},x)^{\top} \cdot f(\wt{A},x)_{j_1}\diag(x) \wt{A}^{\top} f(A,x) \cdot \langle -f(A,x), c(A,x) \rangle \\
  & - ~c_g(\wt{A},x)^{\top} \cdot f(\wt{A},x)_{j_1}\diag(x) \wt{A}^{\top} f(\wt{A},x) \cdot \langle -f(A,x), c(A,x) \rangle |  \\
   + & ~ |c_g(\wt{A},x)^{\top} \cdot f(\wt{A},x)_{j_1}\diag(x) \wt{A}^{\top} f(\wt{A},x) \cdot \langle -f(A,x), c(A,x) \rangle \\
  & - ~c_g(\wt{A},x)^{\top} \cdot f(\wt{A},x)_{j_1}\diag(x) \wt{A}^{\top} f(\wt{A},x) \cdot \langle -f(\wt{A},x), c(A,x) \rangle |  \\
   + & ~ |c_g(\wt{A},x)^{\top} \cdot f(\wt{A},x)_{j_1}\diag(x) \wt{A}^{\top} f(\wt{A},x) \cdot \langle -f(\wt{A},x), c(A,x) \rangle \\
  & - ~c_g(\wt{A},x)^{\top} \cdot f(\wt{A},x)_{j_1}\diag(x) \wt{A}^{\top} f(\wt{A},x) \cdot \langle -f(\wt{A},x), c(\wt{A},x) \rangle | 
\end{align*}
For the first term, we have 
\begin{align*}
   & ~ |c_g(A,x)^{\top} \cdot f(A,x)_{j_1}\diag(x) A^{\top} f(A,x) \cdot \langle -f(A,x), c(A,x) \rangle \\
  & - ~ c_g(\wt{A},x)^{\top} \cdot  f(A,x)_{j_1}\diag(x) A^{\top} f(A,x) \cdot \langle -f(A,x), c(A,x) \rangle |  \\ 
  \leq & ~  \| c_g(A,x)^{\top} - c_g(\wt{A},x)^{\top} \|_2 \cdot | f(A,x)_{j_1} |\cdot \|x \|_2 \cdot \|A^{\top}\|_F \cdot \|f(A,x) \|_2 \cdot \|f(A,x) \|_2 \cdot \|c(A,x) \|_2 \\
  \leq & ~ 40\beta^{-2} \cdot n \cdot R^3 \cdot  \exp(3R^{2}) \cdot \|A - \wt{A}\|_F 
\end{align*}
where the first step follows from Fact~\ref{fac:vector_norm} and Fact~\ref{fac:matrix_norm}, and the second step follows from {\bf Part 8} of Lemma~\ref{lem:basic_lips}, {\bf Part 3, 4, 7} of Lemma~\ref{lem:upper_bound} and $\|x \|_2 \leq R, \|A \|_F \leq R$.

For the second term, we have
\begin{align*}
    & ~ |c_g(\wt{A},x)^{\top} \cdot f(A,x)_{j_1}\diag(x) A^{\top} f(A,x) \cdot \langle -f(A,x), c(A,x) \rangle \\
  & - ~c_g(\wt{A},x)^{\top} \cdot f(\wt{A},x)_{j_1}\diag(x) A^{\top} f(A,x) \cdot \langle -f(A,x), c(A,x) \rangle | \\
  \leq & ~ \| c_g(\wt{A},x)^{\top}\|_2 \cdot |f(A,x)_{j_1} - f(\wt{A},x)_{j_1}|\cdot \| x\|_2 \cdot \| A^{\top}\|_F \cdot \|f(A,x)\|_2 \cdot \|f(A,x) \|_2 \cdot \|c(A,x) \|_2 \\
  \leq & ~ 20\beta^{-2} \cdot n \cdot R^3 \cdot \exp(3R^2) \cdot \|A - \wt{A}\|_F 
\end{align*}
where the first step follows from Fact~\ref{fac:vector_norm} and Fact~\ref{fac:matrix_norm}, and the second step follows from {\bf Part 8} of Lemma~\ref{lem:basic_lips}, {\bf Part 3, 4, 6} of Lemma~\ref{lem:upper_bound}, and $\|x \|_2 \leq R, \|A \|_F \leq R$.

For the third term, we have 
\begin{align*}
    & ~ |c_g(\wt{A},x)^{\top} \cdot f(\wt{A},x)_{j_1}\diag(x) A^{\top} f(A,x) \cdot \langle -f(A,x), c(A,x) \rangle \\
  & - ~c_g(\wt{A},x)^{\top} \cdot f(\wt{A},x)_{j_1}\diag(x) \wt{A}^{\top} f(A,x) \cdot \langle -f(A,x), c(A,x) \rangle |  \\
  \leq & ~ \| c_g(\wt{A},x)^{\top}\|_2 \cdot |f(\wt{A},x)_{j_1}|\cdot \| x\|_2 \cdot \| A^{\top} - \wt{A}^{\top}\|_F \cdot \|f(A,x)\|_2 \cdot \|f(A,x) \|_2 \cdot \|c(A,x) \|_2 \\
  \leq & ~ 10R^2 \cdot \| A^{\top} - \wt{A}^{\top}\|_F
\end{align*}
where the first step follows from Fact~\ref{fac:vector_norm} and Fact~\ref{fac:matrix_norm}, and the second step follows from {\bf Part 3, 4, 6, 7} of Lemma~\ref{lem:upper_bound}, and $\|x \|_2 \leq R, \|A \|_F \leq R$.

For the fourth term, we have 
\begin{align*}
    & ~ |c_g(\wt{A},x)^{\top} \cdot f(\wt{A},x)_{j_1}\diag(x) \wt{A}^{\top} f(A,x) \cdot \langle -f(A,x), c(A,x) \rangle \\
  & - ~c_g(\wt{A},x)^{\top} \cdot f(\wt{A},x)_{j_1}\diag(x) \wt{A}^{\top} f(\wt{A},x) \cdot \langle -f(A,x), c(A,x) \rangle |  \\
  \leq & ~ \| c_g(\wt{A},x)^{\top}\|_2 \cdot |f(\wt{A},x)_{j_1}|\cdot \| x\|_2 \cdot \| \wt{A}^{\top}\|_F \cdot \|f(A,x) - f(\wt{A},x)\|_2 \cdot \|f(A,x) \|_2 \cdot \|c(A,x) \|_2 \\
  \leq & ~ 20\beta^{-2} \cdot n \cdot R^3 \cdot \exp(3R^2) \cdot \|A - \wt{A}\|_F
\end{align*}
where the first step follows from Fact~\ref{fac:vector_norm} and Fact~\ref{fac:matrix_norm}, and the second step follows from {\bf Part 4} of Lemma~\ref{lem:basic_lips}, {\bf Part 3, 4, 6, 7} of Lemma~\ref{lem:upper_bound}, and $\|x \|_2 \leq R, \|A \|_F \leq R$.

For fifth term, it is similar to the fourth term.

For sixth term, we have
\begin{align*}
    & ~ |c_g(\wt{A},x)^{\top} \cdot f(\wt{A},x)_{j_1}\diag(x) \wt{A}^{\top} f(\wt{A},x) \cdot \langle -f(\wt{A},x), c(A,x) \rangle \\
  & - ~c_g(\wt{A},x)^{\top} \cdot f(\wt{A},x)_{j_1}\diag(x) \wt{A}^{\top} f(\wt{A},x) \cdot \langle -f(\wt{A},x), c(\wt{A},x) \rangle |  \\
  \leq & ~ \| c_g(\wt{A},x)^{\top}\|_2 \cdot |f(\wt{A},x)_{j_1}|\cdot \| x\|_2 \cdot \| \wt{A}^{\top}\|_F \cdot \|f(\wt{A},x)\|_2 \cdot \|f(\wt{A},x) \|_2 \cdot \|c(A,x)- c(\wt{A},x) \|_2 \\
  \leq & ~ 10\beta^{-2} \cdot n \cdot R^3 \cdot \exp(3R^2) \cdot \|A - \wt{A}\|_F
\end{align*}
where the first step follows from Fact~\ref{fac:vector_norm} and Fact~\ref{fac:matrix_norm}, and the second step follows from {\bf Part 5} of Lemma~\ref{lem:basic_lips}, {\bf Part 3, 4, 6, 7} of Lemma~\ref{lem:upper_bound}, and $\|x \|_2 \leq R, \|A \|_F \leq R$.

Then, we have
\begin{align*}
   & ~ |G_1(A) - G_1(\wt{A})|  \\
  \leq  &   ~  40\beta^{-2} \cdot n \cdot R^3 \cdot  \exp(3R^{2}) \cdot \|A - \wt{A}\|_F \\
  & +   ~  20\beta^{-2} \cdot n \cdot R^3 \cdot \exp(3R^2) \cdot \|A - \wt{A}\|_F \\
  & +   ~  10R^2 \cdot \| A^{\top} - \wt{A}^{\top}\|_F \\
  & +   ~  20\beta^{-2} \cdot n \cdot R^3 \cdot \exp(3R^2) \cdot \|A - \wt{A}\|_F \\
  & +   ~  20\beta^{-2} \cdot n \cdot R^3 \cdot \exp(3R^2) \cdot \|A - \wt{A}\|_F \\
  & +   ~  10\beta^{-2} \cdot n \cdot R^3 \cdot \exp(3R^2) \cdot \|A - \wt{A}\|_F \\
  \leq & ~ \beta^{-2} \cdot n \cdot \exp(6R^2) \cdot \|A - \wt{A}\|_F
\end{align*}
where the second step follows from $\beta^{-1} \geq 1, n > 1, R \leq \exp(R^2), R^2 \leq \exp(R^2)$ and $R \geq 4$.

For the  $|G_2(A) - G_2(\wt{A})|$, we have
\begin{align*}
   & ~ |G_2(A) - G_2(\wt{A})|  \\
  =  &   ~ |c_g(A,x)^{\top} \cdot f(A,x)_{j_1}\diag(x) A^{\top} f(A,x) \cdot f(A,x)_{j_1} \\
  & - ~ c_g(\wt{A},x)^{\top} \cdot f(\wt{A},x)_{j_1}\diag(x) \wt{A}^{\top} f(\wt{A},x) \cdot f(\wt{A},x)_{j_1}|  \\
  \leq & ~   \beta^{-2} \cdot n \cdot \exp(6R^2) \cdot \|A - \wt{A}\|_F
\end{align*}
where the last step follows from Lemma~\ref{lem:lips_gradient_5}.

Then, we have
\begin{align*}
  & ~ |- c_g(A,x)^{\top} \cdot f(A,x)_{j_1}\diag(x) A^{\top} f(A,x) \cdot (\langle -f(A,x), c(A,x) \rangle + f(A,x)_{j_1})   \\
  & - ~ (- c_g(\wt{A},x)^{\top} \cdot f(\wt{A},x)_{j_1}\diag(x) \wt{A}^{\top} f(\wt{A},x) \cdot (\langle -f(\wt{A},x), c(\wt{A},x) \rangle + f(\wt{A},x)_{j_1}))  | \\
 \leq & ~ |G_1(A) - G_1(\wt{A})| + |G_2(A) -G_2(\wt{A})| \\
 \leq & ~ 2\beta^{-2} \cdot n \cdot \exp(6R^2) \cdot \|A - \wt{A}\|_F
  \end{align*}
\end{proof}

\subsection{Seventh term \texorpdfstring{$- c_g(A,x)^{\top} \cdot f(A,x)_{j_1}\diag(x) A^{\top} ((e_{j_1} - f(A,x)) \circ q(A,x)) $}{}}
\begin{lemma}\label{lem:lips_gradient_7}
If the following conditions hold
\begin{itemize}
    \item Let $A_{j_1,*} \in \R^d$ for all $j_1 \in [n]$
    \item Let $x \in \R^d$
    \item Let $f(A,x) \in \R^n$ be defined as Definition~\ref{def:f}
    \item Let $c(A,x) \in \R^n$ be defined as Definition~\ref{def:c}
    \item Let $g(A,x) \in \R^d$ be defined as Definition~\ref{def:g} 
    \item Let $q(A,x) = c(A,x) + f(A,x) \in \R^n$
    \item Let $c_g(A,x) \in \R^d$ be defined as Definition~\ref{def:c_g}.
\end{itemize}
Then we have
\begin{align*}
& ~  |- c_g(A,x)^{\top} \cdot f(A,x)_{j_1}\diag(x) A^{\top} ((e_{j_1} - f(A,x)) \circ q(A,x)) \\ 
- & ~( - c_g(\wt{A},x)^{\top} \cdot f(\wt{A},x)_{j_1}\diag(x) \wt{A}^{\top} ((e_{j_1} - f(\wt{A},x)) \circ q(\wt{A},x)) )| \\
= & ~ 2\beta^{-2} \cdot n \cdot \exp(6R^2) \cdot \|A - \wt{A} \|_F
\end{align*}
\end{lemma}
\begin{proof}
We define 
\begin{align*}
    G_1(A) := & ~  c_g(A,x)^{\top} \cdot f(A,x)_{j_1}\diag(x) A^{\top} \cdot e_{j_1}  \cdot \diag(q(A,x)) \\
    G_2(A) := & ~  - c_g(A,x)^{\top} \cdot f(A,x)_{j_1}\diag(x) A^{\top} \cdot f(A,x) \cdot \diag(q(A,x))
\end{align*}
      Firstly, we have
      \begin{align*}
          & ~ |- c_g(A,x)^{\top} \cdot f(A,x)_{j_1}\diag(x) A^{\top} ((e_{j_1} - f(A,x)) \circ q(A,x))  \\
  & - ~ (- c_g(\wt{A},x)^{\top} \cdot f(\wt{A},x)_{j_1}\diag(x) \wt{A}^{\top} ((e_{j_1} - f(\wt{A},x)) \circ q(\wt{A},x)) | \\
 = & ~ | c_g(A,x)^{\top} \cdot f(A,x)_{j_1}\diag(x) A^{\top} \cdot (e_{j_1} - f(A,x)) \cdot \diag(q(A,x))   \\
  &  - ~ c_g(\wt{A},x)^{\top} \cdot f(\wt{A},x)_{j_1}\diag(x) \wt{A}^{\top} \cdot  (e_{j_1} - f(\wt{A},x)) \cdot \diag(q(\wt{A},x)) | \\
  = & ~ |G_1(A) - G_2(A) - G_1(\wt{A}) +G_2(\wt{A}) |\\
  \leq & ~ |G_1(A) - G_1(\wt{A})| + |G_2(A) -G_2(\wt{A})| 
  \end{align*}
  For the  $|G_1(A) - G_1(\wt{A})|$, we have
\begin{align*}
   & ~ |G_1(A) - G_1(\wt{A})|  \\
  =  &   ~ |c_g(A,x)^{\top} \cdot f(A,x)_{j_1}\diag(x) A^{\top} \cdot e_{j_1}  \cdot \diag(q(A,x)) \\
  & - ~ c_g(\wt{A},x)^{\top} \cdot f(\wt{A},x)_{j_1}\diag(x) \wt{A}^{\top} \cdot e_{j_1}  \cdot \diag(q(\wt{A},x)) |\\
  \leq  &   ~ |c_g(A,x)^{\top} \cdot f(A,x)_{j_1}\diag(x) A^{\top} \cdot e_{j_1}  \cdot \diag(q(A,x)) \\
  & - ~ c_g(\wt{A},x)^{\top} \cdot f(A,x)_{j_1}\diag(x) A^{\top} \cdot e_{j_1}  \cdot \diag(q(A,x)) |\\
  +  &   ~ |c_g(\wt{A},x)^{\top} \cdot f(A,x)_{j_1}\diag(x) A^{\top} \cdot e_{j_1}  \cdot \diag(q(A,x)) \\
  & - ~ c_g(\wt{A},x)^{\top} \cdot f(\wt{A},x)_{j_1}\diag(x) A^{\top} \cdot e_{j_1}  \cdot \diag(q(A,x)) |\\
  +  &   ~ |c_g(\wt{A},x)^{\top} \cdot f(\wt{A},x)_{j_1}\diag(x) A^{\top} \cdot e_{j_1}  \cdot \diag(q(A,x)) \\
  & - ~ c_g(\wt{A},x)^{\top} \cdot f(\wt{A},x)_{j_1}\diag(x) \wt{A}^{\top} \cdot e_{j_1}  \cdot \diag(q(A,x)) |\\
  +  &   ~ |c_g(\wt{A},x)^{\top} \cdot f(\wt{A},x)_{j_1}\diag(x) \wt{A}^{\top} \cdot e_{j_1}  \cdot \diag(q(A,x)) \\
  & - ~ c_g(\wt{A},x)^{\top} \cdot f(\wt{A},x)_{j_1}\diag(x) \wt{A}^{\top} \cdot e_{j_1}  \cdot \diag(q(\wt{A},x)) |\\
\end{align*}
For the first term, we have
\begin{align*}
   & ~|c_g(A,x)^{\top} \cdot f(A,x)_{j_1}\diag(x) A^{\top} \cdot e_{j_1}  \cdot \diag(q(A,x)) \\
  & - ~ c_g(\wt{A},x)^{\top} \cdot f(A,x)_{j_1}\diag(x) A^{\top} \cdot e_{j_1}  \cdot \diag(q(A,x)) |\\ 
  \leq & ~  \| c_g(A,x)^{\top} - c_g(\wt{A},x)^{\top} \|_2 \cdot | f(A,x)_{j_1} |\cdot \|x \|_2 \cdot \|A^{\top}\|_F \cdot \|e_{j_1}\|_2 \cdot \|q(A,x) \|_2 \\
  \leq & ~ 60 \beta^{-2} \cdot n \cdot R^3 \cdot  \exp(3R^{2}) \cdot \|A - \wt{A}\|_F
\end{align*}
  where the first step follows from Fact~\ref{fac:vector_norm} and Fact~\ref{fac:matrix_norm}, and the second step follows from {\bf Part 8} of Lemma~\ref{lem:basic_lips}, {\bf Part 7, 10} of Lemma~\ref{lem:upper_bound}, and $\|\diag(x) \|_2 \leq \|x \|_2 \leq R, \|\diag(q(A,x)) \| \leq \|q(A,x) \|_2, \|A \|_F \leq R$.

  For the second term, we have
\begin{align*}
   & ~|c_g(\wt{A},x)^{\top} \cdot f(A,x)_{j_1}\diag(x) A^{\top} \cdot e_{j_1}  \cdot \diag(q(A,x)) \\
  & - ~ c_g(\wt{A},x)^{\top} \cdot f(\wt{A},x)_{j_1}\diag(x) A^{\top} \cdot e_{j_1}  \cdot \diag(q(A,x)) |\\ 
  \leq & ~  \|c_g(\wt{A},x)^{\top} \|_2 \cdot | f(A,x)_{j_1} - f(\wt{A},x)_{j_1} |\cdot \|x \|_2 \cdot \|A^{\top}\|_F \cdot \|e_{j_1}\|_2 \cdot \|q(A,x) \|_2 \\
  \leq & ~ 30\beta^{-2} \cdot n \cdot R^3 \cdot \exp(3R^2) \cdot \|A - \wt{A}\|_F
\end{align*}
  where the first step follows from Fact~\ref{fac:vector_norm} and Fact~\ref{fac:matrix_norm}, and the second step follows from {\bf Part 4} of Lemma~\ref{lem:basic_lips}, {\bf Part 6, 10} of Lemma~\ref{lem:upper_bound}, and $\|\diag(x) \|_2 \leq \|x \|_2 \leq R, \|\diag(q(A,x)) \| \leq \|q(A,x) \|_2, \|A \|_F \leq R$.

  For the third term, we have
\begin{align*}
   & ~|c_g(\wt{A},x)^{\top} \cdot f(\wt{A},x)_{j_1}\diag(x) A^{\top} \cdot e_{j_1}  \cdot \diag(q(A,x)) \\
  & - ~ c_g(\wt{A},x)^{\top} \cdot f(\wt{A},x)_{j_1}\diag(x) \wt{A}^{\top} \cdot e_{j_1}  \cdot \diag(q(A,x)) |\\ 
  \leq & ~  \|c_g(\wt{A},x)^{\top} \|_2 \cdot | f(\wt{A},x)_{j_1} |\cdot \|x \|_2 \cdot \|A^{\top} - \wt{A}^{\top}\|_F \cdot \|e_{j_1}\|_2 \cdot \|q(A,x) \|_2 \\
  \leq & ~ 15R \cdot \|A - \wt{A}\|_F
\end{align*}
  where the first step follows from Fact~\ref{fac:vector_norm} and Fact~\ref{fac:matrix_norm}, and the second step follows from {\bf Part 6, 7, 10} of Lemma~\ref{lem:upper_bound}, and $\|\diag(x) \|_2 \leq \|x \|_2 \leq R, \|\diag(q(A,x)) \| \leq \|q(A,x) \|_2, \|A \|_F \leq R$.
  
      For the fourth term, we have
\begin{align*}
   & ~|c_g(\wt{A},x)^{\top} \cdot f(\wt{A},x)_{j_1}\diag(x) \wt{A}^{\top} \cdot e_{j_1}  \cdot \diag(q(A,x)) \\
  & - ~ c_g(\wt{A},x)^{\top} \cdot f(\wt{A},x)_{j_1}\diag(x) \wt{A}^{\top} \cdot e_{j_1}  \cdot \diag(q(\wt{A},x)) |\\ 
  \leq & ~  \|c_g(\wt{A},x)^{\top} \|_2 \cdot | f(\wt{A},x)_{j_1} |\cdot \|x \|_2 \cdot \| \wt{A}^{\top}\|_F \cdot \|e_{j_1}\|_2 \cdot \|q(A,x) -q(\wt{A},x)\|_2 \\
  \leq & ~ 20 \beta^{-2} \cdot n \cdot R^3 \cdot \exp(3R^2) \cdot \|A - \wt{A} \|_F
\end{align*}
  where the first step follows from Fact~\ref{fac:vector_norm} and Fact~\ref{fac:matrix_norm}, and the second step follows from {\bf Part 7} of Lemma~\ref{lem:basic_lips}, {\bf Part 6, 7} of Lemma~\ref{lem:upper_bound}, and $\|\diag(x) \|_2 \leq \|x \|_2 \leq R, \|\diag(q(A,x)) \| \leq \|q(A,x) \|_2, \|A \|_F \leq R$.
  
Then, we have
\begin{align*}
  & ~ |G_1(A) - G_1(\wt{A})|  \\
 \leq & ~ 60\beta^{-2} \cdot n \cdot R^3 \cdot  \exp(3R^{2}) \cdot \|A - \wt{A}\|_F \\
 & + ~ 30\beta^{-2} \cdot n \cdot R^3 \cdot \exp(3R^2) \cdot \|A - \wt{A}\|_F\\
 & + ~ 15R^2 \cdot \|A - \wt{A}\|_F \\
  & + ~ 20 \beta^{-2} \cdot n \cdot R^3 \cdot \exp(3R^2) \cdot \|A - \wt{A} \|_F \\
  \leq & ~ \beta^{-2} \cdot n \cdot \exp(6R^2) \cdot \|A - \wt{A} \|_F
  \end{align*}
  where the second step follows from $\beta^{-1} \geq 1, n > 1, R \leq \exp(R^2), R^2 \leq \exp(R^2)$ and $R \geq 4$.

  For the  $|G_2(A) - G_2(\wt{A})|$, we have
\begin{align*}
   & ~ |G_2(A) - G_2(\wt{A})|  \\
   =  &   ~ | - c_g(A,x)^{\top} \cdot f(A,x)_{j_1}\diag(x) A^{\top} \cdot f(A,x)  \cdot \diag(q(A,x)) \\
  & - ~ (-c_g(\wt{A},x)^{\top} \cdot f(\wt{A},x)_{j_1}\diag(x) \wt{A}^{\top} \cdot f(\wt{A},x)  \cdot \diag(q(\wt{A},x))) |\\
  =  &   ~ | c_g(A,x)^{\top} \cdot f(A,x)_{j_1}\diag(x) A^{\top} \cdot f(A,x)  \cdot \diag(q(A,x)) \\
  & - ~ c_g(\wt{A},x)^{\top} \cdot f(\wt{A},x)_{j_1}\diag(x) \wt{A}^{\top} \cdot f(\wt{A},x)  \cdot \diag(q(\wt{A},x)) |\\
  \leq  &   ~ |c_g(A,x)^{\top} \cdot f(A,x)_{j_1}\diag(x) A^{\top} \cdot f(A,x) \cdot \diag(q(A,x)) \\
  & - ~ c_g(\wt{A},x)^{\top} \cdot f(A,x)_{j_1}\diag(x) A^{\top} \cdot f(A,x)  \cdot \diag(q(A,x)) |\\
  +  &   ~ |c_g(\wt{A},x)^{\top} \cdot f(A,x)_{j_1}\diag(x) A^{\top} \cdot f(A,x)  \cdot \diag(q(A,x)) \\
  & - ~ c_g(\wt{A},x)^{\top} \cdot f(\wt{A},x)_{j_1}\diag(x) A^{\top} \cdot f(A,x)  \cdot \diag(q(A,x)) |\\
  +  &   ~ |c_g(\wt{A},x)^{\top} \cdot f(\wt{A},x)_{j_1}\diag(x) A^{\top} \cdot f(A,x)  \cdot \diag(q(A,x)) \\
  & - ~ c_g(\wt{A},x)^{\top} \cdot f(\wt{A},x)_{j_1}\diag(x) \wt{A}^{\top} \cdot f(A,x)  \cdot \diag(q(A,x)) |\\
  +  &   ~ |c_g(\wt{A},x)^{\top} \cdot f(\wt{A},x)_{j_1}\diag(x) \wt{A}^{\top} \cdot f(A,x)  \cdot \diag(q(A,x)) \\
  & - ~ c_g(\wt{A},x)^{\top} \cdot f(\wt{A},x)_{j_1}\diag(x) \wt{A}^{\top} \cdot f(\wt{A},x)  \cdot \diag(q(A,x)) |\\
  +  &   ~ |c_g(\wt{A},x)^{\top} \cdot f(\wt{A},x)_{j_1}\diag(x) \wt{A}^{\top} \cdot f(\wt{A},x)  \cdot \diag(q(A,x)) \\
  & - ~ c_g(\wt{A},x)^{\top} \cdot f(\wt{A},x)_{j_1}\diag(x) \wt{A}^{\top} \cdot f(\wt{A},x)  \cdot \diag(q(\wt{A},x)) |\\
\end{align*}
For the first term, we have 
\begin{align*}
   & ~ |c_g(A,x)^{\top} \cdot f(A,x)_{j_1}\diag(x) A^{\top} f(A,x) \cdot \diag(q(A,x)) \\
  & - ~ c_g(\wt{A},x)^{\top} \cdot f(A,x)_{j_1}\diag(x) A^{\top} f(A,x) \cdot \diag(q(A,x))| \\
  \leq & ~  \| c_g(A,x)^{\top} - c_g(\wt{A},x)^{\top} \|_2 \cdot | f(A,x)_{j_1} |\cdot \|x \|_2 \cdot \|A^{\top}\|_F \cdot \|f(A,x) \|_2 \cdot \|q(A,x) \|_2  \\
  \leq & ~ 60\beta^{-2} \cdot n \cdot R^3 \cdot  \exp(3R^{2}) \cdot \|A - \wt{A}\|_F 
\end{align*}
where the first step follows from Fact~\ref{fac:vector_norm} and Fact~\ref{fac:matrix_norm}, and the second step follows from {\bf Part 8} of Lemma~\ref{lem:basic_lips}, {\bf Part 3, 7, 10} of Lemma~\ref{lem:upper_bound}, and $\|\diag(x) \|_2 \leq \|x \|_2 \leq R, \|\diag(q(A,x)) \| \leq \|q(A,x) \|_2, \|A \|_F \leq R$.

  For the second term, we have
\begin{align*}
   & ~|c_g(\wt{A},x)^{\top} \cdot f(A,x)_{j_1}\diag(x) A^{\top} \cdot f(A,x)  \cdot \diag(q(A,x)) \\
  & - ~ c_g(\wt{A},x)^{\top} \cdot f(\wt{A},x)_{j_1}\diag(x) A^{\top} \cdot f(A,x)  \cdot \diag(q(A,x)) |\\ 
  \leq & ~  \|c_g(\wt{A},x)^{\top} \|_2 \cdot | f(A,x)_{j_1} - f(\wt{A},x)_{j_1} |\cdot \|x \|_2 \cdot \|A^{\top}\|_F \cdot \|f(A,x)\|_2 \cdot \|q(A,x) \|_2 \\
  \leq & ~ 30\beta^{-2} \cdot n \cdot R^3 \cdot \exp(3R^2) \cdot \|A - \wt{A}\|_F
\end{align*}
where the first step follows from Fact~\ref{fac:vector_norm} and Fact~\ref{fac:matrix_norm}, and the second step follows from {\bf Part 4} of Lemma~\ref{lem:basic_lips}, {\bf Part 3, 6, 10} of Lemma~\ref{lem:upper_bound}, and $\|\diag(x) \|_2 \leq \|x \|_2 \leq R, \|\diag(q(A,x)) \| \leq \|q(A,x) \|_2, \|A \|_F \leq R$.

  For the third term, we have
\begin{align*}
   & ~|c_g(\wt{A},x)^{\top} \cdot f(\wt{A},x)_{j_1}\diag(x) A^{\top} \cdot f(A,x)   \cdot \diag(q(A,x)) \\
  & - ~ c_g(\wt{A},x)^{\top} \cdot f(\wt{A},x)_{j_1}\diag(x) \wt{A}^{\top} \cdot f(A,x)   \cdot \diag(q(A,x)) |\\ 
  \leq & ~  \|c_g(\wt{A},x)^{\top} \|_2 \cdot | f(\wt{A},x)_{j_1} |\cdot \|x \|_2 \cdot \|A^{\top} - \wt{A}^{\top}\|_F \cdot \|f(A,x) \|_2 \cdot \|q(A,x) \|_2 \\
  \leq & ~ 15R^2 \cdot \|A - \wt{A}\|_F
\end{align*}
  where the first step follows from Fact~\ref{fac:vector_norm} and Fact~\ref{fac:matrix_norm}, and the second step follows from {\bf Part 3, 6, 7, 10} of Lemma~\ref{lem:upper_bound}, and $\|\diag(x) \|_2 \leq \|x \|_2 \leq R, \|\diag(q(A,x)) \| \leq \|q(A,x) \|_2, \|A \|_F \leq R$.

  For the fourth term, we have
\begin{align*}
   & ~|c_g(\wt{A},x)^{\top} \cdot f(\wt{A},x)_{j_1}\diag(x) \wt{A}^{\top} \cdot f(A,x)  \cdot \diag(q(A,x)) \\
  & - ~ c_g(\wt{A},x)^{\top} \cdot f(\wt{A},x)_{j_1}\diag(x) \wt{A}^{\top} \cdot f(\wt{A},x)  \cdot \diag(q(\wt{A},x)) |\\ 
  \leq & ~  \|c_g(\wt{A},x)^{\top} \|_2 \cdot | f(\wt{A},x)_{j_1} |\cdot \|x \|_2 \cdot \| \wt{A}^{\top}\|_F \cdot \|f(A,x)-f(\wt{A},x)\|_2 \cdot \|q(A,x)\|_2 \\
  \leq & ~ 30\beta^{-2} \cdot n \cdot R^3 \cdot \exp(3R^2) \cdot \|A - \wt{A}\|_F
\end{align*}
  where the first step follows from Fact~\ref{fac:vector_norm} and Fact~\ref{fac:matrix_norm}, and the second step follows from {\bf Part 4} of Lemma~\ref{lem:basic_lips}, {\bf Part 3, 6, 7} of Lemma~\ref{lem:upper_bound}, and $\|\diag(x) \|_2 \leq \|x \|_2 \leq R, \|\diag(q(A,x)) \| \leq \|q(A,x) \|_2, \|A \|_F \leq R$.

  For the fifth term, we have
\begin{align*}
   & ~|c_g(\wt{A},x)^{\top} \cdot f(\wt{A},x)_{j_1}\diag(x) \wt{A}^{\top} \cdot f(\wt{A},x)  \cdot \diag(q(A,x)) \\
  & - ~ c_g(\wt{A},x)^{\top} \cdot f(\wt{A},x)_{j_1}\diag(x) \wt{A}^{\top} \cdot f(\wt{A},x)  \cdot \diag(q(\wt{A},x)) |\\ 
  \leq & ~  \|c_g(\wt{A},x)^{\top} \|_2 \cdot | f(\wt{A},x)_{j_1} |\cdot \|x \|_2 \cdot \| \wt{A}^{\top}\|_F \cdot \|f(\wt{A},x)\|_2 \cdot \|q(A,x) -q(\wt{A},x)\|_2 \\
  \leq & ~ 40 \beta^{-2} \cdot n \cdot R^2 \cdot \exp(3R^2) \cdot \|A - \wt{A} \|_F
\end{align*}
  where the first step follows from Fact~\ref{fac:vector_norm} and Fact~\ref{fac:matrix_norm}, and the second step follows from {\bf Part 7} of Lemma~\ref{lem:basic_lips}, {\bf Part 6, 7} of Lemma~\ref{lem:upper_bound}, and $\|\diag(x) \|_2 \leq \|x \|_2 \leq 2, \|\diag(q(A,x)) \| \leq \|q(A,x) \|_2, \|A \|_F \leq R$.
  
  Then, we have
\begin{align*}
  & ~ |G_2(A) - G_2(\wt{A})|  \\
 \leq & ~ 120\beta^{-2} \cdot n \cdot R^2 \cdot  \exp(3R^{2}) \cdot \|A - \wt{A}\|_F \\
 & + ~ 60\beta^{-2} \cdot n \cdot R^2 \cdot \exp(3R^2) \cdot \|A - \wt{A}\|_F\\
 & + ~ 30R \cdot \|A - \wt{A}\|_F \\
 & + ~ 60\beta^{-2} \cdot n \cdot R^2 \cdot \exp(3R^2) \cdot \|A - \wt{A}\|_F \\
  & + ~ 40 \beta^{-2} \cdot n \cdot R^2 \cdot \exp(3R^2) \cdot \|A - \wt{A} \|_F \\
  \leq & ~ \beta^{-2} \cdot n \cdot \exp(5R^2) \cdot \|A - \wt{A} \|_F
  \end{align*}
  where the second step follows from $\beta^{-1} \geq 1, n > 1, R \leq \exp(R^2), R^2 \leq \exp(R^2)$ and $R \geq 4$.

Then, we have
      \begin{align*}
          & ~ |- c_g(A,x)^{\top} \cdot f(A,x)_{j_1}\diag(x) A^{\top} ((e_{j_1} - f(A,x)) \circ q(A,x))  \\
  & - ~ (- c_g(\wt{A},x)^{\top} \cdot f(\wt{A},x)_{j_1}\diag(x) \wt{A}^{\top} ((e_{j_1} - f(\wt{A},x)) \circ q(\wt{A},x)) | \\
  \leq & ~ |G_1(A) - G_1(\wt{A})| + |G_2(A) -G_2(\wt{A})| \\
 = & ~ 2\beta^{-2} \cdot n \cdot \exp(6R^2) \cdot \|A - \wt{A} \|_F
  \end{align*}
\end{proof}

\section{Core Tool: Upper bound for basic functions}\label{app:upper_bound_basic}

In this sections, we compute the upper bound of basic functions 

\begin{lemma}\label{lem:upper_bound}
If the following conditions hold
    \begin{itemize}
     \item Let $u(A,x) \in \R^n$ be defined as Definition~\ref{def:u}
    \item Let $\alpha(A,x) \in \R$ be defined as Definition~\ref{def:alpha}
     \item Let $f(A,x) \in \R^n$ be defined as Definition~\ref{def:f}
    \item Let $c(A,x) \in \R^n$ be defined as Definition~\ref{def:c}
    \item Let $g(A,x) \in \R^d$ be defined as Definition~\ref{def:g} 
    \item Let $q(A,x) = c(A,x) + f(A,x) \in \R^n$
    \item Let $c_g(A,x) \in \R^d$ be defined as Definition~\ref{def:c_g}
    \item Let $L_g(A,x) \in \R$ be defined as Definition~\ref{def:l_g}
    \item Let $R \geq 4$ 
    \item $\|A \|_F \leq R, \|A^\top \|_F \leq R, \| x\|_2 \leq R, \|\diag(f(A,x)) \|_F \leq  \|f(A,x) \|_2 \leq 1$ 
    \end{itemize}
Then, we have
\begin{itemize}
    \item {\bf Part 1.} $\| u(A,x)\|_2 \leq \sqrt{n} \exp (R^2) $
    \item  {\bf Part 2.} $ | \alpha(A,x) | \leq  n \exp(R^2) $
    \item  {\bf Part 3.} $\| f(A,x) \|_2 \leq 1$
    \item  {\bf Part 4.} $\|c(A,x) \|_2 \leq 2$ 
    \item  {\bf Part 5.} $\|g(A,x) \|_2 \leq 4 R$ 
    \item  {\bf Part 6.} $\|c_g(A,x) \|_2 \leq 5 R$
    \item  {\bf Part 7.} For $j_0 \in [n]$, we have, $|f(A,x)_{j_0} | \leq 1$
    \item  {\bf Part 8.} For $j_0 \in [n]$, we have, $|u(A,x)_{j_0} | \leq \sqrt{n} \exp(R^2) $
    \item  {\bf Part 9.} For $j_0 \in [n]$, we have, $|c(A,x)_{j_0} | \leq 2 $
    \item  {\bf Part 10.} $\|q(A,x) \|_2 \leq 3$
    \end{itemize}
\end{lemma}
\begin{proof}

    {\bf Proof of Part 1}
    \begin{align*}
        \| u(A,x)\|_2 = & ~ \| \exp (Ax) \|_2\\
        \leq & ~ \sqrt{n}\|\exp(Ax) \|_\infty \\
        \leq & ~ \sqrt{n}\exp\|(Ax) \|_2 \\
        \leq & ~ \sqrt{n} \exp (R^2)
    \end{align*}
where the first step follows from the definition of $u(A,x)$ (see Definition~\ref{def:u}), the second step follows from Fact~\ref{fac:vector_norm}, the third step follows from Fact~\ref{fac:vector_norm}, and the last step follows from $\| A\|_F \leq R$ and $ \| x\|_2 \leq R$
    
    {\bf Proof of Part 2}
    \begin{align*}
    | \alpha(A,x) | = & ~ | \langle u(A,x), {\bf 1}_n \rangle | \\
    \leq & ~ \sqrt{n} \cdot \| u(A,x) \|_2 \\
    \leq & ~ \sqrt{n} \cdot \sqrt{n} \cdot \exp(R^2) \\
    = & ~ n \exp(R^2) 
\end{align*} 
where the first step comes from the definition of $\alpha(A,x)$ (see Definition~\ref{def:alpha}), the second is based on Fact~\ref{fac:vector_norm}, the third step follows from {\bf Part 1}, and the forth step follows from simple algebra.

{\bf Proof of Part 3.}
We have
\begin{align*}
    \| f(A,x) \|_2 \leq & ~ \| f(A, x) \|_1 \\
    = & ~ 1
\end{align*}
where the first step follows from Fact~\ref{fac:vector_norm}, the second step is due to Definition~\ref{def:f} and the fact that $f(A,x)$ is a normalized vector.

{\bf Proof of Part 4.}
\begin{align*}
    \|c(A,x)\|_2 =  & 
    ~\|f(A,x) - b_f \|_2  \\
    \leq & 
    ~ \|f(A,x) \|_2 + \| - b_f\|_2 \\
    \leq & 
~ 1 + 1 \\
    = & ~ 2
\end{align*}
where the first step comes from the definition of $c(A,x)$ (see Definition~\ref{def:c}), the second is based on Fact~\ref{fac:vector_norm}, the third step follows from $\|f(A,x) \|_2 \leq 1$ and $\| b_f \|_2 \leq 1$, and the last step follows from simple algebra.

    {\bf Proof of Part 5.}
We define \begin{align*}
    g_1(A,x) = & ~ - A^\top f(A,x)   \langle c(A, x), f(A, x) \rangle \\
    g_2(A,x) = & ~ A^\top \diag(f(A, x))  c(A,x) \\
    g(A,x) = & ~ g_1(A,x) + g_2(A,x)
\end{align*}

For $g_1(A,x)$, we have
\begin{align*}
    &  ~\|- A^\top f(A,x)   \langle c(A, x), f(A, x) \rangle \|_2 \\
    = & ~ \| A^\top f(A,x)   \langle c(A, x), f(A, x) \rangle \|_2 \\
    \leq & ~ \| A^\top\|_F \cdot \|f(A,x)\|_2 \cdot \|c(A,x)\|_2 \cdot \|f(A,x)\|_2 \\
    \leq & ~  2 R
\end{align*}
where the first step follows from Fact~\ref{fac:vector_norm}, the second step follows from Fact~\ref{fac:vector_norm} and Fact~\ref{fac:matrix_norm}, the last step follows from $\|A^{\top} \|_F \leq R$, {\bf Part 3} and {\bf Part 4}.

For $g_2(A,x)$, we have
\begin{align*}
    &  ~\|A^\top \diag(f(A, x))  c(A,x)\|_2 \\
    \leq & ~ \| A^\top\|_F \cdot \|\diag(f(A,x))\|_F \cdot \|c(A,x)\|_2  \\
    \leq & ~  2  R 
\end{align*}
where the first step follows from Fact~\ref{fac:vector_norm} and Fact~\ref{fac:matrix_norm}, second step follows from $\|\diag(f(A,x)) \|_F \leq 1$, $\|A^{\top} \|_F \leq R$ and {\bf Part 4}.

Then, by combining above two equations, we can get
\begin{align*}
    \|g(A,x) \|_2 \leq & ~ 2 R + 2 R\\
    \leq & ~ 4 R
\end{align*}

{\bf Proof of Part 6.}
\begin{align*}
    \|c_g(A,x) \|_2 = & ~ \|g(A,x) - b_g \|_2 \\
    \leq & ~ \|g(A,x)\|_2 +\|- b_g \|_2 \\
    \leq & ~ 4 R + 1 \\ 
    \leq & ~ 5 R
\end{align*}
where the first step follows from the definition of $c_g(A,x)$ (see Definition~\ref{def:c_g}), the second step follows from Fact~\ref{fac:vector_norm}, the third step follows from {\bf Part 5} and $\| b_f \|_2 \leq 1$, the last step follows from simple algebra.

{\bf Proof of Part 7.}
\begin{align*}
    |f(A,x)_{j_0} | \leq & ~ \sum_{j_0 = 1}^n\sqrt{|f(A,x)_{j_0}|^2} \\
    = & ~ \|f(A,x) \|_2 \\
    \leq & ~ 1
\end{align*}
where the first step follows from simple algebra, the second step follows from the definition of $\ell_2$ norm, and the last step follows from {\bf Part 3}.
{\bf Proof of Part 8.}
\begin{align*}
    |u(A,x)_{j_0} | \leq & ~ \sum_{j_0 = 1}^n\sqrt{|u(A,x)_{j_0}|^2} \\
    = & ~ \|u(A,x) \|_2 \\
    \leq & ~ \sqrt{n} \exp(R^2)
\end{align*}
where the first step follows from simple algebra, the second step follows from the definition of $\ell_2$ norm, and the last step follows from {\bf Part 1}. 

{\bf Proof of Part 9.}
\begin{align*}
    |c(A,x)_{j_0} | \leq & ~ \sum_{j_0 = 1}^n\sqrt{|c(A,x)_{j_0}|^2} \\
    = & ~ \|c(A,x) \|_2 \\
    \leq & ~ 2
\end{align*}
where the first step follows from simple algebra, the second step follows from the definition of $\ell_2$ norm, and the last step follows from {\bf Part 4}.

{\bf Proof of Part 10.}
\begin{align*}
    \|q(A,x) \|_2 = & ~ \|f(A,x) +  c(A,x) \|_2 \\
    \leq & ~ \|f(A,x)\|_2 +\|c(A,x) \|_2 \\
    \leq & ~ 1 + 2 \\ 
    \leq & ~ 3
\end{align*}
where the first step follows from the definition of $q(A,x) = f(A,x) + c(A,x)$, the second step follows from Fact~\ref{fac:vector_norm}, the third step follows from {\bf Part 3} and {\bf Part 4}, the last step follows from simple algebra.
\end{proof} 
\subsection{Upper Bound: Hessian Related}
\begin{lemma}\label{lem:upper_bound_hessian}
    
If the following conditions hold
    \begin{itemize}
     \item Let $f_c(A,x) \in \R^n$ be defined as Definition~\ref{def:f_c}
     \item Let $f_2(A,x) \in \R^n$ be defined as Definition~\ref{def:f_2}
     \item  Let $h(A,x) \in $ be defined as Definition~\ref{def:h}
      \item Let $p_{j_1} (A,x) := \diag(x) A^\top \cdot ( (e_{j_1} - f(A,x)) \circ q(A,x) ) \in \R^d$ for each $j_1 \in [n]$
    \item Let $R \geq 4$ 
    \item $\|A \|_F \leq R, \|A^\top \|_F \leq R, \| x\|_2 \leq R, \|\diag(f(A,x)) \|_F \leq  \|f(A,x) \|_2 \leq 1$ 
    \end{itemize}
    Then, we have
    \begin{itemize}
        \item {\bf Part 1.} $ |f_2(A,x)| \leq 1$ 
        \item {\bf Part 2.} $ |f_c(A,x)| \leq 2$
        \item {\bf Part 3.} $\|h(A,x)\|_2 \leq R^2$
        \item {\bf Part 4.} $\|h_e(A,x)\|_2 \leq 2R^2$
        \item  {\bf Part 4.} $ \| p_{j_1} (A,x)\|_2  \leq 6R^2$
    \end{itemize}
\end{lemma}

\begin{proof}
    {\bf Proof of Part 1.}
    \begin{align*}
        |f_2(A,x)| = & ~  |\langle f(A,x) , f(A,x)\rangle|\\
        \leq & ~ \| f(A,x) \|_2^2 \\
        \leq & ~ 1
    \end{align*}

    {\bf Proof of Part 2.}
    \begin{align*}
        |f_c(A,x)| = & ~  |\langle f(A,x) , c(A,x)\rangle|\\
        \leq & ~ \| f(A,x) \|_2 \| c(A,x)\|_2 \\
        \leq & ~ 2
    \end{align*}

    {\bf Proof of Part 3.}
    \begin{align*}
        \|h(A,x)\|_2 = & ~ \|\diag(x) A^\top f(A,x)\|_2 \\
        \leq & ~ R\|A^{\top} \|_F \|f(A,x) \|_2  \\
        \leq & ~ R^2
    \end{align*}
    
    {\bf Proof of Part 4.}
    \begin{align*}
        \|h_e(A,x)\|_2 = & ~ \|\diag(x) A^\top (e_{j_1}-f( A  ,x))\|_2 \\
        \leq & ~ R\|A^{\top} \|_F (\|e_{j_1} \|_2 +\|- f(A,x) \|_2)  \\
        \leq & ~ 2R^2
    \end{align*}

    {\bf Proof of Part 5.}
    \begin{align*}
        \| p_{j_1} (A,x)\|_2 
        = & ~ \| \diag(x) A^\top \cdot ( (e_{j_1} - f(A,x)) \circ q(A,x) ) \|_2\\
        \leq & ~ \|\diag(x)\|_F \|A^{\top} \|_F (\|e_{j_1} \|_2 +\|- f(A,x) \|_2) \|q(A,x) \|_2 \\
        \leq & ~ 6R^2 
    \end{align*}
\end{proof}
\section{Core Tool: Lipschitz property for basic functions}\label{app:lips_basic}

In this sections, we compute the lipschitz property of basic functions 

\subsection{Lipschitz: Gradient Related}

\begin{lemma}\label{lem:basic_lips}
If the following conditions hold
    \begin{itemize}
     \item Let $u(A,x) \in \R^n$ be defined as Definition~\ref{def:u}
    \item Let $\alpha(A,x) \in \R$ be defined as Definition~\ref{def:alpha}
     \item Let $f(A,x) \in \R^n$ be defined as Definition~\ref{def:f}
    \item Let $c(A,x) \in \R^n$ be defined as Definition~\ref{def:c}
    \item Let $g(A,x) \in \R^d$ be defined as Definition~\ref{def:g} 
    \item Let $q(A,x) = c(A,x) + f(A,x) \in \R^n$
    \item Let $c_g(A,x) \in \R^d$ be defined as Definition~\ref{def:c_g}.
    \item Let $L_g(A,x) \in \R$ be defined as Definition~\ref{def:l_g}
     \item Let $R \geq 4$
    \item $\|A \|_2 \leq R, \|A^{\top} \|_F \leq R, \| x\|_2 \leq R, \|\diag(f(A,x)) \|_F \leq \|f(A,x) \|_2 \leq 1, \| b_g \|_2 \leq 1$ 
    \item Let $\beta$ be the lower bound of $\alpha(A,x)$
    \item Let $\beta \in (0, 0.1)$
   
    \end{itemize}
Then, we have
\begin{itemize}
    \item {\bf Part 1.} $ | u(A, x)_{j_0} - u(\wt{A},x )_{j_0} | \leq R \cdot \exp(R^2) \cdot \| A_{j_0,*} - \wt{A}_{j_0,*} \|_2$
    \item {\bf Part 2.} $ \| u(A, x) - u(\wt{A},x ) \|_2^2 \leq R \cdot \exp(R^2) \cdot \| A - \wt{A} \|_F^2$
    \item {\bf Part 3.} $ |\alpha(A,x)^{-1} - \alpha(\wt{A},x)^{-1}| \leq \beta^{-2} \cdot \sqrt{n}   \exp(2R^2)\|A - \wt{A}\|_F$
    \item {\bf Part 4.} $\|f(A,x) - f(\wt{A},x)\|_2 \leq  2\beta^{-2} \cdot n \cdot \exp(3R^2) \cdot \|A - \wt{A}\|_F $
    \item {\bf Part 5.} $\| c(A,x) - c(\wt{A},x) \|_2 \leq 2\beta^{-2} \cdot n \cdot \exp(3R^2) \cdot \|A - \wt{A}\|_F$
    \item {\bf Part 6.} $ \| g(A,x) - g(\wt{A},x) \|_2 \leq 20\beta^{-2} \cdot n \cdot R \cdot  \exp(3R^{2}) \cdot \|A - \wt{A}\|_F $
    \item {\bf Part 7.} $\| q(A,x) - q(\wt{A},x) \|_2 \leq 4 \beta^{-2} \cdot n \cdot \exp(3R^2) \cdot \|A - \wt{A} \|_F$
    \item {\bf Part 8.} $\| c_g(A,x) - c_g(\wt{A},x) \|_2 \leq 20\beta^{-2} \cdot n \cdot R \cdot  \exp(3R^{2}) \cdot \|A - \wt{A}\|_F$
    \item {\bf Part 9.} $| L_g(A,x) - L_g(\wt{A},x) | \leq 100 \beta^{-2} \cdot n \cdot  \exp(5R^{2}) \cdot \|A - \wt{A}\|_F$
    \item {\bf Part 10.}  For $j_0 \in [n]$, $ |f(A,x)_{j_0} - f(\wt{A},x)_{j_0} | \leq 2  \beta^{-2} \cdot n   \exp(3R^2)\|A - \wt{A}\|_F $
\end{itemize}
\end{lemma}
\begin{proof}
{\bf Proof of Part 1.}
For $j_0 \in [n]$, we have
\begin{align*}
    | u(A, x)_{j_0} - u(\wt{A},x )_{j_0} | = & ~ | \exp( A_{j_0,*} x) - \exp( \wt{A}_{j_0,*} x) | \\ 
\leq & ~ \exp(R^2 ) 
\| x\|_2 \cdot  \| A_{j_0,*} - \wt{A}_{j_0,*} \|_2 \notag \\
\leq & ~  R\exp(R^2 ) 
 \cdot \|  A_{j_0,*} - \wt{A}_{j_0,*} \|_2 
\end{align*}
where the first step follows from definition of $u(A,x)_{j_0}$, the second step follows from Fact~\ref{fac:vector_norm}, and the last step follows from $\|x\|_2 \leq R$.

{\bf Proof of Part 2.}
We have 
\begin{align*}
\| u(A, x) - u(\wt{A},x )\|_2^2 
= & ~ \sum_{j_0=1}^n | u(A, x)_{j_0} - u(\wt{A},x )_{j_0} |^2 \\
\leq & ~ \sum_{j_0=1}^n ( R \exp(R^2) )^2 \|  A_{j_0,*} - \wt{A}_{j_0,*} \|_2^2 \\
= & ~ (R \exp(R^2))^2 \| A - \wt{A} \|_F^2
\end{align*}
where the first step follows from the the definition of $\ell_2- norm$, the second step follows from {\bf Part 1}, and the last step follows from the definition of Frobenius norm.
Taking the square root of both sides we get
\begin{align*}
\| u(A, x) - u(\wt{A},x )\|_2 \leq  R \exp(R^2) \| A - \wt{A} \|_F
\end{align*}

{\bf Proof of Part 3.}
We have
\begin{align*}
    |\alpha(A,x)^{-1} - \alpha(\wt{A},x)^{-1} | = & ~ \alpha(A,x)^{-1} \alpha(\wt{A},x)^{-1} |\alpha(A,x)^{-1} - \alpha(\wt{A},x)^{-1} | 
    \\
    \leq & ~ \beta^{-2} \cdot |\alpha(A,x) - \alpha(\wt{A},x)| \\
    \leq & ~ \beta^{-2} \cdot |\langle u(A,x) , {\bf 1}_n \rangle - \langle u(\wt{A},x), {\bf 1}_n \rangle  | \\
    \leq & ~ \beta^{-2} \cdot \sqrt{n} \| u(A,x) -u(\wt{A},x) \|_2 \\
     \leq & ~   \beta^{-2} \cdot \sqrt{n}  R \exp(R^2)\|A - \wt{A}\|_F \\
     \leq & ~  \beta^{-2} \cdot \sqrt{n}   \exp(2R^2)\|A - \wt{A}\|_F ,
\end{align*}
where the first step is due to simple algebra, the second step is due to the assumption that $\beta \geq \alpha(A,x)$, the third step follows from Definition of $\alpha(A,x)$ (see Definition~\ref{def:alpha}), the fourth step is based on Fact~\ref{fac:circ_rules} and Fact~\ref{fac:vector_norm}, the fifth step is because of {\bf Part 2}, and the sixth step follows from $R \geq 4$ and $\exp(R^2) \geq R$.

{\bf Proof of Part 4.}
\begin{align*}
    \|f(A,x) - f(\wt{A},x) \|_2 = & ~ 
    \|\alpha(A,x)^{-1} u(A,x) -  \alpha(\wt{A},x)^{-1} u(\wt{A},x) \|_2 \\
    \leq & ~ \|\alpha(A,x)^{-1} u(A,x) - \alpha(\wt{A},x)^{-1} u(A,x)\|_2 \\
    + & ~ \|\alpha(\wt{A},x)^{-1} u(A,x) - \alpha(\wt{A},x)^{-1} u(\wt{A},x) \|_2 
\end{align*}
where the first step follows from Definition of $f(A,x)$ (see Definition~\ref{def:f}), the second step follows from triangle inequality.

For the first term above, 
we have \begin{align*}
    \|\alpha(A,x)^{-1} u(A,x) - \alpha(\wt{A},x)^{-1} u(A,x) \|_{2}
    \leq  & ~ \|u(A,x) \|_2 |\alpha(A,x)^{-1} - \alpha(\wt{A},x)^{-1}| \\ 
    \leq & ~ \sqrt{n} \exp(R^2) \cdot \beta^{-2} \cdot \sqrt{n}   \exp(2R^2)\|A - \wt{A}\|_F \\
    \leq & ~\beta^{-2} \cdot n \cdot \exp(3R^2)\|A - \wt{A}\|_F
\end{align*}
where the second step follows from {\bf Part 1} of Lemma~\ref{lem:upper_bound} and {\bf Part 3}, the last step follows from simple algebra.

For the second term above,
we have
 \begin{align*}
     \|\alpha(\wt{A},x)^{-1} u(A,x) - \alpha(\wt{A},x)^{-1} u(\wt{A},x) \|_2 \leq & ~ |\alpha(\wt{A},x)^{-1}|\| u(A,x)  -u(\wt{A},x)  \|_2 \\
     \leq & ~ \beta^{-1} \cdot R \exp(R^2) \cdot \|A - \wt{A} \|_F\\
     \leq & ~ \beta^{-1} \cdot  \exp(2R^2) \cdot \|A - \wt{A} \|_F
     \end{align*}
where the second step follows from {\bf Part 2} of Lemma~\ref{lem:upper_bound} and $\alpha(A,x) \leq \beta$, the last step  follows from $R \geq 4$ and $\exp(R^2) \geq R$.
     
Then, by combining above two equations, we can get 
\begin{align*}
    \|f(A,x) - f(\wt{A},x) \|_2 \leq & ~ \beta^{-2} n \exp(3R^2)\|A - \wt{A}\|_F \\
    + & ~ \beta^{-1} \cdot  \exp(2R^2) \cdot \|A - \wt{A} \|_F \\
    \leq & ~    2\beta^{-2} \cdot n \cdot \exp(3R^2) \cdot \|A - \wt{A}\|_F
\end{align*}
where the second step follows from $\beta \in (0,1)$, $R \geq 4$, and $\exp(R^2) \geq R$. 

{\bf Proof of Part 5.}
\begin{align*}
    \|c(A,x) - c(\wt{A},x)  \|_2 
    \leq & ~ \|(f(A,x) - b_f) -( f(\wt{A},x) - b_f) \|_2 \\
    \leq & ~ \|f(A,x) -f(\wt{A},x) \|_2  \\
    \leq & ~ 2\beta^{-2} \cdot n \cdot \exp(3R^2) \cdot \|A - \wt{A}\|_F
\end{align*}
where the first step follows from Definition of $c(A,x)$ (see Definition~\ref{def:c}), the second step follows from simple algebra, the last step follows from {\bf Part 4}.

{\bf Proof of Part 6.}
We define \begin{align*}
    g_1(A,x) = & ~ - A^\top f(A,x)   \langle c(A, x), f(A, x) \rangle \\
    g_2(A,x) = & ~ A^\top \diag(f(A, x))  c(A,x) \\
    g(A,x) = & ~ g_1(A,x) + g_2(A,x)
\end{align*}

For $g_1(A,x)$, we have
 \begin{align*}
     \| g_1(A,x) -g_1(\wt{A},x) \|_2 = & ~ \|- A^\top f(A,x)   \langle c(A, x), f(A, x) \rangle + \wt{A}^\top f(\wt{A},x)   \langle c(\wt{A}, x), f(\wt{A}, x) \rangle \|_2 \\
     = & ~ \|A^\top f(A,x)   \langle c(A, x), f(A, x) \rangle - \wt{A}^\top f( \wt{A},x)   \langle c( \wt{A}, x), f(\wt{A}, x) \rangle \|_2 \\
     \leq & ~  \|A^\top f(A,x)   \langle c(A, x), f(A, x) \rangle - \wt{A}^\top f(A,x)   \langle c(A, x), f(A, x) \rangle\|_2 \\
     & + ~\|\wt{A}^\top f(A,x)   \langle c(A, x), f(A, x) \rangle - \wt{A}^\top f(\wt{A},x)   \langle c(A, x), f(A, x) \rangle \|_2 \\
     & + ~ \|\wt{A}^\top f(\wt{A},x)   \langle c(A, x), f(A, x) \rangle - \wt{A}^\top f(\wt{A},x)   \langle c(\wt{A}, x), f(A, x) \rangle\|_2 \\
     & + ~ \|\wt{A}^\top f(\wt{A},x)   \langle c(\wt{A}, x), f(A, x) \rangle - \wt{A}^\top f(\wt{A},x)   \langle c(\wt{A}, x), f(\wt{A}, x) \rangle\|_2
 \end{align*}
 where the second step follows from Fact~\ref{fac:vector_norm}, the third step follows from Lemma~\ref{lem:product_rule}.

For the first term, we have
\begin{align*}
    & ~ \|A^\top f(A,x)   \langle c(A, x), f(A, x) \rangle - \wt{A}^\top f(A,x)   \langle c(A, x), f(A, x) \rangle\|_2 \\
   \leq & ~ \|A^{\top} - \wt{A}^{\top} \|_F \cdot \|f(A,x) \|_2 \cdot \|c(A,x) \|_2 \cdot \| f(A,x)\|_2 \\
   \leq & ~ 2 \|A^{\top} - \wt{A}^{\top} \|_F
\end{align*}
where the first step follows from Fact~\ref{fac:vector_norm} and Fact~\ref{fac:matrix_norm}, the second step follows from {\bf Part 3} and {\bf Part 4} of Lemma~\ref{lem:upper_bound}.

For the second term, we have
\begin{align*}
    & ~ \|\wt{A}^\top f(A,x)   \langle c(A, x), f(A, x) \rangle - \wt{A}^\top f(\wt{A},x)   \langle c(A, x), f(A, x) \rangle \|_2 \\
    \leq & ~ \|f(A,x)  - f(\wt{A},x) \|_2 \cdot \|\wt{A}^\top \|_F \cdot \|c(A,x) \|_2 \cdot \|f(A, x) \|_2 \\
    \leq & ~ 4 \beta^{-2} \cdot n \cdot R \cdot  \exp(3R^{2}) \cdot \|A - \wt{A}\|_F 
\end{align*} 
 where the first step follows from Fact~\ref{fac:vector_norm} and Fact~\ref{fac:matrix_norm}, the second step follows from {\bf Part 3} and {\bf Part 4} of Lemma~\ref{lem:upper_bound} and {\bf Part 4}.

For the third  term, we have
\begin{align*}
    & ~ \|\wt{A}^\top f(\wt{A},x)   \langle c(A, x), f(A, x) \rangle - \wt{A}^\top f(\wt{A},x)   \langle c(\wt{A}, x), f(A, x) \rangle\|_2 \\
    \leq & ~ \|\wt{A}^\top \|_F \cdot \|f(\wt{A},x) \|_2 \cdot \|c(A,x) - c(\wt{A}, x) \|_2 \cdot \|f(A,x) \|_2\\
    \leq & ~   2\beta^{-2} \cdot n \cdot R \cdot \exp(3R^2) \cdot \|A - \wt{A}\|_F
\end{align*}
 where the first step follows from Fact~\ref{fac:vector_norm} and Fact~\ref{fac:matrix_norm}, the second step follows from {\bf Part 3} and {\bf Part 4} of Lemma~\ref{lem:upper_bound} and {\bf Part 5}.

For the fourth term, we have
\begin{align*}
    & ~ \|\wt{A}^\top f(\wt{A},x)   \langle c(\wt{A}, x), f(A, x) \rangle - \wt{A}^\top f(\wt{A},x)   \langle c(\wt{A}, x), f(\wt{A}, x) \rangle\|_2 \\
    \leq & ~ \| \wt{A}^\top\|_F \cdot \| f(\wt{A},x)\|_2 \cdot \|c(\wt{A}, x)\|_2 \cdot \|f(A,x)  - f(\wt{A},x) \|_2 \\
    \leq & ~ 4 \beta^{-2} \cdot n \cdot R \cdot  \exp(3R^{2}) \cdot \|A - \wt{A}\|_F 
\end{align*}
 where the first step follows from Fact~\ref{fac:vector_norm} and Fact~\ref{fac:matrix_norm}, the second step follows from {\bf Part 3} and {\bf Part 4} of Lemma~\ref{lem:upper_bound} and {\bf Part 4}.

Then, we have 
\begin{align*}
    \| g_1(A,x) -g_1(\wt{A},x) \|_2 \leq & ~ 2 \|A^{\top} - \wt{A}^{\top} \|_F\\
    & + ~  4 \beta^{-2} \cdot n \cdot R \cdot  \exp(3R^{2}) \cdot \|A - \wt{A}\|_F \\
    & + ~   2\beta^{-2} \cdot n \cdot R \cdot \exp(3R^2) \cdot \|A - \wt{A}\|_F\\
    & + ~   4 \beta^{-2} \cdot n \cdot R \cdot  \exp(3R^{2}) \cdot \|A - \wt{A}\|_F \\
    \leq & ~ 12\beta^{-2} \cdot n \cdot R \cdot  \exp(3R^{2}) \cdot \|A - \wt{A}\|_F  
\end{align*}
where the second step follows from $\beta \in (0, 0.1)$ and $R \geq 4$.

For $g_2(A,x)$, we have 
\begin{align*}
   \| g_2(A,x) - g_2(\wt{A},x)\|= & ~ \|A^\top \diag(f(A, x))  c(A,x) - \wt{A}^\top \diag(f(\wt{A}, x))  c(\wt{A},x)\| \\
   \leq & ~ \|A^\top \diag(f(A, x))  c(A,x) - \wt{A}^\top \diag(f(A, x))  c(A,x)\| \\
    & ~ + \|\wt{A}^\top \diag(f(A, x))  c(A,x) -\wt{A}^\top \diag(f(\wt{A}, x))  c(A,x)\| \\
    & ~ + \|\wt{A}^\top \diag(f(\wt{A}, x))  c(A,x) - \wt{A}^\top \diag(f(\wt{A}, x))  c(\wt{A},x)\|
\end{align*}
where the second step follows from Lemma~\ref{lem:product_rule}.

For the first term, we have
\begin{align*}
    & ~ \|A^\top \diag(f(A, x))  c(A,x) - \wt{A}^\top \diag(f(A, x))  c(A,x)\|_2 \\
   \leq & ~ \|A^{\top} - \wt{A}^{\top} \|_F \cdot \|\diag(f(A,x)) \|_F \cdot \|c(A,x) \|_2  \\
   \leq & ~ 2  \|A^{\top} - \wt{A}^{\top} \|_F
\end{align*}
where the first step follows from Fact~\ref{fac:vector_norm} and Fact~\ref{fac:matrix_norm}, the second step follows from $\|\diag(f(A,x)) \|_F \leq 1$ and {\bf Part 4} of Lemma~\ref{lem:upper_bound}.

For the second term, we have
\begin{align*}
    & ~ \|\wt{A}^\top \diag(f(A, x))  c(A,x) - \wt{A}^\top \diag(f(\wt{A}, x))  c(A,x)\|_2 \\
   \leq & ~ \|\wt{A}^{\top} \|_F \cdot \|\diag(f(A,x))-\diag(f(\wt{A}, x)) \|_F \cdot \|c(A,x) \|_2  \\
   \leq & ~ 4 \cdot \beta ^{-2} \cdot n \cdot R \cdot \exp(3R^2) \cdot \|A - \wt{A}\|_F
\end{align*}
where the first step follows from Fact~\ref{fac:vector_norm} and Fact~\ref{fac:matrix_norm}, the second step follows from {\bf Part 4}, $\|\diag(f(A,x)) -\diag(f(\wt{A},x))  \|_F \leq \|f(A,x) -f(\wt{A},x)  \|_2$, and {\bf Part 4} of Lemma~\ref{lem:upper_bound}.

For the third  term, we have
\begin{align*}
    & ~ \|\wt{A}^\top \diag(f(\wt{A}, x))  c(A,x) - \wt{A}^\top \diag(f(\wt{A}, x))  c(\wt{A},x)\|_2 \\
   \leq & ~ \|\wt{A}^{\top} \|_F \cdot \|\diag(f(\wt{A}, x)) \|_F \cdot \|c(A,x)-c(\wt{A},x) \|_2  \\
   \leq & ~ 2\cdot \beta ^{-2} \cdot n \cdot R \cdot \exp(3R^2) \cdot \|A - \wt{A}\|_F
\end{align*}
where the first step follows from Fact~\ref{fac:vector_norm} and Fact~\ref{fac:matrix_norm}, the second step follows from {\bf Part 5}, $\|\diag(f(A,x)) \|_F \leq 1$, and $\|A^{\top} \|_F \leq R$.

Then, we have
\begin{align*}
    \| g_2(A,x) - g_2(\wt{A},x)\|_2
    \leq & ~ 2  \|A^{\top} - \wt{A}^{\top} \|_F \\
    & + ~ 4 \cdot \beta ^{-2} \cdot n \cdot R \cdot \exp(3R^2) \cdot \|A - \wt{A}\|_F \\
    & + ~ 2\cdot \beta ^{-2} \cdot n \cdot R \cdot \exp(3R^2) \cdot \|A - \wt{A}\|_F \\
    \leq & ~ 8\cdot \beta ^{-2} \cdot n \cdot R \cdot \exp(3R^2) \cdot \|A - \wt{A}\|_F
\end{align*}
where the second step follows from $\beta \in (0, 0.1)$ and $R \geq 4$.

Then, by combining above two equations, we can get
\begin{align*}
    \|g(A,x) - g(\wt{A},x) \|_2 
    \leq & ~ 12\beta^{-2} \cdot n \cdot R \cdot  \exp(3R^{2}) \cdot \|A - \wt{A}\|_F  \\
    & ~ + 8\cdot \beta ^{-2} \cdot n \cdot R \cdot \exp(3R^2) \cdot \|A - \wt{A}\|_F \\
    \leq & ~ 20\beta^{-2} \cdot n \cdot R \cdot  \exp(3R^{2}) \cdot \|A - \wt{A}\|_F 
\end{align*}

{\bf Proof of Part 7.}
\begin{align*}
      \| q(A,x) - q(\wt{A},x) \|_2 
    = & ~ \| c(A,x) + f(A,x) - c(\wt{A},x) - f(\wt{A},x) \|_2 \\
    \leq & ~ \|c(A,x) -  c(\wt{A},x)\|_2 + \|  f(A,x) - f(\wt{A},x)\|_2 \\
    \leq & ~ 4 \cdot \beta^{-2} \cdot n \cdot \exp(3R^2) \cdot \|A - \wt{A} \|_F
\end{align*}
where the first step follows from the definition of $q(A,x)$, the second step follows from Fact~\ref{fac:vector_norm}, and the last step follows from {\bf Part 4} and {\bf Part 5} of Lemma~\ref{lem:basic_lips}.

\iffalse
\begin{align*}
    \| q(A,x) - q(\wt{A},x) \|_2 
    = & ~ \| c(A,x) + f(A,x) + c(\wt{A},x) + f(\wt{A},x) \|_2 \\
    = & ~ \| c(A,x)\|_2 + \|f(A,x)\|_2 + \|c(\wt{A},x)\|_2 + \|f(\wt{A},x) \|_2 \\
    \leq & ~ 4
\end{align*}
where the first step follows from the definition of $q(A,x)$, the second step follows from Fact~\ref{fac:vector_norm}, and the last step follows from {\bf Part 3} and {\bf Part 4} of Lemma~\ref{lem:upper_bound}.
\fi
{\bf Proof of Part 8.}
\begin{align*}
    \| c_g(A,x) - c_g(\wt{A},x) \|_2
    = & ~ \|(g(A,x) - b_g) - (g(\wt{A},x) - b_g) \|_2 \\
    = & ~ \|g(A,x) - g(\wt{A},x) \|_2 \\
    \leq & ~ 20\beta^{-2} \cdot n \cdot R \cdot  \exp(3R^{2}) \cdot \|A - \wt{A}\|_F
\end{align*}
where the first step follows from Definition of $c_g(A,x)$ (see Definition~\ref{def:c_g}), the second step follows from simple algebra, and the last step follows from {\bf Part 6}.

{\bf Proof of Part 9.}
We can show
\begin{align*}
    & ~ | L_g(A,x) - L_g(\wt{A},x) | \\
    = & ~ |0.5\|c_g(A,x)\|_2^2 - 0.5\|c_g(\wt{A},x) \|_2^2 | \\
    = & ~ 0.5|\langle c_g(A,x) ,c_g(A,x) \rangle - \langle c_g(\wt{A},x) ,c_g(\wt{A},x) \rangle| \\
    \leq & ~ 0.5|\langle c_g(A,x) ,c_g(A,x) \rangle - \langle c_g(A,x) ,c_g(\wt{A},x) \rangle +  \langle c_g(A,x) ,c_g(\wt{A},x) \rangle -\langle c_g(\wt{A},x) ,c_g(\wt{A},x) \rangle|\\
    \leq & ~ 0.5 \|c_g(A,x) - c_g(\wt{A},x) \|_2 \cdot \|c_g(A,x)\|_2 + 0.5\|c_g(A,x) - c_g(\wt{A},x) \|_2 \cdot \|c_g(\wt{A},x)\|_2 \\
    \leq & ~ 100 \cdot \beta^{-2} \cdot n \cdot  \exp(5R^{2}) \cdot \|A - \wt{A}\|_F 
\end{align*}
where the first step follows from the definition of $L_g(A,x)$ (see Definition~\ref{def:l_g}), the second step follows from Fact~\ref{fac:vector_norm}, the third step follows from Fact~\ref{fac:vector_norm}, the forth step follows from Fact~\ref{fac:vector_norm}, the fifth step follows from {\bf Part 8} and {\bf Part 6} of Lemma~\ref{lem:upper_bound}.

{\bf Proof of Part 10.}
\begin{align*}
 & ~ | f(A,x)_{j_0} - f(\wt{A},x)_{j_0} | \\
 =  & ~ |\alpha(A,x )^{-1} u(A,x)_{j_0} - \alpha( \wt{A},x )^{-1} u(\wt{A},x)_{j_0} | \\
 \leq & ~ | \alpha(A,x )^{-1} u(A,x)_{j_0} - \alpha( \wt{A},x )^{-1} u(A,x)_{j_0} | + |\alpha(
 \wt{A},x )^{-1} u(\wt{A},x)_{j_0} - \alpha( \wt{A},x )^{-1} u(\wt{A},x)_{j_0} | \\
 \leq & ~| \alpha(A,x )^{-1} - \alpha( \wt{A},x )^{-1}  | |u(A,x)_{j_0} | + |\alpha(\wt{A},x)^{-1} | |u(A,x)_{j_0}-  u(\wt{A},x)_{j_0}| \\
 \leq & ~ \beta^{-2} \cdot n   \exp(3R^2)\|A - \wt{A}\|_F  + \beta^{-1} R \cdot \exp(R^2) \cdot \| A_{j_0,*} - \wt{A}_{j_0,*} \|_2 \\
 \leq & ~ 2  \beta^{-2} \cdot n   \exp(3R^2)\|A - \wt{A}\|_F
\end{align*}
where the  first step follows from the definition of $f(A,x)_{j_0}$ (see Definition~\ref{def:f}), the second step follows from Lemma~\ref{lem:product_rule}, the third step follows from Fact~\ref{fac:vector_norm},   the fourth step follows from {\bf Part 1,3} of Lemma~\ref{lem:basic_lips}, {\bf Part 8} of Lemma~\ref{lem:upper_bound} and $\alpha(A,x) \geq \beta$, and the last step follows from $\beta^{-1} > 1, n > 1, R > 4$ and $\|A - \wt{A} \|_F \geq \| A_{j_0,*} - \wt{A}_{j_0,*}\|_2$.
\end{proof}

\subsection{Lipschitz: Hessian Related}

\begin{lemma}
If the following conditions holds
\begin{itemize}
    \item Let $f_c(A,x) = \langle f(A,x) , c(A,x) \rangle \in \R$
    \item Let $f_2(A,x) = \langle f(A,x), f(A,x) \rangle \in \R$
    \item Let $h(A,x) = \diag(x) A^\top f(A,x) \in \R^d$
    \item Let $p_{j_1} (A,x) := \diag(x) A^\top \cdot ( (e_{j_1} - f(A,x)) \circ q(A,x) ) \in \R^d$ for each $j_1 \in ]n]$
\end{itemize}
Then, we have
\begin{itemize}
    \item {\bf Part 1.}
    \begin{align*}
        |  f_c(A,x) - f_c(\wt{A},x) | \leq  6\beta^{-2} \cdot n \cdot \exp(3R^2) \cdot \|A - \wt{A}\|_F
    \end{align*}
    \item {\bf Part 2.}
    \begin{align*}
        | f_2(A,x) - f_2(\wt{A},x) | \leq 4\beta^{-2} \cdot n \cdot \exp(3R^2) \cdot \|A - \wt{A}\|_F
    \end{align*}
    \item {\bf Part 3.}
    \begin{align*}
        \| h(A,x) - h(\wt{A},x) \|_2 \leq 3\beta^{-2} \cdot n \cdot \exp(4R^2) \cdot \|A - \wt{A}\|_F 
    \end{align*}
    \item {\bf Part 4.}  
    \begin{align*}
        \| h_e(A,x) - h_e(\wt{A},x) \|_2 \leq  3\beta^{-2} \cdot n \cdot \exp(4R^2) \cdot \|A - \wt{A}\|_F 
    \end{align*}
      \item {\bf Part 5.} For each $j_1 \in [n]$
    \begin{align*}
        \| p_{j_1}(A,x) - p_{j_1}(\wt{A},x) \|_2 \leq 13 \beta^{-2} \cdot n \cdot \exp(4R^2) \cdot \|A - \wt{A} \|_F
    \end{align*}
\end{itemize}
\end{lemma}

\begin{proof}

{\bf Proof of Part 1.}
\begin{align*}
     |  f_c(A,x) - f_c(\wt{A},x) | = & ~ \| \langle f(A,x), c(A,x) 
    \rangle  - \langle f(\wt{A},x), c(\wt{A},x) 
    \rangle\| \\
    \leq & ~| \langle f(A,x), c(A,x) 
    \rangle  - \langle f(\wt{A},x), c(A,x) 
    \rangle| + | \langle f(\wt{A},x), c(A,x) 
    \rangle  - \langle f(\wt{A},x), c(\wt{A},x) 
    \rangle| \\
    \leq & ~ \|f(A,x)^{\top} - f(\wt{A},x)^{\top} \|_2 \|c(A,x) \|_2 + \|f(\wt{A},x) \|_2 \| c(A,x)  - c(\wt{A},x) \|_2\\
    \leq  & ~ 6\beta^{-2} \cdot n \cdot \exp(3R^2) \cdot \|A - \wt{A}\|_F
\end{align*}
where the first step follows from the definition of $f_c(A,x)$, the second step follows from Lemma~\ref{lem:product_rule}, the third step follows from Fact~\ref{fac:vector_norm}, and the last step follows from {\bf Part 4, 5} of Lemma~\ref{lem:basic_lips}, and {\bf Part 
3, 4} of Lemma~\ref{lem:upper_bound}.

{\bf Proof of Part 2.}

\begin{align*}
     |  f_2(A,x) - f_2(\wt{A},x) | = & ~ \| \langle f(A,x), f(A,x) 
    \rangle  - \langle f(\wt{A},x), f(\wt{A},x) 
    \rangle\| \\
    \leq & ~| \langle f(A,x), f(A,x) 
    \rangle  - \langle f(\wt{A},x), f(A,x) 
    \rangle| + | \langle f(\wt{A},x), f(A,x) 
    \rangle  - \langle f(\wt{A},x), f(\wt{A},x) 
    \rangle| \\
    \leq & ~ \|f(A,x)^{\top} - f(\wt{A},x)^{\top} \|_2 \|f(A,x) \|_2 + \|f(\wt{A},x) \|_2 \| f(A,x)  - f(\wt{A},x) \|_2\\
    \leq  & ~ 4\beta^{-2} \cdot n \cdot \exp(3R^2) \cdot \|A - \wt{A}\|_F
\end{align*}
where the first step follows from the definition of $f_c(A,x)$, the second step follows from Lemma~\ref{lem:product_rule}, the third step follows from Fact~\ref{fac:vector_norm}, and the last step follows from {\bf Part 4, 5} of Lemma~\ref{lem:basic_lips}, and {\bf Part 
3, 4} of Lemma~\ref{lem:upper_bound}.

{\bf Proof of Part 3.}
\begin{align*}
    & ~\| \diag(x) A^\top f(A,x) - \diag(x) \wt{A}^\top f(\wt{A},x) \|_2
    \\
    \leq & ~ \| \diag(x) \|_F \|  A^\top f(A,x) - \wt{A}^\top f(\wt{A},x)  \|_2 \\
   \leq & ~ R \|A^\top f(A,x) - \wt{A}^\top f(A,x)  \| + R\|\wt{A}^\top f(A,x) - \wt{A}^\top f(\wt{A},x)  \|_2 \\
   \leq & ~ R \|A^{\top}  -  \wt{A}^\top\|_F \| f(A,x)\|_2 + R \|\wt{A}^\top\|_F \|f(A,x) - f(\wt{A},x)\|_F \\
   \leq & ~ R\|A  -  \wt{A}\|_F  + R^2  2\beta^{-2} \cdot n \cdot \exp(3R^2) \cdot \|A - \wt{A}\|_F \\
   \leq & ~ 3\beta^{-2} \cdot n \cdot \exp(4R^2) \cdot \|A - \wt{A}\|_F 
\end{align*}

{\bf Proof of Part 4.}
\begin{align*}
    & ~\| \diag (x) \cdot A^{\top}     \cdot (e_{j_1}- f(A,x) ) - \diag(x) \wt{A}^\top (e_{j_1}-f(\wt{A},x)) \|_2
    \\
    \leq & ~ \| \diag(x) \|_F \|  A^\top (e_{j_1}-f( A  ,x)) - \wt{A}^\top (e_{j_1}-f(\wt{A},x))  \|_2 \\
   \leq & ~ R \|A^\top (e_{j_1}-f( A  ,x)) - \wt{A}^\top (e_{j_1}-f( A  ,x))  \| + R\|\wt{A}^\top (e_{j_1}-f( A  ,x)) - \wt{A}^\top (e_{j_1}-f(\wt{A},x))   \|_2 \\
   \leq & ~ R \|A^{\top}  -  \wt{A}^\top\|_F \|  e_{j_1} \|_2 \| f(A,x)\|_2 + R \|\wt{A}^\top\|_F \|f(A,x) - f(\wt{A},x)\|_F \\
   \leq & ~ R\|A  -  \wt{A}\|_F  + R^2  2\beta^{-2} \cdot n \cdot \exp(3R^2) \cdot \|A - \wt{A}\|_F \\
   \leq & ~ 3\beta^{-2} \cdot n \cdot \exp(4R^2) \cdot \|A - \wt{A}\|_F 
\end{align*}

{\bf Proof of Part 5.}
\begin{align*}
   & ~ \| p_{j_1}(A,x) - p_{j_1}(\wt{A},x) \|_2 \\
    = & ~ \| \diag(x) A^\top \cdot ( (e_{j_1} - f(A,x)) \circ q(A,x) ) - \diag(x) \wt{A}^\top \cdot ( (e_{j_1} - f(\wt{A},x)) \circ q(\wt{A},x) )
\|_2 \\
\leq & ~ \|\diag(x) A^\top \cdot ( (e_{j_1} - f(A,x)) \circ q(A,x) ) - \diag(x) \wt{A}^\top \cdot ( (e_{j_1} - f(A,x)) \circ q(A,x) ) \|_2 \\
 & +  ~ \| \diag(x) \wt{A}^\top \cdot ( (e_{j_1} - f(A,x)) \circ q(A,x) )  - \diag(x) \wt{A}^\top \cdot ( (e_{j_1} - f(\wt{A},x)) \circ q(A,x) ) \|_2 \\
 & + ~ \|\diag(x) \wt{A}^\top \cdot ( (e_{j_1} - f(\wt{A},x)) \circ q(A,x) ) - \diag(x) \wt{A}^\top \cdot ( (e_{j_1} - f(\wt{A},x)) \circ q(\wt{A} ,x) ) \|_2 \\
  \leq &  ~ R \|A^\top  - \wt{A}^\top \|_F\|\cdot \|  e_{j_1} \|_2 \| f(A,x)\|_2 \|q(A,x)  \|_2 \\
 & + ~ R \|\wt{A}^\top  \|_2 \|f(A,x) - f(\wt{A},x) \|_2 \| q(A,x)\|_2 \\
 & + ~ R \|\wt{A}^\top  \|_2 \|  e_{j_1} \|_2 \| f(\wt{A},x) \|_2 \| q(A,x) - q(\wt{A} ,x) \|_2 \\
 \leq & ~ 3 R \|A^\top  - \wt{A}^\top \|_F \\
 & +  ~ 6 R^2  \beta^{-2} \cdot n \cdot \exp(3R^2) \cdot \|A - \wt{A}\|_F \\
 & + ~  4  R^2  \beta^{-2} \cdot n \cdot \exp(3R^2) \cdot \|A - \wt{A} \|_F
 \\
 \leq & ~ 13 \beta^{-2} \cdot n \cdot \exp(4R^2) \cdot \|A - \wt{A} \|_F
\end{align*}

\end{proof}

\newpage
\section{Hessian: Split Hessian into Seven Parts}\label{app:hessian_main}

\subsection{Definitions}
\begin{definition}
    We define the $B_1^{j_1,i_1,j_0,i_0}$ as follows,
    \begin{align*}
        B_1^{j_1,i_1,j_0,i_0} := & ~ \frac{\d}{\d A_{j_1,i_1}}(-  c_g(A,x)^{\top} \cdot f(A,x)_{j_0}  \cdot \langle c(A,x), f(A,x) \rangle \cdot e_{i_0} )
    \end{align*}
    Then, we define $B_{1,1}^{j_1,i_1,j_0,i_0}, \cdots, B_{1,3}^{j_1,i_1,j_0,i_0}$ as follow
    \begin{align*}
         B_{1,1}^{j_1,i_1,j_0,i_0} : = & ~ \frac{\d}{\d A_{j_1,i_1}} (- c_g(A,x)^{\top} ) \cdot  f(A,x)_{j_0} \cdot \langle c(A,x), f(A,x) \rangle \cdot e_{i_0}\\
 B_{1,2}^{j_1,i_1,j_0,i_0} : = & ~ - c_g(A,x)^{\top} \cdot \frac{\d}{\d A_{j_1,i_1}} ( f(A,x)_{j_0} )  \cdot \langle c(A,x), f(A,x) \rangle \cdot e_{i_0}\\
 B_{1,3}^{j_1,i_1,j_0,i_0} : = & ~  -  c_g(A,x)^{\top} \cdot f(A,x)_{j_0}  \cdot  \langle\frac{\d c(A,x)}{\d A_{j_1,i_1}}, f(A,x) \rangle \cdot e_{i_0}\\
  B_{1,4}^{j_1,i_1,j_0,i_0} : = & ~  -  c_g(A,x)^{\top} \cdot f(A,x)_{j_0}  \cdot  \langle c(A,x), \frac{\d f(A,x)}{\d A_{j_1,i_1}} \rangle \cdot e_{i_0}
    \end{align*}
    It is easy to show
    \begin{align*}
        B_1^{j_1,i_1,j_0,i_0} = B_{1,1}^{j_1,i_1,j_0,i_0} +  B_{1,2}^{j_1,i_1,j_0,i_0} + B_{1,3}^{j_1,i_1,j_0,i_0}  + B_{1,4}^{j_1,i_1,j_0,i_0} 
    \end{align*}
\end{definition} 

\begin{definition}
    We define the $B_2^{j_1,i_1,j_0,i_0}$ as follows,
    \begin{align*}
        B_2^{j_1,i_1,j_0,i_0} := & ~ \frac{\d}{\d A_{j_1,i_1}}(- c_g(A,x)^{\top} \cdot f(A,x)_{j_0} \cdot c(A,x)_{j_0} \cdot e_{i_0}   )
    \end{align*}
    Then, we define $B_{2,1}^{j_1,i_1,j_0,i_0}, \cdots, B_{2,3}^{j_1,i_1,j_0,i_0}$ as follow
    \begin{align*}
         B_{2,1}^{j_1,i_1,j_0,i_0} : = & ~ \frac{\d}{\d A_{j_1,i_1}} (- c_g(A,x)^{\top} ) \cdot  f(A,x)_{j_0}  \cdot c(A,x)_{j_0} \cdot e_{i_0}\\
 B_{2,2}^{j_1,i_1,j_0,i_0} : = & ~ - c_g(A,x)^{\top} \cdot \frac{\d}{\d A_{j_1,i_1}} ( f(A,x)_{j_0} )  \cdot c(A,x)_{j_0} \cdot e_{i_0}\\
 B_{2,3}^{j_1,i_1,j_0,i_0} : = & ~  -  c_g(A,x)^{\top} \cdot f(A,x)_{j_0}  \cdot \frac{\d }{\d A_{j_1,i_1}}( c(A,x)_{j_0}) \cdot e_{i_0}
    \end{align*}
    It is easy to show
    \begin{align*}
        B_2^{j_1,i_1,j_0,i_0} = B_{2,1}^{j_1,i_1,j_0,i_0} +  B_{2,2}^{j_1,i_1,j_0,i_0} + B_{2,3}^{j_1,i_1,j_0,i_0} 
    \end{align*}
\end{definition}

\begin{definition} 
    We define the $B_3^{j_1,i_1,j_0,i_0}$ as follows,
    \begin{align*}
        B_3^{j_1,i_1,j_0,i_0} :=  & ~ \frac{\d}{\d A_{j_1,i_1}} (- c_g(A,x)^{\top} \cdot f(A,x)_{j_0} \cdot \langle c(A,x), f(A,x) \rangle \cdot ( (A_{j_0,*})^\top \circ x ))
    \end{align*}
    Then, we define $B_{3,1}^{j_1,i_1,j_0,i_0}, \cdots, B_{3,4}^{j_1,i_1,j_0,i_0}$ as follow
    \begin{align*}
         B_{3,1}^{j_1,i_1,j_0,i_0} : = & ~ \frac{\d}{\d A_{j_1,i_1}} (- c_g(A,x)^{\top} ) \cdot  f(A,x)_{j_0} \cdot \langle c(A,x), f(A,x) \rangle \cdot ( (A_{j_0,*})^\top \circ x )\\
 B_{3,2}^{j_1,i_1,j_0,i_0} : = & ~ - c_g(A,x)^{\top} \cdot \frac{\d}{\d A_{j_1,i_1}} ( f(A,x)_{j_0} )  \cdot \langle c(A,x), f(A,x) \rangle \cdot ( (A_{j_0,*})^\top \circ x )\\
 B_{3,3}^{j_1,i_1,j_0,i_0} : = & ~  -  c_g(A,x)^{\top} \cdot f(A,x)_{j_0}  \cdot \langle\frac{\d  c(A,x)}{\d A_{j_1,i_1}}, f(A,x) \rangle \cdot ( (A_{j_0,*})^\top \circ x )\\
 B_{3,4}^{j_1,i_1,j_0,i_0} : = & ~  -  c_g(A,x)^{\top} \cdot f(A,x)_{j_0}  \cdot \langle c(A,x),\frac{\d  f(A,x) }{\d A_{j_1,i_1}}\rangle \cdot ( (A_{j_0,*})^\top \circ x )\\
 B_{3,5}^{j_1,i_1,j_0,i_0} : = & ~  -  c_g(A,x)^{\top} \cdot f(A,x)_{j_0}  \cdot \langle c(A,x), f(A,x) \rangle \cdot \frac{\d}{\d A_{j_1,i_1}} ( (A_{j_0,*})^\top \circ x)
    \end{align*}
    It is easy to show
    \begin{align*}
        B_3^{j_1,i_1,j_0,i_0} = B_{3,1}^{j_1,i_1,j_0,i_0} +  B_{3,2}^{j_1,i_1,j_0,i_0} + B_{3,3}^{j_1,i_1,j_0,i_0} + B_{3,4}^{j_1,i_1,j_0,i_0} +B_{3,5}^{j_1,i_1,j_0,i_0}
    \end{align*}
\end{definition}

\begin{definition} 
    We define the $B_4^{j_1,i_1,j_0,i_0}$ as follows,
    \begin{align*}
        B_4^{j_1,i_1,j_0,i_0} := & ~ \frac{\d}{\d A_{j_1,i_1}}  (c_g(A,x)^{\top} \cdot f(A,x)_{j_1} \cdot \diag (x) A^{\top} \cdot  f(A,x)  \cdot    \langle c(A, x), f(A, x) \rangle ) 
    \end{align*}
    Then, we define $B_{4,1}^{j_1,i_1,j_0,i_0}, \cdots, B_{4,4}^{j_1,i_1,j_0,i_0}$ as follow
    \begin{align*}
         B_{4,1}^{j_1,i_1,j_0,i_0} : = & ~ \frac{\d}{\d A_{j_1,i_1}} (- c_g(A,x)^{\top} ) \cdot  f(A,x)_{j_0} \cdot \diag (x) A^{\top} \cdot  f(A,x)  \cdot    \langle c(A, x), f(A, x) \rangle\\
 B_{4,2}^{j_1,i_1,j_0,i_0}: = & ~ - c_g(A,x)^{\top} \cdot \frac{\d}{\d A_{j_1,i_1}} ( f(A,x)_{j_0} )  \cdot \diag (x) A^{\top} \cdot  f(A,x)  \cdot    \langle c(A, x), f(A, x) \rangle\\
 B_{4,3}^{j_1,i_1,j_0,i_0}: = & ~ - c_g(A,x)^{\top} \cdot f(A,x)_{j_0} \cdot \diag (x) \cdot \frac{\d}{\d A_{j_1,i_1}} ( A^{\top})  \cdot  f(A,x)  \cdot    \langle c(A, x), f(A, x) \rangle\\
  B_{4,4}^{j_1,i_1,j_0,i_0}: = & ~ - c_g(A,x)^{\top} \cdot f(A,x)_{j_0} \cdot \diag (x) A^{\top} \cdot  \frac{\d}{\d A_{j_1,i_1}} ( f(A,x)  )  \cdot    \langle c(A, x), f(A, x) \rangle\\
   B_{4,5}^{j_1,i_1,j_0,i_0}: = & ~  -  c_g(A,x)^{\top} \cdot f(A,x)_{j_0}  \cdot \diag (x) A^{\top} \cdot  f(A,x)  \cdot  \langle  \frac{\d c(A,x)}{\d A_{j_1,i_1}} , f(A,x) \rangle \\
     B_{4,6}^{j_1,i_1,j_0,i_0}: = & ~  -  c_g(A,x)^{\top} \cdot f(A,x)_{j_0}  \cdot \diag (x) A^{\top} \cdot  f(A,x)  \cdot \langle c(A,x), \frac{\d  f(A,x) }{\d A_{j_1,i_1}}  \rangle
    \end{align*}
    It is easy to show
    \begin{align*}
        B_4^{j_1,i_1,j_0,i_0} = B_{4,1}^{j_1,i_1,j_0,i_0} +  B_{4,2}^{j_1,i_1,j_0,i_0} + B_{4,3}^{j_1,i_1,j_0,i_0} + B_{4,4}^{j_1,i_1,j_0,i_0} + B_{4,5}^{j_1,i_1,j_0,i_0} +B_{4,6}^{j_1,i_1,j_0,i_0}
    \end{align*}
\end{definition}

\begin{definition} 
    We define the $B_5^{j_1,i_1,j_0,i_0}$ as follows,
    \begin{align*}
        B_5^{j_1,i_1,j_0,i_0} := \frac{\d}{\d A_{j_1,i_1}}(c_g(A,x)^{\top} \cdot f(A,x)_{j_0} \cdot \diag (x) A^{\top} \cdot  f(A,x)  \cdot    (\langle -f(A,x), f(A,x) \rangle + f(A,x)_{j_0}))
    \end{align*}
    Then, we define $B_{5,1}^{j_1,i_1,j_0,i_0}, \cdots, B_{5,6}^{j_1,i_1,j_0,i_0}$ as follow
    \begin{align*}
         B_{5,1}^{j_1,i_1,j_0,i_0} := & ~ \frac{\d}{\d A_{j_1,i_1}} (- c_g(A,x)^{\top} )    \cdot f(A,x)_{j_0} \cdot \diag(x) \cdot A^{\top} \cdot f(A,x) \cdot (\langle -f(A,x), f(A,x) \rangle + f(A,x)_{j_0})\\
 B_{5,2}^{j_1,i_1,j_0,i_0} := &   ~ - c_g(A,x)^{\top} \cdot \frac{\d}{\d A_{j_1,i_1}} (f(A,x)_{j_0}) \cdot \diag(x) \cdot A^{\top} \cdot f(A,x) \cdot (\langle -f(A,x), f(A,x) \rangle + f(A,x)_{j_0})\\
 B_{5,3}^{j_1,i_1,j_0,i_0} := &   ~ - c_g(A,x)^{\top} \cdot f(A,x)_{j_0} \cdot \diag(x)  \cdot \frac{\d}{\d A_{j_1,i_1}} (A^{\top}) \cdot  f(A,x) \cdot (\langle -f(A,x), f(A,x) \rangle + f(A,x)_{j_0})\\
 B_{5,4}^{j_1,i_1,j_0,i_0} := &   ~ - c_g(A,x)^{\top} \cdot f(A,x)_{j_0} \cdot \diag(x) \cdot A^{\top} \cdot \frac{\d}{\d A_{j_1,i_1}}(f(A,x)) \cdot (\langle -f(A,x), f(A,x) \rangle + f(A,x)_{j_0})\\
  B_{5,5}^{j_1,i_1,j_0,i_0} : = & ~   c_g(A,x)^{\top} \cdot f(A,x)_{j_0} \cdot \diag(x) \cdot A^{\top} \cdot  f(A,x)  \cdot \langle \frac{\d  f(A,x)}{\d A_{j_1,i_1}},f(A,x) \rangle \\
  B_{5,6}^{j_1,i_1,j_0,i_0} : = & ~   c_g(A,x)^{\top} \cdot f(A,x)_{j_0} \cdot \diag(x) \cdot A^{\top} \cdot  f(A,x)  \cdot \langle f(A,x), \frac{\d f(A,x) }{\d A_{j_1,i_1}} \rangle\\
   B_{5,7}^{j_1,i_1,j_0,i_0} : = & ~  - c_g(A,x)^{\top} \cdot f(A,x)_{j_0} \cdot \diag(x) \cdot A^{\top} \cdot  f(A,x)  \cdot  \frac{\d f(A,x)_{j_0}}{\d A_{j_1,i_1}} 
    \end{align*}
    It is easy to show
    \begin{align*}
        B_5^{j_1,i_1,j_0,i_0} = B_{5,1}^{j_1,i_1,j_0,i_0} +  B_{5,2}^{j_1,i_1,j_0,i_0} + B_{5,3}^{j_1,i_1,j_0,i_0} + B_{5,4}^{j_1,i_1,j_0,i_0} + B_{5,5}^{j_1,i_1,j_0,i_0} + B_{5,6}^{j_1,i_1,j_0,i_0} + B_{5,7}^{j_1,i_1,j_0,i_0}
    \end{align*}
\end{definition}

\begin{definition} 
    We define the $B_6^{j_1,i_1,j_0,i_0}$ as follows,
    \begin{align*}
        B_6^{j_1,i_1,j_0,i_0} & ~ \frac{\d}{\d A_{j_1,i_1}} (- c_g(A,x)^{\top} \cdot  f(A,x)_{j_0}\diag(x) A^{\top} f(A,x) \cdot(\langle -f(A,x), c(A,x) \rangle + f(A,x)_{j_0})) 
    \end{align*}
    Then, we define $B_{6,1}^{j_1,i_1,j_0,i_0}, \cdots, B_{6,7}^{j_1,i_1,j_0,i_0}$ as follow
    \begin{align*}
         B_{6,1}^{j_1,i_1,j_0,i_0}  := & ~ \frac{\d}{\d A_{j_1,i_1}} (- c_g(A,x)^{\top} )    \cdot f(A,x)_{j_0} \cdot \diag(x) \cdot A^{\top} \cdot f(A,x) \cdot (\langle -f(A,x), f(A,x) \rangle + f(A,x)_{j_0})\\
B_{6,2}^{j_1,i_1,j_0,i_0} := &   ~ - c_g(A,x)^{\top} \cdot \frac{\d}{\d A_{j_1,i_1}} (f(A,x)_{j_0}) \cdot \diag(x) \cdot A^{\top} \cdot f(A,x) \cdot (\langle -f(A,x), f(A,x) \rangle + f(A,x)_{j_0})\\
B_{6,3}^{j_1,i_1,j_0,i_0} := &   ~ - c_g(A,x)^{\top} \cdot f(A,x)_{j_0} \cdot \diag(x)  \cdot \frac{\d}{\d A_{j_1,i_1}} (A^{\top}) \cdot  f(A,x) \cdot (\langle -f(A,x), f(A,x) \rangle + f(A,x)_{j_0})\\
B_{6,4}^{j_1,i_1,j_0,i_0}:= &   ~ - c_g(A,x)^{\top} \cdot f(A,x)_{j_0} \cdot \diag(x) \cdot A^{\top} \cdot \frac{\d}{\d A_{j_1,i_1}}(f(A,x)) \cdot (\langle -f(A,x), f(A,x) \rangle + f(A,x)_{j_0})\\
B_{6,5}^{j_1,i_1,j_0,i_0}: = & ~   c_g(A,x)^{\top} \cdot f(A,x)_{j_0} \cdot \diag(x) \cdot A^{\top} \cdot  f(A,x)  \cdot \langle \frac{\d  f(A,x)}{\d A_{j_1,i_1}},f(A,x) \rangle \\
B_{6,6}^{j_1,i_1,j_0,i_0}: = & ~   c_g(A,x)^{\top} \cdot f(A,x)_{j_0} \cdot \diag(x) \cdot A^{\top} \cdot  f(A,x)  \cdot \langle f(A,x), \frac{\d f(A,x) }{\d A_{j_1,i_1}}\rangle \\
B_{6,7}^{j_1,i_1,j_0,i_0}: = & ~  - c_g(A,x)^{\top} \cdot f(A,x)_{j_0} \cdot \diag(x) \cdot A^{\top} \cdot  f(A,x)  \cdot  \frac{\d f(A,x)_{j_0}}{\d A_{j_1,i_1}} 
    \end{align*}
    It is easy to show
    \begin{align*}
        B_6^{j_1,i_1,j_0,i_0} = B_{6,1}^{j_1,i_1,j_0,i_0} +  B_{6,2}^{j_1,i_1,j_0,i_0} + B_{6,3}^{j_1,i_1,j_0,i_0} + B_{6,4}^{j_1,i_1,j_0,i_0} + B_{6,5}^{j_1,i_1,j_0,i_0} + B_{6,6}^{j_1,i_1,j_0,i_0} + B_{6,7}^{j_1,i_1,j_0,i_0}
    \end{align*}
\end{definition}

\begin{definition} 
    We define the $B_7^{j_1,i_1,j_0,i_0}$ as follows,
    \begin{align*}
        B_7^{j_1,i_1,j_0,i_0} : = & ~ \frac{\d}{\d A_{j_1,i_1}}( - c_g(A,x)^{\top} \cdot f(A,x)_{j_0}\diag(x) A^{\top} ((e_{j_0} - f(A,x)) \circ q(A,x)))
    \end{align*}
    Then, we define $B_{7,1}^{j_1,i_1,j_0,i_0}, \cdots, B_{7,6}^{j_1,i_1,j_0,i_0}$ as follow
    \begin{align*}
        B_{7,1}^{j_1,i_1,j_0,i_0} : = & ~ \frac{\d}{\d A_{j_1,i_1}} (- c_g(A,x)^{\top} ) \cdot  f(A,x)_{j_0} \cdot \diag(x) \cdot A^{\top} \cdot ((e_{j_0} - f(A,x)) \circ q(A,x)))\\
B_{7,2}^{j_1,i_1,j_0,i_0} : = & ~- c_g(A,x)^{\top} \cdot \frac{\d}{\d A_{j_1,i_1}} ( f(A,x)_{j_0} )  \cdot \diag(x) \cdot A^{\top}   \cdot((e_{j_0} - f(A,x)) \circ q(A,x)))\\
B_{7,3}^{j_1,i_1,j_0,i_0} : = & ~  - c_g(A,x)^{\top} \cdot f(A,x)_{j_0} \cdot \diag(x) \cdot \frac{\d}{\d A_{j_1,i_1}} (  A^{\top} )   \cdot((e_{j_0} - f(A,x)) \circ q(A,x))) \\
B_{7,4}^{j_1,i_1,j_0,i_0} : = & ~  - c_g(A,x)^{\top} \cdot f(A,x)_{j_0} \cdot \diag(x) \cdot A^{\top} \cdot((e_{j_0} -\frac{\d f(A,x)}{\d A_{j_1,i_1}} ) \circ q(A,x))) \\
B_{7,5}^{j_1,i_1,j_0,i_0} : = & ~   c_g(A,x)^{\top} \cdot f(A,x)_{j_0} \cdot \diag(x) \cdot A^{\top} \cdot \cdot((e_{j_0} - f(A,x)) \circ \frac{\d q(A,x)}{\d A_{j_1,i_1}}) \\
    \end{align*}
    It is easy to show
    \begin{align*}
        B_7^{j_1,i_1,j_0,i_0} = B_{7,1}^{j_1,i_1,j_0,i_0} +  B_{7,2}^{j_1,i_1,j_0,i_0} + B_{7,3}^{j_1,i_1,j_0,i_0} + B_{7,4}^{j_1,i_1,j_0,i_0} + B_{7,5}^{j_1,i_1,j_0,i_0} 
    \end{align*}
\end{definition}

\subsection{Main Results}
\begin{lemma}\label{lem:hessian_main_result}
   If the following conditions hold
    \begin{itemize}
     \item Let $u(A,x) \in \R^n$ be defined as Definition~\ref{def:u}
    \item Let $\alpha(A,x) \in \R$ be defined as Definition~\ref{def:alpha}
     \item Let $f(A,x) \in \R^n$ be defined as Definition~\ref{def:f}
    \item Let $c(A,x) \in \R^n$ be defined as Definition~\ref{def:c}
    \item Let $g(A,x) \in \R^d$ be defined as Definition~\ref{def:g} 
    \item Let $q(A,x) = c(A,x) + f(A,x) \in \R^n$
    \item Let $c_g(A,x) \in \R^d$ be defined as Definition~\ref{def:c_g}.
    \item Let $L_g(A,x) \in \R$ be defined as Definition~\ref{def:l_g}
     \item Let $R \geq 4$
    \item $\|A \|_2 \leq R, \|A^{\top} \|_F \leq R, \| x\|_2 \leq R, \|\diag(f(A,x)) \|_F \leq \|f(A,x) \|_2 \leq 1, \| b_g \|_2 \leq 1$ 
    \item Let $\beta$ be the lower bound of $\alpha(A,x)$
    \item Let $\beta \in (0, 0.1)$
    \item Let $B_1^{j_1,i_1,j_0,i_0}$ be defined as Definition~\ref{def:b_1}.
    \item Let $B_2^{j_1,i_1,j_0,i_0}$ be defined as Definition~\ref{def:b_2}.
   \item Let $B_2^{j_1,i_1,j_0,i_0}$ be defined as Definition~\ref{def:b_3}.
  \item Let $B_2^{j_1,i_1,j_0,i_0}$ be defined as Definition~\ref{def:b_4}.
     \item Let $B_2^{j_1,i_1,j_0,i_0}$ be defined as Definition~\ref{def:b_5}.
    \item Let $B_2^{j_1,i_1,j_0,i_0}$ be defined as Definition~\ref{def:b_6}.
    \item Let $B_2^{j_1,i_1,j_0,i_0}$ be defined as Definition~\ref{def:b_7}.       
    \end{itemize}
Then, For $j_0,j_1 \in [n], i_0,i_1 \in [d]$, we have 
\begin{itemize}
    \item {\bf Part 1.} 
    For $j_0 = j_1$ and $i_0 = i_1$ \begin{align*} 
   & ~ \frac{\d^2 L_g(A,x) }{ \d A_{j_0,i_0} \d A_{j_1,i_1}} \\
   = & ~   B_1^{j_1,i_1,j_1,i_1} + B_2^{j_1,i_1,j_1,i_1} + B_3^{j_1,i_1,j_1,i_1} + B_4^{j_1,i_1,j_1,i_1} + B_5^{j_1,i_1,j_1,i_1} + B_6^{j_1,i_1,j_1,i_1} + B_7^{j_1,i_1,j_1,i_1}
    \end{align*}
    \item {\bf Part 2.}
    For $j_0 = j_1$ and $i_0 \neq i_1$ 
    \begin{align*}
         & ~ \frac{\d^2 L_g(A,x) }{ \d A_{j_0,i_0} \d A_{j_1,i_1}} \\
         = & ~ B_1^{j_1,i_1,j_1,i_0} + B_2^{j_1,i_1,j_1,i_0} + B_3^{j_1,i_1,j_1,i_0} + B_4^{j_1,i_1,j_1,i_0} + B_5^{j_1,i_1,j_1,i_0} + B_6^{j_1,i_1,j_1,i_0} + B_7^{j_1,i_1,j_1,i_0}
    \end{align*}
    \item {\bf Part 3.}
    For $j_0 \neq j_1$ and $i_0 = i_1$,
    \begin{align*}
         & ~\frac{\d^2 L_g(A,x) }{ \d A_{j_0,i_0} \d A_{j_1,i_1}} \\
         = & ~ B_1^{j_1,i_1,j_0,i_1} + B_2^{j_1,i_1,j_0,i_1} + B_3^{j_1,i_1,j_0,i_1} + B_4^{j_1,i_1,j_0,i_1} + B_5^{j_1,i_1,j_0,i_1} + B_6^{j_1,i_1,j_0,i_1} + B_7^{j_1,i_1,j_0,i_1}
    \end{align*}
    \item {\bf Part 4.} For $j_0 \neq j_1$ and $i_0 \neq i_1$,
    \begin{align*}
         & ~\frac{\d^2 L_g(A,x) }{ \d A_{j_0,i_0} \d A_{j_1,i_1}} \\
          = & ~ B_1^{j_1,i_1,j_0,i_0} + B_2^{j_1,i_1,j_0,i_0} + B_3^{j_1,i_1,j_0,i_0} + B_4^{j_1,i_1,j_0,i_0} + B_5^{j_1,i_1,j_0,i_0} + B_6^{j_1,i_1,j_0,i_0} + B_7^{j_1,i_1,j_0,i_0}
    \end{align*}
\end{itemize}

\end{lemma}

\begin{proof}
    {\bf Proof of Part 1.}
    \begin{align*}
        & ~ \frac{\d^2 L_g(A,x) }{ \d A_{j_0,i_0} \d A_{j_1,i_1}} \\
        = & ~ \frac{\d}{\d A_{j_1,i_1}} (\frac{\d L_g(A,x)}{\d A_{j_1,i_1}} )\\
        = & ~ \frac{\d}{\d A_{j_1,i_1}} (-  c_g(A,x)^{\top} \cdot f(A,x)_{j_1}  \cdot \langle c(A,x), f(A,x) \rangle \cdot e_{i_1} \\
        & ~ - c_g(A,x)^{\top} \cdot f(A,x)_{j_1} \cdot c(A,x)_{j_1} \cdot e_{i_1} \\
        & ~ - c_g(A,x)^{\top} \cdot f(A,x)_{j_1} \cdot \langle c(A,x), f(A,x) \rangle \cdot ( (A_{j_1,*})^\top \circ x )\\
        & ~ + c_g(A,x)^{\top} \cdot f(A,x)_{j_1}\diag (x) A^{\top}   f(A,x)  \cdot    \langle c(A, x), f(A, x) \rangle \\
        & ~ - c_g(A,x)^{\top} \cdot f(A,x)_{j_1}\diag(x) A^{\top} f(A,x) \cdot (\langle -f(A,x), f(A,x) \rangle + f(A,x)_{j_1}) \\
        & ~ - c_g(A,x)^{\top} \cdot  f(A,x)_{j_1}\diag(x) A^{\top} f(A,x) \cdot(\langle -f(A,x), c(A,x) \rangle + f(A,x)_{j_1}) \\
        & ~ - c_g(A,x)^{\top} \cdot f(A,x)_{j_1}\diag(x) A^{\top} ((e_{j_1} - f(A,x)) \circ q(A,x)))\\
        = & ~ \frac{\d}{\d A_{j_1,i_1}}(-  c_g(A,x)^{\top} \cdot f(A,x)_{j_1}  \cdot \langle c(A,x), f(A,x) \rangle \cdot e_{i_1} )\\
        & + ~ \frac{\d}{\d A_{j_1,i_1}}(- c_g(A,x)^{\top} \cdot f(A,x)_{j_1} \cdot c(A,x)_{j_1} \cdot e_{i_1} )\\
        & + ~   \frac{\d}{\d A_{j_1,i_1}} (- c_g(A,x)^{\top} \cdot f(A,x)_{j_1} \cdot \langle c(A,x), f(A,x) \rangle \cdot ( (A_{j_1,*})^\top \circ x )) \\
        & + ~ \frac{\d}{\d A_{j_1,i_1}}  (c_g(A,x)^{\top} \cdot f(A,x)_{j_1}\diag (x) A^{\top}   f(A,x)  \cdot    \langle c(A, x), f(A, x) \rangle ) \\
        & + ~ \frac{\d}{\d A_{j_1,i_1}} (- c_g(A,x)^{\top} \cdot f(A,x)_{j_1}\diag(x) A^{\top} f(A,x) \cdot (\langle -f(A,x), f(A,x) \rangle + f(A,x)_{j_1}))\\
        & + ~ \frac{\d}{\d A_{j_1,i_1}} (- c_g(A,x)^{\top} \cdot  f(A,x)_{j_1}\diag(x) A^{\top} f(A,x) \cdot(\langle -f(A,x), c(A,x) \rangle + f(A,x)_{j_1}))\\
        & + ~  \frac{\d}{\d A_{j_1,i_1}}( - c_g(A,x)^{\top} \cdot f(A,x)_{j_1}\diag(x) A^{\top} ((e_{j_1} - f(A,x)) \circ q(A,x)))\\ 
    = & ~    B_1^{j_1,i_1,j_1,i_1} + B_2^{j_1,i_1,j_1,i_1} + B_3^{j_1,i_1,j_1,i_1} + B_4^{j_1,i_1,j_1,i_1} + B_5^{j_1,i_1,j_1,i_1} + B_6^{j_1,i_1,j_1,i_1} + B_7^{j_1,i_1,j_1,i_1}
    \end{align*}
    where the first step follows from the expansion of Hessian, the second step follows from Lemma~\ref{lem:gradient_lg}, and the last step follows from the definition of $ B_1^{j_1,i_1,j_1,i_1}, \cdots, B_7^{j_1,i_1,j_1,i_1}$.

    {\bf Proof of Part 2.}
     \begin{align*}
       & ~ \frac{\d}{\d A_{j_1,i_1}} (\frac{\d L_g(A,x)}{\d A_{j_1,i_0}} )\\
        = & ~ \frac{\d}{\d A_{j_1,i_1}} (-  c_g(A,x)^{\top} \cdot f(A,x)_{j_1}  \cdot \langle c(A,x), f(A,x) \rangle \cdot e_{i_0} \\
        & ~ - c_g(A,x)^{\top} \cdot f(A,x)_{j_1} \cdot c(A,x)_{j_1} \cdot e_{i_0} \\
        & ~ - c_g(A,x)^{\top} \cdot f(A,x)_{j_1} \cdot \langle c(A,x), f(A,x) \rangle \cdot ( (A_{j_1,*})^\top \circ x )\\
        & ~ + c_g(A,x)^{\top} \cdot f(A,x)_{j_1}\diag (x) A^{\top}   f(A,x)  \cdot    \langle c(A, x), f(A, x) \rangle \\
        & ~ - c_g(A,x)^{\top} \cdot f(A,x)_{j_1}\diag(x) A^{\top} f(A,x) \cdot (\langle -f(A,x), f(A,x) \rangle + f(A,x)_{j_1}) \\
        & ~ - c_g(A,x)^{\top} \cdot  f(A,x)_{j_1}\diag(x) A^{\top} f(A,x) \cdot(\langle -f(A,x), c(A,x) \rangle + f(A,x)_{j_1}) \\
        & ~ - c_g(A,x)^{\top} \cdot f(A,x)_{j_1}\diag(x) A^{\top} ((e_{j_1} - f(A,x)) \circ q(A,x)))\\
        = & ~ \frac{\d}{\d A_{j_1,i_1}}(-  c_g(A,x)^{\top} \cdot f(A,x)_{j_1}  \cdot \langle c(A,x), f(A,x) \rangle \cdot e_{i_0} )\\
        & + ~ \frac{\d}{\d A_{j_1,i_1}}(- c_g(A,x)^{\top} \cdot f(A,x)_{j_1} \cdot c(A,x)_{j_1} \cdot e_{i_0} )\\
        & + ~   \frac{\d}{\d A_{j_1,i_1}} (- c_g(A,x)^{\top} \cdot f(A,x)_{j_1} \cdot \langle c(A,x), f(A,x) \rangle \cdot ( (A_{j_1,*})^\top \circ x )) \\
        & + ~ \frac{\d}{\d A_{j_1,i_1}}  (c_g(A,x)^{\top} \cdot f(A,x)_{j_1}\diag (x) A^{\top}   f(A,x)  \cdot    \langle c(A, x), f(A, x) \rangle ) \\
        & + ~ \frac{\d}{\d A_{j_1,i_1}} (- c_g(A,x)^{\top} \cdot f(A,x)_{j_1}\diag(x) A^{\top} f(A,x) \cdot (\langle -f(A,x), f(A,x) \rangle + f(A,x)_{j_1}))\\
        & + ~ \frac{\d}{\d A_{j_1,i_1}} (- c_g(A,x)^{\top} \cdot  f(A,x)_{j_1}\diag(x) A^{\top} f(A,x) \cdot(\langle -f(A,x), c(A,x) \rangle + f(A,x)_{j_1}))\\
        & + ~  \frac{\d}{\d A_{j_1,i_1}}( - c_g(A,x)^{\top} \cdot f(A,x)_{j_1}\diag(x) A^{\top} ((e_{j_1} - f(A,x)) \circ q(A,x))) \\
        = & ~     B_1^{j_1,i_1,j_1,i_0} + B_2^{j_1,i_1,j_1,i_0} + B_3^{j_1,i_1,j_1,i_0} + B_4^{j_1,i_1,j_1,i_0} + B_5^{j_1,i_1,j_1,i_0} + B_6^{j_1,i_1,j_1,i_0} + B_7^{j_1,i_1,j_1,i_0}
\end{align*}

{\bf Proof of Part 3.}
\begin{align*}
       & ~ \frac{\d}{\d A_{j_1,i_1}} (\frac{\d L_g(A,x)}{\d A_{j_0,i_1}} )\\
        = & ~ \frac{\d}{\d A_{j_1,i_1}} (-  c_g(A,x)^{\top} \cdot f(A,x)_{j_0}  \cdot \langle c(A,x), f(A,x) \rangle \cdot e_{i_1} \\
        & ~ - c_g(A,x)^{\top} \cdot f(A,x)_{j_0} \cdot c(A,x)_{j_0} \cdot e_{i_1} \\
        & ~ - c_g(A,x)^{\top} \cdot f(A,x)_{j_0} \cdot \langle c(A,x), f(A,x) \rangle \cdot ( (A_{j_0,*})^\top \circ x )\\
        & ~ + c_g(A,x)^{\top} \cdot f(A,x)_{j_0}\diag (x) A^{\top}   f(A,x)  \cdot    \langle c(A, x), f(A, x) \rangle \\
        & ~ - c_g(A,x)^{\top} \cdot f(A,x)_{j_0}\diag(x) A^{\top} f(A,x) \cdot (\langle -f(A,x), f(A,x) \rangle + f(A,x)_{j_0}) \\
        & ~ - c_g(A,x)^{\top} \cdot  f(A,x)_{j_0}\diag(x) A^{\top} f(A,x) \cdot(\langle -f(A,x), c(A,x) \rangle + f(A,x)_{j_0}) \\
        & ~ - c_g(A,x)^{\top} \cdot f(A,x)_{j_0}\diag(x) A^{\top} ((e_{j_0} - f(A,x)) \circ q(A,x)))\\
    = & ~ \frac{\d}{\d A_{j_1,i_1}}(-  c_g(A,x)^{\top} \cdot f(A,x)_{j_0}  \cdot \langle c(A,x), f(A,x) \rangle \cdot e_{i_1} )\\
        & + ~ \frac{\d}{\d A_{j_1,i_1}}(- c_g(A,x)^{\top} \cdot f(A,x)_{j_0} \cdot c(A,x)_{j_0} \cdot e_{i_1} )\\
        & + ~   \frac{\d}{\d A_{j_1,i_1}} (- c_g(A,x)^{\top} \cdot f(A,x)_{j_0} \cdot \langle c(A,x), f(A,x) \rangle \cdot ( (A_{j_0,*})^\top \circ x )) \\
        & + ~ \frac{\d}{\d A_{j_1,i_1}}  (c_g(A,x)^{\top} \cdot f(A,x)_{j_0}\diag (x) A^{\top}   f(A,x)  \cdot    \langle c(A, x), f(A, x) \rangle ) \\
        & + ~ \frac{\d}{\d A_{j_1,i_1}} (- c_g(A,x)^{\top} \cdot f(A,x)_{j_0}\diag(x) A^{\top} f(A,x) \cdot (\langle -f(A,x), f(A,x) \rangle + f(A,x)_{j_0}))\\
        & + ~ \frac{\d}{\d A_{j_1,i_1}} (- c_g(A,x)^{\top} \cdot  f(A,x)_{j_0}\diag(x) A^{\top} f(A,x) \cdot(\langle -f(A,x), c(A,x) \rangle + f(A,x)_{j_0}))\\
        & + ~  \frac{\d}{\d A_{j_1,i_1}}( - c_g(A,x)^{\top} \cdot f(A,x)_{j_0}\diag(x) A^{\top} ((e_{j_0} - f(A,x)) \circ q(A,x))) \\
   = & ~ B_1^{j_1,i_1,j_0,i_1} + B_2^{j_1,i_1,j_0,i_1} + B_3^{j_1,i_1,j_0,i_1} + B_4^{j_1,i_1,j_0,i_1} + B_5^{j_1,i_1,j_0,i_1} + B_6^{j_1,i_1,j_0,i_1} + B_7^{j_1,i_1,j_0,i_1}
\end{align*}

{\bf Proof of Part 4.}

\begin{align*}
    & ~ \frac{\d}{\d A_{j_1,i_1}} (\frac{\d L_g(A,x)}{\d A_{j_0,i_0}} )\\
        = & ~ \frac{\d}{\d A_{j_1,i_1}} (-  c_g(A,x)^{\top} \cdot f(A,x)_{j_0}  \cdot \langle c(A,x), f(A,x) \rangle \cdot e_{i_0} \\
        & ~ - c_g(A,x)^{\top} \cdot f(A,x)_{j_0} \cdot c(A,x)_{j_0} \cdot e_{i_0} \\
        & ~ - c_g(A,x)^{\top} \cdot f(A,x)_{j_0} \cdot \langle c(A,x), f(A,x) \rangle \cdot ( (A_{j_0,*})^\top \circ x )\\
        & ~ + c_g(A,x)^{\top} \cdot f(A,x)_{j_0}\diag (x) A^{\top}   f(A,x)  \cdot    \langle c(A, x), f(A, x) \rangle \\
        & ~ - c_g(A,x)^{\top} \cdot f(A,x)_{j_0}\diag(x) A^{\top} f(A,x) \cdot (\langle -f(A,x), f(A,x) \rangle + f(A,x)_{j_0}) \\
        & ~ - c_g(A,x)^{\top} \cdot  f(A,x)_{j_0}\diag(x) A^{\top} f(A,x) \cdot(\langle -f(A,x), c(A,x) \rangle + f(A,x)_{j_0}) \\
        & ~ - c_g(A,x)^{\top} \cdot f(A,x)_{j_0}\diag(x) A^{\top} ((e_{j_0} - f(A,x)) \circ q(A,x)))\\
    = & ~ \frac{\d}{\d A_{j_1,i_1}}(-  c_g(A,x)^{\top} \cdot f(A,x)_{j_0}  \cdot \langle c(A,x), f(A,x) \rangle \cdot e_{i_0} )\\
        & + ~ \frac{\d}{\d A_{j_1,i_1}}(- c_g(A,x)^{\top} \cdot f(A,x)_{j_0} \cdot c(A,x)_{j_0} \cdot e_{i_0} )\\
        & + ~   \frac{\d}{\d A_{j_1,i_1}} (- c_g(A,x)^{\top} \cdot f(A,x)_{j_0} \cdot \langle c(A,x), f(A,x) \rangle \cdot ( (A_{j_0,*})^\top \circ x )) \\
        & + ~ \frac{\d}{\d A_{j_1,i_1}}  (c_g(A,x)^{\top} \cdot f(A,x)_{j_0}\diag (x) A^{\top}   f(A,x)  \cdot    \langle c(A, x), f(A, x) \rangle ) \\
        & + ~ \frac{\d}{\d A_{j_1,i_1}} (- c_g(A,x)^{\top} \cdot f(A,x)_{j_0}\diag(x) A^{\top} f(A,x) \cdot (\langle -f(A,x), f(A,x) \rangle + f(A,x)_{j_0}))\\
        & + ~ \frac{\d}{\d A_{j_1,i_1}} (- c_g(A,x)^{\top} \cdot  f(A,x)_{j_0}\diag(x) A^{\top} f(A,x) \cdot(\langle -f(A,x), c(A,x) \rangle + f(A,x)_{j_0}))\\
        & + ~  \frac{\d}{\d A_{j_1,i_1}}( - c_g(A,x)^{\top} \cdot f(A,x)_{j_0}\diag(x) A^{\top} ((e_{j_0} - f(A,x)) \circ q(A,x))) \\
    = & ~B_1^{j_1,i_1,j_0,i_0} + B_2^{j_1,i_1,j_0,i_0} + B_3^{j_1,i_1,j_0,i_0} + B_4^{j_1,i_1,j_0,i_0} + B_5^{j_1,i_1,j_0,i_0} + B_6^{j_1,i_1,j_0,i_0} + B_7^{j_1,i_1,j_0,i_0}
\end{align*} 
\end{proof}

\subsection{Helpful Definitions}
Here we define some notation for making the later Hessian proofs easier
\begin{definition}\label{def:h}
We define $h(A,x) \in \R^d$ as follows
\begin{align*}
    h(A,x) := \underbrace{ \diag(x) }_{d \times d} \underbrace{ A^\top }_{d \times n} \underbrace{ f(A,x) }_{n \times 1}
\end{align*}
\end{definition}
\begin{definition}\label{def:h_e}
We define $h_e(A,x) \in \R^d$ as follows
\begin{align*}
    h_e(A,x) := \diag (x) \cdot A^{\top}     \cdot (e_{j_1}- f(A,x) )
\end{align*}
\end{definition}

\begin{definition}\label{def:f_c}
We define $f_c(A,x) \in \R$ as follows
\begin{align*}
    f_c(A,x) := \langle f(A,x), c(A,x) \rangle
\end{align*}
\end{definition}

\begin{definition}\label{def:f_2}
We define $f_2(A,x) \in \R$ as follows
\begin{align*}
    f_2(A,x) := \langle f(A,x), f(A,x) \rangle
\end{align*}
\end{definition}

\begin{definition}\label{def:p}
We define $p_{j_1}(A,x) \in \R^d$ as follows
\begin{align*}
    p_{j_1}(A,x) :=  \diag(x) \cdot A^\top \cdot   ((e_{j_1} - f(A,x)) \circ q(A,x))
\end{align*}
\end{definition}

\subsection{Hessian PSD}

\begin{lemma}
If the following conditions hold
\begin{itemize}
    \item Let $H \in \R^{nd \times nd}$ be a hessian
    \item Let $H^{j_0,j_1} \in \R^{d \times d}$ denote the $(j_0,j_1)$-th block of $H$.
    \item For all $j_0,j_1 \in [n]$, we have $\| H^{j_0,j_1} \| \leq \lambda$
\end{itemize}
Then, we have
\begin{itemize}
    \item Part 1. $\| H \| \leq n^2 \lambda$
    \item Part 2. $H \succeq - n^2 \lambda \cdot I_{nd}$
\end{itemize}
\end{lemma}
\begin{proof}
The Part 1 follows from definition and simple algebera.

The part 2 follows from Part 1.
\end{proof}

\newpage

\newpage
\section{Gradient Related to Gradient Matching}\label{app:more_gradients}
\begin{lemma}\label{lem:gradients_for_functions}
   If the following conditions hold
    \begin{itemize}
     \item Let $u(A,x) \in \R^n$ be defined as Definition~\ref{def:u}
    \item Let $\alpha(A,x) \in \R$ be defined as Definition~\ref{def:alpha}
     \item Let $f(A,x) \in \R^n$ be defined as Definition~\ref{def:f}
    \item Let $c(A,x) \in \R^n$ be defined as Definition~\ref{def:c}
    \item Let $g(A,x) \in \R^d$ be defined as Definition~\ref{def:g} 
    \item Let $q(A,x) = c(A,x) + f(A,x) \in \R^n$
    \item Let $c_g(A,x) \in \R^d$ be defined as Definition~\ref{def:c_g}.
    \item Let $L_g(A,x) \in \R$ be defined as Definition~\ref{def:l_g}
    \item Let $v \in \R^n$ be a vector 
    \end{itemize}
Then, For $j_0,j_1 \in [n], i_0,i_1 \in [d]$, we have 
\begin{itemize}
    \item {\bf Part 1.} Let $j_0 = j_1$
    \begin{align*}
        \frac{ \d (A_{j_1,*}^{\top} \circ x) }{\d A_{j_1,i_1}} =  e_{i_1} \circ x
    \end{align*}
    \item {\bf Part 2.} Let $j_0 \neq j_1$
    \begin{align*}
        \frac{ \d (A_{j_0,*}^{\top} \circ x) }{\d A_{j_1,i_1}} =  0
    \end{align*}
    \item {\bf Part 3.} Let $j_0 = j_1$ \begin{align*}
       \frac{\d  A^{\top}}{\d A_{j_1,i_1}} = & ~  e_{i_1} \cdot  e_{j_1}^\top
    \end{align*}
     \item {\bf Part 4.} \begin{align*}
       \underbrace{\frac{ \d c_g(A,x)^\top }{ \d A_{j_1,i_1} } }_{1 \times d}
      = & ~ -  \underbrace{e_{i_1}^\top}_{1 \times d} \cdot \underbrace{\langle c(A,x), f(A,x) \rangle}_{scalar} \cdot  \underbrace{f(A,x)_{j_1}}_{scalar}\\
        & ~ - \underbrace{e_{i_1}^\top}_{1 \times d} \cdot \underbrace{c(A,x)_{j_1}}_{scalar} \cdot \underbrace{f(A,x)_{j_1}}_{scalar} \\
        & ~ - \underbrace{f(A,x)_{j_1}}_{scalar} \cdot \underbrace{\langle c(A,x), f(A,x) \rangle}_{scalar} \cdot ( \underbrace{(A_{j_1,*})}_{1 \times d} \circ \underbrace{x^\top}_{1 \times d} )\\
        & ~ +  \underbrace{f(A,x)_{j_1}}_{scalar} \cdot \underbrace{f(A,x)^\top}_{1 \times n}  \cdot \underbrace{A}_{n \times d} \cdot   \underbrace{\diag(x)}_{d \times d} \cdot    \underbrace{\langle c(A,x), f(A,x) \rangle}_{scalar} \\
        & ~ -  \underbrace{f(A,x)_{j_1}}_{scalar} \cdot \underbrace{f(A,x)^\top}_{1 \times n}  \cdot \underbrace{A}_{n \times d} \cdot \underbrace{\diag(x)}_{d \times d} \cdot (\underbrace{\langle -f(A,x), f(A,x) \rangle}_{scalar} + \underbrace{f(A,x)_{j_1}}_{scalar}) \\
        & ~ -   \underbrace{f(A,x)_{j_1}}_{scalar} \cdot \underbrace{f(A,x)^\top}_{1 \times n}  \cdot \underbrace{A}_{n \times d} \cdot  \underbrace{\diag(x)}_{d \times d} \cdot(\underbrace{\langle -f(A,x), c(A,x) \rangle}_{scalar} + \underbrace{f(A,x)_{j_1}}_{scalar}) \\
        & ~ - \underbrace{f(A,x)_{j_1}}_{scalar} \cdot ((\underbrace{e_{j_1}^\top}_{1 \times n} - \underbrace{f(A,x)^\top}_{1 \times n}) \circ \underbrace{q(A,x)^\top}_{1 \times n}) \cdot \underbrace{A}_{n \times d} \cdot  \underbrace{\diag(x)}_{d \times d}   
    \end{align*}
\end{itemize}

\end{lemma}

\begin{proof}
{\bf Proof of Part 1.} For $j_0= j_1$. 
We have
\begin{align*}
     \frac{ \d (A_{j_1,*}^{\top} \circ x) }{\d A_{j_1,i_1}} = & ~ \frac{\d A_{j_1,*}^{\top}}{\d A_{j_1,i_1}} \circ x \\
     = & ~ e_{i_1} \circ x
\end{align*}
where the first step comes from the product rule of derivative, the second step follows from  $A_{j_1,*}$ have all non-zero gradient with respect to $A_{j_1,i_1}$.

{\bf Proof of Part 2.} For $j_0 \neq j_1$, We have
\begin{align*}
     \frac{ \d (A_{j_0,*}^{\top} \circ x) }{\d A_{j_1,i_1}} = & ~ \frac{\d A_{j_0,*}^{\top}}{\d A_{j_1,i_1}} \circ x \\
     = & ~ 0
\end{align*}
where the first step comes from the product rule of derivative, the second step follows from  $A_{j_0,*}$ have all $0$ gradient with respect to $A_{j_1,i_1}$

{\bf Proof of Part 3.}
We have
\begin{align*}
    \frac{\d A^{\top} }{ \d A_{j_1,i_1}}  
    = & ~  \underbrace{ e_{i_1} }_{d \times 1} \underbrace{ e_{j_1}^\top }_{1 \times n} \\
\end{align*}
where the first step follows from only one element of $A_{j_1,i_1}$ have non-zero gradient with respect to $A$, and the last step follows from Fact~\ref{fac:circ_rules}

{\bf Proof of Part 4.}
\begin{align*}
    \frac{ \d c_g(A,x)^\top }{ \d A_{j_1,i_1} }  
    = & ~ ( \frac{\d c_g(A,x) }{ \d A_{j_1,i_1} })^{\top} \\
    = & ~ (  - f(A,x)_{j_1}  \cdot \langle c(A,x), f(A,x) \rangle \cdot e_{i_1})^{\top} \\
        & ~ - (f(A,x)_{j_1} \cdot c(A,x)_{j_1} \cdot e_{i_1})^{\top} \\
        & ~ (- f(A,x)_{j_1} \cdot \langle c(A,x), f(A,x) \rangle \cdot ( (A_{j_1,*})^\top \circ x ) )^{\top}\\
        & ~ + ( f(A,x)_{j_1}\diag (x) A^{\top}   f(A,x)  \cdot    \langle c(A, x), f(A, x) \rangle)^{\top} \\
        & ~ (-  f(A,x)_{j_1}\diag(x) A^{\top} f(A,x) \cdot (\langle -f(A,x), f(A,x)\rangle + f(A,x)_{j_1}))^{\top}  \\
        & ~( -   f(A,x)_{j_1}\diag(x) A^{\top} f(A,x) \cdot(\langle-f(A,x), c(A,x) \rangle + f(A,x)_{j_1}))^{\top}  \\
        & ~ (- f(A,x)_{j_1}\diag(x) A^{\top} ((e_{j_1} - f(A,x)) \circ q(A,x)) )^{\top} \\
     = & ~   - f(A,x)_{j_1}  \cdot \langle c(A,x), f(A,x) \rangle \cdot e_{i_1}^{\top} \\
        & ~ - f(A,x)_{j_1} \cdot c(A,x)_{j_1} \cdot e_{i_1}^{\top} \\
        & ~ - f(A,x)_{j_1} \cdot \langle c(A,x), f(A,x) \rangle \cdot ( (A_{j_1,*})\circ x^\top  ) \\
        & ~ +  f(A,x)_{j_1}f(A,x)^{\top}  A  \diag (x)    \cdot    \langle c(A, x), f(A, x) \rangle \\
        & ~ -  f(A,x)_{j_1}f(A,x)^{\top}  A  \diag (x)  \cdot (\langle -f(A,x), f(A,x)\rangle + f(A,x)_{j_1})  \\
        & ~ -   f(A,x)_{j_1}f(A,x)^{\top}  A  \diag (x)  \cdot(\langle-f(A,x), c(A,x) \rangle + f(A,x)_{j_1})  \\
        & ~ - f(A,x)_{j_1} (e_{j_1}^{\top} - f(A,x)^{\top}) \circ q(A,x)^{\top} A  \diag (x)   \\    
\end{align*}
where the first step follows from the simple algebra, the second step follows from Lemma~\ref{lem:gradient_c_g}, and the last step follows from the simple algebra.
\end{proof}

\newpage

\section{Hessian: First term \texorpdfstring{$B_1^{j_1,i_1,j_0,i_0}$}{}{}}\label{app:hessian_first}

\subsection{Definitions}

\begin{definition}\label{def:b_1}
For $j_0 \neq j_1 \in [n]$ and $i_0 \neq i_1 \in [d]$, we define the $B_1^{j_1,i_1,j_0,i_0}$ as follows,
    \begin{align*}
        B_1^{j_1,i_1,j_0,i_0} := & ~ \frac{\d}{\d A_{j_1,i_1}}(-  c_g(A,x)^{\top} \cdot f(A,x)_{j_0}  \cdot \langle c(A,x), f(A,x) \rangle \cdot e_{i_0} )
    \end{align*}
    Then, we define $B_{1,1}^{j_1,i_1,j_0,i_0}, \cdots, B_{1,3}^{j_1,i_1,j_0,i_0}$ as follow
    \begin{align*}
         B_{1,1}^{j_1,i_1,j_0,i_0} : = & ~ - \frac{\d}{\d A_{j_1,i_1}} ( c_g(A,x)^{\top} ) \cdot  f(A,x)_{j_0} \cdot \langle c(A,x), f(A,x) \rangle \cdot e_{i_0}\\
 B_{1,2}^{j_1,i_1,j_0,i_0} : = & ~ - c_g(A,x)^{\top} \cdot \frac{\d}{\d A_{j_1,i_1}} ( f(A,x)_{j_0} )  \cdot \langle c(A,x), f(A,x) \rangle \cdot e_{i_0}\\
 B_{1,3}^{j_1,i_1,j_0,i_0} : = & ~  -  c_g(A,x)^{\top} \cdot f(A,x)_{j_0}  \cdot  \langle\frac{\d c(A,x)}{\d A_{j_1,i_1}}, f(A,x) \rangle \cdot e_{i_0}\\
  B_{1,4}^{j_1,i_1,j_0,i_0} : = & ~  -  c_g(A,x)^{\top} \cdot f(A,x)_{j_0}  \cdot  \langle c(A,x), \frac{\d f(A,x)}{\d A_{j_1,i_1}} \rangle \cdot e_{i_0}
    \end{align*}
    It is easy to show
    \begin{align*}
        B_1^{j_1,i_1,j_0,i_0} = B_{1,1}^{j_1,i_1,j_0,i_0} +  B_{1,2}^{j_1,i_1,j_0,i_0} + B_{1,3}^{j_1,i_1,j_0,i_0}  + B_{1,4}^{j_1,i_1,j_0,i_0} 
    \end{align*}
    Similarly for $j_1 = j_0$ and $i_0 = i_1$,we have
    \begin{align*}
        B_1^{j_1,i_1,j_1,i_1} = B_{1,1}^{j_1,i_1,j_1,i_1} +  B_{1,2}^{j_1,i_1,j_1,i_1} + B_{1,3}^{j_1,i_1,j_1,i_1}  + B_{1,4}^{j_1,i_1,j_1,i_1} 
         \end{align*}
    For $j_1 = j_0$ and $i_0 \neq i_1$,we have
    \begin{align*}
        B_1^{j_1,i_1,j_1,i_0} = B_{1,1}^{j_1,i_1,j_1,i_0} +  B_{1,2}^{j_1,i_1,j_1,i_0} + B_{1,3}^{j_1,i_1,j_1,i_0}  + B_{1,4}^{j_1,i_1,j_1,i_0} 
    \end{align*}
    For $j_1 \neq j_0$ and $i_0 = i_1$,we have
    \begin{align*}
        B_1^{j_1,i_1,j_0,i_1} = B_{1,1}^{j_1,i_1,j_0,i_1} +  B_{1,2}^{j_1,i_1,j_0,i_1} + B_{1,3}^{j_1,i_1,j_0,i_1}  + B_{1,4}^{j_1,i_1,j_0,i_1} 
    \end{align*}
\end{definition}

\subsection{Case \texorpdfstring{$j_1=j_0, i_1 = i_0$}{}}
\begin{lemma}
For $j_1 = j_0$ and $i_0 = i_1$. If the following conditions hold
    \begin{itemize}
     \item Let $u(A,x) \in \R^n$ be defined as Definition~\ref{def:u}
    \item Let $\alpha(A,x) \in \R$ be defined as Definition~\ref{def:alpha}
     \item Let $f(A,x) \in \R^n$ be defined as Definition~\ref{def:f}
    \item Let $c(A,x) \in \R^n$ be defined as Definition~\ref{def:c}
    \item Let $g(A,x) \in \R^d$ be defined as Definition~\ref{def:g} 
    \item Let $q(A,x) = c(A,x) + f(A,x) \in \R^n$
    \item Let $c_g(A,x) \in \R^d$ be defined as Definition~\ref{def:c_g}.
    \item Let $L_g(A,x) \in \R$ be defined as Definition~\ref{def:l_g}
    \item Let $v \in \R^n$ be a vector 
    \item Let $B_1^{j_1,i_1,j_0,i_0}$ be defined as Definition~\ref{def:b_1}
    \end{itemize}
    Then, For $j_0,j_1 \in [n], i_0,i_1 \in [d]$, we have 
    \begin{itemize}
\item {\bf Part 1.} For $B_{1,1}^{j_1,i_1,j_1,i_1}$, we have 
\begin{align*}
 B_{1,1}^{j_1,i_1,j_1,i_1}  = & ~ \frac{\d}{\d A_{j_1,i_1}} (- c_g(A,x)^{\top} ) \cdot  f(A,x)_{j_1} \cdot \langle c(A,x), f(A,x) \rangle \cdot e_{i_1} \\
 = & ~ B_{1,1,1}^{j_1,i_1,j_1,i_1} + B_{1,1,2}^{j_1,i_1,j_1,i_1} + B_{1,1,3}^{j_1,i_1,j_1,i_1} + B_{1,1,4}^{j_1,i_1,j_1,i_1} + B_{1,1,5}^{j_1,i_1,j_1,i_1} + B_{1,1,6}^{j_1,i_1,j_1,i_1} + B_{1,1,7}^{j_1,i_1,j_1,i_1}
\end{align*} 
\item {\bf Part 2.} For $B_{1,2}^{j_1,i_1,j_1,i_1}$, we have 
\begin{align*}
  B_{1,2}^{j_1,i_1,j_1,i_1} = & ~ - c_g(A,x)^{\top} \cdot \frac{\d}{\d A_{j_1,i_1}} ( f(A,x)_{j_1} )  \cdot \langle c(A,x), f(A,x) \rangle \cdot e_{i_1} \\
    = & ~  B_{1,2,1}^{j_1,i_1,j_1,i_1} + B_{1,2,2}^{j_1,i_1,j_1,i_1}
\end{align*} 
\item {\bf Part 3.} For $B_{1,3}^{j_1,i_1,j_1,i_1}$, we have 
\begin{align*}
  B_{1,3}^{j_1,i_1,j_1,i_1} = & ~ -  c_g(A,x)^{\top} \cdot f(A,x)_{j_1}  \cdot \d \langle \frac{c(A,x)}{\d A_{j_1,i_1}} , f(A,x) \rangle\cdot e_{i_1} \\
     = & ~ B_{1,3,1}^{j_1,i_1,j_1,i_1} + B_{1,3,2}^{j_1,i_1,j_1,i_1}
\end{align*} 
\item {\bf Part 4.} For $B_{1,4}^{j_1,i_1,j_1,i_1}$, we have 
\begin{align*}
 B_{1,4}^{j_1,i_1,j_1,i_1} = & ~ -  c_g(A,x)^{\top} \cdot f(A,x)_{j_1}  \cdot \d \langle c(A,x), \frac{f(A,x)}{\d A_{j_1,i_1}}  \rangle\cdot e_{i_1} \\
     = & ~ B_{1,4,1}^{j_1,i_1,j_1,i_1} + B_{1,4,2}^{j_1,i_1,j_1,i_1}
\end{align*} 
\end{itemize}
\begin{proof}
    {\bf Proof of Part 1.}
    \begin{align*}
    B_{1,1,1}^{j_1,i_1,j_1,i_1} : = & ~ f(A,x)_{j_1}^2 \cdot  e_{i_1}^\top  \cdot \langle c(A,x), f(A,x) \rangle^2 \cdot e_{i_1}\\
    B_{1,1,2}^{j_1,i_1,j_1,i_1} : = & ~  f(A,x)_{j_1}^2 \cdot c(A,x)_{j_1}\cdot e_{i_1}^\top \cdot  \langle c(A,x), f(A,x) \rangle \cdot e_{i_1}\\
    B_{1,1,3}^{j_1,i_1,j_1,i_1} : = & ~ f(A,x)_{j_1}^2 \cdot \langle c(A,x), f(A,x) \rangle \cdot ( (A_{j_1,*}) \circ x^\top  )  \cdot \langle c(A,x), f(A,x) \rangle \cdot e_{i_1}\\
    B_{1,1,4}^{j_1,i_1,j_1,i_1} : = & ~ -  f(A,x)_{j_1}^2 \cdot f(A,x)^\top  \cdot A \cdot \diag(x) \cdot   (\langle c(A,x), f(A,x) \rangle)^2 \cdot  e_{i_1}\\
    B_{1,1,5}^{j_1,i_1,j_1,i_1} : = & ~   f(A,x)_{j_1}^2 \cdot f(A,x)^\top  \cdot A \cdot \diag(x) \cdot (\langle -f(A,x), f(A,x) \rangle + f(A,x)_{j_1})  \cdot \langle c(A,x), f(A,x) \rangle \cdot e_{i_1} \\
    B_{1,1,6}^{j_1,i_1,j_1,i_1} : = & ~    f(A,x)_{j_1}^2 \cdot f(A,x)^\top  \cdot A \cdot  \diag(x) \cdot(\langle -f(A,x), c(A,x) \rangle + f(A,x)_{j_1}) \cdot \langle c(A,x), f(A,x) \rangle \cdot e_{i_1} \\
    B_{1,1,7}^{j_1,i_1,j_1,i_1} : = & ~  f(A,x)_{j_1}^2 \cdot ((e_{j_1}^\top - f(A,x)^\top) \circ q(A,x)^\top) \cdot A \cdot  \diag(x)  \cdot \langle c(A,x), f(A,x) \rangle \cdot e_{i_1}
\end{align*}
Finally, combine them and we have
\begin{align*}
       B_{1,1}^{j_1,i_1,j_1,i_1} = B_{1,1,1}^{j_1,i_1,j_1,i_1} + B_{1,1,2}^{j_1,i_1,j_1,i_1} + B_{1,1,3}^{j_1,i_1,j_1,i_1} + B_{1,1,4}^{j_1,i_1,j_1,i_1} + B_{1,1,5}^{j_1,i_1,j_1,i_1} + B_{1,1,6}^{j_1,i_1,j_1,i_1} + B_{1,1,7}^{j_1,i_1,j_1,i_1}
\end{align*}
{\bf Proof of Part 2.}
    \begin{align*}
    B_{1,2,1}^{j_1,i_1,j_1,i_1} : = & ~ c_g(A,x)^{\top} \cdot f(A,x)_{j_1}^2 \cdot x_{i_1} \cdot \langle c(A,x), f(A,x) \rangle \cdot e_{i_1} \\
    B_{1,2,2}^{j_1,i_1,j_1,i_1} : = & ~  - c_g(A,x)^{\top} \cdot f(A,x)_{j_1} \cdot x_{i_1} \cdot \langle c(A,x), f(A,x) \rangle \cdot e_{i_1}
\end{align*}
Finally, combine them and we have
\begin{align*}
       B_{1,2}^{j_1,i_1,j_1,i_1} = B_{1,2,1}^{j_1,i_1,j_1,i_1} + B_{1,2,2}^{j_1,i_1,j_1,i_1}
\end{align*}
{\bf Proof of Part 3.} 

    \begin{align*}
    B_{1,3,1}^{j_1,i_1,j_1,i_1} : = & ~   c_g(A,x)^{\top} \cdot f(A,x)_{j_1}^2  \cdot x_{i_1} \cdot \langle  f(A,x), f(A,x) \rangle \cdot e_{i_1}\\
    B_{1,3,2}^{j_1,i_1,j_1,i_1} : = & ~  -  c_g(A,x)^{\top} \cdot f(A,x)_{j_1}^3 \cdot x_{i_1}  \cdot e_{i_1}
\end{align*}
Finally, combine them and we have
\begin{align*}
       B_{1,3}^{j_1,i_1,j_1,i_1} = B_{1,3,1}^{j_1,i_1,j_1,i_1} + B_{1,3,2}^{j_1,i_1,j_1,i_1}
\end{align*}

{\bf Proof of Part 4.}

    \begin{align*}
    B_{1,4,1}^{j_1,i_1,j_1,i_1} : = & ~    c_g(A,x)^{\top} \cdot f(A,x)_{j_1}^2 \cdot x_{i_1} \cdot (\langle  f(A,x), c(A,x) \rangle ) \cdot e_{i_1}\\
    B_{1,4,2}^{j_1,i_1,j_1,i_1} : = & ~  -  c_g(A,x)^{\top} \cdot f(A,x)_{j_1}^2 \cdot x_{i_1} \cdot c(A,x)_{j_1} \cdot e_{i_1}
\end{align*}
Finally, combine them and we have
\begin{align*}
       B_{1,4}^{j_1,i_1,j_1,i_1} = B_{1,4,1}^{j_1,i_1,j_1,i_1} + B_{1,4,2}^{j_1,i_1,j_1,i_1}
\end{align*}
\end{proof}
\end{lemma}

\subsection{Case \texorpdfstring{$j_1=j_0, i_1 \neq i_0$}{}}
\begin{lemma}
For $j_1 = j_0$ and $i_1 \neq i_0$. If the following conditions hold
    \begin{itemize}
     \item Let $u(A,x) \in \R^n$ be defined as Definition~\ref{def:u}
    \item Let $\alpha(A,x) \in \R$ be defined as Definition~\ref{def:alpha}
     \item Let $f(A,x) \in \R^n$ be defined as Definition~\ref{def:f}
    \item Let $c(A,x) \in \R^n$ be defined as Definition~\ref{def:c}
    \item Let $g(A,x) \in \R^d$ be defined as Definition~\ref{def:g} 
    \item Let $q(A,x) = c(A,x) + f(A,x) \in \R^n$
    \item Let $c_g(A,x) \in \R^d$ be defined as Definition~\ref{def:c_g}.
    \item Let $L_g(A,x) \in \R$ be defined as Definition~\ref{def:l_g}
    \item Let $v \in \R^n$ be a vector 
    \item Let $B_1^{j_1,i_1,j_0,i_0}$ be defined as Definition~\ref{def:b_1}
    \end{itemize}
    Then, For $j_0=j_1 \in [n], i_0,i_1 \in [d]$, we have 
    \begin{itemize}
\item {\bf Part 1.} For $B_{1,1}^{j_1,i_1,j_1,i_0}$, we have 
\begin{align*}
 B_{1,1}^{j_1,i_1,j_1,i_0}  = & ~ \frac{\d}{\d A_{j_1,i_1}} (- c_g(A,x)^{\top} ) \cdot  f(A,x)_{j_1} \cdot \langle c(A,x), f(A,x) \rangle \cdot e_{i_0} \\
 = & ~ B_{1,1,1}^{j_1,i_1,j_1,i_0} + B_{1,1,2}^{j_1,i_1,j_1,i_0} + B_{1,1,3}^{j_1,i_1,j_1,i_0} + B_{1,1,4}^{j_1,i_1,j_1,i_0} + B_{1,1,5}^{j_1,i_1,j_1,i_0} + B_{1,1,6}^{j_1,i_1,j_1,i_0} + B_{1,1,7}^{j_1,i_1,j_1,i_0}
\end{align*} 
\item {\bf Part 2.} For $B_{1,2}^{j_1,i_1,j_1,i_0}$, we have 
\begin{align*}
  B_{1,2}^{j_1,i_1,j_1,i_0} = & ~ - c_g(A,x)^{\top} \cdot \frac{\d}{\d A_{j_1,i_1}} ( f(A,x)_{j_1} )  \cdot \langle c(A,x), f(A,x) \rangle \cdot e_{i_0} \\
    = & ~  B_{1,2,1}^{j_1,i_1,j_1,i_0} + B_{1,2,2}^{j_1,i_1,j_1,i_0}
\end{align*} 
\item {\bf Part 3.} For $B_{1,3}^{j_1,i_1,j_1,i_0}$, we have 
\begin{align*}
  B_{1,3}^{j_1,i_1,j_1,i_0} = & ~ -  c_g(A,x)^{\top} \cdot f(A,x)_{j_1}  \cdot \d \langle \frac{c(A,x)}{\d A_{j_1,i_1}} , f(A,x) \rangle\cdot e_{i_0} \\
     = & ~ B_{1,3,1}^{j_1,i_1,j_1,i_0} + B_{1,3,2}^{j_1,i_1,j_1,i_0}
\end{align*} 
\item {\bf Part 4.} For $B_{1,4}^{j_1,i_1,j_1,i_0}$, we have 
\begin{align*}
 B_{1,4}^{j_1,i_1,j_1,i_0} = & ~ -  c_g(A,x)^{\top} \cdot f(A,x)_{j_1}  \cdot \d \langle c(A,x), \frac{f(A,x)}{\d A_{j_1,i_1}}  \rangle\cdot e_{i_0} \\
     = & ~ B_{1,4,1}^{j_1,i_1,j_1,i_0} + B_{1,4,2}^{j_1,i_1,j_1,i_0}
\end{align*} 
\end{itemize}
\begin{proof}
    {\bf Proof of Part 1.}
    \begin{align*}
    B_{1,1,1}^{j_1,i_1,j_1,i_0} : = & ~ f(A,x)_{j_1}^2 \cdot  e_{i_1}^\top  \cdot \langle c(A,x), f(A,x) \rangle^2 \cdot e_{i_0}\\
    B_{1,1,2}^{j_1,i_1,j_1,i_0} : = & ~  f(A,x)_{j_1}^2 \cdot c(A,x)_{j_1}\cdot e_{i_1}^\top \cdot  \langle c(A,x), f(A,x) \rangle \cdot e_{i_0}\\
    B_{1,1,3}^{j_1,i_1,j_1,i_0} : = & ~ f(A,x)_{j_1}^2 \cdot \langle c(A,x), f(A,x) \rangle \cdot ( (A_{j_1,*}) \circ x^\top  )  \cdot \langle c(A,x), f(A,x) \rangle \cdot e_{i_0}\\
    B_{1,1,4}^{j_1,i_1,j_1,i_0} : = & ~ -  f(A,x)_{j_1}^2 \cdot f(A,x)^\top  \cdot A \cdot \diag(x) \cdot   (\langle c(A,x), f(A,x) \rangle)^2 \cdot  e_{i_0}\\
    B_{1,1,5}^{j_1,i_1,j_1,i_0} : = & ~   f(A,x)_{j_1}^2 \cdot f(A,x)^\top  \cdot A \cdot \diag(x) \cdot (\langle -f(A,x), f(A,x) \rangle + f(A,x)_{j_1})  \cdot \langle c(A,x), f(A,x) \rangle \cdot e_{i_0} \\
    B_{1,1,6}^{j_1,i_1,j_1,i_0} : = & ~    f(A,x)_{j_1}^2 \cdot f(A,x)^\top  \cdot A \cdot  \diag(x) \cdot(\langle -f(A,x), c(A,x) \rangle + f(A,x)_{j_1}) \cdot \langle c(A,x), f(A,x) \rangle \cdot e_{i_0} \\
    B_{1,1,7}^{j_1,i_1,j_1,i_0} : = & ~  f(A,x)_{j_1}^2 \cdot ((e_{j_1}^\top - f(A,x)^\top) \circ q(A,x)^\top) \cdot A \cdot  \diag(x)  \cdot \langle c(A,x), f(A,x) \rangle \cdot e_{i_0}
\end{align*}
Finally, combine them and we have
\begin{align*}
       B_{1,1}^{j_1,i_1,j_1,i_0} = B_{1,1,1}^{j_1,i_1,j_1,i_0} + B_{1,1,2}^{j_1,i_1,j_1,i_0} + B_{1,1,3}^{j_1,i_1,j_1,i_0} + B_{1,1,4}^{j_1,i_1,j_1,i_0} + B_{1,1,5}^{j_1,i_1,j_1,i_0} + B_{1,1,6}^{j_1,i_1,j_1,i_0} + B_{1,1,7}^{j_1,i_1,j_1,i_0}
\end{align*}
{\bf Proof of Part 2.}
    \begin{align*}
    B_{1,2,1}^{j_1,i_1,j_1,i_0} : = & ~ c_g(A,x)^{\top} \cdot f(A,x)_{j_1}^2 \cdot x_{i_1} \cdot \langle c(A,x), f(A,x) \rangle \cdot e_{i_0} \\
    B_{1,2,2}^{j_1,i_1,j_1,i_0} : = & ~  - c_g(A,x)^{\top} \cdot f(A,x)_{j_1} \cdot x_{i_1} \cdot \langle c(A,x), f(A,x) \rangle \cdot e_{i_0}
\end{align*}
Finally, combine them and we have
\begin{align*}
       B_{1,2}^{j_1,i_1,j_1,i_0} = B_{1,2,1}^{j_1,i_1,j_1,i_0} + B_{1,2,2}^{j_1,i_1,j_1,i_0}
\end{align*}
{\bf Proof of Part 3.}
    \begin{align*}
    B_{1,3,1}^{j_1,i_1,j_1,i_0} : = & ~   c_g(A,x)^{\top} \cdot f(A,x)_{j_1}^2  \cdot x_{i_1} \cdot \langle  f(A,x), f(A,x) \rangle \cdot e_{i_0}\\
    B_{1,3,2}^{j_1,i_1,j_1,i_0} : = & ~  -  c_g(A,x)^{\top} \cdot f(A,x)_{j_1}^3 \cdot x_{i_1}  \cdot e_{i_0}
\end{align*}
Finally, combine them and we have
\begin{align*}
       B_{1,3}^{j_1,i_1,j_1,i_0} = B_{1,3,1}^{j_1,i_1,j_1,i_0} + B_{1,3,2}^{j_1,i_1,j_1,i_0}
\end{align*}
{\bf Proof of Part 4.}
    \begin{align*}
    B_{1,4,1}^{j_1,i_1,j_1,i_0} : = & ~    c_g(A,x)^{\top} \cdot f(A,x)_{j_1}^2 \cdot x_{i_1} \cdot (\langle  f(A,x), c(A,x) \rangle ) \cdot e_{i_0}\\
    B_{1,4,2}^{j_1,i_1,j_1,i_0} : = & ~  -  c_g(A,x)^{\top} \cdot f(A,x)_{j_1}^2 \cdot x_{i_1} \cdot c(A,x)_{j_1} \cdot e_{i_0}
\end{align*}
Finally, combine them and we have
\begin{align*}
       B_{1,4}^{j_1,i_1,j_1,i_0} = B_{1,4,1}^{j_1,i_1,j_1,i_0} + B_{1,4,2}^{j_1,i_1,j_1,i_0}
\end{align*}
\end{proof}
\end{lemma}

\subsection{Constructing \texorpdfstring{$d \times d$}{} matrices for \texorpdfstring{$j_1 = j_0$}{}}

The purpose of the following lemma is to let $i_0$ and $i_1$ disappear.
\begin{lemma}For $j_0,j_1 \in [n]$, a list of $d \times d$ matrices can be expressed as the following sense,
\begin{itemize}
\item {\bf Part 1.}
\begin{align*}
     B_{1,1,1}^{j_1,*,j_1,*} = f(A,x)_{j_1}^2 \cdot  f_c(A,x)^2 \cdot I_d 
\end{align*}
\item {\bf Part 2.}
\begin{align*}
     B_{1,1,2}^{j_1,*,j_1,*} = f(A,x)_{j_1}^2 \cdot c(A,x)_{j_1} \cdot  f_c(A,x) \cdot I_d 
\end{align*}
\item {\bf Part 3.}
\begin{align*}
     B_{1,1,3}^{j_1,*,j_1,*}  = f(A,x)_{j_1}^2 \cdot f_c(A,x)^2 \cdot  {\bf 1}_d  \cdot  ( (A_{j_1,*}) \circ x^\top )
\end{align*}
\item {\bf Part 4.}
\begin{align*}
    B_{1,1,4}^{j_1,*,j_1,*}  =  -  f(A,x)_{j_1}^2 \cdot \langle c(A,x), f(A,x) \rangle^2 \cdot {\bf 1}_d  \cdot  h(A,x)^\top  
\end{align*}
\item {\bf Part 5.}
\begin{align*}
    B_{1,1,5}^{j_1,*,j_1,*}  = & ~   f(A,x)_{j_1}^2 \cdot  (-f_2(A,x) + f(A,x)_{j_1})  \cdot f_c(A,x) \cdot {\bf 1}_{d} \cdot h(A,x)^\top
\end{align*}
\item {\bf Part 6.}
\begin{align*}
    B_{1,1,6}^{j_1,*,j_1,*}  = & ~   f(A,x)_{j_1}^2 \cdot (-f_c(A,x) + f(A,x)_{j_1}) \cdot f_c(A,x) \cdot {\bf 1}_d \cdot h(A,x)^\top
\end{align*}
\item {\bf Part 7.}
\begin{align*}
     B_{1,1,7}^{j_1,*,j_1,*}  =   f(A,x)_{j_1}^2 \cdot f_c(A,x) \cdot  \underbrace{{\bf 1}_d}_{d \times 1} \cdot p_{j_1}(A,x)^\top 
\end{align*}
\item {\bf Part 8.}
\begin{align*}
     B_{1,2,1}^{j_1,*,j_1,*}  =    f(A,x)_{j_1}^2  \cdot f_c(A,x) \cdot \underbrace{ x }_{d \times 1} \cdot \underbrace{ c_g(A,x)^\top }_{1 \times d}
\end{align*}
\item {\bf Part 9.}
\begin{align*}
     B_{1,2,2}^{j_1,*,j_1,*}  =   - f(A,x)_{j_1}  \cdot f_c(A,x) \cdot \underbrace{ x }_{d \times 1} \cdot \underbrace{ c_g(A,x)^\top }_{1 \times d}
\end{align*}
\item {\bf Part 10.}
\begin{align*}
    B_{1,3,1}^{j_1,*,j_1,*}  = f(A,x)_{j_1}^2  \cdot f_2(A,x) \cdot   x \cdot c_g(A,x)^\top  
\end{align*}
\item {\bf Part 11.}
\begin{align*}
    B_{1,3,2}^{j_1,*,j_1,*}  =   -   f(A,x)_{j_1}^3 \cdot   x   \cdot c_g(A,x)^{\top} 
\end{align*}
\item {\bf Part 12.}
\begin{align*}
    B_{1,4,1}^{j_1,*,j_1,*}  = f  (A,x)_{j_1}^2 \cdot  f_c(A,x) \cdot  x  \cdot  c_g(A,x)^{\top} 
\end{align*}
\item {\bf Part 13.}
\begin{align*}
   B_{1,4,2}^{j_1,*,j_1,*}  =  -   f(A,x)_{j_1}^2  \cdot c(A,x)_{j_1} 
 \cdot   x  \cdot c_g(A,x)^{\top}
\end{align*}

\end{itemize}
\begin{proof}
{\bf Proof of Part 1.}
    We have
    \begin{align*}
        B_{1,1,1}^{j_1,i_1,j_1,i_1} = f(A,x)_{j_1}^2 \cdot   \langle c(A,x), f(A,x) \rangle^2 \cdot \underbrace{ e_{i_1}^\top }_{1 \times d}  \cdot \underbrace{ e_{i_1} }_{d \times 1} \\
        B_{1,1,1}^{j_1,i_1,j_1,i_0} = f(A,x)_{j_1}^2 \cdot  \langle c(A,x), f(A,x) \rangle^2 \cdot \underbrace{ e_{i_1}^\top }_{1 \times d} \cdot \underbrace{ e_{i_0} }_{d \times 1}
    \end{align*}

    From the above two equations, we can tell that $B_{1,1,1}^{j_1,*,j_1,*} \in \R^{d \times d}$ is a matrix that only diagonal has entries and off-diagonal are all zeros.
    
    Then we have $B_{1,1,1}^{j_1,*,j_1,*} \in \R^{d \times d}$ can be written as the rescaling of a diagonal matrix,
    \begin{align*}
     B_{1,1,1}^{j_1,*,j_1,*} & ~ = f(A,x)_{j_1}^2 \cdot  \langle c(A,x), f(A,x) \rangle^2 \cdot \underbrace{ I_d }_{d \times d} \\
     & ~ = f(A,x)_{j_1}^2 \cdot  f_c(A,x)^2 \cdot I_d 
\end{align*}
    where the last step is follows from the Definitions~\ref{def:f_c}.

{\bf Proof of Part 2.}
    We have
    \begin{align*}
        B_{1,1,2}^{j_1,i_1,j_1,i_1} = f(A,x)_{j_1}^2 \cdot c(A,x)_{j_1} \cdot \langle c(A,x), f(A,x) \rangle \cdot \underbrace{ e_{i_1}^\top }_{1 \times d} \cdot \underbrace{ e_{i_1} }_{d \times 1} \\
        B_{1,1,2}^{j_1,i_1,j_1,i_0} = f(A,x)_{j_1}^2 \cdot c(A,x)_{j_1} \cdot  \langle c(A,x), f(A,x) \rangle \cdot \underbrace{ e_{i_1}^\top }_{1 \times d} \cdot \underbrace{ e_{i_0} }_{d \times 1 }
    \end{align*}
     From the above two equations, we can tell that $B_{1,1,2}^{j_1,*,j_1,*} \in \R^{d \times d}$ is a matrix that only diagonal has entries and off-diagonal are all zeros.
    
    Then we have $B_{1,1,2}^{j_1,*,j_1,*} \in \R^{d \times d}$ can be written as the rescaling of a diagonal matrix,
\begin{align*}
     B_{1,1,2}^{j_1,*,j_1,*} & ~ = f(A,x)_{j_1}^2 \cdot c(A,x)_{j_1} \cdot  \langle c(A,x), f(A,x) \rangle \cdot I_d \\
     & ~ = f(A,x)_{j_1}^2 \cdot c(A,x)_{j_1} \cdot  f_c(A,x) \cdot I_d
\end{align*}
    where the last step is follows from the Definitions~\ref{def:f_c}.

{\bf Proof of Part 3.}
We have for diagonal entry and off-diagonal entry can be written as follows 
    \begin{align*}
        B_{1,1,3}^{j_1,i_1,j_1,i_1} = \underbrace{ f(A,x)_{j_1}^2 \cdot \langle c(A,x), f(A,x) \rangle^2 }_{ \mathrm{scalar} } \cdot \underbrace{ ( (A_{j_1,*}) \circ x^\top  ) }_{1 \times d} \cdot \underbrace{ e_{i_1} }_{d \times 1} \\
        B_{1,1,3}^{j_1,i_1,j_1,i_0} = \underbrace{ f(A,x)_{j_1}^2 \cdot \langle c(A,x), f(A,x) \rangle^2 }_{ \mathrm{scalar} } \cdot \underbrace{ ( (A_{j_1,*}) \circ x^\top  ) }_{1 \times d}  \cdot \underbrace{ e_{i_0} }_{d \times 1}
    \end{align*}
From the above equation, we can show that matrix $B_{1,1,3}^{j_1,*,j_1,*}$ can be expressed as a rank-$1$ matrix,
\begin{align*}
     B_{1,1,3}^{j_1,*,j_1,*} & ~ = \underbrace{ f(A,x)_{j_1}^2 \cdot \langle c(A,x), f(A,x) \rangle^2 }_{ \mathrm{scalar} } \cdot \underbrace{ {\bf 1}_d }_{d \times 1} \cdot \underbrace{ ( (A_{j_1,*}) \circ x^\top ) }_{1 \times d}\\
     & ~ =  f(A,x)_{j_1}^2 \cdot f_c(A,x)^2 \cdot  {\bf 1}_d  \cdot  ( (A_{j_1,*}) \circ x^\top ) 
\end{align*}
    where the last step is follows from the Definitions~\ref{def:f_c}.

{\bf Proof of Part 4.}
We have for diagonal entry and off-diagonal entry can be written as follows
    \begin{align*}
        B_{1,1,4}^{j_1,i_1,j_1,i_1} = -  \underbrace{ f(A,x)_{j_1}^2 \cdot  \langle c(A,x), f(A,x) \rangle^2 }_{\mathrm{scalar}} \cdot \underbrace{ f(A,x)^\top  \cdot A \cdot \diag(x) }_{1 \times d }\cdot  \underbrace{ e_{i_1} }_{d \times 1} \\
        B_{1,1,4}^{j_1,i_1,j_1,i_0} = -  \underbrace{ f(A,x)_{j_1}^2 \cdot \langle c(A,x), f(A,x) \rangle^2 }_{ \mathrm{scalar} } \cdot \underbrace{ f(A,x)^\top  \cdot A \cdot \diag(x) }_{1 \times d} \cdot \underbrace{ e_{i_0} }_{d \times 1}
    \end{align*}
 From the above equation, we can show that matrix $B_{1,1,4}^{j_1,*,j_1,*}$ can be expressed as a rank-$1$ matrix,
\begin{align*}
    B_{1,1,4}^{j_1,*,j_1,*}  & ~ =  -  f(A,x)_{j_1}^2 \cdot \langle c(A,x), f(A,x) \rangle^2 \cdot \underbrace{ {\bf 1}_d }_{d \times 1} \cdot \underbrace{ f(A,x)^\top  \cdot A \cdot \diag(x) }_{1 \times d}  \\
     & ~ = -  f(A,x)_{j_1}^2 \cdot \langle c(A,x), f(A,x) \rangle^2 \cdot {\bf 1}_d  \cdot  h(A,x)^\top 
\end{align*}
    where the last step is follows from the Definitions~\ref{def:h}.

{\bf Proof of Part 5.}
We have for diagonal entry and off-diagonal entry can be written as follows
    \begin{align*}
        B_{1,1,5}^{j_1,i_1,j_1,i_1} =   f(A,x)_{j_1}^2 \cdot  (\langle -f(A,x), f(A,x) \rangle + f(A,x)_{j_1})  \cdot \langle c(A,x), f(A,x) \rangle \cdot f(A,x)^\top  \cdot A \cdot \diag(x) \cdot e_{i_1} \\
        B_{1,1,5}^{j_1,i_1,j_1,i_0} =   f(A,x)_{j_1}^2 \cdot  (\langle -f(A,x), f(A,x) \rangle + f(A,x)_{j_1})  \cdot \langle c(A,x), f(A,x) \rangle \cdot f(A,x)^\top  \cdot A \cdot \diag(x) \cdot e_{i_0}
    \end{align*}
    From the above equation, we can show that matrix $B_{1,1,5}^{j_1,*,j_1,*}$ can be expressed as a rank-$1$ matrix,
\begin{align*}
    B_{1,1,5}^{j_1,*,j_1,*}  & ~ =     f(A,x)_{j_1}^2 \cdot  (\langle -f(A,x), f(A,x) \rangle + f(A,x)_{j_1})  \cdot \langle c(A,x), f(A,x) \rangle \cdot {\bf 1}_{d} \cdot f(A,x)^\top  \cdot A \cdot \diag(x)\\
     & ~ = f(A,x)_{j_1}^2 \cdot  (-f_2(A,x) + f(A,x)_{j_1})  \cdot f_c(A,x) \cdot {\bf 1}_{d} \cdot h(A,x)^\top
\end{align*}
    where the last step is follows from the Definitions~\ref{def:h}, Definitions~\ref{def:f_c} and Definitions~\ref{def:f_2}.

{\bf Proof of Part 6.}
We have for diagonal entry and off-diagonal entry can be written as follows
    \begin{align*}
        B_{1,1,6}^{j_1,i_1,j_1,i_1} =    f(A,x)_{j_1}^2 \cdot (\langle -f(A,x), c(A,x) \rangle + f(A,x)_{j_1}) \cdot \langle c(A,x), f(A,x) \rangle \cdot f(A,x)^\top  \cdot A \cdot  \diag(x) \cdot e_{i_1}\\
        B_{1,1,6}^{j_1,i_1,j_1,i_0} =    f(A,x)_{j_1}^2 \cdot (\langle -f(A,x), c(A,x) \rangle + f(A,x)_{j_1}) \cdot \langle c(A,x), f(A,x) \rangle \cdot f(A,x)^\top  \cdot A \cdot  \diag(x) \cdot e_{i_0}
    \end{align*}
    From the above equation, we can show that matrix $B_{1,1,6}^{j_1,*,j_1,*}$ can be expressed as a rank-$1$ matrix,
\begin{align*}
    B_{1,1,6}^{j_1,*,j_1,*}  & ~ =   f(A,x)_{j_1}^2 \cdot (\langle -f(A,x), c(A,x) \rangle + f(A,x)_{j_1}) \cdot \langle c(A,x), f(A,x) \rangle \cdot {\bf 1}_d \cdot f(A,x)^\top  \cdot A \cdot  \diag(x)\\
     & ~ = f(A,x)_{j_1}^2 \cdot (-f_c(A,x) + f(A,x)_{j_1}) \cdot f_c(A,x) \cdot {\bf 1}_d \cdot h(A,x)^\top
\end{align*}
    where the last step is follows from the Definitions~\ref{def:h}, Definitions~\ref{def:f_c} and Definitions~\ref{def:f_2}.
    
{\bf Proof of Part 7.}
We have for diagonal entry and off-diagonal entry can be written as follows
    \begin{align*}
         B_{1,1,7}^{j_1,i_1,j_1,i_1} = f(A,x)_{j_1}^2 \cdot \langle c(A,x), f(A,x) \rangle \cdot   ((e_{j_1}^\top - f(A,x)^\top) \circ q(A,x)^\top) \cdot A \cdot  \diag(x)  \cdot e_{i_1}\\
         B_{1,1,7}^{j_1,i_1,j_1,i_0} = f(A,x)_{j_1}^2 \cdot \langle c(A,x), f(A,x) \rangle \cdot   ((e_{j_1}^\top - f(A,x)^\top) \circ q(A,x)^\top) \cdot A \cdot  \diag(x)  \cdot e_{i_0}
    \end{align*}
    From the above equation, we can show that matrix $B_{1,1,7}^{j_1,*,j_1,*}$ can be expressed as a rank-$1$ matrix,
\begin{align*}
     B_{1,1,7}^{j_1,*,j_1,*}  & ~ =   f(A,x)_{j_1}^2 \cdot \langle c(A,x), f(A,x) \rangle \cdot  \underbrace{{\bf 1}_d}_{d \times 1} \cdot \underbrace{\underbrace
     {((e_{j_1}^\top - f(A,x)^\top) \circ q(A,x)^\top)}_{1 \times n} \cdot \underbrace{A}_{n \times d} \cdot  \underbrace{\diag(x)}_{d \times d}}_{1 \times d}  \\
     & ~ =f(A,x)_{j_1}^2 \cdot f_c(A,x) \cdot {\bf 1}_d \cdot p_{j_1}(A,x)^\top
\end{align*}
    where the last step is follows from the Definitions~\ref{def:f_c} and Definitions~\ref{def:p}.
    
{\bf Proof of Part 8.}
We have for diagonal entry and off-diagonal entry can be written as follows
    \begin{align*}
         B_{1,2,1}^{j_1,i_1,j_1,i_1} =   f(A,x)_{j_1}^2  \cdot \langle c(A,x), f(A,x) \rangle \cdot e_{i_1}^\top \cdot  x \cdot c_g(A,x)^\top  \cdot e_{i_1}\\
         B_{1,2,1}^{j_1,i_1,j_1,i_0} = f(A,x)_{j_1}^2  \cdot \langle c(A,x), f(A,x) \rangle \cdot e_{i_1}^\top \cdot  x \cdot c_g(A,x)^\top \cdot e_{i_0}
    \end{align*}
    From the above equation, we can show that matrix $B_{1,2,1}^{j_1,*,j_1,*}$ can be expressed as a rank-$1$ matrix,
\begin{align*}
     B_{1,2,1}^{j_1,*,j_1,*}  & ~ =   f(A,x)_{j_1}^2  \cdot \langle c(A,x), f(A,x) \rangle \cdot \underbrace{ x }_{d \times 1} \cdot \underbrace{ c_g(A,x)^\top }_{1 \times d} \\
     & ~ =f(A,x)_{j_1}^2  \cdot f_c(A,x) \cdot x \cdot  c_g(A,x)^\top 
\end{align*}
    where the last step is follows from the Definitions~\ref{def:f_c}.

{\bf Proof of Part 9.}
We have for diagonal entry and off-diagonal entry can be written as follows
    \begin{align*}
         B_{1,2,2}^{j_1,i_1,j_1,i_1} = -  f(A,x)_{j_1}  \cdot \langle c(A,x), f(A,x) \rangle \cdot e_{i_1}^\top \cdot  x \cdot c_g(A,x)^\top  \cdot e_{i_1}\\
         B_{1,2,2}^{j_1,i_1,j_1,i_0} = -f(A,x)_{j_1}  \cdot \langle c(A,x), f(A,x) \rangle \cdot e_{i_1}^\top \cdot  x \cdot c_g(A,x)^\top \cdot e_{i_0}
    \end{align*}
        From the above equation, we can show that matrix $B_{1,2,2}^{j_1,*,j_1,*}$ can be expressed as a rank-$1$ matrix,
\begin{align*}
     B_{1,2,2}^{j_1,*,j_1,*}  & ~ =   - f(A,x)_{j_1}  \cdot \langle c(A,x), f(A,x) \rangle \cdot \underbrace{ x }_{d \times 1} \cdot \underbrace{ c_g(A,x)^\top }_{1 \times d}\\
     & ~ =- f(A,x)_{j_1}  \cdot f_c(A,x) \cdot  x  \cdot  c_g(A,x)^\top 
\end{align*}
    where the last step is follows from the Definitions~\ref{def:f_c}.

{\bf Proof of Part 10.}
We have for diagonal entry and off-diagonal entry can be written as follows
    \begin{align*}
         B_{1,3,1}^{j_1,i_1,j_1,i_1} =    f(A,x)_{j_1}^2  \cdot \langle  f(A,x), f(A,x) \rangle \cdot  e_{i_1}^\top \cdot  x \cdot c_g(A,x)^\top  \cdot e_{i_1}\\
         B_{1,3,1}^{j_1,i_1,j_1,i_0} =    f(A,x)_{j_1}^2  \cdot \langle  f(A,x), f(A,x) \rangle \cdot  e_{i_1}^\top \cdot  x \cdot c_g(A,x)^\top  \cdot e_{i_0}
    \end{align*}
            From the above equation, we can show that matrix $B_{1,3,1}^{j_1,*,j_1,*}$ can be expressed as a rank-$1$ matrix,
\begin{align*}
    B_{1,3,1}^{j_1,*,j_1,*}  & ~ =     f(A,x)_{j_1}^2  \cdot \langle  f(A,x), f(A,x) \rangle \cdot   x \cdot c_g(A,x)^\top  \\
     & ~ = f(A,x)_{j_1}^2  \cdot f_2(A,x) \cdot   x \cdot c_g(A,x)^\top
\end{align*}
    where the last step is follows from the Definitions~\ref{def:f_2}.

{\bf Proof of Part 11.}
We have for diagonal entry and off-diagonal entry can be written as follows
    \begin{align*}
         B_{1,3,2}^{j_1,i_1,j_1,i_1} =  -   f(A,x)_{j_1}^3 \cdot  e_{i_1}^\top \cdot  x   \cdot c_g(A,x)^{\top} \cdot e_{i_1}\\
         B_{1,3,2}^{j_1,i_1,j_1,i_0} =  -   f(A,x)_{j_1}^3 \cdot  e_{i_1}^\top \cdot  x   \cdot c_g(A,x)^{\top} \cdot e_{i_1}
    \end{align*}
            From the above equation, we can show that matrix $B_{1,3,2}^{j_1,*,j_1,*}$ can be expressed as a rank-$1$ matrix,
\begin{align*}
    B_{1,3,2}^{j_1,i_1,j_1,*}  =   -   f(A,x)_{j_1}^3 \cdot   x   \cdot c_g(A,x)^{\top} 
\end{align*}
{\bf Proof of Part 12.}
We have for diagonal entry and off-diagonal entry can be written as follows
    \begin{align*}
         B_{1,4,1}^{j_1,i_1,j_1,i_1} =   f(A,x)_{j_1}^2 \cdot  \langle f(A,x), c(A,x) \rangle  \cdot e_{i_1}^\top \cdot  x  \cdot  c_g(A,x)^{\top} \cdot  e_{i_1}\\
         B_{1,4,1}^{j_1,i_1,j_1,i_0} =  f(A,x)_{j_1}^2 \cdot  \langle f(A,x), c(A,x) \rangle  \cdot e_{i_1}^\top \cdot  x  \cdot  c_g(A,x)^{\top} \cdot  e_{i_0}
    \end{align*}
            From the above equation, we can show that matrix $B_{1,4,1}^{j_1,*,j_1,*}$ can be expressed as a rank-$1$ matrix,
\begin{align*}
    B_{1,4,1}^{j_1,i_1,j_1,*}  & ~ =   f(A,x)_{j_1}^2 \cdot  \langle f(A,x), c(A,x) \rangle  \cdot  x  \cdot  c_g(A,x)^{\top} \\
     & ~ = f(A,x)_{j_1}^2 \cdot  f_c(A,x) \cdot  x  \cdot  c_g(A,x)^{\top}
\end{align*}
    where the last step is follows from the Definitions~\ref{def:f_c}.

{\bf Proof of Part 13.}
We have for diagonal entry and off-diagonal entry can be written as follows
    \begin{align*}
          B_{1,4,2}^{j_1,i_1,j_1,i_1} = -   f(A,x)_{j_1}^2  \cdot c(A,x)_{j_1} 
 \cdot e_{i_1}^\top \cdot  x  \cdot c_g(A,x)^{\top} \cdot e_{i_1}\\
          B_{1,4,2}^{j_1,i_1,j_1,i_0} =  -   f(A,x)_{j_1}^2  \cdot c(A,x)_{j_1} 
 \cdot e_{i_1}^\top \cdot  x  \cdot c_g(A,x)^{\top} \cdot e_{i_0}
    \end{align*}
            From the above equation, we can show that matrix $B_{1,4,2}^{j_1,*,j_1,*}$ can be expressed as a rank-$1$ matrix,
\begin{align*}
   B_{1,4,2}^{j_1,i_1,j_1,*}  =   -   f(A,x)_{j_1}^2  \cdot c(A,x)_{j_1} 
 \cdot   x  \cdot c_g(A,x)^{\top}
\end{align*}
\end{proof}

\end{lemma}

\subsection{Case \texorpdfstring{$j_1 \neq j_0, i_1 = i_0$}{}}
\begin{lemma}

For $j_1 \neq j_0$ and $i_0 = i_1$. If the following conditions hold
    \begin{itemize}
     \item Let $u(A,x) \in \R^n$ be defined as Definition~\ref{def:u}
    \item Let $\alpha(A,x) \in \R$ be defined as Definition~\ref{def:alpha}
     \item Let $f(A,x) \in \R^n$ be defined as Definition~\ref{def:f}
    \item Let $c(A,x) \in \R^n$ be defined as Definition~\ref{def:c}
    \item Let $g(A,x) \in \R^d$ be defined as Definition~\ref{def:g} 
    \item Let $q(A,x) = c(A,x) + f(A,x) \in \R^n$
    \item Let $c_g(A,x) \in \R^d$ be defined as Definition~\ref{def:c_g}.
    \item Let $L_g(A,x) \in \R$ be defined as Definition~\ref{def:l_g}
    \item Let $v \in \R^n$ be a vector 
    \item Let $B_1^{j_1,i_1,j_0,i_0}$ be defined as Definition~\ref{def:b_1}
    \end{itemize}
    Then, For $j_0,j_1 \in [n], i_0,i_1 \in [d]$, we have 
    \begin{itemize}
\item {\bf Part 1.} For $B_{1,1}^{j_1,i_1,j_0,i_1}$, we have 
\begin{align*}
 B_{1,1}^{j_1,i_1,j_0,i_1}  = & ~ \frac{\d}{\d A_{j_1,i_1}} (- c_g(A,x)^{\top} ) \cdot  f(A,x)_{j_0} \cdot \langle c(A,x), f(A,x) \rangle \cdot e_{i_1} \\
 = & ~ B_{1,1,1}^{j_1,i_1,j_0,i_1} + B_{1,1,2}^{j_1,i_1,j_0,i_1} + B_{1,1,3}^{j_1,i_1,j_0,i_1} + B_{1,1,4}^{j_1,i_1,j_0,i_1} \\ 
 & ~ + B_{1,1,5}^{j_1,i_1,j_0,i_1} + B_{1,1,6}^{j_1,i_1,j_0,i_1} + B_{1,1,7}^{j_1,i_1,j_0,i_1}
\end{align*} 
\item {\bf Part 2.} For $B_{1,2}^{j_1,i_1,j_0,i_1}$, we have 
\begin{align*}
  B_{1,2}^{j_1,i_1,j_0,i_1} = & ~ - c_g(A,x)^{\top} \cdot \frac{\d}{\d A_{j_1,i_1}} ( f(A,x)_{j_0} )  \cdot \langle c(A,x), f(A,x) \rangle \cdot e_{i_1}\\
    = & ~  B_{1,2,1}^{j_1,i_1,j_0,i_1}  
\end{align*} 
\item {\bf Part 3.} For $B_{1,3}^{j_1,i_1,j_0,i_1}$, we have 
\begin{align*}
  B_{1,3}^{j_1,i_1,j_0,i_1} = & ~ -  c_g(A,x)^{\top} \cdot f(A,x)_{j_0}  \cdot \d \langle \frac{c(A,x)}{\d A_{j_1,i_1}} , f(A,x) \rangle\cdot e_{i_1}  \\
     = & ~ B_{1,3,1}^{j_1,i_1,j_0,i_1} + B_{1,3,2}^{j_1,i_1,j_0,i_1}
\end{align*} 
\item {\bf Part 4.} For $B_{1,4}^{j_1,i_1,j_0,i_1}$, we have 
\begin{align*}
 B_{1,4}^{j_1,i_1,j_1,i_1} = & ~ -  c_g(A,x)^{\top} \cdot f(A,x)_{j_0}  \cdot \d \langle c(A,x), \frac{f(A,x)}{\d A_{j_1,i_1}}  \rangle\cdot e_{i_1} \\
     = & ~ B_{1,4,1}^{j_1,i_1,j_0,i_1} + B_{1,4,2}^{j_1,i_1,j_0,i_1}
\end{align*} 
\end{itemize}
\begin{proof}
    {\bf Proof of Part 1.}
    \begin{align*}
    B_{1,1,1}^{j_1,i_1,j_0,i_1} : = & ~  f(A,x)_{j_1} \cdot f(A,x)_{j_0} \cdot  e_{i_1}^\top \cdot \langle c(A,x), f(A,x) \rangle^2 \cdot e_{i_1}\\
    B_{1,1,2}^{j_1,i_1,j_0,i_1} : = & ~  f(A,x)_{j_1} \cdot f(A,x)_{j_0} \cdot c(A,x)_{j_1}\cdot e_{i_1}^\top \cdot  \langle c(A,x), f(A,x) \rangle \cdot e_{i_1}\\
    B_{1,1,3}^{j_1,i_1,j_0,i_1} : = & ~ f(A,x)_{j_1} \cdot f(A,x)_{j_0} \cdot \langle c(A,x), f(A,x) \rangle^2 \cdot ( (A_{j_1,*}) \circ x^\top  )   \cdot e_{i_1}\\
    B_{1,1,4}^{j_1,i_1,j_0,i_1} : = & ~ -  f(A,x)_{j_1} \cdot f(A,x)_{j_0} \cdot f(A,x)^\top  \cdot A \cdot \diag(x) \cdot   (\langle c(A,x), f(A,x) \rangle)^2 \cdot  e_{i_1}\\
    B_{1,1,5}^{j_1,i_1,j_0,i_1} : = & ~   f(A,x)_{j_1} \cdot f(A,x)_{j_0} \cdot f(A,x)^\top  \cdot A \cdot \diag(x) \cdot (\langle -f(A,x), f(A,x) \rangle + f(A,x)_{j_1})  \\
        & ~ \cdot \langle c(A,x), f(A,x) \rangle \cdot e_{i_1} \\
    B_{1,1,6}^{j_1,i_1,j_0,i_1} : = & ~    f(A,x)_{j_1} \cdot f(A,x)_{j_0} \cdot f(A,x)^\top  \cdot A \cdot  \diag(x) \cdot(\langle -f(A,x), c(A,x) \rangle + f(A,x)_{j_1}) \\
        & ~ \cdot \langle c(A,x), f(A,x) \rangle \cdot e_{i_1} \\
    B_{1,1,7}^{j_1,i_1,j_0,i_1} : = & ~ f(A,x)_{j_1} \cdot f(A,x)_{j_0} \cdot ((e_{j_1}^\top - f(A,x)^\top) \circ q(A,x)^\top) \cdot A \cdot  \diag(x)  \cdot \langle c(A,x), f(A,x) \rangle \cdot e_{i_1}
\end{align*}
Finally, combine them and we have
\begin{align*}
       B_{1,1}^{j_1,i_1,j_0,i_1} = B_{1,1,1}^{j_1,i_1,j_0,i_1} + B_{1,1,2}^{j_1,i_1,j_0,i_1} + B_{1,1,3}^{j_1,i_1,j_0,i_1} + B_{1,1,4}^{j_1,i_1,j_0,i_1} + B_{1,1,5}^{j_1,i_1,j_0,i_1} + B_{1,1,6}^{j_1,i_1,j_0,i_1} + B_{1,1,7}^{j_1,i_1,j_0,i_1}
\end{align*}
{\bf Proof of Part 2.}
    \begin{align*}
    B_{1,2,1}^{j_1,i_1,j_0,i_1} : = & ~   c_g(A,x)^{\top} \cdot f(A,x)_{j_1} \cdot  f(A,x)_{j_0} \cdot x_{i_1} \cdot \langle c(A,x), f(A,x) \rangle \cdot e_{i_1} 
\end{align*}
Finally, combine them and we have
\begin{align*}
       B_{1,2}^{j_1,i_1,j_0,i_1} = B_{1,2,1}^{j_1,i_1,j_0,i_1} 
\end{align*}
{\bf Proof of Part 3.} 
    \begin{align*}
    B_{1,3,1}^{j_1,i_1,j_0,i_1} : = & ~    c_g(A,x)^{\top} \cdot f(A,x)_{j_1} \cdot  f(A,x)_{j_0}  \cdot x_{i_1} \cdot \langle  f(A,x), f(A,x) \rangle \cdot e_{i_1}\\
    B_{1,3,2}^{j_1,i_1,j_0,i_1} : = & ~  -  c_g(A,x)^{\top} \cdot f(A,x)_{j_1}^2 \cdot  f(A,x)_{j_0} \cdot x_{i_1}  \cdot e_{i_1}
\end{align*}
Finally, combine them and we have
\begin{align*}
       B_{1,3}^{j_1,i_1,j_0,i_1} = B_{1,3,1}^{j_1,i_1,j_0,i_1} + B_{1,3,2}^{j_1,i_1,j_0,i_1}
\end{align*}
{\bf Proof of Part 4.}
    \begin{align*}
    B_{1,4,1}^{j_1,i_1,j_0,i_1} : = & ~     c_g(A,x)^{\top} \cdot f(A,x)_{j_1} \cdot  f(A,x)_{j_0} \cdot x_{i_1} \cdot \langle  f(A,x), c(A,x) \rangle  \cdot e_{i_1}\\
    B_{1,4,2}^{j_1,i_1,j_0,i_1} : = & ~  -  c_g(A,x)^{\top} \cdot f(A,x)_{j_1} \cdot  f(A,x)_{j_0} \cdot x_{i_1} \cdot c(A,x)_{j_1} \cdot e_{i_1}
\end{align*}
Finally, combine them and we have
\begin{align*}
       B_{1,4}^{j_1,i_1,j_0,i_1} = B_{1,4,1}^{j_1,i_1,j_0,i_1} + B_{1,4,2}^{j_1,i_1,j_0,i_1}
\end{align*}
\end{proof}
\end{lemma}

\subsection{Case \texorpdfstring{$j_1 \neq j_0, i_1 \neq i_0$}{}}
\begin{lemma}
For $j_1 \neq j_0$ and $i_0 \neq i_1$. If the following conditions hold
    \begin{itemize}
     \item Let $u(A,x) \in \R^n$ be defined as Definition~\ref{def:u}
    \item Let $\alpha(A,x) \in \R$ be defined as Definition~\ref{def:alpha}
     \item Let $f(A,x) \in \R^n$ be defined as Definition~\ref{def:f}
    \item Let $c(A,x) \in \R^n$ be defined as Definition~\ref{def:c}
    \item Let $g(A,x) \in \R^d$ be defined as Definition~\ref{def:g} 
    \item Let $q(A,x) = c(A,x) + f(A,x) \in \R^n$
    \item Let $c_g(A,x) \in \R^d$ be defined as Definition~\ref{def:c_g}.
    \item Let $L_g(A,x) \in \R$ be defined as Definition~\ref{def:l_g}
    \item Let $v \in \R^n$ be a vector 
    \item Let $B_1^{j_1,i_1,j_0,i_0}$ be defined as Definition~\ref{def:b_1}
    \end{itemize}
    Then, For $j_0,j_1 \in [n], i_0,i_1 \in [d]$, we have 
    \begin{itemize}
\item {\bf Part 1.} For $B_{1,1}^{j_1,i_1,j_0,i_0}$, we have 
\begin{align*}
 B_{1,1}^{j_1,i_1,j_0,i_0}  = & ~ \frac{\d}{\d A_{j_1,i_1}} (- c_g(A,x)^{\top} ) \cdot  f(A,x)_{j_0} \cdot \langle c(A,x), f(A,x) \rangle \cdot e_{i_0} \\
 = & ~ B_{1,1,1}^{j_1,i_1,j_0,i_0} + B_{1,1,2}^{j_1,i_1,j_0,i_0} + B_{1,1,3}^{j_1,i_1,j_0,i_01} + B_{1,1,4}^{j_1,i_1,j_0,i_0} \\
 & ~ + B_{1,1,5}^{j_1,i_1,j_0,i_0} + B_{1,1,6}^{j_1,i_1,j_0,i_0} + B_{1,1,7}^{j_1,i_1,j_0,i_0}
\end{align*} 
\item {\bf Part 2.} For $B_{1,2}^{j_1,i_1,j_0,i_0}$, we have 
\begin{align*}
  B_{1,2}^{j_1,i_1,j_0,i_0} = & ~ - c_g(A,x)^{\top} \cdot \frac{\d}{\d A_{j_1,i_1}} ( f(A,x)_{j_0} )  \cdot \langle c(A,x), f(A,x) \rangle \cdot e_{i_0}\\
    = & ~  B_{1,2,1}^{j_1,i_1,j_0,i_0}  
\end{align*} 
\item {\bf Part 3.} For $B_{1,3}^{j_1,i_1,j_0,i_0}$, we have 
\begin{align*}
  B_{1,3}^{j_1,i_1,j_0,i_0} = & ~ -  c_g(A,x)^{\top} \cdot f(A,x)_{j_0}  \cdot \d \langle \frac{c(A,x)}{\d A_{j_1,i_1}} , f(A,x) \rangle\cdot e_{i_0}  \\
     = & ~ B_{1,3,1}^{j_1,i_1,j_0,i_0} + B_{1,3,2}^{j_1,i_1,j_0,i_0}
\end{align*} 
\item {\bf Part 4.} For $B_{1,4}^{j_1,i_1,j_0,i_0}$, we have 
\begin{align*}
 B_{1,4}^{j_1,i_1,j_1,i_0} = & ~ -  c_g(A,x)^{\top} \cdot f(A,x)_{j_0}  \cdot \d \langle c(A,x), \frac{f(A,x)}{\d A_{j_1,i_1}}  \rangle\cdot e_{i_0} \\
     = & ~ B_{1,4,1}^{j_1,i_1,j_0,i_0} + B_{1,4,2}^{j_1,i_1,j_0,i_0}
\end{align*} 
\end{itemize}
\begin{proof}
    {\bf Proof of Part 1.}
    \begin{align*}
    B_{1,1,1}^{j_1,i_1,j_0,i_0} : = & ~  f(A,x)_{j_1} \cdot f(A,x)_{j_0} \cdot  e_{i_1}^\top \cdot \langle c(A,x), f(A,x) \rangle^2 \cdot e_{i_0}\\
    B_{1,1,2}^{j_1,i_1,j_0,i_0} : = & ~  f(A,x)_{j_1} \cdot f(A,x)_{j_0} \cdot c(A,x)_{j_1}\cdot e_{i_1}^\top \cdot  \langle c(A,x), f(A,x) \rangle \cdot e_{i_0}\\
    B_{1,1,3}^{j_1,i_1,j_0,i_0} : = & ~ f(A,x)_{j_1} \cdot f(A,x)_{j_0} \cdot \langle c(A,x), f(A,x) \rangle^2 \cdot ( (A_{j_1,*}) \circ x^\top  )   \cdot e_{i_0}\\
    B_{1,1,4}^{j_1,i_1,j_0,i_0} : = & ~ -  f(A,x)_{j_1} \cdot f(A,x)_{j_0} \cdot f(A,x)^\top  \cdot A \cdot \diag(x) \cdot   (\langle c(A,x), f(A,x) \rangle)^2 \cdot  e_{i_0}\\
    B_{1,1,5}^{j_1,i_1,j_0,i_0} : = & ~   f(A,x)_{j_1} \cdot f(A,x)_{j_0} \cdot f(A,x)^\top  \cdot A \cdot \diag(x) \cdot (\langle -f(A,x), f(A,x) \rangle + f(A,x)_{j_1})   \\
        & ~\cdot \langle c(A,x), f(A,x) \rangle \cdot e_{i_0} \\
    B_{1,1,6}^{j_1,i_1,j_0,i_0} : = & ~    f(A,x)_{j_1} \cdot f(A,x)_{j_0} \cdot f(A,x)^\top  \cdot A \cdot  \diag(x) \cdot(\langle -f(A,x), c(A,x) \rangle + f(A,x)_{j_1})  \\
        & ~\cdot \langle c(A,x), f(A,x) \rangle \cdot e_{i_0} \\
    B_{1,1,7}^{j_1,i_1,j_0,i_0} : = & ~ f(A,x)_{j_1} \cdot f(A,x)_{j_0} \cdot ((e_{j_1}^\top - f(A,x)^\top) \circ q(A,x)^\top) \cdot A \cdot  \diag(x)  \cdot \langle c(A,x), f(A,x) \rangle \cdot e_{i_0}
\end{align*}
Finally, combine them and we have
\begin{align*}
       B_{1,1}^{j_1,i_1,j_0,i_0} = B_{1,1,1}^{j_1,i_1,j_0,i_0} + B_{1,1,2}^{j_1,i_1,j_0,i_0} + B_{1,1,3}^{j_1,i_1,j_0,i_0} + B_{1,1,4}^{j_1,i_1,j_0,i_0} + B_{1,1,5}^{j_1,i_1,j_0,i_0} + B_{1,1,6}^{j_1,i_1,j_0,i_0} + B_{1,1,7}^{j_1,i_1,j_0,i_0}
\end{align*}
{\bf Proof of Part 2.}
    \begin{align*}
    B_{1,2,1}^{j_1,i_1,j_0,i_0} : = & ~   c_g(A,x)^{\top} \cdot f(A,x)_{j_1} \cdot  f(A,x)_{j_0} \cdot x_{i_1} \cdot \langle c(A,x), f(A,x) \rangle \cdot e_{i_0} 
\end{align*}
Finally, combine them and we have
\begin{align*}
       B_{1,2}^{j_1,i_1,j_0,i_0} = B_{1,2,1}^{j_1,i_1,j_0,i_0} 
\end{align*}
{\bf Proof of Part 3.} 
    \begin{align*}
    B_{1,3,1}^{j_1,i_1,j_0,i_0} : = & ~    c_g(A,x)^{\top} \cdot f(A,x)_{j_1} \cdot  f(A,x)_{j_0}  \cdot x_{i_1} \cdot \langle  f(A,x), f(A,x) \rangle \cdot e_{i_0}\\
    B_{1,3,2}^{j_1,i_1,j_0,i_0} : = & ~  -  c_g(A,x)^{\top} \cdot f(A,x)_{j_1}^2 \cdot  f(A,x)_{j_0} \cdot x_{i_1}  \cdot e_{i_0}
\end{align*}
Finally, combine them and we have
\begin{align*}
       B_{1,3}^{j_1,i_1,j_0,i_0} = B_{1,3,1}^{j_1,i_1,j_0,i_0} + B_{1,3,2}^{j_1,i_1,j_0,i_0}
\end{align*}
{\bf Proof of Part 4.}
    \begin{align*}
    B_{1,4,1}^{j_1,i_1,j_0,i_0} : = & ~     c_g(A,x)^{\top} \cdot f(A,x)_{j_1} \cdot  f(A,x)_{j_0} \cdot x_{i_1} \cdot \langle  f(A,x), c(A,x) \rangle  \cdot e_{i_0}\\
    B_{1,4,2}^{j_1,i_1,j_0,i_0} : = & ~  -  c_g(A,x)^{\top} \cdot f(A,x)_{j_1} \cdot  f(A,x)_{j_0} \cdot x_{i_1} \cdot c(A,x)_{j_1} \cdot e_{i_0}
\end{align*}
Finally, combine them and we have
\begin{align*}
       B_{1,4}^{j_1,i_1,j_0,i_0} = B_{1,4,1}^{j_1,i_1,j_0,i_0} + B_{1,4,2}^{j_1,i_1,j_0,i_0}
\end{align*}
\end{proof}
\end{lemma}
\subsection{Constructing \texorpdfstring{$d \times d$}{} matrices for \texorpdfstring{$j_1 \neq j_0$}{}}
The purpose of the following lemma is to let $i_0$ and $i_1$ disappear.
\begin{lemma}\label{lem:b_1_j1_j0}
For $j_0,j_1 \in [n]$, a list of $d \times d$ matrices can be expressed as the following sense,
\begin{itemize}
\item {\bf Part 1.}
\begin{align*}
     B_{1,1,1}^{j_1,*,j_0,*} =  f(A,x)_{j_1} \cdot f(A,x)_{j_0} \cdot  f_c(A,x)^2 \cdot I_d 
\end{align*}
\item {\bf Part 2.}
\begin{align*}
     B_{1,1,2}^{j_1,*,j_0,*} =  f(A,x)_{j_1} \cdot f(A,x)_{j_0} \cdot c(A,x)_{j_1}\cdot  f_c(A,x) \cdot  I_d
\end{align*}
\item {\bf Part 3.}
\begin{align*}
     B_{1,1,3}^{j_1,*,j_0,*}  = f(A,x)_{j_1} \cdot f(A,x)_{j_0} \cdot f_c(A,x)^2 \cdot {\bf 1}_d \cdot ( (A_{j_1,*}) \circ x^\top  )
\end{align*}
\item {\bf Part 4.}
\begin{align*}
    B_{1,1,4}^{j_1,*,j_0,*}  =  -  f(A,x)_{j_1} \cdot f(A,x)_{j_0} \cdot f_c(A,x)^2 \cdot {\bf 1}_d  \cdot h(A,x)^\top
\end{align*}
\item {\bf Part 5.}
\begin{align*}
    B_{1,1,5}^{j_1,*,j_0,*}  = & ~ f(A,x)_{j_1} \cdot f(A,x)_{j_0} \cdot f_c(A,x)  \cdot (-f_2(A,x) + f(A,x)_{j_1})   \cdot {\bf 1}_d  \cdot h(A,x)^\top
\end{align*}
\item {\bf Part 6.}
\begin{align*}
    B_{1,1,6}^{j_1,*,j_0,*}  = & ~   f(A,x)_{j_1} \cdot f(A,x)_{j_0}  \cdot(-f_c(A,x) + f(A,x)_{j_1}) \cdot f_c(A,x)   \cdot {\bf 1}_d  \cdot h(A,x)^\top
\end{align*}
\item {\bf Part 7.}
\begin{align*}
     B_{1,1,7}^{j_1,*,j_0,*}  =  f(A,x)_{j_1} \cdot f(A,x)_{j_0} \cdot f_c(A,x) \cdot {\bf 1}_d  \cdot p_{j_1}(A,x)^\top 
\end{align*}
\item {\bf Part 8.}
\begin{align*}
     B_{1,2,1}^{j_1,*,j_0,*}  =    f(A,x)_{j_1} \cdot  f(A,x)_{j_0} \cdot  f_c(A,x) \cdot x \cdot c_g(A,x)^{\top}
\end{align*}
\item {\bf Part 9.}
\begin{align*}
    B_{1,3,1}^{j_1,*,j_0,*}  =  f(A,x)_{j_1} \cdot  f(A,x)_{j_0}   \cdot f_2(A,x) \cdot x \cdot  c_g(A,x)^{\top}
\end{align*}
\item {\bf Part 10.}
\begin{align*}
    B_{1,3,2}^{j_1,*,j_0,*}  =   -   f(A,x)_{j_1}^2 \cdot  f(A,x)_{j_0} \cdot x  \cdot c_g(A,x)^{\top}
\end{align*}
\item {\bf Part 11.}
\begin{align*}
    B_{1,4,1}^{j_1,*,j_0,*}  =   f(A,x)_{j_1} \cdot  f(A,x)_{j_0}  \cdot f_c(A,x)  \cdot x \cdot c_g(A,x)^{\top}
\end{align*}
\item {\bf Part 12.}
\begin{align*}
   B_{1,4,2}^{j_1,*,j_0,*}  =  -   f(A,x)_{j_1} \cdot  f(A,x)_{j_0}  \cdot c(A,x)_{j_1} \cdot  x \cdot c_g(A,x)^{\top} 
\end{align*}

\end{itemize}
\begin{proof}
{\bf Proof of Part 1.}
    We have
    \begin{align*}
        B_{1,1,1}^{j_1,i_1,j_0,i_1} = f(A,x)_{j_1} \cdot f(A,x)_{j_0} \cdot  e_{i_1}^\top \cdot \langle c(A,x), f(A,x) \rangle^2 \cdot e_{i_1}\\
        B_{1,1,1}^{j_1,i_1,j_0,i_0} = f(A,x)_{j_1} \cdot f(A,x)_{j_0} \cdot  e_{i_1}^\top \cdot \langle c(A,x), f(A,x) \rangle^2 \cdot e_{i_0}
    \end{align*}

    From the above two equations, we can tell that $B_{1,1,1}^{j_1,*,j_0,*} \in \R^{d \times d}$ is a matrix that only diagonal has entries and off-diagonal are all zeros.
    
    Then we have $B_{1,1,1}^{j_1,*,j_0,*} \in \R^{d \times d}$ can be written as the rescaling of a diagonal matrix,
    \begin{align*}
     B_{1,1,1}^{j_1,*,j_0,*} & ~ = f(A,x)_{j_1} \cdot f(A,x)_{j_0} \cdot  \langle c(A,x), f(A,x) \rangle^2 \cdot I_d \\
     & ~ =  f(A,x)_{j_1} \cdot f(A,x)_{j_0} \cdot  f_c(A,x)^2 \cdot I_d 
\end{align*}
where the last step is follows from the Definitions~\ref{def:f_c}.

{\bf Proof of Part 2.}
    We have
    \begin{align*}
        B_{1,1,2}^{j_1,i_1,j_0,i_1} =  f(A,x)_{j_1} \cdot f(A,x)_{j_0} \cdot c(A,x)_{j_1}\cdot e_{i_1}^\top \cdot  \langle c(A,x), f(A,x) \rangle \cdot e_{i_1} \\
        B_{1,1,2}^{j_1,i_1,j_0,i_0} =  f(A,x)_{j_1} \cdot f(A,x)_{j_0} \cdot c(A,x)_{j_1}\cdot e_{i_1}^\top \cdot  \langle c(A,x), f(A,x) \rangle \cdot e_{i_0}
    \end{align*}
     From the above two equations, we can tell that $B_{1,1,2}^{j_1,*,j_0,*} \in \R^{d \times d}$ is a matrix that only diagonal has entries and off-diagonal are all zeros.
    
    Then we have $B_{1,1,2}^{j_1,*,j_0,*} \in \R^{d \times d}$ can be written as the rescaling of a diagonal matrix,
\begin{align*}
     B_{1,1,2}^{j_1,*,j_0,*} & ~ = f(A,x)_{j_1} \cdot f(A,x)_{j_0} \cdot c(A,x)_{j_1}\cdot  \langle c(A,x), f(A,x) \rangle \cdot  I_d\\
     & ~ =  f(A,x)_{j_1} \cdot f(A,x)_{j_0} \cdot c(A,x)_{j_1}\cdot  f_c(A,x) \cdot  I_d
\end{align*}
where the last step is follows from the Definitions~\ref{def:f_c}.

{\bf Proof of Part 3.}
We have for diagonal entry and off-diagonal entry can be written as follows 
    \begin{align*}
        B_{1,1,3}^{j_1,i_1,j_0,i_1} =  f(A,x)_{j_1} \cdot f(A,x)_{j_0} \cdot \langle c(A,x), f(A,x) \rangle^2 \cdot ( (A_{j_1,*}) \circ x^\top  )   \cdot e_{i_1} \\
        B_{1,1,3}^{j_1,i_1,j_0,i_0} =  f(A,x)_{j_1} \cdot f(A,x)_{j_0} \cdot \langle c(A,x), f(A,x) \rangle^2 \cdot ( (A_{j_1,*}) \circ x^\top  )   \cdot e_{i_0}
    \end{align*}
From the above equation, we can show that matrix $B_{1,1,3}^{j_1,*,j_0,*}$ can be expressed as a rank-$1$ matrix,
\begin{align*}
     B_{1,1,3}^{j_1,*,j_0,*} & ~ = f(A,x)_{j_1} \cdot f(A,x)_{j_0} \cdot \langle c(A,x), f(A,x) \rangle^2 \cdot {\bf 1}_d \cdot ( (A_{j_1,*}) \circ x^\top  ) \\
     & ~ = f(A,x)_{j_1} \cdot f(A,x)_{j_0} \cdot f_c(A,x)^2 \cdot {\bf 1}_d \cdot ( (A_{j_1,*}) \circ x^\top  )
\end{align*}
where the last step is follows from the Definitions~\ref{def:f_c}.

{\bf Proof of Part 4.}
We have for diagonal entry and off-diagonal entry can be written as follows
    \begin{align*}
        B_{1,1,4}^{j_1,i_1,j_0,i_1} = -  f(A,x)_{j_1} \cdot f(A,x)_{j_0} \cdot f(A,x)^\top  \cdot A \cdot \diag(x) \cdot   (\langle c(A,x), f(A,x) \rangle)^2 \cdot  e_{i_1} \\
        B_{1,1,4}^{j_1,i_1,j_0,i_0} = -  f(A,x)_{j_1} \cdot f(A,x)_{j_0} \cdot f(A,x)^\top  \cdot A \cdot \diag(x) \cdot   (\langle c(A,x), f(A,x) \rangle)^2 \cdot  e_{i_0}
    \end{align*}
 From the above equation, we can show that matrix $B_{1,1,4}^{j_1,*,j_0,*}$ can be expressed as a rank-$1$ matrix,
\begin{align*}
    B_{1,1,4}^{j_1,*,j_0,*}  & ~ =  -  f(A,x)_{j_1} \cdot f(A,x)_{j_0} \cdot    \langle c(A,x), f(A,x) \rangle^2 \cdot {\bf 1}_d  \cdot f(A,x)^\top  \cdot A \cdot \diag(x) \\
     & ~ = -  f(A,x)_{j_1} \cdot f(A,x)_{j_0} \cdot f_c(A,x)^2 \cdot {\bf 1}_d  \cdot h(A,x)^\top
\end{align*}
where the last step is follows from the Definitions~\ref{def:h} and Definitions~\ref{def:f_c}.

{\bf Proof of Part 5.}
We have for diagonal entry and off-diagonal entry can be written as follows
    \begin{align*}
     & ~B_{1,1,5}^{j_1,i_1,j_0,i_1} =   f(A,x)_{j_1} \cdot f(A,x)_{j_0} \cdot f(A,x)^\top  \cdot A \cdot \diag(x) \cdot (\langle -f(A,x), f(A,x) \rangle + f(A,x)_{j_1})  \\
     & ~ \cdot \langle c(A,x), f(A,x) \rangle \cdot e_{i_1} \\
     & ~B_{1,1,5}^{j_1,i_1,j_0,i_0} =  f(A,x)_{j_1} \cdot f(A,x)_{j_0} \cdot f(A,x)^\top  \cdot A \cdot \diag(x) \cdot (\langle -f(A,x), f(A,x) \rangle + f(A,x)_{j_1}) \\
        & ~ \cdot \langle c(A,x), f(A,x) \rangle \cdot e_{i_0}
    \end{align*}
    From the above equation, we can show that matrix $B_{1,1,5}^{j_1,*,j_0,*}$ can be expressed as a rank-$1$ matrix,
\begin{align*}
    B_{1,1,5}^{j_1,*,j_0,*}  & ~ =   f(A,x)_{j_1} \cdot f(A,x)_{j_0} \cdot \langle c(A,x), f(A,x) \rangle  \cdot (\langle -f(A,x), f(A,x) \rangle + f(A,x)_{j_1})    \\
        & ~\cdot {\bf 1}_d  \cdot f(A,x)^\top  \cdot A \cdot \diag(x)\\
     & ~ =f(A,x)_{j_1} \cdot f(A,x)_{j_0} \cdot f_c(A,x)  \cdot (-f_2(A,x) + f(A,x)_{j_1})   \cdot {\bf 1}_d  \cdot h(A,x)^\top
\end{align*}
where the last step is follows from the Definitions~\ref{def:h}, Definitions~\ref{def:f_2} and Definitions~\ref{def:f_c}.

{\bf Proof of Part 6.}
We have for diagonal entry and off-diagonal entry can be written as follows
    \begin{align*}
        & ~ B_{1,1,6}^{j_1,i_1,j_0,i_1} =    f(A,x)_{j_1} \cdot f(A,x)_{j_0} \cdot f(A,x)^\top  \cdot A \cdot  \diag(x) \cdot(\langle -f(A,x), c(A,x) \rangle + f(A,x)_{j_1})  \\
        & ~\cdot \langle c(A,x), f(A,x) \rangle \cdot e_{i_1} \\
         & ~B_{1,1,6}^{j_1,i_1,j_0,i_0} =   f(A,x)_{j_1} \cdot f(A,x)_{j_0} \cdot f(A,x)^\top  \cdot A \cdot  \diag(x) \cdot(\langle -f(A,x), c(A,x) \rangle + f(A,x)_{j_1})  \\
        & ~\cdot \langle c(A,x), f(A,x) \rangle \cdot e_{i_0} 
    \end{align*}
    From the above equation, we can show that matrix $B_{1,1,6}^{j_1,*,j_0,*}$ can be expressed as a rank-$1$ matrix,
\begin{align*}
    B_{1,1,6}^{j_1,*,j_0,*}  & ~ =   f(A,x)_{j_1} \cdot f(A,x)_{j_0}  \cdot(\langle -f(A,x), c(A,x) \rangle + f(A,x)_{j_1}) \cdot \langle c(A,x), f(A,x) \rangle   \\
        & ~ \cdot {\bf 1}_d  \cdot f(A,x)^\top  \cdot A \cdot \diag(x)\\
     & ~ =  f(A,x)_{j_1} \cdot f(A,x)_{j_0}  \cdot(-f_c(A,x) + f(A,x)_{j_1}) \cdot f_c(A,x)   \cdot {\bf 1}_d  \cdot h(A,x)^\top
\end{align*}
where the last step is follows from the Definitions~\ref{def:h} and Definitions~\ref{def:f_c}.

{\bf Proof of Part 7.}
We have for diagonal entry and off-diagonal entry can be written as follows
    \begin{align*}
         B_{1,1,7}^{j_1,i_1,j_0,i_1} = f(A,x)_{j_1} \cdot f(A,x)_{j_0} \cdot ((e_{j_1}^\top - f(A,x)^\top) \circ q(A,x)^\top) \cdot A \cdot  \diag(x)  \cdot \langle c(A,x), f(A,x) \rangle \cdot e_{i_1}\\
         B_{1,1,7}^{j_1,i_1,j_0,i_0} = f(A,x)_{j_1} \cdot f(A,x)_{j_0} \cdot ((e_{j_1}^\top - f(A,x)^\top) \circ q(A,x)^\top) \cdot A \cdot  \diag(x)  \cdot \langle c(A,x), f(A,x) \rangle \cdot e_{i_0}
    \end{align*}
    From the above equation, we can show that matrix $B_{1,1,6}^{j_1,*,j_0,*}$ can be expressed as a rank-$1$ matrix,
\begin{align*}
     B_{1,1,7}^{j_1,*,j_0,*}  & ~ =    f(A,x)_{j_1} \cdot f(A,x)_{j_0} \cdot \langle c(A,x), f(A,x) \rangle \cdot {\bf 1}_d  \cdot ((e_{j_1}^\top - f(A,x)^\top) \circ q(A,x)^\top) \cdot A \cdot  \diag(x)   \\
     & ~ = f(A,x)_{j_1} \cdot f(A,x)_{j_0} \cdot f_c(A,x) \cdot {\bf 1}_d  \cdot p_{j_0}(x)^\top 
\end{align*}
where the last step is follows from the Definitions~\ref{def:f_c} and Definitions~\ref{def:p}.

{\bf Proof of Part 8.}
We have for diagonal entry and off-diagonal entry can be written as follows
    \begin{align*}
         B_{1,2,1}^{j_1,i_1,j_0,i_1} =   c_g(A,x)^{\top} \cdot f(A,x)_{j_1} \cdot  f(A,x)_{j_0} \cdot x_{i_1} \cdot \langle c(A,x), f(A,x) \rangle \cdot e_{i_0} \\
         B_{1,2,1}^{j_1,i_1,j_0,i_0} = c_g(A,x)^{\top} \cdot f(A,x)_{j_1} \cdot  f(A,x)_{j_0} \cdot x_{i_1} \cdot \langle c(A,x), f(A,x) \rangle \cdot e_{i_0} 
    \end{align*}
    From the above equation, we can show that matrix $B_{1,2,1}^{j_1,*,j_0,*}$ can be expressed as a rank-$1$ matrix,
\begin{align*}
     B_{1,2,1}^{j_1,*,j_0,*}  & ~ =  f(A,x)_{j_1} \cdot  f(A,x)_{j_0} \cdot \langle c(A,x), f(A,x) \rangle \cdot x \cdot c_g(A,x)^{\top}  \\
     & ~ =f(A,x)_{j_1} \cdot  f(A,x)_{j_0} \cdot  f_c(A,x) \cdot x \cdot c_g(A,x)^{\top}
\end{align*}
where the last step is follows from the Definitions~\ref{def:f_c}.

{\bf Proof of Part 9.}
We have for diagonal entry and off-diagonal entry can be written as follows
    \begin{align*}
         B_{1,3,1}^{j_1,i_1,j_0,i_1} =    c_g(A,x)^{\top} \cdot f(A,x)_{j_1} \cdot  f(A,x)_{j_0}  \cdot x_{i_1} \cdot \langle  f(A,x), f(A,x) \rangle \cdot e_{i_1}\\
         B_{1,3,1}^{j_1,i_1,j_0,i_0} =    c_g(A,x)^{\top} \cdot f(A,x)_{j_1} \cdot  f(A,x)_{j_0}  \cdot x_{i_1} \cdot \langle  f(A,x), f(A,x) \rangle \cdot e_{i_0}
    \end{align*}
            From the above equation, we can show that matrix $B_{1,3,1}^{j_1,*,j_0,*}$ can be expressed as a rank-$1$ matrix,
\begin{align*}
    B_{1,3,1}^{j_1,*,j_0,*} & ~ =  f(A,x)_{j_1} \cdot  f(A,x)_{j_0}   \cdot \langle  f(A,x), f(A,x) \rangle \cdot x \cdot  c_g(A,x)^{\top}  \\
     & ~ = f(A,x)_{j_1} \cdot  f(A,x)_{j_0}   \cdot f_2(A,x) \cdot x \cdot  c_g(A,x)^{\top}
\end{align*}
where the last step is follows from the Definitions~\ref{def:f_2}.

{\bf Proof of Part 10.}
We have for diagonal entry and off-diagonal entry can be written as follows
    \begin{align*}
         B_{1,3,2}^{j_1,i_1,j_0,i_1} =   -  c_g(A,x)^{\top} \cdot f(A,x)_{j_1}^2 \cdot  f(A,x)_{j_0} \cdot x_{i_1}  \cdot e_{i_0}\\
         B_{1,3,2}^{j_1,i_1,j_0,i_0} =   -  c_g(A,x)^{\top} \cdot f(A,x)_{j_1}^2 \cdot  f(A,x)_{j_0} \cdot x_{i_1}  \cdot e_{i_1}
    \end{align*}
            From the above equation, we can show that matrix $B_{1,3,2}^{j_1,*,j_0,*}$ can be expressed as a rank-$1$ matrix,
\begin{align*}
    B_{1,3,2}^{j_1,*,j_0,*}  =  -   f(A,x)_{j_1}^2 \cdot  f(A,x)_{j_0} \cdot x  \cdot c_g(A,x)^{\top}
\end{align*}
{\bf Proof of Part 11.}
We have for diagonal entry and off-diagonal entry can be written as follows
    \begin{align*}
         B_{1,4,1}^{j_1,i_1,j_0,i_1} =  c_g(A,x)^{\top} \cdot f(A,x)_{j_1} \cdot  f(A,x)_{j_0} \cdot x_{i_1} \cdot \langle  f(A,x), c(A,x) \rangle  \cdot e_{i_1}\\
         B_{1,4,1}^{j_1,i_1,j_0,i_0} =  c_g(A,x)^{\top} \cdot f(A,x)_{j_1} \cdot  f(A,x)_{j_0} \cdot x_{i_1} \cdot \langle  f(A,x), c(A,x) \rangle  \cdot e_{i_0}
    \end{align*}
            From the above equation, we can show that matrix $B_{1,4,1}^{j_1,*,j_0,*}$ can be expressed as a rank-$1$ matrix,
\begin{align*}
    B_{1,4,1}^{j_1,*,j_0,*}  & ~ =  f(A,x)_{j_1} \cdot  f(A,x)_{j_0}  \cdot \langle  f(A,x), c(A,x) \rangle  \cdot x \cdot c_g(A,x)^{\top}   \\
     & ~ = f(A,x)_{j_1} \cdot  f(A,x)_{j_0}  \cdot f_c(A,x)  \cdot x \cdot c_g(A,x)^{\top}
\end{align*}
where the last step is follows from the Definitions~\ref{def:f_c}.

{\bf Proof of Part 12.}
We have for diagonal entry and off-diagonal entry can be written as follows
    \begin{align*}
          B_{1,4,2}^{j_1,i_1,j_0,i_1} = -  c_g(A,x)^{\top} \cdot f(A,x)_{j_1} \cdot  f(A,x)_{j_0} \cdot x_{i_1} \cdot c(A,x)_{j_1} \cdot e_{i_1}\\
          B_{1,4,2}^{j_1,i_1,j_0,i_0} =  -  c_g(A,x)^{\top} \cdot f(A,x)_{j_1} \cdot  f(A,x)_{j_0} \cdot x_{i_1} \cdot c(A,x)_{j_1} \cdot e_{i_0}
    \end{align*}
            From the above equation, we can show that matrix $B_{1,4,2}^{j_1,*,j_0,*}$ can be expressed as a rank-$1$ matrix,
\begin{align*}
   B_{1,4,2}^{j_1,*,j_0,*}  & ~ =   -   f(A,x)_{j_1} \cdot  f(A,x)_{j_0}  \cdot c(A,x)_{j_1} \cdot  x \cdot c_g(A,x)^{\top} 
\end{align*}
\end{proof}

\end{lemma}

\subsection{Expanding \texorpdfstring{$B_1$}{} into many terms}

\begin{lemma}
   If the following conditions hold
    \begin{itemize}
     \item Let $u(A,x) \in \R^n$ be defined as Definition~\ref{def:u}
    \item Let $\alpha(A,x) \in \R$ be defined as Definition~\ref{def:alpha}
     \item Let $f(A,x) \in \R^n$ be defined as Definition~\ref{def:f}
    \item Let $c(A,x) \in \R^n$ be defined as Definition~\ref{def:c}
    \item Let $g(A,x) \in \R^d$ be defined as Definition~\ref{def:g} 
    \item Let $q(A,x) = c(A,x) + f(A,x) \in \R^n$
    \item Let $c_g(A,x) \in \R^d$ be defined as Definition~\ref{def:c_g}.
    \item Let $L_g(A,x) \in \R$ be defined as Definition~\ref{def:l_g}
    \item Let $v \in \R^n$ be a vector 
    \item Let $B_1^{j_1,i_1,j_0,i_0}$ be defined as Definition~\ref{def:b_1}
    \end{itemize}
Then, For $j_0,j_1 \in [n], i_0,i_1 \in [d]$, we have 
\begin{itemize}
    \item {\bf Part 1.}For $j_1 = j_0$ and $i_0 = i_1$
    \begin{align*}
    B_{1}^{j_1,i_1,j_1,i_1} 
    = & ~  B_{1,1}^{j_1,i_1,j_1,i_1}  + B_{1,2}^{j_1,i_1,j_1,i_1}  +B_{1,3}^{j_1,i_1,j_1,i_1} 
       +B_{1,4}^{j_1,i_1,j_1,i_1} 
\end{align*}
\item  {\bf Part 2.} For $j_1 = j_0$ and $i_0 \neq i_1$
\begin{align*}
     B_{1}^{j_1,i_1,j_1,i_0} 
    = & ~  B_{1,1}^{j_1,i_1,j_1,i_0}  + B_{1,2}^{j_1,i_1,j_1,i_0}  +B_{1,3}^{j_1,i_1,j_1,i_0}  +B_{1,4}^{j_1,i_1,j_1,i_0} 
\end{align*}
\item  {\bf Part 3.} For $j_1 \neq j_0$ and $i_0 = i_1$
\begin{align*}
       B_{1}^{j_1,i_1,j_0,i_1} 
    = & ~ B_{1,1}^{j_1,i_1,j_0,i_1}  + B_{1,2}^{j_1,i_1,j_0,i_1}  +B_{1,3}^{j_1,i_1,j_0,i_1} +B_{1,4}^{j_1,i_1,j_0,i_1} 
\end{align*}
    \item {\bf Part 4.} For $j_0 \neq j_1$ and $i_0 \neq i_1$, we have
    \begin{align*}
        B_{1}^{j_1,i_1,j_0,i_0} 
        = & ~ B_{1,1}^{j_1,i_1,j_0,i_0}  + B_{1,2}^{j_1,i_1,j_0,i_0}  +B_{1,3}^{j_1,i_1,j_0,i_0}
     +B_{1,4}^{j_1,i_1,j_0,i_0} 
    \end{align*}
\end{itemize}

\end{lemma}
\begin{proof}
{\bf Proof of Part 1.}
We have
    \begin{align*}
    B_{1}^{j_1,i_1,j_1,i_1}
    = & ~ \frac{\d}{\d A_{j_1,i_1}}(-  c_g(A,x)^{\top} \cdot f(A,x)_{j_1}  \cdot \langle c(A,x), f(A,x) \rangle \cdot e_{i_1}) \\
    = & ~ B_{1,1}^{j_1,i_1,j_1,i_1}  + B_{1,2}^{j_1,i_1,j_1,i_1}  +B_{1,3}^{j_1,i_1,j_1,i_1} +  B_{1,4}^{j_1,i_1,j_1,i_1}
\end{align*}
{\bf Proof of Part 2.}
We have
    \begin{align*}
    B_{1}^{j_1,i_1,j_1,i_0}
    = & ~ \frac{\d}{\d A_{j_1,i_1}}(-  c_g(A,x)^{\top} \cdot f(A,x)_{j_1}  \cdot \langle c(A,x), f(A,x) \rangle \cdot e_{i_0}) \\
    = & ~ B_{1,1}^{j_1,i_1,j_1,i_0} + B_{1,2}^{j_1,i_1,j_1,i_0} +B_{1,3}^{j_1,i_1,j_1,i_0} +  B_{1,4}^{j_1,i_1,j_1,i_0}
\end{align*}

{\bf Proof of Part 3.}
We also have
\begin{align*}
     B_{1,1}^{j_1,i_1,j_0,i_1}  
    = & ~ \frac{\d}{\d A_{j_1,i_1}}(-  c_g(A,x)^{\top} \cdot f(A,x)_{j_0}  \cdot \langle c(A,x), f(A,x) \rangle \cdot e_{i_1}) \\
    = & ~  B_{1,1}^{j_1,i_1,j_0,i_1} +  B_{1,2}^{j_1,i_1,j_0,i_1} +  B_{1,3}^{j_1,i_1,j_0,i_1} +  B_{1,4}^{j_1,i_1,j_0,i_1}
\end{align*}

{\bf Proof of Part 4.}
We also have
\begin{align*}
      B_{1,1}^{j_1,i_1,j_0,i_0}  
    =& ~ \frac{\d}{\d A_{j_1,i_1}}(-  c_g(A,x)^{\top} \cdot f(A,x)_{j_0}  \cdot \langle c(A,x), f(A,x) \rangle \cdot e_{i_0}) \\
    = & ~B_{1,1}^{j_1,i_1,j_0,i_0} +B_{1,2}^{j_1,i_1,j_0,i_0} + B_{1,3}^{j_1,i_1,j_0,i_0}+  B_{1,4}^{j_1,i_1,j_0,i_0}
\end{align*}
\end{proof}

\subsection{Lipschitz Computation}

\begin{lemma}
If the following conditions hold
\begin{itemize}\label{lips: B_1}
    \item Let $B_{1,1,1}^{j_1,*, j_0,*}, \cdots, B_{1,4,2}^{j_1,*, j_0,*} $ be defined as Lemma~\ref{lem:b_1_j1_j0} 
    \item  Let $\|A \|_2 \leq R, \|A^{\top} \|_F \leq R, \| x\|_2 \leq R, \|\diag(f(A,x)) \|_F \leq \|f(A,x) \|_2 \leq 1, \| b_g \|_2 \leq 1$ 
\end{itemize}

Then, we have
\begin{itemize}
    \item {\bf Part 1.}
    \begin{align*}
       \| B_{1,1,1}^{j_1,*,j_0,*} (A) - B_{1,1,1}^{j_1,*,j_0,*} ( \wt{A} ) \|_F \leq \beta^{-2} \cdot n \cdot \sqrt{d}\exp(4R^2) \cdot \|A - \wt{A}\|_F 
    \end{align*}
    \item {\bf Part 2.}
    \begin{align*}
          \| B_{1,1,2}^{j_1,*,j_0,*} (A) - B_{1,1,2}^{j_1,*,j_0,*} ( \wt{A} ) \|_F \leq \beta^{-2} \cdot n \cdot \sqrt{d}  \exp(4R^2)\|A - \wt{A}\|_F
    \end{align*}
    \item {\bf Part 3.}
    \begin{align*}
         \| B_{1,1,3}^{j_1,*,j_0,*} (A) - B_{1,1,3}^{j_1,*,j_0,*} ( \wt{A} ) \|_F \leq  \beta^{-2} \cdot n \cdot \sqrt{d}  \cdot \exp(5R^2)\|A - \wt{A}\|_F
    \end{align*}
    \item {\bf Part 4.}
    \begin{align*}
        \| B_{1,1,4}^{j_1,*,j_0,*} (A) - B_{1,1,4}^{j_1,*,j_0,*} ( \wt{A} ) \|_F \leq \beta^{-2} \cdot n \cdot \sqrt{d} \cdot \exp(5R^2) \cdot \|A - \wt{A}\|_F 
    \end{align*}
    \item {\bf Part 5.}
    \begin{align*}
        \| B_{1,1,5}^{j_1,*,j_0,*} (A) - B_{1,1,5}^{j_1,*,j_0,*} ( \wt{A} ) \|_F \leq \beta^{-2} \cdot n \cdot \sqrt{d}   \cdot \exp(5R^2) \cdot \|A - \wt{A}\|_F 
    \end{align*}
    \item {\bf Part 6.}
    \begin{align*}
        \| B_{1,1,6}^{j_1,*,j_0,*} (A) - B_{1,1,6}^{j_1,*,j_0,*} ( \wt{A} ) \|_F \leq \beta^{-2} \cdot n \cdot \sqrt{d} \cdot \exp(5R^2) \cdot \|A - \wt{A}\|_F 
    \end{align*}
    \item {\bf Part 7.}
    \begin{align*}
        \| B_{1,1,7}^{j_1,*,j_0,*} (A) - B_{1,1,7}^{j_1,*,j_0,*} ( \wt{A} ) \|_F \leq \beta^{-2} \cdot n \cdot \sqrt{d} \cdot \exp(5R^2) \cdot \|A - \wt{A}\|_F 
    \end{align*}
    \item {\bf Part 8.}
    \begin{align*}
        \| B_{1,2,1}^{j_1,*,j_0,*} (A) - B_{1,2,1}^{j_1,*,j_0,*} ( \wt{A} ) \|_F \leq \beta^{-2} \cdot n  \cdot \exp(5R^2) \cdot \|A - \wt{A}\|_F 
    \end{align*}
    \item {\bf Part 9.}
    \begin{align*}
        \| B_{1,3,1}^{j_1,*,j_0,*} (A) - B_{1,3,1}^{j_1,*,j_0,*} ( \wt{A} ) \|_F \leq \beta^{-2} \cdot n  \cdot \exp(5R^2) \cdot \|A - \wt{A}\|_F 
    \end{align*}
    \item {\bf Part 10.}
    \begin{align*}
        \| B_{1,3,2}^{j_1,*,j_0,*} (A) - B_{1,3,2}^{j_1,*,j_0,*} ( \wt{A} ) \|_F \leq \beta^{-2} \cdot n  \cdot \exp(5R^2) \cdot \|A - \wt{A}\|_F 
    \end{align*}
    \item {\bf Part 11.}
    \begin{align*}
        \| B_{1,4,1}^{j_1,*,j_0,*} (A) - B_{1,4,1}^{j_1,*,j_0,*} ( \wt{A} ) \|_F \leq \beta^{-2} \cdot n  \cdot \exp(5R^2) \cdot \|A - \wt{A}\|_F 
    \end{align*}
    \item {\bf Part 12.}
    \begin{align*}
        \| B_{1,4,2}^{j_1,*,j_0,*} (A) - B_{1,4,2}^{j_1,*,j_0,*} ( \wt{A} ) \|_F \leq \beta^{-2} \cdot n  \cdot \exp(5R^2) \cdot \|A - \wt{A}\|_F 
    \end{align*}
    \item{\bf Part 13.}
    \begin{align*}
         \| B_{1}^{j_1,*,j_0,*} (A) - B_{1}^{j_1,*,j_0,*} ( \wt{A} ) \|_F \leq & ~ 12  \beta^{-2} \cdot n \cdot \sqrt{d} \cdot \exp(5R^2)\|A - \wt{A}\|_F
    \end{align*}
\end{itemize}
\end{lemma}
\begin{proof}
{\bf Proof of Part 1.}
\begin{align*}
& ~ \| B_{1,1,1}^{j_1,*,j_0,*} (A) - B_{1,1,1}^{j_1,*,j_0,*} ( \wt{A} ) \|_F \\ \leq 
    & ~ \|  f(A,x)_{j_1} \cdot f(A,x)_{j_0} \cdot  f_c(A,x)^2 \cdot I_d  - f(\wt{A},x)_{j_1} \cdot f(\wt{A},x)_{j_0} \cdot  f_c(A,x)^2 \cdot I_{d}\|_F \\
    \leq & ~ |  f(A,x)_{j_1} - f(\wt{A},x)_{j_1}||f(A,x)_{j_0}| | f_c(A,x) |^2 \| I_d\|_F \\
    & + ~ |f(\wt{A},x)_{j_1}| |f(A,x)_{j_0} - f(\wt{A},x)_{j_0}| | f_c(A,x) |^2 \| I_d\|_F \\
    & + ~  |f(\wt{A},x)_{j_1}| | f(\wt{A},x)_{j_0}| |f_c(A,x) - f_c(\wt{A},x)| | f_c(A,x) |\| I_d\|_F \\
    & + ~ |f(\wt{A},x)_{j_1}| | f(\wt{A},x)_{j_0}|| f_c(\wt{A},x) | |f_c(A,x) - f_c(\wt{A},x)| \| I_d\|_F\\
    \leq & ~ 8  \beta^{-2} \cdot n \cdot \sqrt{d}  \exp(3R^2)\|A - \wt{A}\|_F \\
    & + ~ 8  \beta^{-2} \cdot n \cdot \sqrt{d}  \exp(3R^2)\|A - \wt{A}\|_F \\
    & + ~ 12\beta^{-2} \cdot n \cdot \sqrt{d}\exp(3R^2) \cdot \|A - \wt{A}\|_F \\
    & + ~ 12\beta^{-2} \cdot n \cdot \sqrt{d}\exp(3R^2) \cdot \|A - \wt{A}\|_F \\
    \leq &  ~ 40\beta^{-2} \cdot n \cdot \sqrt{d}\exp(3R^2) \cdot \|A - \wt{A}\|_F \\
    \leq &  ~ \beta^{-2} \cdot n \cdot \sqrt{d}\exp(4R^2) \cdot \|A - \wt{A}\|_F 
\end{align*}

{\bf Proof of Part 2.}
\begin{align*}
    & ~ \| B_{1,1,2}^{j_1,*,j_0,*} (A) - B_{1,1,2}^{j_1,*,j_0,*} ( \wt{A} ) \|_F \\
    \leq & ~  \| f(A,x)_{j_1} \cdot f(A,x)_{j_0} \cdot c(A,x)_{j_1}\cdot  f_c(A,x) \cdot  I_d - f(\wt{A},x)_{j_1} \cdot f(\wt{A},x)_{j_0} \cdot c(\wt{A},x)_{j_1}\cdot  f_c(\wt{A},x) \cdot  I_d\|_F \\
    \leq & ~ | f(A,x)_{j_1} -f(\wt{A},x)_{j_1} | |f(A,x)_{j_0}| |c(A,x)_{j_1}| |f_c(A,x)| \|  I_d \|_F \\
    & + ~ | f(\wt{A},x)_{j_1} | |f(A,x)_{j_0} - f(\wt{A},x)_{j_0}| |c(A,x)_{j_1}| |f_c(A,x)| \|  I_d \|_F \\
    & + ~ | f(\wt{A},x)_{j_1} | |f(\wt{A},x)_{j_0}| |c(A,x)_{j_1} -c(\wt{A},x)_{j_1}| |f_c(A,x)| \|  I_d \|_F \\
    & + ~ | f(\wt{A},x)_{j_1} | |f(\wt{A},x)_{j_0}| |c(\wt{A},x)_{j_1}| |f_c(A,x) - f_c(\wt{A},x)| \|  I_d \|_F \\
    \leq & ~8   \beta^{-2} \cdot n \cdot \sqrt{d}  \exp(3R^2)\|A - \wt{A}\|_F \\
    & + ~  8  \beta^{-2} \cdot n \cdot \sqrt{d}  \exp(3R^2)\|A - \wt{A}\|_F \\
    & + ~  4  \beta^{-2} \cdot n \cdot \sqrt{d}  \exp(3R^2)\|A - \wt{A}\|_F \\
    & + ~  12  \beta^{-2} \cdot n \cdot \sqrt{d}  \exp(3R^2)\|A - \wt{A}\|_F \\
    \leq & ~ 32\beta^{-2} \cdot n \cdot \sqrt{d}  \exp(3R^2)\|A - \wt{A}\|_F\\
    \leq & ~ \beta^{-2} \cdot n \cdot \sqrt{d}  \exp(4R^2)\|A - \wt{A}\|_F
\end{align*}

{\bf Proof of Part 3.}
\begin{align*}
     & ~ \| B_{1,1,3}^{j_1,*,j_0,*} (A) - B_{1,1,3}^{j_1,*,j_0,*} ( \wt{A} ) \|_F \\
     \leq & ~ \| f(A,x)_{j_1} \cdot f(A,x)_{j_0} \cdot f_c(A,x)^2 \cdot {\bf 1}_d \cdot ( (A_{j_1,*}) \circ x^\top  ) \\
     & ~ -  f(\wt{A},x)_{j_1} \cdot f(\wt{A},x)_{j_0} \cdot f_c(\wt{A},x)^2 \cdot {\bf 1}_d \cdot ( (\wt{A}_{j_1,*}) \circ x^\top  )\|_F\\
     \leq & ~ |f(A,x)_{j_1} - f(\wt{A},x)_{j_1}| |f(A,x)_{j_0}| |f_c(A,x)|^2 \cdot \|{\bf 1}_d \|_2\cdot \|A_{j_1,*} \|_2 \| \diag(x)\|_F  \\
     & + ~ | f(\wt{A},x)_{j_1}| |f(A,x)_{j_0} -f(\wt{A},x)_{j_0}  | |f_c(A,x)|^2 \cdot \|{\bf 1}_d \|_2\cdot \|A_{j_1,*} \|_2 \| \diag(x)\|_F  \\
     & + ~ | f(\wt{A},x)_{j_1}| |f(\wt{A},x)_{j_0}  | |f_c(A,x) - f_c(\wt{A},x)||f_c(A,x)| \cdot \|{\bf 1}_d \|_2\cdot \|A_{j_1,*} \|_2 \| \diag(x)\|_F  \\
      & + ~ | f(\wt{A},x)_{j_1}| |f(\wt{A},x)_{j_0}  | |f_c(\wt{A},x)||f_c(A,x) - f_c(\wt{A},x)| \cdot \|{\bf 1}_d \|_2\cdot \|A_{j_1,*} \|_2 \| \diag(x)\|_F  \\
     & + ~ | f(\wt{A},x)_{j_1}| |f(\wt{A},x)_{j_0}  | |f_c(\wt{A},x)||f_c(\wt{A},x)| \cdot \|{\bf 1}_d \|_2\cdot \|A_{j_1,*} - \wt{A}_{j_1,*}\|_2 \| \diag(x)\|_F  \\
    \leq & ~ 8 R^2 \cdot \beta^{-2} \cdot n \cdot \sqrt{d}  \cdot \exp(3R^2)\|A - \wt{A}\|_F \\
    & + ~ 8 R^2 \cdot \beta^{-2} \cdot n \cdot \sqrt{d}  \cdot \exp(3R^2)\|A - \wt{A}\|_F \\
    & + ~ 12 R^2 \cdot \beta^{-2} \cdot n \cdot \sqrt{d}  \cdot \exp(3R^2)\|A - \wt{A}\|_F \\
    & + ~12 R^2 \cdot \beta^{-2} \cdot n \cdot \sqrt{d}  \cdot \exp(3R^2)\|A - \wt{A}\|_F \\
    & + ~ 4 R \cdot \sqrt{d} \cdot \|A - \wt{A} \|_F \\
    \leq & ~ 44  \cdot \beta^{-2} \cdot n \cdot \sqrt{d}  \cdot \exp(4R^2)\|A - \wt{A}\|_F\\
    \leq & ~\beta^{-2} \cdot n \cdot \sqrt{d}  \cdot \exp(5R^2)\|A - \wt{A}\|_F
\end{align*}

{\bf Proof of Part 4.}
\begin{align*}
     & ~ \| B_{1,1,4}^{j_1,*,j_0,*} (A) - B_{1,1,4}^{j_1,*,j_0,*} ( \wt{A} ) \|_F \\
     \leq & ~ \|-  f(A,x)_{j_1}^2 \cdot f_c(A,x)^2 \cdot {\bf 1}_d  \cdot  h(A,x)^\top  -(-f(\wt{A},x)_{j_1}^2 \cdot f_c(\wt{A},x)^2 \cdot {\bf 1}_d  \cdot  h(\wt{A},x)^\top) \|_F \\
      \leq & ~ \| f(A,x)_{j_1}^2 \cdot \langle c(A,x), f(A,x) \rangle^2 \cdot {\bf 1}_d  \cdot  h(A,x)^\top  -f(\wt{A},x)_{j_1}^2 \cdot \langle c(\wt{A},x), f(\wt{A},x) \rangle^2 \cdot {\bf 1}_d  \cdot  h(\wt{A},x)^\top) \|_F \\
      \leq & ~ | f(A,x)_{j_1} - f(\wt{A},x)_{j_1}  | | f(A,x)_{j_1} | \cdot \| f_c(A,x) \|_2^2 \cdot {\bf 1}_d  \cdot  \| h(A,x)^\top \|_2 \\
      & + ~ | f(\wt{A},x)_{j_1}  | | f(A,x)_{j_1} - f(\wt{A},x)_{j_1} | \cdot \|f_c(A,x)\|_2 \cdot \| {\bf 1}_d \|_2 \cdot  \| h(A,x)^\top  \|_2\\
      & + ~ | f(\wt{A},x)_{j_1}  |^2 \cdot \|f_c(A,x) -f_c(\wt{A},x) \|_2 |f_c(A,x)\|_2 \cdot \| {\bf 1}_d \|_2 \cdot \| h(A,x)^\top  \|_2\\
     & + ~ | f(\wt{A},x)_{j_1}  |^2 \cdot \|f_c(\wt{A},x)\|_2  \|f_c(A,x) -f_c(\wt{A},x) \|_2 \cdot \| {\bf 1}_d \|_2 \cdot \| h(A,x)^\top  \|_2\\
    & + ~ | f(\wt{A},x)_{j_1}  |^2 \cdot \|f_c(\wt{A},x)\|_2^2 \cdot \| {\bf 1}_d \|_2 \cdot \| h(A,x)^\top - h(\wt{A},x)^\top \|_2\\
    \leq & ~ 8 R^2 \cdot \beta^{-2} \cdot n \cdot \sqrt{d}  \cdot \exp(3R^2)\|A - \wt{A}\|_F \\
    & + ~ 12 R^2 \cdot \beta^{-2} \cdot n \cdot \sqrt{d}  \cdot \exp(3R^2)\|A - \wt{A}\|_F\\
    & + ~ 12 R^2 \cdot \beta^{-2} \cdot n \cdot \sqrt{d}  \cdot \exp(3R^2)\|A - \wt{A}\|_F\\ 
    & + ~12\beta^{-2} \cdot n \cdot \sqrt{d} \cdot \exp(4R^2) \cdot \|A - \wt{A}\|_F \\
   \leq &  ~\beta^{-2} \cdot n \cdot \sqrt{d} \cdot \exp(5R^2) \cdot \|A - \wt{A}\|_F \\
\end{align*}

{\bf Proof of Part 5.}
\begin{align*}
    & ~ \| B_{1,1,5}^{j_1,*,j_0,*} (A) - B_{1,1,5}^{j_1,*,j_0,*} ( \wt{A} ) \|_F \\
    = & ~ \|f(A,x)_{j_1} \cdot f(A,x)_{j_0} \cdot f_c(A,x)  \cdot (-f_2(A,x) + f(A,x)_{j_1})   \cdot {\bf 1}_d  \cdot h(A,x)^\top\\
    & ~ - f(\wt{A},x)_{j_1} \cdot f(\wt{A},x)_{j_0} \cdot f_c(\wt{A},x)  \cdot (-f_2(\wt{A},x) + f(\wt{A},x)_{j_1})   \cdot {\bf 1}_d  \cdot h(\wt{A},x)^\top \|_F \\
    \leq & ~ |f(A,x)_{j_1} -f(\wt{A},x)_{j_1} |  \cdot |f(A,x)_{j_0}| \cdot |f_c(A,x)| (| -f_2(A,x)| + |f(A,x)_{j_1}|)   \cdot \|{\bf 1}_d \|_2 \cdot \|h(A,x)^\top  \|_2 \\
    & + ~ |f(\wt{A},x)_{j_1} |  \cdot |f(A,x)_{j_0} - f(\wt{A},x)_{j_0}| \cdot |f_c(A,x)|  (| -f_2(A,x)| + |f(A,x)_{j_1}|)  \cdot \|{\bf 1}_d \|_2 \cdot \|h(A,x)^\top \|_2 \\
     & + ~ |f(\wt{A},x)_{j_1} |  \cdot |f(\wt{A},x)_{j_0}| \cdot |f_c(A,x) - f_c(\wt{A},x)| (| -f_2(A,x)| + |f(A,x)_{j_1}|)  \cdot \|{\bf 1}_d \|_2 \cdot \|h(A,x)^\top \|_2 \\
      & + ~ |f(\wt{A},x)_{j_1} |  \cdot |f(\wt{A},x)_{j_0}| \cdot |f_c(\wt{A},x)| (| f_2(A,x) -f_2(\wt{A},x)| + |f(A,x)_{j_1} - f(\wt{A},x)_{j_1}| )  \\ 
      & ~ \cdot \|{\bf 1}_d \|_2 \cdot \|h(A,x)^\top \|_2 \\
       & + ~ |f(\wt{A},x)_{j_1} |  \cdot |f(\wt{A},x)_{j_0}| \cdot |f_c(\wt{A},x)| (| -f_2(A,x)| + |f(A,x)_{j_1}|)  \cdot \|{\bf 1}_d \|_2 \cdot \|h(A,x)^\top - h(\wt{A},x)^\top \|_2 \\
    \leq & ~  8R^2 \cdot  \beta^{-2} \cdot n \cdot \sqrt{d}  \cdot \exp(3R^2)\|A - \wt{A}\|_F \\
    & + ~  8R^2 \cdot  \beta^{-2} \cdot n \cdot \sqrt{d}  \cdot \exp(3R^2)\|A - \wt{A}\|_F \\
     & + ~12R^2 \cdot  \beta^{-2} \cdot n \cdot \sqrt{d}  \cdot \exp(3R^2)\|A - \wt{A}\|_F \\
     & + ~12 R^2 \cdot  \beta^{-2} \cdot n \cdot \sqrt{d}  \cdot \exp(3R^2)\|A - \wt{A}\|_F \\
     & + ~ 16 \beta^{-2} \cdot n \cdot \exp(4R^2) \cdot \|A - \wt{A}\|_F \\
     \leq & ~ 56 \beta^{-2} \cdot n \cdot \sqrt{d}  \cdot \exp(4R^2) \cdot \|A - \wt{A}\|_F \\
     \leq & ~  \beta^{-2} \cdot n \cdot \sqrt{d}  \cdot \exp(5R^2) \cdot \|A - \wt{A}\|_F
\end{align*}

{\bf Proof of Part 6.}
\begin{align*}
& ~ \|B_{1,1,6}^{j_1,*,j_0,*} (A) - B_{1,1,6}^{j_1,*,j_0,*} ( \wt{A} )  \|_F\\
   \leq & ~ \| f(A,x)_{j_1} \cdot f(A,x)_{j_0}  \cdot(-f_c(A,x) + f(A,x)_{j_1}) \cdot f_c(A,x)   \cdot {\bf 1}_d  \cdot h(A,x)^\top \\
   & ~-  f(\wt{A},x)_{j_1} \cdot f(\wt{A},x)_{j_0}  \cdot(-f_c(\wt{A},x) + f(\wt{A},x)_{j_1}) \cdot f_c(\wt{A},x)   \cdot {\bf 1}_d  \cdot h(\wt{A},x)^\top\|_F \\
   \leq & ~ |f(A,x)_{j_1} - f(\wt{A},x)_{j_1}| \cdot |f(A,x)_{j_0}|  \cdot(|-f_c(A,x)| +| f(A,x)_{j_1}|) \cdot |f_c(A,x)|   \cdot \|{\bf 1}_d\|_2  \cdot \|h(A,x)^\top\|_2 \\
   & + ~ |f(\wt{A},x)_{j_1}| \cdot |f(A,x)_{j_0} - f(\wt{A},x)_{j_0}|  \cdot(|-f_c(A,x)| +| f(A,x)_{j_1}|) \cdot |f_c(A,x)|   \cdot \|{\bf 1}_d\|_2  \cdot \|h(A,x)^\top\|_2 \\
   & + ~ |f(\wt{A},x)_{j_1}| \cdot |f(\wt{A},x)_{j_0}|  \cdot(|f_c(A,x) - f_c(\wt{A},x) | +| f(A,x)_{j_1}-f(\wt{A},x)_{j_1}|) \cdot |f_c(A,x)| \\
   & ~ \cdot \|{\bf 1}_d\|_2  \cdot \|h(A,x)^\top\|_2 \\
   & + ~ |f(\wt{A},x)_{j_1}| \cdot |f(\wt{A},x)_{j_0}|  \cdot(|-f_c(\wt{A},x)| +| f(\wt{A},x)_{j_1}|) \cdot |f_c(A,x) -f_c(\wt{A},x)  |   \cdot \|{\bf 1}_d\|_2  \cdot \|h(A,x)^\top\|_2 \\
   & + ~ |f(\wt{A},x)_{j_1}| \cdot |f(\wt{A},x)_{j_0}|  \cdot(|-f_c(\wt{A},x)| +| f(\wt{A},x)_{j_1}|) \cdot f_c(\wt{A},x)  |   \cdot \|{\bf 1}_d\|_2  \cdot \|h(A,x)^\top -h(\wt{A},x)^\top \|_2 \\
   \leq & ~ 12 R^2 \cdot  \beta^{-2} \cdot n \cdot \sqrt{d}  \cdot \exp(3R^2)\|A - \wt{A}\|_F \\
     & + ~  12 R^2 \cdot  \beta^{-2} \cdot n \cdot \sqrt{d}  \cdot \exp(3R^2)\|A - \wt{A}\|_F \\
     & + ~  16 R^2 \cdot  \beta^{-2} \cdot n \cdot \sqrt{d}  \cdot \exp(3R^2)\|A - \wt{A}\|_F \\
     & + ~  18 R^2 \cdot  \beta^{-2} \cdot n \cdot \sqrt{d}  \cdot \exp(3R^2)\|A - \wt{A}\|_F \\
     & + ~  24 \cdot  \beta^{-2} \cdot n \cdot \sqrt{d}  \cdot \exp(4R^2)\|A - \wt{A}\|_F \\
     \leq & ~ 70\beta^{-2} \cdot n \cdot \sqrt{d}  \cdot \exp(4R^2)\|A - \wt{A}\|_F \\
     \leq & ~ \beta^{-2} \cdot n \cdot \sqrt{d}  \cdot \exp(5R^2)\|A - \wt{A}\|_F
\end{align*}

{\bf Proof of Part 7.}
\begin{align*}
& ~ \|B_{1,1,7}^{j_1,*,j_0,*} (A) - B_{1,1,7}^{j_1,*,j_0,*} ( \wt{A} )   \|_F 
\\
  \leq & ~   \|f(A,x)_{j_1} \cdot f(A,x)_{j_0} \cdot f_c(A,x) \cdot {\bf 1}_d  \cdot p_{j_1}(A,x)^\top  - f(\wt{A},x)_{j_1} \cdot f(\wt{A},x)_{j_0} \cdot f_c(\wt{A},x) \cdot {\bf 1}_d  \cdot p_{j_1}(\wt{A},x)^\top  \|_F\\
  \leq & ~ |f(A,x)_{j_1} - f(\wt{A},x)_{j_1}| \cdot |f(A,x)_{j_0} |\cdot |f_c(A,x)| \cdot \|{\bf 1}_d\|_2 \cdot \|p_{j_1}(A,x)^\top\|_2 \\
  & + ~ | f(\wt{A},x)_{j_1}| \cdot |f(A,x)_{j_0} - f(\wt{A},x)_{j_0} |\cdot |f_c(A,x)| \cdot \|{\bf 1}_d\|_2  \cdot \|p_{j_1}(A,x)^\top\|_2 \\
   & + ~ | f(\wt{A},x)_{j_1}| \cdot |f(\wt{A},x)_{j_0} |\cdot |f_c(A,x)- f_c(\wt{A},x)| \cdot \|{\bf 1}_d\|_2  \cdot \|p_{j_1}(A,x)^\top\|_2 \\
   & + ~ | f(\wt{A},x)_{j_1}| \cdot |f(\wt{A},x)_{j_0} |\cdot  |f_c(\wt{A},x)| \cdot \|{\bf 1}_d\|_2 \cdot \|p_{j_1}(A,x)^\top - p_{j_1}(\wt{A},x)^\top\|_2 \\
   \leq & ~ 24R^2 \cdot  \beta^{-2} \cdot n \cdot \sqrt{d}  \cdot \exp(3R^2)\|A - \wt{A}\|_F \\
   & + ~24R^2 \cdot  \beta^{-2} \cdot n \cdot \sqrt{d}  \cdot \exp(3R^2)\|A - \wt{A}\|_F \\
   & + ~36R^2 \cdot  \beta^{-2} \cdot n \cdot \sqrt{d}  \cdot \exp(3R^2)\|A - \wt{A}\|_F \\
   & + ~26 \beta^{-2} \cdot n   \cdot \exp(4R^2) \cdot \|A - \wt{A} \|_F \\
   \leq & ~ 110 \beta^{-2} \cdot n  \cdot \sqrt{d} \cdot \exp(4R^2) \cdot \|A - \wt{A} \|_F\\
   \leq & ~ \beta^{-2} \cdot n \cdot \sqrt{d} \cdot \exp(5R^2) \cdot \|A - \wt{A} \|_F
\end{align*}

{\bf Proof of Part 8.}
\begin{align*}
    & ~ \|B_{1,2,1}^{j_1,*,j_0,*} (A) - B_{1,2,1}^{j_1,*,j_0,*} ( \wt{A} )   \|_F \\
    \leq & ~ \| f(A,x)_{j_1} \cdot  f(A,x)_{j_0} \cdot  f_c(A,x) \cdot x \cdot c_g(A,x)^{\top} -  f(\wt{A},x)_{j_1} \cdot  f(\wt{A},x)_{j_0} \cdot  f_c(\wt{A},x) \cdot x \cdot c_g(\wt{A},x)^{\top} \|_F\\
    \leq & ~ |f(A,x)_{j_1} -f(\wt{A},x)_{j_1}| \cdot  |f(A,x)_{j_0} |\cdot  |f_c(A,x) |\cdot \| x\|_2 \cdot \|c_g(A,x)^{\top}\|_2 \\
    \leq & ~ |f(\wt{A},x)_{j_1}| \cdot  |f(A,x)_{j_0} - f(\wt{A},x)_{j_0} |\cdot  |f_c(A,x) |\cdot \| x\|_2 \cdot \|c_g(A,x)^{\top}\|_2 \\
    \leq & ~ |f(\wt{A},x)_{j_1}| \cdot  |f(\wt{A},x)_{j_0} |\cdot  |f_c(A,x) - f_c(\wt{A},x) |\cdot \| x\|_2 \cdot \|c_g(A,x)^{\top}\|_2 \\
    \leq & ~ |f(\wt{A},x)_{j_1}| \cdot  |f(\wt{A},x)_{j_0} |\cdot  |f_c(\wt{A},x) |\cdot \| x\|_2 \cdot \|c_g(A,x)^{\top} - c_g(\wt{A},x)^{\top}\|_2 \\
    \leq & ~ 20 R^2 \cdot  \beta^{-2} \cdot n  \cdot \exp(3R^2)\|A - \wt{A}\|_F \\
     & + ~ 20 R^2 \cdot  \beta^{-2} \cdot n \cdot   \exp(3R^2)\|A - \wt{A}\|_F \\
     & + ~ 30 R^2 \cdot  \beta^{-2} \cdot n \cdot  \exp(3R^2)\|A - \wt{A}\|_F \\
     & + ~ 40 R^2 \beta^{-2} \cdot n \cdot  \exp(3R^{2}) \cdot \|A - \wt{A}\|_F\\
     \leq & ~ 110  \cdot  \beta^{-2} \cdot n  \cdot \exp(4R^2)\|A - \wt{A}\|_F \\
     \leq & ~  \beta^{-2} \cdot n  \cdot \exp(5R^2)\|A - \wt{A}\|_F \\
\end{align*}

{\bf Proof of Part 9.}
\begin{align*}
     & ~ \| B_{1,3,1}^{j_1,*,j_0,*}(A) -   B_{1,3,1}^{j_1,*,j_0,*}(\wt{A}) \|_F\\
     = & ~  \| f(A,x)_{j_1} \cdot  f(A,x)_{j_0}   \cdot f_2(A,x) \cdot x \cdot  c_g(A,x)^{\top} - f(\wt{A},x)_{j_1} \cdot  f(\wt{A},x)_{j_0}   \cdot f_2(\wt{A},x) \cdot x \cdot  c_g(\wt{A},x)^{\top}  \|_F \\
     \leq & ~ |f(A,x)_{j_1} - f(\wt{A},x)_{j_1}| \cdot  |f(A,x)_{j_0} |  \cdot |f_2(A,x) |\cdot \|x\|_2 \cdot \| c_g(A,x)^{\top}\|_2 \\
     & + ~ |f(\wt{A},x)_{j_1}| \cdot  |f(A,x)_{j_0} -f(\wt{A},x)_{j_0} |  \cdot |f_2(A,x) |\cdot \|x\|_2 \cdot \| c_g(A,x)^{\top}\|_2 \\
      & + ~ |f(\wt{A},x)_{j_1}| \cdot  |f(\wt{A},x)_{j_0} |  \cdot |f_2(A,x) -  f_2(\wt{A},x)|\cdot \|x\|_2 \cdot \| c_g(A,x)^{\top}\|_2 \\
      & + ~ |f(\wt{A},x)_{j_1}| \cdot  |f(\wt{A},x)_{j_0} |  \cdot | f_2(\wt{A},x)|\cdot \|x\|_2 \cdot \| c_g(A,x)^{\top} - c_g(\wt{A},x)^{\top}\|_2 \\
    \leq & ~ 10 R^2 \cdot  \beta^{-2} \cdot n  \cdot \exp(3R^2)\|A - \wt{A}\|_F \\
     & + ~ 10 R^2 \cdot  \beta^{-2} \cdot n \cdot   \exp(3R^2)\|A - \wt{A}\|_F \\
     & + ~ 20 R^2 \cdot  \beta^{-2} \cdot n \cdot  \exp(3R^2)\|A - \wt{A}\|_F \\
     & + ~ 20 R^2 \beta^{-2} \cdot n \cdot  \exp(3R^{2}) \cdot \|A - \wt{A}\|_F\\
     \leq & ~ 60  \cdot  \beta^{-2} \cdot n  \cdot \exp(4R^2)\|A - \wt{A}\|_F \\
     \leq & ~  \beta^{-2} \cdot n  \cdot \exp(5R^2)\|A - \wt{A}\|_F \\
\end{align*}

{\bf Proof of Part 10.}
\begin{align*}
     & ~ \| B_{1,3,2}^{j_1,*,j_0,*}(A) -   B_{1,3,2}^{j_1,*,j_0,*}(\wt{A}) \|_F\\
     \leq & ~ \|  -   f(A,x)_{j_1}^2 \cdot  f(A,x)_{j_0} \cdot x  \cdot c_g(A,x)^{\top} -(  -   f(\wt{A},x)_{j_1}^2 \cdot  f(\wt{A},x)_{j_0} \cdot x  \cdot c_g(\wt{A},x)^{\top})\|_F \\
     \leq & ~ \|     f(A,x)_{j_1}^2 \cdot  f(A,x)_{j_0} \cdot x  \cdot c_g(A,x)^{\top}   -   f(\wt{A},x)_{j_1}^2 \cdot  f(\wt{A},x)_{j_0} \cdot x  \cdot c_g(\wt{A},x)^{\top}\|_F \\
     \leq & ~  |f(A,x)_{j_1} - f(\wt{A},x)_{j_1} | |f(A,x)_{j_1}|  \cdot  |f(A,x)_{j_0} |\cdot \|x\|_2  \cdot \|c_g(A,x)^{\top}\|_2 \\
     & + ~  | f(\wt{A},x)_{j_1} | |f(A,x)_{j_1} - f(\wt{A},x)_{j_1}|  \cdot  |f(A,x)_{j_0} |\cdot\|x\|_2  \cdot \|c_g(A,x)^{\top}\|_2\\
     & + ~  | f(\wt{A},x)_{j_1} | |f(\wt{A},x)_{j_1}|  \cdot  |f(A,x)_{j_0} -f(\wt{A},x)_{j_0}|\cdot\|x\|_2  \cdot \|c_g(A,x)^{\top}\|_2 \\
     & + ~  | f(\wt{A},x)_{j_1} | |f(\wt{A},x)_{j_1}|  \cdot  |f(\wt{A},x)_{j_0}|\cdot\|x\|_2  \cdot \|c_g(A,x)^{\top} - c_g(\wt{A},x)^{\top}\|_2 \\
     \leq & ~ 10 R^2 \cdot  \beta^{-2} \cdot n  \cdot \exp(3R^2)\|A - \wt{A}\|_F \\
     & + ~ 10 R^2 \cdot  \beta^{-2} \cdot n  \cdot \exp(3R^2)\|A - \wt{A}\|_F \\
     & + ~ 10 R^2 \cdot  \beta^{-2} \cdot n  \cdot \exp(3R^2)\|A - \wt{A}\|_F \\
     & + ~ 20 R^2 \beta^{-2} \cdot n \cdot  \exp(3R^{2}) \cdot \|A - \wt{A}\|_F\\
     \leq & ~ 50 \beta^{-2} \cdot n \cdot  \exp(4R^{2}) \cdot \|A - \wt{A}\|_F \\
     \leq & ~  \beta^{-2} \cdot n \cdot  \exp(5R^{2}) \cdot \|A - \wt{A}\|_F
\end{align*}

{\bf Proof of Part 11.}
\begin{align*}
    & ~ \| B_{1,4,1}^{j_1,*,j_0,*}(A) -   B_{1,4,1}^{j_1,*,j_0,*}(\wt{A}) \|_F\\
    = & ~ \| f(A,x)_{j_1} \cdot  f(A,x)_{j_0}  \cdot f_c(A,x)  \cdot x \cdot c_g(A,x)^{\top} -  f(\wt{A},x)_{j_1} \cdot  f(\wt{A},x)_{j_0}  \cdot f_c(\wt{A},x)  \cdot x \cdot c_g(\wt{A},x)^{\top} \|_2\\
    \leq & ~  |f(A,x)_{j_1} - f(\wt{A},x)_{j_1}| \cdot  |f(A,x)_{j_0}|  \cdot| f_c(A,x)|  \cdot \|x\|_2 \cdot \|c_g(A,x)^{\top}\|_2 \\
     & + ~|f(\wt{A},x)_{j_1}| \cdot  |f(A,x)_{j_0} - f(\wt{A},x)_{j_0}|  \cdot| f_c(A,x)|  \cdot \|x\|_2 \cdot \|c_g(A,x)^{\top}\|_2 \\
    & +  ~|f(\wt{A},x)_{j_1}| \cdot  |f(\wt{A},x)_{j_0}|  \cdot| f_c(A,x) - f_c(\wt{A},x)|  \cdot \|x\|_2 \cdot \|c_g(A,x)^{\top}\|_2 \\
    & +  ~|f(\wt{A},x)_{j_1}| \cdot  |f(\wt{A},x)_{j_0}|  \cdot| f_c(A,x) - f_c(\wt{A},x)|  \cdot \|x\|_2 \cdot \|c_g(A,x)^{\top}\|_2 \\
    & +  ~|f(\wt{A},x)_{j_1}| \cdot  |f(\wt{A},x)_{j_0}|  \cdot| f_c(\wt{A},x)|  \cdot \|x\|_2 \cdot \|c_g(A,x)^{\top} - c_g(\wt{A},x)^{\top}\|_2 \\
    \leq & ~ 10 R^2 \cdot  \beta^{-2} \cdot n  \cdot \exp(3R^2)\|A - \wt{A}\|_F \\
     & + ~ 10 R^2 \cdot  \beta^{-2} \cdot n  \cdot \exp(3R^2)\|A - \wt{A}\|_F \\
     & + ~ 30 R^2 \cdot  \beta^{-2} \cdot n  \cdot \exp(3R^2)\|A - \wt{A}\|_F \\
    & + ~ 40 R^2 \cdot  \beta^{-2} \cdot n  \cdot \exp(3R^2)\|A - \wt{A}\|_F \\
     \leq & ~ 90 \cdot  \beta^{-2} \cdot n  \cdot \exp(4R^2)\|A - \wt{A}\|_F \\
      \leq & ~ \beta^{-2} \cdot n  \cdot \exp(5R^2)\|A - \wt{A}\|_F \\
\end{align*}

{\bf Proof of Part 12.}
\begin{align*}
    & ~ \| B_{1,4,2}^{j_1,*,j_0,*}(A) -   B_{1,4,2}^{j_1,*,j_0,*}(\wt{A}) \|_F\\
    \leq & ~ \|f(A,x)_{j_1} \cdot  f(A,x)_{j_0}  \cdot c(A,x)_{j_1} \cdot  x \cdot c_g(A,x)^{\top}  - f(\wt{A},x)_{j_1} \cdot  f(\wt{A},x)_{j_0}  \cdot c(\wt{A},x)_{j_1} \cdot  x \cdot c_g(\wt{A},x)^{\top}  \|_F \\
    \leq & ~ |f(A,x)_{j_1} - f(\wt{A},x)_{j_1}| \cdot  |f(A,x)_{j_0} | \cdot |c(A,x)_{j_1}| \cdot \| x \|_2\cdot  \| c_g(A,x)^{\top} \|_2 \\
     & + ~ |f(\wt{A},x)_{j_1}| \cdot  |f(A,x)_{j_0} - f(\wt{A},x)_{j_0}| \cdot |c(A,x)_{j_1}| \cdot \| x \|_2\cdot  \| c_g(A,x)^{\top} \|_2 \\
     & + ~ |f(\wt{A},x)_{j_1}| \cdot  |f(\wt{A},x)_{j_0}| \cdot |c(A,x)_{j_1} - c(\wt{A},x)_{j_1}| \cdot \| x \|_2\cdot  \| c_g(A,x)^{\top} \|_2 \\
    & + ~ |f(\wt{A},x)_{j_1}| \cdot  |f(\wt{A},x)_{j_0}| \cdot | c(\wt{A},x)_{j_1}| \cdot \| x \|_2\cdot  \| c_g(A,x)^{\top} -   c_g(\wt{A},x)^{\top} \|_2 \\
    \leq & ~ 20 R^2 \cdot  \beta^{-2} \cdot n  \cdot \exp(3R^2)\|A - \wt{A}\|_F \\
    & + ~ 20 R^2 \cdot  \beta^{-2} \cdot n  \cdot \exp(3R^2)\|A - \wt{A}\|_F \\
    & + ~ 10 R^2 \cdot  \beta^{-2} \cdot n  \cdot \exp(3R^2)\|A - \wt{A}\|_F \\
    & + ~ 40 R^2 \cdot  \beta^{-2} \cdot n  \cdot \exp(3R^2)\|A - \wt{A}\|_F \\
    \leq & ~ 90   \beta^{-2} \cdot n  \cdot \exp(4R^2)\|A - \wt{A}\|_F \\
    \leq & ~    \beta^{-2} \cdot n  \cdot \exp(5R^2)\|A - \wt{A}\|_F \\
\end{align*}
{\bf Proof of Part 13}
\begin{align*}
    & ~ \| B_{1}^{j_1,*,j_0,*}(A) -   B_{1}^{j_1,*,j_0,*}(\wt{A})  \|_F \\
    = & ~ \|\sum_{i = 1}^4 B_{1,i}^{j_1,*,j_0,*}(A) -   B_{1,i}^{j_1,*,j_0,*}(\wt{A})  \|_F \\
    \leq & ~ 12  \beta^{-2} \cdot n \cdot \sqrt{d} \cdot \exp(5R^2)\|A - \wt{A}\|_F
\end{align*} 
\end{proof}
\subsection{PSD}

\begin{lemma}\label{psd: B_1}
If the following conditions hold
\begin{itemize}
    \item Let $B_{1,1,1}^{j_1,*, j_0,*}, \cdots, B_{1,4,2}^{j_1,*, j_0,*} $ be defined as Lemma~\ref{lem:b_1_j1_j0} 
    \item  Let $\|A \|_2 \leq R, \|A^{\top} \|_F \leq R, \| x\|_2 \leq R, \|\diag(f(A,x)) \|_F \leq \|f(A,x) \|_2 \leq 1, \| b_g \|_2 \leq 1$ 
\end{itemize}
We have 
\begin{itemize}
    \item {\bf Part 1.} $\| B_{1,1,1}^{j_1,*, j_0,*} \| \leq 4 $ 
\item {\bf Part 2.} $\|B_{1,1,2}^{j_1,*, j_0,*}\| \leq  4  $
    \item {\bf Part 3.} $\|B_{1,1,3}^{j_1,*, j_0,*}\| \leq  4 \sqrt{d}  R^2$
    \item {\bf Part 4.} $\|B_{1,1,4}^{j_1,*, j_0,*} \|\leq  4 \sqrt{d}R^2$
    \item {\bf Part 5.} $\|B_{1,1,5}^{j_1,*, j_0,*} \|\leq 4 \sqrt{d} R^2$
    \item {\bf Part 6.} $\|B_{1,1,6}^{j_1,*, j_0,*} \|\leq  6 \sqrt{d} R^2$
    \item {\bf Part 7.} $\|B_{1,1,7}^{j_1,*, j_0,*} \|\leq  12 \sqrt{d} R^2$
    \item {\bf Part 8.} $\|B_{1,2,1}^{j_1,*, j_0,*}\| \leq  10 R^2$
    \item {\bf Part 9.} $\|B_{1,3,1}^{j_1,*, j_0,*}\| \leq  5 R^2$
    \item {\bf Part 10.} $\|B_{1,3,2}^{j_1,*, j_0,*}\| \leq  5 R^2$
    \item {\bf Part 11.} $\|B_{1,4,1}^{j_1,*, j_0,*} \|\leq 10 R^2$
    \item {\bf Part 12.} $\|B_{1,4,2}^{j_1,*, j_0,*}\| \leq 10 R^2$
    \item {\bf Part 13.} $\|B_{1}^{j_1,*, j_0,*}\| \leq 78\sqrt{d} R^2$
\end{itemize}
\end{lemma}

\begin{proof}
    {\bf Proof of Part 1.}
    \begin{align*}
        \| B_{1,1,1}^{j_1,*, j_0,*} \| 
        = & ~ \| f(A,x)_{j_1} \cdot f(A,x)_{j_0} \cdot  f_c(A,x)^2 \cdot I_d \| \\
        \leq & ~ |f(A,x)_{j_1}| \cdot |f(A,x)_{j_0}| \cdot  |f_c(A,x)|^2 \cdot \|I_d \| \\
        \leq & ~ 4  
    \end{align*}

    {\bf Proof of Part 2.}
    \begin{align*}
        \| B_{1,1,2}^{j_1,*, j_0,*} \|
        = &~
        \| f(A,x)_{j_1} \cdot f(A,x)_{j_0} \cdot c(A,x)_{j_1}\cdot  f_c(A,x) \cdot  I_d \| \\
        \preceq & ~  |f(A,x)_{j_1}| \cdot |f(A,x)_{j_0}| \cdot |c(A,x)_{j_1}|\cdot  |f_c(A,x)| \cdot  \| I_d\| \\
        \preceq & ~ 4  
    \end{align*}

    {\bf Proof of Part 3.}
    \begin{align*}
     \| B_{1,1,3}^{j_1,*, j_0,*} \|
     = & ~
       \| f(A,x)_{j_1} \cdot f(A,x)_{j_0} \cdot f_c(A,x)^2 \cdot \| {\bf 1}_d \cdot ( (A_{j_1,*}) \circ x^\top  ) \| \\
    \leq & ~ |f(A,x)_{j_1}| \cdot |f(A,x)_{j_0}| \cdot |
f_c(A,x)|^2  \cdot \| {\bf 1}_d \|_2 \cdot \| A_{j_1,*}^\top \circ x \|_2 \\
    \leq  & ~ 4\sqrt{d}  \| A_{j_1,*}\|_2 \|\diag(x) \|_{\infty}\\
    \leq &  ~  4R^2\sqrt{d}
    \end{align*}
   
    {\bf Proof of Part 4.}
    \begin{align*}
        \|B_{1,1,4}^{j_1,*, j_0,*}  \|
        = & ~  \| f(A,x)_{j_1} \cdot f(A,x)_{j_0} \cdot f_c(A,x)^2 \cdot {\bf 1}_d  \cdot h(A,x)^\top\|\\
        \leq & ~ | f(A,x)_{j_1} | | \cdot f(A,x)_{j_0} |\cdot |f_c(A,x)|^2 \cdot \|{\bf 1}_d \|_2 \cdot \| h(A,x)^\top \|_2 \\
        \leq & ~ 4 \sqrt{d}R^2
    \end{align*}

    {\bf Proof of Part 5.}
    \begin{align*}
        \|B_{1,1,5}^{j_1,*, j_0,*} \|
        = & ~ \| f(A,x)_{j_1} \cdot f(A,x)_{j_0} \cdot f_c(A,x)  \cdot (-f_2(A,x) + f(A,x)_{j_1})   \cdot {\bf 1}_d  \cdot h(A,x)^\top\| \\
        \leq & ~  | f(A,x)_{j_1}| \cdot |f(A,x)_{j_0}| \cdot |f_c(A,x)|  \cdot (|-f_2(A,x)| + |f(A,x)_{j_1}|)   \cdot \| {\bf 1}_d \|_2  \cdot \|h(A,x)^\top\|_2 \\
        \leq & ~ 4 \sqrt{d} R^2
    \end{align*}

    {\bf Proof of Part 6.}
    \begin{align*}
        \|B_{1,1,6}^{j_1,*, j_0,*} \|
        = & ~ \| f(A,x)_{j_1} \cdot f(A,x)_{j_0}  \cdot(-f_c(A,x) + f(A,x)_{j_1}) \cdot f_c(A,x)   \cdot {\bf 1}_d  \cdot h(A,x)^\top \|\\
        \leq & ~  | f(A,x)_{j_1}| \cdot |f(A,x)_{j_0}| \cdot |f_c(A,x)|  \cdot (|-f_c(A,x)| + |f(A,x)_{j_1}|)   \cdot \| {\bf 1}_d \|_2  \cdot \|h(A,x)^\top\|_2 \\
        \leq & ~ 6 \sqrt{d} R^2
    \end{align*}

    {\bf Proof of Part 7.}
    \begin{align*}
         \|B_{1,1,7}^{j_1,*, j_0,*} \|
         = & ~ \|f(A,x)_{j_1} \cdot f(A,x)_{j_0} \cdot f_c(A,x) \cdot {\bf 1}_d  \cdot p_{j_1}(A,x)^\top  \|\\
         \leq & ~ | f(A,x)_{j_1}| \cdot |f(A,x)_{j_0}| \cdot |f_c(A,x)|\cdot \| {\bf 1}_d \|_2 \cdot \|p_{j_1}(A,x)^\top\|_2 \\
         \leq & ~ 12 \sqrt{d} R^2
    \end{align*}

    {\bf Proof of Part 8.}
    \begin{align*}
        \|B_{1,2,1}^{j_1,*, j_0,*} \|
        = & ~ \| f(A,x)_{j_1} \cdot  f(A,x)_{j_0} \cdot  f_c(A,x) \cdot x \cdot c_g(A,x)^{\top} \| \\
        \leq & ~  | f(A,x)_{j_1}| \cdot |f(A,x)_{j_0}| \cdot |f_c(A,x)| \cdot \|x\|_2 \cdot \|c_g(A,x)^{\top}\|_2\\
        \leq & ~ 10 R^2
    \end{align*}

    {\bf Proof of Part 9.}
    \begin{align*}
     \|B_{1,3,1}^{j_1,*, j_0,*} \|
        = & ~ \| f(A,x)_{j_1} \cdot  f(A,x)_{j_0}   \cdot f_2(A,x) \cdot x \cdot  c_g(A,x)^{\top}\| \\
        \leq & ~ | f(A,x)_{j_1}| \cdot  |f(A,x)_{j_0} |  \cdot |f_2(A,x)| \cdot \|x\|_2 \cdot  \|c_g(A,x)^{\top}|\|_2 \\
        \leq & ~ 5 R^2
    \end{align*}

    {\bf Proof of Part 10.}
    \begin{align*}
        \|B_{1,3,2}^{j_1,*, j_0,*} \|
        = & ~ \| -   f(A,x)_{j_1}^2 \cdot  f(A,x)_{j_0} \cdot x  \cdot c_g(A,x)^{\top} \|\\
        \leq & ~ |f(A,x)_{j_1} |^2 \cdot  |f(A,x)_{j_0}| \cdot \| x\|_2 \cdot \| c_g(A,x)^{\top}\|_2 \\
        \leq & ~ 5 R^2
    \end{align*}

    {\bf Proof of Part 11.}
    \begin{align*}
     \|B_{1,4,1}^{j_1,*, j_0,*} \|
        =- & ~ f(A,x)_{j_1} \cdot  f(A,x)_{j_0}  \cdot f_c(A,x)  \cdot x \cdot c_g(A,x)^{\top} \\
        \leq & ~ |f(A,x)_{j_1}| \cdot  |f(A,x)_{j_0}|  \cdot |f_c(A,x)|  \cdot \|x \|_2 \cdot \| c_g(A,x)^{\top}\|_2 \\
        \leq & ~ 10 R^2
    \end{align*}

    {\bf Proof of Part 12.}
    \begin{align*}
        \|B_{1,4,2}^{j_1,*, j_0,*} \| 
        = & ~ \| -   f(A,x)_{j_1} \cdot  f(A,x)_{j_0}  \cdot c(A,x)_{j_1} \cdot  x \cdot c_g(A,x)^{\top} \|\\
        \leq & ~  |f(A,x)_{j_1}| \cdot  |f(A,x)_{j_0}|  \cdot |c(A,x)_{j_1}|  \cdot \|x \|_2 \cdot \| c_g(A,x)^{\top}\|_2 \\
        \leq & ~  10 R^2
    \end{align*}

    {\bf Proof of Part 13}
\begin{align*}
    & ~ \| B_{1}^{j_1,*,j_0,*}\| \\
    = & ~ \|\sum_{i = 1}^4 B_{1,i}^{j_1,*,j_0,*} \| \\
    \leq & ~ 78\sqrt{d} R^2
\end{align*} 
\end{proof}
\newpage
\section{Hessian: Second term  \texorpdfstring{$B_2^{j_1,i_1,j_0,i_0}$}{}}\label{app:hessian_second}
\subsection{Definitions}
\begin{definition}\label{def:b_2}
    We define the $B_2^{j_1,i_1,j_0,i_0}$ as follows,
    \begin{align*}
        B_2^{j_1,i_1,j_0,i_0} := & ~ \frac{\d}{\d A_{j_1,i_1}}(- c_g(A,x)^{\top} \cdot f(A,x)_{j_0} \cdot c(A,x)_{j_0} \cdot e_{i_0}   )
    \end{align*}
    Then, we define $B_{2,1}^{j_1,i_1,j_0,i_0}, \cdots, B_{2,3}^{j_1,i_1,j_0,i_0}$ as follow
    \begin{align*}
         B_{2,1}^{j_1,i_1,j_0,i_0} : = & ~ \frac{\d}{\d A_{j_1,i_1}} (- c_g(A,x)^{\top} ) \cdot  f(A,x)_{j_0}  \cdot c(A,x)_{j_0} \cdot e_{i_0}\\
 B_{2,2}^{j_1,i_1,j_0,i_0} : = & ~ - c_g(A,x)^{\top} \cdot \frac{\d}{\d A_{j_1,i_1}} ( f(A,x)_{j_0} )  \cdot c(A,x)_{j_0} \cdot e_{i_0}\\
 B_{2,3}^{j_1,i_1,j_0,i_0} : = & ~  -  c_g(A,x)^{\top} \cdot f(A,x)_{j_0}  \cdot \frac{\d }{\d A_{j_1,i_1}}( c(A,x)_{j_0}) \cdot e_{i_0}
    \end{align*}
    It is easy to show
    \begin{align*}
        B_2^{j_1,i_1,j_0,i_0} = B_{2,1}^{j_1,i_1,j_0,i_0} +  B_{2,2}^{j_1,i_1,j_0,i_0} + B_{2,3}^{j_1,i_1,j_0,i_0} 
    \end{align*}
    Similarly for $j_1 = j_0$ and $i_0 = i_1$,we have
    \begin{align*}
        B_2^{j_1,i_1,j_1,i_1} = B_{2,1}^{j_1,i_1,j_1,i_1} +  B_{2,2}^{j_1,i_1,j_1,i_1} + B_{2,3}^{j_1,i_1,j_1,i_1}  
         \end{align*}
    For $j_1 = j_0$ and $i_0 \neq i_1$,we have
    \begin{align*}
        B_2^{j_1,i_1,j_1,i_0} = B_{2,1}^{j_1,i_1,j_1,i_0} +  B_{2,2}^{j_1,i_1,j_1,i_0} + B_{2,3}^{j_1,i_1,j_1,i_0} 
    \end{align*}
    For $j_1 \neq j_0$ and $i_0 = i_1$,we have
    \begin{align*}
        B_2^{j_1,i_1,j_0,i_1} = B_{2,1}^{j_1,i_1,j_0,i_1} +  B_{2,2}^{j_1,i_1,j_0,i_1} + B_{2,3}^{j_1,i_1,j_0,i_1} 
    \end{align*}
\end{definition}

\subsection{Case \texorpdfstring{$j_1=j_0, i_1 = i_0$}{}}
\begin{lemma}
For $j_1 = j_0$ and $i_0 = i_1$. If the following conditions hold
    \begin{itemize}
     \item Let $u(A,x) \in \R^n$ be defined as Definition~\ref{def:u}
    \item Let $\alpha(A,x) \in \R$ be defined as Definition~\ref{def:alpha}
     \item Let $f(A,x) \in \R^n$ be defined as Definition~\ref{def:f}
    \item Let $c(A,x) \in \R^n$ be defined as Definition~\ref{def:c}
    \item Let $g(A,x) \in \R^d$ be defined as Definition~\ref{def:g} 
    \item Let $q(A,x) = c(A,x) + f(A,x) \in \R^n$
    \item Let $c_g(A,x) \in \R^d$ be defined as Definition~\ref{def:c_g}.
    \item Let $L_g(A,x) \in \R$ be defined as Definition~\ref{def:l_g}
    \item Let $v \in \R^n$ be a vector 
    \item Let $B_1^{j_1,i_1,j_0,i_0}$ be defined as Definition~\ref{def:b_1}
    \end{itemize}
    Then, For $j_0,j_1 \in [n], i_0,i_1 \in [d]$, we have 
    \begin{itemize}
\item {\bf Part 1.} For $B_{2,1}^{j_1,i_1,j_1,i_1}$, we have 
\begin{align*}
 B_{2,1}^{j_1,i_1,j_1,i_1}  = & ~ \frac{\d}{\d A_{j_1,i_1}} (- c_g(A,x)^{\top} ) \cdot  f(A,x)_{j_1}  \cdot c(A,x)_{j_1} \cdot e_{i_1} \\
 = & ~ B_{2,1,1}^{j_1,i_1,j_1,i_1} + B_{2,1,2}^{j_1,i_1,j_1,i_1} + B_{2,1,3}^{j_1,i_1,j_1,i_1} + B_{2,1,4}^{j_1,i_1,j_1,i_1} + B_{2,1,5}^{j_1,i_1,j_1,i_1} + B_{2,1,6}^{j_1,i_1,j_1,i_1} + B_{2,1,7}^{j_1,i_1,j_1,i_1}
\end{align*} 
\item {\bf Part 2.} For $B_{2,2}^{j_1,i_1,j_1,i_1}$, we have 
\begin{align*}
  B_{2,2}^{j_1,i_1,j_1,i_1} = & ~ - c_g(A,x)^{\top} \cdot \frac{\d}{\d A_{j_1,i_1}} ( f(A,x)_{j_1} )  \cdot c(A,x)_{j_1} \cdot e_{i_1} \\
    = & ~  B_{2,2,1}^{j_1,i_1,j_1,i_1} + B_{2,2,2}^{j_1,i_1,j_1,i_1}
\end{align*} 
\item {\bf Part 3.} For $B_{2,3}^{j_1,i_1,j_1,i_1}$, we have 
\begin{align*}
  B_{2,3}^{j_1,i_1,j_1,i_1} = & ~ -  c_g(A,x)^{\top} \cdot f(A,x)_{j_1}  \cdot \frac{\d }{\d A_{j_1,i_1}}( c(A,x)_{j_1}) \cdot e_{i_1} \\
     = & ~ B_{2,3,1}^{j_1,i_1,j_1,i_1} + B_{2,3,2}^{j_1,i_1,j_1,i_1}
\end{align*} 
\end{itemize}
\begin{proof}
    {\bf Proof of Part 1.}
    \begin{align*}
    B_{2,1,1}^{j_1,i_1,j_1,i_1} : = & ~  e_{i_1}^\top \cdot \langle c(A,x), f(A,x) \rangle \cdot  f(A,x)_{j_1}^2 \cdot c(A,x)_{j_1} \cdot e_{i_1}\\
    B_{2,1,2}^{j_1,i_1,j_1,i_1} : = & ~ e_{i_1}^\top \cdot c(A,x)_{j_1}^2 \cdot f(A,x)_{j_1}^2 \cdot  e_{i_1}\\
    B_{2,1,3}^{j_1,i_1,j_1,i_1} : = & ~ f(A,x)_{j_1}^2 \cdot \langle c(A,x), f(A,x) \rangle \cdot ( (A_{j_1,*}) \circ x^\top  )  \cdot  c(A,x)_{j_1} \cdot e_{i_1}\\
    B_{2,1,4}^{j_1,i_1,j_1,i_1} : = & ~ -  f(A,x)_{j_1}^2 \cdot f(A,x)^\top  \cdot A \cdot \diag(x) \cdot   \langle c(A,x), f(A,x) \rangle \cdot  c(A,x)_{j_1} \cdot e_{i_1}\\
    B_{2,1,5}^{j_1,i_1,j_1,i_1} : = & ~   f(A,x)_{j_1}^2 \cdot f(A,x)^\top  \cdot A \cdot \diag(x) \cdot (\langle -f(A,x), f(A,x) \rangle + f(A,x)_{j_1})  \cdot c(A,x)_{j_1} \cdot e_{i_1} \\
    B_{2,1,6}^{j_1,i_1,j_1,i_1} : = & ~   f(A,x)_{j_1}^2 \cdot f(A,x)^\top  \cdot A \cdot  \diag(x) \cdot(\langle -f(A,x), c(A,x) \rangle + f(A,x)_{j_1}) \cdot c(A,x)_{j_1} \cdot e_{i_1} \\
    B_{2,1,7}^{j_1,i_1,j_1,i_1} : = & ~ f(A,x)_{j_1}^2 \cdot ((e_{j_1}^\top - f(A,x)^\top) \circ q(A,x)^\top) \cdot A \cdot  \diag(x)  \cdot c(A,x)_{j_1} \cdot e_{i_1}
\end{align*}
Finally, combine them and we have
\begin{align*}
       B_{2,1}^{j_1,i_1,j_1,i_1} = B_{2,1,1}^{j_1,i_1,j_1,i_1} + B_{2,1,2}^{j_1,i_1,j_1,i_1} + B_{2,1,3}^{j_1,i_1,j_1,i_1} + B_{2,1,4}^{j_1,i_1,j_1,i_1} + B_{2,1,5}^{j_1,i_1,j_1,i_1} + B_{2,1,6}^{j_1,i_1,j_1,i_1} + B_{2,1,7}^{j_1,i_1,j_1,i_1}
\end{align*}
{\bf Proof of Part 2.}
    \begin{align*}
    B_{2,2,1}^{j_1,i_1,j_1,i_1} : = & ~  c_g(A,x)^{\top} \cdot f(A,x)_{j_1}^2 \cdot x_{i_1} \cdot c(A,x)_{j_1} \cdot e_{i_1} \\
    B_{2,2,2}^{j_1,i_1,j_1,i_1} : = & ~  - c_g(A,x)^{\top} \cdot f(A,x)_{j_1} \cdot x_{i_1} \cdot c(A,x)_{j_1} \cdot e_{i_1}
\end{align*}
Finally, combine them and we have
\begin{align*}
       B_{2,2}^{j_1,i_1,j_1,i_1} = B_{2,2,1}^{j_1,i_1,j_1,i_1} + B_{2,2,2}^{j_1,i_1,j_1,i_1}
\end{align*}
{\bf Proof of Part 3.} 
    \begin{align*}
    B_{2,3,1}^{j_1,i_1,j_1,i_1} : = & ~    c_g(A,x)^{\top} \cdot f(A,x)_{j_1}^3 \cdot x_{i_1}  \cdot e_{i_1}  \\
    B_{2,3,2}^{j_1,i_1,j_1,i_1} : = & ~  - c_g(A,x)^{\top} \cdot f(A,x)_{j_1}^2 \cdot x_{i_1}  \cdot e_{i_1}
\end{align*}
Finally, combine them and we have
\begin{align*}
       B_{2,3}^{j_1,i_1,j_1,i_1} = B_{2,3,1}^{j_1,i_1,j_1,i_1} + B_{2,3,2}^{j_1,i_1,j_1,i_1}
\end{align*}
\end{proof}
\end{lemma}

\subsection{Case \texorpdfstring{$j_1=j_0, i_1 \neq i_0$}{}}
\begin{lemma}
For $j_1 = j_0$ and $i_1 \neq i_0$. If the following conditions hold
    \begin{itemize}
     \item Let $u(A,x) \in \R^n$ be defined as Definition~\ref{def:u}
    \item Let $\alpha(A,x) \in \R$ be defined as Definition~\ref{def:alpha}
     \item Let $f(A,x) \in \R^n$ be defined as Definition~\ref{def:f}
    \item Let $c(A,x) \in \R^n$ be defined as Definition~\ref{def:c}
    \item Let $g(A,x) \in \R^d$ be defined as Definition~\ref{def:g} 
    \item Let $q(A,x) = c(A,x) + f(A,x) \in \R^n$
    \item Let $c_g(A,x) \in \R^d$ be defined as Definition~\ref{def:c_g}.
    \item Let $L_g(A,x) \in \R$ be defined as Definition~\ref{def:l_g}
    \item Let $v \in \R^n$ be a vector 
    \item Let $B_2^{j_1,i_1,j_0,i_0}$ be defined as Definition~\ref{def:b_1}
    \end{itemize}
    Then, For $j_0=j_1 \in [n], i_0,i_1 \in [d]$, we have 
    \begin{itemize}
\item {\bf Part 1.} For $B_{2,1}^{j_1,i_1,j_1,i_0}$, we have 
\begin{align*}
 B_{2,1}^{j_1,i_1,j_1,i_0}  = & ~ \frac{\d}{\d A_{j_1,i_1}} (- c_g(A,x)^{\top} ) \cdot  f(A,x)_{j_1}  \cdot c(A,x)_{j_1} \cdot e_{i_0} \\
 = & ~ B_{2,1,1}^{j_1,i_1,j_1,i_0} + B_{2,1,2}^{j_1,i_1,j_1,i_0} + B_{2,1,3}^{j_1,i_1,j_1,i_0} + B_{2,1,4}^{j_1,i_1,j_1,i_0} + B_{2,1,5}^{j_1,i_1,j_1,i_0} + B_{2,1,6}^{j_1,i_1,j_1,i_0} + B_{2,1,7}^{j_1,i_1,j_1,i_0}
\end{align*} 
\item {\bf Part 2.} For $B_{2,2}^{j_1,i_1,j_1,i_0}$, we have 
\begin{align*}
  B_{2,2}^{j_1,i_1,j_1,i_0} = & ~ - c_g(A,x)^{\top} \cdot \frac{\d}{\d A_{j_1,i_1}} ( f(A,x)_{j_1} )  \cdot c(A,x)_{j_1} \cdot e_{i_0} \\
    = & ~  B_{2,2,1}^{j_1,i_1,j_1,i_0} + B_{2,2,2}^{j_1,i_1,j_1,i_0}
\end{align*} 
\item {\bf Part 3.} For $B_{2,3}^{j_1,i_1,j_1,i_0}$, we have 
\begin{align*}
  B_{2,3}^{j_1,i_1,j_1,i_0} = & ~ -  c_g(A,x)^{\top} \cdot f(A,x)_{j_1}  \cdot \frac{\d }{\d A_{j_1,i_1}}( c(A,x)_{j_1}) \cdot e_{i_0} \\
     = & ~ B_{2,3,1}^{j_1,i_1,j_1,i_0} + B_{2,3,2}^{j_1,i_1,j_1,i_0}
\end{align*} 
\end{itemize}
\begin{proof}
    {\bf Proof of Part 1.}
    \begin{align*}
    B_{2,1,1}^{j_1,i_1,j_1,i_0} : = & ~  e_{i_1}^\top \cdot \langle c(A,x), f(A,x) \rangle \cdot  f(A,x)_{j_1}^2 \cdot c(A,x)_{j_1} \cdot e_{i_0}\\
    B_{2,1,2}^{j_1,i_1,j_1,i_0} : = & ~ e_{i_1}^\top \cdot c(A,x)_{j_1}^2 \cdot f(A,x)_{j_1}^2 \cdot  e_{i_0}\\
    B_{2,1,3}^{j_1,i_1,j_1,i_0} : = & ~ f(A,x)_{j_1}^2 \cdot \langle c(A,x), f(A,x) \rangle \cdot ( (A_{j_1,*}) \circ x^\top  )  \cdot  c(A,x)_{j_1} \cdot e_{i_0}\\
    B_{2,1,4}^{j_1,i_1,j_1,i_0} : = & ~ -  f(A,x)_{j_1}^2 \cdot f(A,x)^\top  \cdot A \cdot \diag(x) \cdot   \langle c(A,x), f(A,x) \rangle \cdot  c(A,x)_{j_1} \cdot e_{i_0}\\
    B_{2,1,5}^{j_1,i_1,j_1,i_0} : = & ~   f(A,x)_{j_1}^2 \cdot f(A,x)^\top  \cdot A \cdot \diag(x) \cdot (\langle -f(A,x), f(A,x) \rangle + f(A,x)_{j_1})  \cdot c(A,x)_{j_1} \cdot e_{i_0} \\
    B_{2,1,6}^{j_1,i_1,j_1,i_0} : = & ~   f(A,x)_{j_1}^2 \cdot f(A,x)^\top  \cdot A \cdot  \diag(x) \cdot(\langle -f(A,x), c(A,x) \rangle + f(A,x)_{j_1}) \cdot c(A,x)_{j_1} \cdot e_{i_0} \\
    B_{2,1,7}^{j_1,i_1,j_1,i_0} : = & ~ f(A,x)_{j_1}^2 \cdot ((e_{j_1}^\top - f(A,x)^\top) \circ q(A,x)^\top) \cdot A \cdot  \diag(x)  \cdot c(A,x)_{j_1} \cdot e_{i_0}
\end{align*}
Finally, combine them and we have
\begin{align*}
       B_{2,1}^{j_1,i_1,j_1,i_0} = B_{2,1,1}^{j_1,i_1,j_1,i_0} + B_{2,1,2}^{j_1,i_1,j_1,i_0} + B_{2,1,3}^{j_1,i_1,j_1,i_0} + B_{2,1,4}^{j_1,i_1,j_1,i_0} + B_{2,1,5}^{j_1,i_1,j_1,i_0} + B_{2,1,6}^{j_1,i_1,j_1,i_0} + B_{2,1,7}^{j_1,i_1,j_1,i_0}
\end{align*}
{\bf Proof of Part 2.}
    \begin{align*}
    B_{2,2,1}^{j_1,i_1,j_1,i_0} : = & ~  c_g(A,x)^{\top} \cdot f(A,x)_{j_1}^2 \cdot x_{i_1} \cdot c(A,x)_{j_1} \cdot e_{i_0} \\
    B_{2,2,2}^{j_1,i_1,j_1,i_0} : = & ~  - c_g(A,x)^{\top} \cdot f(A,x)_{j_1} \cdot x_{i_1} \cdot c(A,x)_{j_1} \cdot e_{i_0}
\end{align*}
Finally, combine them and we have
\begin{align*}
       B_{2,2}^{j_1,i_1,j_1,i_0} = B_{2,2,1}^{j_1,i_1,j_1,i_0} + B_{2,2,2}^{j_1,i_1,j_1,i_0}
\end{align*}
{\bf Proof of Part 3.} 
    \begin{align*}
    B_{2,3,1}^{j_1,i_1,j_1,i_0} : = & ~    c_g(A,x)^{\top} \cdot f(A,x)_{j_1}^3 \cdot x_{i_1}  \cdot e_{i_0}  \\
    B_{2,3,2}^{j_1,i_1,j_1,i_0} : = & ~  - c_g(A,x)^{\top} \cdot f(A,x)_{j_1}^2 \cdot x_{i_1}  \cdot e_{i_0}
\end{align*}
Finally, combine them and we have
\begin{align*}
       B_{2,3}^{j_1,i_1,j_1,i_0} = B_{2,3,1}^{j_1,i_1,j_1,i_0} + B_{2,3,2}^{j_1,i_1,j_1,i_0}
\end{align*}
\end{proof}
\end{lemma}

\subsection{Constructing \texorpdfstring{$d \times d$}{} matrices for \texorpdfstring{$j_1 = j_0$}{}}

The purpose of the following lemma is to let $i_0$ and $i_1$ disappear.
\begin{lemma}For $j_0,j_1 \in [n]$, a list of $d \times d$ matrices can be expressed as the following sense,
\begin{itemize}
\item {\bf Part 1.}
\begin{align*}
     B_{2,1,1}^{j_1,*,j_1,*} = f(A,x)_{j_1}^2 \cdot c(A,x)_{j_1} \cdot f_c(A,x) \cdot  I_d
\end{align*}
\item {\bf Part 2.}
\begin{align*}
     B_{2,1,2}^{j_1,*,j_1,*} = f(A,x)_{j_1}^2 \cdot c(A,x)_{j_1}^2  \cdot I_d 
\end{align*}
\item {\bf Part 3.}
\begin{align*}
     B_{2,1,3}^{j_1,*,j_1,*}  = f(A,x)_{j_1}^2 \cdot c(A,x)_{j_1} \cdot f_c(A,x) \cdot  {\bf 1}_d  \cdot ( (A_{j_1,*}) \circ x^\top  ) 
\end{align*}
\item {\bf Part 4.}
\begin{align*}
    B_{2,1,4}^{j_1,*,j_1,*}  = -  f(A,x)_{j_1}^2 \cdot  c(A,x)_{j_1} \cdot f_c(A,x) \cdot {\bf 1}_d  \cdot  h(A,x)^\top 
\end{align*}
\item {\bf Part 5.}
\begin{align*}
    B_{2,1,5}^{j_1,*,j_1,*}  = & ~    f(A,x)_{j_1}^2 \cdot  (-f_2(A,x) + f(A,x)_{j_1})  \cdot c(A,x)_{j_1} \cdot {\bf 1}_{d} \cdot h(A,x)^\top
\end{align*}
\item {\bf Part 6.}
\begin{align*}
    B_{2,1,6}^{j_1,*,j_1,*}  = & ~    f(A,x)_{j_1}^2 \cdot (-f_c(A,x) + f(A,x)_{j_1}) \cdot c(A,x)_{j_1} \cdot {\bf 1}_d \cdot h(A,x)^\top
\end{align*}
\item {\bf Part 7.}
\begin{align*}
     B_{2,1,7}^{j_1,*,j_1,*}  =  f(A,x)_{j_1}^2 \cdot c(A,x)_{j_1} \cdot {\bf 1}_d \cdot p_{j_1}(A,x)^\top
\end{align*}
\item {\bf Part 8.}
\begin{align*}
     B_{2,2,1}^{j_1,*,j_1,*}  =  f(A,x)_{j_1}^2  \cdot c(A,x)_{j_1} \cdot   x   \cdot   c_g(A,x)^\top
\end{align*}
\item {\bf Part 9.}
\begin{align*}
     B_{2,2,2}^{j_1,*,j_1,*}  =  - f(A,x)_{j_1}  \cdot   c(A,x)_{j_1} \cdot   x  \cdot   c_g(A,x)^\top  
\end{align*}
\item {\bf Part 10.}
\begin{align*}
    B_{2,3,1}^{j_1,*,j_1,*}  =    f(A,x)_{j_1}^3\cdot   x  \cdot   c_g(A,x)^\top  
\end{align*}
\item {\bf Part 11.}
\begin{align*}
    B_{2,3,2}^{j_1,*,j_1,*}  =   -   f(A,x)_{j_1}^2 \cdot   x   \cdot c_g(A,x)^{\top} 
\end{align*}

\end{itemize}
\begin{proof}
{\bf Proof of Part 1.}
    We have
    \begin{align*}
         B_{2,1,1}^{j_1,i_1,j_1,i_1}  = & ~  e_{i_1}^\top \cdot \langle c(A,x), f(A,x) \rangle \cdot  f(A,x)_{j_1}^2 \cdot c(A,x)_{j_1} \cdot e_{i_1} \\
        B_{2,1,1}^{j_1,i_1,j_1,i_0}  = & ~  e_{i_1}^\top \cdot \langle c(A,x), f(A,x) \rangle \cdot  f(A,x)_{j_1}^2 \cdot c(A,x)_{j_1} \cdot e_{i_0}
    \end{align*}

    From the above two equations, we can tell that $B_{2,1,1}^{j_1,*,j_1,*} \in \R^{d \times d}$ is a matrix that only diagonal has entries and off-diagonal are all zeros.
    
    Then we have $B_{2,1,1}^{j_1,*,j_1,*} \in \R^{d \times d}$ can be written as the rescaling of a diagonal matrix,
    \begin{align*}
     B_{2,1,1}^{j_1,*,j_1,*} & ~ = f(A,x)_{j_1}^2 \cdot c(A,x)_{j_1} \cdot \langle c(A,x), f(A,x) \rangle \cdot  I_d   \\
     & ~ = f(A,x)_{j_1}^2 \cdot c(A,x)_{j_1} \cdot f_c(A,x) \cdot  I_d
\end{align*}
    where the last step is follows from the Definitions~\ref{def:f_c}.

{\bf Proof of Part 2.}
    We have
    \begin{align*}
        B_{2,1,2}^{j_1,i_1,j_1,i_1} = & ~ e_{i_1}^\top \cdot c(A,x)_{j_1}^2 \cdot f(A,x)_{j_1}^2 \cdot  e_{i_1} \\
        B_{2,1,2}^{j_1,i_1,j_1,i_0} = & ~ e_{i_1}^\top \cdot c(A,x)_{j_1}^2 \cdot f(A,x)_{j_1}^2 \cdot  e_{i_0}
    \end{align*}
     From the above two equations, we can tell that $B_{2,1,2}^{j_1,*,j_1,*} \in \R^{d \times d}$ is a matrix that only diagonal has entries and off-diagonal are all zeros.
    
    Then we have $B_{2,1,2}^{j_1,*,j_1,*} \in \R^{d \times d}$ can be written as the rescaling of a diagonal matrix,
\begin{align*}
     B_{2,1,2}^{j_1,*,j_1,*} & ~ = f(A,x)_{j_1}^2 \cdot c(A,x)_{j_1}^2  \cdot I_d 
\end{align*}

{\bf Proof of Part 3.}
We have for diagonal entry and off-diagonal entry can be written as follows 
    \begin{align*}
        B_{2,1,3}^{j_1,i_1,j_1,i_1}  = & ~ f(A,x)_{j_1}^2 \cdot \langle c(A,x), f(A,x) \rangle \cdot ( (A_{j_1,*}) \circ x^\top  )  \cdot  c(A,x)_{j_1} \cdot e_{i_1} \\
        B_{2,1,3}^{j_1,i_1,j_1,i_0}  = & ~ f(A,x)_{j_1}^2 \cdot \langle c(A,x), f(A,x) \rangle \cdot ( (A_{j_1,*}) \circ x^\top  )  \cdot  c(A,x)_{j_1} \cdot e_{i_0}
    \end{align*}
From the above equation, we can show that matrix $B_{2,1,3}^{j_1,*,j_1,*}$ can be expressed as a rank-$1$ matrix,
\begin{align*}
     B_{2,1,3}^{j_1,*,j_1,*} & ~ = f(A,x)_{j_1}^2 \cdot c(A,x)_{j_1} \cdot \langle c(A,x), f(A,x) \rangle \cdot  {\bf 1}_d  \cdot ( (A_{j_1,*}) \circ x^\top  ) \\
     & ~ = f(A,x)_{j_1}^2 \cdot c(A,x)_{j_1} \cdot f_c(A,x) \cdot  {\bf 1}_d  \cdot ( (A_{j_1,*}) \circ x^\top  ) 
\end{align*}
    where the last step is follows from the Definitions~\ref{def:f_c}.

{\bf Proof of Part 4.}
We have for diagonal entry and off-diagonal entry can be written as follows
    \begin{align*}
        B_{2,1,4}^{j_1,i_1,j_1,i_1}  = & ~ -  f(A,x)_{j_1}^2 \cdot f(A,x)^\top  \cdot A \cdot \diag(x) \cdot   \langle c(A,x), f(A,x) \rangle \cdot  c(A,x)_{j_1} \cdot e_{i_1} \\
        B_{2,1,4}^{j_1,i_1,j_1,i_0}  = & ~ -  f(A,x)_{j_1}^2 \cdot f(A,x)^\top  \cdot A \cdot \diag(x) \cdot   \langle c(A,x), f(A,x) \rangle \cdot  c(A,x)_{j_1} \cdot e_{i_0}
    \end{align*}
 From the above equation, we can show that matrix $B_{2,1,4}^{j_1,*,j_1,*}$ can be expressed as a rank-$1$ matrix,
\begin{align*}
    B_{2,1,4}^{j_1,*,j_1,*}  & ~ =  -  f(A,x)_{j_1}^2 \cdot  c(A,x)_{j_1} \cdot  \langle c(A,x), f(A,x) \rangle \cdot  {\bf 1}_d  \cdot  f(A,x)^\top  \cdot A \cdot \diag(x)   \\
     & ~ = -  f(A,x)_{j_1}^2 \cdot  c(A,x)_{j_1} \cdot f_c(A,x) \cdot {\bf 1}_d  \cdot  h(A,x)^\top 
\end{align*}
    where the last step is follows from the Definitions~\ref{def:h} and Definitions~\ref{def:f_c}.

{\bf Proof of Part 5.}
We have for diagonal entry and off-diagonal entry can be written as follows
    \begin{align*}
        B_{2,1,5}^{j_1,i_1,j_1,i_1}  = & ~   f(A,x)_{j_1}^2 \cdot f(A,x)^\top  \cdot A \cdot \diag(x) \cdot (\langle -f(A,x), f(A,x) \rangle + f(A,x)_{j_1})  \cdot c(A,x)_{j_1} \cdot e_{i_1}  \\
        B_{2,1,5}^{j_1,i_1,j_1,i_0}  = & ~   f(A,x)_{j_1}^2 \cdot f(A,x)^\top  \cdot A \cdot \diag(x) \cdot (\langle -f(A,x), f(A,x) \rangle + f(A,x)_{j_1})  \cdot c(A,x)_{j_1} \cdot e_{i_0} 
    \end{align*}
    From the above equation, we can show that matrix $B_{2,1,5}^{j_1,*,j_1,*}$ can be expressed as a rank-$1$ matrix,
\begin{align*}
    B_{2,1,5}^{j_1,*,j_1,*}  & ~ =     f(A,x)_{j_1}^2 \cdot  (\langle -f(A,x), f(A,x) \rangle + f(A,x)_{j_1})  \cdot c(A,x)_{j_1} \cdot {\bf 1}_{d} \cdot f(A,x)^\top  \cdot A \cdot \diag(x)\\
     & ~ = f(A,x)_{j_1}^2 \cdot  (-f_2(A,x) + f(A,x)_{j_1})  \cdot c(A,x)_{j_1} \cdot {\bf 1}_{d} \cdot h(A,x)^\top
\end{align*}
    where the last step is follows from the Definitions~\ref{def:h} and Definitions~\ref{def:f_2}.

{\bf Proof of Part 6.}
We have for diagonal entry and off-diagonal entry can be written as follows
    \begin{align*}
        B_{2,1,6}^{j_1,i_1,j_1,i_1}  = & ~   f(A,x)_{j_1}^2 \cdot f(A,x)^\top  \cdot A \cdot  \diag(x) \cdot(\langle -f(A,x), c(A,x) \rangle + f(A,x)_{j_1}) \cdot c(A,x)_{j_1} \cdot e_{i_1}\\
        B_{2,1,6}^{j_1,i_1,j_1,i_0}  = & ~   f(A,x)_{j_1}^2 \cdot f(A,x)^\top  \cdot A \cdot  \diag(x) \cdot(\langle -f(A,x), c(A,x) \rangle + f(A,x)_{j_1}) \cdot c(A,x)_{j_1} \cdot e_{i_0}
    \end{align*}
    From the above equation, we can show that matrix $B_{2,1,6}^{j_1,*,j_1,*}$ can be expressed as a rank-$1$ matrix,
\begin{align*}
    B_{2,1,6}^{j_1,*,j_1,*}  & ~ =   f(A,x)_{j_1}^2 \cdot (\langle -f(A,x), c(A,x) \rangle + f(A,x)_{j_1}) \cdot c(A,x)_{j_1} \cdot {\bf 1}_d \cdot f(A,x)^\top  \cdot A \cdot  \diag(x)\\
     & ~ = f(A,x)_{j_1}^2 \cdot (-f_c(A,x) + f(A,x)_{j_1}) \cdot c(A,x)_{j_1} \cdot {\bf 1}_d \cdot h(A,x)^\top
\end{align*}
    where the last step is follows from the Definitions~\ref{def:h} and Definitions~\ref{def:f_c}.
    
{\bf Proof of Part 7.}
We have for diagonal entry and off-diagonal entry can be written as follows
    \begin{align*}
         B_{2,1,7}^{j_1,i_1,j_1,i_1}  = & ~ f(A,x)_{j_1}^2 \cdot ((e_{j_1}^\top - f(A,x)^\top) \circ q(A,x)^\top) \cdot A \cdot  \diag(x)  \cdot c(A,x)_{j_1} \cdot e_{i_1}\\
         B_{2,1,7}^{j_1,i_1,j_1,i_0}  = & ~ f(A,x)_{j_1}^2 \cdot ((e_{j_1}^\top - f(A,x)^\top) \circ q(A,x)^\top) \cdot A \cdot  \diag(x)  \cdot c(A,x)_{j_1} \cdot e_{i_0}
    \end{align*}
    From the above equation, we can show that matrix $B_{2,1,7}^{j_1,*,j_1,*}$ can be expressed as a rank-$1$ matrix,
\begin{align*}
     B_{2,1,7}^{j_1,*,j_1,*}  & ~ =   f(A,x)_{j_1}^2 \cdot c(A,x)_{j_1} \cdot  {\bf 1}_d \cdot 
     ((e_{j_1}^\top - f(A,x)^\top) \circ q(A,x)^\top) \cdot A \cdot  \diag(x)  \\
     & ~ =f(A,x)_{j_1}^2 \cdot c(A,x)_{j_1} \cdot {\bf 1}_d \cdot p_{j_1}(A,x)^\top
\end{align*}
    where the last step is follows from the Definitions~\ref{def:p}.
    
{\bf Proof of Part 8.}
We have for diagonal entry and off-diagonal entry can be written as follows
    \begin{align*}
         B_{2,2,1}^{j_1,i_1,j_1,i_1}  = & ~  c_g(A,x)^{\top} \cdot f(A,x)_{j_1}^2 \cdot x_{i_1} \cdot c(A,x)_{j_1} \cdot e_{i_1} \\
         B_{2,2,1}^{j_1,i_1,j_1,i_0}  = & ~  c_g(A,x)^{\top} \cdot f(A,x)_{j_1}^2 \cdot x_{i_1} \cdot c(A,x)_{j_1} \cdot e_{i_0} 
    \end{align*}
    From the above equation, we can show that matrix $B_{2,2,1}^{j_1,*,j_1,*}$ can be expressed as a rank-$1$ matrix,
\begin{align*}
     B_{2,2,1}^{j_1,*,j_1,*}  & ~ =   f(A,x)_{j_1}^2  \cdot c(A,x)_{j_1} \cdot   x   \cdot   c_g(A,x)^\top  
\end{align*}

{\bf Proof of Part 9.}
We have for diagonal entry and off-diagonal entry can be written as follows
    \begin{align*}
          B_{2,2,2}^{j_1,i_1,j_1,i_1}  = & ~  - c_g(A,x)^{\top} \cdot f(A,x)_{j_1} \cdot x_{i_1} \cdot c(A,x)_{j_1} \cdot e_{i_1}\\
          B_{2,2,2}^{j_1,i_1,j_1,i_0}  = & ~  - c_g(A,x)^{\top} \cdot f(A,x)_{j_1} \cdot x_{i_1} \cdot c(A,x)_{j_1} \cdot e_{i_0}
    \end{align*}
        From the above equation, we can show that matrix $B_{2,2,2}^{j_1,*,j_1,*}$ can be expressed as a rank-$1$ matrix,
\begin{align*}
     B_{2,2,2}^{j_1,*,j_1,*}  & ~ =   - f(A,x)_{j_1}  \cdot   c(A,x)_{j_1} \cdot   x  \cdot   c_g(A,x)^\top  
\end{align*}

{\bf Proof of Part 10.}
We have for diagonal entry and off-diagonal entry can be written as follows
    \begin{align*}
         B_{2,3,1}^{j_1,i_1,j_1,i_1}  = & ~    c_g(A,x)^{\top} \cdot f(A,x)_{j_1}^3 \cdot x_{i_1}  \cdot e_{i_1}\\
         B_{2,3,1}^{j_1,i_1,j_1,i_0}  = & ~    c_g(A,x)^{\top} \cdot f(A,x)_{j_1}^3 \cdot x_{i_1}  \cdot e_{i_0}
    \end{align*}
            From the above equation, we can show that matrix $B_{2,3,1}^{j_1,*,j_1,*}$ can be expressed as a rank-$1$ matrix,
\begin{align*}
    B_{2,3,1}^{j_1,*,j_1,*}  & ~ =      f(A,x)_{j_1}^3\cdot   x  \cdot   c_g(A,x)^\top  
\end{align*}

{\bf Proof of Part 11.}
We have for diagonal entry and off-diagonal entry can be written as follows
    \begin{align*}
         B_{2,3,2}^{j_1,i_1,j_1,i_1}  = & ~  - c_g(A,x)^{\top} \cdot f(A,x)_{j_1}^2 \cdot x_{i_1}  \cdot e_{i_1}\\
         B_{2,3,2}^{j_1,i_1,j_1,i_0}  = & ~  - c_g(A,x)^{\top} \cdot f(A,x)_{j_1}^2 \cdot x_{i_1}  \cdot e_{i_0}
    \end{align*}
            From the above equation, we can show that matrix $B_{2,3,2}^{j_1,*,j_1,*}$ can be expressed as a rank-$1$ matrix,
\begin{align*}
    B_{2,3,2}^{j_1,i_1,j_1,*}  =   -   f(A,x)_{j_1}^2 \cdot   x   \cdot c_g(A,x)^{\top} 
\end{align*}
\end{proof}
\end{lemma}

\subsection{Case \texorpdfstring{$j_1 \neq j_0, i_1 = i_0$}{}}
\begin{lemma}
For $j_1 \neq j_0$ and $i_0 = i_1$. If the following conditions hold
    \begin{itemize}
     \item Let $u(A,x) \in \R^n$ be defined as Definition~\ref{def:u}
    \item Let $\alpha(A,x) \in \R$ be defined as Definition~\ref{def:alpha}
     \item Let $f(A,x) \in \R^n$ be defined as Definition~\ref{def:f}
    \item Let $c(A,x) \in \R^n$ be defined as Definition~\ref{def:c}
    \item Let $g(A,x) \in \R^d$ be defined as Definition~\ref{def:g} 
    \item Let $q(A,x) = c(A,x) + f(A,x) \in \R^n$
    \item Let $c_g(A,x) \in \R^d$ be defined as Definition~\ref{def:c_g}.
    \item Let $L_g(A,x) \in \R$ be defined as Definition~\ref{def:l_g}
    \item Let $v \in \R^n$ be a vector 
    \item Let $B_1^{j_1,i_1,j_0,i_0}$ be defined as Definition~\ref{def:b_1}
    \end{itemize}
    Then, For $j_0,j_1 \in [n], i_0,i_1 \in [d]$, we have 
    \begin{itemize}
\item {\bf Part 1.} For $B_{2,1}^{j_1,i_1,j_0,i_1}$, we have 
\begin{align*}
 B_{2,1}^{j_1,i_1,j_0,i_1}  = & ~ \frac{\d}{\d A_{j_1,i_1}} (- c_g(A,x)^{\top} ) \cdot  f(A,x)_{j_0}  \cdot c(A,x)_{j_0} \cdot e_{i_1} \\
 = & ~ B_{2,1,1}^{j_1,i_1,j_0,i_1} + B_{2,1,2}^{j_1,i_1,j_0,i_1} + B_{2,1,3}^{j_1,i_1,j_0,i_1} + B_{2,1,4}^{j_1,i_1,j_0,i_1} + B_{2,1,5}^{j_1,i_1,j_0,i_1} + B_{2,1,6}^{j_1,i_1,j_0,i_1} + B_{2,1,7}^{j_1,i_1,j_0,i_1}
\end{align*} 
\item {\bf Part 2.} For $B_{2,2}^{j_1,i_1,j_0,i_1}$, we have 
\begin{align*}
  B_{2,2}^{j_1,i_1,j_0,i_1} = & ~ - c_g(A,x)^{\top} \cdot \frac{\d}{\d A_{j_1,i_1}} ( f(A,x)_{j_0} )  \cdot c(A,x)_{j_0} \cdot e_{i_1} \\
    = & ~  B_{2,2,1}^{j_1,i_1,j_0,i_1}
\end{align*} 
\item {\bf Part 3.} For $B_{2,3}^{j_1,i_1,j_0,i_1}$, we have 
\begin{align*}
  B_{2,3}^{j_1,i_1,j_0,i_1} = & ~ -  c_g(A,x)^{\top} \cdot f(A,x)_{j_0}  \cdot \frac{\d }{\d A_{j_1,i_1}}( c(A,x)_{j_0}) \cdot e_{i_1} \\
     = & ~ B_{2,3,1}^{j_1,i_1,j_0,i_1} 
\end{align*} 
\end{itemize}
\begin{proof}
    {\bf Proof of Part 1.}
    \begin{align*}
    B_{2,1,1}^{j_1,i_1,j_0,i_1} : = & ~   e_{i_1}^\top \cdot \langle c(A,x), f(A,x) \rangle \cdot  f(A,x)_{j_1} \cdot  f(A,x)_{j_0} \cdot c(A,x)_{j_0} \cdot e_{i_1}\\
    B_{2,1,2}^{j_1,i_1,j_0,i_1} : = & ~ e_{i_1}^\top \cdot c(A,x)_{j_1} \cdot f(A,x)_{j_1} \cdot  f(A,x)_{j_0} \cdot c(A,x)_{j_0} \cdot  e_{i_1}\\
    B_{2,1,3}^{j_1,i_1,j_0,i_1} : = & ~  f(A,x)_{j_1} \cdot \langle c(A,x), f(A,x) \rangle \cdot ( (A_{j_1,*}) \circ x^\top  )  \cdot  f(A,x)_{j_0} \cdot c(A,x)_{j_0} \cdot e_{i_1}\\
    B_{2,1,4}^{j_1,i_1,j_0,i_1} : = & ~ -  f(A,x)_{j_1} \cdot f(A,x)^\top  \cdot A \cdot \diag(x) \cdot   \langle c(A,x), f(A,x) \rangle \cdot  f(A,x)_{j_0} \cdot c(A,x)_{j_0} \cdot e_{i_1}\\
    B_{2,1,5}^{j_1,i_1,j_0,i_1} : = & ~  f(A,x)_{j_1} \cdot f(A,x)^\top  \cdot A \cdot \diag(x) \cdot (\langle -f(A,x), f(A,x) \rangle + f(A,x)_{j_1})  \cdot f(A,x)_{j_0} \cdot c(A,x)_{j_0} \cdot e_{i_1} \\
    B_{2,1,6}^{j_1,i_1,j_0,i_1} : = & ~  f(A,x)_{j_1} \cdot f(A,x)^\top  \cdot A \cdot  \diag(x) \cdot(\langle -f(A,x), c(A,x) \rangle + f(A,x)_{j_1}) \cdot f(A,x)_{j_0} \cdot c(A,x)_{j_0} \cdot e_{i_1} \\
    B_{2,1,7}^{j_1,i_1,j_0,i_1} : = & ~ f(A,x)_{j_1} \cdot ((e_{j_1}^\top - f(A,x)^\top) \circ q(A,x)^\top) \cdot A \cdot  \diag(x)  \cdot f(A,x)_{j_0} \cdot c(A,x)_{j_0} \cdot e_{i_1}
\end{align*}
Finally, combine them and we have
\begin{align*}
       B_{2,1}^{j_1,i_1,j_0,i_1} = B_{2,1,1}^{j_1,i_1,j_0,i_1} + B_{2,1,2}^{j_1,i_1,j_0,i_1} + B_{2,1,3}^{j_1,i_1,j_0,i_1} + B_{2,1,4}^{j_1,i_1,j_0,i_1} + B_{2,1,5}^{j_1,i_1,j_0,i_1} + B_{2,1,6}^{j_1,i_1,j_0,i_1} + B_{2,1,7}^{j_1,i_1,j_0,i_1}
\end{align*}
{\bf Proof of Part 2.}
    \begin{align*}
    B_{2,2,1}^{j_1,i_1,j_0,i_1} : = & ~  c_g(A,x)^{\top} \cdot f(A,x)_{j_1} \cdot  f(A,x)_{j_0} \cdot x_{i_1}  \cdot c(A,x)_{j_0} \cdot e_{i_1} 
\end{align*}
Finally, we have
\begin{align*}
       B_{2,2}^{j_1,i_1,j_0,i_1} = B_{2,2,1}^{j_1,i_1,j_0,i_1}  
\end{align*}
{\bf Proof of Part 3.} 
    \begin{align*}
    B_{2,3,1}^{j_1,i_1,j_0,i_1} : = & ~   c_g(A,x)^{\top} \cdot f(A,x)_{j_1} \cdot  f(A,x)_{j_0}^2 \cdot x_{i_1} \cdot e_{i_1}  
\end{align*}
Finally, we have
\begin{align*}
       B_{2,3}^{j_1,i_1,j_0,i_1} = B_{2,3,1}^{j_1,i_1,j_0,i_1}  
\end{align*}
\end{proof}
\end{lemma}

\subsection{Case \texorpdfstring{$j_1 \neq j_0, i_1 \neq i_0$}{}}
\begin{lemma}
For $j_1 \neq j_0$ and $i_0 \neq i_1$. If the following conditions hold
    \begin{itemize}
     \item Let $u(A,x) \in \R^n$ be defined as Definition~\ref{def:u}
    \item Let $\alpha(A,x) \in \R$ be defined as Definition~\ref{def:alpha}
     \item Let $f(A,x) \in \R^n$ be defined as Definition~\ref{def:f}
    \item Let $c(A,x) \in \R^n$ be defined as Definition~\ref{def:c}
    \item Let $g(A,x) \in \R^d$ be defined as Definition~\ref{def:g} 
    \item Let $q(A,x) = c(A,x) + f(A,x) \in \R^n$
    \item Let $c_g(A,x) \in \R^d$ be defined as Definition~\ref{def:c_g}.
    \item Let $L_g(A,x) \in \R$ be defined as Definition~\ref{def:l_g}
    \item Let $v \in \R^n$ be a vector 
    \item Let $B_1^{j_1,i_1,j_0,i_0}$ be defined as Definition~\ref{def:b_1}
    \end{itemize}
    Then, For $j_0,j_1 \in [n], i_0,i_1 \in [d]$, we have 
    \begin{itemize}
\item {\bf Part 1.} For $B_{2,1}^{j_1,i_1,j_0,i_0}$, we have 
\begin{align*}
 B_{2,1}^{j_1,i_1,j_0,i_0}  = & ~ \frac{\d}{\d A_{j_1,i_1}} (- c_g(A,x)^{\top} ) \cdot  f(A,x)_{j_0}  \cdot c(A,x)_{j_0} \cdot e_{i_0} \\
 = & ~ B_{2,1,1}^{j_1,i_1,j_0,i_0} + B_{2,1,2}^{j_1,i_1,j_0,i_0} + B_{2,1,3}^{j_1,i_1,j_0,i_0} + B_{2,1,4}^{j_1,i_1,j_0,i_0} + B_{2,1,5}^{j_1,i_1,j_0,i_0} + B_{2,1,6}^{j_1,i_1,j_0,i_0} + B_{2,1,7}^{j_1,i_1,j_0,i_0}
\end{align*} 
\item {\bf Part 2.} For $B_{2,2}^{j_1,i_1,j_0,i_0}$, we have 
\begin{align*}
  B_{2,2}^{j_1,i_1,j_0,i_0} = & ~ - c_g(A,x)^{\top} \cdot \frac{\d}{\d A_{j_1,i_1}} ( f(A,x)_{j_0} )  \cdot c(A,x)_{j_0} \cdot e_{i_0} \\
    = & ~  B_{2,2,1}^{j_1,i_1,j_0,i_0}  
\end{align*} 
\item {\bf Part 3.} For $B_{2,3}^{j_1,i_1,j_0,i_0}$, we have 
\begin{align*}
  B_{2,3}^{j_1,i_1,j_0,i_0} = & ~ -  c_g(A,x)^{\top} \cdot f(A,x)_{j_0}  \cdot \frac{\d }{\d A_{j_1,i_1}}( c(A,x)_{j_0}) \cdot e_{i_0} \\
     = & ~ B_{2,3,1}^{j_1,i_1,j_0,i_0} 
\end{align*} 
\end{itemize}
\begin{proof}
    {\bf Proof of Part 1.}
    \begin{align*}
    B_{2,1,1}^{j_1,i_1,j_0,i_0} : = & ~   e_{i_1}^\top \cdot \langle c(A,x), f(A,x) \rangle \cdot  f(A,x)_{j_1} \cdot  f(A,x)_{j_0} \cdot c(A,x)_{j_0} \cdot e_{i_0}\\
    B_{2,1,2}^{j_1,i_1,j_0,i_0} : = & ~ e_{i_1}^\top \cdot c(A,x)_{j_1} \cdot f(A,x)_{j_1} \cdot  f(A,x)_{j_0} \cdot c(A,x)_{j_0} \cdot  e_{i_0}\\
    B_{2,1,3}^{j_1,i_1,j_0,i_0} : = & ~  f(A,x)_{j_1} \cdot \langle c(A,x), f(A,x) \rangle \cdot ( (A_{j_1,*}) \circ x^\top  )  \cdot  f(A,x)_{j_0} \cdot c(A,x)_{j_0} \cdot e_{i_0}\\
    B_{2,1,4}^{j_1,i_1,j_0,i_0} : = & ~ -  f(A,x)_{j_1} \cdot f(A,x)^\top  \cdot A \cdot \diag(x) \cdot   \langle c(A,x), f(A,x) \rangle \cdot  f(A,x)_{j_0} \cdot c(A,x)_{j_0} \cdot e_{i_0}\\
    B_{2,1,5}^{j_1,i_1,j_0,i_0} : = & ~  f(A,x)_{j_1} \cdot f(A,x)^\top  \cdot A \cdot \diag(x) \cdot (\langle -f(A,x), f(A,x) \rangle + f(A,x)_{j_1})  \cdot f(A,x)_{j_0} \cdot c(A,x)_{j_0} \cdot e_{i_0} \\
    B_{2,1,6}^{j_1,i_1,j_0,i_0} : = & ~  f(A,x)_{j_1} \cdot f(A,x)^\top  \cdot A \cdot  \diag(x) \cdot(\langle -f(A,x), c(A,x) \rangle + f(A,x)_{j_1}) \cdot f(A,x)_{j_0} \cdot c(A,x)_{j_0} \cdot e_{i_0} \\
    B_{2,1,7}^{j_1,i_1,j_0,i_0} : = & ~ f(A,x)_{j_1} \cdot ((e_{j_1}^\top - f(A,x)^\top) \circ q(A,x)^\top) \cdot A \cdot  \diag(x)  \cdot f(A,x)_{j_0} \cdot c(A,x)_{j_0} \cdot e_{i_0}
\end{align*}
Finally, combine them and we have
\begin{align*}
       B_{2,1}^{j_1,i_1,j_0,i_0} = B_{2,1,1}^{j_1,i_1,j_0,i_0} + B_{2,1,2}^{j_1,i_1,j_0,i_0} + B_{2,1,3}^{j_1,i_1,j_0,i_0} + B_{2,1,4}^{j_1,i_1,j_0,i_0} + B_{2,1,5}^{j_1,i_1,j_0,i_0} + B_{2,1,6}^{j_1,i_1,j_0,i_0} + B_{2,1,7}^{j_1,i_1,j_0,i_0}
\end{align*}
{\bf Proof of Part 2.}
    \begin{align*}
    B_{2,2,1}^{j_1,i_1,j_0,i_0} : = & ~  c_g(A,x)^{\top} \cdot f(A,x)_{j_1} \cdot  f(A,x)_{j_0} \cdot x_{i_1}  \cdot c(A,x)_{j_0} \cdot e_{i_0} 
\end{align*}
Finally, we have
\begin{align*}
       B_{2,2}^{j_1,i_1,j_0,i_0} = B_{2,2,1}^{j_1,i_1,j_0,i_0}  
\end{align*}
{\bf Proof of Part 3.} 
    \begin{align*}
    B_{2,3,1}^{j_1,i_1,j_0,i_0} : = & ~   c_g(A,x)^{\top} \cdot f(A,x)_{j_1} \cdot  f(A,x)_{j_0}^2 \cdot x_{i_1} \cdot e_{i_0}  
\end{align*}
Finally, we have
\begin{align*}
       B_{2,3}^{j_1,i_1,j_0,i_0} = B_{2,3,1}^{j_1,i_1,j_0,i_0}  
\end{align*}
\end{proof}
\end{lemma}

\subsection{Constructing \texorpdfstring{$d \times d$}{} matrices for \texorpdfstring{$j_1 \neq j_0$}{}}

The purpose of the following lemma is to let $i_0$ and $i_1$ disappear.
\begin{lemma}For $j_0,j_1 \in [n]$, a list of $d \times d$ matrices can be expressed as the following sense,\label{lem:b_2_j1_j0}
\begin{itemize}
\item {\bf Part 1.}
\begin{align*}
     B_{2,1,1}^{j_1,*,j_0,*} = f(A,x)_{j_1} \cdot  f(A,x)_{j_0} \cdot c(A,x)_{j_0} \cdot f_c(A,x) \cdot  I_d
\end{align*}
\item {\bf Part 2.}
\begin{align*}
     B_{2,1,2}^{j_1,*,j_0,*} = c(A,x)_{j_1} \cdot f(A,x)_{j_1} \cdot  f(A,x)_{j_0} \cdot c(A,x)_{j_0} \cdot I_d 
\end{align*}
\item {\bf Part 3.}
\begin{align*}
     B_{2,1,3}^{j_1,*,j_0,*}  = f(A,x)_{j_1} \cdot  f(A,x)_{j_0} \cdot c(A,x)_{j_0} \cdot f_c(A,x) \cdot  {\bf 1}_d  \cdot ( (A_{j_1,*}) \circ x^\top  )
\end{align*}
\item {\bf Part 4.}
\begin{align*}
    B_{2,1,4}^{j_1,*,j_0,*}  = -  f(A,x)_{j_1} \cdot  f(A,x)_{j_0} \cdot c(A,x)_{j_0} \cdot f_c(A,x) \cdot {\bf 1}_d  \cdot  h(A,x)^\top 
\end{align*}
\item {\bf Part 5.}
\begin{align*}
    B_{2,1,5}^{j_1,*,j_0,*}  = & ~    f(A,x)_{j_1} \cdot  f(A,x)_{j_0} \cdot c(A,x)_{j_0}  \cdot  (-f_2(A,x) + f(A,x)_{j_1})   \cdot {\bf 1}_{d} \cdot h(A,x)^\top
\end{align*}
\item {\bf Part 6.}
\begin{align*}
    B_{2,1,6}^{j_1,*,j_0,*}  = & ~   f(A,x)_{j_1} \cdot  f(A,x)_{j_0} \cdot c(A,x)_{j_0}  \cdot (-f_c(A,x) + f(A,x)_{j_1})   \cdot {\bf 1}_d \cdot h(A,x)^\top
\end{align*}
\item {\bf Part 7.}
\begin{align*}
     B_{2,1,7}^{j_1,*,j_0,*}  = f(A,x)_{j_1} \cdot  f(A,x)_{j_0} \cdot c(A,x)_{j_0} \cdot {\bf 1}_d \cdot p_{j_1}(A,x)^\top
\end{align*}
\item {\bf Part 8.}
\begin{align*}
     B_{2,2,1}^{j_1,*,j_0,*}  =  f(A,x)_{j_1} \cdot  f(A,x)_{j_0} \cdot c(A,x)_{j_0} \cdot   x   \cdot   c_g(A,x)^\top 
\end{align*}
\item {\bf Part 9.}
\begin{align*}
    B_{2,3,1}^{j_1,*,j_0,*}  =   f(A,x)_{j_1} \cdot  f(A,x)_{j_0}^2 \cdot   x  \cdot   c_g(A,x)^\top  
\end{align*}

\end{itemize}
\begin{proof}
{\bf Proof of Part 1.}
    We have
    \begin{align*}
         B_{2,1,1}^{j_1,i_1,j_0,i_1}  = & ~   e_{i_1}^\top \cdot \langle c(A,x), f(A,x) \rangle \cdot  f(A,x)_{j_1} \cdot  f(A,x)_{j_0} \cdot c(A,x)_{j_0} \cdot e_{i_1} \\
        B_{2,1,1}^{j_1,i_1,j_0,i_0}  = & ~   e_{i_1}^\top \cdot \langle c(A,x), f(A,x) \rangle \cdot  f(A,x)_{j_1} \cdot  f(A,x)_{j_0} \cdot c(A,x)_{j_0} \cdot e_{i_0}
    \end{align*}

    From the above two equations, we can tell that $B_{2,1,1}^{j_1,*,j_0,*} \in \R^{d \times d}$ is a matrix that only diagonal has entries and off-diagonal are all zeros.
    
    Then we have $B_{2,1,1}^{j_1,*,j_0,*} \in \R^{d \times d}$ can be written as the rescaling of a diagonal matrix,
    \begin{align*}
     B_{2,1,1}^{j_1,*,j_0,*} & ~ =  f(A,x)_{j_1} \cdot  f(A,x)_{j_0} \cdot c(A,x)_{j_0} \cdot \langle c(A,x), f(A,x) \rangle \cdot  I_d   \\
     & ~ =  f(A,x)_{j_1} \cdot  f(A,x)_{j_0} \cdot c(A,x)_{j_0} \cdot f_c(A,x) \cdot  I_d
\end{align*}
    where the last step is follows from the Definitions~\ref{def:f_c}.

{\bf Proof of Part 2.}
    We have
    \begin{align*}
        B_{2,1,2}^{j_1,i_1,j_0,i_1} = & ~e_{i_1}^\top \cdot c(A,x)_{j_1} \cdot f(A,x)_{j_1} \cdot  f(A,x)_{j_0} \cdot c(A,x)_{j_0} \cdot  e_{i_1} \\
        B_{2,1,2}^{j_1,i_1,j_0,i_0} = & ~ e_{i_1}^\top \cdot c(A,x)_{j_1} \cdot f(A,x)_{j_1} \cdot  f(A,x)_{j_0} \cdot c(A,x)_{j_0} \cdot  e_{i_0}
    \end{align*}
     From the above two equations, we can tell that $B_{2,1,2}^{j_1,*,j_0,*} \in \R^{d \times d}$ is a matrix that only diagonal has entries and off-diagonal are all zeros.
    
    Then we have $B_{2,1,2}^{j_1,*,j_0,*} \in \R^{d \times d}$ can be written as the rescaling of a diagonal matrix,
\begin{align*}
     B_{2,1,2}^{j_1,*,j_1,*} & ~ = c(A,x)_{j_1} \cdot f(A,x)_{j_1} \cdot  f(A,x)_{j_0} \cdot c(A,x)_{j_0} \cdot I_d 
\end{align*}

{\bf Proof of Part 3.}
We have for diagonal entry and off-diagonal entry can be written as follows 
    \begin{align*}
        B_{2,1,3}^{j_1,i_1,j_0,i_1}  = & ~ f(A,x)_{j_1} \cdot \langle c(A,x), f(A,x) \rangle \cdot ( (A_{j_1,*}) \circ x^\top  )  \cdot  f(A,x)_{j_0} \cdot c(A,x)_{j_0} \cdot e_{i_1} \\
        B_{2,1,3}^{j_1,i_1,j_0,i_0}  = & ~ f(A,x)_{j_1} \cdot \langle c(A,x), f(A,x) \rangle \cdot ( (A_{j_1,*}) \circ x^\top  )  \cdot  f(A,x)_{j_0} \cdot c(A,x)_{j_0} \cdot e_{i_0}
    \end{align*}
From the above equation, we can show that matrix $B_{2,1,3}^{j_1,*,j_0,*}$ can be expressed as a rank-$1$ matrix,
\begin{align*}
     B_{2,1,3}^{j_1,*,j_0,*} & ~ = f(A,x)_{j_1} \cdot  f(A,x)_{j_0} \cdot c(A,x)_{j_0}  \cdot \langle c(A,x), f(A,x) \rangle \cdot  {\bf 1}_d  \cdot ( (A_{j_1,*}) \circ x^\top  ) \\
     & ~ =f(A,x)_{j_1} \cdot  f(A,x)_{j_0} \cdot c(A,x)_{j_0} \cdot f_c(A,x) \cdot  {\bf 1}_d  \cdot ( (A_{j_1,*}) \circ x^\top  ) 
\end{align*}
    where the last step is follows from the Definitions~\ref{def:f_c}.

{\bf Proof of Part 4.}
We have for diagonal entry and off-diagonal entry can be written as follows
    \begin{align*}
        B_{2,1,4}^{j_1,i_1,j_0,i_1}  = & ~ -  f(A,x)_{j_1} \cdot f(A,x)^\top  \cdot A \cdot \diag(x) \cdot   \langle c(A,x), f(A,x) \rangle \cdot  f(A,x)_{j_0} \cdot c(A,x)_{j_0} \cdot e_{i_1} \\
        B_{2,1,4}^{j_1,i_1,j_0,i_0}  = & ~ -  f(A,x)_{j_1} \cdot f(A,x)^\top  \cdot A \cdot \diag(x) \cdot   \langle c(A,x), f(A,x) \rangle \cdot  f(A,x)_{j_0} \cdot c(A,x)_{j_0} \cdot e_{i_0}
    \end{align*}
 From the above equation, we can show that matrix $B_{2,1,4}^{j_1,*,j_0,*}$ can be expressed as a rank-$1$ matrix,
\begin{align*}
    B_{2,1,4}^{j_1,*,j_0,*}  & ~ =  -  f(A,x)_{j_1} \cdot  f(A,x)_{j_0} \cdot c(A,x)_{j_0} \cdot  \langle c(A,x), f(A,x) \rangle \cdot  {\bf 1}_d  \cdot  f(A,x)^\top  \cdot A \cdot \diag(x)   \\
     & ~ = -  f(A,x)_{j_1} \cdot  f(A,x)_{j_0} \cdot c(A,x)_{j_0} \cdot f_c(A,x) \cdot {\bf 1}_d  \cdot  h(A,x)^\top 
\end{align*}
    where the last step is follows from the Definitions~\ref{def:h} and Definitions~\ref{def:f_c}.

{\bf Proof of Part 5.}
We have for diagonal entry and off-diagonal entry can be written as follows
    \begin{align*}
        B_{2,1,5}^{j_1,i_1,j_0,i_1}  = & ~  f(A,x)_{j_1} \cdot f(A,x)^\top  \cdot A \cdot \diag(x) \cdot (\langle -f(A,x), f(A,x) \rangle + f(A,x)_{j_1})  \cdot f(A,x)_{j_0} \cdot c(A,x)_{j_0} \cdot e_{i_1}  \\
        B_{2,1,5}^{j_1,i_1,j_0,i_0}  = & ~  f(A,x)_{j_1} \cdot f(A,x)^\top  \cdot A \cdot \diag(x) \cdot (\langle -f(A,x), f(A,x) \rangle + f(A,x)_{j_1})  \cdot f(A,x)_{j_0} \cdot c(A,x)_{j_0} \cdot e_{i_0} 
    \end{align*}
    From the above equation, we can show that matrix $B_{2,1,5}^{j_1,*,j_0,*}$ can be expressed as a rank-$1$ matrix,
\begin{align*}
    B_{2,1,5}^{j_1,*,j_0,*}  & ~ =    f(A,x)_{j_1} \cdot  f(A,x)_{j_0} \cdot c(A,x)_{j_0} \cdot  (\langle -f(A,x), f(A,x) \rangle + f(A,x)_{j_1})   \cdot {\bf 1}_{d} \cdot f(A,x)^\top  \cdot A \cdot \diag(x)\\
     & ~ =  f(A,x)_{j_1} \cdot  f(A,x)_{j_0} \cdot c(A,x)_{j_0}  \cdot  (-f_2(A,x) + f(A,x)_{j_1})   \cdot {\bf 1}_{d} \cdot h(A,x)^\top
\end{align*}
    where the last step is follows from the Definitions~\ref{def:h} and Definitions~\ref{def:f_2}.

{\bf Proof of Part 6.}
We have for diagonal entry and off-diagonal entry can be written as follows
    \begin{align*}
        B_{2,1,6}^{j_1,i_1,j_0,i_1}  = & ~   f(A,x)_{j_1} \cdot f(A,x)^\top  \cdot A \cdot  \diag(x) \cdot(\langle -f(A,x), c(A,x) \rangle + f(A,x)_{j_1}) \cdot f(A,x)_{j_0} \cdot c(A,x)_{j_0} \cdot e_{i_1} \\
        B_{2,1,6}^{j_1,i_1,j_0,i_0}  = & ~  f(A,x)_{j_1} \cdot f(A,x)^\top  \cdot A \cdot  \diag(x) \cdot(\langle -f(A,x), c(A,x) \rangle + f(A,x)_{j_1}) \cdot f(A,x)_{j_0} \cdot c(A,x)_{j_0} \cdot e_{i_0} 
    \end{align*}
    From the above equation, we can show that matrix $B_{2,1,6}^{j_1,*,j_0,*}$ can be expressed as a rank-$1$ matrix,
\begin{align*}
    B_{2,1,6}^{j_1,*,j_0,*}  & ~ =   f(A,x)_{j_1} \cdot  f(A,x)_{j_0} \cdot c(A,x)_{j_0}  \cdot (\langle -f(A,x), c(A,x) \rangle + f(A,x)_{j_1})  \cdot {\bf 1}_d \cdot f(A,x)^\top  \cdot A \cdot  \diag(x)\\
     & ~ =  f(A,x)_{j_1} \cdot  f(A,x)_{j_0} \cdot c(A,x)_{j_0}  \cdot (-f_c(A,x) + f(A,x)_{j_1})   \cdot {\bf 1}_d \cdot h(A,x)^\top
\end{align*}
    where the last step is follows from the Definitions~\ref{def:h} and Definitions~\ref{def:f_c}.
    
{\bf Proof of Part 7.}
We have for diagonal entry and off-diagonal entry can be written as follows
    \begin{align*}
         B_{2,1,7}^{j_1,i_1,j_0,i_1}  = & ~  f(A,x)_{j_1} \cdot ((e_{j_1}^\top - f(A,x)^\top) \circ q(A,x)^\top) \cdot A \cdot  \diag(x)  \cdot f(A,x)_{j_0} \cdot c(A,x)_{j_0} \cdot e_{i_1}\\
         B_{2,1,7}^{j_1,i_1,j_0,i_0}  = & ~  f(A,x)_{j_1} \cdot ((e_{j_1}^\top - f(A,x)^\top) \circ q(A,x)^\top) \cdot A \cdot  \diag(x)  \cdot f(A,x)_{j_0} \cdot c(A,x)_{j_0} \cdot e_{i_0}
    \end{align*}
    From the above equation, we can show that matrix $B_{2,1,7}^{j_1,*,j_0,*}$ can be expressed as a rank-$1$ matrix,
\begin{align*}
     B_{2,1,7}^{j_1,*,j_0,*}  & ~ =    f(A,x)_{j_1} \cdot  f(A,x)_{j_0} \cdot c(A,x)_{j_0} \cdot  {\bf 1}_d \cdot 
     ((e_{j_1}^\top - f(A,x)^\top) \circ q(A,x)^\top) \cdot A \cdot  \diag(x)  \\
     & ~ = f(A,x)_{j_1} \cdot  f(A,x)_{j_0} \cdot c(A,x)_{j_0} \cdot {\bf 1}_d \cdot p_{j_1}(A,x)^\top
\end{align*}
    where the last step is follows from the Definitions~\ref{def:p}.
    
{\bf Proof of Part 8.}
We have for diagonal entry and off-diagonal entry can be written as follows
    \begin{align*}
         B_{2,2,1}^{j_1,i_1,j_0,i_1}  = & ~  c_g(A,x)^{\top} \cdot f(A,x)_{j_1} \cdot  f(A,x)_{j_0} \cdot x_{i_1}  \cdot c(A,x)_{j_0} \cdot e_{i_1}  \\
         B_{2,2,1}^{j_1,i_1,j_0,i_0}  = & ~  c_g(A,x)^{\top} \cdot f(A,x)_{j_1} \cdot  f(A,x)_{j_0} \cdot x_{i_1}  \cdot c(A,x)_{j_0} \cdot e_{i_0} 
    \end{align*}
    From the above equation, we can show that matrix $B_{2,2,1}^{j_1,*,j_0,*}$ can be expressed as a rank-$1$ matrix,
\begin{align*}
     B_{2,2,1}^{j_1,*,j_0,*}  & ~ =  f(A,x)_{j_1} \cdot  f(A,x)_{j_0} \cdot c(A,x)_{j_0} \cdot   x   \cdot   c_g(A,x)^\top  
\end{align*}

{\bf Proof of Part 9.}
We have for diagonal entry and off-diagonal entry can be written as follows
    \begin{align*}
         B_{2,3,1}^{j_1,i_1,j_0,i_1}  = & ~     c_g(A,x)^{\top} \cdot f(A,x)_{j_1} \cdot  f(A,x)_{j_0}^2 \cdot x_{i_1} \cdot e_{i_1} \\
         B_{2,3,1}^{j_1,i_1,j_0,i_0}  = & ~     c_g(A,x)^{\top} \cdot f(A,x)_{j_1} \cdot  f(A,x)_{j_0}^2 \cdot x_{i_1} \cdot e_{i_0} 
    \end{align*}
            From the above equation, we can show that matrix $B_{2,3,1}^{j_1,*,j_0,*}$ can be expressed as a rank-$1$ matrix,
\begin{align*}
    B_{2,3,1}^{j_1,*,j_0,*}  & ~ =     f(A,x)_{j_1} \cdot  f(A,x)_{j_0}^2 \cdot   x  \cdot   c_g(A,x)^\top  
\end{align*}

\end{proof}
\end{lemma}

\subsection{Expanding \texorpdfstring{$B_2$}{} into many terms}
\begin{lemma}
   If the following conditions hold
    \begin{itemize}
     \item Let $u(A,x) \in \R^n$ be defined as Definition~\ref{def:u}
    \item Let $\alpha(A,x) \in \R$ be defined as Definition~\ref{def:alpha}
     \item Let $f(A,x) \in \R^n$ be defined as Definition~\ref{def:f}
    \item Let $c(A,x) \in \R^n$ be defined as Definition~\ref{def:c}
    \item Let $g(A,x) \in \R^d$ be defined as Definition~\ref{def:g} 
    \item Let $q(A,x) = c(A,x) + f(A,x) \in \R^n$
    \item Let $c_g(A,x) \in \R^d$ be defined as Definition~\ref{def:c_g}.
    \item Let $L_g(A,x) \in \R$ be defined as Definition~\ref{def:l_g}
    \item Let $v \in \R^n$ be a vector 
    \end{itemize}
Then, For $j_0,j_1 \in [n], i_0,i_1 \in [d]$, we have 
\begin{itemize}
    \item {\bf Part 1.}For $j_1 = j_0$ and $i_0 = i_1$ \begin{align*}
    B_{2}^{j_1,i_1,j_1,i_1}
    = & ~ B_{2,1}^{j_1,i_1,j_1,i_1}  + B_{2,2}^{j_1,i_1,j_1,i_1}  +B_{2,3}^{j_1,i_1,j_1,i_1} 
\end{align*}
\item {\bf Part 2.}For $j_1 = j_0$ and $i_0 \neq i_1$ \begin{align*}
    B_{2}^{j_1,i_1,j_1,i_0}
    = & ~ B_{2,1}^{j_1,i_1,j_1,i_0}  + B_{2,2}^{j_1,i_1,j_1,i_0}  +B_{2,3}^{j_1,i_1,j_1,i_0} 
\end{align*}
\item {\bf Part 3.}For $j_1 \neq j_0$ and $i_0 = i_1$ \begin{align*}
    B_{2}^{j_1,i_1,j_0,i_1}
    = & ~ B_{2,1}^{j_1,i_1,j_0,i_1}  + B_{2,2}^{j_1,i_1,j_0,i_1}  +B_{2,3}^{j_1,i_1,j_0,i_1} 
\end{align*}
\item  {\bf Part 4.} For $j_1 \neq j_0$ and $i_0 \neq i_1$
\begin{align*}
    B_{2}^{j_1,i_1,j_0,i_0}
    = & ~ B_{2,1}^{j_1,i_1,j_0,i_0}  + B_{2,2}^{j_1,i_1,j_0,i_0}  +B_{2,3}^{j_1,i_1,j_0,i_0} 
\end{align*}
\end{itemize}

\end{lemma}
\begin{proof}
{\bf Proof of Part 1.}
we have
    \begin{align*}
    B_{2}^{j_1,i_1,j_1,i_1} = & ~\frac{\d}{\d A_{j_1,i_1}}(- c_g(A,x)^{\top} \cdot f(A,x)_{j_1} \cdot c(A,x)_{j_1} \cdot e_{i_1} )\\
    = & ~ B_{2,1}^{j_1,i_1,j_1,i_1}  + B_{2,2}^{j_1,i_1,j_1,i_1}  +B_{2,3}^{j_1,i_1,j_1,i_1} 
\end{align*}
{\bf Proof of Part 2.}
we have
    \begin{align*}
    B_{2}^{j_1,i_1,j_1,i_0}
    = & ~\frac{\d}{\d A_{j_1,i_1}}(- c_g(A,x)^{\top} \cdot f(A,x)_{j_1} \cdot c(A,x)_{j_1} \cdot e_{i_0} )\\
    = & ~ B_{2,1}^{j_1,i_1,j_1,i_0}  + B_{2,2}^{j_1,i_1,j_1,i_0}  +B_{2,3}^{j_1,i_1,j_1,i_0}
\end{align*}
{\bf Proof of Part 3.}
 we have
    \begin{align*}
    B_{2}^{j_1,i_1,j_0,i_1}
    = & ~\frac{\d}{\d A_{j_1,i_1}}(- c_g(A,x)^{\top} \cdot f(A,x)_{j_0} \cdot c(A,x)_{j_0} \cdot e_{i_0} )\\
    = & ~ B_{2,1}^{j_1,i_1,j_0,i_1}  + B_{2,2}^{j_1,i_1,j_0,i_1}  +B_{2,3}^{j_1,i_1,j_0,i_1} 
\end{align*}

{\bf Proof of Part 4.}
 we have
\begin{align*}
    B_{2}^{j_1,i_1,j_0,i_0}
    =& ~\frac{\d}{\d A_{j_1,i_1}}(- c_g(A,x)^{\top} \cdot f(A,x)_{j_0} \cdot c(A,x)_{j_0} \cdot e_{i_0} )\\
    = & ~B_{2,1}^{j_1,i_1,j_0,i_0}  + B_{2,2}^{j_1,i_1,j_0,i_0}  +B_{2,3}^{j_1,i_1,j_0,i_0} 
\end{align*}
\end{proof}

\subsection{Lipschitz Computation}
\begin{lemma}\label{lips: B_2}
If the following conditions hold
\begin{itemize}
    \item Let $B_{2,1,1}^{j_1,*, j_0,*}, \cdots, B_{2,3,1}^{j_1,*, j_0,*} $ be defined as Lemma~\ref{lem:b_2_j1_j0} 
    \item  Let $\|A \|_2 \leq R, \|A^{\top} \|_F \leq R, \| x\|_2 \leq R, \|\diag(f(A,x)) \|_F \leq \|f(A,x) \|_2 \leq 1, \| b_g \|_2 \leq 1$ 
\end{itemize}
Then, we have
\begin{itemize}
    \item {\bf Part 1.}
    \begin{align*}
       \| B_{2,1,1}^{j_1,*,j_0,*} (A) - B_{2,1,1}^{j_1,*,j_0,*} ( \wt{A} ) \|_F \leq \beta^{-2} \cdot n \cdot \sqrt{d}\exp(4R^2) \cdot \|A - \wt{A}\|_F
    \end{align*}
     \item {\bf Part 2.}
    \begin{align*}
       \| B_{2,1,2}^{j_1,*,j_0,*} (A) - B_{2,1,1}^{j_1,*,j_0,*} ( \wt{A} ) \|_F \leq  \beta^{-2} \cdot n \cdot \sqrt{d}\exp(4R^2) \cdot \|A - \wt{A}\|_F 
    \end{align*}
     \item {\bf Part 3.}
    \begin{align*}
       \| B_{2,1,3}^{j_1,*,j_0,*} (A) - B_{2,1,1}^{j_1,*,j_0,*} ( \wt{A} ) \|_F \leq  \beta^{-2} \cdot n \cdot \sqrt{d}  \cdot \exp(5R^2)\|A - \wt{A}\|_F
    \end{align*}
     \item {\bf Part 4.}
    \begin{align*}
       \| B_{2,1,4}^{j_1,*,j_0,*} (A) - B_{2,1,1}^{j_1,*,j_0,*} ( \wt{A} ) \|_F \leq  \beta^{-2} \cdot n \cdot \sqrt{d} \cdot \exp(5R^2) \cdot \|A - \wt{A}\|_F 
    \end{align*}
     \item {\bf Part 5.}
    \begin{align*}
       \| B_{2,1,5}^{j_1,*,j_0,*} (A) - B_{2,1,1}^{j_1,*,j_0,*} ( \wt{A} ) \|_F \leq  \beta^{-2} \cdot n \cdot \sqrt{d}  \cdot \exp(5R^2) \cdot \|A - \wt{A}\|_F
    \end{align*}
     \item {\bf Part 6.}
    \begin{align*}
       \| B_{2,1,6}^{j_1,*,j_0,*} (A) - B_{2,1,1}^{j_1,*,j_0,*} ( \wt{A} ) \|_F \leq  \beta^{-2} \cdot n \cdot \sqrt{d}  \cdot \exp(5R^2) \cdot \|A - \wt{A}\|_F
    \end{align*}
     \item {\bf Part 7.}
    \begin{align*}
       \| B_{2,1,7}^{j_1,*,j_0,*} (A) - B_{2,1,1}^{j_1,*,j_0,*} ( \wt{A} ) \|_F \leq  \beta^{-2} \cdot n \cdot \sqrt{d}  \cdot \exp(5R^2) \cdot \|A - \wt{A}\|_F
    \end{align*}
    \item {\bf Part 8.}
    \begin{align*}
       \| B_{2,2,1}^{j_1,*,j_0,*} (A) - B_{2,1,1}^{j_1,*,j_0,*} ( \wt{A} ) \|_F \leq  \beta^{-2} \cdot n   \cdot \exp(5R^2) \cdot \|A - \wt{A}\|_F
    \end{align*}
     \item {\bf Part 9.}
    \begin{align*}
       \| B_{2,3,1}^{j_1,*,j_0,*} (A) - B_{2,1,1}^{j_1,*,j_0,*} ( \wt{A} ) \|_F \leq  \beta^{-2} \cdot n   \cdot \exp(5R^2) \cdot \|A - \wt{A}\|_F
    \end{align*}
        \item{\bf Part 10.}
    \begin{align*}
         \| B_{2}^{j_1,*,j_0,*} (A) - B_{2}^{j_1,*,j_0,*} ( \wt{A} ) \|_F \leq & ~ 9 \beta^{-2} \cdot n \cdot \sqrt{d} \cdot \exp(5R^2)\|A - \wt{A}\|_F
    \end{align*}
    \end{itemize}
\end{lemma}
\begin{proof}
{\bf Proof of Part 1.}
\begin{align*}
& ~ \| B_{2,1,1}^{j_1,*,j_0,*} (A) - B_{2,1,1}^{j_1,*,j_0,*} ( \wt{A} ) \| \\ \leq 
    & ~ \|  f(A,x)_{j_1} \cdot  f(A,x)_{j_0} \cdot c(A,x)_{j_0} \cdot f_c(A,x) \cdot  I_d  - f(\wt{A},x)_{j_1} \cdot  f(\wt{A},x)_{j_0} \cdot c(\wt{A},x)_{j_0} \cdot f_c(\wt{A},x) \cdot  I_d \\
    \leq & ~ |  f(A,x)_{j_1} - f(\wt{A},x)_{j_1}|\cdot |f(A,x)_{j_0}|\cdot | c(A,x)_{j_0}|\cdot | f_c(A,x) | \cdot \| I_d\|_F \\
    & + ~ |f(\wt{A},x)_{j_1}|\cdot |f(A,x)_{j_0} - f(\wt{A},x)_{j_0}| \cdot | c(A,x)_{j_0}|\cdot | f_c(A,x) | \cdot \| I_d\|_F\\
    & + ~  |f(\wt{A},x)_{j_1}| \cdot| f(\wt{A},x)_{j_0}| \cdot|c(A,x)_{j_0} - c(\wt{A},x)_{j_0}| \cdot | f_c(A,x) |\| I_d\|_F \\
    & + ~ |f(\wt{A},x)_{j_1}|\cdot | f(\wt{A},x)_{j_0}|\cdot|  c(\wt{A},x)_{j_0} | \cdot|f_c(A,x) - f_c(\wt{A},x)| \cdot\| I_d\|_F\\
    \leq & ~ 8  \beta^{-2} \cdot n \cdot \sqrt{d}  \exp(3R^2)\|A - \wt{A}\|_F \\
    &+ ~ 8  \beta^{-2} \cdot n \cdot \sqrt{d}  \exp(3R^2)\|A - \wt{A}\|_F \\
    & + ~ 4\beta^{-2} \cdot n \cdot \sqrt{d}\exp(3R^2) \cdot \|A - \wt{A}\|_F \\
    & + ~ 12\beta^{-2} \cdot n \cdot \sqrt{d}\exp(3R^2) \cdot \|A - \wt{A}\|_F \\
    \leq &  ~ 32\beta^{-2} \cdot n \cdot \sqrt{d}\exp(3R^2) \cdot \|A - \wt{A}\|_F \\
    \leq &  ~ \beta^{-2} \cdot n \cdot \sqrt{d}\exp(4R^2) \cdot \|A - \wt{A}\|_F 
\end{align*}

{\bf Proof of Part 2.}
\begin{align*}
    & ~ \| B_{2,1,2}^{j_1,*,j_0,*} (A) - B_{2,1,2}^{j_1,*,j_0,*} ( \wt{A} ) \|_F \\
    \leq & ~ \| c(A,x)_{j_1} \cdot f(A,x)_{j_1} \cdot  f(A,x)_{j_0} \cdot c(A,x)_{j_0} \cdot I_d   - c(\wt{A},x)_{j_1} \cdot f(\wt{A},x)_{j_1} \cdot  f(\wt{A},x)_{j_0} \cdot c(\wt{A},x)_{j_0} \cdot I_d\|_F \\
    \leq & ~  |c(A,x)_{j_1} - c(\wt{A},x)_{j_1}| \cdot |f(A,x)_{j_1} |\cdot  |f(A,x)_{j_0} |\cdot |c(A,x)_{j_0} |\cdot \| I_d \|_F \\
    & + ~ | c(\wt{A},x)_{j_1}| \cdot |f(A,x)_{j_1} - f(\wt{A},x)_{j_1}|\cdot  |f(A,x)_{j_0} |\cdot |c(A,x)_{j_0} |\cdot \| I_d \|_F \\
    & + ~ | c(\wt{A},x)_{j_1}| \cdot |f(\wt{A},x)_{j_1}|\cdot  |f(A,x)_{j_0} -f(\wt{A},x)_{j_0} |\cdot |c(A,x)_{j_0} |\cdot \| I_d \|_F \\
    & + ~ | c(\wt{A},x)_{j_1}| \cdot |f(\wt{A},x)_{j_1}|\cdot  |f(\wt{A},x)_{j_0} |\cdot |c(A,x)_{j_0} - c(\wt{A},x)_{j_0} |\cdot \| I_d \|_F \\
    \leq & ~ 4\beta^{-2} \cdot n \cdot \sqrt{d}\exp(3R^2) \cdot \|A - \wt{A}\|_F \\
    & + ~ 8\beta^{-2} \cdot n \cdot \sqrt{d}\exp(3R^2) \cdot \|A - \wt{A}\|_F \\
    & + ~ 8\beta^{-2} \cdot n \cdot \sqrt{d}\exp(3R^2) \cdot \|A - \wt{A}\|_F \\
    & + ~ 4\beta^{-2} \cdot n \cdot \sqrt{d}\exp(3R^2) \cdot \|A - \wt{A}\|_F \\
    \leq & ~ 24 \beta^{-2} \cdot n \cdot \sqrt{d}\exp(3R^2) \cdot \|A - \wt{A}\|_F \\
    \leq & ~ \beta^{-2} \cdot n \cdot \sqrt{d}\exp(4R^2) \cdot \|A - \wt{A}\|_F 
\end{align*}

{\bf Proof of Part 3.}
\begin{align*}
     & ~ \| B_{2,1,3}^{j_1,*,j_0,*} (A) - B_{2,1,3}^{j_1,*,j_0,*} ( \wt{A} ) \|_F \\
      \leq & ~ \| f(A,x)_{j_1} \cdot  f(A,x)_{j_0} \cdot c(A,x)_{j_0} \cdot f_c(A,x) \cdot  {\bf 1}_d  \cdot ( (A_{j_1,*}) \circ x^\top  ) \\
      & - f(\wt{A},x)_{j_1} \cdot  f(\wt{A},x)_{j_0} \cdot c(\wt{A},x)_{j_0} \cdot f_c(\wt{A},x) \cdot  {\bf 1}_d  \cdot ( (\wt{A}_{j_1,*}) \circ x^\top  ) \|\\
   \leq & ~ |f(A,x)_{j_1} - f(\wt{A},x)_{j_1}| \cdot |f(A,x)_{j_0}| \cdot |c(A,x)_{j_0}| \cdot |f_c(A,x)|  \cdot \|{\bf 1}_d \|_2\cdot \|A_{j_1,*} \|_2 \cdot \| \diag(x)\|_F  \\
     & + ~ | f(\wt{A},x)_{j_1}| \cdot |f(A,x)_{j_0} -f(\wt{A},x)_{j_0}  | \cdot |c(A,x)_{j_0}| \cdot |f_c(A,x)| \cdot \|{\bf 1}_d \|_2\cdot \|A_{j_1,*} \|_2 \cdot  \| \diag(x)\|_F  \\
     & + ~ | f(\wt{A},x)_{j_1}| \cdot |f(\wt{A},x)_{j_0}  | \cdot |c(A,x)_{j_0} - c(\wt{A},x)_{j_0}|\cdot |f_c(A,x)| \cdot \|{\bf 1}_d \|_2\cdot \|A_{j_1,*} \|_2 \cdot  \| \diag(x)\|_F  \\
      & + ~ | f(\wt{A},x)_{j_1}|\cdot  |f(\wt{A},x)_{j_0}  |\cdot  |c(\wt{A},x)_{j_0}|\cdot |f_c(A,x) - f_c(\wt{A},x)| \cdot \|{\bf 1}_d \|_2\cdot \|A_{j_1,*} \|_2 \cdot  \| \diag(x)\|_F  \\
     & + ~| f(\wt{A},x)_{j_1}|\cdot  |f(\wt{A},x)_{j_0}  |\cdot  |c(\wt{A},x)_{j_0}|\cdot |f_c(\wt{A},x)| \cdot \|{\bf 1}_d \|_2\cdot \|A_{j_1,*} - \wt{A}_{j_1,*}\|_2 \cdot  \| \diag(x)\|_F  \\
    \leq & ~ 8 R^2 \cdot \beta^{-2} \cdot n \cdot \sqrt{d}  \cdot \exp(3R^2)\|A - \wt{A}\|_F \\
    & + ~ 8 R^2 \cdot \beta^{-2} \cdot n \cdot \sqrt{d}  \cdot \exp(3R^2)\|A - \wt{A}\|_F \\
    & + ~ 4 R^2 \cdot \beta^{-2} \cdot n \cdot \sqrt{d}  \cdot \exp(3R^2)\|A - \wt{A}\|_F \\
    & + ~12 R^2 \cdot \beta^{-2} \cdot n \cdot \sqrt{d}  \cdot \exp(3R^2)\|A - \wt{A}\|_F \\
    & + ~ 4 R \cdot \sqrt{d} \cdot \|A - \wt{A} \|_F \\
    \leq & ~ 36  \cdot \beta^{-2} \cdot n \cdot \sqrt{d}  \cdot \exp(4R^2)\|A - \wt{A}\|_F\\
    \leq & ~\beta^{-2} \cdot n \cdot \sqrt{d}  \cdot \exp(5R^2)\|A - \wt{A}\|_F
\end{align*}

{\bf Proof of Part 4.}
\begin{align*}
     & ~ \| B_{2,1,4}^{j_1,*,j_0,*} (A) - B_{2,1,4}^{j_1,*,j_0,*} ( \wt{A} ) \|_F \\
     \leq & ~ \|-  f(A,x)_{j_1} \cdot  f(A,x)_{j_0} \cdot c(A,x)_{j_0} \cdot f_c(A,x) \cdot {\bf 1}_d  \cdot  h(A,x)^\top  \\
     & ~ -(-  f(\wt{A},x)_{j_1} \cdot  f(\wt{A},x)_{j_0} \cdot c(\wt{A},x)_{j_0} \cdot f_c(\wt{A},x) \cdot {\bf 1}_d  \cdot  h(\wt{A},x)^\top ) \|_F \\
      \leq & ~ \|  f(A,x)_{j_1} \cdot  f(A,x)_{j_0} \cdot c(A,x)_{j_0} \cdot f_c(A,x) \cdot {\bf 1}_d  \cdot  h(A,x)^\top\\
      & ~ - f(\wt{A},x)_{j_1} \cdot  f(\wt{A},x)_{j_0} \cdot c(\wt{A},x)_{j_0} \cdot f_c(\wt{A},x) \cdot {\bf 1}_d  \cdot  h(\wt{A},x)^\top \|_F \\
      \leq & ~ |f(A,x)_{j_1} - f(\wt{A},x)_{j_1}| \cdot |f(A,x)_{j_0}| \cdot |c(A,x)_{j_0}| \cdot |f_c(A,x)| \cdot \| {\bf 1}_d \|_2 \cdot  \| h(A,x)^\top  \|_2\\
      & + ~ | f(\wt{A},x)_{j_1}| \cdot |f(A,x)_{j_0} -f(\wt{A},x)_{j_0}  | \cdot |c(A,x)_{j_0}| \cdot |f_c(A,x)| \cdot \| {\bf 1}_d \|_2 \cdot  \| h(A,x)^\top  \|_2\\
     & + ~ | f(\wt{A},x)_{j_1}| \cdot |f(\wt{A},x)_{j_0}  | \cdot |c(A,x)_{j_0} - c(\wt{A},x)_{j_0}|\cdot |f_c(A,x)| \cdot \| {\bf 1}_d \|_2 \cdot \| h(A,x)^\top  \|_2\\
    & + ~ | f(\wt{A},x)_{j_1}|\cdot  |f(\wt{A},x)_{j_0}  |\cdot  |c(\wt{A},x)_{j_0}|\cdot |f_c(A,x) - f_c(\wt{A},x)| \cdot \| {\bf 1}_d \|_2 \cdot \| h(A,x)^\top  \|_2\\
    & + ~  | f(\wt{A},x)_{j_1}|\cdot  |f(\wt{A},x)_{j_0}  |\cdot  |c(\wt{A},x)_{j_0}|\cdot |  f_c(\wt{A},x)| \cdot \| {\bf 1}_d \|_2 \cdot \| h(A,x)^\top - h(\wt{A},x)^\top \|_2\\
    \leq & ~ 8 R^2 \cdot \beta^{-2} \cdot n \cdot \sqrt{d}  \cdot \exp(3R^2)\|A - \wt{A}\|_F \\
    & + ~ 8 R^2 \cdot \beta^{-2} \cdot n \cdot \sqrt{d}  \cdot \exp(3R^2)\|A - \wt{A}\|_F \\
    & + ~ 4 R^2 \cdot \beta^{-2} \cdot n \cdot \sqrt{d}  \cdot \exp(3R^2)\|A - \wt{A}\|_F\\
    & + ~ 12 R^2 \cdot \beta^{-2} \cdot n \cdot \sqrt{d}  \cdot \exp(3R^2)\|A - \wt{A}\|_F\\ 
    & + ~12\beta^{-2} \cdot n \cdot \sqrt{d} \cdot \exp(4R^2) \cdot \|A - \wt{A}\|_F \\
   \leq &  ~\beta^{-2} \cdot n \cdot \sqrt{d} \cdot \exp(5R^2) \cdot \|A - \wt{A}\|_F \\
\end{align*}

{\bf Proof of Part 5.}
\begin{align*}
    & ~ \| B_{2,1,5}^{j_1,*,j_0,*} (A) - B_{2,1,5}^{j_1,*,j_0,*} ( \wt{A} ) \|_F \\
    = & ~ \| f(A,x)_{j_1} \cdot  f(A,x)_{j_0} \cdot c(A,x)_{j_0}  \cdot  (-f_2(A,x) + f(A,x)_{j_1})   \cdot {\bf 1}_{d} \cdot h(A,x)^\top \\
    & ~ - f(\wt{A},x)_{j_1} \cdot f(\wt{A},x)_{j_0} \cdot c(\wt{A},x)_{j_0}  \cdot (-f_2(\wt{A},x) + f(\wt{A},x)_{j_1})   \cdot {\bf 1}_d  \cdot h(\wt{A},x)^\top \|_F \\
    \leq & ~ |f(A,x)_{j_1} -f(\wt{A},x)_{j_1} |  \cdot |f(A,x)_{j_0}| \cdot |c(A,x)_{j_0}|\cdot  (| -f_2(A,x)| + |f(A,x)_{j_1}|)   \cdot \|{\bf 1}_d \|_2 \cdot \|h(A,x)^\top  \|_2 \\
    & + ~ |f(\wt{A},x)_{j_1} |  \cdot |f(A,x)_{j_0} - f(\wt{A},x)_{j_0}| \cdot |c(A,x)_{j_0}|\cdot   (| -f_2(A,x)| + |f(A,x)_{j_1}|)  \cdot \|{\bf 1}_d \|_2 \cdot \|h(A,x)^\top \|_2 \\
     & + ~ |f(\wt{A},x)_{j_1} |  \cdot |f(\wt{A},x)_{j_0}| \cdot |c(A,x)_{j_0} - c(\wt{A},x)_{j_0}| \cdot (| -f_2(A,x)| + |f(A,x)_{j_1}|)  \cdot \|{\bf 1}_d \|_2 \cdot \|h(A,x)^\top \|_2 \\
      & + ~ |f(\wt{A},x)_{j_1} |  \cdot |f(\wt{A},x)_{j_0}| \cdot |c(\wt{A},x)_{j_0}| \cdot (| f_2(A,x) -f_2(\wt{A},x)| + |f(A,x)_{j_1} - f(\wt{A},x)_{j_1}| )  \cdot \|{\bf 1}_d \|_2 \cdot \|h(A,x)^\top \|_2 \\
       & + ~ |f(\wt{A},x)_{j_1} |  \cdot |f(\wt{A},x)_{j_0}| \cdot |c(\wt{A},x)_{j_0}| \cdot  (| -f_2(\wt{A},x)| + |f(\wt{A},x)_{j_1}|)  \cdot \|{\bf 1}_d \|_2 \cdot \|h(A,x)^\top - h(\wt{A},x)^\top \|_2 \\
    \leq & ~  8R^2 \cdot  \beta^{-2} \cdot n \cdot \sqrt{d}  \cdot \exp(3R^2)\|A - \wt{A}\|_F \\
    & + ~  8R^2 \cdot  \beta^{-2} \cdot n \cdot \sqrt{d}  \cdot \exp(3R^2)\|A - \wt{A}\|_F \\
     & + ~4R^2 \cdot  \beta^{-2} \cdot n \cdot \sqrt{d}  \cdot \exp(3R^2)\|A - \wt{A}\|_F \\
     & + ~12 R^2 \cdot  \beta^{-2} \cdot n \cdot \sqrt{d}  \cdot \exp(3R^2)\|A - \wt{A}\|_F \\
     & + ~ 12 \beta^{-2} \cdot n \cdot \exp(4R^2) \cdot \|A - \wt{A}\|_F \\
     \leq & ~ 44 \beta^{-2} \cdot n \cdot \sqrt{d}  \cdot \exp(4R^2) \cdot \|A - \wt{A}\|_F \\
     \leq & ~  \beta^{-2} \cdot n \cdot \sqrt{d}  \cdot \exp(5R^2) \cdot \|A - \wt{A}\|_F
\end{align*}

{\bf Proof of Part 6.}
\begin{align*}
 & ~ \| B_{2,1,6}^{j_1,*,j_0,*} (A) - B_{2,1,6}^{j_1,*,j_0,*} ( \wt{A} ) \|_F \\
    = & ~ \| f(A,x)_{j_1} \cdot  f(A,x)_{j_0} \cdot c(A,x)_{j_0}  \cdot  (-f_c(A,x) + f(A,x)_{j_1})   \cdot {\bf 1}_{d} \cdot h(A,x)^\top \\
    & ~ - f(\wt{A},x)_{j_1} \cdot f(\wt{A},x)_{j_0} \cdot c(\wt{A},x)_{j_0}  \cdot (-f_c(\wt{A},x) + f(\wt{A},x)_{j_1})   \cdot {\bf 1}_d  \cdot h(\wt{A},x)^\top \|_F \\
    \leq & ~ |f(A,x)_{j_1} -f(\wt{A},x)_{j_1} |  \cdot |f(A,x)_{j_0}| \cdot |c(A,x)_{j_0}|\cdot  (| -f_c(A,x)| + |f(A,x)_{j_1}|)   \cdot \|{\bf 1}_d \|_2 \cdot \|h(A,x)^\top  \|_2 \\
    & + ~ |f(\wt{A},x)_{j_1} |  \cdot |f(A,x)_{j_0} - f(\wt{A},x)_{j_0}| \cdot |c(A,x)_{j_0}|\cdot   (| -f_c(A,x)| + |f(A,x)_{j_1}|)  \cdot \|{\bf 1}_d \|_2 \cdot \|h(A,x)^\top \|_2 \\
     & + ~ |f(\wt{A},x)_{j_1} |  \cdot |f(\wt{A},x)_{j_0}| \cdot |c(A,x)_{j_0} - c(\wt{A},x)_{j_0}| \cdot (| -f_c(A,x)| + |f(A,x)_{j_1}|)  \cdot \|{\bf 1}_d \|_2 \cdot \|h(A,x)^\top \|_2 \\
      & + ~ |f(\wt{A},x)_{j_1} |  \cdot |f(\wt{A},x)_{j_0}| \cdot |c(\wt{A},x)_{j_0}| \cdot (| f_c(A,x) -f_c(\wt{A},x)| + |f(A,x)_{j_1} - f(\wt{A},x)_{j_1}| )  \cdot \|{\bf 1}_d \|_2 \cdot \|h(A,x)^\top \|_2 \\
       & + ~ |f(\wt{A},x)_{j_1} |  \cdot |f(\wt{A},x)_{j_0}| \cdot |c(\wt{A},x)_{j_0}| \cdot  (| -f_c(\wt{A},x)| + |f(\wt{A},x)_{j_1}|)  \cdot \|{\bf 1}_d \|_2 \cdot \|h(A,x)^\top - h(\wt{A},x)^\top \|_2 \\
    \leq & ~  12R^2 \cdot  \beta^{-2} \cdot n \cdot \sqrt{d}  \cdot \exp(3R^2)\|A - \wt{A}\|_F \\
    & + ~  12R^2 \cdot  \beta^{-2} \cdot n \cdot \sqrt{d}  \cdot \exp(3R^2)\|A - \wt{A}\|_F \\
     & + ~6R^2 \cdot  \beta^{-2} \cdot n \cdot \sqrt{d}  \cdot \exp(3R^2)\|A - \wt{A}\|_F \\
     & + ~16 R^2 \cdot  \beta^{-2} \cdot n \cdot \sqrt{d}  \cdot \exp(3R^2)\|A - \wt{A}\|_F \\
     & + ~ 18 \beta^{-2} \cdot n \cdot \exp(4R^2) \cdot \|A - \wt{A}\|_F \\
     \leq & ~ 64 \beta^{-2} \cdot n \cdot \sqrt{d}  \cdot \exp(4R^2) \cdot \|A - \wt{A}\|_F \\
     \leq & ~  \beta^{-2} \cdot n \cdot \sqrt{d}  \cdot \exp(5R^2) \cdot \|A - \wt{A}\|_F
\end{align*}

{\bf Proof of Part 7.}
\begin{align*}
& ~ \|B_{2,1,7}^{j_1,*,j_0,*} (A) - B_{2,1,7}^{j_1,*,j_0,*} ( \wt{A} )   \|_F 
\\
  \leq & ~   \|f(A,x)_{j_1} \cdot f(A,x)_{j_0} \cdot c(A,x)_{j_0} \cdot {\bf 1}_d  \cdot p_{j_1}(A,x)^\top  - f(\wt{A},x)_{j_1} \cdot f(\wt{A},x)_{j_0} \cdot c(\wt{A},x)_{j_0} \cdot {\bf 1}_d  \cdot p_{j_1}(\wt{A},x)^\top  \|_F\\
  \leq & ~ |f(A,x)_{j_1} - f(\wt{A},x)_{j_1}| \cdot |f(A,x)_{j_0} |\cdot |c(A,x)_{j_0}| \cdot \|{\bf 1}_d \|_2  \cdot \|p_{j_1}(A,x)^\top\|_2 \\
  & + ~ | f(\wt{A},x)_{j_1}| \cdot |f(A,x)_{j_0} - f(\wt{A},x)_{j_0} |\cdot |c(A,x)_{j_0}| \cdot \|{\bf 1}_d \|_2  \cdot \|p_{j_1}(A,x)^\top\|_2 \\
   & + ~ | f(\wt{A},x)_{j_1}| \cdot |f(\wt{A},x)_{j_0} |\cdot |c(A,x)_{j_0}- c(\wt{A},x)_{j_0}| \cdot \|{\bf 1}_d \|_2  \cdot \|p_{j_1}(A,x)^\top\|_2 \\
   & + ~ | f(\wt{A},x)_{j_1}| \cdot |f(\wt{A},x)_{j_0} |\cdot  |c(\wt{A},x)_{j_0}| \cdot \|{\bf 1}_d \|_2  \cdot \|p_{j_1}(A,x)^\top - p_{j_1}(\wt{A},x)^\top\|_2 \\
   \leq & ~ 24R^2 \cdot  \beta^{-2} \cdot n \cdot \sqrt{d}  \cdot \exp(3R^2)\|A - \wt{A}\|_F \\
   & + ~24R^2 \cdot  \beta^{-2} \cdot n \cdot \sqrt{d}  \cdot \exp(3R^2)\|A - \wt{A}\|_F \\
   & + ~12R^2 \cdot  \beta^{-2} \cdot n \cdot \sqrt{d}  \cdot \exp(3R^2)\|A - \wt{A}\|_F \\
   & + ~26 \beta^{-2} \cdot n   \cdot \exp(4R^2) \cdot \|A - \wt{A} \|_F \\
   \leq & ~ 86 \beta^{-2} \cdot n  \cdot \sqrt{d} \cdot \exp(4R^2) \cdot \|A - \wt{A} \|_F\\
   \leq & ~ \beta^{-2} \cdot n \cdot \sqrt{d} \cdot \exp(5R^2) \cdot \|A - \wt{A} \|_F
\end{align*}

{\bf Proof of Part 8.}
\begin{align*}
    & ~ \|B_{2,2,1}^{j_1,*,j_0,*} (A) - B_{2,2,1}^{j_1,*,j_0,*} ( \wt{A} )   \|_F \\
    \leq & ~ \| f(A,x)_{j_1} \cdot  f(A,x)_{j_0} \cdot c(A,x)_{j_0} \cdot x \cdot c_g(A,x)^{\top} -  f(\wt{A},x)_{j_1} \cdot  f(\wt{A},x)_{j_0} \cdot  c(\wt{A},x)_{j_0} \cdot x \cdot c_g(\wt{A},x)^{\top} \|_F\\
    \leq & ~ |f(A,x)_{j_1} -f(\wt{A},x)_{j_1}| \cdot  |f(A,x)_{j_0} |\cdot  |c(A,x)_{j_0} |\cdot \| x\|_2 \cdot \|c_g(A,x)^{\top}\|_2 \\
    \leq & ~ |f(\wt{A},x)_{j_1}| \cdot  |f(A,x)_{j_0} - f(\wt{A},x)_{j_0} |\cdot  |c(A,x)_{j_0} |\cdot \| x\|_2 \cdot \|c_g(A,x)^{\top}\|_2 \\
    \leq & ~ |f(\wt{A},x)_{j_1}| \cdot  |f(\wt{A},x)_{j_0} |\cdot  |c(A,x)_{j_0} - c(\wt{A},x)_{j_0} |\cdot \| x\|_2 \cdot \|c_g(A,x)^{\top}\|_2 \\
    \leq & ~ |f(\wt{A},x)_{j_1}| \cdot  |f(\wt{A},x)_{j_0} |\cdot  |c(\wt{A},x)_{j_0} |\cdot \| x\|_2 \cdot \|c_g(A,x)^{\top} - c_g(\wt{A},x)^{\top}\|_2 \\
    \leq & ~ 20 R^2 \cdot  \beta^{-2} \cdot n  \cdot \exp(3R^2)\|A - \wt{A}\|_F \\
     & + ~ 20 R^2 \cdot  \beta^{-2} \cdot n \cdot   \exp(3R^2)\|A - \wt{A}\|_F \\
     & + ~ 10 R^2 \cdot  \beta^{-2} \cdot n \cdot  \exp(3R^2)\|A - \wt{A}\|_F \\
     & + ~ 40 R^2 \beta^{-2} \cdot n \cdot  \exp(3R^{2}) \cdot \|A - \wt{A}\|_F\\
     \leq & ~ 90  \cdot  \beta^{-2} \cdot n  \cdot \exp(4R^2)\|A - \wt{A}\|_F \\
     \leq & ~  \beta^{-2} \cdot n  \cdot \exp(5R^2)\|A - \wt{A}\|_F \\
\end{align*}

{\bf Proof of Part 9.}
\begin{align*}
     & ~ \| B_{2,3,1}^{j_1,*,j_0,*}(A) -   B_{2,3,1}^{j_1,*,j_0,*}(\wt{A}) \|_F\\
     = & ~  \| f(A,x)_{j_1} \cdot  f(A,x)_{j_0}^2 \cdot x \cdot  c_g(A,x)^{\top} - f(\wt{A},x)_{j_1} \cdot  f(\wt{A},x)_{j_0}^2 \cdot x \cdot  c_g(\wt{A},x)^{\top}  \|_F \\
     \leq & ~ |f(A,x)_{j_1} - f(\wt{A},x)_{j_1}| \cdot  |f(A,x)_{j_0} |^2 \cdot \|x\|_2 \cdot \| c_g(A,x)^{\top}\|_2 \\
     & + ~ |f(\wt{A},x)_{j_1}| \cdot  |f(A,x)_{j_0} -f(\wt{A},x)_{j_0} |  \cdot |f(A,x)_{j_0} |\cdot \|x\|_2 \cdot \| c_g(A,x)^{\top}\|_2 \\
      & + ~ |f(\wt{A},x)_{j_1}| \cdot  |f(\wt{A},x)_{j_0} |  \cdot |f(A,x)_{j_0} -f(\wt{A},x)_{j_0} | \cdot \|x\|_2 \cdot \| c_g(A,x)^{\top}\|_2 \\
      & + ~ |f(\wt{A},x)_{j_1}| \cdot  |f(\wt{A},x)_{j_0} |^2 \cdot \|x\|_2 \cdot \| c_g(A,x)^{\top} - c_g(\wt{A},x)^{\top}\|_2 \\
    \leq & ~ 10 R^2 \cdot  \beta^{-2} \cdot n  \cdot \exp(3R^2)\|A - \wt{A}\|_F \\
     & + ~ 10 R^2 \cdot  \beta^{-2} \cdot n \cdot   \exp(3R^2)\|A - \wt{A}\|_F \\
     & + ~ 10 R^2 \cdot  \beta^{-2} \cdot n \cdot  \exp(3R^2)\|A - \wt{A}\|_F \\
     & + ~ 20 R^2 \beta^{-2} \cdot n \cdot  \exp(3R^{2}) \cdot \|A - \wt{A}\|_F\\
     \leq & ~ 50  \cdot  \beta^{-2} \cdot n  \cdot \exp(4R^2)\|A - \wt{A}\|_F \\
     \leq & ~  \beta^{-2} \cdot n  \cdot \exp(5R^2)\|A - \wt{A}\|_F \\
\end{align*}

{\bf Proof of Part 10}
\begin{align*}
    & ~ \| B_{2}^{j_1,*,j_0,*}(A) -   B_{2}^{j_1,*,j_0,*}(\wt{A})  \|_F \\
    = & ~ \|\sum_{i = 1}^3 B_{2,i}^{j_1,*,j_0,*}(A) -   B_{2,i}^{j_1,*,j_0,*}(\wt{A})  \|_F \\
    \leq & ~ 9  \beta^{-2} \cdot n \cdot \sqrt{d} \cdot \exp(5R^2)\|A - \wt{A}\|_F
\end{align*}
\end{proof}

\subsection{PSD}
\begin{lemma}\label{psd: B_2}
If the following conditions hold
\begin{itemize}
    \item Let $B_{2,1,1}^{j_1,*, j_0,*}, \cdots, B_{2,3,1}^{j_1,*, j_0,*} $ be defined as Lemma~\ref{lem:b_2_j1_j0} 
    \item  Let $\|A \|_2 \leq R, \|A^{\top} \|_F \leq R, \| x\|_2 \leq R, \|\diag(f(A,x)) \|_F \leq \|f(A,x) \|_2 \leq 1, \| b_g \|_2 \leq 1$ 
\end{itemize}
We have 
\begin{itemize}
    \item {\bf Part 1.} $\| B_{2,1,1}^{j_1,*, j_0,*} \| \leq 4  $  
    \item {\bf Part 2.} $\|B_{2,1,2}^{j_1,*, j_0,*}\| \preceq 4  $
    \item {\bf Part 3.} $\|B_{2,1,3}^{j_1,*, j_0,*}\| \preceq 4 \sqrt{d} R^2$
    \item {\bf Part 4.} $\|B_{2,1,4}^{j_1,*, j_0,*} \|\preceq 4 \sqrt{d}R^2$
    \item {\bf Part 5.} $\|B_{2,1,5}^{j_1,*, j_0,*} \|\preceq 4 \sqrt{d} R^2$
    \item {\bf Part 6.} $\|B_{2,1,6}^{j_1,*, j_0,*} \|\preceq 6 \sqrt{d} R^2$
    \item {\bf Part 7.} $\|B_{2,1,7}^{j_1,*, j_0,*} \|\preceq 12 \sqrt{d} R^2$
    \item {\bf Part 8.} $\|B_{2,2,1}^{j_1,*, j_0,*}\| \preceq 10 R^2$
    \item {\bf Part 9.} $\|B_{2,3,1}^{j_1,*, j_0,*}\| \preceq 5 R^2$
    \item {\bf Part 10.} $\|B_{2}^{j_1,*, j_0,*}\| \preceq 53\sqrt{d} R^2$
\end{itemize}
\end{lemma}

\begin{proof}
    {\bf Proof of Part 1.}
    \begin{align*}
        \| B_{2,1,1}^{j_1,*, j_0,*} \| 
        = & ~ \| f(A,x)_{j_1} \cdot  f(A,x)_{j_0} \cdot c(A,x)_{j_0} \cdot f_c(A,x) \cdot  I_d \| \\
        \leq & ~ |f(A,x)_{j_1}| \cdot |f(A,x)_{j_0}| \cdot |c(A,x)_{j_0}| \cdot |f_c(A,x)| \cdot \| I_d\|\\
        \leq & ~ 4  
    \end{align*}

    {\bf Proof of Part 2.}
    \begin{align*}
        \| B_{2,1,2}^{j_1,*, j_0,*} \|
        = &~
        \| c(A,x)_{j_1} \cdot f(A,x)_{j_1} \cdot  f(A,x)_{j_0} \cdot c(A,x)_{j_0} \cdot I_d  \| \\
        \preceq & ~  |f(A,x)_{j_1}| \cdot |f(A,x)_{j_0}| \cdot |c(A,x)_{j_1}|\cdot  |c(A,x)_{j_0}| \cdot  \| I_d\| \\
        \preceq & ~ 4 
    \end{align*}

    {\bf Proof of Part 3.}
    \begin{align*}
     \| B_{2,1,3}^{j_1,*, j_0,*} \|
     = & ~
       \| f(A,x)_{j_1} \cdot  f(A,x)_{j_0} \cdot c(A,x)_{j_0} \cdot f_c(A,x) \cdot  {\bf 1}_d  \cdot ( (A_{j_1,*}) \circ x^\top  ) \| \\
    \leq & ~ |f(A,x)_{j_1}| \cdot |f(A,x)_{j_0}| \cdot |c(A,x)_{j_0}|\cdot |f_c(A,x)|   \cdot \| {\bf 1}_d \|_2 \cdot \| A_{j_1,*}^\top \circ x \|_2 \\
    \leq  & ~ 4\sqrt{d}  \| A_{j_1,*}\|_2 \|\diag(x) \|_{\infty}\\
    \leq &  ~  4R^2\sqrt{d}
    \end{align*}
   
    {\bf Proof of Part 4.}
    \begin{align*}
        \|B_{2,1,4}^{j_1,*, j_0,*}  \|
        = & ~  \| f(A,x)_{j_1} \cdot f(A,x)_{j_0} \cdot c(A,x)_{j_0} \cdot f_c(A,x) \cdot {\bf 1}_d  \cdot h(A,x)^\top\|\\
        \leq & ~ | f(A,x)_{j_1} | \cdot|  f(A,x)_{j_0} |\cdot |c(A,x)_{j_0}|\cdot |f_c(A,x)| \cdot \|{\bf 1}_d \|_2 \cdot \| h(A,x)^\top \|_2 \\
        \leq & ~ 4 \sqrt{d}R^2
    \end{align*}

    {\bf Proof of Part 5.}
    \begin{align*}
        \|B_{2,1,5}^{j_1,*, j_0,*} \|
        = & ~ \| f(A,x)_{j_1} \cdot f(A,x)_{j_0} \cdot c(A,x)_{j_0}  \cdot (-f_2(A,x) + f(A,x)_{j_1})   \cdot {\bf 1}_d  \cdot h(A,x)^\top\| \\
        \leq & ~  | f(A,x)_{j_1}| \cdot |f(A,x)_{j_0}| \cdot |c(A,x)_{j_0}|  \cdot (|-f_2(A,x)| + |f(A,x)_{j_1}|)   \cdot \| {\bf 1}_d \|_2  \cdot \|h(A,x)^\top\|_2 \\
        \leq & ~ 4 \sqrt{d} R^2
    \end{align*}

    {\bf Proof of Part 6.}
    \begin{align*}
        \|B_{2,1,6}^{j_1,*, j_0,*} \|
        = & ~ \| f(A,x)_{j_1} \cdot f(A,x)_{j_0} \cdot c(A,x)_{j_0} \cdot(-f_c(A,x) + f(A,x)_{j_1})     \cdot {\bf 1}_d  \cdot h(A,x)^\top \|\\
        \leq & ~  | f(A,x)_{j_1}| \cdot |f(A,x)_{j_0}| \cdot |c(A,x)_{j_0}|  \cdot (|-f_c(A,x)| + |f(A,x)_{j_1}|)   \cdot \| {\bf 1}_d \|_2  \cdot \|h(A,x)^\top\|_2 \\
        \leq & ~ 6 \sqrt{d} R^2
    \end{align*}

    {\bf Proof of Part 7.}
    \begin{align*}
         \|B_{2,1,7}^{j_1,*, j_0,*} \|
         = & ~ \|f(A,x)_{j_1} \cdot f(A,x)_{j_0} \cdot c(A,x)_{j_0} \cdot {\bf 1}_d  \cdot p_{j_1}(A,x)^\top  \|\\
         \leq & ~ | f(A,x)_{j_1}| \cdot |f(A,x)_{j_0}| \cdot |c(A,x)_{j_0}|\cdot \| {\bf 1}_d \|_2 \cdot \|p_{j_1}(A,x)^\top\|_2 \\
         \leq & ~ 12 \sqrt{d} R^2
    \end{align*}

    {\bf Proof of Part 8.}
    \begin{align*}
        \|B_{2,2,1}^{j_1,*, j_0,*} \|
        = & ~ \| f(A,x)_{j_1} \cdot  f(A,x)_{j_0} \cdot  c(A,x)_{j_0} \cdot x \cdot c_g(A,x)^{\top} \| \\
        \leq & ~  | f(A,x)_{j_1}| \cdot |f(A,x)_{j_0}| \cdot |c(A,x)_{j_0}| \cdot \|x\|_2 \cdot \|c_g(A,x)^{\top}\|_2\\
        \leq & ~ 10 R^2
    \end{align*}

    {\bf Proof of Part 9.}
    \begin{align*}
     \|B_{2,3,1}^{j_1,*, j_0,*} \|
        = & ~ \| f(A,x)_{j_1} \cdot  f(A,x)_{j_0}^2 \cdot x \cdot  c_g(A,x)^{\top}\| \\
        \leq & ~ | f(A,x)_{j_1}| \cdot  |f(A,x)_{j_0} | \cdot  |f(A,x)_{j_0} |  \cdot \|x\|_2 \cdot  \|c_g(A,x)^{\top}|\|_2 \\
        \leq & ~ 5 R^2
    \end{align*}

    {\bf Proof of Part 10}
\begin{align*}
    & ~ \| B_{2}^{j_1,*,j_0,*} \|  \\
    = & ~ \|\sum_{i = 1}^3 B_{2,i}^{j_1,*,j_0,*}  \|  \\
    \leq & ~ 53\sqrt{d} R^2
\end{align*}
\end{proof}

\newpage
\section{Hessian: Third term  \texorpdfstring{$B_3^{j_1,i_1,j_0,i_0}$}{}}\label{app:hessian_third}
\subsection{Definitions}
\begin{definition}\label{def:b_3}
 We define the $B_3^{j_1,i_1,j_0,i_0}$ as follows,
    \begin{align*}
        B_3^{j_1,i_1,j_0,i_0} :=  & ~ \frac{\d}{\d A_{j_1,i_1}} (- c_g(A,x)^{\top} \cdot f(A,x)_{j_0} \cdot \langle c(A,x), f(A,x) \rangle \cdot ( (A_{j_0,*})^\top \circ x ))
    \end{align*}
    Then, we define $B_{3,1}^{j_1,i_1,j_0,i_0}, \cdots, B_{3,4}^{j_1,i_1,j_0,i_0}$ as follow
    \begin{align*}
         B_{3,1}^{j_1,i_1,j_0,i_0} : = & ~ \frac{\d}{\d A_{j_1,i_1}} (- c_g(A,x)^{\top} ) \cdot  f(A,x)_{j_0} \cdot \langle c(A,x), f(A,x) \rangle \cdot ( (A_{j_0,*})^\top \circ x )\\
 B_{3,2}^{j_1,i_1,j_0,i_0} : = & ~ - c_g(A,x)^{\top} \cdot \frac{\d}{\d A_{j_1,i_1}} ( f(A,x)_{j_0} )  \cdot \langle c(A,x), f(A,x) \rangle \cdot ( (A_{j_0,*})^\top \circ x )\\
 B_{3,3}^{j_1,i_1,j_0,i_0} : = & ~  -  c_g(A,x)^{\top} \cdot f(A,x)_{j_0}  \cdot \langle\frac{\d  c(A,x)}{\d A_{j_1,i_1}}, f(A,x) \rangle \cdot ( (A_{j_0,*})^\top \circ x )\\
 B_{3,4}^{j_1,i_1,j_0,i_0} : = & ~  -  c_g(A,x)^{\top} \cdot f(A,x)_{j_0}  \cdot \langle c(A,x),\frac{\d  f(A,x) }{\d A_{j_1,i_1}}\rangle \cdot ( (A_{j_0,*})^\top \circ x )\\
 B_{3,5}^{j_1,i_1,j_0,i_0} : = & ~  -  c_g(A,x)^{\top} \cdot f(A,x)_{j_0}  \cdot \langle c(A,x), f(A,x) \rangle \cdot \frac{\d}{\d A_{j_1,i_1}} ( (A_{j_0,*})^\top \circ x)
    \end{align*}
    It is easy to show
    \begin{align*}
        B_3^{j_1,i_1,j_0,i_0} = B_{3,1}^{j_1,i_1,j_0,i_0} +  B_{3,2}^{j_1,i_1,j_0,i_0} + B_{3,3}^{j_1,i_1,j_0,i_0} + B_{3,4}^{j_1,i_1,j_0,i_0} +B_{3,5}^{j_1,i_1,j_0,i_0}
    \end{align*}
    Similarly for $j_1 = j_0$ and $i_0 = i_1$,we have
    \begin{align*}
        B_3^{j_1,i_1,j_1,i_1} = B_{3,1}^{j_1,i_1,j_1,i_1} +  B_{3,2}^{j_1,i_1,j_1,i_1} + B_{3,3}^{j_1,i_1,j_1,i_1}  + B_{3,4}^{j_1,i_1,j_1,i_1} +B_{3,5}^{j_1,i_1,j_1,i_1}
         \end{align*}
    For $j_1 = j_0$ and $i_0 \neq i_1$,we have
    \begin{align*}
        B_3^{j_1,i_1,j_1,i_0} = B_{3,1}^{j_1,i_1,j_1,i_0} +  B_{3,2}^{j_1,i_1,j_1,i_0} + B_{3,3}^{j_1,i_1,j_1,i_0}  + B_{3,4}^{j_1,i_1,j_1,i_0} +B_{3,5}^{j_1,i_1,j_1,i_0}
    \end{align*}
    For $j_1 \neq j_0$ and $i_0 = i_1$,we have
    \begin{align*}
        B_3^{j_1,i_1,j_0,i_1} = B_{3,1}^{j_1,i_1,j_0,i_1} +  B_{3,2}^{j_1,i_1,j_0,i_1} + B_{3,3}^{j_1,i_1,j_0,i_1}  + B_{3,4}^{j_1,i_1,j_0,i_1} +B_{3,5}^{j_1,i_1,j_0,i_1}
    \end{align*}
\end{definition}

\subsection{Case \texorpdfstring{$j_1=j_0, i_1 = i_0$}{}}
\begin{lemma}
For $j_1 = j_0$ and $i_0 = i_1$. If the following conditions hold
    \begin{itemize}
     \item Let $u(A,x) \in \R^n$ be defined as Definition~\ref{def:u}
    \item Let $\alpha(A,x) \in \R$ be defined as Definition~\ref{def:alpha}
     \item Let $f(A,x) \in \R^n$ be defined as Definition~\ref{def:f}
    \item Let $c(A,x) \in \R^n$ be defined as Definition~\ref{def:c}
    \item Let $g(A,x) \in \R^d$ be defined as Definition~\ref{def:g} 
    \item Let $q(A,x) = c(A,x) + f(A,x) \in \R^n$
    \item Let $c_g(A,x) \in \R^d$ be defined as Definition~\ref{def:c_g}.
    \item Let $L_g(A,x) \in \R$ be defined as Definition~\ref{def:l_g}
    \item Let $v \in \R^n$ be a vector 
    \item Let $B_1^{j_1,i_1,j_0,i_0}$ be defined as Definition~\ref{def:b_1}
    \end{itemize}
    Then, For $j_0,j_1 \in [n], i_0,i_1 \in [d]$, we have 
    \begin{itemize}
\item {\bf Part 1.} For $B_{3,1}^{j_1,i_1,j_1,i_1}$, we have 
\begin{align*}
 B_{3,1}^{j_1,i_1,j_1,i_1}  = & ~ \frac{\d}{\d A_{j_1,i_1}} (- c_g(A,x)^{\top} ) \cdot  f(A,x)_{j_1} \cdot \langle c(A,x), f(A,x) \rangle \cdot ( (A_{j_1,*})^\top \circ x ) \\
 = & ~ B_{3,1,1}^{j_1,i_1,j_1,i_1} + B_{3,1,2}^{j_1,i_1,j_1,i_1} + B_{3,1,3}^{j_1,i_1,j_1,i_1} + B_{3,1,4}^{j_1,i_1,j_1,i_1} + B_{3,1,5}^{j_1,i_1,j_1,i_1} + B_{3,1,6}^{j_1,i_1,j_1,i_1} + B_{3,1,7}^{j_1,i_1,j_1,i_1}
\end{align*} 
\item {\bf Part 2.} For $B_{3,2}^{j_1,i_1,j_1,i_1}$, we have 
\begin{align*}
  B_{3,2}^{j_1,i_1,j_1,i_1} = & ~  - c_g(A,x)^{\top} \cdot \frac{\d}{\d A_{j_1,i_1}} ( f(A,x)_{j_1} )  \cdot \langle c(A,x), f(A,x) \rangle \cdot ( (A_{j_1,*})^\top \circ x )  \\
    = & ~  B_{3,2,1}^{j_1,i_1,j_1,i_1} + B_{3,2,2}^{j_1,i_1,j_1,i_1}
\end{align*} 
\item {\bf Part 3.} For $B_{3,3}^{j_1,i_1,j_1,i_1}$, we have 
\begin{align*}
  B_{3,3}^{j_1,i_1,j_1,i_1} = & ~ -  c_g(A,x)^{\top} \cdot f(A,x)_{j_1}  \cdot \langle \d \frac{c(A,x)}{\d A_{j_1,i_1}} , f(A,x) \rangle \cdot ( (A_{j_1,*})^\top \circ x )\\
     = & ~ B_{3,3,1}^{j_1,i_1,j_1,i_1} + B_{3,3,2}^{j_1,i_1,j_1,i_1}
\end{align*} 
\item {\bf Part 4.} For $B_{3,4}^{j_1,i_1,j_1,i_1}$, we have 
\begin{align*}
  B_{3,4}^{j_1,i_1,j_1,i_1} = & ~ -  c_g(A,x)^{\top} \cdot f(A,x)_{j_1}  \cdot \d \langle c(A,x), \frac{f(A,x)}{\d A_{j_1,i_1}}  \rangle \cdot ( (A_{j_1,*})^\top \circ x )\\
     = & ~ B_{3,4,1}^{j_1,i_1,j_1,i_1} + B_{3,4,2}^{j_1,i_1,j_1,i_1}
\end{align*}
\item {\bf Part 5.} For $B_{3,5}^{j_1,i_1,j_1,i_1}$, we have 
\begin{align*}
  B_{3,5}^{j_1,i_1,j_1,i_1} = & ~ -  c_g(A,x)^{\top} \cdot f(A,x)_{j_1}  \cdot \langle c(A,x), f(A,x) \rangle \cdot \frac{\d}{\d A_{j_1,i_1}} ( (A_{j_1,*})^\top \circ x)\\
     = & ~ B_{3,5,1}^{j_1,i_1,j_1,i_1}  
\end{align*}
\end{itemize}
\begin{proof}
    {\bf Proof of Part 1.}
    \begin{align*}
    B_{3,1,1}^{j_1,i_1,j_1,i_1} : = & ~  e_{i_1}^\top \cdot (\langle c(A,x), f(A,x) \rangle)^2 \cdot  f(A,x)_{j_1}^2 \cdot  ( (A_{j_1,*})^\top \circ x )\\
    B_{3,1,2}^{j_1,i_1,j_1,i_1} : = & ~ e_{i_1}^\top \cdot c(A,x)_{j_1}  \cdot f(A,x)_{j_1}^2 \cdot \langle c(A,x), f(A,x) \rangle \cdot ( (A_{j_1,*})^\top \circ x )\\
    B_{3,1,3}^{j_1,i_1,j_1,i_1} : = & ~f(A,x)_{j_1}^2 \cdot (\langle c(A,x), f(A,x) \rangle)^2 \cdot ( (A_{j_1,*}) \circ x^\top  ) \cdot ( (A_{j_1,*})^\top \circ x  )\\
    B_{3,1,4}^{j_1,i_1,j_1,i_1} : = & ~ -  f(A,x)_{j_1}^2 \cdot f(A,x)^\top  \cdot A \cdot \diag(x) \cdot   (\langle c(A,x), f(A,x) \rangle )^2 \cdot ( (A_{j_1,*})^\top \circ x )\\
    B_{3,1,5}^{j_1,i_1,j_1,i_1} : = & ~    f(A,x)_{j_1}^2 \cdot f(A,x)^\top  \cdot A \cdot \diag(x) \cdot (\langle -f(A,x), f(A,x) \rangle + f(A,x)_{j_1}) \\
    & ~\cdot \langle c(A,x), f(A,x) \rangle \cdot ( (A_{j_1,*})^\top \circ x ) \\
    B_{3,1,6}^{j_1,i_1,j_1,i_1} : = & ~  f(A,x)_{j_1}^2 \cdot f(A,x)^\top  \cdot A \cdot  \diag(x) \cdot(\langle -f(A,x), c(A,x) \rangle + f(A,x)_{j_1}) \\
    & ~\cdot \langle c(A,x), f(A,x) \rangle \cdot ( (A_{j_1,*})^\top \circ x ) \\
    B_{3,1,7}^{j_1,i_1,j_1,i_1} : = & ~ f(A,x)_{j_1}^2 \cdot ((e_{j_1}^\top - f(A,x)^\top) \circ q(A,x)^\top) \cdot A \cdot  \diag(x)  \cdot \langle c(A,x), f(A,x) \rangle \cdot ( (A_{j_1,*})^\top \circ x )
\end{align*}
Finally, combine them and we have
\begin{align*}
       B_{3,1}^{j_1,i_1,j_1,i_1} = B_{3,1,1}^{j_1,i_1,j_1,i_1} + B_{3,1,2}^{j_1,i_1,j_1,i_1} + B_{3,1,3}^{j_1,i_1,j_1,i_1} + B_{3,1,4}^{j_1,i_1,j_1,i_1} + B_{3,1,5}^{j_1,i_1,j_1,i_1} + B_{3,1,6}^{j_1,i_1,j_1,i_1} + B_{3,1,7}^{j_1,i_1,j_1,i_1}
\end{align*}
{\bf Proof of Part 2.}
    \begin{align*}
    B_{3,2,1}^{j_1,i_1,j_1,i_1} : = & ~  c_g(A,x)^{\top} \cdot f(A,x)_{j_1}^2 \cdot x_{i_1} \cdot \langle c(A,x), f(A,x) \rangle \cdot ( (A_{j_1,*})^\top \circ x ) \\
    B_{3,2,2}^{j_1,i_1,j_1,i_1} : = & ~ - c_g(A,x)^{\top} \cdot f(A,x)_{j_1} \cdot x_{i_1} \cdot \langle c(A,x), f(A,x) \rangle \cdot ( (A_{j_1,*})^\top \circ x )
\end{align*}
Finally, combine them and we have
\begin{align*}
       B_{3,2}^{j_1,i_1,j_1,i_1} = B_{3,2,1}^{j_1,i_1,j_1,i_1} + B_{3,2,2}^{j_1,i_1,j_1,i_1}
\end{align*}
{\bf Proof of Part 3.} 
    \begin{align*}
    B_{3,3,1}^{j_1,i_1,j_1,i_1} : = & ~   -  c_g(A,x)^{\top} \cdot f(A,x)_{j_1}^2  \cdot x_{i_1} \cdot \langle - f(A,x), f(A,x) \rangle \cdot ( (A_{j_1,*})^\top \circ x )\\
    B_{3,3,2}^{j_1,i_1,j_1,i_1} : = & ~   -  c_g(A,x)^{\top} \cdot f(A,x)_{j_1}^3    \cdot x_{i_1}  \cdot ( (A_{j_1,*})^\top \circ x )\\
\end{align*}
Finally, combine them and we have
\begin{align*}
       B_{3,3}^{j_1,i_1,j_1,i_1} = B_{3,3,1}^{j_1,i_1,j_1,i_1} + B_{3,3,2}^{j_1,i_1,j_1,i_1}
\end{align*}
{\bf Proof of Part 4.} 
    \begin{align*}
    B_{3,4,1}^{j_1,i_1,j_1,i_1} : = & ~   -  c_g(A,x)^{\top} \cdot f(A,x)_{j_1}^2  \cdot x_{i_1} \cdot \langle - f(A,x), c(A,x) \rangle  \cdot ( (A_{j_1,*})^\top \circ x )\\
    B_{3,4,2}^{j_1,i_1,j_1,i_1} : = & ~    -  c_g(A,x)^{\top} \cdot f(A,x)_{j_1}^2  \cdot x_{i_1} \cdot c(A,x)_{j_1} \cdot ( (A_{j_1,*})^\top \circ x )
\end{align*}
Finally, combine them and we have
\begin{align*}
       B_{3,4}^{j_1,i_1,j_1,i_1} = B_{3,4,1}^{j_1,i_1,j_1,i_1} + B_{3,4,2}^{j_1,i_1,j_1,i_1}
\end{align*}
{\bf Proof of Part 5.} 
    \begin{align*}
    B_{3,5,1}^{j_1,i_1,j_1,i_1} : = & ~  -  c_g(A,x)^{\top} \cdot f(A,x)_{j_1}  \cdot \langle c(A,x), f(A,x) \rangle \cdot (e_{i_1} \circ x)
\end{align*}
Finally, combine them and we have
\begin{align*}
       B_{3,5}^{j_1,i_1,j_1,i_1} = B_{3,5,1}^{j_1,i_1,j_1,i_1}  
\end{align*}
\end{proof}
\end{lemma}
\subsection{Case \texorpdfstring{$j_1=j_0, i_1 \neq i_0$}{}}
\begin{lemma}
For $j_1 = j_0$ and $i_0 \neq i_1$. If the following conditions hold
    \begin{itemize}
     \item Let $u(A,x) \in \R^n$ be defined as Definition~\ref{def:u}
    \item Let $\alpha(A,x) \in \R$ be defined as Definition~\ref{def:alpha}
     \item Let $f(A,x) \in \R^n$ be defined as Definition~\ref{def:f}
    \item Let $c(A,x) \in \R^n$ be defined as Definition~\ref{def:c}
    \item Let $g(A,x) \in \R^d$ be defined as Definition~\ref{def:g} 
    \item Let $q(A,x) = c(A,x) + f(A,x) \in \R^n$
    \item Let $c_g(A,x) \in \R^d$ be defined as Definition~\ref{def:c_g}.
    \item Let $L_g(A,x) \in \R$ be defined as Definition~\ref{def:l_g}
    \item Let $v \in \R^n$ be a vector 
    \item Let $B_1^{j_1,i_1,j_0,i_0}$ be defined as Definition~\ref{def:b_1}
    \end{itemize}
    Then, For $j_0,j_1 \in [n], i_0,i_1 \in [d]$, we have 
    \begin{itemize}
\item {\bf Part 1.} For $B_{3,1}^{j_1,i_1,j_1,i_0}$, we have 
\begin{align*}
 B_{3,1}^{j_1,i_1,j_1,i_0}  = & ~ \frac{\d}{\d A_{j_1,i_1}} (- c_g(A,x)^{\top} ) \cdot  f(A,x)_{j_1} \cdot \langle c(A,x), f(A,x) \rangle \cdot ( (A_{j_1,*})^\top \circ x ) \\
 = & ~ B_{3,1,1}^{j_1,i_1,j_1,i_0} + B_{3,1,2}^{j_1,i_1,j_1,i_0} + B_{3,1,3}^{j_1,i_1,j_1,i_0} + B_{3,1,4}^{j_1,i_1,j_1,i_0} + B_{3,1,5}^{j_1,i_1,j_1,i_0} + B_{3,1,6}^{j_1,i_1,j_1,i_0} + B_{3,1,7}^{j_1,i_1,j_1,i_0}
\end{align*} 
\item {\bf Part 2.} For $B_{3,2}^{j_1,i_1,j_1,i_0}$, we have 
\begin{align*}
  B_{3,2}^{j_1,i_1,j_1,i_0} = & ~  - c_g(A,x)^{\top} \cdot \frac{\d}{\d A_{j_1,i_1}} ( f(A,x)_{j_1} )  \cdot \langle c(A,x), f(A,x) \rangle \cdot ( (A_{j_1,*})^\top \circ x )  \\
    = & ~  B_{3,2,1}^{j_1,i_1,j_1,i_0} + B_{3,2,2}^{j_1,i_1,j_1,i_0}
\end{align*} 
\item {\bf Part 3.} For $B_{3,3}^{j_1,i_1,j_1,i_0}$, we have 
\begin{align*}
  B_{3,3}^{j_1,i_1,j_1,i_0} = & ~ -  c_g(A,x)^{\top} \cdot f(A,x)_{j_1}  \cdot \langle \d \frac{c(A,x)}{\d A_{j_1,i_1}} , f(A,x) \rangle \cdot ( (A_{j_1,*})^\top \circ x )\\
     = & ~ B_{3,3,1}^{j_1,i_1,j_1,i_0} + B_{3,3,2}^{j_1,i_1,j_1,i_0}
\end{align*} 
\item {\bf Part 4.} For $B_{3,4}^{j_1,i_1,j_1,i_0}$, we have 
\begin{align*}
  B_{3,4}^{j_1,i_1,j_1,i_0} = & ~ -  c_g(A,x)^{\top} \cdot f(A,x)_{j_1}  \cdot \d \langle c(A,x), \frac{f(A,x)}{\d A_{j_1,i_1}}  \rangle \cdot ( (A_{j_1,*})^\top \circ x )\\
     = & ~ B_{3,4,1}^{j_1,i_1,j_1,i_0} + B_{3,4,2}^{j_1,i_1,j_1,i_0}
\end{align*}
\item {\bf Part 5.} For $B_{3,5}^{j_1,i_1,j_1,i_0}$, we have 
\begin{align*}
  B_{3,5}^{j_1,i_1,j_1,i_0} = & ~ -  c_g(A,x)^{\top} \cdot f(A,x)_{j_1}  \cdot \langle c(A,x), f(A,x) \rangle \cdot \frac{\d}{\d A_{j_1,i_1}} ( (A_{j_1,*})^\top \circ x)\\
     = & ~ B_{3,5,1}^{j_1,i_1,j_1,i_0}  
\end{align*}
\end{itemize}
\begin{proof}
    {\bf Proof of Part 1.}
    \begin{align*}
    B_{3,1,1}^{j_1,i_1,j_1,i_0} : = & ~  e_{i_1}^\top \cdot (\langle c(A,x), f(A,x) \rangle)^2 \cdot  f(A,x)_{j_1}^2 \cdot  ( (A_{j_1,*})^\top \circ x )\\
    B_{3,1,2}^{j_1,i_1,j_1,i_0} : = & ~ e_{i_1}^\top \cdot c(A,x)_{j_1}  \cdot f(A,x)_{j_1}^2 \cdot \langle c(A,x), f(A,x) \rangle \cdot ( (A_{j_1,*})^\top \circ x )\\
    B_{3,1,3}^{j_1,i_1,j_1,i_0} : = & ~f(A,x)_{j_1}^2 \cdot (\langle c(A,x), f(A,x) \rangle)^2 \cdot ( (A_{j_1,*}) \circ x^\top  ) \cdot ( (A_{j_1,*})^\top \circ x  )\\
    B_{3,1,4}^{j_1,i_1,j_1,i_0} : = & ~ -  f(A,x)_{j_1}^2 \cdot f(A,x)^\top  \cdot A \cdot \diag(x) \cdot   (\langle c(A,x), f(A,x) \rangle )^2 \cdot ( (A_{j_1,*})^\top \circ x )\\
    B_{3,1,5}^{j_1,i_1,j_1,i_0} : = & ~    f(A,x)_{j_1}^2 \cdot f(A,x)^\top  \cdot A \cdot \diag(x) \cdot (\langle -f(A,x), f(A,x) \rangle + f(A,x)_{j_1})  \\
    & ~\cdot \langle c(A,x), f(A,x) \rangle \cdot ( (A_{j_1,*})^\top \circ x ) \\
    B_{3,1,6}^{j_1,i_1,j_1,i_0} : = & ~  f(A,x)_{j_1}^2 \cdot f(A,x)^\top  \cdot A \cdot  \diag(x) \cdot(\langle -f(A,x), c(A,x) \rangle + f(A,x)_{j_1}) \\
    & ~\cdot \langle c(A,x), f(A,x) \rangle \cdot ( (A_{j_1,*})^\top \circ x ) \\
    B_{3,1,7}^{j_1,i_1,j_1,i_0} : = & ~ f(A,x)_{j_1}^2 \cdot ((e_{j_1}^\top - f(A,x)^\top) \circ q(A,x)^\top) \cdot A \cdot  \diag(x)  \cdot \langle c(A,x), f(A,x) \rangle \cdot ( (A_{j_1,*})^\top \circ x )
\end{align*}
Finally, combine them and we have
\begin{align*}
       B_{3,1}^{j_1,i_1,j_1,i_0} = B_{3,1,1}^{j_1,i_1,j_1,i_0} + B_{3,1,2}^{j_1,i_1,j_1,i_0} + B_{3,1,3}^{j_1,i_1,j_1,i_0} + B_{3,1,4}^{j_1,i_1,j_1,i_0} + B_{3,1,5}^{j_1,i_1,j_1,i_0} + B_{3,1,6}^{j_1,i_1,j_1,i_0} + B_{3,1,7}^{j_1,i_1,j_1,i_0}
\end{align*}
{\bf Proof of Part 2.}
    \begin{align*}
    B_{3,2,1}^{j_1,i_1,j_1,i_0} : = & ~  c_g(A,x)^{\top} \cdot f(A,x)_{j_1}^2 \cdot x_{i_1} \cdot \langle c(A,x), f(A,x) \rangle \cdot ( (A_{j_1,*})^\top \circ x ) \\
    B_{3,2,2}^{j_1,i_1,j_1,i_0} : = & ~ - c_g(A,x)^{\top} \cdot f(A,x)_{j_1} \cdot x_{i_1} \cdot \langle c(A,x), f(A,x) \rangle \cdot ( (A_{j_1,*})^\top \circ x )
\end{align*}
Finally, combine them and we have
\begin{align*}
       B_{3,2}^{j_1,i_1,j_1,i_0} = B_{3,2,1}^{j_1,i_1,j_1,i_0} + B_{3,2,2}^{j_1,i_1,j_1,i_0}
\end{align*}
{\bf Proof of Part 3.} 
    \begin{align*}
    B_{3,3,1}^{j_1,i_1,j_1,i_0} : = & ~   -  c_g(A,x)^{\top} \cdot f(A,x)_{j_1}^2  \cdot x_{i_1} \cdot \langle - f(A,x), f(A,x) \rangle \cdot ( (A_{j_1,*})^\top \circ x )\\
    B_{3,3,2}^{j_1,i_1,j_1,i_0} : = & ~   -  c_g(A,x)^{\top} \cdot f(A,x)_{j_1}^3    \cdot x_{i_1}  \cdot ( (A_{j_1,*})^\top \circ x )
\end{align*}
Finally, combine them and we have
\begin{align*}
       B_{3,3}^{j_1,i_1,j_1,i_0} = B_{3,3,1}^{j_1,i_1,j_1,i_0} + B_{3,3,2}^{j_1,i_1,j_1,i_0}
\end{align*}
{\bf Proof of Part 4.} 
    \begin{align*}
    B_{3,4,1}^{j_1,i_1,j_1,i_0} : = & ~   -  c_g(A,x)^{\top} \cdot f(A,x)_{j_1}^2  \cdot x_{i_1} \cdot \langle - f(A,x), c(A,x) \rangle  \cdot ( (A_{j_1,*})^\top \circ x )\\
    B_{3,4,2}^{j_1,i_1,j_1,i_0} : = & ~    -  c_g(A,x)^{\top} \cdot f(A,x)_{j_1}^2  \cdot x_{i_1} \cdot c(A,x)_{j_1} \cdot ( (A_{j_1,*})^\top \circ x )
\end{align*}
Finally, combine them and we have
\begin{align*}
       B_{3,4}^{j_1,i_1,j_1,i_0} = B_{3,4,1}^{j_1,i_1,j_1,i_0} + B_{3,4,2}^{j_1,i_1,j_1,i_0}
\end{align*}
{\bf Proof of Part 5.} 
    \begin{align*}
    B_{3,5,1}^{j_1,i_1,j_1,i_0} : = & ~  -  c_g(A,x)^{\top} \cdot f(A,x)_{j_1}  \cdot \langle c(A,x), f(A,x) \rangle \cdot (e_{i_1} \circ x)
\end{align*}
Finally, combine them and we have
\begin{align*}
       B_{3,5}^{j_1,i_1,j_1,i_0} = B_{3,5,1}^{j_1,i_1,j_1,i_0}  
\end{align*}
\end{proof}
\end{lemma}
\subsection{Constructing \texorpdfstring{$d \times d$}{} matrices for \texorpdfstring{$j_1 = j_0$}{}}
The purpose of the following lemma is to let $i_0$ and $i_1$ disappear.
\begin{lemma}For $j_0,j_1 \in [n]$, a list of $d \times d$ matrices can be expressed as the following sense,
\begin{itemize}
\item {\bf Part 1.}
\begin{align*}
B_{3,1,1}^{j_1,*,j_1,*} & ~ =  f_c(A,x)^2 \cdot f(A,x)_{j_1}^2 \cdot  ( (A_{j_1,*})^\top \circ x ) \cdot  {\bf 1}_d^\top 
\end{align*}
\item {\bf Part 2.}
\begin{align*}
B_{3,1,2}^{j_1,*,j_1,*} & ~ =  f(A,x)_{j_1}^2 \cdot c(A,x)_{j_1} \cdot  f_c(A,x) \cdot ( (A_{j_1,*})^\top \circ x ) \cdot  {\bf 1}_d^\top 
\end{align*}
\item {\bf Part 3.}
\begin{align*}
B_{3,1,3}^{j_1,*,j_1,*} & ~ =   f(A,x)_{j_1}^2 \cdot f_c(A,x)^2 \cdot ( (A_{j_1,*}) \circ x^\top  ) \cdot ( (A_{j_1,*})^\top \circ x  ) \cdot I_d
\end{align*}
\item {\bf Part 4.}
\begin{align*}
B_{3,1,4}^{j_1,*,j_1,*}  & ~ =  -  f(A,x)_{j_1}^2 \cdot f_c(A,x)^2  \cdot h(A,x)^\top \cdot ( (A_{j_1,*})^\top \circ x ) \cdot I_d
\end{align*}
\item {\bf Part 5.}
\begin{align*}
B_{3,1,5}^{j_1,*,j_1,*}  & ~ = f(A,x)_{j_1}^2 \cdot (-f_2(A,x) + f(A,x)_{j_1})  \cdot f_c(A,x)  \cdot h(A,x)^\top \cdot ( (A_{j_1,*})^\top \circ x ) \cdot I_d
\end{align*}
\item {\bf Part 6.}
\begin{align*}
B_{3,1,6}^{j_1,*,j_1,*}  & ~ =  f(A,x)_{j_1}^2 \cdot (-f_c(A,x) + f(A,x)_{j_1}) \cdot f_c(A,x)   \cdot h(A,x)^\top \cdot ( (A_{j_1,*})^\top \circ x ) \cdot I_d
\end{align*}
\item {\bf Part 7.}
\begin{align*}
B_{3,1,7}^{j_1,*,j_1,*}  & ~ = f(A,x)_{j_1}^2 \cdot f_c(A,x)  \cdot p_{j_1}(A,x)^\top \cdot  ( (A_{j_1,*})^\top \circ x ) \cdot I_d
\end{align*}
\item {\bf Part 8.}
\begin{align*}
B_{3,2,1}^{j_1,*,j_1,*}  & ~ =   f(A,x)_{j_1}^2  \cdot f_c(A,x) \cdot c_g(A,x)^{\top} \cdot  ( (A_{j_1,*})^\top \circ x )  \cdot x \cdot {\bf 1}_d^{\top}
\end{align*}
\item {\bf Part 9.}
\begin{align*}
B_{3,2,2}^{j_1,*,j_1,*}  & ~ =   - f(A,x)_{j_1}  \cdot f_c(A,x) \cdot  c_g(A,x)^{\top} \cdot  ( (A_{j_1,*})^\top \circ x )  \cdot x \cdot {\bf 1}_d^{\top}
\end{align*}
\item {\bf Part 10.}
\begin{align*}
 B_{3,3,1}^{j_1,*,j_1,*}  & ~ =      f(A,x)_{j_1}^2  \cdot f_2(A,x) \cdot  c_g(A,x)^{\top} \cdot  ( (A_{j_1,*})^\top \circ x )  \cdot x \cdot {\bf 1}_d^{\top}
\end{align*}
\item {\bf Part 11.}
\begin{align*}
B_{3,3,2}^{j_1,*,j_1,*}  =   -   f(A,x)_{j_1}^3 \cdot c_g(A,x)^{\top} \cdot  ( (A_{j_1,*})^\top \circ x )  \cdot x \cdot {\bf 1}_d^{\top}
\end{align*}
\item {\bf Part 12.}
\begin{align*}
B_{3,4,1}^{j_1,*,j_1,*}  & ~ = f(A,x)_{j_1}^2 \cdot  f_c(A,x) \cdot  c_g(A,x)^{\top} \cdot  ( (A_{j_1,*})^\top \circ x )  \cdot x \cdot {\bf 1}_d^{\top}
\end{align*}
\item {\bf Part 13.}
\begin{align*}
 B_{3,4,2}^{j_1,*,j_1,*}  =   -   f(A,x)_{j_1}^2  \cdot c(A,x)_{j_1} 
 \cdot   c_g(A,x)^{\top} \cdot  ( (A_{j_1,*})^\top \circ x )  \cdot x \cdot {\bf 1}_d^{\top}
\end{align*}
\item {\bf Part 14.}
\begin{align*}
B_{3,5,1}^{j_1,*,j_1,*}  =  -   f(A,x)_{j_1}  \cdot f_c(A,x) \cdot   x \cdot  c_g(A,x)^{\top}
\end{align*}

\end{itemize}
\begin{proof}
{\bf Proof of Part 1.}
    We have
    \begin{align*}
        B_{3,1,1}^{j_1,i_1,j_1,i_1}  = & ~  e_{i_1}^\top \cdot (\langle c(A,x), f(A,x) \rangle)^2 \cdot  f(A,x)_{j_1}^2 \cdot  ( (A_{j_1,*})^\top \circ x )\\
        B_{3,1,1}^{j_1,i_1,j_1,i_0}  = & ~  e_{i_1}^\top \cdot (\langle c(A,x), f(A,x) \rangle)^2 \cdot  f(A,x)_{j_1}^2 \cdot  ( (A_{j_1,*})^\top \circ x )
    \end{align*}
    From the above two equations, we can tell that $B_{3,1,1}^{j_1,*,j_1,*} \in \R^{d \times d}$ is a matrix that both the diagonal and off-diagonal have entries.
    
    Then we have $B_{3,1,1}^{j_1,*,j_1,*} \in \R^{d \times d}$ can be written as the rescaling of a diagonal matrix,
    \begin{align*}
     B_{3,1,1}^{j_1,*,j_1,*} & ~ = (\langle c(A,x), f(A,x) \rangle)^2 \cdot  f(A,x)_{j_1}^2 \cdot  ( (A_{j_1,*})^\top \circ x ) \cdot  {\bf 1}_d^\top \\
     & ~ = f_c(A,x)^2 \cdot f(A,x)_{j_1}^2 \cdot  ( (A_{j_1,*})^\top \circ x ) \cdot  {\bf 1}_d^\top
\end{align*}
    where the last step is follows from the Definitions~\ref{def:f_c}. 

{\bf Proof of Part 2.}
    We have
    \begin{align*}
           B_{3,1,2}^{j_1,i_1,j_1,i_1} = & ~ e_{i_1}^\top \cdot c(A,x)_{j_1}  \cdot f(A,x)_{j_1}^2 \cdot \langle c(A,x), f(A,x) \rangle \cdot ( (A_{j_1,*})^\top \circ x )\\
        B_{3,1,2}^{j_1,i_1,j_1,i_0} = & ~ e_{i_1}^\top \cdot c(A,x)_{j_1}  \cdot f(A,x)_{j_1}^2 \cdot \langle c(A,x), f(A,x) \rangle \cdot ( (A_{j_1,*})^\top \circ x )
    \end{align*}
     From the above two equations, we can tell that $B_{3,1,2}^{j_1,*,j_1,*} \in \R^{d \times d}$ is a matrix that only diagonal has entries and off-diagonal are all zeros.
    
    Then we have $B_{3,1,2}^{j_1,*,j_1,*} \in \R^{d \times d}$ can be written as the rescaling of a diagonal matrix,
\begin{align*}
     B_{3,1,2}^{j_1,*,j_1,*} & ~ = f(A,x)_{j_1}^2 \cdot c(A,x)_{j_1} \cdot  \langle c(A,x), f(A,x) \rangle \cdot ( (A_{j_1,*})^\top \circ x ) \cdot  {\bf 1}_d^\top   \\
     & ~ = f(A,x)_{j_1}^2 \cdot c(A,x)_{j_1} \cdot  f_c(A,x) \cdot ( (A_{j_1,*})^\top \circ x ) \cdot  {\bf 1}_d^\top 
\end{align*}
    where the last step is follows from the Definitions~\ref{def:f_c}.

{\bf Proof of Part 3.}
We have for diagonal entry and off-diagonal entry can be written as follows 
    \begin{align*}
        B_{3,1,3}^{j_1,i_1,j_1,i_1} = & ~f(A,x)_{j_1}^2 \cdot (\langle c(A,x), f(A,x) \rangle)^2 \cdot ( (A_{j_1,*}) \circ x^\top  ) \cdot ( (A_{j_1,*})^\top \circ x  )\\
        B_{3,1,3}^{j_1,i_1,j_1,i_0} = & ~f(A,x)_{j_1}^2 \cdot (\langle c(A,x), f(A,x) \rangle)^2 \cdot ( (A_{j_1,*}) \circ x^\top  ) \cdot ( (A_{j_1,*})^\top \circ x  )
    \end{align*}
From the above equation, we can show that matrix $B_{3,1,3}^{j_1,*,j_1,*}$ can be expressed as a rank-$1$ matrix,
\begin{align*}
     B_{3,1,3}^{j_1,*,j_1,*} & ~ = f(A,x)_{j_1}^2 \cdot (\langle c(A,x), f(A,x) \rangle)^2  \cdot ( (A_{j_1,*}) \circ x^\top  ) \cdot ( (A_{j_1,*})^\top \circ x  ) \cdot I_d\\
     & ~ =  f(A,x)_{j_1}^2 \cdot f_c(A,x)^2 \cdot ( (A_{j_1,*}) \circ x^\top  ) \cdot ( (A_{j_1,*})^\top \circ x  ) \cdot I_d
\end{align*}
    where the last step is follows from the Definitions~\ref{def:f_c}.

{\bf Proof of Part 4.}
We have for diagonal entry and off-diagonal entry can be written as follows
    \begin{align*}
        B_{3,1,4}^{j_1,i_1,j_1,i_1}   = & ~ -  f(A,x)_{j_1}^2 \cdot f(A,x)^\top  \cdot A \cdot \diag(x) \cdot   (\langle c(A,x), f(A,x) \rangle )^2 \cdot ( (A_{j_1,*})^\top \circ x )\\
        B_{3,1,4}^{j_1,i_1,j_1,i_0}   = & ~ -  f(A,x)_{j_1}^2 \cdot f(A,x)^\top  \cdot A \cdot \diag(x) \cdot   (\langle c(A,x), f(A,x) \rangle )^2 \cdot ( (A_{j_1,*})^\top \circ x )
    \end{align*}
 From the above equation, we can show that matrix $B_{3,1,4}^{j_1,*,j_1,*}$ can be expressed as a rank-$1$ matrix,
\begin{align*}
    B_{3,1,4}^{j_1,*,j_1,*}  & ~ =  -  f(A,x)_{j_1}^2 \cdot (\langle c(A,x), f(A,x) \rangle )^2  \cdot f(A,x)^\top  \cdot A \cdot \diag(x) \cdot ( (A_{j_1,*})^\top \circ x )\cdot I_d\\
     & ~ =  -  f(A,x)_{j_1}^2 \cdot f_c(A,x)^2  \cdot h(A,x)^\top \cdot ( (A_{j_1,*})^\top \circ x ) \cdot I_d
\end{align*}
    where the last step is follows from the Definitions~\ref{def:h} and Definitions~\ref{def:f_c}.

{\bf Proof of Part 5.}
We have for diagonal entry and off-diagonal entry can be written as follows
    \begin{align*}
         B_{3,1,5}^{j_1,i_1,j_1,i_0} = & ~    f(A,x)_{j_1}^2 \cdot f(A,x)^\top  \cdot A \cdot \diag(x) \cdot (\langle -f(A,x), f(A,x) \rangle + f(A,x)_{j_1})  \cdot \langle c(A,x), f(A,x) \rangle\\
    & ~ \cdot ( (A_{j_1,*})^\top \circ x ) \\
         B_{3,1,5}^{j_1,i_1,j_1,i_0} = & ~    f(A,x)_{j_1}^2 \cdot f(A,x)^\top  \cdot A \cdot \diag(x) \cdot (\langle -f(A,x), f(A,x) \rangle + f(A,x)_{j_1})  \cdot \langle c(A,x), f(A,x) \rangle \\
    & ~\cdot ( (A_{j_1,*})^\top \circ x ) 
    \end{align*}
    From the above equation, we can show that matrix $B_{3,1,5}^{j_1,*,j_1,*}$ can be expressed as a rank-$1$ matrix,
\begin{align*}
    B_{3,1,5}^{j_1,*,j_1,*}  & ~ =  f(A,x)_{j_1}^2 \cdot (\langle -f(A,x), f(A,x) \rangle + f(A,x)_{j_1})  \cdot \langle c(A,x), f(A,x) \rangle  \cdot f(A,x)^\top \\
    & ~ \cdot A \cdot \diag(x) \cdot ( (A_{j_1,*})^\top \circ x ) \cdot I_d\\
     & ~ =  f(A,x)_{j_1}^2 \cdot (-f_2(A,x) + f(A,x)_{j_1})  \cdot f_c(A,x)  \cdot h(A,x)^\top \cdot ( (A_{j_1,*})^\top \circ x ) \cdot I_d
\end{align*}
    where the last step is follows from the Definitions~\ref{def:h}, Definitions~\ref{def:f_c} and Definitions~\ref{def:f_2}.

{\bf Proof of Part 6.}
We have for diagonal entry and off-diagonal entry can be written as follows
    \begin{align*}
        B_{3,1,6}^{j_1,i_1,j_1,i_1}  = & ~  f(A,x)_{j_1}^2 \cdot f(A,x)^\top  \cdot A \cdot  \diag(x) \cdot(\langle -f(A,x), c(A,x) \rangle + f(A,x)_{j_1}) \cdot \langle c(A,x), f(A,x) \rangle \\
    & ~\cdot ( (A_{j_1,*})^\top \circ x ) \\
        B_{3,1,6}^{j_1,i_1,j_1,i_0}  = & ~  f(A,x)_{j_1}^2 \cdot f(A,x)^\top  \cdot A \cdot  \diag(x) \cdot(\langle -f(A,x), c(A,x) \rangle + f(A,x)_{j_1}) \cdot \langle c(A,x), f(A,x) \rangle \\
    & ~\cdot ( (A_{j_1,*})^\top \circ x ) 
    \end{align*}
    From the above equation, we can show that matrix $B_{3,1,6}^{j_1,*,j_1,*}$ can be expressed as a rank-$1$ matrix,
\begin{align*}
    B_{3,1,6}^{j_1,*,j_1,*}  & ~ =   f(A,x)_{j_1}^2 \cdot (\langle -f(A,x), c(A,x) \rangle + f(A,x)_{j_1}) \cdot \langle c(A,x), f(A,x) \rangle  \cdot f(A,x)^\top \\
    & ~ \cdot A \cdot  \diag(x) \cdot ( (A_{j_1,*})^\top \circ x )  \cdot I_d\\
     & ~ = f(A,x)_{j_1}^2 \cdot (-f_c(A,x) + f(A,x)_{j_1}) \cdot f_c(A,x)   \cdot h(A,x)^\top \cdot ( (A_{j_1,*})^\top \circ x ) \cdot I_d
\end{align*}
    where the last step is follows from the Definitions~\ref{def:h} and Definitions~\ref{def:f_c} .
    
{\bf Proof of Part 7.}
We have for diagonal entry and off-diagonal entry can be written as follows
    \begin{align*}
         B_{3,1,7}^{j_1,i_1,j_1,i_1} = & ~ f(A,x)_{j_1}^2 \cdot ((e_{j_1}^\top - f(A,x)^\top) \circ q(A,x)^\top) \cdot A \cdot  \diag(x)  \cdot \langle c(A,x), f(A,x) \rangle \cdot ( (A_{j_1,*})^\top \circ x )\\
         B_{3,1,7}^{j_1,i_1,j_1,i_0} = & ~ f(A,x)_{j_1}^2 \cdot ((e_{j_1}^\top - f(A,x)^\top) \circ q(A,x)^\top) \cdot A \cdot  \diag(x)  \cdot \langle c(A,x), f(A,x) \rangle \cdot ( (A_{j_1,*})^\top \circ x )
    \end{align*}
    From the above equation, we can show that matrix $B_{3,1,7}^{j_1,*,j_1,*}$ can be expressed as a rank-$1$ matrix,
\begin{align*}
     B_{3,1,7}^{j_1,*,j_1,*}  & ~ =   f(A,x)_{j_1}^2 \cdot \langle c(A,x), f(A,x) \rangle  \cdot  ((e_{j_1}^\top - f(A,x)^\top) \circ q(A,x)^\top) \cdot A \cdot  \diag(x) \cdot  ( (A_{j_1,*})^\top \circ x ) \cdot I_d \\
     & ~ =f(A,x)_{j_1}^2 \cdot f_c(A,x)  \cdot p_{j_1}(A,x)^\top \cdot  ( (A_{j_1,*})^\top \circ x ) \cdot I_d
\end{align*}
    where the last step is follows from the Definitions~\ref{def:f_c} and Definitions~\ref{def:p}.
    
{\bf Proof of Part 8.}
We have for diagonal entry and off-diagonal entry can be written as follows
    \begin{align*}
         B_{3,2,1}^{j_1,i_1,j_1,i_0} = & ~  c_g(A,x)^{\top} \cdot f(A,x)_{j_1}^2 \cdot x_{i_1} \cdot \langle c(A,x), f(A,x) \rangle \cdot ( (A_{j_1,*})^\top \circ x ) \\
         B_{3,2,1}^{j_1,i_1,j_1,i_0} = & ~  c_g(A,x)^{\top} \cdot f(A,x)_{j_1}^2 \cdot x_{i_1} \cdot \langle c(A,x), f(A,x) \rangle \cdot ( (A_{j_1,*})^\top \circ x ) 
    \end{align*}
    From the above equation, we can show that matrix $B_{3,2,1}^{j_1,*,j_1,*}$ can be expressed as a rank-$1$ matrix,
\begin{align*}
     B_{3,2,1}^{j_1,*,j_1,*}  & ~ =   f(A,x)_{j_1}^2  \cdot \langle c(A,x), f(A,x) \rangle  \cdot c_g(A,x)^{\top} \cdot  ( (A_{j_1,*})^\top \circ x )  \cdot x \cdot {\bf 1}_d^{\top}\\
     & ~ =f(A,x)_{j_1}^2  \cdot f_c(A,x) \cdot c_g(A,x)^{\top} \cdot  ( (A_{j_1,*})^\top \circ x )  \cdot x \cdot {\bf 1}_d^{\top}
\end{align*}
    where the last step is follows from the Definitions~\ref{def:f_c}.

{\bf Proof of Part 9.}
We have for diagonal entry and off-diagonal entry can be written as follows
    \begin{align*}
         B_{3,2,2}^{j_1,i_1,j_1,i_1} = & ~ - c_g(A,x)^{\top} \cdot f(A,x)_{j_1} \cdot x_{i_1} \cdot \langle c(A,x), f(A,x) \rangle \cdot ( (A_{j_1,*})^\top \circ x )\\
         B_{3,2,2}^{j_1,i_1,j_1,i_0} = & ~ - c_g(A,x)^{\top} \cdot f(A,x)_{j_1} \cdot x_{i_1} \cdot \langle c(A,x), f(A,x) \rangle \cdot ( (A_{j_1,*})^\top \circ x )
    \end{align*}
        From the above equation, we can show that matrix $B_{3,2,2}^{j_1,*,j_1,*}$ can be expressed as a rank-$1$ matrix,
\begin{align*}
     B_{3,2,2}^{j_1,*,j_1,*}  & ~ =   - f(A,x)_{j_1}  \cdot \langle c(A,x), f(A,x) \rangle \cdot  c_g(A,x)^{\top} \cdot  ( (A_{j_1,*})^\top \circ x )  \cdot x \cdot {\bf 1}_d^{\top} \\
     & ~ =- f(A,x)_{j_1}  \cdot f_c(A,x) \cdot  c_g(A,x)^{\top} \cdot  ( (A_{j_1,*})^\top \circ x )  \cdot x \cdot {\bf 1}_d^{\top}
\end{align*}
    where the last step is follows from the Definitions~\ref{def:f_c}.

{\bf Proof of Part 10.}
We have for diagonal entry and off-diagonal entry can be written as follows
    \begin{align*}
         B_{3,3,1}^{j_1,i_1,j_1,i_1}  = & ~   -  c_g(A,x)^{\top} \cdot f(A,x)_{j_1}^2  \cdot x_{i_1} \cdot \langle - f(A,x), f(A,x) \rangle \cdot ( (A_{j_1,*})^\top \circ x )\\
         B_{3,3,1}^{j_1,i_1,j_1,i_0}  = & ~   -  c_g(A,x)^{\top} \cdot f(A,x)_{j_1}^2  \cdot x_{i_1} \cdot \langle - f(A,x), f(A,x) \rangle \cdot ( (A_{j_1,*})^\top \circ x )
    \end{align*}
            From the above equation, we can show that matrix $B_{3,3,1}^{j_1,*,j_1,*}$ can be expressed as a rank-$1$ matrix,
\begin{align*}
    B_{3,3,1}^{j_1,*,j_1,*}  & ~ =     f(A,x)_{j_1}^2  \cdot \langle  f(A,x), f(A,x) \rangle \cdot   c_g(A,x)^{\top} \cdot  ( (A_{j_1,*})^\top \circ x )  \cdot x \cdot {\bf 1}_d^{\top} \\
     & ~ = f(A,x)_{j_1}^2  \cdot f_2(A,x) \cdot  c_g(A,x)^{\top} \cdot  ( (A_{j_1,*})^\top \circ x )  \cdot x \cdot {\bf 1}_d^{\top}
\end{align*}
    where the last step is follows from the Definitions~\ref{def:f_2}.

{\bf Proof of Part 11.}
We have for diagonal entry and off-diagonal entry can be written as follows
    \begin{align*}
         B_{3,3,2}^{j_1,i_1,j_1,i_1}   = & ~   -  c_g(A,x)^{\top} \cdot f(A,x)_{j_1}^3    \cdot x_{i_1}  \cdot ( (A_{j_1,*})^\top \circ x )\\
         B_{3,3,2}^{j_1,i_1,j_1,i_0}   = & ~   -  c_g(A,x)^{\top} \cdot f(A,x)_{j_1}^3    \cdot x_{i_1}  \cdot ( (A_{j_1,*})^\top \circ x )
    \end{align*}
            From the above equation, we can show that matrix $B_{3,3,2}^{j_1,*,j_1,*}$ can be expressed as a rank-$1$ matrix,
\begin{align*}
    B_{3,3,2}^{j_1,*,j_1,*}  =   -   f(A,x)_{j_1}^3 \cdot c_g(A,x)^{\top} \cdot  ( (A_{j_1,*})^\top \circ x )  \cdot x \cdot {\bf 1}_d^{\top}
\end{align*}
{\bf Proof of Part 12.}
We have for diagonal entry and off-diagonal entry can be written as follows
    \begin{align*}
         B_{3,4,1}^{j_1,i_1,j_1,i_0}   = & ~   -  c_g(A,x)^{\top} \cdot f(A,x)_{j_1}^2  \cdot x_{i_1} \cdot \langle - f(A,x), c(A,x) \rangle  \cdot ( (A_{j_1,*})^\top \circ x )\\
         B_{3,4,1}^{j_1,i_1,j_1,i_0}   = & ~   -  c_g(A,x)^{\top} \cdot f(A,x)_{j_1}^2  \cdot x_{i_1} \cdot \langle - f(A,x), c(A,x) \rangle  \cdot ( (A_{j_1,*})^\top \circ x )
    \end{align*}
            From the above equation, we can show that matrix $B_{3,4,1}^{j_1,*,j_1,*}$ can be expressed as a rank-$1$ matrix,
\begin{align*}
    B_{3,4,1}^{j_1,*,j_1,*}  & ~ =   f(A,x)_{j_1}^2 \cdot  \langle f(A,x), c(A,x) \rangle  \cdot  c_g(A,x)^{\top} \cdot  ( (A_{j_1,*})^\top \circ x )  \cdot x \cdot {\bf 1}_d^{\top} \\
     & ~ = f(A,x)_{j_1}^2 \cdot  f_c(A,x) \cdot  c_g(A,x)^{\top} \cdot  ( (A_{j_1,*})^\top \circ x )  \cdot x \cdot {\bf 1}_d^{\top}
\end{align*}
    where the last step is follows from the Definitions~\ref{def:f_c}.

{\bf Proof of Part 13.}
We have for diagonal entry and off-diagonal entry can be written as follows
    \begin{align*}
           B_{3,4,2}^{j_1,i_1,j_1,i_1} = & ~    -  c_g(A,x)^{\top} \cdot f(A,x)_{j_1}^2  \cdot x_{i_1} \cdot c(A,x)_{j_1} \cdot ( (A_{j_1,*})^\top \circ x ) \\
           B_{3,4,2}^{j_1,i_1,j_1,i_0} = & ~    -  c_g(A,x)^{\top} \cdot f(A,x)_{j_1}^2  \cdot x_{i_1} \cdot c(A,x)_{j_1} \cdot ( (A_{j_1,*})^\top \circ x )
    \end{align*}
            From the above equation, we can show that matrix $B_{3,4,2}^{j_1,*,j_1,*}$ can be expressed as a rank-$1$ matrix,
\begin{align*}
   B_{3,4,2}^{j_1,*,j_1,*}  =   -   f(A,x)_{j_1}^2  \cdot c(A,x)_{j_1} 
 \cdot   c_g(A,x)^{\top} \cdot  ( (A_{j_1,*})^\top \circ x )  \cdot x \cdot {\bf 1}_d^{\top}
\end{align*}

{\bf Proof of Part 14.}
We have for diagonal entry and off-diagonal entry can be written as follows
    \begin{align*}
           B_{3,5,1}^{j_1,i_1,j_1,i_1}  = & ~  -  c_g(A,x)^{\top} \cdot f(A,x)_{j_1}  \cdot \langle c(A,x), f(A,x) \rangle \cdot (e_{i_1} \circ x)\\
           B_{3,5,1}^{j_1,i_1,j_1,i_0}  = & ~  -  c_g(A,x)^{\top} \cdot f(A,x)_{j_1}  \cdot \langle c(A,x), f(A,x) \rangle \cdot (e_{i_1} \circ x)
    \end{align*}
            From the above equation, we can show that matrix $B_{3,5,1}^{j_1,*,j_1,*}$ can be expressed as a rank-$1$ matrix,
\begin{align*}
   B_{3,5,1}^{j_1,*,j_1,*} & ~ =   -   f(A,x)_{j_1}  \cdot \langle c(A,x), f(A,x) \rangle \cdot   x \cdot  c_g(A,x)^{\top} \\
   & ~ =  -   f(A,x)_{j_1}  \cdot f_c(A,x) \cdot   x \cdot  c_g(A,x)^{\top}
\end{align*}
    where the last step is follows from the Definitions~\ref{def:f_c}.
\end{proof}
\end{lemma}

\subsection{Case \texorpdfstring{$j_1 \neq j_0, i_1 = i_0$}{}}
\begin{lemma}
For $j_1 \neq j_0$ and $i_0 = i_1$. If the following conditions hold
    \begin{itemize}
     \item Let $u(A,x) \in \R^n$ be defined as Definition~\ref{def:u}
    \item Let $\alpha(A,x) \in \R$ be defined as Definition~\ref{def:alpha}
     \item Let $f(A,x) \in \R^n$ be defined as Definition~\ref{def:f}
    \item Let $c(A,x) \in \R^n$ be defined as Definition~\ref{def:c}
    \item Let $g(A,x) \in \R^d$ be defined as Definition~\ref{def:g} 
    \item Let $q(A,x) = c(A,x) + f(A,x) \in \R^n$
    \item Let $c_g(A,x) \in \R^d$ be defined as Definition~\ref{def:c_g}.
    \item Let $L_g(A,x) \in \R$ be defined as Definition~\ref{def:l_g}
    \item Let $v \in \R^n$ be a vector 
    \item Let $B_1^{j_1,i_1,j_0,i_0}$ be defined as Definition~\ref{def:b_1}
    \end{itemize}
    Then, For $j_0,j_1 \in [n], i_0,i_1 \in [d]$, we have 
    \begin{itemize}
\item {\bf Part 1.} For $B_{3,1}^{j_1,i_1,j_0,i_1}$, we have 
\begin{align*}
 B_{3,1}^{j_1,i_1,j_0,i_1}  = & ~ \frac{\d}{\d A_{j_1,i_1}} (- c_g(A,x)^{\top} ) \cdot  f(A,x)_{j_0} \cdot \langle c(A,x), f(A,x) \rangle \cdot ( (A_{j_0,*})^\top \circ x ) \\
 = & ~ B_{3,1,1}^{j_1,i_1,j_0,i_1} + B_{3,1,2}^{j_1,i_1,j_0,i_1} + B_{3,1,3}^{j_1,i_1,j_0,i_1} + B_{3,1,4}^{j_1,i_1,j_0,i_1} + B_{3,1,5}^{j_1,i_1,j_0,i_1} + B_{3,1,6}^{j_1,i_1,j_0,i_1} + B_{3,1,7}^{j_1,i_1,j_0,i_1}
\end{align*} 
\item {\bf Part 2.} For $B_{3,2}^{j_1,i_1,j_0,i_1}$, we have 
\begin{align*}
  B_{3,2}^{j_1,i_1,j_0,i_1} = & ~  - c_g(A,x)^{\top} \cdot \frac{\d}{\d A_{j_1,i_1}} ( f(A,x)_{j_0} )  \cdot \langle c(A,x), f(A,x) \rangle \cdot ( (A_{j_0,*})^\top \circ x )  \\
    = & ~  B_{3,2,1}^{j_1,i_1,j_0,i_1}  
\end{align*} 
\item {\bf Part 3.} For $B_{3,3}^{j_1,i_1,j_0,i_1}$, we have 
\begin{align*}
  B_{3,3}^{j_1,i_1,j_0,i_1} = & ~ -  c_g(A,x)^{\top} \cdot f(A,x)_{j_0}  \cdot \langle \d \frac{c(A,x)}{\d A_{j_1,i_1}} , f(A,x) \rangle \cdot ( (A_{j_0,*})^\top \circ x )\\
     = & ~ B_{3,3,1}^{j_1,i_1,j_0,i_1} + B_{3,3,2}^{j_1,i_1,j_0,i_1}
\end{align*} 
\item {\bf Part 4.} For $B_{3,4}^{j_1,i_1,j_0,i_1}$, we have 
\begin{align*}
  B_{3,4}^{j_1,i_1,j_0,i_1} = & ~ -  c_g(A,x)^{\top} \cdot f(A,x)_{j_0}  \cdot \d \langle c(A,x), \frac{f(A,x)}{\d A_{j_1,i_1}}  \rangle \cdot ( (A_{j_0,*})^\top \circ x )\\
     = & ~ B_{3,4,1}^{j_1,i_1,j_0,i_1} + B_{3,4,2}^{j_1,i_1,j_0,i_1}
\end{align*}
\item {\bf Part 5.} For $B_{3,5}^{j_1,i_1,j_0,i_1}$, we have 
\begin{align*}
  B_{3,5}^{j_1,i_1,j_0,i_1} = & ~ -  c_g(A,x)^{\top} \cdot f(A,x)_{j_0}  \cdot \langle c(A,x), f(A,x) \rangle \cdot \frac{\d}{\d A_{j_1,i_1}} ( (A_{j_0,*})^\top \circ x)\\
     = & ~ B_{3,5,1}^{j_1,i_1,j_0,i_1}  
\end{align*}
\end{itemize}
\begin{proof}
    {\bf Proof of Part 1.}
    \begin{align*}
    B_{3,1,1}^{j_1,i_1,j_0,i_1} : = & ~   e_{i_1}^\top \cdot (\langle c(A,x), f(A,x) \rangle)^2 \cdot  f(A,x)_{j_1} \cdot  f(A,x)_{j_0} \cdot  ( (A_{j_0,*})^\top \circ x )\\
    B_{3,1,2}^{j_1,i_1,j_0,i_1} : = & ~  e_{i_1}^\top \cdot c(A,x)_{j_1}  \cdot f(A,x)_{j_1} \cdot  f(A,x)_{j_0} \cdot \langle c(A,x), f(A,x) \rangle \cdot ( (A_{j_0,*})^\top \circ x )\\
    B_{3,1,3}^{j_1,i_1,j_0,i_1} : = & ~f(A,x)_{j_1} \cdot  f(A,x)_{j_0} \cdot (\langle c(A,x), f(A,x) \rangle)^2 \cdot  ((A_{j_1,*}) \circ x^\top) \cdot  ((A_{j_0,*}) \circ x^\top) \\
    B_{3,1,4}^{j_1,i_1,j_0,i_1} : = & ~ -  f(A,x)_{j_1} \cdot  f(A,x)_{j_0} \cdot f(A,x)^\top  \cdot A \cdot \diag(x) \cdot   (\langle c(A,x), f(A,x) \rangle )^2 \cdot ( (A_{j_0,*})^\top \circ x )\\
    B_{3,1,5}^{j_1,i_1,j_0,i_1} : = & ~  f(A,x)_{j_1} \cdot  f(A,x)_{j_0} \cdot f(A,x)^\top  \cdot A \cdot \diag(x) \cdot (\langle -f(A,x), f(A,x) \rangle + f(A,x)_{j_1}) \\
    & ~  \cdot \langle c(A,x), f(A,x) \rangle \cdot ( (A_{j_0,*})^\top \circ x ) \\
    B_{3,1,6}^{j_1,i_1,j_0,i_1} : = & ~  f(A,x)_{j_1} \cdot  f(A,x)_{j_0} \cdot f(A,x)^\top  \cdot A \cdot  \diag(x) \cdot(\langle -f(A,x), c(A,x) \rangle + f(A,x)_{j_1})\\
    & ~  \cdot \langle c(A,x), f(A,x) \rangle   \cdot ( (A_{j_0,*})^\top \circ x ) \\
    B_{3,1,7}^{j_1,i_1,j_0,i_1} : = & ~ f(A,x)_{j_1} \cdot  f(A,x)_{j_0} \cdot ((e_{j_1}^\top - f(A,x)^\top) \circ q(A,x)^\top) \cdot A \cdot  \diag(x) \\
    & ~  \cdot \langle c(A,x), f(A,x) \rangle \cdot ( (A_{j_0,*})^\top \circ x )
\end{align*}
Finally, combine them and we have
\begin{align*}
       B_{3,1}^{j_1,i_1,j_0,i_1} = B_{3,1,1}^{j_1,i_1,j_0,i_1} + B_{3,1,2}^{j_1,i_1,j_0,i_1} + B_{3,1,3}^{j_1,i_1,j_0,i_1} + B_{3,1,4}^{j_1,i_1,j_0,i_1} + B_{3,1,5}^{j_1,i_1,j_0,i_1} + B_{3,1,6}^{j_1,i_1,j_0,i_1} + B_{3,1,7}^{j_1,i_1,j_0,i_1}
\end{align*}
{\bf Proof of Part 2.}
    \begin{align*}
    B_{3,2,1}^{j_1,i_1,j_0,i_1} : = & ~ c_g(A,x)^{\top} \cdot f(A,x)_{j_1} \cdot  f(A,x)_{j_0} \cdot x_{i_1} \cdot \langle c(A,x), f(A,x) \rangle \cdot ( (A_{j_0,*})^\top \circ x ) 
\end{align*}
Finally, combine them and we have
\begin{align*}
       B_{3,2}^{j_1,i_1,j_0,i_1} = B_{3,2,1}^{j_1,i_1,j_0,i_1}  
\end{align*}
{\bf Proof of Part 3.} 
    \begin{align*}
    B_{3,3,1}^{j_1,i_1,j_0,i_1} : = & ~  -  c_g(A,x)^{\top} \cdot f(A,x)_{j_1} \cdot f(A,x)_{j_0}  \cdot x_{i_1} \cdot \langle - f(A,x), f(A,x) \rangle \cdot ( (A_{j_0,*})^\top \circ x )\\
    B_{3,3,2}^{j_1,i_1,j_0,i_1} : = & ~   -  c_g(A,x)^{\top} \cdot f(A,x)_{j_1}^2 \cdot f(A,x)_{j_0}    \cdot x_{i_1}  \cdot ( (A_{j_0,*})^\top \circ x )
\end{align*}
Finally, combine them and we have
\begin{align*}
       B_{3,3}^{j_1,i_1,j_0,i_1} = B_{3,3,1}^{j_1,i_1,j_0,i_1} + B_{3,3,2}^{j_1,i_1,j_0,i_1}
\end{align*}
{\bf Proof of Part 4.} 
    \begin{align*}
    B_{3,4,1}^{j_1,i_1,j_0,i_1} : = & ~   -  c_g(A,x)^{\top} \cdot f(A,x)_{j_1} \cdot f(A,x)_{j_0}  \cdot x_{i_1} \cdot \langle - f(A,x), c(A,x) \rangle  \cdot ( (A_{j_0,*})^\top \circ x )\\
    B_{3,4,2}^{j_1,i_1,j_0,i_1} : = & ~    -  c_g(A,x)^{\top} \cdot f(A,x)_{j_1} \cdot f(A,x)_{j_0}  \cdot x_{i_1} \cdot c(A,x)_{j_1} \cdot ( (A_{j_0,*})^\top \circ x )
\end{align*}
Finally, combine them and we have
\begin{align*}
       B_{3,4}^{j_1,i_1,j_0,i_1} = B_{3,4,1}^{j_1,i_1,j_0,i_1} + B_{3,4,2}^{j_1,i_1,j_0,i_1}
\end{align*}
{\bf Proof of Part 5.} 
    \begin{align*}
    B_{3,5,1}^{j_1,i_1,j_0,i_1} : = & ~ 0
\end{align*}
Finally, combine them and we have
\begin{align*}
       B_{3,5}^{j_1,i_1,j_0,i_1} = B_{3,5,1}^{j_1,i_1,j_0,i_1}  
\end{align*}
\end{proof}
\end{lemma}

\subsection{Case \texorpdfstring{$j_1 \neq j_0, i_1 \neq i_0$}{}}
\begin{lemma}
For $j_1 \neq j_0$ and $i_0 \neq i_1$. If the following conditions hold
    \begin{itemize}
     \item Let $u(A,x) \in \R^n$ be defined as Definition~\ref{def:u}
    \item Let $\alpha(A,x) \in \R$ be defined as Definition~\ref{def:alpha}
     \item Let $f(A,x) \in \R^n$ be defined as Definition~\ref{def:f}
    \item Let $c(A,x) \in \R^n$ be defined as Definition~\ref{def:c}
    \item Let $g(A,x) \in \R^d$ be defined as Definition~\ref{def:g} 
    \item Let $q(A,x) = c(A,x) + f(A,x) \in \R^n$
    \item Let $c_g(A,x) \in \R^d$ be defined as Definition~\ref{def:c_g}.
    \item Let $L_g(A,x) \in \R$ be defined as Definition~\ref{def:l_g}
    \item Let $v \in \R^n$ be a vector 
    \item Let $B_1^{j_1,i_1,j_0,i_0}$ be defined as Definition~\ref{def:b_1}
    \end{itemize}
    Then, For $j_0,j_1 \in [n], i_0,i_1 \in [d]$, we have 
    \begin{itemize}
\item {\bf Part 1.} For $B_{3,1}^{j_1,i_1,j_0,i_0}$, we have 
\begin{align*}
 B_{3,1}^{j_1,i_1,j_0,i_0}  = & ~ \frac{\d}{\d A_{j_1,i_1}} (- c_g(A,x)^{\top} ) \cdot  f(A,x)_{j_0} \cdot \langle c(A,x), f(A,x) \rangle \cdot ( (A_{j_0,*})^\top \circ x ) \\
 = & ~ B_{3,1,1}^{j_1,i_1,j_0,i_0} + B_{3,1,2}^{j_1,i_1,j_0,i_0} + B_{3,1,3}^{j_1,i_1,j_0,i_0} + B_{3,1,4}^{j_1,i_1,j_0,i_0} + B_{3,1,5}^{j_1,i_1,j_0,i_0} + B_{3,1,6}^{j_1,i_1,j_0,i_0} + B_{3,1,7}^{j_1,i_1,j_0,i_0}
\end{align*} 
\item {\bf Part 2.} For $B_{3,2}^{j_1,i_1,j_0,i_0}$, we have 
\begin{align*}
  B_{3,2}^{j_1,i_1,j_0,i_0} = & ~  - c_g(A,x)^{\top} \cdot \frac{\d}{\d A_{j_1,i_1}} ( f(A,x)_{j_0} )  \cdot \langle c(A,x), f(A,x) \rangle \cdot ( (A_{j_0,*})^\top \circ x )  \\
    = & ~  B_{3,2,1}^{j_1,i_1,j_0,i_0}  
\end{align*} 
\item {\bf Part 3.} For $B_{3,3}^{j_1,i_1,j_0,i_0}$, we have 
\begin{align*}
  B_{3,3}^{j_1,i_1,j_0,i_0} = & ~ -  c_g(A,x)^{\top} \cdot f(A,x)_{j_0}  \cdot \langle \d \frac{c(A,x)}{\d A_{j_1,i_1}} , f(A,x) \rangle \cdot ( (A_{j_0,*})^\top \circ x )\\
     = & ~ B_{3,3,1}^{j_1,i_1,j_0,i_0} + B_{3,3,2}^{j_1,i_1,j_0,i_0}
\end{align*} 
\item {\bf Part 4.} For $B_{3,4}^{j_1,i_1,j_0,i_0}$, we have 
\begin{align*}
  B_{3,4}^{j_1,i_1,j_0,i_0} = & ~ -  c_g(A,x)^{\top} \cdot f(A,x)_{j_0}  \cdot \d \langle c(A,x), \frac{f(A,x)}{\d A_{j_1,i_1}}  \rangle \cdot ( (A_{j_0,*})^\top \circ x )\\
     = & ~ B_{3,4,1}^{j_1,i_1,j_0,i_0} + B_{3,4,2}^{j_1,i_1,j_0,i_0}
\end{align*}
\item {\bf Part 5.} For $B_{3,5}^{j_1,i_1,j_0,i_0}$, we have 
\begin{align*}
  B_{3,5}^{j_1,i_1,j_0,i_0} = & ~ -  c_g(A,x)^{\top} \cdot f(A,x)_{j_0}  \cdot \langle c(A,x), f(A,x) \rangle \cdot \frac{\d}{\d A_{j_1,i_1}} ( (A_{j_0,*})^\top \circ x)\\
     = & ~ B_{3,5,1}^{j_1,i_1,j_0,i_0}  
\end{align*}
\end{itemize}
\begin{proof}
    {\bf Proof of Part 1.}
    \begin{align*}
    B_{3,1,1}^{j_1,i_1,j_0,i_0} : = & ~   e_{i_1}^\top \cdot (\langle c(A,x), f(A,x) \rangle)^2 \cdot  f(A,x)_{j_1} \cdot  f(A,x)_{j_0} \cdot  ( (A_{j_0,*})^\top \circ x )\\
    B_{3,1,2}^{j_1,i_1,j_0,i_0} : = & ~  e_{i_1}^\top \cdot c(A,x)_{j_1}  \cdot f(A,x)_{j_1} \cdot  f(A,x)_{j_0} \cdot \langle c(A,x), f(A,x) \rangle \cdot ( (A_{j_0,*})^\top \circ x )\\
    B_{3,1,3}^{j_1,i_1,j_0,i_0} : = & ~f(A,x)_{j_1} \cdot  f(A,x)_{j_0} \cdot (\langle c(A,x), f(A,x) \rangle)^2 \cdot  ((A_{j_1,*}) \circ x^\top) \cdot  ((A_{j_0,*})^\top \circ x) \\
    B_{3,1,4}^{j_1,i_1,j_0,i_0} : = & ~ -  f(A,x)_{j_1} \cdot  f(A,x)_{j_0} \cdot f(A,x)^\top  \cdot A \cdot \diag(x) \cdot   (\langle c(A,x), f(A,x) \rangle )^2 \cdot ( (A_{j_0,*})^\top \circ x )\\
    B_{3,1,5}^{j_1,i_1,j_0,i_0} : = & ~  f(A,x)_{j_1} \cdot  f(A,x)_{j_0} \cdot f(A,x)^\top  \cdot A \cdot \diag(x) \cdot (\langle -f(A,x), f(A,x) \rangle + f(A,x)_{j_1})\\
    & ~  \cdot \langle c(A,x), f(A,x) \rangle  \cdot ( (A_{j_0,*})^\top \circ x ) \\
    B_{3,1,6}^{j_1,i_1,j_0,i_0} : = & ~  f(A,x)_{j_1} \cdot  f(A,x)_{j_0} \cdot f(A,x)^\top  \cdot A \cdot  \diag(x) \cdot(\langle -f(A,x), c(A,x) \rangle + f(A,x)_{j_1}) \\
    & ~ \cdot \langle c(A,x), f(A,x) \rangle   \cdot ( (A_{j_0,*})^\top \circ x ) \\
    B_{3,1,7}^{j_1,i_1,j_0,i_0} : = & ~ f(A,x)_{j_1} \cdot  f(A,x)_{j_0} \cdot ((e_{j_1}^\top - f(A,x)^\top) \circ q(A,x)^\top) \cdot A \cdot  \diag(x)\\
    & ~   \cdot \langle c(A,x), f(A,x) \rangle \cdot ( (A_{j_0,*})^\top \circ x )
\end{align*}
Finally, combine them and we have
\begin{align*}
       B_{3,1}^{j_1,i_1,j_0,i_0} = B_{3,1,1}^{j_1,i_1,j_0,i_0} + B_{3,1,2}^{j_1,i_1,j_0,i_0} + B_{3,1,3}^{j_1,i_1,j_0,i_0} + B_{3,1,4}^{j_1,i_1,j_0,i_0} + B_{3,1,5}^{j_1,i_1,j_0,i_0} + B_{3,1,6}^{j_1,i_1,j_0,i_0} + B_{3,1,7}^{j_1,i_1,j_0,i_0}
\end{align*}
{\bf Proof of Part 2.}
    \begin{align*}
    B_{3,2,1}^{j_1,i_1,j_0,i_0} : = & ~ c_g(A,x)^{\top} \cdot f(A,x)_{j_1} \cdot  f(A,x)_{j_0} \cdot x_{i_1} \cdot \langle c(A,x), f(A,x) \rangle \cdot ( (A_{j_0,*})^\top \circ x ) 
\end{align*}
Finally, combine them and we have
\begin{align*}
       B_{3,2}^{j_1,i_1,j_0,i_0} = B_{3,2,1}^{j_1,i_1,j_0,i_0}  
\end{align*}
{\bf Proof of Part 3.} 
    \begin{align*}
    B_{3,3,1}^{j_1,i_1,j_0,i_0} : = & ~  -  c_g(A,x)^{\top} \cdot f(A,x)_{j_1} \cdot f(A,x)_{j_0}  \cdot x_{i_1} \cdot \langle - f(A,x), f(A,x) \rangle \cdot ( (A_{j_0,*})^\top \circ x )\\
    B_{3,3,2}^{j_1,i_1,j_0,i_0} : = & ~   -  c_g(A,x)^{\top} \cdot f(A,x)_{j_1}^2 \cdot f(A,x)_{j_0}    \cdot x_{i_1}  \cdot ( (A_{j_0,*})^\top \circ x )
\end{align*}
Finally, combine them and we have
\begin{align*}
       B_{3,3}^{j_1,i_1,j_0,i_0} = B_{3,3,1}^{j_1,i_1,j_0,i_0} + B_{3,3,2}^{j_1,i_1,j_0,i_0}
\end{align*}
{\bf Proof of Part 4.} 
    \begin{align*}
    B_{3,4,1}^{j_1,i_1,j_0,i_0} : = & ~   -  c_g(A,x)^{\top} \cdot f(A,x)_{j_1} \cdot f(A,x)_{j_0}  \cdot x_{i_1} \cdot \langle - f(A,x), c(A,x) \rangle  \cdot ( (A_{j_0,*})^\top \circ x )\\
    B_{3,4,2}^{j_1,i_1,j_0,i_0} : = & ~    -  c_g(A,x)^{\top} \cdot f(A,x)_{j_1} \cdot f(A,x)_{j_0}  \cdot x_{i_1} \cdot c(A,x)_{j_1} \cdot ( (A_{j_0,*})^\top \circ x )
\end{align*}
Finally, combine them and we have
\begin{align*}
       B_{3,4}^{j_1,i_1,j_0,i_0} = B_{3,4,1}^{j_1,i_1,j_0,i_0} + B_{3,4,2}^{j_1,i_1,j_0,i_0}
\end{align*}
{\bf Proof of Part 5.} 
    \begin{align*}
    B_{3,5,1}^{j_1,i_1,j_0,i_0} : = & ~ 0
\end{align*}
Finally, combine them and we have
\begin{align*}
       B_{3,5}^{j_1,i_1,j_0,i_0} = B_{3,5,1}^{j_1,i_1,j_0,i_0}  
\end{align*}
\end{proof}
\end{lemma}
\subsection{Constructing \texorpdfstring{$d \times d$}{} matrices for \texorpdfstring{$j_1 \neq j_0$}{}}
The purpose of the following lemma is to let $i_0$ and $i_1$ disappear.
\begin{lemma}For $j_0,j_1 \in [n]$, a list of $d \times d$ matrices can be expressed as the following sense,\label{lem:b_3_j1_j0}
\begin{itemize}
\item {\bf Part 1.}
\begin{align*}
B_{3,1,1}^{j_1,*,j_0,*} & ~ = f_c(A,x)^2 \cdot f(A,x)_{j_1} \cdot f(A,x)_{j_0} \cdot  ( (A_{j_0,*})^\top \circ x ) \cdot  {\bf 1}_d^\top
\end{align*}
\item {\bf Part 2.}
\begin{align*}
B_{3,1,2}^{j_1,*,j_0,*} & ~ =  f(A,x)_{j_1} \cdot f(A,x)_{j_0} \cdot c(A,x)_{j_1} \cdot  f_c(A,x) \cdot ( (A_{j_0,*})^\top \circ x ) \cdot  {\bf 1}_d^\top 
\end{align*}
\item {\bf Part 3.}
\begin{align*}
B_{3,1,3}^{j_1,*,j_0,*} & ~ = f(A,x)_{j_1} \cdot  f(A,x)_{j_0} \cdot f_c(A,x)^2 \cdot  ((A_{j_1,*}) \circ x^\top) \cdot  ((A_{j_0,*})^\top \circ x )  \cdot I_d
\end{align*}
\item {\bf Part 4.}
\begin{align*}
B_{3,1,4}^{j_1,*,j_0,*}  & ~ = -  f(A,x)_{j_1} \cdot  f(A,x)_{j_0} \cdot f_c(A,x)^2  \cdot h(A,x)^\top \cdot ( (A_{j_0,*})^\top \circ x ) \cdot I_d 
\end{align*}
\item {\bf Part 5.}
\begin{align*}
B_{3,1,5}^{j_1,*,j_0,*}  & ~ = f(A,x)_{j_1} \cdot  f(A,x)_{j_0} \cdot (-f_2(A,x) + f(A,x)_{j_1})  \cdot f_c(A,x)  \cdot h(A,x)^\top \cdot ( (A_{j_0,*})^\top \circ x ) \cdot I_d
\end{align*}
\item {\bf Part 6.}
\begin{align*}
B_{3,1,6}^{j_1,*,j_0,*}  & ~ =   f(A,x)_{j_1} \cdot  f(A,x)_{j_0} \cdot (-f_c(A,x) + f(A,x)_{j_1}) \cdot f_c(A,x)   \cdot h(A,x)^\top \cdot ( (A_{j_0,*})^\top \circ x ) \cdot I_d
\end{align*}
\item {\bf Part 7.}
\begin{align*}
B_{3,1,7}^{j_1,*,j_0,*}  & ~ = f(A,x)_{j_1} \cdot  f(A,x)_{j_0} \cdot f_c(A,x)   \cdot p_{j_1}(A,x)^\top \cdot  ( (A_{j_0,*})^\top \circ x )\cdot I_d
\end{align*}
\item {\bf Part 8.}
\begin{align*}
B_{3,2,1}^{j_1,*,j_0,*}  & ~ =    f(A,x)_{j_1} \cdot  f(A,x)_{j_0} \cdot f_c(A,x) \cdot c_g(A,x)^{\top} \cdot  ( (A_{j_0,*})^\top \circ x ) \cdot x \cdot {\bf 1}_d^{\top}
\end{align*}
\item {\bf Part 9.}
\begin{align*}
 B_{3,3,1}^{j_1,*,j_0,*}  & ~ =   f(A,x)_{j_1} \cdot f(A,x)_{j_0}  \cdot f_2( A,x) \cdot  c_g(A,x)^{\top} \cdot  ( (A_{j_0,*})^\top \circ x ) \cdot x \cdot {\bf 1}_d^{\top}
\end{align*}
\item {\bf Part 10.}
\begin{align*}
B_{3,3,2}^{j_1,*,j_0,*}  =  -  f(A,x)_{j_1}^2 \cdot f(A,x)_{j_0} \cdot c_g(A,x)^{\top} \cdot  ( (A_{j_0,*})^\top \circ x ) \cdot x \cdot {\bf 1}_d^{\top} 
\end{align*}
\item {\bf Part 11.}
\begin{align*}
B_{3,4,1}^{j_1,*,j_0,*}  & ~ = f(A,x)_{j_1} \cdot f(A,x)_{j_0} \cdot  f_c(A,x) \cdot  c_g(A,x)^{\top} \cdot  ( (A_{j_0,*})^\top \circ x ) \cdot x \cdot {\bf 1}_d^{\top} 
\end{align*}
\item {\bf Part 12.}
\begin{align*}
 B_{3,4,2}^{j_1,*,j_0,*}  =  -   f(A,x)_{j_1} \cdot f(A,x)_{j_0} \cdot c(A,x)_{j_1}  \cdot   c_g(A,x)^{\top} \cdot  ( (A_{j_0,*})^\top \circ x ) \cdot x \cdot {\bf 1}_d^{\top} 
\end{align*}
\item {\bf Part 13.}
\begin{align*}
B_{3,5,1}^{j_1,*,j_0,*}  =  0
\end{align*}

\end{itemize}
\begin{proof}
{\bf Proof of Part 1.}
    We have
    \begin{align*}
        B_{3,1,1}^{j_1,i_1,j_0,i_1}  = & ~  e_{i_1}^\top \cdot (\langle c(A,x), f(A,x) \rangle)^2 \cdot  f(A,x)_{j_1} \cdot f(A,x)_{j_0} \cdot  ( (A_{j_0,*})^\top \circ x )\\
        B_{3,1,1}^{j_1,i_1,j_0,i_0}  = & ~  e_{i_1}^\top \cdot (\langle c(A,x), f(A,x) \rangle)^2 \cdot  f(A,x)_{j_1} \cdot f(A,x)_{j_0} \cdot  ( (A_{j_0,*})^\top \circ x )
    \end{align*}
    From the above two equations, we can tell that $B_{3,1,1}^{j_1,*,j_0,*} \in \R^{d \times d}$ is a matrix that both the diagonal and off-diagonal have entries.
    
    Then we have $B_{3,1,1}^{j_1,*,j_0,*} \in \R^{d \times d}$ can be written as the rescaling of a diagonal matrix,
    \begin{align*}
     B_{3,1,1}^{j_1,*,j_0,*} & ~ = (\langle c(A,x), f(A,x) \rangle)^2 \cdot  f(A,x)_{j_1} \cdot f(A,x)_{j_0} \cdot  ( (A_{j_0,*})^\top \circ x ) \cdot  {\bf 1}_d^\top \\
     & ~ = f_c(A,x)^2 \cdot f(A,x)_{j_1} \cdot f(A,x)_{j_0} \cdot  ( (A_{j_0,*})^\top \circ x ) \cdot  {\bf 1}_d^\top
\end{align*}
    where the last step is follows from the Definitions~\ref{def:f_c}. 

{\bf Proof of Part 2.}
    We have
    \begin{align*}
           B_{3,1,2}^{j_1,i_1,j_0,i_1} = & ~ e_{i_1}^\top \cdot c(A,x)_{j_1}  \cdot f(A,x)_{j_1} \cdot f(A,x)_{j_0} \cdot \langle c(A,x), f(A,x) \rangle \cdot ( (A_{j_0,*})^\top \circ x )\\
        B_{3,1,2}^{j_1,i_1,j_0,i_0} = & ~ e_{i_1}^\top \cdot c(A,x)_{j_1}  \cdot f(A,x)_{j_1} \cdot f(A,x)_{j_0} \cdot \langle c(A,x), f(A,x) \rangle \cdot ( (A_{j_0,*})^\top \circ x )
    \end{align*}
     From the above two equations, we can tell that $B_{3,1,2}^{j_1,*,j_0,*} \in \R^{d \times d}$ is a matrix that only diagonal has entries and off-diagonal are all zeros.
    
    Then we have $B_{3,1,2}^{j_1,*,j_0,*} \in \R^{d \times d}$ can be written as the rescaling of a diagonal matrix,
\begin{align*}
     B_{3,1,2}^{j_1,*,j_0,*} & ~ = f(A,x)_{j_1} \cdot f(A,x)_{j_0} \cdot c(A,x)_{j_1} \cdot  \langle c(A,x), f(A,x) \rangle \cdot ( (A_{j_0,*})^\top \circ x ) \cdot  {\bf 1}_d^\top   \\
     & ~ = f(A,x)_{j_1} \cdot f(A,x)_{j_0} \cdot c(A,x)_{j_1} \cdot  f_c(A,x) \cdot ( (A_{j_0,*})^\top \circ x ) \cdot  {\bf 1}_d^\top 
\end{align*}
    where the last step is follows from the Definitions~\ref{def:f_c}.

{\bf Proof of Part 3.}
We have for diagonal entry and off-diagonal entry can be written as follows 
    \begin{align*}
        B_{3,1,3}^{j_1,i_1,j_0,i_1} = & ~f(A,x)_{j_1} \cdot  f(A,x)_{j_0} \cdot (\langle c(A,x), f(A,x) \rangle)^2 \cdot  ((A_{j_1,*}) \circ x^\top) \cdot  ((A_{j_0,*})^\top \circ x ) \\
        B_{3,1,3}^{j_1,i_1,j_0,i_0} = & ~f(A,x)_{j_1} \cdot  f(A,x)_{j_0} \cdot (\langle c(A,x), f(A,x) \rangle)^2 \cdot  ((A_{j_1,*}) \circ x^\top) \cdot  ((A_{j_0,*})^\top \circ x )
    \end{align*}
From the above equation, we can show that matrix $B_{3,1,3}^{j_1,*,j_0,*}$ can be expressed as a rank-$1$ matrix,
\begin{align*}
     B_{3,1,3}^{j_1,*,j_0,*} & ~ = f(A,x)_{j_1} \cdot  f(A,x)_{j_0} \cdot (\langle c(A,x), f(A,x) \rangle)^2  \cdot ((A_{j_1,*}) \circ x^\top) \cdot ((A_{j_0,*})^\top \circ x )  \cdot I_d\\
     & ~ =  f(A,x)_{j_1} \cdot  f(A,x)_{j_0} \cdot f_c(A,x)^2 \cdot  ((A_{j_1,*}) \circ x^\top) \cdot  ((A_{j_0,*})^\top \circ x ) \cdot I_d
\end{align*}
    where the last step is follows from the Definitions~\ref{def:f_c}.

{\bf Proof of Part 4.}
We have for diagonal entry and off-diagonal entry can be written as follows
    \begin{align*}
        B_{3,1,4}^{j_1,i_1,j_0,i_1}   = & ~ -  f(A,x)_{j_1} \cdot  f(A,x)_{j_0} \cdot f(A,x)^\top  \cdot A \cdot \diag(x) \cdot   (\langle c(A,x), f(A,x) \rangle )^2 \cdot ( (A_{j_0,*})^\top \circ x )\\
        B_{3,1,4}^{j_1,i_1,j_0,i_0}   = & ~ -  f(A,x)_{j_1} \cdot  f(A,x)_{j_0} \cdot f(A,x)^\top  \cdot A \cdot \diag(x) \cdot   (\langle c(A,x), f(A,x) \rangle )^2 \cdot ( (A_{j_0,*})^\top \circ x )
    \end{align*}
 From the above equation, we can show that matrix $B_{3,1,4}^{j_1,*,j_0,*}$ can be expressed as a rank-$1$ matrix,
\begin{align*}
    B_{3,1,4}^{j_1,*,j_0,*}  & ~ =  -  f(A,x)_{j_1} \cdot  f(A,x)_{j_0} \cdot (\langle c(A,x), f(A,x) \rangle )^2  \cdot f(A,x)^\top  \cdot A \cdot \diag(x) \cdot ( (A_{j_0,*})^\top \circ x ) \cdot I_d\\
     & ~ =  -  f(A,x)_{j_1} \cdot  f(A,x)_{j_0} \cdot f_c(A,x)^2  \cdot h(A,x)^\top \cdot ( (A_{j_0,*})^\top \circ x ) \cdot I_d 
\end{align*}
    where the last step is follows from the Definitions~\ref{def:h} and Definitions~\ref{def:f_c}.

{\bf Proof of Part 5.}
We have for diagonal entry and off-diagonal entry can be written as follows
    \begin{align*}
         B_{3,1,5}^{j_1,i_1,j_0,i_0} = & ~    f(A,x)_{j_1} \cdot  f(A,x)_{j_0} \cdot f(A,x)^\top  \cdot A \cdot \diag(x) \cdot (\langle -f(A,x), f(A,x) \rangle + f(A,x)_{j_1})  \\
          & ~ \cdot \langle c(A,x), f(A,x) \rangle \cdot ( (A_{j_0,*})^\top \circ x )   \\
         B_{3,1,5}^{j_1,i_1,j_0,i_0} = & ~    f(A,x)_{j_1} \cdot  f(A,x)_{j_0} \cdot f(A,x)^\top  \cdot A \cdot \diag(x) \cdot (\langle -f(A,x), f(A,x) \rangle + f(A,x)_{j_1})  \\
          & ~ \cdot \langle c(A,x), f(A,x) \rangle  \cdot ( (A_{j_0,*})^\top \circ x ) 
    \end{align*}
    From the above equation, we can show that matrix $B_{3,1,5}^{j_1,*,j_0,*}$ can be expressed as a rank-$1$ matrix,
\begin{align*}
    B_{3,1,5}^{j_1,*,j_0,*}  & ~ =   f(A,x)_{j_1} \cdot  f(A,x)_{j_0} \cdot (\langle -f(A,x), f(A,x) \rangle + f(A,x)_{j_1})  \cdot \langle c(A,x), f(A,x) \rangle  \cdot f(A,x)^\top \\
    & ~ \cdot A \cdot \diag(x) \cdot ( (A_{j_0,*})^\top \circ x ) \cdot I_d  \\
     & ~ =   f(A,x)_{j_1} \cdot  f(A,x)_{j_0} \cdot (-f_2(A,x) + f(A,x)_{j_1})  \cdot f_c(A,x)  \cdot h(A,x)^\top \cdot ( (A_{j_0,*})^\top \circ x ) \cdot I_d
\end{align*}
    where the last step is follows from the Definitions~\ref{def:h}, Definitions~\ref{def:f_c} and Definitions~\ref{def:f_2}.

{\bf Proof of Part 6.}
We have for diagonal entry and off-diagonal entry can be written as follows
    \begin{align*}
        B_{3,1,6}^{j_1,i_1,j_0,i_1}  = & ~  f(A,x)_{j_1} \cdot  f(A,x)_{j_0} \cdot f(A,x)^\top  \cdot A \cdot  \diag(x) \cdot(\langle -f(A,x), c(A,x) \rangle + f(A,x)_{j_1}) \\
          & ~\cdot \langle c(A,x), f(A,x) \rangle \cdot ( (A_{j_0,*})^\top \circ x ) \\
        B_{3,1,6}^{j_1,i_1,j_0,i_0}  = & ~ f(A,x)_{j_1} \cdot  f(A,x)_{j_0} \cdot f(A,x)^\top  \cdot A \cdot  \diag(x) \cdot(\langle -f(A,x), c(A,x) \rangle + f(A,x)_{j_1}) \\
          & ~\cdot \langle c(A,x), f(A,x) \rangle \cdot ( (A_{j_0,*})^\top \circ x ) 
    \end{align*}
    From the above equation, we can show that matrix $B_{3,1,6}^{j_1,*,j_0,*}$ can be expressed as a rank-$1$ matrix,
\begin{align*}
    B_{3,1,6}^{j_1,*,j_0,*}  & ~ =   f(A,x)_{j_1} \cdot  f(A,x)_{j_0} \cdot (\langle -f(A,x), c(A,x) \rangle + f(A,x)_{j_1}) \cdot \langle c(A,x), f(A,x) \rangle    \cdot f(A,x)^\top  \\
          & ~\cdot A \cdot  \diag(x) \cdot ( (A_{j_0,*})^\top \circ x )\\
     & ~ = f(A,x)_{j_1} \cdot  f(A,x)_{j_0} \cdot (-f_c(A,x) + f(A,x)_{j_1}) \cdot f_c(A,x)   \cdot h(A,x)^\top \cdot ( (A_{j_0,*})^\top \circ x ) \cdot I_d
\end{align*}
    where the last step is follows from the Definitions~\ref{def:h} and Definitions~\ref{def:f_c} .
    
{\bf Proof of Part 7.}
We have for diagonal entry and off-diagonal entry can be written as follows
    \begin{align*}
         B_{3,1,7}^{j_1,i_1,j_0,i_1} = & ~ f(A,x)_{j_1} \cdot  f(A,x)_{j_0} \cdot ((e_{j_1}^\top - f(A,x)^\top) \circ q(A,x)^\top) \cdot A \cdot  \diag(x)  \cdot \langle c(A,x), f(A,x) \rangle\\
          & ~ \cdot ( (A_{j_0,*})^\top \circ x )\\
         B_{3,1,7}^{j_1,i_1,j_0,i_0} = & ~ f(A,x)_{j_1} \cdot  f(A,x)_{j_0} \cdot ((e_{j_1}^\top - f(A,x)^\top) \circ q(A,x)^\top) \cdot A \cdot  \diag(x)  \cdot \langle c(A,x), f(A,x) \rangle\\
          & ~ \cdot ( (A_{j_0,*})^\top \circ x )
    \end{align*}
    From the above equation, we can show that matrix $B_{3,1,7}^{j_1,*,j_0,*}$ can be expressed as a rank-$1$ matrix,
\begin{align*}
     B_{3,1,7}^{j_1,*,j_0,*}  & ~ =   f(A,x)_{j_1} \cdot  f(A,x)_{j_0} \cdot \langle c(A,x), f(A,x) \rangle  \cdot  ((e_{j_1}^\top - f(A,x)^\top) \circ q(A,x)^\top) \\
     & ~ \cdot A \cdot  \diag(x) \cdot  ( (A_{j_0,*})^\top \circ x ) \cdot I_d \\
     & ~ =f(A,x)_{j_1} \cdot  f(A,x)_{j_0} \cdot f_c(A,x)   \cdot p_{j_1}(A,x)^\top \cdot  ( (A_{j_0,*})^\top \circ x )\cdot I_d
\end{align*}
    where the last step is follows from the Definitions~\ref{def:f_c} and Definitions~\ref{def:p}.
    
{\bf Proof of Part 8.}
We have for diagonal entry and off-diagonal entry can be written as follows
    \begin{align*}
         B_{3,2,1}^{j_1,i_1,j_0,i_0} = & ~  c_g(A,x)^{\top} \cdot f(A,x)_{j_1} \cdot  f(A,x)_{j_0} \cdot x_{i_1} \cdot \langle c(A,x), f(A,x) \rangle \cdot ( (A_{j_0,*})^\top \circ x )  \\
         B_{3,2,1}^{j_1,i_1,j_0,i_0} = & ~  c_g(A,x)^{\top} \cdot f(A,x)_{j_1} \cdot  f(A,x)_{j_0} \cdot x_{i_1} \cdot \langle c(A,x), f(A,x) \rangle \cdot ( (A_{j_0,*})^\top \circ x ) 
    \end{align*}
    From the above equation, we can show that matrix $B_{3,2,1}^{j_1,*,j_0,*}$ can be expressed as a rank-$1$ matrix,
\begin{align*}
     B_{3,2,1}^{j_1,*,j_0,*}  & ~ =   f(A,x)_{j_1} \cdot  f(A,x)_{j_0}  \cdot \langle c(A,x), f(A,x) \rangle \cdot c_g(A,x)^{\top} \cdot  ( (A_{j_0,*})^\top \circ x ) \cdot x \cdot {\bf 1}_d^{\top} \\
     & ~ = f(A,x)_{j_1} \cdot  f(A,x)_{j_0} \cdot f_c(A,x) \cdot c_g(A,x)^{\top} \cdot  ( (A_{j_0,*})^\top \circ x ) \cdot x \cdot {\bf 1}_d^{\top}
\end{align*}
    where the last step is follows from the Definitions~\ref{def:f_c}.

{\bf Proof of Part 9.}
We have for diagonal entry and off-diagonal entry can be written as follows
    \begin{align*}
         B_{3,3,1}^{j_1,i_1,j_0,i_1} = & ~ -  c_g(A,x)^{\top} \cdot f(A,x)_{j_1} \cdot f(A,x)_{j_0}  \cdot x_{i_1} \cdot \langle - f(A,x), f(A,x) \rangle \cdot ( (A_{j_0,*})^\top \circ x )\\
         B_{3,3,1}^{j_1,i_1,j_0,i_0} = & ~ -  c_g(A,x)^{\top} \cdot f(A,x)_{j_1} \cdot f(A,x)_{j_0}  \cdot x_{i_1} \cdot \langle - f(A,x), f(A,x) \rangle \cdot ( (A_{j_0,*})^\top \circ x )
    \end{align*}
        From the above equation, we can show that matrix $B_{3,3,1}^{j_1,*,j_0,*}$ can be expressed as a rank-$1$ matrix,
\begin{align*}
     B_{3,3,1}^{j_1,*,j_0,*}  & ~ =    f(A,x)_{j_1} \cdot f(A,x)_{j_0}  \cdot \langle  f(A,x), f(A,x) \rangle \cdot  c_g(A,x)^{\top} \cdot  ( (A_{j_0,*})^\top \circ x ) \cdot x \cdot {\bf 1}_d^{\top}   \\
     & ~ = f(A,x)_{j_1} \cdot f(A,x)_{j_0}  \cdot f_2( A,x) \cdot  c_g(A,x)^{\top} \cdot  ( (A_{j_0,*})^\top \circ x ) \cdot x \cdot {\bf 1}_d^{\top} 
\end{align*}
    where the last step is follows from the Definitions~\ref{def:f_2}.

{\bf Proof of Part 10.}
We have for diagonal entry and off-diagonal entry can be written as follows
    \begin{align*}
         B_{3,3,2}^{j_1,i_1,j_0,i_1}   = & ~  -  c_g(A,x)^{\top} \cdot f(A,x)_{j_1}^2 \cdot f(A,x)_{j_0}    \cdot x_{i_1}  \cdot ( (A_{j_0,*})^\top \circ x )\\
         B_{3,3,2}^{j_1,i_1,j_0,i_0}   = & ~  -  c_g(A,x)^{\top} \cdot f(A,x)_{j_1}^2 \cdot f(A,x)_{j_0}    \cdot x_{i_1}  \cdot ( (A_{j_0,*})^\top \circ x )
    \end{align*}
            From the above equation, we can show that matrix $B_{3,3,2}^{j_1,*,j_0,*}$ can be expressed as a rank-$1$ matrix,
\begin{align*}
    B_{3,3,2}^{j_1,*,j_0,*}  =   -  f(A,x)_{j_1}^2 \cdot f(A,x)_{j_0} \cdot c_g(A,x)^{\top} \cdot  ( (A_{j_0,*})^\top \circ x ) \cdot x \cdot {\bf 1}_d^{\top} 
\end{align*}
{\bf Proof of Part 11.}
We have for diagonal entry and off-diagonal entry can be written as follows
    \begin{align*}
         B_{3,4,1}^{j_1,i_1,j_0,i_0}   = & ~  -  c_g(A,x)^{\top} \cdot f(A,x)_{j_1} \cdot f(A,x)_{j_0}  \cdot x_{i_1} \cdot \langle - f(A,x), c(A,x) \rangle  \cdot ( (A_{j_0,*})^\top \circ x )\\
         B_{3,4,1}^{j_1,i_1,j_0,i_0}   = & ~  -  c_g(A,x)^{\top} \cdot f(A,x)_{j_1} \cdot f(A,x)_{j_0}  \cdot x_{i_1} \cdot \langle - f(A,x), c(A,x) \rangle  \cdot ( (A_{j_0,*})^\top \circ x )
    \end{align*}
            From the above equation, we can show that matrix $B_{3,4,1}^{j_1,*,j_0,*}$ can be expressed as a rank-$1$ matrix,
\begin{align*}
    B_{3,4,1}^{j_1,*,j_0,*}  & ~ =  f(A,x)_{j_1} \cdot f(A,x)_{j_0} \cdot  \langle f(A,x), c(A,x) \rangle  \cdot  c_g(A,x)^{\top} \cdot  ( (A_{j_0,*})^\top \circ x ) \cdot x \cdot {\bf 1}_d^{\top}  \\
     & ~ = f(A,x)_{j_1} \cdot f(A,x)_{j_0} \cdot  f_c(A,x) \cdot  c_g(A,x)^{\top} \cdot  ( (A_{j_0,*})^\top \circ x ) \cdot x \cdot {\bf 1}_d^{\top} 
\end{align*}
    where the last step is follows from the Definitions~\ref{def:f_c}.

{\bf Proof of Part 12.}
We have for diagonal entry and off-diagonal entry can be written as follows
    \begin{align*}
           B_{3,4,2}^{j_1,i_1,j_0,i_1} = & ~    -  c_g(A,x)^{\top} \cdot f(A,x)_{j_1} \cdot f(A,x)_{j_0}  \cdot x_{i_1} \cdot c(A,x)_{j_1} \cdot ( (A_{j_0,*})^\top \circ x ) \\
           B_{3,4,2}^{j_1,i_1,j_0,i_0} = & ~    -  c_g(A,x)^{\top} \cdot f(A,x)_{j_1} \cdot f(A,x)_{j_0}  \cdot x_{i_1} \cdot c(A,x)_{j_1} \cdot ( (A_{j_0,*})^\top \circ x )
    \end{align*}
            From the above equation, we can show that matrix $B_{3,4,2}^{j_1,*,j_0,*}$ can be expressed as a rank-$1$ matrix,
\begin{align*}
   B_{3,4,2}^{j_1,*,j_0,*}  =   -   f(A,x)_{j_1} \cdot f(A,x)_{j_0} \cdot c(A,x)_{j_1}  \cdot   c_g(A,x)^{\top} \cdot  ( (A_{j_0,*})^\top \circ x ) \cdot x \cdot {\bf 1}_d^{\top} 
\end{align*}

{\bf Proof of Part 13.}
We have for diagonal entry and off-diagonal entry can be written as follows
    \begin{align*}
           B_{3,5,1}^{j_1,i_1,j_0,i_1}  = & ~ 0\\
           B_{3,5,1}^{j_1,i_1,j_0,i_0}  = & ~ 0
    \end{align*}
            From the above equation, we can show that matrix $B_{3,5,1}^{j_1,*,j_0,*}$ can be expressed as a rank-$1$ matrix,
\begin{align*}
   B_{3,5,1}^{j_1,*,j_1,*}  =  0
\end{align*}
\end{proof}
\end{lemma}
\subsection{Expanding \texorpdfstring{$B_3$}{} into many terms}

\begin{lemma}
   If the following conditions hold
    \begin{itemize}
     \item Let $u(A,x) \in \R^n$ be defined as Definition~\ref{def:u}
    \item Let $\alpha(A,x) \in \R$ be defined as Definition~\ref{def:alpha}
     \item Let $f(A,x) \in \R^n$ be defined as Definition~\ref{def:f}
    \item Let $c(A,x) \in \R^n$ be defined as Definition~\ref{def:c}
    \item Let $g(A,x) \in \R^d$ be defined as Definition~\ref{def:g} 
    \item Let $q(A,x) = c(A,x) + f(A,x) \in \R^n$
    \item Let $c_g(A,x) \in \R^d$ be defined as Definition~\ref{def:c_g}.
    \item Let $L_g(A,x) \in \R$ be defined as Definition~\ref{def:l_g}
    \item Let $v \in \R^n$ be a vector 
    \end{itemize}
Then, For $j_0,j_1 \in [n], i_0,i_1 \in [d]$, we have 
\begin{itemize}
    \item {\bf Part 1.}For $j_1 = j_0$ and $i_0 = i_1$
\begin{align*}
    B_{3}^{j_1,i_1,j_1,i_1}
    = & ~ B_{3,1}^{j_1,i_1,j_1,i_1}  + B_{3,2}^{j_1,i_1,j_1,i_1}  +B_{3,3}^{j_1,i_1,j_1,i_1}  +B_{3,4}^{j_1,i_1,j_1,i_1} +B_{3,5}^{j_1,i_1,j_1,i_1} 
\end{align*}
\item {\bf Part 2.}For $j_1 = j_0$ and $i_0 \neq i_1$
\begin{align*}
    B_{3}^{j_1,i_1,j_1,i_0}
    = & ~ B_{3,1}^{j_1,i_1,j_1,i_0}  + B_{3,2}^{j_1,i_1,j_1,i_0}  +B_{3,3}^{j_1,i_1,j_1,i_0}  +B_{3,4}^{j_1,i_1,j_1,i_0} +B_{3,5}^{j_1,i_1,j_1,i_0} 
\end{align*}
\item {\bf Part 3.}For $j_1 \neq j_0$ and $i_0 = i_1$ \begin{align*}
   B_{3}^{j_1,i_1,j_0,i_1}
    = & ~ B_{3,1}^{j_1,i_1,j_0,i_1}  + B_{3,2}^{j_1,i_1,j_0,i_1}  +B_{3,3}^{j_1,i_1,j_0,i_1}  +B_{3,4}^{j_1,i_1,j_0,i_1} +B_{3,5}^{j_1,i_1,j_0,i_1} 
\end{align*}
\item  {\bf Part 4.} For $j_0 \neq j_1$ and $i_0 \neq i_1$
\begin{align*}
     B_{3}^{j_1,i_1,j_0,i_0}
    = & ~ B_{3,1}^{j_1,i_1,j_0,i_0}  + B_{3,2}^{j_1,i_1,j_0,i_0}  +B_{3,3}^{j_1,i_1,j_0,i_0}  +B_{3,4}^{j_1,i_1,j_0,i_0} +  B_{3,5}^{j_1,i_1,j_0,i_0}
\end{align*}
\end{itemize}
\end{lemma}
\begin{proof}
{\bf Proof of Part 1.} 
we have
\begin{align*}
    B_{3}^{j_1,i_1,j_1,i_1} = & ~\frac{\d}{\d A_{j_1,i_1}} (- c_g(A,x)^{\top} \cdot f(A,x)_{j_1} \cdot \langle c(A,x), f(A,x) \rangle \cdot ( (A_{j_1,*})^\top \circ x ))\\
    = & ~ B_{3,1}^{j_1,i_1,j_1,i_0}  + B_{3,2}^{j_1,i_1,j_1,i_0}  +B_{3,3}^{j_1,i_1,j_1,i_0}  +B_{3,4}^{j_1,i_1,j_1,i_0} +B_{3,5}^{j_1,i_1,j_1,i_1}
\end{align*}
{\bf Proof of Part 2.}  we have
    \begin{align*}
    B_{3}^{j_1,i_1,j_1,i_0} = & ~\frac{\d}{\d A_{j_1,i_1}} (- c_g(A,x)^{\top} \cdot f(A,x)_{j_1} \cdot \langle c(A,x), f(A,x) \rangle \cdot ( (A_{j_1,*})^\top \circ x ))\\
    = & ~ B_{3,1}^{j_1,i_1,j_1,i_0}  + B_{3,2}^{j_1,i_1,j_1,i_0}  +B_{3,3}^{j_1,i_1,j_1,i_0}  +B_{3,4}^{j_1,i_1,j_1,i_0}  +B_{3,5}^{j_1,i_1,j_1,i_0}
\end{align*}
{\bf Proof of Part 3.}
 we have
    \begin{align*}
    B_{3}^{j_1,i_1,j_0,i_1} = & ~\frac{\d}{\d A_{j_1,i_1}} (- c_g(A,x)^{\top} \cdot f(A,x)_{j_0} \cdot \langle c(A,x), f(A,x) \rangle \cdot ( (A_{j_0,*})^\top \circ x ))\\
    = & ~B_{3,1}^{j_1,i_1,j_0,i_1}  + B_{3,2}^{j_1,i_1,j_0,i_1}  +B_{3,3}^{j_1,i_1,j_0,i_1}  +B_{3,4}^{j_1,i_1,j_0,i_1} +B_{3,5}^{j_1,i_1,j_0,i_1} 
\end{align*}
{\bf Proof of Part 4.}
 we have
\begin{align*}
    B_{3}^{j_1,i_1,j_0,i_0}= & ~\frac{\d}{\d A_{j_1,i_1}} (- c_g(A,x)^{\top} \cdot f(A,x)_{j_0} \cdot \langle c(A,x), f(A,x) \rangle \cdot ( (A_{j_0,*})^\top \circ x ))\\
    = & ~B_{3,1}^{j_1,i_1,j_0,i_0}  + B_{3,2}^{j_1,i_1,j_0,i_0}  +B_{3,3}^{j_1,i_1,j_0,i_0}  +B_{3,4}^{j_1,i_1,j_0,i_0}  +  B_{3,5}^{j_1,i_1,j_0,i_0}
\end{align*}
\end{proof}

\subsection{Lipschitz Computation}
\begin{lemma}\label{lips: B_3}
If the following conditions hold
\begin{itemize}
    \item Let $B_{3,1,1}^{j_1,*, j_0,*}, \cdots, B_{3,5,1}^{j_1,*, j_0,*} $ be defined as Lemma~\ref{lem:b_3_j1_j0} 
    \item  Let $\|A \|_2 \leq R, \|A^{\top} \|_F \leq R, \| x\|_2 \leq R, \|\diag(f(A,x)) \|_F \leq \|f(A,x) \|_2 \leq 1, \| b_g \|_2 \leq 1$ 
\end{itemize}
Then, we have
\begin{itemize}
    \item {\bf Part 1.}
    \begin{align*}
       \| B_{3,1,1}^{j_1,*,j_0,*} (A) - B_{3,1,1}^{j_1,*,j_0,*} ( \wt{A} ) \|_F \leq  \beta^{-2} \cdot n \cdot \sqrt{d}\exp(5R^2) \cdot \|A - \wt{A}\|_F
    \end{align*}
     \item {\bf Part 2.}
    \begin{align*}
       \| B_{3,1,2}^{j_1,*,j_0,*} (A) - B_{3,1,2}^{j_1,*,j_0,*} ( \wt{A} ) \|_F \leq   \beta^{-2} \cdot n \cdot \sqrt{d}\exp(5R^2) \cdot \|A - \wt{A}\|_F
    \end{align*}
     \item {\bf Part 3.}
    \begin{align*}
       \| B_{3,1,3}^{j_1,*,j_0,*} (A) - B_{3,1,3}^{j_1,*,j_0,*} ( \wt{A} ) \|_F \leq   \beta^{-2} \cdot n \cdot \sqrt{d}\exp(6R^2) \cdot \|A - \wt{A}\|_F
    \end{align*}
     \item {\bf Part 4.}
    \begin{align*}
       \| B_{3,1,4}^{j_1,*,j_0,*} (A) - B_{3,1,4}^{j_1,*,j_0,*} ( \wt{A} ) \|_F \leq   \beta^{-2} \cdot n \cdot \sqrt{d}\exp(6R^2) \cdot \|A - \wt{A}\|_F
    \end{align*}
     \item {\bf Part 5.}
    \begin{align*}
       \| B_{3,1,5}^{j_1,*,j_0,*} (A) - B_{3,1,5}^{j_1,*,j_0,*} ( \wt{A} ) \|_F \leq   \beta^{-2} \cdot n \cdot \sqrt{d}\exp(6R^2) \cdot \|A - \wt{A}\|_F
    \end{align*}
     \item {\bf Part 6.}
    \begin{align*}
       \| B_{3,1,6}^{j_1,*,j_0,*} (A) - B_{3,1,6}^{j_1,*,j_0,*} ( \wt{A} ) \|_F \leq  \beta^{-2} \cdot n \cdot \sqrt{d}\exp(6R^2) \cdot \|A - \wt{A}\|_F
    \end{align*}
     \item {\bf Part 7.}
    \begin{align*}
       \| B_{3,1,7}^{j_1,*,j_0,*} (A) - B_{3,1,7}^{j_1,*,j_0,*} ( \wt{A} ) \|_F \leq   \beta^{-2} \cdot n \cdot \sqrt{d}\exp(6R^2) \cdot \|A - \wt{A}\|_F
    \end{align*}
    \item {\bf Part 8.}
    \begin{align*}
       \| B_{3,2,1}^{j_1,*,j_0,*} (A) - B_{3,2,1}^{j_1,*,j_0,*} ( \wt{A} ) \|_F \leq  \beta^{-2} \cdot n \cdot \sqrt{d}\exp(6R^2) \cdot \|A - \wt{A}\|_F
    \end{align*}
     \item {\bf Part 9.}
    \begin{align*}
       \| B_{3,3,1}^{j_1,*,j_0,*} (A) - B_{3,3,1}^{j_1,*,j_0,*} ( \wt{A} ) \|_F \leq  \beta^{-2} \cdot n \cdot \sqrt{d}  \exp(6R^2)\|A - \wt{A}\|_F  
    \end{align*}
     \item {\bf Part 10.}
    \begin{align*}
       \| B_{3,3,2}^{j_1,*,j_0,*} (A) - B_{3,3,2}^{j_1,*,j_0,*} ( \wt{A} ) \|_F \leq  \beta^{-2} \cdot n \cdot \sqrt{d}  \exp(6R^2)\|A - \wt{A}\|_F  
    \end{align*}
     \item {\bf Part 11.}
    \begin{align*}
       \| B_{3,4,1}^{j_1,*,j_0,*} (A) - B_{3,4,1}^{j_1,*,j_0,*} ( \wt{A} ) \|_F \leq \beta^{-2} \cdot n \cdot \sqrt{d}  \exp(6R^2)\|A - \wt{A}\|_F   
    \end{align*}
     \item {\bf Part 12.}
    \begin{align*}
       \| B_{3,4,2}^{j_1,*,j_0,*} (A) - B_{3,4,2}^{j_1,*,j_0,*} ( \wt{A} ) \|_F \leq  \beta^{-2} \cdot n \cdot \sqrt{d}  \exp(6R^2)\|A - \wt{A}\|_F  
    \end{align*}
     \item {\bf Part 13.}
    \begin{align*}
       \| B_{3,5,1}^{j_1,*,j_0,*} (A) - B_{3,5,1}^{j_1,*,j_0,*} ( \wt{A} ) \|_F = 0 
    \end{align*}
        \item{\bf Part 14.}
    \begin{align*}
         \| B_{3}^{j_1,*,j_0,*} (A) - B_{3}^{j_1,*,j_0,*} ( \wt{A} ) \|_F \leq & ~ 12  \beta^{-2} \cdot n \cdot \sqrt{d} \cdot \exp(6R^2)\|A - \wt{A}\|_F
    \end{align*}
    \end{itemize}
\end{lemma}
\begin{proof}
{\bf Proof of Part 1.}
\begin{align*}
& ~ \| B_{3,1,1}^{j_1,*,j_0,*} (A) - B_{3,1,1}^{j_1,*,j_0,*} ( \wt{A} ) \| \\ \leq 
    & ~ \|  f_c(A,x)^2 \cdot f(A,x)_{j_1} \cdot f(A,x)_{j_0} \cdot  ( (A_{j_1,*})^\top \circ x ) \cdot  {\bf 1}_d^\top \\
    & ~ - f_c(\wt{A},x)^2 \cdot f(\wt{A},x)_{j_1} \cdot f(\wt{A},x)_{j_0} \cdot ( (\wt{A}_{j_1,*})^\top \circ x ) \cdot  {\bf 1}_d^\top \|_F\\
    \leq & ~ |  f_c(A,x) - f_c(\wt{A},x) |\cdot |f_c(A,x)|\cdot |f(A,x)_{j_1}|\cdot | f(A,x)_{j_0} | \cdot   \| (A_{j_1,*})^\top\|_2 \cdot \| \diag(x)\|_F  \cdot \| {\bf 1}_d^\top\|_2 \\
    & + ~   |f_c(\wt{A},x)|\cdot |  f_c(A,x) - f_c(\wt{A},x) |\cdot |f(A,x)_{j_1}|\cdot | f(A,x)_{j_0} | \cdot   \| (A_{j_1,*})^\top\|_2 \cdot \| \diag(x)\|_F  \cdot \| {\bf 1}_d^\top\|_2\\
    & + ~   |f_c(\wt{A},x)|\cdot |f_c(\wt{A},x)|\cdot  |f(A,x)_{j_1} - f(\wt{A},x)_{j_1}|\cdot | f(A,x)_{j_0} | \cdot   \| (A_{j_1,*})^\top\|_2 \cdot \| \diag(x)\|_F  \cdot \| {\bf 1}_d^\top\|_2 \\
    & + ~   |f_c(\wt{A},x)|\cdot |f_c(\wt{A},x)|\cdot  |f(\wt{A},x)_{j_1}|\cdot | f(A,x)_{j_0} -  f(\wt{A},x)_{j_0} | \cdot   \| (A_{j_1,*})^\top\|_2 \cdot \| \diag(x)\|_F  \cdot \| {\bf 1}_d^\top\|_2\\
    & + ~ |f_c(\wt{A},x)|\cdot |f_c(\wt{A},x)|\cdot  |f(\wt{A},x)_{j_1}|\cdot  | f(\wt{A},x)_{j_0} | \cdot \| (A_{j_1,*})^\top - (\wt{A}_{j_1,*})^\top\|_2 \cdot \| \diag(x)\|_F  \cdot \| {\bf 1}_d^\top\|_2\\
    \leq & ~ 12R^2  \beta^{-2} \cdot n \cdot \sqrt{d}  \exp(3R^2)\|A - \wt{A}\|_F \\
    &+ ~ 12R^2  \beta^{-2} \cdot n \cdot \sqrt{d}  \exp(3R^2)\|A - \wt{A}\|_F \\
    & + ~ 8R^2 \beta^{-2} \cdot n \cdot \sqrt{d}\exp(3R^2) \cdot \|A - \wt{A}\|_F \\
    & + ~ 8R^2\beta^{-2} \cdot n \cdot \sqrt{d}\exp(3R^2) \cdot \|A - \wt{A}\|_F \\
    & + ~ 4 R \cdot \sqrt{d} \cdot \|A - \wt{A} \|_F \\
    \leq &  ~ 44 \beta^{-2} \cdot n \cdot \sqrt{d}\exp(4R^2) \cdot \|A - \wt{A}\|_F \\
    \leq &  ~ \beta^{-2} \cdot n \cdot \sqrt{d}\exp(5R^2) \cdot \|A - \wt{A}\|_F 
\end{align*}

{\bf Proof of Part 2.}
\begin{align*}
    & ~ \| B_{3,1,2}^{j_1,*,j_0,*} (A) - B_{3,1,2}^{j_1,*,j_0,*} ( \wt{A} ) \|_F \\
    \leq & ~ \| f(A,x)_{j_1} \cdot f(A,x)_{j_0} \cdot c(A,x)_{j_1} \cdot  f_c(A,x) \cdot ( (A_{j_1,*})^\top \circ x ) \cdot  {\bf 1}_d^\top \\
    & ~- f(\wt{A},x)_{j_1} \cdot f(\wt{A},x)_{j_0} \cdot c(\wt{A},x)_{j_1} \cdot  f_c(\wt{A},x) \cdot ( (\wt{A}_{j_1,*})^\top \circ x ) \cdot  {\bf 1}_d^\top \|_F\\
 \leq & ~ |  f_c(A,x) - f_c(\wt{A},x) |\cdot |c(A,x)_{j_1}|\cdot |f(A,x)_{j_1}|\cdot | f(A,x)_{j_0} | \cdot   \| (A_{j_1,*})^\top\|_2 \cdot \| \diag(x)\|_F  \cdot \| {\bf 1}_d^\top\|_2 \\
    & + ~   |f_c(\wt{A},x)|\cdot |  c(A,x)_{j_1} - c(\wt{A},x)_{j_1} |\cdot |f(A,x)_{j_1}|\cdot | f(A,x)_{j_0} | \cdot   \| (A_{j_1,*})^\top\|_2 \cdot \| \diag(x)\|_F  \cdot \| {\bf 1}_d^\top\|_2\\
    & + ~   |f_c(\wt{A},x)|\cdot |c(\wt{A},x)_{j_1} |\cdot  |f(A,x)_{j_1} - f(\wt{A},x)_{j_1}|\cdot | f(A,x)_{j_0} | \cdot   \| (A_{j_1,*})^\top\|_2 \cdot \| \diag(x)\|_F  \cdot \| {\bf 1}_d^\top\|_2 \\
    & + ~   |f_c(\wt{A},x)|\cdot |c(\wt{A},x)_{j_1} |\cdot  |f(\wt{A},x)_{j_1}|\cdot | f(A,x)_{j_0} -  f(\wt{A},x)_{j_0} | \cdot   \| (A_{j_1,*})^\top\|_2 \cdot \| \diag(x)\|_F  \cdot \| {\bf 1}_d^\top\|_2\\
    & + ~ |f_c(\wt{A},x)|\cdot |c(\wt{A},x)_{j_1} |\cdot  |f(\wt{A},x)_{j_1}|\cdot  | f(\wt{A},x)_{j_0} | \cdot \| (A_{j_1,*})^\top - (\wt{A}_{j_1,*})^\top\|_2 \cdot \| \diag(x)\|_F  \cdot \| {\bf 1}_d^\top\|_2\\
     \leq & ~ 12R^2  \beta^{-2} \cdot n \cdot \sqrt{d}  \exp(3R^2)\|A - \wt{A}\|_F \\
    &+ ~ 4R^2  \beta^{-2} \cdot n \cdot \sqrt{d}  \exp(3R^2)\|A - \wt{A}\|_F \\
    & + ~ 8R^2 \beta^{-2} \cdot n \cdot \sqrt{d}\exp(3R^2) \cdot \|A - \wt{A}\|_F \\
    & + ~ 8R^2\beta^{-2} \cdot n \cdot \sqrt{d}\exp(3R^2) \cdot \|A - \wt{A}\|_F \\
    & + ~ 4 R \cdot \sqrt{d} \cdot \|A - \wt{A} \|_F \\
    \leq &  ~ 36 \beta^{-2} \cdot n \cdot \sqrt{d}\exp(4R^2) \cdot \|A - \wt{A}\|_F \\
    \leq &  ~ \beta^{-2} \cdot n \cdot \sqrt{d}\exp(5R^2) \cdot \|A - \wt{A}\|_F 
\end{align*}

{\bf Proof of Part 3.}
\begin{align*}
    & ~ \| B_{3,1,3}^{j_1,*,j_0,*} (A) - B_{3,1,3}^{j_1,*,j_0,*} ( \wt{A} ) \|_F \\
    \leq & ~ \| f(A,x)_{j_1} \cdot  f(A,x)_{j_0} \cdot f_c(A,x)^2 \cdot  ((A_{j_1,*}) \circ x^\top) \cdot  ((A_{j_0,*})^\top \circ x ) \cdot I_d \\
    & ~- f(\wt{A},x)_{j_1} \cdot  f(\wt{A},x)_{j_0} \cdot f_c(\wt{A},x)^2 \cdot  ((\wt{A}_{j_1,*}) \circ x^\top) \cdot  ((\wt{A}_{j_0,*})^\top \circ x )  \cdot I_d \|_F\\
 \leq & ~  |f(A,x)_{j_1} - f(\wt{A},x)_{j_1}|\cdot | f(A,x)_{j_0}| \cdot |f_c(A,x)|^2 \cdot \| A_{j_1,*} \|_2 \cdot \| \diag(x)\|_F \cdot  \| (A_{j_0,*})^\top\|_2 \cdot \| \diag(x)\|_F  \cdot \| I_d \|_F\\
    & + ~   |  f(\wt{A},x)_{j_1}| | f(A,x)_{j_0} -  f(\wt{A},x)_{j_0}| \cdot |f_c(A,x)|^2 \cdot \| A_{j_1,*} \|_2 \cdot \| \diag(x)\|_F \cdot  \| (A_{j_0,*})^\top\|_2 \cdot \| \diag(x)\|_F  \cdot \| I_d \|_F\\
    & + ~    |  f(\wt{A},x)_{j_1}||  f(\wt{A},x)_{j_0}| |f_c(A,x) - f_c(\wt{A},x)| \cdot |f_c(A,x)|  \cdot \| A_{j_1,*} \|_2 \cdot \| \diag(x)\|_F \cdot  \| (A_{j_0,*})^\top\|_2 \\
          & ~\cdot \| \diag(x)\|_F  \cdot \| I_d \|_F \\
    & + ~  |  f(\wt{A},x)_{j_1}||  f(\wt{A},x)_{j_0}| |f_c(\wt{A},x)| \cdot |f_c(A,x) - f_c(\wt{A},x)|   \cdot \| A_{j_1,*} \|_2 \cdot \| \diag(x)\|_F \cdot  \| (A_{j_0,*})^\top\|_2 \\
          & ~\cdot \| \diag(x)\|_F  \cdot \| I_d \|_F  \\
    & + ~    |  f(\wt{A},x)_{j_1}||  f(\wt{A},x)_{j_0}| |f_c(\wt{A},x)| \cdot |f_c(\wt{A},x)|  \cdot \| A_{j_1,*} - \wt{A}_{j_1,*} \|_2 \cdot \| \diag(x)\|_F \cdot  \| (A_{j_0,*})^\top\|_2 \\
          & ~ \cdot \| \diag(x)\|_F  \cdot \| I_d \|_F \\
     & + ~    |  f(\wt{A},x)_{j_1}| |  f(\wt{A},x)_{j_0}||f_c(\wt{A},x)| \cdot |f_c(\wt{A},x)|  \cdot \|   \wt{A}_{j_1,*} \|_2 \cdot \| \diag(x)\|_F \cdot  \| (A_{j_0,*})^\top - (\wt{A}_{j_0,*})^\top\|_2 \\
          & ~\cdot \| \diag(x)\|_F  \cdot \| I_d \|_F \\
    \leq & ~ 8R^4  \beta^{-2} \cdot n \cdot \sqrt{d}  \exp(3R^2)\|A - \wt{A}\|_F \\
    &+ ~ 8R^4  \beta^{-2} \cdot n \cdot \sqrt{d}  \exp(3R^2)\|A - \wt{A}\|_F \\
    & + ~ 12R^4 \beta^{-2} \cdot n \cdot \sqrt{d}\exp(3R^2) \cdot \|A - \wt{A}\|_F \\
    & + ~ 12R^4\beta^{-2} \cdot n \cdot \sqrt{d}\exp(3R^2) \cdot \|A - \wt{A}\|_F \\
    & + ~ 4 R^3 \cdot \sqrt{d} \cdot \|A - \wt{A} \|_F \\
    & + ~ 4 R^3 \cdot \sqrt{d} \cdot \|A - \wt{A} \|_F \\
    \leq &  ~ 48 \beta^{-2} \cdot n \cdot \sqrt{d}\exp(5R^2) \cdot \|A - \wt{A}\|_F \\
    \leq &  ~ \beta^{-2} \cdot n \cdot \sqrt{d}\exp(6R^2) \cdot \|A - \wt{A}\|_F 
\end{align*}

{\bf Proof of Part 4.}
\begin{align*}
& ~ \| B_{3,1,4}^{j_1,*,j_0,*} (A) - B_{3,1,4}^{j_1,*,j_0,*} ( \wt{A} ) \| \\ 
\leq  & ~ \| - f(A,x)_{j_1} \cdot  f(A,x)_{j_0} \cdot f_c(A,x)^2  \cdot h(A,x)^\top \cdot ( (A_{j_0,*})^\top \circ x ) \cdot I_d  \\
    & ~- (-f(\wt{A},x)_{j_1} \cdot  f(\wt{A},x)_{j_0} \cdot f_c(\wt{A},x)^2  \cdot h(\wt{A},x)^\top \cdot ( (\wt{A}_{j_0,*})^\top \circ x ) \cdot I_d)\|_F \\
    \leq  & ~ \|  f(A,x)_{j_1} \cdot  f(A,x)_{j_0} \cdot f_c(A,x)^2  \cdot h(A,x)^\top \cdot ( (A_{j_0,*})^\top \circ x ) \cdot I_d  \\
    & ~- f(\wt{A},x)_{j_1} \cdot  f(\wt{A},x)_{j_0} \cdot f_c(\wt{A},x)^2  \cdot h(\wt{A},x)^\top \cdot ( (\wt{A}_{j_0,*})^\top \circ x ) \cdot I_d)\|_F \\
    \leq & ~  |f(A,x)_{j_1} - f(\wt{A},x)_{j_1}|\cdot | f(A,x)_{j_0}| \cdot |f_c(A,x)|^2 \cdot \| h(A,x)^\top \|_2  \cdot  \| (A_{j_0,*})^\top\|_2 \\ 
    & ~ \cdot \| \diag(x)\|_F  \cdot \| I_d \|_F\\
    & + ~   |  f(\wt{A},x)_{j_1}|\cdot | f(A,x)_{j_0} -  f(\wt{A},x)_{j_0}| \cdot |f_c(A,x)|^2 \cdot \| h(A,x)^\top \|_2 \cdot  \| (A_{j_0,*})^\top\|_2 \\
    & ~ \cdot \| \diag(x)\|_F  \cdot \| I_d \|_F\\
    & + ~    |  f(\wt{A},x)_{j_1}|\cdot |  f(\wt{A},x)_{j_0}| \cdot |f_c(A,x) - f_c(\wt{A},x)| \cdot |f_c(A,x)|  \cdot \| h(A,x)^\top \|_2 \cdot  \| (A_{j_0,*})^\top\|_2 \\
    & ~ \cdot \| \diag(x)\|_F  \cdot \| I_d \|_F \\
    & + ~  |  f(\wt{A},x)_{j_1}|\cdot |  f(\wt{A},x)_{j_0}|\cdot |f_c(\wt{A},x)| \cdot |f_c(A,x) - f_c(\wt{A},x)|   \cdot \| h(A,x)^\top \|_2 \cdot  \| (A_{j_0,*})^\top\|_2 \\
    & ~ \cdot \| \diag(x)\|_F  \cdot \| I_d \|_F  \\
    & + ~    |  f(\wt{A},x)_{j_1}|\cdot |  f(\wt{A},x)_{j_0}|\cdot |f_c(\wt{A},x)| \cdot |f_c(\wt{A},x)|  \cdot  \| h(A,x)^\top -h(\wt{A},x)^\top \|_2  \cdot   \| (A_{j_0,*})^\top\|_2 \\
    & ~ \cdot \| \diag(x)\|_F  \cdot \| I_d \|_F \\
       & + ~    |  f(\wt{A},x)_{j_1}|\cdot |  f(\wt{A},x)_{j_0}|\cdot |f_c(\wt{A},x)| \cdot |f_c(\wt{A},x)|  \cdot  \|  h(\wt{A},x)^\top \|_2  \cdot   \| (A_{j_0,*})^\top -  (\wt{A}_{j_0,*})^\top\|_2 \\
       & ~ \cdot \| \diag(x)\|_F  \cdot \| I_d \|_F \\
    \leq & ~ 8R^4  \beta^{-2} \cdot n \cdot \sqrt{d}  \exp(3R^2)\|A - \wt{A}\|_F \\
    &+ ~ 8R^4  \beta^{-2} \cdot n \cdot \sqrt{d}  \exp(3R^2)\|A - \wt{A}\|_F \\
    & + ~ 12R^4\beta^{-2} \cdot n \cdot \sqrt{d}\exp(3R^2) \cdot \|A - \wt{A}\|_F \\
    & + ~ 12R^4\beta^{-2} \cdot n \cdot \sqrt{d}\exp(3R^2) \cdot \|A - \wt{A}\|_F \\
    & + ~ 12R^2\beta^{-2} \cdot n \cdot \sqrt{d}\exp(4R^2) \cdot \|A - \wt{A}\|_F \\
    & + ~ 4R^3 \cdot \sqrt{d} \cdot \|A - \wt{A}\|_F \\
    \leq &  ~ 56\beta^{-2} \cdot n \cdot \sqrt{d}\exp(5R^2) \cdot \|A - \wt{A}\|_F \\
\leq &  ~ \beta^{-2} \cdot n \cdot \sqrt{d}\exp(6R^2) \cdot \|A - \wt{A}\|_F
\end{align*}

{\bf Proof of Part 5.}
\begin{align*}
& ~ \| B_{3,1,5}^{j_1,*,j_0,*} (A) - B_{3,1,5}^{j_1,*,j_0,*} ( \wt{A} ) \| \\ \leq 
    & ~ \| f(A,x)_{j_1} \cdot  f(A,x)_{j_0} \cdot (-f_2(A,x) + f(A,x)_{j_1})  \cdot f_c(A,x)  \cdot h(A,x)^\top \cdot ( (A_{j_0,*})^\top \circ x ) \cdot I_d \\
    & ~- f(\wt{A},x)_{j_1} \cdot  f(\wt{A},x)_{j_0} \cdot (-f_2(\wt{A},x) + f(\wt{A},x)_{j_1})  \cdot f_c(\wt{A},x)  \cdot h(\wt{A},x)^\top \cdot ( (\wt{A}_{j_0,*})^\top \circ x ) \cdot I_d \|_F\\
    \leq & ~ |  f_c(A,x) - f_c(\wt{A},x) |\cdot (| -f_2(A,x)| + |f(A,x)_{j_1}|) \cdot |f(A,x)_{j_1}|\cdot | f(A,x)_{j_0} | \cdot \| h(A,x)^\top \|_2\\
    & ~ \cdot \| (A_{j_0,*})^\top\|_2 \cdot \| \diag(x)\|_F  \cdot \| I_d \|_F \\
    & + ~  |  f_c(\wt{A},x) |\cdot (| f_2(A,x) -  f_2(\wt{A},x)| + |f(A,x)_{j_1}- f(\wt{A},x)_{j_1}|) \cdot |f(A,x)_{j_1}|\cdot | f(A,x)_{j_0} | \cdot \| h(A,x)^\top \|_2 \\
    & ~\cdot  \| (A_{j_0,*})^\top\|_2 \cdot \| \diag(x)\|_F  \cdot \| I_d \|_F \\
    & + ~     |  f_c(\wt{A},x) |\cdot (| f_2(\wt{A},x)| + |  f(\wt{A},x)_{j_1}|) \cdot |f(A,x)_{j_1} - f(\wt{A},x)_{j_1}| \cdot | f(A,x)_{j_0} | \cdot \| h(A,x)^\top \|_2  \\
    & ~\cdot   \| (A_{j_0,*})^\top\|_2 \cdot \| \diag(x)\|_F  \cdot \| I_d \|_F \\
    & + ~     |  f_c(\wt{A},x) |\cdot (| f_2(\wt{A},x)| + |  f(\wt{A},x)_{j_1}|) \cdot | f(\wt{A},x)_{j_1}| \cdot | f(A,x)_{j_0} - f(\wt{A},x)_{j_0} | \cdot \| h(A,x)^\top \|_2\\
    & ~ \cdot   \| (A_{j_0,*})^\top\|_2 \cdot \| \diag(x)\|_F  \cdot \| I_d \|_F \\
     & + ~     |  f_c(\wt{A},x) |\cdot (| f_2(\wt{A},x)| + |  f(\wt{A},x)_{j_1}|) \cdot | f(\wt{A},x)_{j_1}| \cdot |  f(\wt{A},x)_{j_0} | \cdot \| h(A,x)^\top -h(\wt{A},x)^\top \|_2\\
    & ~ \cdot   \| (A_{j_0,*})^\top  \|_2 \cdot \| \diag(x)\|_F  \cdot \| I_d \|_F \\
    & + ~     |  f_c(\wt{A},x) |\cdot (| f_2(\wt{A},x)| + |  f(\wt{A},x)_{j_1}|) \cdot | f(\wt{A},x)_{j_1}| \cdot |  f(\wt{A},x)_{j_0} | \cdot \| h(\wt{A},x)^\top \|_2 \\
    & ~ \cdot   \| (A_{j_0,*})^\top - (\wt{A}_{j_0,*})^\top\|_2 \cdot \| \diag(x)\|_F  \cdot \| I_d \|_F \\
    \leq & ~ 12R^4  \beta^{-2} \cdot n \cdot \sqrt{d}  \exp(3R^2)\|A - \wt{A}\|_F \\
    &+ ~ 12 R^4 \beta^{-2} \cdot n \cdot \sqrt{d}  \exp(3R^2)\|A - \wt{A}\|_F \\
    & + ~ 8 R^4\beta^{-2} \cdot n \cdot \sqrt{d}\exp(3R^2) \cdot \|A - \wt{A}\|_F \\
    & + ~ 8 R^4\beta^{-2} \cdot n \cdot \sqrt{d}\exp(3R^2) \cdot \|A - \wt{A}\|_F \\
    & + ~ 12R^2\beta^{-2} \cdot n \cdot \sqrt{d}\exp(4R^2) \cdot \|A - \wt{A}\|_F \\
    & + ~ 4R^3 \cdot \sqrt{d} \cdot \|A - \wt{A}\|_F\\
    \leq &  ~ 56\beta^{-2} \cdot n \cdot \sqrt{d}\exp(5R^2) \cdot \|A - \wt{A}\|_F \\
\leq &  ~ \beta^{-2} \cdot n \cdot \sqrt{d}\exp(6R^2) \cdot \|A - \wt{A}\|_F
\end{align*}

{\bf Proof of Part 6.}
\begin{align*}
& ~ \| B_{3,1,6}^{j_1,*,j_0,*} (A) - B_{3,1,6}^{j_1,*,j_0,*} ( \wt{A} ) \| \\ \leq 
    & ~ \| f(A,x)_{j_1} \cdot  f(A,x)_{j_0} \cdot (-f_c(A,x) + f(A,x)_{j_1})  \cdot f_c(A,x)  \cdot h(A,x)^\top \cdot ( (A_{j_0,*})^\top \circ x ) \cdot I_d \\
    & ~- f(\wt{A},x)_{j_1} \cdot  f(\wt{A},x)_{j_0} \cdot (-f_c(\wt{A},x) + f(\wt{A},x)_{j_1})  \cdot f_c(\wt{A},x)  \cdot h(\wt{A},x)^\top \cdot ( (\wt{A}_{j_0,*})^\top \circ x ) \cdot I_d\|_F\\
    \leq & ~ |  f_c(A,x) - f_c(\wt{A},x) |\cdot (| -f_c(A,x)| + |f(A,x)_{j_1}|) \cdot |f(A,x)_{j_1}|\cdot | f(A,x)_{j_0} | \cdot \| h(A,x)^\top \|_2\\
    & ~ \cdot \| (A_{j_0,*})^\top\|_2 \cdot \| \diag(x)\|_F  \cdot \| I_d \|_F \\
    & + ~  |  f_c(\wt{A},x) |\cdot (| f_c(A,x) -  f_c(\wt{A},x)| + |f(A,x)_{j_1}- f(\wt{A},x)_{j_1}|) \cdot |f(A,x)_{j_1}|\cdot | f(A,x)_{j_0} | \cdot \| h(A,x)^\top \|_2 \\
    & ~\cdot  \| (A_{j_0,*})^\top\|_2 \cdot \| \diag(x)\|_F  \cdot \| I_d \|_F \\
    & + ~     |  f_c(\wt{A},x) |\cdot (| f_c(\wt{A},x)| + |  f(\wt{A},x)_{j_1}|) \cdot |f(A,x)_{j_1} - f(\wt{A},x)_{j_1}| \cdot | f(A,x)_{j_0} | \cdot \| h(A,x)^\top \|_2  \\
    & ~\cdot   \| (A_{j_0,*})^\top\|_2 \cdot \| \diag(x)\|_F  \cdot \| I_d \|_F \\
    & + ~     |  f_c(\wt{A},x) |\cdot (| f_c(\wt{A},x)| + |  f(\wt{A},x)_{j_1}|) \cdot | f(\wt{A},x)_{j_1}| \cdot | f(A,x)_{j_0} - f(\wt{A},x)_{j_0} | \cdot \| h(A,x)^\top \|_2\\
    & ~ \cdot   \| (A_{j_0,*})^\top\|_2 \cdot \| \diag(x)\|_F  \cdot \| I_d \|_F \\
     & + ~     |  f_c(\wt{A},x) |\cdot (| f_c(\wt{A},x)| + |  f(\wt{A},x)_{j_1}|) \cdot | f(\wt{A},x)_{j_1}| \cdot |  f(\wt{A},x)_{j_0} | \cdot \| h(A,x)^\top -h(\wt{A},x)^\top \|_2\\
    & ~ \cdot   \| (A_{j_0,*})^\top  \|_2 \cdot \| \diag(x)\|_F  \cdot \| I_d \|_F \\
    & + ~     |  f_c(\wt{A},x) |\cdot (| f_c(\wt{A},x)| + |  f(\wt{A},x)_{j_1}|) \cdot | f(\wt{A},x)_{j_1}| \cdot |  f(\wt{A},x)_{j_0} | \cdot \| h(\wt{A},x)^\top \|_2 \\
    & ~ \cdot   \| (A_{j_0,*})^\top - (\wt{A}_{j_0,*})^\top\|_2 \cdot \| \diag(x)\|_F  \cdot \| I_d \|_F \\
    \leq & ~ 18R^4  \beta^{-2} \cdot n \cdot \sqrt{d}  \exp(3R^2)\|A - \wt{A}\|_F \\
    &+ ~ 16 R^4 \beta^{-2} \cdot n \cdot \sqrt{d}  \exp(3R^2)\|A - \wt{A}\|_F \\
    & + ~ 12 R^4\beta^{-2} \cdot n \cdot \sqrt{d}\exp(3R^2) \cdot \|A - \wt{A}\|_F \\
    & + ~ 12 R^4\beta^{-2} \cdot n \cdot \sqrt{d}\exp(3R^2) \cdot \|A - \wt{A}\|_F \\
    & + ~ 18R^2\beta^{-2} \cdot n \cdot \sqrt{d}\exp(4R^2) \cdot \|A - \wt{A}\|_F \\
    & + ~ 6R^3 \cdot \sqrt{d} \cdot \|A - \wt{A}\|_F\\
    \leq &  ~ 82\beta^{-2} \cdot n \cdot \sqrt{d}\exp(5R^2) \cdot \|A - \wt{A}\|_F \\
\leq &  ~ \beta^{-2} \cdot n \cdot \sqrt{d}\exp(6R^2) \cdot \|A - \wt{A}\|_F
\end{align*}

{\bf Proof of Part 7.}
\begin{align*}
& ~ \| B_{3,1,7}^{j_1,*,j_0,*} (A) - B_{3,1,7}^{j_1,*,j_0,*} ( \wt{A} ) \| \\ \leq 
    & ~ \|  f(A,x)_{j_1} \cdot  f(A,x)_{j_0} \cdot f_c(A,x)   \cdot p_{j_1}(A,x)^\top \cdot  ( (A_{j_0,*})^\top \circ x )\cdot I_d \\
    & ~ - f(\wt{A},x)_{j_1} \cdot  f(\wt{A},x)_{j_0} \cdot f_c(\wt{A},x)   \cdot p_{j_1}(\wt{A},x)^\top \cdot  ( (\wt{A}_{j_0,*})^\top \circ x )\cdot I_d \|_F\\
    \leq & ~  |f(A,x)_{j_1} - f(\wt{A},x)_{j_1}|\cdot | f(A,x)_{j_0}| \cdot |f_c(A,x)|  \cdot \| p_{j_1}(A,x)^\top \|_2  \cdot  \| (A_{j_0,*})^\top\|_2 \cdot \| \diag(x)\|_F  \cdot \| I_d \|_F\\
    & + ~   |  f(\wt{A},x)_{j_1}|\cdot | f(A,x)_{j_0} -  f(\wt{A},x)_{j_0}| \cdot |f_c(A,x)|  \cdot \|  p_{j_1}(A,x)^\top \|_2 \cdot  \| (A_{j_0,*})^\top\|_2 \cdot \| \diag(x)\|_F  \cdot \| I_d \|_F\\
    & + ~    |  f(\wt{A},x)_{j_1}|\cdot |  f(\wt{A},x)_{j_0}| \cdot |f_c(A,x) - f_c(\wt{A},x)|    \cdot \| p_{j_1}(A,x)^\top \|_2 \cdot  \| (A_{j_0,*})^\top\|_2 \cdot \| \diag(x)\|_F  \cdot \| I_d \|_F \\
    & + ~  |  f(\wt{A},x)_{j_1}|\cdot |  f(\wt{A},x)_{j_0}|\cdot |f_c(\wt{A},x)|   \cdot \| p_{j_1}(A,x)^\top - p_{j_1}(\wt{A},x)^\top\|_2 \cdot  \| (A_{j_0,*})^\top\|_2 \cdot \| \diag(x)\|_F  \cdot \| I_d \|_F  \\
       & + ~    |  f(\wt{A},x)_{j_1}|\cdot |  f(\wt{A},x)_{j_0}|\cdot |f_c(\wt{A},x)|  \cdot  \|  p_{j_1}(\wt{A},x)^\top \|_2  \cdot   \| (A_{j_0,*})^\top -  (\wt{A}_{j_0,*})^\top\|_2 \cdot \| \diag(x)\|_F  \cdot \| I_d \|_F \\
    \leq & ~ 24R^4  \beta^{-2} \cdot n \cdot \sqrt{d}  \exp(3R^2)\|A - \wt{A}\|_F \\
    &+ ~ 24R^4  \beta^{-2} \cdot n \cdot \sqrt{d}  \exp(3R^2)\|A - \wt{A}\|_F \\
    & + ~ 36R^4\beta^{-2} \cdot n \cdot \sqrt{d}\exp(3R^2) \cdot \|A - \wt{A}\|_F \\
    & + ~ 26R^2\beta^{-2} \cdot n \cdot \sqrt{d}\exp(4R^2) \cdot \|A - \wt{A}\|_F \\
    & + ~ 12R^3 \cdot \sqrt{d} \cdot \|A - \wt{A}\|_F\\
    \leq &  ~ 122\beta^{-2} \cdot n \cdot \sqrt{d}\exp(5R^2) \cdot \|A - \wt{A}\|_F \\
\leq &  ~ \beta^{-2} \cdot n \cdot \sqrt{d}\exp(6R^2) \cdot \|A - \wt{A}\|_F
\end{align*}

{\bf Proof of Part 8.}
\begin{align*}
& ~ \| B_{3,2,1}^{j_1,*,j_0,*} (A) - B_{3,2,1}^{j_1,*,j_0,*} ( \wt{A} ) \|_F \\ \leq 
    & ~ \| f(A,x)_{j_1} \cdot  f(A,x)_{j_0} \cdot f_c(A,x) \cdot c_g(A,x)^{\top} \cdot  ( (A_{j_0,*})^\top \circ x ) \cdot x \cdot {\bf 1}_d^{\top}\\
    & ~ - f(\wt{A},x)_{j_1} \cdot  f(\wt{A},x)_{j_0} \cdot f_c(\wt{A},x) \cdot c_g(\wt{A},x)^{\top} \cdot  ( (\wt{A}_{j_0,*})^\top \circ x ) \cdot x \cdot {\bf 1}_d^{\top} \|_F \\
    \leq & ~  |f(A,x)_{j_1} - f(\wt{A},x)_{j_1}|\cdot | f(A,x)_{j_0}| \cdot |f_c(A,x)|  \cdot \| c_g(A,x)^\top \|_2  \cdot  \| (A_{j_0,*})^\top\|_2 \cdot \| \diag(x)\|_F \cdot \| x \|_2 \cdot \| {\bf 1}_d^{\top} \|_F\\
    & + ~   |  f(\wt{A},x)_{j_1}|\cdot | f(A,x)_{j_0} -  f(\wt{A},x)_{j_0}| \cdot |f_c(A,x)|  \cdot \|   c_g(A,x)^\top \|_2 \cdot  \| (A_{j_0,*})^\top\|_2 \cdot \| \diag(x)\|_F  \cdot \| x \|_2 \cdot \| {\bf 1}_d^{\top} \|_F\\
    & + ~    |  f(\wt{A},x)_{j_1}|\cdot |  f(\wt{A},x)_{j_0}| \cdot |f_c(A,x) - f_c(\wt{A},x)|    \cdot \| c_g(A,x)^\top \|_2 \cdot  \| (A_{j_0,*})^\top\|_2 \cdot \| \diag(x)\|_F  \cdot \| x \|_2 \cdot \| {\bf 1}_d^{\top} \|_F\\
    & + ~  |  f(\wt{A},x)_{j_1}|\cdot |  f(\wt{A},x)_{j_0}|\cdot |f_c(\wt{A},x)|   \cdot \| c_g(A,x)^\top - c_g(\wt{A},x)^\top \|_2 \cdot  \| (A_{j_0,*})^\top\|_2 \cdot \| \diag(x)\|_F  \cdot \| x \|_2 \cdot \| {\bf 1}_d^{\top} \|_F\\
    & + ~    |  f(\wt{A},x)_{j_1}|\cdot |  f(\wt{A},x)_{j_0}|\cdot |f_c(\wt{A},x)|  \cdot  \|  c_g(\wt{A},x)^\top  \|_2  \cdot   \| (A_{j_0,*})^\top -  (\wt{A}_{j_0,*})^\top\|_2 \cdot \| \diag(x)\|_F  \cdot \| x \|_2 \cdot \| {\bf 1}_d^{\top} \|_F\\
    \leq & ~ 20R^4 \beta^{-2} \cdot n \cdot \sqrt{d}  \exp(3R^2)\|A - \wt{A}\|_F \\
    &+ ~ 20R^4  \beta^{-2} \cdot n \cdot \sqrt{d}  \exp(3R^2)\|A - \wt{A}\|_F \\
    & + ~ 30R^4\beta^{-2} \cdot n \cdot \sqrt{d}\exp(3R^2) \cdot \|A - \wt{A}\|_F \\
    & + ~ 40R^4\beta^{-2} \cdot n \cdot \sqrt{d}\exp(3R^2) \cdot \|A - \wt{A}\|_F \\
     & + ~ 10R^3 \cdot \sqrt{d} \cdot \|A - \wt{A}\|_F\\
    \leq &  ~ 120\beta^{-2} \cdot n \cdot \sqrt{d}\exp(5R^2) \cdot \|A - \wt{A}\|_F \\
    \leq &  ~ \beta^{-2} \cdot n \cdot \sqrt{d}\exp(6R^2) \cdot \|A - \wt{A}\|_F
\end{align*}

{\bf Proof of Part 9.}
\begin{align*}
  & ~ \|  B_{3,3,1}^{j_1,*,j_0,*} (A) - B_{3,3,1}^{j_1,*,j_0,*} ( \wt{A} ) \|_F \\
    = & ~ \| f(A,x)_{j_1} \cdot f(A,x)_{j_0}  \cdot f_2( A,x) \cdot  c_g(A,x)^{\top} \cdot  ( (A_{j_0,*})^\top \circ x ) \cdot x \cdot {\bf 1}_d^{\top} \\
    & ~ - f(\wt{A},x)_{j_1} \cdot f(\wt{A},x)_{j_0}  \cdot f_2( \wt{A},x) \cdot  c_g(\wt{A},x)^{\top} \cdot  ( (\wt{A}_{j_0,*})^\top \circ x ) \cdot x \cdot {\bf 1}_d^{\top} \|_F \\
    \leq & ~ |f(A,x)_{j_1} - f(\wt{A},x)_{j_1}| \cdot |f(A,x)_{j_0} | \cdot |f_2( A,x) |\cdot  \| c_g(A,x)^{\top} \|_2 \cdot   \|(A_{j_0,*})^\top \|_2 \|\diag( x)
    \|_F \cdot \| x \|_2 \cdot \|{\bf 1}_d^{\top} \|_2 \\
    & + ~ |f(\wt{A},x)_{j_1}| \cdot |f(A,x)_{j_0} - f(\wt{A},x)_{j_0} | \cdot |f_2( A,x) |\cdot  \| c_g(A,x)^{\top} \|_2 \cdot   \|(A_{j_0,*})^\top \|_2 \|\diag( x)
    \|_F \cdot \| x \|_2 \cdot \|{\bf 1}_d^{\top} \|_2 \\
    & + ~ |f(\wt{A},x)_{j_1}| \cdot |f(\wt{A},x)_{j_0} | \cdot |f_2( A,x)  -f_2( \wt{A},x)|\cdot  \| c_g(A,x)^{\top} \|_2 \cdot   \|(A_{j_0,*})^\top \|_2 \|\diag( x)
    \|_F \cdot \| x \|_2 \cdot \|{\bf 1}_d^{\top} \|_2 \\
    & + ~ |f(\wt{A},x)_{j_1}| \cdot |f(\wt{A},x)_{j_0} | \cdot |f_2( \wt{A},x)|\cdot  \| c_g(A,x)^{\top}  - c_g(\wt{A},x)^{\top} \|_2 \cdot   \|(A_{j_0,*})^\top \|_2 \|\diag( x)
    \|_F \cdot \| x \|_2 \cdot \|{\bf 1}_d^{\top} \|_2 \\
    & + ~ |f(\wt{A},x)_{j_1}| \cdot |f(\wt{A},x)_{j_0} | \cdot |f_2( \wt{A},x)|\cdot  \| c_g(\wt{A},x)^{\top} \|_2 \cdot   \|(A_{j_0,*})^\top -\wt{A}_{j_0,*})^\top \|_2 \|\diag( x)
    \|_F \cdot \| x \|_2 \cdot \|{\bf 1}_d^{\top} \|_2 \\
    \leq & ~ 10R^4 \beta^{-2} \cdot n \cdot \sqrt{d}  \exp(3R^2)\|A - \wt{A}\|_F \\
    & + ~ 10R^4 \beta^{-2} \cdot n \cdot \sqrt{d}  \exp(3R^2)\|A - \wt{A}\|_F \\
    & + ~ 20R^4 \beta^{-2} \cdot n \cdot \sqrt{d}  \exp(3R^2)\|A - \wt{A}\|_F \\
    & + ~ 20R^4 \beta^{-2} \cdot n \cdot \sqrt{d}  \exp(3R^2)\|A - \wt{A}\|_F \\
    & + ~ 5R^3  \sqrt{d} \cdot \|A - \wt{A}\|_F \\
    \leq & ~ 65\beta^{-2} \cdot n \cdot \sqrt{d}  \exp(5R^2)\|A - \wt{A}\|_F \\
    \leq & ~ \beta^{-2} \cdot n \cdot \sqrt{d}  \exp(6R^2)\|A - \wt{A}\|_F 
\end{align*}

{\bf Proof of Part 10.}
\begin{align*}
 & ~ \|  B_{3,3,2}^{j_1,*,j_0,*} (A) - B_{3,3,2}^{j_1,*,j_0,*} ( \wt{A} ) \|_F \\
    = & ~ \| -  f(A,x)_{j_1}^2 \cdot f(A,x)_{j_0} \cdot c_g(A,x)^{\top} \cdot  ( (A_{j_0,*})^\top \circ x ) \cdot x \cdot {\bf 1}_d^{\top} \\
    & -~  (-  f(\wt{A},x)_{j_1}^2 \cdot f(\wt{A},x)_{j_0} \cdot c_g(\wt{A},x)^{\top} \cdot  ( (\wt{A}_{j_0,*})^\top \circ x ) \cdot x \cdot {\bf 1}_d^{\top} )\|_F \\
    \leq  & ~ \|  f(A,x)_{j_1}^2 \cdot f(A,x)_{j_0} \cdot c_g(A,x)^{\top} \cdot  ( (A_{j_0,*})^\top \circ x ) \cdot x \cdot {\bf 1}_d^{\top} \\
    & -~    f(\wt{A},x)_{j_1}^2 \cdot f(\wt{A},x)_{j_0} \cdot c_g(\wt{A},x)^{\top} \cdot  ( (\wt{A}_{j_0,*})^\top \circ x ) \cdot x \cdot {\bf 1}_d^{\top} \|_F \\
    \leq & ~  |f(A,x)_{j_1} -f(\wt{A},x)_{j_1} | \cdot |f(A,x)_{j_1} |\cdot |f(A,x)_{j_0}| \cdot \|c_g(A,x)^{\top}\|_2 \cdot  \| (A_{j_0,*})^\top\|_2  \|\diag (x)\|_F  \cdot \|x\|_2 \cdot \| {\bf 1}_d^{\top} \|_2 \\
    & + ~ |f(\wt{A},x)_{j_1} | \cdot |f(A,x)_{j_1} - f(\wt{A},x)_{j_1} |\cdot |f(A,x)_{j_0}| \cdot \|c_g(A,x)^{\top}\|_2 \cdot  \| (A_{j_0,*})^\top\|_2  \|\diag (x)\|_F  \cdot \|x\|_2 \cdot \| {\bf 1}_d^{\top} \|_2 \\
    & + ~  |f(\wt{A},x)_{j_1} | \cdot | f(\wt{A},x)_{j_1} |\cdot |f(A,x)_{j_0} -f(\wt{A},x)_{j_0}| \cdot \|c_g(A,x)^{\top}\|_2 \cdot  \| (A_{j_0,*})^\top\|_2  \|\diag (x)\|_F  \cdot \|x\|_2 \cdot \| {\bf 1}_d^{\top} \|_2 \\
    & + ~  |f(\wt{A},x)_{j_1} | \cdot | f(\wt{A},x)_{j_1} |\cdot |f(\wt{A},x)_{j_0}| \cdot \|c_g(A,x)^{\top} - c_g(\wt{A},x)^{\top}\|_2 \cdot  \| (A_{j_0,*})^\top\|_2  \|\diag (x)\|_F  \cdot \|x\|_2 \cdot \| {\bf 1}_d^{\top} \|_2 \\
    & + ~  |f(\wt{A},x)_{j_1} | \cdot | f(\wt{A},x)_{j_1} |\cdot |f(\wt{A},x)_{j_0}| \cdot \|c_g(\wt{A},x)^{\top}\|_2 \cdot  \| (A_{j_0,*})^\top - (\wt{A}_{j_0,*})^\top\|_2  \|\diag (x)\|_F  \cdot \|x\|_2 \cdot \| {\bf 1}_d^{\top} \|_2 \\
\leq & ~ 10R^4 \beta^{-2} \cdot n \cdot \sqrt{d}  \exp(3R^2)\|A - \wt{A}\|_F \\
    & + ~ 10R^4 \beta^{-2} \cdot n \cdot \sqrt{d}  \exp(3R^2)\|A - \wt{A}\|_F \\
    & + ~ 10R^4 \beta^{-2} \cdot n \cdot \sqrt{d}  \exp(3R^2)\|A - \wt{A}\|_F \\
    & + ~ 20R^4 \beta^{-2} \cdot n \cdot \sqrt{d}  \exp(3R^2)\|A - \wt{A}\|_F \\
    & + ~ 5R^3  \sqrt{d} \cdot \|A - \wt{A}\|_F \\
    \leq & ~ 55\beta^{-2} \cdot n \cdot \sqrt{d}  \exp(5R^2)\|A - \wt{A}\|_F \\
    \leq & ~ \beta^{-2} \cdot n \cdot \sqrt{d}  \exp(6R^2)\|A - \wt{A}\|_F 
\end{align*}

{\bf Proof of Part 11.}
\begin{align*}
  & ~ \|  B_{3,4,1}^{j_1,*,j_0,*} (A) - B_{3,4,1}^{j_1,*,j_0,*} ( \wt{A} ) \|_F \\
    = & ~ \| f(A,x)_{j_1} \cdot f(A,x)_{j_0}  \cdot f_c( A,x) \cdot  c_g(A,x)^{\top} \cdot  ( (A_{j_0,*})^\top \circ x ) \cdot x \cdot {\bf 1}_d^{\top} \\
    & ~ - f(\wt{A},x)_{j_1} \cdot f(\wt{A},x)_{j_0}  \cdot f_c( \wt{A},x) \cdot  c_g(\wt{A},x)^{\top} \cdot  ( (\wt{A}_{j_0,*})^\top \circ x ) \cdot x \cdot {\bf 1}_d^{\top} \|_F \\
    \leq & ~ |f(A,x)_{j_1} - f(\wt{A},x)_{j_1}| \cdot |f(A,x)_{j_0} | \cdot |f_c( A,x) |\cdot  \| c_g(A,x)^{\top} \|_2 \cdot   \|(A_{j_0,*})^\top \|_2 \|\diag( x)
    \|_F \cdot \| x \|_2 \cdot \|{\bf 1}_d^{\top} \|_2 \\
    & + ~ |f(\wt{A},x)_{j_1}| \cdot |f(A,x)_{j_0} - f(\wt{A},x)_{j_0} | \cdot |f_c( A,x) |\cdot  \| c_g(A,x)^{\top} \|_2 \cdot   \|(A_{j_0,*})^\top \|_2 \|\diag( x)
    \|_F \cdot \| x \|_2 \cdot \|{\bf 1}_d^{\top} \|_2 \\
    & + ~ |f(\wt{A},x)_{j_1}| \cdot |f(\wt{A},x)_{j_0} | \cdot |f_c( A,x)  -f_c( \wt{A},x)|\cdot  \| c_g(A,x)^{\top} \|_2 \cdot   \|(A_{j_0,*})^\top \|_2 \|\diag( x)
    \|_F \cdot \| x \|_2 \cdot \|{\bf 1}_d^{\top} \|_2 \\
    & + ~ |f(\wt{A},x)_{j_1}| \cdot |f(\wt{A},x)_{j_0} | \cdot |f_c( \wt{A},x)|\cdot  \| c_g(A,x)^{\top}  - c_g(\wt{A},x)^{\top} \|_2 \cdot   \|(A_{j_0,*})^\top \|_2 \|\diag( x)
    \|_F \cdot \| x \|_2 \cdot \|{\bf 1}_d^{\top} \|_2 \\
    & + ~ |f(\wt{A},x)_{j_1}| \cdot |f(\wt{A},x)_{j_0} | \cdot |f_c( \wt{A},x)|\cdot  \| c_g(\wt{A},x)^{\top} \|_2 \cdot   \|(A_{j_0,*})^\top -\wt{A}_{j_0,*})^\top \|_2 \|\diag( x)
    \|_F \cdot \| x \|_2 \cdot \|{\bf 1}_d^{\top} \|_2 \\
    \leq & ~ 20R^4 \beta^{-2} \cdot n \cdot \sqrt{d}  \exp(3R^2)\|A - \wt{A}\|_F \\
    & + ~ 20R^4 \beta^{-2} \cdot n \cdot \sqrt{d}  \exp(3R^2)\|A - \wt{A}\|_F \\
    & + ~ 30R^4 \beta^{-2} \cdot n \cdot \sqrt{d}  \exp(3R^2)\|A - \wt{A}\|_F \\
    & + ~ 40R^4 \beta^{-2} \cdot n \cdot \sqrt{d}  \exp(3R^2)\|A - \wt{A}\|_F \\
    & + ~ 10R^3  \sqrt{d} \cdot \|A - \wt{A}\|_F \\
    \leq & ~ 120\beta^{-2} \cdot n \cdot \sqrt{d}  \exp(5R^2)\|A - \wt{A}\|_F \\
    \leq & ~ \beta^{-2} \cdot n \cdot \sqrt{d}  \exp(6R^2)\|A - \wt{A}\|_F 
\end{align*}

{\bf Proof of Part 12.}
\begin{align*}
 & ~ \|  B_{3,4,2}^{j_1,*,j_0,*} (A) - B_{3,4,2}^{j_1,*,j_0,*} ( \wt{A} ) \|_F \\
    = & ~ \| -  f(A,x)_{j_1}  \cdot f(A,x)_{j_0}\cdot  c(A,x)_{j_1} \cdot c_g(A,x)^{\top} \cdot  ( (A_{j_0,*})^\top \circ x ) \cdot x \cdot {\bf 1}_d^{\top} \\
    & -~  (-  f(\wt{A},x)_{j_1}  \cdot f(\wt{A},x)_{j_0}\cdot  c(\wt{A},x)_{j_1}  \cdot c_g(\wt{A},x)^{\top} \cdot  ( (\wt{A}_{j_0,*})^\top \circ x ) \cdot x \cdot {\bf 1}_d^{\top} )\|_F \\
    \leq  & ~ \|   f(A,x)_{j_1}  \cdot f(A,x)_{j_0}\cdot  c(A,x)_{j_1} \cdot c_g(A,x)^{\top} \cdot  ( (A_{j_0,*})^\top \circ x ) \cdot x \cdot {\bf 1}_d^{\top} \\
    & -~    f(\wt{A},x)_{j_1}  \cdot f(\wt{A},x)_{j_0}\cdot  c(\wt{A},x)_{j_1}  \cdot c_g(\wt{A},x)^{\top} \cdot  ( (\wt{A}_{j_0,*})^\top \circ x ) \cdot x \cdot {\bf 1}_d^{\top} \|_F \\
    \leq & ~  |f(A,x)_{j_1} -f(\wt{A},x)_{j_1} | \cdot |c(A,x)_{j_1} |\cdot |f(A,x)_{j_0}| \cdot \|c_g(A,x)^{\top}\|_2 \cdot  \| (A_{j_0,*})^\top\|_2  \|\diag (x)\|_F  \cdot \|x\|_2 \cdot \| {\bf 1}_d^{\top} \|_2 \\
    & + ~ |f(\wt{A},x)_{j_1} | \cdot |c(A,x)_{j_1} - c(\wt{A},x)_{j_1} |\cdot |f(A,x)_{j_0}| \cdot \|c_g(A,x)^{\top}\|_2 \cdot  \| (A_{j_0,*})^\top\|_2  \|\diag (x)\|_F  \cdot \|x\|_2 \cdot \| {\bf 1}_d^{\top} \|_2 \\
    & + ~  |f(\wt{A},x)_{j_1} | \cdot | c(\wt{A},x)_{j_1} |\cdot |f(A,x)_{j_0} -f(\wt{A},x)_{j_0}| \cdot \|c_g(A,x)^{\top}\|_2 \cdot  \| (A_{j_0,*})^\top\|_2  \|\diag (x)\|_F  \cdot \|x\|_2 \cdot \| {\bf 1}_d^{\top} \|_2 \\
    & + ~  |f(\wt{A},x)_{j_1} | \cdot | c(\wt{A},x)_{j_1} |\cdot |f(\wt{A},x)_{j_0}| \cdot \|c_g(A,x)^{\top} - c_g(\wt{A},x)^{\top}\|_2 \cdot  \| (A_{j_0,*})^\top\|_2  \|\diag (x)\|_F  \cdot \|x\|_2 \cdot \| {\bf 1}_d^{\top} \|_2 \\
    & + ~  |f(\wt{A},x)_{j_1} | \cdot | c(\wt{A},x)_{j_1} |\cdot |f(\wt{A},x)_{j_0}| \cdot \|c_g(\wt{A},x)^{\top}\|_2 \cdot  \| (A_{j_0,*})^\top - (\wt{A}_{j_0,*})^\top\|_2  \|\diag (x)\|_F  \cdot \|x\|_2 \cdot \| {\bf 1}_d^{\top} \|_2 \\
\leq & ~ 20R^4 \beta^{-2} \cdot n \cdot \sqrt{d}  \exp(3R^2)\|A - \wt{A}\|_F \\
    & + ~ 10R^4 \beta^{-2} \cdot n \cdot \sqrt{d}  \exp(3R^2)\|A - \wt{A}\|_F \\
    & + ~ 20R^4 \beta^{-2} \cdot n \cdot \sqrt{d}  \exp(3R^2)\|A - \wt{A}\|_F \\
    & + ~ 40R^4 \beta^{-2} \cdot n \cdot \sqrt{d}  \exp(3R^2)\|A - \wt{A}\|_F \\
    & + ~ 10R^3  \sqrt{d} \cdot \|A - \wt{A}\|_F \\
    \leq & ~ 100\beta^{-2} \cdot n \cdot \sqrt{d}  \exp(5R^2)\|A - \wt{A}\|_F \\
    \leq & ~ \beta^{-2} \cdot n \cdot \sqrt{d}  \exp(6R^2)\|A - \wt{A}\|_F 
\end{align*}

{\bf Proof of Part 13.}
\begin{align*}
 & ~ \|  B_{3,5,1}^{j_1,*,j_0,*} (A) - B_{3,5,1}^{j_1,*,j_0,*} ( \wt{A} ) \|_F \\
    = & ~  0 
\end{align*}

{\bf Proof of Part 14.}
\begin{align*}
 & ~ \| B_{3}^{j_1,*,j_0,*} (A) - B_{3}^{j_1,*,j_0,*} ( \wt{A} ) \|_F \\
   = & ~ \|\sum_{i = 1}^5 B_{3,i}^{j_1,*,j_0,*}(A) -   B_{3,i}^{j_1,*,j_0,*}(\wt{A})  \|_F \\
    \leq & ~ 12  \beta^{-2} \cdot n \cdot \sqrt{d} \cdot \exp(6R^2)\|A - \wt{A}\|_F
\end{align*}
\end{proof}

\subsection{PSD}
\begin{lemma}\label{psd: B_3}
If the following conditions hold
\begin{itemize}
    \item Let $B_{3,1,1}^{j_1,*, j_0,*}, \cdots, B_{3,4,2}^{j_1,*, j_0,*} $ be defined as Lemma~\ref{lem:b_3_j1_j0} 
    \item  Let $\|A \|_2 \leq R, \|A^{\top} \|_F \leq R, \| x\|_2 \leq R, \|\diag(f(A,x)) \|_F \leq \|f(A,x) \|_2 \leq 1, \| b_g \|_2 \leq 1$ 
\end{itemize}
We have 
\begin{itemize}
    \item {\bf Part 1.} $\| B_{3,1,1}^{j_1,*, j_0,*} \| \leq 4 \sqrt{d} R^2$  
    \item {\bf Part 2.} $\|B_{3,1,2}^{j_1,*, j_0,*}\| \preceq 4 \sqrt{d} R^2$
    \item {\bf Part 3.} $\|B_{3,1,3}^{j_1,*, j_0,*}\| \preceq 4   R^4$
    \item {\bf Part 4.} $\|B_{3,1,4}^{j_1,*, j_0,*} \|\preceq 4 R^4$
    \item {\bf Part 5.} $\|B_{3,1,5}^{j_1,*, j_0,*} \|\preceq 4  R^4$
    \item {\bf Part 6.} $\|B_{3,1,6}^{j_1,*, j_0,*} \|\preceq 6   R^4$
    \item {\bf Part 7.} $\|B_{3,1,7}^{j_1,*, j_0,*} \|\preceq 12  R^4$
    \item {\bf Part 8.} $\|B_{3,2,1}^{j_1,*, j_0,*}\| \preceq 10\sqrt{d} R^4$
    \item {\bf Part 9.} $\|B_{3,3,1}^{j_1,*, j_0,*}\| \preceq 5 \sqrt{d}R^4$
    \item {\bf Part 10.} $\|B_{3,3,2}^{j_1,*, j_0,*}\| \preceq 5 \sqrt{d}R^4$
    \item {\bf Part 11.} $\|B_{3,4,1}^{j_1,*, j_0,*}\| \preceq 10\sqrt{d} R^4$
    \item {\bf Part 12.} $\|B_{3,4,2}^{j_1,*, j_0,*}\| \preceq 10\sqrt{d} R^4$
    \item {\bf Part 13.} $\|B_{3}^{j_1,*, j_0,*}\| \preceq 78\sqrt{d} R^4$
\end{itemize}
\end{lemma}

\begin{proof}
    {\bf Proof of Part 1.}
    \begin{align*}
        \| B_{3,1,1}^{j_1,*, j_0,*} \| 
        = & ~ \|  f_c(A,x)^2 \cdot f(A,x)_{j_1} \cdot f(A,x)_{j_0} \cdot  ( (A_{j_0,*})^\top \circ x ) \cdot  {\bf 1}_d^\top\| \\
        \leq & ~ |f(A,x)_{j_1}| \cdot |f(A,x)_{j_0}| \cdot |f_c(A,x)|^2 \cdot \| A_{j_0,*}^\top \circ x \|_2 \cdot \| {\bf 1}_d^\top\|_2\\
        \leq & ~ |f(A,x)_{j_1}| \cdot |f(A,x)_{j_0}| \cdot |f_c(A,x)|^2 \cdot \| A_{j_0,*}\|_2 \cdot \|\diag(x) \|_{\infty} \cdot \| {\bf 1}_d^\top\|_2\\
        \leq & ~ 4 \sqrt{d} R^2
    \end{align*}

    {\bf Proof of Part 2.}
    \begin{align*}
        \| B_{3,1,2}^{j_1,*, j_0,*} \|
        = &~
        \| f(A,x)_{j_1} \cdot f(A,x)_{j_0} \cdot c(A,x)_{j_1} \cdot  f_c(A,x) \cdot ( (A_{j_0,*})^\top \circ x ) \cdot  {\bf 1}_d^\top  \| \\
        \preceq & ~  |f(A,x)_{j_1}| \cdot |f(A,x)_{j_0}| \cdot |c(A,x)_{j_1}|\cdot |f_c(A,x)|\cdot \| A_{j_0,*}^\top \circ x \|_2 \cdot \| {\bf 1}_d^\top\|_2\\
        \preceq & ~  |f(A,x)_{j_1}| \cdot |f(A,x)_{j_0}| \cdot |c(A,x)_{j_1}|\cdot |f_c(A,x)|\cdot \| A_{j_0,*}\|_2 \cdot \|\diag(x) \|_{\infty} \cdot \| {\bf 1}_d^\top\|_2\\
        \preceq & ~ 4 \sqrt{d} R^2
    \end{align*}

    {\bf Proof of Part 3.}
    \begin{align*}
     \| B_{3,1,3}^{j_1,*, j_0,*} \|
     = & ~
       \| f(A,x)_{j_1} \cdot  f(A,x)_{j_0} \cdot f_c(A,x)^2 \cdot  ((A_{j_1,*}) \circ x^\top) \cdot  ((A_{j_0,*})^\top \circ x )  \cdot I_d \| \\
    \leq & ~ |f(A,x)_{j_1}| \cdot |f(A,x)_{j_0}| \cdot  |f_c(A,x)|^2  \cdot \| A_{j_1,*} \circ x^\top \|_2   \cdot \| A_{j_0,*}^\top \circ x \|_2 \cdot \| I_d\| \\
    \leq & ~ |f(A,x)_{j_1}| \cdot |f(A,x)_{j_0}| \cdot  |f_c(A,x)|^2  \cdot \| A_{j_1,*}\|_2 \cdot \|\diag(x) \|_{\infty}   \cdot \| A_{j_0,*}\|_2 \cdot \|\diag(x) \|_{\infty} \cdot \| I_d\|\\
    \leq &  ~  4 R^4
    \end{align*}
   
    {\bf Proof of Part 4.}
    \begin{align*}
        \|B_{3,1,4}^{j_1,*, j_0,*}  \|
        = & ~  \|-f(A,x)_{j_1} \cdot  f(A,x)_{j_0} \cdot f_c(A,x)^2  \cdot h(A,x)^\top \cdot ( (A_{j_0,*})^\top \circ x ) \cdot I_d \|\\
        \leq & ~   \| f(A,x)_{j_1} \cdot  f(A,x)_{j_0} \cdot f_c(A,x)^2  \cdot h(A,x)^\top \cdot ( (A_{j_0,*})^\top \circ x ) \cdot I_d \|\\
        \leq & ~ | f(A,x)_{j_1} | \cdot|  f(A,x)_{j_0} |\cdot |f_c(A,x)|^2 \cdot \|h(A,x)^\top\|_2 \cdot \| A_{j_0,*}^\top \circ x \|_2 \cdot \| I_d\| \\
        \leq & ~ | f(A,x)_{j_1} | \cdot|  f(A,x)_{j_0} |\cdot |f_c(A,x)|^2 \cdot \|h(A,x)^\top\|_2 \cdot \| A_{j_0,*}\|_2 \cdot \|\diag(x) \|_{\infty} \cdot \| I_d\| \\
        \leq & ~ 4  R^4
    \end{align*}

    {\bf Proof of Part 5.}
    \begin{align*}
        \|B_{3,1,5}^{j_1,*, j_0,*} \|
        = & ~ \| f(A,x)_{j_1} \cdot  f(A,x)_{j_0} \cdot (-f_2(A,x) + f(A,x)_{j_1})  \cdot f_c(A,x)  \cdot h(A,x)^\top \cdot ( (A_{j_0,*})^\top \circ x ) \cdot I_d\| \\
        \leq & ~  | f(A,x)_{j_1}| \cdot |f(A,x)_{j_0}|    \cdot (|-f_2(A,x)| + |f(A,x)_{j_1}|) \cdot |f_c(A,x)|  \cdot \|h(A,x)^\top\|_2  \cdot ( (A_{j_0,*})^\top \circ x ) \cdot \| I_d\|\\
        \leq & ~  | f(A,x)_{j_1}| \cdot |f(A,x)_{j_0}|    \cdot (|-f_2(A,x)| + |f(A,x)_{j_1}|) \cdot |f_c(A,x)|\\
         & ~\cdot \|h(A,x)^\top\|_2  \cdot \| A_{j_0,*}\|_2 \cdot \|\diag(x) \|_{\infty}  \cdot \| I_d\| \\
        \leq & ~ 4   R^4
    \end{align*}

    {\bf Proof of Part 6.}
    \begin{align*}
        \|B_{3,1,6}^{j_1,*, j_0,*} \|
        = & ~ \| f(A,x)_{j_1} \cdot  f(A,x)_{j_0} \cdot (-f_c(A,x) + f(A,x)_{j_1}) \cdot f_c(A,x)   \cdot h(A,x)^\top \cdot ( (A_{j_0,*})^\top \circ x ) \cdot I_d \|\\
        \leq & ~  | f(A,x)_{j_1}| \cdot |f(A,x)_{j_0}|    \cdot (|-f_c(A,x)| + |f(A,x)_{j_1}|)  \cdot |f_c(A,x)|  \cdot \|h(A,x)^\top\|_2  \cdot ( (A_{j_0,*})^\top \circ x ) \cdot \| I_d\| \\
        \leq & ~  | f(A,x)_{j_1}| \cdot |f(A,x)_{j_0}|    \cdot (|-f_c(A,x)| + |f(A,x)_{j_1}|)  \cdot |f_c(A,x)|  \cdot \|h(A,x)^\top\|_2\\
        & ~\cdot  \| A_{j_0,*}\|_2 \cdot \|\diag(x) \|_{\infty} \cdot \| I_d\| \\
        \leq & ~ 6  R^4
    \end{align*}

    {\bf Proof of Part 7.}
    \begin{align*}
         \|B_{3,1,7}^{j_1,*, j_0,*} \|
         = & ~ \|f(A,x)_{j_1} \cdot  f(A,x)_{j_0} \cdot f_c(A,x)   \cdot p_{j_1}(A,x)^\top \cdot  ( (A_{j_0,*})^\top \circ x )\cdot I_d \|\\
         \leq & ~ | f(A,x)_{j_1}| \cdot |f(A,x)_{j_0}| \cdot |f_c(A,x)|  \|p_{j_1}(A,x)^\top\|_2  \cdot ( (A_{j_0,*})^\top \circ x ) \cdot \| I_d\|\\
         \leq & ~ | f(A,x)_{j_1}| \cdot |f(A,x)_{j_0}| \cdot |f_c(A,x)|  \|p_{j_1}(A,x)^\top\|_2  \cdot \| A_{j_0,*}\|_2 \cdot \|\diag(x) \|_{\infty} \cdot \| I_d\| \\
         \leq & ~ 12   R^4
    \end{align*}

    {\bf Proof of Part 8.}
    \begin{align*}
        \|B_{3,2,1}^{j_1,*, j_0,*} \|
        = & ~ \|  f(A,x)_{j_1} \cdot  f(A,x)_{j_0} \cdot f_c(A,x) \cdot c_g(A,x)^{\top} \cdot  ( (A_{j_0,*})^\top \circ x ) \cdot x \cdot {\bf 1}_d^{\top} \| \\
        \leq & ~  | f(A,x)_{j_1}| \cdot |f(A,x)_{j_0}| \cdot |f_c(A,x)| \cdot \|c_g(A,x)^{\top}\|_2 \cdot ( (A_{j_0,*})^\top \circ x ) \cdot \| x\|_2 \cdot \| {\bf 1}_d^{\top}\|_2\\
        \leq & ~  | f(A,x)_{j_1}| \cdot |f(A,x)_{j_0}| \cdot |f_c(A,x)| \cdot \|c_g(A,x)^{\top}\|_2 \cdot \| A_{j_0,*}\|_2 \cdot \|\diag(x) \|_{\infty} \cdot \| x\|_2 \cdot \| {\bf 1}_d^{\top}\|_2\\
        \leq & ~ 10 \sqrt{d}R^4
    \end{align*}

    {\bf Proof of Part 9.}
    \begin{align*}
     \|B_{3,3,1}^{j_1,*, j_0,*} \|
        = & ~ \|  f(A,x)_{j_1} \cdot f(A,x)_{j_0}  \cdot f_2( A,x) \cdot  c_g(A,x)^{\top} \cdot  ( (A_{j_0,*})^\top \circ x ) \cdot x \cdot {\bf 1}_d^{\top}\| \\
        \leq & ~ | f(A,x)_{j_1}| \cdot  |f(A,x)_{j_0} | \cdot  |f_2( A,x) | \cdot \|c_g(A,x)^{\top}\|_2 \cdot ( (A_{j_0,*})^\top \circ x ) \cdot \| x\|_2 \cdot \| {\bf 1}_d^{\top}\|_2\\
        \leq & ~ | f(A,x)_{j_1}| \cdot  |f(A,x)_{j_0} | \cdot  |f_2( A,x) | \cdot \|c_g(A,x)^{\top}\|_2 \cdot \| A_{j_0,*}\|_2 \cdot \|\diag(x) \|_{\infty} \cdot \| x\|_2 \cdot \| {\bf 1}_d^{\top}\|_2\\
        \leq & ~ 5 \sqrt{d}R^4
    \end{align*}

       {\bf Proof of Part 10.}
    \begin{align*}
     \|B_{3,3,2}^{j_1,*, j_0,*} \|
        = & ~ \|   -  f(A,x)_{j_1}^2 \cdot f(A,x)_{j_0} \cdot c_g(A,x)^{\top} \cdot  ( (A_{j_0,*})^\top \circ x ) \cdot x \cdot {\bf 1}_d^{\top} \| \\
        \leq& ~ \|   f(A,x)_{j_1}^2 \cdot f(A,x)_{j_0} \cdot c_g(A,x)^{\top} \cdot  ( (A_{j_0,*})^\top \circ x ) \cdot x \cdot {\bf 1}_d^{\top} \| \\
        \leq & ~ | f(A,x)_{j_1}|^2 \cdot  |f(A,x)_{j_0} |   \cdot \|c_g(A,x)^{\top}\|_2 \cdot ( (A_{j_0,*})^\top \circ x ) \cdot \| x\|_2 \cdot \| {\bf 1}_d^{\top}\|_2\\
        \leq & ~ | f(A,x)_{j_1}|^2 \cdot  |f(A,x)_{j_0} | \cdot \|c_g(A,x)^{\top}\|_2 \cdot \| A_{j_0,*}\|_2 \cdot \|\diag(x) \|_{\infty} \cdot \| x\|_2 \cdot \| {\bf 1}_d^{\top}\|_2\\
        \leq & ~ 5\sqrt{d} R^4
    \end{align*}

        {\bf Proof of Part 11.}
    \begin{align*}
     \|B_{3,4,1}^{j_1,*, j_0,*} \|
        = & ~ \|  f(A,x)_{j_1} \cdot f(A,x)_{j_0}  \cdot f_c( A,x) \cdot  c_g(A,x)^{\top} \cdot  ( (A_{j_0,*})^\top \circ x ) \cdot x \cdot {\bf 1}_d^{\top}\| \\
        \leq & ~ | f(A,x)_{j_1}| \cdot  |f(A,x)_{j_0} | \cdot  |f_c( A,x) | \cdot \|c_g(A,x)^{\top}\|_2 \cdot ( (A_{j_0,*})^\top \circ x ) \cdot \| x\|_2 \cdot \| {\bf 1}_d^{\top}\|_2\\
        \leq & ~ | f(A,x)_{j_1}| \cdot  |f(A,x)_{j_0} | \cdot  |f_c( A,x) | \cdot \|c_g(A,x)^{\top}\|_2 \cdot \| A_{j_0,*}\|_2 \cdot \|\diag(x) \|_{\infty} \cdot \| x\|_2 \cdot \| {\bf 1}_d^{\top}\|_2\\
        \leq & ~ 10\sqrt{d} R^4
    \end{align*}

       {\bf Proof of Part 12.}
    \begin{align*}
     \|B_{3,4,2}^{j_1,*, j_0,*} \|
        = & ~ \|   -  f(A,x)_{j_1}  \cdot f(A,x)_{j_0}\cdot c(A,x)_{j_1} \cdot c_g(A,x)^{\top} \cdot  ( (A_{j_0,*})^\top \circ x ) \cdot x \cdot {\bf 1}_d^{\top} \| \\
        \leq& ~ \|   f(A,x)_{j_1}  \cdot f(A,x)_{j_0}\cdot c(A,x)_{j_1} \cdot c_g(A,x)^{\top} \cdot  ( (A_{j_0,*})^\top \circ x ) \cdot x \cdot {\bf 1}_d^{\top} \| \\
        \leq & ~ | f(A,x)_{j_1}|  \cdot  |f(A,x)_{j_0} | \cdot  |c(A,x)_{j_1} |  \cdot \|c_g(A,x)^{\top}\|_2 \cdot ( (A_{j_0,*})^\top \circ x ) \cdot \| x\|_2 \cdot \| {\bf 1}_d^{\top}\|_2\\
        \leq & ~ | f(A,x)_{j_1}|  \cdot  |f(A,x)_{j_0} |\cdot  |c(A,x)_{j_1} | \cdot \|c_g(A,x)^{\top}\|_2 \cdot \| A_{j_0,*}\|_2 \cdot \|\diag(x) \|_{\infty} \cdot \| x\|_2 \cdot \| {\bf 1}_d^{\top}\|_2\\
        \leq & ~ 10\sqrt{d} R^4
    \end{align*}

        {\bf Proof of Part 10}
\begin{align*}
    & ~ \| B_{3}^{j_1,*,j_0,*} \|  \\
    = & ~ \|\sum_{i = 1}^4 B_{3,i}^{j_1,*,j_0,*}  \|  \\
    \leq & ~ 78\sqrt{d} R^4
\end{align*}
\end{proof}

\section{Hessian: Fourth term  \texorpdfstring{$B_{4}^{j_1,i_1,j_0,i_0}$}{}}\label{app:hessian_fouth}
\subsection{Definitions}
\begin{definition} \label{def:b_4}
    We define the $B_4^{j_1,i_1,j_0,i_0}$ as follows,
    \begin{align*}
        B_4^{j_1,i_1,j_0,i_0} := & ~ \frac{\d}{\d A_{j_1,i_1}}  (c_g(A,x)^{\top} \cdot f(A,x)_{j_1} \cdot \diag (x) A^{\top} \cdot  f(A,x)  \cdot    \langle c(A, x), f(A, x) \rangle ) 
    \end{align*}
    Then, we define $B_{4,1}^{j_1,i_1,j_0,i_0}, \cdots, B_{4,4}^{j_1,i_1,j_0,i_0}$ as follow
    \begin{align*}
         B_{4,1}^{j_1,i_1,j_0,i_0} : = & ~ \frac{\d}{\d A_{j_1,i_1}} (- c_g(A,x)^{\top} ) \cdot  f(A,x)_{j_0} \cdot \diag (x) A^{\top} \cdot  f(A,x)  \cdot    \langle c(A, x), f(A, x) \rangle\\
 B_{4,2}^{j_1,i_1,j_0,i_0}: = & ~ - c_g(A,x)^{\top} \cdot \frac{\d}{\d A_{j_1,i_1}} ( f(A,x)_{j_0} )  \cdot \diag (x) A^{\top} \cdot  f(A,x)  \cdot    \langle c(A, x), f(A, x) \rangle\\
 B_{4,3}^{j_1,i_1,j_0,i_0}: = & ~ - c_g(A,x)^{\top} \cdot f(A,x)_{j_0} \cdot \diag (x) \cdot \frac{\d}{\d A_{j_1,i_1}} ( A^{\top})  \cdot  f(A,x)  \cdot    \langle c(A, x), f(A, x) \rangle\\
  B_{4,4}^{j_1,i_1,j_0,i_0}: = & ~ - c_g(A,x)^{\top} \cdot f(A,x)_{j_0} \cdot \diag (x) A^{\top} \cdot  \frac{\d}{\d A_{j_1,i_1}} ( f(A,x)  )  \cdot    \langle c(A, x), f(A, x) \rangle\\
   B_{4,5}^{j_1,i_1,j_0,i_0}: = & ~  -  c_g(A,x)^{\top} \cdot f(A,x)_{j_0}  \cdot \diag (x) A^{\top} \cdot  f(A,x)  \cdot  \langle  \frac{\d c(A,x)}{\d A_{j_1,i_1}} , f(A,x) \rangle \\
     B_{4,6}^{j_1,i_1,j_0,i_0}: = & ~  -  c_g(A,x)^{\top} \cdot f(A,x)_{j_0}  \cdot \diag (x) A^{\top} \cdot  f(A,x)  \cdot \langle c(A,x), \frac{\d  f(A,x) }{\d A_{j_1,i_1}}  \rangle
    \end{align*}
    It is easy to show
    \begin{align*}
        B_4^{j_1,i_1,j_0,i_0} = B_{4,1}^{j_1,i_1,j_0,i_0} +  B_{4,2}^{j_1,i_1,j_0,i_0} + B_{4,3}^{j_1,i_1,j_0,i_0} + B_{4,4}^{j_1,i_1,j_0,i_0} + B_{4,5}^{j_1,i_1,j_0,i_0} +B_{4,6}^{j_1,i_1,j_0,i_0}
    \end{align*}
        Similarly for $j_1 = j_0$ and $i_0 = i_1$,we have
    \begin{align*}
        B_4^{j_1,i_1,j_1,i_1} = B_{4,1}^{j_1,i_1,j_1,i_1} +  B_{4,2}^{j_1,i_1,j_1,i_1} + B_{4,3}^{j_1,i_1,j_1,i_1}  + B_{4,4}^{j_1,i_1,j_1,i_1} + B_{4,5}^{j_1,i_1,j_1,i_1} +B_{4,6}^{j_1,i_1,j_1,i_1}
         \end{align*}
    For $j_1 = j_0$ and $i_0 \neq i_1$,we have
    \begin{align*}
        B_4^{j_1,i_1,j_1,i_0} = B_{4,1}^{j_1,i_1,j_1,i_0} +  B_{4,2}^{j_1,i_1,j_1,i_0} + B_{4,3}^{j_1,i_1,j_1,i_0} + B_{4,4}^{j_1,i_1,j_1,i_0} + B_{4,5}^{j_1,i_1,j_1,i_0} +B_{4,6}^{j_1,i_1,j_1,i_0}
    \end{align*}
    For $j_1 \neq j_0$ and $i_0 = i_1$,we have
    \begin{align*}
        B_4^{j_1,i_1,j_0,i_1} = B_{4,1}^{j_1,i_1,j_0,i_1} +  B_{4,2}^{j_1,i_1,j_0,i_1} + B_{4,3}^{j_1,i_1,j_0,i_1} + B_{4,4}^{j_1,i_1,j_0,i_1} + B_{4,5}^{j_1,i_1,j_0,i_1} +B_{4,6}^{j_1,i_1,j_0,i_1}
    \end{align*}
\end{definition}
\subsection{Case \texorpdfstring{$j_1=j_0, i_1 = i_0$}{}}
\begin{lemma}
For $j_1 = j_0$ and $i_0 = i_1$. If the following conditions hold
    \begin{itemize}
     \item Let $u(A,x) \in \R^n$ be defined as Definition~\ref{def:u}
    \item Let $\alpha(A,x) \in \R$ be defined as Definition~\ref{def:alpha}
     \item Let $f(A,x) \in \R^n$ be defined as Definition~\ref{def:f}
    \item Let $c(A,x) \in \R^n$ be defined as Definition~\ref{def:c}
    \item Let $g(A,x) \in \R^d$ be defined as Definition~\ref{def:g} 
    \item Let $q(A,x) = c(A,x) + f(A,x) \in \R^n$
    \item Let $c_g(A,x) \in \R^d$ be defined as Definition~\ref{def:c_g}.
    \item Let $L_g(A,x) \in \R$ be defined as Definition~\ref{def:l_g}
    \item Let $v \in \R^n$ be a vector 
    \item Let $B_1^{j_1,i_1,j_0,i_0}$ be defined as Definition~\ref{def:b_1}
    \end{itemize}
    Then, For $j_0,j_1 \in [n], i_0,i_1 \in [d]$, we have 
    \begin{itemize}
\item {\bf Part 1.} For $B_{4,1}^{j_1,i_1,j_1,i_1}$, we have 
\begin{align*}
 B_{4,1}^{j_1,i_1,j_1,i_1}  = & ~  \frac{\d}{\d A_{j_1,i_1}} (- c_g(A,x)^{\top} ) \cdot  f(A,x)_{j_1} \cdot \diag (x) A^{\top} \cdot  f(A,x)  \cdot    \langle c(A, x), f(A, x) \rangle \\
 = & ~ B_{4,1,1}^{j_1,i_1,j_1,i_1} + B_{4,1,2}^{j_1,i_1,j_1,i_1} + B_{4,1,3}^{j_1,i_1,j_1,i_1} + B_{4,1,4}^{j_1,i_1,j_1,i_1} + B_{4,1,5}^{j_1,i_1,j_1,i_1} + B_{4,1,6}^{j_1,i_1,j_1,i_1} + B_{4,1,7}^{j_1,i_1,j_1,i_1}
\end{align*} 
\item {\bf Part 2.} For $B_{4,2}^{j_1,i_1,j_1,i_1}$, we have 
\begin{align*}
  B_{4,2}^{j_1,i_1,j_1,i_1} = & ~  - c_g(A,x)^{\top} \cdot \frac{\d}{\d A_{j_1,i_1}} ( f(A,x)_{j_1} )  \cdot \diag (x) A^{\top} \cdot  f(A,x)  \cdot    \langle c(A, x), f(A, x) \rangle  \\
    = & ~  B_{4,2,1}^{j_1,i_1,j_1,i_1} + B_{4,2,2}^{j_1,i_1,j_1,i_1}
\end{align*} 
\item {\bf Part 3.} For $B_{4,3}^{j_1,i_1,j_1,i_1}$, we have 
\begin{align*}
  B_{4,3}^{j_1,i_1,j_1,i_1} = & ~ - c_g(A,x)^{\top} \cdot f(A,x)_{j_1} \cdot \diag (x) \cdot \frac{\d}{\d A_{j_1,i_1}} ( A^{\top})  \cdot  f(A,x)  \cdot    \langle c(A, x), f(A, x) \rangle \\
     = & ~ B_{4,3,1}^{j_1,i_1,j_1,i_1}  
\end{align*} 
\item {\bf Part 4.} For $B_{4,4}^{j_1,i_1,j_1,i_1}$, we have 
\begin{align*}
  B_{4,4}^{j_1,i_1,j_1,i_1} = & ~ - c_g(A,x)^{\top} \cdot f(A,x)_{j_1} \cdot \diag (x) A^{\top} \cdot  \frac{\d}{\d A_{j_1,i_1}} ( f(A,x)  )  \cdot    \langle c(A, x), f(A, x) \rangle\\
     = & ~ B_{4,4,1}^{j_1,i_1,j_1,i_1} 
\end{align*}
\item {\bf Part 5.} For $B_{4,5}^{j_1,i_1,j_1,i_1}$, we have 
\begin{align*}
  B_{4,5}^{j_1,i_1,j_1,i_1} = & ~ -  c_g(A,x)^{\top} \cdot f(A,x)_{j_1}  \cdot \diag (x) A^{\top} \cdot  f(A,x)  \cdot  \langle \frac{\d c(A,x)}{\d A_{j_1,i_1}}, f(A,x) \rangle  \\
     = & ~ B_{4,5,1}^{j_1,i_1,j_1,i_1}  + B_{4,5,2}^{j_1,i_1,j_1,i_1}  
\end{align*}
\item {\bf Part 6.} For $B_{4,6}^{j_1,i_1,j_1,i_1}$, we have 
\begin{align*}
  B_{4,6}^{j_1,i_1,j_1,i_1} = & ~ -  c_g(A,x)^{\top} \cdot f(A,x)_{j_1}  \cdot \diag (x) A^{\top} \cdot  f(A,x)  \cdot  \langle c(A,x),  \frac{\d f(A,x)}{\d A_{j_1,i_1}} \rangle  \\
     = & ~ B_{4,6,1}^{j_1,i_1,j_1,i_1}  + B_{4,6,2}^{j_1,i_1,j_1,i_1}  
\end{align*}
\end{itemize}
\begin{proof}
    {\bf Proof of Part 1.}
    \begin{align*}
    B_{4,1,1}^{j_1,i_1,j_1,i_1} : = & ~ e_{i_1}^\top  \cdot  (\langle c(A,x), f(A,x) \rangle)^2  \cdot   f(A,x)_{j_1}^2 \cdot \diag (x) A^{\top} \cdot  f(A,x) \\
    B_{4,1,2}^{j_1,i_1,j_1,i_1} : = & ~ e_{i_1}^\top  \cdot  c(A,x)_{j_1}  \cdot  f(A,x)_{j_1}^2  \cdot \diag (x) A^{\top} \cdot  f(A,x)  \cdot    \langle c(A, x), f(A, x) \rangle\\
    B_{4,1,3}^{j_1,i_1,j_1,i_1} : = & ~f(A,x)_{j_1}^2  \cdot  (\langle c(A,x), f(A,x) \rangle)^2  \cdot (  (A_{j_1,*})  \circ  x^\top  ) \cdot \diag (x) A^{\top} \cdot  f(A,x)  \\
    B_{4,1,4}^{j_1,i_1,j_1,i_1} : = & ~  -   f(A,x)_{j_1}^2  \cdot  f(A,x)^\top   \cdot  A  \cdot   (\diag(x))^2  \cdot  (\langle c(A,x), f(A,x) \rangle)^2 \cdot   A^{\top} \cdot  f(A,x) \\
    B_{4,1,5}^{j_1,i_1,j_1,i_1} : = & ~    f(A,x)_{j_1}^2  \cdot f(A,x)^\top \cdot A \cdot (\diag(x))^2 \cdot (\langle -f(A,x), f(A,x) \rangle + f(A,x)_{j_1}) \cdot A^{\top} \cdot  f(A,x)  \\
    & ~ \cdot    \langle c(A, x), f(A, x) \rangle\\
    B_{4,1,6}^{j_1,i_1,j_1,i_1} : = & ~  f(A,x)_{j_1}^2 \cdot f(A,x)^\top  \cdot A \cdot (\diag(x))^2 \cdot(\langle -f(A,x), c(A,x) \rangle + f(A,x)_{j_1}) \cdot   A^{\top} \cdot  f(A,x) \\
    & ~ \cdot    \langle c(A, x), f(A, x) \rangle\\
    B_{4,1,7}^{j_1,i_1,j_1,i_1} : = & ~ f(A,x)_{j_1}^2 \cdot (( e_{j_1}^\top  - f(A,x)^\top) \circ q(A,x)^\top) \cdot A \cdot  (\diag(x))^2 \cdot A^{\top} \cdot  f(A,x)  \cdot    \langle c(A, x), f(A, x) \rangle
\end{align*}
Finally, combine them and we have
\begin{align*}
       B_{4,1}^{j_1,i_1,j_1,i_1} = B_{4,1,1}^{j_1,i_1,j_1,i_1} + B_{4,1,2}^{j_1,i_1,j_1,i_1} + B_{4,1,3}^{j_1,i_1,j_1,i_1} + B_{4,1,4}^{j_1,i_1,j_1,i_1} + B_{4,1,5}^{j_1,i_1,j_1,i_1} + B_{4,1,6}^{j_1,i_1,j_1,i_1} + B_{4,1,7}^{j_1,i_1,j_1,i_1}
\end{align*}
{\bf Proof of Part 2.}
    \begin{align*}
    B_{4,2,1}^{j_1,i_1,j_1,i_1} : = & ~   c_g(A,x)^{\top} \cdot f(A,x)_{j_1}^2 \cdot x_{i_1} \cdot \diag (x) A^{\top} \cdot  f(A,x)  \cdot    \langle c(A, x), f(A, x) \rangle \\
    B_{4,2,2}^{j_1,i_1,j_1,i_1} : = & ~ - c_g(A,x)^{\top} \cdot f(A,x)_{j_1} \cdot x_{i_1} \cdot \diag (x) A^{\top} \cdot  f(A,x)  \cdot    \langle c(A, x), f(A, x) \rangle
\end{align*}
Finally, combine them and we have
\begin{align*}
       B_{4,2}^{j_1,i_1,j_1,i_1} = B_{4,2,1}^{j_1,i_1,j_1,i_1} + B_{4,2,2}^{j_1,i_1,j_1,i_1}
\end{align*}
{\bf Proof of Part 3.} 
    \begin{align*}
    B_{4,3,1}^{j_1,i_1,j_1,i_1} : = & ~    - c_g(A,x)^{\top} \cdot f(A,x)_{j_1} \cdot \diag (x) \cdot e_{i_1} \cdot  e_{j_1}^\top \cdot  f(A,x)  \cdot    \langle c(A, x), f(A, x) \rangle
\end{align*}
Finally, combine them and we have
\begin{align*}
       B_{4,3}^{j_1,i_1,j_1,i_1} = B_{4,3,1}^{j_1,i_1,j_1,i_1} 
\end{align*}
{\bf Proof of Part 4.} 
    \begin{align*}
    B_{4,4,1}^{j_1,i_1,j_1,i_1} : = & ~  - c_g(A,x)^{\top} \cdot f(A,x)_{j_1}^2  \cdot \diag (x) \cdot A^{\top} \cdot x_{i_1}   \cdot (e_{j_1}- f(A,x) ) \cdot    \langle c(A, x), f(A, x) \rangle
\end{align*}
Finally, combine them and we have
\begin{align*}
       B_{4,4}^{j_1,i_1,j_1,i_1} = B_{4,4,1}^{j_1,i_1,j_1,i_1}  
\end{align*}
{\bf Proof of Part 5.} 
    \begin{align*}
    B_{4,5,1}^{j_1,i_1,j_1,i_1} : = & ~ -  c_g(A,x)^{\top} \cdot f(A,x)_{j_1}^2  \cdot \diag (x) A^{\top} \cdot  f(A,x)  \cdot x_{i_1} \cdot \langle - f(A,x), f(A,x) \rangle\\
    B_{4,5,2}^{j_1,i_1,j_1,i_1} : = & ~ -  c_g(A,x)^{\top} \cdot f(A,x)_{j_1}^3  \cdot \diag (x) A^{\top} \cdot  f(A,x)  \cdot  x_{i_1}
\end{align*}
Finally, combine them and we have
\begin{align*}
       B_{4,5}^{j_1,i_1,j_1,i_1} = B_{4,5,1}^{j_1,i_1,j_1,i_1}  +B_{4,5,2}^{j_1,i_1,j_1,i_1}
\end{align*}
{\bf Proof of Part 6.} 
    \begin{align*}
    B_{4,6,1}^{j_1,i_1,j_1,i_1} : = & ~ -  c_g(A,x)^{\top} \cdot f(A,x)_{j_1}^2  \cdot \diag (x) A^{\top} \cdot  f(A,x)  \cdot x_{i_1} \cdot (\langle - f(A,x), c(A,x) \rangle )\\
    B_{4,6,2}^{j_1,i_1,j_1,i_1} : = & ~ -  c_g(A,x)^{\top} \cdot f(A,x)_{j_1}^2  \cdot \diag (x) A^{\top} \cdot  f(A,x)  \cdot   x_{i_1} \cdot c(A,x)_{j_1} 
\end{align*}
Finally, combine them and we have
\begin{align*}
       B_{4,6}^{j_1,i_1,j_1,i_1} = B_{4,6,1}^{j_1,i_1,j_1,i_1}  +B_{4,6,2}^{j_1,i_1,j_1,i_1}
\end{align*}
\end{proof}
\end{lemma}

\subsection{Case \texorpdfstring{$j_1=j_0, i_1 \neq i_0$}{}}
\begin{lemma}
For $j_1 = j_0$ and $i_0 \neq i_1$. If the following conditions hold
    \begin{itemize}
     \item Let $u(A,x) \in \R^n$ be defined as Definition~\ref{def:u}
    \item Let $\alpha(A,x) \in \R$ be defined as Definition~\ref{def:alpha}
     \item Let $f(A,x) \in \R^n$ be defined as Definition~\ref{def:f}
    \item Let $c(A,x) \in \R^n$ be defined as Definition~\ref{def:c}
    \item Let $g(A,x) \in \R^d$ be defined as Definition~\ref{def:g} 
    \item Let $q(A,x) = c(A,x) + f(A,x) \in \R^n$
    \item Let $c_g(A,x) \in \R^d$ be defined as Definition~\ref{def:c_g}.
    \item Let $L_g(A,x) \in \R$ be defined as Definition~\ref{def:l_g}
    \item Let $v \in \R^n$ be a vector 
    \item Let $B_1^{j_1,i_1,j_0,i_0}$ be defined as Definition~\ref{def:b_1}
    \end{itemize}
    Then, For $j_0,j_1 \in [n], i_0,i_1 \in [d]$, we have 
    \begin{itemize}
\item {\bf Part 1.} For $B_{4,1}^{j_1,i_1,j_1,i_0}$, we have 
\begin{align*}
 B_{4,1}^{j_1,i_1,j_1,i_0}  = & ~  \frac{\d}{\d A_{j_1,i_1}} (- c_g(A,x)^{\top} ) \cdot  f(A,x)_{j_1} \cdot \diag (x) A^{\top} \cdot  f(A,x)  \cdot    \langle c(A, x), f(A, x) \rangle \\
 = & ~ B_{4,1,1}^{j_1,i_1,j_1,i_0} + B_{4,1,2}^{j_1,i_1,j_1,i_0} + B_{4,1,3}^{j_1,i_1,j_1,i_0} + B_{4,1,4}^{j_1,i_1,j_1,i_0} + B_{4,1,5}^{j_1,i_1,j_1,i_0} + B_{4,1,6}^{j_1,i_1,j_1,i_0} + B_{4,1,7}^{j_1,i_1,j_1,i_0}
\end{align*} 
\item {\bf Part 2.} For $B_{4,2}^{j_1,i_1,j_1,i_0}$, we have 
\begin{align*}
  B_{4,2}^{j_1,i_1,j_1,i_0} = & ~  - c_g(A,x)^{\top} \cdot \frac{\d}{\d A_{j_1,i_1}} ( f(A,x)_{j_1} )  \cdot \diag (x) A^{\top} \cdot  f(A,x)  \cdot    \langle c(A, x), f(A, x) \rangle  \\
    = & ~  B_{4,2,1}^{j_1,i_1,j_1,i_0} + B_{4,2,2}^{j_1,i_1,j_1,i_0}
\end{align*} 
\item {\bf Part 3.} For $B_{4,3}^{j_1,i_1,j_1,i_0}$, we have 
\begin{align*}
  B_{4,3}^{j_1,i_1,j_1,i_0} = & ~ - c_g(A,x)^{\top} \cdot f(A,x)_{j_1} \cdot \diag (x) \cdot \frac{\d}{\d A_{j_1,i_1}} ( A^{\top})  \cdot  f(A,x)  \cdot    \langle c(A, x), f(A, x) \rangle \\
     = & ~ B_{4,3,1}^{j_1,i_1,j_1,i_0}  
\end{align*} 
\item {\bf Part 4.} For $B_{4,4}^{j_1,i_1,j_1,i_0}$, we have 
\begin{align*}
  B_{4,4}^{j_1,i_1,j_1,i_0} = & ~ - c_g(A,x)^{\top} \cdot f(A,x)_{j_1} \cdot \diag (x) A^{\top} \cdot  \frac{\d}{\d A_{j_1,i_1}} ( f(A,x)  )  \cdot    \langle c(A, x), f(A, x) \rangle\\
     = & ~ B_{4,4,1}^{j_1,i_1,j_1,i_0} 
\end{align*}
\item {\bf Part 5.} For $B_{4,5}^{j_1,i_1,j_1,i_0}$, we have 
\begin{align*}
  B_{4,5}^{j_1,i_1,j_1,i_0} = & ~ -  c_g(A,x)^{\top} \cdot f(A,x)_{j_1}  \cdot \diag (x) A^{\top} \cdot  f(A,x)  \cdot  \langle \frac{\d c(A,x)}{\d A_{j_1,i_1}}, f(A,x) \rangle  \\
     = & ~ B_{4,5,1}^{j_1,i_1,j_1,i_0}  + B_{4,5,2}^{j_1,i_1,j_1,i_0}  
\end{align*}
\item {\bf Part 6.} For $B_{4,6}^{j_1,i_1,j_1,i_0}$, we have 
\begin{align*}
  B_{4,6}^{j_1,i_1,j_1,i_0} = & ~ -  c_g(A,x)^{\top} \cdot f(A,x)_{j_1}  \cdot \diag (x) A^{\top} \cdot  f(A,x)  \cdot  \langle c(A,x),  \frac{\d f(A,x)}{\d A_{j_1,i_1}} \rangle  \\
     = & ~ B_{4,6,1}^{j_1,i_1,j_1,i_0}  + B_{4,6,2}^{j_1,i_1,j_1,i_0}  
\end{align*}
\end{itemize}
\begin{proof}
    {\bf Proof of Part 1.}
    \begin{align*}
    B_{4,1,1}^{j_1,i_1,j_1,i_0} : = & ~ e_{i_1}^\top  \cdot  (\langle c(A,x), f(A,x) \rangle)^2  \cdot   f(A,x)_{j_1}^2 \cdot \diag (x) A^{\top} \cdot  f(A,x) \\
    B_{4,1,2}^{j_1,i_1,j_1,i_0} : = & ~ e_{i_1}^\top  \cdot  c(A,x)_{j_1}  \cdot  f(A,x)_{j_1}^2  \cdot \diag (x) A^{\top} \cdot  f(A,x)  \cdot    \langle c(A, x), f(A, x) \rangle\\
    B_{4,1,3}^{j_1,i_1,j_1,i_0} : = & ~f(A,x)_{j_1}^2  \cdot  (\langle c(A,x), f(A,x) \rangle)^2  \cdot (  (A_{j_1,*})  \circ  x^\top  ) \cdot \diag (x) A^{\top} \cdot  f(A,x)  \\
    B_{4,1,4}^{j_1,i_1,j_1,i_0} : = & ~  -   f(A,x)_{j_1}^2  \cdot  f(A,x)^\top   \cdot  A  \cdot   (\diag(x))^2  \cdot  (\langle c(A,x), f(A,x) \rangle)^2 \cdot   A^{\top} \cdot  f(A,x) \\
    B_{4,1,5}^{j_1,i_1,j_1,i_0} : = & ~    f(A,x)_{j_1}^2  \cdot f(A,x)^\top \cdot A \cdot (\diag(x))^2 \cdot (\langle -f(A,x), f(A,x) \rangle + f(A,x)_{j_1}) \cdot A^{\top} \cdot  f(A,x) \\
    & ~ \cdot    \langle c(A, x), f(A, x) \rangle\\
    B_{4,1,6}^{j_1,i_1,j_1,i_0} : = & ~  f(A,x)_{j_1}^2 \cdot f(A,x)^\top  \cdot A \cdot (\diag(x))^2 \cdot(\langle -f(A,x), c(A,x) \rangle + f(A,x)_{j_1}) \cdot   A^{\top} \cdot  f(A,x)  \\
    & ~ \cdot    \langle c(A, x), f(A, x) \rangle\\
    B_{4,1,7}^{j_1,i_1,j_1,i_0} : = & ~ f(A,x)_{j_1}^2 \cdot (( e_{j_1}^\top  - f(A,x)^\top) \circ q(A,x)^\top) \cdot A \cdot  (\diag(x))^2 \cdot A^{\top} \cdot  f(A,x)  \cdot    \langle c(A, x), f(A, x) \rangle
\end{align*}
Finally, combine them and we have
\begin{align*}
       B_{4,1}^{j_1,i_1,j_1,i_0} = B_{4,1,1}^{j_1,i_1,j_1,i_0} + B_{4,1,2}^{j_1,i_1,j_1,i_0} + B_{4,1,3}^{j_1,i_1,j_1,i_0} + B_{4,1,4}^{j_1,i_1,j_1,i_0} + B_{4,1,5}^{j_1,i_1,j_1,i_0} + B_{4,1,6}^{j_1,i_1,j_1,i_0} + B_{4,1,7}^{j_1,i_1,j_1,i_0}
\end{align*}
{\bf Proof of Part 2.}
    \begin{align*}
    B_{4,2,1}^{j_1,i_1,j_1,i_0} : = & ~   c_g(A,x)^{\top} \cdot f(A,x)_{j_1}^2 \cdot x_{i_1} \cdot \diag (x) A^{\top} \cdot  f(A,x)  \cdot    \langle c(A, x), f(A, x) \rangle \\
    B_{4,2,2}^{j_1,i_1,j_1,i_0} : = & ~ - c_g(A,x)^{\top} \cdot f(A,x)_{j_1} \cdot x_{i_1} \cdot \diag (x) A^{\top} \cdot  f(A,x)  \cdot    \langle c(A, x), f(A, x) \rangle
\end{align*}
Finally, combine them and we have
\begin{align*}
       B_{4,2}^{j_1,i_1,j_1,i_0} = B_{4,2,1}^{j_1,i_1,j_1,i_0} + B_{4,2,2}^{j_1,i_1,j_1,i_0}
\end{align*}
{\bf Proof of Part 3.} 
    \begin{align*}
    B_{4,3,1}^{j_1,i_1,j_1,i_0} : = & ~    - c_g(A,x)^{\top} \cdot f(A,x)_{j_1} \cdot \diag (x) \cdot e_{i_1} \cdot  e_{j_1}^\top \cdot  f(A,x)  \cdot    \langle c(A, x), f(A, x) \rangle
\end{align*}
Finally, combine them and we have
\begin{align*}
       B_{4,3}^{j_1,i_1,j_1,i_0} = B_{4,3,1}^{j_1,i_1,j_1,i_0} 
\end{align*}
{\bf Proof of Part 4.} 
    \begin{align*}
    B_{4,4,1}^{j_1,i_1,j_1,i_0} : = & ~  - c_g(A,x)^{\top} \cdot f(A,x)_{j_1}^2  \cdot \diag (x) \cdot A^{\top} \cdot x_{i_1}   \cdot (e_{j_1}- f(A,x) ) \cdot    \langle c(A, x), f(A, x) \rangle
\end{align*}
Finally, combine them and we have
\begin{align*}
       B_{4,4}^{j_1,i_1,j_1,i_0} = B_{4,4,1}^{j_1,i_1,j_1,i_0}  
\end{align*}
{\bf Proof of Part 5.} 
    \begin{align*}
    B_{4,5,1}^{j_1,i_1,j_1,i_0} : = & ~ -  c_g(A,x)^{\top} \cdot f(A,x)_{j_1}^2  \cdot \diag (x) A^{\top} \cdot  f(A,x)  \cdot x_{i_1} \cdot \langle - f(A,x), f(A,x) \rangle\\
    B_{4,5,2}^{j_1,i_1,j_1,i_0} : = & ~ -  c_g(A,x)^{\top} \cdot f(A,x)_{j_1}^3  \cdot \diag (x) A^{\top} \cdot  f(A,x)  \cdot  x_{i_1}
\end{align*}
Finally, combine them and we have
\begin{align*}
       B_{4,5}^{j_1,i_1,j_1,i_0} = B_{4,5,1}^{j_1,i_1,j_1,i_0}  +B_{4,5,2}^{j_1,i_1,j_1,i_0}
\end{align*}
{\bf Proof of Part 6.} 
    \begin{align*}
    B_{4,6,1}^{j_1,i_1,j_1,i_0} : = & ~ -  c_g(A,x)^{\top} \cdot f(A,x)_{j_1}^2  \cdot \diag (x) A^{\top} \cdot  f(A,x)  \cdot x_{i_1} \cdot (\langle - f(A,x), c(A,x) \rangle )\\
    B_{4,6,2}^{j_1,i_1,j_1,i_0} : = & ~ -  c_g(A,x)^{\top} \cdot f(A,x)_{j_1}^2  \cdot \diag (x) A^{\top} \cdot  f(A,x)  \cdot   x_{i_1} \cdot c(A,x)_{j_1} 
\end{align*}
Finally, combine them and we have
\begin{align*}
       B_{4,6}^{j_1,i_1,j_1,i_0} = B_{4,6,1}^{j_1,i_1,j_1,i_0}  +B_{4,6,2}^{j_1,i_1,j_1,i_0}
\end{align*}
\end{proof}
\end{lemma}

\subsection{Constructing \texorpdfstring{$d \times d$}{} matrices for \texorpdfstring{$j_1 = j_0$}{}}
The purpose of the following lemma is to let $i_0$ and $i_1$ disappear.
\begin{lemma}For $j_0,j_1 \in [n]$, a list of $d \times d$ matrices can be expressed as the following sense,
\begin{itemize}
\item {\bf Part 1.}
\begin{align*}
B_{4,1,1}^{j_1,*,j_1,*} & ~ =  f_c(A,x)^2 \cdot f(A,x)_{j_1}^2 \cdot  h(A,x) \cdot  {\bf 1}_d^\top
\end{align*}
\item {\bf Part 2.}
\begin{align*}
B_{4,1,2}^{j_1,*,j_1,*} & ~ =   f(A,x)_{j_1}^2 \cdot c(A,x)_{j_1} \cdot  f_c(A,x) \cdot h(A,x) \cdot  {\bf 1}_d^\top 
\end{align*}
\item {\bf Part 3.}
\begin{align*}
B_{4,1,3}^{j_1,*,j_1,*} & ~ =   f(A,x)_{j_1}^2 \cdot f_c(A,x)^2  \cdot ( (A_{j_1,*}) \circ x^\top  ) \cdot  h(A,x) \cdot I_d
\end{align*}
\item {\bf Part 4.}
\begin{align*}
B_{4,1,4}^{j_1,*,j_1,*}  & ~ =  -  f(A,x)_{j_1}^2 \cdot f_c(A,x)^2 \cdot h(A,x)^\top \cdot h(A,x) \cdot I_d
\end{align*}
\item {\bf Part 5.}
\begin{align*}
B_{4,1,5}^{j_1,*,j_1,*}  & ~ = f(A,x)_{j_1}^2 \cdot (-f_2(A,x) + f(A,x)_{j_1})  \cdot f_c(A,x) \cdot  h(A,x)^\top \cdot h(A,x) \cdot I_d 
\end{align*}
\item {\bf Part 6.}
\begin{align*}
B_{4,1,6}^{j_1,*,j_1,*}  & ~ =     f(A,x)_{j_1}^2 \cdot (-f_c(A,x) + f(A,x)_{j_1}) \cdot f_c(A,x) \cdot  h(A,x)^\top \cdot h(A,x) \cdot I_d
\end{align*}
\item {\bf Part 7.}
\begin{align*}
B_{4,1,7}^{j_1,*,j_1,*}  & ~ =  f(A,x)_{j_1}^2 \cdot f_c(A,x)  \cdot p_{j_1}(A,x)^\top \cdot  h(A,x) \cdot I_d 
\end{align*}
\item {\bf Part 8.}
\begin{align*}
B_{4,2,1}^{j_1,*,j_1,*}  & ~ =    f(A,x)_{j_1}^2  \cdot f_c(A,x) \cdot  c_g(A,x)^\top  \cdot  h(A,x)  \cdot x \cdot {\bf 1}_d^{\top} 
\end{align*}
\item {\bf Part 9.}
\begin{align*}
B_{4,2,2}^{j_1,*,j_1,*}  & ~ =   - f(A,x)_{j_1}  \cdot f_c(A,x) \cdot   c_g(A,x)^\top  \cdot  h(A,x)  \cdot x \cdot {\bf 1}_d^{\top} 
\end{align*}
\item {\bf Part 10.}
\begin{align*}
 B_{4,3,1}^{j_1,*,j_1,*}  & ~ =     -   f(A,x)_{j_1}^2  \cdot    f_c(A,x)   \cdot {\bf 1}_d \cdot c_g(A,x)^{\top} \cdot \diag (x)
\end{align*}
\item {\bf Part 11.}
\begin{align*}
B_{4,4,1}^{j_1,*,j_1,*}  =    -   f(A,x)_{j_1}^2   \cdot f_c(A,x) \cdot  c_g(A,x)^\top \cdot h_e(A,x) \cdot x \cdot {\bf 1}_d^{\top}
\end{align*}
\item {\bf Part 12.}
\begin{align*}
B_{4,5,1}^{j_1,*,j_1,*}  & ~ =   f(A,x)_{j_1}^2 \cdot  f_2(A,x)  \cdot  c_g(A,x)^{\top} \cdot  h(A,x) \cdot x \cdot {\bf 1}_d^{\top} 
\end{align*}
\item {\bf Part 13.}
\begin{align*}
 B_{4,5,2}^{j_1,*,j_1,*}  =   f(A,x)_{j_1}^3 \cdot  c_g(A,x)^{\top} \cdot h(A,x) \cdot x \cdot {\bf 1}_d^{\top}  
\end{align*}
\item {\bf Part 14.}
\begin{align*}
B_{4,6,1}^{j_1,*,j_1,*}  =  f(A,x)_{j_1}^2 \cdot  f_c(A,x)  \cdot  c_g(A,x)^{\top} \cdot  h(A,x) \cdot x \cdot {\bf 1}_d^{\top}  
\end{align*}
\item {\bf Part 15.}
\begin{align*}
B_{4,6,2}^{j_1,*,j_1,*}  =   f(A,x)_{j_1}^2 \cdot c(A,x)_{j_1} \cdot  c_g(A,x)^{\top} \cdot h(A,x) \cdot x \cdot {\bf 1}_d^{\top}  
\end{align*}

\end{itemize}
\begin{proof}
{\bf Proof of Part 1.}
    We have
    \begin{align*}
        B_{4,1,1}^{j_1,i_1,j_1,i_1}  = & ~ e_{i_1}^\top  \cdot  (\langle c(A,x), f(A,x) \rangle)^2  \cdot   f(A,x)_{j_1}^2 \cdot \diag (x) A^{\top} \cdot  f(A,x)\\
        B_{4,1,1}^{j_1,i_1,j_1,i_0}  = & ~ e_{i_1}^\top  \cdot  (\langle c(A,x), f(A,x) \rangle)^2  \cdot   f(A,x)_{j_1}^2 \cdot \diag (x) A^{\top} \cdot  f(A,x)
    \end{align*}
    From the above two equations, we can tell that $B_{4,1,1}^{j_1,*,j_1,*} \in \R^{d \times d}$ is a matrix that both the diagonal and off-diagonal have entries.
    
    Then we have $B_{4,1,1}^{j_1,*,j_1,*} \in \R^{d \times d}$ can be written as the rescaling of a diagonal matrix,
    \begin{align*}
     B_{4,1,1}^{j_1,*,j_1,*} & ~ = (\langle c(A,x), f(A,x) \rangle)^2 \cdot  f(A,x)_{j_1}^2 \cdot \diag (x) A^{\top} \cdot  f(A,x) \cdot  {\bf 1}_d^\top \\
     & ~ = f_c(A,x)^2 \cdot f(A,x)_{j_1}^2 \cdot  h(A,x) \cdot  {\bf 1}_d^\top
\end{align*}
    where the last step is follows from the Definitions~\ref{def:h} and Definitions~\ref{def:f_c}. 

{\bf Proof of Part 2.}
    We have
    \begin{align*}
           B_{4,1,2}^{j_1,i_1,j_1,i_1} = & ~ e_{i_1}^\top  \cdot  c(A,x)_{j_1}  \cdot  f(A,x)_{j_1}^2  \cdot \diag (x) A^{\top} \cdot  f(A,x)  \cdot    \langle c(A, x), f(A, x) \rangle\\
        B_{4,1,2}^{j_1,i_1,j_1,i_0} = & ~ e_{i_1}^\top  \cdot  c(A,x)_{j_1}  \cdot  f(A,x)_{j_1}^2  \cdot \diag (x) A^{\top} \cdot  f(A,x)  \cdot    \langle c(A, x), f(A, x) \rangle
    \end{align*}
     From the above two equations, we can tell that $B_{4,1,2}^{j_1,*,j_1,*} \in \R^{d \times d}$ is a matrix that only diagonal has entries and off-diagonal are all zeros.
    
    Then we have $B_{4,1,2}^{j_1,*,j_1,*} \in \R^{d \times d}$ can be written as the rescaling of a diagonal matrix,
\begin{align*}
     B_{4,1,2}^{j_1,*,j_1,*} & ~ = f(A,x)_{j_1}^2 \cdot c(A,x)_{j_1} \cdot  \langle c(A,x), f(A,x) \rangle \cdot \diag (x) A^{\top} \cdot  f(A,x)  \cdot  {\bf 1}_d^\top   \\
     & ~ = f(A,x)_{j_1}^2 \cdot c(A,x)_{j_1} \cdot  f_c(A,x) \cdot h(A,x) \cdot  {\bf 1}_d^\top 
\end{align*}
    where the last step is follows from the Definitions~\ref{def:h} and Definitions~\ref{def:f_c}.

{\bf Proof of Part 3.}
We have for diagonal entry and off-diagonal entry can be written as follows 
    \begin{align*}
        B_{4,1,3}^{j_1,i_1,j_1,i_1} = & ~f(A,x)_{j_1}^2  \cdot  (\langle c(A,x), f(A,x) \rangle)^2  \cdot (  (A_{j_1,*})  \circ  x^\top  ) \cdot \diag (x) A^{\top} \cdot  f(A,x) \\
        B_{4,1,3}^{j_1,i_1,j_1,i_0} = & ~f(A,x)_{j_1}^2  \cdot  (\langle c(A,x), f(A,x) \rangle)^2  \cdot (  (A_{j_1,*})  \circ  x^\top  ) \cdot \diag (x) A^{\top} \cdot  f(A,x) 
    \end{align*}
From the above equation, we can show that matrix $B_{4,1,3}^{j_1,*,j_1,*}$ can be expressed as a rank-$1$ matrix,
\begin{align*}
     B_{4,1,3}^{j_1,*,j_1,*} & ~ = f(A,x)_{j_1}^2 \cdot (\langle c(A,x), f(A,x) \rangle)^2 \cdot ( (A_{j_1,*}) \circ x^\top  ) \cdot  \diag (x) A^{\top} \cdot  f(A,x) \cdot I_d \\
     & ~ =  f(A,x)_{j_1}^2 \cdot f_c(A,x)^2  \cdot ( (A_{j_1,*}) \circ x^\top  ) \cdot  h(A,x) \cdot I_d
\end{align*}
    where the last step is follows from the Definitions~\ref{def:h} and Definitions~\ref{def:f_c}.

{\bf Proof of Part 4.}
We have for diagonal entry and off-diagonal entry can be written as follows
    \begin{align*}
        B_{4,1,4}^{j_1,i_1,j_1,i_1}   = & ~ -   f(A,x)_{j_1}^2  \cdot  f(A,x)^\top   \cdot  A  \cdot   (\diag(x))^2  \cdot  (\langle c(A,x), f(A,x) \rangle)^2 \cdot   A^{\top} \cdot  f(A,x) \\
        B_{4,1,4}^{j_1,i_1,j_1,i_0}   = & ~ -   f(A,x)_{j_1}^2  \cdot  f(A,x)^\top   \cdot  A  \cdot   (\diag(x))^2  \cdot  (\langle c(A,x), f(A,x) \rangle)^2 \cdot   A^{\top} \cdot  f(A,x) 
    \end{align*}
 From the above equation, we can show that matrix $B_{4,1,4}^{j_1,*,j_1,*}$ can be expressed as a rank-$1$ matrix,
\begin{align*}
    B_{4,1,4}^{j_1,*,j_1,*}  & ~ =  -  f(A,x)_{j_1}^2 \cdot (\langle c(A,x), f(A,x) \rangle )^2 \cdot f(A,x)^\top  \cdot A \cdot (\diag (x))^2 \cdot A^{\top} \cdot  f(A,x)\cdot I_d \\
     & ~ =  -  f(A,x)_{j_1}^2 \cdot f_c(A,x)^2 \cdot h(A,x)^\top \cdot h(A,x) \cdot I_d
\end{align*}
    where the last step is follows from the Definitions~\ref{def:h} and Definitions~\ref{def:f_c}.

{\bf Proof of Part 5.}
We have for diagonal entry and off-diagonal entry can be written as follows
    \begin{align*}
         B_{4,1,5}^{j_1,i_1,j_1,i_0} = & ~    f(A,x)_{j_1}^2  \cdot f(A,x)^\top \cdot A \cdot (\diag(x))^2 \cdot (\langle -f(A,x), f(A,x) \rangle + f(A,x)_{j_1}) \cdot A^{\top} \cdot  f(A,x) \\& ~ \cdot    \langle c(A, x), f(A, x) \rangle \\
         B_{4,1,5}^{j_1,i_1,j_1,i_0} = & ~    f(A,x)_{j_1}^2  \cdot f(A,x)^\top \cdot A \cdot (\diag(x))^2 \cdot (\langle -f(A,x), f(A,x) \rangle + f(A,x)_{j_1}) \cdot A^{\top} \cdot  f(A,x) \\& ~ \cdot    \langle c(A, x), f(A, x) \rangle
    \end{align*}
    From the above equation, we can show that matrix $B_{4,1,5}^{j_1,*,j_1,*}$ can be expressed as a rank-$1$ matrix,
\begin{align*}
    B_{4,1,5}^{j_1,*,j_1,*}  & ~ =  f(A,x)_{j_1}^2 \cdot (\langle -f(A,x), f(A,x) \rangle + f(A,x)_{j_1})  \cdot \langle c(A,x), f(A,x) \rangle \cdot  f(A,x)^\top \\
    & ~ \cdot A \cdot (\diag (x))^2 \cdot A^{\top} \cdot  f(A,x)\cdot I_d\\
     & ~ =  f(A,x)_{j_1}^2 \cdot (-f_2(A,x) + f(A,x)_{j_1})  \cdot f_c(A,x) \cdot  h(A,x)^\top \cdot h(A,x) \cdot I_d
\end{align*}
    where the last step is follows from the Definitions~\ref{def:h}, Definitions~\ref{def:f_c} and Definitions~\ref{def:f_2}.

{\bf Proof of Part 6.}
We have for diagonal entry and off-diagonal entry can be written as follows
    \begin{align*}
        B_{4,1,6}^{j_1,i_1,j_1,i_1}  = & ~   f(A,x)_{j_1}^2 \cdot f(A,x)^\top  \cdot A \cdot (\diag(x))^2 \cdot(\langle -f(A,x), c(A,x) \rangle + f(A,x)_{j_1}) \cdot   A^{\top} \cdot  f(A,x)  \\& ~ \cdot    \langle c(A, x), f(A, x) \rangle\\
        B_{4,1,6}^{j_1,i_1,j_1,i_0}  = & ~   f(A,x)_{j_1}^2 \cdot f(A,x)^\top  \cdot A \cdot (\diag(x))^2 \cdot(\langle -f(A,x), c(A,x) \rangle + f(A,x)_{j_1}) \cdot   A^{\top} \cdot  f(A,x) \\& ~ \cdot    \langle c(A, x), f(A, x) \rangle
    \end{align*}
    From the above equation, we can show that matrix $B_{4,1,6}^{j_1,*,j_1,*}$ can be expressed as a rank-$1$ matrix,
\begin{align*}
    B_{4,1,6}^{j_1,*,j_1,*}  & ~ =   f(A,x)_{j_1}^2 \cdot (\langle -f(A,x), c(A,x) \rangle + f(A,x)_{j_1}) \cdot \langle c(A,x), f(A,x) \rangle \cdot  f(A,x)^\top \\
    & ~ \cdot A \cdot (\diag (x))^2 \cdot A^{\top} \cdot  f(A,x)\cdot I_d\\
     & ~ = f(A,x)_{j_1}^2 \cdot (-f_c(A,x) + f(A,x)_{j_1}) \cdot f_c(A,x) \cdot  h(A,x)^\top \cdot h(A,x) \cdot I_d
\end{align*}
    where the last step is follows from the Definitions~\ref{def:h} and Definitions~\ref{def:f_c} .
    
{\bf Proof of Part 7.}
We have for diagonal entry and off-diagonal entry can be written as follows
    \begin{align*}
         B_{4,1,7}^{j_1,i_1,j_1,i_1} = & ~ f(A,x)_{j_1}^2 \cdot (( e_{j_1}^\top  - f(A,x)^\top) \circ q(A,x)^\top) \cdot A \cdot  (\diag(x))^2 \cdot A^{\top} \cdot  f(A,x)  \cdot    \langle c(A, x), f(A, x) \rangle\\
         B_{4,1,7}^{j_1,i_1,j_1,i_0} = & ~ f(A,x)_{j_1}^2 \cdot (( e_{j_1}^\top  - f(A,x)^\top) \circ q(A,x)^\top) \cdot A \cdot  (\diag(x))^2 \cdot A^{\top} \cdot  f(A,x)  \cdot    \langle c(A, x), f(A, x) \rangle
    \end{align*}
    From the above equation, we can show that matrix $B_{4,1,7}^{j_1,*,j_1,*}$ can be expressed as a rank-$1$ matrix,
\begin{align*}
     B_{4,1,7}^{j_1,*,j_1,*}  & ~ =   f(A,x)_{j_1}^2 \cdot \langle c(A,x), f(A,x) \rangle  \cdot  ((e_{j_1}^\top - f(A,x)^\top) \circ q(A,x)^\top) \cdot A \cdot  (\diag (x))^2 \cdot A^{\top} \cdot  f(A,x) \cdot I_d \\
     & ~ =f(A,x)_{j_1}^2 \cdot f_c(A,x)  \cdot p_{j_1}(A,x)^\top \cdot  h(A,x) \cdot I_d 
\end{align*}
    where the last step is follows from the Definitions~\ref{def:h}, Definitions~\ref{def:f_c} and Definitions~\ref{def:p}.
    
{\bf Proof of Part 8.}
We have for diagonal entry and off-diagonal entry can be written as follows
    \begin{align*}
         B_{4,2,1}^{j_1,i_1,j_1,i_0} = & ~  c_g(A,x)^{\top} \cdot f(A,x)_{j_1}^2 \cdot x_{i_1} \cdot \diag (x) A^{\top} \cdot  f(A,x)  \cdot    \langle c(A, x), f(A, x) \rangle \\
         B_{4,2,1}^{j_1,i_1,j_1,i_0} = & ~  c_g(A,x)^{\top} \cdot f(A,x)_{j_1}^2 \cdot x_{i_1} \cdot \diag (x) A^{\top} \cdot  f(A,x)  \cdot    \langle c(A, x), f(A, x) \rangle
    \end{align*}
    From the above equation, we can show that matrix $B_{4,2,1}^{j_1,*,j_1,*}$ can be expressed as a rank-$1$ matrix,
\begin{align*}
     B_{4,2,1}^{j_1,*,j_1,*}  & ~ =   f(A,x)_{j_1}^2  \cdot \langle c(A,x), f(A,x) \rangle \cdot c_g(A,x)^{\top} \cdot   \diag (x) \cdot A^{\top} \cdot  f(A,x) \cdot x \cdot {\bf 1}_d^{\top}  \\
     & ~ =f(A,x)_{j_1}^2  \cdot f_c(A,x) \cdot  c_g(A,x)^\top  \cdot  h(A,x)  \cdot x \cdot {\bf 1}_d^{\top} 
\end{align*}
    where the last step is follows from the Definitions~\ref{def:h} and Definitions~\ref{def:f_c}.

{\bf Proof of Part 9.}
We have for diagonal entry and off-diagonal entry can be written as follows
    \begin{align*}
         B_{4,2,2}^{j_1,i_1,j_1,i_1} = & ~ - c_g(A,x)^{\top} \cdot f(A,x)_{j_1} \cdot x_{i_1} \cdot \diag (x) A^{\top} \cdot  f(A,x)  \cdot    \langle c(A, x), f(A, x) \rangle\\
         B_{4,2,2}^{j_1,i_1,j_1,i_0} = & ~ - c_g(A,x)^{\top} \cdot f(A,x)_{j_1} \cdot x_{i_1} \cdot \diag (x) A^{\top} \cdot  f(A,x)  \cdot    \langle c(A, x), f(A, x) \rangle
    \end{align*}
        From the above equation, we can show that matrix $B_{4,2,2}^{j_1,*,j_1,*}$ can be expressed as a rank-$1$ matrix,
\begin{align*}
     B_{4,2,2}^{j_1,*,j_1,*}  & ~ =   - f(A,x)_{j_1}  \cdot \langle c(A,x), f(A,x) \rangle \cdot  c_g(A,x)^{\top} \cdot   \diag (x) \cdot A^{\top} \cdot  f(A,x) \cdot x \cdot {\bf 1}_d^{\top}  \\
     & ~ =- f(A,x)_{j_1}  \cdot f_c(A,x) \cdot   c_g(A,x)^\top  \cdot  h(A,x)  \cdot x \cdot {\bf 1}_d^{\top} 
\end{align*}
    where the last step is follows from the Definitions~\ref{def:h} and Definitions~\ref{def:f_c}.

{\bf Proof of Part 10.}
We have for diagonal entry and off-diagonal entry can be written as follows
    \begin{align*}
         B_{4,3,1}^{j_1,i_1,j_1,i_1}  = & ~   - \underbrace{c_g(A,x)^{\top}}_{1 \times d} \cdot f(A,x)_{j_1} \cdot \underbrace{\diag (x)}_{d \times d} \cdot \underbrace{e_{i_1}}_{d \times 1} \cdot  \underbrace{e_{j_1}^\top}_{1 \times n} \cdot  \underbrace{f(A,x)}_{n \times 1}  \cdot    \langle c(A, x), f(A, x) \rangle\\
         B_{4,3,1}^{j_1,i_1,j_1,i_0}  = & ~  - c_g(A,x)^{\top} \cdot f(A,x)_{j_1} \cdot \diag (x) \cdot e_{i_1} \cdot  e_{j_1}^\top \cdot  f(A,x)  \cdot    \langle c(A, x), f(A, x) \rangle
    \end{align*}
            From the above equation, we can show that matrix $B_{4,3,1}^{j_1,*,j_1,*}$ can be expressed as a rank-$1$ matrix,
\begin{align*}
    B_{4,3,1}^{j_1,*,j_1,*}  & ~ =  -   f(A,x)_{j_1}^2  \cdot    \langle c(A, x), f(A, x) \rangle   \cdot {\bf 1}_d \cdot c_g(A,x)^{\top} \cdot \diag (x)  \\
     & ~ =    -   f(A,x)_{j_1}^2  \cdot    f_c(A,x)   \cdot {\bf 1}_d \cdot c_g(A,x)^{\top} \cdot \diag (x)
\end{align*} 
    where the last step is follows from the  Definitions~\ref{def:f_c}.

{\bf Proof of Part 11.}
We have for diagonal entry and off-diagonal entry can be written as follows
    \begin{align*}
         B_{4,4,1}^{j_1,i_1,j_1,i_1}   = & ~   -  c_g(A,x)^{\top}  \cdot f(A,x)_{j_1}^2  \cdot  \diag (x)  \cdot  A^{\top}  \cdot x_{i_1}   \cdot (e_{j_1}- f(A,x) ) \cdot    \langle c(A, x), f(A, x) \rangle\\
         B_{4,4,1}^{j_1,i_1,j_1,i_0}   = & ~  - c_g(A,x)^{\top} \cdot f(A,x)_{j_1}^2  \cdot \diag (x) \cdot A^{\top} \cdot x_{i_1}   \cdot (e_{j_1}- f(A,x) ) \cdot    \langle c(A, x), f(A, x) \rangle
    \end{align*}
            From the above equation, we can show that matrix $B_{4,4,1}^{j_1,*,j_1,*}$ can be expressed as a rank-$1$ matrix,
\begin{align*}
    B_{4,4,1}^{j_1,*,j_1,*} & ~ =   -   f(A,x)_{j_1}^2   \cdot \langle c(A, x), f(A, x) \rangle \cdot  c_g(A,x)^\top \cdot  \diag (x) \cdot A^{\top}     \cdot (e_{j_1}- f(A,x) )  \cdot x \cdot {\bf 1}_d^{\top} \\
     & ~ =  -   f(A,x)_{j_1}^2   \cdot f_c(A,x) \cdot  c_g(A,x)^\top \cdot h_e(A,x) \cdot x \cdot {\bf 1}_d^{\top}
\end{align*}
    where the last step is follows from the Definitions~\ref{def:h_e} and Definitions~\ref{def:f_c}.

{\bf Proof of Part 12.}
We have for diagonal entry and off-diagonal entry can be written as follows
    \begin{align*}
         B_{4,5,1}^{j_1,i_1,j_1,i_0}   = & ~    -  c_g(A,x)^{\top} \cdot f(A,x)_{j_1}^2  \cdot \diag (x) A^{\top} \cdot  f(A,x)  \cdot x_{i_1} \cdot \langle - f(A,x), f(A,x) \rangle\\
         B_{4,5,1}^{j_1,i_1,j_1,i_0}   = & ~   -  c_g(A,x)^{\top} \cdot f(A,x)_{j_1}^2  \cdot \diag (x) A^{\top} \cdot  f(A,x)  \cdot x_{i_1} \cdot \langle - f(A,x), f(A,x) \rangle
    \end{align*}
            From the above equation, we can show that matrix $B_{4,5,1}^{j_1,*,j_1,*}$ can be expressed as a rank-$1$ matrix,
\begin{align*}
    B_{4,5,1}^{j_1,*,j_1,*}  & ~ =   f(A,x)_{j_1}^2 \cdot  \langle f(A,x), f(A,x) \rangle  \cdot  c_g(A,x)^{\top} \cdot  \diag (x) A^{\top} \cdot  f(A,x) \cdot x \cdot {\bf 1}_d^{\top}  \\
     & ~ =  f(A,x)_{j_1}^2 \cdot  f_2(A,x)  \cdot  c_g(A,x)^{\top} \cdot  h(A,x) \cdot x \cdot {\bf 1}_d^{\top} 
\end{align*}
    where the last step is follows from the Definitions~\ref{def:h} and Definitions~\ref{def:f_2}.

{\bf Proof of Part 13.}
We have for diagonal entry and off-diagonal entry can be written as follows
    \begin{align*}
         B_{4,5,2}^{j_1,i_1,j_1,i_0}   = & ~    -  c_g(A,x)^{\top} \cdot f(A,x)_{j_1}^3  \cdot \diag (x) A^{\top} \cdot  f(A,x)  \cdot  x_{i_1}\\
         B_{4,5,2}^{j_1,i_1,j_1,i_0}   = & ~   -  c_g(A,x)^{\top} \cdot f(A,x)_{j_1}^3  \cdot \diag (x) A^{\top} \cdot  f(A,x)  \cdot  x_{i_1}
    \end{align*}
            From the above equation, we can show that matrix $B_{4,5,2}^{j_1,*,j_1,*}$ can be expressed as a rank-$1$ matrix,
\begin{align*}
    B_{4,5,2}^{j_1,*,j_1,*}  & ~ =   f(A,x)_{j_1}^3 \cdot  c_g(A,x)^{\top} \cdot  \diag (x) A^{\top} \cdot  f(A,x) \cdot x \cdot {\bf 1}_d^{\top}  \\
     & ~ =   f(A,x)_{j_1}^3 \cdot  c_g(A,x)^{\top} \cdot h(A,x) \cdot x \cdot {\bf 1}_d^{\top} 
\end{align*}
    where the last step is follows from the Definitions~\ref{def:h}.

    {\bf Proof of Part 14.}
We have for diagonal entry and off-diagonal entry can be written as follows
    \begin{align*}
         B_{4,6,1}^{j_1,i_1,j_1,i_0}   = & ~     -  c_g(A,x)^{\top} \cdot f(A,x)_{j_1}^2  \cdot \diag (x) A^{\top} \cdot  f(A,x)  \cdot x_{i_1} \cdot (\langle - f(A,x), c(A,x) \rangle )\\
         B_{4,6,1}^{j_1,i_1,j_1,i_0}   = & ~   -  c_g(A,x)^{\top} \cdot f(A,x)_{j_1}^2  \cdot \diag (x) A^{\top} \cdot  f(A,x)  \cdot x_{i_1} \cdot (\langle - f(A,x), c(A,x) \rangle )
    \end{align*}
            From the above equation, we can show that matrix $B_{4,6,1}^{j_1,*,j_1,*}$ can be expressed as a rank-$1$ matrix,
\begin{align*}
    B_{4,6,1}^{j_1,*,j_1,*}  & ~ =   f(A,x)_{j_1}^2 \cdot  \langle f(A,x), c(A,x) \rangle  \cdot  c_g(A,x)^{\top} \cdot  \diag (x) A^{\top} \cdot  f(A,x) \cdot x \cdot {\bf 1}_d^{\top}  \\
     & ~ =  f(A,x)_{j_1}^2 \cdot  f_c(A,x)  \cdot  c_g(A,x)^{\top} \cdot  h(A,x) \cdot x \cdot {\bf 1}_d^{\top} 
\end{align*}
    where the last step is follows from the Definitions~\ref{def:h} and Definitions~\ref{def:f_c}.

 {\bf Proof of Part 15.}
We have for diagonal entry and off-diagonal entry can be written as follows
    \begin{align*}
         B_{4,6,2}^{j_1,i_1,j_1,i_0}   = & ~   -  c_g(A,x)^{\top} \cdot f(A,x)_{j_1}^2  \cdot \diag (x) A^{\top} \cdot  f(A,x)  \cdot   x_{i_1} \cdot c(A,x)_{j_1} \\
         B_{4,6,2}^{j_1,i_1,j_1,i_0}   = & ~   -  c_g(A,x)^{\top} \cdot f(A,x)_{j_1}^2  \cdot \diag (x) A^{\top} \cdot  f(A,x)  \cdot   x_{i_1} \cdot c(A,x)_{j_1} 
    \end{align*}
            From the above equation, we can show that matrix $B_{4,6,2}^{j_1,*,j_1,*}$ can be expressed as a rank-$1$ matrix,
\begin{align*}
    B_{4,6,2}^{j_1,*,j_1,*}  & ~ =   f(A,x)_{j_1}^2 \cdot c(A,x)_{j_1} \cdot  c_g(A,x)^{\top} \cdot  \diag (x) A^{\top} \cdot  f(A,x) \cdot x \cdot {\bf 1}_d^{\top}  \\
     & ~ =   f(A,x)_{j_1}^2 \cdot c(A,x)_{j_1} \cdot  c_g(A,x)^{\top} \cdot h(A,x) \cdot x \cdot {\bf 1}_d^{\top} 
\end{align*}
    where the last step is follows from the Definitions~\ref{def:h}.
\end{proof}
\end{lemma}

\subsection{Case \texorpdfstring{$j_1 \neq j_0, i_1 = i_0$}{}}
\begin{lemma}
For $j_1 \neq j_0$ and $i_0 = i_1$. If the following conditions hold
    \begin{itemize}
     \item Let $u(A,x) \in \R^n$ be defined as Definition~\ref{def:u}
    \item Let $\alpha(A,x) \in \R$ be defined as Definition~\ref{def:alpha}
     \item Let $f(A,x) \in \R^n$ be defined as Definition~\ref{def:f}
    \item Let $c(A,x) \in \R^n$ be defined as Definition~\ref{def:c}
    \item Let $g(A,x) \in \R^d$ be defined as Definition~\ref{def:g} 
    \item Let $q(A,x) = c(A,x) + f(A,x) \in \R^n$
    \item Let $c_g(A,x) \in \R^d$ be defined as Definition~\ref{def:c_g}.
    \item Let $L_g(A,x) \in \R$ be defined as Definition~\ref{def:l_g}
    \item Let $v \in \R^n$ be a vector 
    \item Let $B_1^{j_1,i_1,j_0,i_0}$ be defined as Definition~\ref{def:b_1}
    \end{itemize}
    Then, For $j_0,j_1 \in [n], i_0,i_1 \in [d]$, we have 
    \begin{itemize}
\item {\bf Part 1.} For $B_{4,1}^{j_1,i_1,j_0,i_1}$, we have 
\begin{align*}
 B_{4,1}^{j_1,i_1,j_0,i_1}  = & ~  \frac{\d}{\d A_{j_1,i_1}} (- c_g(A,x)^{\top} ) \cdot  f(A,x)_{j_0} \cdot \diag (x) A^{\top} \cdot  f(A,x)  \cdot    \langle c(A, x), f(A, x) \rangle \\
 = & ~ B_{4,1,1}^{j_1,i_1,j_0,i_1} + B_{4,1,2}^{j_1,i_1,j_0,i_1} + B_{4,1,3}^{j_1,i_1,j_0,i_1} + B_{4,1,4}^{j_1,i_1,j_0,i_1} + B_{4,1,5}^{j_1,i_1,j_0,i_1} + B_{4,1,6}^{j_1,i_1,j_0,i_1} + B_{4,1,7}^{j_1,i_1,j_0,i_1}
\end{align*} 
\item {\bf Part 2.} For $B_{4,2}^{j_1,i_1,j_0,i_1}$, we have 
\begin{align*}
  B_{4,2}^{j_1,i_1,j_0,i_1} = & ~  - c_g(A,x)^{\top} \cdot \frac{\d}{\d A_{j_1,i_1}} ( f(A,x)_{j_0} )  \cdot \diag (x) A^{\top} \cdot  f(A,x)  \cdot    \langle c(A, x), f(A, x) \rangle  \\
    = & ~  B_{4,2,1}^{j_1,i_1,j_0,i_1}  
\end{align*} 
\item {\bf Part 3.} For $B_{4,3}^{j_1,i_1,j_0,i_1}$, we have 
\begin{align*}
  B_{4,3}^{j_1,i_1,j_0,i_1} = & ~ - c_g(A,x)^{\top} \cdot f(A,x)_{j_0} \cdot \diag (x) \cdot \frac{\d}{\d A_{j_1,i_1}} ( A^{\top})  \cdot  f(A,x)  \cdot    \langle c(A, x), f(A, x) \rangle \\
     = & ~ B_{4,3,1}^{j_1,i_1,j_0,i_1}  
\end{align*} 
\item {\bf Part 4.} For $B_{4,4}^{j_1,i_1,j_0,i_1}$, we have 
\begin{align*}
  B_{4,4}^{j_1,i_1,j_0,i_1} = & ~ - c_g(A,x)^{\top} \cdot f(A,x)_{j_0} \cdot \diag (x) A^{\top} \cdot  \frac{\d}{\d A_{j_1,i_1}} ( f(A,x)  )  \cdot    \langle c(A, x), f(A, x) \rangle\\
     = & ~ B_{4,4,1}^{j_1,i_1,j_0,i_1} 
\end{align*}
\item {\bf Part 5.} For $B_{4,5}^{j_1,i_1,j_0,i_1}$, we have 
\begin{align*}
  B_{4,5}^{j_1,i_1,j_0,i_1} = & ~ -  c_g(A,x)^{\top} \cdot f(A,x)_{j_0}  \cdot \diag (x) A^{\top} \cdot  f(A,x)  \cdot  \langle \frac{\d c(A,x)}{\d A_{j_1,i_1}}, f(A,x) \rangle  \\
     = & ~ B_{4,5,1}^{j_1,i_1,j_0,i_1}  + B_{4,5,2}^{j_1,i_1,j_0,i_1}  
\end{align*}
\item {\bf Part 6.} For $B_{4,6}^{j_1,i_1,j_0,i_1}$, we have 
\begin{align*}
  B_{4,6}^{j_1,i_1,j_0,i_1} = & ~ -  c_g(A,x)^{\top} \cdot f(A,x)_{j_0}  \cdot \diag (x) A^{\top} \cdot  f(A,x)  \cdot  \langle c(A,x),  \frac{\d f(A,x)}{\d A_{j_1,i_1}} \rangle  \\
     = & ~ B_{4,6,1}^{j_1,i_1,j_0,i_1}  + B_{4,6,2}^{j_1,i_1,j_0,i_1}  
\end{align*}
\end{itemize}
\begin{proof}
    {\bf Proof of Part 1.}
    \begin{align*}
    B_{4,1,1}^{j_1,i_1,j_0,i_1} : = & ~ e_{i_1}^\top  \cdot  (\langle c(A,x), f(A,x) \rangle)^2  \cdot  f(A,x)_{j_1} \cdot f(A,x)_{j_0} \cdot \diag (x) A^{\top} \cdot  f(A,x) \\
    B_{4,1,2}^{j_1,i_1,j_0,i_1} : = & ~ e_{i_1}^\top  \cdot  c(A,x)_{j_1}  \cdot  f(A,x)_{j_1} \cdot f(A,x)_{j_0}  \cdot \diag (x) A^{\top} \cdot  f(A,x)  \cdot    \langle c(A, x), f(A, x) \rangle\\
    B_{4,1,3}^{j_1,i_1,j_0,i_1} : = & ~f(A,x)_{j_1} \cdot f(A,x)_{j_0}  \cdot  (\langle c(A,x), f(A,x) \rangle)^2  \cdot (  (A_{j_1,*})  \circ  x^\top  ) \cdot \diag (x) A^{\top} \cdot  f(A,x)  \\
    B_{4,1,4}^{j_1,i_1,j_0,i_1} : = & ~  -   f(A,x)_{j_1} \cdot f(A,x)_{j_0}  \cdot  f(A,x)^\top   \cdot  A  \cdot   (\diag(x))^2  \cdot  (\langle c(A,x), f(A,x) \rangle)^2 \cdot   A^{\top} \cdot  f(A,x) \\
    B_{4,1,5}^{j_1,i_1,j_0,i_1} : = & ~    f(A,x)_{j_1} \cdot f(A,x)_{j_0}  \cdot f(A,x)^\top \cdot A \cdot (\diag(x))^2 \cdot (\langle -f(A,x), f(A,x) \rangle + f(A,x)_{j_1}) \\
    & ~ \cdot A^{\top} \cdot  f(A,x)  \cdot    \langle c(A, x), f(A, x) \rangle\\
    B_{4,1,6}^{j_1,i_1,j_0,i_1} : = & ~  f(A,x)_{j_1} \cdot f(A,x)_{j_0} \cdot f(A,x)^\top  \cdot A \cdot (\diag(x))^2 \cdot(\langle -f(A,x), c(A,x) \rangle + f(A,x)_{j_1}) \\
    & ~ \cdot   A^{\top} \cdot  f(A,x)  \cdot    \langle c(A, x), f(A, x) \rangle\\
    B_{4,1,7}^{j_1,i_1,j_0,i_1} : = & ~ f(A,x)_{j_1} \cdot f(A,x)_{j_0} \cdot (( e_{j_1}^\top  - f(A,x)^\top) \circ q(A,x)^\top) \cdot A \cdot  (\diag(x))^2 \cdot A^{\top} \cdot  f(A,x) \\
          & ~ \cdot    \langle c(A, x), f(A, x) \rangle
\end{align*}
Finally, combine them and we have
\begin{align*}
       B_{4,1}^{j_1,i_1,j_0,i_1} = B_{4,1,1}^{j_1,i_1,j_0,i_1} + B_{4,1,2}^{j_1,i_1,j_0,i_1} + B_{4,1,3}^{j_1,i_1,j_0,i_1} + B_{4,1,4}^{j_1,i_1,j_0,i_1} + B_{4,1,5}^{j_1,i_1,j_0,i_1} + B_{4,1,6}^{j_1,i_1,j_0,i_1} + B_{4,1,7}^{j_1,i_1,j_0,i_1}
\end{align*}
{\bf Proof of Part 2.}
    \begin{align*}
    B_{4,2,1}^{j_1,i_1,j_0,i_1} : = & ~   c_g(A,x)^{\top} \cdot f(A,x)_{j_1} \cdot f(A,x)_{j_0} \cdot x_{i_1} \cdot \diag (x) A^{\top} \cdot  f(A,x)  \cdot    \langle c(A, x), f(A, x) \rangle 
\end{align*}
Finally, combine them and we have
\begin{align*}
       B_{4,2}^{j_1,i_1,j_0,i_1} = B_{4,2,1}^{j_1,i_1,j_0,i_1}  
\end{align*}
{\bf Proof of Part 3.} 
    \begin{align*}
    B_{4,3,1}^{j_1,i_1,j_0,i_1} : = & ~    - c_g(A,x)^{\top} \cdot f(A,x)_{j_0} \cdot \diag (x) \cdot e_{i_1} \cdot  e_{j_1}^\top \cdot  f(A,x)  \cdot    \langle c(A, x), f(A, x) \rangle
\end{align*}
Finally, combine them and we have
\begin{align*}
       B_{4,3}^{j_1,i_1,j_0,i_1} = B_{4,3,1}^{j_1,i_1,j_0,i_1} 
\end{align*}
{\bf Proof of Part 4.} 
    \begin{align*}
    B_{4,4,1}^{j_1,i_1,j_0,i_1} : = & ~  - c_g(A,x)^{\top} \cdot f(A,x)_{j_1} \cdot f(A,x)_{j_0}   \cdot \diag (x) \cdot A^{\top} \cdot x_{i_1}   \cdot (e_{j_1}- f(A,x) ) \cdot    \langle c(A, x), f(A, x) \rangle
\end{align*}
Finally, combine them and we have
\begin{align*}
       B_{4,4}^{j_1,i_1,j_0,i_1} = B_{4,4,1}^{j_1,i_1,j_0,i_1}  
\end{align*}
{\bf Proof of Part 5.} 
    \begin{align*}
    B_{4,5,1}^{j_1,i_1,j_0,i_1} : = & ~ -  c_g(A,x)^{\top} \cdot f(A,x)_{j_1} \cdot f(A,x)_{j_0}  \cdot \diag (x) A^{\top} \cdot  f(A,x)  \cdot x_{i_1} \cdot \langle - f(A,x), f(A,x) \rangle\\
    B_{4,5,2}^{j_1,i_1,j_0,i_1} : = & ~ -  c_g(A,x)^{\top} \cdot f(A,x)_{j_1}^2  \cdot f(A,x)_{j_0}  \cdot \diag (x) A^{\top} \cdot  f(A,x)  \cdot  x_{i_1}
\end{align*}
Finally, combine them and we have
\begin{align*}
       B_{4,5}^{j_1,i_1,j_0,i_1} = B_{4,5,1}^{j_1,i_1,j_0,i_1}  +B_{4,5,2}^{j_1,i_1,j_0,i_1}
\end{align*}
{\bf Proof of Part 6.} 
    \begin{align*}
    B_{4,6,1}^{j_1,i_1,j_0,i_1} : = & ~ -  c_g(A,x)^{\top} \cdot f(A,x)_{j_1} \cdot f(A,x)_{j_0}  \cdot \diag (x) A^{\top} \cdot  f(A,x)  \cdot x_{i_1} \cdot (\langle - f(A,x), c(A,x) \rangle )\\
    B_{4,6,2}^{j_1,i_1,j_0,i_1} : = & ~ -  c_g(A,x)^{\top} \cdot f(A,x)_{j_1} \cdot f(A,x)_{j_0}  \cdot \diag (x) A^{\top} \cdot  f(A,x)  \cdot   x_{i_1} \cdot c(A,x)_{j_1} 
\end{align*}
Finally, combine them and we have
\begin{align*}
       B_{4,6}^{j_1,i_1,j_0,i_1} = B_{4,6,1}^{j_1,i_1,j_0,i_1}  +B_{4,6,2}^{j_1,i_1,j_0,i_1}
\end{align*}
\end{proof}
\end{lemma}

\subsection{Case \texorpdfstring{$j_1 \neq j_0, i_1 \neq i_0$}{}}
\begin{lemma}
For $j_1 \neq j_0$ and $i_0 \neq i_1$. If the following conditions hold
    \begin{itemize}
     \item Let $u(A,x) \in \R^n$ be defined as Definition~\ref{def:u}
    \item Let $\alpha(A,x) \in \R$ be defined as Definition~\ref{def:alpha}
     \item Let $f(A,x) \in \R^n$ be defined as Definition~\ref{def:f}
    \item Let $c(A,x) \in \R^n$ be defined as Definition~\ref{def:c}
    \item Let $g(A,x) \in \R^d$ be defined as Definition~\ref{def:g} 
    \item Let $q(A,x) = c(A,x) + f(A,x) \in \R^n$
    \item Let $c_g(A,x) \in \R^d$ be defined as Definition~\ref{def:c_g}.
    \item Let $L_g(A,x) \in \R$ be defined as Definition~\ref{def:l_g}
    \item Let $v \in \R^n$ be a vector 
    \item Let $B_1^{j_1,i_1,j_0,i_0}$ be defined as Definition~\ref{def:b_1}
    \end{itemize}
    Then, For $j_0,j_1 \in [n], i_0,i_1 \in [d]$, we have 
    \begin{itemize}
\item {\bf Part 1.} For $B_{4,1}^{j_1,i_1,j_0,i_0}$, we have 
\begin{align*}
 B_{4,1}^{j_1,i_1,j_0,i_0}  = & ~  \frac{\d}{\d A_{j_1,i_1}} (- c_g(A,x)^{\top} ) \cdot  f(A,x)_{j_0} \cdot \diag (x) A^{\top} \cdot  f(A,x)  \cdot    \langle c(A, x), f(A, x) \rangle \\
 = & ~ B_{4,1,1}^{j_1,i_1,j_0,i_0} + B_{4,1,2}^{j_1,i_1,j_0,i_0} + B_{4,1,3}^{j_1,i_1,j_0,i_0} + B_{4,1,4}^{j_1,i_1,j_0,i_0} + B_{4,1,5}^{j_1,i_1,j_0,i_0} + B_{4,1,6}^{j_1,i_1,j_0,i_0} + B_{4,1,7}^{j_1,i_1,j_0,i_0}
\end{align*} 
\item {\bf Part 2.} For $B_{4,2}^{j_1,i_1,j_0,i_0}$, we have 
\begin{align*}
  B_{4,2}^{j_1,i_1,j_0,i_0} = & ~  - c_g(A,x)^{\top} \cdot \frac{\d}{\d A_{j_1,i_1}} ( f(A,x)_{j_0} )  \cdot \diag (x) A^{\top} \cdot  f(A,x)  \cdot    \langle c(A, x), f(A, x) \rangle  \\
    = & ~  B_{4,2,1}^{j_1,i_1,j_0,i_0}  
\end{align*} 
\item {\bf Part 3.} For $B_{4,3}^{j_1,i_1,j_0,i_0}$, we have 
\begin{align*}
  B_{4,3}^{j_1,i_1,j_0,i_0} = & ~ - c_g(A,x)^{\top} \cdot f(A,x)_{j_0} \cdot \diag (x) \cdot \frac{\d}{\d A_{j_1,i_1}} ( A^{\top})  \cdot  f(A,x)  \cdot    \langle c(A, x), f(A, x) \rangle \\
     = & ~ B_{4,3,1}^{j_1,i_1,j_0,i_0}  
\end{align*} 
\item {\bf Part 4.} For $B_{4,4}^{j_1,i_1,j_0,i_0}$, we have 
\begin{align*}
  B_{4,4}^{j_1,i_1,j_0,i_0} = & ~ - c_g(A,x)^{\top} \cdot f(A,x)_{j_0} \cdot \diag (x) A^{\top} \cdot  \frac{\d}{\d A_{j_1,i_1}} ( f(A,x)  )  \cdot    \langle c(A, x), f(A, x) \rangle\\
     = & ~ B_{4,4,1}^{j_1,i_1,j_0,i_0} 
\end{align*}
\item {\bf Part 5.} For $B_{4,5}^{j_1,i_1,j_0,i_0}$, we have 
\begin{align*}
  B_{4,5}^{j_1,i_1,j_0,i_0} = & ~ -  c_g(A,x)^{\top} \cdot f(A,x)_{j_0}  \cdot \diag (x) A^{\top} \cdot  f(A,x)  \cdot  \langle \frac{\d c(A,x)}{\d A_{j_1,i_1}}, f(A,x) \rangle  \\
     = & ~ B_{4,5,1}^{j_1,i_1,j_0,i_0}  + B_{4,5,2}^{j_1,i_1,j_0,i_0}  
\end{align*}
\item {\bf Part 6.} For $B_{4,6}^{j_1,i_1,j_0,i_0}$, we have 
\begin{align*}
  B_{4,6}^{j_1,i_1,j_0,i_0} = & ~ -  c_g(A,x)^{\top} \cdot f(A,x)_{j_0}  \cdot \diag (x) A^{\top} \cdot  f(A,x)  \cdot  \langle c(A,x),  \frac{\d f(A,x)}{\d A_{j_1,i_1}} \rangle  \\
     = & ~ B_{4,6,1}^{j_1,i_1,j_0,i_0}  + B_{4,6,2}^{j_1,i_1,j_0,i_0}  
\end{align*}
\end{itemize}
\begin{proof}
    {\bf Proof of Part 1.}
    \begin{align*}
    B_{4,1,1}^{j_1,i_1,j_0,i_0} : = & ~ e_{i_1}^\top  \cdot  (\langle c(A,x), f(A,x) \rangle)^2  \cdot  f(A,x)_{j_1} \cdot f(A,x)_{j_0} \cdot \diag (x) A^{\top} \cdot  f(A,x) \\
    B_{4,1,2}^{j_1,i_1,j_0,i_0} : = & ~ e_{i_1}^\top  \cdot  c(A,x)_{j_1}  \cdot  f(A,x)_{j_1} \cdot f(A,x)_{j_0}  \cdot \diag (x) A^{\top} \cdot  f(A,x)  \cdot    \langle c(A, x), f(A, x) \rangle\\
    B_{4,1,3}^{j_1,i_1,j_0,i_0} : = & ~f(A,x)_{j_1} \cdot f(A,x)_{j_0}  \cdot  (\langle c(A,x), f(A,x) \rangle)^2  \cdot (  (A_{j_1,*})  \circ  x^\top  ) \cdot \diag (x) A^{\top} \cdot  f(A,x)  \\
    B_{4,1,4}^{j_1,i_1,j_0,i_0} : = & ~  -   f(A,x)_{j_1} \cdot f(A,x)_{j_0}  \cdot  f(A,x)^\top   \cdot  A  \cdot   (\diag(x))^2  \cdot  (\langle c(A,x), f(A,x) \rangle)^2 \cdot   A^{\top} \cdot  f(A,x) \\
    B_{4,1,5}^{j_1,i_1,j_0,i_0} : = & ~    f(A,x)_{j_1} \cdot f(A,x)_{j_0}  \cdot f(A,x)^\top \cdot A \cdot (\diag(x))^2 \cdot (\langle -f(A,x), f(A,x) \rangle + f(A,x)_{j_1}) \\
    & ~ \cdot A^{\top} \cdot  f(A,x)  \cdot    \langle c(A, x), f(A, x) \rangle\\
    B_{4,1,6}^{j_1,i_1,j_0,i_0} : = & ~  f(A,x)_{j_1} \cdot f(A,x)_{j_0} \cdot f(A,x)^\top  \cdot A \cdot (\diag(x))^2 \cdot(\langle -f(A,x), c(A,x) \rangle + f(A,x)_{j_1})\\
    & ~ \cdot   A^{\top} \cdot  f(A,x)  \cdot    \langle c(A, x), f(A, x) \rangle\\
    B_{4,1,7}^{j_1,i_1,j_0,i_0} : = & ~ f(A,x)_{j_1} \cdot f(A,x)_{j_0} \cdot (( e_{j_1}^\top  - f(A,x)^\top) \circ q(A,x)^\top) \cdot A \cdot  (\diag(x))^2 \cdot A^{\top} \cdot  f(A,x) \\
          & ~ \cdot    \langle c(A, x), f(A, x) \rangle
\end{align*}
Finally, combine them and we have
\begin{align*}
       B_{4,1}^{j_1,i_1,j_0,i_0} = B_{4,1,1}^{j_1,i_1,j_0,i_0} + B_{4,1,2}^{j_1,i_1,j_0,i_0} + B_{4,1,3}^{j_1,i_1,j_0,i_0} + B_{4,1,4}^{j_1,i_1,j_0,i_0} + B_{4,1,5}^{j_1,i_1,j_0,i_0} + B_{4,1,6}^{j_1,i_1,j_0,i_0} + B_{4,1,7}^{j_1,i_1,j_0,i_0}
\end{align*}
{\bf Proof of Part 2.}
    \begin{align*}
    B_{4,2,1}^{j_1,i_1,j_0,i_0} : = & ~   c_g(A,x)^{\top} \cdot f(A,x)_{j_1} \cdot f(A,x)_{j_0} \cdot x_{i_1} \cdot \diag (x) A^{\top} \cdot  f(A,x)  \cdot    \langle c(A, x), f(A, x) \rangle 
\end{align*}
Finally, combine them and we have
\begin{align*}
       B_{4,2}^{j_1,i_1,j_0,i_0} = B_{4,2,1}^{j_1,i_1,j_0,i_0}  
\end{align*}
{\bf Proof of Part 3.} 
    \begin{align*}
    B_{4,3,1}^{j_1,i_1,j_0,i_0} : = & ~    - c_g(A,x)^{\top} \cdot f(A,x)_{j_0} \cdot \diag (x) \cdot e_{i_1} \cdot  e_{j_1}^\top \cdot  f(A,x)  \cdot    \langle c(A, x), f(A, x) \rangle
\end{align*}
Finally, combine them and we have
\begin{align*}
       B_{4,3}^{j_1,i_1,j_0,i_0} = B_{4,3,1}^{j_1,i_1,j_0,i_0} 
\end{align*}
{\bf Proof of Part 4.} 
    \begin{align*}
    B_{4,4,1}^{j_1,i_1,j_0,i_0} : = & ~  - c_g(A,x)^{\top} \cdot f(A,x)_{j_1} \cdot f(A,x)_{j_0}   \cdot \diag (x) \cdot A^{\top} \cdot x_{i_1}   \cdot (e_{j_1}- f(A,x) ) \cdot    \langle c(A, x), f(A, x) \rangle
\end{align*}
Finally, combine them and we have
\begin{align*}
       B_{4,4}^{j_1,i_1,j_0,i_0} = B_{4,4,1}^{j_1,i_1,j_0,i_0}  
\end{align*}
{\bf Proof of Part 5.} 
    \begin{align*}
    B_{4,5,1}^{j_1,i_1,j_0,i_0} : = & ~ -  c_g(A,x)^{\top} \cdot f(A,x)_{j_1} \cdot f(A,x)_{j_0}  \cdot \diag (x) A^{\top} \cdot  f(A,x)  \cdot x_{i_1} \cdot \langle - f(A,x), f(A,x) \rangle\\
    B_{4,5,2}^{j_1,i_1,j_0,i_0} : = & ~ -  c_g(A,x)^{\top} \cdot f(A,x)_{j_1}^2  \cdot f(A,x)_{j_0}  \cdot \diag (x) A^{\top} \cdot  f(A,x)  \cdot  x_{i_1}
\end{align*}
Finally, combine them and we have
\begin{align*}
       B_{4,5}^{j_1,i_1,j_0,i_0} = B_{4,5,1}^{j_1,i_1,j_0,i_0}  +B_{4,5,2}^{j_1,i_1,j_0,i_0}
\end{align*}
{\bf Proof of Part 6.} 
    \begin{align*}
    B_{4,6,1}^{j_1,i_1,j_0,i_0} : = & ~ -  c_g(A,x)^{\top} \cdot f(A,x)_{j_1} \cdot f(A,x)_{j_0}  \cdot \diag (x) A^{\top} \cdot  f(A,x)  \cdot x_{i_1} \cdot (\langle - f(A,x), c(A,x) \rangle )\\
    B_{4,6,2}^{j_1,i_1,j_0,i_0} : = & ~ -  c_g(A,x)^{\top} \cdot f(A,x)_{j_1} \cdot f(A,x)_{j_0}  \cdot \diag (x) A^{\top} \cdot  f(A,x)  \cdot   x_{i_1} \cdot c(A,x)_{j_1} 
\end{align*}
Finally, combine them and we have
\begin{align*}
       B_{4,6}^{j_1,i_1,j_0,i_0} = B_{4,6,1}^{j_1,i_1,j_0,i_0}  +B_{4,6,2}^{j_1,i_1,j_0,i_0}
\end{align*}
\end{proof}
\end{lemma}

\subsection{Constructing \texorpdfstring{$d \times d$}{} matrices for \texorpdfstring{$j_1 \neq j_0$}{}}
The purpose of the following lemma is to let $i_0$ and $i_1$ disappear.
\begin{lemma}For $j_0,j_1 \in [n]$, a list of $d \times d$ matrices can be expressed as the following sense,\label{lem:b_4_j1_j0}
\begin{itemize}
\item {\bf Part 1.}
\begin{align*}
B_{4,1,1}^{j_1,*,j_0,*} & ~ =  f_c(A,x)^2 \cdot f(A,x)_{j_1} \cdot f(A,x)_{j_0} \cdot  h(A,x) \cdot  {\bf 1}_d^\top
\end{align*}
\item {\bf Part 2.}
\begin{align*}
B_{4,1,2}^{j_1,*,j_0,*} & ~ =   f(A,x)_{j_1} \cdot f(A,x)_{j_0} \cdot c(A,x)_{j_1} \cdot  f_c(A,x) \cdot h(A,x) \cdot  {\bf 1}_d^\top 
\end{align*}
\item {\bf Part 3.}
\begin{align*}
B_{4,1,3}^{j_1,*,j_0,*} & ~ =   f(A,x)_{j_1} \cdot f(A,x)_{j_0} \cdot f_c(A,x)^2  \cdot ( (A_{j_1,*}) \circ x^\top  ) \cdot  h(A,x) \cdot I_d
\end{align*}
\item {\bf Part 4.}
\begin{align*}
B_{4,1,4}^{j_1,*,j_0,*}  & ~ =  -  f(A,x)_{j_1} \cdot f(A,x)_{j_0} \cdot f_c(A,x)^2 \cdot h(A,x)^\top \cdot h(A,x) \cdot I_d
\end{align*}
\item {\bf Part 5.}
\begin{align*}
B_{4,1,5}^{j_1,*,j_0,*}  & ~ = f(A,x)_{j_1} \cdot f(A,x)_{j_0} \cdot (-f_2(A,x) + f(A,x)_{j_1})  \cdot f_c(A,x) \cdot  h(A,x)^\top \cdot h(A,x) \cdot I_d 
\end{align*}
\item {\bf Part 6.}
\begin{align*}
B_{4,1,6}^{j_1,*,j_0,*}  & ~ =     f(A,x)_{j_1} \cdot f(A,x)_{j_0} \cdot (-f_c(A,x) + f(A,x)_{j_1}) \cdot f_c(A,x) \cdot  h(A,x)^\top \cdot h(A,x) \cdot I_d
\end{align*}
\item {\bf Part 7.}
\begin{align*}
B_{4,1,7}^{j_1,*,j_0,*}  & ~ =  f(A,x)_{j_1} \cdot f(A,x)_{j_0} \cdot f_c(A,x)  \cdot p_{j_1}(A,x)^\top \cdot  h(A,x) \cdot I_d 
\end{align*}
\item {\bf Part 8.}
\begin{align*}
B_{4,2,1}^{j_1,*,j_0,*}  & ~ =    f(A,x)_{j_1} \cdot f(A,x)_{j_0}  \cdot f_c(A,x) \cdot  c_g(A,x)^\top  \cdot  h(A,x)  \cdot x \cdot {\bf 1}_d^{\top} 
\end{align*}
\item {\bf Part 9.}
\begin{align*}
 B_{4,3,1}^{j_1,*,j_0,*}  & ~ =     -  f(A,x)_{j_1} \cdot   f(A,x)_{j_0}  \cdot    f_c(A,x)   \cdot {\bf 1}_d \cdot c_g(A,x)^{\top} \cdot \diag (x)
\end{align*}
\item {\bf Part 10.}
\begin{align*}
B_{4,4,1}^{j_1,*,j_0,*}  =    -   f(A,x)_{j_1} \cdot f(A,x)_{j_0}   \cdot f_c(A,x) \cdot  c_g(A,x)^\top \cdot h_e(A,x) \cdot x \cdot {\bf 1}_d^{\top}
\end{align*}
\item {\bf Part 11.}
\begin{align*}
B_{4,5,1}^{j_1,*,j_0,*}  & ~ =   f(A,x)_{j_1} \cdot f(A,x)_{j_0} \cdot  f_2(A,x)  \cdot  c_g(A,x)^{\top} \cdot  h(A,x) \cdot x \cdot {\bf 1}_d^{\top} 
\end{align*}
\item {\bf Part 12.}
\begin{align*}
 B_{4,5,2}^{j_1,*,j_0,*}  =   f(A,x)_{j_1}^2 \cdot f(A,x)_{j_0} \cdot  c_g(A,x)^{\top} \cdot h(A,x) \cdot x \cdot {\bf 1}_d^{\top}  
\end{align*}
\item {\bf Part 13.}
\begin{align*}
B_{4,6,1}^{j_1,*,j_0,*}  =  f(A,x)_{j_1} \cdot f(A,x)_{j_0} \cdot  f_c(A,x)  \cdot  c_g(A,x)^{\top} \cdot  h(A,x) \cdot x \cdot {\bf 1}_d^{\top}  
\end{align*}
\item {\bf Part 14.}
\begin{align*}
B_{4,6,2}^{j_1,*,j_0,*}  =   f(A,x)_{j_1} \cdot f(A,x)_{j_0} \cdot c(A,x)_{j_1} \cdot  c_g(A,x)^{\top} \cdot h(A,x) \cdot x \cdot {\bf 1}_d^{\top}  
\end{align*}

\end{itemize}
\begin{proof}
{\bf Proof of Part 1.}
    We have
    \begin{align*}
        B_{4,1,1}^{j_1,i_1,j_0,i_1}  = & ~ e_{i_1}^\top  \cdot  (\langle c(A,x), f(A,x) \rangle)^2  \cdot    f(A,x)_{j_1} \cdot f(A,x)_{j_0} \cdot \diag (x) A^{\top} \cdot  f(A,x)\\
        B_{4,1,1}^{j_1,i_1,j_0,i_0}  = & ~ e_{i_1}^\top  \cdot  (\langle c(A,x), f(A,x) \rangle)^2  \cdot    f(A,x)_{j_1} \cdot f(A,x)_{j_0} \cdot \diag (x) A^{\top} \cdot  f(A,x)
    \end{align*}
    From the above two equations, we can tell that $B_{4,1,1}^{j_1,*,j_1,*} \in \R^{d \times d}$ is a matrix that both the diagonal and off-diagonal have entries.
    
    Then we have $B_{4,1,1}^{j_1,*,j_0,*} \in \R^{d \times d}$ can be written as the rescaling of a diagonal matrix,
    \begin{align*}
     B_{4,1,1}^{j_1,*,j_0,*} & ~ = (\langle c(A,x), f(A,x) \rangle)^2 \cdot   f(A,x)_{j_1} \cdot f(A,x)_{j_0} \cdot \diag (x) A^{\top} \cdot  f(A,x) \cdot  {\bf 1}_d^\top \\
     & ~ = f_c(A,x)^2 \cdot  f(A,x)_{j_1} \cdot f(A,x)_{j_0} \cdot  h(A,x) \cdot  {\bf 1}_d^\top
\end{align*}
    where the last step is follows from the Definitions~\ref{def:h} and Definitions~\ref{def:f_c}. 

{\bf Proof of Part 2.}
    We have
    \begin{align*}
           B_{4,1,2}^{j_1,i_1,j_0,i_1} = & ~ e_{i_1}^\top  \cdot  c(A,x)_{j_1}  \cdot  f(A,x)_{j_1} \cdot f(A,x)_{j_0}  \cdot \diag (x) A^{\top} \cdot  f(A,x)  \cdot    \langle c(A, x), f(A, x) \rangle\\
        B_{4,1,2}^{j_1,i_1,j_0,i_0} = & ~ e_{i_1}^\top  \cdot  c(A,x)_{j_1}  \cdot  f(A,x)_{j_1} \cdot f(A,x)_{j_0}  \cdot \diag (x) A^{\top} \cdot  f(A,x)  \cdot    \langle c(A, x), f(A, x) \rangle
    \end{align*}
     From the above two equations, we can tell that $B_{4,1,2}^{j_1,*,j_1,*} \in \R^{d \times d}$ is a matrix that only diagonal has entries and off-diagonal are all zeros.
    
    Then we have $B_{4,1,2}^{j_1,*,j_0,*} \in \R^{d \times d}$ can be written as the rescaling of a diagonal matrix,
\begin{align*}
     B_{4,1,2}^{j_1,*,j_0,*} & ~ = f(A,x)_{j_1} \cdot f(A,x)_{j_0} \cdot c(A,x)_{j_1} \cdot  \langle c(A,x), f(A,x) \rangle \cdot \diag (x) A^{\top} \cdot  f(A,x)  \cdot  {\bf 1}_d^\top   \\
     & ~ = f(A,x)_{j_1} \cdot f(A,x)_{j_0} \cdot c(A,x)_{j_1} \cdot  f_c(A,x) \cdot h(A,x) \cdot  {\bf 1}_d^\top 
\end{align*}
    where the last step is follows from the Definitions~\ref{def:h} and Definitions~\ref{def:f_c}.

{\bf Proof of Part 3.}
We have for diagonal entry and off-diagonal entry can be written as follows 
    \begin{align*}
        B_{4,1,3}^{j_1,i_1,j_0,i_1} = & ~f(A,x)_{j_1} \cdot f(A,x)_{j_0}  \cdot  (\langle c(A,x), f(A,x) \rangle)^2  \cdot (  (A_{j_1,*})  \circ  x^\top  ) \cdot \diag (x) A^{\top} \cdot  f(A,x) \\
        B_{4,1,3}^{j_1,i_1,j_0,i_0} = & ~f(A,x)_{j_1} \cdot f(A,x)_{j_0}  \cdot  (\langle c(A,x), f(A,x) \rangle)^2  \cdot (  (A_{j_1,*})  \circ  x^\top  ) \cdot \diag (x) A^{\top} \cdot  f(A,x) 
    \end{align*}
From the above equation, we can show that matrix $B_{4,1,3}^{j_1,*,j_0,*}$ can be expressed as a rank-$1$ matrix,
\begin{align*}
     B_{4,1,3}^{j_1,*,j_0,*} & ~ = f(A,x)_{j_1} \cdot f(A,x)_{j_0} \cdot (\langle c(A,x), f(A,x) \rangle)^2 \cdot ( (A_{j_1,*}) \circ x^\top  ) \cdot  \diag (x) A^{\top} \cdot  f(A,x) \cdot I_d \\
     & ~ =  f(A,x)_{j_1} \cdot f(A,x)_{j_0} \cdot f_c(A,x)^2  \cdot ( (A_{j_1,*}) \circ x^\top  ) \cdot  h(A,x) \cdot I_d
\end{align*}
    where the last step is follows from the Definitions~\ref{def:h} and Definitions~\ref{def:f_c}.

{\bf Proof of Part 4.}
We have for diagonal entry and off-diagonal entry can be written as follows
    \begin{align*}
        B_{4,1,4}^{j_1,i_1,j_0,i_1}   = & ~ -   f(A,x)_{j_1} \cdot f(A,x)_{j_0}  \cdot  f(A,x)^\top   \cdot  A  \cdot   (\diag(x))^2  \cdot  (\langle c(A,x), f(A,x) \rangle)^2 \cdot   A^{\top} \cdot  f(A,x) \\
        B_{4,1,4}^{j_1,i_1,j_0,i_0}   = & ~ -   f(A,x)_{j_1} \cdot f(A,x)_{j_0}  \cdot  f(A,x)^\top   \cdot  A  \cdot   (\diag(x))^2  \cdot  (\langle c(A,x), f(A,x) \rangle)^2 \cdot   A^{\top} \cdot  f(A,x) 
    \end{align*}
 From the above equation, we can show that matrix $B_{4,1,4}^{j_1,*,j_0,*}$ can be expressed as a rank-$1$ matrix,
\begin{align*}
    B_{4,1,4}^{j_1,*,j_0,*}  & ~ =  -  f(A,x)_{j_1} \cdot f(A,x)_{j_0} \cdot (\langle c(A,x), f(A,x) \rangle )^2 \cdot f(A,x)^\top  \cdot A \cdot (\diag (x))^2 \cdot A^{\top} \cdot  f(A,x)\cdot I_d \\
     & ~ =  -  f(A,x)_{j_1} \cdot f(A,x)_{j_0} \cdot f_c(A,x)^2 \cdot h(A,x)^\top \cdot h(A,x) \cdot I_d
\end{align*}
    where the last step is follows from the Definitions~\ref{def:h} and Definitions~\ref{def:f_c}.

{\bf Proof of Part 5.}
We have for diagonal entry and off-diagonal entry can be written as follows
    \begin{align*}
         B_{4,1,5}^{j_1,i_1,j_0,i_0} = & ~    f(A,x)_{j_1} \cdot f(A,x)_{j_0} \cdot f(A,x)^\top \cdot A \cdot (\diag(x))^2 \cdot (\langle -f(A,x), f(A,x) \rangle + f(A,x)_{j_1}) \cdot A^{\top} \\
         & ~ \cdot  f(A,x)  \cdot    \langle c(A, x), f(A, x) \rangle \\
         B_{4,1,5}^{j_1,i_1,j_0,i_0} = & ~    f(A,x)_{j_1} \cdot f(A,x)_{j_0}  \cdot f(A,x)^\top \cdot A \cdot (\diag(x))^2 \cdot (\langle -f(A,x), f(A,x) \rangle + f(A,x)_{j_1}) \cdot A^{\top}\\
         & ~ \cdot  f(A,x)  \cdot  \langle c(A, x), f(A, x) \rangle
    \end{align*}
    From the above equation, we can show that matrix $B_{4,1,5}^{j_1,*,j_0,*}$ can be expressed as a rank-$1$ matrix,
\begin{align*}
    B_{4,1,5}^{j_1,*,j_0,*}  & ~ =  f(A,x)_{j_1} \cdot f(A,x)_{j_0} \cdot (\langle -f(A,x), f(A,x) \rangle + f(A,x)_{j_1})  \cdot \langle c(A,x), f(A,x) \rangle \cdot  f(A,x)^\top  \cdot A \\
        & ~\cdot (\diag (x))^2 \cdot A^{\top} \cdot  f(A,x)\cdot I_d\\
     & ~ =  f(A,x)_{j_1} \cdot f(A,x)_{j_0} \cdot (-f_2(A,x) + f(A,x)_{j_1})  \cdot f_c(A,x) \cdot  h(A,x)^\top \cdot h(A,x) \cdot I_d
\end{align*}
    where the last step is follows from the Definitions~\ref{def:h}, Definitions~\ref{def:f_c} and Definitions~\ref{def:f_2}.

{\bf Proof of Part 6.}
We have for diagonal entry and off-diagonal entry can be written as follows
    \begin{align*}
        B_{4,1,6}^{j_1,i_1,j_0,i_1}  = & ~   f(A,x)_{j_1} \cdot f(A,x)_{j_0} \cdot f(A,x)^\top  \cdot A \cdot (\diag(x))^2 \cdot(\langle -f(A,x), c(A,x) \rangle + f(A,x)_{j_1}) \cdot   A^{\top}\\
        & ~ \cdot  f(A,x)  \cdot    \langle c(A, x), f(A, x) \rangle\\
        B_{4,1,6}^{j_1,i_1,j_0,i_0}  = & ~   f(A,x)_{j_1} \cdot f(A,x)_{j_0} \cdot f(A,x)^\top  \cdot A \cdot (\diag(x))^2 \cdot(\langle -f(A,x), c(A,x) \rangle + f(A,x)_{j_1}) \cdot   A^{\top}\\
        & ~ \cdot  f(A,x)  \cdot    \langle c(A, x), f(A, x) \rangle
    \end{align*}
    From the above equation, we can show that matrix $B_{4,1,6}^{j_1,*,j_0,*}$ can be expressed as a rank-$1$ matrix,
\begin{align*}
    B_{4,1,6}^{j_1,*,j_0,*}  & ~ =   f(A,x)_{j_1} \cdot f(A,x)_{j_0} \cdot (\langle -f(A,x), c(A,x) \rangle + f(A,x)_{j_1}) \cdot \langle c(A,x), f(A,x) \rangle\\
        & ~ \cdot  f(A,x)^\top  \cdot A \cdot (\diag (x))^2 \cdot A^{\top} \cdot  f(A,x)\cdot I_d\\
     & ~ = f(A,x)_{j_1} \cdot f(A,x)_{j_0} \cdot (-f_c(A,x) + f(A,x)_{j_1}) \cdot f_c(A,x) \cdot  h(A,x)^\top \cdot h(A,x) \cdot I_d
\end{align*}
    where the last step is follows from the Definitions~\ref{def:h} and Definitions~\ref{def:f_c} .
    
{\bf Proof of Part 7.}
We have for diagonal entry and off-diagonal entry can be written as follows
    \begin{align*}
         B_{4,1,7}^{j_1,i_1,j_0,i_1} = & ~ f(A,x)_{j_1} \cdot f(A,x)_{j_0} \cdot (( e_{j_1}^\top  - f(A,x)^\top) \circ q(A,x)^\top) \cdot A \cdot  (\diag(x))^2 \cdot A^{\top} \cdot  f(A,x) \\
        & ~ \cdot    \langle c(A, x), f(A, x) \rangle\\
         B_{4,1,7}^{j_1,i_1,j_0,i_0} = & ~ f(A,x)_{j_1} \cdot f(A,x)_{j_0} \cdot (( e_{j_1}^\top  - f(A,x)^\top) \circ q(A,x)^\top) \cdot A \cdot  (\diag(x))^2 \cdot A^{\top} \cdot  f(A,x) \\
        & ~ \cdot    \langle c(A, x), f(A, x) \rangle
    \end{align*}
    From the above equation, we can show that matrix $B_{4,1,7}^{j_1,*,j_0,*}$ can be expressed as a rank-$1$ matrix,
\begin{align*}
     B_{4,1,7}^{j_1,*,j_0,*}  & ~ =   f(A,x)_{j_1} \cdot f(A,x)_{j_0} \cdot \langle c(A,x), f(A,x) \rangle  \cdot  ((e_{j_1}^\top - f(A,x)^\top) \circ q(A,x)^\top) \\
     & ~ \cdot A \cdot  (\diag (x))^2 \cdot A^{\top} \cdot  f(A,x) \cdot I_d \\
     & ~ =f(A,x)_{j_1} \cdot f(A,x)_{j_0} \cdot f_c(A,x)  \cdot p_{j_1}(A,x)^\top \cdot  h(A,x) \cdot I_d 
\end{align*}
    where the last step is follows from the Definitions~\ref{def:h}, Definitions~\ref{def:f_c} and Definitions~\ref{def:p}.
    
{\bf Proof of Part 8.}
We have for diagonal entry and off-diagonal entry can be written as follows
    \begin{align*}
         B_{4,2,1}^{j_1,i_1,j_0,i_0} = & ~  c_g(A,x)^{\top} \cdot f(A,x)_{j_1} \cdot f(A,x)_{j_0} \cdot x_{i_1} \cdot \diag (x) A^{\top} \cdot  f(A,x)  \cdot    \langle c(A, x), f(A, x) \rangle \\
         B_{4,2,1}^{j_1,i_1,j_0,i_0} = & ~  c_g(A,x)^{\top} \cdot f(A,x)_{j_1} \cdot f(A,x)_{j_0} \cdot x_{i_1} \cdot \diag (x) A^{\top} \cdot  f(A,x)  \cdot    \langle c(A, x), f(A, x) \rangle
    \end{align*}
    From the above equation, we can show that matrix $B_{4,2,1}^{j_1,*,j_0,*}$ can be expressed as a rank-$1$ matrix,
\begin{align*}
     B_{4,2,1}^{j_1,*,j_0,*}  & ~ =   f(A,x)_{j_1} \cdot f(A,x)_{j_0}  \cdot \langle c(A,x), f(A,x) \rangle \cdot c_g(A,x)^{\top} \cdot   \diag (x) \cdot A^{\top} \cdot  f(A,x) \cdot x \cdot {\bf 1}_d^{\top}  \\
     & ~ =f(A,x)_{j_1} \cdot f(A,x)_{j_0}  \cdot f_c(A,x) \cdot  c_g(A,x)^\top  \cdot  h(A,x)  \cdot x \cdot {\bf 1}_d^{\top} 
\end{align*}
    where the last step is follows from the Definitions~\ref{def:h} and Definitions~\ref{def:f_c}.

{\bf Proof of Part 9.}
We have for diagonal entry and off-diagonal entry can be written as follows
    \begin{align*}
         B_{4,3,1}^{j_1,i_1,j_0,i_1}  = & ~   -  c_g(A,x)^{\top}  \cdot f(A,x)_{j_0} \cdot  \diag (x)  \cdot e_{i_1}  \cdot   e_{j_1}^\top  \cdot  f(A,x)   \cdot    \langle c(A, x), f(A, x) \rangle\\
         B_{4,3,1}^{j_1,i_1,j_0,i_0}  = & ~  - c_g(A,x)^{\top} \cdot f(A,x)_{j_0} \cdot \diag (x) \cdot e_{i_1} \cdot  e_{j_1}^\top \cdot  f(A,x)  \cdot    \langle c(A, x), f(A, x) \rangle
    \end{align*}
            From the above equation, we can show that matrix $B_{4,3,1}^{j_1,*,j_1,*}$ can be expressed as a rank-$1$ matrix,
\begin{align*}
    B_{4,3,1}^{j_1,*,j_1,*}  & ~ =  -  f(A,x)_{j_1} \cdot f(A,x)_{j_0}   \cdot    \langle c(A, x), f(A, x) \rangle   \cdot {\bf 1}_d \cdot c_g(A,x)^{\top} \cdot \diag (x)  \\
     & ~ =    -  f(A,x)_{j_1} \cdot f(A,x)_{j_0}   \cdot    f_c(A,x)   \cdot {\bf 1}_d \cdot c_g(A,x)^{\top} \cdot \diag (x)
\end{align*} 
    where the last step is follows from the  Definitions~\ref{def:f_c}.

{\bf Proof of Part 10.}
We have for diagonal entry and off-diagonal entry can be written as follows
    \begin{align*}
         B_{4,4,1}^{j_1,i_1,j_0,i_1}   = & ~   -  c_g(A,x)^{\top}  \cdot f(A,x)_{j_1} \cdot f(A,x)_{j_0}  \cdot  \diag (x)  \cdot  A^{\top}  \cdot x_{i_1}   \cdot (e_{j_1}- f(A,x) ) \cdot    \langle c(A, x), f(A, x) \rangle\\
         B_{4,4,1}^{j_1,i_1,j_0,i_0}   = & ~  - c_g(A,x)^{\top} \cdot f(A,x)_{j_1} \cdot f(A,x)_{j_0}  \cdot \diag (x) \cdot A^{\top} \cdot x_{i_1}   \cdot (e_{j_1}- f(A,x) ) \cdot    \langle c(A, x), f(A, x) \rangle
    \end{align*}
            From the above equation, we can show that matrix $B_{4,4,1}^{j_1,*,j_0,*}$ can be expressed as a rank-$1$ matrix,
\begin{align*}
    B_{4,4,1}^{j_1,*,j_0,*} & ~ =   -   f(A,x)_{j_1} \cdot f(A,x)_{j_0}   \cdot \langle c(A, x), f(A, x) \rangle \cdot  c_g(A,x)^\top \cdot  \diag (x) \cdot A^{\top}     \cdot (e_{j_1}- f(A,x) )  \cdot x \cdot {\bf 1}_d^{\top} \\
     & ~ =  -  f(A,x)_{j_1} \cdot f(A,x)_{j_0}   \cdot f_c(A,x) \cdot  c_g(A,x)^\top \cdot h_e(A,x) \cdot x \cdot {\bf 1}_d^{\top}
\end{align*}
    where the last step is follows from the Definitions~\ref{def:h_e} and Definitions~\ref{def:f_c}.

{\bf Proof of Part 11.}
We have for diagonal entry and off-diagonal entry can be written as follows
    \begin{align*}
         B_{4,5,1}^{j_1,i_1,j_0,i_0}   = & ~    -  c_g(A,x)^{\top} \cdot f(A,x)_{j_1} \cdot f(A,x)_{j_0}  \cdot \diag (x) A^{\top} \cdot  f(A,x)  \cdot x_{i_1} \cdot \langle - f(A,x), f(A,x) \rangle\\
         B_{4,5,1}^{j_1,i_1,j_0,i_0}   = & ~   -  c_g(A,x)^{\top} \cdot f(A,x)_{j_1} \cdot f(A,x)_{j_0}  \cdot \diag (x) A^{\top} \cdot  f(A,x)  \cdot x_{i_1} \cdot \langle - f(A,x), f(A,x) \rangle
    \end{align*}
            From the above equation, we can show that matrix $B_{4,5,1}^{j_1,*,j_0,*}$ can be expressed as a rank-$1$ matrix,
\begin{align*}
    B_{4,5,1}^{j_1,*,j_0,*}  & ~ =   f(A,x)_{j_1} \cdot f(A,x)_{j_0} \cdot  \langle f(A,x), f(A,x) \rangle  \cdot  c_g(A,x)^{\top} \cdot  \diag (x) A^{\top} \cdot  f(A,x) \cdot x \cdot {\bf 1}_d^{\top}  \\
     & ~ =  f(A,x)_{j_1} \cdot f(A,x)_{j_0} \cdot  f_2(A,x)  \cdot  c_g(A,x)^{\top} \cdot  h(A,x) \cdot x \cdot {\bf 1}_d^{\top} 
\end{align*}
    where the last step is follows from the Definitions~\ref{def:h} and Definitions~\ref{def:f_2}.

{\bf Proof of Part 12.}
We have for diagonal entry and off-diagonal entry can be written as follows
    \begin{align*}
         B_{4,5,2}^{j_1,i_1,j_0,i_0}   = & ~    -  c_g(A,x)^{\top} \cdot f(A,x)_{j_1}^2 \cdot f(A,x)_{j_0}  \cdot \diag (x) A^{\top} \cdot  f(A,x)  \cdot  x_{i_1}\\
         B_{4,5,2}^{j_1,i_1,j_0,i_0}   = & ~   -  c_g(A,x)^{\top} \cdot f(A,x)_{j_1}^2 \cdot f(A,x)_{j_0}  \cdot \diag (x) A^{\top} \cdot  f(A,x)  \cdot  x_{i_1}
    \end{align*}
            From the above equation, we can show that matrix $B_{4,5,2}^{j_1,*,j_0,*}$ can be expressed as a rank-$1$ matrix,
\begin{align*}
    B_{4,5,2}^{j_1,*,j_0,*}  & ~ =   f(A,x)_{j_1}^2 \cdot f(A,x)_{j_0} \cdot  c_g(A,x)^{\top} \cdot  \diag (x) A^{\top} \cdot  f(A,x) \cdot x \cdot {\bf 1}_d^{\top}  \\
     & ~ =   f(A,x)_{j_1}^2 \cdot f(A,x)_{j_0} \cdot  c_g(A,x)^{\top} \cdot h(A,x) \cdot x \cdot {\bf 1}_d^{\top} 
\end{align*}
    where the last step is follows from the Definitions~\ref{def:h}.

    {\bf Proof of Part 13.}
We have for diagonal entry and off-diagonal entry can be written as follows
    \begin{align*}
         B_{4,6,1}^{j_1,i_1,j_0,i_0}   = & ~     -  c_g(A,x)^{\top} \cdot  f(A,x)_{j_1} \cdot f(A,x)_{j_0}  \cdot \diag (x) A^{\top} \cdot  f(A,x)  \cdot x_{i_1} \cdot (\langle - f(A,x), c(A,x) \rangle )\\
         B_{4,6,1}^{j_1,i_1,j_0,i_0}   = & ~   -  c_g(A,x)^{\top} \cdot  f(A,x)_{j_1} \cdot f(A,x)_{j_0}  \cdot \diag (x) A^{\top} \cdot  f(A,x)  \cdot x_{i_1} \cdot (\langle - f(A,x), c(A,x) \rangle )
    \end{align*}
            From the above equation, we can show that matrix $B_{4,6,1}^{j_1,*,j_0,*}$ can be expressed as a rank-$1$ matrix,
\begin{align*}
    B_{4,6,1}^{j_1,*,j_0,*}  & ~ =    f(A,x)_{j_1} \cdot f(A,x)_{j_0} \cdot  \langle f(A,x), c(A,x) \rangle  \cdot  c_g(A,x)^{\top} \cdot  \diag (x) A^{\top} \cdot  f(A,x) \cdot x \cdot {\bf 1}_d^{\top}  \\
     & ~ =   f(A,x)_{j_1} \cdot f(A,x)_{j_0} \cdot  f_c(A,x)  \cdot  c_g(A,x)^{\top} \cdot  h(A,x) \cdot x \cdot {\bf 1}_d^{\top} 
\end{align*}
    where the last step is follows from the Definitions~\ref{def:h} and Definitions~\ref{def:f_c}.

 {\bf Proof of Part 14.}
We have for diagonal entry and off-diagonal entry can be written as follows
    \begin{align*}
         B_{4,6,2}^{j_1,i_1,j_0,i_0}   = & ~   -  c_g(A,x)^{\top} \cdot  f(A,x)_{j_1} \cdot f(A,x)_{j_0}  \cdot \diag (x) A^{\top} \cdot  f(A,x)  \cdot   x_{i_1} \cdot c(A,x)_{j_1} \\
         B_{4,6,2}^{j_1,i_1,j_0,i_0}   = & ~   -  c_g(A,x)^{\top} \cdot  f(A,x)_{j_1} \cdot f(A,x)_{j_0}  \cdot \diag (x) A^{\top} \cdot  f(A,x)  \cdot   x_{i_1} \cdot c(A,x)_{j_1} 
    \end{align*}
            From the above equation, we can show that matrix $B_{4,6,2}^{j_1,*,j_0,*}$ can be expressed as a rank-$1$ matrix,
\begin{align*}
    B_{4,6,2}^{j_1,*,j_0,*}  & ~ =    f(A,x)_{j_1} \cdot f(A,x)_{j_0} \cdot c(A,x)_{j_1} \cdot  c_g(A,x)^{\top} \cdot  \diag (x) A^{\top} \cdot  f(A,x) \cdot x \cdot {\bf 1}_d^{\top}  \\
     & ~ =    f(A,x)_{j_1} \cdot f(A,x)_{j_0} \cdot c(A,x)_{j_1} \cdot  c_g(A,x)^{\top} \cdot h(A,x) \cdot x \cdot {\bf 1}_d^{\top} 
\end{align*}
    where the last step is follows from the Definitions~\ref{def:h}.
\end{proof}
\end{lemma}
\subsection{Expanding \texorpdfstring{$B_4$}{} into many terms}

\begin{lemma}
   If the following conditions hold
    \begin{itemize}
     \item Let $u(A,x) \in \R^n$ be defined as Definition~\ref{def:u}
    \item Let $\alpha(A,x) \in \R$ be defined as Definition~\ref{def:alpha}
     \item Let $f(A,x) \in \R^n$ be defined as Definition~\ref{def:f}
    \item Let $c(A,x) \in \R^n$ be defined as Definition~\ref{def:c}
    \item Let $g(A,x) \in \R^d$ be defined as Definition~\ref{def:g} 
    \item Let $q(A,x) = c(A,x) + f(A,x) \in \R^n$
    \item Let $c_g(A,x) \in \R^d$ be defined as Definition~\ref{def:c_g}.
    \item Let $L_g(A,x) \in \R$ be defined as Definition~\ref{def:l_g}
    \item Let $v \in \R^n$ be a vector 
    \end{itemize}
Then, For $j_0,j_1 \in [n], i_0,i_1 \in [d]$, we have 
\begin{itemize}
    \item {\bf Part 1.}For $j_1 = j_0$ and $i_0 = i_1$
\begin{align*}
    B_{4}^{j_1,i_1,j_1,i_1}
    = & ~ B_{4,1}^{j_1,i_1,j_1,i_1}  + B_{4,2}^{j_1,i_1,j_1,i_1}  +B_{4,3}^{j_1,i_1,j_1,i_1}  +B_{4,4}^{j_1,i_1,j_1,i_1} +B_{4,5}^{j_1,i_1,j_1,i_1} +B_{4,6}^{j_1,i_1,j_1,i_1} 
\end{align*}
\item {\bf Part 2.}For $j_1 = j_0$ and $i_0 \neq i_1$
\begin{align*}
     B_{4}^{j_1,i_1,j_1,i_0}
    = & ~ B_{4,1}^{j_1,i_1,j_1,i_0}  + B_{4,2}^{j_1,i_1,j_1,i_0}  +B_{4,3}^{j_1,i_1,j_1,i_0}  +B_{4,4}^{j_1,i_1,j_1,i_0} +B_{4,5}^{j_1,i_1,j_1,i_0} +B_{4,6}^{j_1,i_1,j_1,i_0} 
\end{align*}
\item {\bf Part 3.}For $j_1 \neq j_0$ and $i_0 = i_1$ \begin{align*}
   B_{4}^{j_1,i_1,j_0,i_1}
    = & ~ B_{4,1}^{j_1,i_1,j_0,i_1}  + B_{4,2}^{j_1,i_1,j_0,i_1}  +B_{4,3}^{j_1,i_1,j_0,i_1}  +B_{4,4}^{j_1,i_1,j_0,i_1} +B_{4,5}^{j_1,i_1,j_0,i_1} +B_{4,6}^{j_1,i_1,j_0,i_1}
\end{align*}
\item {\bf Part 4.} For $j_0 \neq j_1$ and $i_0 \neq i_1$,
\begin{align*}
 B_{4}^{j_1,i_1,j_0,i_0}
    = & ~ B_{4,1}^{j_1,i_1,j_0,i_0}  + B_{4,2}^{j_1,i_1,j_0,i_0}  +B_{4,3}^{j_1,i_1,j_0,i_0}  +B_{4,4}^{j_1,i_1,j_0,i_0} +B_{4,5}^{j_1,i_1,j_0,i_0} +B_{4,6}^{j_1,i_1,j_0,i_0}
\end{align*}
\end{itemize}
\end{lemma}
\begin{proof}
{\bf Proof of Part 1.} we have
    \begin{align*}
    B_{4}^{j_1,i_1,j_1,i_1}
    = & ~ \frac{\d}{\d A_{j_1,i_1}}  (c_g(A,x)^{\top} \cdot f(A,x)_{j_1}\diag (x) \cdot A^{\top}   f(A,x)  \cdot    \langle c(A, x), f(A, x) \rangle ) \\
    = & ~B_{4,1}^{j_1,i_1,j_1,i_1}  + B_{4,2}^{j_1,i_1,j_1,i_1}  +B_{4,3}^{j_1,i_1,j_1,i_1}  +B_{4,4}^{j_1,i_1,j_1,i_1} +B_{4,5}^{j_1,i_1,j_1,i_1} +B_{4,6}^{j_1,i_1,j_1,i_1}
\end{align*}
 {\bf Proof of Part 2.} we have
    \begin{align*}
    B_{4}^{j_1,i_1,j_1,i_0}
    = & ~ \frac{\d}{\d A_{j_1,i_1}}  (c_g(A,x)^{\top} \cdot f(A,x)_{j_1}\diag (x) \cdot A^{\top}   f(A,x)  \cdot    \langle c(A, x), f(A, x) \rangle ) \\
    = & ~ B_{4,1}^{j_1,i_1,j_1,i_0}  + B_{4,2}^{j_1,i_1,j_1,i_0}  +B_{4,3}^{j_1,i_1,j_1,i_0}  +B_{4,4}^{j_1,i_1,j_1,i_0} +B_{4,5}^{j_1,i_1,j_1,i_0} +B_{4,6}^{j_1,i_1,j_1,i_0} 
\end{align*}
 {\bf Proof of Part 3.}  we have
\begin{align*}
     B_{4}^{j_1,i_1,j_0,i_1}
    = & ~ \frac{\d}{\d A_{j_1,i_1}}  (c_g(A,x)^{\top} \cdot f(A,x)_{j_0} \cdot \diag (x) \cdot A^{\top}   f(A,x)  \cdot    \langle c(A, x), f(A, x) \rangle ) \\
    = & ~B_{4,1}^{j_1,i_1,j_0,i_1}  + B_{4,2}^{j_1,i_1,j_0,i_1}  +B_{4,3}^{j_1,i_1,j_0,i_1}  +B_{4,4}^{j_1,i_1,j_0,i_1} +B_{4,5}^{j_1,i_1,j_0,i_1} +B_{4,6}^{j_1,i_1,j_0,i_1}
\end{align*}
{\bf Proof of Part 4.}
      we can have
\begin{align*}
     B_{4}^{j_1,i_1,j_0,i_0}
     =& ~ \frac{\d}{\d A_{j_1,i_1}}  (c_g(A,x)^{\top} \cdot f(A,x)_{j_0} \cdot \diag (x) \cdot A^{\top}   f(A,x)  \cdot    \langle c(A, x), f(A, x) \rangle ) \\
    = & ~B_{4,1}^{j_1,i_1,j_0,i_0}  + B_{4,2}^{j_1,i_1,j_0,i_0}  +B_{4,3}^{j_1,i_1,j_0,i_0}  +B_{4,4}^{j_1,i_1,j_0,i_0} +B_{4,5}^{j_1,i_1,j_0,i_0} +B_{4,6}^{j_1,i_1,j_0,i_0}
\end{align*}
\end{proof}
\subsection{Lipschitz Computation}
\begin{lemma}\label{lips: B_4}
If the following conditions hold
\begin{itemize}
    \item Let $B_{4,1,1}^{j_1,*, j_0,*}, \cdots, B_{4,6,2}^{j_1,*, j_0,*} $ be defined as Lemma~\ref{lem:b_4_j1_j0} 
    \item  Let $\|A \|_2 \leq R, \|A^{\top} \|_F \leq R, \| x\|_2 \leq R, \|\diag(f(A,x)) \|_F \leq \|f(A,x) \|_2 \leq 1, \| b_g \|_2 \leq 1$ 
\end{itemize}
Then, we have
\begin{itemize}
    \item {\bf Part 1.}
    \begin{align*}
       \| B_{4,1,1}^{j_1,*,j_0,*} (A) - B_{4,1,1}^{j_1,*,j_0,*} ( \wt{A} ) \|_F \leq  \beta^{-2} \cdot n \cdot \sqrt{d}\exp(5R^2) \cdot \|A - \wt{A}\|_F
    \end{align*}
     \item {\bf Part 2.}
    \begin{align*}
       \| B_{4,1,2}^{j_1,*,j_0,*} (A) - B_{4,1,2}^{j_1,*,j_0,*} ( \wt{A} ) \|_F \leq   \beta^{-2} \cdot n \cdot \sqrt{d}\exp(5R^2) \cdot \|A - \wt{A}\|_F
    \end{align*}
     \item {\bf Part 3.}
    \begin{align*}
       \| B_{4,1,3}^{j_1,*,j_0,*} (A) - B_{4,1,3}^{j_1,*,j_0,*} ( \wt{A} ) \|_F \leq   \beta^{-2} \cdot n \cdot \sqrt{d}\exp(6R^2) \cdot \|A - \wt{A}\|_F
    \end{align*}
     \item {\bf Part 4.}
    \begin{align*}
       \| B_{4,1,4}^{j_1,*,j_0,*} (A) - B_{4,1,4}^{j_1,*,j_0,*} ( \wt{A} ) \|_F \leq   \beta^{-2} \cdot n \cdot \sqrt{d}\exp(6R^2) \cdot \|A - \wt{A}\|_F
    \end{align*}
     \item {\bf Part 5.}
    \begin{align*}
       \| B_{4,1,5}^{j_1,*,j_0,*} (A) - B_{4,1,5}^{j_1,*,j_0,*} ( \wt{A} ) \|_F \leq   \beta^{-2} \cdot n \cdot \sqrt{d}\exp(6R^2) \cdot \|A - \wt{A}\|_F
    \end{align*}
     \item {\bf Part 6.}
    \begin{align*}
       \| B_{4,1,6}^{j_1,*,j_0,*} (A) - B_{4,1,6}^{j_1,*,j_0,*} ( \wt{A} ) \|_F \leq  \beta^{-2} \cdot n \cdot \sqrt{d}\exp(6R^2) \cdot \|A - \wt{A}\|_F
    \end{align*}
     \item {\bf Part 7.}
    \begin{align*}
       \| B_{4,1,7}^{j_1,*,j_0,*} (A) - B_{4,1,7}^{j_1,*,j_0,*} ( \wt{A} ) \|_F \leq   \beta^{-2} \cdot n \cdot \sqrt{d}\exp(6R^2) \cdot \|A - \wt{A}\|_F
    \end{align*}
    \item {\bf Part 8.}
    \begin{align*}
       \| B_{4,2,1}^{j_1,*,j_0,*} (A) - B_{4,2,1}^{j_1,*,j_0,*} ( \wt{A} ) \|_F \leq  \beta^{-2} \cdot n \cdot \sqrt{d}\exp(6R^2) \cdot \|A - \wt{A}\|_F
    \end{align*}
     \item {\bf Part 9.}
    \begin{align*}
       \| B_{4,3,1}^{j_1,*,j_0,*} (A) - B_{4,3,1}^{j_1,*,j_0,*} ( \wt{A} ) \|_F \leq  \beta^{-2} \cdot n \cdot \sqrt{d}  \exp(5R^2)\|A - \wt{A}\|_F  
    \end{align*}
     \item {\bf Part 10.}
    \begin{align*}
       \| B_{4,4,1}^{j_1,*,j_0,*} (A) - B_{4,4,1}^{j_1,*,j_0,*} ( \wt{A} ) \|_F \leq  \beta^{-2} \cdot n \cdot \sqrt{d}  \exp(6R^2)\|A - \wt{A}\|_F  
    \end{align*}
     \item {\bf Part 11.}
    \begin{align*}
       \| B_{4,5,1}^{j_1,*,j_0,*} (A) - B_{4,5,1}^{j_1,*,j_0,*} ( \wt{A} ) \|_F \leq \beta^{-2} \cdot n \cdot \sqrt{d}  \exp(6R^2)\|A - \wt{A}\|_F   
    \end{align*}
     \item {\bf Part 12.}
    \begin{align*}
       \| B_{4,5,2}^{j_1,*,j_0,*} (A) - B_{4,5,2}^{j_1,*,j_0,*} ( \wt{A} ) \|_F \leq  \beta^{-2} \cdot n \cdot \sqrt{d}  \exp(6R^2)\|A - \wt{A}\|_F  
    \end{align*}
     \item {\bf Part 13.}
    \begin{align*}
       \| B_{4,6,1}^{j_1,*,j_0,*} (A) - B_{4,6,1}^{j_1,*,j_0,*} ( \wt{A} ) \|_F \leq    \beta^{-2} \cdot n \cdot \sqrt{d}  \exp(6R^2)\|A - \wt{A}\|_F 
    \end{align*}
         \item {\bf Part 14.}
    \begin{align*}
       \| B_{4,6,2}^{j_1,*,j_0,*} (A) - B_{4,6,2}^{j_1,*,j_0,*} ( \wt{A} ) \|_F \leq  \beta^{-2} \cdot n \cdot \sqrt{d}  \exp(6R^2)\|A - \wt{A}\|_F   
    \end{align*}
        \item{\bf Part 15.}
    \begin{align*}
         \| B_{4}^{j_1,*,j_0,*} (A) - B_{4}^{j_1,*,j_0,*} ( \wt{A} ) \|_F \leq & ~   14\beta^{-2} \cdot n \cdot \sqrt{d}  \exp(6R^2)\|A - \wt{A}\|_F   
    \end{align*}
    \end{itemize}
\end{lemma}
\begin{proof}
{\bf Proof of Part 1.}
\begin{align*}
& ~ \| B_{4,1,1}^{j_1,*,j_0,*} (A) - B_{4,1,1}^{j_1,*,j_0,*} ( \wt{A} ) \| \\ \leq 
    & ~ \| f_c(A,x)^2 \cdot f(A,x)_{j_1} \cdot f(A,x)_{j_0} \cdot  h(A,x) \cdot  {\bf 1}_d^\top  - f_c(\wt{A},x)^2 \cdot f(\wt{A},x)_{j_1} \cdot f(\wt{A},x)_{j_0} \cdot  h(\wt{A},x) \cdot  {\bf 1}_d^\top \|_F\\
    \leq & ~ |  f_c(A,x) - f_c(\wt{A},x) |\cdot |f_c(A,x)|\cdot |f(A,x)_{j_1}|\cdot | f(A,x)_{j_0} | \cdot   \| h(A,x)\|_2  \cdot \| {\bf 1}_d^\top\|_2 \\
    & + ~   |f_c(\wt{A},x)|\cdot |  f_c(A,x) - f_c(\wt{A},x) |\cdot |f(A,x)_{j_1}|\cdot | f(A,x)_{j_0} | \cdot  \| h(A,x)\|_2  \cdot \| {\bf 1}_d^\top\|_2 \\
    & + ~   |f_c(\wt{A},x)|\cdot |f_c(\wt{A},x)|\cdot  |f(A,x)_{j_1} - f(\wt{A},x)_{j_1}|\cdot | f(A,x)_{j_0} | \cdot   \| h(A,x)\|_2  \cdot \| {\bf 1}_d^\top\|_2 \\
    & + ~   |f_c(\wt{A},x)|\cdot |f_c(\wt{A},x)|\cdot  |f(\wt{A},x)_{j_1}|\cdot | f(A,x)_{j_0} -  f(\wt{A},x)_{j_0} | \cdot   \| h(A,x)\|_2  \cdot \| {\bf 1}_d^\top\|_2 \\
    & + ~ |f_c(\wt{A},x)|\cdot |f_c(\wt{A},x)|\cdot  |f(\wt{A},x)_{j_1}|\cdot  | f(\wt{A},x)_{j_0} | \cdot \| h(A,x) - h(\wt{A},x)\|_2  \cdot \| {\bf 1}_d^\top\|_2 \\
    \leq & ~ 12R^2  \beta^{-2} \cdot n \cdot \sqrt{d}  \exp(3R^2)\|A - \wt{A}\|_F \\
    &+ ~ 12R^2  \beta^{-2} \cdot n \cdot \sqrt{d}  \exp(3R^2)\|A - \wt{A}\|_F \\
    & + ~ 8R^2 \beta^{-2} \cdot n \cdot \sqrt{d}\exp(3R^2) \cdot \|A - \wt{A}\|_F \\
    & + ~ 8R^2\beta^{-2} \cdot n \cdot \sqrt{d}\exp(3R^2) \cdot \|A - \wt{A}\|_F \\
    & + ~ 12 \beta^{-2} \cdot n \cdot \sqrt{d}\exp(4R^2) \cdot \|A - \wt{A}\|_F\\
    \leq &  ~ 52 \beta^{-2} \cdot n \cdot \sqrt{d}\exp(4R^2) \cdot \|A - \wt{A}\|_F \\
    \leq &  ~ \beta^{-2} \cdot n \cdot \sqrt{d}\exp(5R^2) \cdot \|A - \wt{A}\|_F 
\end{align*}

{\bf Proof of Part 2.}
\begin{align*}
    & ~ \| B_{4,1,2}^{j_1,*,j_0,*} (A) - B_{4,1,2}^{j_1,*,j_0,*} ( \wt{A} ) \|_F \\
    \leq & ~ \| f(A,x)_{j_1} \cdot f(A,x)_{j_0} \cdot c(A,x)_{j_1} \cdot  f_c(A,x) \cdot h(A,x) \cdot  {\bf 1}_d^\top  \\
    & ~- f(\wt{A},x)_{j_1} \cdot f(\wt{A},x)_{j_0} \cdot c(\wt{A},x)_{j_1} \cdot  f_c(\wt{A},x) \cdot h(\wt{A},x) \cdot  {\bf 1}_d^\top  \|_F\\
 \leq & ~ |  f_c(A,x) - f_c(\wt{A},x) |\cdot |c(A,x)_{j_1}|\cdot |f(A,x)_{j_1}|\cdot | f(A,x)_{j_0} | \cdot  \| h(A,x)\|_2  \cdot \| {\bf 1}_d^\top\|_2  \\
    & + ~   |f_c(\wt{A},x)|\cdot |  c(A,x)_{j_1} - c(\wt{A},x)_{j_1} |\cdot |f(A,x)_{j_1}|\cdot | f(A,x)_{j_0} | \cdot  \| h(A,x)\|_2  \cdot \| {\bf 1}_d^\top\|_2 \\
    & + ~   |f_c(\wt{A},x)|\cdot |c(\wt{A},x)_{j_1} |\cdot  |f(A,x)_{j_1} - f(\wt{A},x)_{j_1}|\cdot | f(A,x)_{j_0} | \cdot  \| h(A,x)\|_2  \cdot \| {\bf 1}_d^\top\|_2  \\
    & + ~   |f_c(\wt{A},x)|\cdot |c(\wt{A},x)_{j_1} |\cdot  |f(\wt{A},x)_{j_1}|\cdot | f(A,x)_{j_0} -  f(\wt{A},x)_{j_0} | \cdot \| h(A,x)\|_2  \cdot \| {\bf 1}_d^\top\|_2\\
    & + ~ |f_c(\wt{A},x)|\cdot |c(\wt{A},x)_{j_1} |\cdot  |f(\wt{A},x)_{j_1}|\cdot  | f(\wt{A},x)_{j_0} | \cdot  \| h(A,x)- h(\wt{A},x)\|_2  \cdot \| {\bf 1}_d^\top\|_2 \\
     \leq & ~ 12R^2  \beta^{-2} \cdot n \cdot \sqrt{d}  \exp(3R^2)\|A - \wt{A}\|_F \\
    &+ ~ 4R^2  \beta^{-2} \cdot n \cdot \sqrt{d}  \exp(3R^2)\|A - \wt{A}\|_F \\
    & + ~ 8R^2 \beta^{-2} \cdot n \cdot \sqrt{d}\exp(3R^2) \cdot \|A - \wt{A}\|_F \\
    & + ~ 8R^2\beta^{-2} \cdot n \cdot \sqrt{d}\exp(3R^2) \cdot \|A - \wt{A}\|_F \\
    & + ~ 12 \beta^{-2} \cdot n \cdot \sqrt{d}\exp(4R^2) \cdot \|A - \wt{A}\|_F \\
    \leq &  ~ 44 \beta^{-2} \cdot n \cdot \sqrt{d}\exp(4R^2) \cdot \|A - \wt{A}\|_F \\
    \leq &  ~ \beta^{-2} \cdot n \cdot \sqrt{d}\exp(5R^2) \cdot \|A - \wt{A}\|_F 
\end{align*}

{\bf Proof of Part 3.}
\begin{align*}
    & ~ \| B_{4,1,3}^{j_1,*,j_0,*} (A) - B_{4,1,3}^{j_1,*,j_0,*} ( \wt{A} ) \|_F \\
    \leq & ~ \| f(A,x)_{j_1} \cdot  f(A,x)_{j_0} \cdot f_c(A,x)^2 \cdot  ((A_{j_1,*}) \circ x^\top) \cdot  h(A,x) \cdot I_d \\
    & ~- f(\wt{A},x)_{j_1} \cdot  f(\wt{A},x)_{j_0} \cdot f_c(\wt{A},x)^2 \cdot  ((\wt{A}_{j_1,*}) \circ x^\top) \cdot  h(\wt{A},x)  \cdot I_d \|_F\\
 \leq & ~  |f(A,x)_{j_1} - f(\wt{A},x)_{j_1}|\cdot | f(A,x)_{j_0}| \cdot |f_c(A,x)|^2 \cdot \| A_{j_1,*} \|_2 \cdot \| \diag(x)\|_F \cdot \|h(A,x)\|_2  \cdot \| I_d \|_F\\
    & + ~   |  f(\wt{A},x)_{j_1}|\cdot | f(A,x)_{j_0} -  f(\wt{A},x)_{j_0}| \cdot |f_c(A,x)|^2 \cdot \| A_{j_1,*} \|_2 \cdot \| \diag(x)\|_F \cdot   \|h(A,x)\|_2  \cdot \| I_d \|_F\\
    & + ~    |  f(\wt{A},x)_{j_1}|\cdot |  f(\wt{A},x)_{j_0}| \cdot |f_c(A,x) - f_c(\wt{A},x)| \cdot |f_c(A,x)|  \cdot \| A_{j_1,*} \|_2 \cdot \| \diag(x)\|_F \cdot  \|h(A,x)\|_2 \cdot \| I_d \|_F \\
    & + ~  |  f(\wt{A},x)_{j_1}|\cdot |  f(\wt{A},x)_{j_0}|\cdot |f_c(\wt{A},x)| \cdot |f_c(A,x) - f_c(\wt{A},x)|   \cdot \| A_{j_1,*} \|_2 \cdot \| \diag(x)\|_F \cdot  \|h(A,x)\|_2 \cdot \| I_d \|_F  \\
    & + ~    |  f(\wt{A},x)_{j_1}|\cdot |  f(\wt{A},x)_{j_0}|\cdot |f_c(\wt{A},x)| \cdot |f_c(\wt{A},x)|  \cdot \| A_{j_1,*} - \wt{A}_{j_1,*} \|_2 \cdot \| \diag(x)\|_F \cdot  \|h(A,x)\|_2  \cdot \| I_d \|_F \\
     & + ~    |  f(\wt{A},x)_{j_1}|\cdot |  f(\wt{A},x)_{j_0}|\cdot |f_c(\wt{A},x)| \cdot |f_c(\wt{A},x)|  \cdot \|   \wt{A}_{j_1,*} \|_2 \cdot \| \diag(x)\|_F \cdot  \|h(A,x) - h(\wt{A},x)\|_2  \cdot \| I_d \|_F \\
    \leq & ~ 8R^4  \beta^{-2} \cdot n \cdot \sqrt{d}  \exp(3R^2)\|A - \wt{A}\|_F \\
    &+ ~ 8R^4  \beta^{-2} \cdot n \cdot \sqrt{d}  \exp(3R^2)\|A - \wt{A}\|_F \\
    & + ~ 12R^4 \beta^{-2} \cdot n \cdot \sqrt{d}\exp(3R^2) \cdot \|A - \wt{A}\|_F \\
    & + ~ 12R^4\beta^{-2} \cdot n \cdot \sqrt{d}\exp(3R^2) \cdot \|A - \wt{A}\|_F \\
    & + ~ 4 R^3 \cdot \sqrt{d} \cdot \|A - \wt{A} \|_F \\
    & + ~ 12R^2 \beta^{-2} \cdot n \cdot \sqrt{d}\exp(4R^2) \cdot \|A - \wt{A}\|_F\\
    \leq &  ~ 56 \beta^{-2} \cdot n \cdot \sqrt{d}\exp(5R^2) \cdot \|A - \wt{A}\|_F \\
    \leq &  ~ \beta^{-2} \cdot n \cdot \sqrt{d}\exp(6R^2) \cdot \|A - \wt{A}\|_F 
\end{align*}

{\bf Proof of Part 4.}
\begin{align*}
& ~ \| B_{4,1,4}^{j_1,*,j_0,*} (A) - B_{4,1,4}^{j_1,*,j_0,*} ( \wt{A} ) \| \\ 
\leq  & ~ \| - f(A,x)_{j_1} \cdot  f(A,x)_{j_0} \cdot f_c(A,x)^2  \cdot h(A,x)^\top \cdot h(A,x) \cdot I_d  \\
    & ~- (-f(\wt{A},x)_{j_1} \cdot  f(\wt{A},x)_{j_0} \cdot f_c(\wt{A},x)^2  \cdot h(\wt{A},x)^\top \cdot h(\wt{A},x) \cdot I_d)\|_F \\
    \leq  & ~ \|  f(A,x)_{j_1} \cdot  f(A,x)_{j_0} \cdot f_c(A,x)^2  \cdot h(A,x)^\top \cdot h(A,x) \cdot I_d  \\
    & ~- f(\wt{A},x)_{j_1} \cdot  f(\wt{A},x)_{j_0} \cdot f_c(\wt{A},x)^2  \cdot h(\wt{A},x)^\top \cdot h(\wt{A},x) \cdot I_d\|_F \\
    \leq & ~  |f(A,x)_{j_1} - f(\wt{A},x)_{j_1}|\cdot | f(A,x)_{j_0}| \cdot |f_c(A,x)|^2 \cdot \| h(A,x)^\top \|_2  \cdot  \| h(A,x)  \|_2  \cdot \| I_d \|_F\\
    & + ~   |  f(\wt{A},x)_{j_1}|\cdot | f(A,x)_{j_0} -  f(\wt{A},x)_{j_0}| \cdot |f_c(A,x)|^2 \cdot \| h(A,x)^\top \|_2 \cdot \| h(A,x)  \|_2  \cdot \| I_d \|_F\\
    & + ~    |  f(\wt{A},x)_{j_1}|\cdot |  f(\wt{A},x)_{j_0}| \cdot |f_c(A,x) - f_c(\wt{A},x)| \cdot |f_c(A,x)|  \cdot \| h(A,x)^\top \|_2 \cdot \| h(A,x)  \|_2 \cdot \| I_d \|_F \\
    & + ~  |  f(\wt{A},x)_{j_1}|\cdot |  f(\wt{A},x)_{j_0}|\cdot |f_c(\wt{A},x)| \cdot |f_c(A,x) - f_c(\wt{A},x)|   \cdot \| h(A,x)^\top \|_2 \cdot \| h(A,x)  \|_2  \cdot \| I_d \|_F  \\
    & + ~    |  f(\wt{A},x)_{j_1}|\cdot |  f(\wt{A},x)_{j_0}|\cdot |f_c(\wt{A},x)| \cdot |f_c(\wt{A},x)|  \cdot  \| h(A,x)^\top -h(\wt{A},x)^\top \|_2  \cdot  \| h(A,x)  \|_2 \cdot \| I_d \|_F \\
       & + ~    |  f(\wt{A},x)_{j_1}|\cdot |  f(\wt{A},x)_{j_0}|\cdot |f_c(\wt{A},x)| \cdot |f_c(\wt{A},x)|  \cdot  \|  h(\wt{A},x)^\top \|_2  \cdot  \| h(A,x) -  h(\wt{A},x) \|_2  \cdot \| I_d \|_F \\
    \leq & ~ 8R^4  \beta^{-2} \cdot n \cdot \sqrt{d}  \exp(3R^2)\|A - \wt{A}\|_F \\
    &+ ~ 8R^4  \beta^{-2} \cdot n \cdot \sqrt{d}  \exp(3R^2)\|A - \wt{A}\|_F \\
    & + ~ 12R^4\beta^{-2} \cdot n \cdot \sqrt{d}\exp(3R^2) \cdot \|A - \wt{A}\|_F \\
    & + ~ 12R^4\beta^{-2} \cdot n \cdot \sqrt{d}\exp(3R^2) \cdot \|A - \wt{A}\|_F \\
    & + ~ 12R^2\beta^{-2} \cdot n \cdot \sqrt{d}\exp(4R^2) \cdot \|A - \wt{A}\|_F \\
    & + ~ 12R^2\beta^{-2} \cdot n \cdot \sqrt{d}\exp(4R^2) \cdot \|A - \wt{A}\|_F \\
    \leq &  ~ 64\beta^{-2} \cdot n \cdot \sqrt{d}\exp(5R^2) \cdot \|A - \wt{A}\|_F \\
\leq &  ~ \beta^{-2} \cdot n \cdot \sqrt{d}\exp(6R^2) \cdot \|A - \wt{A}\|_F
\end{align*}

{\bf Proof of Part 5.}
\begin{align*}
& ~ \| B_{4,1,5}^{j_1,*,j_0,*} (A) - B_{4,1,5}^{j_1,*,j_0,*} ( \wt{A} ) \| \\ \leq 
    & ~ \| f(A,x)_{j_1} \cdot  f(A,x)_{j_0} \cdot (-f_2(A,x) + f(A,x)_{j_1})  \cdot f_c(A,x)  \cdot h(A,x)^\top \cdot h(A,x) \cdot I_d \\
    & ~- f(\wt{A},x)_{j_1} \cdot  f(\wt{A},x)_{j_0} \cdot (-f_2(\wt{A},x) + f(\wt{A},x)_{j_1})  \cdot f_c(\wt{A},x)  \cdot h(\wt{A},x)^\top \cdot  h(\wt{A},x) \cdot I_d \|_F\\
    \leq & ~ |  f_c(A,x) - f_c(\wt{A},x) |\cdot (| -f_2(A,x)| + |f(A,x)_{j_1}|) \cdot |f(A,x)_{j_1}|\cdot | f(A,x)_{j_0} | \cdot \| h(A,x)^\top \|_2\\
    & ~ \cdot \| h(A,x) \|_2  \cdot \| I_d \|_F \\
    & + ~  |  f_c(\wt{A},x) |\cdot (| f_2(A,x) -  f_2(\wt{A},x)| + |f(A,x)_{j_1}- f(\wt{A},x)_{j_1}|) \cdot |f(A,x)_{j_1}|\cdot | f(A,x)_{j_0} | \cdot \| h(A,x)^\top \|_2 \\
    & ~\cdot \| h(A,x) \|_2 \cdot \| I_d \|_F \\
    & + ~     |  f_c(\wt{A},x) |\cdot (| f_2(\wt{A},x)| + |  f(\wt{A},x)_{j_1}|) \cdot |f(A,x)_{j_1} - f(\wt{A},x)_{j_1}| \cdot | f(A,x)_{j_0} | \cdot \| h(A,x)^\top \|_2  \\
    & ~\cdot  \| h(A,x) \|_2  \cdot \| I_d \|_F \\
    & + ~     |  f_c(\wt{A},x) |\cdot (| f_2(\wt{A},x)| + |  f(\wt{A},x)_{j_1}|) \cdot | f(\wt{A},x)_{j_1}| \cdot | f(A,x)_{j_0} - f(\wt{A},x)_{j_0} | \cdot \| h(A,x)^\top \|_2\\
    & ~ \cdot  \| h(A,x) \|_2  \cdot \| I_d \|_F \\
     & + ~     |  f_c(\wt{A},x) |\cdot (| f_2(\wt{A},x)| + |  f(\wt{A},x)_{j_1}|) \cdot | f(\wt{A},x)_{j_1}| \cdot |  f(\wt{A},x)_{j_0} | \cdot \| h(A,x)^\top -h(\wt{A},x)^\top \|_2\\
    & ~ \cdot   \| h(A,x) \|_2 \cdot \| I_d \|_F \\
    & + ~     |  f_c(\wt{A},x) |\cdot (| f_2(\wt{A},x)| + |  f(\wt{A},x)_{j_1}|) \cdot | f(\wt{A},x)_{j_1}| \cdot |  f(\wt{A},x)_{j_0} | \cdot \| h(\wt{A},x)^\top \|_2 \\
    & ~ \cdot  \| h(A,x)- h(\wt{A},x) \|_2  \cdot \| I_d \|_F \\
    \leq & ~ 12R^4  \beta^{-2} \cdot n \cdot \sqrt{d}  \exp(3R^2)\|A - \wt{A}\|_F \\
    &+ ~ 12 R^4 \beta^{-2} \cdot n \cdot \sqrt{d}  \exp(3R^2)\|A - \wt{A}\|_F \\
    & + ~ 8 R^4\beta^{-2} \cdot n \cdot \sqrt{d}\exp(3R^2) \cdot \|A - \wt{A}\|_F \\
    & + ~ 8 R^4\beta^{-2} \cdot n \cdot \sqrt{d}\exp(3R^2) \cdot \|A - \wt{A}\|_F \\
    & + ~ 12R^2\beta^{-2} \cdot n \cdot \sqrt{d}\exp(4R^2) \cdot \|A - \wt{A}\|_F \\
    & + ~ 12R^2\beta^{-2} \cdot n \cdot \sqrt{d}\exp(4R^2) \cdot \|A - \wt{A}\|_F\\
    \leq &  ~ 64\beta^{-2} \cdot n \cdot \sqrt{d}\exp(5R^2) \cdot \|A - \wt{A}\|_F \\
\leq &  ~ \beta^{-2} \cdot n \cdot \sqrt{d}\exp(6R^2) \cdot \|A - \wt{A}\|_F
\end{align*}

{\bf Proof of Part 6.}
\begin{align*}
& ~ \| B_{4,1,6}^{j_1,*,j_0,*} (A) - B_{4,1,6}^{j_1,*,j_0,*} ( \wt{A} ) \| \\ \leq 
    & ~ \| f(A,x)_{j_1} \cdot  f(A,x)_{j_0} \cdot (-f_c(A,x) + f(A,x)_{j_1})  \cdot f_c(A,x)  \cdot h(A,x)^\top \cdot h(A,x) \cdot I_d \\
    & ~- f(\wt{A},x)_{j_1} \cdot  f(\wt{A},x)_{j_0} \cdot (-f_c(\wt{A},x) + f(\wt{A},x)_{j_1})  \cdot f_c(\wt{A},x)  \cdot h(\wt{A},x)^\top \cdot h(\wt{A},x) \cdot I_d\|_F\\
    \leq & ~ |  f_c(A,x) - f_c(\wt{A},x) |\cdot (| -f_c(A,x)| + |f(A,x)_{j_1}|) \cdot |f(A,x)_{j_1}|\cdot | f(A,x)_{j_0} | \cdot \| h(A,x)^\top \|_2\\
    & ~ \cdot \| h(A,x) \|_2  \cdot \| I_d \|_F \\
    & + ~  |  f_c(\wt{A},x) |\cdot (| f_c(A,x) -  f_c(\wt{A},x)| + |f(A,x)_{j_1}- f(\wt{A},x)_{j_1}|) \cdot |f(A,x)_{j_1}|\cdot | f(A,x)_{j_0} | \cdot \| h(A,x)^\top \|_2 \\
    & ~\cdot  \| h(A,x) \|_2  \cdot \| I_d \|_F \\
    & + ~     |  f_c(\wt{A},x) |\cdot (| f_c(\wt{A},x)| + |  f(\wt{A},x)_{j_1}|) \cdot |f(A,x)_{j_1} - f(\wt{A},x)_{j_1}| \cdot | f(A,x)_{j_0} | \cdot \| h(A,x)^\top \|_2  \\
    & ~\cdot  \| h(A,x) \|_2  \cdot \| I_d \|_F \\
    & + ~     |  f_c(\wt{A},x) |\cdot (| f_c(\wt{A},x)| + |  f(\wt{A},x)_{j_1}|) \cdot | f(\wt{A},x)_{j_1}| \cdot | f(A,x)_{j_0} - f(\wt{A},x)_{j_0} | \cdot \| h(A,x)^\top \|_2\\
    & ~ \cdot  \| h(A,x) \|_2 \cdot \| I_d \|_F \\
     & + ~     |  f_c(\wt{A},x) |\cdot (| f_c(\wt{A},x)| + |  f(\wt{A},x)_{j_1}|) \cdot | f(\wt{A},x)_{j_1}| \cdot |  f(\wt{A},x)_{j_0} | \cdot \| h(A,x)^\top -h(\wt{A},x)^\top \|_2\\
    & ~ \cdot  \| h(A,x) \|_2 \cdot \| I_d \|_F \\
    & + ~     |  f_c(\wt{A},x) |\cdot (| f_c(\wt{A},x)| + |  f(\wt{A},x)_{j_1}|) \cdot | f(\wt{A},x)_{j_1}| \cdot |  f(\wt{A},x)_{j_0} | \cdot \| h(\wt{A},x)^\top \|_2 \\
    & ~ \cdot  \| h(A,x) - h(\wt{A},x) \|_2  \cdot \| I_d \|_F \\
    \leq & ~ 18R^4  \beta^{-2} \cdot n \cdot \sqrt{d}  \exp(3R^2)\|A - \wt{A}\|_F \\
    &+ ~ 16 R^4 \beta^{-2} \cdot n \cdot \sqrt{d}  \exp(3R^2)\|A - \wt{A}\|_F \\
    & + ~ 12 R^4\beta^{-2} \cdot n \cdot \sqrt{d}\exp(3R^2) \cdot \|A - \wt{A}\|_F \\
    & + ~ 12 R^4\beta^{-2} \cdot n \cdot \sqrt{d}\exp(3R^2) \cdot \|A - \wt{A}\|_F \\
    & + ~ 18R^2\beta^{-2} \cdot n \cdot \sqrt{d}\exp(4R^2) \cdot \|A - \wt{A}\|_F \\
    & + ~ 18R^2\beta^{-2} \cdot n \cdot \sqrt{d}\exp(4R^2) \cdot \|A - \wt{A}\|_F \\
    \leq &  ~ 94\beta^{-2} \cdot n \cdot \sqrt{d}\exp(5R^2) \cdot \|A - \wt{A}\|_F \\
\leq &  ~ \beta^{-2} \cdot n \cdot \sqrt{d}\exp(6R^2) \cdot \|A - \wt{A}\|_F
\end{align*}

{\bf Proof of Part 7.}
\begin{align*}
& ~ \| B_{4,1,7}^{j_1,*,j_0,*} (A) - B_{4,1,7}^{j_1,*,j_0,*} ( \wt{A} ) \| \\ \leq 
    & ~ \|  f(A,x)_{j_1} \cdot  f(A,x)_{j_0} \cdot f_c(A,x)   \cdot p_{j_1}(A,x)^\top \cdot  h(A,x) \cdot I_d \\
    & ~ - f(\wt{A},x)_{j_1} \cdot  f(\wt{A},x)_{j_0} \cdot f_c(\wt{A},x)   \cdot p_{j_1}(\wt{A},x)^\top \cdot  h(\wt{A},x) \cdot I_d \|_F\\
    \leq & ~  |f(A,x)_{j_1} - f(\wt{A},x)_{j_1}|\cdot | f(A,x)_{j_0}| \cdot |f_c(A,x)|  \cdot \| p_{j_1}(A,x)^\top \|_2  \cdot \| h(A,x) \|_2  \cdot \| I_d \|_F\\
    & + ~   |  f(\wt{A},x)_{j_1}|\cdot | f(A,x)_{j_0} -  f(\wt{A},x)_{j_0}| \cdot |f_c(A,x)|  \cdot \|  p_{j_1}(A,x)^\top \|_2 \cdot   \| h(A,x) \|_2   \cdot \| I_d \|_F\\
    & + ~    |  f(\wt{A},x)_{j_1}|\cdot |  f(\wt{A},x)_{j_0}| \cdot |f_c(A,x) - f_c(\wt{A},x)|    \cdot \| p_{j_1}(A,x)^\top \|_2 \cdot  \| h(A,x) \|_2  \cdot \| I_d \|_F \\
    & + ~  |  f(\wt{A},x)_{j_1}|\cdot |  f(\wt{A},x)_{j_0}|\cdot |f_c(\wt{A},x)|   \cdot \| p_{j_1}(A,x)^\top - p_{j_1}(\wt{A},x)^\top\|_2 \cdot  \| h(A,x) \|_2  \cdot \| I_d \|_F  \\
       & + ~    |  f(\wt{A},x)_{j_1}|\cdot |  f(\wt{A},x)_{j_0}|\cdot |f_c(\wt{A},x)|  \cdot  \|  p_{j_1}(\wt{A},x)^\top \|_2  \cdot   \| h(A,x)- h(\wt{A},x) \|_2   \cdot \| I_d \|_F \\
    \leq & ~ 24R^4  \beta^{-2} \cdot n \cdot \sqrt{d}  \exp(3R^2)\|A - \wt{A}\|_F \\
    &+ ~ 24R^4  \beta^{-2} \cdot n \cdot \sqrt{d}  \exp(3R^2)\|A - \wt{A}\|_F \\
    & + ~ 36R^4\beta^{-2} \cdot n \cdot \sqrt{d}\exp(3R^2) \cdot \|A - \wt{A}\|_F \\
    & + ~ 26R^2\beta^{-2} \cdot n \cdot \sqrt{d}\exp(4R^2) \cdot \|A - \wt{A}\|_F \\
    & + ~ 36R^2\beta^{-2} \cdot n \cdot \sqrt{d}\exp(4R^2) \cdot \|A - \wt{A}\|_F \\
    \leq &  ~ 146\beta^{-2} \cdot n \cdot \sqrt{d}\exp(5R^2) \cdot \|A - \wt{A}\|_F \\
\leq &  ~ \beta^{-2} \cdot n \cdot \sqrt{d}\exp(6R^2) \cdot \|A - \wt{A}\|_F
\end{align*}

{\bf Proof of Part 8.}
\begin{align*}
& ~ \| B_{4,2,1}^{j_1,*,j_0,*} (A) - B_{4,2,1}^{j_1,*,j_0,*} ( \wt{A} ) \| \\ \leq 
    & ~ \| f(A,x)_{j_1} \cdot  f(A,x)_{j_0} \cdot f_c(A,x) \cdot c_g(A,x)^{\top} \cdot h(A,x) \cdot x \cdot {\bf 1}_d^{\top}\\
    & ~ - f(\wt{A},x)_{j_1} \cdot  f(\wt{A},x)_{j_0} \cdot f_c(\wt{A},x) \cdot c_g(\wt{A},x)^{\top} \cdot  h(\wt{A},x) \cdot x \cdot {\bf 1}_d^{\top} \|_F \\
    \leq & ~  |f(A,x)_{j_1} - f(\wt{A},x)_{j_1}|\cdot | f(A,x)_{j_0}| \cdot |f_c(A,x)|  \cdot \| c_g(A,x)^\top \|_2  \cdot  \|h(A,x)\|_2 \cdot \| x \|_2 \cdot \| {\bf 1}_d^{\top} \|_F\\
    & + ~   |  f(\wt{A},x)_{j_1}|\cdot | f(A,x)_{j_0} -  f(\wt{A},x)_{j_0}| \cdot |f_c(A,x)|  \cdot \|   c_g(A,x)^\top \|_2 \cdot  \|h(A,x)\|_2   \cdot \| x \|_2 \cdot \| {\bf 1}_d^{\top} \|_F\\
    & + ~    |  f(\wt{A},x)_{j_1}|\cdot |  f(\wt{A},x)_{j_0}| \cdot |f_c(A,x) - f_c(\wt{A},x)|    \cdot \| c_g(A,x)^\top \|_2 \cdot  \|h(A,x)\|_2  \cdot \| x \|_2 \cdot \| {\bf 1}_d^{\top} \|_F\\
    & + ~  |  f(\wt{A},x)_{j_1}|\cdot |  f(\wt{A},x)_{j_0}|\cdot |f_c(\wt{A},x)|   \cdot \| c_g(A,x)^\top - c_g(\wt{A},x)^\top \|_2 \cdot  \|h(A,x)\|_2  \cdot \| x \|_2 \cdot \| {\bf 1}_d^{\top} \|_F\\
    & + ~    |  f(\wt{A},x)_{j_1}|\cdot |  f(\wt{A},x)_{j_0}|\cdot |f_c(\wt{A},x)|  \cdot  \|  c_g(\wt{A},x)^\top  \|_2  \cdot   \|h(A,x) - h(\wt{A},x)\|_2  \cdot \| x \|_2 \cdot \| {\bf 1}_d^{\top} \|_F\\
    \leq & ~ 20R^4 \beta^{-2} \cdot n \cdot \sqrt{d}  \exp(3R^2)\|A - \wt{A}\|_F \\
    &+ ~ 20R^4  \beta^{-2} \cdot n \cdot \sqrt{d}  \exp(3R^2)\|A - \wt{A}\|_F \\
    & + ~ 30R^4\beta^{-2} \cdot n \cdot \sqrt{d}\exp(3R^2) \cdot \|A - \wt{A}\|_F \\
    & + ~ 40R^4\beta^{-2} \cdot n \cdot \sqrt{d}\exp(3R^2) \cdot \|A - \wt{A}\|_F \\
     & + ~ 30R^2\beta^{-2} \cdot n \cdot \sqrt{d}\exp(4R^2) \cdot \|A - \wt{A}\|_F\\
    \leq &  ~ 140\beta^{-2} \cdot n \cdot \sqrt{d}\exp(5R^2) \cdot \|A - \wt{A}\|_F \\
    \leq &  ~ \beta^{-2} \cdot n \cdot \sqrt{d}\exp(6R^2) \cdot \|A - \wt{A}\|_F
\end{align*}

{\bf Proof of Part 9.}
\begin{align*}
  & ~ \|  B_{4,3,1}^{j_1,*,j_0,*} (A) - B_{4,3,1}^{j_1,*,j_0,*} ( \wt{A} ) \|_F \\
   = & ~ \| -  f(A,x)_{j_1} \cdot   f(A,x)_{j_0}  \cdot    f_c(A,x)   \cdot {\bf 1}_d \cdot c_g(A,x)^{\top} \cdot \diag (x)\\
    & ~ -(-  f(\wt{A},x)_{j_1} \cdot   f(\wt{A},x)_{j_0}  \cdot    f_c(\wt{A},x)   \cdot {\bf 1}_d \cdot c_g(\wt{A},x)^{\top} \cdot \diag (x))\|_F \\
    \leq & ~ \|    f(A,x)_{j_1} \cdot   f(A,x)_{j_0}  \cdot    f_c(A,x)   \cdot {\bf 1}_d \cdot c_g(A,x)^{\top} \cdot \diag (x)\\
    & ~ -   f(\wt{A},x)_{j_1} \cdot   f(\wt{A},x)_{j_0}  \cdot    f_c(\wt{A},x)   \cdot {\bf 1}_d \cdot c_g(\wt{A},x)^{\top} \cdot \diag (x) \|_F \\
    \leq & ~  |f(A,x)_{j_1} - f(\wt{A},x)_{j_1}|\cdot | f(A,x)_{j_0}| \cdot |f_c(A,x)|  \cdot \| c_g(A,x)^\top \|_2  \cdot  \|  {\bf 1}_d\|_2 \cdot \| \diag(x)\|_F  \\
    & + ~   |  f(\wt{A},x)_{j_1}|\cdot | f(A,x)_{j_0} -  f(\wt{A},x)_{j_0}| \cdot |f_c(A,x)|  \cdot \|   c_g(A,x)^\top \|_2 \cdot \|  {\bf 1}_d\|_2 \cdot \| \diag(x)\|_F\\
    & + ~    |  f(\wt{A},x)_{j_1}|\cdot |  f(\wt{A},x)_{j_0}| \cdot |f_c(A,x) - f_c(\wt{A},x)|    \cdot \| c_g(A,x)^\top \|_2 \cdot \|  {\bf 1}_d\|_2 \cdot \| \diag(x)\|_F\\
    & + ~  |  f(\wt{A},x)_{j_1}|\cdot |  f(\wt{A},x)_{j_0}|\cdot |f_c(\wt{A},x)|   \cdot \| c_g(A,x)^\top - c_g(\wt{A},x)^\top \|_2 \cdot  \|  {\bf 1}_d\|_2 \cdot \| \diag(x)\|_F\\
    \leq & ~ 20R^2 \beta^{-2} \cdot n \cdot \sqrt{d}  \exp(3R^2)\|A - \wt{A}\|_F \\
    &+ ~ 20R^2  \beta^{-2} \cdot n \cdot \sqrt{d}  \exp(3R^2)\|A - \wt{A}\|_F \\
    & + ~ 30R^2\beta^{-2} \cdot n \cdot \sqrt{d}\exp(3R^2) \cdot \|A - \wt{A}\|_F \\
    & + ~ 40R^2\beta^{-2} \cdot n \cdot \sqrt{d}\exp(3R^2) \cdot \|A - \wt{A}\|_F \\
    \leq &  ~ 110\beta^{-2} \cdot n \cdot \sqrt{d}\exp(4R^2) \cdot \|A - \wt{A}\|_F \\
    \leq &  ~ \beta^{-2} \cdot n \cdot \sqrt{d}\exp(5R^2) \cdot \|A - \wt{A}\|_F
\end{align*}

{\bf Proof of Part 10.}
\begin{align*}
  & ~ \|  B_{4,4,1}^{j_1,*,j_0,*} (A) - B_{4,4,1}^{j_1,*,j_0,*} ( \wt{A} ) \|_F \\
    = & ~ \| -f(A,x)_{j_1} \cdot f(A,x)_{j_0}  \cdot f_c( A,x) \cdot  c_g(A,x)^{\top} \cdot  h_e(A,x) \cdot x \cdot {\bf 1}_d^{\top} \\
    & ~ - (-f(\wt{A},x)_{j_1} \cdot f(\wt{A},x)_{j_0}  \cdot f_c( \wt{A},x) \cdot  c_g(\wt{A},x)^{\top} \cdot  h_e(\wt{A},x) \cdot x \cdot {\bf 1}_d^{\top}) \|_F \\
    \leq & ~ \| f(A,x)_{j_1} \cdot f(A,x)_{j_0}  \cdot f_c( A,x) \cdot  c_g(A,x)^{\top} \cdot  h_e(A,x) \cdot x \cdot {\bf 1}_d^{\top} \\
    & ~ - f(\wt{A},x)_{j_1} \cdot f(\wt{A},x)_{j_0}  \cdot f_c( \wt{A},x) \cdot  c_g(\wt{A},x)^{\top} \cdot  h_e(\wt{A},x) \cdot x \cdot {\bf 1}_d^{\top} \|_F \\
    \leq & ~ |f(A,x)_{j_1} - f(\wt{A},x)_{j_1}| \cdot |f(A,x)_{j_0} | \cdot |f_c( A,x) |\cdot  \| c_g(A,x)^{\top} \|_2 \cdot  \| h_e(A,x) \|_2 \cdot \| x \|_2 \cdot \|{\bf 1}_d^{\top} \|_2 \\
    & + ~ |f(\wt{A},x)_{j_1}| \cdot |f(A,x)_{j_0} - f(\wt{A},x)_{j_0} | \cdot |f_c( A,x) |\cdot  \| c_g(A,x)^{\top} \|_2 \cdot   \| h_e(A,x) \|_2 \cdot \| x \|_2 \cdot \|{\bf 1}_d^{\top} \|_2 \\
    & + ~ |f(\wt{A},x)_{j_1}| \cdot |f(\wt{A},x)_{j_0} | \cdot |f_c( A,x)  -f_c( \wt{A},x)|\cdot  \| c_g(A,x)^{\top} \|_2 \cdot   \| h_e(A,x) \|_2 \cdot \| x \|_2 \cdot \|{\bf 1}_d^{\top} \|_2 \\
    & + ~ |f(\wt{A},x)_{j_1}| \cdot |f(\wt{A},x)_{j_0} | \cdot |f_c( \wt{A},x)|\cdot  \| c_g(A,x)^{\top}  - c_g(\wt{A},x)^{\top} \|_2 \cdot  \| h_e(A,x) \|_2 \cdot \| x \|_2 \cdot \|{\bf 1}_d^{\top} \|_2 \\
    & + ~ |f(\wt{A},x)_{j_1}| \cdot |f(\wt{A},x)_{j_0} | \cdot |f_c( \wt{A},x)|\cdot  \| c_g(\wt{A},x)^{\top} \|_2 \cdot   \| h_e(A,x) - h_e(\wt{A},x) \|_2 \cdot \| x \|_2 \cdot \|{\bf 1}_d^{\top} \|_2 \\
    \leq & ~ 40R^4 \beta^{-2} \cdot n \cdot \sqrt{d}  \exp(3R^2)\|A - \wt{A}\|_F \\
    & + ~ 40R^4 \beta^{-2} \cdot n \cdot \sqrt{d}  \exp(3R^2)\|A - \wt{A}\|_F \\
    & + ~ 60R^4 \beta^{-2} \cdot n \cdot \sqrt{d}  \exp(3R^2)\|A - \wt{A}\|_F \\
    & + ~ 80R^4 \beta^{-2} \cdot n \cdot \sqrt{d}  \exp(3R^2)\|A - \wt{A}\|_F \\
    & + ~ 30R^2  \beta^{-2} \cdot n \cdot \sqrt{d}  \exp(4R^2)\|A - \wt{A}\|_F \\
    \leq & ~ 250\beta^{-2} \cdot n \cdot \sqrt{d}  \exp(5R^2)\|A - \wt{A}\|_F \\
    \leq & ~ \beta^{-2} \cdot n \cdot \sqrt{d}  \exp(6R^2)\|A - \wt{A}\|_F 
\end{align*}

{\bf Proof of Part 11.}
\begin{align*}
  & ~ \|  B_{4,5,1}^{j_1,*,j_0,*} (A) - B_{4,5,1}^{j_1,*,j_0,*} ( \wt{A} ) \|_F \\
    =   & ~ \| f(A,x)_{j_1} \cdot f(A,x)_{j_0}  \cdot f_2( A,x) \cdot  c_g(A,x)^{\top} \cdot  h(A,x) \cdot x \cdot {\bf 1}_d^{\top} \\
    & ~ - f(\wt{A},x)_{j_1} \cdot f(\wt{A},x)_{j_0}  \cdot f_2( \wt{A},x) \cdot  c_g(\wt{A},x)^{\top} \cdot  h(\wt{A},x) \cdot x \cdot {\bf 1}_d^{\top} \|_F \\
    \leq & ~ |f(A,x)_{j_1} - f(\wt{A},x)_{j_1}| \cdot |f(A,x)_{j_0} | \cdot |f_2( A,x) |\cdot  \| c_g(A,x)^{\top} \|_2 \cdot  \| h(A,x) \|_2 \cdot \| x \|_2 \cdot \|{\bf 1}_d^{\top} \|_2 \\
    & + ~ |f(\wt{A},x)_{j_1}| \cdot |f(A,x)_{j_0} - f(\wt{A},x)_{j_0} | \cdot |f_2( A,x) |\cdot  \| c_g(A,x)^{\top} \|_2 \cdot   \| h(A,x) \|_2 \cdot \| x \|_2 \cdot \|{\bf 1}_d^{\top} \|_2 \\
    & + ~ |f(\wt{A},x)_{j_1}| \cdot |f(\wt{A},x)_{j_0} | \cdot |f_2( A,x)  -f_2( \wt{A},x)|\cdot  \| c_g(A,x)^{\top} \|_2 \cdot   \| h(A,x) \|_2 \cdot \| x \|_2 \cdot \|{\bf 1}_d^{\top} \|_2 \\
    & + ~ |f(\wt{A},x)_{j_1}| \cdot |f(\wt{A},x)_{j_0} | \cdot |f_2( \wt{A},x)|\cdot  \| c_g(A,x)^{\top}  - c_g(\wt{A},x)^{\top} \|_2 \cdot  \| h(A,x) \|_2 \cdot \| x \|_2 \cdot \|{\bf 1}_d^{\top} \|_2 \\
    & + ~ |f(\wt{A},x)_{j_1}| \cdot |f(\wt{A},x)_{j_0} | \cdot |f_2( \wt{A},x)|\cdot  \| c_g(\wt{A},x)^{\top} \|_2 \cdot   \| h(A,x) - h(\wt{A},x) \|_2 \cdot \| x \|_2 \cdot \|{\bf 1}_d^{\top} \|_2 \\
    \leq & ~ 10R^4 \beta^{-2} \cdot n \cdot \sqrt{d}  \exp(3R^2)\|A - \wt{A}\|_F \\
    & + ~ 10R^4 \beta^{-2} \cdot n \cdot \sqrt{d}  \exp(3R^2)\|A - \wt{A}\|_F \\
    & + ~ 20R^4 \beta^{-2} \cdot n \cdot \sqrt{d}  \exp(3R^2)\|A - \wt{A}\|_F \\
    & + ~ 20R^4 \beta^{-2} \cdot n \cdot \sqrt{d}  \exp(3R^2)\|A - \wt{A}\|_F \\
    & + ~ 15R^2  \beta^{-2} \cdot n \cdot \sqrt{d}  \exp(4R^2)\|A - \wt{A}\|_F \\
    \leq & ~ 75\beta^{-2} \cdot n \cdot \sqrt{d}  \exp(5R^2)\|A - \wt{A}\|_F \\
    \leq & ~ \beta^{-2} \cdot n \cdot \sqrt{d}  \exp(6R^2)\|A - \wt{A}\|_F 
\end{align*}

{\bf Proof of Part 12.}
\begin{align*}
  & ~ \|  B_{4,5,2}^{j_1,*,j_0,*} (A) - B_{4,5,2}^{j_1,*,j_0,*} ( \wt{A} ) \|_F \\
    =   & ~ \| f(A,x)_{j_1}^2 \cdot f(A,x)_{j_0}    \cdot  c_g(A,x)^{\top} \cdot  h(A,x) \cdot x \cdot {\bf 1}_d^{\top} \\
    & ~ - f(\wt{A},x)_{j_1}^2 \cdot f(\wt{A},x)_{j_0}  \cdot   c_g(\wt{A},x)^{\top} \cdot  h(\wt{A},x) \cdot x \cdot {\bf 1}_d^{\top} \|_F \\
    \leq & ~ |f(A,x)_{j_1} - f(\wt{A},x)_{j_1}| \cdot |f(A,x)_{j_1} | \cdot |f(A,x)_{j_0} |  \cdot  \| c_g(A,x)^{\top} \|_2 \cdot  \| h(A,x) \|_2 \cdot \| x \|_2 \cdot \|{\bf 1}_d^{\top} \|_2 \\
     & + ~ |f(\wt{A},x)_{j_1}| \cdot |f(A,x)_{j_1} - f(\wt{A},x)_{j_1}|\cdot |f(A,x)_{j_0} |\cdot  \| c_g(A,x)^{\top} \|_2 \cdot   \| h(A,x) \|_2 \cdot \| x \|_2 \cdot \|{\bf 1}_d^{\top} \|_2 \\
    & + ~ |f(\wt{A},x)_{j_1}| \cdot |f(\wt{A},x)_{j_1}| \cdot |f(A,x)_{j_0} - f(\wt{A},x)_{j_0} |  \cdot  \| c_g(A,x)^{\top} \|_2 \cdot   \| h(A,x) \|_2 \cdot \| x \|_2 \cdot \|{\bf 1}_d^{\top} \|_2 \\
    & + ~ |f(\wt{A},x)_{j_1}| \cdot |f(\wt{A},x)_{j_1}| \cdot |f(\wt{A},x)_{j_0} | \cdot   \| c_g(A,x)^{\top}  - c_g(\wt{A},x)^{\top} \|_2 \cdot  \| h(A,x) \|_2 \cdot \| x \|_2 \cdot \|{\bf 1}_d^{\top} \|_2 \\
    & + ~ |f(\wt{A},x)_{j_1}| \cdot |f(\wt{A},x)_{j_1}| \cdot |f(\wt{A},x)_{j_0} | \cdot    \| c_g(\wt{A},x)^{\top} \|_2 \cdot   \| h(A,x) - h(\wt{A},x) \|_2 \cdot \| x \|_2 \cdot \|{\bf 1}_d^{\top} \|_2 \\
    \leq & ~ 10R^4 \beta^{-2} \cdot n \cdot \sqrt{d}  \exp(3R^2)\|A - \wt{A}\|_F \\
    & + ~ 10R^4 \beta^{-2} \cdot n \cdot \sqrt{d}  \exp(3R^2)\|A - \wt{A}\|_F \\
    & + ~ 10R^4 \beta^{-2} \cdot n \cdot \sqrt{d}  \exp(3R^2)\|A - \wt{A}\|_F \\
    & + ~ 20R^4 \beta^{-2} \cdot n \cdot \sqrt{d}  \exp(3R^2)\|A - \wt{A}\|_F \\
    & + ~ 15R^2  \beta^{-2} \cdot n \cdot \sqrt{d}  \exp(4R^2)\|A - \wt{A}\|_F \\
    \leq & ~ 65\beta^{-2} \cdot n \cdot \sqrt{d}  \exp(5R^2)\|A - \wt{A}\|_F \\
    \leq & ~ \beta^{-2} \cdot n \cdot \sqrt{d}  \exp(6R^2)\|A - \wt{A}\|_F 
\end{align*}

{\bf Proof of Part 13.}
\begin{align*}
  & ~ \|  B_{4,6,1}^{j_1,*,j_0,*} (A) - B_{4,6,1}^{j_1,*,j_0,*} ( \wt{A} ) \|_F \\
    =   & ~ \| f(A,x)_{j_1} \cdot f(A,x)_{j_0}  \cdot f_c( A,x) \cdot  c_g(A,x)^{\top} \cdot  h(A,x) \cdot x \cdot {\bf 1}_d^{\top} \\
    & ~ - f(\wt{A},x)_{j_1} \cdot f(\wt{A},x)_{j_0}  \cdot f_c( \wt{A},x) \cdot  c_g(\wt{A},x)^{\top} \cdot  h(\wt{A},x) \cdot x \cdot {\bf 1}_d^{\top} \|_F \\
    \leq & ~ |f(A,x)_{j_1} - f(\wt{A},x)_{j_1}| \cdot |f(A,x)_{j_0} | \cdot |f_c( A,x) |\cdot  \| c_g(A,x)^{\top} \|_2 \cdot  \| h(A,x) \|_2 \cdot \| x \|_2 \cdot \|{\bf 1}_d^{\top} \|_2 \\
    & + ~ |f(\wt{A},x)_{j_1}| \cdot |f(A,x)_{j_0} - f(\wt{A},x)_{j_0} | \cdot |f_c( A,x) |\cdot  \| c_g(A,x)^{\top} \|_2 \cdot   \| h(A,x) \|_2 \cdot \| x \|_2 \cdot \|{\bf 1}_d^{\top} \|_2 \\
    & + ~ |f(\wt{A},x)_{j_1}| \cdot |f(\wt{A},x)_{j_0} | \cdot |f_c( A,x)  -f_c( \wt{A},x)|\cdot  \| c_g(A,x)^{\top} \|_2 \cdot   \| h(A,x) \|_2 \cdot \| x \|_2 \cdot \|{\bf 1}_d^{\top} \|_2 \\
    & + ~ |f(\wt{A},x)_{j_1}| \cdot |f(\wt{A},x)_{j_0} | \cdot |f_c( \wt{A},x)|\cdot  \| c_g(A,x)^{\top}  - c_g(\wt{A},x)^{\top} \|_2 \cdot  \| h(A,x) \|_2 \cdot \| x \|_2 \cdot \|{\bf 1}_d^{\top} \|_2 \\
    & + ~ |f(\wt{A},x)_{j_1}| \cdot |f(\wt{A},x)_{j_0} | \cdot |f_c( \wt{A},x)|\cdot  \| c_g(\wt{A},x)^{\top} \|_2 \cdot   \| h(A,x) - h(\wt{A},x) \|_2 \cdot \| x \|_2 \cdot \|{\bf 1}_d^{\top} \|_2 \\
    \leq & ~ 20R^4 \beta^{-2} \cdot n \cdot \sqrt{d}  \exp(3R^2)\|A - \wt{A}\|_F \\
    & + ~ 20R^4 \beta^{-2} \cdot n \cdot \sqrt{d}  \exp(3R^2)\|A - \wt{A}\|_F \\
    & + ~ 30R^4 \beta^{-2} \cdot n \cdot \sqrt{d}  \exp(3R^2)\|A - \wt{A}\|_F \\
    & + ~ 40R^4 \beta^{-2} \cdot n \cdot \sqrt{d}  \exp(3R^2)\|A - \wt{A}\|_F \\
    & + ~ 30R^2  \beta^{-2} \cdot n \cdot \sqrt{d}  \exp(4R^2)\|A - \wt{A}\|_F \\
    \leq & ~ 140\beta^{-2} \cdot n \cdot \sqrt{d}  \exp(5R^2)\|A - \wt{A}\|_F \\
    \leq & ~ \beta^{-2} \cdot n \cdot \sqrt{d}  \exp(6R^2)\|A - \wt{A}\|_F 
\end{align*}

{\bf Proof of Part 14.}
\begin{align*}
  & ~ \|  B_{4,6,2}^{j_1,*,j_0,*} (A) - B_{4,6,2}^{j_1,*,j_0,*} ( \wt{A} ) \|_F \\
    =   & ~ \| f(A,x)_{j_1} \cdot f(A,x)_{j_0}  \cdot  c( A,x)_{j_1} \cdot  c_g(A,x)^{\top} \cdot  h(A,x) \cdot x \cdot {\bf 1}_d^{\top} \\
    & ~ - f(\wt{A},x)_{j_1} \cdot f(\wt{A},x)_{j_0}  \cdot c(\wt{A},x)_{j_1}  \cdot  c_g(\wt{A},x)^{\top} \cdot  h(\wt{A},x) \cdot x \cdot {\bf 1}_d^{\top} \|_F \\
    \leq & ~ |f(A,x)_{j_1} - f(\wt{A},x)_{j_1}| \cdot |f(A,x)_{j_0} | \cdot |c( A,x)_{j_1} |\cdot  \| c_g(A,x)^{\top} \|_2 \cdot  \| h(A,x) \|_2 \cdot \| x \|_2 \cdot \|{\bf 1}_d^{\top} \|_2 \\
    & + ~ |f(\wt{A},x)_{j_1}| \cdot |f(A,x)_{j_0} - f(\wt{A},x)_{j_0} | \cdot |c( A,x)_{j_1} |\cdot  \| c_g(A,x)^{\top} \|_2 \cdot   \| h(A,x) \|_2 \cdot \| x \|_2 \cdot \|{\bf 1}_d^{\top} \|_2 \\
    & + ~ |f(\wt{A},x)_{j_1}| \cdot |f(\wt{A},x)_{j_0} | \cdot |c( A,x)_{j_1} - c(\wt{A},x)_{j_1} |\cdot  \| c_g(A,x)^{\top} \|_2 \cdot   \| h(A,x) \|_2 \cdot \| x \|_2 \cdot \|{\bf 1}_d^{\top} \|_2 \\
    & + ~ |f(\wt{A},x)_{j_1}| \cdot |f(\wt{A},x)_{j_0} | \cdot |c(\wt{A},x)_{j_1}|\cdot  \| c_g(A,x)^{\top}  - c_g(\wt{A},x)^{\top} \|_2 \cdot  \| h(A,x) \|_2 \cdot \| x \|_2 \cdot \|{\bf 1}_d^{\top} \|_2 \\
    & + ~ |f(\wt{A},x)_{j_1}| \cdot |f(\wt{A},x)_{j_0} | \cdot |c(\wt{A},x)_{j_1}|\cdot  \| c_g(\wt{A},x)^{\top} \|_2 \cdot   \| h(A,x) - h(\wt{A},x) \|_2 \cdot \| x \|_2 \cdot \|{\bf 1}_d^{\top} \|_2 \\
    \leq & ~ 20R^4 \beta^{-2} \cdot n \cdot \sqrt{d}  \exp(3R^2)\|A - \wt{A}\|_F \\
    & + ~ 20R^4 \beta^{-2} \cdot n \cdot \sqrt{d}  \exp(3R^2)\|A - \wt{A}\|_F \\
    & + ~ 10R^4 \beta^{-2} \cdot n \cdot \sqrt{d}  \exp(3R^2)\|A - \wt{A}\|_F \\
    & + ~ 40R^4 \beta^{-2} \cdot n \cdot \sqrt{d}  \exp(3R^2)\|A - \wt{A}\|_F \\
    & + ~ 30R^2  \beta^{-2} \cdot n \cdot \sqrt{d}  \exp(4R^2)\|A - \wt{A}\|_F \\
    \leq & ~ 120\beta^{-2} \cdot n \cdot \sqrt{d}  \exp(5R^2)\|A - \wt{A}\|_F \\
    \leq & ~ \beta^{-2} \cdot n \cdot \sqrt{d}  \exp(6R^2)\|A - \wt{A}\|_F 
\end{align*}

{\bf Proof of Part 15.}
\begin{align*}
 & ~ \| B_{4}^{j_1,*,j_0,*} (A) - B_{4}^{j_1,*,j_0,*} ( \wt{A} ) \|_F \\
   = & ~ \|\sum_{i = 1}^6 B_{4,i}^{j_1,*,j_0,*}(A) -   B_{4,i}^{j_1,*,j_0,*}(\wt{A})  \|_F \\
    \leq & ~ 14  \beta^{-2} \cdot n \cdot \sqrt{d} \cdot \exp(6R^2)\|A - \wt{A}\|_F
\end{align*}

\end{proof}

\subsection{PSD}
\begin{lemma}\label{psd: B_4}
If the following conditions hold
\begin{itemize}
    \item Let $B_{4,1,1}^{j_1,*, j_0,*}, \cdots, B_{4,6,2}^{j_1,*, j_0,*} $ be defined as Lemma~\ref{lem:b_4_j1_j0} 
    \item  Let $\|A \|_2 \leq R, \|A^{\top} \|_F \leq R, \| x\|_2 \leq R, \|\diag(f(A,x)) \|_F \leq \|f(A,x) \|_2 \leq 1, \| b_g \|_2 \leq 1$ 
\end{itemize}
We have 
\begin{itemize}
    \item {\bf Part 1.} $\| B_{4,1,1}^{j_1,*, j_0,*} \| \leq 4 \sqrt{d} R^2$  
    \item {\bf Part 2.} $\|B_{4,1,2}^{j_1,*, j_0,*}\| \preceq 4 \sqrt{d} R^2$
    \item {\bf Part 3.} $\|B_{4,1,3}^{j_1,*, j_0,*}\| \preceq 4  R^4$
    \item {\bf Part 4.} $\|B_{4,1,4}^{j_1,*, j_0,*} \|\preceq 4 R^4$
    \item {\bf Part 5.} $\|B_{4,1,5}^{j_1,*, j_0,*} \|\preceq 4  R^4$
    \item {\bf Part 6.} $\|B_{4,1,6}^{j_1,*, j_0,*} \|\preceq 6  R^4$
    \item {\bf Part 7.} $\|B_{4,1,7}^{j_1,*, j_0,*} \|\preceq 12  R^4$
    \item {\bf Part 8.} $\|B_{4,2,1}^{j_1,*, j_0,*}\| \preceq 10\sqrt{d} R^4$
    \item {\bf Part 9.} $\|B_{4,3,1}^{j_1,*, j_0,*}\| \preceq 10 \sqrt{d}R^2$
    \item {\bf Part 10.} $\|B_{4,4,1}^{j_1,*, j_0,*}\| \preceq 20\sqrt{d} R^4$
    \item {\bf Part 11.} $\|B_{4,5,1}^{j_1,*, j_0,*}\| \preceq 5\sqrt{d} R^4$
    \item {\bf Part 12.} $\|B_{4,5,2}^{j_1,*, j_0,*}\| \preceq 5\sqrt{d}R^4$
    \item {\bf Part 13.} $\|B_{4,6,1}^{j_1,*, j_0,*}\| \preceq 10\sqrt{d}R^4$
    \item {\bf Part 14.} $\|B_{4,6,2}^{j_1,*, j_0,*}\| \preceq 10\sqrt{d} R^4$
    \item {\bf Part 15.} $\|B_{4}^{j_1,*, j_0,*}\| \preceq 108\sqrt{d} R^4$
\end{itemize}
\end{lemma}

\begin{proof}
    {\bf Proof of Part 1.}
    \begin{align*}
        \| B_{4,1,1}^{j_1,*, j_0,*} \| 
        = & ~ \|  f_c(A,x)^2 \cdot f(A,x)_{j_1} \cdot f(A,x)_{j_0} \cdot  h(A,x)\cdot  {\bf 1}_d^\top\| \\
        \leq & ~ |f(A,x)_{j_1}| \cdot |f(A,x)_{j_0}| \cdot |f_c(A,x)|^2 \cdot \|h(A,x)\|_2 \cdot \| {\bf 1}_d^\top\|_2\\
        \leq & ~ 4 \sqrt{d} R^2
    \end{align*}

    {\bf Proof of Part 2.}
    \begin{align*}
        \| B_{4,1,2}^{j_1,*, j_0,*} \|
        = &~
        \| f(A,x)_{j_1} \cdot f(A,x)_{j_0} \cdot c(A,x)_{j_1} \cdot  f_c(A,x) \cdot h(A,x) \cdot  {\bf 1}_d^\top  \| \\
        \preceq & ~  |f(A,x)_{j_1}| \cdot |f(A,x)_{j_0}| \cdot |c(A,x)_{j_1}|\cdot |f_c(A,x)|\cdot \|h(A,x)\|_2 \cdot \| {\bf 1}_d^\top\|_2\\
        \preceq & ~ 4 \sqrt{d} R^2
    \end{align*}

    {\bf Proof of Part 3.}
    \begin{align*}
     \| B_{4,1,3}^{j_1,*, j_0,*} \|
     = & ~
       \| f(A,x)_{j_1} \cdot  f(A,x)_{j_0} \cdot f_c(A,x)^2 \cdot  ((A_{j_1,*}) \circ x^\top) \cdot h(A,x)  \cdot I_d \| \\
    \leq & ~ |f(A,x)_{j_1}| \cdot |f(A,x)_{j_0}| \cdot  |f_c(A,x)|^2    \cdot \| A_{j_1,*}^\top \circ x \|_2 \cdot \|h(A,x)\|_2 \cdot \| I_d\| \\
    \leq & ~ |f(A,x)_{j_1}| \cdot |f(A,x)_{j_0}| \cdot  |f_c(A,x)|^2    \cdot \| A_{j_1,*}\|_2 \cdot \|\diag(x) \|_{\infty} \cdot \|h(A,x)\|_2 \cdot \| I_d\| \\
    \leq &  ~  4 R^4
    \end{align*}
   
    {\bf Proof of Part 4.}
    \begin{align*}
        \|B_{4,1,4}^{j_1,*, j_0,*}  \|
        = & ~  \|-f(A,x)_{j_1} \cdot  f(A,x)_{j_0} \cdot f_c(A,x)^2  \cdot h(A,x)^\top \cdot  h(A,x) \cdot I_d \|\\
        \leq & ~   \| f(A,x)_{j_1} \cdot  f(A,x)_{j_0} \cdot f_c(A,x)^2  \cdot h(A,x)^\top \cdot  h(A,x) \cdot I_d \|\\
        \leq & ~ | f(A,x)_{j_1} | \cdot|  f(A,x)_{j_0} |\cdot |f_c(A,x)|^2 \cdot \|h(A,x)^\top\|_2 \cdot  \|h(A,x)\|_2  \cdot \| I_d\|\\
        \leq & ~ | f(A,x)_{j_1} | \cdot|  f(A,x)_{j_0} |\cdot |f_c(A,x)|^2 \cdot \|h(A,x)^\top\|_2 \cdot  \|h(A,x)\|_2  \cdot \| I_d\| \\
        \leq & ~ 4 R^4
    \end{align*}

    {\bf Proof of Part 5.}
    \begin{align*}
        \|B_{4,1,5}^{j_1,*, j_0,*} \|
        = & ~ \| f(A,x)_{j_1} \cdot  f(A,x)_{j_0} \cdot (-f_2(A,x) + f(A,x)_{j_1})  \cdot f_c(A,x)  \cdot h(A,x)^\top \cdot  h(A,x) \cdot I_d\| \\
        \leq & ~  | f(A,x)_{j_1}| \cdot |f(A,x)_{j_0}|    \cdot (|-f_2(A,x)| + |f(A,x)_{j_1}|) \cdot |f_c(A,x)|  \cdot \|h(A,x)^\top\|_2  \cdot \|h(A,x)\|_2 \cdot \| I_d\|\\
        \leq & ~ 4 R^4
    \end{align*}

    {\bf Proof of Part 6.}
    \begin{align*}
        \|B_{4,1,6}^{j_1,*, j_0,*} \|
        = & ~ \| f(A,x)_{j_1} \cdot  f(A,x)_{j_0} \cdot (-f_c(A,x) + f(A,x)_{j_1}) \cdot f_c(A,x)   \cdot h(A,x)^\top \cdot  h(A,x) \cdot I_d \|\\
        \leq & ~  | f(A,x)_{j_1}| \cdot |f(A,x)_{j_0}|    \cdot (|-f_c(A,x)| + |f(A,x)_{j_1}|)  \cdot |f_c(A,x)|  \cdot \|h(A,x)^\top\|_2  \cdot \|h(A,x)\|_2 \cdot \| I_d\| \\
        \leq & ~ 6 R^4
    \end{align*}

    {\bf Proof of Part 7.}
    \begin{align*}
         \|B_{4,1,7}^{j_1,*, j_0,*} \|
         = & ~ \|f(A,x)_{j_1} \cdot  f(A,x)_{j_0} \cdot f_c(A,x)   \cdot p_{j_1}(A,x)^\top \cdot   h(A,x) \cdot I_d \|\\
         \leq & ~ | f(A,x)_{j_1}| \cdot |f(A,x)_{j_0}| \cdot |f_c(A,x)| \cdot \|p_{j_1}(A,x)^\top\|_2  \cdot \|h(A,x)\|_2 \cdot \| I_d\|\\
         \leq & ~ 12  R^4
    \end{align*}

    {\bf Proof of Part 8.}
    \begin{align*}
        \|B_{4,2,1}^{j_1,*, j_0,*} \|
        = & ~ \|  f(A,x)_{j_1} \cdot  f(A,x)_{j_0} \cdot f_c(A,x) \cdot c_g(A,x)^{\top} \cdot h(A,x) \cdot x \cdot {\bf 1}_d^{\top} \| \\
        \leq & ~  | f(A,x)_{j_1}| \cdot |f(A,x)_{j_0}| \cdot |f_c(A,x)| \cdot \|c_g(A,x)^{\top}\|_2 \cdot \|h(A,x)\|_2 \cdot \| x\|_2 \cdot \| {\bf 1}_d^{\top}\|_2\\
        \leq & ~ 10 \sqrt{d} R^4
    \end{align*}

    {\bf Proof of Part 9.}
    \begin{align*}
     \|B_{4,3,1}^{j_1,*, j_0,*} \|
        = & ~ \|   -  f(A,x)_{j_1} \cdot   f(A,x)_{j_0}  \cdot    f_c(A,x)   \cdot {\bf 1}_d \cdot c_g(A,x)^{\top} \cdot \diag (x)\| \\
        \leq & ~ \|   f(A,x)_{j_1} \cdot   f(A,x)_{j_0}  \cdot    f_c(A,x)   \cdot {\bf 1}_d \cdot c_g(A,x)^{\top} \cdot \diag (x)\| \\
        \leq & ~ | f(A,x)_{j_1}| \cdot  |f(A,x)_{j_0} | \cdot  |f_c( A,x) | \cdot \| {\bf 1}_d\|_2 \cdot \|c_g(A,x)^{\top}\|_2 \cdot \| \diag (x)\|  \\
        \leq & ~ 10 \sqrt{d} R^2
    \end{align*}

        {\bf Proof of Part 10.}
    \begin{align*}
     \|B_{4,4,1}^{j_1,*, j_0,*} \|
        = & ~ \|  -f(A,x)_{j_1} \cdot f(A,x)_{j_0}  \cdot f_c( A,x) \cdot  c_g(A,x)^{\top} \cdot h_e(A,x) \cdot x \cdot {\bf 1}_d^{\top}\| \\
        \leq & ~ \|  f(A,x)_{j_1} \cdot f(A,x)_{j_0}  \cdot f_c( A,x) \cdot  c_g(A,x)^{\top} \cdot h_e(A,x) \cdot x \cdot {\bf 1}_d^{\top}\| \\
        \leq & ~ | f(A,x)_{j_1}| \cdot  |f(A,x)_{j_0} | \cdot  |f_c( A,x) | \cdot |c_g(A,x)^{\top}| \cdot \| h_e(A,x) \|_2 \cdot \| x\|_2 \cdot \| {\bf 1}_d^{\top}\|_2\\
        \leq & ~ 20 \sqrt{d}R^4
    \end{align*}

       {\bf Proof of Part 11.}
    \begin{align*}
     \|B_{4,5,1}^{j_1,*, j_0,*} \|
        = & ~ \|  f(A,x)_{j_1} \cdot f(A,x)_{j_0} \cdot  f_2(A,x)  \cdot  c_g(A,x)^{\top} \cdot  h(A,x) \cdot x \cdot {\bf 1}_d^{\top}  \| \\
        \leq & ~ | f(A,x)_{j_1}|  \cdot  |f(A,x)_{j_0} | \cdot  |f_2(A,x) |  \cdot \|c_g(A,x)^{\top}\|_2 \cdot \|h(A,x)\|_2 \cdot \| x\|_2 \cdot \| {\bf 1}_d^{\top}\|_2\\
        \leq & ~ 5\sqrt{d} R^4
    \end{align*}

           {\bf Proof of Part 12.}
    \begin{align*}
     \|B_{4,5,2}^{j_1,*, j_0,*} \|
        = & ~ \|  f(A,x)_{j_1}^2 \cdot f(A,x)_{j_0}    \cdot  c_g(A,x)^{\top} \cdot  h(A,x) \cdot x \cdot {\bf 1}_d^{\top}  \| \\
        \leq & ~ | f(A,x)_{j_1}|^2  \cdot  |f(A,x)_{j_0} |  \cdot \|c_g(A,x)^{\top}\|_2 \cdot \|h(A,x)\|_2 \cdot \| x\|_2 \cdot \| {\bf 1}_d^{\top}\|_2\\
        \leq & ~ 5\sqrt{d} R^4
    \end{align*}

           {\bf Proof of Part 13.}
    \begin{align*}
     \|B_{4,6,1}^{j_1,*, j_0,*} \|
        = & ~ \|  f(A,x)_{j_1} \cdot f(A,x)_{j_0} \cdot  f_c(A,x)  \cdot  c_g(A,x)^{\top} \cdot  h(A,x) \cdot x \cdot {\bf 1}_d^{\top}  \| \\
        \leq & ~ | f(A,x)_{j_1}|  \cdot  |f(A,x)_{j_0} | \cdot  |f_c(A,x) |  \cdot \|c_g(A,x)^{\top}\|_2 \cdot \|h(A,x)\|_2 \cdot \| x\|_2 \cdot \| {\bf 1}_d^{\top}\|_2\\
        \leq & ~ 10\sqrt{d} R^4
    \end{align*}

           {\bf Proof of Part 14.}
    \begin{align*}
     \|B_{4,6,2}^{j_1,*, j_0,*} \|
        = & ~ \|  f(A,x)_{j_1}  \cdot f(A,x)_{j_0}    \cdot  c(A,x)_{j_1} \cdot c_g(A,x)^{\top} \cdot  h(A,x) \cdot x \cdot {\bf 1}_d^{\top}  \| \\
        \leq & ~ | f(A,x)_{j_1}|   \cdot  |f(A,x)_{j_0} |  \cdot |c(A,x)_{j_1} |  \cdot\|c_g(A,x)^{\top}\|_2 \cdot \|h(A,x)\|_2 \cdot \| x\|_2 \cdot \| {\bf 1}_d^{\top}\|_2\\
        \leq & ~ 10\sqrt{d} R^4
    \end{align*}

            {\bf Proof of Part 15}
\begin{align*}
    & ~ \| B_{4}^{j_1,*,j_0,*} \|  \\
    = & ~ \|\sum_{i = 1}^6 B_{4,i}^{j_1,*,j_0,*}  \|  \\
    \leq & ~  108\sqrt{d} R^4
\end{align*}
\end{proof}

\newpage
\section{Hessian: Fifth term  \texorpdfstring{$H_5^{j_1,i_1,j_0,i_0}$}{}}\label{app:hessian_fifth}
\subsection{Definitions}
\begin{definition}\label{def:b_5} 
    We define the $B_5^{j_1,i_1,j_0,i_0}$ as follows,
    \begin{align*}
        B_5^{j_1,i_1,j_0,i_0} := \frac{\d}{\d A_{j_1,i_1}}(c_g(A,x)^{\top} \cdot f(A,x)_{j_0} \cdot \diag (x) A^{\top} \cdot  f(A,x)  \cdot    (\langle -f(A,x), f(A,x) \rangle + f(A,x)_{j_0}))
    \end{align*}
    Then, we define $B_{5,1}^{j_1,i_1,j_0,i_0}, \cdots, B_{5,6}^{j_1,i_1,j_0,i_0}$ as follow
    \begin{align*}
         B_{5,1}^{j_1,i_1,j_0,i_0} := & ~ \frac{\d}{\d A_{j_1,i_1}} (- c_g(A,x)^{\top} )    \cdot f(A,x)_{j_0} \cdot \diag(x) \cdot A^{\top} \cdot f(A,x) \cdot (\langle -f(A,x), f(A,x) \rangle + f(A,x)_{j_0})\\
 B_{5,2}^{j_1,i_1,j_0,i_0} := &   ~ - c_g(A,x)^{\top} \cdot \frac{\d}{\d A_{j_1,i_1}} (f(A,x)_{j_0}) \cdot \diag(x) \cdot A^{\top} \cdot f(A,x) \cdot (\langle -f(A,x), f(A,x) \rangle + f(A,x)_{j_0})\\
 B_{5,3}^{j_1,i_1,j_0,i_0} := &   ~ - c_g(A,x)^{\top} \cdot f(A,x)_{j_0} \cdot \diag(x)  \cdot \frac{\d}{\d A_{j_1,i_1}} (A^{\top}) \cdot  f(A,x) \cdot (\langle -f(A,x), f(A,x) \rangle + f(A,x)_{j_0})\\
 B_{5,4}^{j_1,i_1,j_0,i_0} := &   ~ - c_g(A,x)^{\top} \cdot f(A,x)_{j_0} \cdot \diag(x) \cdot A^{\top} \cdot \frac{\d}{\d A_{j_1,i_1}}(f(A,x)) \cdot (\langle -f(A,x), f(A,x) \rangle + f(A,x)_{j_0})\\
  B_{5,5}^{j_1,i_1,j_0,i_0} : = & ~   c_g(A,x)^{\top} \cdot f(A,x)_{j_0} \cdot \diag(x) \cdot A^{\top} \cdot  f(A,x)  \cdot \langle \frac{\d  f(A,x)}{\d A_{j_1,i_1}},f(A,x) \rangle \\
  B_{5,6}^{j_1,i_1,j_0,i_0} : = & ~   c_g(A,x)^{\top} \cdot f(A,x)_{j_0} \cdot \diag(x) \cdot A^{\top} \cdot  f(A,x)  \cdot \langle f(A,x), \frac{\d f(A,x) }{\d A_{j_1,i_1}} \rangle\\
   B_{5,7}^{j_1,i_1,j_0,i_0} : = & ~  - c_g(A,x)^{\top} \cdot f(A,x)_{j_0} \cdot \diag(x) \cdot A^{\top} \cdot  f(A,x)  \cdot  \frac{\d f(A,x)_{j_0}}{\d A_{j_1,i_1}} 
    \end{align*}
    It is easy to show
    \begin{align*}
        B_5^{j_1,i_1,j_0,i_0} = B_{5,1}^{j_1,i_1,j_0,i_0} +  B_{5,2}^{j_1,i_1,j_0,i_0} + B_{5,3}^{j_1,i_1,j_0,i_0} + B_{5,4}^{j_1,i_1,j_0,i_0} + B_{5,5}^{j_1,i_1,j_0,i_0} + B_{5,6}^{j_1,i_1,j_0,i_0} + B_{5,7}^{j_1,i_1,j_0,i_0}
    \end{align*}
       Similarly for $j_1 = j_0$ and $i_0 = i_1$,we have
    \begin{align*}
        B_5^{j_1,i_1,j_1,i_1} = B_{5,1}^{j_1,i_1,j_1,i_1} +  B_{5,2}^{j_1,i_1,j_1,i_1} + B_{5,3}^{j_1,i_1,j_1,i_1} + B_{5,4}^{j_1,i_1,j_1,i_1} + B_{5,5}^{j_1,i_1,j_1,i_1} + B_{5,6}^{j_1,i_1,j_1,i_1} + B_{5,7}^{j_1,i_1,j_1,i_1}
         \end{align*}
    For $j_1 = j_0$ and $i_0 \neq i_1$,we have
    \begin{align*}
       B_5^{j_1,i_1,j_1,i_0} = B_{5,1}^{j_1,i_1,j_1,i_0} +  B_{5,2}^{j_1,i_1,j_1,i_0} + B_{5,3}^{j_1,i_1,j_1,i_0} + B_{5,4}^{j_1,i_1,j_1,i_0} + B_{5,5}^{j_1,i_1,j_1,i_0} + B_{5,6}^{j_1,i_1,j_1,i_0} + B_{5,7}^{j_1,i_1,j_1,i_0}
    \end{align*}
    For $j_1 \neq j_0$ and $i_0 = i_1$,we have
    \begin{align*}
       B_5^{j_1,i_1,j_0,i_1} = B_{5,1}^{j_1,i_1,j_0,i_1} +  B_{5,2}^{j_1,i_1,j_0,i_1} + B_{5,3}^{j_1,i_1,j_0,i_1} + B_{5,4}^{j_1,i_1,j_0,i_1} + B_{5,5}^{j_1,i_1,j_0,i_1} + B_{5,6}^{j_1,i_1,j_0,i_1} + B_{5,7}^{j_1,i_1,j_0,i_1}
    \end{align*}
\end{definition}
\subsection{Case \texorpdfstring{$j_1=j_0, i_1 = i_0$}{}}
\begin{lemma}
For $j_1 = j_0$ and $i_0 = i_1$. If the following conditions hold
    \begin{itemize}
     \item Let $u(A,x) \in \R^n$ be defined as Definition~\ref{def:u}
    \item Let $\alpha(A,x) \in \R$ be defined as Definition~\ref{def:alpha}
     \item Let $f(A,x) \in \R^n$ be defined as Definition~\ref{def:f}
    \item Let $c(A,x) \in \R^n$ be defined as Definition~\ref{def:c}
    \item Let $g(A,x) \in \R^d$ be defined as Definition~\ref{def:g} 
    \item Let $q(A,x) = c(A,x) + f(A,x) \in \R^n$
    \item Let $c_g(A,x) \in \R^d$ be defined as Definition~\ref{def:c_g}.
    \item Let $L_g(A,x) \in \R$ be defined as Definition~\ref{def:l_g}
    \item Let $v \in \R^n$ be a vector 
    \item Let $B_1^{j_1,i_1,j_0,i_0}$ be defined as Definition~\ref{def:b_1}
    \end{itemize}
    Then, For $j_0,j_1 \in [n], i_0,i_1 \in [d]$, we have 
    \begin{itemize}
\item {\bf Part 1.} For $B_{5,1}^{j_1,i_1,j_1,i_1}$, we have 
\begin{align*}
 B_{5,1}^{j_1,i_1,j_1,i_1}  = & ~  \frac{\d}{\d A_{j_1,i_1}} (- c_g(A,x)^{\top} ) \cdot  f(A,x)_{j_1} \cdot \diag (x) A^{\top} \cdot  f(A,x)  \cdot  (\langle -f(A,x), f(A,x) \rangle + f(A,x)_{j_1})\\
 = & ~ B_{5,1,1}^{j_1,i_1,j_1,i_1} + B_{5,1,2}^{j_1,i_1,j_1,i_1} + B_{5,1,3}^{j_1,i_1,j_1,i_1} + B_{5,1,4}^{j_1,i_1,j_1,i_1} + B_{5,1,5}^{j_1,i_1,j_1,i_1} + B_{5,1,6}^{j_1,i_1,j_1,i_1} + B_{5,1,7}^{j_1,i_1,j_1,i_1}
\end{align*} 
\item {\bf Part 2.} For $B_{5,2}^{j_1,i_1,j_1,i_1}$, we have 
\begin{align*}
  & ~B_{5,2}^{j_1,i_1,j_1,i_1} \\
  = & ~ - c_g(A,x)^{\top} \cdot \frac{\d}{\d A_{j_1,i_1}} ( f(A,x)_{j_1} )  \cdot \diag(x) \cdot A^{\top} \cdot f(A,x) \cdot(\langle -f(A,x), f(A,x) \rangle + f(A,x)_{j_1}) \\
    = & ~  B_{5,2,1}^{j_1,i_1,j_1,i_1} + B_{5,2,2}^{j_1,i_1,j_1,i_1}
\end{align*} 
\item {\bf Part 3.} For $B_{5,3}^{j_1,i_1,j_1,i_1}$, we have 
\begin{align*}
 & ~ B_{5,3}^{j_1,i_1,j_1,i_1} \\= & ~ - c_g(A,x)^{\top} \cdot f(A,x)_{j_1} \cdot \diag(x) \cdot \frac{\d}{\d A_{j_1,i_1}} (  A^{\top} ) \cdot f(A,x) \cdot(\langle -f(A,x), f(A,x) \rangle + f(A,x)_{j_1}) \\
     = & ~ B_{5,3,1}^{j_1,i_1,j_1,i_1}  
\end{align*} 
\item {\bf Part 4.} For $B_{5,4}^{j_1,i_1,j_1,i_1}$, we have 
\begin{align*}
  B_{5,4}^{j_1,i_1,j_1,i_1} = & ~ - c_g(A,x)^{\top} \cdot f(A,x)_{j_1} \cdot \diag(x) \cdot A^{\top} \cdot \frac{\d f(A,x)}{\d A_{j_1,i_1}} \cdot(\langle -f(A,x), f(A,x) \rangle + f(A,x)_{j_1}) \\
     = & ~ B_{5,4,1}^{j_1,i_1,j_1,i_1} 
\end{align*}
\item {\bf Part 5.} For $B_{5,5}^{j_1,i_1,j_1,i_1}$, we have 
\begin{align*}
  B_{5,5}^{j_1,i_1,j_1,i_1} = & ~  c_g(A,x)^{\top} \cdot f(A,x)_{j_1} \cdot \diag(x) \cdot A^{\top} \cdot  f(A,x)  \cdot \langle f(A,x), \frac{\d f(A,x) }{\d A_{j_1,i_1}}\rangle \\
     = & ~ B_{5,5,1}^{j_1,i_1,j_1,i_1}  + B_{5,5,2}^{j_1,i_1,j_1,i_1}  
\end{align*}
\item {\bf Part 6.} For $B_{5,6}^{j_1,i_1,j_1,i_1}$, we have 
\begin{align*}
  B_{5,6}^{j_1,i_1,j_1,i_1} = & ~ c_g(A,x)^{\top} \cdot f(A,x)_{j_1} \cdot \diag(x) \cdot A^{\top} \cdot  f(A,x)  \cdot \langle \frac{\d  f(A,x) }{\d A_{j_1,i_1}},f(A,x)\rangle \\
     = & ~ B_{5,6,1}^{j_1,i_1,j_1,i_1}  + B_{5,6,2}^{j_1,i_1,j_1,i_1}  
\end{align*}
\item {\bf Part 7.} For $B_{5,7}^{j_1,i_1,j_1,i_1}$, we have 
\begin{align*}
  B_{5,7}^{j_1,i_1,j_1,i_1} = & ~ - c_g(A,x)^{\top} \cdot f(A,x)_{j_1} \cdot \diag(x) \cdot A^{\top} \cdot  f(A,x)  \cdot  \frac{\d f(A,x)_{j_1}}{\d A_{j_1,i_1}} \\
     = & ~ B_{5,7,1}^{j_1,i_1,j_1,i_1}  + B_{5,7,2}^{j_1,i_1,j_1,i_1}  
\end{align*}
\end{itemize}
\begin{proof}
    {\bf Proof of Part 1.}
    \begin{align*}
    B_{5,1,1}^{j_1,i_1,j_1,i_1} : = & ~ e_{i_1}^\top \cdot \langle c(A,x), f(A,x) \rangle \cdot  f(A,x)_{j_1}^2 \cdot \diag(x) \cdot A^{\top} \cdot f(A,x) \cdot(\langle -f(A,x), f(A,x) \rangle + f(A,x)_{j_1})\\
    B_{5,1,2}^{j_1,i_1,j_1,i_1} : = & ~   e_{i_1}^\top \cdot c(A,x)_{j_1}\cdot f(A,x)_{j_1}^2 \cdot \diag(x) \cdot A^{\top} \cdot f(A,x) \cdot(\langle -f(A,x), f(A,x) \rangle + f(A,x)_{j_1})\\
    B_{5,1,3}^{j_1,i_1,j_1,i_1} : = & ~ f(A,x)_{j_1}^2 \cdot \langle c(A,x), f(A,x) \rangle \cdot ( (A_{j_1,*}) \circ x^\top  )  \cdot \diag(x) \cdot A^{\top} \cdot f(A,x)\\& ~ \cdot (\langle -f(A,x), f(A,x) \rangle + f(A,x)_{j_1})\\
    B_{5,1,4}^{j_1,i_1,j_1,i_1} : = & ~   -  f(A,x)_{j_1}^2 \cdot f(A,x)^\top  \cdot A \cdot (\diag(x))^2 \\
    & ~ \cdot   \langle c(A,x), f(A,x) \rangle \cdot A^{\top} \cdot f(A,x) \cdot(\langle -f(A,x), f(A,x) \rangle + f(A,x)_{j_1})\\
    B_{5,1,5}^{j_1,i_1,j_1,i_1} : = & ~    f(A,x)_{j_1}^2 \cdot f(A,x)^\top  \cdot A \cdot (\diag(x))^2 \cdot (\langle (-f(A,x)), f(A,x) \rangle + f(A,x)_{j_1})^2  \cdot A^{\top} \cdot f(A,x) \\
    B_{5,1,6}^{j_1,i_1,j_1,i_1} : = & ~  f(A,x)_{j_1}^2 \cdot f(A,x)^\top  \cdot A \cdot  (\diag(x))^2 \cdot(\langle -f(A,x), c(A,x) \rangle + f(A,x)_{j_1}) \cdot A^{\top} \cdot f(A,x) \\
        & ~ \cdot(\langle -f(A,x), f(A,x) \rangle + f(A,x)_{j_1})\\
    B_{5,1,7}^{j_1,i_1,j_1,i_1} : = & ~   f(A,x)_{j_1}^2 \cdot ((e_{j_1}^\top - f(A,x)^\top) \circ q(A,x)^\top) \cdot A \cdot  (\diag(x))^2 \cdot A^{\top} \cdot f(A,x) \\
    & ~ \cdot(\langle -f(A,x), f(A,x) \rangle + f(A,x)_{j_1})
\end{align*}
Finally, combine them and we have
\begin{align*}
       B_{5,1}^{j_1,i_1,j_1,i_1} = B_{5,1,1}^{j_1,i_1,j_1,i_1} + B_{5,1,2}^{j_1,i_1,j_1,i_1} + B_{5,1,3}^{j_1,i_1,j_1,i_1} + B_{5,1,4}^{j_1,i_1,j_1,i_1} + B_{5,1,5}^{j_1,i_1,j_1,i_1} + B_{5,1,6}^{j_1,i_1,j_1,i_1} + B_{5,1,7}^{j_1,i_1,j_1,i_1}
\end{align*}
{\bf Proof of Part 2.}
    \begin{align*}
    B_{5,2,1}^{j_1,i_1,j_1,i_1} : = & ~   f(A,x)_{j_1}^2 \cdot x_{i_1} \cdot c_g(A,x)^{\top} \cdot \diag(x) \cdot A^{\top} \cdot f(A,x) \cdot(\langle -f(A,x), f(A,x) \rangle + f(A,x)_{j_1}) \\
    B_{5,2,2}^{j_1,i_1,j_1,i_1} : = & ~ - f(A,x)_{j_1} \cdot x_{i_1} \cdot c_g(A,x)^{\top} \cdot \diag(x) \cdot A^{\top} \cdot f(A,x) \cdot(\langle -f(A,x), f(A,x) \rangle + f(A,x)_{j_1})
\end{align*}
Finally, combine them and we have
\begin{align*}
       B_{5,2}^{j_1,i_1,j_1,i_1} = B_{5,2,1}^{j_1,i_1,j_1,i_1} + B_{5,2,2}^{j_1,i_1,j_1,i_1}
\end{align*}
{\bf Proof of Part 3.} 
    \begin{align*}
    B_{5,3,1}^{j_1,i_1,j_1,i_1} : = & ~    -  c_g(A,x)^{\top} \cdot f(A,x)_{j_1} \cdot \diag(x) \cdot e_{i_1} \cdot e_{j_1}^\top \cdot f(A,x) \cdot(\langle -f(A,x), f(A,x) \rangle + f(A,x)_{j_1})
\end{align*}
Finally, combine them and we have
\begin{align*}
       B_{5,3}^{j_1,i_1,j_1,i_1} = B_{5,3,1}^{j_1,i_1,j_1,i_1} 
\end{align*}
{\bf Proof of Part 4.} 
    \begin{align*}
    B_{5,4,1}^{j_1,i_1,j_1,i_1} : = & ~ - c_g(A,x)^{\top} \cdot f(A,x)_{j_1}^2 \cdot \diag(x) \cdot A^{\top}\cdot x_i \cdot  (e_{j_1}- f(A,x) ) \cdot (\langle -f(A,x), f(A,x) \rangle + f(A,x)_{j_1})
\end{align*}
Finally, combine them and we have
\begin{align*}
       B_{5,4}^{j_1,i_1,j_1,i_1} = B_{5,4,1}^{j_1,i_1,j_1,i_1}  
\end{align*}
{\bf Proof of Part 5.} 
    \begin{align*}
    B_{5,5,1}^{j_1,i_1,j_1,i_1} : = & ~ c_g(A,x)^{\top} \cdot f(A,x)_{j_1}^2 \cdot \diag(x) \cdot A^{\top} \cdot  f(A,x)  \cdot  x_{i_1} \cdot \langle - f(A,x), f(A,x) \rangle\\
    B_{5,5,2}^{j_1,i_1,j_1,i_1} : = & ~ c_g(A,x)^{\top} \cdot f(A,x)_{j_1}^3 \cdot \diag(x) \cdot A^{\top} \cdot  f(A,x)  \cdot  x_{i_1}  
\end{align*}
Finally, combine them and we have
\begin{align*}
       B_{5,5}^{j_1,i_1,j_1,i_1} = B_{5,5,1}^{j_1,i_1,j_1,i_1}  +B_{5,5,2}^{j_1,i_1,j_1,i_1}
\end{align*}
{\bf Proof of Part 6.} 
    \begin{align*}
     B_{5,6,1}^{j_1,i_1,j_1,i_1} : = & ~ c_g(A,x)^{\top} \cdot f(A,x)_{j_1}^2 \cdot \diag(x) \cdot A^{\top} \cdot  f(A,x)  \cdot  x_{i_1} \cdot \langle - f(A,x), f(A,x) \rangle\\
    B_{5,6,2}^{j_1,i_1,j_1,i_1} : = & ~ c_g(A,x)^{\top} \cdot f(A,x)_{j_1}^3 \cdot \diag(x) \cdot A^{\top} \cdot  f(A,x)  \cdot  x_{i_1}  
\end{align*}
Finally, combine them and we have
\begin{align*}
       B_{5,6}^{j_1,i_1,j_1,i_1} = B_{5,6,1}^{j_1,i_1,j_1,i_1}  +B_{5,6,2}^{j_1,i_1,j_1,i_1}
\end{align*}
{\bf Proof of Part 7.} 
    \begin{align*}
     B_{5,7,1}^{j_1,i_1,j_1,i_1} : = & ~  f(A,x)_{j_1}^3 \cdot  x_{i_1}\cdot c_g(A,x)^{\top} \cdot \diag(x) A^{\top} \cdot  f(A,x)   \\
    B_{5,7,2}^{j_1,i_1,j_1,i_1} : = & ~ -  f(A,x)_{j_1}^2 \cdot  x_{i_1} c_g(A,x)^{\top} \cdot \diag(x) \cdot A^{\top} \cdot  f(A,x)  
\end{align*}
Finally, combine them and we have
\begin{align*}
       B_{5,7}^{j_1,i_1,j_1,i_1} = B_{5,7,1}^{j_1,i_1,j_1,i_1}  +B_{5,7,2}^{j_1,i_1,j_1,i_1}
\end{align*}
\end{proof}
\end{lemma}

\subsection{Case \texorpdfstring{$j_1=j_0, i_1 \neq i_0$}{}}
\begin{lemma}
For $j_1 = j_0$ and $i_0 \neq i_1$. If the following conditions hold
    \begin{itemize}
     \item Let $u(A,x) \in \R^n$ be defined as Definition~\ref{def:u}
    \item Let $\alpha(A,x) \in \R$ be defined as Definition~\ref{def:alpha}
     \item Let $f(A,x) \in \R^n$ be defined as Definition~\ref{def:f}
    \item Let $c(A,x) \in \R^n$ be defined as Definition~\ref{def:c}
    \item Let $g(A,x) \in \R^d$ be defined as Definition~\ref{def:g} 
    \item Let $q(A,x) = c(A,x) + f(A,x) \in \R^n$
    \item Let $c_g(A,x) \in \R^d$ be defined as Definition~\ref{def:c_g}.
    \item Let $L_g(A,x) \in \R$ be defined as Definition~\ref{def:l_g}
    \item Let $v \in \R^n$ be a vector 
    \item Let $B_1^{j_1,i_1,j_0,i_0}$ be defined as Definition~\ref{def:b_1}
    \end{itemize}
    Then, For $j_0,j_1 \in [n], i_0,i_1 \in [d]$, we have 
    \begin{itemize}
\item {\bf Part 1.} For $B_{5,1}^{j_1,i_1,j_1,i_0}$, we have 
\begin{align*}
 B_{5,1}^{j_1,i_1,j_1,i_0}  = & ~  \frac{\d}{\d A_{j_1,i_1}} (- c_g(A,x)^{\top} ) \cdot  f(A,x)_{j_1} \cdot \diag (x) A^{\top} \cdot  f(A,x)  \cdot  (\langle -f(A,x), f(A,x) \rangle + f(A,x)_{j_1})\\
 = & ~ B_{5,1,1}^{j_1,i_1,j_1,i_0} + B_{5,1,2}^{j_1,i_1,j_1,i_0} + B_{5,1,3}^{j_1,i_1,j_1,i_0} + B_{5,1,4}^{j_1,i_1,j_1,i_0} + B_{5,1,5}^{j_1,i_1,j_1,i_0} + B_{5,1,6}^{j_1,i_1,j_1,i_0} + B_{5,1,7}^{j_1,i_1,j_1,i_0}
\end{align*} 
\item {\bf Part 2.} For $B_{5,2}^{j_1,i_1,j_1,i_0}$, we have 
\begin{align*}
 & ~ B_{5,2}^{j_1,i_1,j_1,i_0} \\= & ~ - c_g(A,x)^{\top} \cdot \frac{\d}{\d A_{j_1,i_1}} ( f(A,x)_{j_1} )  \cdot \diag(x) \cdot A^{\top} \cdot f(A,x) \cdot(\langle -f(A,x), f(A,x) \rangle + f(A,x)_{j_1}) \\
    = & ~  B_{5,2,1}^{j_1,i_1,j_1,i_0} + B_{5,2,2}^{j_1,i_1,j_1,i_0}
\end{align*} 
\item {\bf Part 3.} For $B_{5,3}^{j_1,i_1,j_1,i_0}$, we have 
\begin{align*}
  & ~B_{5,3}^{j_1,i_1,j_1,i_0} \\= & ~ - c_g(A,x)^{\top} \cdot f(A,x)_{j_1} \cdot \diag(x) \cdot \frac{\d}{\d A_{j_1,i_1}} (  A^{\top} ) \cdot f(A,x) \cdot(\langle -f(A,x), f(A,x) \rangle + f(A,x)_{j_1}) \\
     = & ~ B_{5,3,1}^{j_1,i_1,j_1,i_0}  
\end{align*} 
\item {\bf Part 4.} For $B_{5,4}^{j_1,i_1,j_1,i_0}$, we have 
\begin{align*}
  B_{5,4}^{j_1,i_1,j_1,i_0} = & ~ - c_g(A,x)^{\top} \cdot f(A,x)_{j_1} \cdot \diag(x) \cdot A^{\top} \cdot \frac{\d f(A,x)}{\d A_{j_1,i_1}} \cdot(\langle -f(A,x), f(A,x) \rangle + f(A,x)_{j_1}) \\
     = & ~ B_{5,4,1}^{j_1,i_1,j_1,i_0} 
\end{align*}
\item {\bf Part 5.} For $B_{5,5}^{j_1,i_1,j_1,i_0}$, we have 
\begin{align*}
  B_{5,5}^{j_1,i_1,j_1,i_0} = & ~  c_g(A,x)^{\top} \cdot f(A,x)_{j_1} \cdot \diag(x) \cdot A^{\top} \cdot  f(A,x)  \cdot \langle f(A,x), \frac{\d f(A,x) }{\d A_{j_1,i_1}}\rangle \\
     = & ~ B_{5,5,1}^{j_1,i_1,j_1,i_0}  + B_{5,5,2}^{j_1,i_1,j_1,i_0}  
\end{align*}
\item {\bf Part 6.} For $B_{5,6}^{j_1,i_1,j_1,i_0}$, we have 
\begin{align*}
  B_{5,6}^{j_1,i_1,j_1,i_0} = & ~ c_g(A,x)^{\top} \cdot f(A,x)_{j_1} \cdot \diag(x) \cdot A^{\top} \cdot  f(A,x)  \cdot \langle \frac{\d  f(A,x) }{\d A_{j_1,i_1}},f(A,x)\rangle \\
     = & ~ B_{5,6,1}^{j_1,i_1,j_1,i_0}  + B_{5,6,2}^{j_1,i_1,j_1,i_0}  
\end{align*}
\item {\bf Part 7.} For $B_{5,7}^{j_1,i_1,j_1,i_0}$, we have 
\begin{align*}
  B_{5,7}^{j_1,i_1,j_1,i_0} = & ~ - c_g(A,x)^{\top} \cdot f(A,x)_{j_1} \cdot \diag(x) \cdot A^{\top} \cdot  f(A,x)  \cdot  \frac{\d f(A,x)_{j_1}}{\d A_{j_1,i_1}} \\
     = & ~ B_{5,7,1}^{j_1,i_1,j_1,i_0}  + B_{5,7,2}^{j_1,i_1,j_1,i_0}  
\end{align*}
\end{itemize}
\begin{proof}
    {\bf Proof of Part 1.}
    \begin{align*}
    B_{5,1,1}^{j_1,i_1,j_1,i_0} : = & ~ e_{i_1}^\top \cdot \langle c(A,x), f(A,x) \rangle \cdot  f(A,x)_{j_1}^2 \cdot \diag(x) \cdot A^{\top} \cdot f(A,x) \cdot(\langle -f(A,x), f(A,x) \rangle + f(A,x)_{j_1})\\
    B_{5,1,2}^{j_1,i_1,j_1,i_0} : = & ~   e_{i_1}^\top \cdot c(A,x)_{j_1}\cdot f(A,x)_{j_1}^2 \cdot \diag(x) \cdot A^{\top} \cdot f(A,x) \\
    & ~ \cdot(\langle -f(A,x), f(A,x) \rangle + f(A,x)_{j_1})\\
    B_{5,1,3}^{j_1,i_1,j_1,i_0} : = & ~ f(A,x)_{j_1}^2 \cdot \langle c(A,x), f(A,x) \rangle \cdot ( (A_{j_1,*}) \circ x^\top  )  \cdot \diag(x) \cdot A^{\top} \cdot f(A,x) \\& ~\cdot (\langle -f(A,x), f(A,x) \rangle + f(A,x)_{j_1})\\
    B_{5,1,4}^{j_1,i_1,j_1,i_0} : = & ~   -  f(A,x)_{j_1}^2 \cdot f(A,x)^\top  \cdot A \cdot (\diag(x))^2 \cdot   \langle c(A,x), f(A,x) \rangle \cdot A^{\top} \cdot f(A,x) \\
    & ~ \cdot(\langle -f(A,x), f(A,x) \rangle + f(A,x)_{j_1})\\
    B_{5,1,5}^{j_1,i_1,j_1,i_0} : = & ~    f(A,x)_{j_1}^2 \cdot f(A,x)^\top  \cdot A \cdot (\diag(x))^2 \cdot (\langle (-f(A,x)), f(A,x) \rangle + f(A,x)_{j_1})^2  \cdot A^{\top} \cdot f(A,x) \\
    B_{5,1,6}^{j_1,i_1,j_1,i_0} : = & ~  f(A,x)_{j_1}^2 \cdot f(A,x)^\top  \cdot A \cdot  (\diag(x))^2 \cdot(\langle -f(A,x), c(A,x) \rangle + f(A,x)_{j_1}) \cdot A^{\top} \cdot f(A,x) \\
        & ~ \cdot(\langle -f(A,x), f(A,x) \rangle + f(A,x)_{j_1})\\
    B_{5,1,7}^{j_1,i_1,j_1,i_0} : = & ~   f(A,x)_{j_1}^2 \cdot ((e_{j_1}^\top - f(A,x)^\top) \circ q(A,x)^\top) \cdot A \cdot  (\diag(x))^2 \cdot A^{\top} \cdot f(A,x)\\
    & ~ \cdot(\langle -f(A,x), f(A,x) \rangle + f(A,x)_{j_1})
\end{align*}
Finally, combine them and we have
\begin{align*}
       B_{5,1}^{j_1,i_1,j_1,i_0} = B_{5,1,1}^{j_1,i_1,j_1,i_0} + B_{5,1,2}^{j_1,i_1,j_1,i_0} + B_{5,1,3}^{j_1,i_1,j_1,i_0} + B_{5,1,4}^{j_1,i_1,j_1,i_0} + B_{5,1,5}^{j_1,i_1,j_1,i_0} + B_{5,1,6}^{j_1,i_1,j_1,i_0} + B_{5,1,7}^{j_1,i_1,j_1,i_0}
\end{align*}
{\bf Proof of Part 2.}
    \begin{align*}
    B_{5,2,1}^{j_1,i_1,j_1,i_0} : = & ~   f(A,x)_{j_1}^2 \cdot x_{i_1} \cdot c_g(A,x)^{\top} \cdot \diag(x) \cdot A^{\top} \cdot f(A,x) \cdot(\langle -f(A,x), f(A,x) \rangle + f(A,x)_{j_1}) \\
    B_{5,2,2}^{j_1,i_1,j_1,i_0} : = & ~ - f(A,x)_{j_1} \cdot x_{i_1} \cdot c_g(A,x)^{\top} \cdot \diag(x) \cdot A^{\top} \cdot f(A,x) \cdot(\langle -f(A,x), f(A,x) \rangle + f(A,x)_{j_1})
\end{align*}
Finally, combine them and we have
\begin{align*}
       B_{5,2}^{j_1,i_1,j_1,i_0} = B_{5,2,1}^{j_1,i_1,j_1,i_0} + B_{5,2,2}^{j_1,i_1,j_1,i_0}
\end{align*}
{\bf Proof of Part 3.} 
    \begin{align*}
    B_{5,3,1}^{j_1,i_1,j_1,i_0} : = & ~    -  c_g(A,x)^{\top} \cdot f(A,x)_{j_1} \cdot \diag(x) \cdot e_{i_1} \cdot e_{j_1}^\top \cdot f(A,x) \cdot(\langle -f(A,x), f(A,x) \rangle + f(A,x)_{j_1})
\end{align*}
Finally, combine them and we have
\begin{align*}
       B_{5,3}^{j_1,i_1,j_1,i_0} = B_{5,3,1}^{j_1,i_1,j_1,i_0} 
\end{align*}
{\bf Proof of Part 4.} 
    \begin{align*}
    B_{5,4,1}^{j_1,i_1,j_1,i_0} : = & ~ - c_g(A,x)^{\top} \cdot f(A,x)_{j_1}^2 \cdot \diag(x) \cdot A^{\top}\cdot x_i \cdot  (e_{j_1}- f(A,x) ) \cdot (\langle -f(A,x), f(A,x) \rangle + f(A,x)_{j_1})\\
\end{align*}
Finally, combine them and we have
\begin{align*}
       B_{5,4}^{j_1,i_1,j_1,i_0} = B_{5,4,1}^{j_1,i_1,j_1,i_0}  
\end{align*}
{\bf Proof of Part 5.} 
    \begin{align*}
    B_{5,5,1}^{j_1,i_1,j_1,i_0} : = & ~ c_g(A,x)^{\top} \cdot f(A,x)_{j_1}^2 \cdot \diag(x) \cdot A^{\top} \cdot  f(A,x)  \cdot  x_{i_1} \cdot \langle - f(A,x), f(A,x) \rangle\\
    B_{5,5,2}^{j_1,i_1,j_1,i_0} : = & ~ c_g(A,x)^{\top} \cdot f(A,x)_{j_1}^3 \cdot \diag(x) \cdot A^{\top} \cdot  f(A,x)  \cdot  x_{i_1}  
\end{align*}
Finally, combine them and we have
\begin{align*}
       B_{5,5}^{j_1,i_1,j_1,i_0} = B_{5,5,1}^{j_1,i_1,j_1,i_0}  +B_{5,5,2}^{j_1,i_1,j_1,i_0}
\end{align*}
{\bf Proof of Part 6.} 
    \begin{align*}
     B_{5,6,1}^{j_1,i_1,j_1,i_0} : = & ~ c_g(A,x)^{\top} \cdot f(A,x)_{j_1}^2 \cdot \diag(x) \cdot A^{\top} \cdot  f(A,x)  \cdot  x_{i_1} \cdot \langle - f(A,x), f(A,x) \rangle\\
    B_{5,6,2}^{j_1,i_1,j_1,i_0} : = & ~ c_g(A,x)^{\top} \cdot f(A,x)_{j_1}^3 \cdot \diag(x) \cdot A^{\top} \cdot  f(A,x)  \cdot  x_{i_1}  
\end{align*}
Finally, combine them and we have
\begin{align*}
       B_{5,6}^{j_1,i_1,j_1,i_0} = B_{5,6,1}^{j_1,i_1,j_1,i_0}  +B_{5,6,2}^{j_1,i_1,j_1,i_0}
\end{align*}
{\bf Proof of Part 7.} 
    \begin{align*}
     B_{5,7,1}^{j_1,i_1,j_1,i_0} : = & ~  f(A,x)_{j_1}^3 \cdot  x_{i_1}\cdot c_g(A,x)^{\top} \cdot \diag(x) A^{\top} \cdot  f(A,x)   \\
    B_{5,7,2}^{j_1,i_1,j_1,i_0} : = & ~ -  f(A,x)_{j_1}^2 \cdot  x_{i_1} c_g(A,x)^{\top} \cdot \diag(x) \cdot A^{\top} \cdot  f(A,x)  
\end{align*}
Finally, combine them and we have
\begin{align*}
       B_{5,7}^{j_1,i_1,j_1,i_0} = B_{5,7,1}^{j_1,i_1,j_1,i_0}  +B_{5,7,2}^{j_1,i_1,j_1,i_0}
\end{align*}
\end{proof}
\end{lemma}
\subsection{Constructing \texorpdfstring{$d \times d$}{} matrices for \texorpdfstring{$j_1 = j_0$}{}}
The purpose of the following lemma is to let $i_0$ and $i_1$ disappear.
\begin{lemma}For $j_0,j_1 \in [n]$, a list of $d \times d$ matrices can be expressed as the following sense,
\begin{itemize}
\item {\bf Part 1.}
\begin{align*}
B_{5,1,1}^{j_1,*,j_1,*} & ~ =   f_c(A,x) \cdot  f(A,x)_{j_1}^2 \cdot  (-f_2(A,x) + f(A,x)_{j_1}) \cdot h(A,x) \cdot {\bf 1}_d^\top
\end{align*}
\item {\bf Part 2.}
\begin{align*}
B_{5,1,2}^{j_1,*,j_1,*} & ~ =    c(A,x)_{j_1}\cdot f(A,x)_{j_1}^2 \cdot  (-f_2(A,x) + f(A,x)_{j_1}) \cdot h(A,x) \cdot {\bf 1}_d^\top 
\end{align*}
\item {\bf Part 3.}
\begin{align*}
B_{5,1,3}^{j_1,*,j_1,*} & ~ =    f(A,x)_{j_1}^2 \cdot f_c(A,x) \cdot (-f_2(A,x) + f(A,x)_{j_1}) \cdot ( (A_{j_1,*}) \circ x^\top  )  \cdot h(A,x)\cdot I_d
\end{align*}
\item {\bf Part 4.}
\begin{align*}
B_{5,1,4}^{j_1,*,j_1,*}  & ~ =   -  f(A,x)_{j_1}^2 \cdot f_c(A,x)  \cdot(-f_2(A,x) + f(A,x)_{j_1}) \cdot h(A,x)^\top \cdot h(A,x) \cdot I_d
\end{align*}
\item {\bf Part 5.}
\begin{align*}
B_{5,1,5}^{j_1,*,j_1,*}  & ~ =   f(A,x)_{j_1}^2 \cdot (-f_2(A,x) + f(A,x)_{j_1})^2 \cdot h(A,x)^\top \cdot h(A,x) \cdot I_d
\end{align*}
\item {\bf Part 6.}
\begin{align*}
B_{5,1,6}^{j_1,*,j_1,*}  & ~ =     f(A,x)_{j_1}^2 \cdot (-f_2(A,x) + f(A,x)_{j_1})\cdot(-f_c(A,x) + f(A,x)_{j_1}) \cdot h(A,x)^\top \cdot h(A,x) \cdot I_d
\end{align*}
\item {\bf Part 7.}
\begin{align*}
B_{5,1,7}^{j_1,*,j_1,*}  & ~ =   f(A,x)_{j_1}^2 \cdot (-f_2(A,x) + f(A,x)_{j_1})\cdot p_{j_1}(A,x)^\top \cdot h(A,x) \cdot I_d
\end{align*}
\item {\bf Part 8.}
\begin{align*}
B_{5,2,1}^{j_1,*,j_1,*}  & ~ =     f(A,x)_{j_1}^2 \cdot  (-f_2(A,x) + f(A,x)_{j_1}) \cdot c_g(A,x)^{\top} \cdot h(A,x) \cdot x \cdot {\bf 1}_d^{\top} 
\end{align*}
\item {\bf Part 9.}
\begin{align*}
B_{5,2,2}^{j_1,*,j_1,*}  & ~ =    -f(A,x)_{j_1}  \cdot  (-f_2(A,x) + f(A,x)_{j_1}) \cdot c_g(A,x)^{\top} \cdot h(A,x) \cdot x \cdot {\bf 1}_d^{\top} 
\end{align*}
\item {\bf Part 10.}
\begin{align*}
 B_{5,3,1}^{j_1,*,j_1,*}  & ~ =      - f(A,x)_{j_1}^2 \cdot (-f_2(A,x) + f(A,x)_{j_1})\cdot {\bf 1}_d \cdot c_g(A,x)^{\top}   \cdot \diag(x)  
\end{align*}
\item {\bf Part 11.}
\begin{align*}
B_{5,4,1}^{j_1,*,j_1,*}  =     - f(A,x)_{j_1}^2 \cdot (-f_2(A,x) + f(A,x)_{j_1}) \cdot c_g(A,x)^{\top}  \cdot  h_e(A,x)\cdot x \cdot {\bf 1}_d^{\top}
\end{align*}
\item {\bf Part 12.}
\begin{align*}
B_{5,5,1}^{j_1,*,j_1,*}  & ~ =    -  f(A,x)_{j_1}^2 \cdot f_2(A,x) \cdot c_g(A,x)^{\top} \cdot h(A,x)  \cdot x \cdot {\bf 1}_d^{\top}
\end{align*}
\item {\bf Part 13.}
\begin{align*}
 B_{5,5,2}^{j_1,*,j_1,*}  =   f(A,x)_{j_1}^3  \cdot c_g(A,x)^{\top} \cdot h(A,x)  \cdot x \cdot {\bf 1}_d^{\top}
\end{align*}
\item {\bf Part 14.}
\begin{align*}
B_{5,6,1}^{j_1,*,j_1,*}  =   -  f(A,x)_{j_1}^2 \cdot f_2(A,x) \cdot c_g(A,x)^{\top} \cdot h(A,x)  \cdot x \cdot {\bf 1}_d^{\top}
\end{align*}
\item {\bf Part 15.}
\begin{align*}
B_{5,6,2}^{j_1,*,j_1,*}  =    f(A,x)_{j_1}^3  \cdot c_g(A,x)^{\top} \cdot h(A,x)  \cdot x \cdot {\bf 1}_d^{\top}
\end{align*}
\item {\bf Part 16.}
\begin{align*}
B_{5,7,1}^{j_1,*,j_1,*}  =   f(A,x)_{j_1}^3  \cdot c_g(A,x)^{\top} \cdot h(A,x)  \cdot x \cdot {\bf 1}_d^{\top}
\end{align*}
\item {\bf Part 17.}
\begin{align*}
B_{5,7,2}^{j_1,*,j_1,*}  =    f(A,x)_{j_1}^2  \cdot c_g(A,x)^{\top} \cdot h(A,x)  \cdot x \cdot {\bf 1}_d^{\top}
\end{align*}

\end{itemize}
\begin{proof}
{\bf Proof of Part 1.}
    We have
    \begin{align*}
        B_{5,1,1}^{j_1,i_1,j_1,i_1}  = & ~e_{i_1}^\top \cdot \langle c(A,x), f(A,x) \rangle \cdot  f(A,x)_{j_1}^2 \cdot \diag(x) \cdot A^{\top} \cdot f(A,x) \cdot(\langle -f(A,x), f(A,x) \rangle + f(A,x)_{j_1})\\
        B_{5,1,1}^{j_1,i_1,j_1,i_0}  = & ~ e_{i_1}^\top \cdot \langle c(A,x), f(A,x) \rangle \cdot  f(A,x)_{j_1}^2 \cdot \diag(x) \cdot A^{\top} \cdot f(A,x) \cdot(\langle -f(A,x), f(A,x) \rangle + f(A,x)_{j_1})
    \end{align*}
    From the above two equations, we can tell that $B_{5,1,1}^{j_1,*,j_1,*} \in \R^{d \times d}$ is a matrix that both the diagonal and off-diagonal have entries.
    
    Then we have $B_{5,1,1}^{j_1,*,j_1,*} \in \R^{d \times d}$ can be written as the rescaling of a diagonal matrix,
    \begin{align*}
     B_{5,1,1}^{j_1,*,j_1,*} & ~ = \langle c(A,x), f(A,x) \rangle \cdot  f(A,x)_{j_1}^2 \cdot  (\langle -f(A,x), f(A,x) \rangle + f(A,x)_{j_1}) \cdot \diag(x) \cdot A^{\top} \cdot f(A,x) \cdot {\bf 1}_d^\top \\
     & ~ = f_c(A,x) \cdot  f(A,x)_{j_1}^2 \cdot  (-f_2(A,x) + f(A,x)_{j_1}) \cdot h(A,x) \cdot {\bf 1}_d^\top
\end{align*}
    where the last step is follows from the Definitions~\ref{def:h}, Definitions~\ref{def:f_2} and Definitions~\ref{def:f_c}. 

{\bf Proof of Part 2.}
    We have
    \begin{align*}
           B_{5,1,2}^{j_1,i_1,j_1,i_1} = & ~ e_{i_1}^\top \cdot c(A,x)_{j_1}\cdot f(A,x)_{j_1}^2 \cdot \diag(x) \cdot A^{\top} \cdot f(A,x) \cdot(\langle -f(A,x), f(A,x) \rangle + f(A,x)_{j_1})\\
        B_{5,1,2}^{j_1,i_1,j_1,i_0} = & ~ e_{i_1}^\top \cdot c(A,x)_{j_1}\cdot f(A,x)_{j_1}^2 \cdot \diag(x) \cdot A^{\top} \cdot f(A,x) \cdot(\langle -f(A,x), f(A,x) \rangle + f(A,x)_{j_1})
    \end{align*}
     From the above two equations, we can tell that $B_{5,1,2}^{j_1,*,j_1,*} \in \R^{d \times d}$ is a matrix that only diagonal has entries and off-diagonal are all zeros.
    
    Then we have $B_{5,1,2}^{j_1,*,j_1,*} \in \R^{d \times d}$ can be written as the rescaling of a diagonal matrix,
\begin{align*}
     B_{5,1,2}^{j_1,*,j_1,*} & ~ = c(A,x)_{j_1}\cdot f(A,x)_{j_1}^2 \cdot  (\langle -f(A,x), f(A,x) \rangle + f(A,x)_{j_1}) \cdot \diag(x) \cdot A^{\top} \cdot f(A,x) \cdot {\bf 1}_d^\top \\
     & ~ =  c(A,x)_{j_1}\cdot f(A,x)_{j_1}^2 \cdot  (-f_2(A,x) + f(A,x)_{j_1}) \cdot h(A,x) \cdot {\bf 1}_d^\top 
\end{align*}
    where the last step is follows from the Definitions~\ref{def:h} and Definitions~\ref{def:f_2}.

{\bf Proof of Part 3.}
We have for diagonal entry and off-diagonal entry can be written as follows 
    \begin{align*}
        B_{5,1,3}^{j_1,i_1,j_1,i_1} = & ~f(A,x)_{j_1}^2 \cdot \langle c(A,x), f(A,x) \rangle \cdot ( (A_{j_1,*}) \circ x^\top  )  \cdot \diag(x) \cdot A^{\top}\\& ~ \cdot f(A,x) \cdot (\langle -f(A,x), f(A,x) \rangle + f(A,x)_{j_1}) \\
        B_{5,1,3}^{j_1,i_1,j_1,i_0} = & ~f(A,x)_{j_1}^2 \cdot \langle c(A,x), f(A,x) \rangle \cdot ( (A_{j_1,*}) \circ x^\top  )  \cdot \diag(x) \cdot A^{\top}\\& ~ \cdot f(A,x) \cdot (\langle -f(A,x), f(A,x) \rangle + f(A,x)_{j_1})
    \end{align*}
From the above equation, we can show that matrix $B_{5,1,3}^{j_1,*,j_1,*}$ can be expressed as a rank-$1$ matrix,
\begin{align*}
     B_{5,1,3}^{j_1,*,j_1,*} & ~ = f(A,x)_{j_1}^2 \cdot \langle c(A,x), f(A,x) \rangle \cdot (\langle -f(A,x), f(A,x) \rangle + f(A,x)_{j_1}) \cdot ( (A_{j_1,*}) \circ x^\top  )\\
     & ~ \cdot \diag(x) \cdot A^{\top} \cdot f(A,x) \cdot I_d\\
     & ~ =  f(A,x)_{j_1}^2 \cdot f_c(A,x) \cdot (-f_2(A,x) + f(A,x)_{j_1}) \cdot ( (A_{j_1,*}) \circ x^\top  )  \cdot h(A,x)\cdot I_d
\end{align*}
    where the last step is follows from the Definitions~\ref{def:h}, Definitions~\ref{def:f_2} and Definitions~\ref{def:f_c}.

{\bf Proof of Part 4.}
We have for diagonal entry and off-diagonal entry can be written as follows
    \begin{align*}
        B_{5,1,4}^{j_1,i_1,j_1,i_1}   = & ~  -  f(A,x)_{j_1}^2 \cdot f(A,x)^\top  \cdot A \cdot (\diag(x))^2 \cdot   \langle c(A,x), f(A,x) \rangle \cdot A^{\top} \cdot f(A,x) \\
        & ~ \cdot(\langle -f(A,x), f(A,x) \rangle + f(A,x)_{j_1}) \\
        B_{5,1,4}^{j_1,i_1,j_1,i_0}   = & ~ -    f(A,x)_{j_1}^2 \cdot f(A,x)^\top  \cdot A \cdot (\diag(x))^2 \cdot   \langle c(A,x), f(A,x) \rangle \cdot A^{\top} \cdot f(A,x) \\
        & ~ \cdot(\langle -f(A,x), f(A,x) \rangle + f(A,x)_{j_1})
    \end{align*}
 From the above equation, we can show that matrix $B_{5,1,4}^{j_1,*,j_1,*}$ can be expressed as a rank-$1$ matrix,
\begin{align*}
    B_{5,1,4}^{j_1,*,j_1,*}  & ~ = -  f(A,x)_{j_1}^2 \cdot \langle c(A,x), f(A,x) \rangle  \cdot(\langle -f(A,x), f(A,x) \rangle + f(A,x)_{j_1}) \cdot f(A,x)^\top  \\& ~\cdot A \cdot (\diag(x))^2 \cdot  A^{\top} \cdot f(A,x) \cdot I_d\\
     & ~ =   -  f(A,x)_{j_1}^2 \cdot f_c(A,x)  \cdot(-f_2(A,x) + f(A,x)_{j_1}) \cdot h(A,x)^\top \cdot h(A,x) \cdot I_d
\end{align*}
   where the last step is follows from the Definitions~\ref{def:h}, Definitions~\ref{def:f_2} and Definitions~\ref{def:f_c}.

{\bf Proof of Part 5.}
We have for diagonal entry and off-diagonal entry can be written as follows
    \begin{align*}
         B_{5,1,5}^{j_1,i_1,j_1,i_0} = & ~    f(A,x)_{j_1}^2 \cdot f(A,x)^\top  \cdot A \cdot (\diag(x))^2 \cdot (\langle -f(A,x), f(A,x) \rangle + f(A,x)_{j_1})^2  \cdot A^{\top} \cdot f(A,x)  \\
         B_{5,1,5}^{j_1,i_1,j_1,i_0} = & ~    f(A,x)_{j_1}^2 \cdot f(A,x)^\top  \cdot A \cdot (\diag(x))^2 \cdot (\langle -f(A,x), f(A,x) \rangle + f(A,x)_{j_1})^2  \cdot A^{\top} \cdot f(A,x) 
    \end{align*}
    From the above equation, we can show that matrix $B_{5,1,5}^{j_1,*,j_1,*}$ can be expressed as a rank-$1$ matrix,
\begin{align*}
    B_{5,1,5}^{j_1,*,j_1,*}  & ~ =  f(A,x)_{j_1}^2 \cdot (\langle -f(A,x), f(A,x) \rangle + f(A,x)_{j_1})^2 \cdot f(A,x)^\top  \cdot A \cdot (\diag(x))^2 \cdot  A^{\top} \cdot f(A,x) \cdot I_d\\
     & ~ =    f(A,x)_{j_1}^2 \cdot (-f_2(A,x) + f(A,x)_{j_1})^2 \cdot h(A,x)^\top \cdot h(A,x) \cdot I_d
\end{align*}
    where the last step is follows from the Definitions~\ref{def:h} and Definitions~\ref{def:f_2}.

{\bf Proof of Part 6.}
We have for diagonal entry and off-diagonal entry can be written as follows
    \begin{align*}
        B_{5,1,6}^{j_1,i_1,j_1,i_1}  = & ~   f(A,x)_{j_1}^2 \cdot f(A,x)^\top  \cdot A \cdot  (\diag(x))^2 \cdot(\langle -f(A,x), c(A,x) \rangle + f(A,x)_{j_1}) \cdot A^{\top} \cdot f(A,x) \\
        & ~ \cdot(\langle -f(A,x), f(A,x) \rangle + f(A,x)_{j_1})\\
        B_{5,1,6}^{j_1,i_1,j_1,i_0}  = & ~   f(A,x)_{j_1}^2 \cdot f(A,x)^\top  \cdot A \cdot  (\diag(x))^2 \cdot(\langle -f(A,x), c(A,x) \rangle + f(A,x)_{j_1}) \cdot A^{\top} \cdot f(A,x) \\
        & ~ \cdot(\langle -f(A,x), f(A,x) \rangle + f(A,x)_{j_1})
    \end{align*}
    From the above equation, we can show that matrix $B_{5,1,6}^{j_1,*,j_1,*}$ can be expressed as a rank-$1$ matrix,
\begin{align*}
    B_{5,1,6}^{j_1,*,j_1,*}  & ~ =    f(A,x)_{j_1}^2 \cdot (\langle -f(A,x), f(A,x) \rangle + f(A,x)_{j_1})\cdot(\langle -f(A,x), c(A,x) \rangle + f(A,x)_{j_1}) \\
    & ~\cdot f(A,x)^\top  \cdot A \cdot (\diag(x))^2 \cdot  A^{\top} \cdot f(A,x) \cdot I_d\\
     & ~ =   f(A,x)_{j_1}^2 \cdot (-f_2(A,x) + f(A,x)_{j_1})\cdot(-f_c(A,x) + f(A,x)_{j_1}) \cdot h(A,x)^\top \cdot h(A,x) \cdot I_d
\end{align*}
    where the last step is follows from the Definitions~\ref{def:h}, Definitions~\ref{def:f_2} and Definitions~\ref{def:f_c}.
    
{\bf Proof of Part 7.}
We have for diagonal entry and off-diagonal entry can be written as follows
    \begin{align*}
         B_{5,1,7}^{j_1,i_1,j_1,i_1} = & ~  f(A,x)_{j_1}^2 \cdot ((e_{j_1}^\top - f(A,x)^\top) \circ q(A,x)^\top) \cdot A \cdot  (\diag(x))^2 \cdot A^{\top} \cdot f(A,x) \\
         & ~ \cdot(\langle -f(A,x), f(A,x) \rangle + f(A,x)_{j_1})\\
         B_{5,1,7}^{j_1,i_1,j_1,i_0} = & ~  f(A,x)_{j_1}^2 \cdot ((e_{j_1}^\top - f(A,x)^\top) \circ q(A,x)^\top) \cdot A \cdot  (\diag(x))^2 \cdot A^{\top} \cdot f(A,x) \\
         & ~ \cdot(\langle -f(A,x), f(A,x) \rangle + f(A,x)_{j_1})
    \end{align*}
    From the above equation, we can show that matrix $B_{5,1,7}^{j_1,*,j_1,*}$ can be expressed as a rank-$1$ matrix,
\begin{align*}
     B_{5,1,7}^{j_1,*,j_1,*}  & ~ =   f(A,x)_{j_1}^2 \cdot (\langle -f(A,x), f(A,x) \rangle + f(A,x)_{j_1})\cdot ((e_{j_1}^\top - f(A,x)^\top) \circ q(A,x)^\top) \cdot A \cdot  (\diag(x))^2 \\
     &~ \cdot A^{\top} \cdot f(A,x)  \cdot I_d\\
     & ~ = f(A,x)_{j_1}^2 \cdot (-f_2(A,x) + f(A,x)_{j_1})\cdot p_{j_1}(A,x)^\top \cdot h(A,x) \cdot I_d
\end{align*}
    where the last step is follows from the Definitions~\ref{def:h}, Definitions~\ref{def:f_2} and Definitions~\ref{def:p}.

    {\bf Proof of Part 8.}
We have for diagonal entry and off-diagonal entry can be written as follows
    \begin{align*}
         B_{5,2,1}^{j_1,i_1,j_1,i_1} = & ~   f(A,x)_{j_1}^2 \cdot x_{i_1} \cdot c_g(A,x)^{\top} \cdot \diag(x) \cdot A^{\top} \cdot f(A,x) \cdot(\langle -f(A,x), f(A,x) \rangle + f(A,x)_{j_1})\\
         B_{5,2,1}^{j_1,i_1,j_1,i_0} = & ~  f(A,x)_{j_1}^2 \cdot x_{i_1} \cdot c_g(A,x)^{\top} \cdot \diag(x) \cdot A^{\top} \cdot f(A,x) \cdot(\langle -f(A,x), f(A,x) \rangle + f(A,x)_{j_1})
    \end{align*}
    From the above equation, we can show that matrix $B_{5,2,1}^{j_1,*,j_1,*}$ can be expressed as a rank-$1$ matrix,
\begin{align*}
     B_{5,2,1}^{j_1,*,j_1,*}  & ~ =   f(A,x)_{j_1}^2 \cdot  (\langle -f(A,x), f(A,x) \rangle + f(A,x)_{j_1}) \cdot c_g(A,x)^{\top} \cdot \diag(x) \cdot A^{\top} \cdot f(A,x) \cdot x \cdot {\bf 1}_d^{\top}  \\ 
     & ~ =  f(A,x)_{j_1}^2 \cdot  (-f_2(A,x) + f(A,x)_{j_1}) \cdot c_g(A,x)^{\top} \cdot h(A,x) \cdot x \cdot {\bf 1}_d^{\top} 
\end{align*}
    where the last step is follows from the Definitions~\ref{def:h}, Definitions~\ref{def:f_2}.
    {\bf Proof of Part 9.}
We have for diagonal entry and off-diagonal entry can be written as follows
    \begin{align*}
         B_{5,2,2}^{j_1,i_1,j_1,i_1} = & ~   - f(A,x)_{j_1} \cdot x_{i_1} \cdot c_g(A,x)^{\top} \cdot \diag(x) \cdot A^{\top} \cdot f(A,x) \cdot(\langle -f(A,x), f(A,x) \rangle + f(A,x)_{j_1})\\
         B_{5,2,2}^{j_1,i_1,j_1,i_0} = & ~ - f(A,x)_{j_1} \cdot x_{i_1} \cdot c_g(A,x)^{\top} \cdot \diag(x) \cdot A^{\top} \cdot f(A,x) \cdot(\langle -f(A,x), f(A,x) \rangle + f(A,x)_{j_1})
    \end{align*}
    From the above equation, we can show that matrix $B_{5,2,2}^{j_1,*,j_1,*}$ can be expressed as a rank-$1$ matrix,
\begin{align*}
     B_{5,2,2}^{j_1,*,j_1,*}  & ~ =  - f(A,x)_{j_1}  \cdot  (\langle -f(A,x), f(A,x) \rangle + f(A,x)_{j_1}) \cdot c_g(A,x)^{\top} \cdot \diag(x) \cdot A^{\top} \cdot f(A,x) \cdot x \cdot {\bf 1}_d^{\top}  \\ 
     & ~ =  -f(A,x)_{j_1}  \cdot  (-f_2(A,x) + f(A,x)_{j_1}) \cdot c_g(A,x)^{\top} \cdot h(A,x) \cdot x \cdot {\bf 1}_d^{\top} 
\end{align*}
    where the last step is follows from the Definitions~\ref{def:h}, Definitions~\ref{def:f_2}.

   {\bf Proof of Part 10.}
We have for diagonal entry and off-diagonal entry can be written as follows
    \begin{align*}
         B_{5,3,1}^{j_1,i_1,j_1,i_1} = & ~   -  c_g(A,x)^{\top} \cdot f(A,x)_{j_1} \cdot \diag(x) \cdot e_{i_1} \cdot e_{j_1}^\top \cdot f(A,x) \cdot(\langle -f(A,x), f(A,x) \rangle + f(A,x)_{j_1})\\
         B_{5,3,1}^{j_1,i_1,j_1,i_0} = & ~ -  c_g(A,x)^{\top} \cdot f(A,x)_{j_1} \cdot \diag(x) \cdot e_{i_1} \cdot e_{j_1}^\top \cdot f(A,x) \cdot(\langle -f(A,x), f(A,x) \rangle + f(A,x)_{j_1})
    \end{align*}
    From the above equation, we can show that matrix $B_{5,3,1}^{j_1,*,j_1,*}$ can be expressed as a rank-$1$ matrix,
\begin{align*}
     B_{5,3,1}^{j_1,*,j_1,*}  & ~ = - f(A,x)_{j_1}^2 \cdot (\langle -f(A,x), f(A,x) \rangle + f(A,x)_{j_1})\cdot {\bf 1}_d \cdot c_g(A,x)^{\top}   \cdot \diag(x)  \\ 
     & ~ =   - f(A,x)_{j_1}^2 \cdot (-f_2(A,x) + f(A,x)_{j_1})\cdot {\bf 1}_d \cdot c_g(A,x)^{\top}   \cdot \diag(x)  
\end{align*}
    where the last step is follows from the Definitions~\ref{def:f_2}.

    {\bf Proof of Part 11.}
We have for diagonal entry and off-diagonal entry can be written as follows
    \begin{align*}
         B_{5,4,1}^{j_1,i_1,j_1,i_1} = & ~   - c_g(A,x)^{\top} \cdot f(A,x)_{j_1}^2 \cdot \diag(x) \cdot A^{\top}\cdot x_i \cdot  (e_{j_1}- f(A,x) ) \cdot (\langle -f(A,x), f(A,x) \rangle + f(A,x)_{j_1})\\
         B_{5,4,1}^{j_1,i_1,j_1,i_0} = & ~  - c_g(A,x)^{\top} \cdot f(A,x)_{j_1}^2 \cdot \diag(x) \cdot A^{\top}\cdot x_i \cdot  (e_{j_1}- f(A,x) ) \cdot (\langle -f(A,x), f(A,x) \rangle + f(A,x)_{j_1})
    \end{align*}
    From the above equation, we can show that matrix $B_{5,4,1}^{j_1,*,j_1,*}$ can be expressed as a rank-$1$ matrix,
\begin{align*}
     B_{5,4,1}^{j_1,*,j_1,*}  & ~ = - f(A,x)_{j_1}^2 \cdot (\langle -f(A,x), f(A,x) \rangle + f(A,x)_{j_1}) \cdot c_g(A,x)^{\top}  \cdot \diag(x) \cdot A^{\top}  \cdot  (e_{j_1}- f(A,x) ) \cdot x \cdot {\bf 1}_d^{\top}\\ 
     & ~ =   - f(A,x)_{j_1}^2 \cdot (-f_2(A,x) + f(A,x)_{j_1}) \cdot c_g(A,x)^{\top}  \cdot  h_e(A,x)\cdot x \cdot {\bf 1}_d^{\top}
\end{align*}
    where the last step is follows from the Definitions~\ref{def:f_2} and Definitions~\ref{def:h_e}.

        {\bf Proof of Part 12.}
We have for diagonal entry and off-diagonal entry can be written as follows
    \begin{align*}
         B_{5,5,1}^{j_1,i_1,j_1,i_1} = & ~  c_g(A,x)^{\top} \cdot f(A,x)_{j_1}^2 \cdot \diag(x) \cdot A^{\top} \cdot  f(A,x)  \cdot  x_{i_1} \cdot \langle - f(A,x), f(A,x) \rangle\\
         B_{5,5,1}^{j_1,i_1,j_1,i_0} = & ~ c_g(A,x)^{\top} \cdot f(A,x)_{j_1}^2 \cdot \diag(x) \cdot A^{\top} \cdot  f(A,x)  \cdot  x_{i_1} \cdot \langle - f(A,x), f(A,x) \rangle
    \end{align*}
    From the above equation, we can show that matrix $B_{5,5,1}^{j_1,*,j_1,*}$ can be expressed as a rank-$1$ matrix,
\begin{align*}
     B_{5,5,1}^{j_1,*,j_1,*}  & ~ = f(A,x)_{j_1}^2 \cdot \langle - f(A,x), f(A,x) \rangle \cdot c_g(A,x)^{\top} \cdot \diag(x) \cdot A^{\top} \cdot  f(A,x)  \cdot x \cdot {\bf 1}_d^{\top}\\ 
     & ~ =  -  f(A,x)_{j_1}^2 \cdot f_2(A,x) \cdot c_g(A,x)^{\top} \cdot h(A,x)  \cdot x \cdot {\bf 1}_d^{\top}
\end{align*}
    where the last step is follows from the Definitions~\ref{def:h} and Definitions~\ref{def:f_2}.

    {\bf Proof of Part 13.}
We have for diagonal entry and off-diagonal entry can be written as follows
    \begin{align*}
         B_{5,5,2}^{j_1,i_1,j_1,i_1} = & ~  c_g(A,x)^{\top} \cdot f(A,x)_{j_1}^3 \cdot \diag(x) \cdot A^{\top} \cdot  f(A,x)  \cdot  x_{i_1}  \\
         B_{5,5,2}^{j_1,i_1,j_1,i_0} = & ~ c_g(A,x)^{\top} \cdot f(A,x)_{j_1}^3 \cdot \diag(x) \cdot A^{\top} \cdot  f(A,x)  \cdot  x_{i_1}  
    \end{align*}
    From the above equation, we can show that matrix $B_{5,5,2}^{j_1,*,j_1,*}$ can be expressed as a rank-$1$ matrix,
\begin{align*}
     B_{5,5,2}^{j_1,*,j_1,*}  & ~ = f(A,x)_{j_1}^3   \cdot c_g(A,x)^{\top} \cdot \diag(x) \cdot A^{\top} \cdot  f(A,x)  \cdot x \cdot {\bf 1}_d^{\top}\\ 
     & ~ =  f(A,x)_{j_1}^3  \cdot c_g(A,x)^{\top} \cdot h(A,x)  \cdot x \cdot {\bf 1}_d^{\top}
\end{align*}
    where the last step is follows from the Definitions~\ref{def:h}.

        {\bf Proof of Part 14.}
We have for diagonal entry and off-diagonal entry can be written as follows
    \begin{align*}
         B_{5,6,1}^{j_1,i_1,j_1,i_1} = & ~  c_g(A,x)^{\top} \cdot f(A,x)_{j_1}^2 \cdot \diag(x) \cdot A^{\top} \cdot  f(A,x)  \cdot  x_{i_1} \cdot \langle - f(A,x), f(A,x) \rangle\\
         B_{5,6,1}^{j_1,i_1,j_1,i_0} = & ~ c_g(A,x)^{\top} \cdot f(A,x)_{j_1}^2 \cdot \diag(x) \cdot A^{\top} \cdot  f(A,x)  \cdot  x_{i_1} \cdot \langle - f(A,x), f(A,x) \rangle
    \end{align*}
    From the above equation, we can show that matrix $B_{5,6,1}^{j_1,*,j_1,*}$ can be expressed as a rank-$1$ matrix,
\begin{align*}
     B_{5,6,1}^{j_1,*,j_1,*}  & ~ = f(A,x)_{j_1}^2 \cdot \langle - f(A,x), f(A,x) \rangle \cdot c_g(A,x)^{\top} \cdot \diag(x) \cdot A^{\top} \cdot  f(A,x)  \cdot x \cdot {\bf 1}_d^{\top}\\ 
     & ~ =  -  f(A,x)_{j_1}^2 \cdot f_2(A,x) \cdot c_g(A,x)^{\top} \cdot h(A,x)  \cdot x \cdot {\bf 1}_d^{\top}
\end{align*}
    where the last step is follows from the Definitions~\ref{def:h} and Definitions~\ref{def:f_2}.

    {\bf Proof of Part 15.}
We have for diagonal entry and off-diagonal entry can be written as follows
    \begin{align*}
         B_{5,6,2}^{j_1,i_1,j_1,i_1} = & ~  c_g(A,x)^{\top} \cdot f(A,x)_{j_1}^3 \cdot \diag(x) \cdot A^{\top} \cdot  f(A,x)  \cdot  x_{i_1}  \\
         B_{5,6,2}^{j_1,i_1,j_1,i_0} = & ~ c_g(A,x)^{\top} \cdot f(A,x)_{j_1}^3 \cdot \diag(x) \cdot A^{\top} \cdot  f(A,x)  \cdot  x_{i_1}  
    \end{align*}
    From the above equation, we can show that matrix $B_{5,6,2}^{j_1,*,j_1,*}$ can be expressed as a rank-$1$ matrix,
\begin{align*}
     B_{5,6,2}^{j_1,*,j_1,*}  & ~ = f(A,x)_{j_1}^3   \cdot c_g(A,x)^{\top} \cdot \diag(x) \cdot A^{\top} \cdot  f(A,x)  \cdot x \cdot {\bf 1}_d^{\top}\\ 
     & ~ =  f(A,x)_{j_1}^3  \cdot c_g(A,x)^{\top} \cdot h(A,x)  \cdot x \cdot {\bf 1}_d^{\top}
\end{align*}
    where the last step is follows from the Definitions~\ref{def:h}.

    {\bf Proof of Part 16.}
We have for diagonal entry and off-diagonal entry can be written as follows
    \begin{align*}
         B_{5,7,1}^{j_1,i_1,j_1,i_1} = & ~  f(A,x)_{j_1}^3 \cdot  x_{i_1}\cdot c_g(A,x)^{\top} \cdot \diag(x) A^{\top} \cdot  f(A,x)  \\
         B_{5,7,1}^{j_1,i_1,j_1,i_0} = & ~ f(A,x)_{j_1}^3 \cdot  x_{i_1}\cdot c_g(A,x)^{\top} \cdot \diag(x) A^{\top} \cdot  f(A,x) 
    \end{align*}
    From the above equation, we can show that matrix $B_{5,7,1}^{j_1,*,j_1,*}$ can be expressed as a rank-$1$ matrix,
\begin{align*}
     B_{5,7,1}^{j_1,*,j_1,*}  & ~ = f(A,x)_{j_1}^3   \cdot c_g(A,x)^{\top} \cdot \diag(x) \cdot A^{\top} \cdot  f(A,x)  \cdot x \cdot {\bf 1}_d^{\top}\\ 
     & ~ =  f(A,x)_{j_1}^3  \cdot c_g(A,x)^{\top} \cdot h(A,x)  \cdot x \cdot {\bf 1}_d^{\top}
\end{align*}
    where the last step is follows from the Definitions~\ref{def:h}.

       {\bf Proof of Part 17.}
We have for diagonal entry and off-diagonal entry can be written as follows
    \begin{align*}
         B_{5,7,2}^{j_1,i_1,j_1,i_1} = & ~  -  f(A,x)_{j_1}^2 \cdot  x_{i_1} c_g(A,x)^{\top} \cdot \diag(x) \cdot A^{\top} \cdot  f(A,x)   \\
         B_{5,7,2}^{j_1,i_1,j_1,i_0} = & ~ -  f(A,x)_{j_1}^2 \cdot  x_{i_1} c_g(A,x)^{\top} \cdot \diag(x) \cdot A^{\top} \cdot  f(A,x) 
    \end{align*}
    From the above equation, we can show that matrix $B_{5,7,2}^{j_1,*,j_1,*}$ can be expressed as a rank-$1$ matrix,
\begin{align*}
     B_{5,7,2}^{j_1,*,j_1,*}  & ~ = f(A,x)_{j_1}^2   \cdot c_g(A,x)^{\top} \cdot \diag(x) \cdot A^{\top} \cdot  f(A,x)  \cdot x \cdot {\bf 1}_d^{\top}\\ 
     & ~ =  f(A,x)_{j_1}^2  \cdot c_g(A,x)^{\top} \cdot h(A,x)  \cdot x \cdot {\bf 1}_d^{\top}
\end{align*}
    where the last step is follows from the Definitions~\ref{def:h}.
\end{proof}
\end{lemma}

\subsection{Case \texorpdfstring{$j_1 \neq j_0, i_1 = i_0$}{}}
\begin{lemma}
For $j_1 \neq j_0$ and $i_0 = i_1$. If the following conditions hold
    \begin{itemize}
     \item Let $u(A,x) \in \R^n$ be defined as Definition~\ref{def:u}
    \item Let $\alpha(A,x) \in \R$ be defined as Definition~\ref{def:alpha}
     \item Let $f(A,x) \in \R^n$ be defined as Definition~\ref{def:f}
    \item Let $c(A,x) \in \R^n$ be defined as Definition~\ref{def:c}
    \item Let $g(A,x) \in \R^d$ be defined as Definition~\ref{def:g} 
    \item Let $q(A,x) = c(A,x) + f(A,x) \in \R^n$
    \item Let $c_g(A,x) \in \R^d$ be defined as Definition~\ref{def:c_g}.
    \item Let $L_g(A,x) \in \R$ be defined as Definition~\ref{def:l_g}
    \item Let $v \in \R^n$ be a vector 
    \item Let $B_1^{j_1,i_1,j_0,i_0}$ be defined as Definition~\ref{def:b_1}
    \end{itemize}
    Then, For $j_0,j_1 \in [n], i_0,i_1 \in [d]$, we have 
    \begin{itemize}
\item {\bf Part 1.} For $B_{5,1}^{j_1,i_1,j_0,i_1}$, we have 
\begin{align*}
 B_{5,1}^{j_1,i_1,j_0,i_1}  = & ~  \frac{\d}{\d A_{j_1,i_1}} (- c_g(A,x)^{\top} ) \cdot  f(A,x)_{j_0} \cdot \diag (x) A^{\top} \cdot  f(A,x)  \cdot  (\langle -f(A,x), f(A,x) \rangle + f(A,x)_{j_0})\\
 = & ~ B_{5,1,1}^{j_1,i_1,j_0,i_1} + B_{5,1,2}^{j_1,i_1,j_0,i_1} + B_{5,1,3}^{j_1,i_1,j_0,i_1} + B_{5,1,4}^{j_1,i_1,j_0,i_1} + B_{5,1,5}^{j_1,i_1,j_0,i_1} + B_{5,1,6}^{j_1,i_1,j_0,i_1} + B_{5,1,7}^{j_1,i_1,j_0,i_1}
\end{align*} 
\item {\bf Part 2.} For $B_{5,2}^{j_1,i_1,j_0,i_1}$, we have 
\begin{align*}
  & ~ B_{5,2}^{j_1,i_1,j_0,i_1} \\= & ~ - c_g(A,x)^{\top} \cdot \frac{\d}{\d A_{j_1,i_1}} ( f(A,x)_{j_0} )  \cdot \diag(x) \cdot A^{\top} \cdot f(A,x) \cdot(\langle -f(A,x), f(A,x) \rangle + f(A,x)_{j_0}) \\
    = & ~  B_{5,2,1}^{j_1,i_1,j_0,i_1}  
\end{align*} 
\item {\bf Part 3.} For $B_{5,3}^{j_1,i_1,j_0,i_1}$, we have 
\begin{align*}
   & ~ B_{5,3}^{j_1,i_1,j_0,i_1} \\ = & ~ - c_g(A,x)^{\top} \cdot f(A,x)_{j_0} \cdot \diag(x) \cdot \frac{\d}{\d A_{j_1,i_1}} (  A^{\top} ) \cdot f(A,x) \cdot(\langle -f(A,x), f(A,x) \rangle + f(A,x)_{j_0}) \\
     = & ~ B_{5,3,1}^{j_1,i_1,j_0,i_1}  
\end{align*} 
\item {\bf Part 4.} For $B_{5,4}^{j_1,i_1,j_0,i_1}$, we have 
\begin{align*}
  B_{5,4}^{j_1,i_1,j_0,i_1} = & ~ - c_g(A,x)^{\top} \cdot f(A,x)_{j_0} \cdot \diag(x) \cdot A^{\top} \cdot \frac{\d f(A,x)}{\d A_{j_1,i_1}} \cdot(\langle -f(A,x), f(A,x) \rangle + f(A,x)_{j_0}) \\
     = & ~ B_{5,4,1}^{j_1,i_1,j_0,i_1} 
\end{align*}
\item {\bf Part 5.} For $B_{5,5}^{j_1,i_1,j_0,i_1}$, we have 
\begin{align*}
  B_{5,5}^{j_1,i_1,j_0,i_1} = & ~  c_g(A,x)^{\top} \cdot f(A,x)_{j_0} \cdot \diag(x) \cdot A^{\top} \cdot  f(A,x)  \cdot \langle f(A,x), \frac{\d f(A,x) }{\d A_{j_1,i_1}}\rangle \\
     = & ~ B_{5,5,1}^{j_1,i_1,j_0,i_1}  + B_{5,5,2}^{j_1,i_1,j_0,i_1}  
\end{align*}
\item {\bf Part 6.} For $B_{5,6}^{j_1,i_1,j_0,i_1}$, we have 
\begin{align*}
  B_{5,6}^{j_1,i_1,j_0,i_1} = & ~ c_g(A,x)^{\top} \cdot f(A,x)_{j_0} \cdot \diag(x) \cdot A^{\top} \cdot  f(A,x)  \cdot \langle \frac{\d  f(A,x) }{\d A_{j_1,i_1}},f(A,x)\rangle \\
     = & ~ B_{5,6,1}^{j_1,i_1,j_0,i_1}  + B_{5,6,2}^{j_1,i_1,j_0,i_1}  
\end{align*}
\item {\bf Part 7.} For $B_{5,7}^{j_1,i_1,j_0,i_1}$, we have 
\begin{align*}
  B_{5,7}^{j_1,i_1,j_0,i_1} = & ~ - c_g(A,x)^{\top} \cdot f(A,x)_{j_1} \cdot \diag(x) \cdot A^{\top} \cdot  f(A,x)  \cdot  \frac{\d f(A,x)_{j_0}}{\d A_{j_1,i_1}} \\
     = & ~ B_{5,7,1}^{j_1,i_1,j_0,i_1}  
\end{align*}
\end{itemize}
\begin{proof}
    {\bf Proof of Part 1.}
    \begin{align*}
    B_{5,1,1}^{j_1,i_1,j_0,i_1} : = & ~ e_{i_1}^\top \cdot \langle c(A,x), f(A,x) \rangle \cdot  f(A,x)_{j_1} \cdot f(A,x)_{j_0} \cdot \diag(x) \cdot A^{\top}  \\
        & ~\cdot f(A,x) \cdot(\langle -f(A,x), f(A,x) \rangle + f(A,x)_{j_0})\\
    B_{5,1,2}^{j_1,i_1,j_0,i_1} : = & ~   e_{i_1}^\top \cdot c(A,x)_{j_1}\cdot f(A,x)_{j_1} \cdot f(A,x)_{j_0} \cdot \diag(x) \cdot A^{\top} \cdot f(A,x) \cdot(\langle -f(A,x), f(A,x) \rangle + f(A,x)_{j_0})\\
    B_{5,1,3}^{j_1,i_1,j_0,i_1} : = & ~ f(A,x)_{j_1} \cdot f(A,x)_{j_0} \cdot \langle c(A,x), f(A,x) \rangle \cdot ( (A_{j_1,*}) \circ x^\top  )  \cdot \diag(x) \cdot A^{\top} \cdot f(A,x) \cdot \\
    & ~(\langle -f(A,x), f(A,x) \rangle + f(A,x)_{j_0})\\
    B_{5,1,4}^{j_1,i_1,j_0,i_1} : = & ~   -  f(A,x)_{j_1} \cdot f(A,x)_{j_0} \cdot f(A,x)^\top  \cdot A \cdot (\diag(x))^2 \cdot   \langle c(A,x), f(A,x) \rangle \cdot A^{\top} \cdot f(A,x)\\
    & ~ \cdot(\langle -f(A,x), f(A,x) \rangle + f(A,x)_{j_0})\\
    B_{5,1,5}^{j_1,i_1,j_0,i_1} : = & ~    f(A,x)_{j_1} \cdot f(A,x)_{j_0} \cdot f(A,x)^\top  \cdot A \cdot (\diag(x))^2 \cdot (\langle (-f(A,x)), f(A,x) \rangle + f(A,x)_{j_1})   \cdot A^{\top} \cdot f(A,x) \\
    & ~ \cdot(\langle -f(A,x), f(A,x) \rangle + f(A,x)_{j_0})\\
    B_{5,1,6}^{j_1,i_1,j_0,i_1} : = & ~  f(A,x)_{j_1} \cdot f(A,x)_{j_0} \cdot f(A,x)^\top  \cdot A \cdot  (\diag(x))^2 \cdot(\langle -f(A,x), c(A,x) \rangle + f(A,x)_{j_1}) \cdot A^{\top} \cdot f(A,x) \\
        & ~ \cdot(\langle -f(A,x), f(A,x) \rangle + f(A,x)_{j_0})\\
    B_{5,1,7}^{j_1,i_1,j_0,i_1} : = & ~   f(A,x)_{j_1} \cdot f(A,x)_{j_0} \cdot ((e_{j_1}^\top - f(A,x)^\top) \circ q(A,x)^\top) \cdot A \cdot  (\diag(x))^2 \cdot A^{\top} \cdot f(A,x)\\
    & ~\cdot(\langle -f(A,x), f(A,x) \rangle + f(A,x)_{j_0})
\end{align*}
Finally, combine them and we have
\begin{align*}
       B_{5,1}^{j_1,i_1,j_0,i_1} = B_{5,1,1}^{j_1,i_1,j_0,i_1} + B_{5,1,2}^{j_1,i_1,j_0,i_1} + B_{5,1,3}^{j_1,i_1,j_0,i_1} + B_{5,1,4}^{j_1,i_1,j_0,i_1} + B_{5,1,5}^{j_1,i_1,j_0,i_1} + B_{5,1,6}^{j_1,i_1,j_0,i_1} + B_{5,1,7}^{j_1,i_1,j_0,i_1}
\end{align*}
{\bf Proof of Part 2.}
    \begin{align*}
    B_{5,2,1}^{j_1,i_1,j_0,i_1} : = & ~   f(A,x)_{j_1} \cdot f(A,x)_{j_0} \cdot x_{i_1} \cdot c_g(A,x)^{\top} \cdot \diag(x) \cdot A^{\top} \cdot f(A,x) \cdot(\langle -f(A,x), f(A,x) \rangle + f(A,x)_{j_0})  
\end{align*}
Finally, combine them and we have
\begin{align*}
       B_{5,2}^{j_1,i_1,j_0,i_1} = B_{5,2,1}^{j_1,i_1,j_0,i_1}  
\end{align*}
{\bf Proof of Part 3.} 
    \begin{align*}
    B_{5,3,1}^{j_1,i_1,j_0,i_1} : = & ~    -  c_g(A,x)^{\top} \cdot f(A,x)_{j_0} \cdot \diag(x) \cdot e_{i_1} \cdot e_{j_1}^\top \cdot f(A,x) \cdot(\langle -f(A,x), f(A,x) \rangle + f(A,x)_{j_0})
\end{align*}
Finally, combine them and we have
\begin{align*}
       B_{5,3}^{j_1,i_1,j_0,i_1} = B_{5,3,1}^{j_1,i_1,j_0,i_1} 
\end{align*}
{\bf Proof of Part 4.} 
    \begin{align*}
    B_{5,4,1}^{j_1,i_1,j_0,i_1} : = & ~ - c_g(A,x)^{\top} \cdot f(A,x)_{j_1} \cdot f(A,x)_{j_0} \cdot \diag(x) \cdot A^{\top}\cdot x_i \cdot  (e_{j_1}- f(A,x) ) \\
     & ~ \cdot (\langle -f(A,x), f(A,x) \rangle + f(A,x)_{j_0})
\end{align*}
Finally, combine them and we have
\begin{align*}
       B_{5,4}^{j_1,i_1,j_0,i_1} = B_{5,4,1}^{j_1,i_1,j_0,i_1}  
\end{align*}
{\bf Proof of Part 5.} 
    \begin{align*}
    B_{5,5,1}^{j_1,i_1,j_0,i_1} : = & ~ c_g(A,x)^{\top} \cdot f(A,x)_{j_1} \cdot f(A,x)_{j_0} \cdot \diag(x) \cdot A^{\top} \cdot  f(A,x)  \cdot  x_{i_1} \cdot \langle - f(A,x), f(A,x) \rangle\\
    B_{5,5,2}^{j_1,i_1,j_0,i_1} : = & ~ c_g(A,x)^{\top} \cdot f(A,x)_{j_1}^2 \cdot f(A,x)_{j_0} \cdot \diag(x) \cdot A^{\top} \cdot  f(A,x)  \cdot  x_{i_1}  
\end{align*}
Finally, combine them and we have
\begin{align*}
       B_{5,5}^{j_1,i_1,j_0,i_1} = B_{5,5,1}^{j_1,i_1,j_0,i_1}  +B_{5,5,2}^{j_1,i_1,j_0,i_1}
\end{align*}
{\bf Proof of Part 6.} 
    \begin{align*}
     B_{5,6,1}^{j_1,i_1,j_0,i_1} : = & ~ c_g(A,x)^{\top} \cdot f(A,x)_{j_1} \cdot f(A,x)_{j_0} \cdot \diag(x) \cdot A^{\top} \cdot  f(A,x)  \cdot  x_{i_1} \cdot \langle - f(A,x), f(A,x) \rangle\\
    B_{5,6,2}^{j_1,i_1,j_0,i_1} : = & ~ c_g(A,x)^{\top} \cdot f(A,x)_{j_1}^2 \cdot f(A,x)_{j_0} \cdot \diag(x) \cdot A^{\top} \cdot  f(A,x)  \cdot  x_{i_1}  
\end{align*}
Finally, combine them and we have
\begin{align*}
       B_{5,6}^{j_1,i_1,j_0,i_1} = B_{5,6,1}^{j_1,i_1,j_0,i_1}  +B_{5,6,2}^{j_1,i_1,j_0,i_1}
\end{align*}
{\bf Proof of Part 7.} 
    \begin{align*}
     B_{5,7,1}^{j_1,i_1,j_0,i_1} : = & ~  f(A,x)_{j_1} \cdot f(A,x)_{j_0}^2 \cdot  x_{i_1}\cdot c_g(A,x)^{\top} \cdot \diag(x) A^{\top} \cdot  f(A,x)   
\end{align*}
Finally, combine them and we have
\begin{align*}
       B_{5,7}^{j_1,i_1,j_0,i_1} = B_{5,7,1}^{j_1,i_1,j_0,i_1}  
\end{align*}
\end{proof}
\end{lemma}

\subsection{Case \texorpdfstring{$j_1 \neq j_0, i_1 \neq i_0$}{}}
\begin{lemma}
For $j_1 \neq j_0$ and $i_0 \neq i_1$. If the following conditions hold
    \begin{itemize}
     \item Let $u(A,x) \in \R^n$ be defined as Definition~\ref{def:u}
    \item Let $\alpha(A,x) \in \R$ be defined as Definition~\ref{def:alpha}
     \item Let $f(A,x) \in \R^n$ be defined as Definition~\ref{def:f}
    \item Let $c(A,x) \in \R^n$ be defined as Definition~\ref{def:c}
    \item Let $g(A,x) \in \R^d$ be defined as Definition~\ref{def:g} 
    \item Let $q(A,x) = c(A,x) + f(A,x) \in \R^n$
    \item Let $c_g(A,x) \in \R^d$ be defined as Definition~\ref{def:c_g}.
    \item Let $L_g(A,x) \in \R$ be defined as Definition~\ref{def:l_g}
    \item Let $v \in \R^n$ be a vector 
    \item Let $B_1^{j_1,i_1,j_0,i_0}$ be defined as Definition~\ref{def:b_1}
    \end{itemize}
    Then, For $j_0,j_1 \in [n], i_0,i_1 \in [d]$, we have 
    \begin{itemize}
\item {\bf Part 1.} For $B_{5,1}^{j_1,i_1,j_0,i_0}$, we have 
\begin{align*}
 B_{5,1}^{j_1,i_1,j_0,i_0}  = & ~  \frac{\d}{\d A_{j_1,i_1}} (- c_g(A,x)^{\top} ) \cdot  f(A,x)_{j_0} \cdot \diag (x) A^{\top} \cdot  f(A,x)  \cdot  (\langle -f(A,x), f(A,x) \rangle + f(A,x)_{j_0})\\
 = & ~ B_{5,1,1}^{j_1,i_1,j_0,i_0} + B_{5,1,2}^{j_1,i_1,j_0,i_0} + B_{5,1,3}^{j_1,i_1,j_0,i_0} + B_{5,1,4}^{j_1,i_1,j_0,i_0} + B_{5,1,5}^{j_1,i_1,j_0,i_0} + B_{5,1,6}^{j_1,i_1,j_0,i_0} + B_{5,1,7}^{j_1,i_1,j_0,i_0}
\end{align*} 
\item {\bf Part 2.} For $B_{5,2}^{j_1,i_1,j_0,i_0}$, we have 
\begin{align*}
  B_{5,2}^{j_1,i_1,j_0,i_0} = & ~ - c_g(A,x)^{\top} \cdot \frac{\d}{\d A_{j_1,i_1}} ( f(A,x)_{j_0} )  \cdot \diag(x) \cdot A^{\top} \cdot f(A,x) \cdot(\langle -f(A,x), f(A,x) \rangle + f(A,x)_{j_0}) \\
    = & ~  B_{5,2,1}^{j_1,i_1,j_0,i_0}  
\end{align*} 
\item {\bf Part 3.} For $B_{5,3}^{j_1,i_1,j_0,i_0}$, we have 
\begin{align*}
  B_{5,3}^{j_1,i_1,j_0,i_0} = & ~ - c_g(A,x)^{\top} \cdot f(A,x)_{j_0} \cdot \diag(x) \cdot \frac{\d}{\d A_{j_1,i_1}} (  A^{\top} ) \cdot f(A,x) \cdot(\langle -f(A,x), f(A,x) \rangle + f(A,x)_{j_0}) \\
     = & ~ B_{5,3,1}^{j_1,i_1,j_0,i_0}  
\end{align*} 
\item {\bf Part 4.} For $B_{5,4}^{j_1,i_1,j_0,i_0}$, we have 
\begin{align*}
  B_{5,4}^{j_1,i_1,j_0,i_0} = & ~ - c_g(A,x)^{\top} \cdot f(A,x)_{j_0} \cdot \diag(x) \cdot A^{\top} \cdot \frac{\d f(A,x)}{\d A_{j_1,i_1}} \cdot(\langle -f(A,x), f(A,x) \rangle + f(A,x)_{j_0}) \\
     = & ~ B_{5,4,1}^{j_1,i_1,j_0,i_0} 
\end{align*}
\item {\bf Part 5.} For $B_{5,5}^{j_1,i_1,j_0,i_0}$, we have 
\begin{align*}
  B_{5,5}^{j_1,i_1,j_0,i_0} = & ~  c_g(A,x)^{\top} \cdot f(A,x)_{j_0} \cdot \diag(x) \cdot A^{\top} \cdot  f(A,x)  \cdot \langle f(A,x), \frac{\d f(A,x) }{\d A_{j_1,i_1}}\rangle \\
     = & ~ B_{5,5,1}^{j_1,i_1,j_0,i_0}  + B_{5,5,2}^{j_1,i_1,j_0,i_0}  
\end{align*}
\item {\bf Part 6.} For $B_{5,6}^{j_1,i_1,j_0,i_0}$, we have 
\begin{align*}
  B_{5,6}^{j_1,i_1,j_0,i_0} = & ~ c_g(A,x)^{\top} \cdot f(A,x)_{j_0} \cdot \diag(x) \cdot A^{\top} \cdot  f(A,x)  \cdot \langle \frac{\d  f(A,x) }{\d A_{j_1,i_1}},f(A,x)\rangle \\
     = & ~ B_{5,6,1}^{j_1,i_1,j_0,i_0}  + B_{5,6,2}^{j_1,i_1,j_0,i_0}  
\end{align*}
\item {\bf Part 7.} For $B_{5,7}^{j_1,i_1,j_0,i_0}$, we have 
\begin{align*}
  B_{5,7}^{j_1,i_1,j_0,i_0} = & ~ - c_g(A,x)^{\top} \cdot f(A,x)_{j_1} \cdot \diag(x) \cdot A^{\top} \cdot  f(A,x)  \cdot  \frac{\d f(A,x)_{j_0}}{\d A_{j_1,i_1}} \\
     = & ~ B_{5,7,1}^{j_1,i_1,j_0,i_0}  
\end{align*}
\end{itemize}
\begin{proof}
    {\bf Proof of Part 1.}
    \begin{align*}
    B_{5,1,1}^{j_1,i_1,j_0,i_0} : = & ~ e_{i_1}^\top \cdot \langle c(A,x), f(A,x) \rangle \cdot  f(A,x)_{j_1} \cdot f(A,x)_{j_0} \cdot \diag(x) \cdot A^{\top} \cdot f(A,x) \\
    & ~ \cdot(\langle -f(A,x), f(A,x) \rangle + f(A,x)_{j_0})\\
    B_{5,1,2}^{j_1,i_1,j_0,i_0} : = & ~   e_{i_1}^\top \cdot c(A,x)_{j_1}\cdot f(A,x)_{j_1} \cdot f(A,x)_{j_0} \cdot \diag(x) \cdot A^{\top} \cdot f(A,x) \cdot(\langle -f(A,x), f(A,x) \rangle + f(A,x)_{j_0})\\
    B_{5,1,3}^{j_1,i_1,j_0,i_0} : = & ~ f(A,x)_{j_1} \cdot f(A,x)_{j_0} \cdot \langle c(A,x), f(A,x) \rangle \cdot ( (A_{j_1,*}) \circ x^\top  )  \cdot \diag(x) \cdot A^{\top} \cdot f(A,x) \cdot \\
    & ~(\langle -f(A,x), f(A,x) \rangle + f(A,x)_{j_0})\\
    B_{5,1,4}^{j_1,i_1,j_0,i_0} : = & ~   -  f(A,x)_{j_1} \cdot f(A,x)_{j_0} \cdot f(A,x)^\top  \cdot A \cdot (\diag(x))^2 \cdot   \langle c(A,x), f(A,x) \rangle \cdot A^{\top} \cdot f(A,x)\\
    & ~ \cdot(\langle -f(A,x), f(A,x) \rangle + f(A,x)_{j_0})\\
    B_{5,1,5}^{j_1,i_1,j_0,i_0} : = & ~    f(A,x)_{j_1} \cdot f(A,x)_{j_0} \cdot f(A,x)^\top  \cdot A \cdot (\diag(x))^2 \cdot (\langle (-f(A,x)), f(A,x) \rangle + f(A,x)_{j_1})   \cdot A^{\top} \cdot f(A,x) \\
    & ~ \cdot(\langle -f(A,x), f(A,x) \rangle + f(A,x)_{j_0})\\
    B_{5,1,6}^{j_1,i_1,j_0,i_0} : = & ~  f(A,x)_{j_1} \cdot f(A,x)_{j_0} \cdot f(A,x)^\top  \cdot A \cdot  (\diag(x))^2 \cdot(\langle -f(A,x), c(A,x) \rangle + f(A,x)_{j_1}) \cdot A^{\top} \cdot f(A,x) \\
        & ~ \cdot(\langle -f(A,x), f(A,x) \rangle + f(A,x)_{j_0})\\
    B_{5,1,7}^{j_1,i_1,j_0,i_0} : = & ~   f(A,x)_{j_1} \cdot f(A,x)_{j_0} \cdot ((e_{j_1}^\top - f(A,x)^\top) \circ q(A,x)^\top) \cdot A \cdot  (\diag(x))^2 \cdot A^{\top} \cdot f(A,x)\\
    & ~\cdot(\langle -f(A,x), f(A,x) \rangle + f(A,x)_{j_0})
\end{align*}
Finally, combine them and we have
\begin{align*}
       B_{5,1}^{j_1,i_1,j_0,i_0} = B_{5,1,1}^{j_1,i_1,j_0,i_0} + B_{5,1,2}^{j_1,i_1,j_0,i_0} + B_{5,1,3}^{j_1,i_1,j_0,i_0} + B_{5,1,4}^{j_1,i_1,j_0,i_0} + B_{5,1,5}^{j_1,i_1,j_0,i_0} + B_{5,1,6}^{j_1,i_1,j_0,i_0} + B_{5,1,7}^{j_1,i_1,j_0,i_0}
\end{align*}
{\bf Proof of Part 2.}
    \begin{align*}
    B_{5,2,1}^{j_1,i_1,j_0,i_0} : = & ~   f(A,x)_{j_1} \cdot f(A,x)_{j_0} \cdot x_{i_1} \cdot c_g(A,x)^{\top} \cdot \diag(x) \cdot A^{\top} \cdot f(A,x) \cdot(\langle -f(A,x), f(A,x) \rangle + f(A,x)_{j_0})  
\end{align*}
Finally, combine them and we have
\begin{align*}
       B_{5,2}^{j_1,i_1,j_0,i_0} = B_{5,2,1}^{j_1,i_1,j_0,i_0}  
\end{align*}
{\bf Proof of Part 3.} 
    \begin{align*}
    B_{5,3,1}^{j_1,i_1,j_0,i_0} : = & ~    -  c_g(A,x)^{\top} \cdot f(A,x)_{j_0} \cdot \diag(x) \cdot e_{i_1} \cdot e_{j_1}^\top \cdot f(A,x) \cdot(\langle -f(A,x), f(A,x) \rangle + f(A,x)_{j_0})
\end{align*}
Finally, combine them and we have
\begin{align*}
       B_{5,3}^{j_1,i_1,j_0,i_0} = B_{5,3,1}^{j_1,i_1,j_0,i_0} 
\end{align*}
{\bf Proof of Part 4.} 
    \begin{align*}
    B_{5,4,1}^{j_1,i_1,j_0,i_0} : = & ~ - c_g(A,x)^{\top} \cdot f(A,x)_{j_1} \cdot f(A,x)_{j_0} \cdot \diag(x) \cdot A^{\top}\cdot x_i \cdot  (e_{j_1}- f(A,x) ) \\
    & ~ \cdot (\langle -f(A,x), f(A,x) \rangle + f(A,x)_{j_0})
\end{align*}
Finally, combine them and we have
\begin{align*}
       B_{5,4}^{j_1,i_1,j_0,i_0} = B_{5,4,1}^{j_1,i_1,j_0,i_0}  
\end{align*}
{\bf Proof of Part 5.} 
    \begin{align*}
    B_{5,5,1}^{j_1,i_1,j_0,i_0} : = & ~ c_g(A,x)^{\top} \cdot f(A,x)_{j_1} \cdot f(A,x)_{j_0} \cdot \diag(x) \cdot A^{\top} \cdot  f(A,x)  \cdot  x_{i_1} \cdot \langle - f(A,x), f(A,x) \rangle\\
    B_{5,5,2}^{j_1,i_1,j_0,i_0} : = & ~ c_g(A,x)^{\top} \cdot f(A,x)_{j_1}^2 \cdot f(A,x)_{j_0} \cdot \diag(x) \cdot A^{\top} \cdot  f(A,x)  \cdot  x_{i_1}  
\end{align*}
Finally, combine them and we have
\begin{align*}
       B_{5,5}^{j_1,i_1,j_0,i_0} = B_{5,5,1}^{j_1,i_1,j_0,i_0}  +B_{5,5,2}^{j_1,i_1,j_0,i_0}
\end{align*}
{\bf Proof of Part 6.} 
    \begin{align*}
     B_{5,6,1}^{j_1,i_1,j_0,i_0} : = & ~ c_g(A,x)^{\top} \cdot f(A,x)_{j_1} \cdot f(A,x)_{j_0} \cdot \diag(x) \cdot A^{\top} \cdot  f(A,x)  \cdot  x_{i_1} \cdot \langle - f(A,x), f(A,x) \rangle\\
    B_{5,6,2}^{j_1,i_1,j_0,i_0} : = & ~ c_g(A,x)^{\top} \cdot f(A,x)_{j_1}^2 \cdot f(A,x)_{j_0} \cdot \diag(x) \cdot A^{\top} \cdot  f(A,x)  \cdot  x_{i_1}  
\end{align*}
Finally, combine them and we have
\begin{align*}
       B_{5,6}^{j_1,i_1,j_0,i_0} = B_{5,6,1}^{j_1,i_1,j_0,i_0}  +B_{5,6,2}^{j_1,i_1,j_0,i_0}
\end{align*}
{\bf Proof of Part 7.} 
    \begin{align*}
     B_{5,7,1}^{j_1,i_1,j_0,i_0} : = & ~  f(A,x)_{j_1} \cdot f(A,x)_{j_0}^2 \cdot  x_{i_1}\cdot c_g(A,x)^{\top} \cdot \diag(x) A^{\top} \cdot  f(A,x)   
\end{align*}
Finally, combine them and we have
\begin{align*}
       B_{5,7}^{j_1,i_1,j_0,i_0} = B_{5,7,1}^{j_1,i_1,j_0,i_0}  
\end{align*}
\end{proof}
\end{lemma}

\subsection{Constructing \texorpdfstring{$d \times d$}{} matrices for \texorpdfstring{$j_1 \neq j_0$}{}}
The purpose of the following lemma is to let $i_0$ and $i_1$ disappear.
\begin{lemma}For $j_0,j_1 \in [n]$, a list of $d \times d$ matrices can be expressed as the following sense,\label{lem:b_5_j1_j0}
\begin{itemize}
\item {\bf Part 1.}
\begin{align*}
B_{5,1,1}^{j_1,*,j_0,*} & ~ =   f_c(A,x) \cdot   f(A,x)_{j_1} \cdot f(A,x)_{j_0} \cdot  (-f_2(A,x) + f(A,x)_{j_0}) \cdot h(A,x) \cdot {\bf 1}_d^\top
\end{align*}
\item {\bf Part 2.}
\begin{align*}
B_{5,1,2}^{j_1,*,j_0,*} & ~ =    c(A,x)_{j_1}\cdot  f(A,x)_{j_1} \cdot f(A,x)_{j_0} \cdot  (-f_2(A,x) + f(A,x)_{j_0}) \cdot h(A,x) \cdot {\bf 1}_d^\top 
\end{align*}
\item {\bf Part 3.}
\begin{align*}
B_{5,1,3}^{j_1,*,j_0,*} & ~ =     f(A,x)_{j_1} \cdot f(A,x)_{j_0} \cdot f_c(A,x) \cdot (-f_2(A,x) + f(A,x)_{j_0}) \cdot ( (A_{j_1,*}) \circ x^\top  )  \cdot h(A,x)\cdot I_d
\end{align*}
\item {\bf Part 4.}
\begin{align*}
B_{5,1,4}^{j_1,*,j_0,*}  & ~ =   -   f(A,x)_{j_1} \cdot f(A,x)_{j_0} \cdot f_c(A,x)  \cdot(-f_2(A,x) + f(A,x)_{j_0}) \cdot h(A,x)^\top \cdot h(A,x) \cdot I_d
\end{align*}
\item {\bf Part 5.}
\begin{align*}
B_{5,1,5}^{j_1,*,j_0,*}  & ~ =    f(A,x)_{j_1} \cdot f(A,x)_{j_0} \cdot (-f_2(A,x) + f(A,x)_{j_1}) \cdot(-f_2(A,x) + f(A,x)_{j_0}) \cdot h(A,x)^\top \cdot h(A,x) \cdot I_d
\end{align*}
\item {\bf Part 6.}
\begin{align*}
B_{5,1,6}^{j_1,*,j_0,*}  & ~ =      f(A,x)_{j_1} \cdot f(A,x)_{j_0} \cdot (-f_2(A,x) + f(A,x)_{j_0})\cdot(-f_c(A,x) + f(A,x)_{j_1}) \cdot h(A,x)^\top \cdot h(A,x) \cdot I_d
\end{align*}
\item {\bf Part 7.}
\begin{align*}
B_{5,1,7}^{j_1,*,j_0,*}  & ~ =    f(A,x)_{j_1} \cdot f(A,x)_{j_0} \cdot (-f_2(A,x) + f(A,x)_{j_0})\cdot p_{j_1}(A,x)^\top \cdot h(A,x) \cdot I_d
\end{align*}
\item {\bf Part 8.}
\begin{align*}
B_{5,2,1}^{j_1,*,j_0,*}  & ~ =      f(A,x)_{j_1} \cdot f(A,x)_{j_0} \cdot  (-f_2(A,x) + f(A,x)_{j_0}) \cdot c_g(A,x)^{\top} \cdot h(A,x) \cdot x \cdot {\bf 1}_d^{\top} 
\end{align*}
\item {\bf Part 9.}
\begin{align*}
 B_{5,3,1}^{j_1,*,j_0,*}  & ~ =      -  f(A,x)_{j_1} \cdot f(A,x)_{j_0} \cdot (-f_2(A,x) + f(A,x)_{j_0})\cdot {\bf 1}_d \cdot c_g(A,x)^{\top}   \cdot \diag(x)  
\end{align*}
\item {\bf Part 10.}
\begin{align*}
B_{5,4,1}^{j_1,*,j_0,*}  =     -  f(A,x)_{j_1} \cdot f(A,x)_{j_0} \cdot (-f_2(A,x) + f(A,x)_{j_0}) \cdot c_g(A,x)^{\top}  \cdot  h_e(A,x)\cdot x \cdot {\bf 1}_d^{\top}
\end{align*}
\item {\bf Part 11.}
\begin{align*}
B_{5,5,1}^{j_1,*,j_0,*}  & ~ =    -  f(A,x)_{j_1} \cdot f(A,x)_{j_0} \cdot f_2(A,x) \cdot c_g(A,x)^{\top} \cdot h(A,x)  \cdot x \cdot {\bf 1}_d^{\top}
\end{align*}
\item {\bf Part 12.}
\begin{align*}
 B_{5,5,2}^{j_1,*,j_0,*}  =     f(A,x)_{j_1}^2 \cdot f(A,x)_{j_0}  \cdot c_g(A,x)^{\top} \cdot h(A,x)  \cdot x \cdot {\bf 1}_d^{\top}
\end{align*}
\item {\bf Part 13.}
\begin{align*}
B_{5,6,1}^{j_1,*,j_0,*}  =   -  f(A,x)_{j_1} \cdot f(A,x)_{j_0} \cdot f_2(A,x) \cdot c_g(A,x)^{\top} \cdot h(A,x)  \cdot x \cdot {\bf 1}_d^{\top}
\end{align*}
\item {\bf Part 14.}
\begin{align*}
B_{5,6,2}^{j_1,*,j_0,*}  =    f(A,x)_{j_1}^2 \cdot f(A,x)_{j_0}  \cdot c_g(A,x)^{\top} \cdot h(A,x)  \cdot x \cdot {\bf 1}_d^{\top}
\end{align*}
\item {\bf Part 15.}
\begin{align*}
B_{5,7,1}^{j_1,*,j_0,*}  =    f(A,x)_{j_1}\cdot f(A,x)_{j_0}^2   \cdot c_g(A,x)^{\top} \cdot h(A,x)  \cdot x \cdot {\bf 1}_d^{\top}
\end{align*}

\end{itemize}
\begin{proof}
{\bf Proof of Part 1.}
    We have
    \begin{align*}
        B_{5,1,1}^{j_1,i_1,j_0,i_1}  = & ~e_{i_1}^\top \cdot \langle c(A,x), f(A,x) \rangle \cdot  f(A,x)_{j_1} \cdot f(A,x)_{j_0} \cdot \diag(x) \cdot A^{\top}  \\
        & ~\cdot f(A,x) \cdot(\langle -f(A,x), f(A,x) \rangle + f(A,x)_{j_0})\\
        B_{5,1,1}^{j_1,i_1,j_0,i_0}  = & ~ e_{i_1}^\top \cdot \langle c(A,x), f(A,x) \rangle \cdot  f(A,x)_{j_1} \cdot f(A,x)_{j_0} \cdot \diag(x) \cdot A^{\top}  \\
        & ~\cdot f(A,x) \cdot(\langle -f(A,x), f(A,x) \rangle + f(A,x)_{j_0})
    \end{align*}
    From the above two equations, we can tell that $B_{5,1,1}^{j_1,*,j_0,*} \in \R^{d \times d}$ is a matrix that both the diagonal and off-diagonal have entries.
    
    Then we have $B_{5,1,1}^{j_1,*,j_0,*} \in \R^{d \times d}$ can be written as the rescaling of a diagonal matrix,
    \begin{align*}
     B_{5,1,1}^{j_1,*,j_0,*} & ~ = \langle c(A,x), f(A,x) \rangle \cdot  f(A,x)_{j_1} \cdot f(A,x)_{j_0} \\
        & ~ \cdot  (\langle -f(A,x), f(A,x) \rangle + f(A,x)_{j_0}) \cdot \diag(x) \cdot A^{\top} \cdot f(A,x) \cdot {\bf 1}_d^\top \\
     & ~ = f_c(A,x) \cdot  f(A,x)_{j_1} \cdot f(A,x)_{j_0} \cdot  (-f_2(A,x) + f(A,x)_{j_0}) \cdot h(A,x) \cdot {\bf 1}_d^\top
\end{align*}
    where the last step is follows from the Definitions~\ref{def:h}, Definitions~\ref{def:f_2} and Definitions~\ref{def:f_c}. 

{\bf Proof of Part 2.}
    We have
    \begin{align*}
           B_{5,1,2}^{j_1,i_1,j_0,i_1} = & ~ e_{i_1}^\top \cdot c(A,x)_{j_1}\cdot f(A,x)_{j_1} \cdot f(A,x)_{j_0} \cdot \diag(x) \cdot A^{\top} \cdot f(A,x) \cdot(\langle -f(A,x), f(A,x) \rangle + f(A,x)_{j_0})\\
        B_{5,1,2}^{j_1,i_1,j_0,i_0} = & ~ e_{i_1}^\top \cdot c(A,x)_{j_1}\cdot f(A,x)_{j_1} \cdot f(A,x)_{j_0} \cdot \diag(x) \cdot A^{\top} \cdot f(A,x) \cdot(\langle -f(A,x), f(A,x) \rangle + f(A,x)_{j_0})
    \end{align*}
     From the above two equations, we can tell that $B_{5,1,2}^{j_1,*,j_0,*} \in \R^{d \times d}$ is a matrix that only diagonal has entries and off-diagonal are all zeros.
    
    Then we have $B_{5,1,2}^{j_1,*,j_0,*} \in \R^{d \times d}$ can be written as the rescaling of a diagonal matrix,
\begin{align*}
     B_{5,1,2}^{j_1,*,j_0,*} & ~ = c(A,x)_{j_1}\cdot f(A,x)_{j_1} \cdot f(A,x)_{j_0} \cdot  (\langle -f(A,x), f(A,x) \rangle + f(A,x)_{j_0}) \cdot \diag(x) \cdot A^{\top} \cdot f(A,x) \cdot {\bf 1}_d^\top \\
     & ~ =  c(A,x)_{j_1}\cdot f(A,x)_{j_1} \cdot f(A,x)_{j_0} \cdot  (-f_2(A,x) + f(A,x)_{j_0}) \cdot h(A,x) \cdot {\bf 1}_d^\top 
\end{align*}
    where the last step is follows from the Definitions~\ref{def:h} and Definitions~\ref{def:f_2}.

{\bf Proof of Part 3.}
We have for diagonal entry and off-diagonal entry can be written as follows 
    \begin{align*}
        B_{5,1,3}^{j_1,i_1,j_0,i_1} = & ~f(A,x)_{j_1} \cdot f(A,x)_{j_0} \cdot \langle c(A,x), f(A,x) \rangle \cdot ( (A_{j_1,*}) \circ x^\top  )  \cdot \diag(x) \cdot A^{\top} \cdot f(A,x) \\
     & ~\cdot (\langle -f(A,x), f(A,x) \rangle + f(A,x)_{j_0}) \\
        B_{5,1,3}^{j_1,i_1,j_0,i_0} = & ~f(A,x)_{j_1} \cdot f(A,x)_{j_0} \cdot \langle c(A,x), f(A,x) \rangle \cdot ( (A_{j_1,*}) \circ x^\top  )  \cdot \diag(x) \cdot A^{\top} \cdot f(A,x) \\
     & ~\cdot (\langle -f(A,x), f(A,x) \rangle + f(A,x)_{j_0})
    \end{align*}
From the above equation, we can show that matrix $B_{5,1,3}^{j_1,*,j_0,*}$ can be expressed as a rank-$1$ matrix,
\begin{align*}
     B_{5,1,3}^{j_1,*,j_0,*} & ~ = f(A,x)_{j_1} \cdot f(A,x)_{j_0} \cdot \langle c(A,x), f(A,x) \rangle \cdot (\langle -f(A,x), f(A,x) \rangle + f(A,x)_{j_0}) \cdot ( (A_{j_1,*}) \circ x^\top  )  \\
     & ~\cdot \diag(x) \cdot A^{\top} \cdot f(A,x) \cdot I_d\\
     & ~ =  f(A,x)_{j_1} \cdot f(A,x)_{j_0} \cdot f_c(A,x) \cdot (-f_2(A,x) + f(A,x)_{j_0}) \cdot ( (A_{j_1,*}) \circ x^\top  )  \cdot h(A,x)\cdot I_d
\end{align*}
    where the last step is follows from the Definitions~\ref{def:h}, Definitions~\ref{def:f_2} and Definitions~\ref{def:f_c}.

{\bf Proof of Part 4.}
We have for diagonal entry and off-diagonal entry can be written as follows
    \begin{align*}
        B_{5,1,4}^{j_1,i_1,j_0,i_1}   = & ~  -   f(A,x)_{j_1} \cdot f(A,x)_{j_0} \cdot f(A,x)^\top  \cdot A \cdot (\diag(x))^2 \cdot   \langle c(A,x), f(A,x) \rangle \cdot A^{\top} \cdot f(A,x) \\
     & ~\cdot(\langle -f(A,x), f(A,x) \rangle + f(A,x)_{j_0}) \\
        B_{5,1,4}^{j_1,i_1,j_0,i_0}   = & ~ -     f(A,x)_{j_1} \cdot f(A,x)_{j_0} \cdot f(A,x)^\top  \cdot A \cdot (\diag(x))^2 \cdot   \langle c(A,x), f(A,x) \rangle \cdot A^{\top} \cdot f(A,x) \\
     & ~\cdot(\langle -f(A,x), f(A,x) \rangle + f(A,x)_{j_0})
    \end{align*}
 From the above equation, we can show that matrix $B_{5,1,4}^{j_1,*,j_0,*}$ can be expressed as a rank-$1$ matrix,
\begin{align*}
    B_{5,1,4}^{j_1,*,j_0,*}  & ~ = -   f(A,x)_{j_1} \cdot f(A,x)_{j_0} \cdot \langle c(A,x), f(A,x) \rangle  \cdot(\langle -f(A,x), f(A,x) \rangle + f(A,x)_{j_0}) \\
    & ~\cdot f(A,x)^\top  \cdot A \cdot (\diag(x))^2 \cdot  A^{\top} \cdot f(A,x)\cdot I_d\\
     & ~ =   -  f(A,x)_{j_1} \cdot f(A,x)_{j_0} \cdot f_c(A,x)  \cdot(-f_2(A,x) + f(A,x)_{j_0}) \cdot h(A,x)^\top \cdot h(A,x) \cdot I_d
\end{align*}
   where the last step is follows from the Definitions~\ref{def:h}, Definitions~\ref{def:f_2} and Definitions~\ref{def:f_c}.

{\bf Proof of Part 5.}
We have for diagonal entry and off-diagonal entry can be written as follows
    \begin{align*}
         B_{5,1,5}^{j_1,i_1,j_0,i_0} = & ~    f(A,x)_{j_1} \cdot f(A,x)_{j_0} \cdot f(A,x)^\top  \cdot A \cdot (\diag(x))^2 \cdot (\langle -f(A,x), f(A,x) \rangle + f(A,x)_{j_1})   \\
     & ~\cdot (\langle -f(A,x), f(A,x) \rangle + f(A,x)_{j_0})   \cdot A^{\top} \cdot f(A,x)  \\
         B_{5,1,5}^{j_1,i_1,j_0,i_0} = & ~    f(A,x)_{j_1} \cdot f(A,x)_{j_0} \cdot f(A,x)^\top  \cdot A \cdot (\diag(x))^2 \cdot (\langle -f(A,x), f(A,x) \rangle + f(A,x)_{j_1}) \\
     & ~\cdot (\langle -f(A,x), f(A,x) \rangle + f(A,x)_{j_0})  \cdot A^{\top} \cdot f(A,x) 
    \end{align*}
    From the above equation, we can show that matrix $B_{5,1,5}^{j_1,*,j_0,*}$ can be expressed as a rank-$1$ matrix,
\begin{align*}
    B_{5,1,5}^{j_1,*,j_0,*}  & ~ =  f(A,x)_{j_1} \cdot f(A,x)_{j_0} \cdot (\langle -f(A,x), f(A,x) \rangle + f(A,x)_{j_1}) \cdot (\langle -f(A,x), f(A,x) \rangle + f(A,x)_{j_0}) \\
     & ~\cdot f(A,x)^\top  \cdot A \cdot (\diag(x))^2 \cdot  A^{\top} \cdot f(A,x) \cdot I_d\\
     & ~ =    f(A,x)_{j_1} \cdot f(A,x)_{j_0} \cdot (-f_2(A,x) + f(A,x)_{j_1}) \cdot (-f_2(A,x) + f(A,x)_{j_0}) \cdot h(A,x)^\top \cdot h(A,x) \cdot I_d
\end{align*}
    where the last step is follows from the Definitions~\ref{def:h} and Definitions~\ref{def:f_2}.

{\bf Proof of Part 6.}
We have for diagonal entry and off-diagonal entry can be written as follows
    \begin{align*}
        B_{5,1,6}^{j_1,i_1,j_0,i_1}  = & ~   f(A,x)_{j_1} \cdot f(A,x)_{j_0}  \cdot f(A,x)^\top  \cdot A \cdot  (\diag(x))^2 \cdot(\langle -f(A,x), c(A,x) \rangle + f(A,x)_{j_1}) \cdot A^{\top} \cdot f(A,x) \\
        & ~ \cdot(\langle -f(A,x), f(A,x) \rangle + f(A,x)_{j_0})\\
        B_{5,1,6}^{j_1,i_1,j_0,i_0}  = & ~   f(A,x)_{j_1} \cdot f(A,x)_{j_0}  \cdot f(A,x)^\top  \cdot A \cdot  (\diag(x))^2 \cdot(\langle -f(A,x), c(A,x) \rangle + f(A,x)_{j_1}) \cdot A^{\top} \cdot f(A,x) \\
        & ~ \cdot(\langle -f(A,x), f(A,x) \rangle + f(A,x)_{j_0})
    \end{align*}
    From the above equation, we can show that matrix $B_{5,1,6}^{j_1,*,j_0,*}$ can be expressed as a rank-$1$ matrix,
\begin{align*}
    B_{5,1,6}^{j_1,*,j_0,*}  & ~ =   f(A,x)_{j_1} \cdot f(A,x)_{j_0}  \cdot (\langle -f(A,x), f(A,x) \rangle + f(A,x)_{j_0})\cdot(\langle -f(A,x), c(A,x) \rangle + f(A,x)_{j_1}) \\
    & ~\cdot f(A,x)^\top  \cdot A \cdot (\diag(x))^2 \cdot  A^{\top} \cdot f(A,x) \cdot I_d\\
     & ~ =   f(A,x)_{j_1} \cdot f(A,x)_{j_0}  \cdot (-f_2(A,x) + f(A,x)_{j_0})\cdot(-f_c(A,x) + f(A,x)_{j_1}) \cdot h(A,x)^\top \cdot h(A,x) \cdot I_d
\end{align*}
    where the last step is follows from the Definitions~\ref{def:h}, Definitions~\ref{def:f_2} and Definitions~\ref{def:f_c}.
    
{\bf Proof of Part 7.}
We have for diagonal entry and off-diagonal entry can be written as follows
    \begin{align*}
         B_{5,1,7}^{j_1,i_1,j_0,i_1} = & ~  f(A,x)_{j_1} \cdot f(A,x)_{j_0} \cdot ((e_{j_1}^\top - f(A,x)^\top) \circ q(A,x)^\top) \cdot A \cdot  (\diag(x))^2 \cdot A^{\top} \cdot f(A,x) \\
    & ~\cdot(\langle -f(A,x), f(A,x) \rangle + f(A,x)_{j_0})\\
         B_{5,1,7}^{j_1,i_1,j_0,i_0} = & ~  f(A,x)_{j_1} \cdot f(A,x)_{j_0} \cdot ((e_{j_1}^\top - f(A,x)^\top) \circ q(A,x)^\top) \cdot A \cdot  (\diag(x))^2 \cdot A^{\top} \cdot f(A,x) \\
    & ~\cdot(\langle -f(A,x), f(A,x) \rangle + f(A,x)_{j_0})
    \end{align*}
    From the above equation, we can show that matrix $B_{5,1,7}^{j_1,*,j_0,*}$ can be expressed as a rank-$1$ matrix,
\begin{align*}
     B_{5,1,7}^{j_1,*,j_0,*}  & ~ =   f(A,x)_{j_1} \cdot f(A,x)_{j_0} \cdot (\langle -f(A,x), f(A,x) \rangle + f(A,x)_{j_0}) \\
     &~\cdot ((e_{j_1}^\top - f(A,x)^\top) \circ q(A,x)^\top) \cdot A \cdot  (\diag(x))^2 \cdot A^{\top} \cdot f(A,x) \cdot I_d\\
     & ~ = f(A,x)_{j_1} \cdot f(A,x)_{j_0} \cdot (-f_2(A,x) + f(A,x)_{j_0})\cdot p_{j_1}(A,x)^\top \cdot h(A,x) \cdot I_d
\end{align*}
    where the last step is follows from the Definitions~\ref{def:h}, Definitions~\ref{def:f_2} and Definitions~\ref{def:p}.

    {\bf Proof of Part 8.}
We have for diagonal entry and off-diagonal entry can be written as follows
    \begin{align*}
         B_{5,2,1}^{j_1,i_1,j_0,i_1} = & ~   f(A,x)_{j_1} \cdot f(A,x)_{j_0} \cdot x_{i_1} \cdot c_g(A,x)^{\top} \cdot \diag(x) \cdot A^{\top} \cdot f(A,x) \cdot(\langle -f(A,x), f(A,x) \rangle + f(A,x)_{j_0})\\
         B_{5,2,1}^{j_1,i_1,j_0,i_0} = & ~  f(A,x)_{j_1} \cdot f(A,x)_{j_0} \cdot x_{i_1} \cdot c_g(A,x)^{\top} \cdot \diag(x) \cdot A^{\top} \cdot f(A,x) \cdot(\langle -f(A,x), f(A,x) \rangle + f(A,x)_{j_0})
    \end{align*}
    From the above equation, we can show that matrix $B_{5,2,1}^{j_1,*,j_0,*}$ can be expressed as a rank-$1$ matrix,
\begin{align*}
     B_{5,2,1}^{j_1,*,j_0,*}  & ~ =   f(A,x)_{j_1} \cdot f(A,x)_{j_0} \cdot  (\langle -f(A,x), f(A,x) \rangle + f(A,x)_{j_0}) \cdot c_g(A,x)^{\top}\\
     & ~ \cdot \diag(x) \cdot A^{\top} \cdot f(A,x) \cdot x \cdot {\bf 1}_d^{\top}  \\ 
     & ~ =  f(A,x)_{j_1} \cdot f(A,x)_{j_0} \cdot  (-f_2(A,x) + f(A,x)_{j_0}) \cdot c_g(A,x)^{\top} \cdot h(A,x) \cdot x \cdot {\bf 1}_d^{\top} 
\end{align*}
    where the last step is follows from the Definitions~\ref{def:h}, Definitions~\ref{def:f_2}.

   {\bf Proof of Part 9.}
We have for diagonal entry and off-diagonal entry can be written as follows
    \begin{align*}
         B_{5,3,1}^{j_1,i_1,j_0,i_1} = & ~   -  c_g(A,x)^{\top} \cdot f(A,x)_{j_0} \cdot \diag(x) \cdot e_{i_1} \cdot e_{j_1}^\top \cdot f(A,x) \cdot(\langle -f(A,x), f(A,x) \rangle + f(A,x)_{j_0})\\
         B_{5,3,1}^{j_1,i_1,j_0,i_0} = & ~ -  c_g(A,x)^{\top} \cdot f(A,x)_{j_0} \cdot \diag(x) \cdot e_{i_1} \cdot e_{j_1}^\top \cdot f(A,x) \cdot(\langle -f(A,x), f(A,x) \rangle + f(A,x)_{j_0})
    \end{align*}
    From the above equation, we can show that matrix $B_{5,3,1}^{j_1,*,j_0,*}$ can be expressed as a rank-$1$ matrix,
\begin{align*}
     B_{5,3,1}^{j_1,*,j_0,*}  & ~ = - f(A,x)_{j_1} \cdot f(A,x)_{j_0} \cdot (\langle -f(A,x), f(A,x) \rangle + f(A,x)_{j_0})\cdot {\bf 1}_d \cdot c_g(A,x)^{\top}   \cdot \diag(x)  \\ 
     & ~ =   - f(A,x)_{j_1} \cdot f(A,x)_{j_0} \cdot (-f_2(A,x) + f(A,x)_{j_0})\cdot {\bf 1}_d \cdot c_g(A,x)^{\top}   \cdot \diag(x)  
\end{align*}
    where the last step is follows from the Definitions~\ref{def:f_2}.

    {\bf Proof of Part 10.}
We have for diagonal entry and off-diagonal entry can be written as follows
    \begin{align*}
         B_{5,4,1}^{j_1,i_1,j_0,i_1} = & ~   - c_g(A,x)^{\top} \cdot  f(A,x)_{j_1} \cdot f(A,x)_{j_0} \cdot \diag(x) \cdot A^{\top}\cdot x_i \cdot  (e_{j_1}- f(A,x) ) \\
     & ~\cdot (\langle -f(A,x), f(A,x) \rangle + f(A,x)_{j_0})\\
         B_{5,4,1}^{j_1,i_1,j_0,i_0} = & ~  - c_g(A,x)^{\top} \cdot  f(A,x)_{j_1} \cdot f(A,x)_{j_0} \cdot \diag(x) \cdot A^{\top}\cdot x_i \cdot  (e_{j_1}- f(A,x) ) \\
     & ~\cdot (\langle -f(A,x), f(A,x) \rangle + f(A,x)_{j_0})
    \end{align*}
    From the above equation, we can show that matrix $B_{5,4,1}^{j_1,*,j_0,*}$ can be expressed as a rank-$1$ matrix,
\begin{align*}
     B_{5,4,1}^{j_1,*,j_0,*}  & ~ = -  f(A,x)_{j_1} \cdot f(A,x)_{j_0} \cdot (\langle -f(A,x), f(A,x) \rangle + f(A,x)_{j_0}) \cdot c_g(A,x)^{\top} \\
     & ~ \cdot \diag(x) \cdot A^{\top}  \cdot  (e_{j_1}- f(A,x) ) \cdot x \cdot {\bf 1}_d^{\top}\\ 
     & ~ =   -  f(A,x)_{j_1} \cdot f(A,x)_{j_0} \cdot (-f_2(A,x) + f(A,x)_{j_0}) \cdot c_g(A,x)^{\top}  \cdot  h_e(A,x)\cdot x \cdot {\bf 1}_d^{\top}
\end{align*}
    where the last step is follows from the Definitions~\ref{def:f_2} and Definitions~\ref{def:h_e}.

        {\bf Proof of Part 11.}
We have for diagonal entry and off-diagonal entry can be written as follows
    \begin{align*}
         B_{5,5,1}^{j_1,i_1,j_0,i_1} = & ~  c_g(A,x)^{\top} \cdot f(A,x)_{j_1} \cdot f(A,x)_{j_0} \cdot \diag(x) \cdot A^{\top} \cdot  f(A,x)  \cdot  x_{i_1} \cdot \langle - f(A,x), f(A,x) \rangle\\
         B_{5,5,1}^{j_1,i_1,j_0,i_0} = & ~ c_g(A,x)^{\top} \cdot f(A,x)_{j_1} \cdot f(A,x)_{j_0} \cdot \diag(x) \cdot A^{\top} \cdot  f(A,x)  \cdot  x_{i_1} \cdot \langle - f(A,x), f(A,x) \rangle
    \end{align*}
    From the above equation, we can show that matrix $B_{5,5,1}^{j_1,*,j_0,*}$ can be expressed as a rank-$1$ matrix,
\begin{align*}
     B_{5,5,1}^{j_1,*,j_0,*}  & ~ = f(A,x)_{j_1} \cdot f(A,x)_{j_0} \cdot \langle - f(A,x), f(A,x) \rangle \cdot c_g(A,x)^{\top} \cdot \diag(x) \cdot A^{\top} \cdot  f(A,x)  \cdot x \cdot {\bf 1}_d^{\top}\\ 
     & ~ =  -  f(A,x)_{j_1} \cdot f(A,x)_{j_0} \cdot f_2(A,x) \cdot c_g(A,x)^{\top} \cdot h(A,x)  \cdot x \cdot {\bf 1}_d^{\top}
\end{align*}
    where the last step is follows from the Definitions~\ref{def:h} and Definitions~\ref{def:f_2}.

    {\bf Proof of Part 12.}
We have for diagonal entry and off-diagonal entry can be written as follows
    \begin{align*}
         B_{5,5,2}^{j_1,i_1,j_0,i_1} = & ~  c_g(A,x)^{\top} \cdot f(A,x)_{j_1}^2 \cdot f(A,x)_{j_0} \cdot \diag(x) \cdot A^{\top} \cdot  f(A,x)  \cdot  x_{i_1}  \\
         B_{5,5,2}^{j_1,i_1,j_0,i_0} = & ~ c_g(A,x)^{\top} \cdot f(A,x)_{j_1}^2 \cdot f(A,x)_{j_0} \cdot \diag(x) \cdot A^{\top} \cdot  f(A,x)  \cdot  x_{i_1}  
    \end{align*}
    From the above equation, we can show that matrix $B_{5,5,2}^{j_1,*,j_0,*}$ can be expressed as a rank-$1$ matrix,
\begin{align*}
     B_{5,5,2}^{j_1,*,j_0,*}  & ~ = f(A,x)_{j_1}^2 \cdot f(A,x)_{j_0}   \cdot c_g(A,x)^{\top} \cdot \diag(x) \cdot A^{\top} \cdot  f(A,x)  \cdot x \cdot {\bf 1}_d^{\top}\\ 
     & ~ =  f(A,x)_{j_1}^2 \cdot f(A,x)_{j_0}  \cdot c_g(A,x)^{\top} \cdot h(A,x)  \cdot x \cdot {\bf 1}_d^{\top}
\end{align*}
    where the last step is follows from the Definitions~\ref{def:h}.

        {\bf Proof of Part 13.}
We have for diagonal entry and off-diagonal entry can be written as follows
    \begin{align*}
         B_{5,6,1}^{j_1,i_1,j_0,i_1} = & ~  c_g(A,x)^{\top} \cdot f(A,x)_{j_1} \cdot f(A,x)_{j_0} \cdot \diag(x) \cdot A^{\top} \cdot  f(A,x)  \cdot  x_{i_1} \cdot \langle - f(A,x), f(A,x) \rangle\\
         B_{5,6,1}^{j_1,i_1,j_0,i_0} = & ~ c_g(A,x)^{\top} \cdot f(A,x)_{j_1} \cdot f(A,x)_{j_0} \cdot \diag(x) \cdot A^{\top} \cdot  f(A,x)  \cdot  x_{i_1} \cdot \langle - f(A,x), f(A,x) \rangle
    \end{align*}
    From the above equation, we can show that matrix $B_{5,6,1}^{j_1,*,j_0,*}$ can be expressed as a rank-$1$ matrix,
\begin{align*}
     B_{5,6,1}^{j_1,*,j_0,*}  & ~ = f(A,x)_{j_1} \cdot f(A,x)_{j_0} \cdot \langle - f(A,x), f(A,x) \rangle \cdot c_g(A,x)^{\top} \cdot \diag(x) \cdot A^{\top} \cdot  f(A,x)  \cdot x \cdot {\bf 1}_d^{\top}\\ 
     & ~ =  -  f(A,x)_{j_1} \cdot f(A,x)_{j_0} \cdot f_2(A,x) \cdot c_g(A,x)^{\top} \cdot h(A,x)  \cdot x \cdot {\bf 1}_d^{\top}
\end{align*}
    where the last step is follows from the Definitions~\ref{def:h} and Definitions~\ref{def:f_2}.

    {\bf Proof of Part 14.}
We have for diagonal entry and off-diagonal entry can be written as follows
    \begin{align*}
         B_{5,6,2}^{j_1,i_1,j_0,i_1} = & ~  c_g(A,x)^{\top} \cdot f(A,x)_{j_1}^2 \cdot f(A,x)_{j_0} \cdot \diag(x) \cdot A^{\top} \cdot  f(A,x)  \cdot  x_{i_1}   \\
         B_{5,6,2}^{j_1,i_1,j_0,i_0} = & ~c_g(A,x)^{\top} \cdot f(A,x)_{j_1}^2 \cdot f(A,x)_{j_0} \cdot \diag(x) \cdot A^{\top} \cdot  f(A,x)  \cdot  x_{i_1} 
    \end{align*}
    From the above equation, we can show that matrix $B_{5,6,2}^{j_1,*,j_0,*}$ can be expressed as a rank-$1$ matrix,
\begin{align*}
     B_{5,6,2}^{j_1,*,j_0,*}  & ~ = f(A,x)_{j_1}^2 \cdot f(A,x)_{j_0}   \cdot c_g(A,x)^{\top} \cdot \diag(x) \cdot A^{\top} \cdot  f(A,x)  \cdot x \cdot {\bf 1}_d^{\top}\\ 
     & ~ =  f(A,x)_{j_1}^2 \cdot f(A,x)_{j_0}  \cdot c_g(A,x)^{\top} \cdot h(A,x)  \cdot x \cdot {\bf 1}_d^{\top}
\end{align*}
    where the last step is follows from the Definitions~\ref{def:h}.

    {\bf Proof of Part 15.}
We have for diagonal entry and off-diagonal entry can be written as follows
    \begin{align*}
         B_{5,7,1}^{j_1,i_1,j_0,i_1} = & ~  f(A,x)_{j_1}\cdot f(A,x)_{j_0}^2  \cdot  x_{i_1}\cdot c_g(A,x)^{\top} \cdot \diag(x) A^{\top} \cdot  f(A,x)  \\
         B_{5,7,1}^{j_1,i_1,j_0,i_0} = & ~ f(A,x)_{j_1}\cdot f(A,x)_{j_0}^2  \cdot  x_{i_1}\cdot c_g(A,x)^{\top} \cdot \diag(x) A^{\top} \cdot  f(A,x) 
    \end{align*}
    From the above equation, we can show that matrix $B_{5,7,1}^{j_1,*,j_0,*}$ can be expressed as a rank-$1$ matrix,
\begin{align*}
     B_{5,7,1}^{j_1,*,j_0,*}  & ~ = f(A,x)_{j_1}\cdot f(A,x)_{j_0}^2    \cdot c_g(A,x)^{\top} \cdot \diag(x) \cdot A^{\top} \cdot  f(A,x)  \cdot x \cdot {\bf 1}_d^{\top}\\ 
     & ~ =  f(A,x)_{j_1}\cdot f(A,x)_{j_0}^2   \cdot c_g(A,x)^{\top} \cdot h(A,x)  \cdot x \cdot {\bf 1}_d^{\top}
\end{align*}
    where the last step is follows from the Definitions~\ref{def:h}.
\end{proof}
\end{lemma}

\subsection{Expanding \texorpdfstring{$B_5$}{} into many terms}
\begin{lemma}
   If the following conditions hold
    \begin{itemize}
     \item Let $u(A,x) \in \R^n$ be defined as Definition~\ref{def:u}
    \item Let $\alpha(A,x) \in \R$ be defined as Definition~\ref{def:alpha}
     \item Let $f(A,x) \in \R^n$ be defined as Definition~\ref{def:f}
    \item Let $c(A,x) \in \R^n$ be defined as Definition~\ref{def:c}
    \item Let $g(A,x) \in \R^d$ be defined as Definition~\ref{def:g} 
    \item Let $q(A,x) = c(A,x) + f(A,x) \in \R^n$
    \item Let $c_g(A,x) \in \R^d$ be defined as Definition~\ref{def:c_g}.
    \item Let $L_g(A,x) \in \R$ be defined as Definition~\ref{def:l_g}
    \item Let $v \in \R^n$ be a vector 
    \end{itemize}
Then, For $j_0,j_1 \in [n], i_0,i_1 \in [d]$, we have 
\begin{itemize}
    \item {\bf Part 1.}For $j_1 = j_0$ and $i_0 = i_1$
\begin{align*} 
   B_5^{j_1,i_1,j_1,i_1} = B_{5,1}^{j_1,i_1,j_1,i_1} +  B_{5,2}^{j_1,i_1,j_1,i_1} + B_{5,3}^{j_1,i_1,j_1,i_1} + B_{5,4}^{j_1,i_1,j_1,i_1} + B_{5,5}^{j_1,i_1,j_1,i_1} + B_{5,6}^{j_1,i_1,j_1,i_1} + B_{5,7}^{j_1,i_1,j_1,i_1}
\end{align*}
\item {\bf Part 2.}For $j_1 = j_0$ and $i_0 \neq i_1$
\begin{align*}
 B_5^{j_1,i_1,j_1,i_0} = B_{5,1}^{j_1,i_1,j_1,i_0} +  B_{5,2}^{j_1,i_1,j_1,i_0} + B_{5,3}^{j_1,i_1,j_1,i_0} + B_{5,4}^{j_1,i_1,j_1,i_0} + B_{5,5}^{j_1,i_1,j_1,i_0} + B_{5,6}^{j_1,i_1,j_1,i_0} + B_{5,7}^{j_1,i_1,j_1,i_0}
\end{align*}
\item {\bf Part 3.}For $j_1 \neq j_0$ and $i_0 = i_1$ \begin{align*}
   B_5^{j_1,i_1,j_0,i_1} = B_{5,1}^{j_1,i_1,j_0,i_1} +  B_{5,2}^{j_1,i_1,j_0,i_1} + B_{5,3}^{j_1,i_1,j_0,i_1} + B_{5,4}^{j_1,i_1,j_0,i_1} + B_{5,5}^{j_1,i_1,j_0,i_1} + B_{5,6}^{j_1,i_1,j_0,i_1} + B_{5,7}^{j_1,i_1,j_0,i_1}
\end{align*}
\item {\bf Part 4.} For $j_1 \neq j_0$ and $i_1 \neq i_0$
\begin{align*}
B_5^{j_1,i_1,j_0,i_0} = B_{5,1}^{j_1,i_1,j_0,i_0} +  B_{5,2}^{j_1,i_1,j_0,i_0} + B_{5,3}^{j_1,i_1,j_0,i_0} + B_{5,4}^{j_1,i_1,j_0,i_0} + B_{5,5}^{j_1,i_1,j_0,i_0} + B_{5,6}^{j_1,i_1,j_0,i_0} + B_{5,7}^{j_1,i_1,j_0,i_0}
\end{align*}
\end{itemize}
\end{lemma}
\begin{proof}
{\bf Proof of Part 1.} we have
    \begin{align*}
    B_{5,1}^{j_1,i_1,j_1,i_1} = & ~ \frac{\d}{\d A_{j_1,i_1}}(c_g(A,x)^{\top} \cdot f(A,x)_{j_1} \cdot \diag (x) A^{\top} \cdot  f(A,x)  \cdot    (\langle -f(A,x), f(A,x) \rangle + f(A,x)_{j_1}))\\
    & ~B_{5,1}^{j_1,i_1,j_1,i_1} +  B_{5,2}^{j_1,i_1,j_1,i_1} + B_{5,3}^{j_1,i_1,j_1,i_1} + B_{5,4}^{j_1,i_1,j_1,i_1} + B_{5,5}^{j_1,i_1,j_1,i_1} + B_{5,6}^{j_1,i_1,j_1,i_1} + B_{5,7}^{j_1,i_1,j_1,i_1}
\end{align*}
{\bf Proof of Part 2.} we have
    \begin{align*}
    B_{5,1}^{j_1,i_1,j_1,i_0} = & ~ \frac{\d}{\d A_{j_1,i_1}}(c_g(A,x)^{\top} \cdot f(A,x)_{j_1} \cdot \diag (x) A^{\top} \cdot  f(A,x)  \cdot    (\langle -f(A,x), f(A,x) \rangle + f(A,x)_{j_1}))\\
     & ~=B_{5,1}^{j_1,i_1,j_1,i_0} +  B_{5,2}^{j_1,i_1,j_1,i_0} + B_{5,3}^{j_1,i_1,j_1,i_0} + B_{5,4}^{j_1,i_1,j_1,i_0} + B_{5,5}^{j_1,i_1,j_1,i_0} + B_{5,6}^{j_1,i_1,j_1,i_0} + B_{5,7}^{j_1,i_1,j_1,i_0}
\end{align*}
{\bf Proof of Part 3.}  we have
\begin{align*}
        B_{5,1}^{j_1,i_1,j_0,i_1} = & ~ \frac{\d}{\d A_{j_1,i_1}}(c_g(A,x)^{\top} \cdot f(A,x)_{j_0} \cdot \diag (x) A^{\top} \cdot  f(A,x)  \cdot    (\langle -f(A,x), f(A,x) \rangle + f(A,x)_{j_0}))\\
   & ~= B_{5,1}^{j_1,i_1,j_0,i_1} +  B_{5,2}^{j_1,i_1,j_0,i_1} + B_{5,3}^{j_1,i_1,j_0,i_1} + B_{5,4}^{j_1,i_1,j_0,i_1} + B_{5,5}^{j_1,i_1,j_0,i_1} + B_{5,6}^{j_1,i_1,j_0,i_1} + B_{5,7}^{j_1,i_1,j_0,i_1}
\end{align*}
{\bf Proof of Part 4.}
 we have
\begin{align*}
  B_{5,1}^{j_1,i_1,j_0,i_0} = & ~ \frac{\d}{\d A_{j_1,i_1}}(c_g(A,x)^{\top} \cdot f(A,x)_{j_0} \cdot \diag (x) A^{\top} \cdot  f(A,x)  \cdot    (\langle -f(A,x), f(A,x) \rangle + f(A,x)_{j_0}))\\
   & ~= B_{5,1}^{j_1,i_1,j_0,i_0} +  B_{5,2}^{j_1,i_1,j_0,i_0} + B_{5,3}^{j_1,i_1,j_0,i_0} + B_{5,4}^{j_1,i_1,j_0,i_0} + B_{5,5}^{j_1,i_1,j_0,i_0} + B_{5,6}^{j_1,i_1,j_0,i_0} + B_{5,7}^{j_1,i_1,j_0,i_0}
\end{align*}
\end{proof}
\subsection{Lipschitz Computation}
\begin{lemma}\label{lips: B_5}
If the following conditions hold
\begin{itemize}
    \item Let $B_{5,1,1}^{j_1,*, j_0,*}, \cdots, B_{5,7,1}^{j_1,*, j_0,*} $ be defined as Lemma~\ref{lem:b_5_j1_j0} 
    \item  Let $\|A \|_2 \leq R, \|A^{\top} \|_F \leq R, \| x\|_2 \leq R, \|\diag(f(A,x)) \|_F \leq \|f(A,x) \|_2 \leq 1, \| b_g \|_2 \leq 1$ 
\end{itemize}
Then, we have
\begin{itemize}
    \item {\bf Part 1.}
    \begin{align*}
       \| B_{5,1,1}^{j_1,*,j_0,*} (A) - B_{5,1,1}^{j_1,*,j_0,*} ( \wt{A} ) \|_F \leq  \beta^{-2} \cdot n \cdot \sqrt{d}\exp(5R^2) \cdot \|A - \wt{A}\|_F
    \end{align*}
     \item {\bf Part 2.}
    \begin{align*}
       \| B_{5,1,2}^{j_1,*,j_0,*} (A) - B_{5,1,2}^{j_1,*,j_0,*} ( \wt{A} ) \|_F \leq   \beta^{-2} \cdot n \cdot \sqrt{d}\exp(5R^2) \cdot \|A - \wt{A}\|_F
    \end{align*}
     \item {\bf Part 3.}
    \begin{align*}
       \| B_{5,1,3}^{j_1,*,j_0,*} (A) - B_{5,1,3}^{j_1,*,j_0,*} ( \wt{A} ) \|_F \leq   \beta^{-2} \cdot n \cdot \sqrt{d}\exp(6R^2) \cdot \|A - \wt{A}\|_F
    \end{align*}
     \item {\bf Part 4.}
    \begin{align*}
       \| B_{5,1,4}^{j_1,*,j_0,*} (A) - B_{5,1,4}^{j_1,*,j_0,*} ( \wt{A} ) \|_F \leq   \beta^{-2} \cdot n \cdot \sqrt{d}\exp(6R^2) \cdot \|A - \wt{A}\|_F
    \end{align*}
     \item {\bf Part 5.}
    \begin{align*}
       \| B_{5,1,5}^{j_1,*,j_0,*} (A) - B_{5,1,5}^{j_1,*,j_0,*} ( \wt{A} ) \|_F \leq   \beta^{-2} \cdot n \cdot \sqrt{d}\exp(6R^2) \cdot \|A - \wt{A}\|_F
    \end{align*}
     \item {\bf Part 6.}
    \begin{align*}
       \| B_{5,1,6}^{j_1,*,j_0,*} (A) - B_{5,1,6}^{j_1,*,j_0,*} ( \wt{A} ) \|_F \leq  \beta^{-2} \cdot n \cdot \sqrt{d}\exp(6R^2) \cdot \|A - \wt{A}\|_F
    \end{align*}
     \item {\bf Part 7.}
    \begin{align*}
       \| B_{5,1,7}^{j_1,*,j_0,*} (A) - B_{5,1,7}^{j_1,*,j_0,*} ( \wt{A} ) \|_F \leq   \beta^{-2} \cdot n \cdot \sqrt{d}\exp(6R^2) \cdot \|A - \wt{A}\|_F
    \end{align*}
    \item {\bf Part 8.}
    \begin{align*}
       \| B_{5,2,1}^{j_1,*,j_0,*} (A) - B_{5,2,1}^{j_1,*,j_0,*} ( \wt{A} ) \|_F \leq  \beta^{-2} \cdot n \cdot \sqrt{d}\exp(6R^2) \cdot \|A - \wt{A}\|_F
    \end{align*}
     \item {\bf Part 9.}
    \begin{align*}
       \| B_{5,3,1}^{j_1,*,j_0,*} (A) - B_{5,3,1}^{j_1,*,j_0,*} ( \wt{A} ) \|_F \leq  \beta^{-2} \cdot n \cdot \sqrt{d}  \exp(5R^2)\|A - \wt{A}\|_F  
    \end{align*}
     \item {\bf Part 10.}
    \begin{align*}
       \| B_{5,4,1}^{j_1,*,j_0,*} (A) - B_{5,4,1}^{j_1,*,j_0,*} ( \wt{A} ) \|_F \leq  \beta^{-2} \cdot n \cdot \sqrt{d}  \exp(6R^2)\|A - \wt{A}\|_F  
    \end{align*}
     \item {\bf Part 11.}
    \begin{align*}
       \| B_{5,5,1}^{j_1,*,j_0,*} (A) - B_{5,5,1}^{j_1,*,j_0,*} ( \wt{A} ) \|_F \leq \beta^{-2} \cdot n \cdot \sqrt{d}  \exp(6R^2)\|A - \wt{A}\|_F   
    \end{align*}
     \item {\bf Part 12.}
    \begin{align*}
       \| B_{5,5,2}^{j_1,*,j_0,*} (A) - B_{5,5,2}^{j_1,*,j_0,*} ( \wt{A} ) \|_F \leq  \beta^{-2} \cdot n \cdot \sqrt{d}  \exp(6R^2)\|A - \wt{A}\|_F  
    \end{align*}
     \item {\bf Part 13.}
    \begin{align*}
       \| B_{5,6,1}^{j_1,*,j_0,*} (A) - B_{5,6,1}^{j_1,*,j_0,*} ( \wt{A} ) \|_F \leq    \beta^{-2} \cdot n \cdot \sqrt{d}  \exp(6R^2)\|A - \wt{A}\|_F 
    \end{align*}
         \item {\bf Part 14.}
    \begin{align*}
       \| B_{5,6,2}^{j_1,*,j_0,*} (A) - B_{5,6,2}^{j_1,*,j_0,*} ( \wt{A} ) \|_F \leq  \beta^{-2} \cdot n \cdot \sqrt{d}  \exp(6R^2)\|A - \wt{A}\|_F   
    \end{align*}
             \item {\bf Part 15.}
    \begin{align*}
       \| B_{5,7,1}^{j_1,*,j_0,*} (A) - B_{5,7,1}^{j_1,*,j_0,*} ( \wt{A} ) \|_F \leq  \beta^{-2} \cdot n \cdot \sqrt{d}  \exp(6R^2)\|A - \wt{A}\|_F   
    \end{align*}
        \item{\bf Part 16.}
    \begin{align*}
         \| B_{5}^{j_1,*,j_0,*} (A) - B_{5}^{j_1,*,j_0,*} ( \wt{A} ) \|_F \leq & ~   15\beta^{-2} \cdot n \cdot \sqrt{d}  \exp(6R^2)\|A - \wt{A}\|_F   
    \end{align*}
    \end{itemize}
\end{lemma}
\begin{proof}
{\bf Proof of Part 1.}
\begin{align*}
& ~ \| B_{5,1,1}^{j_1,*,j_0,*} (A) - B_{5,1,1}^{j_1,*,j_0,*} ( \wt{A} ) \| \\ \leq 
    & ~ \| f_c(A,x)  \cdot f(A,x)_{j_1} \cdot f(A,x)_{j_0} \cdot (-f_2(A,x) + f(A,x)_{j_0}) \cdot  h(A,x) \cdot  {\bf 1}_d^\top \\
    & ~ - f_c(\wt{A},x)  \cdot f(\wt{A},x)_{j_1} \cdot f(\wt{A},x)_{j_0} \cdot  (-f_2(\wt{A},x) + f(\wt{A},x)_{j_0}) \cdot h(\wt{A},x) \cdot  {\bf 1}_d^\top \|_F\\
    \leq & ~ |  f_c(A,x) - f_c(\wt{A},x) | \cdot |f(A,x)_{j_1}|\cdot | f(A,x)_{j_0} | \cdot   ( |-f_2(A,x)| +  |f(A,x)_{j_0}|) \cdot \| h(A,x)\|_2  \cdot \| {\bf 1}_d^\top\|_2 \\
    & + ~   |f_c(\wt{A},x)| \cdot  |f(A,x)_{j_1} - f(\wt{A},x)_{j_1}|\cdot | f(A,x)_{j_0} | \cdot  ( |-f_2(A,x)| +  |f(A,x)_{j_0}|) \cdot  \| h(A,x)\|_2  \cdot \| {\bf 1}_d^\top\|_2 \\
    & + ~   |f_c(\wt{A},x)| \cdot  |f(\wt{A},x)_{j_1}|\cdot | f(A,x)_{j_0} -  f(\wt{A},x)_{j_0} | \cdot ( |-f_2(A,x)| +  |f(A,x)_{j_0}|) \cdot   \| h(A,x)\|_2  \cdot \| {\bf 1}_d^\top\|_2 \\
     & + ~   |f_c(\wt{A},x)| \cdot  |f(\wt{A},x)_{j_1}|\cdot | f(\wt{A},x)_{j_0} | \cdot ( | f_2(A,x) -  f_2(\wt{A},x)| +  |f(A,x)_{j_0} - f(\wt{A},x)_{j_0}|) \cdot   \| h(A,x)\|_2  \cdot \| {\bf 1}_d^\top\|_2 \\
    & + ~ |f_c(\wt{A},x)|\cdot   |f(\wt{A},x)_{j_1}|\cdot  | f(\wt{A},x)_{j_0} | \cdot ( |-f_2(\wt{A},x)| +  |f(\wt{A},x)_{j_0}|) \cdot \| h(A,x) - h(\wt{A},x)\|_2  \cdot \| {\bf 1}_d^\top\|_2 \\
    \leq & ~ 12R^2  \beta^{-2} \cdot n \cdot \sqrt{d}  \exp(3R^2)\|A - \wt{A}\|_F \\
    &+ ~ 8R^2  \beta^{-2} \cdot n \cdot \sqrt{d}  \exp(3R^2)\|A - \wt{A}\|_F \\
    & + ~ 8R^2 \beta^{-2} \cdot n \cdot \sqrt{d}\exp(3R^2) \cdot \|A - \wt{A}\|_F \\
    & + ~ 12R^2\beta^{-2} \cdot n \cdot \sqrt{d}\exp(3R^2) \cdot \|A - \wt{A}\|_F \\
    & + ~ 12 \beta^{-2} \cdot n \cdot \sqrt{d}\exp(4R^2) \cdot \|A - \wt{A}\|_F\\
    \leq &  ~ 52 \beta^{-2} \cdot n \cdot \sqrt{d}\exp(4R^2) \cdot \|A - \wt{A}\|_F \\
    \leq &  ~ \beta^{-2} \cdot n \cdot \sqrt{d}\exp(5R^2) \cdot \|A - \wt{A}\|_F 
\end{align*}

{\bf Proof of Part 2.}
\begin{align*}
& ~ \| B_{5,1,2}^{j_1,*,j_0,*} (A) - B_{5,1,2}^{j_1,*,j_0,*} ( \wt{A} ) \| \\ \leq 
    & ~ \|c(A,x)_{j_1}  \cdot f(A,x)_{j_1} \cdot f(A,x)_{j_0} \cdot (-f_2(A,x) + f(A,x)_{j_0}) \cdot  h(A,x) \cdot  {\bf 1}_d^\top \\
    & ~ - c(\wt{A},x)_{j_1} \cdot f(\wt{A},x)_{j_1} \cdot f(\wt{A},x)_{j_0} \cdot  (-f_2(\wt{A},x) + f(\wt{A},x)_{j_0}) \cdot h(\wt{A},x) \cdot  {\bf 1}_d^\top \|_F\\
    \leq & ~ |  c(A,x)_{j_1} - c(\wt{A},x)_{j_1} | \cdot |f(A,x)_{j_1}|\cdot | f(A,x)_{j_0} | \cdot   ( |-f_2(A,x)| +  |f(A,x)_{j_0}|) \cdot \| h(A,x)\|_2  \cdot \| {\bf 1}_d^\top\|_2 \\
    & + ~   | c(\wt{A},x)_{j_1}| \cdot  |f(A,x)_{j_1} - f(\wt{A},x)_{j_1}|\cdot | f(A,x)_{j_0} | \cdot  ( |-f_2(A,x)| +  |f(A,x)_{j_0}|) \cdot  \| h(A,x)\|_2  \cdot \| {\bf 1}_d^\top\|_2 \\
    & + ~   | c(\wt{A},x)_{j_1}| \cdot  |f(\wt{A},x)_{j_1}|\cdot | f(A,x)_{j_0} -  f(\wt{A},x)_{j_0} | \cdot ( |-f_2(A,x)| +  |f(A,x)_{j_0}|) \cdot   \| h(A,x)\|_2  \cdot \| {\bf 1}_d^\top\|_2 \\
     & + ~   | c(\wt{A},x)_{j_1}| \cdot  |f(\wt{A},x)_{j_1}|\cdot | f(\wt{A},x)_{j_0} | \cdot ( | f_2(A,x) -  f_2(\wt{A},x)| +  |f(A,x)_{j_0} - f(\wt{A},x)_{j_0}|) \cdot   \| h(A,x)\|_2  \cdot \| {\bf 1}_d^\top\|_2 \\
    & + ~ | c(\wt{A},x)_{j_1}|\cdot   |f(\wt{A},x)_{j_1}|\cdot  | f(\wt{A},x)_{j_0} | \cdot ( |-f_2(\wt{A},x)| +  |f(\wt{A},x)_{j_0}|) \cdot \| h(A,x) - h(\wt{A},x)\|_2  \cdot \| {\bf 1}_d^\top\|_2 \\
    \leq & ~ 4R^2  \beta^{-2} \cdot n \cdot \sqrt{d}  \exp(3R^2)\|A - \wt{A}\|_F \\
    &+ ~ 8R^2  \beta^{-2} \cdot n \cdot \sqrt{d}  \exp(3R^2)\|A - \wt{A}\|_F \\
    & + ~ 8R^2 \beta^{-2} \cdot n \cdot \sqrt{d}\exp(3R^2) \cdot \|A - \wt{A}\|_F \\
    & + ~ 12R^2\beta^{-2} \cdot n \cdot \sqrt{d}\exp(3R^2) \cdot \|A - \wt{A}\|_F \\
    & + ~ 12 \beta^{-2} \cdot n \cdot \sqrt{d}\exp(4R^2) \cdot \|A - \wt{A}\|_F\\
    \leq &  ~ 44 \beta^{-2} \cdot n \cdot \sqrt{d}\exp(4R^2) \cdot \|A - \wt{A}\|_F \\
    \leq &  ~ \beta^{-2} \cdot n \cdot \sqrt{d}\exp(5R^2) \cdot \|A - \wt{A}\|_F 
\end{align*}

{\bf Proof of Part 3.}
\begin{align*}
& ~ \| B_{5,1,3}^{j_1,*,j_0,*} (A) - B_{5,1,3}^{j_1,*,j_0,*} ( \wt{A} ) \| \\ \leq 
    & ~ \| f_c(A,x)  \cdot f(A,x)_{j_1} \cdot f(A,x)_{j_0} \cdot (-f_2(A,x) + f(A,x)_{j_0}) \cdot ( (A_{j_1,*}) \circ x^\top  )  \cdot h(A,x) \cdot  I_d  \\
    & ~ - f_c(\wt{A},x)  \cdot f(\wt{A},x)_{j_1} \cdot f(\wt{A},x)_{j_0} \cdot  (-f_2(\wt{A},x) + f(\wt{A},x)_{j_0}) \cdot( (\wt{A}_{j_1,*}) \circ x^\top  )  \cdot h(\wt{A},x) \cdot  I_d  \|_F\\
    \leq & ~ |  f_c(A,x) - f_c(\wt{A},x) | \cdot |f(A,x)_{j_1}|\cdot | f(A,x)_{j_0} | \cdot   ( |-f_2(A,x)| +  |f(A,x)_{j_0}|)\\
    & ~ \cdot \| A_{j_1,*} \|_2 \cdot \| \diag(x) \|_F  \cdot \| h(A,x)\|_2  \cdot \|I_d \|_F \\
    & + ~   |f_c(\wt{A},x)| \cdot  |f(A,x)_{j_1} - f(\wt{A},x)_{j_1}|\cdot | f(A,x)_{j_0} | \cdot  ( |-f_2(A,x)| +  |f(A,x)_{j_0}|) \\
    & ~\cdot \| A_{j_1,*} \|_2 \cdot \| \diag(x) \|_F  \cdot \| h(A,x)\|_2  \cdot \|I_d \|_F  \\
    & + ~   |f_c(\wt{A},x)| \cdot  |f(\wt{A},x)_{j_1}|\cdot | f(A,x)_{j_0} -  f(\wt{A},x)_{j_0} | \cdot ( |-f_2(A,x)| +  |f(A,x)_{j_0}|) \\
    & ~\cdot \| A_{j_1,*} \|_2 \cdot \| \diag(x) \|_F  \cdot  \| h(A,x)\|_2  \cdot \|I_d \|_F  \\
     & + ~   |f_c(\wt{A},x)| \cdot  |f(\wt{A},x)_{j_1}|\cdot | f(\wt{A},x)_{j_0} | \cdot ( | f_2(A,x) -  f_2(\wt{A},x)| +  |f(A,x)_{j_0} - f(\wt{A},x)_{j_0}|)\\
    & ~ \cdot \| A_{j_1,*} \|_2 \cdot \| \diag(x) \|_F  \cdot  \| h(A,x)\|_2  \cdot \|I_d \|_F  \\
       & + ~ |f_c(\wt{A},x)|\cdot   |f(\wt{A},x)_{j_1}|\cdot  | f(\wt{A},x)_{j_0} | \cdot ( |-f_2(\wt{A},x)| +  |f(\wt{A},x)_{j_0}|)\\
    & ~ \cdot \|  A_{j_1,*} - \wt{A}_{j_1,*} \|_2 \cdot \| \diag(x) \|_F  \cdot \| h(A,x) \|_2  \cdot \|I_d \|_F  \\
    & + ~ |f_c(\wt{A},x)|\cdot   |f(\wt{A},x)_{j_1}|\cdot  | f(\wt{A},x)_{j_0} | \cdot ( |-f_2(\wt{A},x)| +  |f(\wt{A},x)_{j_0}|) \\
    & ~\cdot \| \wt{A}_{j_1,*} \|_2 \cdot \| \diag(x) \|_F  \cdot \| h(A,x) - h(\wt{A},x)\|_2  \cdot \|I_d \|_F  \\
    \leq & ~ 12R^4  \beta^{-2} \cdot n \cdot \sqrt{d}  \exp(3R^2)\|A - \wt{A}\|_F \\
    &+ ~ 8R^4  \beta^{-2} \cdot n \cdot \sqrt{d}  \exp(3R^2)\|A - \wt{A}\|_F \\
    & + ~ 8R^4 \beta^{-2} \cdot n \cdot \sqrt{d}\exp(3R^2) \cdot \|A - \wt{A}\|_F \\
    & + ~ 12R^4\beta^{-2} \cdot n \cdot \sqrt{d}\exp(3R^2) \cdot \|A - \wt{A}\|_F \\
    & + ~ 4R^3  \sqrt{d} \cdot \|A - \wt{A}\|_F \\
    & + ~ 12R^2 \beta^{-2} \cdot n \cdot \sqrt{d}\exp(4R^2) \cdot \|A - \wt{A}\|_F\\
    \leq &  ~ 56 \beta^{-2} \cdot n \cdot \sqrt{d}\exp(5R^2) \cdot \|A - \wt{A}\|_F \\
    \leq &  ~ \beta^{-2} \cdot n \cdot \sqrt{d}\exp(6R^2) \cdot \|A - \wt{A}\|_F 
\end{align*}

{\bf Proof of Part 4.}
\begin{align*}
& ~ \| B_{5,1,4}^{j_1,*,j_0,*} (A) - B_{5,1,4}^{j_1,*,j_0,*} ( \wt{A} ) \| \\ 
\leq  & ~ \|- f_c(A,x)  \cdot f(A,x)_{j_1} \cdot f(A,x)_{j_0} \cdot (-f_2(A,x) + f(A,x)_{j_0}) \cdot h(A,x)^\top  \cdot h(A,x) \cdot  I_d \\
    & ~ - (-f_c(\wt{A},x)  \cdot f(\wt{A},x)_{j_1} \cdot f(\wt{A},x)_{j_0} \cdot  (-f_2(\wt{A},x) + f(\wt{A},x)_{j_0}) \cdot h(\wt{A},x)^\top \cdot h(\wt{A},x) \cdot  I_d \|_F)\\
\leq  & ~ \| f_c(A,x)  \cdot f(A,x)_{j_1} \cdot f(A,x)_{j_0} \cdot (-f_2(A,x) + f(A,x)_{j_0}) \cdot h(A,x)^\top \cdot h(A,x) \cdot  I_d \\
    & ~ - f_c(\wt{A},x)  \cdot f(\wt{A},x)_{j_1} \cdot f(\wt{A},x)_{j_0} \cdot  (-f_2(\wt{A},x) + f(\wt{A},x)_{j_0}) \cdot h(\wt{A},x)^\top  \cdot h(\wt{A},x) \cdot I_d \|_F\\
    \leq & ~ |  f_c(A,x) - f_c(\wt{A},x) | \cdot |f(A,x)_{j_1}|\cdot | f(A,x)_{j_0} | \cdot   ( |-f_2(A,x)| +  |f(A,x)_{j_0}|)\\
    & ~ \cdot \| h(A,x)^\top\|_2   \cdot \| h(A,x)\|_2  \cdot \|I_d \|_F  \\
    & + ~   |f_c(\wt{A},x)| \cdot  |f(A,x)_{j_1} - f(\wt{A},x)_{j_1}|\cdot | f(A,x)_{j_0} | \cdot  ( |-f_2(A,x)| +  |f(A,x)_{j_0}|) \\
    & ~\cdot \| h(A,x)^\top\|_2 \cdot \| h(A,x)\|_2  \cdot \|I_d \|_F \\
    & + ~   |f_c(\wt{A},x)| \cdot  |f(\wt{A},x)_{j_1}|\cdot | f(A,x)_{j_0} -  f(\wt{A},x)_{j_0} | \cdot ( |-f_2(A,x)| +  |f(A,x)_{j_0}|) \\
    & ~\cdot \| h(A,x)^\top\|_2  \cdot  \| h(A,x)\|_2  \cdot \|I_d \|_F \\
     & + ~   |f_c(\wt{A},x)| \cdot  |f(\wt{A},x)_{j_1}|\cdot | f(\wt{A},x)_{j_0} | \cdot ( | f_2(A,x) -  f_2(\wt{A},x)| +  |f(A,x)_{j_0} - f(\wt{A},x)_{j_0}|)\\
    & ~ \cdot \| h(A,x)^\top\|_2  \cdot  \| h(A,x)\|_2  \cdot\|I_d \|_F  \\
       & + ~ |f_c(\wt{A},x)|\cdot   |f(\wt{A},x)_{j_1}|\cdot  | f(\wt{A},x)_{j_0} | \cdot ( |-f_2(\wt{A},x)| +  |f(\wt{A},x)_{j_0}|)\\
    & ~ \cdot \| h(A,x)^\top -  h(\wt{A},x)^\top\|_2 \cdot \| h(A,x) \|_2  \cdot \|I_d \|_F  \\
    & + ~ |f_c(\wt{A},x)|\cdot   |f(\wt{A},x)_{j_1}|\cdot  | f(\wt{A},x)_{j_0} | \cdot ( |-f_2(\wt{A},x)| +  |f(\wt{A},x)_{j_0}|) \\
    & ~\cdot \| h(\wt{A},x)^\top\|_2  \cdot \| h(A,x) - h(\wt{A},x)\|_2  \cdot \|I_d \|_F  \\
    \leq & ~ 12R^4  \beta^{-2} \cdot n \cdot \sqrt{d}  \exp(3R^2)\|A - \wt{A}\|_F \\
    &+ ~ 8R^4  \beta^{-2} \cdot n \cdot \sqrt{d}  \exp(3R^2)\|A - \wt{A}\|_F \\
    & + ~ 8R^4 \beta^{-2} \cdot n \cdot \sqrt{d}\exp(3R^2) \cdot \|A - \wt{A}\|_F \\
    & + ~ 12R^4\beta^{-2} \cdot n \cdot \sqrt{d}\exp(3R^2) \cdot \|A - \wt{A}\|_F \\
    & + ~ 12R^2 \beta^{-2} \cdot n \cdot \sqrt{d}\exp(4R^2) \cdot \|A - \wt{A}\|_F\\
    & + ~ 12R^2 \beta^{-2} \cdot n \cdot \sqrt{d}\exp(4R^2) \cdot \|A - \wt{A}\|_F\\
    \leq &  ~ 64 \beta^{-2} \cdot n \cdot \sqrt{d}\exp(5R^2) \cdot \|A - \wt{A}\|_F \\
    \leq &  ~ \beta^{-2} \cdot n \cdot \sqrt{d}\exp(6R^2) \cdot \|A - \wt{A}\|_F 
\end{align*}

{\bf Proof of Part 5.}
\begin{align*}
& ~ \| B_{5,1,5}^{j_1,*,j_0,*} (A) - B_{5,1,5}^{j_1,*,j_0,*} ( \wt{A} ) \| \\ 
\leq  & ~ \| (-f_2(A,x) + f(A,x)_{j_1})  \cdot f(A,x)_{j_1} \cdot f(A,x)_{j_0} \cdot (-f_2(A,x) + f(A,x)_{j_0}) \cdot h(A,x)^\top  \cdot h(A,x) \cdot  I_d \\
    & ~ - (-f_2(\wt{A},x) + f(\wt{A},x)_{j_1})  \cdot f(\wt{A},x)_{j_1} \cdot f(\wt{A},x)_{j_0} \cdot  (-f_2(\wt{A},x) + f(\wt{A},x)_{j_0}) \cdot h(\wt{A},x)^\top \cdot h(\wt{A},x) \cdot  I_d \|_F\\
    \leq & ~ (\|f_2(A,x) - f_2(\wt{A},x)\| + \|f(A,x)_{j_1}- f(\wt{A},x)_{j_1}\|) \cdot |f(A,x)_{j_1}|\cdot | f(A,x)_{j_0} | \cdot   ( |-f_2(A,x)| +  |f(A,x)_{j_0}|)\\
    & ~ \cdot \| h(A,x)^\top\|_2   \cdot \| h(A,x)\|_2  \cdot \|I_d \|_F \\
    & + ~   (\|  f_2(\wt{A},x)\| + \|  f(\wt{A},x)_{j_1}\|) \cdot  |f(A,x)_{j_1} - f(\wt{A},x)_{j_1}|\cdot | f(A,x)_{j_0} | \cdot  ( |-f_2(A,x)| +  |f(A,x)_{j_0}|) \\
    & ~\cdot \| h(A,x)^\top\|_2 \cdot \| h(A,x)\|_2  \cdot \|I_d \|_F \\
    & + ~   (\|  f_2(\wt{A},x)\| + \|  f(\wt{A},x)_{j_1}\|) \cdot  |f(\wt{A},x)_{j_1}|\cdot | f(A,x)_{j_0} -  f(\wt{A},x)_{j_0} | \cdot ( |-f_2(A,x)| +  |f(A,x)_{j_0}|) \\
    & ~\cdot \| h(A,x)^\top\|_2  \cdot  \| h(A,x)\|_2  \cdot \|I_d \|_F \\
     & + ~   (\|  f_2(\wt{A},x)\| + \|  f(\wt{A},x)_{j_1}\|)  \cdot  |f(\wt{A},x)_{j_1}|\cdot | f(\wt{A},x)_{j_0} | \cdot ( | f_2(A,x) -  f_2(\wt{A},x)| +  |f(A,x)_{j_0} - f(\wt{A},x)_{j_0}|)\\
    & ~ \cdot \| h(A,x)^\top\|_2  \cdot  \| h(A,x)\|_2  \cdot \|I_d \|_F \\
       & + ~ (\|  f_2(\wt{A},x)\| + \|  f(\wt{A},x)_{j_1}\|)  \cdot   |f(\wt{A},x)_{j_1}|\cdot  | f(\wt{A},x)_{j_0} | \cdot ( |-f_2(\wt{A},x)| +  |f(\wt{A},x)_{j_0}|)\\
    & ~ \cdot \| h(A,x)^\top -  h(\wt{A},x)^\top\|_2 \cdot \| h(A,x) \|_2  \cdot \|I_d \|_F \\
    & + ~(\|  f_2(\wt{A},x)\| + \|  f(\wt{A},x)_{j_1}\|) \cdot   |f(\wt{A},x)_{j_1}|\cdot  | f(\wt{A},x)_{j_0} | \cdot ( |-f_2(\wt{A},x)| +  |f(\wt{A},x)_{j_0}|) \\
    & ~\cdot \| h(\wt{A},x)^\top\|_2  \cdot \| h(A,x) - h(\wt{A},x)\|_2  \cdot \|I_d \|_F \\
    \leq & ~ 12R^4  \beta^{-2} \cdot n \cdot \sqrt{d}  \exp(3R^2)\|A - \wt{A}\|_F \\
    &+ ~ 8R^4  \beta^{-2} \cdot n \cdot \sqrt{d}  \exp(3R^2)\|A - \wt{A}\|_F \\
    & + ~ 8R^4 \beta^{-2} \cdot n \cdot \sqrt{d}\exp(3R^2) \cdot \|A - \wt{A}\|_F \\
    & + ~ 12R^4\beta^{-2} \cdot n \cdot \sqrt{d}\exp(3R^2) \cdot \|A - \wt{A}\|_F \\
    & + ~ 12R^2 \beta^{-2} \cdot n \cdot \sqrt{d}\exp(4R^2) \cdot \|A - \wt{A}\|_F\\
    & + ~ 12R^2 \beta^{-2} \cdot n \cdot \sqrt{d}\exp(4R^2) \cdot \|A - \wt{A}\|_F\\
    \leq &  ~ 64 \beta^{-2} \cdot n \cdot \sqrt{d}\exp(5R^2) \cdot \|A - \wt{A}\|_F \\
    \leq &  ~ \beta^{-2} \cdot n \cdot \sqrt{d}\exp(6R^2) \cdot \|A - \wt{A}\|_F 
\end{align*}

{\bf Proof of Part 6.}
\begin{align*}
& ~ \| B_{5,1,6}^{j_1,*,j_0,*} (A) - B_{5,1,6}^{j_1,*,j_0,*} ( \wt{A} ) \| \\ 
\leq  & ~ \| (-f_c(A,x) + f(A,x)_{j_1})  \cdot f(A,x)_{j_1} \cdot f(A,x)_{j_0} \cdot (-f_2(A,x) + f(A,x)_{j_0}) \cdot h(A,x)^\top  \cdot h(A,x) \cdot  I_d \\
    & ~ - (-f_c(\wt{A},x) + f(\wt{A},x)_{j_1})  \cdot f(\wt{A},x)_{j_1} \cdot f(\wt{A},x)_{j_0} \cdot  (-f_2(\wt{A},x) + f(\wt{A},x)_{j_0}) \cdot h(\wt{A},x)^\top \cdot h(\wt{A},x) \cdot I_d \|_F\\
    \leq & ~ (\|f_c(A,x) - f_c(\wt{A},x)\| + \|f(A,x)_{j_1}- f(\wt{A},x)_{j_1}\|) \cdot |f(A,x)_{j_1}|\cdot | f(A,x)_{j_0} | \cdot   ( |-f_2(A,x)| +  |f(A,x)_{j_0}|)\\
    & ~ \cdot \| h(A,x)^\top\|_2   \cdot \| h(A,x)\|_2  \cdot \|I_d\|_F \\
    & + ~   (\|  f_c(\wt{A},x)\| + \|  f(\wt{A},x)_{j_1}\|) \cdot  |f(A,x)_{j_1} - f(\wt{A},x)_{j_1}|\cdot | f(A,x)_{j_0} | \cdot  ( |-f_2(A,x)| +  |f(A,x)_{j_0}|) \\
    & ~\cdot \| h(A,x)^\top\|_2 \cdot \| h(A,x)\|_2  \cdot \|I_d\|_F \\
    & + ~   (\|  f_c(\wt{A},x)\| + \|  f(\wt{A},x)_{j_1}\|) \cdot  |f(\wt{A},x)_{j_1}|\cdot | f(A,x)_{j_0} -  f(\wt{A},x)_{j_0} | \cdot ( |-f_2(A,x)| +  |f(A,x)_{j_0}|) \\
    & ~\cdot \| h(A,x)^\top\|_2  \cdot  \| h(A,x)\|_2  \cdot \|I_d\|_F \\
     & + ~   (\|  f_c(\wt{A},x)\| + \|  f(\wt{A},x)_{j_1}\|)  \cdot  |f(\wt{A},x)_{j_1}|\cdot | f(\wt{A},x)_{j_0} | \cdot ( | f_2(A,x) -  f_2(\wt{A},x)| +  |f(A,x)_{j_0} - f(\wt{A},x)_{j_0}|)\\
    & ~ \cdot \| h(A,x)^\top\|_2  \cdot  \| h(A,x)\|_2  \cdot \|I_d\|_F \\
       & + ~ (\|  f_c(\wt{A},x)\| + \|  f(\wt{A},x)_{j_1}\|)  \cdot   |f(\wt{A},x)_{j_1}|\cdot  | f(\wt{A},x)_{j_0} | \cdot ( |-f_2(\wt{A},x)| +  |f(\wt{A},x)_{j_0}|)\\
    & ~ \cdot \| h(A,x)^\top -  h(\wt{A},x)^\top\|_2 \cdot \| h(A,x) \|_2  \cdot \|I_d\|_F \\
    & + ~(\|  f_c(\wt{A},x)\| + \|  f(\wt{A},x)_{j_1}\|) \cdot   |f(\wt{A},x)_{j_1}|\cdot  | f(\wt{A},x)_{j_0} | \cdot ( |-f_2(\wt{A},x)| +  |f(\wt{A},x)_{j_0}|) \\
    & ~\cdot \| h(\wt{A},x)^\top\|_2  \cdot \| h(A,x) - h(\wt{A},x)\|_2  \cdot \|I_d\|_F \\
    \leq & ~ 16R^4  \beta^{-2} \cdot n \cdot \sqrt{d}  \exp(3R^2)\|A - \wt{A}\|_F \\
    &+ ~ 12R^4  \beta^{-2} \cdot n \cdot \sqrt{d}  \exp(3R^2)\|A - \wt{A}\|_F \\
    & + ~ 12R^4 \beta^{-2} \cdot n \cdot \sqrt{d}\exp(3R^2) \cdot \|A - \wt{A}\|_F \\
    & + ~ 18R^4\beta^{-2} \cdot n \cdot \sqrt{d}\exp(3R^2) \cdot \|A - \wt{A}\|_F \\
    & + ~ 18R^2 \beta^{-2} \cdot n \cdot \sqrt{d}\exp(4R^2) \cdot \|A - \wt{A}\|_F\\
    & + ~ 18R^2 \beta^{-2} \cdot n \cdot \sqrt{d}\exp(4R^2) \cdot \|A - \wt{A}\|_F\\
    \leq &  ~ 94 \beta^{-2} \cdot n \cdot \sqrt{d}\exp(5R^2) \cdot \|A - \wt{A}\|_F \\
    \leq &  ~ \beta^{-2} \cdot n \cdot \sqrt{d}\exp(6R^2) \cdot \|A - \wt{A}\|_F 
\end{align*}

{\bf Proof of Part 7.}
\begin{align*}
& ~ \| B_{5,1,7}^{j_1,*,j_0,*} (A) - B_{5,1,7}^{j_1,*,j_0,*} ( \wt{A} ) \| \\ \leq 
    & ~ \|  f(A,x)_{j_1} \cdot  f(A,x)_{j_0} \cdot (-f_2(A,x) + f(A,x)_{j_0})    \cdot p_{j_1}(A,x)^\top \cdot  h(A,x) \cdot I_d \\
    & ~ - f(\wt{A},x)_{j_1} \cdot  f(\wt{A},x)_{j_0} \cdot (-f_2(\wt{A},x) + f(\wt{A},x)_{j_0})    \cdot p_{j_1}(\wt{A},x)^\top \cdot  h(\wt{A},x) \cdot I_d \|_F\\
    \leq & ~  |f(A,x)_{j_1} - f(\wt{A},x)_{j_1}|\cdot | f(A,x)_{j_0}| \cdot ( |-f_2(A,x)| +  |f(A,x)_{j_0}|)  \cdot \| p_{j_1}(A,x)^\top \|_2  \cdot \| h(A,x) \|_2  \cdot \| I_d \|_F\\
    & + ~   |  f(\wt{A},x)_{j_1}|\cdot | f(A,x)_{j_0} -  f(\wt{A},x)_{j_0}| \cdot ( |-f_2(A,x)| +  |f(A,x)_{j_0}|)  \cdot \|  p_{j_1}(A,x)^\top \|_2 \cdot   \| h(A,x) \|_2   \cdot \| I_d \|_F\\
    & + ~    |  f(\wt{A},x)_{j_1}|\cdot |  f(\wt{A},x)_{j_0}| \cdot ( |f_2(A,x) - f_2(\wt{A},x)| +  |f(A,x)_{j_0} - f(\wt{A},x)_{j_0}|)   \cdot \| p_{j_1}(A,x)^\top \|_2 \\
    & ~\cdot  \| h(A,x) \|_2  \cdot \| I_d \|_F \\
    & + ~  |  f(\wt{A},x)_{j_1}|\cdot |  f(\wt{A},x)_{j_0}|\cdot ( | f_2(\wt{A},x)| +  | f(\wt{A},x)_{j_0}|)   \cdot \| p_{j_1}(A,x)^\top - p_{j_1}(\wt{A},x)^\top\|_2 \cdot  \| h(A,x) \|_2  \cdot \| I_d \|_F  \\
       & + ~    |  f(\wt{A},x)_{j_1}|\cdot |  f(\wt{A},x)_{j_0}|\cdot ( | f_2(\wt{A},x)| +  | f(\wt{A},x)_{j_0}|)   \cdot  \|  p_{j_1}(\wt{A},x)^\top \|_2  \cdot   \| h(A,x)- h(\wt{A},x) \|_2   \cdot \| I_d \|_F \\
    \leq & ~ 24R^4  \beta^{-2} \cdot n \cdot \sqrt{d}  \exp(3R^2)\|A - \wt{A}\|_F \\
    &+ ~ 24R^4  \beta^{-2} \cdot n \cdot \sqrt{d}  \exp(3R^2)\|A - \wt{A}\|_F \\
    & + ~ 36R^4\beta^{-2} \cdot n \cdot \sqrt{d}\exp(3R^2) \cdot \|A - \wt{A}\|_F \\
    & + ~ 26R^2\beta^{-2} \cdot n \cdot \sqrt{d}\exp(4R^2) \cdot \|A - \wt{A}\|_F \\
    & + ~ 36R^2\beta^{-2} \cdot n \cdot \sqrt{d}\exp(4R^2) \cdot \|A - \wt{A}\|_F \\
    \leq &  ~ 146\beta^{-2} \cdot n \cdot \sqrt{d}\exp(5R^2) \cdot \|A - \wt{A}\|_F \\
\leq &  ~ \beta^{-2} \cdot n \cdot \sqrt{d}\exp(6R^2) \cdot \|A - \wt{A}\|_F
\end{align*}

{\bf Proof of Part 8.}
\begin{align*}
& ~ \| B_{5,2,1}^{j_1,*,j_0,*} (A) - B_{5,2,1}^{j_1,*,j_0,*} ( \wt{A} ) \| \\ \leq 
    & ~ \| f(A,x)_{j_1} \cdot  f(A,x)_{j_0} \cdot (-f_2(A,x) + f(A,x)_{j_0})\cdot c_g(A,x)^{\top} \cdot h(A,x) \cdot x \cdot {\bf 1}_d^{\top}\\
    & ~ - f(\wt{A},x)_{j_1} \cdot  f(\wt{A},x)_{j_0} \cdot (-f_2(\wt{A},x) + f(\wt{A},x)_{j_0}) \cdot c_g(\wt{A},x)^{\top} \cdot  h(\wt{A},x) \cdot x \cdot {\bf 1}_d^{\top} \|_F \\
    \leq & ~  |f(A,x)_{j_1} - f(\wt{A},x)_{j_1}|\cdot | f(A,x)_{j_0}| \cdot ( |- f_2(A,x)| +  | f(A,x)_{j_0}|)  \cdot \| c_g(A,x)^\top \|_2  \cdot  \|h(A,x)\|_2 \cdot \| x \|_2 \cdot \| {\bf 1}_d^{\top} \|_F\\
    & + ~   |  f(\wt{A},x)_{j_1}|\cdot | f(A,x)_{j_0} -  f(\wt{A},x)_{j_0}| \cdot  ( |- f_2(A,x)| +  | f(A,x)_{j_0}|)  \cdot \|   c_g(A,x)^\top \|_2 \cdot  \|h(A,x)\|_2   \cdot \| x \|_2 \cdot \| {\bf 1}_d^{\top} \|_F\\
    & + ~    |  f(\wt{A},x)_{j_1}|\cdot |  f(\wt{A},x)_{j_0}| \cdot ( |f_2(A,x) - f_2(\wt{A},x)| +  |f(A,x)_{j_0} - f(\wt{A},x)_{j_0}|)  \cdot \| c_g(A,x)^\top \|_2  \\
     & ~\cdot  \|h(A,x)\|_2  \cdot \| x \|_2 \cdot \| {\bf 1}_d^{\top} \|_F\\
    & + ~  |  f(\wt{A},x)_{j_1}|\cdot |  f(\wt{A},x)_{j_0}|\cdot ( | f_2(\wt{A},x)| +  | f(\wt{A},x)_{j_0}|)    \cdot \| c_g(A,x)^\top - c_g(\wt{A},x)^\top \|_2 \cdot  \|h(A,x)\|_2  \cdot \| x \|_2 \cdot \| {\bf 1}_d^{\top} \|_F\\
    & + ~    |  f(\wt{A},x)_{j_1}|\cdot |  f(\wt{A},x)_{j_0}|\cdot ( | f_2(\wt{A},x)| +  | f(\wt{A},x)_{j_0}|)   \cdot  \|  c_g(\wt{A},x)^\top  \|_2  \cdot   \|h(A,x) - h(\wt{A},x)\|_2  \cdot \| x \|_2 \cdot \| {\bf 1}_d^{\top} \|_F\\
    \leq & ~ 20R^4 \beta^{-2} \cdot n \cdot \sqrt{d}  \exp(3R^2)\|A - \wt{A}\|_F \\
    &+ ~ 20R^4  \beta^{-2} \cdot n \cdot \sqrt{d}  \exp(3R^2)\|A - \wt{A}\|_F \\
    & + ~ 30R^4\beta^{-2} \cdot n \cdot \sqrt{d}\exp(3R^2) \cdot \|A - \wt{A}\|_F \\
    & + ~ 40R^4\beta^{-2} \cdot n \cdot \sqrt{d}\exp(3R^2) \cdot \|A - \wt{A}\|_F \\
     & + ~ 30R^2\beta^{-2} \cdot n \cdot \sqrt{d}\exp(4R^2) \cdot \|A - \wt{A}\|_F\\
    \leq &  ~ 140\beta^{-2} \cdot n \cdot \sqrt{d}\exp(5R^2) \cdot \|A - \wt{A}\|_F \\
    \leq &  ~ \beta^{-2} \cdot n \cdot \sqrt{d}\exp(6R^2) \cdot \|A - \wt{A}\|_F
\end{align*}

{\bf Proof of Part 9.}
\begin{align*}
  & ~ \|  B_{5,3,1}^{j_1,*,j_0,*} (A) - B_{5,3,1}^{j_1,*,j_0,*} ( \wt{A} ) \|_F \\
   = & ~ \| -  f(A,x)_{j_1} \cdot   f(A,x)_{j_0}  \cdot   (-f_2(A,x) + f(A,x)_{j_0})   \cdot {\bf 1}_d \cdot c_g(A,x)^{\top} \cdot \diag (x)\\
    & ~ -(-  f(\wt{A},x)_{j_1} \cdot   f(\wt{A},x)_{j_0}  \cdot  (-f_2(\wt{A},x) + f(\wt{A},x)_{j_0})   \cdot {\bf 1}_d \cdot c_g(\wt{A},x)^{\top} \cdot \diag (x))\|_F \\
    \leq & ~ \|    f(A,x)_{j_1} \cdot   f(A,x)_{j_0}  \cdot    (-f_2(A,x) + f(A,x)_{j_0})   \cdot {\bf 1}_d \cdot c_g(A,x)^{\top} \cdot \diag (x)\\
    & ~ -   f(\wt{A},x)_{j_1} \cdot   f(\wt{A},x)_{j_0}  \cdot   (-f_2(\wt{A},x) + f(\wt{A},x)_{j_0})  \cdot {\bf 1}_d \cdot c_g(\wt{A},x)^{\top} \cdot \diag (x) \|_F \\
    \leq & ~  |f(A,x)_{j_1} - f(\wt{A},x)_{j_1}|\cdot | f(A,x)_{j_0}| \cdot ( |- f_2(A,x)| +  | f(A,x)_{j_0}|)  \cdot \| c_g(A,x)^\top \|_2  \cdot  \|  {\bf 1}_d\|_2 \cdot \| \diag(x)\|_F  \\
    & + ~   |  f(\wt{A},x)_{j_1}|\cdot | f(A,x)_{j_0} -  f(\wt{A},x)_{j_0}| \cdot ( |- f_2(A,x)| +  | f(A,x)_{j_0}|)  \cdot \|   c_g(A,x)^\top \|_2 \cdot \|  {\bf 1}_d\|_2 \cdot \| \diag(x)\|_F\\
    & + ~    |  f(\wt{A},x)_{j_1}|\cdot |  f(\wt{A},x)_{j_0}| \cdot ( |f_2(A,x) - f_2(\wt{A},x)| +  |f(A,x)_{j_0} - f(\wt{A},x)_{j_0}|)   \cdot \| c_g(A,x)^\top \|_2 \cdot \|  {\bf 1}_d\|_2 \\
    & ~ \cdot \| \diag(x)\|_F\\
    & + ~  |  f(\wt{A},x)_{j_1}|\cdot |  f(\wt{A},x)_{j_0}|\cdot ( | f_2(\wt{A},x)| +  | f(\wt{A},x)_{j_0}|)    \cdot \| c_g(A,x)^\top - c_g(\wt{A},x)^\top \|_2 \cdot  \|  {\bf 1}_d\|_2 \cdot \| \diag(x)\|_F\\
    \leq & ~ 20R^2 \beta^{-2} \cdot n \cdot \sqrt{d}  \exp(3R^2)\|A - \wt{A}\|_F \\
    &+ ~ 20R^2  \beta^{-2} \cdot n \cdot \sqrt{d}  \exp(3R^2)\|A - \wt{A}\|_F \\
    & + ~ 30R^2\beta^{-2} \cdot n \cdot \sqrt{d}\exp(3R^2) \cdot \|A - \wt{A}\|_F \\
    & + ~ 40R^2\beta^{-2} \cdot n \cdot \sqrt{d}\exp(3R^2) \cdot \|A - \wt{A}\|_F \\
    \leq &  ~ 110\beta^{-2} \cdot n \cdot \sqrt{d}\exp(4R^2) \cdot \|A - \wt{A}\|_F \\
    \leq &  ~ \beta^{-2} \cdot n \cdot \sqrt{d}\exp(5R^2) \cdot \|A - \wt{A}\|_F
\end{align*}

{\bf Proof of Part 10.}
\begin{align*}
  & ~ \|  B_{5,4,1}^{j_1,*,j_0,*} (A) - B_{5,4,1}^{j_1,*,j_0,*} ( \wt{A} ) \|_F \\
    = & ~ \| -f(A,x)_{j_1} \cdot f(A,x)_{j_0}  \cdot (-f_2(A,x) + f(A,x)_{j_0}) \cdot  c_g(A,x)^{\top} \cdot  h_e(A,x) \cdot x \cdot {\bf 1}_d^{\top} \\
    & ~ - (-f(\wt{A},x)_{j_1} \cdot f(\wt{A},x)_{j_0}  \cdot (-f_2(\wt{A},x) + f(\wt{A},x)_{j_0})  \cdot  c_g(\wt{A},x)^{\top} \cdot  h_e(\wt{A},x) \cdot x \cdot {\bf 1}_d^{\top}) \|_F \\
    \leq & ~ \| f(A,x)_{j_1} \cdot f(A,x)_{j_0}  \cdot (-f_2(A,x) + f(A,x)_{j_0}) \cdot  c_g(A,x)^{\top} \cdot  h_e(A,x) \cdot x \cdot {\bf 1}_d^{\top} \\
    & ~ - f(\wt{A},x)_{j_1} \cdot f(\wt{A},x)_{j_0}  \cdot (-f_2(\wt{A},x) + f(\wt{A},x)_{j_0})  \cdot  c_g(\wt{A},x)^{\top} \cdot  h_e(\wt{A},x) \cdot x \cdot {\bf 1}_d^{\top} \|_F \\
    \leq & ~ |f(A,x)_{j_1} - f(\wt{A},x)_{j_1}| \cdot |f(A,x)_{j_0} | \cdot ( |- f_2(A,x)| +  | f(A,x)_{j_0}|) \cdot  \| c_g(A,x)^{\top} \|_2 \cdot  \| h_e(A,x) \|_2 \cdot \| x \|_2 \cdot \|{\bf 1}_d^{\top} \|_2 \\
    & + ~ |f(\wt{A},x)_{j_1}| \cdot |f(A,x)_{j_0} - f(\wt{A},x)_{j_0} | \cdot ( |- f_2(A,x)| +  | f(A,x)_{j_0}|) \cdot  \| c_g(A,x)^{\top} \|_2 \cdot   \| h_e(A,x) \|_2 \cdot \| x \|_2 \\
    & ~ \cdot \|{\bf 1}_d^{\top} \|_2 \\
    & + ~ |f(\wt{A},x)_{j_1}| \cdot |f(\wt{A},x)_{j_0} | \cdot ( |f_2(A,x) - f_2(\wt{A},x)| +  |f(A,x)_{j_0} - f(\wt{A},x)_{j_0}|) \cdot  \| c_g(A,x)^{\top} \|_2 \\
     & ~\cdot   \| h_e(A,x) \|_2 \cdot \| x \|_2 \cdot \|{\bf 1}_d^{\top} \|_2 \\
    & + ~ |f(\wt{A},x)_{j_1}| \cdot |f(\wt{A},x)_{j_0} | \cdot  ( | f_2(\wt{A},x)| +  | f(\wt{A},x)_{j_0}|) \cdot  \| c_g(A,x)^{\top}  - c_g(\wt{A},x)^{\top} \|_2 \cdot  \| h_e(A,x) \|_2 \cdot \| x \|_2 \cdot \|{\bf 1}_d^{\top} \|_2 \\
    & + ~ |f(\wt{A},x)_{j_1}| \cdot |f(\wt{A},x)_{j_0} | \cdot ( | f_2(\wt{A},x)| +  | f(\wt{A},x)_{j_0}|)  \cdot  \| c_g(\wt{A},x)^{\top} \|_2 \cdot   \| h_e(A,x) - h_e(\wt{A},x) \|_2 \cdot \| x \|_2 \cdot \|{\bf 1}_d^{\top} \|_2 \\
    \leq & ~ 40R^4 \beta^{-2} \cdot n \cdot \sqrt{d}  \exp(3R^2)\|A - \wt{A}\|_F \\
    & + ~ 40R^4 \beta^{-2} \cdot n \cdot \sqrt{d}  \exp(3R^2)\|A - \wt{A}\|_F \\
    & + ~ 60R^4 \beta^{-2} \cdot n \cdot \sqrt{d}  \exp(3R^2)\|A - \wt{A}\|_F \\
    & + ~ 80R^4 \beta^{-2} \cdot n \cdot \sqrt{d}  \exp(3R^2)\|A - \wt{A}\|_F \\
    & + ~ 30R^2  \beta^{-2} \cdot n \cdot \sqrt{d}  \exp(4R^2)\|A - \wt{A}\|_F \\
    \leq & ~ 250\beta^{-2} \cdot n \cdot \sqrt{d}  \exp(5R^2)\|A - \wt{A}\|_F \\
    \leq & ~ \beta^{-2} \cdot n \cdot \sqrt{d}  \exp(6R^2)\|A - \wt{A}\|_F 
\end{align*}

{\bf Proof of Part 11.}
\begin{align*}
  & ~ \|  B_{5,5,1}^{j_1,*,j_0,*} (A) - B_{5,5,1}^{j_1,*,j_0,*} ( \wt{A} ) \|_F \\
    =   & ~ \|- f(A,x)_{j_1} \cdot f(A,x)_{j_0}  \cdot f_2( A,x) \cdot  c_g(A,x)^{\top} \cdot  h(A,x) \cdot x \cdot {\bf 1}_d^{\top} \\
    & ~ -( -f(\wt{A},x)_{j_1} \cdot f(\wt{A},x)_{j_0}  \cdot f_2( \wt{A},x) \cdot  c_g(\wt{A},x)^{\top} \cdot  h(\wt{A},x) \cdot x \cdot {\bf 1}_d^{\top}) \|_F \\
    \leq    & ~ \| f(A,x)_{j_1} \cdot f(A,x)_{j_0}  \cdot f_2( A,x) \cdot  c_g(A,x)^{\top} \cdot  h(A,x) \cdot x \cdot {\bf 1}_d^{\top} \\
    & ~ - f(\wt{A},x)_{j_1} \cdot f(\wt{A},x)_{j_0}  \cdot f_2( \wt{A},x) \cdot  c_g(\wt{A},x)^{\top} \cdot  h(\wt{A},x) \cdot x \cdot {\bf 1}_d^{\top} \|_F \\
    \leq & ~ |f(A,x)_{j_1} - f(\wt{A},x)_{j_1}| \cdot |f(A,x)_{j_0} | \cdot |f_2( A,x) |\cdot  \| c_g(A,x)^{\top} \|_2 \cdot  \| h(A,x) \|_2 \cdot \| x \|_2 \cdot \|{\bf 1}_d^{\top} \|_2 \\
    & + ~ |f(\wt{A},x)_{j_1}| \cdot |f(A,x)_{j_0} - f(\wt{A},x)_{j_0} | \cdot |f_2( A,x) |\cdot  \| c_g(A,x)^{\top} \|_2 \cdot   \| h(A,x) \|_2 \cdot \| x \|_2 \cdot \|{\bf 1}_d^{\top} \|_2 \\
    & + ~ |f(\wt{A},x)_{j_1}| \cdot |f(\wt{A},x)_{j_0} | \cdot |f_2( A,x)  -f_2( \wt{A},x)|\cdot  \| c_g(A,x)^{\top} \|_2 \cdot   \| h(A,x) \|_2 \cdot \| x \|_2 \cdot \|{\bf 1}_d^{\top} \|_2 \\
    & + ~ |f(\wt{A},x)_{j_1}| \cdot |f(\wt{A},x)_{j_0} | \cdot |f_2( \wt{A},x)|\cdot  \| c_g(A,x)^{\top}  - c_g(\wt{A},x)^{\top} \|_2 \cdot  \| h(A,x) \|_2 \cdot \| x \|_2 \cdot \|{\bf 1}_d^{\top} \|_2 \\
    & + ~ |f(\wt{A},x)_{j_1}| \cdot |f(\wt{A},x)_{j_0} | \cdot |f_2( \wt{A},x)|\cdot  \| c_g(\wt{A},x)^{\top} \|_2 \cdot   \| h(A,x) - h(\wt{A},x) \|_2 \cdot \| x \|_2 \cdot \|{\bf 1}_d^{\top} \|_2 \\
    \leq & ~ 10R^4 \beta^{-2} \cdot n \cdot \sqrt{d}  \exp(3R^2)\|A - \wt{A}\|_F \\
    & + ~ 10R^4 \beta^{-2} \cdot n \cdot \sqrt{d}  \exp(3R^2)\|A - \wt{A}\|_F \\
    & + ~ 20R^4 \beta^{-2} \cdot n \cdot \sqrt{d}  \exp(3R^2)\|A - \wt{A}\|_F \\
    & + ~ 20R^4 \beta^{-2} \cdot n \cdot \sqrt{d}  \exp(3R^2)\|A - \wt{A}\|_F \\
    & + ~ 15R^2  \beta^{-2} \cdot n \cdot \sqrt{d}  \exp(4R^2)\|A - \wt{A}\|_F \\
    \leq & ~ 75\beta^{-2} \cdot n \cdot \sqrt{d}  \exp(5R^2)\|A - \wt{A}\|_F \\
    \leq & ~ \beta^{-2} \cdot n \cdot \sqrt{d}  \exp(6R^2)\|A - \wt{A}\|_F 
\end{align*}

{\bf Proof of Part 12.}
\begin{align*}
  & ~ \|  B_{5,5,2}^{j_1,*,j_0,*} (A) - B_{5,5,2}^{j_1,*,j_0,*} ( \wt{A} ) \|_F \\
    =   & ~ \| f(A,x)_{j_1}^2 \cdot f(A,x)_{j_0}    \cdot  c_g(A,x)^{\top} \cdot  h(A,x) \cdot x \cdot {\bf 1}_d^{\top} \\
    & ~ - f(\wt{A},x)_{j_1}^2 \cdot f(\wt{A},x)_{j_0}  \cdot   c_g(\wt{A},x)^{\top} \cdot  h(\wt{A},x) \cdot x \cdot {\bf 1}_d^{\top} \|_F \\
    \leq & ~ |f(A,x)_{j_1} - f(\wt{A},x)_{j_1}| \cdot |f(A,x)_{j_1} | \cdot |f(A,x)_{j_0} |  \cdot  \| c_g(A,x)^{\top} \|_2 \cdot  \| h(A,x) \|_2 \cdot \| x \|_2 \cdot \|{\bf 1}_d^{\top} \|_2 \\
     & + ~ |f(\wt{A},x)_{j_1}| \cdot |f(A,x)_{j_1} - f(\wt{A},x)_{j_1}|\cdot |f(A,x)_{j_0} |\cdot  \| c_g(A,x)^{\top} \|_2 \cdot   \| h(A,x) \|_2 \cdot \| x \|_2 \cdot \|{\bf 1}_d^{\top} \|_2 \\
    & + ~ |f(\wt{A},x)_{j_1}| \cdot |f(\wt{A},x)_{j_1}| \cdot |f(A,x)_{j_0} - f(\wt{A},x)_{j_0} |  \cdot  \| c_g(A,x)^{\top} \|_2 \cdot   \| h(A,x) \|_2 \cdot \| x \|_2 \cdot \|{\bf 1}_d^{\top} \|_2 \\
    & + ~ |f(\wt{A},x)_{j_1}| \cdot |f(\wt{A},x)_{j_1}| \cdot |f(\wt{A},x)_{j_0} | \cdot   \| c_g(A,x)^{\top}  - c_g(\wt{A},x)^{\top} \|_2 \cdot  \| h(A,x) \|_2 \cdot \| x \|_2 \cdot \|{\bf 1}_d^{\top} \|_2 \\
    & + ~ |f(\wt{A},x)_{j_1}| \cdot |f(\wt{A},x)_{j_1}| \cdot |f(\wt{A},x)_{j_0} | \cdot    \| c_g(\wt{A},x)^{\top} \|_2 \cdot   \| h(A,x) - h(\wt{A},x) \|_2 \cdot \| x \|_2 \cdot \|{\bf 1}_d^{\top} \|_2 \\
    \leq & ~ 10R^4 \beta^{-2} \cdot n \cdot \sqrt{d}  \exp(3R^2)\|A - \wt{A}\|_F \\
    & + ~ 10R^4 \beta^{-2} \cdot n \cdot \sqrt{d}  \exp(3R^2)\|A - \wt{A}\|_F \\
    & + ~ 10R^4 \beta^{-2} \cdot n \cdot \sqrt{d}  \exp(3R^2)\|A - \wt{A}\|_F \\
    & + ~ 20R^4 \beta^{-2} \cdot n \cdot \sqrt{d}  \exp(3R^2)\|A - \wt{A}\|_F \\
    & + ~ 15R^2  \beta^{-2} \cdot n \cdot \sqrt{d}  \exp(4R^2)\|A - \wt{A}\|_F \\
    \leq & ~ 65\beta^{-2} \cdot n \cdot \sqrt{d}  \exp(5R^2)\|A - \wt{A}\|_F \\
    \leq & ~ \beta^{-2} \cdot n \cdot \sqrt{d}  \exp(6R^2)\|A - \wt{A}\|_F 
\end{align*}

{\bf Proof of Part 13.}
\begin{align*}
  & ~ \|  B_{5,6,1}^{j_1,*,j_0,*} (A) - B_{5,6,1}^{j_1,*,j_0,*} ( \wt{A} ) \|_F \\
    =   & ~ \|- f(A,x)_{j_1} \cdot f(A,x)_{j_0}  \cdot f_2( A,x) \cdot  c_g(A,x)^{\top} \cdot  h(A,x) \cdot x \cdot {\bf 1}_d^{\top} \\
    & ~ -( -f(\wt{A},x)_{j_1} \cdot f(\wt{A},x)_{j_0}  \cdot f_2( \wt{A},x) \cdot  c_g(\wt{A},x)^{\top} \cdot  h(\wt{A},x) \cdot x \cdot {\bf 1}_d^{\top}) \|_F \\
    = & ~ \|  B_{5,5,1}^{j_1,*,j_0,*} (A) - B_{5,5,1}^{j_1,*,j_0,*} ( \wt{A} ) \|_F \\
    \leq & ~ \beta^{-2} \cdot n \cdot \sqrt{d}  \exp(6R^2)\|A - \wt{A}\|_F 
\end{align*}

{\bf Proof of Part 14.}
\begin{align*}
  & ~ \|  B_{5,6,2}^{j_1,*,j_0,*} (A) - B_{5,6,2}^{j_1,*,j_0,*} ( \wt{A} ) \|_F \\
    =   & ~ \| f(A,x)_{j_1}^2 \cdot f(A,x)_{j_0}    \cdot  c_g(A,x)^{\top} \cdot  h(A,x) \cdot x \cdot {\bf 1}_d^{\top} \\
    & ~ - f(\wt{A},x)_{j_1}^2 \cdot f(\wt{A},x)_{j_0}  \cdot   c_g(\wt{A},x)^{\top} \cdot  h(\wt{A},x) \cdot x \cdot {\bf 1}_d^{\top} \|_F \\
    = & ~ \|  B_{5,5,2}^{j_1,*,j_0,*} (A) - B_{5,5,2}^{j_1,*,j_0,*} ( \wt{A} ) \|_F \\
    \leq & ~ \beta^{-2} \cdot n \cdot \sqrt{d}  \exp(6R^2)\|A - \wt{A}\|_F 
\end{align*}

{\bf Proof of Part 15.}
\begin{align*}
  & ~ \|  B_{5,7,1}^{j_1,*,j_0,*} (A) - B_{5,7,1}^{j_1,*,j_0,*} ( \wt{A} ) \|_F \\
    =   & ~ \| f(A,x)_{j_0}^2 \cdot f(A,x)_{j_1}    \cdot  c_g(A,x)^{\top} \cdot  h(A,x) \cdot x \cdot {\bf 1}_d^{\top} \\
    & ~ - f(\wt{A},x)_{j_0}^2 \cdot f(\wt{A},x)_{j_1}  \cdot   c_g(\wt{A},x)^{\top} \cdot  h(\wt{A},x) \cdot x \cdot {\bf 1}_d^{\top} \|_F \\
    \leq & ~ |f(A,x)_{j_0} - f(\wt{A},x)_{j_0}| \cdot |f(A,x)_{j_0} | \cdot |f(A,x)_{j_1} |  \cdot  \| c_g(A,x)^{\top} \|_2 \cdot  \| h(A,x) \|_2 \cdot \| x \|_2 \cdot \|{\bf 1}_d^{\top} \|_2 \\
     & + ~ |f(\wt{A},x)_{j_0}| \cdot |f(A,x)_{j_0} - f(\wt{A},x)_{j_0}|\cdot |f(A,x)_{j_1} |\cdot  \| c_g(A,x)^{\top} \|_2 \cdot   \| h(A,x) \|_2 \cdot \| x \|_2 \cdot \|{\bf 1}_d^{\top} \|_2 \\
    & + ~ |f(\wt{A},x)_{j_0}| \cdot |f(\wt{A},x)_{j_0}| \cdot |f(A,x)_{j_1} - f(\wt{A},x)_{j_1} |  \cdot  \| c_g(A,x)^{\top} \|_2 \cdot   \| h(A,x) \|_2 \cdot \| x \|_2 \cdot \|{\bf 1}_d^{\top} \|_2 \\
    & + ~ |f(\wt{A},x)_{j_0}| \cdot |f(\wt{A},x)_{j_0}| \cdot |f(\wt{A},x)_{j_1} | \cdot   \| c_g(A,x)^{\top}  - c_g(\wt{A},x)^{\top} \|_2 \cdot  \| h(A,x) \|_2 \cdot \| x \|_2 \cdot \|{\bf 1}_d^{\top} \|_2 \\
    & + ~ |f(\wt{A},x)_{j_0}| \cdot |f(\wt{A},x)_{j_0}| \cdot |f(\wt{A},x)_{j_1} | \cdot    \| c_g(\wt{A},x)^{\top} \|_2 \cdot   \| h(A,x) - h(\wt{A},x) \|_2 \cdot \| x \|_2 \cdot \|{\bf 1}_d^{\top} \|_2 \\
    \leq & ~ 10R^4 \beta^{-2} \cdot n \cdot \sqrt{d}  \exp(3R^2)\|A - \wt{A}\|_F \\
    & + ~ 10R^4 \beta^{-2} \cdot n \cdot \sqrt{d}  \exp(3R^2)\|A - \wt{A}\|_F \\
    & + ~ 10R^4 \beta^{-2} \cdot n \cdot \sqrt{d}  \exp(3R^2)\|A - \wt{A}\|_F \\
    & + ~ 20R^4 \beta^{-2} \cdot n \cdot \sqrt{d}  \exp(3R^2)\|A - \wt{A}\|_F \\
    & + ~ 15R^2  \beta^{-2} \cdot n \cdot \sqrt{d}  \exp(4R^2)\|A - \wt{A}\|_F \\
    \leq & ~ 65\beta^{-2} \cdot n \cdot \sqrt{d}  \exp(5R^2)\|A - \wt{A}\|_F \\
    \leq & ~ \beta^{-2} \cdot n \cdot \sqrt{d}  \exp(6R^2)\|A - \wt{A}\|_F 
\end{align*}

{\bf Proof of Part 16.}
\begin{align*}
 & ~ \| B_{5}^{j_1,*,j_0,*} (A) - B_{5}^{j_1,*,j_0,*} ( \wt{A} ) \|_F \\
   = & ~ \|\sum_{i = 1}^7 B_{5,i}^{j_1,*,j_0,*}(A) -   B_{5,i}^{j_1,*,j_0,*}(\wt{A})  \|_F \\
    \leq & ~ 15  \beta^{-2} \cdot n \cdot \sqrt{d} \cdot \exp(6R^2)\|A - \wt{A}\|_F
\end{align*}
\end{proof}

\subsection{PSD}
\begin{lemma}
If the following conditions hold
\begin{itemize}
    \item Let $B_{5,1,1}^{j_1,*, j_0,*}, \cdots, B_{5,7,1}^{j_1,*, j_0,*} $ be defined as Lemma~\ref{lem:b_5_j1_j0} \label{psd: B_5}
    \item  Let $\|A \|_2 \leq R, \|A^{\top} \|_F \leq R, \| x\|_2 \leq R, \|\diag(f(A,x)) \|_F \leq \|f(A,x) \|_2 \leq 1, \| b_g \|_2 \leq 1$ 
\end{itemize}
We have 
\begin{itemize}
    \item {\bf Part 1.} $\| B_{5,1,1}^{j_1,*, j_0,*} \| \leq 4 \sqrt{d} R^2$  
    \item {\bf Part 2.} $\|B_{5,1,2}^{j_1,*, j_0,*}\| \preceq 4 \sqrt{d} R^2$
    \item {\bf Part 3.} $\|B_{5,1,3}^{j_1,*, j_0,*}\| \preceq 4  R^4$
    \item {\bf Part 4.} $\|B_{5,1,4}^{j_1,*, j_0,*} \|\preceq 4 R^4$
    \item {\bf Part 5.} $\|B_{5,1,5}^{j_1,*, j_0,*} \|\preceq 4  R^4$
    \item {\bf Part 6.} $\|B_{5,1,6}^{j_1,*, j_0,*} \|\preceq 6  R^4$
    \item {\bf Part 7.} $\|B_{5,1,7}^{j_1,*, j_0,*} \|\preceq 12  R^4$
    \item {\bf Part 8.} $\|B_{5,2,1}^{j_1,*, j_0,*}\| \preceq 10\sqrt{d} R^4$
    \item {\bf Part 9.} $\|B_{5,3,1}^{j_1,*, j_0,*}\| \preceq 10 \sqrt{d}R^2$
    \item {\bf Part 10.} $\|B_{5,4,1}^{j_1,*, j_0,*}\| \preceq 20\sqrt{d} R^4$
    \item {\bf Part 11.} $\|B_{5,5,1}^{j_1,*, j_0,*}\| \preceq 5\sqrt{d} R^4$
    \item {\bf Part 12.} $\|B_{5,5,2}^{j_1,*, j_0,*}\| \preceq 5\sqrt{d}R^4$
    \item {\bf Part 13.} $\|B_{5,6,1}^{j_1,*, j_0,*}\| \preceq 5\sqrt{d}R^4$
    \item {\bf Part 14.} $\|B_{5,6,2}^{j_1,*, j_0,*}\| \preceq 5\sqrt{d} R^4$
    \item {\bf Part 15.} $\|B_{5,7,1}^{j_1,*, j_0,*}\| \preceq 5\sqrt{d} R^4$
    \item {\bf Part 16.} $\|B_{5}^{j_1,*, j_0,*}\| \preceq 103\sqrt{d} R^4$
\end{itemize}
\end{lemma}

\begin{proof}
    {\bf Proof of Part 1.}
    \begin{align*}
        \| B_{5,1,1}^{j_1,*, j_0,*} \| 
        = & ~ \|  f_c(A,x)  \cdot f(A,x)_{j_1} \cdot f(A,x)_{j_0} \cdot  (-f_2(A,x) + f(A,x)_{j_0}) \cdot h(A,x)\cdot  {\bf 1}_d^\top\| \\
        \leq & ~ |f(A,x)_{j_1}| \cdot |f(A,x)_{j_0}| \cdot |f_c(A,x)|  \cdot  (|-f_2(A,x)| + |f(A,x)_{j_0}|) \cdot \|h(A,x)\|_2 \cdot \| {\bf 1}_d^\top\|_2\\
        \leq & ~ 4 \sqrt{d} R^2
    \end{align*}

    {\bf Proof of Part 2.}
    \begin{align*}
        \| B_{5,1,2}^{j_1,*, j_0,*} \|
        = &~
        \| f(A,x)_{j_1} \cdot f(A,x)_{j_0} \cdot c(A,x)_{j_1} \cdot  (-f_2(A,x) + f(A,x)_{j_0}) \cdot h(A,x) \cdot  {\bf 1}_d^\top  \| \\
        \preceq & ~  |f(A,x)_{j_1}| \cdot |f(A,x)_{j_0}| \cdot |c(A,x)_{j_1}|\cdot (|-f_2(A,x)| + |f(A,x)_{j_0}|) \cdot \|h(A,x)\|_2 \cdot \| {\bf 1}_d^\top\|_2\\
        \preceq & ~ 4 \sqrt{d} R^2
    \end{align*}

    {\bf Proof of Part 3.}
    \begin{align*}
     & ~ \| B_{5,1,3}^{j_1,*, j_0,*} \| \\
     = & ~
       \| f(A,x)_{j_1} \cdot  f(A,x)_{j_0} \cdot f_c(A,x)\cdot  (-f_2(A,x) + f(A,x)_{j_0}) \cdot  ((A_{j_1,*}) \circ x^\top) \cdot h(A,x)  \cdot I_d \| \\
    \leq & ~ |f(A,x)_{j_1}| \cdot |f(A,x)_{j_0}| \cdot  |f_c(A,x)|\cdot (|-f_2(A,x)| + |f(A,x)_{j_0}|)     \cdot \| A_{j_1,*}^\top \circ x \|_2 \cdot \|h(A,x)\|_2 \cdot \| I_d\| \\
    \leq & ~ |f(A,x)_{j_1}| \cdot |f(A,x)_{j_0}| \cdot  |f_c(A,x)|\cdot (|-f_2(A,x)| + |f(A,x)_{j_0}|)    \cdot \| A_{j_1,*}\|_2 \cdot \|\diag(x) \|_{\infty} \cdot \|h(A,x)\|_2 \cdot \| I_d\| \\
    \leq &  ~  4 R^4
    \end{align*}
   
    {\bf Proof of Part 4.}
    \begin{align*}
        \|B_{5,1,4}^{j_1,*, j_0,*}  \|
        = & ~  \|-f(A,x)_{j_1} \cdot  f(A,x)_{j_0} \cdot f_c(A,x)\cdot  (-f_2(A,x) + f(A,x)_{j_0})  \cdot h(A,x)^\top \cdot  h(A,x) \cdot I_d \|\\
        \leq & ~   \| f(A,x)_{j_1} \cdot  f(A,x)_{j_0} \cdot f_c(A,x)\cdot (|-f_2(A,x)| + |f(A,x)_{j_0}|)  \cdot h(A,x)^\top \cdot  h(A,x) \cdot I_d \|\\
        \leq & ~ | f(A,x)_{j_1} | \cdot|  f(A,x)_{j_0} |\cdot |f_c(A,x)|\cdot (|-f_2(A,x)| + |f(A,x)_{j_0}|) \cdot \|h(A,x)^\top\|_2 \cdot  \|h(A,x)\|_2  \cdot \| I_d\|\\
        \leq & ~ | f(A,x)_{j_1} | \cdot|  f(A,x)_{j_0} |\cdot |f_c(A,x)|\cdot (|-f_2(A,x)| + |f(A,x)_{j_0}|) \cdot \|h(A,x)^\top\|_2 \cdot  \|h(A,x)\|_2  \cdot \| I_d\| \\
        \leq & ~ 4 R^4
    \end{align*}

    {\bf Proof of Part 5.}
    \begin{align*}
       & ~  \|B_{5,1,5}^{j_1,*, j_0,*} \|\\
        = & ~ \| f(A,x)_{j_1} \cdot  f(A,x)_{j_0} \cdot (-f_2(A,x) + f(A,x)_{j_1})  \cdot (-f_2(A,x) + f(A,x)_{j_0})  \cdot h(A,x)^\top \cdot  h(A,x) \cdot I_d\| \\
        \leq & ~  | f(A,x)_{j_1}| \cdot |f(A,x)_{j_0}|    \cdot (|-f_2(A,x)| + |f(A,x)_{j_1}|) \cdot (|-f_2(A,x)| + |f(A,x)_{j_0}|)   \\
        & ~\cdot \|h(A,x)^\top\|_2  \cdot \|h(A,x)\|_2 \cdot \| I_d\|\\
        \leq & ~ 4 R^4
    \end{align*}

    {\bf Proof of Part 6.}
    \begin{align*}
       & ~  \|B_{5,1,6}^{j_1,*, j_0,*} \|\\
        = & ~ \| f(A,x)_{j_1} \cdot  f(A,x)_{j_0} \cdot (-f_c(A,x) + f(A,x)_{j_1}) \cdot (-f_2(A,x) + f(A,x)_{j_0})   \cdot h(A,x)^\top \cdot  h(A,x) \cdot I_d \|\\
        \leq & ~  | f(A,x)_{j_1}| \cdot |f(A,x)_{j_0}|    \cdot (|-f_c(A,x)| + |f(A,x)_{j_1}|)  \cdot (|-f_2(A,x)| + |f(A,x)_{j_0}|)  \\
        & ~ \cdot \|h(A,x)^\top\|_2  \cdot \|h(A,x)\|_2 \cdot \| I_d\| \\
        \leq & ~ 6 R^4
    \end{align*}

    {\bf Proof of Part 7.}
    \begin{align*}
         \|B_{5,1,7}^{j_1,*, j_0,*} \|
         = & ~ \|f(A,x)_{j_1} \cdot  f(A,x)_{j_0} \cdot (-f_2(A,x) + f(A,x)_{j_0})   \cdot p_{j_1}(A,x)^\top \cdot   h(A,x) \cdot I_d \|\\
         \leq & ~ | f(A,x)_{j_1}| \cdot |f(A,x)_{j_0}| \cdot (|-f_2(A,x)| + |f(A,x)_{j_0}|) \cdot \|p_{j_1}(A,x)^\top\|_2  \cdot \|h(A,x)\|_2 \cdot \| I_d\|\\
         \leq & ~ 12  R^4
    \end{align*}

    {\bf Proof of Part 8.}
    \begin{align*}
        \|B_{5,2,1}^{j_1,*, j_0,*} \|
        = & ~ \|  f(A,x)_{j_1} \cdot  f(A,x)_{j_0} \cdot (-f_2(A,x) + f(A,x)_{j_0}) \cdot c_g(A,x)^{\top} \cdot h(A,x) \cdot x \cdot {\bf 1}_d^{\top} \| \\
        \leq & ~  | f(A,x)_{j_1}| \cdot |f(A,x)_{j_0}| \cdot (|-f_2(A,x)| + |f(A,x)_{j_0}|) \cdot \|c_g(A,x)^{\top}\|_2 \cdot \|h(A,x)\|_2 \cdot \| x\|_2 \cdot \| {\bf 1}_d^{\top}\|_2\\
        \leq & ~ 10 \sqrt{d} R^4
    \end{align*}

    {\bf Proof of Part 9.}
    \begin{align*}
     \|B_{5,3,1}^{j_1,*, j_0,*} \|
        = & ~ \|   -  f(A,x)_{j_1} \cdot   f(A,x)_{j_0}  \cdot    (-f_2(A,x) + f(A,x)_{j_0})   \cdot {\bf 1}_d \cdot c_g(A,x)^{\top} \cdot \diag (x)\| \\
        \leq & ~ \|   f(A,x)_{j_1} \cdot   f(A,x)_{j_0}  \cdot  (-f_2(A,x) + f(A,x)_{j_0})   \cdot {\bf 1}_d \cdot c_g(A,x)^{\top} \cdot \diag (x)\| \\
        \leq & ~ | f(A,x)_{j_1}| \cdot  |f(A,x)_{j_0} | \cdot  (|-f_2(A,x)| + |f(A,x)_{j_0}|) \cdot \| {\bf 1}_d\|_2 \cdot \|c_g(A,x)^{\top}\|_2 \cdot \| \diag (x)\|  \\
        \leq & ~ 10 \sqrt{d} R^2
    \end{align*}

        {\bf Proof of Part 10.}
    \begin{align*}
     \|B_{5,4,1}^{j_1,*, j_0,*} \|
        = & ~ \|  -f(A,x)_{j_1} \cdot f(A,x)_{j_0}  \cdot  (-f_2(A,x) + f(A,x)_{j_0}) \cdot  c_g(A,x)^{\top} \cdot h_e(A,x) \cdot x \cdot {\bf 1}_d^{\top}\| \\
        \leq & ~ \|  f(A,x)_{j_1} \cdot f(A,x)_{j_0}  \cdot  (-f_2(A,x) + f(A,x)_{j_0}) \cdot  c_g(A,x)^{\top} \cdot h_e(A,x) \cdot x \cdot {\bf 1}_d^{\top}\| \\
        \leq & ~ | f(A,x)_{j_1}| \cdot  |f(A,x)_{j_0} | \cdot  (|-f_2(A,x)| + |f(A,x)_{j_0}|) \cdot |c_g(A,x)^{\top}| \cdot \| h_e(A,x) \|_2 \cdot \| x\|_2 \cdot \| {\bf 1}_d^{\top}\|_2\\
        \leq & ~ 20 \sqrt{d}R^4
    \end{align*}

       {\bf Proof of Part 11.}
    \begin{align*}
     \|B_{5,5,1}^{j_1,*, j_0,*} \|
        = & ~ \| - f(A,x)_{j_1} \cdot f(A,x)_{j_0} \cdot  f_2(A,x)  \cdot  c_g(A,x)^{\top} \cdot  h(A,x) \cdot x \cdot {\bf 1}_d^{\top}  \| \\
        \leq& ~ \|  f(A,x)_{j_1} \cdot f(A,x)_{j_0} \cdot  f_2(A,x)  \cdot  c_g(A,x)^{\top} \cdot  h(A,x) \cdot x \cdot {\bf 1}_d^{\top}  \| \\
        \leq & ~ | f(A,x)_{j_1}|  \cdot  |f(A,x)_{j_0} | \cdot  |f_2(A,x) |  \cdot \|c_g(A,x)^{\top}\|_2 \cdot \|h(A,x)\|_2 \cdot \| x\|_2 \cdot \| {\bf 1}_d^{\top}\|_2\\
        \leq & ~ 5\sqrt{d} R^4
    \end{align*}

           {\bf Proof of Part 12.}
    \begin{align*}
     \|B_{5,5,2}^{j_1,*, j_0,*} \|
        = & ~ \|  f(A,x)_{j_1}^2 \cdot f(A,x)_{j_0}    \cdot  c_g(A,x)^{\top} \cdot  h(A,x) \cdot x \cdot {\bf 1}_d^{\top}  \| \\
        \leq & ~ | f(A,x)_{j_1}|^2  \cdot  |f(A,x)_{j_0} |  \cdot \|c_g(A,x)^{\top}\|_2 \cdot \|h(A,x)\|_2 \cdot \| x\|_2 \cdot \| {\bf 1}_d^{\top}\|_2\\
        \leq & ~ 5\sqrt{d} R^4
    \end{align*}

           {\bf Proof of Part 13.}
    \begin{align*}
     \|B_{5,6,1}^{j_1,*, j_0,*} \|
        = & ~ \| - f(A,x)_{j_1} \cdot f(A,x)_{j_0} \cdot  f_2(A,x)  \cdot  c_g(A,x)^{\top} \cdot  h(A,x) \cdot x \cdot {\bf 1}_d^{\top}  \| \\
        \leq& ~ \|  f(A,x)_{j_1} \cdot f(A,x)_{j_0} \cdot  f_2(A,x)  \cdot  c_g(A,x)^{\top} \cdot  h(A,x) \cdot x \cdot {\bf 1}_d^{\top}  \| \\
        \leq & ~ | f(A,x)_{j_1}|  \cdot  |f(A,x)_{j_0} | \cdot  |f_2(A,x) |  \cdot \|c_g(A,x)^{\top}\|_2 \cdot \|h(A,x)\|_2 \cdot \| x\|_2 \cdot \| {\bf 1}_d^{\top}\|_2\\
        \leq & ~ 5\sqrt{d} R^4
    \end{align*}

           {\bf Proof of Part 14.}
    \begin{align*}
     \|B_{5,6,2}^{j_1,*, j_0,*} \|
       = & ~ \|  f(A,x)_{j_1}^2 \cdot f(A,x)_{j_0}    \cdot  c_g(A,x)^{\top} \cdot  h(A,x) \cdot x \cdot {\bf 1}_d^{\top}  \| \\
        \leq & ~ | f(A,x)_{j_1}|^2  \cdot  |f(A,x)_{j_0} |  \cdot \|c_g(A,x)^{\top}\|_2 \cdot \|h(A,x)\|_2 \cdot \| x\|_2 \cdot \| {\bf 1}_d^{\top}\|_2\\
        \leq & ~ 5\sqrt{d} R^4
    \end{align*}
    
           {\bf Proof of Part 15.}
    \begin{align*}
     \|B_{5,7,1}^{j_1,*, j_0,*} \|
       = & ~ \|  f(A,x)_{j_1}\cdot f(A,x)_{j_0}^2   \cdot c_g(A,x)^{\top} \cdot h(A,x)  \cdot x \cdot {\bf 1}_d^{\top}  \| \\
        \leq & ~ | f(A,x)_{j_1}|  \cdot  |f(A,x)_{j_0} |^2  \cdot \|c_g(A,x)^{\top}\|_2 \cdot \|h(A,x)\|_2 \cdot \| x\|_2 \cdot \| {\bf 1}_d^{\top}\|_2\\
        \leq & ~ 5\sqrt{d} R^4
    \end{align*}

    {\bf Proof of Part 16.}
\begin{align*}
 & ~ \| B_{5}^{j_1,*,j_0,*}\| \\
   = & ~ \|\sum_{i = 1}^7 B_{5,i}^{j_1,*,j_0,*}\| \\
    \leq & ~ 103\sqrt{d} R^4
\end{align*}
\end{proof}

\newpage

\section{Hessian: Sixth term  \texorpdfstring{$B_6^{j_1,i_1,j_0,i_0}$}{}}\label{app:hessian_sixth}
\subsection{Definitions}
\begin{definition} \label{def:b_6}
    We define the $B_6^{j_1,i_1,j_0,i_0}$ as follows,
    \begin{align*}
        B_6^{j_1,i_1,j_0,i_0} & ~ = \frac{\d}{\d A_{j_1,i_1}} (- c_g(A,x)^{\top} \cdot  f(A,x)_{j_0}\diag(x) A^{\top} f(A,x) \cdot(\langle -f(A,x), c(A,x) \rangle + f(A,x)_{j_0})) 
    \end{align*}
    Then, we define $B_{6,1}^{j_1,i_1,j_0,i_0}, \cdots, B_{6,7}^{j_1,i_1,j_0,i_0}$ as follow
    \begin{align*}
         B_{6,1}^{j_1,i_1,j_0,i_0}  := & ~ \frac{\d}{\d A_{j_1,i_1}} (- c_g(A,x)^{\top} )    \cdot f(A,x)_{j_0} \cdot \diag(x) \cdot A^{\top} \cdot f(A,x) \cdot (\langle -f(A,x), f(A,x) \rangle + f(A,x)_{j_0})\\
B_{6,2}^{j_1,i_1,j_0,i_0} := &   ~ - c_g(A,x)^{\top} \cdot \frac{\d}{\d A_{j_1,i_1}} (f(A,x)_{j_0}) \cdot \diag(x) \cdot A^{\top} \cdot f(A,x) \cdot (\langle -f(A,x), c(A,x) \rangle + f(A,x)_{j_0})\\
B_{6,3}^{j_1,i_1,j_0,i_0} := &   ~ - c_g(A,x)^{\top} \cdot f(A,x)_{j_0} \cdot \diag(x)  \cdot \frac{\d}{\d A_{j_1,i_1}} (A^{\top}) \cdot  f(A,x) \cdot (\langle -f(A,x), c(A,x) \rangle + f(A,x)_{j_0})\\
B_{6,4}^{j_1,i_1,j_0,i_0}:= &   ~ - c_g(A,x)^{\top} \cdot f(A,x)_{j_0} \cdot \diag(x) \cdot A^{\top} \cdot \frac{\d}{\d A_{j_1,i_1}}(f(A,x)) \cdot (\langle -f(A,x), c(A,x) \rangle + f(A,x)_{j_0})\\
B_{6,5}^{j_1,i_1,j_0,i_0}: = & ~   c_g(A,x)^{\top} \cdot f(A,x)_{j_0} \cdot \diag(x) \cdot A^{\top} \cdot  f(A,x)  \cdot \langle \frac{\d  f(A,x)}{\d A_{j_1,i_1}},c(A,x) \rangle \\
B_{6,6}^{j_1,i_1,j_0,i_0}: = & ~   c_g(A,x)^{\top} \cdot f(A,x)_{j_0} \cdot \diag(x) \cdot A^{\top} \cdot  f(A,x)  \cdot \langle f(A,x), \frac{\d c(A,x) }{\d A_{j_1,i_1}}\rangle \\
B_{6,7}^{j_1,i_1,j_0,i_0}: = & ~  - c_g(A,x)^{\top} \cdot f(A,x)_{j_0} \cdot \diag(x) \cdot A^{\top} \cdot  f(A,x)  \cdot  \frac{\d f(A,x)_{j_1}}{\d A_{j_1,i_1}} 
    \end{align*}
    It is easy to show
    \begin{align*}
        B_6^{j_1,i_1,j_0,i_0} = B_{6,1}^{j_1,i_1,j_0,i_0} +  B_{6,2}^{j_1,i_1,j_0,i_0} + B_{6,3}^{j_1,i_1,j_0,i_0} + B_{6,4}^{j_1,i_1,j_0,i_0} + B_{6,5}^{j_1,i_1,j_0,i_0} + B_{6,6}^{j_1,i_1,j_0,i_0} + B_{6,7}^{j_1,i_1,j_0,i_0}
    \end{align*}
           Similarly for $j_1 = j_0$ and $i_0 = i_1$,we have
    \begin{align*}
        B_6^{j_1,i_1,j_1,i_1} = B_{6,1}^{j_1,i_1,j_1,i_1} +  B_{6,2}^{j_1,i_1,j_1,i_1} + B_{6,3}^{j_1,i_1,j_1,i_1} + B_{6,4}^{j_1,i_1,j_1,i_1} + B_{6,5}^{j_1,i_1,j_1,i_1} + B_{6,6}^{j_1,i_1,j_1,i_1} + B_{6,7}^{j_1,i_1,j_1,i_1}
         \end{align*}
    For $j_1 = j_0$ and $i_0 \neq i_1$,we have
    \begin{align*}
       B_6^{j_1,i_1,j_1,i_0} = B_{6,1}^{j_1,i_1,j_1,i_0} +  B_{6,2}^{j_1,i_1,j_1,i_0} + B_{6,3}^{j_1,i_1,j_1,i_0} + B_{6,4}^{j_1,i_1,j_1,i_0} + B_{6,5}^{j_1,i_1,j_1,i_0} + B_{6,6}^{j_1,i_1,j_1,i_0} + B_{6,7}^{j_1,i_1,j_1,i_0}
    \end{align*}
    For $j_1 \neq j_0$ and $i_0 = i_1$,we have
    \begin{align*}
       B_6^{j_1,i_1,j_0,i_1} = B_{6,1}^{j_1,i_1,j_0,i_1} +  B_{6,2}^{j_1,i_1,j_0,i_1} + B_{6,3}^{j_1,i_1,j_0,i_1} + B_{6,4}^{j_1,i_1,j_0,i_1} + B_{6,5}^{j_1,i_1,j_0,i_1} + B_{6,6}^{j_1,i_1,j_0,i_1} + B_{6,7}^{j_1,i_1,j_0,i_1}
    \end{align*}
\end{definition}

\subsection{Case \texorpdfstring{$j_1=j_0, i_1 = i_0$}{}}
\begin{lemma}
For $j_1 = j_0$ and $i_0 = i_1$. If the following conditions hold
    \begin{itemize}
     \item Let $u(A,x) \in \R^n$ be defined as Definition~\ref{def:u}
    \item Let $\alpha(A,x) \in \R$ be defined as Definition~\ref{def:alpha}
     \item Let $f(A,x) \in \R^n$ be defined as Definition~\ref{def:f}
    \item Let $c(A,x) \in \R^n$ be defined as Definition~\ref{def:c}
    \item Let $g(A,x) \in \R^d$ be defined as Definition~\ref{def:g} 
    \item Let $q(A,x) = c(A,x) + f(A,x) \in \R^n$
    \item Let $c_g(A,x) \in \R^d$ be defined as Definition~\ref{def:c_g}.
    \item Let $L_g(A,x) \in \R$ be defined as Definition~\ref{def:l_g}
    \item Let $v \in \R^n$ be a vector 
    \item Let $B_1^{j_1,i_1,j_0,i_0}$ be defined as Definition~\ref{def:b_1}
    \end{itemize}
    Then, For $j_0,j_1 \in [n], i_0,i_1 \in [d]$, we have 
    \begin{itemize}
\item {\bf Part 1.} For $B_{6,1}^{j_1,i_1,j_1,i_1}$, we have 
\begin{align*}
 B_{6,1}^{j_1,i_1,j_1,i_1}  = & ~  \frac{\d}{\d A_{j_1,i_1}} (- c_g(A,x)^{\top} ) \cdot  f(A,x)_{j_1} \cdot \diag (x) A^{\top} \cdot  f(A,x)  \cdot  (\langle -f(A,x), c(A,x) \rangle + f(A,x)_{j_1})\\
 = & ~ B_{6,1,1}^{j_1,i_1,j_1,i_1} + B_{6,1,2}^{j_1,i_1,j_1,i_1} + B_{6,1,3}^{j_1,i_1,j_1,i_1} + B_{6,1,4}^{j_1,i_1,j_1,i_1} + B_{6,1,5}^{j_1,i_1,j_1,i_1} + B_{6,1,6}^{j_1,i_1,j_1,i_1} + B_{6,1,7}^{j_1,i_1,j_1,i_1}
\end{align*} 
\item {\bf Part 2.} For $B_{6,2}^{j_1,i_1,j_1,i_1}$, we have 
\begin{align*}
  B_{6,2}^{j_1,i_1,j_1,i_1} = & ~ - c_g(A,x)^{\top} \cdot \frac{\d}{\d A_{j_1,i_1}} ( f(A,x)_{j_1} )  \cdot \diag(x) \cdot A^{\top} \cdot f(A,x) \cdot(\langle -f(A,x), c(A,x) \rangle + f(A,x)_{j_1}) \\
    = & ~  B_{6,2,1}^{j_1,i_1,j_1,i_1} + B_{6,2,2}^{j_1,i_1,j_1,i_1}
\end{align*} 
\item {\bf Part 3.} For $B_{6,3}^{j_1,i_1,j_1,i_1}$, we have 
\begin{align*}
  B_{6,3}^{j_1,i_1,j_1,i_1} = & ~ - c_g(A,x)^{\top} \cdot f(A,x)_{j_1} \cdot \diag(x) \cdot \frac{\d}{\d A_{j_1,i_1}} (  A^{\top} ) \cdot f(A,x) \cdot(\langle -f(A,x), c(A,x) \rangle + f(A,x)_{j_1}) \\
     = & ~ B_{6,3,1}^{j_1,i_1,j_1,i_1}  
\end{align*} 
\item {\bf Part 4.} For $B_{6,4}^{j_1,i_1,j_1,i_1}$, we have 
\begin{align*}
  B_{6,4}^{j_1,i_1,j_1,i_1} = & ~ - c_g(A,x)^{\top} \cdot f(A,x)_{j_1} \cdot \diag(x) \cdot A^{\top} \cdot \frac{\d f(A,x)}{\d A_{j_1,i_1}} \cdot(\langle -f(A,x), c(A,x) \rangle + f(A,x)_{j_1}) \\
     = & ~ B_{6,4,1}^{j_1,i_1,j_1,i_1} 
\end{align*}
\item {\bf Part 5.} For $B_{6,5}^{j_1,i_1,j_1,i_1}$, we have 
\begin{align*}
  B_{6,5}^{j_1,i_1,j_1,i_1} = & ~  c_g(A,x)^{\top} \cdot f(A,x)_{j_1} \cdot \diag(x) \cdot A^{\top} \cdot  f(A,x)  \cdot \langle \frac{\d  f(A,x) }{\d A_{j_1,i_1}},c(A,x)\rangle \\
     = & ~ B_{6,5,1}^{j_1,i_1,j_1,i_1}  + B_{6,5,2}^{j_1,i_1,j_1,i_1}  
\end{align*}
\item {\bf Part 6.} For $B_{5,6}^{j_1,i_1,j_1,i_1}$, we have 
\begin{align*}
  B_{6,6}^{j_1,i_1,j_1,i_1} = & ~ c_g(A,x)^{\top} \cdot f(A,x)_{j_1} \cdot \diag(x) \cdot A^{\top} \cdot  f(A,x)  \cdot \langle f(A,x), \frac{\d c(A,x) }{\d A_{j_1,i_1}}\rangle \\
     = & ~ B_{6,6,1}^{j_1,i_1,j_1,i_1}  + B_{6,6,2}^{j_1,i_1,j_1,i_1}  
\end{align*}
\item {\bf Part 7.} For $B_{6,7}^{j_1,i_1,j_1,i_1}$, we have 
\begin{align*}
  B_{6,7}^{j_1,i_1,j_1,i_1} = & ~ - c_g(A,x)^{\top} \cdot f(A,x)_{j_1} \cdot \diag(x) \cdot A^{\top} \cdot  f(A,x)  \cdot  \frac{\d f(A,x)_{j_1}}{\d A_{j_1,i_1}} \\
     = & ~ B_{6,7,1}^{j_1,i_1,j_1,i_1}  + B_{6,7,2}^{j_1,i_1,j_1,i_1}  
\end{align*}
\end{itemize}
\begin{proof}
    {\bf Proof of Part 1.}
    \begin{align*}
    B_{6,1,1}^{j_1,i_1,j_1,i_1} : = & ~ e_{i_1}^\top \cdot \langle c(A,x), f(A,x) \rangle \cdot  f(A,x)_{j_1}^2 \cdot \diag(x) \cdot A^{\top} \cdot f(A,x) \cdot(\langle -f(A,x), c(A,x) \rangle + f(A,x)_{j_1})\\
    B_{6,1,2}^{j_1,i_1,j_1,i_1} : = & ~   e_{i_1}^\top \cdot c(A,x)_{j_1}\cdot f(A,x)_{j_1}^2 \cdot \diag(x) \cdot A^{\top} \cdot f(A,x) \cdot(\langle -f(A,x), c(A,x) \rangle + f(A,x)_{j_1})\\
    B_{6,1,3}^{j_1,i_1,j_1,i_1} : = & ~ f(A,x)_{j_1}^2 \cdot \langle c(A,x), f(A,x) \rangle \cdot ( (A_{j_1,*}) \circ x^\top  )  \cdot \diag(x)  \\
        & ~ \cdot A^{\top} \cdot f(A,x) \cdot (\langle -f(A,x), c(A,x) \rangle + f(A,x)_{j_1})\\
    B_{6,1,4}^{j_1,i_1,j_1,i_1} : = & ~   -  f(A,x)_{j_1}^2 \cdot f(A,x)^\top  \cdot A \cdot (\diag(x))^2 \cdot   \langle c(A,x), f(A,x) \rangle \cdot A^{\top} \cdot f(A,x) \\
    & ~ \cdot(\langle -f(A,x), c(A,x) \rangle + f(A,x)_{j_1})\\
    B_{6,1,5}^{j_1,i_1,j_1,i_1} : = & ~    f(A,x)_{j_1}^2 \cdot f(A,x)^\top  \cdot A \cdot (\diag(x))^2 \cdot (\langle (-f(A,x)), f(A,x) \rangle + f(A,x)_{j_1})   \cdot A^{\top} \cdot f(A,x) \\
    &~\cdot (\langle -f(A,x), c(A,x) \rangle + f(A,x)_{j_1}) \\
    B_{6,1,6}^{j_1,i_1,j_1,i_1} : = & ~  f(A,x)_{j_1}^2 \cdot f(A,x)^\top  \cdot A \cdot  (\diag(x))^2 \cdot(\langle -f(A,x), c(A,x) \rangle + f(A,x)_{j_1})^2 \cdot A^{\top} \cdot f(A,x) \\
    B_{6,1,7}^{j_1,i_1,j_1,i_1} : = & ~   f(A,x)_{j_1}^2 \cdot ((e_{j_1}^\top - f(A,x)^\top) \circ q(A,x)^\top) \cdot A \cdot  (\diag(x))^2 \cdot A^{\top}  \\
        & ~\cdot f(A,x) \cdot(\langle -f(A,x), c(A,x) \rangle + f(A,x)_{j_1})
\end{align*}
Finally, combine them and we have
\begin{align*}
       B_{6,1}^{j_1,i_1,j_1,i_1} =  B_{6,1,1}^{j_1,i_1,j_1,i_1} + B_{6,1,2}^{j_1,i_1,j_1,i_1} + B_{6,1,3}^{j_1,i_1,j_1,i_1} + B_{6,1,4}^{j_1,i_1,j_1,i_1} + B_{6,1,5}^{j_1,i_1,j_1,i_1} + B_{6,1,6}^{j_1,i_1,j_1,i_1} + B_{6,1,7}^{j_1,i_1,j_1,i_1}
\end{align*}
{\bf Proof of Part 2.}
    \begin{align*}
    B_{6,2,1}^{j_1,i_1,j_1,i_1} : = & ~   f(A,x)_{j_1}^2 \cdot x_{i_1} \cdot c_g(A,x)^{\top} \cdot \diag(x) \cdot A^{\top} \cdot f(A,x) \cdot(\langle -f(A,x), c(A,x) \rangle + f(A,x)_{j_1}) \\
    B_{6,2,2}^{j_1,i_1,j_1,i_1} : = & ~ - f(A,x)_{j_1} \cdot x_{i_1} \cdot c_g(A,x)^{\top} \cdot \diag(x) \cdot A^{\top} \cdot f(A,x) \cdot(\langle -f(A,x), c(A,x) \rangle + f(A,x)_{j_1})
\end{align*}
Finally, combine them and we have
\begin{align*}
       B_{6,2}^{j_1,i_1,j_1,i_1} = B_{6,2,1}^{j_1,i_1,j_1,i_1} + B_{6,2,2}^{j_1,i_1,j_1,i_1}
\end{align*}
{\bf Proof of Part 3.} 
    \begin{align*}
    B_{6,3,1}^{j_1,i_1,j_1,i_1} : = & ~    -  c_g(A,x)^{\top} \cdot f(A,x)_{j_1} \cdot \diag(x) \cdot e_{i_1} \cdot e_{j_1}^\top \cdot f(A,x) \cdot(\langle -f(A,x), c(A,x) \rangle + f(A,x)_{j_1})
\end{align*}
Finally, combine them and we have
\begin{align*}
       B_{6,3}^{j_1,i_1,j_1,i_1} = B_{6,3,1}^{j_1,i_1,j_1,i_1} 
\end{align*}
{\bf Proof of Part 4.} 
    \begin{align*}
    B_{6,4,1}^{j_1,i_1,j_1,i_1} : = & ~ - c_g(A,x)^{\top} \cdot f(A,x)_{j_1}^2 \cdot \diag(x) \cdot A^{\top}\cdot x_i \cdot  (e_{j_1}- f(A,x) ) \cdot (\langle -f(A,x), c(A,x) \rangle + f(A,x)_{j_1})
\end{align*}
Finally, combine them and we have
\begin{align*}
       B_{6,4}^{j_1,i_1,j_1,i_1} = B_{6,4,1}^{j_1,i_1,j_1,i_1}  
\end{align*}
{\bf Proof of Part 5.} 
    \begin{align*}
    B_{6,5,1}^{j_1,i_1,j_1,i_1} : = & ~ c_g(A,x)^{\top} \cdot f(A,x)_{j_1}^2 \cdot \diag(x) \cdot A^{\top} \cdot  f(A,x)  \cdot  x_{i_1} \cdot \langle - f(A,x), f(A,x) \rangle\\
    B_{6,5,2}^{j_1,i_1,j_1,i_1} : = & ~ c_g(A,x)^{\top} \cdot f(A,x)_{j_1}^3 \cdot \diag(x) \cdot A^{\top} \cdot  f(A,x)  \cdot  x_{i_1}  
\end{align*}
Finally, combine them and we have
\begin{align*}
       B_{6,5}^{j_1,i_1,j_1,i_1} = B_{6,5,1}^{j_1,i_1,j_1,i_1}  +B_{6,5,2}^{j_1,i_1,j_1,i_1}
\end{align*}
{\bf Proof of Part 6.} 
    \begin{align*}
     B_{6,6,1}^{j_1,i_1,j_1,i_1} : = & ~ c_g(A,x)^{\top} \cdot f(A,x)_{j_1}^2 \cdot \diag(x) \cdot A^{\top} \cdot  f(A,x)  \cdot  x_{i_1} \cdot \langle - f(A,x), c(A,x) \rangle\\
    B_{6,6,2}^{j_1,i_1,j_1,i_1} : = & ~ c_g(A,x)^{\top} \cdot f(A,x)_{j_1}^2 \cdot c(A,x)_{j_1} \cdot \diag(x) \cdot A^{\top} \cdot  f(A,x)  \cdot  x_{i_1}  
\end{align*}
Finally, combine them and we have
\begin{align*}
       B_{6,6}^{j_1,i_1,j_1,i_1} = B_{6,6,1}^{j_1,i_1,j_1,i_1}  +B_{6,6,2}^{j_1,i_1,j_1,i_1}
\end{align*}
{\bf Proof of Part 7.} 
    \begin{align*}
     B_{6,7,1}^{j_1,i_1,j_1,i_1} : = & ~  f(A,x)_{j_1}^3 \cdot  x_{i_1}\cdot c_g(A,x)^{\top} \cdot \diag(x) A^{\top} \cdot  f(A,x)   \\
    B_{6,7,2}^{j_1,i_1,j_1,i_1} : = & ~ -  f(A,x)_{j_1}^2 \cdot  x_{i_1} c_g(A,x)^{\top} \cdot \diag(x) \cdot A^{\top} \cdot  f(A,x)  
\end{align*}
Finally, combine them and we have
\begin{align*}
       B_{6,7}^{j_1,i_1,j_1,i_1} = B_{6,7,1}^{j_1,i_1,j_1,i_1}  +B_{6,7,2}^{j_1,i_1,j_1,i_1}
\end{align*}
\end{proof}
\end{lemma}

\subsection{Case \texorpdfstring{$j_1=j_0, i_1 \neq i_0$}{}}
\begin{lemma}
For $j_1 = j_0$ and $i_0 \neq i_1$. If the following conditions hold
    \begin{itemize}
     \item Let $u(A,x) \in \R^n$ be defined as Definition~\ref{def:u}
    \item Let $\alpha(A,x) \in \R$ be defined as Definition~\ref{def:alpha}
     \item Let $f(A,x) \in \R^n$ be defined as Definition~\ref{def:f}
    \item Let $c(A,x) \in \R^n$ be defined as Definition~\ref{def:c}
    \item Let $g(A,x) \in \R^d$ be defined as Definition~\ref{def:g} 
    \item Let $q(A,x) = c(A,x) + f(A,x) \in \R^n$
    \item Let $c_g(A,x) \in \R^d$ be defined as Definition~\ref{def:c_g}.
    \item Let $L_g(A,x) \in \R$ be defined as Definition~\ref{def:l_g}
    \item Let $v \in \R^n$ be a vector 
    \item Let $B_1^{j_1,i_1,j_0,i_0}$ be defined as Definition~\ref{def:b_1}
    \end{itemize}
    Then, For $j_0,j_1 \in [n], i_0,i_1 \in [d]$, we have 
    \begin{itemize}
\item {\bf Part 1.} For $B_{6,1}^{j_1,i_1,j_1,i_1}$, we have 
\begin{align*}
 B_{6,1}^{j_1,i_1,j_1,i_0}  = & ~  \frac{\d}{\d A_{j_1,i_1}} (- c_g(A,x)^{\top} ) \cdot  f(A,x)_{j_1} \cdot \diag (x) A^{\top} \cdot  f(A,x)  \cdot  (\langle -f(A,x), c(A,x) \rangle + f(A,x)_{j_1})\\
 = & ~ B_{6,1,1}^{j_1,i_1,j_1,i_0} + B_{6,1,2}^{j_1,i_1,j_1,i_0} + B_{6,1,3}^{j_1,i_1,j_1,i_0} + B_{6,1,4}^{j_1,i_1,j_1,i_0} + B_{6,1,5}^{j_1,i_1,j_1,i_0} + B_{6,1,6}^{j_1,i_1,j_1,i_0} + B_{6,1,7}^{j_1,i_1,j_1,i_0}
\end{align*} 
\item {\bf Part 2.} For $B_{6,2}^{j_1,i_1,j_1,i_0}$, we have 
\begin{align*}
  B_{6,2}^{j_1,i_1,j_1,i_0} = & ~ - c_g(A,x)^{\top} \cdot \frac{\d}{\d A_{j_1,i_1}} ( f(A,x)_{j_1} )  \cdot \diag(x) \cdot A^{\top} \cdot f(A,x) \cdot(\langle -f(A,x), c(A,x) \rangle + f(A,x)_{j_1}) \\
    = & ~  B_{6,2,1}^{j_1,i_1,j_1,i_0} + B_{6,2,2}^{j_1,i_1,j_1,i_0}
\end{align*} 
\item {\bf Part 3.} For $B_{6,3}^{j_1,i_1,j_1,i_0}$, we have 
\begin{align*}
  B_{6,3}^{j_1,i_1,j_1,i_0} = & ~ - c_g(A,x)^{\top} \cdot f(A,x)_{j_1} \cdot \diag(x) \cdot \frac{\d}{\d A_{j_1,i_1}} (  A^{\top} ) \cdot f(A,x) \cdot(\langle -f(A,x), c(A,x) \rangle + f(A,x)_{j_1}) \\
     = & ~ B_{6,3,1}^{j_1,i_1,j_1,i_0}  
\end{align*} 
\item {\bf Part 4.} For $B_{6,4}^{j_1,i_1,j_1,i_0}$, we have 
\begin{align*}
  B_{6,4}^{j_1,i_1,j_1,i_1} = & ~ - c_g(A,x)^{\top} \cdot f(A,x)_{j_1} \cdot \diag(x) \cdot A^{\top} \cdot \frac{\d f(A,x)}{\d A_{j_1,i_1}} \cdot(\langle -f(A,x), c(A,x) \rangle + f(A,x)_{j_1}) \\
     = & ~ B_{6,4,1}^{j_1,i_1,j_1,i_0} 
\end{align*}
\item {\bf Part 5.} For $B_{6,5}^{j_1,i_1,j_1,i_0}$, we have 
\begin{align*}
  B_{6,5}^{j_1,i_1,j_1,i_0} = & ~  c_g(A,x)^{\top} \cdot f(A,x)_{j_1} \cdot \diag(x) \cdot A^{\top} \cdot  f(A,x)  \cdot \langle \frac{\d  f(A,x) }{\d A_{j_1,i_1}},c(A,x)\rangle \\
     = & ~ B_{6,5,1}^{j_1,i_1,j_1,i_0}  + B_{6,5,2}^{j_1,i_1,j_1,i_0}  
\end{align*}
\item {\bf Part 6.} For $B_{5,6}^{j_1,i_1,j_1,i_0}$, we have 
\begin{align*}
  B_{6,6}^{j_1,i_1,j_1,i_0} = & ~ c_g(A,x)^{\top} \cdot f(A,x)_{j_1} \cdot \diag(x) \cdot A^{\top} \cdot  f(A,x)  \cdot \langle f(A,x), \frac{\d c(A,x) }{\d A_{j_1,i_1}}\rangle \\
     = & ~ B_{6,6,1}^{j_1,i_1,j_1,i_0}  + B_{6,6,2}^{j_1,i_1,j_1,i_0}  
\end{align*}
\item {\bf Part 7.} For $B_{6,7}^{j_1,i_1,j_1,i_0}$, we have 
\begin{align*}
  B_{6,7}^{j_1,i_1,j_1,i_0} = & ~ - c_g(A,x)^{\top} \cdot f(A,x)_{j_1} \cdot \diag(x) \cdot A^{\top} \cdot  f(A,x)  \cdot  \frac{\d f(A,x)_{j_1}}{\d A_{j_1,i_1}} \\
     = & ~ B_{6,7,1}^{j_1,i_1,j_1,i_0}  + B_{6,7,2}^{j_1,i_1,j_1,i_0}  
\end{align*}
\end{itemize}
\begin{proof}
    {\bf Proof of Part 1.}
    \begin{align*}
    B_{6,1,1}^{j_1,i_1,j_1,i_0} : = & ~ e_{i_1}^\top \cdot \langle c(A,x), f(A,x) \rangle \cdot  f(A,x)_{j_1}^2 \cdot \diag(x) \cdot A^{\top} \cdot f(A,x) \cdot(\langle -f(A,x), c(A,x) \rangle + f(A,x)_{j_1})\\
    B_{6,1,2}^{j_1,i_1,j_1,i_0} : = & ~   e_{i_1}^\top \cdot c(A,x)_{j_1}\cdot f(A,x)_{j_1}^2 \cdot \diag(x) \cdot A^{\top} \cdot f(A,x) \cdot(\langle -f(A,x), c(A,x) \rangle + f(A,x)_{j_1})\\
    B_{6,1,3}^{j_1,i_1,j_1,i_0} : = & ~ f(A,x)_{j_1}^2 \cdot \langle c(A,x), f(A,x) \rangle \cdot ( (A_{j_1,*}) \circ x^\top  )  \cdot \diag(x) \cdot A^{\top}  \\
        & ~\cdot f(A,x) \cdot (\langle -f(A,x), c(A,x) \rangle + f(A,x)_{j_1})\\
    B_{6,1,4}^{j_1,i_1,j_1,i_0} : = & ~   -  f(A,x)_{j_1}^2 \cdot f(A,x)^\top  \cdot A \cdot (\diag(x))^2 \cdot   \langle c(A,x), f(A,x) \rangle \cdot A^{\top} \cdot f(A,x) \\
    & ~ \cdot(\langle -f(A,x), c(A,x) \rangle + f(A,x)_{j_1})\\
    B_{6,1,5}^{j_1,i_1,j_1,i_0} : = & ~    f(A,x)_{j_1}^2 \cdot f(A,x)^\top  \cdot A \cdot (\diag(x))^2 \cdot (\langle (-f(A,x)), f(A,x) \rangle + f(A,x)_{j_1})   \cdot A^{\top} \cdot f(A,x) \\
    &~\cdot (\langle -f(A,x), c(A,x) \rangle + f(A,x)_{j_1}) \\
    B_{6,1,6}^{j_1,i_1,j_1,i_0} : = & ~  f(A,x)_{j_1}^2 \cdot f(A,x)^\top  \cdot A \cdot  (\diag(x))^2 \cdot(\langle -f(A,x), c(A,x) \rangle + f(A,x)_{j_1})^2 \cdot A^{\top} \cdot f(A,x) \\
    B_{6,1,7}^{j_1,i_1,j_1,i_0} : = & ~   f(A,x)_{j_1}^2 \cdot ((e_{j_1}^\top - f(A,x)^\top) \circ q(A,x)^\top) \cdot A \cdot  (\diag(x))^2 \cdot A^{\top} \cdot f(A,x)\\
    & ~ \cdot(\langle -f(A,x), c(A,x) \rangle + f(A,x)_{j_1})
\end{align*}
Finally, combine them and we have
\begin{align*}
       B_{6,1}^{j_1,i_1,j_1,i_0} = B_{6,1,1}^{j_1,i_1,j_1,i_0} + B_{6,1,2}^{j_1,i_1,j_1,i_0} + B_{6,1,3}^{j_1,i_1,j_1,i_0} + B_{6,1,4}^{j_1,i_1,j_1,i_0} + B_{6,1,5}^{j_1,i_1,j_1,i_0} + B_{6,1,6}^{j_1,i_1,j_1,i_0} + B_{6,1,7}^{j_1,i_1,j_1,i_0}
\end{align*}
{\bf Proof of Part 2.}
    \begin{align*}
    B_{6,2,1}^{j_1,i_1,j_1,i_0} : = & ~   f(A,x)_{j_1}^2 \cdot x_{i_1} \cdot c_g(A,x)^{\top} \cdot \diag(x) \cdot A^{\top} \cdot f(A,x) \cdot(\langle -f(A,x), c(A,x) \rangle + f(A,x)_{j_1}) \\
    B_{6,2,2}^{j_1,i_1,j_1,i_0} : = & ~ - f(A,x)_{j_1} \cdot x_{i_1} \cdot c_g(A,x)^{\top} \cdot \diag(x) \cdot A^{\top} \cdot f(A,x) \cdot(\langle -f(A,x), c(A,x) \rangle + f(A,x)_{j_1})
\end{align*}
Finally, combine them and we have
\begin{align*}
       B_{6,2}^{j_1,i_1,j_1,i_0} = B_{6,2,1}^{j_1,i_1,j_1,i_0} + B_{6,2,2}^{j_1,i_1,j_1,i_0}
\end{align*}
{\bf Proof of Part 3.} 
    \begin{align*}
    B_{6,3,1}^{j_1,i_1,j_1,i_0} : = & ~    -  c_g(A,x)^{\top} \cdot f(A,x)_{j_1} \cdot \diag(x) \cdot e_{i_1} \cdot e_{j_1}^\top \cdot f(A,x) \cdot(\langle -f(A,x), c(A,x) \rangle + f(A,x)_{j_1})
\end{align*}
Finally, combine them and we have
\begin{align*}
       B_{6,3}^{j_1,i_1,j_1,i_0} = B_{6,3,1}^{j_1,i_1,j_1,i_0} 
\end{align*}
{\bf Proof of Part 4.} 
    \begin{align*}
    B_{6,4,1}^{j_1,i_1,j_1,i_0} : = & ~ - c_g(A,x)^{\top} \cdot f(A,x)_{j_1}^2 \cdot \diag(x) \cdot A^{\top}\cdot x_i \cdot  (e_{j_1}- f(A,x) ) \cdot (\langle -f(A,x), c(A,x) \rangle + f(A,x)_{j_1})
\end{align*}
Finally, combine them and we have
\begin{align*}
       B_{6,4}^{j_1,i_1,j_1,i_0} = B_{6,4,1}^{j_1,i_1,j_1,i_0}  
\end{align*}
{\bf Proof of Part 5.} 
    \begin{align*}
    B_{6,5,1}^{j_1,i_1,j_1,i_0} : = & ~ c_g(A,x)^{\top} \cdot f(A,x)_{j_1}^2 \cdot \diag(x) \cdot A^{\top} \cdot  f(A,x)  \cdot  x_{i_1} \cdot \langle - f(A,x), f(A,x) \rangle\\
    B_{6,5,2}^{j_1,i_1,j_1,i_0} : = & ~ c_g(A,x)^{\top} \cdot f(A,x)_{j_1}^3 \cdot \diag(x) \cdot A^{\top} \cdot  f(A,x)  \cdot  x_{i_1}  
\end{align*}
Finally, combine them and we have
\begin{align*}
       B_{6,5}^{j_1,i_1,j_1,i_0} = B_{6,5,1}^{j_1,i_1,j_1,i_0}  +B_{6,5,2}^{j_1,i_1,j_1,i_0}
\end{align*}
{\bf Proof of Part 6.} 
    \begin{align*}
     B_{6,6,1}^{j_1,i_1,j_1,i_0} : = & ~ c_g(A,x)^{\top} \cdot f(A,x)_{j_1}^2 \cdot \diag(x) \cdot A^{\top} \cdot  f(A,x)  \cdot  x_{i_1} \cdot \langle - f(A,x), c(A,x) \rangle\\
    B_{6,6,2}^{j_1,i_1,j_1,i_0} : = & ~ c_g(A,x)^{\top} \cdot f(A,x)_{j_1}^2 \cdot c(A,x)_{j_1} \cdot \diag(x) \cdot A^{\top} \cdot  f(A,x)  \cdot  x_{i_1}  
\end{align*}
Finally, combine them and we have
\begin{align*}
       B_{6,6}^{j_1,i_1,j_1,i_0} = B_{6,6,1}^{j_1,i_1,j_1,i_0}  +B_{6,6,2}^{j_1,i_1,j_1,i_0}
\end{align*}
{\bf Proof of Part 7.} 
    \begin{align*}
     B_{6,7,1}^{j_1,i_1,j_1,i_0} : = & ~  f(A,x)_{j_1}^3 \cdot  x_{i_1}\cdot c_g(A,x)^{\top} \cdot \diag(x) A^{\top} \cdot  f(A,x)   \\
    B_{6,7,2}^{j_1,i_1,j_1,i_0} : = & ~ -  f(A,x)_{j_1}^2 \cdot  x_{i_1} c_g(A,x)^{\top} \cdot \diag(x) \cdot A^{\top} \cdot  f(A,x)  
\end{align*}
Finally, combine them and we have
\begin{align*}
       B_{6,7}^{j_1,i_1,j_1,i_0} = B_{6,7,1}^{j_1,i_1,j_1,i_0}  +B_{6,7,2}^{j_1,i_1,j_1,i_0}
\end{align*}
\end{proof}
\end{lemma}

\subsection{Constructing \texorpdfstring{$d \times d$}{} matrices for \texorpdfstring{$j_1 = j_0$}{}}
The purpose of the following lemma is to let $i_0$ and $i_1$ disappear.
\begin{lemma}For $j_0,j_1 \in [n]$, a list of $d \times d$ matrices can be expressed as the following sense,
\begin{itemize}
\item {\bf Part 1.}
\begin{align*}
B_{6,1,1}^{j_1,*,j_1,*} & ~ =   f_c(A,x) \cdot  f(A,x)_{j_1}^2 \cdot  (-f_c(A,x) + f(A,x)_{j_1}) \cdot h(A,x) \cdot {\bf 1}_d^\top
\end{align*}
\item {\bf Part 2.}
\begin{align*}
B_{6,1,2}^{j_1,*,j_1,*} & ~ =    c(A,x)_{j_1}\cdot f(A,x)_{j_1}^2 \cdot  (-f_c(A,x) + f(A,x)_{j_1}) \cdot h(A,x) \cdot {\bf 1}_d^\top 
\end{align*}
\item {\bf Part 3.}
\begin{align*}
B_{6,1,3}^{j_1,*,j_1,*} & ~ =    f(A,x)_{j_1}^2 \cdot f_c(A,x) \cdot (-f_c(A,x) + f(A,x)_{j_1}) \cdot ( (A_{j_1,*}) \circ x^\top  )  \cdot h(A,x)\cdot I_d
\end{align*}
\item {\bf Part 4.}
\begin{align*}
B_{6,1,4}^{j_1,*,j_1,*}  & ~ =   -  f(A,x)_{j_1}^2 \cdot f_c(A,x)  \cdot(-f_c(A,x) + f(A,x)_{j_1}) \cdot h(A,x)^\top \cdot h(A,x) \cdot I_d
\end{align*}
\item {\bf Part 5.}
\begin{align*}
B_{6,1,5}^{j_1,*,j_1,*}  & ~ =   f(A,x)_{j_1}^2 \cdot (-f_2(A,x) + f(A,x)_{j_1}) \cdot (-f_c(A,x) + f(A,x)_{j_1}) \cdot h(A,x)^\top \cdot h(A,x) \cdot I_d
\end{align*}
\item {\bf Part 6.}
\begin{align*}
B_{6,1,6}^{j_1,*,j_1,*}  & ~ =     f(A,x)_{j_1}^2 \cdot  (-f_c(A,x) + f(A,x)_{j_1})^2 \cdot h(A,x)^\top \cdot h(A,x) \cdot I_d
\end{align*}
\item {\bf Part 7.}
\begin{align*}
B_{6,1,7}^{j_1,*,j_1,*}  & ~ =   f(A,x)_{j_1}^2 \cdot (-f_c(A,x) + f(A,x)_{j_1})\cdot p_{j_1}(A,x)^\top \cdot h(A,x) \cdot I_d
\end{align*}
\item {\bf Part 8.}
\begin{align*}
B_{6,2,1}^{j_1,*,j_1,*}  & ~ =     f(A,x)_{j_1}^2 \cdot  (-f_c(A,x) + f(A,x)_{j_1}) \cdot c_g(A,x)^{\top} \cdot h(A,x) \cdot x \cdot {\bf 1}_d^{\top} 
\end{align*}
\item {\bf Part 9.}
\begin{align*}
B_{6,2,2}^{j_1,*,j_1,*}  & ~ =    -f(A,x)_{j_1}  \cdot  (-f_c(A,x) + f(A,x)_{j_1}) \cdot c_g(A,x)^{\top} \cdot h(A,x) \cdot x \cdot {\bf 1}_d^{\top} 
\end{align*}
\item {\bf Part 10.}
\begin{align*}
 B_{6,3,1}^{j_1,*,j_1,*}  & ~ =      - f(A,x)_{j_1}^2 \cdot (-f_c(A,x) + f(A,x)_{j_1})\cdot {\bf 1}_d \cdot c_g(A,x)^{\top}   \cdot \diag(x)  
\end{align*}
\item {\bf Part 11.}
\begin{align*}
B_{6,4,1}^{j_1,*,j_1,*}  =     - f(A,x)_{j_1}^2 \cdot (-f_c(A,x) + f(A,x)_{j_1}) \cdot c_g(A,x)^{\top}  \cdot  h_e(A,x)\cdot x \cdot {\bf 1}_d^{\top}
\end{align*}
\item {\bf Part 12.}
\begin{align*}
B_{6,5,1}^{j_1,*,j_1,*}  & ~ =    -  f(A,x)_{j_1}^2 \cdot f_2(A,x) \cdot c_g(A,x)^{\top} \cdot h(A,x)  \cdot x \cdot {\bf 1}_d^{\top}
\end{align*}
\item {\bf Part 13.}
\begin{align*}
 B_{6,5,2}^{j_1,*,j_1,*}  =   f(A,x)_{j_1}^3  \cdot c_g(A,x)^{\top} \cdot h(A,x)  \cdot x \cdot {\bf 1}_d^{\top}
\end{align*}
\item {\bf Part 14.}
\begin{align*}
B_{6,6,1}^{j_1,*,j_1,*}  =   -  f(A,x)_{j_1}^2 \cdot f_c(A,x) \cdot c_g(A,x)^{\top} \cdot h(A,x)  \cdot x \cdot {\bf 1}_d^{\top}
\end{align*}
\item {\bf Part 15.}
\begin{align*}
B_{6,6,2}^{j_1,*,j_1,*}  =    f(A,x)_{j_1}^2 \cdot c(A,x)_{j_1} \cdot c_g(A,x)^{\top} \cdot h(A,x)  \cdot x \cdot {\bf 1}_d^{\top}
\end{align*}
\item {\bf Part 16.}
\begin{align*}
B_{6,7,1}^{j_1,*,j_1,*}  =   f(A,x)_{j_1}^3  \cdot c_g(A,x)^{\top} \cdot h(A,x)  \cdot x \cdot {\bf 1}_d^{\top}
\end{align*}
\item {\bf Part 17.}
\begin{align*}
B_{6,7,2}^{j_1,*,j_1,*}  =    f(A,x)_{j_1}^2  \cdot c_g(A,x)^{\top} \cdot h(A,x)  \cdot x \cdot {\bf 1}_d^{\top}
\end{align*}

\end{itemize}
\begin{proof}
{\bf Proof of Part 1.}
    We have
    \begin{align*}
        B_{6,1,1}^{j_1,i_1,j_1,i_1}  = & ~e_{i_1}^\top \cdot \langle c(A,x), f(A,x) \rangle \cdot  f(A,x)_{j_1}^2 \cdot \diag(x) \cdot A^{\top} \cdot f(A,x) \cdot(\langle -f(A,x), c(A,x) \rangle + f(A,x)_{j_1})\\
        B_{6,1,1}^{j_1,i_1,j_1,i_0}  = & ~ e_{i_1}^\top \cdot \langle c(A,x), f(A,x) \rangle \cdot  f(A,x)_{j_1}^2 \cdot \diag(x) \cdot A^{\top} \cdot f(A,x) \cdot(\langle -f(A,x), c(A,x) \rangle + f(A,x)_{j_1})
    \end{align*}
    From the above two equations, we can tell that $B_{6,1,1}^{j_1,*,j_1,*} \in \R^{d \times d}$ is a matrix that both the diagonal and off-diagonal have entries.
    
    Then we have $B_{6,1,1}^{j_1,*,j_1,*} \in \R^{d \times d}$ can be written as the rescaling of a diagonal matrix,
    \begin{align*}
     B_{6,1,1}^{j_1,*,j_1,*} & ~ = \langle c(A,x), f(A,x) \rangle \cdot  f(A,x)_{j_1}^2 \cdot  (\langle -f(A,x), c(A,x) \rangle + f(A,x)_{j_1}) \cdot \diag(x) \cdot A^{\top} \cdot f(A,x) \cdot {\bf 1}_d^\top \\
     & ~ = f_c(A,x) \cdot  f(A,x)_{j_1}^2 \cdot  (-f_c(A,x) + f(A,x)_{j_1}) \cdot h(A,x) \cdot {\bf 1}_d^\top
\end{align*}
    where the last step is follows from the Definitions~\ref{def:h} and Definitions~\ref{def:f_c}. 

{\bf Proof of Part 2.}
    We have
    \begin{align*}
           B_{6,1,2}^{j_1,i_1,j_1,i_1} = & ~ e_{i_1}^\top \cdot c(A,x)_{j_1}\cdot f(A,x)_{j_1}^2 \cdot \diag(x) \cdot A^{\top} \cdot f(A,x) \cdot(\langle -f(A,x), c(A,x) \rangle + f(A,x)_{j_1})\\
        B_{6,1,2}^{j_1,i_1,j_1,i_0} = & ~ e_{i_1}^\top \cdot c(A,x)_{j_1}\cdot f(A,x)_{j_1}^2 \cdot \diag(x) \cdot A^{\top} \cdot f(A,x) \cdot(\langle -f(A,x), c(A,x) \rangle + f(A,x)_{j_1})
    \end{align*}
     From the above two equations, we can tell that $B_{6,1,2}^{j_1,*,j_1,*} \in \R^{d \times d}$ is a matrix that only diagonal has entries and off-diagonal are all zeros.
    
    Then we have $B_{6,1,2}^{j_1,*,j_1,*} \in \R^{d \times d}$ can be written as the rescaling of a diagonal matrix,
\begin{align*}
     B_{6,1,2}^{j_1,*,j_1,*} & ~ = c(A,x)_{j_1}\cdot f(A,x)_{j_1}^2 \cdot  (\langle -f(A,x), c(A,x) \rangle + f(A,x)_{j_1}) \cdot \diag(x) \cdot A^{\top} \cdot f(A,x) \cdot {\bf 1}_d^\top \\
     & ~ =  c(A,x)_{j_1}\cdot f(A,x)_{j_1}^2 \cdot  (-f_c(A,x) + f(A,x)_{j_1}) \cdot h(A,x) \cdot {\bf 1}_d^\top 
\end{align*}
    where the last step is follows from the Definitions~\ref{def:h} and Definitions~\ref{def:f_c}.

{\bf Proof of Part 3.}
We have for diagonal entry and off-diagonal entry can be written as follows 
    \begin{align*}
        B_{6,1,3}^{j_1,i_1,j_1,i_1} = & ~f(A,x)_{j_1}^2 \cdot \langle c(A,x), f(A,x) \rangle \cdot ( (A_{j_1,*}) \circ x^\top  )  \cdot \diag(x) \cdot A^{\top}  \\
        & ~\cdot f(A,x) \cdot (\langle -f(A,x), c(A,x) \rangle + f(A,x)_{j_1}) \\
        B_{6,1,3}^{j_1,i_1,j_1,i_0} = & ~f(A,x)_{j_1}^2 \cdot \langle c(A,x), f(A,x) \rangle \cdot ( (A_{j_1,*}) \circ x^\top  )  \cdot \diag(x) \cdot A^{\top}  \\
        & ~\cdot f(A,x) \cdot (\langle -f(A,x), c(A,x) \rangle + f(A,x)_{j_1})
    \end{align*}
From the above equation, we can show that matrix $B_{6,1,3}^{j_1,*,j_1,*}$ can be expressed as a rank-$1$ matrix,
\begin{align*}
     B_{6,1,3}^{j_1,*,j_1,*} & ~ = f(A,x)_{j_1}^2 \cdot \langle c(A,x), f(A,x) \rangle \cdot (\langle -f(A,x), c(A,x) \rangle + f(A,x)_{j_1}) \cdot ( (A_{j_1,*}) \circ x^\top  )\\ 
     & ~ \cdot \diag(x) \cdot A^{\top} \cdot f(A,x) \cdot I_d\\
     & ~ =  f(A,x)_{j_1}^2 \cdot f_c(A,x) \cdot (-f_c(A,x) + f(A,x)_{j_1}) \cdot ( (A_{j_1,*}) \circ x^\top  )  \cdot h(A,x)\cdot I_d
\end{align*}
    where the last step is follows from the Definitions~\ref{def:h} and Definitions~\ref{def:f_c}.

{\bf Proof of Part 4.}
We have for diagonal entry and off-diagonal entry can be written as follows
    \begin{align*}
        B_{6,1,4}^{j_1,i_1,j_1,i_1}   = & ~  -  f(A,x)_{j_1}^2 \cdot f(A,x)^\top  \cdot A \cdot (\diag(x))^2 \cdot   \langle c(A,x), f(A,x) \rangle \cdot A^{\top} \cdot f(A,x) \\ 
     & ~\cdot(\langle -f(A,x), c(A,x) \rangle + f(A,x)_{j_1}) \\
        B_{6,1,4}^{j_1,i_1,j_1,i_0}   = & ~ -    f(A,x)_{j_1}^2 \cdot f(A,x)^\top  \cdot A \cdot (\diag(x))^2 \cdot   \langle c(A,x), f(A,x) \rangle \cdot A^{\top} \cdot f(A,x) \\ 
     & ~\cdot(\langle -f(A,x), c(A,x) \rangle + f(A,x)_{j_1})
    \end{align*}
 From the above equation, we can show that matrix $B_{6,1,4}^{j_1,*,j_1,*}$ can be expressed as a rank-$1$ matrix,
\begin{align*}
    B_{6,1,4}^{j_1,*,j_1,*}  & ~ = -  f(A,x)_{j_1}^2 \cdot \langle c(A,x), f(A,x) \rangle  \cdot(\langle -f(A,x), c(A,x) \rangle + f(A,x)_{j_1}) \cdot f(A,x)^\top\\ 
     & ~  \cdot A \cdot (\diag(x))^2 \cdot  A^{\top} \cdot f(A,x)\cdot I_d\\
     & ~ =   -  f(A,x)_{j_1}^2 \cdot f_c(A,x)  \cdot(-f_c(A,x) + f(A,x)_{j_1}) \cdot h(A,x)^\top \cdot h(A,x) \cdot I_d
\end{align*}
   where the last step is follows from the Definitions~\ref{def:h} and Definitions~\ref{def:f_c}.

{\bf Proof of Part 5.}
We have for diagonal entry and off-diagonal entry can be written as follows
    \begin{align*}
         B_{6,1,5}^{j_1,i_1,j_1,i_0} = & ~    f(A,x)_{j_1}^2 \cdot f(A,x)^\top  \cdot A \cdot (\diag(x))^2 \cdot (\langle -f(A,x), f(A,x) \rangle + f(A,x)_{j_1})   \cdot A^{\top} \cdot f(A,x)  \\
         &~ \cdot(\langle -f(A,x), c(A,x) \rangle + f(A,x)_{j_1}) \\
         B_{6,1,5}^{j_1,i_1,j_1,i_0} = & ~    f(A,x)_{j_1}^2 \cdot f(A,x)^\top  \cdot A \cdot (\diag(x))^2 \cdot (\langle -f(A,x), f(A,x) \rangle + f(A,x)_{j_1})   \cdot A^{\top} \cdot f(A,x) \\
         &~ \cdot(\langle -f(A,x), c(A,x) \rangle + f(A,x)_{j_1})
    \end{align*}
    From the above equation, we can show that matrix $B_{6,1,5}^{j_1,*,j_1,*}$ can be expressed as a rank-$1$ matrix,
\begin{align*}
    B_{6,1,5}^{j_1,*,j_1,*}  & ~ =  f(A,x)_{j_1}^2 \cdot (\langle -f(A,x), f(A,x) \rangle + f(A,x)_{j_1})  \cdot f(A,x)^\top  \cdot A \cdot (\diag(x))^2 \cdot  A^{\top} \cdot f(A,x) \cdot I_d\\
         &~ \cdot(\langle -f(A,x), c(A,x) \rangle + f(A,x)_{j_1})\\
     & ~ =    f(A,x)_{j_1}^2 \cdot (-f_2(A,x) + f(A,x)_{j_1})\cdot (-f_c(A,x) + f(A,x)_{j_1})  \cdot h(A,x)^\top \cdot h(A,x) \cdot I_d
\end{align*}
    where the last step is follows from the Definitions~\ref{def:h}, Definitions~\ref{def:f_c} and Definitions~\ref{def:f_2}.

{\bf Proof of Part 6.}
We have for diagonal entry and off-diagonal entry can be written as follows
    \begin{align*}
        B_{6,1,6}^{j_1,i_1,j_1,i_1}  = & ~   f(A,x)_{j_1}^2 \cdot f(A,x)^\top  \cdot A \cdot  (\diag(x))^2 \cdot(\langle -f(A,x), c(A,x) \rangle + f(A,x)_{j_1})^2 \cdot A^{\top} \cdot f(A,x) \\
        B_{6,1,6}^{j_1,i_1,j_1,i_0}  = & ~   f(A,x)_{j_1}^2 \cdot f(A,x)^\top  \cdot A \cdot  (\diag(x))^2 \cdot(\langle -f(A,x), c(A,x) \rangle + f(A,x)_{j_1})^2 \cdot A^{\top} \cdot f(A,x) 
    \end{align*}
    From the above equation, we can show that matrix $B_{6,1,6}^{j_1,*,j_1,*}$ can be expressed as a rank-$1$ matrix,
\begin{align*}
    B_{6,1,6}^{j_1,*,j_1,*}  & ~ =    f(A,x)_{j_1}^2  \cdot(\langle -f(A,x), c(A,x) \rangle + f(A,x)_{j_1})^2  \cdot f(A,x)^\top  \cdot A \cdot (\diag(x))^2 \cdot  A^{\top} \cdot f(A,x) \cdot I_d\\
     & ~ =   f(A,x)_{j_1}^2  \cdot(-f_c(A,x) + f(A,x)_{j_1})^2 \cdot h(A,x)^\top \cdot h(A,x) \cdot I_d
\end{align*}
    where the last step is follows from the Definitions~\ref{def:h} and Definitions~\ref{def:f_c}.
    
{\bf Proof of Part 7.}
We have for diagonal entry and off-diagonal entry can be written as follows
    \begin{align*}
         B_{6,1,7}^{j_1,i_1,j_1,i_1} = & ~  f(A,x)_{j_1}^2 \cdot ((e_{j_1}^\top - f(A,x)^\top) \circ q(A,x)^\top) \cdot A \cdot  (\diag(x))^2 \cdot A^{\top} \cdot f(A,x) \\ 
     & ~\cdot(\langle -f(A,x), c(A,x) \rangle + f(A,x)_{j_1})\\
         B_{6,1,7}^{j_1,i_1,j_1,i_0} = & ~  f(A,x)_{j_1}^2 \cdot ((e_{j_1}^\top - f(A,x)^\top) \circ q(A,x)^\top) \cdot A \cdot  (\diag(x))^2 \cdot A^{\top} \cdot f(A,x) \\ 
     & ~\cdot(\langle -f(A,x), c(A,x) \rangle + f(A,x)_{j_1})
    \end{align*}
    From the above equation, we can show that matrix $B_{6,1,7}^{j_1,*,j_1,*}$ can be expressed as a rank-$1$ matrix,
\begin{align*}
     B_{6,1,7}^{j_1,*,j_1,*}  & ~ =   f(A,x)_{j_1}^2 \cdot (\langle -f(A,x), c(A,x) \rangle + f(A,x)_{j_1})\cdot ((e_{j_1}^\top - f(A,x)^\top) \circ q(A,x)^\top) \cdot A \\ 
     & ~\cdot  (\diag(x))^2 \cdot A^{\top} \cdot f(A,x) \cdot I_d\\
     & ~ = f(A,x)_{j_1}^2 \cdot (-f_c(A,x) + f(A,x)_{j_1})\cdot p_{j_1}(A,x)^\top \cdot h(A,x) \cdot I_d
\end{align*}
    where the last step is follows from the Definitions~\ref{def:h}, Definitions~\ref{def:f_c} and Definitions~\ref{def:p}.

    {\bf Proof of Part 8.}
We have for diagonal entry and off-diagonal entry can be written as follows
    \begin{align*}
         B_{6,2,1}^{j_1,i_1,j_1,i_1} = & ~   f(A,x)_{j_1}^2 \cdot x_{i_1} \cdot c_g(A,x)^{\top} \cdot \diag(x) \cdot A^{\top} \cdot f(A,x) \cdot(\langle -f(A,x), c(A,x) \rangle + f(A,x)_{j_1})\\
         B_{6,2,1}^{j_1,i_1,j_1,i_0} = & ~  f(A,x)_{j_1}^2 \cdot x_{i_1} \cdot c_g(A,x)^{\top} \cdot \diag(x) \cdot A^{\top} \cdot f(A,x) \cdot(\langle -f(A,x), c(A,x) \rangle + f(A,x)_{j_1})
    \end{align*}
    From the above equation, we can show that matrix $B_{6,2,1}^{j_1,*,j_1,*}$ can be expressed as a rank-$1$ matrix,
\begin{align*}
     B_{6,2,1}^{j_1,*,j_1,*}  & ~ =   f(A,x)_{j_1}^2 \cdot  (\langle -f(A,x), cf(A,x) \rangle + f(A,x)_{j_1}) \cdot c_g(A,x)^{\top} \cdot \diag(x) \cdot A^{\top} \cdot f(A,x) \cdot x \cdot {\bf 1}_d^{\top}  \\ 
     & ~ =  f(A,x)_{j_1}^2 \cdot  (-f_c(A,x) + f(A,x)_{j_1}) \cdot c_g(A,x)^{\top} \cdot h(A,x) \cdot x \cdot {\bf 1}_d^{\top} 
\end{align*}
    where the last step is follows from the Definitions~\ref{def:h}, Definitions~\ref{def:f_c}.
    
    {\bf Proof of Part 9.}
We have for diagonal entry and off-diagonal entry can be written as follows
    \begin{align*}
         B_{6,2,2}^{j_1,i_1,j_1,i_1} = & ~   - f(A,x)_{j_1} \cdot x_{i_1} \cdot c_g(A,x)^{\top} \cdot \diag(x) \cdot A^{\top} \cdot f(A,x) \cdot(\langle -f(A,x), c(A,x) \rangle + f(A,x)_{j_1})\\
         B_{6,2,2}^{j_1,i_1,j_1,i_0} = & ~ - f(A,x)_{j_1} \cdot x_{i_1} \cdot c_g(A,x)^{\top} \cdot \diag(x) \cdot A^{\top} \cdot f(A,x) \cdot(\langle -f(A,x), c(A,x) \rangle + f(A,x)_{j_1})
    \end{align*}
    From the above equation, we can show that matrix $B_{6,2,2}^{j_1,*,j_1,*}$ can be expressed as a rank-$1$ matrix,
\begin{align*}
     B_{6,2,2}^{j_1,*,j_1,*}  & ~ =  - f(A,x)_{j_1}  \cdot  (\langle -f(A,x), c(A,x) \rangle + f(A,x)_{j_1}) \cdot c_g(A,x)^{\top} \cdot \diag(x) \cdot A^{\top} \cdot f(A,x) \cdot x \cdot {\bf 1}_d^{\top}  \\ 
     & ~ =  -f(A,x)_{j_1}  \cdot  (-f_c(A,x) + f(A,x)_{j_1}) \cdot c_g(A,x)^{\top} \cdot h(A,x) \cdot x \cdot {\bf 1}_d^{\top} 
\end{align*}
    where the last step is follows from the Definitions~\ref{def:h}, Definitions~\ref{def:f_c}.

   {\bf Proof of Part 10.}
We have for diagonal entry and off-diagonal entry can be written as follows
    \begin{align*}
         B_{6,3,1}^{j_1,i_1,j_1,i_1} = & ~   -  c_g(A,x)^{\top} \cdot f(A,x)_{j_1} \cdot \diag(x) \cdot e_{i_1} \cdot e_{j_1}^\top \cdot f(A,x) \cdot(\langle -f(A,x), c(A,x) \rangle + f(A,x)_{j_1})\\
         B_{6,3,1}^{j_1,i_1,j_1,i_0} = & ~ -  c_g(A,x)^{\top} \cdot f(A,x)_{j_1} \cdot \diag(x) \cdot e_{i_1} \cdot e_{j_1}^\top \cdot f(A,x) \cdot(\langle -f(A,x), c(A,x) \rangle + f(A,x)_{j_1})
    \end{align*}
    From the above equation, we can show that matrix $B_{6,3,1}^{j_1,*,j_1,*}$ can be expressed as a rank-$1$ matrix,
\begin{align*}
     B_{6,3,1}^{j_1,*,j_1,*}  & ~ = - f(A,x)_{j_1}^2 \cdot (\langle -f(A,x), c(A,x) \rangle + f(A,x)_{j_1})\cdot {\bf 1}_d \cdot c_g(A,x)^{\top}   \cdot \diag(x)  \\ 
     & ~ =   - f(A,x)_{j_1}^2 \cdot (-f_c(A,x) + f(A,x)_{j_1})\cdot {\bf 1}_d \cdot c_g(A,x)^{\top}   \cdot \diag(x)  
\end{align*}
    where the last step is follows from the Definitions~\ref{def:f_c}.

    {\bf Proof of Part 11.}
We have for diagonal entry and off-diagonal entry can be written as follows
    \begin{align*}
         B_{6,4,1}^{j_1,i_1,j_1,i_1} = & ~   - c_g(A,x)^{\top} \cdot f(A,x)_{j_1}^2 \cdot \diag(x) \cdot A^{\top}\cdot x_i \cdot  (e_{j_1}- f(A,x) ) \cdot (\langle -f(A,x), c(A,x) \rangle + f(A,x)_{j_1})\\
         B_{6,4,1}^{j_1,i_1,j_1,i_0} = & ~  - c_g(A,x)^{\top} \cdot f(A,x)_{j_1}^2 \cdot \diag(x) \cdot A^{\top}\cdot x_i \cdot  (e_{j_1}- f(A,x) ) \cdot (\langle -f(A,x), c(A,x) \rangle + f(A,x)_{j_1})
    \end{align*}
    From the above equation, we can show that matrix $B_{6,4,1}^{j_1,*,j_1,*}$ can be expressed as a rank-$1$ matrix,
\begin{align*}
     B_{6,4,1}^{j_1,*,j_1,*}  & ~ = - f(A,x)_{j_1}^2 \cdot (\langle -f(A,x), c(A,x) \rangle + f(A,x)_{j_1}) \cdot c_g(A,x)^{\top}  \cdot \diag(x) \cdot A^{\top}  \cdot  (e_{j_1}- f(A,x) ) \cdot x \cdot {\bf 1}_d^{\top}\\ 
     & ~ =   - f(A,x)_{j_1}^2 \cdot (-f_c(A,x) + f(A,x)_{j_1}) \cdot c_g(A,x)^{\top}  \cdot  h_e(A,x)\cdot x \cdot {\bf 1}_d^{\top}
\end{align*}
    where the last step is follows from the Definitions~\ref{def:f_c} and Definitions~\ref{def:h_e}.

        {\bf Proof of Part 12.}
We have for diagonal entry and off-diagonal entry can be written as follows
    \begin{align*}
         B_{6,5,1}^{j_1,i_1,j_1,i_1} = & ~  c_g(A,x)^{\top} \cdot f(A,x)_{j_1}^2 \cdot \diag(x) \cdot A^{\top} \cdot  f(A,x)  \cdot  x_{i_1} \cdot \langle - f(A,x), f(A,x) \rangle\\
         B_{6,5,1}^{j_1,i_1,j_1,i_0} = & ~ c_g(A,x)^{\top} \cdot f(A,x)_{j_1}^2 \cdot \diag(x) \cdot A^{\top} \cdot  f(A,x)  \cdot  x_{i_1} \cdot \langle - f(A,x), f(A,x) \rangle
    \end{align*}
    From the above equation, we can show that matrix $B_{6,5,1}^{j_1,*,j_1,*}$ can be expressed as a rank-$1$ matrix,
\begin{align*}
     B_{6,5,1}^{j_1,*,j_1,*}  & ~ = f(A,x)_{j_1}^2 \cdot \langle - f(A,x), f(A,x) \rangle \cdot c_g(A,x)^{\top} \cdot \diag(x) \cdot A^{\top} \cdot  f(A,x)  \cdot x \cdot {\bf 1}_d^{\top}\\ 
     & ~ =  -  f(A,x)_{j_1}^2 \cdot f_2(A,x) \cdot c_g(A,x)^{\top} \cdot h(A,x)  \cdot x \cdot {\bf 1}_d^{\top}
\end{align*}
    where the last step is follows from the Definitions~\ref{def:h} and Definitions~\ref{def:f_2}.

    {\bf Proof of Part 13.}
We have for diagonal entry and off-diagonal entry can be written as follows
    \begin{align*}
         B_{6,5,2}^{j_1,i_1,j_1,i_1} = & ~  c_g(A,x)^{\top} \cdot f(A,x)_{j_1}^3 \cdot \diag(x) \cdot A^{\top} \cdot  f(A,x)  \cdot  x_{i_1}  \\
         B_{6,5,2}^{j_1,i_1,j_1,i_0} = & ~ c_g(A,x)^{\top} \cdot f(A,x)_{j_1}^3 \cdot \diag(x) \cdot A^{\top} \cdot  f(A,x)  \cdot  x_{i_1}  
    \end{align*}
    From the above equation, we can show that matrix $B_{6,5,2}^{j_1,*,j_1,*}$ can be expressed as a rank-$1$ matrix,
\begin{align*}
     B_{6,5,2}^{j_1,*,j_1,*}  & ~ = f(A,x)_{j_1}^3   \cdot c_g(A,x)^{\top} \cdot \diag(x) \cdot A^{\top} \cdot  f(A,x)  \cdot x \cdot {\bf 1}_d^{\top}\\ 
     & ~ =  f(A,x)_{j_1}^3  \cdot c_g(A,x)^{\top} \cdot h(A,x)  \cdot x \cdot {\bf 1}_d^{\top}
\end{align*}
    where the last step is follows from the Definitions~\ref{def:h}.

        {\bf Proof of Part 14.}
We have for diagonal entry and off-diagonal entry can be written as follows
    \begin{align*}
         B_{6,6,1}^{j_1,i_1,j_1,i_1} = & ~  c_g(A,x)^{\top} \cdot f(A,x)_{j_1}^2 \cdot \diag(x) \cdot A^{\top} \cdot  f(A,x)  \cdot  x_{i_1} \cdot \langle - f(A,x), c(A,x) \rangle\\
         B_{6,6,1}^{j_1,i_1,j_1,i_0} = & ~ c_g(A,x)^{\top} \cdot f(A,x)_{j_1}^2 \cdot \diag(x) \cdot A^{\top} \cdot  f(A,x)  \cdot  x_{i_1} \cdot \langle - f(A,x), c(A,x) \rangle
    \end{align*}
    From the above equation, we can show that matrix $B_{6,6,1}^{j_1,*,j_1,*}$ can be expressed as a rank-$1$ matrix,
\begin{align*}
     B_{6,6,1}^{j_1,*,j_1,*}  & ~ = f(A,x)_{j_1}^2 \cdot \langle - f(A,x), c(A,x) \rangle \cdot c_g(A,x)^{\top} \cdot \diag(x) \cdot A^{\top} \cdot  f(A,x)  \cdot x \cdot {\bf 1}_d^{\top}\\ 
     & ~ =  -  f(A,x)_{j_1}^2 \cdot f_c(A,x) \cdot c_g(A,x)^{\top} \cdot h(A,x)  \cdot x \cdot {\bf 1}_d^{\top}
\end{align*}
    where the last step is follows from the Definitions~\ref{def:h} and Definitions~\ref{def:f_c}.

    {\bf Proof of Part 15.}
We have for diagonal entry and off-diagonal entry can be written as follows
    \begin{align*}
         B_{6,6,2}^{j_1,i_1,j_1,i_1} = & ~  c_g(A,x)^{\top} \cdot f(A,x)_{j_1}^2 \cdot c(A,x)_{j_1} \cdot \diag(x) \cdot A^{\top} \cdot  f(A,x)  \cdot  x_{i_1}  \\
         B_{6,6,2}^{j_1,i_1,j_1,i_0} = & ~ c_g(A,x)^{\top} \cdot f(A,x)_{j_1}^2 \cdot c(A,x)_{j_1} \cdot \diag(x) \cdot A^{\top} \cdot  f(A,x)  \cdot  x_{i_1}  
    \end{align*}
    From the above equation, we can show that matrix $B_{5,6,2}^{j_1,*,j_1,*}$ can be expressed as a rank-$1$ matrix,
\begin{align*}
     B_{5,6,2}^{j_1,*,j_1,*}  & ~ = f(A,x)_{j_1}^2 \cdot   c(A,x)_{j_1}  \cdot c_g(A,x)^{\top} \cdot \diag(x) \cdot A^{\top} \cdot  f(A,x)  \cdot x \cdot {\bf 1}_d^{\top}\\ 
     & ~ =  f(A,x)_{j_1}^2 \cdot   c(A,x)_{j_1}   \cdot c_g(A,x)^{\top} \cdot h(A,x)  \cdot x \cdot {\bf 1}_d^{\top}
\end{align*}
    where the last step is follows from the Definitions~\ref{def:h}.

    {\bf Proof of Part 16.}
We have for diagonal entry and off-diagonal entry can be written as follows
    \begin{align*}
         B_{6,7,1}^{j_1,i_1,j_1,i_1} = & ~  f(A,x)_{j_1}^3 \cdot  x_{i_1}\cdot c_g(A,x)^{\top} \cdot \diag(x) A^{\top} \cdot  f(A,x)  \\
         B_{6,7,1}^{j_1,i_1,j_1,i_0} = & ~ f(A,x)_{j_1}^3 \cdot  x_{i_1}\cdot c_g(A,x)^{\top} \cdot \diag(x) A^{\top} \cdot  f(A,x) 
    \end{align*}
    From the above equation, we can show that matrix $B_{6,7,1}^{j_1,*,j_1,*}$ can be expressed as a rank-$1$ matrix,
\begin{align*}
     B_{6,7,1}^{j_1,*,j_1,*}  & ~ = f(A,x)_{j_1}^3   \cdot c_g(A,x)^{\top} \cdot \diag(x) \cdot A^{\top} \cdot  f(A,x)  \cdot x \cdot {\bf 1}_d^{\top}\\ 
     & ~ =  f(A,x)_{j_1}^3  \cdot c_g(A,x)^{\top} \cdot h(A,x)  \cdot x \cdot {\bf 1}_d^{\top}
\end{align*}
    where the last step is follows from the Definitions~\ref{def:h}.

       {\bf Proof of Part 17.}
We have for diagonal entry and off-diagonal entry can be written as follows
    \begin{align*}
         B_{6,7,2}^{j_1,i_1,j_1,i_1} = & ~  -  f(A,x)_{j_1}^2 \cdot  x_{i_1} c_g(A,x)^{\top} \cdot \diag(x) \cdot A^{\top} \cdot  f(A,x)   \\
         B_{6,7,2}^{j_1,i_1,j_1,i_0} = & ~ -  f(A,x)_{j_1}^2 \cdot  x_{i_1} c_g(A,x)^{\top} \cdot \diag(x) \cdot A^{\top} \cdot  f(A,x) 
    \end{align*}
    From the above equation, we can show that matrix $B_{6,7,2}^{j_1,*,j_1,*}$ can be expressed as a rank-$1$ matrix,
\begin{align*}
     B_{6,7,2}^{j_1,*,j_1,*}  & ~ = f(A,x)_{j_1}^2   \cdot c_g(A,x)^{\top} \cdot \diag(x) \cdot A^{\top} \cdot  f(A,x)  \cdot x \cdot {\bf 1}_d^{\top}\\ 
     & ~ =  f(A,x)_{j_1}^2  \cdot c_g(A,x)^{\top} \cdot h(A,x)  \cdot x \cdot {\bf 1}_d^{\top}
\end{align*}
    where the last step is follows from the Definitions~\ref{def:h}.
\end{proof}
\end{lemma}

\subsection{Case \texorpdfstring{$j_1 \neq j_0, i_1 = i_0$}{}}
\begin{lemma}
For $j_1 \neq j_0$ and $i_0 = i_1$. If the following conditions hold
    \begin{itemize}
     \item Let $u(A,x) \in \R^n$ be defined as Definition~\ref{def:u}
    \item Let $\alpha(A,x) \in \R$ be defined as Definition~\ref{def:alpha}
     \item Let $f(A,x) \in \R^n$ be defined as Definition~\ref{def:f}
    \item Let $c(A,x) \in \R^n$ be defined as Definition~\ref{def:c}
    \item Let $g(A,x) \in \R^d$ be defined as Definition~\ref{def:g} 
    \item Let $q(A,x) = c(A,x) + f(A,x) \in \R^n$
    \item Let $c_g(A,x) \in \R^d$ be defined as Definition~\ref{def:c_g}.
    \item Let $L_g(A,x) \in \R$ be defined as Definition~\ref{def:l_g}
    \item Let $v \in \R^n$ be a vector 
    \item Let $B_1^{j_1,i_1,j_0,i_0}$ be defined as Definition~\ref{def:b_1}
    \end{itemize}
    Then, For $j_0,j_1 \in [n], i_0,i_1 \in [d]$, we have 
    \begin{itemize}
\item {\bf Part 1.} For $B_{6,1}^{j_1,i_1,j_0,i_1}$, we have 
\begin{align*}
 B_{6,1}^{j_1,i_1,j_0,i_1}  = & ~  \frac{\d}{\d A_{j_1,i_1}} (- c_g(A,x)^{\top} ) \cdot  f(A,x)_{j_0} \cdot \diag (x) A^{\top} \cdot  f(A,x)  \cdot  (\langle -f(A,x), c(A,x) \rangle + f(A,x)_{j_0})\\
 = & ~ B_{6,1,1}^{j_1,i_1,j_0,i_1} + B_{6,1,2}^{j_1,i_1,j_0,i_1} + B_{6,1,3}^{j_1,i_1,j_0,i_1} + B_{6,1,4}^{j_1,i_1,j_0,i_1} + B_{6,1,5}^{j_1,i_1,j_0,i_1} + B_{6,1,6}^{j_1,i_1,j_0,i_1} + B_{6,1,7}^{j_1,i_1,j_0,i_1}
\end{align*} 
\item {\bf Part 2.} For $B_{6,2}^{j_1,i_1,j_0,i_1}$, we have 
\begin{align*}
  B_{6,2}^{j_1,i_1,j_0,i_1} = & ~ - c_g(A,x)^{\top} \cdot \frac{\d}{\d A_{j_1,i_1}} ( f(A,x)_{j_0} )  \cdot \diag(x) \cdot A^{\top} \cdot f(A,x) \cdot(\langle -f(A,x), c(A,x) \rangle + f(A,x)_{j_0}) \\
    = & ~  B_{6,2,1}^{j_1,i_1,j_0,i_1}  
\end{align*} 
\item {\bf Part 3.} For $B_{6,3}^{j_1,i_1,j_0,i_1}$, we have 
\begin{align*}
  B_{6,3}^{j_1,i_1,j_0,i_1} = & ~ - c_g(A,x)^{\top} \cdot f(A,x)_{j_0} \cdot \diag(x) \cdot \frac{\d}{\d A_{j_1,i_1}} (  A^{\top} ) \cdot f(A,x) \cdot(\langle -f(A,x), c(A,x) \rangle + f(A,x)_{j_0}) \\
     = & ~ B_{6,3,1}^{j_1,i_1,j_0,i_1}  
\end{align*} 
\item {\bf Part 4.} For $B_{6,4}^{j_1,i_1,j_0,i_1}$, we have 
\begin{align*}
  B_{6,4}^{j_1,i_1,j_0,i_1} = & ~ - c_g(A,x)^{\top} \cdot f(A,x)_{j_0} \cdot \diag(x) \cdot A^{\top} \cdot \frac{\d f(A,x)}{\d A_{j_1,i_1}} \cdot(\langle -f(A,x), c(A,x) \rangle + f(A,x)_{j_0}) \\
     = & ~ B_{6,4,1}^{j_1,i_1,j_0,i_1} 
\end{align*}
\item {\bf Part 5.} For $B_{6,5}^{j_1,i_1,j_0,i_1}$, we have 
\begin{align*}
  B_{6,5}^{j_1,i_1,j_0,i_1} = & ~ c_g(A,x)^{\top} \cdot f(A,x)_{j_0} \cdot \diag(x) \cdot A^{\top} \cdot  f(A,x)  \cdot \langle \frac{\d  f(A,x) }{\d A_{j_1,i_1}},c(A,x)\rangle  \\
     = & ~ B_{5,5,1}^{j_1,i_1,j_0,i_1}  + B_{5,5,2}^{j_1,i_1,j_0,i_1}  
\end{align*}
\item {\bf Part 6.} For $B_{6,6}^{j_1,i_1,j_0,i_1}$, we have 
\begin{align*}
  B_{6,6}^{j_1,i_1,j_0,i_1} = & ~ c_g(A,x)^{\top} \cdot f(A,x)_{j_0} \cdot \diag(x) \cdot A^{\top} \cdot  f(A,x)  \cdot \langle f(A,x), \frac{\d c(A,x) }{\d A_{j_1,i_1}}\rangle \\
     = & ~ B_{6,6,1}^{j_1,i_1,j_0,i_1}  + B_{6,6,2}^{j_1,i_1,j_0,i_1}  
\end{align*}
\item {\bf Part 7.} For $B_{6,7}^{j_1,i_1,j_0,i_1}$, we have 
\begin{align*}
  B_{6,7}^{j_1,i_1,j_0,i_1} = & ~ - c_g(A,x)^{\top} \cdot f(A,x)_{j_0} \cdot \diag(x) \cdot A^{\top} \cdot  f(A,x)  \cdot  \frac{\d f(A,x)_{j_0}}{\d A_{j_1,i_1}} \\
     = & ~ B_{6,7,1}^{j_1,i_1,j_0,i_1}  
\end{align*}
\end{itemize}
\begin{proof}
    {\bf Proof of Part 1.}
    \begin{align*}
    B_{6,1,1}^{j_1,i_1,j_0,i_1} : = & ~ e_{i_1}^\top \cdot \langle c(A,x), f(A,x) \rangle \cdot  f(A,x)_{j_1} \cdot f(A,x)_{j_0} \cdot \diag(x) \cdot A^{\top} \\
        & ~ \cdot f(A,x) \cdot(\langle -f(A,x), c(A,x) \rangle + f(A,x)_{j_0})\\
    B_{6,1,2}^{j_1,i_1,j_0,i_1} : = & ~   e_{i_1}^\top \cdot c(A,x)_{j_1}\cdot f(A,x)_{j_1} \cdot f(A,x)_{j_0} \cdot \diag(x) \cdot A^{\top} \cdot f(A,x) \cdot(\langle -f(A,x), c(A,x) \rangle + f(A,x)_{j_0})\\
    B_{6,1,3}^{j_1,i_1,j_0,i_1} : = & ~ f(A,x)_{j_1} \cdot f(A,x)_{j_0} \cdot \langle c(A,x), f(A,x) \rangle \cdot ( (A_{j_1,*}) \circ x^\top  )  \cdot \diag(x) \cdot A^{\top} \cdot f(A,x) \cdot \\
    & ~(\langle -f(A,x), c(A,x) \rangle + f(A,x)_{j_0})\\
    B_{6,1,4}^{j_1,i_1,j_0,i_1} : = & ~   -  f(A,x)_{j_1} \cdot f(A,x)_{j_0} \cdot f(A,x)^\top  \cdot A \cdot (\diag(x))^2 \cdot   \langle c(A,x), f(A,x) \rangle \cdot A^{\top} \cdot f(A,x)\\
    & ~ \cdot(\langle -f(A,x), c(A,x) \rangle + f(A,x)_{j_0})\\
    B_{6,1,5}^{j_1,i_1,j_0,i_1} : = & ~    f(A,x)_{j_1} \cdot f(A,x)_{j_0} \cdot f(A,x)^\top  \cdot A \cdot (\diag(x))^2 \cdot (\langle (-f(A,x)), f(A,x) \rangle + f(A,x)_{j_1})   \cdot A^{\top} \cdot f(A,x) \\
    & ~ \cdot(\langle -f(A,x), c(A,x) \rangle + f(A,x)_{j_0})\\
    B_{6,1,6}^{j_1,i_1,j_0,i_1} : = & ~  f(A,x)_{j_1} \cdot f(A,x)_{j_0} \cdot f(A,x)^\top  \cdot A \cdot  (\diag(x))^2 \cdot(\langle -f(A,x), c(A,x) \rangle + f(A,x)_{j_1}) \cdot A^{\top} \cdot f(A,x) \\
        & ~ \cdot(\langle -f(A,x), c(A,x) \rangle + f(A,x)_{j_0})\\
    B_{6,1,7}^{j_1,i_1,j_0,i_1} : = & ~   f(A,x)_{j_1} \cdot f(A,x)_{j_0} \cdot ((e_{j_1}^\top - f(A,x)^\top) \circ q(A,x)^\top) \cdot A \cdot  (\diag(x))^2 \cdot A^{\top} \cdot f(A,x)\\
    & ~\cdot(\langle -f(A,x), c(A,x) \rangle + f(A,x)_{j_0})
\end{align*}
Finally, combine them and we have
\begin{align*}
       B_{6,1}^{j_1,i_1,j_0,i_1} =  B_{6,1,1}^{j_1,i_1,j_0,i_1} + B_{6,1,2}^{j_1,i_1,j_0,i_1} + B_{6,1,3}^{j_1,i_1,j_0,i_1} + B_{6,1,4}^{j_1,i_1,j_0,i_1} + B_{6,1,5}^{j_1,i_1,j_0,i_1} + B_{6,1,6}^{j_1,i_1,j_0,i_1} + B_{6,1,7}^{j_1,i_1,j_0,i_1}
\end{align*}
{\bf Proof of Part 2.}
    \begin{align*}
    B_{6,2,1}^{j_1,i_1,j_0,i_1} : = & ~   f(A,x)_{j_1} \cdot f(A,x)_{j_0} \cdot x_{i_1} \cdot c_g(A,x)^{\top} \cdot \diag(x) \cdot A^{\top} \cdot f(A,x) \cdot(\langle -f(A,x), c(A,x) \rangle + f(A,x)_{j_0})  
\end{align*}
Finally, combine them and we have
\begin{align*}
       B_{6,2}^{j_1,i_1,j_0,i_1} = B_{6,2,1}^{j_1,i_1,j_0,i_1}  
\end{align*}
{\bf Proof of Part 3.} 
    \begin{align*}
    B_{6,3,1}^{j_1,i_1,j_0,i_1} : = & ~    -  c_g(A,x)^{\top} \cdot f(A,x)_{j_0} \cdot \diag(x) \cdot e_{i_1} \cdot e_{j_1}^\top \cdot f(A,x) \cdot(\langle -f(A,x), c(A,x) \rangle + f(A,x)_{j_0})
\end{align*}
Finally, combine them and we have
\begin{align*}
       B_{6,3}^{j_1,i_1,j_0,i_1} = B_{6,3,1}^{j_1,i_1,j_0,i_1} 
\end{align*}
{\bf Proof of Part 4.} 
    \begin{align*}
    B_{6,4,1}^{j_1,i_1,j_0,i_1} : = & ~ - c_g(A,x)^{\top} \cdot f(A,x)_{j_1} \cdot f(A,x)_{j_0} \cdot \diag(x) \cdot A^{\top}\cdot x_i \cdot  (e_{j_1}- f(A,x) )  \\
        & ~\cdot (\langle -f(A,x), c(A,x) \rangle + f(A,x)_{j_0})
\end{align*}
Finally, combine them and we have
\begin{align*}
       B_{6,4}^{j_1,i_1,j_0,i_1} = B_{6,4,1}^{j_1,i_1,j_0,i_1}  
\end{align*}
{\bf Proof of Part 5.} 
    \begin{align*}
    B_{6,5,1}^{j_1,i_1,j_0,i_1} : = & ~ c_g(A,x)^{\top} \cdot f(A,x)_{j_1} \cdot f(A,x)_{j_0} \cdot \diag(x) \cdot A^{\top} \cdot  f(A,x)  \cdot  x_{i_1} \cdot \langle - f(A,x), f(A,x) \rangle\\
    B_{6,5,2}^{j_1,i_1,j_0,i_1} : = & ~ c_g(A,x)^{\top} \cdot f(A,x)_{j_1}^2 \cdot f(A,x)_{j_0} \cdot \diag(x) \cdot A^{\top} \cdot  f(A,x)  \cdot  x_{i_1}  
\end{align*}
Finally, combine them and we have
\begin{align*}
       B_{6,5}^{j_1,i_1,j_0,i_1} = B_{6,5,1}^{j_1,i_1,j_0,i_1}  +B_{6,5,2}^{j_1,i_1,j_0,i_1}
\end{align*}
{\bf Proof of Part 6.} 
    \begin{align*}
     B_{6,6,1}^{j_1,i_1,j_0,i_1} : = & ~ c_g(A,x)^{\top} \cdot f(A,x)_{j_1} \cdot f(A,x)_{j_0} \cdot \diag(x) \cdot A^{\top} \cdot  f(A,x)  \cdot  x_{i_1} \cdot \langle - f(A,x), c(A,x) \rangle\\
    B_{6,6,2}^{j_1,i_1,j_0,i_1} : = & ~ c_g(A,x)^{\top} \cdot f(A,x)_{j_1} \cdot c(A,x)_{j_1} \cdot f(A,x)_{j_0} \cdot \diag(x) \cdot A^{\top} \cdot  f(A,x)  \cdot  x_{i_1}  
\end{align*}
Finally, combine them and we have
\begin{align*}
       B_{6,6}^{j_1,i_1,j_0,i_1} = B_{6,6,1}^{j_1,i_1,j_0,i_1}  +B_{6,6,2}^{j_1,i_1,j_0,i_1}
\end{align*}
{\bf Proof of Part 7.} 
    \begin{align*}
     B_{6,7,1}^{j_1,i_1,j_0,i_1} : = & ~  f(A,x)_{j_1} \cdot f(A,x)_{j_0}^2 \cdot  x_{i_1}\cdot c_g(A,x)^{\top} \cdot \diag(x) A^{\top} \cdot  f(A,x)   
\end{align*}
Finally, combine them and we have
\begin{align*}
       B_{6,7}^{j_1,i_1,j_0,i_1} = B_{6,7,1}^{j_1,i_1,j_0,i_1}  
\end{align*}
\end{proof}
\end{lemma}

\subsection{Case \texorpdfstring{$j_1 \neq j_0, i_1 \neq i_0$}{}}
\begin{lemma}
For $j_1 \neq j_0$ and $i_0 \neq i_1$. If the following conditions hold
    \begin{itemize}
     \item Let $u(A,x) \in \R^n$ be defined as Definition~\ref{def:u}
    \item Let $\alpha(A,x) \in \R$ be defined as Definition~\ref{def:alpha}
     \item Let $f(A,x) \in \R^n$ be defined as Definition~\ref{def:f}
    \item Let $c(A,x) \in \R^n$ be defined as Definition~\ref{def:c}
    \item Let $g(A,x) \in \R^d$ be defined as Definition~\ref{def:g} 
    \item Let $q(A,x) = c(A,x) + f(A,x) \in \R^n$
    \item Let $c_g(A,x) \in \R^d$ be defined as Definition~\ref{def:c_g}.
    \item Let $L_g(A,x) \in \R$ be defined as Definition~\ref{def:l_g}
    \item Let $v \in \R^n$ be a vector 
    \item Let $B_1^{j_1,i_1,j_0,i_0}$ be defined as Definition~\ref{def:b_1}
    \end{itemize}
    Then, For $j_0,j_1 \in [n], i_0,i_1 \in [d]$, we have 
    \begin{itemize}
\item {\bf Part 1.} For $B_{6,1}^{j_1,i_1,j_0,i_0}$, we have 
\begin{align*}
 B_{6,1}^{j_1,i_1,j_0,i_0}  = & ~  \frac{\d}{\d A_{j_1,i_1}} (- c_g(A,x)^{\top} ) \cdot  f(A,x)_{j_0} \cdot \diag (x) A^{\top} \cdot  f(A,x)  \cdot  (\langle -f(A,x), c(A,x) \rangle + f(A,x)_{j_0})\\
 = & ~ B_{6,1,1}^{j_1,i_1,j_0,i_0} + B_{6,1,2}^{j_1,i_1,j_0,i_0} + B_{6,1,3}^{j_1,i_1,j_0,i_0} + B_{6,1,4}^{j_1,i_1,j_0,i_0} + B_{6,1,5}^{j_1,i_1,j_0,i_0} + B_{6,1,6}^{j_1,i_1,j_0,i_0} + B_{6,1,7}^{j_1,i_1,j_0,i_0}
\end{align*} 
\item {\bf Part 2.} For $B_{6,2}^{j_1,i_1,j_0,i_0}$, we have 
\begin{align*}
  B_{6,2}^{j_1,i_1,j_0,i_0} = & ~ - c_g(A,x)^{\top} \cdot \frac{\d}{\d A_{j_1,i_1}} ( f(A,x)_{j_0} )  \cdot \diag(x) \cdot A^{\top} \cdot f(A,x) \cdot(\langle -f(A,x), c(A,x) \rangle + f(A,x)_{j_0}) \\
    = & ~  B_{6,2,1}^{j_1,i_1,j_0,i_0}  
\end{align*} 
\item {\bf Part 3.} For $B_{6,3}^{j_1,i_1,j_0,i_0}$, we have 
\begin{align*}
  B_{6,3}^{j_1,i_1,j_0,i_0} = & ~ - c_g(A,x)^{\top} \cdot f(A,x)_{j_0} \cdot \diag(x) \cdot \frac{\d}{\d A_{j_1,i_1}} (  A^{\top} ) \cdot f(A,x) \cdot(\langle -f(A,x), c(A,x) \rangle + f(A,x)_{j_0}) \\
     = & ~ B_{6,3,1}^{j_1,i_1,j_0,i_0}  
\end{align*} 
\item {\bf Part 4.} For $B_{6,4}^{j_1,i_1,j_0,i_0}$, we have 
\begin{align*}
  B_{6,4}^{j_1,i_1,j_0,i_0} = & ~ - c_g(A,x)^{\top} \cdot f(A,x)_{j_0} \cdot \diag(x) \cdot A^{\top} \cdot \frac{\d f(A,x)}{\d A_{j_1,i_1}} \cdot(\langle -f(A,x), c(A,x) \rangle + f(A,x)_{j_0}) \\
     = & ~ B_{6,4,1}^{j_1,i_1,j_0,i_0} 
\end{align*}
\item {\bf Part 5.} For $B_{6,5}^{j_1,i_1,j_0,i_0}$, we have 
\begin{align*}
  B_{6,5}^{j_1,i_1,j_0,i_0} = & ~ c_g(A,x)^{\top} \cdot f(A,x)_{j_0} \cdot \diag(x) \cdot A^{\top} \cdot  f(A,x)  \cdot \langle \frac{\d  f(A,x) }{\d A_{j_1,i_1}},c(A,x)\rangle  \\
     = & ~ B_{5,5,1}^{j_1,i_1,j_0,i_0}  + B_{5,5,2}^{j_1,i_1,j_0,i_0}  
\end{align*}
\item {\bf Part 6.} For $B_{6,6}^{j_1,i_1,j_0,i_0}$, we have 
\begin{align*}
  B_{6,6}^{j_1,i_1,j_0,i_0} = & ~ c_g(A,x)^{\top} \cdot f(A,x)_{j_0} \cdot \diag(x) \cdot A^{\top} \cdot  f(A,x)  \cdot \langle f(A,x), \frac{\d c(A,x) }{\d A_{j_1,i_1}}\rangle \\
     = & ~ B_{6,6,1}^{j_1,i_1,j_0,i_0}  + B_{6,6,2}^{j_1,i_1,j_0,i_0}  
\end{align*}
\item {\bf Part 7.} For $B_{6,7}^{j_1,i_1,j_0,i_0}$, we have 
\begin{align*}
  B_{6,7}^{j_1,i_1,j_0,i_0} = & ~ - c_g(A,x)^{\top} \cdot f(A,x)_{j_0} \cdot \diag(x) \cdot A^{\top} \cdot  f(A,x)  \cdot  \frac{\d f(A,x)_{j_0}}{\d A_{j_1,i_1}} \\
     = & ~ B_{6,7,1}^{j_1,i_1,j_0,i_0}  
\end{align*}
\end{itemize}
\begin{proof}
    {\bf Proof of Part 1.}
    \begin{align*}
    B_{6,1,1}^{j_1,i_1,j_0,i_0} : = & ~ e_{i_1}^\top \cdot \langle c(A,x), f(A,x) \rangle \cdot  f(A,x)_{j_1} \cdot f(A,x)_{j_0} \cdot \diag(x) \cdot A^{\top}  \\
        & ~\cdot f(A,x) \cdot(\langle -f(A,x), c(A,x) \rangle + f(A,x)_{j_0})\\
    B_{6,1,2}^{j_1,i_1,j_0,i_0} : = & ~   e_{i_1}^\top \cdot c(A,x)_{j_1}\cdot f(A,x)_{j_1} \cdot f(A,x)_{j_0} \cdot \diag(x) \cdot A^{\top} \cdot f(A,x) \cdot(\langle -f(A,x), c(A,x) \rangle + f(A,x)_{j_0})\\
    B_{6,1,3}^{j_1,i_1,j_0,i_0} : = & ~ f(A,x)_{j_1} \cdot f(A,x)_{j_0} \cdot \langle c(A,x), f(A,x) \rangle \cdot ( (A_{j_1,*}) \circ x^\top  )  \cdot \diag(x) \cdot A^{\top} \cdot f(A,x) \cdot \\
    & ~(\langle -f(A,x), c(A,x) \rangle + f(A,x)_{j_0})\\
    B_{6,1,4}^{j_1,i_1,j_0,i_0} : = & ~   -  f(A,x)_{j_1} \cdot f(A,x)_{j_0} \cdot f(A,x)^\top  \cdot A \cdot (\diag(x))^2 \cdot   \langle c(A,x), f(A,x) \rangle \cdot A^{\top} \cdot f(A,x)\\
    & ~ \cdot(\langle -f(A,x), c(A,x) \rangle + f(A,x)_{j_0})\\
    B_{6,1,5}^{j_1,i_1,j_0,i_0} : = & ~    f(A,x)_{j_1} \cdot f(A,x)_{j_0} \cdot f(A,x)^\top  \cdot A \cdot (\diag(x))^2 \cdot (\langle (-f(A,x)), f(A,x) \rangle + f(A,x)_{j_1})   \cdot A^{\top} \cdot f(A,x) \\
    & ~ \cdot(\langle -f(A,x), c(A,x) \rangle + f(A,x)_{j_0})\\
    B_{6,1,6}^{j_1,i_1,j_0,i_0} : = & ~  f(A,x)_{j_1} \cdot f(A,x)_{j_0} \cdot f(A,x)^\top  \cdot A \cdot  (\diag(x))^2 \cdot(\langle -f(A,x), c(A,x) \rangle + f(A,x)_{j_1}) \cdot A^{\top} \cdot f(A,x) \\
        & ~ \cdot(\langle -f(A,x), c(A,x) \rangle + f(A,x)_{j_0})\\
    B_{6,1,7}^{j_1,i_1,j_0,i_0} : = & ~   f(A,x)_{j_1} \cdot f(A,x)_{j_0} \cdot ((e_{j_1}^\top - f(A,x)^\top) \circ q(A,x)^\top) \cdot A \cdot  (\diag(x))^2 \cdot A^{\top} \cdot f(A,x)\\
    & ~\cdot(\langle -f(A,x), c(A,x) \rangle + f(A,x)_{j_0})
\end{align*}
Finally, combine them and we have
\begin{align*}
       B_{6,1}^{j_1,i_1,j_0,i_0} =  B_{6,1,1}^{j_1,i_1,j_0,i_0} + B_{6,1,2}^{j_1,i_1,j_0,i_0} + B_{6,1,3}^{j_1,i_1,j_0,i_0} + B_{6,1,4}^{j_1,i_1,j_0,i_0} + B_{6,1,5}^{j_1,i_1,j_0,i_0} + B_{6,1,6}^{j_1,i_1,j_0,i_0} + B_{6,1,7}^{j_1,i_1,j_0,i_0}
\end{align*}
{\bf Proof of Part 2.}
    \begin{align*}
    B_{6,2,1}^{j_1,i_1,j_0,i_0} : = & ~   f(A,x)_{j_1} \cdot f(A,x)_{j_0} \cdot x_{i_1} \cdot c_g(A,x)^{\top} \cdot \diag(x) \cdot A^{\top} \cdot f(A,x) \cdot(\langle -f(A,x), c(A,x) \rangle + f(A,x)_{j_0})  
\end{align*}
Finally, combine them and we have
\begin{align*}
       B_{6,2}^{j_1,i_1,j_0,i_0} = B_{6,2,1}^{j_1,i_1,j_0,i_0}  
\end{align*}
{\bf Proof of Part 3.} 
    \begin{align*}
    B_{6,3,1}^{j_1,i_1,j_0,i_0} : = & ~    -  c_g(A,x)^{\top} \cdot f(A,x)_{j_0} \cdot \diag(x) \cdot e_{i_1} \cdot e_{j_1}^\top \cdot f(A,x) \cdot(\langle -f(A,x), c(A,x) \rangle + f(A,x)_{j_0})
\end{align*}
Finally, combine them and we have
\begin{align*}
       B_{6,3}^{j_1,i_1,j_0,i_0} = B_{6,3,1}^{j_1,i_1,j_0,i_0} 
\end{align*}
{\bf Proof of Part 4.} 
    \begin{align*}
    B_{6,4,1}^{j_1,i_1,j_0,i_0} : = & ~ - c_g(A,x)^{\top} \cdot f(A,x)_{j_1} \cdot f(A,x)_{j_0} \cdot \diag(x) \cdot A^{\top}\cdot x_i \\ 
     & ~\cdot  (e_{j_1}- f(A,x) ) \cdot (\langle -f(A,x), c(A,x) \rangle + f(A,x)_{j_0})
\end{align*}
Finally, combine them and we have
\begin{align*}
       B_{6,4}^{j_1,i_1,j_0,i_0} = B_{6,4,1}^{j_1,i_1,j_0,i_0}  
\end{align*}
{\bf Proof of Part 5.} 
    \begin{align*}
    B_{6,5,1}^{j_1,i_1,j_0,i_0} : = & ~ c_g(A,x)^{\top} \cdot f(A,x)_{j_1} \cdot f(A,x)_{j_0} \cdot \diag(x) \cdot A^{\top} \cdot  f(A,x)  \cdot  x_{i_1} \cdot \langle - f(A,x), f(A,x) \rangle\\
    B_{6,5,2}^{j_1,i_1,j_0,i_0} : = & ~ c_g(A,x)^{\top} \cdot f(A,x)_{j_1}^2 \cdot f(A,x)_{j_0} \cdot \diag(x) \cdot A^{\top} \cdot  f(A,x)  \cdot  x_{i_1}  
\end{align*}
Finally, combine them and we have
\begin{align*}
       B_{6,5}^{j_1,i_1,j_0,i_0} = B_{6,5,1}^{j_1,i_1,j_0,i_0}  +B_{6,5,2}^{j_1,i_1,j_0,i_0}
\end{align*}
{\bf Proof of Part 6.} 
    \begin{align*}
     B_{6,6,1}^{j_1,i_1,j_0,i_0} : = & ~ c_g(A,x)^{\top} \cdot f(A,x)_{j_1} \cdot f(A,x)_{j_0} \cdot \diag(x) \cdot A^{\top} \cdot  f(A,x)  \cdot  x_{i_1} \cdot \langle - f(A,x), c(A,x) \rangle\\
    B_{6,6,2}^{j_1,i_1,j_0,i_0} : = & ~ c_g(A,x)^{\top} \cdot f(A,x)_{j_1} \cdot c(A,x)_{j_1} \cdot f(A,x)_{j_0} \cdot \diag(x) \cdot A^{\top} \cdot  f(A,x)  \cdot  x_{i_1}  
\end{align*}
Finally, combine them and we have
\begin{align*}
       B_{6,6}^{j_1,i_1,j_0,i_0} = B_{6,6,1}^{j_1,i_1,j_0,i_0}  +B_{6,6,2}^{j_1,i_1,j_0,i_0}
\end{align*}
{\bf Proof of Part 7.} 
    \begin{align*}
     B_{6,7,1}^{j_1,i_1,j_0,i_0} : = & ~  f(A,x)_{j_1} \cdot f(A,x)_{j_0}^2 \cdot  x_{i_1}\cdot c_g(A,x)^{\top} \cdot \diag(x) A^{\top} \cdot  f(A,x)   
\end{align*}
Finally, combine them and we have
\begin{align*}
       B_{6,7}^{j_1,i_1,j_0,i_0} = B_{6,7,1}^{j_1,i_1,j_0,i_0}  
\end{align*}
\end{proof}
\end{lemma}

\subsection{Constructing \texorpdfstring{$d \times d$}{} matrices for \texorpdfstring{$j_1 \neq j_0$}{}}
The purpose of the following lemma is to let $i_0$ and $i_1$ disappear.
\begin{lemma}For $j_0,j_1 \in [n]$, a list of $d \times d$ matrices can be expressed as the following sense,\label{lem:b_6_j1_j0}
\begin{itemize}
\item {\bf Part 1.}
\begin{align*}
B_{6,1,1}^{j_1,*,j_0,*} & ~ =   f_c(A,x) \cdot   f(A,x)_{j_1} \cdot f(A,x)_{j_0} \cdot  (-f_c(A,x) + f(A,x)_{j_0}) \cdot h(A,x) \cdot {\bf 1}_d^\top
\end{align*}
\item {\bf Part 2.}
\begin{align*}
B_{6,1,2}^{j_1,*,j_0,*} & ~ =    c(A,x)_{j_1}\cdot  f(A,x)_{j_1} \cdot f(A,x)_{j_0} \cdot  (-f_c(A,x) + f(A,x)_{j_0}) \cdot h(A,x) \cdot {\bf 1}_d^\top 
\end{align*}
\item {\bf Part 3.}
\begin{align*}
B_{6,1,3}^{j_1,*,j_0,*} & ~ =     f(A,x)_{j_1} \cdot f(A,x)_{j_0} \cdot f_c(A,x) \cdot (-f_c(A,x) + f(A,x)_{j_0}) \cdot ( (A_{j_1,*}) \circ x^\top  )  \cdot h(A,x)\cdot I_d
\end{align*}
\item {\bf Part 4.}
\begin{align*}
B_{6,1,4}^{j_1,*,j_0,*}  & ~ =   -   f(A,x)_{j_1} \cdot f(A,x)_{j_0} \cdot f_c(A,x)  \cdot(-f_c(A,x) + f(A,x)_{j_0}) \cdot h(A,x)^\top \cdot h(A,x) \cdot I_d
\end{align*}
\item {\bf Part 5.}
\begin{align*}
B_{6,1,5}^{j_1,*,j_0,*}  & ~ =    f(A,x)_{j_1} \cdot f(A,x)_{j_0} \cdot (-f_2(A,x) + f(A,x)_{j_1}) \cdot(-f_c(A,x) + f(A,x)_{j_0}) \cdot h(A,x)^\top \cdot h(A,x) \cdot I_d
\end{align*}
\item {\bf Part 6.}
\begin{align*}
B_{6,1,6}^{j_1,*,j_0,*}  & ~ =      f(A,x)_{j_1} \cdot f(A,x)_{j_0} \cdot (-f_c(A,x) + f(A,x)_{j_0})\cdot(-f_c(A,x) + f(A,x)_{j_1}) \cdot h(A,x)^\top \cdot h(A,x) \cdot I_d
\end{align*}
\item {\bf Part 7.}
\begin{align*}
B_{6,1,7}^{j_1,*,j_0,*}  & ~ =    f(A,x)_{j_1} \cdot f(A,x)_{j_0} \cdot (-f_c(A,x) + f(A,x)_{j_0})\cdot p_{j_1}(A,x)^\top \cdot h(A,x) \cdot I_d
\end{align*}
\item {\bf Part 8.}
\begin{align*}
B_{6,2,1}^{j_1,*,j_0,*}  & ~ =      f(A,x)_{j_1} \cdot f(A,x)_{j_0} \cdot  (-f_c(A,x) + f(A,x)_{j_0}) \cdot c_g(A,x)^{\top} \cdot h(A,x) \cdot x \cdot {\bf 1}_d^{\top} 
\end{align*}
\item {\bf Part 9.}
\begin{align*}
 B_{6,3,1}^{j_1,*,j_0,*}  & ~ =      -  f(A,x)_{j_1} \cdot f(A,x)_{j_0} \cdot (-f_c(A,x) + f(A,x)_{j_0})\cdot {\bf 1}_d \cdot c_g(A,x)^{\top}   \cdot \diag(x)  
\end{align*}
\item {\bf Part 10.}
\begin{align*}
B_{6,4,1}^{j_1,*,j_0,*}  =     -  f(A,x)_{j_1} \cdot f(A,x)_{j_0} \cdot (-f_c(A,x) + f(A,x)_{j_0}) \cdot c_g(A,x)^{\top}  \cdot  h_e(A,x)\cdot x \cdot {\bf 1}_d^{\top}
\end{align*}
\item {\bf Part 11.}
\begin{align*}
B_{6,5,1}^{j_1,*,j_0,*}  & ~ =    -  f(A,x)_{j_1} \cdot f(A,x)_{j_0} \cdot f_2(A,x) \cdot c_g(A,x)^{\top} \cdot h(A,x)  \cdot x \cdot {\bf 1}_d^{\top}
\end{align*}
\item {\bf Part 13.}
\begin{align*}
 B_{6,5,2}^{j_1,*,j_0,*}  =     f(A,x)_{j_1}^2 \cdot f(A,x)_{j_0}  \cdot c_g(A,x)^{\top} \cdot h(A,x)  \cdot x \cdot {\bf 1}_d^{\top}
\end{align*}
\item {\bf Part 13.}
\begin{align*}
B_{6,6,1}^{j_1,*,j_0,*}  =   -  f(A,x)_{j_1} \cdot f(A,x)_{j_0} \cdot f_c(A,x) \cdot c_g(A,x)^{\top} \cdot h(A,x)  \cdot x \cdot {\bf 1}_d^{\top}
\end{align*}
\item {\bf Part 14.}
\begin{align*}
B_{6,6,2}^{j_1,*,j_0,*}  =    f(A,x)_{j_1} \cdot c(A,x)_{j_1} \cdot f(A,x)_{j_0}  \cdot c_g(A,x)^{\top} \cdot h(A,x)  \cdot x \cdot {\bf 1}_d^{\top}
\end{align*}
\item {\bf Part 15.}
\begin{align*}
B_{6,7,1}^{j_1,*,j_0,*}  =    f(A,x)_{j_1}\cdot f(A,x)_{j_0}^2   \cdot c_g(A,x)^{\top} \cdot h(A,x)  \cdot x \cdot {\bf 1}_d^{\top}
\end{align*}

\end{itemize}
\begin{proof}
{\bf Proof of Part 1.}
    We have
    \begin{align*}
        B_{6,1,1}^{j_1,i_1,j_0,i_1}  = & ~e_{i_1}^\top \cdot \langle c(A,x), f(A,x) \rangle \cdot  f(A,x)_{j_1} \cdot f(A,x)_{j_0} \cdot \diag(x) \cdot A^{\top} \\
        & ~ \cdot f(A,x) \cdot(\langle -f(A,x), c(A,x) \rangle + f(A,x)_{j_0})\\
        B_{6,1,1}^{j_1,i_1,j_0,i_0}  = & ~ e_{i_1}^\top \cdot \langle c(A,x), f(A,x) \rangle \cdot  f(A,x)_{j_1} \cdot f(A,x)_{j_0} \cdot \diag(x) \cdot A^{\top} \\
        & ~ \cdot f(A,x) \cdot(\langle -f(A,x), c(A,x) \rangle + f(A,x)_{j_0})
    \end{align*}
    From the above two equations, we can tell that $B_{6,1,1}^{j_1,*,j_0,*} \in \R^{d \times d}$ is a matrix that both the diagonal and off-diagonal have entries.
    
    Then we have $B_{6,1,1}^{j_1,*,j_0,*} \in \R^{d \times d}$ can be written as the rescaling of a diagonal matrix,
    \begin{align*}
     B_{6,1,1}^{j_1,*,j_0,*} & ~ = \langle c(A,x), f(A,x) \rangle \cdot  f(A,x)_{j_1} \cdot f(A,x)_{j_0} \\
        & ~ \cdot  (\langle -f(A,x), c(A,x) \rangle + f(A,x)_{j_0}) \cdot \diag(x) \cdot A^{\top} \cdot f(A,x) \cdot {\bf 1}_d^\top \\
     & ~ = f_c(A,x) \cdot  f(A,x)_{j_1} \cdot f(A,x)_{j_0} \cdot  (-f_c(A,x) + f(A,x)_{j_0}) \cdot h(A,x) \cdot {\bf 1}_d^\top
\end{align*}
    where the last step is follows from the Definitions~\ref{def:h} and Definitions~\ref{def:f_c}. 

{\bf Proof of Part 2.}
    We have
    \begin{align*}
           B_{6,1,2}^{j_1,i_1,j_0,i_1} = & ~ e_{i_1}^\top \cdot c(A,x)_{j_1}\cdot f(A,x)_{j_1} \cdot f(A,x)_{j_0} \cdot \diag(x) \cdot A^{\top} \cdot f(A,x) \cdot(\langle -f(A,x), c(A,x) \rangle + f(A,x)_{j_0})\\
        B_{6,1,2}^{j_1,i_1,j_0,i_0} = & ~ e_{i_1}^\top \cdot c(A,x)_{j_1}\cdot f(A,x)_{j_1} \cdot f(A,x)_{j_0} \cdot \diag(x) \cdot A^{\top} \cdot f(A,x) \cdot(\langle -f(A,x), c(A,x) \rangle + f(A,x)_{j_0})
    \end{align*}
     From the above two equations, we can tell that $B_{6,1,2}^{j_1,*,j_0,*} \in \R^{d \times d}$ is a matrix that only diagonal has entries and off-diagonal are all zeros.
    
    Then we have $B_{6,1,2}^{j_1,*,j_0,*} \in \R^{d \times d}$ can be written as the rescaling of a diagonal matrix,
\begin{align*}
     B_{6,1,2}^{j_1,*,j_0,*} & ~ = c(A,x)_{j_1}\cdot f(A,x)_{j_1} \cdot f(A,x)_{j_0} \cdot  (\langle -f(A,x), c(A,x) \rangle + f(A,x)_{j_0}) \cdot \diag(x) \cdot A^{\top} \cdot f(A,x) \cdot {\bf 1}_d^\top \\
     & ~ =  c(A,x)_{j_1}\cdot f(A,x)_{j_1} \cdot f(A,x)_{j_0} \cdot  (-f_c(A,x) + f(A,x)_{j_0}) \cdot h(A,x) \cdot {\bf 1}_d^\top 
\end{align*}
    where the last step is follows from the Definitions~\ref{def:h} and Definitions~\ref{def:f_c}.

{\bf Proof of Part 3.}
We have for diagonal entry and off-diagonal entry can be written as follows 
    \begin{align*}
        B_{6,1,3}^{j_1,i_1,j_0,i_1} = & ~f(A,x)_{j_1} \cdot f(A,x)_{j_0} \cdot \langle c(A,x), f(A,x) \rangle \cdot ( (A_{j_1,*}) \circ x^\top  )  \cdot \diag(x) \cdot A^{\top} \cdot f(A,x) \\
     & ~\cdot (\langle -f(A,x), c(A,x) \rangle + f(A,x)_{j_0}) \\
        B_{6,1,3}^{j_1,i_1,j_0,i_0} = & ~f(A,x)_{j_1} \cdot f(A,x)_{j_0} \cdot \langle c(A,x), f(A,x) \rangle \cdot ( (A_{j_1,*}) \circ x^\top  )  \cdot \diag(x) \cdot A^{\top} \cdot f(A,x) \\
     & ~\cdot (\langle -f(A,x), c(A,x) \rangle + f(A,x)_{j_0})
    \end{align*}
From the above equation, we can show that matrix $B_{6,1,3}^{j_1,*,j_0,*}$ can be expressed as a rank-$1$ matrix,
\begin{align*}
     B_{6,1,3}^{j_1,*,j_0,*} & ~ = f(A,x)_{j_1} \cdot f(A,x)_{j_0} \cdot \langle c(A,x), f(A,x) \rangle \cdot (\langle -f(A,x), c(A,x) \rangle + f(A,x)_{j_0}) \cdot ( (A_{j_1,*}) \circ x^\top  )  \\
     & ~\cdot \diag(x) \cdot A^{\top} \cdot f(A,x) \cdot I_d\\
     & ~ =  f(A,x)_{j_1} \cdot f(A,x)_{j_0} \cdot f_c(A,x) \cdot (-f_c(A,x) + f(A,x)_{j_0}) \cdot ( (A_{j_1,*}) \circ x^\top  )  \cdot h(A,x)\cdot I_d
\end{align*}
    where the last step is follows from the Definitions~\ref{def:h} and Definitions~\ref{def:f_c}.

{\bf Proof of Part 4.}
We have for diagonal entry and off-diagonal entry can be written as follows
    \begin{align*}
        B_{6,1,4}^{j_1,i_1,j_0,i_1}   = & ~  -   f(A,x)_{j_1} \cdot f(A,x)_{j_0} \cdot f(A,x)^\top  \cdot A \cdot (\diag(x))^2 \cdot   \langle c(A,x), f(A,x) \rangle \cdot A^{\top} \cdot f(A,x) \\
     & ~\cdot(\langle -f(A,x), c(A,x) \rangle + f(A,x)_{j_0}) \\
        B_{6,1,4}^{j_1,i_1,j_0,i_0}   = & ~ -     f(A,x)_{j_1} \cdot f(A,x)_{j_0} \cdot f(A,x)^\top  \cdot A \cdot (\diag(x))^2 \cdot   \langle c(A,x), f(A,x) \rangle \cdot A^{\top} \cdot f(A,x) \\
     & ~\cdot(\langle -f(A,x), c(A,x) \rangle + f(A,x)_{j_0})
    \end{align*}
 From the above equation, we can show that matrix $B_{5,1,4}^{j_1,*,j_0,*}$ can be expressed as a rank-$1$ matrix,
\begin{align*}
    B_{6,1,4}^{j_1,*,j_0,*}  & ~ = -   f(A,x)_{j_1} \cdot f(A,x)_{j_0} \cdot \langle c(A,x), f(A,x) \rangle  \cdot(\langle -f(A,x), c(A,x) \rangle + f(A,x)_{j_0}) \\
    & ~\cdot f(A,x)^\top  \cdot A \cdot (\diag(x))^2 \cdot  A^{\top} \cdot f(A,x)\cdot I_d\\
     & ~ =   -  f(A,x)_{j_1} \cdot f(A,x)_{j_0} \cdot f_c(A,x)  \cdot(-f_c(A,x) + f(A,x)_{j_0}) \cdot h(A,x)^\top \cdot h(A,x) \cdot I_d
\end{align*}
   where the last step is follows from the Definitions~\ref{def:h} and Definitions~\ref{def:f_c}.

{\bf Proof of Part 5.}
We have for diagonal entry and off-diagonal entry can be written as follows
    \begin{align*}
         B_{6,1,5}^{j_1,i_1,j_0,i_0} = & ~    f(A,x)_{j_1} \cdot f(A,x)_{j_0} \cdot f(A,x)^\top  \cdot A \cdot (\diag(x))^2 \cdot (\langle -f(A,x), f(A,x) \rangle + f(A,x)_{j_1})   \\
     & ~\cdot (\langle -f(A,x), c(A,x) \rangle + f(A,x)_{j_0})   \cdot A^{\top} \cdot f(A,x)  \\
         B_{6,1,5}^{j_1,i_1,j_0,i_0} = & ~    f(A,x)_{j_1} \cdot f(A,x)_{j_0} \cdot f(A,x)^\top  \cdot A \cdot (\diag(x))^2 \cdot (\langle -f(A,x), f(A,x) \rangle + f(A,x)_{j_1}) \\
     & ~\cdot (\langle -f(A,x), c(A,x) \rangle + f(A,x)_{j_0})  \cdot A^{\top} \cdot f(A,x) 
    \end{align*}
    From the above equation, we can show that matrix $B_{6,1,5}^{j_1,*,j_0,*}$ can be expressed as a rank-$1$ matrix,
\begin{align*}
    B_{6,1,5}^{j_1,*,j_0,*}  & ~ =  f(A,x)_{j_1} \cdot f(A,x)_{j_0} \cdot (\langle -f(A,x), f(A,x) \rangle + f(A,x)_{j_1}) \cdot (\langle -f(A,x), c(A,x) \rangle + f(A,x)_{j_0}) \\ 
     & ~\cdot f(A,x)^\top  \cdot A \cdot (\diag(x))^2 \cdot  A^{\top} \cdot f(A,x) \cdot I_d\\
     & ~ =    f(A,x)_{j_1} \cdot f(A,x)_{j_0} \cdot (-f_2(A,x) + f(A,x)_{j_1}) \cdot (-f_c(A,x) + f(A,x)_{j_0}) \cdot h(A,x)^\top \cdot h(A,x) \cdot I_d
\end{align*}
    where the last step is follows from the Definitions~\ref{def:h}, Definitions~\ref{def:f_c} and Definitions~\ref{def:f_2}.

{\bf Proof of Part 6.}
We have for diagonal entry and off-diagonal entry can be written as follows
    \begin{align*}
        B_{6,1,6}^{j_1,i_1,j_0,i_1}  = & ~   f(A,x)_{j_1} \cdot f(A,x)_{j_0}  \cdot f(A,x)^\top  \cdot A \cdot  (\diag(x))^2 \cdot(\langle -f(A,x), c(A,x) \rangle + f(A,x)_{j_1}) \cdot A^{\top} \cdot f(A,x) \\
        & ~ \cdot(\langle -f(A,x), c(A,x) \rangle + f(A,x)_{j_0})\\
        B_{6,1,6}^{j_1,i_1,j_0,i_0}  = & ~   f(A,x)_{j_1} \cdot f(A,x)_{j_0}  \cdot f(A,x)^\top  \cdot A \cdot  (\diag(x))^2 \cdot(\langle -f(A,x), c(A,x) \rangle + f(A,x)_{j_1}) \cdot A^{\top} \cdot f(A,x) \\
        & ~ \cdot(\langle -f(A,x), c(A,x) \rangle + f(A,x)_{j_0})
    \end{align*}
    From the above equation, we can show that matrix $B_{6,1,6}^{j_1,*,j_0,*}$ can be expressed as a rank-$1$ matrix,
\begin{align*}
    B_{6,1,6}^{j_1,*,j_0,*}  & ~ =   f(A,x)_{j_1} \cdot f(A,x)_{j_0}  \cdot (\langle -f(A,x), c(A,x) \rangle + f(A,x)_{j_0})\cdot(\langle -f(A,x), c(A,x) \rangle + f(A,x)_{j_1}) \\
    & ~\cdot f(A,x)^\top  \cdot A \cdot (\diag(x))^2 \cdot  A^{\top} \cdot f(A,x) \cdot I_d\\
     & ~ =   f(A,x)_{j_1} \cdot f(A,x)_{j_0}  \cdot (-f_c(A,x) + f(A,x)_{j_0})\cdot(-f_c(A,x) + f(A,x)_{j_1}) \cdot h(A,x)^\top \cdot h(A,x) \cdot I_d
\end{align*}
    where the last step is follows from the Definitions~\ref{def:h} and Definitions~\ref{def:f_c}.
    
{\bf Proof of Part 7.}
We have for diagonal entry and off-diagonal entry can be written as follows
    \begin{align*}
         B_{6,1,7}^{j_1,i_1,j_0,i_1} = & ~  f(A,x)_{j_1} \cdot f(A,x)_{j_0} \cdot ((e_{j_1}^\top - f(A,x)^\top) \circ q(A,x)^\top) \cdot A \cdot  (\diag(x))^2 \cdot A^{\top} \cdot f(A,x) \\
    & ~\cdot(\langle -f(A,x), c(A,x) \rangle + f(A,x)_{j_0})\\
         B_{6,1,7}^{j_1,i_1,j_0,i_0} = & ~  f(A,x)_{j_1} \cdot f(A,x)_{j_0} \cdot ((e_{j_1}^\top - f(A,x)^\top) \circ q(A,x)^\top) \cdot A \cdot  (\diag(x))^2 \cdot A^{\top} \cdot f(A,x) \\
    & ~\cdot(\langle -f(A,x), c(A,x) \rangle + f(A,x)_{j_0})
    \end{align*}
    From the above equation, we can show that matrix $B_{6,1,7}^{j_1,*,j_0,*}$ can be expressed as a rank-$1$ matrix,
\begin{align*}
     B_{6,1,7}^{j_1,*,j_0,*}  & ~ =   f(A,x)_{j_1} \cdot f(A,x)_{j_0} \cdot (\langle -f(A,x), c(A,x) \rangle + f(A,x)_{j_0}) \\
     &~\cdot ((e_{j_1}^\top - f(A,x)^\top) \circ q(A,x)^\top) \cdot A \cdot  (\diag(x))^2 \cdot A^{\top} \cdot f(A,x) \cdot I_d\\
     & ~ = f(A,x)_{j_1} \cdot f(A,x)_{j_0} \cdot (-f_c(A,x) + f(A,x)_{j_0})\cdot p_{j_1}(A,x)^\top \cdot h(A,x) \cdot I_d
\end{align*}
    where the last step is follows from the Definitions~\ref{def:h}, Definitions~\ref{def:f_c} and Definitions~\ref{def:p}.

    {\bf Proof of Part 8.}
We have for diagonal entry and off-diagonal entry can be written as follows
    \begin{align*}
         B_{6,2,1}^{j_1,i_1,j_0,i_1} = & ~   f(A,x)_{j_1} \cdot f(A,x)_{j_0} \cdot x_{i_1} \cdot c_g(A,x)^{\top} \cdot \diag(x) \cdot A^{\top} \cdot f(A,x) \cdot(\langle -f(A,x), c(A,x) \rangle + f(A,x)_{j_0})\\
         B_{6,2,1}^{j_1,i_1,j_0,i_0} = & ~  f(A,x)_{j_1} \cdot f(A,x)_{j_0} \cdot x_{i_1} \cdot c_g(A,x)^{\top} \cdot \diag(x) \cdot A^{\top} \cdot f(A,x) \cdot(\langle -f(A,x), c(A,x) \rangle + f(A,x)_{j_0})
    \end{align*}
    From the above equation, we can show that matrix $B_{6,2,1}^{j_1,*,j_0,*}$ can be expressed as a rank-$1$ matrix,
\begin{align*}
     B_{6,2,1}^{j_1,*,j_0,*}  & ~ =   f(A,x)_{j_1} \cdot f(A,x)_{j_0} \cdot  (\langle -f(A,x), c(A,x) \rangle + f(A,x)_{j_0}) \cdot c_g(A,x)^{\top} \cdot \diag(x) \cdot A^{\top} \cdot f(A,x) \cdot x \cdot {\bf 1}_d^{\top}  \\ 
     & ~ =  f(A,x)_{j_1} \cdot f(A,x)_{j_0} \cdot  (-f_c(A,x) + f(A,x)_{j_0}) \cdot c_g(A,x)^{\top} \cdot h(A,x) \cdot x \cdot {\bf 1}_d^{\top} 
\end{align*}
    where the last step is follows from the Definitions~\ref{def:h}, Definitions~\ref{def:f_c}.

   {\bf Proof of Part 9.}
We have for diagonal entry and off-diagonal entry can be written as follows
    \begin{align*}
         B_{6,3,1}^{j_1,i_1,j_0,i_1} = & ~   -  c_g(A,x)^{\top} \cdot f(A,x)_{j_0} \cdot \diag(x) \cdot e_{i_1} \cdot e_{j_1}^\top \cdot f(A,x) \cdot(\langle -f(A,x), c(A,x) \rangle + f(A,x)_{j_0})\\
         B_{6,3,1}^{j_1,i_1,j_0,i_0} = & ~ -  c_g(A,x)^{\top} \cdot f(A,x)_{j_0} \cdot \diag(x) \cdot e_{i_1} \cdot e_{j_1}^\top \cdot f(A,x) \cdot(\langle -f(A,x), c(A,x) \rangle + f(A,x)_{j_0})
    \end{align*}
    From the above equation, we can show that matrix $B_{6,3,1}^{j_1,*,j_0,*}$ can be expressed as a rank-$1$ matrix,
\begin{align*}
     B_{6,3,1}^{j_1,*,j_0,*}  & ~ = - f(A,x)_{j_1} \cdot f(A,x)_{j_0} \cdot (\langle -f(A,x), c(A,x) \rangle + f(A,x)_{j_0})\cdot {\bf 1}_d \cdot c_g(A,x)^{\top}   \cdot \diag(x)  \\ 
     & ~ =   - f(A,x)_{j_1} \cdot f(A,x)_{j_0} \cdot (-f_c(A,x) + f(A,x)_{j_0})\cdot {\bf 1}_d \cdot c_g(A,x)^{\top}   \cdot \diag(x)  
\end{align*}
    where the last step is follows from the Definitions~\ref{def:f_c}.

    {\bf Proof of Part 10.}
We have for diagonal entry and off-diagonal entry can be written as follows
    \begin{align*}
         B_{6,4,1}^{j_1,i_1,j_0,i_1} = & ~   - c_g(A,x)^{\top} \cdot  f(A,x)_{j_1} \cdot f(A,x)_{j_0} \cdot \diag(x) \cdot A^{\top}\cdot x_i \cdot \\
        & ~  (e_{j_1}- f(A,x) ) \cdot (\langle -f(A,x), c(A,x) \rangle + f(A,x)_{j_0})\\
         B_{6,4,1}^{j_1,i_1,j_0,i_0} = & ~  - c_g(A,x)^{\top} \cdot  f(A,x)_{j_1} \cdot f(A,x)_{j_0} \cdot \diag(x) \cdot A^{\top}\cdot x_i \cdot \\
        & ~  (e_{j_1}- f(A,x) ) \cdot (\langle -f(A,x), c(A,x) \rangle + f(A,x)_{j_0})
    \end{align*}
    From the above equation, we can show that matrix $B_{6,4,1}^{j_1,*,j_0,*}$ can be expressed as a rank-$1$ matrix,
\begin{align*}
     B_{6,4,1}^{j_1,*,j_0,*}  & ~ = -  f(A,x)_{j_1} \cdot f(A,x)_{j_0} \cdot (\langle -f(A,x), c(A,x) \rangle + f(A,x)_{j_0}) \cdot c_g(A,x)^{\top} \\ 
     & ~  \cdot \diag(x) \cdot A^{\top}  \cdot  (e_{j_1}- f(A,x) ) \cdot x \cdot {\bf 1}_d^{\top}\\ 
     & ~ =   -  f(A,x)_{j_1} \cdot f(A,x)_{j_0} \cdot (-f_c(A,x) + f(A,x)_{j_0}) \cdot c_g(A,x)^{\top}  \cdot  h_e(A,x)\cdot x \cdot {\bf 1}_d^{\top}
\end{align*}
    where the last step is follows from the Definitions~\ref{def:f_c} and Definitions~\ref{def:h_e}.

        {\bf Proof of Part 11.}
We have for diagonal entry and off-diagonal entry can be written as follows
    \begin{align*}
         B_{6,5,1}^{j_1,i_1,j_0,i_1} = & ~  c_g(A,x)^{\top} \cdot f(A,x)_{j_1} \cdot f(A,x)_{j_0} \cdot \diag(x) \cdot A^{\top} \cdot  f(A,x)  \cdot  x_{i_1} \cdot \langle - f(A,x), f(A,x) \rangle\\
         B_{6,5,1}^{j_1,i_1,j_0,i_0} = & ~ c_g(A,x)^{\top} \cdot f(A,x)_{j_1} \cdot f(A,x)_{j_0} \cdot \diag(x) \cdot A^{\top} \cdot  f(A,x)  \cdot  x_{i_1} \cdot \langle - f(A,x), f(A,x) \rangle
    \end{align*}
    From the above equation, we can show that matrix $B_{6,5,1}^{j_1,*,j_0,*}$ can be expressed as a rank-$1$ matrix,
\begin{align*}
     B_{6,5,1}^{j_1,*,j_0,*}  & ~ = f(A,x)_{j_1} \cdot f(A,x)_{j_0} \cdot \langle - f(A,x), f(A,x) \rangle \cdot c_g(A,x)^{\top} \cdot \diag(x) \cdot A^{\top} \cdot  f(A,x)  \cdot x \cdot {\bf 1}_d^{\top}\\ 
     & ~ =  -  f(A,x)_{j_1} \cdot f(A,x)_{j_0} \cdot f_2(A,x) \cdot c_g(A,x)^{\top} \cdot h(A,x)  \cdot x \cdot {\bf 1}_d^{\top}
\end{align*}
    where the last step is follows from the Definitions~\ref{def:h} and Definitions~\ref{def:f_2}.

    {\bf Proof of Part 12.}
We have for diagonal entry and off-diagonal entry can be written as follows
    \begin{align*}
         B_{6,5,2}^{j_1,i_1,j_0,i_1} = & ~  c_g(A,x)^{\top} \cdot f(A,x)_{j_1}^2 \cdot f(A,x)_{j_0} \cdot \diag(x) \cdot A^{\top} \cdot  f(A,x)  \cdot  x_{i_1}  \\
         B_{6,5,2}^{j_1,i_1,j_0,i_0} = & ~ c_g(A,x)^{\top} \cdot f(A,x)_{j_1}^2 \cdot f(A,x)_{j_0} \cdot \diag(x) \cdot A^{\top} \cdot  f(A,x)  \cdot  x_{i_1}  
    \end{align*}
    From the above equation, we can show that matrix $B_{6,5,2}^{j_1,*,j_0,*}$ can be expressed as a rank-$1$ matrix,
\begin{align*}
     B_{6,5,2}^{j_1,*,j_0,*}  & ~ = f(A,x)_{j_1}^2 \cdot f(A,x)_{j_0}   \cdot c_g(A,x)^{\top} \cdot \diag(x) \cdot A^{\top} \cdot  f(A,x)  \cdot x \cdot {\bf 1}_d^{\top}\\ 
     & ~ =  f(A,x)_{j_1}^2 \cdot f(A,x)_{j_0}  \cdot c_g(A,x)^{\top} \cdot h(A,x)  \cdot x \cdot {\bf 1}_d^{\top}
\end{align*}
    where the last step is follows from the Definitions~\ref{def:h}.

        {\bf Proof of Part 13.}
We have for diagonal entry and off-diagonal entry can be written as follows
    \begin{align*}
         B_{6,6,1}^{j_1,i_1,j_0,i_1} = & ~  c_g(A,x)^{\top} \cdot f(A,x)_{j_1} \cdot f(A,x)_{j_0} \cdot \diag(x) \cdot A^{\top} \cdot  f(A,x)  \cdot  x_{i_1} \cdot \langle - f(A,x), c(A,x) \rangle\\
         B_{6,6,1}^{j_1,i_1,j_0,i_0} = & ~ c_g(A,x)^{\top} \cdot f(A,x)_{j_1} \cdot f(A,x)_{j_0} \cdot \diag(x) \cdot A^{\top} \cdot  f(A,x)  \cdot  x_{i_1} \cdot \langle - f(A,x), c(A,x) \rangle
    \end{align*}
    From the above equation, we can show that matrix $B_{6,6,1}^{j_1,*,j_0,*}$ can be expressed as a rank-$1$ matrix,
\begin{align*}
     B_{6,6,1}^{j_1,*,j_0,*}  & ~ = f(A,x)_{j_1} \cdot f(A,x)_{j_0} \cdot \langle - f(A,x), c(A,x) \rangle \cdot c_g(A,x)^{\top} \cdot \diag(x) \cdot A^{\top} \cdot  f(A,x)  \cdot x \cdot {\bf 1}_d^{\top}\\ 
     & ~ =  -  f(A,x)_{j_1} \cdot f(A,x)_{j_0} \cdot f_c(A,x) \cdot c_g(A,x)^{\top} \cdot h(A,x)  \cdot x \cdot {\bf 1}_d^{\top}
\end{align*}
    where the last step is follows from the Definitions~\ref{def:h} and Definitions~\ref{def:f_2}.

    {\bf Proof of Part 14.}
We have for diagonal entry and off-diagonal entry can be written as follows
    \begin{align*}
         B_{6,6,2}^{j_1,i_1,j_0,i_1} = & ~  c_g(A,x)^{\top} \cdot f(A,x)_{j_1} \cdot c(A,x)_{j_1} \cdot f(A,x)_{j_0} \cdot \diag(x) \cdot A^{\top} \cdot  f(A,x)  \cdot  x_{i_1}   \\
         B_{6,6,2}^{j_1,i_1,j_0,i_0} = & ~c_g(A,x)^{\top} \cdot f(A,x)_{j_1} \cdot c(A,x)_{j_1} \cdot f(A,x)_{j_0} \cdot \diag(x) \cdot A^{\top} \cdot  f(A,x)  \cdot  x_{i_1} 
    \end{align*}
    From the above equation, we can show that matrix $B_{6,6,2}^{j_1,*,j_0,*}$ can be expressed as a rank-$1$ matrix,
\begin{align*}
     B_{6,6,2}^{j_1,*,j_0,*}  & ~ = f(A,x)_{j_1} \cdot c(A,x)_{j_1} \cdot f(A,x)_{j_0}   \cdot c_g(A,x)^{\top} \cdot \diag(x) \cdot A^{\top} \cdot  f(A,x)  \cdot x \cdot {\bf 1}_d^{\top}\\ 
     & ~ =  f(A,x)_{j_1} \cdot c(A,x)_{j_1} \cdot f(A,x)_{j_0}  \cdot c_g(A,x)^{\top} \cdot h(A,x)  \cdot x \cdot {\bf 1}_d^{\top}
\end{align*}
    where the last step is follows from the Definitions~\ref{def:h}.

    {\bf Proof of Part 15.}
We have for diagonal entry and off-diagonal entry can be written as follows
    \begin{align*}
         B_{6,7,1}^{j_1,i_1,j_0,i_1} = & ~  f(A,x)_{j_1}\cdot f(A,x)_{j_0}^2  \cdot  x_{i_1}\cdot c_g(A,x)^{\top} \cdot \diag(x) A^{\top} \cdot  f(A,x)  \\
         B_{6,7,1}^{j_1,i_1,j_0,i_0} = & ~ f(A,x)_{j_1}\cdot f(A,x)_{j_0}^2  \cdot  x_{i_1}\cdot c_g(A,x)^{\top} \cdot \diag(x) A^{\top} \cdot  f(A,x) 
    \end{align*}
    From the above equation, we can show that matrix $B_{6,7,1}^{j_1,*,j_0,*}$ can be expressed as a rank-$1$ matrix,
\begin{align*}
     B_{6,7,1}^{j_1,*,j_0,*}  & ~ = f(A,x)_{j_1}\cdot f(A,x)_{j_0}^2    \cdot c_g(A,x)^{\top} \cdot \diag(x) \cdot A^{\top} \cdot  f(A,x)  \cdot x \cdot {\bf 1}_d^{\top}\\ 
     & ~ =  f(A,x)_{j_1}\cdot f(A,x)_{j_0}^2   \cdot c_g(A,x)^{\top} \cdot h(A,x)  \cdot x \cdot {\bf 1}_d^{\top}
\end{align*}
    where the last step is follows from the Definitions~\ref{def:h}.
\end{proof}
\end{lemma}

\subsection{Expanding \texorpdfstring{$B_6$}{} into many terms}

\begin{lemma}
   If the following conditions hold
    \begin{itemize}
     \item Let $u(A,x) \in \R^n$ be defined as Definition~\ref{def:u}
    \item Let $\alpha(A,x) \in \R$ be defined as Definition~\ref{def:alpha}
     \item Let $f(A,x) \in \R^n$ be defined as Definition~\ref{def:f}
    \item Let $c(A,x) \in \R^n$ be defined as Definition~\ref{def:c}
    \item Let $g(A,x) \in \R^d$ be defined as Definition~\ref{def:g} 
    \item Let $q(A,x) = c(A,x) + f(A,x) \in \R^n$
    \item Let $c_g(A,x) \in \R^d$ be defined as Definition~\ref{def:c_g}.
    \item Let $L_g(A,x) \in \R$ be defined as Definition~\ref{def:l_g}
    \item Let $v \in \R^n$ be a vector 
    \end{itemize}
Then, For $j_0,j_1 \in [n], i_0,i_1 \in [d]$, we have 
\begin{itemize}
    \item {\bf Part 1.}For $j_1 = j_0$ and $i_0 = i_1$
\begin{align*}
    B_6^{j_1,i_1,j_1,i_1} = B_{6,1}^{j_1,i_1,j_1,i_1} +  B_{6,2}^{j_1,i_1,j_1,i_1} + B_{6,3}^{j_1,i_1,j_1,i_1} + B_{6,4}^{j_1,i_1,j_1,i_1} + B_{6,5}^{j_1,i_1,j_1,i_1} + B_{6,6}^{j_1,i_1,j_1,i_1} + B_{6,7}^{j_1,i_1,j_1,i_1}
\end{align*}
\item {\bf Part 2.}For $j_1 = j_0$ and $i_0 \neq i_1$
\begin{align*}
      B_6^{j_1,i_1,j_1,i_0} = B_{6,1}^{j_1,i_1,j_1,i_0} +  B_{6,2}^{j_1,i_1,j_1,i_0} + B_{6,3}^{j_1,i_1,j_1,i_0} + B_{6,4}^{j_1,i_1,j_1,i_0} + B_{6,5}^{j_1,i_1,j_1,i_0} + B_{6,6}^{j_1,i_1,j_1,i_0} + B_{6,7}^{j_1,i_1,j_1,i_0}
\end{align*}
\item {\bf Part 3.}For $j_1 \neq j_0$ and $i_0 = i_1$ 
\begin{align*}
    B_6^{j_1,i_1,j_0,i_1} = B_{6,1}^{j_1,i_1,j_0,i_1} +  B_{6,2}^{j_1,i_1,j_0,i_1} + B_{6,3}^{j_1,i_1,j_0,i_1} + B_{6,4}^{j_1,i_1,j_0,i_1} + B_{6,5}^{j_1,i_1,j_0,i_1} + B_{6,6}^{j_1,i_1,j_0,i_1} + B_{6,7}^{j_1,i_1,j_0,i_1}
\end{align*}
\item {\bf Part 4.} For $j_0 \neq j_1$ and $i_0 \neq i_1$
\begin{align*}
  B_6^{j_1,i_1,j_0,i_0} = B_{6,1}^{j_1,i_1,j_0,i_0} +  B_{6,2}^{j_1,i_1,j_0,i_0} + B_{6,3}^{j_1,i_1,j_0,i_0} + B_{6,4}^{j_1,i_1,j_0,i_0} + B_{6,5}^{j_1,i_1,j_0,i_0} + B_{6,6}^{j_1,i_1,j_0,i_0} + B_{6,7}^{j_1,i_1,j_0,i_0}
\end{align*}
\end{itemize}
\end{lemma}
\begin{proof}
{\bf Proof of Part 1.} we have
    \begin{align*}
     B_6^{j_1,i_1,j_1,i_1} & ~ = \frac{\d}{\d A_{j_1,i_1}} (- c_g(A,x)^{\top} \cdot  f(A,x)_{j_1}\diag(x) A^{\top} f(A,x) \cdot(\langle -f(A,x), c(A,x) \rangle + f(A,x)_{j_1})) \\
     & ~ =  B_{6,1}^{j_1,i_1,j_1,i_1} +  B_{6,2}^{j_1,i_1,j_1,i_1} + B_{6,3}^{j_1,i_1,j_1,i_1} + B_{6,4}^{j_1,i_1,j_1,i_1} + B_{6,5}^{j_1,i_1,j_1,i_1} + B_{6,6}^{j_1,i_1,j_1,i_1} + B_{6,7}^{j_1,i_1,j_1,i_1}
\end{align*}
{\bf Proof of Part 2.}   we have
    \begin{align*}
    B_6^{j_1,i_1,j_1,i_0} & ~ = \frac{\d}{\d A_{j_1,i_1}} (- c_g(A,x)^{\top} \cdot  f(A,x)_{j_1}\diag(x) A^{\top} f(A,x) \cdot(\langle -f(A,x), c(A,x) \rangle + f(A,x)_{j_1})) \\
     & ~ =  B_{6,1}^{j_1,i_1,j_1,i_0} +  B_{6,2}^{j_1,i_1,j_1,i_0} + B_{6,3}^{j_1,i_1,j_1,i_0} + B_{6,4}^{j_1,i_1,j_1,i_0} + B_{6,5}^{j_1,i_1,j_1,i_0} + B_{6,6}^{j_1,i_1,j_1,i_0} + B_{6,7}^{j_1,i_1,j_1,i_0}
\end{align*}
{\bf Proof of Part 3.} we have
    \begin{align*}
   B_6^{j_1,i_1,j_0,i_1} & ~ = \frac{\d}{\d A_{j_1,i_1}} (- c_g(A,x)^{\top} \cdot  f(A,x)_{j_0}\diag(x) A^{\top} f(A,x) \cdot(\langle -f(A,x), c(A,x) \rangle + f(A,x)_{j_0})) \\
     & ~ =  B_{6,1}^{j_1,i_1,j_0,i_1} +  B_{6,2}^{j_1,i_1,j_0,i_1} + B_{6,3}^{j_1,i_1,j_0,i_1} + B_{6,4}^{j_1,i_1,j_0,i_1} + B_{6,5}^{j_1,i_1,j_0,i_1} + B_{6,6}^{j_1,i_1,j_0,i_1} + B_{6,7}^{j_1,i_1,j_0,i_1}
\end{align*}
{\bf Proof of Part 4.}
 we have
\begin{align*}
     B_6^{j_1,i_1,j_0,i_0} & ~ = \frac{\d}{\d A_{j_1,i_1}} (- c_g(A,x)^{\top} \cdot  f(A,x)_{j_0}\diag(x) A^{\top} f(A,x) \cdot(\langle -f(A,x), c(A,x) \rangle + f(A,x)_{j_0})) \\
     & ~ = B_{6,1}^{j_1,i_1,j_0,i_0} +  B_{6,2}^{j_1,i_1,j_0,i_0} + B_{6,3}^{j_1,i_1,j_0,i_0} + B_{6,4}^{j_1,i_1,j_0,i_0} + B_{6,5}^{j_1,i_1,j_0,i_0} + B_{6,6}^{j_1,i_1,j_0,i_0} + B_{6,7}^{j_1,i_1,j_0,i_0}
\end{align*}
\end{proof}
\subsection{Lipschitz Computation}
\begin{lemma}\label{lips: B_6}
If the following conditions hold
\begin{itemize}
    \item Let $B_{6,1,1}^{j_1,*, j_0,*}, \cdots, B_{6,7,1}^{j_1,*, j_0,*} $ be defined as Lemma~\ref{lem:b_6_j1_j0} 
    \item  Let $\|A \|_2 \leq R, \|A^{\top} \|_F \leq R, \| x\|_2 \leq R, \|\diag(f(A,x)) \|_F \leq \|f(A,x) \|_2 \leq 1, \| b_g \|_2 \leq 1$ 
\end{itemize}
Then, we have
\begin{itemize}
    \item {\bf Part 1.}
    \begin{align*}
       \| B_{6,1,1}^{j_1,*,j_0,*} (A) - B_{6,1,1}^{j_1,*,j_0,*} ( \wt{A} ) \|_F \leq  \beta^{-2} \cdot n \cdot \sqrt{d}\exp(5R^2) \cdot \|A - \wt{A}\|_F
    \end{align*}
     \item {\bf Part 2.}
    \begin{align*}
       \| B_{6,1,2}^{j_1,*,j_0,*} (A) - B_{6,1,2}^{j_1,*,j_0,*} ( \wt{A} ) \|_F \leq   \beta^{-2} \cdot n \cdot \sqrt{d}\exp(5R^2) \cdot \|A - \wt{A}\|_F
    \end{align*}
     \item {\bf Part 3.}
    \begin{align*}
       \| B_{6,1,3}^{j_1,*,j_0,*} (A) - B_{6,1,3}^{j_1,*,j_0,*} ( \wt{A} ) \|_F \leq   \beta^{-2} \cdot n \cdot \sqrt{d}\exp(6R^2) \cdot \|A - \wt{A}\|_F
    \end{align*}
     \item {\bf Part 4.}
    \begin{align*}
       \| B_{6,1,4}^{j_1,*,j_0,*} (A) - B_{6,1,4}^{j_1,*,j_0,*} ( \wt{A} ) \|_F \leq   \beta^{-2} \cdot n \cdot \sqrt{d}\exp(6R^2) \cdot \|A - \wt{A}\|_F
    \end{align*}
     \item {\bf Part 5.}
    \begin{align*}
       \| B_{6,1,5}^{j_1,*,j_0,*} (A) - B_{6,1,5}^{j_1,*,j_0,*} ( \wt{A} ) \|_F \leq   \beta^{-2} \cdot n \cdot \sqrt{d}\exp(6R^2) \cdot \|A - \wt{A}\|_F
    \end{align*}
     \item {\bf Part 6.}
    \begin{align*}
       \| B_{6,1,6}^{j_1,*,j_0,*} (A) - B_{6,1,6}^{j_1,*,j_0,*} ( \wt{A} ) \|_F \leq  \beta^{-2} \cdot n \cdot \sqrt{d}\exp(6R^2) \cdot \|A - \wt{A}\|_F
    \end{align*}
     \item {\bf Part 7.}
    \begin{align*}
       \| B_{6,1,7}^{j_1,*,j_0,*} (A) - B_{6,1,7}^{j_1,*,j_0,*} ( \wt{A} ) \|_F \leq   \beta^{-2} \cdot n \cdot \sqrt{d}\exp(6R^2) \cdot \|A - \wt{A}\|_F
    \end{align*}
    \item {\bf Part 8.}
    \begin{align*}
       \| B_{6,2,1}^{j_1,*,j_0,*} (A) - B_{6,2,1}^{j_1,*,j_0,*} ( \wt{A} ) \|_F \leq  \beta^{-2} \cdot n \cdot \sqrt{d}\exp(6R^2) \cdot \|A - \wt{A}\|_F
    \end{align*}
     \item {\bf Part 9.}
    \begin{align*}
       \| B_{6,3,1}^{j_1,*,j_0,*} (A) - B_{6,3,1}^{j_1,*,j_0,*} ( \wt{A} ) \|_F \leq  \beta^{-2} \cdot n \cdot \sqrt{d}  \exp(5R^2)\|A - \wt{A}\|_F  
    \end{align*}
     \item {\bf Part 10.}
    \begin{align*}
       \| B_{6,4,1}^{j_1,*,j_0,*} (A) - B_{6,4,1}^{j_1,*,j_0,*} ( \wt{A} ) \|_F \leq  \beta^{-2} \cdot n \cdot \sqrt{d}  \exp(6R^2)\|A - \wt{A}\|_F  
    \end{align*}
     \item {\bf Part 11.}
    \begin{align*}
       \| B_{6,5,1}^{j_1,*,j_0,*} (A) - B_{6,5,1}^{j_1,*,j_0,*} ( \wt{A} ) \|_F \leq \beta^{-2} \cdot n \cdot \sqrt{d}  \exp(6R^2)\|A - \wt{A}\|_F   
    \end{align*}
     \item {\bf Part 12.}
    \begin{align*}
       \| B_{6,5,2}^{j_1,*,j_0,*} (A) - B_{6,5,2}^{j_1,*,j_0,*} ( \wt{A} ) \|_F \leq  \beta^{-2} \cdot n \cdot \sqrt{d}  \exp(6R^2)\|A - \wt{A}\|_F  
    \end{align*}
     \item {\bf Part 13.}
    \begin{align*}
       \| B_{6,6,1}^{j_1,*,j_0,*} (A) - B_{6,6,1}^{j_1,*,j_0,*} ( \wt{A} ) \|_F \leq    \beta^{-2} \cdot n \cdot \sqrt{d}  \exp(6R^2)\|A - \wt{A}\|_F 
    \end{align*}
         \item {\bf Part 14.}
    \begin{align*}
       \| B_{6,6,2}^{j_1,*,j_0,*} (A) - B_{6,6,2}^{j_1,*,j_0,*} ( \wt{A} ) \|_F \leq  \beta^{-2} \cdot n \cdot \sqrt{d}  \exp(6R^2)\|A - \wt{A}\|_F   
    \end{align*}
             \item {\bf Part 15.}
    \begin{align*}
       \| B_{6,7,1}^{j_1,*,j_0,*} (A) - B_{6,7,1}^{j_1,*,j_0,*} ( \wt{A} ) \|_F \leq  \beta^{-2} \cdot n \cdot \sqrt{d}  \exp(6R^2)\|A - \wt{A}\|_F   
    \end{align*}
        \item{\bf Part 15.}
    \begin{align*}
         \| B_{6}^{j_1,*,j_0,*} (A) - B_{6}^{j_1,*,j_0,*} ( \wt{A} ) \|_F \leq & ~   15\beta^{-2} \cdot n \cdot \sqrt{d}  \exp(6R^2)\|A - \wt{A}\|_F   
    \end{align*}
    \end{itemize}
\end{lemma}
\begin{proof}
{\bf Proof of Part 1.}
\begin{align*}
& ~ \| B_{6,1,1}^{j_1,*,j_0,*} (A) - B_{6,1,1}^{j_1,*,j_0,*} ( \wt{A} ) \| \\ \leq 
    & ~ \| f_c(A,x)  \cdot f(A,x)_{j_1} \cdot f(A,x)_{j_0} \cdot (-f_c(A,x) + f(A,x)_{j_0}) \cdot  h(A,x) \cdot  {\bf 1}_d^\top \\
    & ~ - f_c(\wt{A},x)  \cdot f(\wt{A},x)_{j_1} \cdot f(\wt{A},x)_{j_0} \cdot  (-f_c(\wt{A},x) + f(\wt{A},x)_{j_0}) \cdot h(\wt{A},x) \cdot  {\bf 1}_d^\top \|_F\\
    \leq & ~ |  f_c(A,x) - f_c(\wt{A},x) | \cdot |f(A,x)_{j_1}|\cdot | f(A,x)_{j_0} | \cdot   ( |-f_c(A,x)| +  |f(A,x)_{j_0}|) \cdot \| h(A,x)\|_2  \cdot \| {\bf 1}_d^\top\|_2 \\
    & + ~   |f_c(\wt{A},x)| \cdot  |f(A,x)_{j_1} - f(\wt{A},x)_{j_1}|\cdot | f(A,x)_{j_0} | \cdot  ( |-f_c(A,x)| +  |f(A,x)_{j_0}|) \cdot  \| h(A,x)\|_2  \cdot \| {\bf 1}_d^\top\|_2 \\
    & + ~   |f_c(\wt{A},x)| \cdot  |f(\wt{A},x)_{j_1}|\cdot | f(A,x)_{j_0} -  f(\wt{A},x)_{j_0} | \cdot ( |-f_c(A,x)| +  |f(A,x)_{j_0}|) \cdot   \| h(A,x)\|_2  \cdot \| {\bf 1}_d^\top\|_2 \\
     & + ~   |f_c(\wt{A},x)| \cdot  |f(\wt{A},x)_{j_1}|\cdot | f(\wt{A},x)_{j_0} | \cdot ( | f_c(A,x) -  f_c(\wt{A},x)| +  |f(A,x)_{j_0} - f(\wt{A},x)_{j_0}|) \cdot   \| h(A,x)\|_2  \cdot \| {\bf 1}_d^\top\|_2 \\
    & + ~ |f_c(\wt{A},x)|\cdot   |f(\wt{A},x)_{j_1}|\cdot  | f(\wt{A},x)_{j_0} | \cdot ( |-f_c(\wt{A},x)| +  |f(\wt{A},x)_{j_0}|) \cdot \| h(A,x) - h(\wt{A},x)\|_2  \cdot \| {\bf 1}_d^\top\|_2 \\
    \leq & ~ 18R^2  \beta^{-2} \cdot n \cdot \sqrt{d}  \exp(3R^2)\|A - \wt{A}\|_F \\
    &+ ~ 12R^2  \beta^{-2} \cdot n \cdot \sqrt{d}  \exp(3R^2)\|A - \wt{A}\|_F \\
    & + ~ 12R^2 \beta^{-2} \cdot n \cdot \sqrt{d}\exp(3R^2) \cdot \|A - \wt{A}\|_F \\
    & + ~ 16R^2\beta^{-2} \cdot n \cdot \sqrt{d}\exp(3R^2) \cdot \|A - \wt{A}\|_F \\
    & + ~ 18 \beta^{-2} \cdot n \cdot \sqrt{d}\exp(4R^2) \cdot \|A - \wt{A}\|_F\\
    \leq &  ~ 76 \beta^{-2} \cdot n \cdot \sqrt{d}\exp(4R^2) \cdot \|A - \wt{A}\|_F \\
    \leq &  ~ \beta^{-2} \cdot n \cdot \sqrt{d}\exp(5R^2) \cdot \|A - \wt{A}\|_F 
\end{align*}

{\bf Proof of Part 2.}
\begin{align*}
& ~ \| B_{6,1,2}^{j_1,*,j_0,*} (A) - B_{6,1,2}^{j_1,*,j_0,*} ( \wt{A} ) \| \\ \leq 
    & ~ \|c(A,x)_{j_1}  \cdot f(A,x)_{j_1} \cdot f(A,x)_{j_0} \cdot (-f_c(A,x) + f(A,x)_{j_0}) \cdot  h(A,x) \cdot  {\bf 1}_d^\top \\
    & ~ - c(\wt{A},x)_{j_1} \cdot f(\wt{A},x)_{j_1} \cdot f(\wt{A},x)_{j_0} \cdot  (-f_c(\wt{A},x) + f(\wt{A},x)_{j_0}) \cdot h(\wt{A},x) \cdot  {\bf 1}_d^\top \|_F\\
    \leq & ~ |  c(A,x)_{j_1} - c(\wt{A},x)_{j_1} | \cdot |f(A,x)_{j_1}|\cdot | f(A,x)_{j_0} | \cdot   ( |-f_c(A,x)| +  |f(A,x)_{j_0}|) \cdot \| h(A,x)\|_2  \cdot \| {\bf 1}_d^\top\|_2 \\
    & + ~   | c(\wt{A},x)_{j_1}| \cdot  |f(A,x)_{j_1} - f(\wt{A},x)_{j_1}|\cdot | f(A,x)_{j_0} | \cdot  ( |-f_c(A,x)| +  |f(A,x)_{j_0}|) \cdot  \| h(A,x)\|_2  \cdot \| {\bf 1}_d^\top\|_2 \\
    & + ~   | c(\wt{A},x)_{j_1}| \cdot  |f(\wt{A},x)_{j_1}|\cdot | f(A,x)_{j_0} -  f(\wt{A},x)_{j_0} | \cdot ( |-f_c(A,x)| +  |f(A,x)_{j_0}|) \cdot   \| h(A,x)\|_2  \cdot \| {\bf 1}_d^\top\|_2 \\
     & + ~   | c(\wt{A},x)_{j_1}| \cdot  |f(\wt{A},x)_{j_1}|\cdot | f(\wt{A},x)_{j_0} | \cdot ( | f_c(A,x) -  f_c(\wt{A},x)| +  |f(A,x)_{j_0} - f(\wt{A},x)_{j_0}|) \cdot   \| h(A,x)\|_2  \cdot \| {\bf 1}_d^\top\|_2 \\
    & + ~ | c(\wt{A},x)_{j_1}|\cdot   |f(\wt{A},x)_{j_1}|\cdot  | f(\wt{A},x)_{j_0} | \cdot ( |-f_c(\wt{A},x)| +  |f(\wt{A},x)_{j_0}|) \cdot \| h(A,x) - h(\wt{A},x)\|_2  \cdot \| {\bf 1}_d^\top\|_2 \\
    \leq & ~ 6R^2  \beta^{-2} \cdot n \cdot \sqrt{d}  \exp(3R^2)\|A - \wt{A}\|_F \\
    &+ ~ 12R^2  \beta^{-2} \cdot n \cdot \sqrt{d}  \exp(3R^2)\|A - \wt{A}\|_F \\
    & + ~ 12R^2 \beta^{-2} \cdot n \cdot \sqrt{d}\exp(3R^2) \cdot \|A - \wt{A}\|_F \\
    & + ~ 16R^2\beta^{-2} \cdot n \cdot \sqrt{d}\exp(3R^2) \cdot \|A - \wt{A}\|_F \\
    & + ~ 18 \beta^{-2} \cdot n \cdot \sqrt{d}\exp(4R^2) \cdot \|A - \wt{A}\|_F\\
    \leq &  ~ 64 \beta^{-2} \cdot n \cdot \sqrt{d}\exp(4R^2) \cdot \|A - \wt{A}\|_F \\
    \leq &  ~ \beta^{-2} \cdot n \cdot \sqrt{d}\exp(5R^2) \cdot \|A - \wt{A}\|_F 
\end{align*}

{\bf Proof of Part 3.}
\begin{align*}
& ~ \| B_{6,1,3}^{j_1,*,j_0,*} (A) - B_{6,1,3}^{j_1,*,j_0,*} ( \wt{A} ) \| \\ \leq 
    & ~ \| f_c(A,x)  \cdot f(A,x)_{j_1} \cdot f(A,x)_{j_0} \cdot (-f_c(A,x) + f(A,x)_{j_0}) \cdot ( (A_{j_1,*}) \circ x^\top  )  \cdot h(A,x) \cdot  I_d  \\
    & ~ - f_c(\wt{A},x)  \cdot f(\wt{A},x)_{j_1} \cdot f(\wt{A},x)_{j_0} \cdot  (-f_c(\wt{A},x) + f(\wt{A},x)_{j_0}) \cdot( (\wt{A}_{j_1,*}) \circ x^\top  )  \cdot h(\wt{A},x) \cdot  I_d  \|_F\\
    \leq & ~ |  f_c(A,x) - f_c(\wt{A},x) | \cdot |f(A,x)_{j_1}|\cdot | f(A,x)_{j_0} | \cdot   ( |-f_c(A,x)| +  |f(A,x)_{j_0}|)\\
    & ~ \cdot \| A_{j_1,*} \|_2 \cdot \| \diag(x) \|_F  \cdot \| h(A,x)\|_2  \cdot \|I_d \|_F \\
    & + ~   |f_c(\wt{A},x)| \cdot  |f(A,x)_{j_1} - f(\wt{A},x)_{j_1}|\cdot | f(A,x)_{j_0} | \cdot  ( |-f_c(A,x)| +  |f(A,x)_{j_0}|) \\
    & ~\cdot \| A_{j_1,*} \|_2 \cdot \| \diag(x) \|_F  \cdot \| h(A,x)\|_2  \cdot \|I_d \|_F  \\
    & + ~   |f_c(\wt{A},x)| \cdot  |f(\wt{A},x)_{j_1}|\cdot | f(A,x)_{j_0} -  f(\wt{A},x)_{j_0} | \cdot ( |-f_c(A,x)| +  |f(A,x)_{j_0}|) \\
    & ~\cdot \| A_{j_1,*} \|_2 \cdot \| \diag(x) \|_F  \cdot  \| h(A,x)\|_2  \cdot \|I_d \|_F  \\
     & + ~   |f_c(\wt{A},x)| \cdot  |f(\wt{A},x)_{j_1}|\cdot | f(\wt{A},x)_{j_0} | \cdot ( | f_c(A,x) -  f_c(\wt{A},x)| +  |f(A,x)_{j_0} - f(\wt{A},x)_{j_0}|)\\
    & ~ \cdot \| A_{j_1,*} \|_2 \cdot \| \diag(x) \|_F  \cdot  \| h(A,x)\|_2  \cdot \|I_d \|_F  \\
       & + ~ |f_c(\wt{A},x)|\cdot   |f(\wt{A},x)_{j_1}|\cdot  | f(\wt{A},x)_{j_0} | \cdot ( |-f_c(\wt{A},x)| +  |f(\wt{A},x)_{j_0}|)\\
    & ~ \cdot \|  A_{j_1,*} - \wt{A}_{j_1,*} \|_2 \cdot \| \diag(x) \|_F  \cdot \| h(A,x) \|_2  \cdot \|I_d \|_F  \\
    & + ~ |f_c(\wt{A},x)|\cdot   |f(\wt{A},x)_{j_1}|\cdot  | f(\wt{A},x)_{j_0} | \cdot ( |-f_c(\wt{A},x)| +  |f(\wt{A},x)_{j_0}|) \\
    & ~\cdot \| \wt{A}_{j_1,*} \|_2 \cdot \| \diag(x) \|_F  \cdot \| h(A,x) - h(\wt{A},x)\|_2  \cdot \|I_d \|_F  \\
    \leq & ~ 18R^4  \beta^{-2} \cdot n \cdot \sqrt{d}  \exp(3R^2)\|A - \wt{A}\|_F \\
    &+ ~ 12R^4  \beta^{-2} \cdot n \cdot \sqrt{d}  \exp(3R^2)\|A - \wt{A}\|_F \\
    & + ~ 12R^4 \beta^{-2} \cdot n \cdot \sqrt{d}\exp(3R^2) \cdot \|A - \wt{A}\|_F \\
    & + ~ 16R^4\beta^{-2} \cdot n \cdot \sqrt{d}\exp(3R^2) \cdot \|A - \wt{A}\|_F \\
    & + ~ 6R^3  \sqrt{d} \cdot \|A - \wt{A}\|_F \\
    & + ~ 18R^2 \beta^{-2} \cdot n \cdot \sqrt{d}\exp(4R^2) \cdot \|A - \wt{A}\|_F\\
    \leq &  ~ 82 \beta^{-2} \cdot n \cdot \sqrt{d}\exp(5R^2) \cdot \|A - \wt{A}\|_F \\
    \leq &  ~ \beta^{-2} \cdot n \cdot \sqrt{d}\exp(6R^2) \cdot \|A - \wt{A}\|_F 
\end{align*}

{\bf Proof of Part 4.}
\begin{align*}
& ~ \| B_{6,1,4}^{j_1,*,j_0,*} (A) - B_{6,1,4}^{j_1,*,j_0,*} ( \wt{A} ) \| \\ 
\leq  & ~ \|- f_c(A,x)  \cdot f(A,x)_{j_1} \cdot f(A,x)_{j_0} \cdot (-f_c(A,x) + f(A,x)_{j_0}) \cdot h(A,x)^\top  \cdot h(A,x) \cdot  I_d \\
    & ~ - (-f_c(\wt{A},x)  \cdot f(\wt{A},x)_{j_1} \cdot f(\wt{A},x)_{j_0} \cdot  (-f_c(\wt{A},x) + f(\wt{A},x)_{j_0}) \cdot h(\wt{A},x)^\top \cdot h(\wt{A},x) \cdot  I_d \|_F)\\
\leq  & ~ \| f_c(A,x)  \cdot f(A,x)_{j_1} \cdot f(A,x)_{j_0} \cdot (-f_c(A,x) + f(A,x)_{j_0}) \cdot h(A,x)^\top \cdot h(A,x) \cdot  I_d \\
    & ~ - f_c(\wt{A},x)  \cdot f(\wt{A},x)_{j_1} \cdot f(\wt{A},x)_{j_0} \cdot  (-f_c(\wt{A},x) + f(\wt{A},x)_{j_0}) \cdot h(\wt{A},x)^\top  \cdot h(\wt{A},x) \cdot I_d \|_F\\
    \leq & ~ |  f_c(A,x) - f_c(\wt{A},x) | \cdot |f(A,x)_{j_1}|\cdot | f(A,x)_{j_0} | \cdot   ( |-f_c(A,x)| +  |f(A,x)_{j_0}|)\\
    & ~ \cdot \| h(A,x)^\top\|_2   \cdot \| h(A,x)\|_2  \cdot \|I_d \|_F  \\
    & + ~   |f_c(\wt{A},x)| \cdot  |f(A,x)_{j_1} - f(\wt{A},x)_{j_1}|\cdot | f(A,x)_{j_0} | \cdot  ( |-f_c(A,x)| +  |f(A,x)_{j_0}|) \\
    & ~\cdot \| h(A,x)^\top\|_2 \cdot \| h(A,x)\|_2  \cdot \|I_d \|_F \\
    & + ~   |f_c(\wt{A},x)| \cdot  |f(\wt{A},x)_{j_1}|\cdot | f(A,x)_{j_0} -  f(\wt{A},x)_{j_0} | \cdot ( |-f_c(A,x)| +  |f(A,x)_{j_0}|) \\
    & ~\cdot \| h(A,x)^\top\|_2  \cdot  \| h(A,x)\|_2  \cdot \|I_d \|_F \\
     & + ~   |f_c(\wt{A},x)| \cdot  |f(\wt{A},x)_{j_1}|\cdot | f(\wt{A},x)_{j_0} | \cdot ( | f_c(A,x) -  f_c(\wt{A},x)| +  |f(A,x)_{j_0} - f(\wt{A},x)_{j_0}|)\\
    & ~ \cdot \| h(A,x)^\top\|_2  \cdot  \| h(A,x)\|_2  \cdot\|I_d \|_F  \\
       & + ~ |f_c(\wt{A},x)|\cdot   |f(\wt{A},x)_{j_1}|\cdot  | f(\wt{A},x)_{j_0} | \cdot ( |-f_c(\wt{A},x)| +  |f(\wt{A},x)_{j_0}|)\\
    & ~ \cdot \| h(A,x)^\top -  h(\wt{A},x)^\top\|_2 \cdot \| h(A,x) \|_2  \cdot \|I_d \|_F  \\
    & + ~ |f_c(\wt{A},x)|\cdot   |f(\wt{A},x)_{j_1}|\cdot  | f(\wt{A},x)_{j_0} | \cdot ( |-f_c(\wt{A},x)| +  |f(\wt{A},x)_{j_0}|) \\
    & ~\cdot \| h(\wt{A},x)^\top\|_2  \cdot \| h(A,x) - h(\wt{A},x)\|_2  \cdot \|I_d \|_F  \\
    \leq & ~ 18R^4  \beta^{-2} \cdot n \cdot \sqrt{d}  \exp(3R^2)\|A - \wt{A}\|_F \\
    &+ ~ 12R^4  \beta^{-2} \cdot n \cdot \sqrt{d}  \exp(3R^2)\|A - \wt{A}\|_F \\
    & + ~ 12R^4 \beta^{-2} \cdot n \cdot \sqrt{d}\exp(3R^2) \cdot \|A - \wt{A}\|_F \\
    & + ~ 16R^4\beta^{-2} \cdot n \cdot \sqrt{d}\exp(3R^2) \cdot \|A - \wt{A}\|_F \\
    & + ~ 18R^2 \beta^{-2} \cdot n \cdot \sqrt{d}\exp(4R^2) \cdot \|A - \wt{A}\|_F\\
    & + ~ 18R^2 \beta^{-2} \cdot n \cdot \sqrt{d}\exp(4R^2) \cdot \|A - \wt{A}\|_F\\
    \leq &  ~ 94 \beta^{-2} \cdot n \cdot \sqrt{d}\exp(5R^2) \cdot \|A - \wt{A}\|_F \\
    \leq &  ~ \beta^{-2} \cdot n \cdot \sqrt{d}\exp(6R^2) \cdot \|A - \wt{A}\|_F 
\end{align*}

{\bf Proof of Part 5.}
\begin{align*}
& ~ \| B_{6,1,5}^{j_1,*,j_0,*} (A) - B_{6,1,5}^{j_1,*,j_0,*} ( \wt{A} ) \| \\ 
\leq  & ~ \| (-f_2(A,x) + f(A,x)_{j_1})  \cdot f(A,x)_{j_1} \cdot f(A,x)_{j_0} \cdot (-f_c(A,x) + f(A,x)_{j_0}) \cdot h(A,x)^\top  \cdot h(A,x) \cdot  I_d \\
    & ~ - (-f_2(\wt{A},x) + f(\wt{A},x)_{j_1})  \cdot f(\wt{A},x)_{j_1} \cdot f(\wt{A},x)_{j_0} \cdot  (-f_c(\wt{A},x) + f(\wt{A},x)_{j_0}) \cdot h(\wt{A},x)^\top \cdot h(\wt{A},x) \cdot  I_d \|_F\\
    \leq & ~ (\|f_2(A,x) - f_2(\wt{A},x)\| + \|f(A,x)_{j_1}- f(\wt{A},x)_{j_1}\|) \cdot |f(A,x)_{j_1}|\cdot | f(A,x)_{j_0} | \cdot   ( |-f_c(A,x)| +  |f(A,x)_{j_0}|)\\
    & ~ \cdot \| h(A,x)^\top\|_2   \cdot \| h(A,x)\|_2  \cdot \|I_d \|_F \\
    & + ~   (\|  f_2(\wt{A},x)\| + \|  f(\wt{A},x)_{j_1}\|) \cdot  |f(A,x)_{j_1} - f(\wt{A},x)_{j_1}|\cdot | f(A,x)_{j_0} | \cdot  ( |-f_c(A,x)| +  |f(A,x)_{j_0}|) \\
    & ~\cdot \| h(A,x)^\top\|_2 \cdot \| h(A,x)\|_2  \cdot \|I_d \|_F \\
    & + ~   (\|  f_2(\wt{A},x)\| + \|  f(\wt{A},x)_{j_1}\|) \cdot  |f(\wt{A},x)_{j_1}|\cdot | f(A,x)_{j_0} -  f(\wt{A},x)_{j_0} | \cdot ( |-f_c(A,x)| +  |f(A,x)_{j_0}|) \\
    & ~\cdot \| h(A,x)^\top\|_2  \cdot  \| h(A,x)\|_2  \cdot \|I_d \|_F \\
     & + ~   (\|  f_2(\wt{A},x)\| + \|  f(\wt{A},x)_{j_1}\|)  \cdot  |f(\wt{A},x)_{j_1}|\cdot | f(\wt{A},x)_{j_0} | \cdot ( | f_c(A,x) -  f_c(\wt{A},x)| +  |f(A,x)_{j_0} - f(\wt{A},x)_{j_0}|)\\
    & ~ \cdot \| h(A,x)^\top\|_2  \cdot  \| h(A,x)\|_2  \cdot \|I_d \|_F \\
       & + ~ (\|  f_2(\wt{A},x)\| + \|  f(\wt{A},x)_{j_1}\|)  \cdot   |f(\wt{A},x)_{j_1}|\cdot  | f(\wt{A},x)_{j_0} | \cdot ( |-f_c(\wt{A},x)| +  |f(\wt{A},x)_{j_0}|)\\
    & ~ \cdot \| h(A,x)^\top -  h(\wt{A},x)^\top\|_2 \cdot \| h(A,x) \|_2  \cdot \|I_d \|_F \\
    & + ~(\|  f_2(\wt{A},x)\| + \|  f(\wt{A},x)_{j_1}\|) \cdot   |f(\wt{A},x)_{j_1}|\cdot  | f(\wt{A},x)_{j_0} | \cdot ( |-f_c(\wt{A},x)| +  |f(\wt{A},x)_{j_0}|) \\
    & ~\cdot \| h(\wt{A},x)^\top\|_2  \cdot \| h(A,x) - h(\wt{A},x)\|_2  \cdot \|I_d \|_F \\
    \leq & ~ 18R^4  \beta^{-2} \cdot n \cdot \sqrt{d}  \exp(3R^2)\|A - \wt{A}\|_F \\
    &+ ~ 12R^4  \beta^{-2} \cdot n \cdot \sqrt{d}  \exp(3R^2)\|A - \wt{A}\|_F \\
    & + ~ 12R^4 \beta^{-2} \cdot n \cdot \sqrt{d}\exp(3R^2) \cdot \|A - \wt{A}\|_F \\
    & + ~ 16R^4\beta^{-2} \cdot n \cdot \sqrt{d}\exp(3R^2) \cdot \|A - \wt{A}\|_F \\
    & + ~ 18R^2 \beta^{-2} \cdot n \cdot \sqrt{d}\exp(4R^2) \cdot \|A - \wt{A}\|_F\\
    & + ~ 18R^2 \beta^{-2} \cdot n \cdot \sqrt{d}\exp(4R^2) \cdot \|A - \wt{A}\|_F\\
    \leq &  ~ 94 \beta^{-2} \cdot n \cdot \sqrt{d}\exp(5R^2) \cdot \|A - \wt{A}\|_F \\
    \leq &  ~ \beta^{-2} \cdot n \cdot \sqrt{d}\exp(6R^2) \cdot \|A - \wt{A}\|_F 
\end{align*}

{\bf Proof of Part 6.}
\begin{align*}
& ~ \| B_{6,1,6}^{j_1,*,j_0,*} (A) - B_{6,1,6}^{j_1,*,j_0,*} ( \wt{A} ) \| \\ 
\leq  & ~ \| (-f_c(A,x) + f(A,x)_{j_1})  \cdot f(A,x)_{j_1} \cdot f(A,x)_{j_0} \cdot (-f_c(A,x) + f(A,x)_{j_0}) \cdot h(A,x)^\top  \cdot h(A,x) \cdot  I_d \\
    & ~ - (-f_c(\wt{A},x) + f(\wt{A},x)_{j_1})  \cdot f(\wt{A},x)_{j_1} \cdot f(\wt{A},x)_{j_0} \cdot  (-f_c(\wt{A},x) + f(\wt{A},x)_{j_0}) \cdot h(\wt{A},x)^\top \cdot h(\wt{A},x) \cdot I_d \|_F\\
    \leq & ~ (\|f_c(A,x) - f_c(\wt{A},x)\| + \|f(A,x)_{j_1}- f(\wt{A},x)_{j_1}\|) \cdot |f(A,x)_{j_1}|\cdot | f(A,x)_{j_0} | \cdot   ( |-f_c(A,x)| +  |f(A,x)_{j_0}|)\\
    & ~ \cdot \| h(A,x)^\top\|_2   \cdot \| h(A,x)\|_2  \cdot \|I_d\|_F \\
    & + ~   (\|  f_c(\wt{A},x)\| + \|  f(\wt{A},x)_{j_1}\|) \cdot  |f(A,x)_{j_1} - f(\wt{A},x)_{j_1}|\cdot | f(A,x)_{j_0} | \cdot  ( |-f_c(A,x)| +  |f(A,x)_{j_0}|) \\
    & ~\cdot \| h(A,x)^\top\|_2 \cdot \| h(A,x)\|_2  \cdot \|I_d\|_F \\
    & + ~   (\|  f_c(\wt{A},x)\| + \|  f(\wt{A},x)_{j_1}\|) \cdot  |f(\wt{A},x)_{j_1}|\cdot | f(A,x)_{j_0} -  f(\wt{A},x)_{j_0} | \cdot ( |-f_c(A,x)| +  |f(A,x)_{j_0}|) \\
    & ~\cdot \| h(A,x)^\top\|_2  \cdot  \| h(A,x)\|_2  \cdot \|I_d\|_F \\
     & + ~   (\|  f_c(\wt{A},x)\| + \|  f(\wt{A},x)_{j_1}\|)  \cdot  |f(\wt{A},x)_{j_1}|\cdot | f(\wt{A},x)_{j_0} | \cdot ( | f_c(A,x) -  f_c(\wt{A},x)| +  |f(A,x)_{j_0} - f(\wt{A},x)_{j_0}|)\\
    & ~ \cdot \| h(A,x)^\top\|_2  \cdot  \| h(A,x)\|_2  \cdot \|I_d\|_F \\
       & + ~ (\|  f_c(\wt{A},x)\| + \|  f(\wt{A},x)_{j_1}\|)  \cdot   |f(\wt{A},x)_{j_1}|\cdot  | f(\wt{A},x)_{j_0} | \cdot ( |-f_c(\wt{A},x)| +  |f(\wt{A},x)_{j_0}|)\\
    & ~ \cdot \| h(A,x)^\top -  h(\wt{A},x)^\top\|_2 \cdot \| h(A,x) \|_2  \cdot \|I_d\|_F \\
    & + ~(\|  f_c(\wt{A},x)\| + \|  f(\wt{A},x)_{j_1}\|) \cdot   |f(\wt{A},x)_{j_1}|\cdot  | f(\wt{A},x)_{j_0} | \cdot ( |-f_c(\wt{A},x)| +  |f(\wt{A},x)_{j_0}|) \\
    & ~\cdot \| h(\wt{A},x)^\top\|_2  \cdot \| h(A,x) - h(\wt{A},x)\|_2  \cdot \|I_d\|_F \\
    \leq & ~ 24R^4  \beta^{-2} \cdot n \cdot \sqrt{d}  \exp(3R^2)\|A - \wt{A}\|_F \\
    &+ ~ 18R^4  \beta^{-2} \cdot n \cdot \sqrt{d}  \exp(3R^2)\|A - \wt{A}\|_F \\
    & + ~ 18R^4 \beta^{-2} \cdot n \cdot \sqrt{d}\exp(3R^2) \cdot \|A - \wt{A}\|_F \\
    & + ~ 24R^4\beta^{-2} \cdot n \cdot \sqrt{d}\exp(3R^2) \cdot \|A - \wt{A}\|_F \\
    & + ~ 27R^2 \beta^{-2} \cdot n \cdot \sqrt{d}\exp(4R^2) \cdot \|A - \wt{A}\|_F\\
    & + ~ 27R^2 \beta^{-2} \cdot n \cdot \sqrt{d}\exp(4R^2) \cdot \|A - \wt{A}\|_F\\
    \leq &  ~ 138 \beta^{-2} \cdot n \cdot \sqrt{d}\exp(5R^2) \cdot \|A - \wt{A}\|_F \\
    \leq &  ~ \beta^{-2} \cdot n \cdot \sqrt{d}\exp(6R^2) \cdot \|A - \wt{A}\|_F 
\end{align*}

{\bf Proof of Part 7.}
\begin{align*}
& ~ \| B_{6,1,7}^{j_1,*,j_0,*} (A) - B_{6,1,7}^{j_1,*,j_0,*} ( \wt{A} ) \| \\ \leq 
    & ~ \|  f(A,x)_{j_1} \cdot  f(A,x)_{j_0} \cdot (-f_c(A,x) + f(A,x)_{j_0})    \cdot p_{j_1}(A,x)^\top \cdot  h(A,x) \cdot I_d \\
    & ~ - f(\wt{A},x)_{j_1} \cdot  f(\wt{A},x)_{j_0} \cdot (-f_c(\wt{A},x) + f(\wt{A},x)_{j_0})    \cdot p_{j_1}(\wt{A},x)^\top \cdot  h(\wt{A},x) \cdot I_d \|_F\\
    \leq & ~  |f(A,x)_{j_1} - f(\wt{A},x)_{j_1}|\cdot | f(A,x)_{j_0}| \cdot ( |-f_c(A,x)| +  |f(A,x)_{j_0}|)  \cdot \| p_{j_1}(A,x)^\top \|_2  \cdot \| h(A,x) \|_2  \cdot \| I_d \|_F\\
    & + ~   |  f(\wt{A},x)_{j_1}|\cdot | f(A,x)_{j_0} -  f(\wt{A},x)_{j_0}| \cdot ( |-f_c(A,x)| +  |f(A,x)_{j_0}|)  \cdot \|  p_{j_1}(A,x)^\top \|_2 \cdot   \| h(A,x) \|_2   \cdot \| I_d \|_F\\
    & + ~    |  f(\wt{A},x)_{j_1}|\cdot |  f(\wt{A},x)_{j_0}| \cdot ( |f_c(A,x) - f_c(\wt{A},x)| +  |f(A,x)_{j_0} - f(\wt{A},x)_{j_0}|)   \cdot \| p_{j_1}(A,x)^\top \|_2 \cdot  \| h(A,x) \|_2  \cdot \| I_d \|_F \\
    & + ~  |  f(\wt{A},x)_{j_1}|\cdot |  f(\wt{A},x)_{j_0}|\cdot ( | f_c(\wt{A},x)| +  | f(\wt{A},x)_{j_0}|)   \cdot \| p_{j_1}(A,x)^\top - p_{j_1}(\wt{A},x)^\top\|_2 \cdot  \| h(A,x) \|_2  \cdot \| I_d \|_F  \\
       & + ~    |  f(\wt{A},x)_{j_1}|\cdot |  f(\wt{A},x)_{j_0}|\cdot ( | f_c(\wt{A},x)| +  | f(\wt{A},x)_{j_0}|)   \cdot  \|  p_{j_1}(\wt{A},x)^\top \|_2  \cdot   \| h(A,x)- h(\wt{A},x) \|_2   \cdot \| I_d \|_F \\
    \leq & ~ 36R^4  \beta^{-2} \cdot n \cdot \sqrt{d}  \exp(3R^2)\|A - \wt{A}\|_F \\
    &+ ~ 36R^4  \beta^{-2} \cdot n \cdot \sqrt{d}  \exp(3R^2)\|A - \wt{A}\|_F \\
    & + ~ 48R^4\beta^{-2} \cdot n \cdot \sqrt{d}\exp(3R^2) \cdot \|A - \wt{A}\|_F \\
    & + ~ 39R^2\beta^{-2} \cdot n \cdot \sqrt{d}\exp(4R^2) \cdot \|A - \wt{A}\|_F \\
    & + ~ 54R^2\beta^{-2} \cdot n \cdot \sqrt{d}\exp(4R^2) \cdot \|A - \wt{A}\|_F \\
    \leq &  ~ 213\beta^{-2} \cdot n \cdot \sqrt{d}\exp(5R^2) \cdot \|A - \wt{A}\|_F \\
\leq &  ~ \beta^{-2} \cdot n \cdot \sqrt{d}\exp(6R^2) \cdot \|A - \wt{A}\|_F
\end{align*}

{\bf Proof of Part 8.}
\begin{align*}
& ~ \| B_{6,2,1}^{j_1,*,j_0,*} (A) - B_{6,2,1}^{j_1,*,j_0,*} ( \wt{A} ) \| \\ \leq 
    & ~ \| f(A,x)_{j_1} \cdot  f(A,x)_{j_0} \cdot (-f_c(A,x) + f(A,x)_{j_0})\cdot c_g(A,x)^{\top} \cdot h(A,x) \cdot x \cdot {\bf 1}_d^{\top}\\
    & ~ - f(\wt{A},x)_{j_1} \cdot  f(\wt{A},x)_{j_0} \cdot (-f_c(\wt{A},x) + f(\wt{A},x)_{j_0}) \cdot c_g(\wt{A},x)^{\top} \cdot  h(\wt{A},x) \cdot x \cdot {\bf 1}_d^{\top} \|_F \\
    \leq & ~  |f(A,x)_{j_1} - f(\wt{A},x)_{j_1}|\cdot | f(A,x)_{j_0}| \cdot ( |- f_c(A,x)| +  | f(A,x)_{j_0}|)  \cdot \| c_g(A,x)^\top \|_2  \cdot  \|h(A,x)\|_2 \cdot \| x \|_2 \cdot \| {\bf 1}_d^{\top} \|_F\\
    & + ~   |  f(\wt{A},x)_{j_1}|\cdot | f(A,x)_{j_0} -  f(\wt{A},x)_{j_0}| \cdot  ( |- f_c(A,x)| +  | f(A,x)_{j_0}|)  \cdot \|   c_g(A,x)^\top \|_2 \cdot  \|h(A,x)\|_2   \cdot \| x \|_2 \cdot \| {\bf 1}_d^{\top} \|_F\\
    & + ~    |  f(\wt{A},x)_{j_1}|\cdot |  f(\wt{A},x)_{j_0}| \cdot ( |f_c(A,x) - f_c(\wt{A},x)| +  |f(A,x)_{j_0} - f(\wt{A},x)_{j_0}|)  \cdot \| c_g(A,x)^\top \|_2 \\ 
     & ~\cdot  \|h(A,x)\|_2  \cdot \| x \|_2 \cdot \| {\bf 1}_d^{\top} \|_F\\
    & + ~  |  f(\wt{A},x)_{j_1}|\cdot |  f(\wt{A},x)_{j_0}|\cdot ( | f_c(\wt{A},x)| +  | f(\wt{A},x)_{j_0}|)    \cdot \| c_g(A,x)^\top - c_g(\wt{A},x)^\top \|_2 \cdot  \|h(A,x)\|_2  \cdot \| x \|_2 \cdot \| {\bf 1}_d^{\top} \|_F\\
    & + ~    |  f(\wt{A},x)_{j_1}|\cdot |  f(\wt{A},x)_{j_0}|\cdot ( | f_c(\wt{A},x)| +  | f(\wt{A},x)_{j_0}|)   \cdot  \|  c_g(\wt{A},x)^\top  \|_2  \cdot   \|h(A,x) - h(\wt{A},x)\|_2  \cdot \| x \|_2 \cdot \| {\bf 1}_d^{\top} \|_F\\
    \leq & ~ 30R^4 \beta^{-2} \cdot n \cdot \sqrt{d}  \exp(3R^2)\|A - \wt{A}\|_F \\
    &+ ~ 30R^4  \beta^{-2} \cdot n \cdot \sqrt{d}  \exp(3R^2)\|A - \wt{A}\|_F \\
    & + ~ 40R^4\beta^{-2} \cdot n \cdot \sqrt{d}\exp(3R^2) \cdot \|A - \wt{A}\|_F \\
    & + ~ 60R^4\beta^{-2} \cdot n \cdot \sqrt{d}\exp(3R^2) \cdot \|A - \wt{A}\|_F \\
     & + ~ 45R^2\beta^{-2} \cdot n \cdot \sqrt{d}\exp(4R^2) \cdot \|A - \wt{A}\|_F\\
    \leq &  ~ 205\beta^{-2} \cdot n \cdot \sqrt{d}\exp(5R^2) \cdot \|A - \wt{A}\|_F \\
    \leq &  ~ \beta^{-2} \cdot n \cdot \sqrt{d}\exp(6R^2) \cdot \|A - \wt{A}\|_F
\end{align*}

{\bf Proof of Part 9.}
\begin{align*}
  & ~ \|  B_{6,3,1}^{j_1,*,j_0,*} (A) - B_{6,3,1}^{j_1,*,j_0,*} ( \wt{A} ) \|_F \\
   = & ~ \| -  f(A,x)_{j_1} \cdot   f(A,x)_{j_0}  \cdot   (-f_c(A,x) + f(A,x)_{j_0})   \cdot {\bf 1}_d \cdot c_g(A,x)^{\top} \cdot \diag (x)\\
    & ~ -(-  f(\wt{A},x)_{j_1} \cdot   f(\wt{A},x)_{j_0}  \cdot  (-f_c(\wt{A},x) + f(\wt{A},x)_{j_0})   \cdot {\bf 1}_d \cdot c_g(\wt{A},x)^{\top} \cdot \diag (x))\|_F \\
    \leq & ~ \|    f(A,x)_{j_1} \cdot   f(A,x)_{j_0}  \cdot    (-f_c(A,x) + f(A,x)_{j_0})   \cdot {\bf 1}_d \cdot c_g(A,x)^{\top} \cdot \diag (x)\\
    & ~ -   f(\wt{A},x)_{j_1} \cdot   f(\wt{A},x)_{j_0}  \cdot   (-f_c(\wt{A},x) + f(\wt{A},x)_{j_0})  \cdot {\bf 1}_d \cdot c_g(\wt{A},x)^{\top} \cdot \diag (x) \|_F \\
    \leq & ~  |f(A,x)_{j_1} - f(\wt{A},x)_{j_1}|\cdot | f(A,x)_{j_0}| \cdot ( |- f_c(A,x)| +  | f(A,x)_{j_0}|)  \cdot \| c_g(A,x)^\top \|_2  \cdot  \|  {\bf 1}_d\|_2 \cdot \| \diag(x)\|_F  \\
    & + ~   |  f(\wt{A},x)_{j_1}|\cdot | f(A,x)_{j_0} -  f(\wt{A},x)_{j_0}| \cdot ( |- f_c(A,x)| +  | f(A,x)_{j_0}|)  \cdot \|   c_g(A,x)^\top \|_2 \cdot \|  {\bf 1}_d\|_2 \cdot \| \diag(x)\|_F\\
    & + ~    |  f(\wt{A},x)_{j_1}|\cdot |  f(\wt{A},x)_{j_0}| \cdot ( |f_c(A,x) - f_c(\wt{A},x)| +  |f(A,x)_{j_0} - f(\wt{A},x)_{j_0}|)   \cdot \| c_g(A,x)^\top \|_2 \cdot \|  {\bf 1}_d\|_2 \cdot \| \diag(x)\|_F\\
    & + ~  |  f(\wt{A},x)_{j_1}|\cdot |  f(\wt{A},x)_{j_0}|\cdot ( | f_c(\wt{A},x)| +  | f(\wt{A},x)_{j_0}|)    \cdot \| c_g(A,x)^\top - c_g(\wt{A},x)^\top \|_2 \cdot  \|  {\bf 1}_d\|_2 \cdot \| \diag(x)\|_F\\
    \leq & ~ 30R^2 \beta^{-2} \cdot n \cdot \sqrt{d}  \exp(3R^2)\|A - \wt{A}\|_F \\
    &+ ~ 30R^2  \beta^{-2} \cdot n \cdot \sqrt{d}  \exp(3R^2)\|A - \wt{A}\|_F \\
    & + ~ 40R^2\beta^{-2} \cdot n \cdot \sqrt{d}\exp(3R^2) \cdot \|A - \wt{A}\|_F \\
    & + ~ 60R^2\beta^{-2} \cdot n \cdot \sqrt{d}\exp(3R^2) \cdot \|A - \wt{A}\|_F \\
    \leq &  ~ 160\beta^{-2} \cdot n \cdot \sqrt{d}\exp(4R^2) \cdot \|A - \wt{A}\|_F \\
    \leq &  ~ \beta^{-2} \cdot n \cdot \sqrt{d}\exp(5R^2) \cdot \|A - \wt{A}\|_F
\end{align*}

{\bf Proof of Part 10.}
\begin{align*}
  & ~ \|  B_{6,4,1}^{j_1,*,j_0,*} (A) - B_{6,4,1}^{j_1,*,j_0,*} ( \wt{A} ) \|_F \\
    = & ~ \| -f(A,x)_{j_1} \cdot f(A,x)_{j_0}  \cdot (-f_c(A,x) + f(A,x)_{j_0}) \cdot  c_g(A,x)^{\top} \cdot  h_e(A,x) \cdot x \cdot {\bf 1}_d^{\top} \\
    & ~ - (-f(\wt{A},x)_{j_1} \cdot f(\wt{A},x)_{j_0}  \cdot (-f_c(\wt{A},x) + f(\wt{A},x)_{j_0})  \cdot  c_g(\wt{A},x)^{\top} \cdot  h_e(\wt{A},x) \cdot x \cdot {\bf 1}_d^{\top}) \|_F \\
    \leq & ~ \| f(A,x)_{j_1} \cdot f(A,x)_{j_0}  \cdot (-f_c(A,x) + f(A,x)_{j_0}) \cdot  c_g(A,x)^{\top} \cdot  h_e(A,x) \cdot x \cdot {\bf 1}_d^{\top} \\
    & ~ - f(\wt{A},x)_{j_1} \cdot f(\wt{A},x)_{j_0}  \cdot (-f_c(\wt{A},x) + f(\wt{A},x)_{j_0})  \cdot  c_g(\wt{A},x)^{\top} \cdot  h_e(\wt{A},x) \cdot x \cdot {\bf 1}_d^{\top} \|_F \\
    \leq & ~ |f(A,x)_{j_1} - f(\wt{A},x)_{j_1}| \cdot |f(A,x)_{j_0} | \cdot ( |- f_c(A,x)| +  | f(A,x)_{j_0}|) \cdot  \| c_g(A,x)^{\top} \|_2 \cdot  \| h_e(A,x) \|_2 \cdot \| x \|_2 \cdot \|{\bf 1}_d^{\top} \|_2 \\
    & + ~ |f(\wt{A},x)_{j_1}| \cdot |f(A,x)_{j_0} - f(\wt{A},x)_{j_0} | \cdot ( |- f_c(A,x)| +  | f(A,x)_{j_0}|) \cdot  \| c_g(A,x)^{\top} \|_2 \cdot   \| h_e(A,x) \|_2 \cdot \| x \|_2 \cdot \|{\bf 1}_d^{\top} \|_2 \\
    & + ~ |f(\wt{A},x)_{j_1}| \cdot |f(\wt{A},x)_{j_0} | \cdot ( |f_c(A,x) - f_c(\wt{A},x)| +  |f(A,x)_{j_0} - f(\wt{A},x)_{j_0}|) \cdot  \| c_g(A,x)^{\top} \|_2 \\ 
     & ~\cdot   \| h_e(A,x) \|_2 \cdot \| x \|_2 \cdot \|{\bf 1}_d^{\top} \|_2 \\
    & + ~ |f(\wt{A},x)_{j_1}| \cdot |f(\wt{A},x)_{j_0} | \cdot  ( | f_c(\wt{A},x)| +  | f(\wt{A},x)_{j_0}|) \cdot  \| c_g(A,x)^{\top}  - c_g(\wt{A},x)^{\top} \|_2 \cdot  \| h_e(A,x) \|_2 \cdot \| x \|_2 \cdot \|{\bf 1}_d^{\top} \|_2 \\
    & + ~ |f(\wt{A},x)_{j_1}| \cdot |f(\wt{A},x)_{j_0} | \cdot ( | f_c(\wt{A},x)| +  | f(\wt{A},x)_{j_0}|)  \cdot  \| c_g(\wt{A},x)^{\top} \|_2 \cdot   \| h_e(A,x) - h_e(\wt{A},x) \|_2 \cdot \| x \|_2 \cdot \|{\bf 1}_d^{\top} \|_2 \\
    \leq & ~ 60R^4 \beta^{-2} \cdot n \cdot \sqrt{d}  \exp(3R^2)\|A - \wt{A}\|_F \\
    & + ~ 60R^4 \beta^{-2} \cdot n \cdot \sqrt{d}  \exp(3R^2)\|A - \wt{A}\|_F \\
    & + ~ 80R^4 \beta^{-2} \cdot n \cdot \sqrt{d}  \exp(3R^2)\|A - \wt{A}\|_F \\
    & + ~ 120R^4 \beta^{-2} \cdot n \cdot \sqrt{d}  \exp(3R^2)\|A - \wt{A}\|_F \\
    & + ~ 45R^2  \beta^{-2} \cdot n \cdot \sqrt{d}  \exp(4R^2)\|A - \wt{A}\|_F \\
    \leq & ~ 365\beta^{-2} \cdot n \cdot \sqrt{d}  \exp(5R^2)\|A - \wt{A}\|_F \\
    \leq & ~ \beta^{-2} \cdot n \cdot \sqrt{d}  \exp(6R^2)\|A - \wt{A}\|_F 
\end{align*}

{\bf Proof of Part 11.}
\begin{align*}
  & ~ \|  B_{6,5,1}^{j_1,*,j_0,*} (A) - B_{6,5,1}^{j_1,*,j_0,*} ( \wt{A} ) \|_F \\
    =   & ~ \|- f(A,x)_{j_1} \cdot f(A,x)_{j_0}  \cdot f_2( A,x) \cdot  c_g(A,x)^{\top} \cdot  h(A,x) \cdot x \cdot {\bf 1}_d^{\top} \\
    & ~ -( -f(\wt{A},x)_{j_1} \cdot f(\wt{A},x)_{j_0}  \cdot f_2( \wt{A},x) \cdot  c_g(\wt{A},x)^{\top} \cdot  h(\wt{A},x) \cdot x \cdot {\bf 1}_d^{\top}) \|_F \\
    \leq    & ~ \| f(A,x)_{j_1} \cdot f(A,x)_{j_0}  \cdot f_2( A,x) \cdot  c_g(A,x)^{\top} \cdot  h(A,x) \cdot x \cdot {\bf 1}_d^{\top} \\
    & ~ - f(\wt{A},x)_{j_1} \cdot f(\wt{A},x)_{j_0}  \cdot f_2( \wt{A},x) \cdot  c_g(\wt{A},x)^{\top} \cdot  h(\wt{A},x) \cdot x \cdot {\bf 1}_d^{\top} \|_F \\
    \leq & ~ |f(A,x)_{j_1} - f(\wt{A},x)_{j_1}| \cdot |f(A,x)_{j_0} | \cdot |f_2( A,x) |\cdot  \| c_g(A,x)^{\top} \|_2 \cdot  \| h(A,x) \|_2 \cdot \| x \|_2 \cdot \|{\bf 1}_d^{\top} \|_2 \\
    & + ~ |f(\wt{A},x)_{j_1}| \cdot |f(A,x)_{j_0} - f(\wt{A},x)_{j_0} | \cdot |f_2( A,x) |\cdot  \| c_g(A,x)^{\top} \|_2 \cdot   \| h(A,x) \|_2 \cdot \| x \|_2 \cdot \|{\bf 1}_d^{\top} \|_2 \\
    & + ~ |f(\wt{A},x)_{j_1}| \cdot |f(\wt{A},x)_{j_0} | \cdot |f_2( A,x)  -f_2( \wt{A},x)|\cdot  \| c_g(A,x)^{\top} \|_2 \cdot   \| h(A,x) \|_2 \cdot \| x \|_2 \cdot \|{\bf 1}_d^{\top} \|_2 \\
    & + ~ |f(\wt{A},x)_{j_1}| \cdot |f(\wt{A},x)_{j_0} | \cdot |f_2( \wt{A},x)|\cdot  \| c_g(A,x)^{\top}  - c_g(\wt{A},x)^{\top} \|_2 \cdot  \| h(A,x) \|_2 \cdot \| x \|_2 \cdot \|{\bf 1}_d^{\top} \|_2 \\
    & + ~ |f(\wt{A},x)_{j_1}| \cdot |f(\wt{A},x)_{j_0} | \cdot |f_2( \wt{A},x)|\cdot  \| c_g(\wt{A},x)^{\top} \|_2 \cdot   \| h(A,x) - h(\wt{A},x) \|_2 \cdot \| x \|_2 \cdot \|{\bf 1}_d^{\top} \|_2 \\
    \leq & ~ 10R^4 \beta^{-2} \cdot n \cdot \sqrt{d}  \exp(3R^2)\|A - \wt{A}\|_F \\
    & + ~ 10R^4 \beta^{-2} \cdot n \cdot \sqrt{d}  \exp(3R^2)\|A - \wt{A}\|_F \\
    & + ~ 20R^4 \beta^{-2} \cdot n \cdot \sqrt{d}  \exp(3R^2)\|A - \wt{A}\|_F \\
    & + ~ 20R^4 \beta^{-2} \cdot n \cdot \sqrt{d}  \exp(3R^2)\|A - \wt{A}\|_F \\
    & + ~ 15R^2  \beta^{-2} \cdot n \cdot \sqrt{d}  \exp(4R^2)\|A - \wt{A}\|_F \\
    \leq & ~ 75\beta^{-2} \cdot n \cdot \sqrt{d}  \exp(5R^2)\|A - \wt{A}\|_F \\
    \leq & ~ \beta^{-2} \cdot n \cdot \sqrt{d}  \exp(6R^2)\|A - \wt{A}\|_F 
\end{align*}

{\bf Proof of Part 12.}
\begin{align*}
  & ~ \|  B_{6,5,2}^{j_1,*,j_0,*} (A) - B_{6,5,2}^{j_1,*,j_0,*} ( \wt{A} ) \|_F \\
    =   & ~ \| f(A,x)_{j_1}^2 \cdot f(A,x)_{j_0}    \cdot  c_g(A,x)^{\top} \cdot  h(A,x) \cdot x \cdot {\bf 1}_d^{\top} \\
    & ~ - f(\wt{A},x)_{j_1}^2 \cdot f(\wt{A},x)_{j_0}  \cdot   c_g(\wt{A},x)^{\top} \cdot  h(\wt{A},x) \cdot x \cdot {\bf 1}_d^{\top} \|_F \\
    \leq & ~ |f(A,x)_{j_1} - f(\wt{A},x)_{j_1}| \cdot |f(A,x)_{j_1} | \cdot |f(A,x)_{j_0} |  \cdot  \| c_g(A,x)^{\top} \|_2 \cdot  \| h(A,x) \|_2 \cdot \| x \|_2 \cdot \|{\bf 1}_d^{\top} \|_2 \\
     & + ~ |f(\wt{A},x)_{j_1}| \cdot |f(A,x)_{j_1} - f(\wt{A},x)_{j_1}|\cdot |f(A,x)_{j_0} |\cdot  \| c_g(A,x)^{\top} \|_2 \cdot   \| h(A,x) \|_2 \cdot \| x \|_2 \cdot \|{\bf 1}_d^{\top} \|_2 \\
    & + ~ |f(\wt{A},x)_{j_1}| \cdot |f(\wt{A},x)_{j_1}| \cdot |f(A,x)_{j_0} - f(\wt{A},x)_{j_0} |  \cdot  \| c_g(A,x)^{\top} \|_2 \cdot   \| h(A,x) \|_2 \cdot \| x \|_2 \cdot \|{\bf 1}_d^{\top} \|_2 \\
    & + ~ |f(\wt{A},x)_{j_1}| \cdot |f(\wt{A},x)_{j_1}| \cdot |f(\wt{A},x)_{j_0} | \cdot   \| c_g(A,x)^{\top}  - c_g(\wt{A},x)^{\top} \|_2 \cdot  \| h(A,x) \|_2 \cdot \| x \|_2 \cdot \|{\bf 1}_d^{\top} \|_2 \\
    & + ~ |f(\wt{A},x)_{j_1}| \cdot |f(\wt{A},x)_{j_1}| \cdot |f(\wt{A},x)_{j_0} | \cdot    \| c_g(\wt{A},x)^{\top} \|_2 \cdot   \| h(A,x) - h(\wt{A},x) \|_2 \cdot \| x \|_2 \cdot \|{\bf 1}_d^{\top} \|_2 \\
    \leq & ~ 10R^4 \beta^{-2} \cdot n \cdot \sqrt{d}  \exp(3R^2)\|A - \wt{A}\|_F \\
    & + ~ 10R^4 \beta^{-2} \cdot n \cdot \sqrt{d}  \exp(3R^2)\|A - \wt{A}\|_F \\
    & + ~ 10R^4 \beta^{-2} \cdot n \cdot \sqrt{d}  \exp(3R^2)\|A - \wt{A}\|_F \\
    & + ~ 20R^4 \beta^{-2} \cdot n \cdot \sqrt{d}  \exp(3R^2)\|A - \wt{A}\|_F \\
    & + ~ 15R^2  \beta^{-2} \cdot n \cdot \sqrt{d}  \exp(4R^2)\|A - \wt{A}\|_F \\
    \leq & ~ 65\beta^{-2} \cdot n \cdot \sqrt{d}  \exp(5R^2)\|A - \wt{A}\|_F \\
    \leq & ~ \beta^{-2} \cdot n \cdot \sqrt{d}  \exp(6R^2)\|A - \wt{A}\|_F 
\end{align*}

{\bf Proof of Part 13.}
\begin{align*}
  & ~ \|  B_{6,6,1}^{j_1,*,j_0,*} (A) - B_{6,6,1}^{j_1,*,j_0,*} ( \wt{A} ) \|_F \\
    =   & ~ \|- f(A,x)_{j_1} \cdot f(A,x)_{j_0}  \cdot f_c( A,x) \cdot  c_g(A,x)^{\top} \cdot  h(A,x) \cdot x \cdot {\bf 1}_d^{\top} \\
    & ~ -( -f(\wt{A},x)_{j_1} \cdot f(\wt{A},x)_{j_0}  \cdot f_c( \wt{A},x) \cdot  c_g(\wt{A},x)^{\top} \cdot  h(\wt{A},x) \cdot x \cdot {\bf 1}_d^{\top}) \|_F \\
    \leq    & ~ \| f(A,x)_{j_1} \cdot f(A,x)_{j_0}  \cdot f_c( A,x) \cdot  c_g(A,x)^{\top} \cdot  h(A,x) \cdot x \cdot {\bf 1}_d^{\top} \\
    & ~ - f(\wt{A},x)_{j_1} \cdot f(\wt{A},x)_{j_0}  \cdot f_c( \wt{A},x) \cdot  c_g(\wt{A},x)^{\top} \cdot  h(\wt{A},x) \cdot x \cdot {\bf 1}_d^{\top} \|_F \\
    \leq & ~ |f(A,x)_{j_1} - f(\wt{A},x)_{j_1}| \cdot |f(A,x)_{j_0} | \cdot |f_c( A,x) |\cdot  \| c_g(A,x)^{\top} \|_2 \cdot  \| h(A,x) \|_2 \cdot \| x \|_2 \cdot \|{\bf 1}_d^{\top} \|_2 \\
    & + ~ |f(\wt{A},x)_{j_1}| \cdot |f(A,x)_{j_0} - f(\wt{A},x)_{j_0} | \cdot |f_c( A,x) |\cdot  \| c_g(A,x)^{\top} \|_2 \cdot   \| h(A,x) \|_2 \cdot \| x \|_2 \cdot \|{\bf 1}_d^{\top} \|_2 \\
    & + ~ |f(\wt{A},x)_{j_1}| \cdot |f(\wt{A},x)_{j_0} | \cdot |f_c( A,x)  -f_c( \wt{A},x)|\cdot  \| c_g(A,x)^{\top} \|_2 \cdot   \| h(A,x) \|_2 \cdot \| x \|_2 \cdot \|{\bf 1}_d^{\top} \|_2 \\
    & + ~ |f(\wt{A},x)_{j_1}| \cdot |f(\wt{A},x)_{j_0} | \cdot |f_c( \wt{A},x)|\cdot  \| c_g(A,x)^{\top}  - c_g(\wt{A},x)^{\top} \|_2 \cdot  \| h(A,x) \|_2 \cdot \| x \|_2 \cdot \|{\bf 1}_d^{\top} \|_2 \\
    & + ~ |f(\wt{A},x)_{j_1}| \cdot |f(\wt{A},x)_{j_0} | \cdot |f_c( \wt{A},x)|\cdot  \| c_g(\wt{A},x)^{\top} \|_2 \cdot   \| h(A,x) - h(\wt{A},x) \|_2 \cdot \| x \|_2 \cdot \|{\bf 1}_d^{\top} \|_2 \\
    \leq & ~ 20R^4 \beta^{-2} \cdot n \cdot \sqrt{d}  \exp(3R^2)\|A - \wt{A}\|_F \\
    & + ~ 20R^4 \beta^{-2} \cdot n \cdot \sqrt{d}  \exp(3R^2)\|A - \wt{A}\|_F \\
    & + ~ 30R^4 \beta^{-2} \cdot n \cdot \sqrt{d}  \exp(3R^2)\|A - \wt{A}\|_F \\
    & + ~ 40R^4 \beta^{-2} \cdot n \cdot \sqrt{d}  \exp(3R^2)\|A - \wt{A}\|_F \\
    & + ~ 30R^2  \beta^{-2} \cdot n \cdot \sqrt{d}  \exp(4R^2)\|A - \wt{A}\|_F \\
    \leq & ~ 140\beta^{-2} \cdot n \cdot \sqrt{d}  \exp(5R^2)\|A - \wt{A}\|_F \\
    \leq & ~ \beta^{-2} \cdot n \cdot \sqrt{d}  \exp(6R^2)\|A - \wt{A}\|_F 
\end{align*}

{\bf Proof of Part 14.}
\begin{align*}
  & ~ \|  B_{6,6,2}^{j_1,*,j_0,*} (A) - B_{6,6,2}^{j_1,*,j_0,*} ( \wt{A} ) \|_F \\
    =   & ~ \| f(A,x)_{j_1} \cdot c(A,x)_{j_1} \cdot f(A,x)_{j_0}    \cdot  c_g(A,x)^{\top} \cdot  h(A,x) \cdot x \cdot {\bf 1}_d^{\top} \\
    & ~ - f(\wt{A},x)_{j_1}\cdot c(\wt{A},x)_{j_1} \cdot f(\wt{A},x)_{j_0}  \cdot   c_g(\wt{A},x)^{\top} \cdot  h(\wt{A},x) \cdot x \cdot {\bf 1}_d^{\top} \|_F \\
    \leq & ~ |f(A,x)_{j_1} - f(\wt{A},x)_{j_1}| \cdot | c(A,x)_{j_1} | \cdot |f(A,x)_{j_0} |  \cdot  \| c_g(A,x)^{\top} \|_2 \cdot  \| h(A,x) \|_2 \cdot \| x \|_2 \cdot \|{\bf 1}_d^{\top} \|_2 \\
     & + ~ |f(\wt{A},x)_{j_1}| \cdot |c(A,x)_{j_1} - c(\wt{A},x)_{j_1}|\cdot |f(A,x)_{j_0} |\cdot  \| c_g(A,x)^{\top} \|_2 \cdot   \| h(A,x) \|_2 \cdot \| x \|_2 \cdot \|{\bf 1}_d^{\top} \|_2 \\
    & + ~ |f(\wt{A},x)_{j_1}| \cdot |c(\wt{A},x)_{j_1}| \cdot |f(A,x)_{j_0} - f(\wt{A},x)_{j_0} |  \cdot  \| c_g(A,x)^{\top} \|_2 \cdot   \| h(A,x) \|_2 \cdot \| x \|_2 \cdot \|{\bf 1}_d^{\top} \|_2 \\
    & + ~ |f(\wt{A},x)_{j_1}| \cdot |c(\wt{A},x)_{j_1}| \cdot |f(\wt{A},x)_{j_0} | \cdot   \| c_g(A,x)^{\top}  - c_g(\wt{A},x)^{\top} \|_2 \cdot  \| h(A,x) \|_2 \cdot \| x \|_2 \cdot \|{\bf 1}_d^{\top} \|_2 \\
    & + ~ |f(\wt{A},x)_{j_1}| \cdot |c(\wt{A},x)_{j_1}| \cdot |f(\wt{A},x)_{j_0} | \cdot    \| c_g(\wt{A},x)^{\top} \|_2 \cdot   \| h(A,x) - h(\wt{A},x) \|_2 \cdot \| x \|_2 \cdot \|{\bf 1}_d^{\top} \|_2 \\
    \leq & ~ 20R^4 \beta^{-2} \cdot n \cdot \sqrt{d}  \exp(3R^2)\|A - \wt{A}\|_F \\
    & + ~ 10R^4 \beta^{-2} \cdot n \cdot \sqrt{d}  \exp(3R^2)\|A - \wt{A}\|_F \\
    & + ~ 20R^4 \beta^{-2} \cdot n \cdot \sqrt{d}  \exp(3R^2)\|A - \wt{A}\|_F \\
    & + ~ 40R^4 \beta^{-2} \cdot n \cdot \sqrt{d}  \exp(3R^2)\|A - \wt{A}\|_F \\
    & + ~ 30R^2  \beta^{-2} \cdot n \cdot \sqrt{d}  \exp(4R^2)\|A - \wt{A}\|_F \\
    \leq & ~ 120\beta^{-2} \cdot n \cdot \sqrt{d}  \exp(5R^2)\|A - \wt{A}\|_F \\
    \leq & ~ \beta^{-2} \cdot n \cdot \sqrt{d}  \exp(6R^2)\|A - \wt{A}\|_F 
\end{align*}

{\bf Proof of Part 15.}
\begin{align*}
  & ~ \|  B_{6,7,1}^{j_1,*,j_0,*} (A) - B_{6,7,1}^{j_1,*,j_0,*} ( \wt{A} ) \|_F \\
    =   & ~ \| f(A,x)_{j_0}^2 \cdot f(A,x)_{j_1}    \cdot  c_g(A,x)^{\top} \cdot  h(A,x) \cdot x \cdot {\bf 1}_d^{\top} \\
    & ~ - f(\wt{A},x)_{j_0}^2 \cdot f(\wt{A},x)_{j_1}  \cdot   c_g(\wt{A},x)^{\top} \cdot  h(\wt{A},x) \cdot x \cdot {\bf 1}_d^{\top} \|_F \\
    \leq & ~ |f(A,x)_{j_0} - f(\wt{A},x)_{j_0}| \cdot |f(A,x)_{j_0} | \cdot |f(A,x)_{j_1} |  \cdot  \| c_g(A,x)^{\top} \|_2 \cdot  \| h(A,x) \|_2 \cdot \| x \|_2 \cdot \|{\bf 1}_d^{\top} \|_2 \\
     & + ~ |f(\wt{A},x)_{j_0}| \cdot |f(A,x)_{j_0} - f(\wt{A},x)_{j_0}|\cdot |f(A,x)_{j_1} |\cdot  \| c_g(A,x)^{\top} \|_2 \cdot   \| h(A,x) \|_2 \cdot \| x \|_2 \cdot \|{\bf 1}_d^{\top} \|_2 \\
    & + ~ |f(\wt{A},x)_{j_0}| \cdot |f(\wt{A},x)_{j_0}| \cdot |f(A,x)_{j_1} - f(\wt{A},x)_{j_1} |  \cdot  \| c_g(A,x)^{\top} \|_2 \cdot   \| h(A,x) \|_2 \cdot \| x \|_2 \cdot \|{\bf 1}_d^{\top} \|_2 \\
    & + ~ |f(\wt{A},x)_{j_0}| \cdot |f(\wt{A},x)_{j_0}| \cdot |f(\wt{A},x)_{j_1} | \cdot   \| c_g(A,x)^{\top}  - c_g(\wt{A},x)^{\top} \|_2 \cdot  \| h(A,x) \|_2 \cdot \| x \|_2 \cdot \|{\bf 1}_d^{\top} \|_2 \\
    & + ~ |f(\wt{A},x)_{j_0}| \cdot |f(\wt{A},x)_{j_0}| \cdot |f(\wt{A},x)_{j_1} | \cdot    \| c_g(\wt{A},x)^{\top} \|_2 \cdot   \| h(A,x) - h(\wt{A},x) \|_2 \cdot \| x \|_2 \cdot \|{\bf 1}_d^{\top} \|_2 \\
    \leq & ~ 10R^4 \beta^{-2} \cdot n \cdot \sqrt{d}  \exp(3R^2)\|A - \wt{A}\|_F \\
    & + ~ 10R^4 \beta^{-2} \cdot n \cdot \sqrt{d}  \exp(3R^2)\|A - \wt{A}\|_F \\
    & + ~ 10R^4 \beta^{-2} \cdot n \cdot \sqrt{d}  \exp(3R^2)\|A - \wt{A}\|_F \\
    & + ~ 20R^4 \beta^{-2} \cdot n \cdot \sqrt{d}  \exp(3R^2)\|A - \wt{A}\|_F \\
    & + ~ 15R^2  \beta^{-2} \cdot n \cdot \sqrt{d}  \exp(4R^2)\|A - \wt{A}\|_F \\
    \leq & ~ 65\beta^{-2} \cdot n \cdot \sqrt{d}  \exp(5R^2)\|A - \wt{A}\|_F \\
    \leq & ~ \beta^{-2} \cdot n \cdot \sqrt{d}  \exp(6R^2)\|A - \wt{A}\|_F 
\end{align*}

{\bf Proof of Part 16.}
\begin{align*}
 & ~ \| B_{6}^{j_1,*,j_0,*} (A) - B_{6}^{j_1,*,j_0,*} ( \wt{A} ) \|_F \\
   = & ~ \|\sum_{i = 1}^7 B_{4,i}^{j_1,*,j_0,*}(A) -   B_{4,i}^{j_1,*,j_0,*}(\wt{A})  \|_F \\
    \leq & ~ 15  \beta^{-2} \cdot n \cdot \sqrt{d} \cdot \exp(6R^2)\|A - \wt{A}\|_F
\end{align*}
\end{proof}
\subsection{PSD}
\begin{lemma}\label{psd: B_6}
If the following conditions hold
\begin{itemize}
    \item Let $B_{6,1,1}^{j_1,*, j_0,*}, \cdots, B_{5,7,1}^{j_1,*, j_0,*} $ be defined as Lemma~\ref{lem:b_6_j1_j0} 
    \item  Let $\|A \|_2 \leq R, \|A^{\top} \|_F \leq R, \| x\|_2 \leq R, \|\diag(f(A,x)) \|_F \leq \|f(A,x) \|_2 \leq 1, \| b_g \|_2 \leq 1$ 
\end{itemize}
We have 
\begin{itemize}
    \item {\bf Part 1.} $\| B_{6,1,1}^{j_1,*, j_0,*} \| \leq 6 \sqrt{d} R^2$  
    \item {\bf Part 2.} $\|B_{6,1,2}^{j_1,*, j_0,*}\| \preceq 6 \sqrt{d} R^2$
    \item {\bf Part 3.} $\|B_{6,1,3}^{j_1,*, j_0,*}\| \preceq 6  R^4$
    \item {\bf Part 4.} $\|B_{6,1,4}^{j_1,*, j_0,*} \|\preceq 6 R^4$
    \item {\bf Part 5.} $\|B_{6,1,5}^{j_1,*, j_0,*} \|\preceq 6  R^4$
    \item {\bf Part 6.} $\|B_{6,1,6}^{j_1,*, j_0,*} \|\preceq 9  R^4$
    \item {\bf Part 7.} $\|B_{6,1,7}^{j_1,*, j_0,*} \|\preceq 18  R^4$
    \item {\bf Part 8.} $\|B_{6,2,1}^{j_1,*, j_0,*}\| \preceq 15\sqrt{d} R^4$
    \item {\bf Part 9.} $\|B_{6,3,1}^{j_1,*, j_0,*}\| \preceq 15 \sqrt{d}R^2$
    \item {\bf Part 10.} $\|B_{6,4,1}^{j_1,*, j_0,*}\| \preceq 30\sqrt{d} R^4$
    \item {\bf Part 11.} $\|B_{6,5,1}^{j_1,*, j_0,*}\| \preceq 5\sqrt{d} R^4$
    \item {\bf Part 12.} $\|B_{6,5,2}^{j_1,*, j_0,*}\| \preceq 5\sqrt{d}R^4$
    \item {\bf Part 13.} $\|B_{6,6,1}^{j_1,*, j_0,*}\| \preceq 10\sqrt{d}R^4$
    \item {\bf Part 14.} $\|B_{6,6,2}^{j_1,*, j_0,*}\| \preceq 10\sqrt{d} R^4$
    \item {\bf Part 15.} $\|B_{6,7,1}^{j_1,*, j_0,*}\| \preceq 5\sqrt{d} R^4$
    \item {\bf Part 16.} $\|B_{6}^{j_1,*, j_0,*}\| \preceq 147\sqrt{d} R^4$
\end{itemize}
\end{lemma}

\begin{proof}
    {\bf Proof of Part 1.}
    \begin{align*}
        \| B_{6,1,1}^{j_1,*, j_0,*} \| 
        = & ~ \|  f_c(A,x)  \cdot f(A,x)_{j_1} \cdot f(A,x)_{j_0} \cdot  (-f_c(A,x) + f(A,x)_{j_0}) \cdot h(A,x)\cdot  {\bf 1}_d^\top\| \\
        \leq & ~ |f(A,x)_{j_1}| \cdot |f(A,x)_{j_0}| \cdot |f_c(A,x)|  \cdot  (|-f_c(A,x)| + |f(A,x)_{j_0}|) \cdot \|h(A,x)\|_2 \cdot \| {\bf 1}_d^\top\|_2\\
        \leq & ~ 6 \sqrt{d} R^2
    \end{align*}

    {\bf Proof of Part 2.}
    \begin{align*}
        \| B_{6,1,2}^{j_1,*, j_0,*} \|
        = &~
        \| f(A,x)_{j_1} \cdot f(A,x)_{j_0} \cdot c(A,x)_{j_1} \cdot  (-f_c(A,x) + f(A,x)_{j_0}) \cdot h(A,x) \cdot  {\bf 1}_d^\top  \| \\
        \preceq & ~  |f(A,x)_{j_1}| \cdot |f(A,x)_{j_0}| \cdot |c(A,x)_{j_1}|\cdot (|-f_c(A,x)| + |f(A,x)_{j_0}|) \cdot \|h(A,x)\|_2 \cdot \| {\bf 1}_d^\top\|_2\\
        \preceq & ~ 6 \sqrt{d} R^2
    \end{align*}

    {\bf Proof of Part 3.}
    \begin{align*}
     & ~\| B_{6,1,3}^{j_1,*, j_0,*} \| \\
     = & ~ \| f(A,x)_{j_1} \cdot  f(A,x)_{j_0} \cdot f_c(A,x)\cdot  (-f_c(A,x) + f(A,x)_{j_0}) \cdot  ((A_{j_1,*}) \circ x^\top) \cdot h(A,x)  \cdot I_d \| \\
    \leq & ~ |f(A,x)_{j_1}| \cdot |f(A,x)_{j_0}| \cdot  |f_c(A,x)|\cdot (|-f_c(A,x)| + |f(A,x)_{j_0}|)     \cdot \| A_{j_1,*}^\top \circ x \|_2 \cdot \|h(A,x)\|_2 \cdot \| I_d\| \\
    \leq & ~ |f(A,x)_{j_1}| \cdot |f(A,x)_{j_0}| \cdot  |f_c(A,x)|\cdot (|-f_c(A,x)| + |f(A,x)_{j_0}|)    \cdot \| A_{j_1,*}\|_2 \cdot \|\diag(x) \|_{\infty} \cdot \|h(A,x)\|_2 \cdot \| I_d\| \\
    \leq &  ~  6 R^4
    \end{align*}
   
    {\bf Proof of Part 4.}
    \begin{align*}
        \|B_{6,1,4}^{j_1,*, j_0,*}  \|
        = & ~  \|-f(A,x)_{j_1} \cdot  f(A,x)_{j_0} \cdot f_c(A,x)\cdot  (-f_c(A,x) + f(A,x)_{j_0})  \cdot h(A,x)^\top \cdot  h(A,x) \cdot I_d \|\\
        \leq & ~   \| f(A,x)_{j_1} \cdot  f(A,x)_{j_0} \cdot f_c(A,x)\cdot (|-f_c(A,x)| + |f(A,x)_{j_0}|)  \cdot h(A,x)^\top \cdot  h(A,x) \cdot I_d \|\\
        \leq & ~ | f(A,x)_{j_1} | \cdot|  f(A,x)_{j_0} |\cdot |f_c(A,x)|\cdot (|-f_c(A,x)| + |f(A,x)_{j_0}|) \cdot \|h(A,x)^\top\|_2 \cdot  \|h(A,x)\|_2  \cdot \| I_d\|\\
        \leq & ~ | f(A,x)_{j_1} | \cdot|  f(A,x)_{j_0} |\cdot |f_c(A,x)|\cdot (|-f_c(A,x)| + |f(A,x)_{j_0}|) \cdot \|h(A,x)^\top\|_2 \cdot  \|h(A,x)\|_2  \cdot \| I_d\| \\
        \leq & ~ 6 R^4
    \end{align*}

    {\bf Proof of Part 5.}
    \begin{align*}
        & ~ \|B_{6,1,5}^{j_1,*, j_0,*} \| \\
        = & ~ \| f(A,x)_{j_1} \cdot  f(A,x)_{j_0} \cdot (-f_2(A,x) + f(A,x)_{j_1})  \cdot (-f_c(A,x) + f(A,x)_{j_0})  \cdot h(A,x)^\top \cdot  h(A,x) \cdot I_d\| \\
        \leq & ~  | f(A,x)_{j_1}| \cdot |f(A,x)_{j_0}|    \cdot (|-f_2(A,x)| + |f(A,x)_{j_1}|) \cdot (|-f_c(A,x)| + |f(A,x)_{j_0}|)  \cdot \|h(A,x)^\top\|_2  \\
        & ~ \cdot \|h(A,x)\|_2 \cdot \| I_d\|\\
        \leq & ~ 6 R^4
    \end{align*}

    {\bf Proof of Part 6.}
    \begin{align*}
        & ~ \|B_{6,1,6}^{j_1,*, j_0,*} \| \\
        = & ~ \| f(A,x)_{j_1} \cdot  f(A,x)_{j_0} \cdot (-f_c(A,x) + f(A,x)_{j_1}) \cdot (-f_c(A,x) + f(A,x)_{j_0})   \cdot h(A,x)^\top \cdot  h(A,x) \cdot I_d \|\\
        \leq & ~  | f(A,x)_{j_1}| \cdot |f(A,x)_{j_0}|    \cdot (|-f_c(A,x)| + |f(A,x)_{j_1}|)  \cdot (|-f_c(A,x)| + |f(A,x)_{j_0}|)  \cdot \|h(A,x)^\top\|_2   \\
        & ~\cdot \|h(A,x)\|_2 \cdot \| I_d\| \\
        \leq & ~ 9 R^4
    \end{align*}

    {\bf Proof of Part 7.}
    \begin{align*}
         \|B_{6,1,7}^{j_1,*, j_0,*} \|
         = & ~ \|f(A,x)_{j_1} \cdot  f(A,x)_{j_0} \cdot (-f_c(A,x) + f(A,x)_{j_0})   \cdot p_{j_1}(A,x)^\top \cdot   h(A,x) \cdot I_d \|\\
         \leq & ~ | f(A,x)_{j_1}| \cdot |f(A,x)_{j_0}| \cdot (|-f_c(A,x)| + |f(A,x)_{j_0}|) \cdot \|p_{j_1}(A,x)^\top\|_2  \cdot \|h(A,x)\|_2 \cdot \| I_d\|\\
         \leq & ~ 18  R^4
    \end{align*}

    {\bf Proof of Part 8.}
    \begin{align*}
        \|B_{6,2,1}^{j_1,*, j_0,*} \|
        = & ~ \|  f(A,x)_{j_1} \cdot  f(A,x)_{j_0} \cdot (-f_c(A,x) + f(A,x)_{j_0}) \cdot c_g(A,x)^{\top} \cdot h(A,x) \cdot x \cdot {\bf 1}_d^{\top} \| \\
        \leq & ~  | f(A,x)_{j_1}| \cdot |f(A,x)_{j_0}| \cdot (|-f_c(A,x)| + |f(A,x)_{j_0}|) \cdot \|c_g(A,x)^{\top}\|_2 \cdot \|h(A,x)\|_2 \cdot \| x\|_2 \cdot \| {\bf 1}_d^{\top}\|_2\\
        \leq & ~ 15 \sqrt{d} R^4
    \end{align*}

    {\bf Proof of Part 9.}
    \begin{align*}
     \|B_{6,3,1}^{j_1,*, j_0,*} \|
        = & ~ \|   -  f(A,x)_{j_1} \cdot   f(A,x)_{j_0}  \cdot    (-f_c(A,x) + f(A,x)_{j_0})   \cdot {\bf 1}_d \cdot c_g(A,x)^{\top} \cdot \diag (x)\| \\
        \leq & ~ \|   f(A,x)_{j_1} \cdot   f(A,x)_{j_0}  \cdot  (-f_c(A,x) + f(A,x)_{j_0})   \cdot {\bf 1}_d \cdot c_g(A,x)^{\top} \cdot \diag (x)\| \\
        \leq & ~ | f(A,x)_{j_1}| \cdot  |f(A,x)_{j_0} | \cdot  (|-f_c(A,x)| + |f(A,x)_{j_0}|) \cdot \| {\bf 1}_d\|_2 \cdot \|c_g(A,x)^{\top}\|_2 \cdot \| \diag (x)\|  \\
        \leq & ~ 15 \sqrt{d} R^2
    \end{align*}

        {\bf Proof of Part 10.}
    \begin{align*}
     \|B_{6,4,1}^{j_1,*, j_0,*} \|
        = & ~ \|  -f(A,x)_{j_1} \cdot f(A,x)_{j_0}  \cdot  (-f_c(A,x) + f(A,x)_{j_0}) \cdot  c_g(A,x)^{\top} \cdot h_e(A,x) \cdot x \cdot {\bf 1}_d^{\top}\| \\
        \leq & ~ \|  f(A,x)_{j_1} \cdot f(A,x)_{j_0}  \cdot  (-f_c(A,x) + f(A,x)_{j_0}) \cdot  c_g(A,x)^{\top} \cdot h_e(A,x) \cdot x \cdot {\bf 1}_d^{\top}\| \\
        \leq & ~ | f(A,x)_{j_1}| \cdot  |f(A,x)_{j_0} | \cdot  (|-f_c(A,x)| + |f(A,x)_{j_0}|) \cdot |c_g(A,x)^{\top}| \cdot \| h_e(A,x) \|_2 \cdot \| x\|_2 \cdot \| {\bf 1}_d^{\top}\|_2\\
        \leq & ~ 30 \sqrt{d}R^4
    \end{align*}

       {\bf Proof of Part 11.}
    \begin{align*}
     \|B_{6,5,1}^{j_1,*, j_0,*} \|
        = & ~ \| - f(A,x)_{j_1} \cdot f(A,x)_{j_0} \cdot  f_2(A,x)  \cdot  c_g(A,x)^{\top} \cdot  h(A,x) \cdot x \cdot {\bf 1}_d^{\top}  \| \\
        \leq& ~ \|  f(A,x)_{j_1} \cdot f(A,x)_{j_0} \cdot  f_2(A,x)  \cdot  c_g(A,x)^{\top} \cdot  h(A,x) \cdot x \cdot {\bf 1}_d^{\top}  \| \\
        \leq & ~ | f(A,x)_{j_1}|  \cdot  |f(A,x)_{j_0} | \cdot  |f_2(A,x) |  \cdot \|c_g(A,x)^{\top}\|_2 \cdot \|h(A,x)\|_2 \cdot \| x\|_2 \cdot \| {\bf 1}_d^{\top}\|_2\\
        \leq & ~ 5\sqrt{d} R^4
    \end{align*}

           {\bf Proof of Part 12.}
    \begin{align*}
     \|B_{6,5,2}^{j_1,*, j_0,*} \|
        = & ~ \|  f(A,x)_{j_1}^2 \cdot f(A,x)_{j_0}    \cdot  c_g(A,x)^{\top} \cdot  h(A,x) \cdot x \cdot {\bf 1}_d^{\top}  \| \\
        \leq & ~ | f(A,x)_{j_1}|^2  \cdot  |f(A,x)_{j_0} |  \cdot \|c_g(A,x)^{\top}\|_2 \cdot \|h(A,x)\|_2 \cdot \| x\|_2 \cdot \| {\bf 1}_d^{\top}\|_2\\
        \leq & ~ 5\sqrt{d} R^4
    \end{align*}

           {\bf Proof of Part 13.}
    \begin{align*}
     \|B_{6,6,1}^{j_1,*, j_0,*} \|
        = & ~ \| - f(A,x)_{j_1} \cdot f(A,x)_{j_0} \cdot  f_c(A,x)  \cdot  c_g(A,x)^{\top} \cdot  h(A,x) \cdot x \cdot {\bf 1}_d^{\top}  \| \\
        \leq& ~ \|  f(A,x)_{j_1} \cdot f(A,x)_{j_0} \cdot  f_c(A,x)  \cdot  c_g(A,x)^{\top} \cdot  h(A,x) \cdot x \cdot {\bf 1}_d^{\top}  \| \\
        \leq & ~ | f(A,x)_{j_1}|  \cdot  |f(A,x)_{j_0} | \cdot  |f_c(A,x) |  \cdot \|c_g(A,x)^{\top}\|_2 \cdot \|h(A,x)\|_2 \cdot \| x\|_2 \cdot \| {\bf 1}_d^{\top}\|_2\\
        \leq & ~ 10\sqrt{d} R^4
    \end{align*}

           {\bf Proof of Part 14.}
    \begin{align*}
     \|B_{6,6,2}^{j_1,*, j_0,*} \|
       = & ~ \|  f(A,x)_{j_1} \cdot c(A,x)_{j_1} \cdot f(A,x)_{j_0}    \cdot  c_g(A,x)^{\top} \cdot  h(A,x) \cdot x \cdot {\bf 1}_d^{\top}  \| \\
        \leq & ~ | f(A,x)_{j_1}|\cdot  |c(A,x)_{j_1} |   \cdot  |f(A,x)_{j_0} |  \cdot \|c_g(A,x)^{\top}\|_2 \cdot \|h(A,x)\|_2 \cdot \| x\|_2 \cdot \| {\bf 1}_d^{\top}\|_2\\
        \leq & ~ 10\sqrt{d} R^4
    \end{align*}
    
           {\bf Proof of Part 15.}
    \begin{align*}
     \|B_{6,7,1}^{j_1,*, j_0,*} \|
       = & ~ \|  f(A,x)_{j_1}\cdot f(A,x)_{j_0}^2   \cdot c_g(A,x)^{\top} \cdot h(A,x)  \cdot x \cdot {\bf 1}_d^{\top}  \| \\
        \leq & ~ | f(A,x)_{j_1}|  \cdot  |f(A,x)_{j_0} |^2  \cdot \|c_g(A,x)^{\top}\|_2 \cdot \|h(A,x)\|_2 \cdot \| x\|_2 \cdot \| {\bf 1}_d^{\top}\|_2\\
        \leq & ~ 5\sqrt{d} R^4
    \end{align*}

        {\bf Proof of Part 16.}
\begin{align*}
 & ~ \| B_{6}^{j_1,*,j_0,*}\| \\
   = & ~ \|\sum_{i = 1}^7 B_{6,i}^{j_1,*,j_0,*}\| \\
    \leq & ~ 147\sqrt{d} R^4
\end{align*}
\end{proof}

\section{Hessian: Seventh term  \texorpdfstring{$B_7^{j_1,i_1,j_0,i_0}$}{}}\label{app:hessian_seventh}
\subsection{Definitions}
\begin{definition} \label{def:b_7}
    We define the $B_7^{j_1,i_1,j_0,i_0}$ as follows,
    \begin{align*}
        B_7^{j_1,i_1,j_0,i_0} : = & ~ \frac{\d}{\d A_{j_1,i_1}}( - c_g(A,x)^{\top} \cdot f(A,x)_{j_0}\diag(x) A^{\top} ((e_{j_0} - f(A,x)) \circ q(A,x)))
    \end{align*}
    Then, we define $B_{7,1}^{j_1,i_1,j_0,i_0}, \cdots, B_{7,6}^{j_1,i_1,j_0,i_0}$ as follow
    \begin{align*}
        B_{7,1}^{j_1,i_1,j_0,i_0} : = & ~ \frac{\d}{\d A_{j_1,i_1}} (- c_g(A,x)^{\top} ) \cdot  f(A,x)_{j_0} \cdot \diag(x) \cdot A^{\top} \cdot ((e_{j_0} - f(A,x)) \circ q(A,x)))\\
B_{7,2}^{j_1,i_1,j_0,i_0} : = & ~- c_g(A,x)^{\top} \cdot \frac{\d}{\d A_{j_1,i_1}} ( f(A,x)_{j_0} )  \cdot \diag(x) \cdot A^{\top}   \cdot((e_{j_0} - f(A,x)) \circ q(A,x)))\\
B_{7,3}^{j_1,i_1,j_0,i_0} : = & ~  - c_g(A,x)^{\top} \cdot f(A,x)_{j_0} \cdot \diag(x) \cdot \frac{\d}{\d A_{j_1,i_1}} (  A^{\top} )   \cdot((e_{j_0} - f(A,x)) \circ q(A,x))) \\
B_{7,4}^{j_1,i_1,j_0,i_0} : = & ~  - c_g(A,x)^{\top} \cdot f(A,x)_{j_0} \cdot \diag(x) \cdot A^{\top} \cdot((e_{j_0} -\frac{\d f(A,x)}{\d A_{j_1,i_1}} ) \circ q(A,x))) \\
B_{7,5}^{j_1,i_1,j_0,i_0} : = & ~  - c_g(A,x)^{\top} \cdot f(A,x)_{j_0} \cdot \diag(x) \cdot A^{\top} \cdot((e_{j_0} - f(A,x)) \circ \frac{\d q(A,x)}{\d A_{j_1,i_1}}) \\
    \end{align*}
    It is easy to show
    \begin{align*}
        B_7^{j_1,i_1,j_0,i_0} = B_{7,1}^{j_1,i_1,j_0,i_0} +  B_{7,2}^{j_1,i_1,j_0,i_0} + B_{7,3}^{j_1,i_1,j_0,i_0} + B_{7,4}^{j_1,i_1,j_0,i_0} + B_{7,5}^{j_1,i_1,j_0,i_0} 
    \end{align*}
    Similarly for $j_1 = j_0$ and $i_0 = i_1$,we have
    \begin{align*}
        B_7^{j_1,i_1,j_1,i_1} = B_{7,1}^{j_1,i_1,j_1,i_1} +  B_{7,2}^{j_1,i_1,j_1,i_1} + B_{7,3}^{j_1,i_1,j_1,i_1} + B_{7,4}^{j_1,i_1,j_1,i_1} + B_{7,5}^{j_1,i_1,j_1,i_1} 
         \end{align*}
    For $j_1 = j_0$ and $i_0 \neq i_1$,we have
    \begin{align*}
       B_7^{j_1,i_1,j_1,i_0} = B_{7,1}^{j_1,i_1,j_1,i_0} +  B_{7,2}^{j_1,i_1,j_1,i_0} + B_{7,3}^{j_1,i_1,j_1,i_0} + B_{7,4}^{j_1,i_1,j_1,i_0} + B_{7,5}^{j_1,i_1,j_1,i_0}  
    \end{align*}
    For $j_1 \neq j_0$ and $i_0 = i_1$,we have
    \begin{align*}
       B_7^{j_1,i_1,j_0,i_1} = B_{7,1}^{j_1,i_1,j_0,i_1} +  B_{7,2}^{j_1,i_1,j_0,i_1} + B_{7,3}^{j_1,i_1,j_0,i_1} + B_{7,4}^{j_1,i_1,j_0,i_1} + B_{7,5}^{j_1,i_1,j_0,i_1} 
    \end{align*}
\end{definition}
\subsection{Case \texorpdfstring{$j_1=j_0, i_1 = i_0$}{}}
\begin{lemma}
For $j_1 = j_0$ and $i_0 = i_1$. If the following conditions hold
    \begin{itemize}
     \item Let $u(A,x) \in \R^n$ be defined as Definition~\ref{def:u}
    \item Let $\alpha(A,x) \in \R$ be defined as Definition~\ref{def:alpha}
     \item Let $f(A,x) \in \R^n$ be defined as Definition~\ref{def:f}
    \item Let $c(A,x) \in \R^n$ be defined as Definition~\ref{def:c}
    \item Let $g(A,x) \in \R^d$ be defined as Definition~\ref{def:g} 
    \item Let $q(A,x) = c(A,x) + f(A,x) \in \R^n$
    \item Let $c_g(A,x) \in \R^d$ be defined as Definition~\ref{def:c_g}.
    \item Let $L_g(A,x) \in \R$ be defined as Definition~\ref{def:l_g}
    \item Let $v \in \R^n$ be a vector 
    \item Let $B_1^{j_1,i_1,j_0,i_0}$ be defined as Definition~\ref{def:b_1}
    \end{itemize}
    Then, For $j_0,j_1 \in [n], i_0,i_1 \in [d]$, we have 
    \begin{itemize}
\item {\bf Part 1.} For $B_{7,1}^{j_1,i_1,j_1,i_1}$, we have 
\begin{align*}
 B_{7,1}^{j_1,i_1,j_1,i_1}  = & ~ \frac{\d}{\d A_{j_1,i_1}} (- c_g(A,x)^{\top} ) \cdot f(A,x)_{j_1} \cdot \diag(x) \cdot A^{\top} \cdot ((e_{j_1} - f(A,x)) \circ q(A,x))\\
 = & ~ B_{7,1,1}^{j_1,i_1,j_1,i_1} + B_{7,1,2}^{j_1,i_1,j_1,i_1} + B_{7,1,3}^{j_1,i_1,j_1,i_1} + B_{7,1,4}^{j_1,i_1,j_1,i_1} + B_{7,1,5}^{j_1,i_1,j_1,i_1}  
\end{align*} 
\item {\bf Part 2.} For $B_{7,2}^{j_1,i_1,j_1,i_1}$, we have 
\begin{align*}
  B_{7,2}^{j_1,i_1,j_1,i_1} = & ~ - c_g(A,x)^{\top} \cdot \frac{\d}{\d A_{j_1,i_1}} ( f(A,x)_{j_1} )  \cdot \diag(x) \cdot A^{\top} \cdot ((e_{j_1} - f(A,x)) \circ q(A,x)) \\
    = & ~  B_{7,2,1}^{j_1,i_1,j_1,i_1} + B_{7,2,2}^{j_1,i_1,j_1,i_1}
\end{align*} 
\item {\bf Part 3.} For $B_{7,3}^{j_1,i_1,j_1,i_1}$, we have 
\begin{align*}
  B_{7,3}^{j_1,i_1,j_1,i_1} = & ~ - c_g(A,x)^{\top} \cdot f(A,x)_{j_1} \cdot \diag(x) \cdot \frac{\d}{\d A_{j_1,i_1}} (  A^{\top} ) \cdot ((e_{j_1} - f(A,x)) \circ q(A,x))  \\
     = & ~ B_{7,3,1}^{j_1,i_1,j_1,i_1}  
\end{align*} 
\item {\bf Part 4.} For $B_{7,4}^{j_1,i_1,j_1,i_1}$, we have 
\begin{align*}
  B_{7,4}^{j_1,i_1,j_1,i_1} = & ~ - c_g(A,x)^{\top} \cdot f(A,x)_{j_1} \cdot \diag(x) \cdot A^{\top} \cdot ( ( \frac{\d}{\d A_{j_1,i_1}} (e_{j_1} - f(A,x)) ) \circ q(A,x))  \\
     = & ~ B_{7,4,1}^{j_1,i_1,j_1,i_1} 
\end{align*}
\item {\bf Part 5.} For $B_{6,5}^{j_1,i_1,j_1,i_1}$, we have 
\begin{align*}
  B_{7,5}^{j_1,i_1,j_1,i_1} = & ~  - c_g(A,x)^{\top} \cdot f(A,x)_{j_1} \cdot \diag(x) \cdot A^{\top} \cdot ((e_{j_1} - f(A,x)) \circ \frac{\d}{\d A_{j_1,i_1}} ( q(A,x))) \\
     = & ~ B_{7,5,1}^{j_1,i_1,j_1,i_1}     
\end{align*} 
\end{itemize}
\end{lemma}
\begin{proof}
    {\bf Proof of Part 1.}
    \begin{align*}
    B_{7,1,1}^{j_1,i_1,j_1,i_1} : = & ~  e_{i_1}^\top \cdot \langle c(A,x), f(A,x) \rangle \cdot  f(A,x)_{j_1}^2 \cdot \diag(x) \cdot A^{\top} \cdot ((e_{j_1} - f(A,x)) \circ q(A,x))\\
    B_{7,1,2}^{j_1,i_1,j_1,i_1} : = & ~  e_{i_1}^\top \cdot c(A,x)_{j_1}\cdot f(A,x)_{j_1}^2 \cdot \diag(x) \cdot A^{\top} \cdot ((e_{j_1} - f(A,x)) \circ q(A,x))\\
    B_{7,1,3}^{j_1,i_1,j_1,i_1} : = & ~ f(A,x)_{j_1}^2 \cdot \langle c(A,x), f(A,x) \rangle \cdot ( (A_{j_1,*}) \circ x^\top  )  \cdot \diag(x) \cdot A^{\top} \cdot ((e_{j_1} - f(A,x)) \circ q(A,x))\\
    B_{7,1,4}^{j_1,i_1,j_1,i_1} : = & ~  -  f(A,x)_{j_1}^2 \cdot f(A,x)^\top  \cdot A \cdot (\diag(x))^2 \cdot   \langle c(A,x), f(A,x) \rangle \cdot A^{\top} \cdot ((e_{j_1} - f(A,x)) \circ q(A,x))\\
    B_{7,1,5}^{j_1,i_1,j_1,i_1} : = & ~     f(A,x)_{j_1}^2 \cdot f(A,x)^\top  \cdot A \cdot (\diag(x))^2 \cdot (\langle -f(A,x), f(A,x) \rangle + f(A,x)_{j_1})   \\
        & ~\cdot A^{\top} \cdot ((e_{j_1} - f(A,x)) \circ q(A,x)) \\
    B_{7,1,6}^{j_1,i_1,j_1,i_1} : = & ~  f(A,x)_{j_1}^2 \cdot f(A,x)^\top  \cdot A \cdot  (\diag(x))^2 \cdot(\langle -f(A,x), c(A,x) \rangle + f(A,x)_{j_1})  \\
        & ~\cdot A^{\top} \cdot ((e_{j_1} - f(A,x)) \circ q(A,x)) \\
    B_{7,1,7}^{j_1,i_1,j_1,i_1} : = & ~  f(A,x)_{j_1}^2 \cdot ((e_{j_1}^\top - f(A,x)^\top) \circ q(A,x)^\top) \cdot A \cdot  (\diag(x))^2 \cdot A^{\top} \cdot ((e_{j_1} - f(A,x)) \circ q(A,x))
\end{align*}
Finally, combine them and we have
\begin{align*}
       B_{7,1}^{j_1,i_1,j_1,i_1} = B_{7,1,1}^{j_1,i_1,j_1,i_1} + B_{7,1,2}^{j_1,i_1,j_1,i_1} + B_{7,1,3}^{j_1,i_1,j_1,i_1} + B_{7,1,4}^{j_1,i_1,j_1,i_1} + B_{7,1,5}^{j_1,i_1,j_1,i_1}  
\end{align*}
{\bf Proof of Part 2.}
    \begin{align*}
    B_{7,2,1}^{j_1,i_1,j_1,i_1} : = & ~   c_g(A,x)^{\top} \cdot f(A,x)_{j_1}^2 \cdot x_{i_1} \cdot \diag(x) \cdot A^{\top} \cdot ((e_{j_1} - f(A,x)) \circ q(A,x))  \\
    B_{7,2,2}^{j_1,i_1,j_1,i_1} : = & ~  - c_g(A,x)^{\top} \cdot f(A,x)_{j_1} \cdot x_{i_1} \cdot   \diag(x) \cdot A^{\top} \cdot ((e_{j_1} - f(A,x)) \circ q(A,x))
\end{align*}
Finally, combine them and we have
\begin{align*}
       B_{7,2}^{j_1,i_1,j_1,i_1} = B_{7,2,1}^{j_1,i_1,j_1,i_1} + B_{7,2,2}^{j_1,i_1,j_1,i_1}
\end{align*}
{\bf Proof of Part 3.} 
    \begin{align*}
    B_{7,3,1}^{j_1,i_1,j_1,i_1} : = & ~   -  c_g(A,x)^{\top} \cdot f(A,x)_{j_1} \cdot \diag(x) \cdot e_{i_1} \cdot e_{j_1}^\top \cdot ((e_{j_1} - f(A,x)) \circ q(A,x))
\end{align*}
Finally, combine them and we have
\begin{align*}
       B_{7,3}^{j_1,i_1,j_1,i_1} = B_{7,3,1}^{j_1,i_1,j_1,i_1} 
\end{align*}
{\bf Proof of Part 4.} 
    \begin{align*}
    B_{7,4,1}^{j_1,i_1,j_1,i_1} : = & ~  c_g(A,x)^{\top} \cdot f(A,x)_{j_1}^2 \cdot   \diag(x) \cdot A^{\top} \cdot x_{i_1}  \cdot (    (e_{j_1}- f(A,x) )  \circ q(A,x)) 
\end{align*}
Finally, combine them and we have
\begin{align*}
       B_{7,4}^{j_1,i_1,j_1,i_1} = B_{7,4,1}^{j_1,i_1,j_1,i_1}  
\end{align*}
{\bf Proof of Part 5.} 
    \begin{align*}
    B_{7,5,1}^{j_1,i_1,j_1,i_1} : = & ~-2 c_g(A,x)^{\top}  \cdot f(A,x)_{j_1}^2 \cdot \diag(x) \cdot A^{\top} \cdot x_{i_1}  \cdot ((e_{j_1} - f(A,x)) \circ  ( (e_{j_1} - f(A,x)) ) ) 
\end{align*}
Finally, combine them and we have
\begin{align*}
       B_{7,5}^{j_1,i_1,j_1,i_1} = B_{7,5,1}^{j_1,i_1,j_1,i_1}   
\end{align*}
\end{proof}
\subsection{Case \texorpdfstring{$j_1=j_0, i_1 \neq i_0$}{}}
\begin{lemma}
For $j_1 = j_0$ and $i_0 \neq i_1$. If the following conditions hold
    \begin{itemize}
     \item Let $u(A,x) \in \R^n$ be defined as Definition~\ref{def:u}
    \item Let $\alpha(A,x) \in \R$ be defined as Definition~\ref{def:alpha}
     \item Let $f(A,x) \in \R^n$ be defined as Definition~\ref{def:f}
    \item Let $c(A,x) \in \R^n$ be defined as Definition~\ref{def:c}
    \item Let $g(A,x) \in \R^d$ be defined as Definition~\ref{def:g} 
    \item Let $q(A,x) = c(A,x) + f(A,x) \in \R^n$
    \item Let $c_g(A,x) \in \R^d$ be defined as Definition~\ref{def:c_g}.
    \item Let $L_g(A,x) \in \R$ be defined as Definition~\ref{def:l_g}
    \item Let $v \in \R^n$ be a vector 
    \item Let $B_1^{j_1,i_1,j_0,i_0}$ be defined as Definition~\ref{def:b_1}
    \end{itemize}
    Then, For $j_0,j_1 \in [n], i_0,i_1 \in [d]$, we have 
    \begin{itemize}
\item {\bf Part 1.} For $B_{7,1}^{j_1,i_1,j_1,i_0}$, we have 
\begin{align*}
 B_{7,1}^{j_1,i_1,j_1,i_0}  = & ~ \frac{\d}{\d A_{j_1,i_1}} (- c_g(A,x)^{\top} ) \cdot f(A,x)_{j_1} \cdot \diag(x) \cdot A^{\top} \cdot ((e_{j_1} - f(A,x)) \circ q(A,x))\\
 = & ~ B_{7,1,1}^{j_1,i_1,j_1,i_0} + B_{7,1,2}^{j_1,i_1,j_1,i_0} + B_{7,1,3}^{j_1,i_1,j_1,i_0} + B_{7,1,4}^{j_1,i_1,j_1,i_0} + B_{7,1,5}^{j_1,i_1,j_1,i_0}  
\end{align*} 
\item {\bf Part 2.} For $B_{7,2}^{j_1,i_1,j_1,i_0}$, we have 
\begin{align*}
  B_{7,2}^{j_1,i_1,j_1,i_0} = & ~ - c_g(A,x)^{\top} \cdot \frac{\d}{\d A_{j_1,i_1}} ( f(A,x)_{j_1} )  \cdot \diag(x) \cdot A^{\top} \cdot ((e_{j_1} - f(A,x)) \circ q(A,x)) \\
    = & ~  B_{7,2,1}^{j_1,i_1,j_1,i_0} + B_{7,2,2}^{j_1,i_1,j_1,i_0}
\end{align*} 
\item {\bf Part 3.} For $B_{7,3}^{j_1,i_1,j_1,i_0}$, we have 
\begin{align*}
  B_{7,3}^{j_1,i_1,j_1,i_0} = & ~ - c_g(A,x)^{\top} \cdot f(A,x)_{j_1} \cdot \diag(x) \cdot \frac{\d}{\d A_{j_1,i_1}} (  A^{\top} ) \cdot ((e_{j_1} - f(A,x)) \circ q(A,x))  \\
     = & ~ B_{7,3,1}^{j_1,i_1,j_1,i_0}  
\end{align*} 
\item {\bf Part 4.} For $B_{7,4}^{j_1,i_1,j_1,i_0}$, we have 
\begin{align*}
  B_{7,4}^{j_1,i_1,j_1,i_0} = & ~ - c_g(A,x)^{\top} \cdot f(A,x)_{j_1} \cdot \diag(x) \cdot A^{\top} \cdot ( ( \frac{\d}{\d A_{j_1,i_1}} (e_{j_1} - f(A,x)) ) \circ q(A,x))  \\
     = & ~ B_{7,4,1}^{j_1,i_1,j_1,i_0} 
\end{align*}
\item {\bf Part 5.} For $B_{6,5}^{j_1,i_1,j_1,i_0}$, we have 
\begin{align*}
  B_{7,5}^{j_1,i_1,j_1,i_0} = & ~  - c_g(A,x)^{\top} \cdot f(A,x)_{j_1} \cdot \diag(x) \cdot A^{\top} \cdot ((e_{j_1} - f(A,x)) \circ \frac{\d}{\d A_{j_1,i_1}} ( q(A,x))) \\
     = & ~ B_{7,5,1}^{j_1,i_1,j_1,i_0}     
\end{align*} 
\end{itemize}
\end{lemma}
\begin{proof}
    {\bf Proof of Part 1.}
    \begin{align*}
    B_{7,1,1}^{j_1,i_1,j_1,i_0} : = & ~  e_{i_1}^\top \cdot \langle c(A,x), f(A,x) \rangle \cdot  f(A,x)_{j_1}^2 \cdot \diag(x) \cdot A^{\top} \cdot ((e_{j_1} - f(A,x)) \circ q(A,x))\\
    B_{7,1,2}^{j_1,i_1,j_1,i_0} : = & ~  e_{i_1}^\top \cdot c(A,x)_{j_1}\cdot f(A,x)_{j_1}^2 \cdot \diag(x) \cdot A^{\top} \cdot ((e_{j_1} - f(A,x)) \circ q(A,x))\\
    B_{7,1,3}^{j_1,i_1,j_1,i_0} : = & ~ f(A,x)_{j_1}^2 \cdot \langle c(A,x), f(A,x) \rangle \cdot ( (A_{j_1,*}) \circ x^\top  )  \cdot \diag(x) \cdot A^{\top} \cdot ((e_{j_1} - f(A,x)) \circ q(A,x))\\
    B_{7,1,4}^{j_1,i_1,j_1,i_0} : = & ~  -  f(A,x)_{j_1}^2 \cdot f(A,x)^\top  \cdot A \cdot (\diag(x))^2 \cdot   \langle c(A,x), f(A,x) \rangle \cdot A^{\top} \cdot ((e_{j_1} - f(A,x)) \circ q(A,x))\\
    B_{7,1,5}^{j_1,i_1,j_1,i_0} : = & ~     f(A,x)_{j_1}^2 \cdot f(A,x)^\top  \cdot A \cdot (\diag(x))^2 \cdot (\langle -f(A,x), f(A,x) \rangle + f(A,x)_{j_1})   \\
        & ~\cdot A^{\top} \cdot ((e_{j_1} - f(A,x)) \circ q(A,x)) \\
    B_{7,1,6}^{j_1,i_1,j_1,i_0} : = & ~  f(A,x)_{j_1}^2 \cdot f(A,x)^\top  \cdot A \cdot  (\diag(x))^2 \cdot(\langle -f(A,x), c(A,x) \rangle + f(A,x)_{j_1})  \\
        & ~\cdot A^{\top} \cdot ((e_{j_1} - f(A,x)) \circ q(A,x)) \\
    B_{7,1,7}^{j_1,i_1,j_1,i_0} : = & ~  f(A,x)_{j_1}^2 \cdot ((e_{j_1}^\top - f(A,x)^\top) \circ q(A,x)^\top) \cdot A \cdot  (\diag(x))^2 \cdot A^{\top} \cdot ((e_{j_1} - f(A,x)) \circ q(A,x))
\end{align*}
Finally, combine them and we have
\begin{align*}
       B_{7,1}^{j_1,i_1,j_1,i_0} = B_{7,1,1}^{j_1,i_1,j_1,i_0} + B_{7,1,2}^{j_1,i_1,j_1,i_0} + B_{7,1,3}^{j_1,i_1,j_1,i_0} + B_{7,1,4}^{j_1,i_1,j_1,i_0} + B_{7,1,5}^{j_1,i_1,j_1,i_0}  
\end{align*}
{\bf Proof of Part 2.}
    \begin{align*}
    B_{7,2,1}^{j_1,i_1,j_1,i_0} : = & ~   c_g(A,x)^{\top} \cdot f(A,x)_{j_1}^2 \cdot x_{i_1} \cdot \diag(x) \cdot A^{\top} \cdot ((e_{j_1} - f(A,x)) \circ q(A,x))  \\
    B_{7,2,2}^{j_1,i_1,j_1,i_0} : = & ~  - c_g(A,x)^{\top} \cdot f(A,x)_{j_1} \cdot x_{i_1} \cdot   \diag(x) \cdot A^{\top} \cdot ((e_{j_1} - f(A,x)) \circ q(A,x))
\end{align*}
Finally, combine them and we have
\begin{align*}
       B_{7,2}^{j_1,i_1,j_1,i_0} = B_{7,2,1}^{j_1,i_1,j_1,i_0} + B_{7,2,2}^{j_1,i_1,j_1,i_0}
\end{align*}
{\bf Proof of Part 3.} 
    \begin{align*}
    B_{7,3,1}^{j_1,i_1,j_1,i_0} : = & ~   -  c_g(A,x)^{\top} \cdot f(A,x)_{j_1} \cdot \diag(x) \cdot e_{i_1} \cdot e_{j_1}^\top \cdot ((e_{j_1} - f(A,x)) \circ q(A,x))
\end{align*}
Finally, combine them and we have
\begin{align*}
       B_{7,3}^{j_1,i_1,j_1,i_0} = B_{7,3,1}^{j_1,i_1,j_1,i_0} 
\end{align*}
{\bf Proof of Part 4.} 
    \begin{align*}
    B_{7,4,1}^{j_1,i_1,j_1,i_0} : = & ~  c_g(A,x)^{\top} \cdot f(A,x)_{j_1}^2 \cdot   \diag(x) \cdot A^{\top} \cdot x_{i_1}  \cdot (    (e_{j_1}- f(A,x) )  \circ q(A,x)) 
\end{align*}
Finally, combine them and we have
\begin{align*}
       B_{7,4}^{j_1,i_1,j_1,i_0} = B_{7,4,1}^{j_1,i_1,j_1,i_0}  
\end{align*}
{\bf Proof of Part 5.} 
    \begin{align*}
    B_{7,5,1}^{j_1,i_1,j_1,i_0} : = & ~-2 c_g(A,x)^{\top}  \cdot f(A,x)_{j_1}^2 \cdot \diag(x) \cdot A^{\top} \cdot x_{i_1}  \cdot ((e_{j_1} - f(A,x)) \circ  ( (e_{j_1} - f(A,x)) ) ) 
\end{align*}
Finally, combine them and we have
\begin{align*}
       B_{7,5}^{j_1,i_1,j_1,i_0} = B_{7,5,1}^{j_1,i_1,j_1,i_0}   
\end{align*}
\end{proof}
\subsection{Constructing \texorpdfstring{$d \times d$}{} matrices for \texorpdfstring{$j_1 = j_0$}{}}
The purpose of the following lemma is to let $i_0$ and $i_1$ disappear.
\begin{lemma}For $j_0,j_1 \in [n]$, a list of $d \times d$ matrices can be expressed as the following sense,
\begin{itemize}
\item {\bf Part 1.}
\begin{align*}
B_{7,1,1}^{j_1,*,j_1,*} & ~ =    f_c(A,x) \cdot  f(A,x)_{j_1}^2 \cdot  p_{j_1}(A,x) \cdot {\bf 1}_d^\top
\end{align*}
\item {\bf Part 2.}
\begin{align*}
B_{7,1,2}^{j_1,*,j_1,*} & ~ =      c(A,x)_{j_1}\cdot f(A,x)_{j_1}^2 \cdot  p_{j_1}(A,x) \cdot {\bf 1}_d^\top 
\end{align*}
\item {\bf Part 3.}
\begin{align*}
B_{7,1,3}^{j_1,*,j_1,*} & ~ =    f(A,x)_{j_1}^2 \cdot f_c(A,x) \cdot ( (A_{j_1,*}) \circ x^\top  )  \cdot   p_{j_1}(A,x)  \cdot I_d
\end{align*}
\item {\bf Part 4.}
\begin{align*}
B_{7,1,4}^{j_1,*,j_1,*}  & ~ =    -  f(A,x)_{j_1}^2 \cdot f_c(A,x)  \cdot h(A,x)^\top \cdot p_{j_1}(A,x) \cdot I_d
\end{align*}
\item {\bf Part 5.}
\begin{align*}
B_{7,1,5}^{j_1,*,j_1,*}  & ~ =     f(A,x)_{j_1}^2 \cdot (-f_2(A,x) + f(A,x)_{j_1})  \cdot h(A,x)^\top \cdot p_{j_1}(A,x) \cdot I_d
\end{align*}
\item {\bf Part 6.}
\begin{align*}
B_{7,1,6}^{j_1,*,j_1,*}  & ~ =      f(A,x)_{j_1}^2  \cdot(-f_c(A,x) + f(A,x)_{j_1})  \cdot h(A,x)^\top \cdot  p_{j_1}(A,x) \cdot I_d
\end{align*}
\item {\bf Part 7.}
\begin{align*}
B_{7,1,7}^{j_1,*,j_1,*}  & ~ =   f(A,x)_{j_1}^2  \cdot p_{j_1}(A,x)^\top \cdot p_{j_1}(A,x) \cdot I_d
\end{align*}
\item {\bf Part 8.}
\begin{align*}
B_{7,2,1}^{j_1,*,j_1,*}  & ~ =    f(A,x)_{j_1}^2  \cdot c_g(A,x)^{\top} \cdot p_{j_1}(A,x)\cdot x \cdot {\bf 1}_d^{\top} 
\end{align*}
\item {\bf Part 9.}
\begin{align*}
B_{7,2,2}^{j_1,*,j_1,*}  & ~ =    -f(A,x)_{j_1}  \cdot  c_g(A,x)^{\top} \cdot p_{j_1}(A,x) \cdot x \cdot {\bf 1}_d^{\top} 
\end{align*}
\item {\bf Part 10.}
\begin{align*}
 B_{7,3,1}^{j_1,*,j_1,*}  & ~ =     - f(A,x)_{j_1}\cdot e_{j_1}^\top \cdot ((e_{j_1} - f(A,x)) \circ q(A,x)) \cdot {\bf 1}_d \cdot c_g(A,x)^{\top}   \cdot \diag(x)   
\end{align*}
\item {\bf Part 11.}
\begin{align*}
B_{7,4,1}^{j_1,*,j_1,*}  =     f(A,x)_{j_1}^2 \cdot c_g(A,x)^{\top}  \cdot  p_{j_1}(A,x)\cdot x \cdot {\bf 1}_d^{\top}
\end{align*}
\item {\bf Part 12.}
\begin{align*}
B_{7,5,1}^{j_1,*,j_1,*}  & ~ =  -2f(A,x)_{j_1}^2 \cdot   c_g(A,x)^{\top} \cdot \diag(x) \cdot A^{\top} \cdot  ((e_{j_1} - f(A,x)) \circ  ( (e_{j_1} - f(A,x)) ) )   \cdot x \cdot {\bf 1}_d^{\top}
\end{align*} 

\end{itemize}
\end{lemma}
\begin{proof}
{\bf Proof of Part 1.}
    We have
    \begin{align*}
        B_{7,1,1}^{j_1,i_1,j_1,i_1}  = & ~ e_{i_1}^\top \cdot \langle c(A,x), f(A,x) \rangle \cdot  f(A,x)_{j_1}^2 \cdot \diag(x) \cdot A^{\top} \cdot ((e_{j_1} - f(A,x)) \circ q(A,x))\\
        B_{7,1,1}^{j_1,i_1,j_1,i_0}   = & ~  e_{i_1}^\top \cdot \langle c(A,x), f(A,x) \rangle \cdot  f(A,x)_{j_1}^2 \cdot \diag(x) \cdot A^{\top} \cdot ((e_{j_1} - f(A,x)) \circ q(A,x))
    \end{align*}
    From the above two equations, we can tell that $B_{7,1,1}^{j_1,*,j_1,*} \in \R^{d \times d}$ is a matrix that both the diagonal and off-diagonal have entries.
    
    Then we have $B_{7,1,1}^{j_1,*,j_1,*} \in \R^{d \times d}$ can be written as the rescaling of a diagonal matrix,
    \begin{align*}
     B_{7,1,1}^{j_1,*,j_1,*} & ~ = \langle c(A,x), f(A,x) \rangle \cdot  f(A,x)_{j_1}^2 \cdot \diag(x) \cdot A^{\top} \cdot ((e_{j_1} - f(A,x)) \circ q(A,x)) \cdot {\bf 1}_d^\top\\
     & ~ = f_c(A,x) \cdot  f(A,x)_{j_1}^2 \cdot  p_{j_1}(A,x) \cdot {\bf 1}_d^\top
\end{align*}
    where the last step is follows from the Definitions~\ref{def:f_c} and Definitions~\ref{def:p}.

{\bf Proof of Part 2.}
    We have
    \begin{align*}
           B_{7,1,2}^{j_1,i_1,j_1,i_1} = & ~ e_{i_1}^\top \cdot c(A,x)_{j_1}\cdot f(A,x)_{j_1}^2 \cdot \diag(x) \cdot A^{\top} \cdot ((e_{j_1} - f(A,x)) \circ q(A,x))\\
        B_{7,1,2}^{j_1,i_1,j_1,i_0} = & ~ e_{i_1}^\top \cdot c(A,x)_{j_1}\cdot f(A,x)_{j_1}^2 \cdot \diag(x) \cdot A^{\top} \cdot ((e_{j_1} - f(A,x)) \circ q(A,x))
    \end{align*}
     From the above two equations, we can tell that $B_{7,1,2}^{j_1,*,j_1,*} \in \R^{d \times d}$ is a matrix that only diagonal has entries and off-diagonal are all zeros.
    
    Then we have $B_{7,1,2}^{j_1,*,j_1,*} \in \R^{d \times d}$ can be written as the rescaling of a diagonal matrix,
\begin{align*}
     B_{7,1,2}^{j_1,*,j_1,*} & ~ = c(A,x)_{j_1}\cdot f(A,x)_{j_1}^2 \cdot  \diag(x) \cdot A^{\top} \cdot ((e_{j_1} - f(A,x)) \circ q(A,x)) \cdot {\bf 1}_d^\top\\
     & ~ =  c(A,x)_{j_1}\cdot f(A,x)_{j_1}^2 \cdot  p_{j_1}(A,x) \cdot {\bf 1}_d^\top 
\end{align*}
    where the last step is follows from the Definitions~\ref{def:p}.

{\bf Proof of Part 3.}
We have for diagonal entry and off-diagonal entry can be written as follows 
    \begin{align*}
        B_{7,1,3}^{j_1,i_1,j_1,i_1} = & ~f(A,x)_{j_1}^2 \cdot \langle c(A,x), f(A,x) \rangle \cdot ( (A_{j_1,*}) \circ x^\top  )  \cdot \diag(x) \cdot A^{\top} \cdot ((e_{j_1} - f(A,x)) \circ q(A,x)) \\
        B_{7,1,3}^{j_1,i_1,j_1,i_0} = & ~f(A,x)_{j_1}^2 \cdot \langle c(A,x), f(A,x) \rangle \cdot ( (A_{j_1,*}) \circ x^\top  )  \cdot \diag(x) \cdot A^{\top} \cdot ((e_{j_1} - f(A,x)) \circ q(A,x))
    \end{align*}
From the above equation, we can show that matrix $B_{7,1,3}^{j_1,*,j_1,*}$ can be expressed as a rank-$1$ matrix,
\begin{align*}
     B_{7,1,3}^{j_1,*,j_1,*} & ~ = f(A,x)_{j_1}^2 \cdot \langle c(A,x), f(A,x) \rangle \cdot ( (A_{j_1,*}) \circ x^\top  )  \cdot \diag(x) \cdot A^{\top} \cdot ((e_{j_1} - f(A,x)) \circ q(A,x)) \cdot I_d\\
     & ~ =  f(A,x)_{j_1}^2 \cdot f_c(A,x) \cdot ( (A_{j_1,*}) \circ x^\top  )  \cdot   p_{j_1}(A,x)  \cdot I_d
\end{align*}
    where the last step is follows from the Definitions~\ref{def:f_c} and Definitions~\ref{def:p}.

{\bf Proof of Part 4.}
We have for diagonal entry and off-diagonal entry can be written as follows
    \begin{align*}
        B_{7,1,4}^{j_1,i_1,j_1,i_1}   = & ~  -  f(A,x)_{j_1}^2 \cdot f(A,x)^\top  \cdot A \cdot (\diag(x))^2 \cdot   \langle c(A,x), f(A,x) \rangle \cdot A^{\top} \cdot ((e_{j_1} - f(A,x)) \circ q(A,x))\\
        B_{7,1,4}^{j_1,i_1,j_1,i_0}   = & ~  -  f(A,x)_{j_1}^2 \cdot f(A,x)^\top  \cdot A \cdot (\diag(x))^2 \cdot   \langle c(A,x), f(A,x) \rangle \cdot A^{\top} \cdot ((e_{j_1} - f(A,x)) \circ q(A,x))
    \end{align*}
 From the above equation, we can show that matrix $B_{7,1,4}^{j_1,*,j_1,*}$ can be expressed as a rank-$1$ matrix,
\begin{align*}
    B_{7,1,4}^{j_1,*,j_1,*}  & ~ = -  f(A,x)_{j_1}^2 \cdot \langle c(A,x), f(A,x) \rangle    \cdot f(A,x)^\top  \cdot A \cdot (\diag(x))^2 \cdot  A^{\top}  \cdot ((e_{j_1} - f(A,x)) \circ q(A,x)) \cdot I_d\\
     & ~ =   -  f(A,x)_{j_1}^2 \cdot f_c(A,x)  \cdot h(A,x)^\top \cdot p_{j_1}(A,x) \cdot I_d
\end{align*}
   where the last step is follows from the Definitions~\ref{def:h}, Definitions~\ref{def:f_c} and Definitions~\ref{def:p}.

{\bf Proof of Part 5.}
We have for diagonal entry and off-diagonal entry can be written as follows
    \begin{align*}
         B_{7,1,5}^{j_1,i_1,j_1,i_0} = & ~     f(A,x)_{j_1}^2 \cdot f(A,x)^\top  \cdot A \cdot (\diag(x))^2 \cdot (\langle -f(A,x), f(A,x) \rangle + f(A,x)_{j_1}) \\
        & ~  \cdot A^{\top} \cdot ((e_{j_1} - f(A,x)) \circ q(A,x))  \\
         B_{7,1,5}^{j_1,i_1,j_1,i_0} = & ~    f(A,x)_{j_1}^2 \cdot f(A,x)^\top  \cdot A \cdot (\diag(x))^2 \cdot (\langle -f(A,x), f(A,x) \rangle + f(A,x)_{j_1}) \\
        & ~  \cdot A^{\top} \cdot ((e_{j_1} - f(A,x)) \circ q(A,x)) 
    \end{align*}
    From the above equation, we can show that matrix $B_{7,1,5}^{j_1,*,j_1,*}$ can be expressed as a rank-$1$ matrix,
\begin{align*}
   & ~ B_{7,1,5}^{j_1,*,j_1,*} \\= & ~   f(A,x)_{j_1}^2 \cdot (\langle -f(A,x), f(A,x) \rangle + f(A,x)_{j_1})  \cdot f(A,x)^\top  \cdot A \cdot (\diag(x))^2 \cdot  A^{\top} \cdot ((e_{j_1} - f(A,x)) \circ q(A,x)) \cdot I_d\\
    = & ~     f(A,x)_{j_1}^2 \cdot (-f_2(A,x) + f(A,x)_{j_1})  \cdot h(A,x)^\top \cdot p_{j_1}(A,x) \cdot I_d
\end{align*}
    where the last step is follows from the Definitions~\ref{def:h}, Definitions~\ref{def:f_2} and Definitions~\ref{def:p}.

{\bf Proof of Part 6.}
We have for diagonal entry and off-diagonal entry can be written as follows
    \begin{align*}
        B_{7,1,6}^{j_1,i_1,j_1,i_1}  = & ~    f(A,x)_{j_1}^2 \cdot f(A,x)^\top  \cdot A \cdot  (\diag(x))^2 \cdot(\langle -f(A,x), c(A,x) \rangle + f(A,x)_{j_1}) \\
        & ~ \cdot A^{\top} \cdot ((e_{j_1} - f(A,x)) \circ q(A,x))\\
        B_{7,1,6}^{j_1,i_1,j_1,i_0}  = & ~   f(A,x)_{j_1}^2 \cdot f(A,x)^\top  \cdot A \cdot  (\diag(x))^2 \cdot(\langle -f(A,x), c(A,x) \rangle + f(A,x)_{j_1})  \\
        & ~\cdot A^{\top} \cdot ((e_{j_1} - f(A,x)) \circ q(A,x))
    \end{align*}
    From the above equation, we can show that matrix $B_{7,1,6}^{j_1,*,j_1,*}$ can be expressed as a rank-$1$ matrix,
\begin{align*}
   & ~ B_{7,1,6}^{j_1,*,j_1,*} \\ = & ~     f(A,x)_{j_1}^2  \cdot(\langle -f(A,x), c(A,x) \rangle + f(A,x)_{j_1})   \cdot f(A,x)^\top  \cdot A \cdot (\diag(x))^2 \cdot  A^{\top} \cdot ((e_{j_1} - f(A,x)) \circ q(A,x)) \cdot I_d\\
   =  & ~    f(A,x)_{j_1}^2  \cdot(-f_c(A,x) + f(A,x)_{j_1})  \cdot h(A,x)^\top \cdot  p_{j_1}(A,x) \cdot I_d
\end{align*}
      where the last step is follows from the Definitions~\ref{def:h}, Definitions~\ref{def:f_c} and Definitions~\ref{def:p}.
    
{\bf Proof of Part 7.}
We have for diagonal entry and off-diagonal entry can be written as follows
    \begin{align*}
         B_{7,1,7}^{j_1,i_1,j_1,i_1} = & ~  f(A,x)_{j_1}^2 \cdot ((e_{j_1}^\top - f(A,x)^\top) \circ q(A,x)^\top) \cdot A \cdot  (\diag(x))^2 \cdot A^{\top} \cdot ((e_{j_1} - f(A,x)) \circ q(A,x))\\
         B_{7,1,7}^{j_1,i_1,j_1,i_0} = & ~  f(A,x)_{j_1}^2 \cdot ((e_{j_1}^\top - f(A,x)^\top) \circ q(A,x)^\top) \cdot A \cdot  (\diag(x))^2 \cdot A^{\top} \cdot ((e_{j_1} - f(A,x)) \circ q(A,x))
    \end{align*}
    From the above equation, we can show that matrix $B_{7,1,7}^{j_1,*,j_1,*}$ can be expressed as a rank-$1$ matrix,
\begin{align*}
     B_{7,1,7}^{j_1,*,j_1,*}  & ~ =    f(A,x)_{j_1}^2 \cdot ((e_{j_1}^\top - f(A,x)^\top) \circ q(A,x)^\top) \cdot A \cdot  (\diag(x))^2 \cdot A^{\top} \cdot ((e_{j_1} - f(A,x)) \circ q(A,x)) \cdot I_d\\
     & ~ = f(A,x)_{j_1}^2  \cdot p_{j_1}(A,x)^\top \cdot p_{j_1}(A,x) \cdot I_d
\end{align*}
    where the last step is follows from the Definitions~\ref{def:p}.

    {\bf Proof of Part 8.}
We have for diagonal entry and off-diagonal entry can be written as follows
    \begin{align*}
         B_{7,2,1}^{j_1,i_1,j_1,i_1} = & ~   c_g(A,x)^{\top} \cdot f(A,x)_{j_1}^2 \cdot x_{i_1} \cdot \diag(x) \cdot A^{\top} \cdot ((e_{j_1} - f(A,x)) \circ q(A,x)) \\
         B_{7,2,1}^{j_1,i_1,j_1,i_0} = & ~   c_g(A,x)^{\top} \cdot f(A,x)_{j_1}^2 \cdot x_{i_1} \cdot \diag(x) \cdot A^{\top} \cdot ((e_{j_1} - f(A,x)) \circ q(A,x)) 
    \end{align*}
    From the above equation, we can show that matrix $B_{7,2,1}^{j_1,*,j_1,*}$ can be expressed as a rank-$1$ matrix,
\begin{align*}
     B_{7,2,1}^{j_1,*,j_1,*}  & ~ =   f(A,x)_{j_1}^2  \cdot c_g(A,x)^{\top} \cdot \diag(x) \cdot A^{\top} \cdot ((e_{j_1} - f(A,x)) \circ q(A,x))  \cdot x \cdot {\bf 1}_d^{\top}  \\ 
     & ~ =  f(A,x)_{j_1}^2  \cdot c_g(A,x)^{\top} \cdot p_{j_1}(A,x)\cdot x \cdot {\bf 1}_d^{\top} 
\end{align*}
    where the last step is follows from the Definitions~\ref{def:p}.
    
    {\bf Proof of Part 9.}
We have for diagonal entry and off-diagonal entry can be written as follows
    \begin{align*}
         B_{7,2,2}^{j_1,i_1,j_1,i_1} = & ~   - c_g(A,x)^{\top} \cdot f(A,x)_{j_1} \cdot x_{i_1} \cdot   \diag(x) \cdot A^{\top} \cdot ((e_{j_1} - f(A,x)) \circ q(A,x))\\
         B_{7,2,2}^{j_1,i_1,j_1,i_0} = & ~ - c_g(A,x)^{\top} \cdot f(A,x)_{j_1} \cdot x_{i_1} \cdot   \diag(x) \cdot A^{\top} \cdot ((e_{j_1} - f(A,x)) \circ q(A,x))
    \end{align*}
    From the above equation, we can show that matrix $B_{7,2,2}^{j_1,*,j_1,*}$ can be expressed as a rank-$1$ matrix,
\begin{align*}
     B_{7,2,2}^{j_1,*,j_1,*}  & ~ =  - f(A,x)_{j_1}  \cdot c_g(A,x)^{\top} \cdot \diag(x) \cdot A^{\top} \cdot ((e_{j_1} - f(A,x)) \circ q(A,x))\cdot x \cdot {\bf 1}_d^{\top}  \\ 
     & ~ =  -f(A,x)_{j_1}  \cdot  c_g(A,x)^{\top} \cdot p_{j_1}(A,x) \cdot x \cdot {\bf 1}_d^{\top} 
\end{align*}
    where the last step is follows from the Definitions~\ref{def:p}.

   {\bf Proof of Part 10.}
We have for diagonal entry and off-diagonal entry can be written as follows
    \begin{align*}
         B_{7,3,1}^{j_1,i_1,j_1,i_1} = & ~   -  c_g(A,x)^{\top} \cdot f(A,x)_{j_1} \cdot \diag(x) \cdot e_{i_1} \cdot e_{j_1}^\top \cdot ((e_{j_1} - f(A,x)) \circ q(A,x))\\
         B_{7,3,1}^{j_1,i_1,j_1,i_0} = & ~ -  c_g(A,x)^{\top} \cdot f(A,x)_{j_1} \cdot \diag(x) \cdot e_{i_1} \cdot e_{j_1}^\top \cdot ((e_{j_1} - f(A,x)) \circ q(A,x))
    \end{align*}
    From the above equation, we can show that matrix $B_{7,3,1}^{j_1,*,j_1,*}$ can be expressed as a rank-$1$ matrix,
\begin{align*}
     B_{7,3,1}^{j_1,*,j_1,*}  & ~ = - f(A,x)_{j_1}\cdot e_{j_1}^\top \cdot ((e_{j_1} - f(A,x)) \circ q(A,x)) \cdot {\bf 1}_d \cdot c_g(A,x)^{\top}   \cdot \diag(x)    
\end{align*}

    {\bf Proof of Part 11.}
We have for diagonal entry and off-diagonal entry can be written as follows
    \begin{align*}
         B_{7,4,1}^{j_1,i_1,j_1,i_1} = & ~    c_g(A,x)^{\top} \cdot f(A,x)_{j_1}^2 \cdot   \diag(x) \cdot A^{\top} \cdot x_{i_1}  \cdot (    (e_{j_1}- f(A,x) )  \circ q(A,x)) \\
         B_{7,4,1}^{j_1,i_1,j_1,i_0} = & ~  c_g(A,x)^{\top} \cdot f(A,x)_{j_1}^2 \cdot   \diag(x) \cdot A^{\top} \cdot x_{i_1}  \cdot (    (e_{j_1}- f(A,x) )  \circ q(A,x)) 
    \end{align*}
    From the above equation, we can show that matrix $B_{7,4,1}^{j_1,*,j_1,*}$ can be expressed as a rank-$1$ matrix,
\begin{align*}
     B_{7,4,1}^{j_1,*,j_1,*}  & ~ =  f(A,x)_{j_1}^2  \cdot c_g(A,x)^{\top}  \cdot \diag(x) \cdot A^{\top}  \cdot  (    (e_{j_1}- f(A,x) )  \circ q(A,x))  \cdot x \cdot {\bf 1}_d^{\top}\\ 
     & ~ =    f(A,x)_{j_1}^2 \cdot c_g(A,x)^{\top}  \cdot  p_{j_1}(A,x)\cdot x \cdot {\bf 1}_d^{\top}
\end{align*}
    where the last step is follows from the Definitions~\ref{def:p}.

        {\bf Proof of Part 12.}
We have for diagonal entry and off-diagonal entry can be written as follows
    \begin{align*}
         B_{7,5,1}^{j_1,i_1,j_1,i_1} = & ~ -2 c_g(A,x)^{\top}  \cdot f(A,x)_{j_1}^2 \cdot \diag(x) \cdot A^{\top} \cdot x_{i_1}  \cdot ((e_{j_1} - f(A,x)) \circ  ( (e_{j_1} - f(A,x)) ) ) \\
         B_{7,5,1}^{j_1,i_1,j_1,i_0} = & ~ -2 c_g(A,x)^{\top}  \cdot f(A,x)_{j_1}^2 \cdot \diag(x) \cdot A^{\top} \cdot x_{i_1}  \cdot ((e_{j_1} - f(A,x)) \circ  ( (e_{j_1} - f(A,x)) ) ) 
    \end{align*}
    From the above equation, we can show that matrix $B_{7,5,1}^{j_1,*,j_1,*}$ can be expressed as a rank-$1$ matrix,
\begin{align*}
     B_{7,5,1}^{j_1,*,j_1,*}  & ~ = -2f(A,x)_{j_1}^2 \cdot   c_g(A,x)^{\top} \cdot \diag(x) \cdot A^{\top} \cdot  ((e_{j_1} - f(A,x)) \circ  ( (e_{j_1} - f(A,x)) ) )   \cdot x \cdot {\bf 1}_d^{\top}
\end{align*}
    where the last step is follows from the Definitions~\ref{def:h} and Definitions~\ref{def:f_2}.

\end{proof}
    \subsection{Case \texorpdfstring{$j_1 \neq j_0, i_1 = i_0$}{}}
\begin{lemma}
For $j_1 \neq j_0$ and $i_0 = i_1$. If the following conditions hold
    \begin{itemize}
     \item Let $u(A,x) \in \R^n$ be defined as Definition~\ref{def:u}
    \item Let $\alpha(A,x) \in \R$ be defined as Definition~\ref{def:alpha}
     \item Let $f(A,x) \in \R^n$ be defined as Definition~\ref{def:f}
    \item Let $c(A,x) \in \R^n$ be defined as Definition~\ref{def:c}
    \item Let $g(A,x) \in \R^d$ be defined as Definition~\ref{def:g} 
    \item Let $q(A,x) = c(A,x) + f(A,x) \in \R^n$
    \item Let $c_g(A,x) \in \R^d$ be defined as Definition~\ref{def:c_g}.
    \item Let $L_g(A,x) \in \R$ be defined as Definition~\ref{def:l_g}
    \item Let $v \in \R^n$ be a vector 
    \item Let $B_1^{j_1,i_1,j_0,i_0}$ be defined as Definition~\ref{def:b_1}
    \end{itemize}
    Then, For $j_0,j_1 \in [n], i_0,i_1 \in [d]$, we have 
    \begin{itemize}
\item {\bf Part 1.} For $B_{7,1}^{j_1,i_1,j_0,i_1}$, we have 
\begin{align*}
 B_{7,1}^{j_1,i_1,j_0,i_1}  = & ~ \frac{\d}{\d A_{j_1,i_1}} (- c_g(A,x)^{\top} ) \cdot f(A,x)_{j_0} \cdot \diag(x) \cdot A^{\top} \cdot ((e_{j_0} - f(A,x)) \circ q(A,x))\\
 = & ~ B_{7,1,1}^{j_1,i_1,j_0,i_1} + B_{7,1,2}^{j_1,i_1,j_0,i_1} + B_{7,1,3}^{j_1,i_1,j_0,i_1} + B_{7,1,4}^{j_1,i_1,j_0,i_1} + B_{7,1,5}^{j_1,i_1,j_0,i_1}  
\end{align*} 
\item {\bf Part 2.} For $B_{7,2}^{j_1,i_1,j_0,i_1}$, we have 
\begin{align*}
  B_{7,2}^{j_1,i_1,j_0,i_1} = & ~ - c_g(A,x)^{\top} \cdot \frac{\d}{\d A_{j_1,i_1}} ( f(A,x)_{j_0} )  \cdot \diag(x) \cdot A^{\top} \cdot ((e_{j_0} - f(A,x)) \circ q(A,x)) \\
    = & ~  B_{7,2,1}^{j_1,i_1,j_0,i_1} 
\end{align*} 
\item {\bf Part 3.} For $B_{7,3}^{j_1,i_1,j_0,i_1}$, we have 
\begin{align*}
  B_{7,3}^{j_1,i_1,j_0,i_1} = & ~ - c_g(A,x)^{\top} \cdot f(A,x)_{j_0} \cdot \diag(x) \cdot \frac{\d}{\d A_{j_1,i_1}} (  A^{\top} ) \cdot ((e_{j_0} - f(A,x)) \circ q(A,x))  \\
     = & ~ B_{7,3,1}^{j_1,i_1,j_0,i_1}  
\end{align*} 
\item {\bf Part 4.} For $B_{7,4}^{j_1,i_1,j_0,i_1}$, we have 
\begin{align*}
  B_{7,4}^{j_1,i_1,j_0,i_1} = & ~ - c_g(A,x)^{\top} \cdot f(A,x)_{j_0} \cdot \diag(x) \cdot A^{\top} \cdot ( ( \frac{\d}{\d A_{j_1,i_1}} (e_{j_0} - f(A,x)) ) \circ q(A,x))  \\
     = & ~ B_{7,4,1}^{j_1,i_1,j_0,i_1} 
\end{align*}
\item {\bf Part 5.} For $B_{6,5}^{j_1,i_1,j_0,i_1}$, we have 
\begin{align*}
  B_{7,5}^{j_1,i_1,j_0,i_1} = & ~  - c_g(A,x)^{\top} \cdot f(A,x)_{j_0} \cdot \diag(x) \cdot A^{\top} \cdot ((e_{j_0} - f(A,x)) \circ \frac{\d}{\d A_{j_1,i_1}} ( q(A,x))) \\
     = & ~ B_{7,5,1}^{j_1,i_1,j_0,i_1}     
\end{align*} 
\end{itemize}
\end{lemma}
\begin{proof}
    {\bf Proof of Part 1.}
    \begin{align*}
    B_{7,1,1}^{j_1,i_1,j_0,i_1} : = & ~  e_{i_1}^\top \cdot \langle c(A,x), f(A,x) \rangle \cdot  f(A,x)_{j_1} \cdot f(A,x)_{j_0} \cdot \diag(x) \cdot A^{\top} \cdot ((e_{j_0} - f(A,x)) \circ q(A,x))\\
    B_{7,1,2}^{j_1,i_1,j_0,i_1} : = & ~  e_{i_1}^\top \cdot c(A,x)_{j_1}\cdot  f(A,x)_{j_1} \cdot f(A,x)_{j_0}  \cdot \diag(x) \cdot A^{\top} \cdot ((e_{j_0} - f(A,x)) \circ q(A,x))\\
    B_{7,1,3}^{j_1,i_1,j_0,i_1} : = & ~  f(A,x)_{j_1} \cdot f(A,x)_{j_0}  \cdot \langle c(A,x), f(A,x) \rangle \cdot ( (A_{j_1,*}) \circ x^\top  )  \cdot \diag(x) \cdot A^{\top} \cdot ((e_{j_0} - f(A,x)) \circ q(A,x))\\
    B_{7,1,4}^{j_1,i_1,j_0,i_1} : = & ~  -   f(A,x)_{j_1} \cdot f(A,x)_{j_0}  \cdot f(A,x)^\top  \cdot A \cdot (\diag(x))^2 \cdot   \langle c(A,x), f(A,x) \rangle \cdot A^{\top} \cdot ((e_{j_0} - f(A,x)) \circ q(A,x))\\
    B_{7,1,5}^{j_1,i_1,j_0,i_1} : = & ~      f(A,x)_{j_1} \cdot f(A,x)_{j_0}  \cdot f(A,x)^\top  \cdot A \cdot (\diag(x))^2 \cdot (\langle -f(A,x), f(A,x) \rangle + f(A,x)_{j_1}) \\
    & ~ \cdot A^{\top} \cdot ((e_{j_0} - f(A,x)) \circ q(A,x)) \\
    B_{7,1,6}^{j_1,i_1,j_0,i_1} : = & ~   f(A,x)_{j_1} \cdot f(A,x)_{j_0}  \cdot f(A,x)^\top  \cdot A \cdot  (\diag(x))^2 \cdot(\langle -f(A,x), c(A,x) \rangle + f(A,x)_{j_1}) \\
    & ~\cdot A^{\top} \cdot ((e_{j_0} - f(A,x)) \circ q(A,x)) \\
    B_{7,1,7}^{j_1,i_1,j_0,i_1} : = & ~  f(A,x)_{j_1} \cdot f(A,x)_{j_0}  \cdot ((e_{j_1}^\top - f(A,x)^\top) \circ q(A,x)^\top) \cdot A \cdot  (\diag(x))^2 \cdot A^{\top} \cdot ((e_{j_0} - f(A,x)) \circ q(A,x))
\end{align*}
Finally, combine them and we have
\begin{align*}
       B_{7,1}^{j_1,i_1,j_0,i_1} =  B_{7,1,1}^{j_1,i_1,j_0,i_1} + B_{7,1,2}^{j_1,i_1,j_0,i_1} + B_{7,1,3}^{j_1,i_1,j_0,i_1} + B_{7,1,4}^{j_1,i_1,j_0,i_1} + B_{7,1,5}^{j_1,i_1,j_0,i_1}  
\end{align*}
{\bf Proof of Part 2.}
    \begin{align*}
    B_{7,2,1}^{j_1,i_1,j_0,i_1} : = & ~   c_g(A,x)^{\top} \cdot f(A,x)_{j_1} \cdot f(A,x)_{j_0} \cdot x_{i_1} \cdot \diag(x) \cdot A^{\top} \cdot ((e_{j_0} - f(A,x)) \circ q(A,x))  
\end{align*}
Finally, combine them and we have
\begin{align*}
       B_{7,2}^{j_1,i_1,j_0,i_1} = B_{7,2,1}^{j_1,i_1,j_0,i_1} 
\end{align*}
{\bf Proof of Part 3.} 
    \begin{align*}
    B_{7,3,1}^{j_1,i_1,j_0,i_1} : = & ~   -  c_g(A,x)^{\top} \cdot f(A,x)_{j_0} \cdot \diag(x) \cdot e_{i_1} \cdot e_{j_1}^\top \cdot ((e_{j_0} - f(A,x)) \circ q(A,x))
\end{align*}
Finally, combine them and we have
\begin{align*}
       B_{7,3}^{j_1,i_1,j_0,i_1} = B_{7,3,1}^{j_1,i_1,j_0,i_1} 
\end{align*}
{\bf Proof of Part 4.} 
    \begin{align*}
    B_{7,4,1}^{j_1,i_1,j_0,i_1} : = & ~  c_g(A,x)^{\top} \cdot f(A,x)_{j_1} \cdot f(A,x)_{j_0} \cdot   \diag(x) \cdot A^{\top} \cdot x_{i_1}  \cdot (    (e_{j_1}- f(A,x) )  \circ q(A,x)) 
\end{align*}
Finally, combine them and we have
\begin{align*}
       B_{7,4}^{j_1,i_1,j_0,i_1} = B_{7,4,1}^{j_1,i_1,j_0,i_1}  
\end{align*}
{\bf Proof of Part 5.} 
    \begin{align*}
    B_{7,5,1}^{j_1,i_1,j_0,i_1} : = & ~-2 c_g(A,x)^{\top}  \cdot f(A,x)_{j_1} \cdot f(A,x)_{j_0} \cdot \diag(x) \cdot A^{\top} \cdot x_{i_1}  \cdot ((e_{j_0} - f(A,x)) \circ  ( (e_{j_1} - f(A,x)) ) ) 
\end{align*}
Finally, combine them and we have
\begin{align*}
       B_{7,5}^{j_1,i_1,j_0,i_1} = B_{7,5,1}^{j_1,i_1,j_0,i_1}   
\end{align*}
\end{proof}
    \subsection{Case \texorpdfstring{$j_1 \neq j_0, i_1 \neq i_0$}{}}
\begin{lemma}
For $j_1 \neq j_0$ and $i_0 \neq i_1$. If the following conditions hold
    \begin{itemize}
     \item Let $u(A,x) \in \R^n$ be defined as Definition~\ref{def:u}
    \item Let $\alpha(A,x) \in \R$ be defined as Definition~\ref{def:alpha}
     \item Let $f(A,x) \in \R^n$ be defined as Definition~\ref{def:f}
    \item Let $c(A,x) \in \R^n$ be defined as Definition~\ref{def:c}
    \item Let $g(A,x) \in \R^d$ be defined as Definition~\ref{def:g} 
    \item Let $q(A,x) = c(A,x) + f(A,x) \in \R^n$
    \item Let $c_g(A,x) \in \R^d$ be defined as Definition~\ref{def:c_g}.
    \item Let $L_g(A,x) \in \R$ be defined as Definition~\ref{def:l_g}
    \item Let $v \in \R^n$ be a vector 
    \item Let $B_1^{j_1,i_1,j_0,i_0}$ be defined as Definition~\ref{def:b_1}
    \end{itemize}
    Then, For $j_0,j_1 \in [n], i_0,i_1 \in [d]$, we have 
    \begin{itemize}
\item {\bf Part 1.} For $B_{7,1}^{j_1,i_1,j_0,i_0}$, we have 
\begin{align*}
 B_{7,1}^{j_1,i_1,j_0,i_0}  = & ~ \frac{\d}{\d A_{j_1,i_1}} (- c_g(A,x)^{\top} ) \cdot f(A,x)_{j_0} \cdot \diag(x) \cdot A^{\top} \cdot ((e_{j_0} - f(A,x)) \circ q(A,x))\\
 = & ~ B_{7,1,1}^{j_1,i_1,j_0,i_0} + B_{7,1,2}^{j_1,i_1,j_0,i_0} + B_{7,1,3}^{j_1,i_1,j_0,i_0} + B_{7,1,4}^{j_1,i_1,j_0,i_0} + B_{7,1,5}^{j_1,i_1,j_0,i_0}  
\end{align*} 
\item {\bf Part 2.} For $B_{7,2}^{j_1,i_1,j_0,i_0}$, we have 
\begin{align*}
  B_{7,2}^{j_1,i_1,j_0,i_0} = & ~ - c_g(A,x)^{\top} \cdot \frac{\d}{\d A_{j_1,i_1}} ( f(A,x)_{j_0} )  \cdot \diag(x) \cdot A^{\top} \cdot ((e_{j_0} - f(A,x)) \circ q(A,x)) \\
    = & ~  B_{7,2,1}^{j_1,i_1,j_0,i_0} 
\end{align*} 
\item {\bf Part 3.} For $B_{7,3}^{j_1,i_1,j_0,i_0}$, we have 
\begin{align*}
  B_{7,3}^{j_1,i_1,j_0,i_0} = & ~ - c_g(A,x)^{\top} \cdot f(A,x)_{j_0} \cdot \diag(x) \cdot \frac{\d}{\d A_{j_1,i_1}} (  A^{\top} ) \cdot ((e_{j_0} - f(A,x)) \circ q(A,x))  \\
     = & ~ B_{7,3,1}^{j_1,i_1,j_0,i_0}  
\end{align*} 
\item {\bf Part 4.} For $B_{7,4}^{j_1,i_1,j_0,i_0}$, we have 
\begin{align*}
  B_{7,4}^{j_1,i_1,j_0,i_0} = & ~ - c_g(A,x)^{\top} \cdot f(A,x)_{j_0} \cdot \diag(x) \cdot A^{\top} \cdot ( ( \frac{\d}{\d A_{j_1,i_1}} (e_{j_0} - f(A,x)) ) \circ q(A,x))  \\
     = & ~ B_{7,4,1}^{j_1,i_1,j_0,i_0} 
\end{align*}
\item {\bf Part 5.} For $B_{6,5}^{j_1,i_1,j_0,i_0}$, we have 
\begin{align*}
  B_{7,5}^{j_1,i_1,j_0,i_0} = & ~  - c_g(A,x)^{\top} \cdot f(A,x)_{j_0} \cdot \diag(x) \cdot A^{\top} \cdot ((e_{j_0} - f(A,x)) \circ \frac{\d}{\d A_{j_1,i_1}} ( q(A,x))) \\
     = & ~ B_{7,5,1}^{j_1,i_1,j_0,i_0}     
\end{align*} 
\end{itemize}
\end{lemma}
\begin{proof}
    {\bf Proof of Part 1.}
    \begin{align*}
    B_{7,1,1}^{j_1,i_1,j_0,i_0} : = & ~  e_{i_1}^\top \cdot \langle c(A,x), f(A,x) \rangle \cdot  f(A,x)_{j_1} \cdot f(A,x)_{j_0} \cdot \diag(x) \cdot A^{\top} \cdot ((e_{j_0} - f(A,x)) \circ q(A,x))\\
    B_{7,1,2}^{j_1,i_1,j_0,i_0} : = & ~  e_{i_1}^\top \cdot c(A,x)_{j_1}\cdot  f(A,x)_{j_1} \cdot f(A,x)_{j_0}  \cdot \diag(x) \cdot A^{\top} \cdot ((e_{j_0} - f(A,x)) \circ q(A,x))\\
    B_{7,1,3}^{j_1,i_1,j_0,i_0} : = & ~  f(A,x)_{j_1} \cdot f(A,x)_{j_0}  \cdot \langle c(A,x), f(A,x) \rangle \cdot ( (A_{j_1,*}) \circ x^\top  )  \cdot \diag(x) \cdot A^{\top} \cdot ((e_{j_0} - f(A,x)) \circ q(A,x))\\
    B_{7,1,4}^{j_1,i_1,j_0,i_0} : = & ~  -   f(A,x)_{j_1} \cdot f(A,x)_{j_0}  \cdot f(A,x)^\top  \cdot A \cdot (\diag(x))^2 \cdot   \langle c(A,x), f(A,x) \rangle \cdot A^{\top} \cdot ((e_{j_0} - f(A,x)) \circ q(A,x))\\
    B_{7,1,5}^{j_1,i_1,j_0,i_0} : = & ~      f(A,x)_{j_1} \cdot f(A,x)_{j_0}  \cdot f(A,x)^\top  \cdot A \cdot (\diag(x))^2 \cdot (\langle -f(A,x), f(A,x) \rangle + f(A,x)_{j_1}) \\
    & ~\cdot A^{\top} \cdot ((e_{j_0} - f(A,x)) \circ q(A,x)) \\
    B_{7,1,6}^{j_1,i_1,j_0,i_0} : = & ~   f(A,x)_{j_1} \cdot f(A,x)_{j_0}  \cdot f(A,x)^\top  \cdot A \cdot  (\diag(x))^2 \cdot(\langle -f(A,x), c(A,x) \rangle + f(A,x)_{j_1}) \\
    & ~\cdot A^{\top} \cdot ((e_{j_0} - f(A,x)) \circ q(A,x)) \\
    B_{7,1,7}^{j_1,i_1,j_0,i_0} : = & ~  f(A,x)_{j_1} \cdot f(A,x)_{j_0}  \cdot ((e_{j_1}^\top - f(A,x)^\top) \circ q(A,x)^\top) \cdot A \cdot  (\diag(x))^2 \cdot A^{\top} \cdot ((e_{j_0} - f(A,x)) \circ q(A,x))
\end{align*}
Finally, combine them and we have
\begin{align*}
       B_{7,1}^{j_1,i_1,j_0,i_0} =  B_{7,1,1}^{j_1,i_1,j_0,i_0} + B_{7,1,2}^{j_1,i_1,j_0,i_0} + B_{7,1,3}^{j_1,i_1,j_0,i_0} + B_{7,1,4}^{j_1,i_1,j_0,i_0} + B_{7,1,5}^{j_1,i_1,j_0,i_0}  
\end{align*}
{\bf Proof of Part 2.}
    \begin{align*}
    B_{7,2,1}^{j_1,i_1,j_0,i_0} : = & ~   c_g(A,x)^{\top} \cdot f(A,x)_{j_1} \cdot f(A,x)_{j_0} \cdot x_{i_1} \cdot \diag(x) \cdot A^{\top} \cdot ((e_{j_0} - f(A,x)) \circ q(A,x))  
\end{align*}
Finally, combine them and we have
\begin{align*}
       B_{7,2}^{j_1,i_1,j_0,i_0} = B_{7,2,1}^{j_1,i_1,j_0,i_0} 
\end{align*}
{\bf Proof of Part 3.} 
    \begin{align*}
    B_{7,3,1}^{j_1,i_1,j_0,i_0} : = & ~   -  c_g(A,x)^{\top} \cdot f(A,x)_{j_0} \cdot \diag(x) \cdot e_{i_1} \cdot e_{j_1}^\top \cdot ((e_{j_0} - f(A,x)) \circ q(A,x))
\end{align*}
Finally, combine them and we have
\begin{align*}
       B_{7,3}^{j_1,i_1,j_0,i_0} = B_{7,3,1}^{j_1,i_1,j_0,i_0} 
\end{align*}
{\bf Proof of Part 4.} 
    \begin{align*}
    B_{7,4,1}^{j_1,i_1,j_0,i_0} : = & ~  c_g(A,x)^{\top} \cdot f(A,x)_{j_1} \cdot f(A,x)_{j_0} \cdot   \diag(x) \cdot A^{\top} \cdot x_{i_1}  \cdot (    (e_{j_1}- f(A,x) )  \circ q(A,x)) 
\end{align*}
Finally, combine them and we have
\begin{align*}
       B_{7,4}^{j_1,i_1,j_0,i_0} = B_{7,4,1}^{j_1,i_1,j_0,i_0}  
\end{align*}
{\bf Proof of Part 5.} 
    \begin{align*}
    B_{7,5,1}^{j_1,i_1,j_0,i_0} : = & ~-2 c_g(A,x)^{\top}  \cdot f(A,x)_{j_1} \cdot f(A,x)_{j_0} \cdot \diag(x) \cdot A^{\top} \cdot x_{i_1}  \cdot ((e_{j_0} - f(A,x)) \circ  ( (e_{j_1} - f(A,x)) ) ) 
\end{align*}
Finally, combine them and we have
\begin{align*}
       B_{7,5}^{j_1,i_1,j_0,i_0} = B_{7,5,1}^{j_1,i_1,j_0,i_0}   
\end{align*}
\end{proof}

\subsection{Constructing \texorpdfstring{$d \times d$}{} matrices for \texorpdfstring{$j_1 \neq j_0$}{}}
The purpose of the following lemma is to let $i_0$ and $i_1$ disappear.
\begin{lemma}For $j_0,j_1 \in [n]$, a list of $d \times d$ matrices can be expressed as the following sense,\label{lem:b_7_j1_j0}
\begin{itemize}
\item {\bf Part 1.}
\begin{align*}
B_{7,1,1}^{j_1,*,j_0,*} & ~ =    f_c(A,x) \cdot  f(A,x)_{j_1} \cdot f(A,x)_{j_0} \cdot  p_{j_0}(A,x) \cdot {\bf 1}_d^\top
\end{align*}
\item {\bf Part 2.}
\begin{align*}
B_{7,1,2}^{j_1,*,j_0,*} & ~ =      c(A,x)_{j_1}\cdot f(A,x)_{j_1} \cdot f(A,x)_{j_0} \cdot  p_{j_0}(A,x) \cdot {\bf 1}_d^\top 
\end{align*}
\item {\bf Part 3.}
\begin{align*}
B_{7,1,3}^{j_1,*,j_0,*} & ~ =   f(A,x)_{j_1} \cdot f(A,x)_{j_0} \cdot f_c(A,x) \cdot ( (A_{j_1,*}) \circ x^\top  )  \cdot   p_{j_0}(A,x)  \cdot I_d
\end{align*}
\item {\bf Part 4.}
\begin{align*}
B_{7,1,4}^{j_1,*,j_0,*}  & ~ =    -  f(A,x)_{j_1} \cdot f(A,x)_{j_0} \cdot f_c(A,x)  \cdot h(A,x)^\top \cdot p_{j_0}(A,x) \cdot I_d
\end{align*}
\item {\bf Part 5.}
\begin{align*}
B_{7,1,5}^{j_1,*,j_0,*}  & ~ =    f(A,x)_{j_1} \cdot f(A,x)_{j_0} \cdot (-f_2(A,x) + f(A,x)_{j_1})  \cdot h(A,x)^\top \cdot p_{j_0}(A,x) \cdot I_d
\end{align*}
\item {\bf Part 6.}
\begin{align*}
B_{7,1,6}^{j_1,*,j_0,*}  & ~ =    f(A,x)_{j_1} \cdot f(A,x)_{j_0}  \cdot(-f_c(A,x) + f(A,x)_{j_1})  \cdot h(A,x)^\top \cdot  p_{j_0}(A,x) \cdot I_d
\end{align*}
\item {\bf Part 7.}
\begin{align*}
B_{7,1,7}^{j_1,*,j_0,*}  & ~ =   f(A,x)_{j_1} \cdot f(A,x)_{j_0} \cdot p_{j_1}(A,x)^\top \cdot p_{j_0}(A,x) \cdot I_d
\end{align*}
\item {\bf Part 8.}
\begin{align*}
B_{7,2,1}^{j_1,*,j_0,*}  & ~ =    f(A,x)_{j_1} \cdot f(A,x)_{j_0}  \cdot c_g(A,x)^{\top} \cdot p_{j_0}(A,x)\cdot x \cdot {\bf 1}_d^{\top} 
\end{align*}
\item {\bf Part 9.}
\begin{align*}
 B_{7,3,1}^{j_1,*,j_0,*}  & ~ =     - f(A,x)_{j_0}\cdot e_{j_1}^\top \cdot ((e_{j_0} - f(A,x)) \circ q(A,x)) \cdot {\bf 1}_d \cdot c_g(A,x)^{\top}   \cdot \diag(x)   
\end{align*}
\item {\bf Part 10.}
\begin{align*}
B_{7,4,1}^{j_1,*,j_0,*}  =     f(A,x)_{j_1} \cdot f(A,x)_{j_0} \cdot c_g(A,x)^{\top}  \cdot  p_{j_1}(A,x)\cdot x \cdot {\bf 1}_d^{\top}
\end{align*}
\item {\bf Part 11.}
\begin{align*}
B_{7,5,1}^{j_1,*,j_0,*}  & ~ =  -2f(A,x)_{j_1} \cdot f(A,x)_{j_0} \cdot   c_g(A,x)^{\top} \cdot \diag(x) \cdot A^{\top} \cdot  ((e_{j_0} - f(A,x)) \circ  ( (e_{j_1} - f(A,x)) ) )   \cdot x \cdot {\bf 1}_d^{\top}
\end{align*} 

\end{itemize}
\end{lemma}
\begin{proof}
{\bf Proof of Part 1.}
    We have
    \begin{align*}
        B_{7,1,1}^{j_1,i_1,j_0,i_1}  = & ~ e_{i_1}^\top \cdot \langle c(A,x), f(A,x) \rangle \cdot  f(A,x)_{j_1} \cdot f(A,x)_{j_0} \cdot \diag(x) \cdot A^{\top} \cdot ((e_{j_0} - f(A,x)) \circ q(A,x))\\
        B_{7,1,1}^{j_1,i_1,j_0,i_0}   = & ~  e_{i_1}^\top \cdot \langle c(A,x), f(A,x) \rangle \cdot  f(A,x)_{j_1} \cdot f(A,x)_{j_0} \cdot \diag(x) \cdot A^{\top} \cdot ((e_{j_0} - f(A,x)) \circ q(A,x))
    \end{align*}
    From the above two equations, we can tell that $B_{7,1,1}^{j_1,*,j_0,*} \in \R^{d \times d}$ is a matrix that both the diagonal and off-diagonal have entries.
    
    Then we have $B_{7,1,1}^{j_1,*,j_0,*} \in \R^{d \times d}$ can be written as the rescaling of a diagonal matrix,
    \begin{align*}
     B_{7,1,1}^{j_1,*,j_0,*} & ~ = \langle c(A,x), f(A,x) \rangle \cdot  f(A,x)_{j_1} \cdot f(A,x)_{j_0} \cdot \diag(x) \cdot A^{\top} \cdot ((e_{j_0} - f(A,x)) \circ q(A,x)) \cdot {\bf 1}_d^\top\\
     & ~ = f_c(A,x) \cdot  f(A,x)_{j_1} \cdot f(A,x)_{j_0} \cdot  p_{j_0}(A,x) \cdot {\bf 1}_d^\top
\end{align*}
    where the last step is follows from the Definitions~\ref{def:f_c} and Definitions~\ref{def:p}.

{\bf Proof of Part 2.}
    We have
    \begin{align*}
        B_{7,1,2}^{j_1,i_1,j_0,i_1} = & ~ e_{i_1}^\top \cdot c(A,x)_{j_1}\cdot f(A,x)_{j_1} \cdot f(A,x)_{j_0} \cdot \diag(x) \cdot A^{\top} \cdot ((e_{j_0} - f(A,x)) \circ q(A,x))\\
        B_{7,1,2}^{j_1,i_1,j_0,i_0} = & ~ e_{i_1}^\top \cdot c(A,x)_{j_1}\cdot f(A,x)_{j_1} \cdot f(A,x)_{j_0} \cdot \diag(x) \cdot A^{\top} \cdot ((e_{j_0} - f(A,x)) \circ q(A,x))
    \end{align*}
     From the above two equations, we can tell that $B_{7,1,2}^{j_1,*,j_0,*} \in \R^{d \times d}$ is a matrix that only diagonal has entries and off-diagonal are all zeros.
    
    Then we have $B_{7,1,2}^{j_1,*,j_0,*} \in \R^{d \times d}$ can be written as the rescaling of a diagonal matrix,
\begin{align*}
     B_{7,1,2}^{j_1,*,j_0,*} & ~ = c(A,x)_{j_1}\cdot f(A,x)_{j_1} \cdot f(A,x)_{j_0} \cdot  \diag(x) \cdot A^{\top} \cdot ((e_{j_0} - f(A,x)) \circ q(A,x)) \cdot {\bf 1}_d^\top\\
     & ~ =  c(A,x)_{j_1}\cdot f(A,x)_{j_1} \cdot f(A,x)_{j_0} \cdot  p_{j_0}(A,x) \cdot {\bf 1}_d^\top 
\end{align*}
    where the last step is follows from the Definitions~\ref{def:p}.

{\bf Proof of Part 3.}
We have for diagonal entry and off-diagonal entry can be written as follows 
    \begin{align*}
        B_{7,1,3}^{j_1,i_1,j_0,i_1} = & ~f(A,x)_{j_1} \cdot f(A,x)_{j_0} \cdot \langle c(A,x), f(A,x) \rangle \cdot ( (A_{j_1,*}) \circ x^\top  )  \cdot \diag(x) \cdot A^{\top} \cdot ((e_{j_0} - f(A,x)) \circ q(A,x)) \\
        B_{7,1,3}^{j_1,i_1,j_0,i_0} = & ~f(A,x)_{j_1} \cdot f(A,x)_{j_0} \cdot \langle c(A,x), f(A,x) \rangle \cdot ( (A_{j_1,*}) \circ x^\top  )  \cdot \diag(x) \cdot A^{\top} \cdot ((e_{j_0} - f(A,x)) \circ q(A,x))
    \end{align*}
From the above equation, we can show that matrix $B_{7,1,3}^{j_1,*,j_0,*}$ can be expressed as a rank-$1$ matrix,
\begin{align*}
     B_{7,1,3}^{j_1,*,j_0,*} & ~ = f(A,x)_{j_1} \cdot f(A,x)_{j_0} \cdot \langle c(A,x), f(A,x) \rangle \cdot ( (A_{j_1,*}) \circ x^\top  )  \cdot \diag(x)  \\
        & ~\cdot A^{\top} \cdot ((e_{j_0} - f(A,x)) \circ q(A,x)) \cdot I_d\\
     & ~ =  f(A,x)_{j_1} \cdot f(A,x)_{j_0} \cdot f_c(A,x) \cdot ( (A_{j_1,*}) \circ x^\top  )  \cdot   p_{j_0}(A,x)  \cdot I_d
\end{align*}
    where the last step is follows from the Definitions~\ref{def:f_c} and Definitions~\ref{def:p}.

{\bf Proof of Part 4.}
We have for diagonal entry and off-diagonal entry can be written as follows
    \begin{align*}
        B_{7,1,4}^{j_1,i_1,j_0,i_1}   = & ~  -  f(A,x)_{j_1} \cdot f(A,x)_{j_0} \cdot f(A,x)^\top  \cdot A \cdot (\diag(x))^2 \cdot   \langle c(A,x), f(A,x) \rangle  \\
        & ~\cdot A^{\top} \cdot ((e_{j_0} - f(A,x)) \circ q(A,x))\\
        B_{7,1,4}^{j_1,i_1,j_0,i_0}   = & ~  -  f(A,x)_{j_1} \cdot f(A,x)_{j_0} \cdot f(A,x)^\top  \cdot A \cdot (\diag(x))^2 \cdot   \langle c(A,x), f(A,x) \rangle  \\
        & ~\cdot A^{\top} \cdot ((e_{j_0} - f(A,x)) \circ q(A,x))
    \end{align*}
 From the above equation, we can show that matrix $B_{7,1,4}^{j_1,*,j_0,*}$ can be expressed as a rank-$1$ matrix,
\begin{align*}
    B_{7,1,4}^{j_1,*,j_1,*}  & ~ = -  f(A,x)_{j_1} \cdot f(A,x)_{j_0} \cdot \langle c(A,x), f(A,x) \rangle    \cdot f(A,x)^\top  \cdot A \cdot (\diag(x))^2  \\
        & ~\cdot  A^{\top}  \cdot ((e_{j_0} - f(A,x)) \circ q(A,x)) \cdot I_d\\
     & ~ =   -  f(A,x)_{j_1} \cdot f(A,x)_{j_0} \cdot f_c(A,x)  \cdot h(A,x)^\top \cdot p_{j_0}(A,x) \cdot I_d
\end{align*}
   where the last step is follows from the Definitions~\ref{def:h}, Definitions~\ref{def:f_c} and Definitions~\ref{def:p}.

{\bf Proof of Part 5.}
We have for diagonal entry and off-diagonal entry can be written as follows
    \begin{align*}
         B_{7,1,5}^{j_1,i_1,j_0,i_0} = & ~     f(A,x)_{j_1} \cdot f(A,x)_{j_0} \cdot f(A,x)^\top  \cdot A \cdot (\diag(x))^2 \cdot (\langle -f(A,x), f(A,x) \rangle + f(A,x)_{j_1}) \\
         & ~\cdot A^{\top} \cdot ((e_{j_0} - f(A,x)) \circ q(A,x))  \\
         B_{7,1,5}^{j_1,i_1,j_0,i_0} = & ~    f(A,x)_{j_1} \cdot f(A,x)_{j_0} \cdot f(A,x)^\top  \cdot A \cdot (\diag(x))^2 \cdot (\langle -f(A,x), f(A,x) \rangle + f(A,x)_{j_1})  \\
         & ~\cdot A^{\top} \cdot ((e_{j_0} - f(A,x)) \circ q(A,x)) 
    \end{align*}
    From the above equation, we can show that matrix $B_{7,1,5}^{j_1,*,j_0,*}$ can be expressed as a rank-$1$ matrix,
\begin{align*}
    B_{7,1,5}^{j_1,*,j_0,*}  & ~ =  f(A,x)_{j_1} \cdot f(A,x)_{j_0} \cdot (\langle -f(A,x), f(A,x) \rangle + f(A,x)_{j_1})  \cdot f(A,x)^\top  \cdot A \cdot (\diag(x))^2 \\
    & ~\cdot  A^{\top} \cdot ((e_{j_0} - f(A,x)) \circ q(A,x)) \cdot I_d\\
     & ~ =    f(A,x)_{j_1} \cdot f(A,x)_{j_0} \cdot (-f_2(A,x) + f(A,x)_{j_1})  \cdot h(A,x)^\top \cdot p_{j_0}(A,x) \cdot I_d
\end{align*}
    where the last step is follows from the Definitions~\ref{def:h}, Definitions~\ref{def:f_2} and Definitions~\ref{def:p}.

{\bf Proof of Part 6.}
We have for diagonal entry and off-diagonal entry can be written as follows
    \begin{align*}
        B_{7,1,6}^{j_1,i_1,j_0,i_1}  = & ~    f(A,x)_{j_1} \cdot f(A,x)_{j_0} \cdot f(A,x)^\top  \cdot A \cdot  (\diag(x))^2 \cdot(\langle -f(A,x), c(A,x) \rangle + f(A,x)_{j_1}) \\
    & ~\cdot A^{\top} \cdot ((e_{j_0} - f(A,x)) \circ q(A,x))\\
        B_{7,1,6}^{j_1,i_1,j_0,i_0}  = & ~   f(A,x)_{j_1} \cdot f(A,x)_{j_0} \cdot f(A,x)^\top  \cdot A \cdot  (\diag(x))^2 \cdot(\langle -f(A,x), c(A,x) \rangle + f(A,x)_{j_1}) \\
    & ~\cdot A^{\top} \cdot ((e_{j_0} - f(A,x)) \circ q(A,x))
    \end{align*}
    From the above equation, we can show that matrix $B_{7,1,6}^{j_1,*,j_0,*}$ can be expressed as a rank-$1$ matrix,
\begin{align*}
    B_{7,1,6}^{j_1,*,j_0,*}  & ~ =    f(A,x)_{j_1} \cdot f(A,x)_{j_0}  \cdot(\langle -f(A,x), c(A,x) \rangle + f(A,x)_{j_1})   \cdot f(A,x)^\top  \cdot A \cdot (\diag(x))^2 \\
    & ~\cdot  A^{\top} \cdot ((e_{j_0} - f(A,x)) \circ q(A,x)) \cdot I_d\\
     & ~ =  f(A,x)_{j_1} \cdot f(A,x)_{j_0} \cdot(-f_c(A,x) + f(A,x)_{j_1})  \cdot h(A,x)^\top \cdot  p_{j_0}(A,x) \cdot I_d
\end{align*}
      where the last step is follows from the Definitions~\ref{def:h}, Definitions~\ref{def:f_c} and Definitions~\ref{def:p}.
    
{\bf Proof of Part 7.}
We have for diagonal entry and off-diagonal entry can be written as follows
    \begin{align*}
         B_{7,1,7}^{j_1,i_1,j_0,i_1} = & ~  f(A,x)_{j_1} \cdot f(A,x)_{j_0}\cdot ((e_{j_1}^\top - f(A,x)^\top) \circ q(A,x)^\top) \cdot A \cdot  (\diag(x))^2  \\
        & ~\cdot A^{\top} \cdot ((e_{j_0} - f(A,x)) \circ q(A,x))\\
         B_{7,1,7}^{j_1,i_1,j_0,i_0} = & ~  f(A,x)_{j_1} \cdot f(A,x)_{j_0} \cdot ((e_{j_1}^\top - f(A,x)^\top) \circ q(A,x)^\top) \cdot A \cdot  (\diag(x))^2  \\
        & ~\cdot A^{\top} \cdot ((e_{j_0} - f(A,x)) \circ q(A,x))
    \end{align*}
    From the above equation, we can show that matrix $B_{7,1,7}^{j_1,*,j_0,*}$ can be expressed as a rank-$1$ matrix,
\begin{align*}
     B_{7,1,7}^{j_1,*,j_0,*}  & ~ =   f(A,x)_{j_1} \cdot f(A,x)_{j_0} \cdot ((e_{j_1}^\top - f(A,x)^\top) \circ q(A,x)^\top) \cdot A \cdot  (\diag(x))^2  \\
        & ~\cdot A^{\top} \cdot ((e_{j_0} - f(A,x)) \circ q(A,x)) \cdot I_d\\
     & ~ = f(A,x)_{j_1} \cdot f(A,x)_{j_0}  \cdot p_{j_1}(A,x)^\top \cdot p_{j_0}(A,x) \cdot I_d
\end{align*}
    where the last step is follows from the Definitions~\ref{def:p}.

    {\bf Proof of Part 8.}
We have for diagonal entry and off-diagonal entry can be written as follows
    \begin{align*}
         B_{7,2,1}^{j_1,i_1,j_0,i_1} = & ~   c_g(A,x)^{\top} \cdot f(A,x)_{j_1} \cdot f(A,x)_{j_0} \cdot x_{i_1} \cdot \diag(x) \cdot A^{\top} \cdot ((e_{j_0} - f(A,x)) \circ q(A,x)) \\
         B_{7,2,1}^{j_1,i_1,j_0,i_0} = & ~   c_g(A,x)^{\top} \cdot f(A,x)_{j_1} \cdot f(A,x)_{j_0} \cdot x_{i_1} \cdot \diag(x) \cdot A^{\top} \cdot ((e_{j_0} - f(A,x)) \circ q(A,x)) 
    \end{align*}
    From the above equation, we can show that matrix $B_{7,2,1}^{j_1,*,j_1,*}$ can be expressed as a rank-$1$ matrix,
\begin{align*}
     B_{7,2,1}^{j_1,*,j_0,*}  & ~ =  f(A,x)_{j_1} \cdot f(A,x)_{j_0}  \cdot c_g(A,x)^{\top} \cdot \diag(x) \cdot A^{\top} \cdot ((e_{j_0} - f(A,x)) \circ q(A,x))  \cdot x \cdot {\bf 1}_d^{\top}  \\ 
     & ~ =  f(A,x)_{j_1} \cdot f(A,x)_{j_0}  \cdot c_g(A,x)^{\top} \cdot p_{j_0}(A,x)\cdot x \cdot {\bf 1}_d^{\top} 
\end{align*}
    where the last step is follows from the Definitions~\ref{def:p}.

   {\bf Proof of Part 9.}
We have for diagonal entry and off-diagonal entry can be written as follows
    \begin{align*}
         B_{7,3,1}^{j_1,i_1,j_0,i_1} = & ~   -  c_g(A,x)^{\top} \cdot f(A,x)_{j_0} \cdot \diag(x) \cdot e_{i_1} \cdot e_{j_1}^\top \cdot ((e_{j_0} - f(A,x)) \circ q(A,x))\\
         B_{7,3,1}^{j_1,i_1,j_0,i_0} = & ~ -  c_g(A,x)^{\top} \cdot f(A,x)_{j_0} \cdot \diag(x) \cdot e_{i_1} \cdot e_{j_1}^\top \cdot ((e_{j_0} - f(A,x)) \circ q(A,x))
    \end{align*}
    From the above equation, we can show that matrix $B_{7,3,1}^{j_1,*,j_0,*}$ can be expressed as a rank-$1$ matrix,
\begin{align*}
     B_{7,3,1}^{j_1,*,j_0,*}  & ~ = - f(A,x)_{j_0}\cdot e_{j_1}^\top \cdot ((e_{j_0} - f(A,x)) \circ q(A,x)) \cdot {\bf 1}_d \cdot c_g(A,x)^{\top}   \cdot \diag(x)    
\end{align*}

    {\bf Proof of Part 10.}
We have for diagonal entry and off-diagonal entry can be written as follows
    \begin{align*}
         B_{7,4,1}^{j_1,i_1,j_0,i_1} = & ~    c_g(A,x)^{\top} \cdot f(A,x)_{j_1} \cdot f(A,x)_{j_0} \cdot   \diag(x) \cdot A^{\top} \cdot x_{i_1}  \cdot (    (e_{j_1}- f(A,x) )  \circ q(A,x)) \\
         B_{7,4,1}^{j_1,i_1,j_0,i_0} = & ~  c_g(A,x)^{\top} \cdot f(A,x)_{j_1} \cdot f(A,x)_{j_0} \cdot   \diag(x) \cdot A^{\top} \cdot x_{i_1}  \cdot (    (e_{j_1}- f(A,x) )  \circ q(A,x)) 
    \end{align*}
    From the above equation, we can show that matrix $B_{7,4,1}^{j_1,*,j_0,*}$ can be expressed as a rank-$1$ matrix,
\begin{align*}
     B_{7,4,1}^{j_1,*,j_0,*}  & ~ =  f(A,x)_{j_1} \cdot f(A,x)_{j_0}  \cdot c_g(A,x)^{\top}  \cdot \diag(x) \cdot A^{\top}  \cdot  (    (e_{j_1}- f(A,x) )  \circ q(A,x))  \cdot x \cdot {\bf 1}_d^{\top}\\ 
     & ~ =    f(A,x)_{j_1} \cdot f(A,x)_{j_0} \cdot c_g(A,x)^{\top}  \cdot  p_{j_1}(A,x)\cdot x \cdot {\bf 1}_d^{\top}
\end{align*}
    where the last step is follows from the Definitions~\ref{def:p}.

        {\bf Proof of Part 11.}
We have for diagonal entry and off-diagonal entry can be written as follows
    \begin{align*}
         B_{7,5,1}^{j_1,i_1,j_0,i_1} = & ~ -2 c_g(A,x)^{\top}  \cdot f(A,x)_{j_1} \cdot f(A,x)_{j_0} \cdot \diag(x) \cdot A^{\top} \cdot x_{i_1}  \cdot ((e_{j_0} - f(A,x)) \circ  ( (e_{j_1} - f(A,x)) ) ) \\
         B_{7,5,1}^{j_1,i_1,j_0,i_0} = & ~ -2 c_g(A,x)^{\top}  \cdot f(A,x)_{j_1} \cdot f(A,x)_{j_0} \cdot \diag(x) \cdot A^{\top} \cdot x_{i_1}  \cdot ((e_{j_0} - f(A,x)) \circ  ( (e_{j_1} - f(A,x)) ) ) 
    \end{align*}
    From the above equation, we can show that matrix $B_{7,5,1}^{j_1,*,j_0,*}$ can be expressed as a rank-$1$ matrix,
\begin{align*}
     B_{7,5,1}^{j_1,*,j_0,*}  & ~ = -2f(A,x)_{j_1} \cdot f(A,x)_{j_0} \cdot   c_g(A,x)^{\top} \cdot \diag(x) \cdot A^{\top} \cdot  ((e_{j_0} - f(A,x)) \circ  ( (e_{j_1} - f(A,x)) ) )   \cdot x \cdot {\bf 1}_d^{\top}
\end{align*}
    where the last step is follows from the Definitions~\ref{def:h} and Definitions~\ref{def:f_2}.
\end{proof}
\subsection{Expanding \texorpdfstring{$B_7$}{} into many terms}

\begin{lemma}
   If the following conditions hold
    \begin{itemize}
     \item Let $u(A,x) \in \R^n$ be defined as Definition~\ref{def:u}
    \item Let $\alpha(A,x) \in \R$ be defined as Definition~\ref{def:alpha}
     \item Let $f(A,x) \in \R^n$ be defined as Definition~\ref{def:f}
    \item Let $c(A,x) \in \R^n$ be defined as Definition~\ref{def:c}
    \item Let $g(A,x) \in \R^d$ be defined as Definition~\ref{def:g} 
    \item Let $q(A,x) = c(A,x) + f(A,x) \in \R^n$
    \item Let $c_g(A,x) \in \R^d$ be defined as Definition~\ref{def:c_g}.
    \item Let $L_g(A,x) \in \R$ be defined as Definition~\ref{def:l_g}
    \item Let $v \in \R^n$ be a vector 
    \end{itemize}
Then, For $j_0,j_1 \in [n], i_0,i_1 \in [d]$, we have 
\begin{itemize}
    \item {\bf Part 1.}For $j_1 = j_0$ and $i_0 = i_1$
\begin{align*}
B_7^{j_1,i_1,j_1,i_1} = B_{7,1}^{j_1,i_1,j_1,i_1} +  B_{7,2}^{j_1,i_1,j_1,i_1} + B_{7,3}^{j_1,i_1,j_1,i_1} + B_{7,4}^{j_1,i_1,j_1,i_1} + B_{7,5}^{j_1,i_1,j_1,i_1} 
\end{align*}
\item {\bf Part 2.}For $j_1 = j_0$ and $i_0 \neq i_1$
\begin{align*}
 B_7^{j_1,i_1,j_1,i_0} = B_{7,1}^{j_1,i_1,j_1,i_0} +  B_{7,2}^{j_1,i_1,j_1,i_0} + B_{7,3}^{j_1,i_1,j_1,i_0} + B_{7,4}^{j_1,i_1,j_1,i_0} + B_{7,5}^{j_1,i_1,j_1,i_0} 
\end{align*}
\item {\bf Part 3.}For $j_1 \neq j_0$ and $i_0 = i_1$ 
\begin{align*}
     B_7^{j_1,i_1,j_0,i_1} = B_{7,1}^{j_1,i_1,j_0,i_1} +  B_{7,2}^{j_1,i_1,j_0,i_1} + B_{7,3}^{j_1,i_1,j_0,i_1} + B_{7,4}^{j_1,i_1,j_0,i_1} + B_{7,5}^{j_1,i_1,j_0,i_1}
\end{align*}
\item {\bf Part 4.} For $j_0 \neq j_1$ and $i_0 \neq i_1$
\begin{align*}
B_7^{j_1,i_1,j_0,i_0} = B_{7,1}^{j_1,i_1,j_0,i_0} +  B_{7,2}^{j_1,i_1,j_0,i_0} + B_{7,3}^{j_1,i_1,j_0,i_0} + B_{7,4}^{j_1,i_1,j_0,i_0} + B_{7,5}^{j_1,i_1,j_0,i_0} 
\end{align*}
\end{itemize}
\end{lemma}

\begin{proof}
{\bf Proof of Part 1.}
 we have
    \begin{align*}
   B_7^{j_1,i_1,j_1,i_1} = & ~ \frac{\d}{\d A_{j_1,i_1}}( - c_g(A,x)^{\top} \cdot f(A,x)_{j_1}\diag(x) A^{\top} ((e_{j_1} - f(A,x)) \circ q(A,x))) \\
    = &  ~   B_{7,1}^{j_1,i_1,j_1,i_1} +  B_{7,2}^{j_1,i_1,j_1,i_1} + B_{7,3}^{j_1,i_1,j_1,i_1} + B_{7,4}^{j_1,i_1,j_1,i_1} + B_{7,5}^{j_1,i_1,j_1,i_1} 
\end{align*}
{\bf Proof of Part 2.} 
we have
    \begin{align*}
    B_7^{j_1,i_1,j_1,i_0} =& ~ \frac{\d}{\d A_{j_1,i_1}}( - c_g(A,x)^{\top} \cdot f(A,x)_{j_1}\diag(x) A^{\top} ((e_{j_1} - f(A,x)) \circ q(A,x))) \\
    = &  ~    B_{7,1}^{j_1,i_1,j_1,i_0} +  B_{7,2}^{j_1,i_1,j_1,i_0} + B_{7,3}^{j_1,i_1,j_1,i_0} + B_{7,4}^{j_1,i_1,j_1,i_0} + B_{7,5}^{j_1,i_1,j_1,i_0} 
\end{align*}
{\bf Proof of Part 3.} 
 we have
    \begin{align*}
    B_7^{j_1,i_1,j_0,i_1} = & ~ \frac{\d}{\d A_{j_1,i_1}}( - c_g(A,x)^{\top} \cdot f(A,x)_{j_0}\diag(x) A^{\top} ((e_{j_0} - f(A,x)) \circ q(A,x))) \\
    = &  ~  B_{7,1}^{j_1,i_1,j_0,i_1} +  B_{7,2}^{j_1,i_1,j_0,i_1} + B_{7,3}^{j_1,i_1,j_0,i_1} + B_{7,4}^{j_1,i_1,j_0,i_1} + B_{7,5}^{j_1,i_1,j_0,i_1}
\end{align*}

{\bf Proof of Part 4.}
  we have
\begin{align*}
       B_7^{j_1,i_1,j_0,i_0} = & ~ \frac{\d}{\d A_{j_1,i_1}}( - c_g(A,x)^{\top} \cdot f(A,x)_{j_0}\diag(x) A^{\top} ((e_{j_0} - f(A,x)) \circ q(A,x))) \\
    = &  ~   B_{7,1}^{j_1,i_1,j_0,i_0} +  B_{7,2}^{j_1,i_1,j_0,i_0} + B_{7,3}^{j_1,i_1,j_0,i_0} + B_{7,4}^{j_1,i_1,j_0,i_0} + B_{7,5}^{j_1,i_1,j_0,i_0} 
\end{align*}
\end{proof}

\subsection{Lipschitz Computation}
\begin{lemma}\label{lips: B_7}
If the following conditions hold
\begin{itemize}
    \item Let $B_{7,1,1}^{j_1,*, j_0,*}, \cdots, B_{7,5,1}^{j_1,*, j_0,*} $ be defined as Lemma~\ref{lem:b_7_j1_j0} 
    \item  Let $\|A \|_2 \leq R, \|A^{\top} \|_F \leq R, \| x\|_2 \leq R, \|\diag(f(A,x)) \|_F \leq \|f(A,x) \|_2 \leq 1, \| b_g \|_2 \leq 1$ 
\end{itemize}
Then, we have
\begin{itemize}
    \item {\bf Part 1.}
    \begin{align*}
       \| B_{7,1,1}^{j_1,*,j_0,*} (A) - B_{7,1,1}^{j_1,*,j_0,*} ( \wt{A} ) \|_F \leq  \beta^{-2} \cdot n \cdot \sqrt{d}\exp(5R^2) \cdot \|A - \wt{A}\|_F
    \end{align*}
     \item {\bf Part 2.}
    \begin{align*}
       \| B_{7,1,2}^{j_1,*,j_0,*} (A) - B_{7,1,2}^{j_1,*,j_0,*} ( \wt{A} ) \|_F \leq   \beta^{-2} \cdot n \cdot \sqrt{d}\exp(5R^2) \cdot \|A - \wt{A}\|_F
    \end{align*}
     \item {\bf Part 3.}
    \begin{align*}
       \| B_{7,1,3}^{j_1,*,j_0,*} (A) - B_{7,1,3}^{j_1,*,j_0,*} ( \wt{A} ) \|_F \leq   \beta^{-2} \cdot n \cdot \sqrt{d}\exp(6R^2) \cdot \|A - \wt{A}\|_F
    \end{align*}
     \item {\bf Part 4.}
    \begin{align*}
       \| B_{7,1,4}^{j_1,*,j_0,*} (A) - B_{7,1,4}^{j_1,*,j_0,*} ( \wt{A} ) \|_F \leq   \beta^{-2} \cdot n \cdot \sqrt{d}\exp(6R^2) \cdot \|A - \wt{A}\|_F
    \end{align*}
     \item {\bf Part 5.}
    \begin{align*}
       \| B_{7,1,5}^{j_1,*,j_0,*} (A) - B_{7,1,5}^{j_1,*,j_0,*} ( \wt{A} ) \|_F \leq   \beta^{-2} \cdot n \cdot \sqrt{d}\exp(6R^2) \cdot \|A - \wt{A}\|_F
    \end{align*}
     \item {\bf Part 6.}
    \begin{align*}
       \| B_{7,1,6}^{j_1,*,j_0,*} (A) - B_{7,1,6}^{j_1,*,j_0,*} ( \wt{A} ) \|_F \leq  \beta^{-2} \cdot n \cdot \sqrt{d}\exp(6R^2) \cdot \|A - \wt{A}\|_F
    \end{align*}
     \item {\bf Part 7.}
    \begin{align*}
       \| B_{7,1,7}^{j_1,*,j_0,*} (A) - B_{7,1,7}^{j_1,*,j_0,*} ( \wt{A} ) \|_F \leq   \beta^{-2} \cdot n \cdot \sqrt{d}\exp(6R^2) \cdot \|A - \wt{A}\|_F
    \end{align*}
    \item {\bf Part 8.}
    \begin{align*}
       \| B_{7,2,1}^{j_1,*,j_0,*} (A) - B_{7,2,1}^{j_1,*,j_0,*} ( \wt{A} ) \|_F \leq  \beta^{-2} \cdot n \cdot \sqrt{d}\exp(6R^2) \cdot \|A - \wt{A}\|_F
    \end{align*}
     \item {\bf Part 9.}
    \begin{align*}
       \| B_{7,3,1}^{j_1,*,j_0,*} (A) - B_{7,3,1}^{j_1,*,j_0,*} ( \wt{A} ) \|_F \leq  \beta^{-2} \cdot n \cdot \sqrt{d}  \exp(5R^2)\|A - \wt{A}\|_F  
    \end{align*}
     \item {\bf Part 10.}
    \begin{align*}
       \| B_{7,4,1}^{j_1,*,j_0,*} (A) - B_{7,4,1}^{j_1,*,j_0,*} ( \wt{A} ) \|_F \leq  \beta^{-2} \cdot n \cdot \sqrt{d}  \exp(6R^2)\|A - \wt{A}\|_F  
    \end{align*}
     \item {\bf Part 11.}
    \begin{align*}
       \| B_{7,5,1}^{j_1,*,j_0,*} (A) - B_{7,5,1}^{j_1,*,j_0,*} ( \wt{A} ) \|_F \leq \beta^{-2} \cdot n \cdot \sqrt{d}  \exp(6R^2)\|A - \wt{A}\|_F   
    \end{align*}
        \item{\bf Part 12.}
    \begin{align*}
         \| B_{7}^{j_1,*,j_0,*} (A) - B_{7}^{j_1,*,j_0,*} ( \wt{A} ) \|_F \leq & ~   11\beta^{-2} \cdot n \cdot \sqrt{d}  \exp(6R^2)\|A - \wt{A}\|_F   
    \end{align*}
    \end{itemize}
\end{lemma}

\begin{proof}
{\bf Proof of Part 1.}
\begin{align*}
& ~ \| B_{7,1,1}^{j_1,*,j_0,*} (A) - B_{7,1,1}^{j_1,*,j_0,*} ( \wt{A} ) \| \\ \leq 
    & ~ \| f_c(A,x)  \cdot f(A,x)_{j_1} \cdot f(A,x)_{j_0} \cdot p_{j_0}(A,x)  \cdot  {\bf 1}_d^\top \\
    & ~ - f_c(\wt{A},x)  \cdot f(\wt{A},x)_{j_1} \cdot f(\wt{A},x)_{j_0} \cdot p_{j_0}(\wt{A},x)  \cdot  {\bf 1}_d^\top \|_F\\
    \leq & ~ |  f_c(A,x) - f_c(\wt{A},x) | \cdot |f(A,x)_{j_1}|\cdot | f(A,x)_{j_0} | \cdot    \| p_{j_0}(A,x)\|_2  \cdot \| {\bf 1}_d^\top\|_2 \\
    & + ~   |f_c(\wt{A},x)| \cdot  |f(A,x)_{j_1} - f(\wt{A},x)_{j_1}|\cdot | f(A,x)_{j_0} | \cdot  \| p_{j_0}(A,x)\|_2 \cdot \| {\bf 1}_d^\top\|_2 \\
    & + ~   |f_c(\wt{A},x)| \cdot  |f(\wt{A},x)_{j_1}|\cdot | f(A,x)_{j_0} -  f(\wt{A},x)_{j_0} | \cdot \| p_{j_0}(A,x)\|_2  \cdot \| {\bf 1}_d^\top\|_2 \\
     & + ~   |f_c(\wt{A},x)| \cdot  |f(\wt{A},x)_{j_1}|\cdot | f(\wt{A},x)_{j_0} | \cdot \| p_{j_0}(A,x) - p_{j_0}(\wt{A},x)\|_2  \cdot \| {\bf 1}_d^\top\|_2 \\
    \leq & ~ 36R^2  \beta^{-2} \cdot n \cdot \sqrt{d}  \exp(3R^2)\|A - \wt{A}\|_F \\
    &+ ~ 24R^2  \beta^{-2} \cdot n \cdot \sqrt{d}  \exp(3R^2)\|A - \wt{A}\|_F \\
    & + ~ 24R^2 \beta^{-2} \cdot n \cdot \sqrt{d}\exp(3R^2) \cdot \|A - \wt{A}\|_F \\
    & + ~ 26 \beta^{-2} \cdot n \cdot \sqrt{d}\exp(4R^2) \cdot \|A - \wt{A}\|_F\\
    \leq &  ~ 110 \beta^{-2} \cdot n \cdot \sqrt{d}\exp(4R^2) \cdot \|A - \wt{A}\|_F \\
    \leq &  ~ \beta^{-2} \cdot n \cdot \sqrt{d}\exp(5R^2) \cdot \|A - \wt{A}\|_F 
\end{align*}

{\bf Proof of Part 2.}
\begin{align*}
& ~ \| B_{7,1,2}^{j_1,*,j_0,*} (A) - B_{7,1,2}^{j_1,*,j_0,*} ( \wt{A} ) \| \\ \leq 
    & ~ \| c(A,x)_{j_1}  \cdot f(A,x)_{j_1} \cdot f(A,x)_{j_0} \cdot p_{j_0}(A,x)  \cdot  {\bf 1}_d^\top \\
    & ~ - c(\wt{A},x)_{j_1}   \cdot f(\wt{A},x)_{j_1} \cdot f(\wt{A},x)_{j_0} \cdot p_{j_0}(\wt{A},x)  \cdot  {\bf 1}_d^\top \|_F\\
    \leq & ~ |  c(A,x)_{j_1}  - c(\wt{A},x)_{j_1}  | \cdot |f(A,x)_{j_1}|\cdot | f(A,x)_{j_0} | \cdot    \| p_{j_0}(A,x)\|_2  \cdot \| {\bf 1}_d^\top\|_2 \\
    & + ~   | c(\wt{A},x)_{j_1}  | \cdot  |f(A,x)_{j_1} - f(\wt{A},x)_{j_1}|\cdot | f(A,x)_{j_0} | \cdot  \| p_{j_0}(A,x)\|_2 \cdot \| {\bf 1}_d^\top\|_2 \\
    & + ~   | c(\wt{A},x)_{j_1}  |  \cdot  |f(\wt{A},x)_{j_1}|\cdot | f(A,x)_{j_0} -  f(\wt{A},x)_{j_0} | \cdot \| p_{j_0}(A,x)\|_2  \cdot \| {\bf 1}_d^\top\|_2 \\
     & + ~   | c(\wt{A},x)_{j_1}  |  \cdot  |f(\wt{A},x)_{j_1}|\cdot | f(\wt{A},x)_{j_0} | \cdot \| p_{j_0}(A,x) - p_{j_0}(\wt{A},x)\|_2  \cdot \| {\bf 1}_d^\top\|_2 \\
    \leq & ~ 12R^2  \beta^{-2} \cdot n \cdot \sqrt{d}  \exp(3R^2)\|A - \wt{A}\|_F \\
    &+ ~ 24R^2  \beta^{-2} \cdot n \cdot \sqrt{d}  \exp(3R^2)\|A - \wt{A}\|_F \\
    & + ~ 24R^2 \beta^{-2} \cdot n \cdot \sqrt{d}\exp(3R^2) \cdot \|A - \wt{A}\|_F \\
    & + ~ 26 \beta^{-2} \cdot n \cdot \sqrt{d}\exp(4R^2) \cdot \|A - \wt{A}\|_F\\
    \leq &  ~ 86 \beta^{-2} \cdot n \cdot \sqrt{d}\exp(4R^2) \cdot \|A - \wt{A}\|_F \\
    \leq &  ~ \beta^{-2} \cdot n \cdot \sqrt{d}\exp(5R^2) \cdot \|A - \wt{A}\|_F 
\end{align*}

{\bf Proof of Part 3.}
\begin{align*}
    & ~ \| B_{7,1,3}^{j_1,*,j_0,*} (A) - B_{7,1,3}^{j_1,*,j_0,*} ( \wt{A} ) \|_F \\
    = & ~ \| f(A,x)_{j_1} \cdot f(A,x)_{j_0} \cdot f_c(A,x) \cdot ( (A_{j_1,*}) \circ x^\top  )  \cdot   p_{j_0}(A,x)  \cdot I_d \\
    & - ~  f(\wt{A},x)_{j_1} \cdot f(\wt{A},x)_{j_0} \cdot f_c(\wt{A},x) \cdot  \wt{A}_{j_1,*}) \circ x^\top  )  \cdot   p_{j_0}(\wt{A},x)  \cdot I_d \|_F\\
    \leq & ~ |f(A,x)_{j_1} - f(\wt{A},x)_{j_1}| \cdot |f(A,x)_{j_0} |\cdot |f_c(A,x)| \cdot  \|A_{j_1,*}\|_2 \| \diag(x) \|_F  \cdot  \| p_{j_0}(A,x) \|_2 \cdot \|I_d\|_F \\
    & + ~ |f(\wt{A},x)_{j_1}| \cdot |f(A,x)_{j_0} - f(\wt{A},x)_{j_0}|\cdot |f_c(A,x)| \cdot  \|A_{j_1,*}\|_2 \| \diag(x) \|_F  \cdot  \| p_{j_0}(A,x) \|_2 \cdot \|I_d\|_F \\
    & + ~ |f(\wt{A},x)_{j_1}| \cdot | f(\wt{A},x)_{j_0}|\cdot |f_c(A,x) -f_c(\wt{A},x)| \cdot  \|A_{j_1,*}\|_2 \| \diag(x) \|_F  \cdot  \| p_{j_0}(A,x) \|_2 \cdot \|I_d\|_F \\
    & + ~ |f(\wt{A},x)_{j_1}| \cdot | f(\wt{A},x)_{j_0}|\cdot |f_c(\wt{A},x)| \cdot  \|A_{j_1,*} - \wt{A}_{j_1,*}  \|_2 \| \diag(x) \|_F  \cdot  \| p_{j_0}(A,x) \|_2 \cdot \|I_d\|_F \\
    & + ~ |f(\wt{A},x)_{j_1}| \cdot | f(\wt{A},x)_{j_0}|\cdot |f_c(\wt{A},x)| \cdot  \|\wt{A}_{j_1,*}  \|_2 \| \diag(x) \|_F  \cdot  \| p_{j_0}(A,x) - p_{j_0}(\wt{A},x) \|_2 \cdot \|I_d\|_F \\
    \leq & ~ 24 R^4 \beta^{-2} \cdot n \cdot \sqrt{d}\exp(3R^2) \cdot \|A - \wt{A}\|_F \\
    & + ~ 24 R^4 \beta^{-2} \cdot n \cdot \sqrt{d}\exp(3R^2) \cdot \|A - \wt{A}\|_F \\
    & + ~ 36 R^4 \beta^{-2} \cdot n \cdot \sqrt{d}\exp(3R^2) \cdot \|A - \wt{A}\|_F \\
    & + ~ 12 R^3 \cdot \sqrt{d}\|A - \wt{A}\|_F \\
    & + ~ 26 R^2 \beta^{-2} \cdot n \cdot \sqrt{d}\exp(4R^2) \cdot \|A - \wt{A}\|_F \\
    \leq & ~ 122 \beta^{-2} \cdot n \cdot \sqrt{d}\exp(5R^2) \cdot \|A - \wt{A}\|_F \\
    \leq & ~  \beta^{-2} \cdot n \cdot \sqrt{d}\exp(6R^2) \cdot \|A - \wt{A}\|_F 
\end{align*}

{\bf Proof of Part 4.}
\begin{align*}
    & ~ \| B_{7,1,4}^{j_1,*,j_0,*} (A) - B_{7,1,4}^{j_1,*,j_0,*} ( \wt{A} ) \|_F \\
    \leq & ~ \|-  f(A,x)_{j_1} \cdot f(A,x)_{j_0} \cdot f_c(A,x)  \cdot h(A,x)^\top \cdot p_{j_0}(A,x) \cdot I_d \\
    & ~ -(-  f(A,x)_{j_1} \cdot f(A,x)_{j_0} \cdot f_c(A,x)  \cdot h(A,x)^\top \cdot p_{j_0}(A,x) \cdot I_d)\|_F\\
    \leq & ~ \|  f(A,x)_{j_1} \cdot f(A,x)_{j_0} \cdot f_c(A,x)  \cdot h(A,x)^\top \cdot p_{j_0}(A,x) \cdot I_d \\
    & ~ -  f(A,x)_{j_1} \cdot f(A,x)_{j_0} \cdot f_c(A,x)  \cdot h(A,x)^\top \cdot p_{j_0}(A,x) \cdot I_d\|_F\\
    \leq & ~  |f(A,x)_{j_1} - f(\wt{A},x)_{j_1}| \cdot |f(A,x)_{j_0} |\cdot |f_c(A,x)  |\cdot \|h(A,x)^\top\|_2 \cdot \|p_{j_0}(A,x)\|_2 \cdot \|I_d  \|_F \\
    & + ~ |f(\wt{A},x)_{j_1}| \cdot |f(A,x)_{j_0}  -f(\wt{A},x)_{j_0}|\cdot |f_c(A,x)  |\cdot \|h(A,x)^\top\|_2 \cdot \|p_{j_0}(A,x)\|_2 \cdot \|I_d  \|_F \\
    & + ~ |f(\wt{A},x)_{j_1}| \cdot |f(\wt{A},x)_{j_0}|\cdot |f_c(A,x)  - f_c(\wt{A},x) |\cdot \|h(A,x)^\top\|_2 \cdot \|p_{j_0}(A,x)\|_2 \cdot \|I_d  \|_F \\
    & + ~ |f(\wt{A},x)_{j_1}| \cdot |f(\wt{A},x)_{j_0}|\cdot | f_c(\wt{A},x) |\cdot \|h(A,x)^\top - h(\wt{A},x)^\top\|_2 \cdot \|p_{j_0}(A,x)\|_2 \cdot \|I_d  \|_F \\
    & + ~ |f(\wt{A},x)_{j_1}| \cdot |f(\wt{A},x)_{j_0}|\cdot | f_c(\wt{A},x) |\cdot \|h(\wt{A},x)^\top\|_2 \cdot \|p_{j_0}(A,x) - p_{j_0}(\wt{A},x)\|_2 \cdot \|I_d  \|_F \\
    \leq & ~ 24R^4 \beta^{-2} \cdot n \cdot \sqrt{d}\exp(3R^2) \cdot \|A - \wt{A}\|_F \\
    & +  ~ 24R^4 \beta^{-2} \cdot n \cdot \sqrt{d}\exp(3R^2) \cdot \|A - \wt{A}\|_F \\
    & +  ~ 36R^4 \beta^{-2} \cdot n \cdot \sqrt{d}\exp(3R^2) \cdot \|A - \wt{A}\|_F \\
    & +  ~ 36R^4 \beta^{-2} \cdot n \cdot \sqrt{d}\exp(3R^2) \cdot \|A - \wt{A}\|_F \\
    & +  ~ 18R^2 \beta^{-2} \cdot n \cdot \sqrt{d}\exp(4R^2) \cdot \|A - \wt{A}\|_F \\
    \leq & ~ 138\beta^{-2} \cdot n \cdot \sqrt{d}\exp(5R^2) \cdot \|A - \wt{A}\|_F \\
    \leq & ~\beta^{-2} \cdot n \cdot \sqrt{d}\exp(6R^2) \cdot \|A - \wt{A}\|_F 
\end{align*}

{\bf Proof of Part 5.}
\begin{align*}
    & ~ \| B_{7,1,5}^{j_1,*,j_0,*} (A) - B_{7,1,5}^{j_1,*,j_0,*} ( \wt{A} ) \|_F \\
    = & ~  \| f(A,x)_{j_1} \cdot f(A,x)_{j_0} \cdot (-f_2(A,x) + f(A,x)_{j_1})  \cdot h(A,x)^\top \cdot p_{j_0}(A,x) \cdot I_d \\
    & - ~ f(\wt{A},x)_{j_1} \cdot f(\wt{A},x)_{j_0} \cdot (-f_2(\wt{A},x) + f(\wt{A},x)_{j_1})  \cdot h(\wt{A},x)^\top \cdot p_{j_0}(\wt{A},x) \cdot I_d\|_F \\
    \leq & ~ |f(A,x)_{j_1} - f(\wt{A},x)_{j_1}| \cdot |f(A,x)_{j_0}| \cdot (|-f_2(A,x)| + |f(A,x)_{j_1}|)  \cdot \|h(A,x)^\top\|_2 \cdot \|p_{j_0}(A,x)\|_2 \cdot \|I_d\|_F \\
    & + ~ | f(\wt{A},x)_{j_1}| \cdot |f(A,x)_{j_0}- f(\wt{A},x)_{j_0}| \cdot (|-f_2(A,x)| + |f(A,x)_{j_1}|)  \cdot \|h(A,x)^\top\|_2 \cdot \|p_{j_0}(A,x)\|_2 \cdot \|I_d\|_F \\
    & + ~ | f(\wt{A},x)_{j_1}| \cdot | f(\wt{A},x)_{j_0}| \cdot (|f_2(A,x) - f_2(\wt{A},x)| + |f(A,x)_{j_1} - f(\wt{A},x)_{j_1}|)  \cdot \|h(A,x)^\top\|_2 \cdot \|p_{j_0}(A,x)\|_2 \cdot \|I_d\|_F \\
    & + ~ | f(\wt{A},x)_{j_1}| \cdot | f(\wt{A},x)_{j_0}| \cdot (|f_2(\wt{A},x)| + | f(\wt{A},x)_{j_1}|)  \cdot \|h(A,x)^\top - h(\wt{A},x)^\top\|_2 \cdot \|p_{j_0}(A,x)\|_2 \cdot \|I_d\|_F \\
    & + ~ | f(\wt{A},x)_{j_1}| \cdot | f(\wt{A},x)_{j_0}| \cdot (|f_2(\wt{A},x)| + | f(\wt{A},x)_{j_1}|)  \cdot \|h(\wt{A},x)^\top\|_2 \cdot \|p_{j_0}(A,x) - p_{j_0}(\wt{A},x)\|_2 \cdot \|I_d\|_F \\
    \leq & ~ 24R^4 \beta^{-2} \cdot n \cdot \sqrt{d}\exp(3R^2) \cdot \|A - \wt{A}\|_F \\
    & + ~ 24R^4 \beta^{-2} \cdot n \cdot \sqrt{d}\exp(3R^2) \cdot \|A - \wt{A}\|_F \\
    & + ~ 36 R^4 \beta^{-2} \cdot n \cdot \sqrt{d}\exp(3R^2) \cdot \|A - \wt{A}\|_F \\
    & + ~ 18 R^2 \beta^{-2} \cdot n \cdot \sqrt{d}\exp(4R^2) \cdot \|A - \wt{A}\|_F \\
    & + ~ 26 R^2 \beta^{-2} \cdot n \cdot \sqrt{d}\exp(4R^2) \cdot \|A - \wt{A}\|_F \\
    \leq & ~ 128 \beta^{-2} \cdot n \cdot \sqrt{d}\exp(5R^2) \cdot \|A - \wt{A}\|_F \\
    \leq & ~ \beta^{-2} \cdot n \cdot \sqrt{d}\exp(6R^2) \cdot \|A - \wt{A}\|_F 
\end{align*}

{\bf Proof of Part 6.}
\begin{align*}
     & ~ \| B_{7,1,6}^{j_1,*,j_0,*} (A) - B_{7,1,6}^{j_1,*,j_0,*} ( \wt{A} ) \|_F \\
    = & ~ \|f(A,x)_{j_1} \cdot f(A,x)_{j_0} \cdot(-f_c(A,x) + f(A,x)_{j_1})  \cdot h(A,x)^\top \cdot  p_{j_0}(A,x) \cdot I_d\|_F\\
    & - ~ f(\wt{A},x)_{j_1} \cdot f(\wt{A},x)_{j_0} \cdot(-f_c(\wt{A},x) + f(\wt{A},x)_{j_1})  \cdot h(\wt{A},x)^\top \cdot  p_{j_0}(\wt{A},x) \cdot I_d\|_F\\
    \leq & ~ |f(A,x)_{j_1} -f(\wt{A},x)_{j_1}| \cdot |f(A,x)_{j_0}| \cdot(|-f_c(A,x)| + |f(A,x)_{j_1}|)  \cdot \|h(A,x)^\top\|_2 \cdot \| p_{j_0}(A,x) \|_2 \cdot \|I_d\|_F \\
    & + ~ |f(\wt{A},x)_{j_1}| \cdot |f(A,x)_{j_0} - f(\wt{A},x)_{j_0}| \cdot(|-f_c(A,x)| + |f(A,x)_{j_1}|)  \cdot \|h(A,x)^\top\|_2 \cdot \| p_{j_0}(A,x) \|_2 \cdot \|I_d\|_F \\
    & + ~ |f(\wt{A},x)_{j_1}| \cdot |f(\wt{A},x)_{j_0}| \cdot(|f_c(A,x) - f_c(\wt{A},x)| + |f(A,x)_{j_1}  - f(\wt{A},x)_{j_1}|)  \cdot \|h(A,x)^\top\|_2 \cdot \| p_{j_0}(A,x) \|_2 \cdot \|I_d\|_F \\
    & + ~ |f(\wt{A},x)_{j_1}| \cdot |f(\wt{A},x)_{j_0}| \cdot(|f_c(\wt{A},x)| + |f(\wt{A},x)_{j_1}|)  \cdot \|h(A,x)^\top - h(\wt{A},x)^\top\|_2 \cdot \| p_{j_0}(A,x) \|_2 \cdot \|I_d\|_F \\
    & + ~ |f(\wt{A},x)_{j_1}| \cdot |f(\wt{A},x)_{j_0}| \cdot(|f_c(\wt{A},x)| + |f(\wt{A},x)_{j_1}|)  \cdot \|h(\wt{A},x)^\top\|_2 \cdot \| p_{j_0}(A,x)  - p_{j_0}(\wt{A},x)\|_2 \cdot \|I_d\|_F \\
    \leq & ~ 36R^4 \beta^{-2} \cdot n \cdot \sqrt{d}\exp(3R^2) \cdot \|A - \wt{A}\|_F \\
    & + ~ 36R^4 \beta^{-2} \cdot n \cdot \sqrt{d}\exp(3R^2) \cdot \|A - \wt{A}\|_F \\
    & + ~ 48R^4 \beta^{-2} \cdot n \cdot \sqrt{d}\exp(3R^2) \cdot \|A - \wt{A}\|_F \\
    & + ~ 54R^2 \beta^{-2} \cdot n \cdot \sqrt{d}\exp(4R^2) \cdot \|A - \wt{A}\|_F \\
    & + ~ 39R^2 \beta^{-2} \cdot n \cdot \sqrt{d}\exp(4R^2) \cdot \|A - \wt{A}\|_F \\
    \leq & ~ 213\beta^{-2} \cdot n \cdot \sqrt{d}\exp(5R^2) \cdot \|A - \wt{A}\|_F \\
    \leq & ~ \beta^{-2} \cdot n \cdot \sqrt{d}\exp(6R^2) \cdot \|A - \wt{A}\|_F 
\end{align*}

{\bf Proof of Part 7.}
\begin{align*}
     & ~ \| B_{7,1,7}^{j_1,*,j_0,*} (A) - B_{7,1,7}^{j_1,*,j_0,*} ( \wt{A} ) \|_F \\
     = & ~ \| f(A,x)_{j_1} \cdot f(A,x)_{j_0}  \cdot p_{j_1}(A,x)^\top \cdot p_{j_0}(A,x) \cdot I_d -  f(\wt{A},x)_{j_1} \cdot f(\wt{A},x)_{j_0}  \cdot p_{j_1}(\wt{A},x)^\top \cdot p_{j_0}(\wt{A},x) \cdot I_d \|_F \\
     \leq & ~ |f(A,x)_{j_1} - f(\wt{A},x)_{j_1}| \cdot |f(A,x)_{j_0}|  \cdot \| p_{j_1}(A,x)^\top\|_2 \cdot \|p_{j_0}(A,x) \|_2 \cdot \|I_d\|_F \\
     & + ~ |f(\wt{A},x)_{j_1}| \cdot |f(A,x)_{j_0} -f(\wt{A},x)_{j_0}|  \cdot \| p_{j_1}(A,x)^\top\|_2 \cdot \|p_{j_0}(A,x) \|_2 \cdot \|I_d\|_F \\
     & + ~ |f(\wt{A},x)_{j_1}| \cdot |f(\wt{A},x)_{j_0}|  \cdot \| p_{j_1}(A,x)^\top - p_{j_1}(\wt{A},x)^\top\|_2 \cdot \|p_{j_0}(A,x) \|_2 \cdot \|I_d\|_F \\
     & + ~ |f(\wt{A},x)_{j_1}| \cdot |f(\wt{A},x)_{j_0}|  \cdot \|  p_{j_1}(\wt{A},x)^\top\|_2 \cdot \|p_{j_0}(A,x)  -p_{j_0}(\wt{A},x) \|_2 \cdot \|I_d\|_F \\
     \leq & ~ 72R^4 \beta^{-2} \cdot n \cdot \sqrt{d}\exp(3R^2) \cdot \|A - \wt{A}\|_F \\
     & + ~ 72R^4 \beta^{-2} \cdot n \cdot \sqrt{d}\exp(3R^2) \cdot \|A - \wt{A}\|_F \\
     & + ~ 78R^2\beta^{-2} \cdot n \cdot \sqrt{d}\exp(4R^2) \cdot \|A - \wt{A}\|_F \\
     & + ~ 78R^2\beta^{-2} \cdot n \cdot \sqrt{d}\exp(4R^2) \cdot \|A - \wt{A}\|_F \\
     \leq & ~ 300 \beta^{-2} \cdot n \cdot \sqrt{d}\exp(5R^2) \cdot \|A - \wt{A}\|_F \\
     \leq & ~ \beta^{-2} \cdot n \cdot \sqrt{d}\exp(6R^2) \cdot \|A - \wt{A}\|_F
\end{align*}

{\bf Proof of Part 8.}
\begin{align*}
    & ~ \| B_{7,2,1}^{j_1,*,j_0,*} (A) - B_{7,2,1}^{j_1,*,j_0,*} ( \wt{A} ) \|_F \\
    =  & ~ \| f(A,x)_{j_1} \cdot f(A,x)_{j_0}  \cdot c_g(A,x)^{\top} \cdot p_{j_0}(A,x)\cdot x \cdot {\bf 1}_d^{\top}  -  f(\wt{A},x)_{j_1} \cdot f(\wt{A},x)_{j_0}  \cdot c_g(\wt{A},x)^{\top} \cdot p_{j_0}(\wt{A},x)\cdot x \cdot {\bf 1}_d^{\top}  \|_F\\
    \leq & ~ |f(A,x)_{j_1} - f(\wt{A},x)_{j_1}| \cdot |f(A,x)_{j_0}|  \cdot 
    \| c_g(A,x)^{\top}\|_2 \cdot \|p_{j_0}(A,x)\|_2\cdot \|x\|_2 \cdot \|{\bf 1}_d^{\top}\|_2 \\
     & + ~ |f(\wt{A},x)_{j_1}| \cdot |f(A,x)_{j_0} - f(\wt{A},x)_{j_0}|  \cdot 
    \| c_g(A,x)^{\top}\|_2 \cdot \|p_{j_0}(A,x)\|_2\cdot \|x\|_2 \cdot \|{\bf 1}_d^{\top}\|_2 \\
    & + ~ |f(\wt{A},x)_{j_1}| \cdot | f(\wt{A},x)_{j_0}|  \cdot 
    \| c_g(A,x)^{\top} - c_g(\wt{A},x)^{\top}\|_2 \cdot \|p_{j_0}(A,x)\|_2\cdot \|x\|_2 \cdot \|{\bf 1}_d^{\top}\|_2 \\
    & + ~ |f(\wt{A},x)_{j_1}| \cdot | f(\wt{A},x)_{j_0}|  \cdot 
    \| c_g(\wt{A},x)^{\top}\|_2 \cdot \|p_{j_0}(A,x) -p_{j_0}(\wt{A},x)\|_2\cdot \|x\|_2 \cdot \|{\bf 1}_d^{\top}\|_2 \\
    \leq & ~ 60R^4 \beta^{-2} \cdot n \cdot \sqrt{d}\exp(3R^2) \cdot \|A - \wt{A}\|_F \\
    & + ~ 60R^4 \beta^{-2} \cdot n \cdot \sqrt{d}\exp(3R^2) \cdot \|A - \wt{A}\|_F \\
    & + ~ 120 R^4 \beta^{-2} \cdot n \cdot \sqrt{d}\exp(3R^2) \cdot \|A - \wt{A}\|_F \\
    & + ~ 65 R^2 \beta^{-2} \cdot n \cdot \sqrt{d}\exp(4R^2) \cdot \|A - \wt{A}\|_F \\
    \leq & ~ 305 \beta^{-2} \cdot n \cdot \sqrt{d}\exp(5R^2) \cdot \|A - \wt{A}\|_F \\
    \leq & ~\beta^{-2} \cdot n \cdot \sqrt{d}\exp(5R^2) \cdot \|A - \wt{A}\|_F \\
\end{align*}

{\bf Proof of Part 9.}
\begin{align*}
    & ~ \| B_{7,3,1}^{j_1,*,j_0,*} (A) - B_{7,3,1}^{j_1,*,j_0,*} ( \wt{A} ) \|_F \\
    \leq & ~ \|- f(A,x)_{j_0}\cdot e_{j_1}^\top \cdot ((e_{j_0} - f(A,x)) \circ q(A,x)) \cdot {\bf 1}_d \cdot c_g(A,x)^{\top}   \cdot \diag(x) \\
    & - ~ (- f(A,x)_{j_0}\cdot e_{j_1}^\top \cdot ((e_{j_0} - f(A,x)) \circ q(A,x)) \cdot {\bf 1}_d \cdot c_g(A,x)^{\top}   \cdot \diag(x) ) \|_F \\
    \leq & ~ \| f(A,x)_{j_0}\cdot e_{j_1}^\top \cdot ((e_{j_0} - f(A,x)) \circ q(A,x)) \cdot {\bf 1}_d \cdot c_g(A,x)^{\top}   \cdot \diag(x) \\
    & - ~ f(A,x)_{j_0}\cdot e_{j_1}^\top \cdot ((e_{j_0} - f(A,x)) \circ q(A,x)) \cdot {\bf 1}_d \cdot c_g(A,x)^{\top}   \cdot \diag(x)  \|_F \\
    \leq & ~ |f(A,x)_{j_0} - f(\wt{A},x)_{j_0}| \cdot \| e_{j_1}^\top \| \cdot (\|e_{j_0}^{\top}\|_2 + \|-f(A,x)^{\top}\|_2) \| \diag(q(A,x) )\|_F \cdot \|{\bf 1}_d \|_2\cdot \|c_g(A,x)^{\top}\|_2   \cdot \|\diag(x)\|_F\\
    & + ~ | f(\wt{A},x)_{j_0}| \cdot \| e_{j_1}^\top \| \cdot \|f(A,x)^{\top} -f(\wt{A},x)^{\top}\|_2 \| \diag(q(A,x) )\|_F \cdot \|{\bf 1}_d \|_2\cdot \|c_g(A,x)^{\top}\|_2   \cdot \|\diag(x)\|_F\\
    & + ~ |f(\wt{A},x)_{j_0}| \cdot \| e_{j_1}^\top \| \cdot (\|e_{j_0}^{\top}\|_2 + \|-f(\wt{A},x)^{\top}\|_2) \| \diag(q(A,x) ) - \diag(q(\wt{A},x) )\|_F \\
    & ~ \cdot \|{\bf 1}_d \|_2\cdot \|c_g(A,x)^{\top}\|_2   \cdot \|\diag(x)\|_F\\
    & + ~ |f(\wt{A},x)_{j_0}| \cdot \| e_{j_1}^\top \| \cdot (\|e_{j_0}^{\top}\|_2 + \|-f(\wt{A},x)^{\top}\|_2)\| \diag(q(\wt{A},x) )\|_F \cdot \|{\bf 1}_d \|_2 \\
        & ~\cdot \|c_g(A,x)^{\top} - c_g(\wt{A},x)^{\top}\|_2   \cdot \|\diag(x)\|_F\\
    \leq & ~ 60R^2  \beta^{-2} \cdot n \cdot \sqrt{d}\exp(3R^2) \cdot \|A - \wt{A}\|_F \\
    & + ~ 30R^2  \beta^{-2} \cdot n \cdot \sqrt{d}\exp(3R^2) \cdot \|A - \wt{A}\|_F \\
    & + ~ 40R^2   \beta^{-2} \cdot n \cdot \sqrt{d}\exp(3R^2) \cdot \|A - \wt{A}\|_F \\
    & + ~ 120R^2  \beta^{-2} \cdot n \cdot \sqrt{d}\exp(3R^2) \cdot \|A - \wt{A}\|_F \\
    \leq & ~ 250 \beta^{-2} \cdot n \cdot \sqrt{d}\exp(4R^2) \cdot \|A - \wt{A}\|_F \\
    \leq & ~ \beta^{-2} \cdot n \cdot \sqrt{d}\exp(5R^2) \cdot \|A - \wt{A}\|_F \\
\end{align*}

{\bf Proof of Part 10}
\begin{align*}
    & ~ \| B_{7,4,1}^{j_1,*,j_0,*} (A) - B_{7,4,1}^{j_1,*,j_0,*} ( \wt{A} ) \|_F \\
    \leq & ~  \| f(A,x)_{j_1} \cdot f(A,x)_{j_0} \cdot c_g(A,x)^{\top}  \cdot  p_{j_1}(A,x)\cdot x \cdot {\bf 1}_d^{\top} - f(\wt{A},x)_{j_1} \cdot f(\wt{A},x)_{j_0} \cdot c_g(\wt{A},x)^{\top}  \cdot  p_{j_1}(\wt{A},x)\cdot x \cdot {\bf 1}_d^{\top} \|_F \\
    \leq & ~ |f(A,x)_{j_1} - f(\wt{A},x)_{j_1}| \cdot |f(A,x)_{j_0} |\cdot \|c_g(A,x)^{\top} \|_2 \cdot \| p_{j_1}(A,x)\|_2\cdot \|x\|_2\cdot \|{\bf 1}_d^{\top}\|_2 \\
    &  + ~|f(\wt{A},x)_{j_1}| \cdot |f(A,x)_{j_0}  -f(\wt{A},x)_{j_0}|\cdot \|c_g(A,x)^{\top} \|_2 \cdot \| p_{j_1}(A,x)\|_2\cdot \|x\|_2\cdot \|{\bf 1}_d^{\top}\|_2 \\
    &  + ~|f(\wt{A},x)_{j_1}| \cdot |f(\wt{A},x)_{j_0}|\cdot \|c_g(A,x)^{\top} - c_g(\wt{A},x)^{\top} \|_2 \cdot \| p_{j_1}(A,x)\|_2\cdot \|x\|_2\cdot \|{\bf 1}_d^{\top}\|_2 \\
    &  + ~|f(\wt{A},x)_{j_1}| \cdot |f(\wt{A},x)_{j_0}|\cdot \| c_g(\wt{A},x)^{\top} \|_2 \cdot \| p_{j_1}(A,x) - p_{j_1}(\wt{A},x)\|_2\cdot \|x\|_2\cdot \|{\bf 1}_d^{\top}\|_2 \\
    \leq & ~  60R^4  \beta^{-2} \cdot n \cdot \sqrt{d}\exp(3R^2) \cdot \|A - \wt{A}\|_F \\
    & + ~  60R^4  \beta^{-2} \cdot n \cdot \sqrt{d}\exp(3R^2) \cdot \|A - \wt{A}\|_F \\
    & + ~  120 R^4  \beta^{-2} \cdot n \cdot \sqrt{d}\exp(3R^2) \cdot \|A - \wt{A}\|_F \\
    &+ ~  65R^2 \beta^{-2} \cdot n \cdot \sqrt{d}\exp(4R^2) \cdot \|A - \wt{A}\|_F \\
    \leq & ~305\beta^{-2} \cdot n \cdot \sqrt{d}\exp(5R^2) \cdot \|A - \wt{A}\|_F \\ 
    \leq & ~\beta^{-2} \cdot n \cdot \sqrt{d}\exp(6R^2) \cdot \|A - \wt{A}\|_F 
\end{align*}

{\bf Proof of Part 11.}
\begin{align*}
    & ~ \| B_{7,5,1}^{j_1,*,j_0,*} (A) - B_{7,5,1}^{j_1,*,j_0,*} ( \wt{A} ) \|_F \\
    \leq & ~ \|-2f(A,x)_{j_1} \cdot f(A,x)_{j_0} \cdot   c_g(A,x)^{\top} \cdot \diag(x) \cdot A^{\top} \cdot  ((e_{j_0} - f(A,x)) \circ  ( (e_{j_1} - f(A,x)) ) )   \cdot x \cdot {\bf 1}_d^{\top} \\
    - & (-2f(A,x)_{j_1} \cdot f(A,x)_{j_0} \cdot   c_g(A,x)^{\top} \cdot \diag(x) \cdot A^{\top} \cdot  ((e_{j_0} - f(A,x)) \circ  ( (e_{j_1} - f(A,x)) ) )   \cdot x \cdot {\bf 1}_d^{\top} )\|_F \\
    \leq & ~ \|2f(A,x)_{j_1} \cdot f(A,x)_{j_0} \cdot   c_g(A,x)^{\top} \cdot \diag(x) \cdot A^{\top} \cdot  ((e_{j_0} - f(A,x)) \circ  ( (e_{j_1} - f(A,x)) ) )   \cdot x \cdot {\bf 1}_d^{\top} \\
    - & 2f(A,x)_{j_1} \cdot f(A,x)_{j_0} \cdot   c_g(A,x)^{\top} \cdot \diag(x) \cdot A^{\top} \cdot  ((e_{j_0} - f(A,x)) \circ  ( (e_{j_1} - f(A,x)) ) )   \cdot x \cdot {\bf 1}_d^{\top} \|_F \\
    \leq & ~ 2|f(A,x)_{j_1}-f(\wt{A},x)_{j_1} | \cdot|f(A,x)_{j_0}| \cdot\|c_g(A,x)^{\top}\|_2 \cdot\|\diag(x)\|_F \cdot\|A^{\top}\|_2   \\
    & \cdot  ~ (\|e_{j_0}\|_2 + \| f(A,x)\|_2) \cdot(\|e_{j_0}\|_2 + \|f(A,x)\|_2)  \cdot \|x\|_2 \cdot \|{\bf 1}_d^{\top}\|_2 \\
    & + ~ 2|f(\wt{A},x)_{j_1} | \cdot|f(A,x)_{j_0} - f(\wt{A},x)_{j_0}| \cdot\|c_g(A,x)^{\top}\|_2 \cdot\|\diag(x)\|_F \cdot\|A^{\top}\|_2   \\
    & \cdot  ~ (\|e_{j_0}\|_2 + \| f(A,x)\|_2) \cdot(\|e_{j_0}\|_2 + \|f(A,x)\|_2)  \cdot \|x\|_2 \cdot \|{\bf 1}_d^{\top}\|_2 \\
    & + ~ 2|f(\wt{A},x)_{j_1} | \cdot|f(\wt{A},x)_{j_0}| \cdot\|c_g(A,x)^{\top} - c_g(\wt{A},x)^{\top}\|_2 \cdot\|\diag(x)\|_F \cdot\|A^{\top}\|_2   \\
    & \cdot  ~ (\|e_{j_0}\|_2 + \| f(A,x)\|_2) \cdot(\|e_{j_0}\|_2 + \|f(A,x)\|_2)  \cdot \|x\|_2 \cdot \|{\bf 1}_d^{\top}\|_2 \\
    & + ~ 2|f(\wt{A},x)_{j_1} | \cdot|f(\wt{A},x)_{j_0}| \cdot\|c_g(\wt{A},x)^{\top}\|_2 \cdot\|\diag(x)\|_F \cdot\|A^{\top} - \wt{A}^{\top}\|_2   \\
    & \cdot  ~ (\|e_{j_0}\|_2 + \| f(A,x)\|_2) \cdot(\|e_{j_0}\|_2 + \|f(A,x)\|_2)  \cdot \|x\|_2 \cdot \|{\bf 1}_d^{\top}\|_2 \\
    & + ~ 2|f(\wt{A},x)_{j_1} | \cdot|f(\wt{A},x)_{j_0}| \cdot\|c_g(\wt{A},x)^{\top}\|_2 \cdot\|\diag(x)\|_F \cdot\|\wt{A}^{\top}\|_2   \\
    & \cdot  ~ (\| f(A,x) - f(\wt{A},x)\|_2) \cdot(\|e_{j_0}\|_2 + \|f(A,x)\|_2)  \cdot \|x\|_2 \cdot \|{\bf 1}_d^{\top}\|_2 \\
    & + ~ 2|f(\wt{A},x)_{j_1} | \cdot|f(\wt{A},x)_{j_0}| \cdot\|c_g(\wt{A},x)^{\top}\|_2 \cdot\|\diag(x)\|_F \cdot\|\wt{A}^{\top}\|_2   \\
    & \cdot  ~ (\|e_{j_0} \|_2+\| f(\wt{A},x)\|_2) \cdot(\|f(A,x) - f(\wt{A},x)\|_2)  \cdot \|x\|_2 \cdot \|{\bf 1}_d^{\top}\|_2 \\
    \leq & ~ 80R^4 \beta^{-2} \cdot n \cdot \sqrt{d}\exp(3R^2) \cdot \|A - \wt{A}\|_F \\
    & + ~ 80R^4 \beta^{-2} \cdot n \cdot \sqrt{d}\exp(3R^2) \cdot \|A - \wt{A}\|_F \\
    & + ~160 R^4 \beta^{-2} \cdot n \cdot \sqrt{d}   \cdot  \exp(3R^{2}) \cdot \|A - \wt{A}\|_F \\
    & + ~40R^3 \cdot \sqrt{d}  \|A - \wt{A}\|_F \\
    & + ~40 R^4   \beta^{-2} \cdot n \cdot \sqrt{d}   \cdot  \exp(3R^{2}) \cdot \|A - \wt{A}\|_F \\
    & + ~40 R^4   \beta^{-2} \cdot n \cdot \sqrt{d}   \cdot  \exp(3R^{2}) \cdot \|A - \wt{A}\|_F \\
    \leq & ~ 360\beta^{-2} \cdot n \cdot \sqrt{d}   \cdot  \exp(5R^{2}) \cdot \|A - \wt{A}\|_F \\
    \leq & ~ \beta^{-2} \cdot n \cdot \sqrt{d}   \cdot  \exp(6R^{2}) \cdot \|A - \wt{A}\|_F 
\end{align*}

{\bf Proof of Part 12.}
\begin{align*}
 & ~ \| B_{7}^{j_1,*,j_0,*} (A) - B_{7}^{j_1,*,j_0,*} ( \wt{A} ) \|_F \\
   = & ~ \|\sum_{i = 1}^5 B_{7,i}^{j_1,*,j_0,*}(A) -   B_{7,i}^{j_1,*,j_0,*}(\wt{A})  \|_F \\
    \leq & ~ 11  \beta^{-2} \cdot n \cdot \sqrt{d} \cdot \exp(6R^2)\|A - \wt{A}\|_F
\end{align*}
\end{proof}
\subsection{PSD}
\begin{lemma}\label{psd: B_7}
If the following conditions hold
\begin{itemize}
    \item Let $B_{7,1,1}^{j_1,*, j_0,*}, \cdots, B_{7,5,1}^{j_1,*, j_0,*} $ be defined as Lemma~\ref{lem:b_7_j1_j0} 
    \item  Let $\|A \|_2 \leq R, \|A^{\top} \|_F \leq R, \| x\|_2 \leq R, \|\diag(f(A,x)) \|_F \leq \|f(A,x) \|_2 \leq 1, \| b_g \|_2 \leq 1$ 
\end{itemize}
We have 
\begin{itemize}
    \item {\bf Part 1.} $\| B_{7,1,1}^{j_1,*, j_0,*} \| \leq 12 \sqrt{d} R^2$  
    \item {\bf Part 2.} $\|B_{7,1,2}^{j_1,*, j_0,*}\| \preceq 12 \sqrt{d} R^2$
    \item {\bf Part 3.} $\|B_{7,1,3}^{j_1,*, j_0,*}\| \preceq 12  R^4$
    \item {\bf Part 4.} $\|B_{7,1,4}^{j_1,*, j_0,*} \|\preceq 12 R^4$
    \item {\bf Part 5.} $\|B_{7,1,5}^{j_1,*, j_0,*} \|\preceq 12  R^4$
    \item {\bf Part 6.} $\|B_{7,1,6}^{j_1,*, j_0,*} \|\preceq 18  R^4$
    \item {\bf Part 7.} $\|B_{7,1,7}^{j_1,*, j_0,*} \|\preceq 36  R^4$
    \item {\bf Part 8.} $\|B_{7,2,1}^{j_1,*, j_0,*}\| \preceq 30\sqrt{d} R^4$
    \item {\bf Part 9.} $\|B_{7,3,1}^{j_1,*, j_0,*}\| \preceq 30 \sqrt{d}R^2$
    \item {\bf Part 10.} $\|B_{7,4,1}^{j_1,*, j_0,*}\| \preceq 30\sqrt{d} R^4$
    \item {\bf Part 11.} $\|B_{7,5,1}^{j_1,*, j_0,*}\| \preceq 40\sqrt{d} R^4$
    \item {\bf Part 12.} $\|B_{7}^{j_1,*, j_0,*}\| \preceq 244\sqrt{d} R^4$
\end{itemize}
\end{lemma}

\begin{proof}
    {\bf Proof of Part 1.}
    \begin{align*}
        \| B_{7,1,1}^{j_1,*, j_0,*} \| 
        = & ~ \|  f_c(A,x)  \cdot f(A,x)_{j_1} \cdot f(A,x)_{j_0} \cdot  p_{j_0}(A,x)\cdot  {\bf 1}_d^\top\| \\
        \leq & ~ |f(A,x)_{j_1}| \cdot |f(A,x)_{j_0}| \cdot |f_c(A,x)|  \cdot  \|p_{j_0}(A,x)\|_2 \cdot \| {\bf 1}_d^\top\|_2\\
        \leq & ~ 12 \sqrt{d} R^2
    \end{align*}

    {\bf Proof of Part 2.}
    \begin{align*}
        \| B_{7,1,2}^{j_1,*, j_0,*} \|
        = &~
        \| f(A,x)_{j_1} \cdot f(A,x)_{j_0} \cdot c(A,x)_{j_1} \cdot  p_{j_0}(A,x) \cdot  {\bf 1}_d^\top  \| \\
        \preceq & ~  |f(A,x)_{j_1}| \cdot |f(A,x)_{j_0}| \cdot |c(A,x)_{j_1}|\cdot \|p_{j_0}(A,x)\|_2 \cdot \| {\bf 1}_d^\top\|_2\\
        \preceq & ~ 12 \sqrt{d} R^2
    \end{align*}

    {\bf Proof of Part 3.}
    \begin{align*}
     \| B_{7,1,3}^{j_1,*, j_0,*} \|
     = & ~
       \| f(A,x)_{j_1} \cdot  f(A,x)_{j_0} \cdot f_c(A,x) \cdot  ((A_{j_1,*}) \circ x^\top) \cdot p_{j_0}(A,x)  \cdot I_d \| \\
    \leq & ~ |f(A,x)_{j_1}| \cdot |f(A,x)_{j_0}| \cdot  |f_c(A,x)|     \cdot \| A_{j_1,*}^\top \circ x \|_2 \cdot \|p_{j_0}(A,x)\|_2 \cdot \| I_d\| \\
    \leq & ~ |f(A,x)_{j_1}| \cdot |f(A,x)_{j_0}| \cdot  |f_c(A,x)|     \cdot \| A_{j_1,*}\|_2 \cdot \|\diag(x) \|_{\infty} \cdot \|p_{j_0}(A,x)\|_2 \cdot \| I_d\| \\
    \leq &  ~  12 R^4
    \end{align*}
   
    {\bf Proof of Part 4.}
    \begin{align*}
        \|B_{7,1,4}^{j_1,*, j_0,*}  \|
        = & ~  \|-f(A,x)_{j_1} \cdot  f(A,x)_{j_0} \cdot f_c(A,x)   \cdot h(A,x)^\top \cdot p_{j_0}(A,x)  \cdot I_d \|\\
        \leq & ~   \| f(A,x)_{j_1} \cdot  f(A,x)_{j_0} \cdot f_c(A,x)   \cdot h(A,x)^\top \cdot  p_{j_0}(A,x) \cdot I_d \|\\
        \leq & ~ | f(A,x)_{j_1} | \cdot|  f(A,x)_{j_0} |\cdot |f_c(A,x)|  \cdot \|h(A,x)^\top\|_2 \cdot  \|p_{j_0}(A,x)\|_2  \cdot \| I_d\|\\
        \leq & ~ | f(A,x)_{j_1} | \cdot|  f(A,x)_{j_0} |\cdot |f_c(A,x)|  \cdot \|h(A,x)^\top\|_2 \cdot  \|p_{j_0}(A,x)\|_2 \cdot \| I_d\| \\
        \leq & ~ 12 R^4
    \end{align*}

    {\bf Proof of Part 5.}
    \begin{align*}
        \|B_{7,1,5}^{j_1,*, j_0,*} \|
        = & ~ \| f(A,x)_{j_1} \cdot  f(A,x)_{j_0} \cdot (-f_2(A,x) + f(A,x)_{j_1})    \cdot h(A,x)^\top \cdot  p_{j_0}(A,x) \cdot I_d\| \\
        \leq & ~  | f(A,x)_{j_1}| \cdot |f(A,x)_{j_0}|    \cdot (|-f_2(A,x)| + |f(A,x)_{j_1}|)  \cdot \|h(A,x)^\top\|_2  \cdot \|p_{j_0}(A,x)\|_2 \cdot \| I_d\|\\
        \leq & ~ 12 R^4
    \end{align*}

    {\bf Proof of Part 6.}
    \begin{align*}
        \|B_{7,1,6}^{j_1,*, j_0,*} \|
        = & ~ \| f(A,x)_{j_1} \cdot  f(A,x)_{j_0} \cdot (-f_c(A,x) + f(A,x)_{j_1})   \cdot h(A,x)^\top \cdot  p_{j_0}(A,x) \cdot I_d \|\\
        \leq & ~  | f(A,x)_{j_1}| \cdot |f(A,x)_{j_0}|    \cdot (|-f_c(A,x)| + |f(A,x)_{j_1}|)   \cdot \|h(A,x)^\top\|_2  \cdot \|p_{j_0}(A,x)\|_2 \cdot \| I_d\| \\
        \leq & ~ 18 R^4
    \end{align*}

    {\bf Proof of Part 7.}
    \begin{align*}
         \|B_{7,1,7}^{j_1,*, j_0,*} \|
         = & ~ \|f(A,x)_{j_1} \cdot  f(A,x)_{j_0}    \cdot p_{j_1}(A,x)^\top \cdot   p_{j_0}(A,x) \cdot I_d \|\\
         \leq & ~ | f(A,x)_{j_1}| \cdot |f(A,x)_{j_0}|  \cdot \|p_{j_1}(A,x)^\top\|_2  \cdot \|p_{j_0}(A,x)\|_2 \cdot \| I_d\|\\
         \leq & ~ 36  R^4
    \end{align*}

    {\bf Proof of Part 8.}
    \begin{align*}
        \|B_{7,2,1}^{j_1,*, j_0,*} \|
        = & ~ \|  f(A,x)_{j_1} \cdot  f(A,x)_{j_0}  \cdot c_g(A,x)^{\top} \cdot p_{j_0}(A,x) \cdot x \cdot {\bf 1}_d^{\top} \| \\
        \leq & ~  | f(A,x)_{j_1}| \cdot |f(A,x)_{j_0}|   \cdot \|c_g(A,x)^{\top}\|_2 \cdot \|p_{j_0}(A,x)\|_2 \cdot \| x\|_2 \cdot \| {\bf 1}_d^{\top}\|_2\\
        \leq & ~ 30 \sqrt{d} R^4
    \end{align*}

    {\bf Proof of Part 9.}
    \begin{align*}
    & ~\|B_{7,3,1}^{j_1,*, j_0,*} \|    \\
        = & ~ \|     - f(A,x)_{j_0}\cdot e_{j_1}^\top \cdot ((e_{j_0} - f(A,x)) \circ q(A,x)) \cdot {\bf 1}_d \cdot c_g(A,x)^{\top}   \cdot \diag(x)  \| \\
        \leq & ~ \|     f(A,x)_{j_0}\cdot e_{j_1}^\top \cdot ((e_{j_0} - f(A,x)) \circ q(A,x)) \cdot {\bf 1}_d \cdot c_g(A,x)^{\top}   \cdot \diag(x)  \| \\
        \leq & ~ | f(A,x)_{j_0}| \cdot  \|e_{j_1}^\top\|_2 \cdot  (\|e_{j_0}\|_2 + \|-f(A,x)\|_2) \cdot \| \diag(q(A,x))\|_{\infty} \cdot \| {\bf 1}_d\|_2 \cdot \|c_g(A,x)^{\top}\|_2 \cdot \| \diag (x)\|_{\infty}  \\
        \leq & ~ 30 \sqrt{d} R^2
    \end{align*}

        {\bf Proof of Part 10.}
    \begin{align*}
     \|B_{7,4,1}^{j_1,*, j_0,*} \|
        = & ~   \|  f(A,x)_{j_1} \cdot f(A,x)_{j_0}   \cdot  c_g(A,x)^{\top} \cdot p_{j_1}(A,x) \cdot x \cdot {\bf 1}_d^{\top}\| \\
        \leq & ~ | f(A,x)_{j_1}| \cdot  |f(A,x)_{j_0} | \cdot |c_g(A,x)^{\top}| \cdot \|p_{j_1}(A,x)  \|_2 \cdot \| x\|_2 \cdot \| {\bf 1}_d^{\top}\|_2\\
        \leq & ~ 30 \sqrt{d}R^4
    \end{align*}

       {\bf Proof of Part 11.}
    \begin{align*}
     & ~\|B_{7,5,1}^{j_1,*, j_0,*} \|\\
        = & ~ \|  -2f(A,x)_{j_1} \cdot f(A,x)_{j_0} \cdot   c_g(A,x)^{\top} \cdot \diag(x) \cdot A^{\top} \cdot  ((e_{j_0} - f(A,x)) \circ  ( (e_{j_1} - f(A,x)) ) )   \cdot x \cdot {\bf 1}_d^{\top}  \| \\
        \leq& ~ \|   2f(A,x)_{j_1} \cdot f(A,x)_{j_0} \cdot   c_g(A,x)^{\top} \cdot \diag(x) \cdot A^{\top} \cdot  ((e_{j_0} - f(A,x)) \circ  ( (e_{j_1} - f(A,x)) ) )   \cdot x \cdot {\bf 1}_d^{\top}  \| \\
        \leq & ~2 | f(A,x)_{j_1}|  \cdot  |f(A,x)_{j_0} |   \cdot \|c_g(A,x)^{\top}\|_2 \cdot \|\diag(x)\|_{\infty} \cdot \| A^{\top}\|_F \cdot(\|e_{j_0}\|_2 + \|-f(A,x)\|_2)\\
        & ~ \cdot \|\diag(e_{j_1} - f(A,x)) \|_{\infty}   \cdot \| x\|_2 \cdot \| {\bf 1}_d^{\top}\|_2\\
        \leq & ~ 40\sqrt{d} R^4
    \end{align*}

  {\bf Proof of Part 12.}
\begin{align*}
 & ~ \| B_{7}^{j_1,*,j_0,*}\| \\
   = & ~ \|\sum_{i = 1}^5 B_{7,i}^{j_1,*,j_0,*} \| \\
    \leq & ~  244\sqrt{d} R^4
\end{align*}
\end{proof}

\section{Lipschitz Property for Hessian}\label{app:lipschitz_hessian}
\begin{lemma}[Formal version of Lemma~\ref{lem:lipschitz:informal}]\label{lem:lipschitz:formal}
    If the following conditions hold
\begin{itemize}
 \item Let $\| B_{1}^{j_1,*,j_0,*} (A) - B_{1}^{j_1,*,j_0,*} ( \wt{A} ) \|_F$ be defined as {\bf Part 16} of Lemma~\ref{lips: B_1} 
 \item Let $\| B_{2}^{j_1,*,j_0,*} (A) - B_{2}^{j_1,*,j_0,*} ( \wt{A} ) \|_F$ be defined as {\bf Part 16} of Lemma~\ref{lips: B_2} 
 \item Let $\| B_{3}^{j_1,*,j_0,*} (A) - B_{3}^{j_1,*,j_0,*} ( \wt{A} ) \|_F$ be defined as {\bf Part 16} of Lemma~\ref{lips: B_3} 
 \item Let $\| B_{4}^{j_1,*,j_0,*} (A) - B_{4}^{j_1,*,j_0,*} ( \wt{A} ) \|_F$ be defined as {\bf Part 16} of Lemma~\ref{lips: B_4} 
 \item Let $\| B_{5}^{j_1,*,j_0,*} (A) - B_{5}^{j_1,*,j_0,*} ( \wt{A} ) \|_F$ be defined as {\bf Part 16} of Lemma~\ref{lips: B_5} 
 \item Let $\| B_{6}^{j_1,*,j_0,*} (A) - B_{6}^{j_1,*,j_0,*} ( \wt{A} ) \|_F$ be defined as {\bf Part 15} of Lemma~\ref{lips: B_6} 
    \item Let $\| B_{7}^{j_1,*,j_0,*} (A) - B_{7}^{j_1,*,j_0,*} ( \wt{A} ) \|_F$ be defined as {\bf Part 12} of Lemma~\ref{lips: B_7} 
\end{itemize}
Then, we have
\begin{itemize}
    \item {\bf Part 1.}
    \begin{align*}
       \| H^{j_1,*,j_0,*} (A) -  H^{j_1,*,j_0,*} ( \wt{A} ) \|_F 
       \leq & ~  \beta^{-2} \cdot n \cdot \sqrt{d} \cdot \exp(7R^2)\|A - \wt{A}\|_F 
    \end{align*}
\end{itemize}
\end{lemma}
\begin{proof}
    {\bf Proof of Part 1.}
\begin{align*}
    & ~ \| H^{j_1,*,j_0,*} (A) - H^{j_1,*,j_0,*} ( \wt{A} ) \|_F \\
       \leq& ~ \| B_{1}^{j_1,*,j_0,*} (A) - B_{1}^{j_1,*,j_0,*} ( \wt{A} ) \|_F 
          +\| B_{2}^{j_1,*,j_0,*} (A) - B_{2}^{j_1,*,j_0,*} ( \wt{A} ) \|_F \\
        & ~ +\| B_{3}^{j_1,*,j_0,*} (A) - B_{3}^{j_1,*,j_0,*} ( \wt{A} ) \|_F 
        +\| B_{4}^{j_1,*,j_0,*} (A) - B_{4}^{j_1,*,j_0,*} ( \wt{A} ) \|_F \\
       & ~ +\| B_{5}^{j_1,*,j_0,*} (A) - B_{5}^{j_1,*,j_0,*} ( \wt{A} ) \|_F 
        + \| B_{6}^{j_1,*,j_0,*} (A) - B_{6}^{j_1,*,j_0,*} ( \wt{A} ) \|_F\\
       & ~ +\| B_{7}^{j_1,*,j_0,*} (A) - B_{7}^{j_1,*,j_0,*} ( \wt{A} ) \|_F\\
        \leq& ~ 12  \beta^{-2} \cdot n \cdot \sqrt{d} \cdot \exp(5R^2)\|A - \wt{A}\|_F 
         +9 \beta^{-2} \cdot n \cdot \sqrt{d}  \exp(5R^2)\|A - \wt{A}\|_F\\
        & ~ +12\beta^{-2} \cdot n \cdot \sqrt{d} \exp(6R^2)\|A- \wt{A}\|_F 
         +14\beta^{-2} \cdot n \cdot \sqrt{d}  \exp(6R^2)\|A - \wt{A}\|_F   \\
        & ~ + 15\beta^{-2} \cdot n \cdot \sqrt{d}  \exp(6R^2)\|A - \wt{A}\|_F   
         +15\beta^{-2} \cdot n \cdot \sqrt{d}  \exp(6R^2)\|A - \wt{A}\|_F   \\
        & ~ +11\beta^{-2} \cdot n \cdot \sqrt{d}  \exp(6R^2)\|A - \wt{A}\|_F   \\
        \leq & ~88\beta^{-2} \cdot n \cdot \sqrt{d}  \exp(6R^2)\|A - \wt{A}\|_F \\
        \leq& ~   \beta^{-2} \cdot n \cdot \sqrt{d} \cdot \exp(7R^2)\|A - \wt{A}\|_F 
\end{align*}
\end{proof}

\section{Positive Definite Property for Hessian}\label{app:psd_hessian}
\begin{lemma}[Formal version of Lemma~\ref{lem:psd:informal}]\label{lem:psd:formal}
    If the following conditions hold
\begin{itemize}
 \item Let $\| B_{1}^{j_1,*,j_0,*} \|$ be defined as {\bf Part 13} of Lemma~\ref{psd: B_1} 
 \item Let $\| B_{2}^{j_1,*,j_0,*} \|$ be defined as {\bf Part 10} of Lemma~\ref{psd: B_2} 
 \item Let $\| B_{3}^{j_1,*,j_0,*} \|$ be defined as {\bf Part 13} of Lemma~\ref{psd: B_3} 
 \item Let $\| B_{4}^{j_1,*,j_0,*} \|$ be defined as {\bf Part 15} of Lemma~\ref{psd: B_4} 
 \item Let $\| B_{5}^{j_1,*,j_0,*} \|$ be defined as {\bf Part 16} of Lemma~\ref{psd: B_5} 
 \item Let $\| B_{6}^{j_1,*,j_0,*} \|$ be defined as {\bf Part 16} of Lemma~\ref{psd: B_6} 
 \item Let $\| B_{7}^{j_1,*,j_0,*} \|$ be defined as {\bf Part 12} of Lemma~\ref{psd: B_7} 
\end{itemize}
Then, we have
\begin{itemize}
    \item {\bf Part 1.}
    \begin{align*}
       \| H^{j_1,*,j_0,*} \|
       \leq & ~  \sqrt{d} R^5 
    \end{align*}
    \item {\bf Part 2.} 
    \begin{align*}
        \|  H  \| \leq &~  n^2 \sqrt{d} R^5 
    \end{align*}
    \item {\bf Part 3.}
    \begin{align*}
        H \succeq - n^2 \sqrt{d} R^5 \cdot I_{nd}
    \end{align*}
    \item  {\bf Part 4.} Let $l = -  n^2 \sqrt{d} R^5 + 2\tau$, where $\tau \leq \frac{n^2 \sqrt{d} R^5}{2}$
    \begin{align*}
        \nabla_A^2 L_g(A,x) + \nabla_A^2 L_{\rm reg}
       \succeq & ~  l \cdot I_{nd}
    \end{align*}
\end{itemize}
\end{lemma}
\begin{proof}

"
    {\bf Proof of Part 1.}
\begin{align*}
    & ~ \| H^{j_1,*,j_0,*} \| \\
       \leq& ~ \| B_{1}^{j_1,*,j_0,*} \|
          +\| B_{2}^{j_1,*,j_0,*}  \| 
          +\| B_{3}^{j_1,*,j_0,*} \|
        +\| B_{4}^{j_1,*,j_0,*}   \|
        +\| B_{5}^{j_1,*,j_0,*}  \|
        + \| B_{6}^{j_1,*,j_0,*}   \|
        +\| B_{7}^{j_1,*,j_0,*} \|\\
        \leq& ~  78\sqrt{d} R^2 
         + 53\sqrt{d} R^2
         +78\sqrt{d} R^4 
         + 108\sqrt{d} R^4   
         + 103\sqrt{d} R^4   
         +147\sqrt{d} R^4 
         +244\sqrt{d} R^4   \\
        \leq & ~811\sqrt{d} R^4 \\
        \leq& ~   \sqrt{d} R^5 
\end{align*}

{\bf Proof of Part 2.} 
\begin{align*}
    \| H \| \leq & ~ n^2 \| H^{j_1,*,j_0,*} \| \\
    \leq & ~  n^2 \sqrt{d} R^5 
\end{align*}

{\bf Proof of Part 3.}
\begin{align*}
    H \succeq ~ -  n^2 \sqrt{d} R^5 \cdot I_{nd \times nd}
\end{align*}

{\bf Proof of Part 4.}
\begin{align*}
    \nabla_A^2 L_g(A,x) + \nabla_A^2 L_{\rm reg} 
    = & ~ H + 2\tau \cdot I_{nd \times nd} \\
    \succeq  & ~ l \cdot I_{nd}
\end{align*}
\end{proof}

\section{Approximate Newton Method: Main Result of Attack}\label{app:main_result}

Now we show our main result as follows:

\begin{theorem}[Main result, formal version of Theorem~\ref{thm:main_result:informal}]\label{thm:main_result:formal}
Given an observed gradient vector $b_g \in \R^d$. We let $L_g(A, x)$ be defined as Definition~\ref{def:attack}. We denote the optimal solution for $L_g(A, x)$ as $A^*$.

Next, we choose a good initial point $A_0$ that is close enough to $A^*$. Assume that there exists a scalar $R > 1$ such that $\|A_t\| \leq F$ for any $t \in [T]$.

Then, for any accuracy parameter $\epsilon \in (0,0.1)$ and a failure probability $\delta \in (0,0.1)$, an algorithm based on the Newton method can be employed to recover the initial data. The result of this algorithm guarantees that within $T = O(\log(\|A_0 - A^*\|_F / \epsilon))$ executions, it outputs a matrix $\Tilde{A} \in \mathbb{R}^{n \times d}$ satisfying $\|\wt{A} - A^*\|_F \leq \epsilon$ with a probability of at least $1 - \delta$.
\end{theorem}

\subsection{Definition and Update Rule}\label{sec:newton:definitions}

In this section, we introduce the basic definitions. We focus on analyzing the following function:
\begin{align*}
    \min_{x \in \R^d } L(x).
\end{align*}

Here, we give the definition of $(l,M)$-good Loss function.

\begin{definition}[$l$-local Minimum]\label{def:local_min}

    Let $L : \R^d \rightarrow \R$ be a function.

    Let $l > 0$ be a positive real number. 

    There exists $x^* \in \R^d$ such that
    \begin{align*}
        \nabla L(x^*) = {\bf 0}_d
    \end{align*}
    and
    \begin{align*}
        \nabla^2 L(x^*) \succeq l \cdot I_d.
    \end{align*}
\end{definition}

\begin{definition}[Hessian is $M$-Lipschitz]\label{def:hessian_lipschitz}
    Let $L : \R^d \rightarrow \R$ be a function.

    Let $M > 0$ be a positive real number. 

    The Hessian of $L$ is $M$-Lipschitz if
    \begin{align*}
        \| \nabla^2 L(y) - \nabla^2 L(x) \| \leq M \cdot \| y - x \|_2 
    \end{align*}
\end{definition}

\begin{definition}[Good Initialization Point]\label{def:good_point}
    Let $L : \R^d \rightarrow \R$ be a function.

    Let $x_0 \in \R^d$ be the initialization point.

    Let $x^* \in \R^d$.

    Let $r_0 = \|x_0 - x^*\|_2 \in \R$.

    Let $M > 0$ be a positive real number. 

    If $r_0$ satisfy $r_0 M \leq 0.1 l$, then we say $x_0$ is a good initialization point .
\end{definition}

\begin{definition}[$(l,M)$-good Loss function]\label{def:f_ass}

Let $L : \R^d \rightarrow \R$ be a function.

$L$ is called a $(l,M)$-good function if it satisfies $l$-local Minimum (Definition~\ref{def:local_min}), Hessian is $M$-Lipschitz (Definition~\ref{def:hessian_lipschitz}), and good initialization point (Definition~\ref{def:good_point}).

\end{definition}

Then, we define the gradient and Hessian.

\begin{definition}[Gradient and Hessian]\label{def:gradient_hessian}
Let $L : \R^d \rightarrow \R$ be a function.

The function $g : \R^d \rightarrow \R^d$, which is defined as
\begin{align*}
    g(x) := \nabla L(x),
\end{align*}
is called the gradient of the function $L$.

The function $H : \R^d \rightarrow \R^{d \times d}$, which is defined as
\begin{align*}
    H(x) := \nabla^2 L(x),
\end{align*}
is called the Hessian of the function $L$.

\end{definition}

After that, we introduce the definition of the exact update of Newton's method.
\begin{definition}[Exact update of the Newton method]\label{def:exact_update_variant}
\begin{align*}
    x_{t+1} = x_t - H(x_t)^{-1} \cdot g(x_t)
\end{align*}
\end{definition}

\subsection{Approximate of Hessian and Update Rule}\label{sec:newton:approximation}

In this section, we introduce the approximate of Hessian and update rule. We define the approximate Hessian as follows.

\begin{definition}[Approximate Hessian]\label{def:wt_H}

Let $H$ be defined as in Definition~\ref{def:gradient_hessian}.

We define $\wt{H}: \R^d \rightarrow \R^{d \times d}$ to be a function satisfying:
\begin{align*}
 (1-\epsilon_0) \cdot H(x_t) \preceq \wt{H}(x_t) \preceq (1+\epsilon_0) \cdot H(x_t) .
\end{align*}
\end{definition}

$\wt{H}$ is the approximated Hessian. We apply Lemma~4.5 in \cite{dsw22} to get the approximated Hessian.

\begin{lemma}[\cite{syyz22,dsw22}]\label{lem:subsample}

Let $\epsilon_0 = 0.01 \in \R$, which is called the constant precision parameter. 

Let $A \in \R^{n \times d}$.

Let $D \in \R^{n \times n}$ be an arbitrary positive diagonal matrix.

There is an algorithm that can run in a time complexity of approximately
\begin{align*}
O( (\nnz(A) + d^{\omega} ) \poly(\log(n/\delta)) ).
\end{align*}

This algorithm produces a sparse diagonal matrix $\wt{D}$ which is in $\R^{n \times n}$.

The key property of is that it satisfies the following equation:
\begin{align*}
(1- \epsilon_0) A^\top D A \preceq A^\top \wt{D} A \preceq (1+\epsilon_0) A^\top D A.
\end{align*}

It is worth noting that $\omega$ represents the exponent related to matrix multiplication, with an approximate value of $\omega \approx 2.373$ (see \cite{lg14,w12,aw21}).

\end{lemma}

Following the standard in the literature on Approximate Newton Hessian, as in \cite{a00, jkl+20, bpsw21, szz21, hjs+22, lsz23}, we take into account the following.

\begin{definition}[Approximate update]\label{def:update_x_k+1}

We examine the following procedure:
\begin{align*}
    x_{t+1} = x_t  - \wt{H}(x_t)^{-1} \cdot  g(x_t)  .
\end{align*}
\end{definition}

We present a technique from previous research.
\begin{lemma}[Iterative shrinking Lemma, Lemma 6.9 of \cite{lsz23}]\label{lem:one_step_shrinking}

Let $L : \R^d \rightarrow \R$ be a $(l,M)$-good function, as defined in Definition~\ref{def:f_ass}.

Let $\epsilon_0$ be a positive real number in $(0,0.1)$.

Let $x_t, x^* \in \R^d$ and $r_t:= \| x_t - x^* \|_2 \in \R$.

Let $M > 0$ be a positive real number and $\ov{r}_t: = M \cdot r_t$.

Then, 
\begin{align*}
r_{t+1} \leq 2 \cdot (\epsilon_0 + \ov{r}_t/( l - \ov{r}_t ) ) \cdot r_t.
\end{align*}

\end{lemma}

To represent the total number of iterations of the algorithm, let's use the symbol $T$. In order to utilize Lemma~\ref{lem:one_step_shrinking}, we need the following induction hypothesis. This lemma is a well-established concept in the literature in \cite{lsz23}.

\begin{lemma}[Induction hypothesis, Lemma 6.10 on page 34 of \cite{lsz23}]\label{lem:newton_induction}

Let $i$ be an arbitrary element in $[t]$.

Let $x_i, x^* \in \R^d$ and $r_i:= \| x_i - x^* \|_2 \in \R$.

Let $\epsilon_0$ be a positive real number in $(0,0.1)$.

Let $0.4 \cdot r_{i-1} \geq r_{i}$.

Let $0.1 l \geq M \cdot r_i$, where $M > 0$.

Then,
\begin{itemize}
    \item $0.4 r_t \geq r_{t+1}$
    \item $0.1 l \geq M \cdot r_{t+1}$
\end{itemize}

\end{lemma}
\section{Differential Privacy of Laplacian Perturbation}\label{app:diff_privacy}

\subsection{Definitions}

\begin{definition}[Laplacian distribution noise vector]\label{def:laplace_noise}
    We say a noise $\xi \in \R^{d}$ vector is sampled from ${\rm Laplace}(\mu, b)$, where the $i$-th element of $\xi_i$ that satisfies 
    \begin{align*}
        \Pr[x=\xi_i] = \frac{1}{2b} \exp( - \frac{| x - \mu |}{b})
    \end{align*}
\end{definition}

\begin{definition}[$\ell_1$ sensitivity]\label{def:Delta}
    Let $g(A, x) \in \R^d$ be defined as Definition~\ref{def:g}, given $A \in \R^{n \times d}$ and $A' \in \R^{n \times d}$ are neighbouring batch data. We define
    \begin{align*}
        \Delta = \max_{A, A'} \| g(A, x) - g(A', x) \|_1
    \end{align*}
\end{definition}

\begin{definition}[Perturbed gradient]\label{def:M}
    Let $g(A, x) \in \R^d$ be defined as Definition~\ref{def:g}, we sample $\xi$ from ${\rm Laplace}(\mu, b)$ as Definition~\ref{def:laplace_noise}, we define
    \begin{align*}
        \mathcal{M}(A) := g(A, x) + \xi
    \end{align*}
\end{definition}

\subsection{Differential Privacy}

\begin{theorem}[Formal version of Theorem~\ref{thm:differential_privacy:informal}]\label{thm:differential_privacy:formal}
    We denote $\varepsilon > 0$ as our privacy cost, let $\Delta$ be denoted as Definition~\ref{def:Delta}. We sample $\xi \in \R^d$ from ${\rm Laplace}(0, \frac{\Delta}{\varepsilon})$ as Definition~\ref{def:laplace_noise}. Let perturbed gradient be defined as $\mathcal{M}(A) \in \R^d$, denote $b_g \in \R^d$ as the observed gradient. Suppose given $A \in \R^{n \times d}$ and $A' \in \R^{n \times d}$ are neighboring batch data. We have
    \begin{align*}
        \frac{\Pr[\mathcal{M}(A) = b_g]}{\Pr[\mathcal{M}(A') = b_g]} \leq \exp(\varepsilon)
    \end{align*}
\end{theorem}

\begin{proof}
    We have
    \begin{align*}
        \frac{\Pr[\mathcal{M}(A) = b_g]}{\Pr[\mathcal{M}(A') = b_g]}
        = & ~ \frac{\prod_{i=1}^d \frac{\varepsilon}{2\Delta} \exp( - \frac{ \varepsilon | {b_g}_i - g(A, x)_i |}{\Delta})}{\prod_{i=1}^d \frac{\varepsilon}{2\Delta} \exp( - \frac{ \varepsilon | {b_g}_i - g(A', x)_i |}{\Delta})} \\
        = & ~ \frac{\prod_{i=1}^d  \exp( - \frac{ \varepsilon | {b_g}_i - g(A, x)_i |}{\Delta})}{\prod_{i=1}^d  \exp( - \frac{ \varepsilon | {b_g}_i - g(A', x)_i |}{\Delta})} \\
        = & ~ \frac{\exp( - \frac{ \varepsilon \sum_{i=1}^d | {b_g}_i - g(A, x)_i |}{\Delta})}{\exp( - \frac{ \varepsilon  \sum_{i=1}^d | {b_g}_i - g(A', x)_i |}{\Delta})} \\
        = & ~ \exp( - \frac{ \varepsilon \sum_{i=1}^d | {b_g}_i - g(A, x)_i |}{\Delta} + \frac{ \varepsilon  \sum_{i=1}^d | {b_g}_i - g(A', x)_i |}{\Delta}) \\
        = & ~ \exp( \frac{ \varepsilon \sum_{i=1}^d (| {b_g}_i - g(A, x)_i | - | {b_g}_i - g(A', x)_i | )}{\Delta} ) \\
        \leq & ~ \exp( \frac{ \varepsilon  \| g(A, x) - g(A', x) \|_1}{\Delta} ) \\
        = & ~ \exp( \frac{ \varepsilon  \Delta}{\Delta} ) \\
        = & ~  \exp(\varepsilon)
    \end{align*}
    where the first step follows from Lemma~\ref{lem:pr_M}, the second, third, fourth and fifth steps follow from simple algebras, the sixth step follows from triangle inequality, the seventh step follows from Definition~\ref{def:Delta}, the last step follows from simple algebra.
\end{proof}

\begin{lemma}\label{lem:pr_M}
    We denote $\varepsilon > 0$ as our privacy cost, let $\Delta$ be denoted as Definition~\ref{def:Delta}. We sample $\xi \in \R^d$ from ${\rm Laplace}(0, \frac{\Delta}{\varepsilon})$ as Definition~\ref{def:laplace_noise}. Let perturbed gradient be defined as $\mathcal{M}(A) \in \R^d$, denote $b_g \in \R^d$ as the observed gradient, we have
    \begin{align*}
        \Pr[\mathcal{M}(A) = b_g] = \prod_{i=1}^d \frac{\varepsilon}{2\Delta} \exp( - \frac{ \varepsilon | {b_g}_i - \mathcal{M}(A)_i |}{\Delta})
    \end{align*}
\end{lemma}

\begin{proof}
    We have
    \begin{align*}
        \Pr[\mathcal{M}(A) = b_g]
        = & ~ \Pr[g(A, x) + \xi = b_g] \\
        = & ~ \Pr[\xi = b_g - g(A, x)] \\
        = & ~ \prod_{i=1}^d \Pr[\xi_i = {b_g}_i - g(A, x)_i] \\
        = & ~ \prod_{i=1}^d \frac{\varepsilon}{2\Delta} \exp( - \frac{ \varepsilon | {b_g}_i - g(A, x)_i |}{\Delta})
    \end{align*}
    where the first step follows from Definition~\ref{def:M}, the second and third steps follow from simple algebras, the last step follows from Definition~\ref{def:laplace_noise} and $\xi \in {\rm Laplace}(0, \frac{\Delta}{\varepsilon})$.
\end{proof}

\subsection{Computation of \texorpdfstring{$\ell_1$ sensitivity}{}}

\begin{lemma}[Formal version of Lemma~\ref{lem:ell_sensitivity:informal}]\label{lem:ell_sensitivity:formal}
    Let $\Delta$ be defined as Definition~\ref{def:Delta}, we have
    \begin{align*}
        \Delta \leq & ~ 20 \beta^{-2} n^{1.5} d \exp(10R^2) 
    \end{align*}
\end{lemma}

\begin{proof}
    We have
    \begin{align*}
        \Delta
        = & ~ \max_{A, A'} \| g(A, x) - g(A', x) \|_1 \\
        \leq & ~ \max_{A, A'} \sqrt{d} \| g(A, x) - g(A', x) \|_2 \\
        \leq & ~ 10 \beta^{-2} n^{1.5} d \exp(6R^2) \max_{A, A'} \| A - A' \|_2 \\
        \leq & ~ 10 \beta^{-2} n^{1.5} d \exp(6R^2) \max_{A, A'} \{\| A \|_2 + \| A' \|_2\} \\
        \leq & ~ 20 \beta^{-2} n^{1.5} d \exp(10R^2) 
    \end{align*}
    where the first step follows from Definition~\ref{def:Delta}, the second step follows from Fact~\ref{fac:vector_norm}, the third step follows from Lemma~\ref{lem:lipschitz_grad}, the fourth step follows from Fact~\ref{fac:vector_norm}, the last step follows from simple algebra.
\end{proof}

\section{Convergence of Perturbed Gradient}\label{app:convergence}

\subsection{Definitions}

\begin{definition}[SGD]
    We run SGD algorithm for $T$ steps to optimize $x$. Let $\mathcal{M}(A)$ be defined as Definition~\ref{def:M}, we initialize $x_0$, denote the learning rate as $\eta$. For $t in [T]$, given gradient $g(A_t, x_t)$, then we can optimize $x_t$ as follows:
    \begin{align*}
        x_{t+1} = x_t - \eta g(A_t, x_t)
    \end{align*}
\end{definition}

\begin{definition}[Loss function: normal version]\label{def:L}
    Let loss function for SGD algorithm $L_f(A, x)$ be defined as Definition~\ref{def:L_exp}, we define
    \begin{align*}
        L(x) = \frac{1}{N} \sum_{i=1}^N L_f(A_i, x) 
    \end{align*}
    where $A_i$ is the $i$-th batch data, $N$ represents the size of dataset.
\end{definition}

\begin{definition}[Perturbed gradient: normal version]\label{def:M_1}
    Let loss function $L(x)$ be defined as Definition~\ref{def:L}. Let $\xi$ be defined as Definition~\ref{def:laplace_noise}, let $g(A, x)$ be defined as Definition~\ref{def:g}, we define perturbed gradient as follows:
    \begin{align*}
        \mathcal{M}(x) = & ~ \frac{1}{N} \sum_{i=1}^N \mathcal{M}(A_i, x) \\
        = & ~ \frac{1}{N} \sum_{i=1}^N g(A_i, x) + \xi^{(i)}
    \end{align*}
    where $A_i$ is the $i$-th batch data, $N$ represents the size of dataset, $\xi^{(i)}$ is the $i$-th noise sampled from ${\rm Laplace}(0, b)$.
\end{definition}

\subsection{Convergence Guarantee}

\begin{theorem}[Formal version of Theorem~\ref{thm:training_convergence:informal}]\label{thm:training_convergence:formal}
    Suppose for all $A_i \in \R^{n \times d}$ and $b \in \R^d$ in dataset satisfies $\| A \|_F \leq R$ and $\|b \|_1 \leq 1$. Denote $\beta \in (0, 0.1)$ and $R \geq 4$. Let $L(x)$ be defined as Definition~\ref{def:L}. Let $\mathcal{M}(x)$ be defined as Definition~\ref{def:M_1}. Denote $\gamma := 7R\beta^{-2}n^{1.5}\exp(3R^2)$. Denote $\epsilon > 0$ as the training error. We run SGD algorithm with learning rate $\eta = \sqrt{\frac{L(x_1) - L(x^*)}{T(8\gamma R^2 + d \gamma (\frac{\Delta}{\varepsilon})^2)}}$, at most 
    \begin{align*}
        T = \frac{4 (L(x_1) - L(x^*)) \cdot (8\gamma R^2 + d \gamma (\frac{\Delta}{\varepsilon})^2)}{\epsilon^2}
    \end{align*}
    iterations, we have
    \begin{align*}
        \min_t \{ (\nabla L(x_t))^2 \} \leq \epsilon
    \end{align*}
\end{theorem}

\begin{proof}
    We have
    \begin{align}\label{eq:ineq_nabla_L}
        L(x_{t+1}) 
        \leq & ~ L(x_t) + \langle \nabla  L(x_t), x_{t+1} - x_t \rangle + \frac{\gamma}{2} \| x_{t+1} - x_t \|_2^2 \notag \\
        = & ~ L(x_t) - \langle \nabla  L(x_t), \eta \mathcal{M}(A_t, x_t) \rangle + \frac{\gamma}{2} \| \eta \mathcal{M}(A_t, x_t) \|_2^2 \notag \\
        = & ~ \E[L(x_t) - \langle \nabla  L(x_t), \eta \mathcal{M}(A_t, x_t) \rangle + \frac{\gamma}{2} \| \eta \mathcal{M}(A_t, x_t) \|_2^2] \notag  \\
        = & ~ L(x_t) - \E[\langle \nabla  L(x_t), \eta \mathcal{M}(A_t, x_t) \rangle + \E[\frac{\gamma}{2} \| \eta \mathcal{M}(A_t, x_t) \|_2^2] \notag \\
        = & ~ L(x_t) - \eta \| \nabla L(x_t) \|_2^2 + \E[\frac{\gamma}{2} \| \eta (g(A_t, x_t) + \xi^{(t)}) \|_2^2] \notag \\
        = & ~ L(x_t) - \eta \| \nabla L(x_t) \|_2^2 + \E[\frac{\gamma}{2}\eta^2 ( \| g(A_t, x_t) \|_2^2 + 2\langle g(A_t, x_t), \xi_t \rangle + \|\xi^{(t)} \|_2^2)] \notag \\
        = & ~ L(x_t) - \eta \| \nabla L(x_t) \|_2^2 + \E[\frac{\gamma}{2}\eta^2  \| g(A_t, x_t) \|_2^2] + \E[\gamma\eta^2\langle g(A_t, x_t), \xi^{(t)} \rangle] + \E[\frac{\gamma}{2}\eta^2\|\xi^{(t)} \|_2^2] \notag \\
        = & ~ L(x_t) - \eta \| \nabla L(x_t) \|_2^2 + \E[\frac{\gamma}{2}\eta^2  \| g(A_t, x_t) \|_2^2] + \E[\frac{\gamma}{2}\eta^2\|\xi^{(t)} \|_2^2] \notag \\
        = & ~ L(x_t) - \eta \| \nabla L(x_t) \|_2^2 + \E[\frac{\gamma}{2}\eta^2  \| g(A_t, x_t) \|_2^2] + d \gamma\eta^2 (\frac{\Delta}{\varepsilon})^2 \notag \\
        \leq & ~ L(x_t) - \eta \| \nabla L(x_t) \|_2^2 + 8\gamma\eta^2 R^2 + d \gamma\eta^2 (\frac{\Delta}{\varepsilon})^2
    \end{align}
    where the first step follows from Corollary~\ref{cor:lipschitz}, the second step follows from $x_{t+1} = x_t - \eta \mathcal{M}(A)$, the third step follows from expectation rule, the fourth step follows from simple algebra, the fifth step follows from Definition~\ref{def:M_1} and $\E[g(A_t, x)] = L(x)$, the sixth and seventh steps follow from simple algebras, the eighth step follows from $\E[\xi] = 0$, the ninth step follows from Lemma~\ref{lem:E_xi}, the last step follows from Part 5 of Lemma~\ref{lem:upper_bound}.

    Hence, we can show that
    \begin{align*}
        \min_t \{ \| \nabla L(x_t) \|_2^2 \} 
        \leq & ~ \frac{1}{T} \sum_{t=1}^T  \| \nabla L(x_t) \|_2^2  \\
        \leq & ~ \frac{1}{T} \sum_{t=1}^T \big( \frac{1}{\eta}(L(x_t) - L(x_{t+1})) + 8\gamma\eta R^2 + d \gamma\eta (\frac{\Delta}{\varepsilon})^2 \big) \\
        = & ~ \frac{1}{\eta T}(L(x_1) - L(x^*)) + 8\gamma\eta R^2 + d \gamma\eta (\frac{\Delta}{\varepsilon})^2  \\
        = & ~ 2 \frac{1}{\sqrt{T}}\sqrt{ (L(x_1) - L(x^*)) \cdot (8\gamma R^2 + d \gamma (\frac{\Delta}{\varepsilon})^2) }
    \end{align*}
    where the first step follows from simple algebra, the second step follows from Eq.~\eqref{eq:ineq_nabla_L}, the third step follows from simple algebra, the last step follows from $\eta = \sqrt{\frac{L(x_1) - L(x^*)}{T(8\gamma R^2 + d \gamma (\frac{\Delta}{\varepsilon})^2)}}$.
\end{proof}

\section{Core Tool: Helpful Lemmas for Convergence Proof}\label{app:core_tool}

\subsection{Lipschitz Property for \texorpdfstring{$g(A_i, x)$}{}}

\begin{lemma}\label{lem:lipschitz_g}
    If the given conditions are satisfied
    \begin{itemize}
        \item Let $A \in \R^{n \times d}$
        \item Let $b \in \R^{n}$ satisfy that $\| b \|_1 \leq 1$
        \item Let $\beta \in (0, 0.1)$
        \item Let $R \geq 4$
        \item $\| A \| \leq R$
        \item Let $R_f := \beta^{-2} n^{1.5} \exp(3R^2)$ 
        \item Let $g(A, x)$ be defined as Definition~\ref{def:g}
    \end{itemize}
    For all $x, y \in \R^d$, we have
    \begin{align*}
        \| g(A, x) - g(A, y) \|_2 \leq 7R R_f\| x - y \|_2
    \end{align*}
\end{lemma}

\begin{proof}
    We have
    \begin{align*}
        & ~ \| g(A, x) - g(A, y) \|_2 \\
        = & ~ \| A^\top \cdot \Big( -f(A,x) \langle c(A, x), f(A, x) \rangle +  \diag(f(A, x))  c(A,x) \Big) \\
        & ~ - A^\top \cdot \Big( -f(A,y) \langle c(A, y), f(A, y) \rangle +  \diag(f(A, y))  c(A,y) \Big) \|_2 \\
        \leq & ~ \| A \|_F \cdot \| f(A,x) \langle c(A, x), f(A, x) \rangle +  \diag(f(A, x))  c(A,x)\\
        & ~ +f(A,y) \langle c(A, y), f(A, y) \rangle -  \diag(f(A, y))  c(A,y)  \|_2 \\
        \leq & ~ \| A \|_F \cdot ( \| f(A,x) \langle c(A, x), f(A, x) \rangle - f(A,y) \langle c(A, y), f(A, y) \rangle\|_2 \\
        & ~ +  \|\diag(f(A, x))  c(A,x) -  \diag(f(A, y))  c(A,y)  \|_2 ) \\
        \leq & ~ \| A \|_F \cdot ( 5R_f \| x - y \|_2 \\
        & ~ +  \|\diag(f(A, x))  c(A,x) -  \diag(f(A, y))  c(A,y)  \|_2 ) \\
        \leq & ~ \| A \|_F \cdot ( 5R_f \| x - y \|_2 +  2R_f \| x - y \|_2 ) \\
        \leq & ~ \| A \|_F \cdot 7R_f \| x - y \|_2 \\
        \leq & ~ 7 RR_f \| x - y \|_2
    \end{align*}
    where the first step follows from Definition~\ref{def:g}, the second and third steps follow from Fact~\ref{fac:vector_norm}, the fourth step follows from Lemma~\ref{lem:lipschitz_f_c_f}, the fifth step follows from Lemma~\ref{lem:lipschitz_diag_f_c}, the last step follows from $\|A\|_F \leq R$.
\end{proof}

\begin{lemma}\label{lem:lipschitz_f_c_f}
    If the given conditions are satisfied
    \begin{itemize}
        \item Let $A \in \R^{n \times d}$
        \item Let $b \in \R^{n}$ satisfy that $\| b \|_1 \leq 1$
        \item Let $\beta \in (0, 0.1)$
        \item Let $R \geq 4$
        \item $\| A \| \leq R$
        \item Let $R_f := \beta^{-2} n^{1.5} \exp(3R^2)$ 
        \item Denote $f(A, x)$ as Definition~\ref{def:f}
        \item Denote $c(A, x)$ as Definition~\ref{def:c}
    \end{itemize}
    For all $x, y \in \R^d$, we have
    \begin{align*}
        \| f(A,x) \langle c(A, x), f(A, x) \rangle - f(A,y) \langle c(A, y), f(A, y) \rangle\|_2 \leq 5R_f \| x - y \|_2
    \end{align*}
\end{lemma}

\begin{proof}
    We have
    \begin{align*}
        & ~ \| f(A,x) \langle c(A, x), f(A, x) \rangle - f(A,y) \langle c(A, y), f(A, y) \rangle\|_2 \\
        \leq & ~ \| f(A,x) \langle c(A, x), f(A, x) \rangle - f(A,y) \langle c(A, x), f(A, x) \rangle\|_2 \\
        & ~ + \| f(A,y) \langle c(A, x), f(A, x) \rangle - f(A,y) \langle c(A, y), f(A, x) \rangle\|_2 \\
        & ~ + \| f(A,y) \langle c(A, y), f(A, x) \rangle - f(A,y) \langle c(A, y), f(A, y) \rangle\|_2 \\
        \leq & ~ \langle c(A, x), f(A, x) \rangle \| f(A,x) - f(A,y) \|_2 \\
        & ~ + \| f(A,y) \langle c(A, x), f(A, x) \rangle - f(A,y) \langle c(A, y), f(A, x) \rangle\|_2 \\
        & ~ + \| f(A,y) \langle c(A, y), f(A, x) \rangle - f(A,y) \langle c(A, y), f(A, y) \rangle\|_2 \\
        \leq & ~ \langle c(A, x), f(A, x) \rangle \| f(A,x) - f(A,y) \|_2 \\
        & ~ + \|f(A,y)\|_2 \cdot \|f(A, x) \|_2 \cdot \| c(A, x) - c(A, y) \|_2 \\
        & ~ + \| f(A,y) \langle c(A, y), f(A, x) \rangle - f(A,y) \langle c(A, y), f(A, y) \rangle\|_2 \\
        \leq & ~ \langle c(A, x), f(A, x) \rangle \| f(A,x) - f(A,y) \|_2 \\
        & ~ + \|f(A,y)\|_2 \cdot \|f(A, x) \|_2 \cdot \| c(A, x) - c(A, y) \|_2 \\
        & ~ + \|f(A,y)\|_2 \cdot \|c(A, y) \|_2 \cdot \| f(A, x) - f(A, y) \|_2 \\
        \leq & ~ \| c(A, x)\|_2\cdot\| f(A, x) \|_2 \cdot \| f(A,x) - f(A,y) \|_2 \\
        & ~ + \|f(A,y)\|_2 \cdot \|f(A, x) \|_2 \cdot \| c(A, x) - c(A, y) \|_2 \\
        & ~ + \|f(A,y)\|_2 \cdot \|c(A, y) \|_2 \cdot \| f(A, x) - f(A, y) \|_2 \\
        \leq & ~ 2 \cdot \| f(A,x) - f(A,y) \|_2 \\
        & ~ + 1 \cdot \| c(A, x) - c(A, y) \|_2 \\
        & ~ + 2 \cdot \| f(A, x) - f(A, y) \|_2 \\
        \leq & ~ 4R_f \| x - y \|_2 \\
        & ~ + 1 \cdot \| c(A, x) - c(A, y) \|_2 \\
        \leq & ~ 5R_f \| x - y \|_2
    \end{align*}
    where the first step follows from Lemma~\ref{lem:product_rule}, the second, third, fourth and fifth steps follow from Fact~\ref{fac:vector_norm}, the sixth step follows from Part 3 and Part 4 of Lemma~\ref{lem:upper_bound}, the last two steps follow from Lemma~\ref{lem:lischitz_f_c}. 
\end{proof}

\begin{lemma}\label{lem:lipschitz_diag_f_c}
    If the given conditions are satisfied
    \begin{itemize}
        \item Let $A \in \R^{n \times d}$
        \item Let $b \in \R^{n}$ satisfy that $\| b \|_1 \leq 1$
        \item Let $\beta \in (0, 0.1)$
        \item Let $R \geq 4$
        \item $\| A \| \leq R$
        \item Let $R_f := \beta^{-2} n^{1.5} \exp(3R^2)$ 
        \item Denote $f(A, x)$ as Definition~\ref{def:f}
        \item Denote $c(A, x)$ as Definition~\ref{def:c}
    \end{itemize}
    For all $x, y \in \R^d$, we have
    \begin{align*}
        \|\diag(f(A, x))  c(A,x) -  \diag(f(A, y))  c(A,y)  \|_2 \leq 2R_f \| x - y \|_2
    \end{align*}
\end{lemma}

\begin{proof}
    We have
    \begin{align*}
        & ~ \|\diag(f(A, x))  c(A,x) -  \diag(f(A, y))  c(A,y)  \|_2 \\
        \leq & ~ \|\diag(f(A, x))  c(A,x) -  \diag(f(A, y))  c(A,x)  \|_2 \\
        & ~ +  \|\diag(f(A, y))  c(A,x) -  \diag(f(A, y))  c(A,y)  \|_2 \\
        \leq & ~ \|c(A,x)\|_2 \cdot \|\diag(f(A, x)) -  \diag(f(A, y))  \|_F \\
        & ~ +  \|\diag(f(A, y))  c(A,x) -  \diag(f(A, y))  c(A,y)  \|_2 \\
        \leq & ~ \|c(A,x)\|_2 \cdot \|\diag(f(A, x)) -  \diag(f(A, y))  \|_F \\
        & ~ +  \| \diag(f(A, y)) \|_F \cdot \|  c(A,x) -  c(A,y)  \|_2 \\
        = & ~ \|c(A,x)\|_2 \cdot \|\diag(f(A, x)) -  \diag(f(A, y))  \|_F \\
        & ~ +  \| f(A, y) \|_2 \cdot \|  c(A,x) -  c(A,y)  \|_2 \\
        = & ~ \|c(A,x)\|_2 \cdot \|\diag(f(A, x) - f(A, y))  \|_F \\
        & ~ +  \| f(A, y) \|_2 \cdot \|  c(A,x) -  c(A,y)  \|_2 \\
        = & ~ \|c(A,x)\|_2 \cdot \| f(A, x) - f(A, y)  \|_2 \\
        & ~ +  \| f(A, y) \|_2 \cdot \|  c(A,x) -  c(A,y)  \|_2 \\
        \leq & ~ 2 \cdot \| f(A, x) - f(A, y)  \|_2 \\
        & ~ +  \| f(A, y) \|_2 \cdot \|  c(A,x) -  c(A,y)  \|_2 \\
        \leq & ~ 2 \cdot \| f(A, x) - f(A, y)  \|_2 +  1 \cdot \|  c(A,x) -  c(A,y)  \|_2 \\
        \leq & ~ 2 R_f \| x - y \|_2
    \end{align*}
    where the first step follows from Lemma~\ref{lem:product_rule}, the second and third steps follow from Fact~\ref{fac:matrix_norm}, the fourth step follows from Fact~\ref{fac:circ_rules}, the fifth step follows from Fact~\ref{fac:vector_norm}, the sixth and seventh steps follow from Part 3 and Part 4 of Lemma~\ref{lem:upper_bound}, the last two steps follow from Lemma~\ref{lem:lischitz_f_c}.
\end{proof}

\begin{lemma}\label{lem:lischitz_f_c}
    If the given conditions are satisfied
    \begin{itemize}
        \item Let $A \in \R^{n \times d}$
        \item Let $b \in \R^{n}$ satisfy that $\| b \|_1 \leq 1$
        \item Let $\beta \in (0, 0.1)$
        \item Let $R \geq 4$
        \item $\| A \| \leq R$
        \item Let $R_f := \beta^{-2} n^{1.5} \exp(3R^2)$ 
        \item Denote $f(A, x)$ as Definition~\ref{def:f}
        \item Denote $c(A, x)$ as Definition~\ref{def:c}
    \end{itemize}
    For all $x, y \in \R^d$, we have
    \begin{itemize}
        \item Part 1. (see Part 4 of Lemma 7.2 in \cite{dls23})
        \begin{align*}
            \|f(A, x) - f(A, y)\|_2 \leq R_f \cdot \|x - y\|_2
        \end{align*}
        \item Part 2. (see Part 5 of Lemma 7.2 in \cite{dls23})
        \begin{align*}
            \|c(A, x) - c(A, y)\|_2 \leq R_f \cdot \|x - y\|_2
        \end{align*}
        \end{itemize}
\end{lemma}

\subsection{Lipschitz Property for \texorpdfstring{$\mathcal{M}(x)$}{}}

\begin{lemma}
    If the given conditions are satisfied
    \begin{itemize}
        \item Let $A \in \R^{n \times d}$
        \item Let $b \in \R^{n}$ satisfy that $\| b \|_1 \leq 1$
        \item Let $\beta \in (0, 0.1)$
        \item Let $R \geq 4$
        \item $\| A \| \leq R$
        \item Let $R_f := \beta^{-2} n^{1.5} \exp(3R^2)$ 
        \item Let $\mathcal{M}(x)$ be define as Definition~\ref{def:M_1}
    \end{itemize}
    For all $x, y \in \R^d$, we have
    \begin{align*}
        \| \mathcal{M}(x) - \mathcal{M}(y) \|_2 \leq 7 R R_f \| x - y \|_2
    \end{align*}
\end{lemma}

\begin{proof}
    We have
    \begin{align*}
        \| \mathcal{M}(x) - \mathcal{M}(y) \|_2
        = & ~ \|\frac{1}{N} \sum_{i=1}^N g(A_i, x) + \xi^{(i)} - \frac{1}{N} \sum_{i=1}^N g(A_i, y) - \xi^{(i)} \|_2 \\
        = & ~ \|\frac{1}{N} \sum_{i=1}^N g(A_i, x) - \frac{1}{N} \sum_{i=1}^N g(A_i, y) \|_2 \\
        = & ~ \frac{1}{N}\| \sum_{i=1}^N g(A_i, x) - \sum_{i=1}^N g(A_i, y) \|_2 \\
        \leq & ~ \frac{1}{N} \sum_{i=1}^N  \| g(A_i, x) - g(A_i, y) \|_2 \\
        \leq & ~ \frac{1}{N} \sum_{i=1}^N  7 R R_f \| x - y \|_2 \\
        = & ~ 7 R R_f \| x - y \|_2
    \end{align*}
    where the first step follows from Definition~\ref{def:M_1}, the second step follows from simple algebra, the third and fourth steps follow from Fact~\ref{fac:vector_norm}, the last step follows from Lemma~\ref{lem:lipschitz_g}.
\end{proof}

\subsection{Corollary of Lipschitz Property}

\begin{corollary}\label{cor:lipschitz}
    If a loss function $L(x)$ satisfies $\| \nabla L(x) - \nabla L(y) \|_2 \leq \gamma \| x - y\|_2$, then we have
    \begin{align*}
        L(x) - ( L(y) + \langle \nabla L(y) (x - y) \rangle \leq \frac{1}{2} \gamma \| x - y \|_2^2
    \end{align*}
\end{corollary}

\subsection{Computation of \texorpdfstring{$\| \xi \|_2$}{}}

\begin{lemma}\label{lem:E_xi}
    For all $\xi \in \R^d$ be sampled as ${\rm Laplace}(0, \frac{\Delta}{\varepsilon})$ as Definition~\ref{def:laplace_noise}, we have
    \begin{align*}
        \E[\| \xi \|_2] = \sqrt{2 d (\frac{\Delta}{\varepsilon})^2}
    \end{align*}
\end{lemma}

\begin{proof}
    We have
    \begin{align*}
        \E[\| \xi \|_2]
        = & ~ \sqrt{\sum_{i=1}^d \E[\xi_i^2]} \\
        = & ~ \sqrt{\sum_{i=1}^d \int_{-\infty}^{+\infty} \xi_i^2 \Pr[\xi_i]\d \xi_i} \\
        = & ~ \sqrt{\sum_{i=1}^d 2(\frac{\Delta}{\varepsilon})^2} \\
        = & ~ \sqrt{ 2d(\frac{\Delta}{\varepsilon})^2}
    \end{align*}
    where the first step follows from simple norm rule, the second step follows from simple probability rule, the third step follows from simple integral rule and probability rule, the last step follows from simple algebra.
\end{proof}

\ifdefined\isarxiv
%\section*{Acknowledgments}
\bibliographystyle{alpha}
\bibliography{ref}
\fi
%%%% Cut-line between first 10 pages and appendix

%%% some writing rules

%% Writing rule for creating tags.
%% Tags :
%% Theorem    \ref{thm:bla_bla}
%% Lemma      \ref{lem:bla_bla}
%% Claim      \ref{cla:bla_bla}
%% Corollary  \ref{cor:bla_bla}
%% Fact       \ref{fac:bla_bla}
%% Definition \ref{def:bla_bla}
%% Section    \ref{sec:bla_bla}
%% Subsection \ref{sub:bla_bla}
%% Equation   \ref{eq:bla_bla}

\end{document}